\documentclass[twoside]{memoir}

\usepackage{makeidx}         
\usepackage{graphicx}        
\usepackage{multicol}        
\usepackage[bottom]{footmisc}

\usepackage{amsthm,amsfonts,amssymb,amsmath}
\usepackage{xspace}       
\usepackage{stmaryrd}     
\usepackage{bbm}
\usepackage{mathrsfs}  
\usepackage{xurl}
\usepackage{ulem}
\usepackage{multirow}
\usepackage{makecell}

\usepackage[utf8]{inputenc}

\usepackage{fancyvrb}

\usepackage[shortlabels]{enumitem}
\setlist[itemize]{leftmargin=*} 
\setlist[enumerate]{leftmargin=*}

\setsecnumdepth{subsection}

\newtheorem{theorem}{Theorem}[chapter] 

\newtheorem{lemma}[theorem]{Lemma}
\newtheorem{proposition}[theorem]{Proposition}
\newtheorem{corollary}[theorem]{Corollary}
\newtheorem{definition}[theorem]{Definition}

\theoremstyle{remark} 
\newtheorem{remark}{Remark}[chapter]
\newtheorem{example}{Example}[chapter]
\newtheorem{conjecture}{Conjecture}[chapter]

\theoremstyle{definition}
\newtheorem{myexercise}{Exercise}[chapter]

\newtheorem*{hint}{Hint}

\DeclareMathOperator{\range}{range}
\DeclareMathOperator{\card}{card}
\DeclareMathOperator{\MSE}{MSE}
\DeclareMathOperator{\Bias}{Bias}
\DeclareMathOperator{\Col}{Col}

\newcommand{\R}{\mathbb{R}}
\newcommand{\C}{\mathbb{C}}

\newcommand{\Z}{\mathbb{Z}}
\newcommand{\N}{\mathbb{N}}
\newcommand{\symS}{\mathbb{S}}
\newcommand{\Hsp}{\mathbb{H}}
\newcommand{\Tn}{\mathcal{T}}
\newcommand{\Hilb}{\mathbb{H}}
\newcommand{\calX}{\mathcal{X}}
\newcommand{\gramK}{{G_K}}
\newcommand{\Pcol}{P_{\Col(X)}}
\newcommand{\Lag}{\mathcal{L}}

\usepackage{mybackref}

\def\C{\mathbb{C}}
\def\N{\mathbb{N}}

\def\R{\mathbb{R}}

\def\Z{\mathbb{Z}}
\def\E{\mathbb{E}}  
\def\P{\mathbb{P}}  

\def\eps{\varepsilon}
\def\epsilon{\varepsilon}

\DeclareMathOperator{\rank}{rank}
\DeclareMathOperator{\diag}{diag}
\DeclareMathOperator{\trace}{trace}
\DeclareMathOperator{\Tr}{trace}
\DeclareMathOperator{\Vol}{Vol}
\DeclareMathOperator{\Var}{Var}
\DeclareMathOperator{\erfc}{erfc}
\DeclareMathOperator{\err}{err}

\newcommand{\RR}{\mathbb{R}}

\newcommand{\NN}{\mathbb{N}}
\newcommand{\EE}{\mathbb{E}}
\newcommand{\CC}{\mathbb{C}}
\newcommand{\PP}{\mathbb{P}}
\newcommand{\SSS}{\mathbb{S}}

\newcommand{\AAAA}{\mathcal{A}}

\newcommand{\NNN}{\mathcal{N}}
\newcommand{\CNNN}{\mathbb{C}\mathcal{N}}
\newcommand{\RRR}{\mathcal{R}}

\newcommand{\OOO}{\mathcal{O}}
\newcommand{\LLL}{\mathcal{L}}
\newcommand{\LL}{\mathcal{L}}

\newcommand{\MMM}{\mathcal{M}}

\theoremstyle{definition} 
\newtheorem{fact}{Fact}[chapter]
\newcommand{\Repart}{\operatorname{Re}}
\newcommand{\Impart}{\operatorname{Im}}
\newcommand{\cov}{\operatorname{Cov}}
\newcommand{\Cov}{\operatorname{Cov}}
\newcommand{\vol}{\operatorname{vol}}
\newcommand{\cut}{\operatorname{cut}}

\newcommand{\Prob}{{\P}}

\newcommand{\sign}{\operatorname{sign}}
\newcommand{\tr}{\operatorname{Tr}}
\newcommand{\argmin}{\operatorname{argmin}}

\newcommand{\Id}{\operatorname{I}}

\newcommand{\abs}[1]{|#1|}
\newcommand{\defeq}{\mathrel{\mathop:}=}

\newcommand{\proofb}[1]{{\noindent\emph{Proof.}} #1 \hfill$\Box$\linebreak}
\newcommand{\argmax}{\operatornamewithlimits{argmax}}
\newcommand{\ceil}[1]{\lceil #1 \rceil}
\newcommand{\floor}[1]{\lfloor #1 \rfloor}

\newcommand{\1}{\mathbf{1}}
\newcommand{\polylog}{\mathrm{polylog}}
\newcommand{\Ncut}{\operatorname{Ncut}}
\newcommand{\supp}{\operatorname{supp}}
\newcommand{\ess}{\operatorname{ess}}

\newcommand{\Gp}{\mathcal{G}_+}
\newcommand{\Gm}{\mathcal{G}_-}
\newcommand{\Dp}{D_{\mathcal{G}}^+}
\newcommand{\Dm}{D_{\mathcal{G}}^-}
\newcommand{\pp}{\frac{\alpha\log(n)}{n}}
\newcommand{\qq}{\frac{\beta\log(n)}{n}}
\newcommand{\NC}{\mathrm{NC}}
\newcommand{\Dphi}{\mathcal{D}} 

\newcommand{\ddiag}{\mathrm{ddiag}}
\newcommand{\phiLap}{\rho}
\newcommand{\mb}{\mathbf}
\newcommand{\vecc}{\mathrm{vec}}

\usepackage{enumerate}

\usepackage{mathtools}
\usepackage[chapter]{algorithm}
\usepackage{subcaption}
\usepackage{tikz}
\usetikzlibrary{arrows.meta}
\usepackage[noend]{algpseudocode}

\usepackage{makeidx}         
\usepackage{multicol}        

\nouppercaseheads

\makeevenhead{headings}{\thepage}{}{\normalfont\leftmark}
\makeoddhead{headings}{\normalfont\rightmark}{}{\thepage}

\newcommand{\Rade}{\mathfrak{R}}

\newcommand{\norm}[1]{{\left\lVert{#1}\right\rVert}}
\newcommand{\brap}[1]{\left( #1 \right)}
\newcommand{\bras}[1]{\left[ #1 \right]}
\newcommand{\brac}[1]{\left\{ #1 \right\}}
\newcommand{\bran}[1]{\left\langle #1 \right\rangle}

\newcommand{\Cleak}{\xi}
\DeclareMathOperator{\mse}{mse}

\newcommand{\level}{}
\newcommand{\sep}{}

\newcommand{\DD}{\mathbb{D}}

\definecolor{AfonsoBlue}{RGB}{30,65,123}
\newcommand{\notezerox}[1]{{\texttt{\textbf{#1}}}}

\newcommand{\asb}[1]{}
\newcommand{\amit}[1]{}
\newcommand{\tos}[1]{}
\newcommand{\ffv}[1]{}

\begin{document}

\thispagestyle{empty}
\begin{center}
\includegraphics[width=\textwidth]{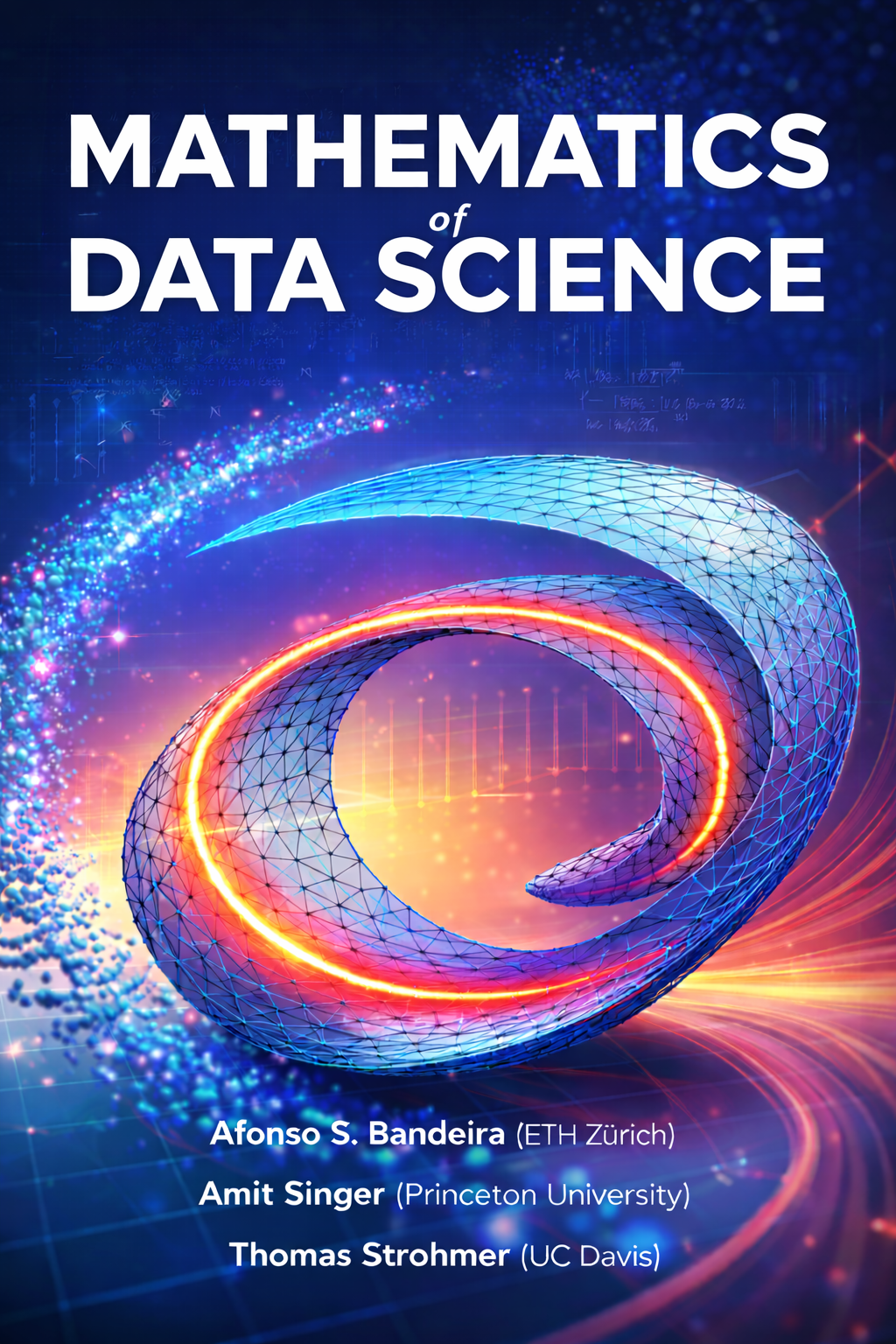}
\end{center}
\newpage

\thispagestyle{empty}
\author{Afonso S. Bandeira (ETH Z\"{u}rich) \\ Amit Singer (Princeton University)\\ Thomas Strohmer (UC Davis)}
\title{\vspace{1cm}  \textbf{Mathematics of Data Science} \notezerox{\\ \textcolor{white}{.} \hfill\Large{Preprint: version 1.1-}}}
\maketitle
\thispagestyle{empty}

\frontmatter


\tableofcontents

\mainmatter

\setcounter{chapter}{-1}

\chapter{Notes on this Version and Current Status}

\notezerox{
This is a preprint of a book in preparation by the authors.\\
}

\notezerox{ 
We anticipate the focus and content of this manuscript
not to change except for the correction of typos and other errors and other minor changes such as including additional homework problems. It can
 already by used as textbook for a graduate course in Mathematics of Data Science; it has been used as such by the authors at their home institutions.\\
}

\notezerox{
We welcome suggestions and comments, and would like to learn about any possible errors and typos.   
}

\notezerox{
Please contact the authors at \url{bandeira@math.ethz.ch}, \\\url{strohmer@math.ucdavis.edu}, or \url{amits@math.princeton.edu}. In fact, this version was improved using many suggestions sent from, and typos found by, many students and readers. Thank you! \\
}

\notezerox{
Thank you,
}

\notezerox{
Afonso, Thomas, and Amit.
}


\normalem

\chapter{Introduction}\label{c:introduction}

\section{Origins}

The idea of this book emerged from the classes on ``mathematics of data science'' that the authors have been teaching at their home
institutions Princeton University (AS), UC Davis (TS), and MIT, NYU, and ETH (ASB)\footnote{Lecture notes from these classes also
contain a list of mathematical open problems~\cite{Afonso_10L42P}, which has now helped to inspire a mathematical blog of open problems at~\url{https://randomstrasse101.math.ethz.ch/}. The open problems of 2024 are available at~\cite{RandomstrasseProblems2024}.}. The lecture notes eventually became the basis to this book. While there are specialized books and many advanced research papers on each topic covered in this book, we believe it fills in the gap of being a single source that provides an accessible entry point at an advanced undergraduate, or graduate, level to mathematics of data science. The book can serve readers as a segue to those more advanced and specialized topis. In that light, one of the main purposes of this book is to be used as a textbook at both the (advanced) undergraduate and graduate levels. This book also differs from most texts in the way that it interweaves mathematical theory with some of its applications in data science.

\section{The data science revolution}

The data science revolution is one of the most profound shifts in modern history, reshaping industries, economies, and societies with unprecedented speed and scale.
At its core, the data science revolution is driven by the exponential growth in data, advances in computing power, and the development of sophisticated analytical techniques.

Data science and its cousin artificial intelligence have  become a cornerstone of modern decision-making and innovation, penetrating almost all aspects of our life.
In healthcare, for example, data science is driving personalized medicine and predictive analytics. By analyzing patient data, healthcare providers can tailor treatments to individual needs, predict potential health issues before they arise, and improve patient outcomes.
In finance, data science is revolutionizing risk management, fraud detection, and investment strategies. Algorithms analyze market trends, assess creditworthiness, and identify unusual transactions to mitigate risks and enhance decision-making.

Retailers leverage data science to understand consumer behavior, optimize supply chains, and personalize marketing efforts.  The transportation industry benefits from data science through advancements in autonomous vehicles, route optimization, and predictive maintenance. Companies use data to enhance logistics, reduce costs, and improve safety.
The energy sector uses data science to optimize resource allocation, predict equipment failures, and enhance energy efficiency. Data-driven insights help in managing renewable energy sources and reducing environmental impact.

From the statistical techniques that form the basis of data analysis to the complex algorithms that power machine learning and artificial intelligence, mathematics is at the heart of data science. Understanding the connection between data science and mathematics is crucial for appreciating the depth and breadth of this revolution.

\section{Mathematics as a foundation of data science}

Data science is an inherently interdisciplinary field that merges techniques from mathematics, statistics, computer science, signal processing, and domain-specific knowledge to extract insights from data. However, at its core, data science is fundamentally about understanding and leveraging patterns in data. This understanding is built on a complex and nuanced foundation of mathematical principles that enable the development of models and algorithms capable of making sense of complex datasets. This book aims to unravel many of these mathematical principles and provide a thorough grounding in the essentials of data science from a rigorous, mathematical perspective.

Mathematics provides the language and framework for formalizing the processes involved in data analysis, from data collection and preprocessing to modeling and evaluation. It offers the tools necessary for designing effective algorithms, making precise predictions, visualizing complex datasets, quantifying uncertainty, and interpreting the results of data-driven experiments. 
As the field of data science keeps evolving at a rapid speed, new algorithms and techniques are constantly being developed. A strong mathematical foundation enables practitioners to quickly understand and adopt these innovations because they can grasp the underlying principles rather than just following black-box implementations.  Moreover, with a solid grasp of mathematics, learning new methods becomes easier.

Also, when algorithms fail to produce correct results in real world applications, we would like to know why they failed. Is it because of 
 mistakes in the experimental setup, did we gather the wrong kind of data, do we have insufficient data, is it because of corrupted measurements, calibration errors, or  incorrect modeling assumptions, or perhaps due to fundamentally unrealistic expectations, or is it due to a deficiency of the algorithm itself? If it is the latter, can it be fixed by a better initialization, a more careful tuning of the parameters, or by choosing a different algorithm? Or is a more fundamental modification required, such as developing a different model, including additional prior information, taking more measurements, or a better compensation of calibration errors? How reliable are our  algorithms to small changes in the input data and
can we explain why an algorithm arrives at a certain prediction?

As we are dealing with larger datasets and we aim for faster and faster throughput,  it becomes increasingly important to address the aforementioned challenges in a systematic and principled manner.
Thus, a rigorous and thorough study of computational algorithms both from a theoretical and numerical viewpoint is not a luxury, but  emerges as
an imperative ingredient towards effective data-driven discovery.

While a thorough mathematical framework can never completely avoid some guessing and trial-and-error, it can dramatically reduce  the time and resources we spend on trial-and-error. Good mathematical theory can help in predicting if a problem is solvable at all in feasible time. Moreover, fundamental mathematical analysis often leads to new discoveries such as faster or more robust algorithms, and it may reveal hitherto unknown connections to other, already well studied problems.

At the heart of data science lies a rich tapestry of mathematical concepts. Key areas include statistics and  probability, linear algebra, optimization,  numerical algorithms, analysis, and graph theory,  but this list is by no means exhaustive. This book emphasizes the geometric viewpoint of data sets. Specifically, it is a very powerful way to think about a dataset as a point cloud in a high dimensional space for a wide range of analysis tasks such as clustering, classification, dimension reduction, prediction, denoising, outlier detection, and more. In this geometric viewpoint, linear models of data are interpreted as linear subspaces in a Euclidean space, while non-linear models can be viewed as manifolds embedded in the ambient space. Equally powerful is the representation of data points and their interconnections and similarities as a graph and learning using graph algorithms and analysis. In particular, the graph representation provides a powerful way to extend classical methods for function approximation and interpolation to the setting of complex and high dimensional setting of modern data sets. The computational machinery provided by classical harmonic analysis on Euclidean spaces such as Fourier analysis can be extended in this way to more complex geometries and networks.

As data science continues to evolve, new mathematical techniques and approaches will emerge, expanding the possibilities for analyzing and interpreting data. This book provides a solid foundation upon which readers can build their expertise, adapt to new developments, and contribute to the advancement of the field.

Data science also opens a new playground for mathematics, motivating new questions and new areas of mathematical inquiry. For this reason, another goal of this book is to motivate mathematicians (and, in particular, mathematics students) to engage with these topics and to create new mathematics motivated by questions arising from data and computation.

\section{Whom is this book for?}

Many of the concepts, algorithms, and methods presented in this book do not only belong in the toolbox of a well trained data scientist, but are also found in the standard repertoire of researchers and practitioners of machine learning and AI.

The purpose of this book is twofold.  On the one hand. the book is designed to guide readers through many of the mathematical foundations of data science in a structured and accessible manner. It assumes a basic familiarity with mathematical concepts and aims to build upon this foundation to explore more advanced topics relevant to data science.
On the other hand, the selection of data science topics for this book is also guided by our desire to demonstrate how data science leads to deep and exciting mathematical problems, and how (sometimes seemingly unrelated) mathematical concepts can provide key insight into data science tasks. This mutual enrichment is akin to how physics and mathematics have had a vibrant cross-fertilization over many centuries.

While the tools and methods of data science and machine learning that are currently {\em en vogue} tend to change quickly, a solid grasp of the underlying mathematical principles provides an enduring foundation that allows one to adapt to, and even shape, the innovations of  future data science. We sincerely hope that this book can make a contribution to this endeavor. 

\emph{What prerequisites should a reader have?}  An undergraduate level knowledge of linear algebra and probability is assumed.

\section{Book organization and some omissions}

\emph{What is covered in this book?}
This book contains the material that one might want to teach in either a year-long course or a one-semester course on the mathematical foundations of data science.

The book is separated into  sixteen chapters, and an effort was made to intertwine mathematical techniques and motivating applications. It starts with a description of the curious phenomena of high dimensional geometry, taking the opportunity to introduce and recall some probability theory in Chapter~\ref{c:surprises}. It proceeds to discussing linear dimension reduction techniques (such as PCA and truncated SVD) and doing a recap of several linear algebra concepts in Chapter~\ref{c:svd}. Linear regression and least squares are the central topics in Chapter~\ref{c:linreg_ls}.

We move on to graph theory (including spectral graph theory, and spectral clustering in Chapter~\ref{c:graphs}) and non-linear dimension reduction (including diffusion maps in Chapter~\ref{c:diffusion}), continuing to highlight how central the spectrum of matrices formed from data are to understanding the data geometry. 
Chapter~\ref{c:johnson} focuses on dimension reduction via randomized methods, including a brief introduction to randomized linear algebra. In Chapter~\ref{c:optimization} we dive into the tools of optimization that play a key role in data science, such as convex optimization and gradient descent.
We explore several classical classification methods in Chapter~\ref{c:classification} and present a gentle introduction to Deep Learning in Chapter~\ref{c:deeplearning}.

In Chapter~\ref{c:convergence} we analyze the large sample limit 
of diffusion maps and other non-linear dimensionality reduction methods.
The concept and power of convex relaxations is illustrated via the community detection problem in Chapter~\ref{c:community}. 
Chapters~\ref{c:probability-gaussiananalysis} and~\ref{c:probability-matrixconcentration} are devoted to probability, the first to Gaussian analysis and concentration of measure, and the second to matrix concentration inequalities. 
Finally, Chapters~\ref{c:cs} and~\ref{c:lowrank} deal with sparse data models, the first one is dedicated to Compressive Sensing, while the second one covers low-rank matrix recovery.

\emph{What is not covered in this book? }
This book does not cover everything related to mathematics of data science. Indeed, there are several topics not included to keep the book at a manageable length. 
Perhaps the most notable omissions are relevant topics from statistics such as statistical inference and parameter estimation (e.g., Bayesian inference, etc.). There are already excellent textbooks dedicated to these topics, which are also being typically taught in separate courses. Such courses in statistics are of course recommended to those interested in data science, and can be taken either before, after, or in parallel to a course based on this book. The book also does not cover more specialized (but definitely very important) topics like data privacy, which is deeply connected to the mathematics of data science. The book is also at fault to not discuss optimal transport which has gained a lot of attention in mathematical data science. And we do not dive too deep into learning theory (such as VC dimension), as there are already plenty of resources on this topic. Notwithstanding these deliberate omissions, we believe that this book covers most of the foundational mathematical pillars of data science.

\section{Some related resources}


At~{\footnotesize{\url{https://video.ethz.ch/lectures/d-math/2025/autumn/401-4944-20L}}} you can access videos of lectures given by one of the authors on parts of this book (or see~{\footnotesize{\url{https://people.math.ethz.ch/~abandeira/videos.html}}} for more videos on related topics).
For the readers that find themselves looking for more problems or exercises: you might enjoy this note~\cite{JigSawStatCompGaps2025} which presents many exciting new results in theoretical computer science and high dimensional statistics via mathematical exercises. For open problems related to many topics of this book, you can visit~{\footnotesize{\url{https://randomstrasse101.math.ethz.ch/}}} or~\cite{RandomstrasseProblems2024} (or~\cite{Afonso_10L42P} for a less updated list).

\section*{Acknowledgment}

We want to thank our students, colleagues, and readers of earlier drafts of this book, who provided valuable feedback, alerted us to errors, and pointed out typos. Please keep doing so! We are grateful to Junda (Albie) Sheng and Stefan Broecker for generating some of the figures and to Yuan Ni for providing code for some numerical simulations. We are very thankful for the invaluable help that teaching assistants provided while we teach our courses on this topic, including the designing of countless problems and exercises on these topics. A particular thank you to Pedro Abdalla, Kevin Lucca, Anastasia Kireeva, Antoine Maillard, Petar Niz\'{i}c-Nikolac, and Almut R\"odder, who have greatly helped with the teaching of 
classes on Mathematics of Data Science at ETH; many exercises in the book originated as homework problems in these classes~\cite{ExercisesInMDS2025}. Amit Singer acknowledges support from the Simons Foundation Math+X Investigator Award, AFOSR FA9550-20-1-0266, AFOSR FA9550-23-1-0249, NSF DMS 2009753, NSF DMS 2510039, and NIH/NIGMS R01GM136780-01.
Thomas Strohmer  acknowledges support from NSF DMS-2208356, NIH R01HL16351, P41EB032840, and DE-SC0023490. 


\chapter{Curses, Blessings, and Surprises in High Dimensions}
\label{c:surprises}

Most of the material in this book will naturally ``live'' in the context of high dimensions, either motivated by the analysis of high dimensional data or by associating with networks a space whose dimension is the number of network nodes, 
to name a few. This chapter is about High Dimensions. It discusses the curse of dimensionality, but also many of its blessings. The first is caused by the exponential increase in volume associated with adding extra dimensions to Euclidean space. The latter is a manifestation of an intriguing phenomenon called the concentration of measure. This concentration phenomenon will give rise to many surprising facts about high dimensional geometry that we will discuss.
Since several of the results discussed in this chapter require basic tools from probability, we will also review some fundamental probabilistic concepts.

\subsubsection{The curse of dimensionality}

The {\em curse of dimensionality} refers to the fact that many algorithmic approaches to problems in
$\R^d$ become {\em exponentially} more difficult as the dimension $d$ grows. The expression ``curse of dimensionality'' was coined by Richard Bellman to describe the problem caused by the exponential increase in volume associated with adding extra dimensions to Euclidean space~\cite{bellman1957}.

For instance, if we want to sample the unit interval such that the distance between adjacent points is at most 0.01,
then 100 evenly-spaced sample points suffice; an equivalent sampling of a five-dimensional unit hypercube with a grid with a spacing of 0.01 between adjacent points would require $10^{10}$ sample points. Thus, a modest increase in dimensions results in a dramatic increase in required data points to cover the space at the same density.

Intimately connected to the curse of dimensionality is the problem of {\em overfitting} and {\em underfitting}. Here, overfitting refers to the issue that an algorithm may show good performance on the training data, but poor generalization to other data. Underfitting in turn, corresponds to poor performance on the training data (and poor generalization to other data). This problem manifests itself in many machine learning algorithms.

We will discuss a toy example from image classification in more detail to illustrate the underlying issues.
Assume we want to classify images into two groups, cars and bicycles, say. From the vast number of images depicting cars or bicycles, we are only able to obtain a small number of training images, say five images of cars and five images of bicycles. We want to train a simple linear classifier based on these ten labeled training images to correctly classify the remaining unlabeled car/bicycle images. We start with a simple feature, e.g.\ the amount of red pixels in each image.
However, this is unlikely to give a linear separation of the training data. We add more features and eventually the training images become linearly separable. This might suggest that increasing the number of features until perfect classification of the training data is achieved, is a sound strategy. However, as we {\em linearly increase} the dimension of the feature space, the density of our training data {\em decreases exponentially} with the feature dimension.

In other words, to maintain a comparable density of our training data, we would need to increase the size of the dataset exponentially -- the curse of dimensionality.
Thus, we risk producing a model that could be very good at predicting the target class on the training set, but it may fail miserably when faced with new data.  This means that our model does not {\em generalize} from the training data to the test data, leading to large {\em generalization error}.  The generalization error is an important concept in data science and machine learning, since it measures how accurately an algorithm can predict outcomes for new, unseen data.  As such, it indicates the data science model's ability to perform well beyond the training data, and is closely related to the concept of overfitting.

\section{Geometry of spheres and cubes in high dimension}

When we peel an orange, then after having removed the rind we are still left with the majority of the orange. Suppose now we peel a $d$-dimensional orange for large $d$, then after removing the orange peel we would be left with essentially nothing.
The reason for this -- from a healthy nutrition viewpoint discouraging -- fact is that for a $d$-dimensional unit ball almost all of its volume is concentrated near the boundary sphere. This is just one of many surprising phenomena in high dimensions. Many of these surprises are actually a manifestation of some form of concentration of measure that we will analyze in more detail in the next section (and then these surprises are no longer so startling).

When introducing data analysis concepts, we typically use only a few dimensions in order to facilitate visualization.
However, our intuition about space, which is naturally based on two and three dimensions, can often be misleading in high dimensions.
Many properties of even very basic objects become counterintuitive in higher dimensions. Understanding these paradoxical properties is essential
in data analysis as it allows us to avoid pitfalls in the design of algorithms and statistical methods for high-dimensional data.
It is therefore instructive to analyze the shape and properties of some basic geometric forms that we understand very well
in dimensions two and three, in high dimensions.

To that end, we will look at some of the
properties of the sphere and the cube as the dimension increases.
The $d$-dimensional hyperball of radius $R$ is defined by
$$B^d(R) =\{x \in \R^d : x_1^2 + \dots +x_d^2 \le R^2\},$$ the $d$-dimensional hypersphere (or $d$-sphere) of radius $R$ is given by
$$S^{d-1}(R) =\{x \in \R^d : x_1^2 + \dots +x_d^2 = R^2\},$$
and the $d$-dimensional hypercube with side length $2R$  is the subset of $\R^d$ defined as the $d$-fold product of
intervals $[-R, R]$:
$$C^d(R) = \underbrace{ [-R,R] \times  \dots  \times [-R,R]}_{\text{$d$ times}}.$$
If there is no danger of confusion, we may write $B^d$ for the unit ball $B^d(1)$, $S^{d-1}$ for the unit sphere $S^{d-1}(1)$, and
$C^d$ for the unit hypercube $C^d(\frac{1}{2})$.

\subsection{High dimensional geometry and high dimensional probability} 

In the rest of this section we will establish a variety of (at first) surprising properties of high dimensional geometry. To illustrate the connections between geometry and probability we will derive some of the results first using analytical tools (computing volumes and integrals) and then using tools from Probability, namely concentration inequalities (which the reader might agree gives more elegant arguments). The goal here is not to get the best bounds, but to illustrate these properties and showcase the duality between geometry and probability.

\subsubsection{Volume of the hyperball}

\begin{proposition}
The volume of $B^d(R)$ is given by
\begin{equation}
\label{volsphere}
\Vol(B^d(R)) = \frac{\pi^{\frac{d}{2}} R^d}{\frac{d}{2}\ \Gamma(\frac{d}{2})}.
\end{equation}
\end{proposition}

\begin{proof}
The volume of $B^{d}(R)$ is given by
\begin{equation}\label{volume_dsphere}
\Vol(B^d(R)) = \int_0^R s_d r^{d-1} dr =  \frac{s_d R^d}{d},
\end{equation}
where $s_d$ denotes the (hyper-)surface area of a unit $d$-sphere. A unit $d$-sphere must satisfy
$$s_d \int_0^{\infty} e^{-r^2} r^{d-1} dr = \underbrace{\int_{-\infty}^{\infty} \cdots \int_{-\infty}^{\infty}}_{\text{$d$ times}} e^{-(x_1^2 + \cdots +x_d^2)}  dx_1 \cdots d x_d = \Big( \int_{-\infty}^{\infty} e^{-x^2} dx \Big)^d.$$
Recall that the Gamma function is given by
$$\Gamma(n) = \int_0^\infty r^{n-1} e^{-r}  dr = 2 \int_0^\infty e^{-r^2} r^{2n-1} dr,$$
hence
$\frac{1}{2} s_d \Gamma(\frac{d}{2}) = \Big[ \Gamma(\frac{1}{2}) \Big]^d = \big( \pi^{\frac{1}{2}}\big)^d,$
and thus
$s_d = \frac{2 \pi^{\frac{d}{2}} }{\Gamma(\frac{d}{2})}.$
Plugging this expression into~\eqref{volume_dsphere} gives
\begin{equation}
\label{volsphere}
\Vol(B^d(R)) = \frac{\pi^{\frac{d}{2}} R^d}{\frac{d}{2}\ \Gamma(\frac{d}{2})}.
\end{equation}
\end{proof}

Recall that for even $d$, we have $\Gamma(d/2) = (d/2 - 1)!$. To understand the asymptotic behavior $\Vol(B^d(R))$ we can use the celebrated Stirling's Formula,
\begin{align*}
  \Gamma(n) \approx \sqrt{\frac{2\pi}{n}}\left(\frac{n}{e}\right)^n
\end{align*}
we obtain an approximation for the volume of the unit $d$-ball for large $d$
\begin{align}\label{approxvolume}
  \Vol(B^d) \approx \frac{1}{\sqrt{d\pi}} \Big( \frac{2\pi e}{d} \Big)^{\frac{d}{2}},
\end{align}
where $\approx$ means that the quotient goes to $1$ as the relevant parameter, in this case $d$, goes to $\infty$.

\begin{figure}
\centering
\includegraphics[width=0.8\textwidth]{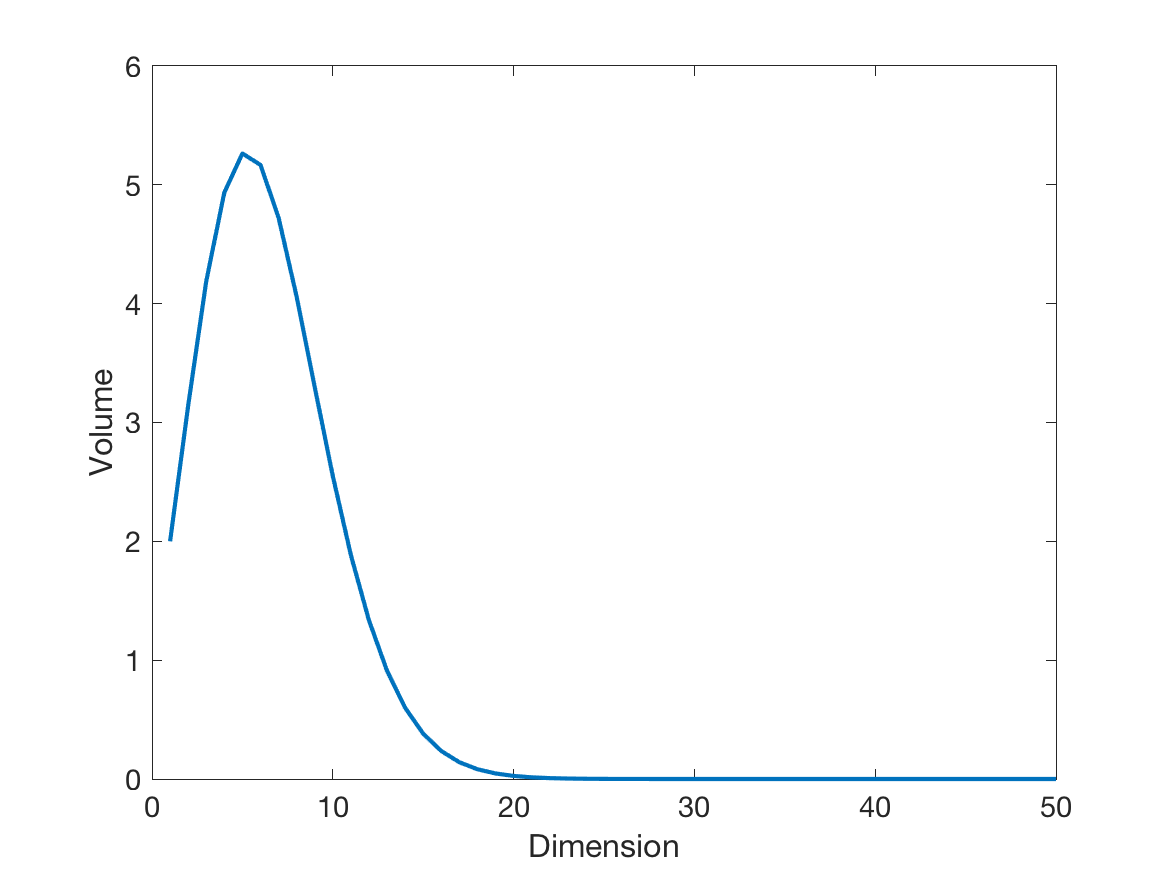}
\caption{The volume of the unit $d$-ball using the exact formula in equation~\eqref{volsphere}. The volume reaches its maximum for $d=5$ and decreases rapidly to zero with increasing dimension $d$.}
\label{fig:volume_sphere}
\end{figure}

Since the denominator in the parenthesis of equation~\eqref{approxvolume} goes to infinity,
the volume of the unit $d$-sphere goes rapidly to $0$ as the dimension $d$ increases to infinity, see also Figure~\ref{fig:volume_sphere}. Notice that this would still be the case for any fixed radius $R$.

Thus, unit spheres in high dimensions have almost no volume---compare this to the unit cube, which has volume $1$ in any dimension. For $B^d(R)$ to have volume equal to 1, its radius $R$ must be
approximately (asymptotically) equal to $\sqrt{\frac{d}{2\pi e}}$.

\subsubsection{Concentration of the volume of a ball near its equator}

If we take an orange and cut it into slices, then the slices near the center are larger since the sphere is wider there. This effect increases dramatically (exponentially with the dimension) with increasing dimension.
Assume we want to cut off a slab around the ``equator\footnote{To define the ``equator'' of a the $d$-dimensional ball, we need to pick a ``north pole'' as reference. Without loss of generality we could pick the unit vector in the $x_1$-direction as defining ``north''.}'' of the $d$-unit ball such that 99\% of its volume is contained inside the slab. In two dimensions the width of the slab has to be almost 2, so that 99\% of the volume are captured by the slab. But as the dimension increases the width of the slab gets rapidly smaller. Indeed, in high dimensions only a very thin slab is required, since nearly all the volume of the unit ball lies a very small distance away from the equator.  The following theorem makes the considerations above precise.


\begin{fact}\label{fact:spheremostlyinequator}
Almost all the volume of $B^d(R)$ lies near its equator.
\end{fact}

\begin{proof}
It suffices to prove the result for the unit $d$-ball.
Without loss of generality we pick as ``north'' the direction $x_1$. The intersection of the sphere with the plane $x_1=0$  forms our equator, which is formally given by the $d-1$-dimensional region
$\{x: \|x\| \le 1, x_1=0\}$, where $\|x\|$ is the Euclidean norm\footnote{Throughout the book,  $\|\cdot \|$ denotes the Euclidean norm if we are dealing with vectors and it denotes the operator norm if we are dealing with matrices or operators.} of $x$. This intersection is a sphere of dimension $d-1$ with volume $\Vol(B^{d-1})$ given by
the $(d-1)$-analog of formula~\eqref{volsphere} with $R=1$.

We now compute the volume of $B^d$ that lies between $x_1=0$ and $x_1=p_0$.
Let $P_0=\{x: \|x\|\le 1, x_1\ge p_0\}$ be the ``polar cap'', i.e., part of the sphere above the slab of width $2p_0$ around the equator.
To compute the volume of the cap $P$ we will integrate over all slices of the cap from $p_0$ to 1.  
Each such slice will be a sphere of dimension $d-1$ and radius $\sqrt{1-p^2}$, hence its  volume is
$(1-p^2)^{\frac{d-1}{2}} \Vol(B^{d-1})$. Therefore
$$\Vol(P) = \int_{p_0}^1 (1-p^2)^{\frac{d-1}{2}} \Vol(B^{d-1}) \, dp = \Vol(B^{d-1})   \int_{p_0}^1 (1-p^2)^{\frac{d-1}{2}} \, dp.$$
Using $e^x \ge 1+x$ for all $x$ we can upper bound  this integral by
\begin{eqnarray*}
\Vol(P) &\le& \Vol(B^{d-1}) \int_{p_0}^\infty e^{-\frac{d-1}{2}p^2} \, dp = \Vol(B^{d-1})\sqrt{\frac{2}{d-1}} \int_{p_0 \sqrt{\frac{d-1}{2}}}^{\infty} e^{-u^2}\,du \\
&=& \Vol(B^{d-1}) \sqrt{\frac{\pi}{2(d-1)}}\erfc\left(p_0 \sqrt{\frac{d-1}{2}} \right),
\end{eqnarray*}
where $\erfc(x) = \frac{2}{\sqrt{\pi}}\int_x^{\infty} e^{-u^2}\,du$ is the complementary error function. The upper bound $\erfc(x) \leq \frac{e^{-x^2}}{\sqrt{\pi} x}$ gives
\begin{equation}\label{eq:chapter2:volBvolP:1}
\Vol(P) \le \Vol(B^{d-1}) \sqrt{\frac{\pi}{2(d-1)}}\frac{e^{- \frac{d-1}{2}p_0^2}}{\sqrt{\pi} p_0 \sqrt{\frac{d-1}{2}}} =  \frac{\Vol(B^{d-1})}{d-1}\frac{e^{- \frac{d-1}{2}p_0^2}}{ p_0}
\end{equation}

Recall, from~\eqref{volsphere} that $\Vol(B^{d}) = \frac{\pi^{\frac{d}{2}} }{\frac{d}{2}\ \Gamma(\frac{d}{2})}$, so, for $d$ large enough (since $\frac{\Gamma(\frac{d}{2})}{\Gamma(\frac{d-1}{2})} \approx \sqrt{\frac{d}2}$),
\begin{equation}\label{eq:chapter2:volBvolP:2}
\Vol(B^{d-1}) = \frac{\pi^{-1/2}}{\frac{d-1}{d}} \frac{\Gamma(\frac{d}{2})}{\Gamma(\frac{d-1}{2})}\Vol(B^{d}) \leq \frac{d-1}2\Vol(B^{d}).
\end{equation}
Finally, combining~\eqref{eq:chapter2:volBvolP:1} and~\eqref{eq:chapter2:volBvolP:2} shows that the ratio between the volume of the polar caps and the entire hypersphere is bounded by
$$\frac{2 \Vol(P)}{\Vol(B^{d})} \le (d-1)\frac{ \Vol(P)}{\Vol(B^{d-1})} \le \frac{1}{p_0}\exp\left(-\frac{d-1}{2}p_0^2\right).$$
The expression above shows that this ratio decreases exponentially as both $d$ and $p$
increase, proving our claim that the volume of the sphere concentrates strongly around its equator.

\end{proof}


\subsection{Concentration of the volume of a ball on shells}

We consider two concentric balls $B^d(1)$ and $B^d(1-\eps)$. Using equation~\eqref{volsphere}, the ratio of their volumes is
\begin{align}
\frac{\Vol\big(B^d(1-\eps)\big)}{\Vol\big(B^d(1)\big)} = (1-\eps)^d.
\end{align}
Clearly, for every $\eps$ this ratio tends to zero as $d \to \infty$. This implies that the spherical shell given by the region between $B^d(1)$ and $B^d(1-\eps)$ will contain most of the volume of $B^d(1)$  for large enough $d$ even if $\eps$ is very small. How quickly does the volume concentrate at the surface of $B^d(1)$? We choose $\eps$ as a function of $d$, e.g. $\eps = \frac{t}{d}$, then
\[
\frac{\Vol(B^d(1-\eps))}{\Vol(B^d(1))} = \Big(1-\frac{t}{d}\Big)^d \to e^{-t}.
\]
Thus, almost all the volume of $B^d(R)$ is contained in an annulus of width $R/d$.

Therefore, if we peel a $d$-dimensional orange---and even if we peel it very carefully so that we remove only a very thin layer of its peel---we will have removed most of the orange and are left with almost nothing.

\subsubsection{Geometry of the hypercube} \label{ss:hypercube}

We have seen that most of the volume of the hypersphere is concentrated near its surface. A similar result also holds for the hypercube, and it is indeed a tendency for high-dimensional geometric objects. Yet, the hypercube exhibits
an even more interesting volume concentration behavior, which we will establish below.

We start with a basic observation.
\begin{proposition}
The hypercube $C^d$ has volume 1 and diameter $\sqrt{d}$.
\end{proposition}
The above proposition, although mathematically  trivial, hints already at a somewhat counterintuitive behavior of the cube in high dimensions. Its corners seem to get ``stretched out'' more and more, while the rest of the cube must ``shrink'' to keep the volume constant. This property becomes even more striking when we compare the cube with the sphere as the dimension increases.

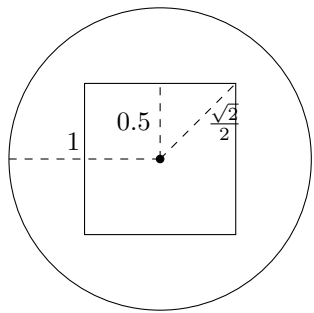
\begin{figure}[ht!]
  \centering
  \begin{tikzpicture}[scale=0.2]
  \draw (-5,5) -- (5,5) -- (5,-5) -- (-5,-5) -- (-5,5);
  \node[draw,circle,fill,scale=0.3] (origin) at (0,0) {};
  \draw[dashed] (0,0) -- node [left,midway] (half) { 0.5} (0,5);
  \draw[dashed] (0,0) -- node [right,midway] (root) {$\frac{\sqrt{2}}{2}$} (5,5);
  \draw[dashed] (0,0) -- node [above,near end] (one) {\qquad 1} (-10,0);
  \draw (0,0) circle (10);
  \end{tikzpicture}
  \caption{$2$-dimensional unit sphere and unit cube, centered at the origin.}
  \label{2dim}
\end{figure}

\begin{figure}[ht!]
  \centering
  \begin{tikzpicture}[scale=0.15]
  \draw (2,5) -- (17.07,17.07) -- (5,2) -- ( 5,-2) -- (17.07,-17.07) -- (2,-5) -- (-2,-5) -- (-17.07,-17.07) -- (-5,-2) -- (-5,2) -- (-17.07,17.07) -- (-2,5) -- (2,5);
  \node[draw,circle,fill,scale=0.3] (origin) at (0,0) {};
  \draw[dashed] (0,0) -- node [left,near end] (half) {0.5} (0,5);
    \draw[dashed] (0,0) -- node [right,near end] (half) {1} (10,0);

  \draw[dashed] (0,0) -- node [right,near end] (one) {\,$\sqrt{d}/2$} (17.07,17.07);
  \draw (0,0) circle (10);
  \end{tikzpicture}
  \caption{Projections of the $d$-dimensional unit sphere and unit cube, centered at the origin (4 of the $2^d$ vertices of the hypercube are shown).}
  \label{ddim}
\end{figure}
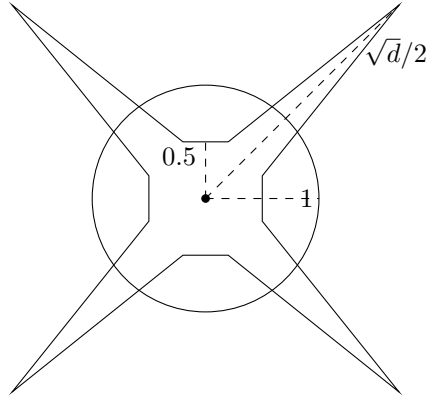

In two dimensions (Figure~\ref{2dim}), the unit square is completely contained in the unit sphere.  The distance from the center to a vertex (radius of the circumscribed sphere) is $\frac{\sqrt{2}}{2}$ and the apothem (radius of the inscribed sphere) is $\frac{1}{2}$.  In four dimensions (Figure~\ref{4dim}), the distance from the center to a vertex is $1$, so the vertices of the cube touch the surface of the sphere.  However, the apothem is still $\frac{1}{2}$.  The result, when projected in two dimensions no longer appears convex, however all hypercubes are convex.  This is part of the strangeness of higher dimensions - hypercubes are both convex and ``pointy.''  In dimensions greater than $4$ the distance from the center to a vertex is $\frac{\sqrt{d}}{2} > 1$, and thus the vertices of the hypercube extend far outside the sphere, cf.~Figure~\ref{ddim}.

\begin{figure}[ht!]
  \centering
  \begin{tikzpicture}[scale=0.2]
  \draw (2,5) -- (7.07,7.07) -- (5,2) -- ( 5,-2) -- (7.07,-7.07) -- (2,-5) -- (-2,-5) -- (-7.07,-7.07) -- (-5,-2) -- (-5,2) -- (-7.07,7.07) -- (-2,5) -- (2,5);
  \node[draw,circle,fill,scale=0.3] (origin) at (0,0) {};
  \draw[dashed] (0,0) -- node [right,near end] (half) {0.5} (0,5);
  \draw[dashed] (0,0) -- node [right,midway] (one) {\,1} (7.07,7.07);
  \draw (0,0) circle (10);
  \end{tikzpicture}
  \caption{Projections of the $4$-dimensional unit sphere and unit cube, centered at the origin (4 of the 16 vertices of the hypercube are shown).}
  \label{4dim}
\end{figure}
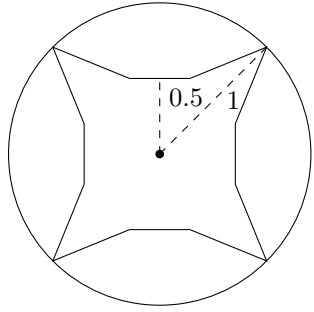

The considerations above suggest the following observation:


\begin{fact}\label{fact:hypercubemostlycorners}
Most of the volume of the high-dimensional cube is located in its corners.
\end{fact}


\begin{proof}
Let $C^d$ denote the unit hypercube. The subset $Q_R$ of $C^d$ of points $x\in\RR^d$ whose distance to the center is smaller that $R$ ($Q_R = \{x\in C^d: \|x\|\leq R\}$) is given by $Q = C^d\cap B^d(R)$. Thus
\[
\frac{\vol(Q_R)}{\vol(C^d)} = \vol(Q) \le B^d(R) \approx \frac{1}{\sqrt{d\pi}} \Big( \frac{2\pi e R^2}{d} \Big)^{\frac{d}{2}},
\]
which goes to zero as long as $R<\sqrt{\frac{d}{2\pi e}}$. This shows that most of the volume of $C^d$ lies in points at distance of order $\sqrt{d}$ of the center.

\end{proof}

\section{Basic concepts from probability}

We briefly review some fundamental concepts from probability theory, which are helpful or necessary to understand the blessings of dimensionality and some of the surprises encountered in high dimensions. More advanced probabilistic concepts
will be presented in Chapter~\ref{c:probability-gaussiananalysis}.
We assume that the reader is familiar with elementary probability as is covered in introductory probability courses (see, for example~\cite{durrett2019probability,ross2014introduction}).

The two most basic concepts in probability associated with a random variable $X$ are {\em expectation} (or {\em mean}) and {\em variance}, denoted by
$$\E[X] \qquad \text{and} \qquad \Var(X) := \E [ (X- \E [X])^2 ],$$
respectively. 

A useful identity is the {\em Law of Total Variance}. This fundamental theorem in probability theory decomposes the variance of a random variable into two distinct components: the variance that can be explained by another variable and the variance that remains as ``noise.'' Let $X$ and $Y$ be random variables on the same probability space, and assume that the variance of $X$ is finite. The Law of Total Variance states
\begin{eqnarray}\label{eq:totalvariance}
\Var(X) = \E[\Var(X \mid Y)] + \Var(\E[X \mid Y]).\end{eqnarray}
The two terms have a natural interpretation:
\begin{itemize}
\item $\E[\Var(X \mid Y)]$ is the unexplained variance --- the variability in $X$ that
persists even after knowing $Y$.
\item
$\Var(\E[X \mid Y])$ is the is the variance of the conditional mean --- the variability in $X$ explained by $Y$. 
\end{itemize}

An important tool to describe probability distributions is the {\em moment generating function} (MGF) of $X$, defined by
\begin{equation}\label{eq:MomentGeneratingFunction}    
M_X(t) = \E [e^{tX}], \quad t\in \R,
\end{equation}
the choice of nomenclature can be easily justified by expanding $M_X(t)$ in a series.
The $p$-th moment of $X$ is defined by $\E[X^p]$ for $p >0$ and the $p$-th absolute moment is $\E[|X|^p]$.

We can introduce $L^p$-norms of random variables by taking the $p$-th root of moments, i.e.,
$$\|X\|_{L^p} := \big(\E[|X|^p]\big)^{\frac{1}{p}}, \qquad p \in [0,\infty],$$
with the usual extension to $p=\infty$ by setting
$$\|X\|_{\infty} := \ess \sup |X|.$$

Let $(\Omega,\Sigma,\P)$ be a probability space, where $\Sigma$ denotes a $\sigma$-algebra on the sample space
$\Omega$ and $\P$ is a probability measure on $(\Omega,\Sigma).$ For fixed $p$ the vector space
$L^p(\Omega,\Sigma,\P)$ consists of all random variables $X$ on $\Omega$ with finite $L^p$-norm, i.e.,
$$L^p(\Omega,\Sigma,\P) = \{X : \|X\|_{L^p} < \infty\}.$$
The random variable can be viewed as a (measurable) map $X:\Omega\to\RR$ (or $\CC$). We will usually not mention the underlying probability space. For example, we will often simply write $L^p$ for $L^p(\Omega,\Sigma,\P)$.

The case $p=2$ deserves special attention since $L^2$ is a Hilbert space with inner product and norm
$$\langle X, Y \rangle_{L^2} = \E[XY], \qquad \|X\|_{L^2} = \big(\E [X^2] \big)^{\frac{1}{2}},$$
respectively. Note that the {\em standard deviation} $\sigma(X):=\sqrt{\Var(X)}$ of $X$ can be written as
$$\sigma(X) = \| X - \E[X] \|_{L^2}.$$
The {\em covariance} of the random variables $X$ and $Y$ is
\begin{equation}
\label{covariance}
\cov(X,Y) = \E [ (X - \E[X]) (Y - \E [Y])]  = \langle X - \E[X], Y - \E [Y] \rangle_{L^2} .
\end{equation}

We recall a few classical inequalities for random variables. {\em H\"older's inequality} states that for random variables $X$ and $Y$ on a common probability space and $p,q \ge 1$ with $1/p +1/q =1$, there holds
\begin{equation}
\label{hoelder}
| \E [XY] | \le  \|X\|_{L^p} \|Y\|_{L^q}. 
\end{equation}
The special case $p=q=2$ is the {\em Cauchy-Schwarz inequality}
\begin{equation}
\label{cauchy-schwarz}
 | \E [XY] | \le \sqrt{ \E [|X|^2] \E [|Y|^2]}.
\end{equation}

{\em Jensen's inequality} states that for any random variable $X$ and a convex function $\phi: \R \to \R$, we have
\begin{equation}
\label{Jensen}
\phi(\E[X]) \le \E [\phi(X)].
\end{equation}

Since $\phi(x) = x^{q/p}$ is a convex function for $q\geq p \geq 0$, it follows immediately from Jensen's inequality that
$$\| X\|_{L^p} \le \|X\|_{L^q} \qquad \text{for $0 \le p \le q < \infty$.}$$

{\em Minkovskii's inequality} states that for any $p \in [0,\infty]$ and any random variables $X, Y$, we have
\begin{equation}
\label{minkovski}
\|X + Y\|_{L^p} \le  \|X \|_{L^p} + \|Y\|_{L^p},
\end{equation}
which can be viewed as the {\em triangle inequality}.

\bigskip

The {\em cumulative distribution function} of $X$ is defined by
$$F_X (t) =\P(X \le t), \quad t \in \R.$$
We have $\P\{X > t\} = 1 -F_X(t),$
where the function $t \mapsto \P \{ |X| \ge t \}$ is called the {\em tail} of $X$.
The following lemma establishes a close connection between expectation and tails.
\begin{proposition}[Integral identity]
\label{integralidentity}
Let $X$ be a non-negative random variable. Then
$$
\E[X] = \int_0^{\infty} \P \{X > t \} \, dt.$$
The two sides of this identity are either finite or infinite simultaneously.

\end{proposition}

Given an event $E$ with non-zero probability,$\P(\cdot |E)$ denotes conditional probability, furthermore for a random variable $X$ we use $\E[X|E]$ to denote the conditional expectation.

\subsection{Tail bounds}

{\em Markov's inequality} is a fundamental tool to bound the tail of a random variable in terms of its expectation.

\begin{proposition}\label{prop:markov}
For any non-negative real random variable $X$  we have
\begin{equation}
\label{markov}
\P \{ X \ge t \} \le \frac{ \E[X] }{ t } \qquad \text{for all $t  > 0$.}
\end{equation}
\end{proposition}

We provide two versions of the same proof, one using the language of conditional expectations.

\begin{proof}

Let ${\mathcal I}$ denote the event $\{X \ge t\}$. Then
$$\E[X] = \sum_{s  \in S} p(s) X(s) = \sum_{s \in {\mathcal I}} p(s) X(s) + \sum_{s \in {\mathcal I}^c} p(s) X(s),$$
where $p(s)$ denotes the probability of $s$; in case of continuous variables this should be replaced with the density function and $\sum$ with an integral.

Since $X$ is non-negative, it holds $\sum_{s \in {\mathcal I}^c} p(s) X(s) \ge 0$ and
$$\E[X] \ge \sum_{s \in {\mathcal I}} p(s) X(s) \ge t \sum_{s \in {\mathcal I}} p(s) = t \P\{{\mathcal I}\}.$$ 
\end{proof}

\begin{proof}[Using the language of conditional expectation]
$$\E[X] = \P(X<t) \E[X | X<t] + \P(X>t) \E[X | X\geq t], $$
where we take the product to be zero if the probability is zero.

Since $X$ is non-negative, it holds $ \P(X<t) \E[X | X<t] \geq 0$. Also,  $\E[X | X \geq t] > t$. Hence,
$$\E[X] \ge \P(X>t) \E[X | X>t] \ge t \P(X \geq t).$$

\end{proof}

An important consequence of Markov's inequality is  {\em Chebyshev's inequality}.
\begin{corollary}\label{chebyshev}
Let $X$ be a random variable with mean $\mu$ and variance $\sigma^2$. Then, for any $t>0$
\begin{equation}
\label{chebyshev}
\P \{ | X - \mu | \ge t \} \le \frac{ \sigma^2 }{ t^2}.
\end{equation}
\end{corollary}

Chebyshev's inequality, which follows by applying Markov's inequality to the non-negative random variable $Y = (X - \E[X])^2$, is a form of concentration inequality, as it guarantees that $X$ must be close to its mean $\mu$ whenever the variance of $X$ is small. Both, Markov's and Chebyshev's inequality are sharp, i.e., in general they cannot be improved.

Markov's inequality only requires the existence of the first moment. We can say a bit more if in addition the random variable $X$ has a moment generating function
in a neighborhood around zero, that is, there is a constant $b>0$ such that $\E [ e^{\lambda (X-\mu)}]$ exists for all $\lambda \in [0,b]$.
In this case we can apply Markov's inequality to the random variable $Y = e^{\lambda (X-\mu)}$ and obtain the generic {\em Chernoff bound}
\begin{equation}\label{basicchernoff}
\P \{ X -\mu \ge t \} = \P \{ e^{\lambda (X-\mu)} \ge e^{\lambda t} \} \le \frac{ \E[e^{\lambda (X-\mu)}] }{ e^{\lambda t} }.
\end{equation}
In particular, optimizing over $\lambda$ in order to obtain the tightest bound in~\eqref{basicchernoff} gives
$$
\log \P \{ X -\mu \ge t \} \le - \sup_{ \lambda \in [0,b]}  \{ \lambda t - \log \E[e^{\lambda (X-\mu)} ]  \}.
$$

\bigskip
{\em Gaussian random variables} are among the most important random variables.
A Gaussian random variable $X$ with mean $\mu$ and standard deviation $\sigma$ has a probability density function given by
\begin{equation} \label{def:gaussian}
\psi(t) = \frac{1}{\sqrt{2 \pi \sigma^2}} \exp\left( - \frac{(t-\mu)^2}{2\sigma^2}  \right).
\end{equation}
We write $X \sim {\mathcal N}(\mu,\sigma^2)$.
We call a Gaussian random variable $X$ with $\E[X] = 0$ and $\E[X^2] = 1$  a {\em standard Gaussian} or {\em standard normal}  (random
variable). In this case we have the following tail bound.

\begin{proposition}[Gaussian tail bounds] \label{prop:gaussian}
Let $X \sim {\mathcal N}(\mu,\sigma^2)$. Then for all $t>0$
\begin{equation}\label{gaussiantail}
 \P (X \ge \mu + t) \le e^{-t^2/2\sigma^2}.
\end{equation}
\end{proposition}

\begin{proof}
Without loss of generality we consider $\mu=0$ and $\sigma=1$, as it is straightforward to extend to the general case with a change of variables. We use the moment-generating function $\lambda \mapsto \E[e^{\lambda X}]$. A simple calculation gives
\begin{equation*}
\E[e^{\lambda X}] =   \frac{1}{\sqrt{2 \pi}} \int_{-\infty}^{\infty} e^{ \lambda x- x^2/2} \, dx =
\frac{1}{\sqrt{2 \pi}}   e^{  \lambda^2/2} \int_{-\infty}^{\infty} e^{ -(x- \lambda)^2/2} \, dx = e^{ \lambda^2/2},
\end{equation*}
where we have used the fact that $ \int_{-\infty}^{\infty} e^{ -(x- \lambda)^2/2} \, dx$  is just the entire Gaussian integral shifted and therefore its value is $\sqrt{2\pi}$.
We now apply  Chernoff's bound~\eqref{basicchernoff} and obtain $\P(X \geq t) \le \E[e^{\lambda X}]  e^{-\lambda t}$. Minimizing this expression over $\lambda$ gives $\lambda=t$ and thus $\P (X \geq t) \le e^{-t^2/2}$.
\end{proof}

\subsection{Sub-gaussian random variables}

\begin{definition}\label{def:subgaussian}
A random variable $X$ with mean $\mu = \E[X]$  is called {\em sub-Gaussian} if there is a positive number $\sigma$ such that
$$\E[e^{\lambda(X-\mu)}] \le e^{\sigma^2 \lambda^2/2}, \qquad \text{for all $\lambda \in \R$.}$$
\end{definition}
Note that $\sigma^2$ is not necessarily the variance of $X$. If $X$ satisfies the above definition, we also say that $X$ is sub-Gaussian with parameter $\sigma$, or $X$ is $(\mu,\sigma)$ sub-Gaussian in case we want to emphasize $\mu$ as well. Clearly, owing to the symmetry in the definition, $-X$ is sub-Gaussian if and only if $X$ is sub-Gaussian.
Obviously, any Gaussian random variable with variance $\sigma^2$ is sub-Gaussian with parameter $\sigma$.
We refer to~\cite{vershynin2018} for other, equivalent, definitions of sub-Gaussian random variables.

Combining the moment condition in Definition~\ref{def:subgaussian} with calculations similar to those that lead us to the Gaussian tail bounds in~\ref{prop:gaussian}, yields the following concentration inequality for sub-Gaussian random variables.
\begin{proposition}[Sub-Gaussian tail bounds] \label{prop:subgaussian}
Assume $X$ is sub-Gaussian with parameter $\sigma$. Then for all $t>0$
\begin{equation}\label{subgaussiantail}
 \P (|X - \mu| \ge t) \le 2e^{-t^2/2\sigma^2} \qquad \text{for all $t \in \R$.}
\end{equation}
\end{proposition}

An important example of non-Gaussian, but sub-Gaussian random variables are {\em Rademacher random  variables}. A Rademacher random variable  $\eps$ takes on the values $\pm 1$ with equal probability and is sub-Gaussian with parameter $\sigma$. 
Indeed, any bounded random variable is sub-Gaussian (see Remark~\ref{remark:hoeffdingsubgaussian}). 

While many important random variables have a sub-Gaussian distribution, this class does not include several frequently occurring distributions with heavier tails. A classical example is the  {\em chi-squared distribution}, which we will discuss at the end of this chapter.

\subsection{Sub-exponential random variables}

Relaxing slightly the  condition on the moment-generating function in Definition~\ref{def:subgaussian}  leads to the class of {\em sub-exponential} random variables.
\begin{definition}\label{def:subexponential}
A random variable $X$ with mean $\mu = \E[X]$  is called {\em sub-exponential} if there are parameters $\nu, b$ such that
$$\E[e^{\lambda(X-\mu)}] \le e^{\nu^2 \lambda^2/2}, \qquad \text{for all $|\lambda| \le \frac{1}{b}$.}$$
\end{definition}

Clearly, a sub-Gaussian random variable is sub-exponential (set $\nu = \sigma$ and $b=0$, where $1/b$ is interpreted as $+\infty$).
However, the converse is not true. Take for example $X \sim {\mathcal N}(0,1)$ and consider the random variable $Z=X^2$. For $\lambda < \frac{1}{2}$ it holds that
\begin{equation}\label{subexp2}
\E [ e^{\lambda(Z-1)} ] = \frac{1}{\sqrt{2\pi}} \int_{-\infty}^{\infty} e^{\lambda (x^2-1)} e^{-x^2/2} dx = \frac{ e^{-\lambda} }{\sqrt{1-2\lambda}}.
\end{equation}
However, for $\lambda \ge \frac{1}{2}$ the moment-generating function does not exist, which implies that $X^2$ is not sub-Gaussian.
But $X^2$ is sub-exponential. Indeed, a brief computation shows that
$$\frac{ e^{-\lambda} }{\sqrt{1-2\lambda}} \le e^{2\lambda^2} = e^{4\lambda^2/2}, \qquad \text{for all $|\lambda| \le 1/4$,}$$
which in turn implies that $X^2$ is sub-exponential with parameters $(\nu,b)=(2,4)$.

Following  a similar procedure that yielded sub-Gaussian tail bounds produces concentration inequalities for sub-exponential random variables. However, in this case we see two different types of concentration emerging, depending on the value of $t$. In particular, we see a sub-exponential tail behavior.
\begin{proposition}[Sub-exponential tail bounds] \label{prop:subexponential}
Assume $X$, with mean $\mu$, is sub-exponential with parameters $(\nu,b)$. Then
\begin{equation}\label{subexponentialtail}
 \P (X \ge \mu + t) \le \begin{cases}
 e^{-t^2/2\nu^2} & \text{if $0 \le t \le \frac{\nu^2}{b}$,} \\
 e^{-t/2b} & \text{if $t > \frac{\nu^2}{b}$.}
\end{cases}
\end{equation}
\end{proposition}

Both the sub-Gaussian property and  the sub-exponential property is preserved under
summation for independent random variables, and the associated parameters transform in
a simple manner (see Remark~\ref{remark:hoeffdingsubgaussian}).\footnote{There is an alternative way of defining sub-gaussian and sub-exponential random variables, through the so called $\psi_p$-norm. Given $p>0$ and $X$ centered, $\|X\|_{\psi_p} = \defeq \inf\left\{C>0:\ \EE\exp\left(\frac{|X|}{C}\right)^p\leq 2\right\}$. Sub-gaussian random variables are those for which the $\psi_2$-norm is finite, while sub-exponential random variables are those for which the $\psi_1$-norm is finite (see, e.g.,~\cite{vershynin2018}).}

\section{Concentration of measure, a ``blessing of dimensionality''}

Suppose we wish to predict the outcome of an event of interest. One natural approach would be to compute
the expected value of the object. However, how can we tell how good the expected value
is to the actual outcome of the event? Without further information of how well the actual
outcome concentrates around its expectation, the expected value is of little use. We would like
to have an estimate for the probability that the actual outcome deviates from its expectation
by a certain amount. This is exactly the role that  {\em concentration inequalities} play in probability and statistics.

The  concentration  of  measure  phenomenon plays a central role in the asymptotic geometry of Banach spaces and in high dimensional probability; see the work of Vitali Milman and Ledoux's monograph~\cite{milman1986asymptotic,ledoux2001concentration}.

The celebrated law of large numbers of classical probability theory is the most well known form
of {\em concentration of measure}; it states that sums of independent random variables are,
under very mild conditions, close to their expectation with a large probability. We will see
various quantitative versions of such concentration inequalities throughout this course. Some
deal with sums of scalar random variables, others with sums of random vectors or sums of
random matrices.
Such concentration inequalities are instances of what is sometimes called {\em Blessings of dimensionality} (cf.~\cite{donoho2000high}). This expression refers to the fact that certain random fluctuations can be well
controlled in high dimensions, while it would be very complicated to make such predictive
statements in moderate dimensions.

\subsection{Large deviation inequalities}\label{s:largedeviations}

Concentration and large deviations inequalities are among the most useful tools when understanding the performance of some algorithms.
We start with two of the most fundamental results in probability. We refer to Sections 1.7 and 2.4 in~\cite{durrett2019probability} for the proofs and variations.

\begin{theorem}[Strong law of large numbers] \label{th:stronglaw}
Let $X_1,X_2,\dots $ be a sequence of i.i.d.\ random variables with mean $\mu$. Denote
$$S_n: = X_1+\dots + X_n.$$
Then, as $n \to \infty$
\begin{equation}
\frac{S_n}{n} \to \mu \qquad \text{almost surely}.
\end{equation}
\end{theorem}

The celebrated {\em central limit theorem} tells us that the limiting distribution of a sum of i.i.d.\ random variables is always Gaussian.
The best known version is probably due to Lindeberg-L\'{e}vy.

\begin{theorem}[Lindeberg-L\'{e}vy Central limit theorem] \label{th:centrallimit}
Let $X_1,X_2,\dots $ be a sequence of i.i.d.\ random variables with mean $\mu$ and variance $\sigma^2$. Denote
$$S_n: = X_1+\dots + X_n,$$
and consider the normalized random variable $Z_n$ with mean zero and variance one, given by
$$
Z_n: = \frac{ S_n - \E [S_n] }{ \sqrt{\Var{S_n}} } = \frac{1}{\sigma\sqrt{n}} \sum_{i=1}^n (X_i - \mu).
$$
Then, as $n \to \infty$
\begin{equation}
Z_n \to {\mathcal N}(0,1) \qquad \text{in distribution}.
\end{equation}
\end{theorem}

The strong law of large numbers and the central limit theorem give us qualitative statements about the behavior of a sum of i.i.d.\ random variables. In many applications it is desirable to be able to quantify how such a sum deviates around its mean. This is where concentration inequalities
come into play.

The intuitive idea is that if we have a sum of independent random variables
\[
X = X_1 + \cdots + X_n,
\]
where $X_i$ are i.i.d.\ centered random variables, then while the value of $X$ can be of order\footnote{Recall the {\em big-$\OOO$} and {\em little-o} notations:
Let $f(n)$ and $g(n)$ be real-valued functions defined for sufficiently large $n$. We say that
$
f(n) = \OOO(g(n)) \quad \text{as } n \to \infty
$
if there exist constants $C > 0$ and $n_0$ such that
$
|f(n)| \le C,|g(n)| \quad \text{for all } n \ge n_0,
$
i.e, $f$ grows at most on the order of $g$, up to a constant factor.
We say that
$
f(n) = o(g(n)) \quad \text{as } n \to \infty
$
if
$
\lim_{n \to \infty} \frac{f(n)}{g(n)} = 0,
$
i.e., $f$ grows strictly slower than $g$.}
$\OOO(n)$ it will very likely be of order $\OOO(\sqrt{n})$ (note that this is the order of its standard deviation). The inequalities that follow are ways of very precisely controlling the probability of $X$ being larger (or smaller) than $\OOO(\sqrt{n})$. While we could use, for example, Chebyshev's inequality for this, in the inequalities that follow the probabilities will be exponentially small, rather than just quadratically small, which will be crucial in many applications to come.
Moreover,  unlike the classical central limit theorem, the concentration inequalities below are {\em non-asymptotic} in the sense that they hold
for all fixed $n$ and not just for $n \to \infty$ (but the larger the $n$,  the stronger the inequalities become).

\begin{theorem}[Hoeffding's Inequality]\label{thm:hoeffding1}
 Let $X_1,X_2,\ldots,X_n$ be independent bounded random variables with $|X_i| \leq a_i$ and $\mathbb{E}[X_i]=0$. Then,
 $$\Prob\left\{\left|\sum_{i=1}^n X_i\right|\geq t\right\}\le 2\exp\left(-\displaystyle{\frac{t^2}{2 \sum_{i=1}^n a_i^2}}\right).$$
\end{theorem}

The inequality implies that fluctuations larger than $\OOO\left(\sqrt{n}\right)$ have small probability. For example, if $a_i=a$ for all $i$, setting $t=a\sqrt{2n\log n}$ yields that the probability is at most~$\frac{2}{n}$.

\begin{proof}
We prove the result for the case $|X_i| \le a$, the extension to the case $|X_i| \leq a_i$ is straightforward.
We first get a probability bound for the event $\sum_{i=1}^n X_i > t$. The proof, again, will follow from Markov. Since we want an exponentially small probability, we use a classical trick that involves exponentiating with any $\lambda>0$ and then choosing the optimal $\lambda$.
\begin{eqnarray}
\Prob\left\{ \sum_{i=1}^n X_i \geq t \right\} 
&=&  \Prob\left\{e^{\lambda\sum_{i=1}^n X_i} \geq e^{\lambda  t} \right\} \nonumber \\
&\leq & \frac{\mathbb{E}[e^{\lambda\sum_{i=1}^n X_i}]}{e^{ t\lambda}} \nonumber \\
&=& e^{- t \lambda} \prod_{i=1}^n \mathbb{E}[e^{\lambda X_i}], \label{eq:4:WhereBernsteinIsGoingToImprove}\label{eq:4:Hof}
\end{eqnarray}
where the penultimate step follows from Markov's inequality and the last equality follows from independence of the $X_i$'s.

We now use the fact that $|X_i|\leq a$ to bound $\mathbb{E}[e^{\lambda X_i}]$. Because the function $f(x) = e^{ \lambda x}$ is convex,
\begin{equation*}
e^{\lambda x} \leq \frac{a+x}{2a}e^{\lambda a} + \frac{a-x}{2a}e^{-\lambda a},
\end{equation*}
for all $x\in[-a,a]$.

Since, for all $i$, $\mathbb{E}[X_i]=0$ we get
\begin{equation*}
\mathbb{E}[e^{\lambda X_i}] \leq \EE\left[ \frac{a+X_i}{2a}e^{\lambda a} + \frac{a-X_i}{2a}e^{-\lambda a} \right] = \frac{1}{2}\left(e^{\lambda a} + e^{- \lambda a} \right) = \cosh (\lambda a)
\end{equation*}

Note that\footnote{This follows immediately from the Taylor expansions:
$\cosh(x) = \sum_{n=0}^\infty \frac{x^{2n}}{(2n)!}$,
$e^{x^2/2} = \sum_{n=0}^\infty \frac{x^{2n}}{2^n n!}$, and $(2n)! \geq 2^n n!$. }

$$\cosh(x) \leq e^{x^2/2}, \quad \text{for all } x\in \mathbb{R}$$
Hence,
\begin{equation}\label{eq:boundedimpliessubgaussian}
\mathbb{E}[e^{\lambda X_i}]  \leq e^{(\lambda a)^2/2}.
\end{equation}

Together with (\ref{eq:4:Hof}), this gives
   \begin{eqnarray*}
\Prob\left\{  \sum_{i=1}^n X_i \geq t \right\} &\leq& e^{- t \lambda} \prod_{i=1}^n e^{(\lambda a)^2/2} \\
&=& e^{- t \lambda} e^{n (\lambda a)^2/2}
\end{eqnarray*}

This inequality holds for any choice of $\lambda\geq 0$, so we choose the value of $\lambda$ that minimizes
$$\min_\lambda \left\{ n \frac{(\lambda a)^2}2 - t \lambda \right\}$$

Differentiating readily shows that the minimizer is given by
$$\lambda = \frac{t}{n a^2},$$
which satisfies $\lambda >0$. For this choice of $\lambda$,
$$n(\lambda a)^2/2 - t \lambda = \frac1{n}\left(\frac{t^2}{2a^2} - \frac{t^2}{a^2}\right) = -\frac{t^2}{2na^2}$$

Thus,
\begin{eqnarray*}
\Prob\left\{ \sum_{i=1}^n X_i \geq t \right\} &\leq& e^{-\frac{t^2}{2n a^2}}
\end{eqnarray*}

By using the same argument on $ \sum_{i=1}^n \left( - X_i\right)$, and union bounding over the two events we get,
\begin{eqnarray*}
\Prob\left\{  \left|\sum_{i=1}^n X_i\right| \geq t \right\} &\leq& 2e^{-\frac{t^2}{2na^2}}
\end{eqnarray*}
\end{proof}

\begin{remark}[Hoeffding's inequality and sub-Gaussianity]\label{remark:hoeffdingsubgaussian}
It is useful to dissect the proof of Hoeffding's inequality. Notice that the proof was carried out by showing two things: (i) that the sum of independent sub-Gaussian random variables is sub-Gaussian with parameter whose square is the sum of the squares of the sub-Gaussian parameters of the original random variables (this is effectively what is proven in~\eqref{eq:4:Hof}), and (ii) that bounded random variables are sub-Gaussian (shown in~\eqref{eq:boundedimpliessubgaussian}).
\end{remark}

\begin{remark}\label{remark:4:HvsB}
Hoeffding's inequality is suboptimal in a sense we now describe. Let us say that we have random variables $r_1,\dots, r_n$ i.i.d. distributed as
\[
r_i = \left\{ \begin{array}{rl}
  -1 & \text{ with probability } p/2   \\
 0 & \text{ with probability }  1-p  \\
  1 & \text{ with probability } p/2.
            \end{array}    \right.
\]

Then, $\EE(r_i)=0$ and $|r_i|\leq 1$ so Hoeffding's inequality gives:

\[
 \Prob\left\{\left|\sum_{i=1}^n r_i\right|>t\right\}\le 2\exp\left(-\displaystyle{\frac{t^2}{2n}}\right).
\]

Intuitively, the smaller $p$ is, the more concentrated $\left|\sum_{i=1}^n r_i\right|$ should be, however Hoeffding's inequality does not capture this behaviour.

\end{remark}

A natural way to attempt to capture this behaviour is by noting that the variance of $\sum_{i=1}^n r_i$ depends on $p$ as $\mathrm{Var}(r_i) = p$. The inequality that follows, Bernstein's inequality, uses the variance of the summands to improve over Hoeffding's inequality.

The way this is going to be achieved is by strengthening the proof above, more specifically in step~\eqref{eq:4:WhereBernsteinIsGoingToImprove} we will use the bound on the variance to get a better estimate on $\mathbb{E}[e^{\lambda X_i}]$ essentially by realizing that if $X_i$ is centered, $\EE X_i^2 = \sigma^2$, and $|X_i|\leq a$ then, for $k\geq 2$, $\EE X_i^k \leq \EE |X_i|^k \leq a^{k-2}\EE |X_i|^{2} \leq  a^{k-2} \sigma^2= \left( \frac{\sigma^2}{a^2} \right)a^k$.

\begin{theorem}[Bernstein's Inequality]\label{thm:Bernstein}

Let $X_1,X_2,\ldots,X_n$ be independent centered bounded random variables satisfying $|X_i| \leq a$ and 
 $\mathbb{E}[X_i^2]=\sigma^2$. Then,
$$\Prob \left\{\left| \sum_{i=1}^n X_i \right| \geq t \right\} \leq 2\exp\left(-\frac{t^2}{2n\sigma^2 + \frac{2}{3}at}\right).$$
\end{theorem}

\begin{remark}

Before proving Bernstein's inequality, note that for the example of Remark~\ref{remark:4:HvsB} we get
\[
 \Prob\left\{\left|\sum_{i=1}^n r_i\right|\geq t\right\}\leq 2\exp\left(-\frac{t^2}{2np + \frac{2}{3}t}\right),
\]
which exhibits a dependence on $p$ and, for small values of $p$ (and appropriate values of $t$) is considerably smaller than what Hoeffding's inequality gives.
\end{remark}

\proofb{

As before, we will prove
$$\Prob \left\{ \sum_{i=1}^n X_i \geq t \right\} \leq \exp\left(-\frac{t^2}{2n\sigma^2 + \frac{2}{3}at}\right),$$
and then union bound with the same result for $- \sum_{i=1}^n X_i$, to prove the Theorem.

For any $\lambda >0$ we have
\begin{eqnarray*}
\Prob\left\{ \sum_{i=1}^n X_i \geq t \right\} &=& \Prob\{e^{\lambda \sum X_i} \geq e^{\lambda t} \} \\
&\leq & \frac{\mathbb{E}[e^{\lambda \sum X_i}]}{e^{\lambda t}} \\
&=& e^{-\lambda t} \prod_{i=1}^n \mathbb{E}[e^{\lambda X_i}]
\end{eqnarray*}

The following calculations reveal the source of the improvement over Hoeffding's inequality.
\begin{eqnarray*}
\mathbb{E}[e^{\lambda X_i}] &=& \mathbb{E}\left[1 + \lambda X_i + \sum_{m=2}^\infty \frac{\lambda ^m X_i^m}{m!} \right] \\
&\leq & 1 + \sum_{m=2}^\infty \frac{\lambda ^m a^{m-2} \sigma^2}{m!} \\
&=& 1 + \frac{\sigma^2}{a^2} \sum_{m=2}^\infty \frac{(\lambda a)^m}{m!} \\
&=& 1+ \frac{\sigma^2}{a^2} \left(e^{\lambda a} - 1 - \lambda a \right)
\end{eqnarray*}
Therefore,
\begin{equation*}
\Prob\left\{ \sum_{i=1}^n X_i \geq t \right\} \leq e^{-\lambda t} \left[1+ \frac{\sigma^2}{a^2} \left(e^{\lambda a} - 1 - \lambda a \right) \right]^n
\end{equation*}

We will use a few simple inequalities (that can be easily proved with calculus) such as\footnote{In fact $y=1+x$ is a tangent line to the graph of $f(x)=e^x$.} $1+x \leq e^x,\  \text{for all } x\in\mathbb{R}$.

This means that,
$$ 1 + \frac{\sigma^2}{a^2}\left(e^{\lambda a} - 1 - \lambda a \right) \leq  e^{\frac{\sigma^2}{a^2}(e^{\lambda a} - 1 - \lambda a)},$$

which readily implies
$$\Prob\left\{ \sum_{i=1}^n X_i \geq t \right\} \leq e^{-\lambda t} e^{\frac{n\sigma^2}{a^2}(e^{\lambda a} - 1 - \lambda a)}.$$

As before, we try to find the value of $\lambda >0$ that minimizes
$$\min_{\lambda}\left\{  -\lambda t + \frac{n\sigma^2}{a^2}(e^{\lambda a}-1-\lambda a) \right\}$$
Differentiation gives
$$-t + \frac{n\sigma^2}{a^2} (ae^{\lambda a} - a) = 0$$
which implies that the optimal choice of $\lambda$ is given by
$$\lambda ^* = \frac{1}{a}\log\left(1 + \frac{at}{n\sigma^2}\right)$$

If we set
\begin{equation}
u = \frac{at}{n\sigma^2},
\end{equation}
then $\lambda^* = \frac{1}{a}\log(1 + u)$.

Now, the value of the minimum is given by
\[
 -\lambda^* t + \frac{n\sigma^2}{a^2} (e^{\lambda^*a} -1 -\lambda^*a) = -\frac{n\sigma^2}{a^2}\left[(1+u)\log(1+u) - u  \right].
\]

This means that
\begin{eqnarray*}
\Prob\left\{ \sum_{i=1}^n X_i \geq t \right\} &\leq& \exp\left(-\frac{n\sigma^2}{a^2}\left\{(1+u)\log(1+u)-u\right\} \right)
\end{eqnarray*}

The rest of the proof follows by noting that, for every $u>0$,
\begin{equation}
\left(1+u\right)\log(1+u) - u \geq \frac{u}{\frac2u+\frac{2}{3}},
\end{equation}
which implies:

\begin{eqnarray*}
\Prob\left\{ \sum_{i=1}^n X_i \geq t \right\} &\leq& \exp\left(-\frac{n\sigma^2}{a^2}\frac{u}{\frac{2}{u}+\frac{2}{3}} \right)\\
&=& \exp\left(-\frac{t^2}{2n\sigma^2 + \frac{2}{3}at} \right).
\end{eqnarray*}

}

We refer to~\cite{vershynin2018} for several useful variations of Bernstein's inequality.

\subsection{The geometry of the hypercube revisited} \label{ss:hypercube1}

Equipped with the probabilistic tools from the previous sections, we will give an alternative proof for Fact~\ref{fact:hypercubemostlycorners}.

\begin{theorem}\label{thm:hypercubemostlycorners}[See Fact~\ref{fact:hypercubemostlycorners}]
Let $x$ be a random, uniformly drawn, point in $C^d$. We have that $\|x\|\geq \frac{\sqrt{d}}4$ with high probability (meaning with probability $1-o(1)$).
\end{theorem}

The proof of this statement  will be based on a probabilistic argument, thereby illustrating (again) the nice and fruitful connection between geometry and probability in high dimension. Pick a point at random in $C^d$, which corresponds to $\left[-\frac12,\frac12\right]^d$.  We want to upperbound the probability that the point is also in the sphere or radius $R$.

  Let $x = (x_1, \dots, x_d) \in \R^d$ and each $x_i \in \left[-\frac12,\frac12\right]$ is chosen uniformly at random.  The event that
$x$ also lies in the sphere of radius $R$ corresponds to the inequality
$$\|x\|_2 = \sqrt{\sum_{i=1}^d x_i^2} \leq R.$$
Let $z_i = x_i^2$ and note that
  \begin{align*}
    \E [z_i] = \int_{-1/2}^{1/2} t^2 \, dt = 2\frac{(1/2)^3}{3}=\frac1{12}, 
  \end{align*}
  which implies $\E[\|x\|_2^2] = \frac{d}{12}$.
  Since the random variables $z_i - \E[z_i]=z_i-\frac1{12}$ are independent, centered, and bounded ($|z_i - \E[z_i]|\leq \max\left\{\frac{1}{12},\frac{1}{4}-\frac{1}{12}\right\}=\frac{2}{12}$), we can use Hoeffding's inequality (Theorem~\ref{thm:hoeffding1}). This gives 
  \begin{align*}
    \P(\|x\|_2^2 \leq R^2) &= \P\left(\sum_{i=1}^d x_i^2 \leq R^2\right)
    = \P\left(\sum_{i=1}^d (z_i - \E[z_i]) \leq R^2 - \frac{d}{12}\right) \\
    &\leq \exp\left[-\frac{\left(R^2-\frac{d}{12}\right)^2}{2d\left( \frac2{12} \right)^2}\right] =  \exp\left[-\frac{\left(12R^2-d\right)^2}{8d}\right]. 
  \end{align*}
Taking $R=\frac{\sqrt{d}}4$ gives $\P\left(\|x\|_2 \leq \frac{\sqrt{d}}4\right)\leq \exp\left[-c_0d\right]$ for a universal constant $c_0$ (which can be computed in this case to be $c_0=\frac{\left(1-\frac{12}{16}\right)^2}{8}=\frac1{128}$, but that's not important). 

\bigskip

Since we now have gained a better understanding of the properties of the cube in high dimensions, we can use this knowledge to our advantage. For instance, this ``pointiness'' of the hypercube (in form of the $\ell_1$-ball) turns out to very useful  in the areas of compressive sensing and sparse recovery, see Chapter~\ref{c:cs}.

\subsection{Tail bounds for the chi-squared distribution}

If $X_1,\dots,X_n$ are independent, standard normal random variables, then the sum of their squares,
$Z = \sum_{k=1}^n X_k^2$ is distributed according to the {\em chi-squared distribution} with $n$ degrees of freedom. We denote this  by
$Z \sim \chi^2(n)$. Its probability density function is
$$
\phi(t) =
\begin{cases}
 \frac{t^{\frac{n}{2}-1} e^{-\frac{n}{2}}}{{ 2^{\frac{n}{2}} \Gamma(\frac{n}{2})}},  &   t > 0. \\
0, & \text{else.}
\end{cases}
$$
Since the  $X_k^2, k=1,\dots,n$ are subexponential with parameters $(2,4)$ and independent, $Z = \sum_{k=1}^n X_k^2$ is subexponential with
parameters $(2\sqrt{n},4)$ (analogously to Remark~\ref{remark:hoeffdingsubgaussian}). Therefore, using~\eqref{subexponentialtail}, we obtain the $\chi^2$ tail bound
\begin{equation}\label{chi-tail}
\P \left(   \left| \frac{1}{n} \sum_{k=1}^n X_k^2 - 1 \right| \ge t \right) \le
\begin{cases}
2 e^{-nt^2/8} & \text{for $t \in (0,1)$.} \\
2e^{-nt/8} & \text{if $t \ge 1$.}  \\
\end{cases}
\end{equation}
A very useful variation of this bound, commonly known as ``Lemma~1 in Laurent and Massart~\cite{Laurent_Massart_TailChiSquare}'' is the following Theorem.

\begin{theorem}[~\cite{Laurent_Massart_TailChiSquare}]\label{thm:chisquare_concentration}
 Let $X_1,\dots,X_n$ be i.i.d. standard Gaussian random variables ($\NNN(0,1)$), and $a_1,\dots,a_n$ non-negative numbers and not all zero. Let
 \[
 Z = \sum_{k=1}^n a_k\left( X_k^2  - 1  \right).
 \]
 The following inequalities hold for any $x>0$:
 \begin{itemize}
 \item $\Prob\left\{ Z \geq 2\|a\|_2 \sqrt{x} + 2\|a\|_\infty x  \right\} \leq \exp(-x)$,
 \item $\Prob\left\{ Z \leq -2\|a\|_2 \sqrt{x}  \right\} \leq \exp(-x)$,
 \end{itemize}
 where $\|a\|_2^2 = \sum_{k=1}^n a_k^2$ and $\|a\|_\infty = \max_{1\leq k\leq n}|a_k|$.

\end{theorem}
Note that if $a_k=1$, for all $k$, then $Z$ is a $\chi_2$ random variable with $n$ degrees of freedom, so this theorem immediately gives a deviation inequality for $\chi_2$ random variables, readily comparable to~\eqref{chi-tail}.

\subsection{How to generate random points on a sphere}

How can we sample a point uniformly at random from $S^{d-1}$? The first approach that may come to mind is
the following method to generate random points on a unit circle. Independently generate each coordinate  uniformly at random from the interval $[-1,1]$. This yields points that are distributed uniformly at random
in a square that contains the unit ball. We could now project all points onto the unit ball. However,
the resulting distribution will not be uniform since more points fall on a line from the origin to a
vertex of the square, than fall on a line from the origin to the midpoint of an edge due to the
difference in length of the diagonal of the square to its side length.

To remedy this problem, we could
discard all points outside the unit ball and project the remaining points onto the sphere. However, if we generalize this technique  to higher dimensions, the analysis in the previous section has shown that
the ratio of the volume of $S^{d-1}(1)$ to the volume of $C^d(1)$ decreases rapidly. This makes this process not practical, since almost all the generated points will be discarded in this process and we end up with essentially no points inside (and thus, after projection, on) the sphere.

Instead we can proceed as follows: Recall that the multivariate Gaussian distribution is rotation invariant about the origin. This means that we can simply construct a vector in $\R^d$ whose entries are independently drawn from a univariate Gaussian distribution, and then normalize it to lie on the sphere. This gives a distribution of points that is uniform over the sphere.

\medskip
Having a method of generating  points uniformly at random on $S^{d-1}$ at our disposal, we can now give a probabilistic proof that points on  $S^{d-1}$ concentrate near its equator.

\begin{theorem}\label{theorem:spheremostlyinequator}
Let $x$ be a point randomly, and uniformly, drawn in the sphere in $d$ dimensions $S^{d-1}$. For any $\rho>0$ and $d$ large enough, $|x_1|\geq \sqrt{\frac{\rho}d}$ has probability $\geq 1-e^{-\frac{\rho}3}$.
\end{theorem}

Without loss of generality, we pick the first canonical basis vector $e_1$ to represents the  ``north pole'', and the intersection of the sphere with the plane $x_1=0$  forms our equator. 
We create a random vector  by sampling $(z_1,\dots,z_d) \sim {\mathcal N}(0,I_d)$ and normalize the vector to get $x = (x_1,\dots, x_d) = \frac{1}{\sqrt{\sum_{k=1}^d z_k^2}} (z_1,\dots,z_d)$. Because $x$ is on the sphere, it holds that $\|x\|_2=1$. 

We are interested in understanding the random variable $x_1^2$. Notice that $1=\sum_{k=1}^d x_k^2 = \E\left[\sum_{k=1}^d x_k^2\right] =d\,\E x_1^2$ since all coordinates are identically distributed (but not independent), this means that $\EE x_k^2=\frac1{d}$ for all coordinates $x_k$. Our goal is to upperbound the probability that $|x_1|=\frac{|z_1|}{\sqrt{\sum_{k=1}^d z_k^2}}$ is large, we will do that by union bounding the events that $z_1$ is large (in absolute value) and that $\sum_{k=1}^d z_k^2$ is small (which can be done with \eqref{chi-tail}). We will not attempt to derive the strongest possible bound, but rather to showcase an useful argument. Let $\rho>0$ and $0<\eps<1$ be parameters that we will set later (perhaps depending on $d$). We have

\begin{align*}
\P\left( |x_1|  \geq \sqrt{\frac{\rho}{d}} \right) & \leq \P\left( |z_1| \geq \sqrt{(1-\eps)\rho} \,\text{ or }\, \sum_{k=1}^d z_k^2 \leq (1-\eps)d \right) \\
& \leq \P\left( |z_1| \geq \sqrt{(1-\eps)\rho} \right) + \P\left( \frac1d\sum_{k=1}^d z_k^2 \leq 1 - \eps \right) \\
& \leq 2\exp\left[ -\frac{(1-\eps)\rho}{2} \right] + 2\exp\left[ -\frac{d\eps^2}2 \right],
\end{align*}
where in the last inequality we used the Gaussian tail bound (Proposition~\ref{prop:gaussian}) in the first term and the $\chi^2$ bound \eqref{chi-tail} in the second.

For $d$ large enough, taking $\eps$ sufficiently small we have $$\P\left( |x_1|  \geq \sqrt{\frac{\rho}{d}} \right) \leq \exp\left[ -\frac{\rho}{3} \right],$$ where the $3$ could be replaced by any real number larger than $2$.

A more general version of this phenomenon is that $k$ coordinates of a uniformly random vector in $\SSS^{d-1}$ have $\ell_2$ norm squared that concentrates well around $\frac{k}{d}$. We omit the the proof (it can be found in~\cite{Dasgupta_Gupta_JLsimpleproof}, and the ideas are similar to the ones used to obtain other inequalities in this Chapter).

\begin{lemma}[\cite{Dasgupta_Gupta_JLsimpleproof}]\label{lemma_COM}
 Let $x_1, \ldots,x_d$ be i.i.d standard Gaussian random variables and $X = (y_1,\ldots,y_d)$. Let $g:\RR^d\to\RR^k$ be the projection into the first $d$ coordinates and $y = g\left(\frac{x}{\|x\|}\right) = \frac1{\|x\|}(x_1,\ldots,x_k)$ and $L = \|Z\|^2$. It is clear that $\EE L = \frac{k}d$. In fact, $L$ is very concentrated around its mean
\begin{itemize}
 \item If $\beta<1$,
\[
 \Pr\left[ L \leq \beta \frac{k}d\right] \leq \exp\left( \frac{k}2(1-\beta +\log\beta) \right).
\]
 \item If $\beta>1$,
\[
 \Pr\left[ L \geq \beta \frac{k}d\right] \leq \exp\left( \frac{k}2(1-\beta +\log\beta) \right).
\]
\end{itemize}

\end{lemma}

\begin{remark}[Random vectors in high dimensions]
The results shown in this chapter give us valuable insight about the behavior of random vectors in high dimensions. The results showing that most of the ball is near the boundary, together with the concentration of the norm of the gaussian vector $z\sim\NNN(0,I)$ give us the crucial insight that lengths of random high dimensional vectors tend to concentrate. The theorems that establish that most of the sphere is near the equator give us a perhaps even more valuable insight: two random vectors in high dimensions are almost always almost orthogonal. Indeed, we can take one of them to be the ``north pole'' without loss of generality and (in the case in which they are uniformly drawn in the sphere) Theorem~\ref{theorem:spheremostlyinequator} states that their inner product is in the order of $\frac{1}{\sqrt{d}}$ smaller than the product of their norms, which corresponds to an angle very close to perpendicular.
\end{remark}

We will dive deeper into the topic of probability in high dimensions in Chapters~\ref{c:probability-gaussiananalysis} and~\ref{c:probability-matrixconcentration}. There are several excellent texts about different aspects of high dimensional probability, two notable examples are~\cite{vanHandel_LectureNotesProb_14} and \cite{vershynin2018}.

\section*{Exercises}\label{sec:ex2}
\addcontentsline{toc}{section}{Exercises}

\begin{myexercise}[\level\level\sep Drawing points in the unit ball]
    \label{prob:unit_ball_annulus}
    Let $d \geq 2$ and $x_1, \ldots, x_n \in \R^d$ be $n \geq d$ points sampled uniformly at random from the unit ball $B^d$. The goal of this exercise is to show that with high probability all points will be contained in an annulus of width $2 \ln n / d$ ($A$ events), and every pair of points will be nearly orthogonal ($O$ events). Namely, for any distinct numbers $i, j \in [n]$ define the events
    \begin{equation*}
        A_i = \brac{\norm{x_i} \geq 1 - \frac{2\ln n}{d}},
        \qquad\qquad
        O_{i,j} = \brac{\abs{\bran{x_i, x_j}} \leq \frac{ \sqrt{7 \ln n}}{\sqrt{d-1}}}.
    \end{equation*}
    Prove that there exists a constant $C > 0$ (independent of both $n$ and $d$) such that
    \begin{equation*}
        \P\brap{A_i \text{ for all } i, O_{i,j} \text{ for all } i \neq j} \geq 1 - \frac{C}{n}.
    \end{equation*}
\end{myexercise}

\begin{myexercise}[\level\sep Integral identity and Chebyshev's inequality]
    \label{prob:integral_identity}
    Let $X$ be a non-negative integrable random variable.
    \begin{enumerate}[(a)]
        \item Prove the integral identity:
        \begin{equation*}
            \mathbb{E}[X] = \int_{0}^{\infty} \P(X>t) \, dt.
        \end{equation*}
        
        \item Generalize it for an arbitrary integrable random variable $Y$ (not necessarily non-negative).
        
        \item Assume that $Y$ has a finite absolute moment $\E |Y|^p$ for some $p>0$. Find an integral expression for $\E |Y|^p$.

        \item Let $X$ be any random variable with finite expected value and finite $p$-th central moment $\E\abs{X-\E X}^p$, for some $p \ge 1$. Prove that for any $t > 0$:
        \begin{equation*}
            \P\brap{\abs{X-\E X} \geq t} \leq \frac{\E\abs{X-\E X}^p}{t^p}.
        \end{equation*}
    \end{enumerate}
\end{myexercise}

\begin{myexercise}[\level\sep MGF]
    \label{prob:MGF}
    Let $X$ be a real random variables. Recall the definition of the MGF:
    \begin{equation*}
        M_X(\lambda) = \E\bras{e^{\lambda X}}.
    \end{equation*}
    \begin{enumerate}[(a)]
        \item Show that $M_{X+X'}(\lambda) = M_{X}(\lambda)M_{X'}(\lambda)$, where $X'$ is another random variable independent of $X$.
        
        \item Show that $M_X(\lambda) \geq \exp\brap{\lambda\, \E[X]}$.

        \item Let $Z$ be a standard Gaussian random variable, i.e. the one having the probability density function
        \begin{equation*}
            f_Z(z) = \frac{1}{\sqrt{2\pi}} e^{-\frac{z^2}{2}}.
        \end{equation*}
        Show that $M_Z(\lambda) = e^{\lambda^2 / 2}$.
    \end{enumerate}
\end{myexercise}

\begin{myexercise}[\level\level\sep Cantelli's inequality]
    \label{prob:cantelli}
    Let $X$ be a real random variable with finite mean and finite variance $\Var(X)$. Then for any $t > 0$,
    \begin{equation*}
        \P(X - \E X \geq t) \leq \frac{\Var(X)}{\Var(X) + t^2}.
    \end{equation*}
\end{myexercise}
\begin{hint}
    Work with the shifted random variable $Y = X - \E X + u$, for arbitrary shift $u$.
\end{hint}

\begin{myexercise}[\level\level\sep Paley–Zygmund inequality]
    \label{prob:paley_zygmund}
    Let $X$ be a non-negative random variable with finite variance. Prove that for any $0 \leq \theta \leq 1$:
    \begin{equation*}
        \P\brap{X > \theta \, \E\bras{X}} \geq (1 - \theta)^2 \frac{(\E\bras{X})^2}{\E\bras{X^2}}.
    \end{equation*}
\end{myexercise}
\begin{hint}
    Note $\E\bras{X} = \E\bras{X \1_{\brac{X \leq \theta \, \E\bras{X}}}} + \E\bras{X \1_{\brac{X > \theta \, \E\bras{X}}}}$. How could one obtain the term $\E\bras{X^2}$?
\end{hint}

\begin{myexercise}[\level\sep log-MGF and Cramér transform]
    \label{prob:cramer_transform}
    Let $X$ be a centered (i.e. $\E\bras{X} = 0$) random variable. We define the logarithm of the moment generating function (log-MGF) for $\lambda \in \R$ as
    \begin{equation*}
        \psi_X(\lambda) = \log\E\bras{e^{\lambda X}}.
    \end{equation*}
    Suppose that it exists in an open neighbourhood around zero. The Cramér transform of $X$ is defined for $t \in \R$:
    \begin{equation*}
        \psi_X^*(t) = \sup_{\lambda \in \R} \brap{\lambda t - \psi_X(\lambda)}.
    \end{equation*}
    \begin{enumerate}[(a)]
        \item Prove that for any $t \geq 0$, one has
        \begin{equation*}
            \P\brap{X \geq t} \leq \exp\brap{-\psi_X^*(t)}.
        \end{equation*}
        \item Suppose that $X_1, \ldots, X_n$ are $n$ i.i.d. copies of $X$, and let $S_n = X_1 + \ldots + X_n$. Prove that for any $t \in \R$:
        \begin{equation*}
            \psi_{S_n}^*(t) = n \psi_X^*(t/n).
        \end{equation*}
    \end{enumerate}
\end{myexercise}

\begin{myexercise}[\level\level\sep Chernoff bound for polynomial vs. exponential moments]
    \label{prob:poly_vs_expo_moments}
    Let $X$ be a non-negative real random variable whose MGF is finite over $\R$. Fix $t > 0$. Show that
    \begin{equation*}
        \inf_{p\in\N \cup \brac{0}} \frac{\E\bras{X^p}}{t^p} \leq\inf_{\lambda>0} \frac{\E\bras{e^{\lambda X}}}{e^{\lambda t}}.
    \end{equation*}
\end{myexercise}

\begin{myexercise}[\level\level\sep Poisson tail bound]
    \label{prob:poisson_tail}
    Let $X$ be a Poisson random variable with parameter $\mu \in (0, \infty)$. Prove that for any $t > 0$:
    \begin{equation*}
        \P\brap{X > \mu + t} \leq \exp\brap{-\mu\, h(t / \mu)},
    \end{equation*}
    where $h(x) = \brap{1+x}\log\brap{1+x} - x$.
\end{myexercise}

\begin{myexercise}[\level\level\sep Weak bound for Komlós Conjecture]
    \label{prob:komlos_weak}
    Let $A\in \mathbb{R}^{n \times n}$ matrix whose columns $a_1, \ldots, a_n$ satisfy $\|a_i\|_{2} = 1$ for all $i \in \{1, \ldots, n\}$. Prove that there exists an absolute constant $C>0$ such that
    \begin{equation*}
        \min_{\eps \in \{-1,+1\}^n}\norm{A\eps}_{\infty} \leq C\sqrt{n}.
    \end{equation*}
\end{myexercise}
\begin{hint}
    Introduce randomness on $\eps$ and take expectation.
\end{hint}

\begin{myexercise}[\level\sep Properties of subgaussian random variables]
    \label{prob:subgaussian_properties}
    \begin{enumerate}[(a)]
        \item Show that if $X$ and $X'$ are independent mean-zero subgaussian random variables with variance parameters $\sigma$ and $\sigma'$ respectively, then $X + X'$ is a subgaussian random variable with variance parameter $\sqrt{\sigma^2 + \sigma'^2}$.

        \item Is this still true if we drop the independence assumption?

        \item Show that for any $\lambda \in \R$ it holds
        \begin{equation*}
            \Var(X) \leq \frac{2}{\lambda^2}\brap{\exp\brap{\frac{\lambda^2 \sigma^2}{2}}-1}.
        \end{equation*}

        \item Using L'Hôpital's rule, or otherwise, taking $\lambda \to 0$ deduce that
        \begin{equation*}
            \Var(X) \leq \sigma^2.
        \end{equation*}
    \end{enumerate} 
\end{myexercise}
\begin{hint}
    This inequality might be useful: $2 + t^2 \leq e^{t} + e^{-t}$.
\end{hint}

\begin{myexercise}[\level\level\sep Hoeffding's lemma]
    \label{prob:hoeffding_lemma}
    Let $X$ be a random variables such that $X \in [a, b]$ a.s., for some real numbers $a \leq b$. In the proof of Theorem~\ref{thm:hoeffding1} it was shown that $X$ is $(b-a)/2$-subgaussian. We will show here that this is optimal, and prove a slighly weaker version using symmetrization.
    \begin{enumerate}
        \item Let $X'$ be an independent copy of $X$, and set $Y = X - X'$. This is usually called the \textit{symmetrization} of $X$. Show that for any $\lambda \in \R$, the following inequality between MGFs holds:
        \begin{equation*}
            \E\bras{e^{\lambda (X - \E X)}} \leq \E\bras{e^{\lambda Y}}.
        \end{equation*}
        
        \item Show that
        \begin{equation*}
            \E\bras{e^{\lambda Y}} = \E\bras{\cosh(\lambda Y)}.
        \end{equation*}

        \item Using approximation $\cosh(x) \leq e^{x^2/2}$ (no proof needed) conclude that $X$ is $(b-a)$-subgaussian.

        \item Using Problem~\ref{prob:subgaussian_properties} (no proof needed) show that for any real numbers $a \leq b$, there is a random variable $X$ such that $X \in [a, b]$ a.s., and for any $\sigma < (b-a)/2$, $X$ is not subgaussian with parameter $\sigma$.
    \end{enumerate}
\end{myexercise}
\begin{remark}
    Note that in Definition~\ref{def:subgaussian} the subgaussianity condition is given in terms of $X-\mu$, where $\mu$ is the mean of $X$.
\end{remark}

\begin{myexercise}[\level\sep Hoeffding's inequality for subgaussians]
    \label{prob:hoeffding_subg}
    Let $X_1, \ldots, X_n$ be independent random variables such that $X_i$ is subgaussian with parameter $\sigma_i$, and let $S_n = X_1 + \ldots + X_n$ be the sum. Fix $t > 0$.
    \begin{enumerate}[(a)]
        \item Show that the Cramér transform of $S_n - \E S_n$ is lower bounded by
        \begin{equation*}
            \psi_{S_n - \E S_n}^*(t) \geq \frac{t^2}{2 \sum_{i=1}^n \sigma_i^2}.
        \end{equation*}
        \item Using Problem~\ref{prob:cramer_transform} deduce that
        \begin{equation*}
            \P\brap{\abs{S_n - \E S_n} > t} \leq 2\exp\brap{-\frac{t^2}{2\sum_{i=1}^n \sigma_i^2}}.
        \end{equation*}
    \end{enumerate}
\end{myexercise}

\begin{myexercise}[\level\level\level\sep Bernstein's inequality - bounded moments]
    \label{prob:bernstein_moments}
    Let $X_1, \dots, X_n$ be independent centered random variables such that for all $i \in [n]$ and integers $m \geq 2$, one has
    \begin{equation*}
        \E\abs{X_i}^m \leq \frac{\sigma_i^2 R^{m-2}}{2} m!,
    \end{equation*}
    where $R>0$ and $\sigma_i>0$ are constants that may depend only on distribution of $X_i$. 
    \begin{enumerate}[(a)]
        \item Prove that, for all $t > 0$, 
        \begin{equation*}
            \P\brap{\abs{ \sum_{i=1}^n X_i} > t} \leq 2 \exp \brap{ - \frac{t^2}{2(\nu^2 + Rt)}},
        \end{equation*}
        where $\nu^2 = \sum_{i=1}^n \sigma_i^2$.
        
        \item Deduce Bernstein's inequality for bounded variables (Theorem 2.17 in lecture notes). Namely, show that for independent centered random variables $X_1, \dots, X_n$ satisfying $|X_i|\le a$ a.s. and $\E X_i^2 \le \sigma^2$, it holds
        \begin{equation*}
            \P\brap{\abs{ \sum_{i=1}^n X_i} > t} \leq 2 \exp \brap{ -\frac{t^2}{2n\sigma^2 + \frac{2}{3} at}}
        \end{equation*}
        for any $t > 0$.
    \end{enumerate}
\end{myexercise}
\begin{hint}
    After applying the Chernoff bound, choose $\lambda = \frac{t}{\nu^2 + Rt}$.
\end{hint}


\chapter{Singular Value Decomposition and Principal Component Analysis}
\label{c:svd}

Data is most often represented as a matrix, even network data and graphs are often naturally represented by their adjacency matrix. For this reason Linear Algebra is one of the key tools in data analysis. Perhaps more surprising is the fact that spectral properties of matrices representing data play a crucial role in data analysis. We illustrate this importance with a discussion of the Singular Value Decomposition (SVD) and Principal Component Analysis (PCA) that are often used for data compression, denoising, and dimension reduction. Tools from random matrix theory are then utilized to better understand the performance of SVD and PCA in the high dimensional regime.

\section{Singular Value Decomposition}

The SVD is like a mathematical Swiss Army knife, with its many uses across different areas including linear algebra, operator theory, functional analysis, optimization, numerical analysis, statistics, and information theory. The SVD is also one of the most useful tools for analyzing data. We recommend~\cite{HJ90} and~\cite{Golub_MatrixComputations} as base references in linear algebra and for the SVD.

Given a matrix $A\in\RR^{m\times n}$, the SVD of $A$ is defined as
\begin{equation}\label{eq:SVD:PCAnotes}
 A = U \Sigma V^\top  ,
\end{equation}
where $U\in O(m)$, $V\in O(n)$ are orthogonal matrices (meaning that $U^\top  U = UU^\top   =\Id_{m\times m}$ and $V^\top  V = VV^\top   = \Id_{n\times n}$) and $\Sigma\in\RR^{m\times n}$ is a matrix with non-negative entries on its diagonal and otherwise zero entries.

The columns of $U$ and $V$, denoted by $u_k$ and $v_k$, are referred to as the left and right singular vectors of $A$, respectively and the diagonal elements of $\Sigma$ are the singular values of $A$, denoted
by $\sigma_1 \geq \sigma_2 \geq \sigma_r > 0$. 
In particular, $\operatorname{rank}(A)=r$, that is, the number of non-zero singular values $r$ is the rank of $A$.
Through the SVD, any matrix can be written as a sum of rank-1 matrices
\begin{equation}\label{eq:svdrank1sum}
A = \sum_{k=1}^r \sigma_k u_k v_k^\top  .
\end{equation}

Furthermore, it is straightforward to show that the left singular vectors $u_1,\dots,u_m$ are eigenvectors of the matrix $A A^\top  $, the 
right singular vectors $v_1,\dots,v_n$ are eigenvectors of the matrix $A^\top   A$, and the squared singular values $\sigma_1,\dots,\sigma_r$ are the non-zero eigenvalues of the matrices $A A^\top  $ and $A^\top   A$, respectively.

\begin{remark}
The SVD of a complex valued matrix $A \in \mathbb{C}^{m\times n}$ is given by $A = U \Sigma V^*$, where $U \in \mathbb{C}^{m\times m}$ and $V \in \mathbb{C}^{n\times n}$ are unitary matrices ($UU^* = U^* U = \Id_{m\times m}$ and $VV^* = V^*V = \Id_{n\times n}$) and $\Sigma$ is the same as in the real case (i.e., $\Sigma\in\RR^{m\times n}$ is a matrix with non-negative entries on its diagonal and otherwise zero entries). We focus our presentation on the SVD of real valued matrices as they are more common in data science, but everything can be straightforwardly extended to the complex case.
\end{remark}

\begin{remark}
 Say $m\leq n$, it is easy to see that we can also think of the SVD as having $U\in\RR^{m\times n}$ where $UU^\top  =\Id$, $\Sigma \in\RR^{n\times n}$ a diagonal matrix with non-negative entries and $V\in O(n)$. Furthermore, if $\rank A = r$, then the ``thin'' SVD of $A$ is given by $A=U\Sigma V^\top  $, where $U \in \R^{m \times r}$, $\Sigma \in \R^{r 
 \times r}$, and $V \in \R^{n \times r}$.
\end{remark}

\subsection{Matrix norms}

A very powerful modeling tool in data science is given by low rank matrices. In fact, we will devote the whole Chapter~\ref{c:lowrank} to this topic. As already suggested in the expansion~\eqref{eq:svdrank1sum} the SVD will play an important role in this, being used to provide low rank approximation of data matrices.

In order to be able to talk about low rank approximations of matrices, we need a notion of distance between matrices. Just like with vectors, the distance between matrices can be measured using a suitable norm of the difference. One popular norm is the Frobenius norm, or the Hilbert-Schmidt norm, defined as
\begin{equation}
\|A\|_F = \sqrt{\sum_{i,j} a_{ij}^2},
\end{equation}
where $a_{ij}$ are the entries of the matrix $A$. This norm is simply the Euclidean norm of a vector of length $mn$ comprised of the matrix elements. The Frobenius norm can also be expressed in terms of the singular values. To see this, first express the Frobenius norm in terms of the trace of $A^\top   A$ as
\begin{equation}\label{eq:SVD:frob}
\|A\|_F^2 = \sum_{i,j} a_{ij}^2 = \tr(A^\top   A),
\end{equation}
where we recall that the trace of a square matrix $A$ is defined as
\begin{equation}
\tr(A) = \sum_{i} a_{ii}.
\end{equation}
A particularly important property of the trace is that for any $A$ of size $m \times n$ and $B$ of size $n\times m$
\begin{equation}\label{eq:SVD:trace}
 \tr(AB) = \tr(BA).
\end{equation}
Note that this implies that, e.g., $\tr(ABC) = \tr(CAB)$, but it does not imply that, e.g., $\tr(ABC) = \tr(ACB)$ which is not true in general.
Now, plugging the SVD (\ref{eq:SVD:PCAnotes}) into (\ref{eq:SVD:frob}) gives
\begin{equation}\label{eq:SVD:MF}
\|A\|_F^2 = \tr(A^\top   A) = \tr(V \Sigma^\top   U^\top   U \Sigma V^\top  ) = \tr(\Sigma^\top   \Sigma) = \sum_{k=1}^r \sigma_k^2,
\end{equation}
where we used the orthogonality of $U$ and $V$ and the trace property (\ref{eq:SVD:trace}). We conclude that the Frobenius norm of a matrix equals the Euclidean norm of its vector of singular values.

A different way to define the size of a matrix is by viewing it as an operator and measuring by how much it can dilate vectors. For example, the operator 2-norm is defined as
\begin{equation}
\|A\|_2 = \sup_{\|x\|=1}\|A x\|.
\end{equation}
This operator norm can also be succinctly expressed in terms of the singular values. Indeed, for any $x \in \mathbb{R}^n$
\begin{equation}
Ax = \sum_{k=1}^r \sigma_k u_k (v_k^\top   x).
\end{equation}
Using the orthogonality of the left singular vectors $u_k$ we get
\begin{equation}
\|Ax \|^2 = \sum_{k=1}^r \sigma_k^2 \langle v_k , x \rangle^2 \leq \sigma_1^2 \sum_{k=1}^r \langle v_k, x \rangle^2 \leq \sigma_1^2 \sum_{k=1}^n \langle v_k, x \rangle^2 = \sigma_1^2 \|x\|^2,
\end{equation}
where the last equality is due to the orthogonality of the right singular vectors $v_k$. Moreover, we get equality by choosing $x=v_1$. We conclude that the 2-norm is simply the largest singular value
\begin{equation}
\|A\|_2 = \sigma_1.
\end{equation}

Before proceeding we briefly recall H\"older's inequality
\begin{proposition}\label{prop:holderineq}
For $p,q\in(0,\infty)$ such that $\frac1p+\frac1q=1$, and $x,y\in\RR^n$, we have
\[
\sum_{k=1}^n|x_ky_k| \leq \left( \sum_{k=1}^n|x_k|^p \right)^{\frac1p} \left( \sum_{k=1}^n|y_k|^q \right)^{\frac1q}.
\]
The following useful form is equivalent:
\begin{equation}\label{eq:HolderRS}
\left(\sum_{k=1}^n|c_k||a_k|^r|b_k|^s\right)^{r+s} \leq \left( \sum_{k=1}^n|c_k||a_k|^{r+s} \right)^{r} \left( \sum_{k=1}^n|c_k||b_k|^{r+s} \right)^{s},
\end{equation}
for non-negative $r$ and $s$.
\end{proposition}

\subsection{Existence of the SVD}
The matrix 2-norm plays an important role in proving that any matrix has an SVD.
\begin{theorem}
If $A$ is a real $m \times n$ matrix, then there
exist orthogonal matrices
$$U = [u_1, \ldots , u_m] \in O(m) \;\; \text{and} \;\; V = [v_1,\ldots, v_n] \in O(n)$$
such that
$$U^\top  AV = \operatorname{diag}(\sigma_1,\ldots,\sigma_p) \in \mathbb{R}^{m\times n},\;\; p = \min\{m,n\}$$
where $\sigma_1 \geq \sigma_2 \geq \cdots \geq \sigma_p \geq 0$.
\end{theorem}

\begin{proof}
We will use induction to prove the existence of SVD. Let $x \in \mathbb{R}^n$ and $y\in \mathbb{R}^m$ be
unit 2-norm vectors that satisfy $Ax = \sigma y$ where $\sigma = \|A\|_2$, the 2-norm of $A$. There
exist matrices $V_2 \in \mathbb{R}^{n\times (n-1)}$ and $U_2 \in \mathbb{R}^{m\times (m-1)}$ such that $V =
[x, V_2] \in \mathbb{R}^{n\times n}$ and $U = [y, U_2] \in \mathbb{R}^{m\times m}$ are orthogonal matrices (this follows from a simple result in linear algebra that given a basis of a subspace, it is possible to extend it to a basis of the whole space). Consider now the matrix $U^\top   A V$ which has the following structure
$$U^\top   A V = \left[\begin{array}{cc}
                      \sigma & w^\top   \\
                      0 & B \\
                    \end{array}
 \right] = A_1.$$
A short explanation: we can write $U^\top  AV = [y, U_2]^\top   A [x, V_2]$. Thus, the top left element in
the resulting matrix is $y^\top   A x = y^\top   \sigma y = \sigma$ (because $Ax = \sigma y$ and $\|y\|=1$).
The leftmost column underneath $\sigma$ is all 0's, as we have $U_2^\top   Ax = \sigma U_2^\top  
y = 0$ by orthogonality of $U_2$ and $y$. The topmost column to the right of $\sigma$ is some vector $w^\top  $, and the bottom right of
the matrix is represented by a smaller matrix $B\in \mathbb{R}^{(m-1)\times (n-1)}$.
Since
$$\left\|A_1 \left[
          \begin{array}{c}
            \sigma \\
            w \\
          \end{array}
        \right]
 \right\|^2 = \left\|\left[
          \begin{array}{c}
            \sigma^2 + \|w\|^2 \\
            Bw \\
          \end{array}
        \right]\right\|^2 \geq (\sigma^2 + \|w\|^2)^2$$
we have $\|A_1\|_2^2 \geq (\sigma^2 + \|w\|^2)$. However, $\|A_1\|^2  = \|A\|_2^2 = \sigma^2$, because the matrix 2-norm is invariant under orthogonal transformations. Therefore, $\|w\|^2 = 0$ and $$U^\top   A V = \left[\begin{array}{cc}
                      \sigma & 0 \\
                      0 & B \\
                    \end{array}
 \right].$$ The proof is completed by applying induction on the smaller
matrix $B$, resulting in the next largest $\sigma$ on the diagonal in each iteration, and we
eventually end up with $U^\top  AV = \operatorname{diag}(\sigma_1, \ldots, \sigma_p)$ with $\sigma_1 \geq \cdots \geq \sigma_p \geq 0$.
\end{proof}

\subsection{Low rank matrix approximation}\label{ss:lowranksvd}

A very important property of the SVD is that it provides the best low rank approximation of a matrix, when the approximation error is measured in terms of either the operator norm or the Frobenius norm. Specifically, for any $0 \leq s \leq r$ consider the rank-$s$ matrix $A_s = \sum_{k=1}^s \sigma_k u_k v_k^\top  $. Then, among all matrices of rank $s$, $A_s$ best approximates $A$ in terms of the operator norm (or Frobenius norm) error. When measuring error using the operator norm, the approximation error equals $\sigma_{s+1}$, which is the largest singular value among the remaining $r-s$ smallest singular values
\begin{equation}
\|A - A_s \|_2 = \inf_{B \in \mathbb{R}^{m\times n}, \operatorname{rank}(B) \leq s} \|A - B \|_2 = \sigma_{s+1}.
\end{equation}
A similar result holds for the Frobenius norm, with the approximation error given in terms of the remaining $r-s$ smallest singular values as
\begin{equation}
\|A - A_s \|_F = \inf_{B \in \mathbb{R}^{m\times n}, \operatorname{rank}(B) \leq s} \|A - B \|_F = \sqrt{\sum_{k=s+1}^r \sigma_k^2}.
\end{equation}
In fact, $A_s$ is the best low rank approximation for any univariate matrix norm satisfying $\|U A V\| = \| A \|$ for any $U \in O(m), V\in O(n)$, that is, norms that are invariant to multiplication by orthogonal matrices \cite{mirsky1960symmetric}.

We prove the low rank approximation property for the operator norm case\footnote{Both, Theorem~\ref{th:lowrankoperator} and Theorem~\ref{th:rank_k_Frob} are also known as Eckart-Young-Mirsky Theorem~\cite{Golub_MatrixComputations}.}.
\begin{theorem}\label{th:lowrankoperator}
Let the SVD of a matrix $A\in \mathbb{R}^{m\times n}$ be $A = U \Sigma V^\top  $. If $s < r = \operatorname{rank}(A)$ and we have the rank-$s$ matrix $A_s$,
$$A_s = \sum_{k=1}^s \sigma_k u_kv_k^\top  $$
then
$$\min_{\operatorname{rank}(B) \leq s} \|A - B \|_2 = \|A - A_s \|_2 =  \sigma_{s+1}.$$
\end{theorem}
\begin{proof}
Since $A_s$ and $A$ have the same left and right singular vectors and the same $s$ leading singular values, it follows that $U^\top   A_s V = \operatorname{diag}(\sigma_1,\ldots,\sigma_s,0,\ldots,0)$ and $U^\top   (A - A_s) V = \operatorname{diag}(0,\ldots,0,\sigma_{s+1},\ldots,\sigma_p)$, where $p=\min\{m,n\}$. This implies $\|A-A_s\|_2 = \sigma_{s+1}$, the largest remaining singular value.
Now, suppose we have a matrix $B \in \mathbb{R}^{m\times n}$ with $\operatorname{rank}(B) = s$. We can find an orthonormal
basis for the null space of $B$, i.e. orthonormal vectors $x_1, \ldots, x_{n-s}$ such that $\operatorname{null}(B) =
\operatorname{span}\{x_1,\ldots,x_{n-s}\}$. By a dimension argument, we have that
$$\operatorname{span}\{x_1,\ldots,x_{n-s}\}\cap \operatorname{span}\{v_1,\ldots,v_{s+1}\} \neq \{0 \},$$
since these are two subspaces of $\mathbb{R}^n$ of dimension $n-s$ and $s+1$. Take a unit vector $z$ in this intersection; for this vector, we know that $Bz = 0$ and $Az = \sum_{k=1}^{s+1} \sigma_k u_k (v_k^\top   z)$. Therefore,
$$\|A-B\|_2^2 \geq \|(A-B)z\|^2 = \|Az\|^2 =  \sum_{k=1}^{s+1} \sigma_k^2 (v_k^\top   z)^2 \geq \sigma_{s+1}^2 \sum_{k=1}^{s+1} (v_k^\top   z)^2 = \sigma_{s+1}^2$$
which completes the proof.
\end{proof}

For the Frobenius norm case, the low rank approximation property follows from von Neumann's trace inequality \cite{vonNeumann1937some}:
\begin{theorem}
\label{th:vonNeumann}
If $A$ and $B$ are complex-valued $n\times n$ matrices with singular values $\sigma_1(A),\ldots, \sigma_n(A)$ and $\sigma_1(B),\ldots, \sigma_n(B)$, then $$|\tr(AB)| \leq \sum_{k=1}^n \sigma_k(A) \sigma_k(B).$$
\end{theorem}
The reader is referred to \cite{mirsky1975trace} for a proof of Theorem \ref{th:vonNeumann}. We are now ready to prove that the SVD provides the best low rank approximation in the Frobenius norm.
\begin{theorem}\label{th:rank_k_Frob}
Let the SVD of a matrix $A\in \mathbb{R}^{m\times n}$ be $A = U \Sigma V^\top  $. If $s < r = \operatorname{rank}(A)$ and we have the rank-$s$ matrix $A_s$,
$$A_s = \sum_{k=1}^s \sigma_k u_k v_k^\top  $$
then
$$\min_{\operatorname{rank}(B) \leq s} \|A - B \|_F = \|A - A_s \|_F =  \sqrt{\sum_{k=s+1}^r \sigma_k^2}.$$
\end{theorem}
\begin{proof}
Let $B$ be any matrix with $\operatorname{rank}(B) = s < r = \operatorname{rank}(A)$. Notice that
$$\|A - B\|_F^2 = \|A\|_F^2 + \|B\|_F^2 - 2\tr(A^\top   B) = \sum_{k=1}^r \sigma_k^2(A) + \sum_{k=1}^s \sigma_k^2(B) - 2\tr(A^\top   B).$$
By von Neumann's trace inequality (whose statement for square matrices implies the general case of rectangular matrices)
$$\tr(A^\top   B) \leq \sum_{k=1}^s \sigma_k(A) \sigma_k(B).$$
Therefore,
\begin{eqnarray*}
\|A - B\|_F^2 &\geq&  \sum_{k=1}^r \sigma_k^2(A) + \sum_{k=1}^s \sigma_k^2(B) - 2\sum_{k=1}^s \sigma_k(A) \sigma_k(B) \\
&=& \sum_{k=s+1}^r \sigma_k^2(A) + \sum_{k=1}^s (\sigma_k(A) - \sigma_k(B))^2 \\
&\geq& \sum_{k=s+1}^r \sigma_k^2(A),
\end{eqnarray*}
which establishes the lower bound. Finally, the lower bound is clearly attained for $B = A_s$.
\end{proof}

The low rank approximation property has a wide ranging implication on data compression. The storage size of an $m\times n$ data matrix is $mn$. If that matrix is of rank $r$, then storage size reduces from $mn$ to $(n+m+1)r$ (for storing $r$ left and right singular vectors and values). For $r \ll \min\{n,m\}$ this reduction can be quite dramatic. For example, if $r=10$ and $n=m=10^6$, then storage reduces from $10^{12}$ entries to just $2\cdot 10^7$. But even if the matrix is not precisely of rank $r$, but only approximately, in the sense that $\sigma_{r+1} \ll \sigma_1$, then we are guaranteed by the above approximation results to incur only a small approximation error due to compression using the top $r$ singular vectors and values. In many cases, the singular values of large data matrices decay very quickly, motivating this type of low rank approximation which oftentimes is the only way to handle massive data sets that otherwise cannot be stored and/or manipulated efficiently. Remarkably, even treating an image as a matrix of pixel intensity values and compressing it this way yields good image compression and de-noising algorithms (as it mitigates the noise corresponding to singular values that are truncated). This is illustrated using a photo of size $1224 \times 1632$ pixels taken at Prospect Garden in Princeton University on April 28, 2015 by Amit Singer.

\begin{figure}
\begin{center}
\subcaptionbox{}{
\includegraphics[width = 0.42\textwidth, angle=270]{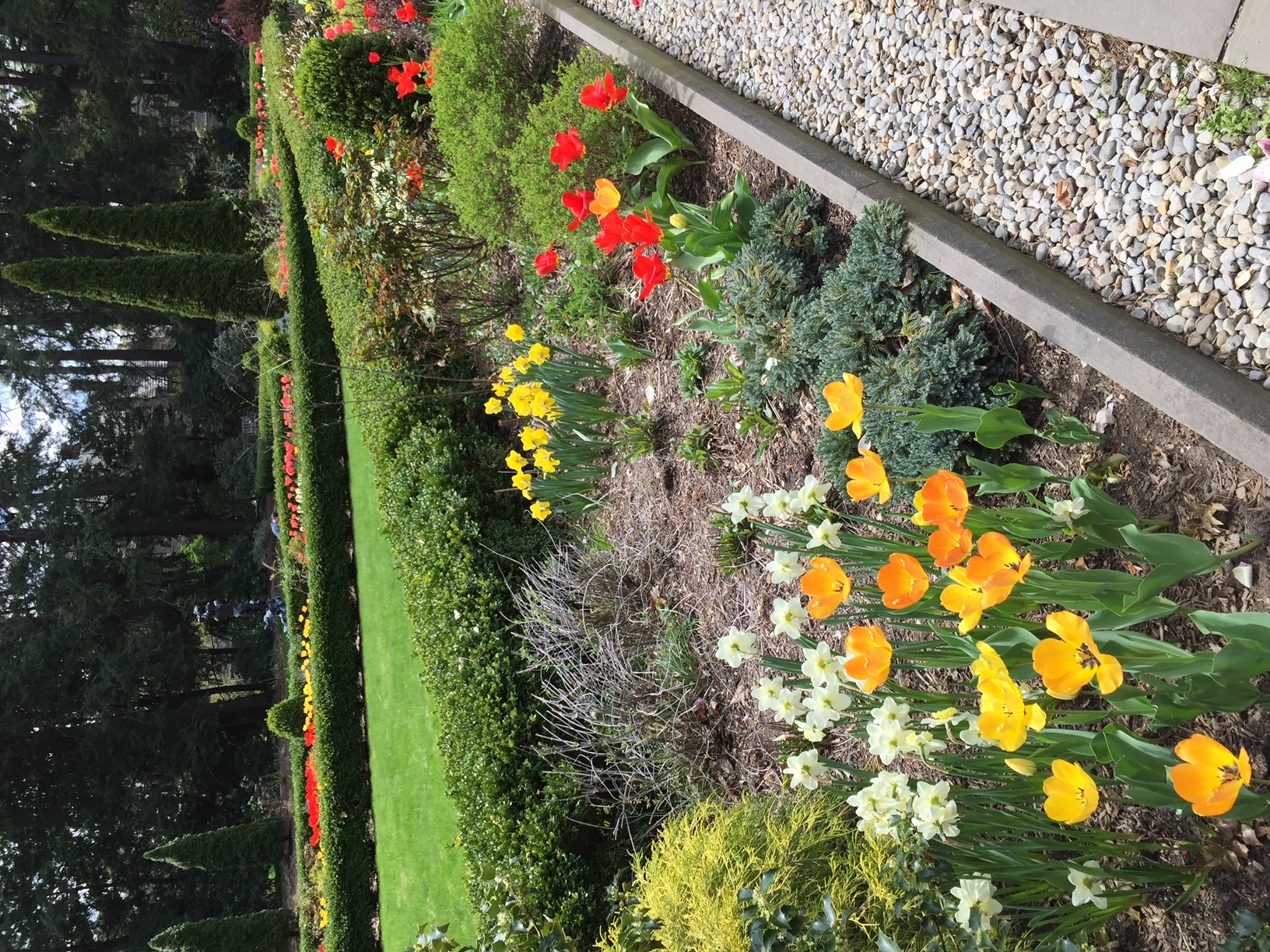}}
\subcaptionbox{}{
\includegraphics[width = 0.42\textwidth, angle=270]{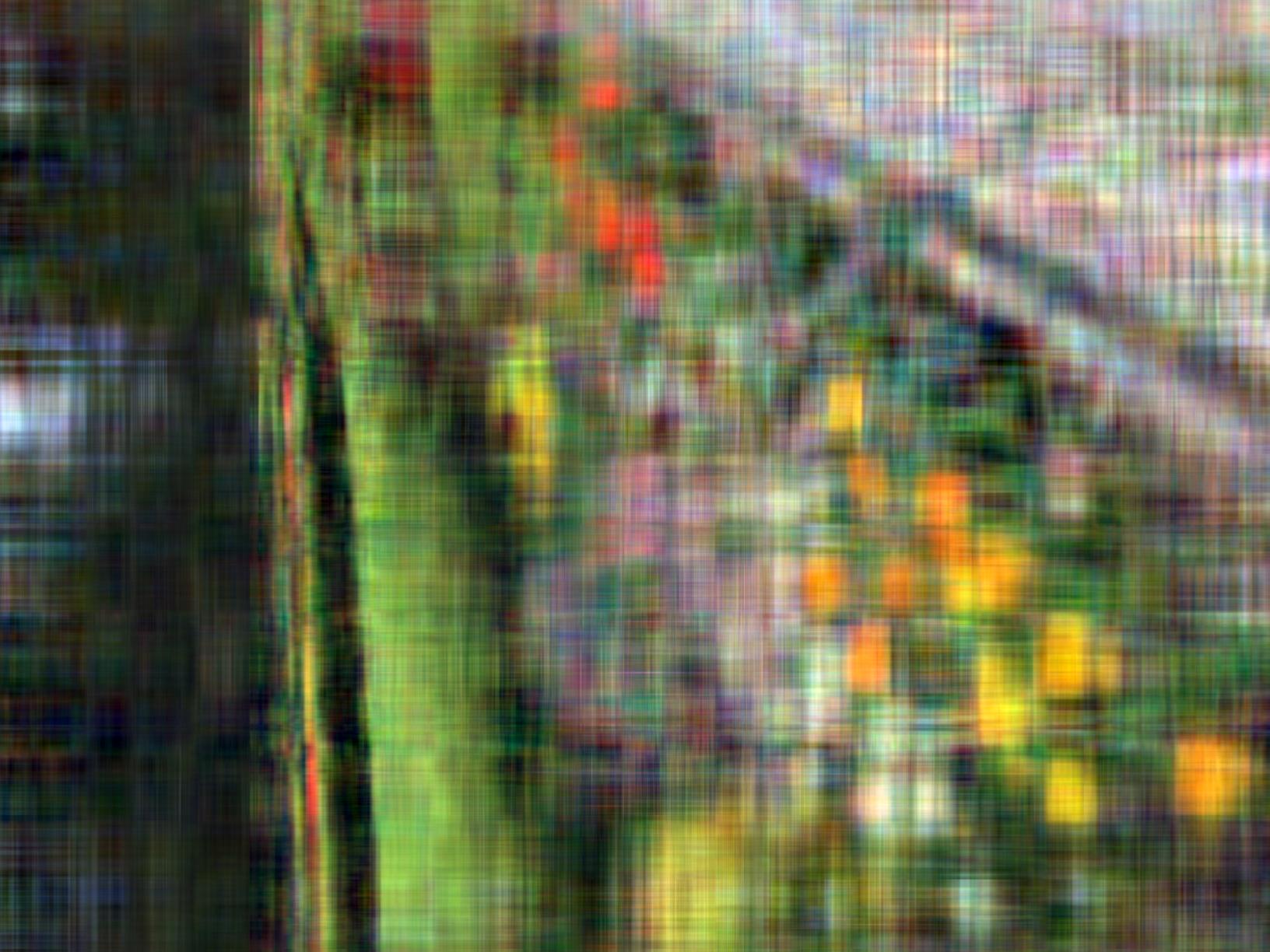}}
\subcaptionbox{}{
\includegraphics[width = 0.42\textwidth, angle=270]{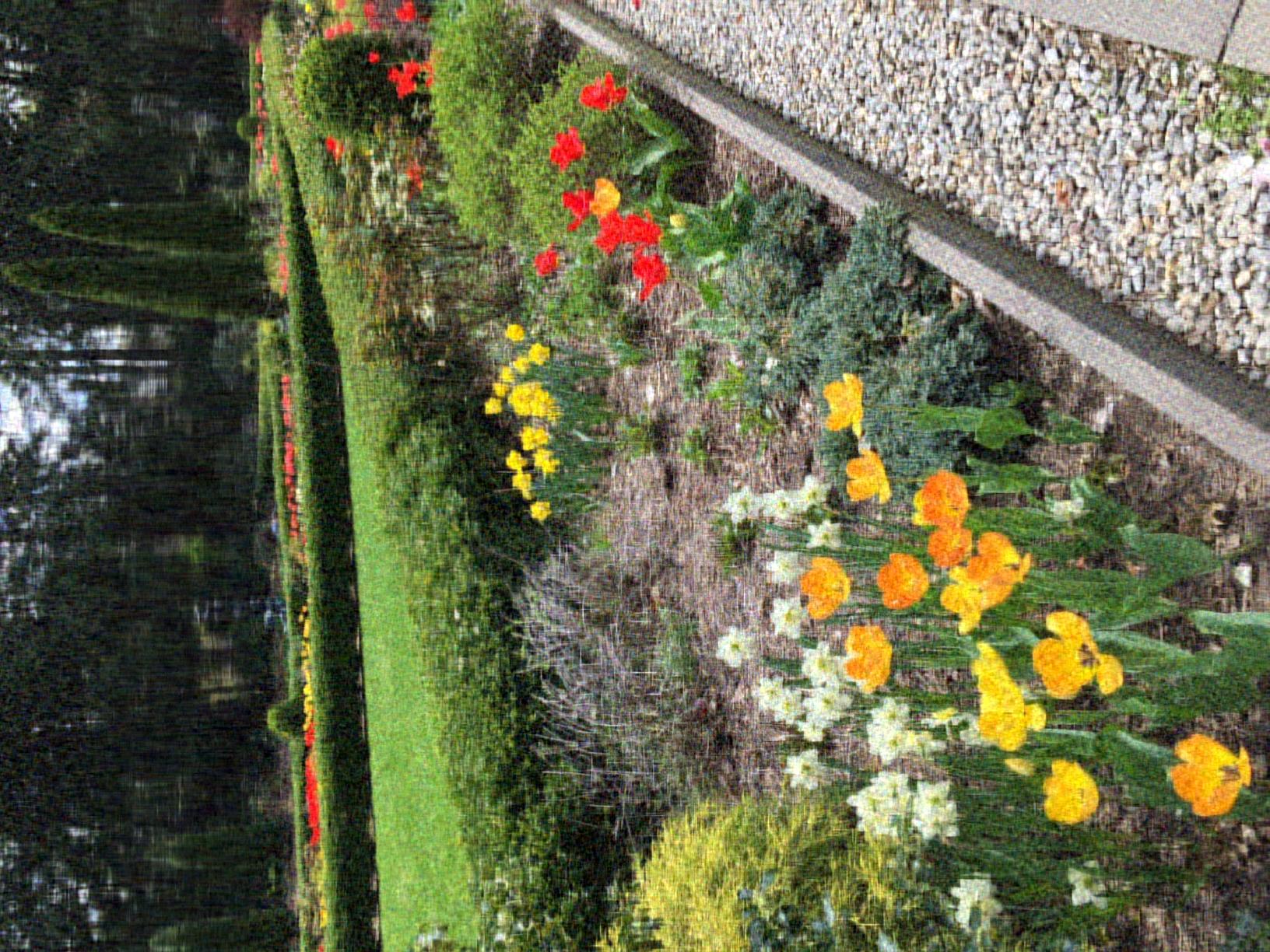}}
\caption{The Prospect Garden in Princeton University. (a) Original image, (b) rank-10 approximation, (c) rank-100 approximation.}
\label{fig:princeton}
\end{center}
\end{figure}

\begin{figure}
\begin{center}
\includegraphics[width = 0.7\textwidth]{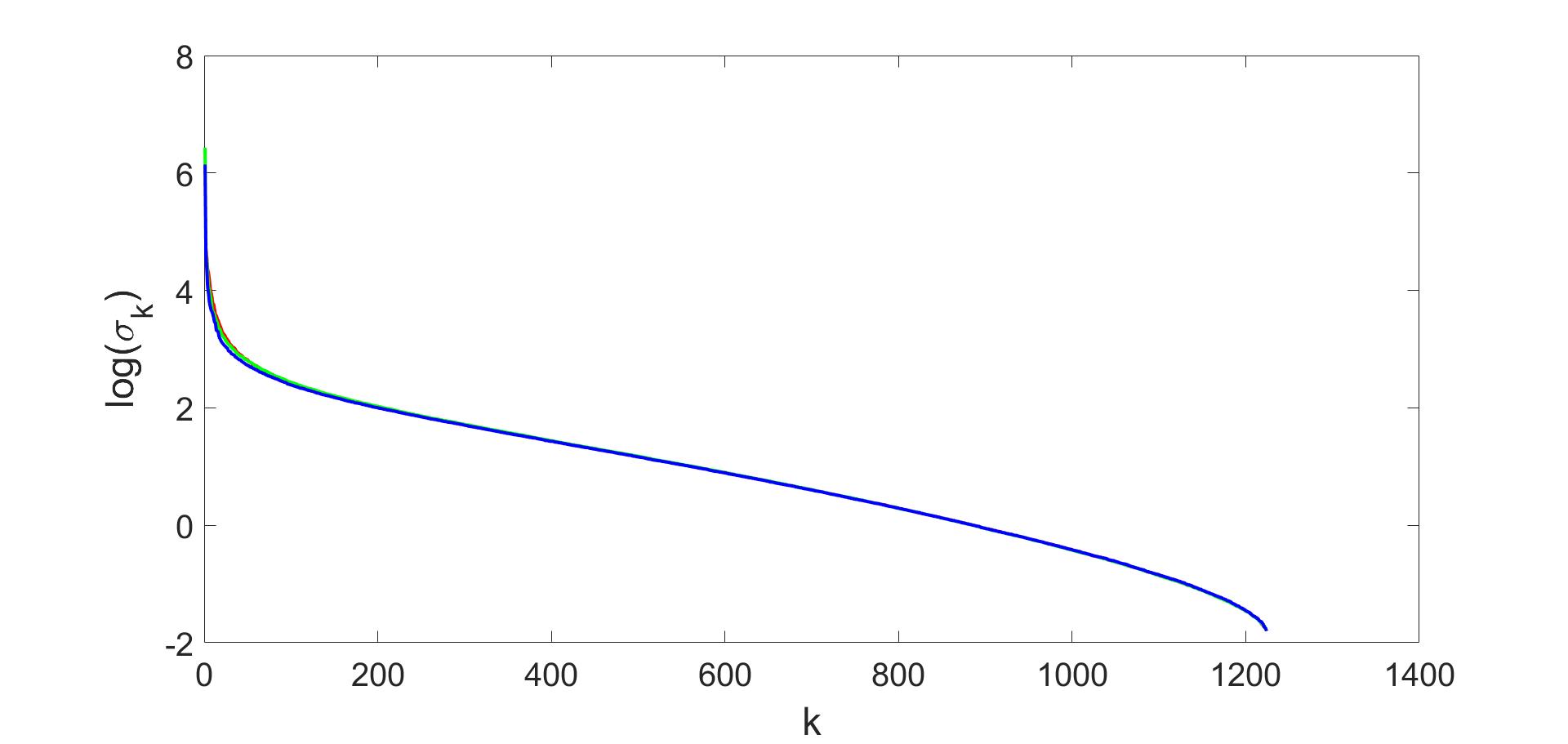}
\caption{The decay of the singular values of the three matrices, one for each RGB color channel for the photo shown in Figure~\ref{fig:princeton}.
Notice that the singular values of the three matrices (shown in red, green, and blue) are very similar. We use a logarithmic scale due to the rapid decay of singular values.}
\label{fig:rapiddecay}
\end{center}
\end{figure}

The original photo, see Figure~\ref{fig:princeton}(a),  is considered as three matrices (one for each RGB color channel). The main features of the image appear already in the rank-10 approximation, Figure~\ref{fig:princeton}(b), although many fine details are clearly missing and there are also some visible artifacts (such as striping, smearing, etc.). The rank-100 approximation, Figure~\ref{fig:princeton}(c), has exceptional resemblance with the original photo. Of course there are still differences between the two images (such as artifacts on the curbstone) but these are less visible to the naked eye on first inspection. Note that compression using $r=10$ and $r=100$ require less than $1\%$ and $10\%$, respectively, storage of the original image. The decay of the singular values of the three matrices is illustrated in Figure~\ref{fig:rapiddecay}. We use a logarithmic scale due to the rapid decay of singular values. Notice that the singular values of the three matrices are very similar.

\begin{remark}
The computational complexity of computing the SVD of a matrix of size $m\times n$ with $m \geq n$ is $\OOO(mn^2)$. This cubic scaling could be prohibitive for massive data matrices. In Chapter~\ref{s:RSVD} we discuss numerical algorithms that use randomization to efficiently compute the low rank approximation of such large matrices more efficiently.
\end{remark}

\begin{remark}
The SVD plays a critical role in sensitivity analysis of linear systems of the form $Ax=b$. In particular, the error is bounded in terms of the condition number $\kappa(A)$ which is the ratio of the largest singular value and the smallest singular value. A rigorous treatment of this important topic is beyond the scope of this book and can be found in other texts on numerical linear algebra~\cite{Golub_MatrixComputations}.
\end{remark}

\subsection{Spectral Decomposition}

If $M\in\RR^{n\times n}$ is symmetric then it admits a spectral decomposition
\[
 M = V\Lambda V^\top  ,
\]
where $V\in O(n)$ is a matrix whose columns $v_k$ are the eigenvectors of $M$ and $\Lambda$ is a diagonal matrix whose diagonal elements $\lambda_k$ are the eigenvalues of $M$. Similarly, we can write
\[
 M = \sum_{k=1}^n\lambda_k v_k v_k^\top  .
\]

When all of the eigenvalues of $M$ are non-negative we say that $M$ is positive semidefinite and write $M\succeq 0$. In that case we can write
\[
 M = \left( V \Lambda^{1/2} \right) \left( V \Lambda^{1/2} \right)^\top  .
\]
A decomposition of $M$ of the form $M=UU^\top  $ (such as the one above) is called a Cholesky decomposition (oftentimes, in a Cholesky decomposition, the matrix $U$ is required to be triangular, here we do not make this requirement).

For symmetric matrices, the operator 2-norm is also known as the spectral norm, given by
\[
 \|M\| = \max_k \left|\lambda_k(M)\right|.
\]

\subsection{Quadratic forms}

In both this and following chapters, we will be interested in solving problems of the type
\[
 \max_{\substack{V\in \RR^{n\times d} \\ V^\top  V = \Id_{d\times d}}} \tr\left(V^\top   M V \right),
\]
where $M$ is a symmetric $n\times n$ matrix.

Note that this is equivalent to
\begin{equation}\label{eq:0:evalue:vard}
 \max_{\substack{v_1,\dots,v_d \in \RR^{n} \\ v_i^\top  v_j = \delta_{ij}}} \sum_{k=1}^d v_k^\top   M v_k,
\end{equation}
where $\delta$ is the Kronecker delta ($\delta_{ij}=1$ for $i=j$ and $\delta_{ij}=0$ otherwise).

When $d=1$ this reduces to the more familiar
\begin{equation}\label{eq:0:evalue:var}
 \max_{\substack{v\in \RR^n \\ \|v\|_2 = 1}} v^\top   M v.
\end{equation}

It is easy to see (for example, using the spectral decomposition of $M$) that~\eqref{eq:0:evalue:var} is maximized by the leading eigenvector of $M$ and
\[
\max_{\substack{v\in \RR^n \\ \|v\|_2 = 1}} v^\top   M v = \lambda_{\max} (M).
\]

Furthermore~\eqref{eq:0:evalue:vard} is maximized by
taking $v_1,\dots,v_d$ to be the $d$ leading eigenvectors of $M$ and its value is simply the sum of the $d$ largest eigenvalues of $M$. This follows, for example, from a Theorem of Fan (see page 3 of~\cite{Fan_variational_evalue_UAU}).  A fortunate consequence is that the solution to~\eqref{eq:0:evalue:vard} can be computed sequentially: we can first solve for $d=1$, computing $v_1$, then update the solution for $d=2$ by simply computing $v_2$ without changing the first vector.

\begin{remark}
All of the tools and results above have natural analogues when the matrices have complex entries (and are Hermitian instead of symmetric).
\end{remark}

\subsection{SVD and the Moore-Penrose Pseudoinverse}
\label{ss:pinv}

The SVD is intimately connected to the Moore-Penrose pseudoinverse  of a matrix~\cite{ben2003generalized}. The Moore-Penrose pseudoinverse (named after mathematicians E.~H.~Moore and Roger Penrose) of $A$, denoted as $A^\dagger$, is a generalization of the matrix inverse to matrices that may not be square, or if square, may not have full rank. For an invertible matrix, the pseudoinverse is identical to the standard inverse. However, for a rectangular or rank-deficient square matrix, the pseudoinverse provides a unique solution that satisfies a set of four specific conditions known as the Penrose conditions,
 that uniquely define the pseudoinverse $A^\dagger$ for any matrix $A$. These conditions are a generalization of the properties of a standard matrix inverse. A matrix $A^\dagger$ is the pseudoinverse of $A$ if and only if it satisfies all four of the following equations:
\begin{align}
\label{eq:pinvproperties}
\begin{split}
  AA^\dagger A & =  A, \\
  A^\dagger AA^\dagger  & =  A^\dagger, \\
  (AA^{\dagger})^{T} & =  AA^\dagger, \\
  (A^{\dagger} A)^\top   & =  A^\dagger A.
  \end{split}
\end{align}
These conditions ensure that the pseudoinverse acts ``as much as possible'' as an inverse for matrices that are not invertible, including rectangular or rank-deficient square matrices.  The above four conditions are equivalent to the matrices $A A^\dagger$ and $A^\dagger A$ being orthogonal projectors onto the range of $A$ and the range of $A^\top  $ respectively.

The primary purpose of the pseudoinverse is to find  ``best-fit'' solution to a system of linear equations $Ax=b$ when a unique solution does not exist. In cases where the system has multiple solutions, the pseudoinverse provides the one with the minimum Euclidean norm. For systems with no exact solution (an overdetermined system), it provides the least-squares solution, which minimizes the error $\|Ax - b\|$. This makes the pseudoinverse a tremendously useful tool in fields like numerical mathematics, statistics, machine learning, and signal processing. We will explore this in more detail for example in Chapter~\ref{ss:ridge}.

The SVD provides a straightforward and computationally stable way to calculate $A^\dagger$. 
Given $A = U \Sigma V^\top  $, we can represent $\Sigma$ in block form as
$$\Sigma = 
\begin{bmatrix} \Sigma_r & \mathbf{0}_{r \times (n-r)} \\ \mathbf{0}_{(m-r) \times r} & \mathbf{0}_{(m-r) \times (n-r)} 
\end{bmatrix},
$$
where $\Sigma_r = \text{diag}(\sigma_1, \sigma_2, \dots, \sigma_r)$ is an $r \times r$ diagonal matrix containing the non-zero singular values. The pseudoinverse of $A$ is given by the formula. 
\begin{equation}\label{pinv}
A^\dagger = V \Sigma^\dagger U^\top  ,
\end{equation}
where the matrix $\Sigma^\dagger$ is given by
$$\Sigma^\dagger = 
\begin{bmatrix} \Sigma_r^{-1} & \mathbf{0}_{r \times (m-r)} \\ \mathbf{0}_{(n-r) \times r} & \mathbf{0}_{(n-r) \times (m-r)} 
\end{bmatrix},
$$
with $\Sigma_r^{-1} = \text{diag}\left(\frac{1}{\sigma_1}, \frac{1}{\sigma_2}, \dots, \frac{1}{\sigma_r}\right)$.

This method is particularly powerful because it works for any matrix---square or rectangular, full rank or rank-deficient---and is numerically robust, avoiding the issues with ill-conditioned matrices that can plague other inversion methods. Indeed, in the practical computation of $\Sigma^\dagger$ one usually chooses a threshold $\tau$, such that all singular values less than $\tau$ are treated as zero. 

The SVD and the pseudoinverse  play a central role in linear regression and least squares problems, see Chapter~\ref{c:linreg_ls}.

\section{Principal Component Analysis and dimension reduction}\label{s:PCA}

When faced with a high dimensional dataset, a natural approach is to attempt to reduce its dimension, either by projecting it to a lower dimensional space or by finding a better representation for the data using a small number of meaningful features. Beyond data compression and visualization, dimension reduction facilitates downstream analysis such as clustering and regression that perform significantly better in lower dimensions. We will explore a few different ways of reducing the dimension, both linearly and non-linearly.

We start with the classical Principal Component Analysis (PCA). PCA continues to be one of the most effective and simplest tools for exploratory data analysis. It dates back to a 1901 paper by Karl Pearson~\cite{Pearson_PCA_1901} and was rediscovered several times by different research communities where it is also known under different names such as the proper orthogonal decomposition (POD) and the Karhunen–Lo\`eve transform (KLT) among others.

Suppose we have $n$ data points $x_1,\dots,x_n$ in $\RR^p$, and we are interested in (linearly) projecting the data to $d< p$ dimensions. This is particularly useful if, say, one wants to visualize the data in two or three dimensions ($d=2, 3$). There are a couple of seemingly different criteria we can use to choose this projection:

\begin{enumerate}
 \item Find the $d$-dimensional affine subspace for which the projections of $x_1,\dots,x_n$ on it best approximate the original points $x_1,\dots,x_n$.

 \item Find the $d$-dimensional projection of $x_1,\dots,x_n$ that preserves as much variance of the data as possible.
\end{enumerate}

As we will see below, these two approaches are equivalent and they correspond to Principal Component Analysis.

Before proceeding, we recall a couple of simple statistical quantities associated with $x_1,\dots,x_n$, that will reappear below.

Given $x_1,\dots,x_n$ we define its sample mean as
\begin{equation}
 \mu_n = \frac1{n}\sum_{k=1}^n x_k,
\end{equation}
and its sample covariance as
\begin{equation}
 \Sigma_n = \frac1{n-1}\sum_{k=1}^n \left(x_k-\mu_n\right)\left(x_k-\mu_n\right)^\top  .
\end{equation}

\begin{remark}
If $x_1,\dots, x_n$ are independently sampled from a distribution, $\mu_n$ and $\Sigma_n$ are unbiased estimators for the mean and the covariance of the distribution, respectively.
\end{remark}

\subsubsection{PCA as the best $d$-dimensional affine fit}

We start with the first interpretation of PCA and then show that it is equivalent to the second. We are trying to approximate each $x_k$ by
\begin{equation}\label{eq:PCA:01}
 x_k \approx \mu + \sum_{i=1}^d \left(\beta_k\right)_iv_i,
\end{equation}
where $v_1,\dots,v_d$ is an orthonormal basis for the $d$-dimensional subspace, $\mu \in \RR^p$ represents the translation, and $\beta_k \in \mathbb{R}^d$ corresponds to the coefficients of $x_k$. Without loss of generality we take
\begin{equation}\label{eq:bk_sum_0}
\sum_{k=1}^n \beta_k = 0,
\end{equation}
as any joint translation of $\beta_k$ can be absorbed into $\mu$.

If we represent the subspace by $V = [v_1\cdots v_d]\in\RR^{p\times d}$ then we can rewrite~\eqref{eq:PCA:01} as
\begin{equation}\label{eq:PCA:01}
 x_k \approx \mu + V\beta_k,
\end{equation}
where $V^\top  V=\Id_{d\times d}$, because the vectors $v_i$ are orthonormal.

We will measure goodness of fit in terms of least squares and attempt to solve
\begin{equation}\label{eq:PCA:goodnessoffit0}
 \min_{\substack{\mu,\ V,\ \beta_k \\ V^\top  V=\Id}} \sum_{k=1}^n \left\| x_k - \left( \mu + V\beta_k \right)  \right\|_2^2
\end{equation}

We start by optimizing for $\mu$. It is easy to see that the first order condition for $\mu$ corresponds to
\[
 \nabla_\mu \sum_{k=1}^n \left\| x_k - \left( \mu + V\beta_k \right)  \right\|_2^2 = 0 \Longleftrightarrow \sum_{k=1}^n \left( x_k - \left( \mu + V\beta_k \right) \right) = 0.
\]
Thus, the optimal value $\mu^\ast$ of $\mu$ satisfies
\[
\left( \sum_{k=1}^n x_k \right) - n\mu^\ast - V\left( \sum_{k=1}^n \beta_k \right)=0.
\]
Since we assumed in~\eqref{eq:bk_sum_0} that $\sum_{k=1}^n \beta_k = 0$, we have that the optimal $\mu$ is given by
\[
 \mu^\ast = \frac1n \sum_{k=1}^n x_k = \mu_n,
\]
the sample mean.

We can then proceed to finding the solution for \eqref{eq:PCA:goodnessoffit0} by solving
\begin{equation}\label{eq:PCA:goodnessoffit1}
 \min_{\substack{V,\ \beta_k \\ V^\top  V=\Id}} \sum_{k=1}^n \left\| x_k - \mu_n -  V\beta_k   \right\|_2^2.
\end{equation}

Let us proceed by optimizing for $\beta_k$. The problem almost fully decouples in each $k$, the only constraint coupling them being~\eqref{eq:bk_sum_0}. We will ignore this constraint, solve the decoupled problems, and verify that it is automatically satisfied. Hence we focus on, for each $k$,
\begin{equation}\label{eq:PCA:goodnessoffit12}
 \min_{\beta_k} \left\| x_k - \mu_n - V\beta_k   \right\|_2^2 =
 \min_{\beta_k} \Big\| x_k - \mu_n - \sum_{i=1}^d \left(\beta_k \right)_i v_i   \Big\|_2^2.
\end{equation}
Since $v_1,\dots,v_d$ are orthonormal, it is easy to see that the solution is given by $\left(\beta_k^\ast \right)_i = v_i^\top  \left( x_k - \mu_n\right)$ which can be succinctly written as $\beta_k^* = V^\top  \left(x_k - \mu_n\right)$, which satisfies~\eqref{eq:bk_sum_0}. Thus, \eqref{eq:PCA:goodnessoffit1} is equivalent to
\begin{equation}\label{eq:PCA:goodnessoffit13}
 \min_{V^\top  V=\Id} \sum_{k=1}^n \left\| \left( x_k - \mu_n \right) -  VV^\top  \left(x_k - \mu_n\right)   \right\|_2^2.
\end{equation}

Note that
\begin{eqnarray*}
 \left\| \left( x_k - \mu_n \right) -  VV^\top  \left(x_k - \mu_n\right)   \right\|_2^2 & = & \left( x_k - \mu_n \right)^\top  \left( x_k - \mu_n \right) \\ & &  - 2\left( x_k - \mu_n \right)^\top  VV^\top  \left(x_k - \mu_n\right) \\ & &  +\left(x_k - \mu_n\right)^\top  V\left(V^\top  V\right)V^\top  \left(x_k - \mu_n\right) \\
 & = & \left( x_k - \mu_n \right)^\top  \left( x_k - \mu_n \right) \\ & &  - \left( x_k - \mu_n \right)^\top  VV^\top  \left(x_k - \mu_n\right).
\end{eqnarray*}

Since $\left( x_k - \mu_n \right)^\top  \left( x_k - \mu_n \right)$ does not depend on $V$, minimizing \eqref{eq:PCA:goodnessoffit13} is equivalent to
\begin{equation}\label{eq:PCA:goodnessoffit13_max}
 \max_{V^\top  V=\Id} \sum_{k=1}^n \left( x_k - \mu_n \right)^\top  VV^\top  \left(x_k - \mu_n\right).
\end{equation}

A few algebraic manipulations using properties of the trace yields:
\begin{eqnarray*}
 \sum_{k=1}^n \left( x_k - \mu_n \right)^\top  VV^\top  \left(x_k - \mu_n\right) & = &\sum_{k=1}^n \tr\left[  \left( x_k - \mu_n \right)^\top  VV^\top  \left(x_k - \mu_n\right)  \right] \\
 & = & \sum_{k=1}^n\tr\left[  V^\top  \left(x_k - \mu_n\right)\left( x_k - \mu_n \right)^\top  V  \right] \\
 & = & \tr\left[  V^\top  \sum_{k=1}^n\left(x_k - \mu_n\right)\left( x_k - \mu_n \right)^\top  V  \right] \\
 & = & (n-1)\tr\left[  V^\top   \Sigma_n V  \right].
\end{eqnarray*}
This means that the solution to \eqref{eq:PCA:goodnessoffit13_max} is given by the solution of
\begin{equation}\label{eq:PCA:goodnessoffit13_min}
 \max_{V^\top  V=\Id} \tr\left[  V^\top   \Sigma_n V  \right].
\end{equation}
As we saw above (recall \eqref{eq:0:evalue:vard}) the solution is given by $V=\left[v_1,\cdots,v_d\right]$ where $v_1,\dots,v_d$ correspond to the $d$ leading eigenvectors of $\Sigma_n$.

\subsubsection{PCA as the $d$-dimensional projection that preserves the most variance}

We now show that the alternative interpretation of PCA, of finding the $d$-dimensional projection of $x_1,\dots,x_n$ that preserves the most variance, also arrives to the optimization problem \eqref{eq:PCA:goodnessoffit13_min}.
We aim to find an orthonormal basis $v_1,\dots,v_d$ (organized as $V=\left[v_1,\dots,v_d\right]$ with $V^\top  V = \Id_{d\times d}$) of a $d$-dimensional space such that the projection of $x_1,\dots,x_n$ onto this subspace has the most variance. Equivalently we can ask for the points
\[
 \left\{   \left[\begin{array}{c} v_1^\top  x_k  \\ \vdots \\ v_d^\top  x_k\end{array}\right]   \right\}_{k=1}^n,
\]
to have as much variance as possible. Hence, we are interested in solving
\begin{equation}\label{eq:PCA:maxvariance_min}
\max_{V^\top  V=\Id} \sum_{k=1}^n\left\|  V^\top  x_k - \frac1n \sum_{r=1}^n V^\top  x_r   \right\|^2.
\end{equation}
Note that
\[
\sum_{k=1}^n\left\|  V^\top  x_k- \frac1n \sum_{r=1}^n V^\top  x_r   \right\|^2 = \sum_{k=1}^n\left\|  V^\top  \left(x_k-\mu_n\right)   \right\|^2 = (n-1) \tr\left(V^\top  \Sigma_n V\right),
\]
showing that \eqref{eq:PCA:maxvariance_min} is equivalent to \eqref{eq:PCA:goodnessoffit13_min} and that the two interpretations of PCA are indeed equivalent.

\subsubsection{Finding the Principal Components}

When given a dataset $x_1,\dots,x_n\in\RR^p$, in order to compute the Principal Components one needs to compute the leading eigenvectors of
\[
\Sigma_n = \frac1{n-1}\sum_{k=1}^n \left(x_k-\mu_n\right)\left(x_k-\mu_n\right)^\top  .
\]
A na\"ive way of doing this is to construct $\Sigma_n$ (which takes $\OOO(np^2)$ work) and then finding its spectral decomposition (which takes $\OOO(p^3)$ work). This means that the computational complexity of this procedure is $\OOO\left( \max \left\{ np^2, p^3  \right\} \right)$ (e.g.\ see~\cite{HJ90} or~\cite{Golub_MatrixComputations}).

An alternative is to use the Singular Value Decomposition~\eqref{eq:SVD:PCAnotes}. Let $X = \left[x_1 \cdots x_n  \right]$, and recall that
\[
\Sigma_n = \frac{1}{n-1} \left( X - \mu_n\1^\top   \right) \left( X - \mu_n\1^\top   \right)^\top  .
\]

Let us take the SVD of $\frac{1}{\sqrt{n-1}}\left(X - \mu_n\1^\top  \right) = U_L D U_R^\top  $ with $U_L \in O(p)$, $D$ diagonal, and $U_R^\top  U_R = \Id$. Then,
\[
\Sigma_n = \frac{1}{n-1} \left( X - \mu_n\1^\top   \right) \left( X - \mu_n\1^\top   \right)^\top   = U_L D U_R^\top   U_R D U_L^\top   = U_L D^2 U_L^\top  ,
\]
meaning that $U_L$ corresponds to the eigenvectors of $\Sigma_n$. Computing the SVD of $X - \mu_n\1^\top  $ takes $\OOO(\min \{n^2p,p^2n\})$ work but if one is interested in simply computing the top $d$ eigenvectors then this computational cost reduces to $\OOO(dnp)$. This can be further improved with randomized algorithms. In Chapter~\ref{s:RSVD} we will discuss randomized algorithms that compute an approximate solution in $\OOO\left( pn\log d +(p+n)d^2  \right)$ time, see also~\cite{HMT11,Rokhlin_Szlam_Tygert_randPCA,Musco_Musco_BlockLanczos}.

Numerical stability is another important reason why computing the principal components using the SVD is preferable. Since the eigenvalues of $\Sigma_n$ are
proportional to the squares of the singular values of $X-\mu_n \1^\top  $, problems arise when the ratio of singular
values exceeds $10^8$, causing the ratio of the corresponding eigenvalues of $\Sigma_n$ to be larger than $10^{16}$. In this case, the smaller eigenvalue would be rounded to zero (due to machine precision), which is certainly
not desirable.

\subsubsection{Example: The eigenspace of handwritten digits (MNIST)}

Every aspiring data scientist will at some point experiment with the MNIST dataset, a collection of handwritten digits created by researchers Burges, Cortes, and LeCun from data by the National Institute of Standards and Technology (NIST)~\cite{lecun1998mnist}. The most common version of MNIST has 60,000 training images and 10,000 test images, each image containing $28 \times 28$ grayscale pixels. MNIST has become a highly influential, frequently used benchmark in the data science/machine learning community\footnote{See~\cite{hardt2022patterns} for more information about MNIST. The book also contains a lot of useful information about other benchmark datasets.}.

We will apply PCA to the MNIST dataset to analyze whether a high-dimensional space (784 dimensions in this case) can be compressed into a low-dimensional {\em latent space} that is still able to capture the ``essence'' of these handwritten digits. To process the data via PCA, we first flatten each image into a vector $x\in\R^{784}$. The resulting data matrix $X$ has a size of $60000\times784$. In this matrix, each row is a single handwritten digit (an observation), while each column represents the intensity of a specific pixel location (a feature).

\begin{figure}[h]
\begin{center}
\includegraphics[width = 0.9\textwidth]{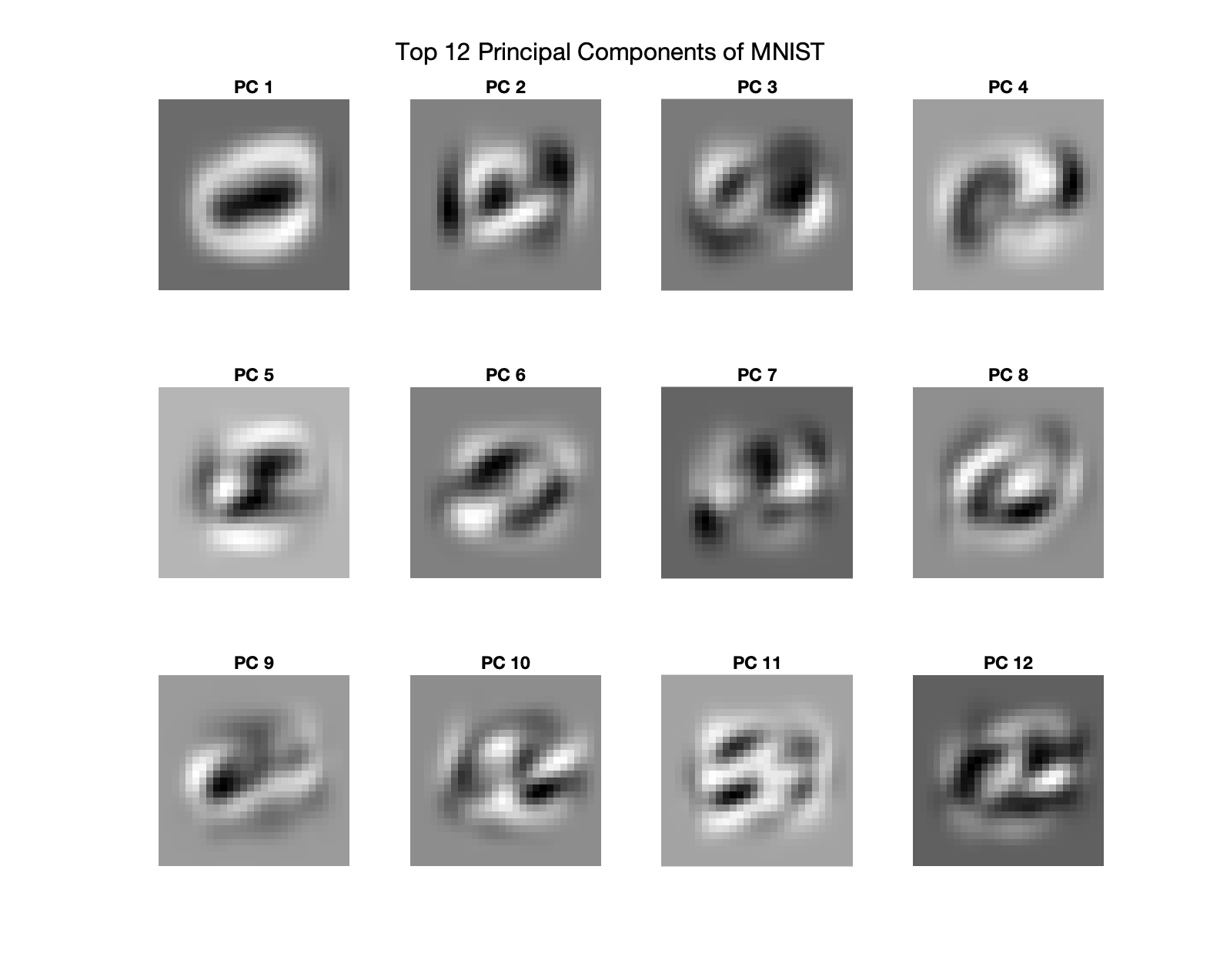}
\caption{The first twelve principal components of the MNIST images.}
\label{fig:pcamnist}
\end{center}
\end{figure}

Before performing PCA, we compute the mean digit $\bar{x}$, which represents the ``average'' handwriting across the entire dataset. By subtracting this mean from every row $X_{\text{centered}} =X - \bar{x}$, we shift the coordinate system so that the origin lies at the center of the data cloud. This ensures that the first principal component truly represents the direction of maximum variance. When we compute the SVD of $X_{\text{centered}}$, the right singular vectors (the columns of V) are the Principal Components. We can reshape these $784 \times 1$ vectors back into $28 \times 28$ images, sometimes called {\em eigen-digits}, to visualize what the algorithm has ``learned.'' The first twelve eigen-digits are depicted in Figure~\ref{fig:pcamnist}.

\subsubsection{Which $d$ should we pick?}

\begin{figure}[h]
 \begin{center}
 \subcaptionbox{}{\includegraphics[width = 55mm,height=38.6mm]{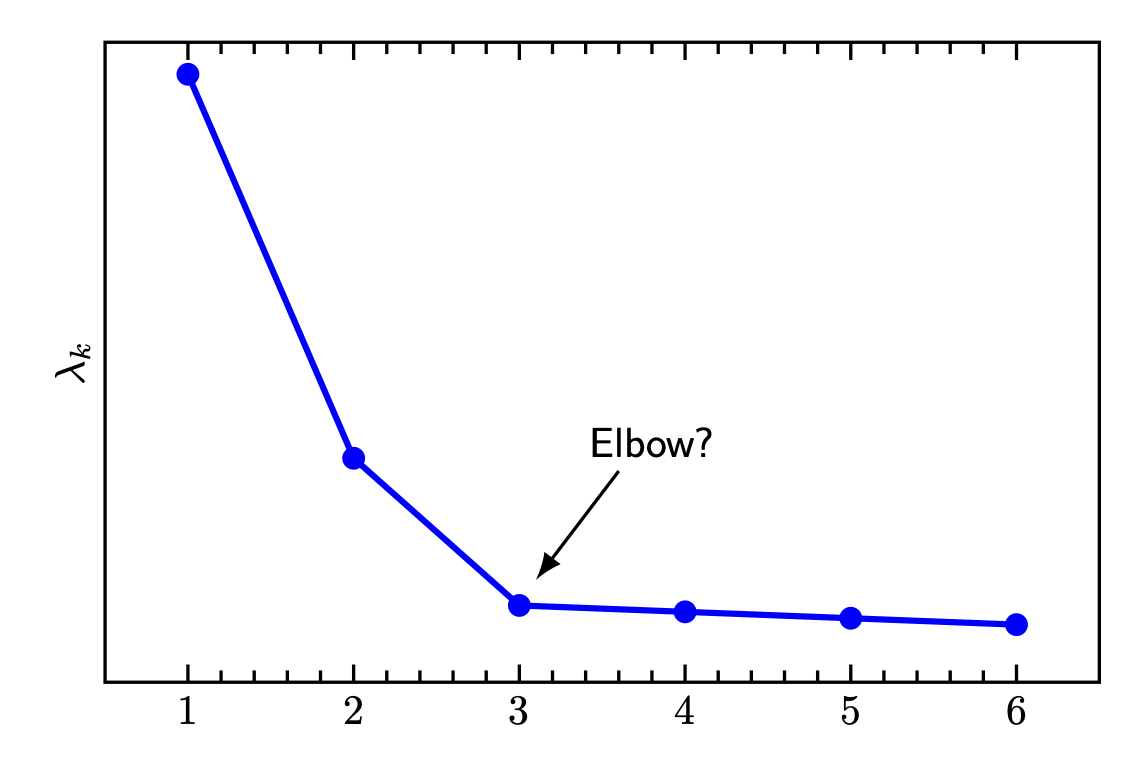}}
 \quad 
 \subcaptionbox{}{ \includegraphics[width = 57mm,height=39mm]{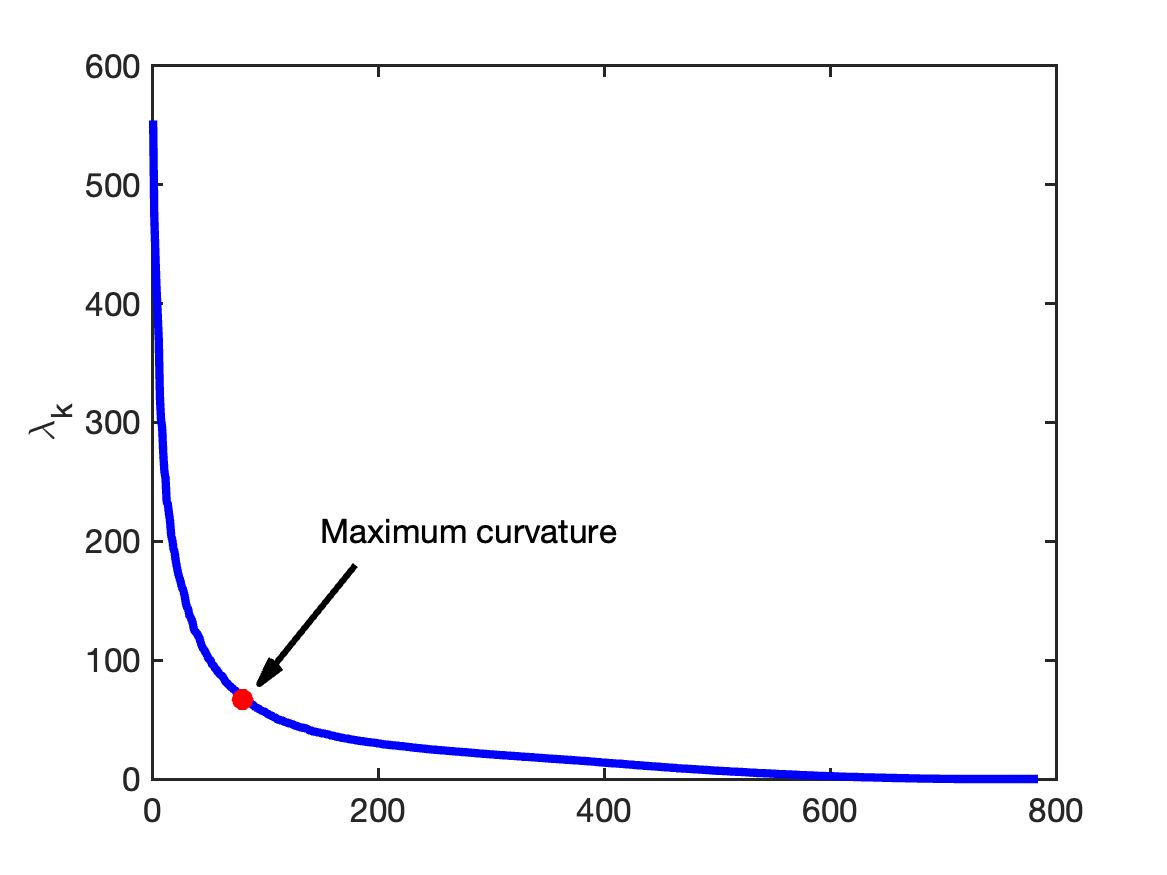}}
    \caption{(a) A plot of the values of the ordered eigenvalues (depicted is an idealized scenario) helps to identify a reasonable choice of the number of principal components. (b) The ordered eigenvalues for the MNIST example.}
    \label{fig:elbow}   
\end{center}
\end{figure}

Given a dataset, if the objective is to visualize it then picking $d=2$ or $d=3$ might make the most sense. However, PCA is useful for many other purposes, for example: \begin{enumerate}
\item Denoising: often times the data belongs to a lower dimensional space but is corrupted by high dimensional noise. In such cases, PCA helps reduce the noise while keeping the signal.
\item Downstream analysis: One may be interested in running an algorithm (clustering, regression, etc.) that would be too computationally expensive or too statistically insignificant to run in high dimensions. Dimension reduction using PCA may help there.
\end{enumerate}
In these applications (and many others) it is not clear how to pick $d$. A fairly popular heuristic is to try to choose the cut-off at a component that has significantly more variance than the one immediately after. Since the total variance is $\tr(\Sigma_n) = \sum_{k=1}^p \lambda_k$, the proportion of variance in the $i$'th component is nothing but $\frac{\lambda_i}{\tr(\Sigma_n)}$. A plot of the values of the ordered eigenvalues, also known as a scree plot, helps to identify a reasonable choice of $d$. It is common to then try to identify an ``elbow'' on the scree plot to choose the cut-off, as illustrated in Figure~\ref{fig:elbow}(a). Figure~\ref{fig:elbow}(b) shows the scree plot for the MNIST example, where the elbow is less well defined, as often happens in practice. In this case, the point of maximum curvature of the graph can serve as a surrogate for the elbow.

\begin{figure}[h]
\begin{center}
\includegraphics[width = 0.7\textwidth]{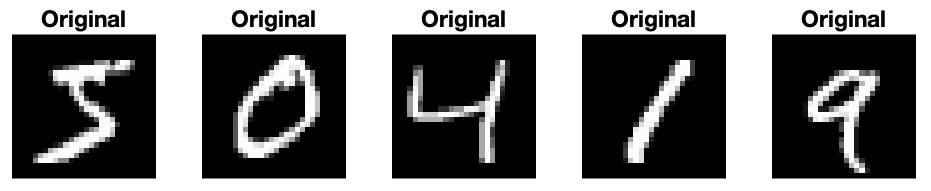}
\includegraphics[width = 0.7\textwidth]{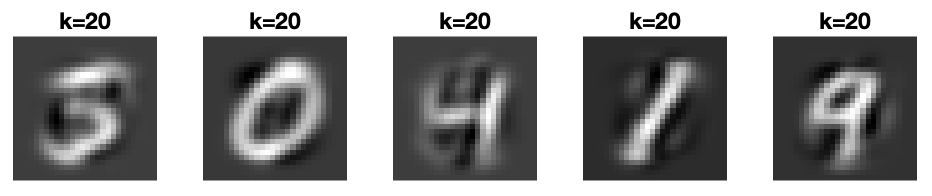}
\includegraphics[width = 0.7\textwidth]{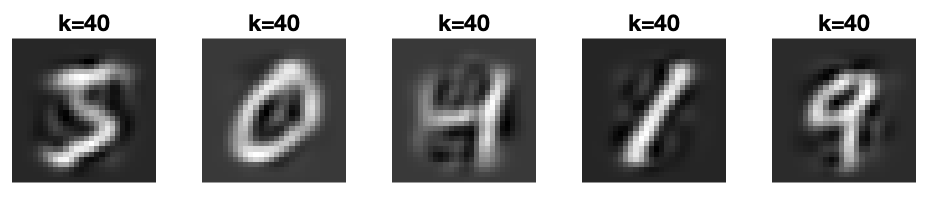}
\includegraphics[width = 0.7\textwidth]{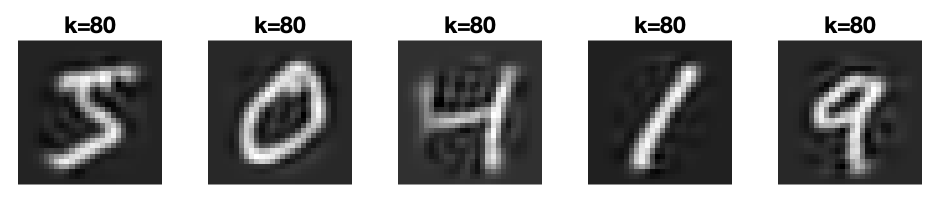}
\includegraphics[width = 0.7\textwidth]{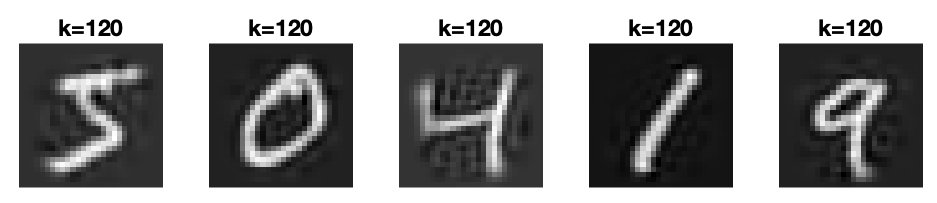}
\includegraphics[width = 0.7\textwidth]{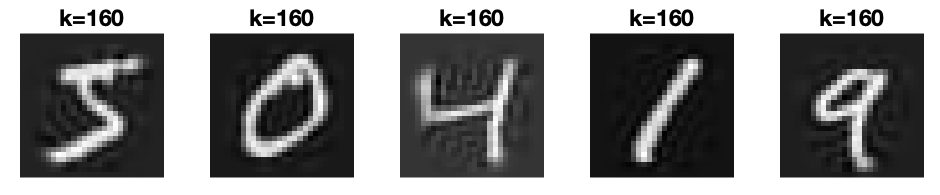}
\caption{MNIST approximation via PCA with different cutoffs.}
\label{fig:pcamnist1}
\end{center}
\end{figure}

Figure~\ref{fig:pcamnist1} shows the choice of different values of $k=d$ for the PCA approximation of MNIST, illustrated via some typical example digits.

In the next section we will look into Random Matrix Theory to better understand the behavior of the eigenvalues of $\Sigma_n$ and gain insight into choosing cut-off values.

\section{PCA in high dimensions and Mar\v{c}enko-Pastur law}

Let us assume that the data points $x_1,\dots,x_n\in\RR^p$ are independent draws of a zero mean Gaussian random variable $g\sim \NNN(0,\Sigma)$ with some covariance matrix $\Sigma \in \RR^{p\times p}$. In this case, when we use PCA we are hoping to find a low dimensional structure in the distribution, which should correspond to the large eigenvalues of $\Sigma$ (and their corresponding eigenvectors). For that reason, and since PCA depends on the spectral properties of $\Sigma_n$, we would like to understand whether the spectral properties of the sample covariance matrix $\Sigma_n$ (eigenvalues and eigenvectors) are close to those of $\Sigma$, also known as the population covariance.

Since $\EE\Sigma_n = \Sigma$, if $p$ is fixed and $n\to \infty$ the law of large numbers guarantees that indeed $\Sigma_n\to \Sigma$. However, in many modern applications it is not uncommon to have $p$ of the same order as $n$ (or, sometimes, even larger). For example, if our dataset is composed of images then $n$ is the number of images and $p$ the number of pixels per image; it is conceivable that the number of pixels and the number of images are comparable. Unfortunately, in that case, it is no longer clear that $\Sigma_n\to \Sigma$. Dealing with this type of difficulties is an important goal of high dimensional statistics.

For simplicity we will try to understand the spectral properties of
\[
S_n = \frac1n X X^\top  ,
\]
where $x_1,\dots,x_n$ are the columns of $X$. Since $x\sim \NNN(0,\Sigma)$ we know that $\mu_n \to 0$ (and, clearly, $\frac{n}{n-1}\to 1$), hence the spectral properties of $S_n$ will be essentially the same as $\Sigma_n$.\footnote{In this case, $S_n$ is actually the maximum likelihood estimator for $\Sigma$.}

\begin{figure}[h]
 \begin{center}
\includegraphics[width = 0.49\textwidth]{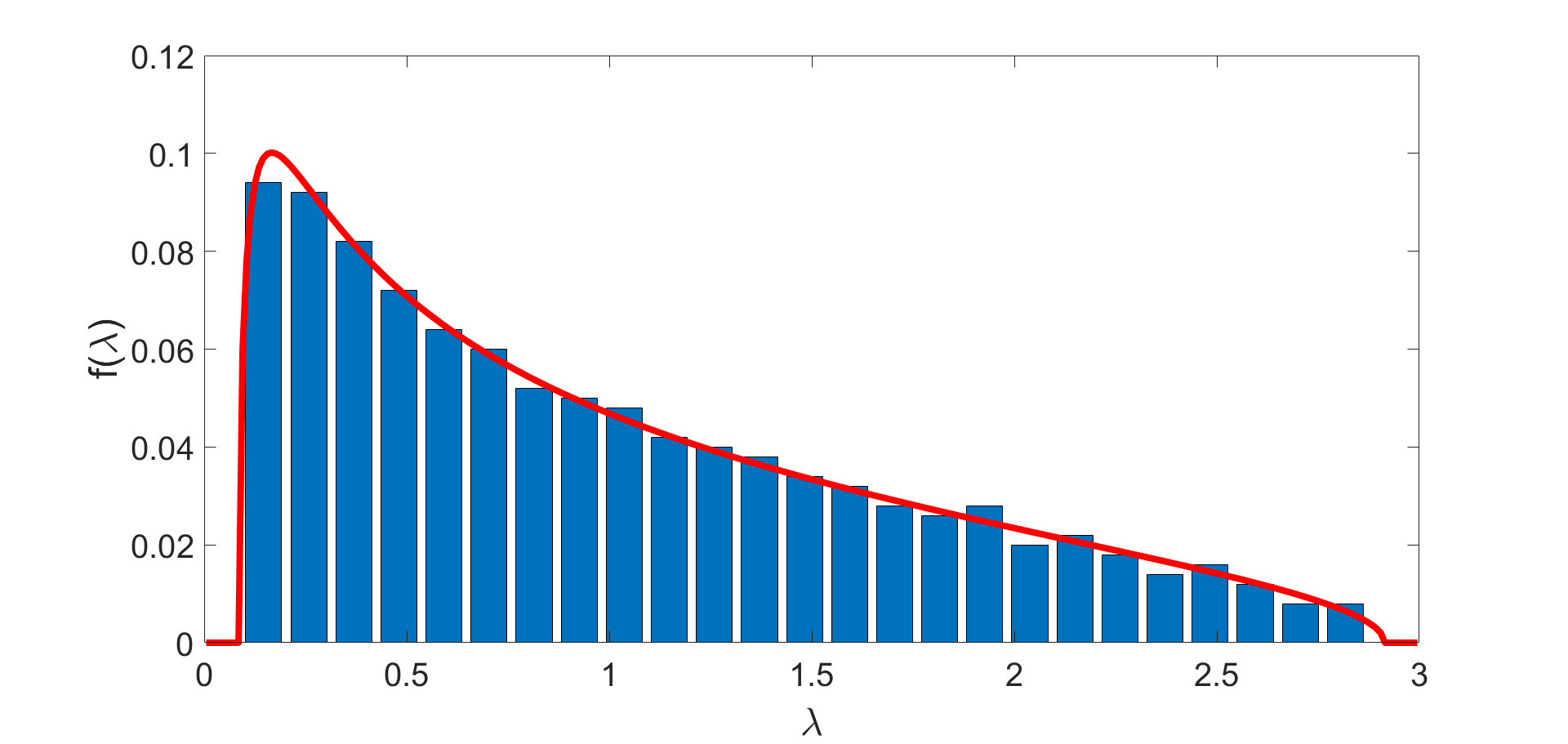}
\includegraphics[width = 0.49\textwidth]{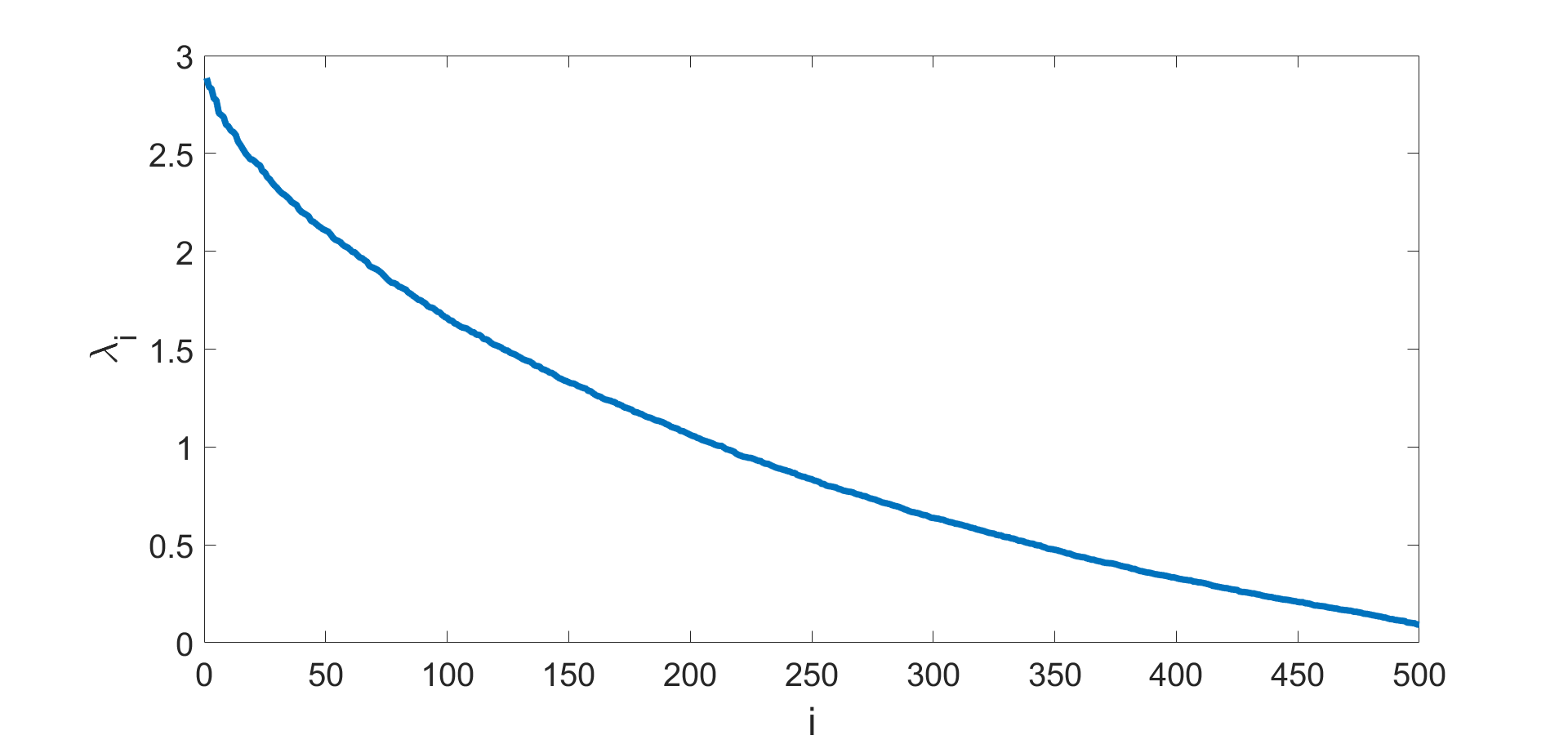}
\end{center}
\caption{A normalized histogram (left panel) and a scree plot (right panel) of the eigenvalues of a realization of $S_n$ (when $\Sigma = \Id$) for $p = 500$ and $n=1000$. The red line is the eigenvalue distribution predicted by the Mar\v{c}enko-Pastur distribution}
\label{fig:pastur1}
\end{figure}

Let us start by looking into a simple example, $\Sigma = \Id$. In that case, the distribution has no low dimensional structure, as the distribution is rotation invariant. Figure~\ref{fig:pastur1} depicts a normalized histogram (left panel) and a scree plot (right panel) of the eigenvalues of a realization of $S_n$ (when $\Sigma = \Id$) for $p = 500$ and $n=1000$. The red line is the eigenvalue distribution predicted by the Mar\v{c}enko-Pastur distribution~\eqref{eq:MPdist_PCA1}, that we will discuss below.

As one can see, there are many eigenvalues considerably larger than $1$, as well as many eigenvalues significantly smaller than $1$. Notice that, if given this profile of eigenvalues of $\Sigma_n$ one could potentially be led to believe that the data has low dimensional structure, when in truth the distribution it was drawn from is isotropic.

Understanding the distribution of eigenvalues of random matrices is at the core of Random Matrix Theory (there are many good books on Random Matrix Theory, e.g.~\cite{bai2010spectral,Tao_topicsRMT,Anderson_Guionnet_Zeitouni_IntroRandomMatrices}). This particular limiting distribution was first established in 1967 by Mar\v{c}enko and Pastur~\cite{Marchenko_Pastur_1967} and is now referred to as the Mar\v{c}enko-Pastur distribution. They showed that, if $p$ and $n$ both grow indefinitely to $\infty$ with their ratio fixed $p/n = \gamma \leq 1$, the sample distribution of the eigenvalues of $S_n$ (like the histogram above), in the limit, will be

\begin{equation}\label{eq:MPdist_PCA1}
dF_\gamma(\lambda) = \frac1{2\pi} \frac{\sqrt{\left( \gamma_+ - \lambda \right)\left( \lambda - \gamma_- \right)}}{\gamma \lambda} 1_{\left[ \gamma_-, \gamma_+  \right]}(\lambda) d\lambda,
\end{equation}
with support $\left[ \gamma_-, \gamma_+  \right]$, where $\gamma_- = (1-\sqrt{\gamma})^2$, $\gamma_+= (1+\sqrt{\gamma})^2$, and $\gamma = p/n$. This is plotted as the red line in the figure above.

\begin{remark}
We will not provide the proof of the Mar\v{c}enko-Pastur law here (you can see, for example,~\cite{Bai_MarchenkPastur_manyproofs} for several different proofs of it), but an approach to a proof is using the so-called moment method. The central idea is to note that one can compute moments of the eigenvalue distribution in two different ways and note that (in the limit) for any $k$,
\[
\frac1p \EE \tr \left[ \left( \frac1n XX^\top   \right)^k \right] =  \frac1p \EE \tr \left( S_n^k \right) =\EE\frac1p \sum_{i=1}^p \lambda_i^k(S_n) = \int_{\gamma_-}^{\gamma_+} \lambda^k dF_\gamma(\lambda),
\]
and that the quantities $\frac1p \EE \tr \left[ \left( \frac1n XX^\top   \right)^k \right]$ can be estimated (these estimates rely essentially on combinatorics). The distribution $dF_\gamma(\lambda)$ can then be computed from its moments.
\end{remark}

\subsection{Spike models and the BBP phase transition}

What if there actually is some (linear) low dimensional structure in the data? When can we expect to capture it with PCA? A particularly simple, yet relevant, example to analyze is when the covariance matrix $\Sigma$ is a rank-1 perturbation of the identity matrix and is of the form $\Sigma = I + \beta uu^\top  $, where $u$ is a unit norm vector and $\beta > 0$. This is a particular case of a spike model.

One way to think about this spike model is that each data point $x$ consists of a signal part $\sqrt{\beta}g_0 u$ where $g_0$ is a scalar standard Gaussian $\NNN(0,1)$ (i.e. a normally distributed multiple of a fixed vector $\sqrt{\beta}u$) and a noise part $g\sim\NNN(0,I)$ (independent of $g_0$). Then $x = g + \sqrt{\beta}g_0 u$ is a Gaussian random variable
\begin{equation}
 \label{eq:spikymodel}  
 x\sim \NNN(0,I + \beta uu^\top  ).
\end{equation}
Whereas the signal part $\sqrt{\beta}g_0 u$ resides on a central line in the direction of $u$, the noise part is high dimensional and isotropic. We therefore refer to $\beta$ as the signal-to-noise ratio (SNR). Indeed, $\beta$ is the ratio of the signal variance (in the $u$-direction) to the noise variance (in each direction).

A natural question is whether this rank-$1$ perturbation can be seen in $S_n = \frac{1}{n} XX^\top  $, where the columns of $X$ are now given by~\eqref{eq:spikymodel}. In other words, is it possible to detect the direction of the line $u$ from noisy measurements in high dimension? Let us build some intuition with an example. Figure~\ref{fig:pastur2} shows the histogram of the eigenvalues of a random realization of $S_n$ for $p=500$, $n=1000$, $u$ is the first element of the canonical basis $u=e_1$, and $\beta = 1.5$:

\begin{figure}
\begin{center}
\includegraphics[width = 0.9\textwidth]{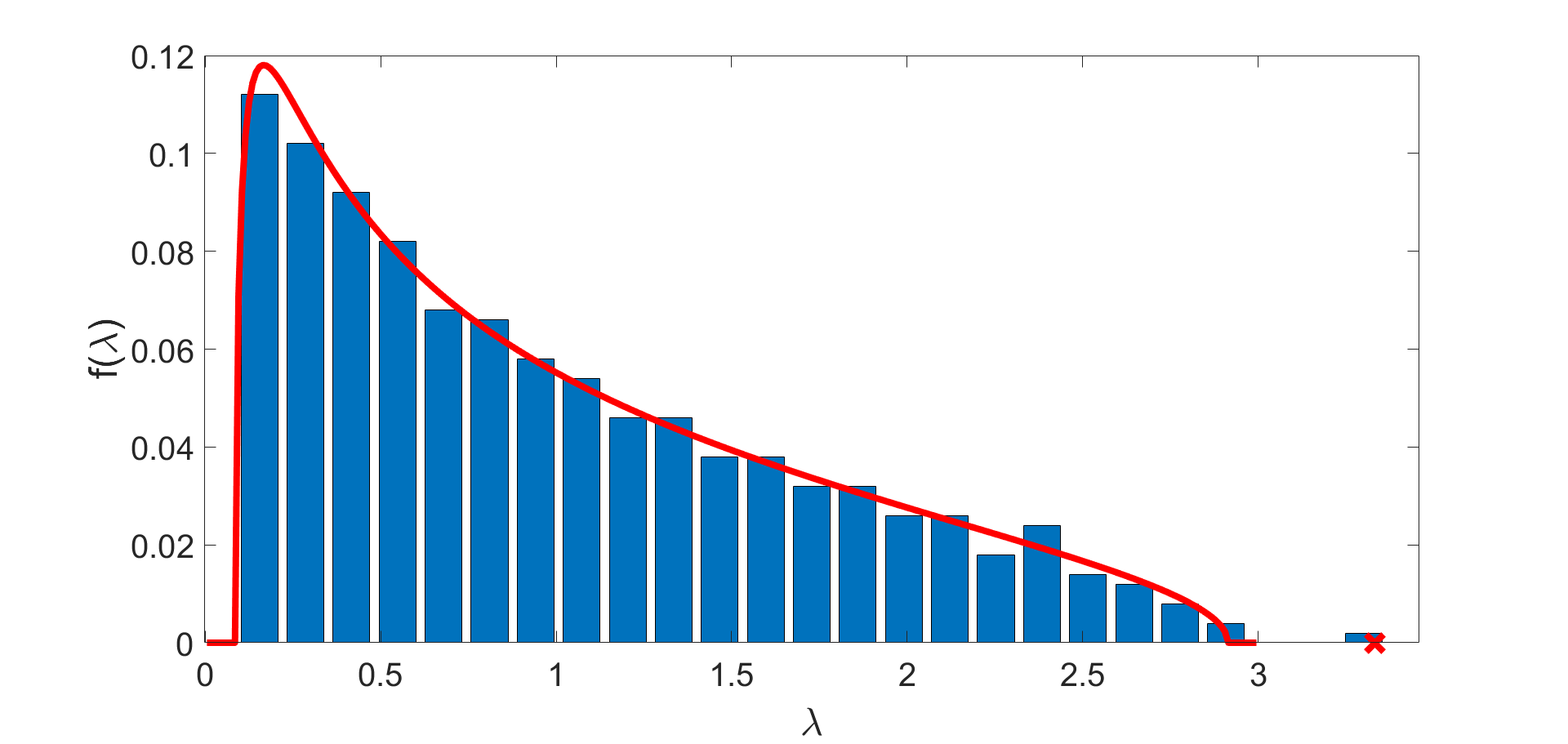}
\end{center}
\caption{The histogram of the eigenvalues of a random realization of the spike model $S_n$ for $p=500$, $n=1000$,  $u$ is the first element of the canonical basis $u=e_1$, and $\beta = 1.5$}
\label{fig:pastur2}
\end{figure}

The histogram suggests that there is an eigenvalue of $S_n$ that ``pops out'' of the support of the Mar\v{c}enko-Pastur distribution (below we will estimate the location of this eigenvalue, and that estimate corresponds to the red ``x''). It is worth noting that the largest eigenvalues of $\Sigma$ is simply $1+\beta = 2.5$ while the largest eigenvalue of $S_n$ appears considerably larger than that. 

\begin{figure}[h]
\begin{center}
\includegraphics[width = 0.9\textwidth]{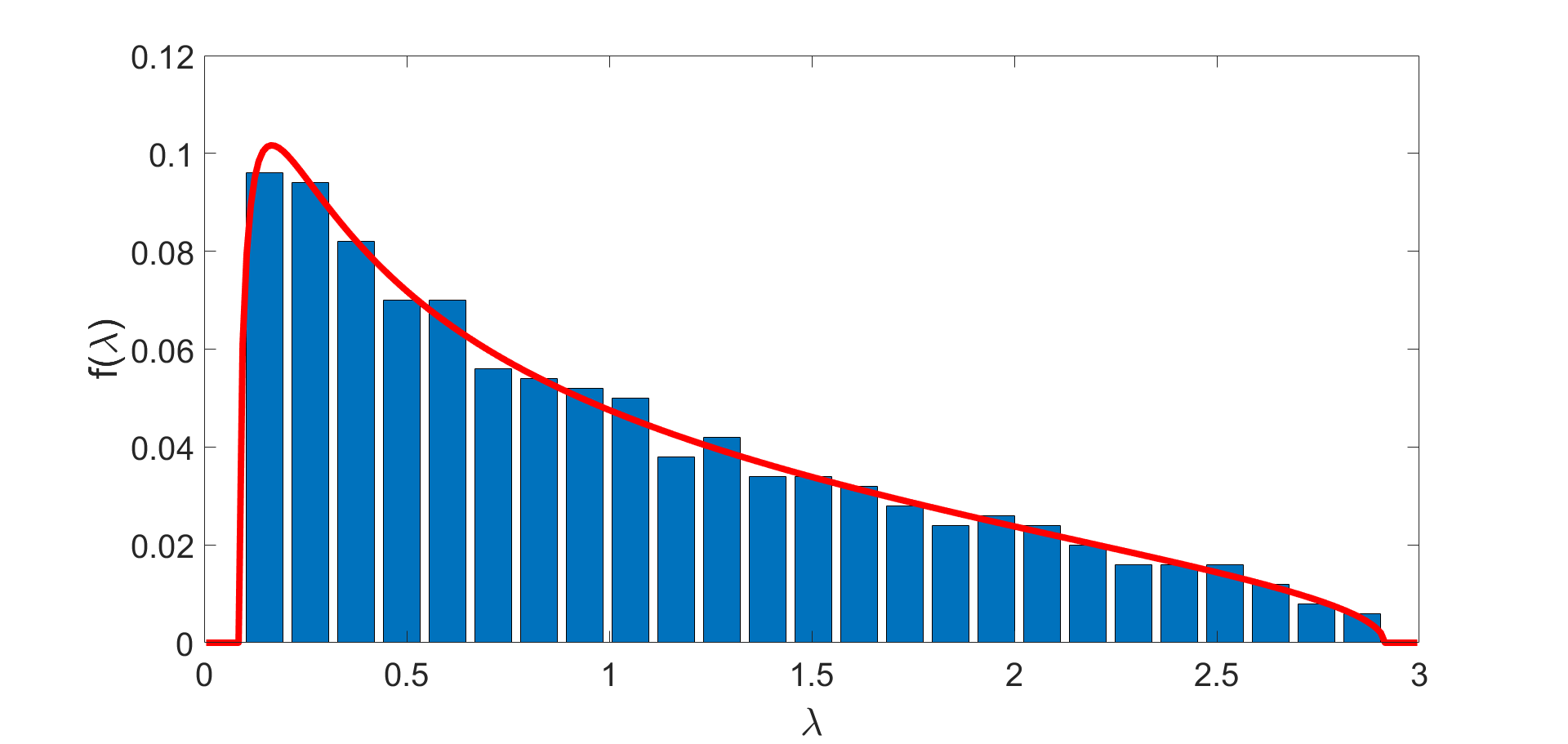}
\end{center}
\caption{The histogram of the eigenvalues of a random realization of the spike model $S_n$ for $p=500$, $n=1000$, $u$ is the first element of the canonical basis $u=e_1$, and $\beta = 0.5$}
\label{fig:pastur3}
\end{figure}

Let us try now the same experiment with $\beta = 0.5$, see Figure~\ref{fig:pastur3} for the result. It appears that, for $\beta = 0.5$, the histogram of the eigenvalues is indistinguishable from when $\Sigma = \Id$. In particular, no eigenvalue is separated from the Mar\v{c}enko-Pastur distribution.

This motivates the following question:

\smallskip
\noindent
{\em  Question:}
For which values of $\gamma$ and $\beta$ do we expect to see an eigenvalue of $S_n$ popping out of the support of the Mar\v{c}enko-Pastur distribution, and what is the limiting value that we expect it to take?

\medskip

As we will see below, there is a critical value of $\beta$, denoted $\beta_c$, below which we do not expect to see a change in the distribution of eigenvalues and above which we expect one of the eigenvalues to pop outside of the support. This phenomenon is known as the BBP phase transition (after Baik, Ben Arous, and P\'ech\'e~\cite{BPP_MC_BPP_2005}). There are many very nice papers about this and similar phenomena, including~\cite{Paul_Marchenko_Pastur_BPP_TechReport,Johnston_MC_BPP,BPP_MC_BPP_2005,Paul_Marchenko_Pastur_BPP_ActualPaper,Baik_Silverstein_BPP_05,Karoui_BPP_05,BenaychGeorges_Nadakuditi_PCA,BenaychGeorges_Nadakuditi_SVD}.\footnote{Notice that the Mar\v{c}enko-Pastur theorem does not imply that all eigenvalues are actually in the support of the Mar\v{c}enko-Pastur distribution, it just rules out that a non-vanishing proportion are. However, it is possible to show that indeed, in the limit, all eigenvalues will be in the support (see, for example,~\cite{Paul_Marchenko_Pastur_BPP_TechReport}).}

In what follows we will find the critical value $\beta_c$ and estimate the location of the largest eigenvalue of $S_n$ for any $\beta$. While the argument we will use can be made precise (and is borrowed from~\cite{Paul_Marchenko_Pastur_BPP_TechReport}) we will be ignoring a few details for the sake of exposition. In other words, the argument below can be transformed into a rigorous proof, but it is not one at the present form.

We want to understand the behavior of the leading eigenvalue of the sample covariance matrix
\[
S_n = \frac1n\sum_{i=1}^n x_ix_i^\top  .
\]
Since $x \sim \NNN(0,I+\beta uu^\top  )$ we can write $x = (I + \beta uu^\top  )^{1/2} z$ where $z \sim \NNN(0,I)$ is an isotropic Gaussian. Then,
\[
S_n = \frac{1}{n} \sum_{i=1}^n (I + \beta uu^\top  )^{1/2} z_i z_i^\top   (I + \beta uu^\top  )^{1/2} = (I + \beta uu^\top  )^{1/2} Z_n (I + \beta uu^\top  )^{1/2},
\]
where $Z_n = \frac{1}{n} \sum_{i=1}^n z_i z_i^\top  $ is the sample covariance matrix of independent isotropic Gaussians. The matrices $S_n = (I + \beta uu^\top  )^{1/2} Z_n (I + \beta uu^\top  )^{1/2}$ and $Z_n (I + \beta uu^\top  )$ are related by a similarity transformation, and therefore have exactly the same eigenvalues. Hence, it suffices to find the leading eigenvalue of the matrix $Z_n(I + \beta uu^\top  )$, which is a rank-1 perturbation of $Z_n$ (indeed, $Z_n(I + \beta uu^\top  ) = Z_n + \beta Z_n uu^\top  $). We already know that the eigenvalues of $Z_n$ follow the Mar\v{c}enko-Pastur distribution, so we are left to understand the effect of a rank-1 perturbation on its eigenvalues.

To find the leading eigenvalue $\lambda$ of $Z_n(I + \beta uu^\top  )$, let $v$ be the corresponding eigenvector, that is,
\[
Z_n(I + \beta uu^\top  ) v = \lambda v.
\]
Subtract $Z_n v$ from both sides to get
\[
\beta Z_n uu^\top   v = (\lambda I - Z_n)v.
\]
Assuming $\lambda$ is not an eigenvalue of $Z_n$, we can multiply by $(\lambda I - Z_n)^{-1}$ to get\footnote{Intuitively, $\lambda$ is larger than all the eigenvalues of $Z_n$, because it corresponds to a perturbation of $Z_n$ by a positive definite matrix $\beta uu^\top  $; yet, a formal justification is beyond the present discussion.}
\[
\beta (\lambda I - Z_n)^{-1} Z_n uu^\top   v = v.
\]
Our assumption also implies that $u^\top   v \neq 0$, for otherwise $v=0$.
Multiplying by $u^\top  $ gives
\[
\beta u^\top   (\lambda I - Z_n)^{-1} Z_n u (u^\top   v) = u^\top   v.
\]
Dividing by $\beta u^\top   v$ (which is not 0 as explained above) yields
\begin{equation}\label{eq:SVD:spike1}
u^\top   (\lambda I - Z_n)^{-1} Z_n u = \frac{1}{\beta}.
\end{equation}
Suppose $w_1,\ldots,w_p$ are orthonormal eigenvectors of $Z_n$ (with corresponding eigenvalues $\lambda_1,\ldots, \lambda_p$), and expand $u$ in that basis:
\[ u = \sum_{i=1}^p \alpha_i w_i.\]
Plugging this expansion in (\ref{eq:SVD:spike1}) gives
\begin{equation}
\sum_{i=1}^p \frac{\lambda_i}{\lambda-\lambda_i} \alpha_i^2 = \frac{1}{\beta}\label{eq:SVD:spike2}
\end{equation}
For large $p$, each $\alpha_i^2$ concentrates around its mean value $\mathbb{E}[\alpha_i^2]=\frac{1}{p}$ (again, this statement can be made rigorous), and \eqref{eq:SVD:spike2} becomes
\begin{equation}
\lim_{p\to \infty} \frac{1}{p}\sum_{i=1}^p \frac{\lambda_i}{\lambda-\lambda_i} = \frac{1}{\beta}
\end{equation}
Since the eigenvalues $\lambda_1,\ldots,\lambda_p$ follow the Mar\v{c}enko-Pastur distribution, the limit on the left hand side can be replaced by the integral
\begin{equation}
\int_{\gamma_{-}}^{\gamma_{+}} \frac{t }{\lambda-t}\,d F_\gamma(t) = \frac{1}{\beta}
\end{equation}
Using an integral table (or an integral software), we find that
\begin{equation}
\frac{1}{\beta} = \int_{\gamma_{-}}^{\gamma_{+}} \frac{t }{\lambda-t}\,d F_\gamma(t) =\frac{1}{4 \gamma} \left[2 \lambda -(\gamma_{-}+\gamma_{+})- 2 \sqrt{(\lambda-\gamma_{-})(\lambda-\gamma_{+})}\right]. \label{eq:SVD:Integral}
\end{equation}

For $\lambda=\gamma_{+}$, that is, when the top eigenvalue touches the right edge of the Mar\v{c}enko-Pastur distribution,  \eqref{eq:SVD:Integral} becomes $\frac{1}{4 \gamma} (\gamma_{+}-\gamma_{-})$. This is the critical point that one gets the pop out of the top eigenvalue from the bulk of the Mar\v{c}enko-Pastur distribution. To calculate the critical value $\beta_c$, we recall that $\gamma_{-}=(1-\sqrt{\gamma})^2$ and $\gamma_{+}=(1+\sqrt{\gamma})^2$, hence
\begin{equation}
\frac{1}{\beta_c} = \frac{1}{4 \gamma} \left((1+\sqrt{\gamma})^2-(1-\sqrt{\gamma})^2\right).
\end{equation}
Therefore, the critical SNR is
\begin{equation}
\label{eq:SVD:critical}
\beta_{c}= \sqrt{\gamma}=\sqrt{\frac{p}{n}}.
\end{equation}
When $\beta > \sqrt{\frac{p}{n}}$ one can observe the pop out of the top eigenvalue from the bulk.

Eq.~(\ref{eq:SVD:critical}) illustrates the interplay of the SNR $\beta$, the number of samples $n$, and the dimension $p$. Low SNR, small sample size, and high dimensionality are all obstacles for detecting linear structure in noisy high dimensional data.

More generally, inverting the relationship between $\beta$ and $\lambda$ given by \eqref{eq:SVD:Integral} (which simply amounts to solving a quadratic), we find that the largest eigenvalue $\lambda$ of the sample covariance matrix $S_n$ has the limiting value
\begin{equation}
\label{eq:SVD:pop}
\lambda \to \left\{\begin{array}{ccc}
                     (\beta + 1)\left(1 + \frac{\gamma}{\beta} \right) & &  \text{for }\;\; \beta \geq \sqrt{\gamma}, \\
                      &  & \\
                     (1+\sqrt{\gamma})^2 & & \text{for }\;\; \beta < \sqrt{\gamma}.
                   \end{array}
  \right.
\end{equation}
In the finite sample case $\lambda$ will be fluctuating around that value.

Notice that the critical SNR value, $\beta_{c} = \sqrt{\gamma}$ is buried deep inside the support of the Mar\v{c}enko-Pastur distribution, because $\sqrt{\gamma} < \gamma_{+} = (1+\sqrt{\gamma})^2$. In other words, the SNR does not have to be greater than the operator norm of the noise matrix in order for it to pop out. We see that the noise effectively pushes the eigenvalue to the right (indeed, $\lambda > \beta$).

The asymptotic squared correlation $|\langle u, v\rangle |^2$ between the top eigenvector $v$ of the sample covariance matrix and true signal vector $u$ can be calculated in a similar fashion. The limiting correlation value turns out to be
\begin{equation}
\label{eq:SVD:corr}
|\langle v, u \rangle|^2 \to \left\{\begin{array}{ccc}
                                      \frac{1-\frac{\gamma}{\beta^2}}{1+\frac{\gamma}{\beta^2}} & & \text{for }\;\; \beta \geq \sqrt{\gamma} \\
                                       & & \\
                                      0 & & \text{for }\;\; \beta < \sqrt{\gamma}
                                    \end{array}
 \right.
\end{equation}
Notice that the correlation value tends to 1 as $\beta \to \infty$, but is strictly less than 1 for any finite SNR.

\subsection{Wigner matrices}\label{subsection:WignerMatrices}

Another very important random matrix model is the Wigner matrix (and it will make appearances in Chapters~\ref{c:probability-gaussiananalysis},~\ref{c:probability-matrixconcentration} and~\ref{c:community}). Given an integer $n$, a standard Gaussian Wigner matrix $W\in\RR^{n\times n}$ is a symmetric matrix with independent $\NNN(0,1)$ off-diagonal entries (except for the fact that $W_{ij}=W_{ji}$) and jointly independent $\NNN(0,2)$ diagonal entries. In the limit, the eigenvalues of $\frac1{\sqrt{n}}W$ are distributed according to the so-called semi-circular law
\begin{equation}\label{eq:WignerSC}
 dSC(x) = \frac1{2\pi}\sqrt{4-x^2}1_{[-2,2]}(x)dx,
\end{equation}
and there is also a BBP like transition for this matrix ensemble~\cite{Feral_Peche_BPPWigner}. More precisely, if $v$ is a unit-norm vector in $\RR^n$ and $\xi\geq 0$ then the largest eigenvalue of $\frac1{\sqrt{n}}W+\xi vv^\top  $ satisfies

\begin{itemize}
 \item If $\xi \leq 1$ then
 \[
  \lambda_{\max}\left(\frac1{\sqrt{n}}W+\xi vv^\top   \right) \to 2,
 \]
 \item and if $\xi > 1$ then
 \begin{equation}\label{eq:1spike_Wigner:PCA1}
  \lambda_{\max}\left(\frac1{\sqrt{n}}W+\xi vv^\top   \right) \to \xi + \frac1{\xi}.
 \end{equation}
\end{itemize}

The typical correlation, with $v$, of the leading eigenvector $v_{\max}$ of $\frac1{\sqrt{n}}W+\xi vv^\top  $ is also known:
\begin{itemize}
 \item If $\xi \leq 1$ then
 \[
  \left| \left\langle v_{\max},v \right\rangle \right|^2 \to 0,
 \]
 \item and if $\xi > 1$ then
 \[
    \left| \left\langle v_{\max},v \right\rangle \right|^2 \to 1-\frac{1}{\xi^2}.
 \]
\end{itemize}

From a statistical viewpoint, a central question is to understand for different distributions of matrices, when it is possible to detect and estimate a spike in a random matrix~\cite{AlexAmeliaAfonsoAnkur_PCA}. When the underlying random matrix corresponds to the adjacency matrix of a random graph and the spike to a bias on the distribution of the graph edges, corresponding to structural properties of the graph, the estimates above are able to predict important phase transitions in community detection in networks, as we will see in Chapter~\ref{c:community}.

\subsection{Rank and covariance estimation}
The spike model and random matrix theory thus offer a principled way for determining the number of principal components, or equivalently of the rank of the hidden linear structure: simply count the number of eigenvalues to the right of the Mar\v{c}enko-Pastur distribution. In practice, this approach for rank estimation is often too simplistic for several reasons. First, for actual datasets, $n$ and $p$ are finite, and one needs to take into account non-asymptotic corrections and finite sample fluctuations \cite{kritchman2008determining,kritchman2009non}. Second, the noise may be heteroskedastic (that is, noise variance is different in different directions). Moreover, the noise statistics could also be unknown and it can be non-Gaussian \cite{liu2018pca}. In some situations it might be possible to estimate the noise statistics directly from the data and to homogenize the noise (a procedure sometimes known as ``whitening'') \cite{leeb2018optimal}. These situations call for careful analysis, and many open problems remain in the field.

Another popular method for rank estimation is using permutation methods. In permutation methods, each column of the data matrix is randomly permutated, so that the low-rank linear structure in the data is destroyed through scrambling, while only the noise is preserved. The process can be repeated multiple times, and the statistics of the singular values of the scrambled data matrices are then used to determine the rank. In particular, only singular values of the original (unscrambled) data matrix that are larger than the largest singular value of the scrambled matrices (taking fluctuations into account of course) are considered as corresponding to signal and are counted towards the rank. The mathematical analysis of permutation methods is another active field of research \cite{dobriban2017permutation,dobriban2018deterministic}.

In some applications, the objective is to estimate the low rank covariance matrix of the clean signal $\Sigma$ from the noisy measurements. We saw that in the spike model, the eigenvalues of the sample covariance matrix are inflated due to noise. It is therefore required to shrink the computed eigenvalues of $S_n$ in order to obtain a better estimate of the eigenvalues of $\Sigma$. That is, if $$S_n = \sum_{i=1}^p \lambda_i v_i v_i^\top  $$ is the spectral decomposition of $S_n$, then we seek an estimator of $\Sigma$, denoted $\hat{\Sigma}$ of the form $$\hat{\Sigma} = \sum_{i=1}^p \eta(\lambda_i) v_i v_i^\top  .$$
The scalar nonlinearity $\eta : \mathbb{R}^+ \to \mathbb{R}^+$ is known as the shrinkage function.  An obvious shrinkage procedure is to estimate $\beta = \eta(\lambda)$ from the computed $\lambda$ by inverting (\ref{eq:SVD:pop}) (and setting $\beta=0$ for $\lambda < \gamma_{+}$). It turns out that this particular shrinker is optimal in terms of the operator norm loss. However, for other loss functions (such as the Frobenius norm loss), the optimal shrinkage function takes a different form \cite{donoho2018optimal}. The reason why the shrinker depends on the loss function is that the eigenvectors of $S_n$ are not perfectly correlated with those of $\Sigma$ but rather make some non-trivial angle, as in (\ref{eq:SVD:corr}). In other words, the eigenvectors are noisy, and it may require more aggressive shrinkage to account for that error in the eigenvector. It can be shown that the eigenvector $v$ of the sample covariance is uniformly distributed in a cone around $u$ whose opening angle is given by (\ref{eq:SVD:corr}). While we can improve the estimation of the eigenvalue via shrinkage, it is however unclear how to improve the estimation of the eigenvector (without any a priori knowledge about it). Finally, we remark that eigenvalue shrinkage also plays an important role in denoising.

\section*{Exercises}
\addcontentsline{toc}{section}{Exercises}

\begin{myexercise}
The $p$-norm of a matrix $A$ is defined as
$$\|A\|_p = \sup_{x\neq 0} \frac{\|Ax\|_p}{\|x\|_p} = \sup_{\|x\|_p=1} \|Ax\|_p.$$
Prove the following properties of the matrix $p$-norm:
\begin{enumerate}
\item $\|AB\|_p \leq \|A\|_p \|B\|_p$ for any $A \in \mathbb{R}^{m\times n}$ and $B\in \mathbb{R}^{n\times k}$.
\item $\|A\|_1 = \max_{1\leq j \leq n} \sum_{i=1}^m |a_{ij}|$.
\item $\|A\|_\infty =  \max_{1 \leq i \leq m} \sum_{j=1}^n |a_{ij}|$. \\
Notice that the last two properties imply $\|A\|_\infty = \|A^\top  \|_1$.
\item $\|A\|_2^2 \leq \|A\|_1 \|A\|_\infty$.
\end{enumerate}
\end{myexercise}

\begin{myexercise}
The Schatten $p$-norm of an $m\times n$ matrix $A$ is given by the $\ell_p$ norm ($p\geq 1$) of its vector of singular values, namely,
$$\|A\|_p = \left(\sum_{k=1}^{\min\{m,n \}} \sigma_k^p\right)^{1/p}.$$
A matrix norm $\| \cdot \|$ that satisfies $\|QAZ\| = \|A\|$ for all $Q$ and $Z$ orthogonal matrices is called a unitarily invariant norm.
\begin{enumerate}
\item Show that the Schatten $p$-norm is unitarily invariant.
\item Note that the case $p=1$ is sometimes called the nuclear norm of the matrix, the case $p=2$ is the Frobenius norm, and $p=\infty$ is the operator norm. Verify that the nuclear norm of $A$, denoted $\|A\|_*$, is given by
$$\|A\|_{*} = \tr\left(\sqrt{A^\top   A} \right).$$
\end{enumerate}   
\end{myexercise}

\begin{myexercise}
Show that any unitarily invariant matrix norm is dependent only on the singular values of the matrix.   
\end{myexercise}

\begin{myexercise}
Given a square $n\times n$ matrix $A$ whose SVD is $A=U\Sigma V^\top  $, show that its closest (in the Frobenius norm) orthogonal matrix $O$ is given by $W=UV^\top  $. That is, show that
    $$\|A - UV^\top  \|_F = \min_{WW^\top  =W^\top  W=\Id} \|A-W\|_F,$$ where
    $A=U\Sigma V^\top  $.
    In other words, $W$ is obtained from the SVD of $A$ by dropping the diagonal matrix $\Sigma$. Use this observation to conclude what is the optimal rotation (and possibly reflection) that aligns two sets of points $p_1,p_2,\ldots,p_n$ and $q_1,\ldots,q_n$ in $\mathbb{R}^d$, that is, find $W$ that minimizes $\sum_{i=1}^n \|Wp_i-q_i\|^2$. This is known as the orthogonal Procrustes problem, see also  \cite{arun1987least,fan1955some,keller1975closest}.    
\end{myexercise}

\begin{myexercise}[\level\sep Low rank approximation]\label{prob:low_rank_inequality}
    Let $A \in \R^{m \times n}$ and $k$ be an integer such that $1 \leq k \leq \rank(A)$.
    \begin{enumerate}[(a)]
        \item Prove that there exists a matrix $B \in \R^{m \times n}$ of rank $k$ such that
        \begin{equation*}
            \norm{A-B} \leq \frac{\norm{A}_F}{\sqrt{k}}.
        \end{equation*}
        
        \item Does the statement (a) hold if the operator norm on the left hand side is replaced with the Frobenius norm $\|A - B \|_F$? 
    \end{enumerate}
\end{myexercise}

\begin{myexercise}
For the spike model, show that the asymptotic squared correlation $|\langle v,u \rangle|^2$ between the eigenvector $v$ of the sample covariance matrix and the ``true" signal vector $u$ is given by (\ref{eq:SVD:corr}).
\begin{hint}
    Use MAPLE or an integral table to calculate the integral $$\int_{\gamma_-}^{\gamma_+} \frac{t^2 \,dF_\gamma(t)}{(\lambda-t)^2},$$
    where $dF_\gamma(t)$ is the Mar\v{c}enko-Pastur measure.
\end{hint}

\end{myexercise}

\begin{myexercise}
Confirm Wigner's semi-circle law using computer simulations (take, e.g., $n=400$).
\end{myexercise}

\begin{myexercise}
Show that the asymptotic largest eigenvalue of a rank-1 perturbation of a Wigner matrix is given by (\ref{eq:1spike_Wigner:PCA1}).    
\end{myexercise}

\begin{myexercise}
 Let $dSC(x) = \frac1{2\pi}\sqrt{4-x^2}1_{[-2,2]}(x)$ be the semi-circle distribution. Prove that if $X$ is distributed according to $dSC(x)$ then
    $$\mathbb{E}[X^k] = \left\{\begin{array}{cc}
   C_{\frac{k}{2}} & \text{if}\;\; k \;\;\text{even}, \\
0 & \text{if}\;\;k\;\;\text{odd},
  \end{array} \right.$$
where $C_n$ is the $n$-th Catalan number, $C_n = \frac{1}{n+1} {2n \choose n}$.    
\end{myexercise}

\begin{myexercise}[\level\sep Equivalent definitions of spectral/operator norms]
    Given a matrix $M \in \R^{m \times n}$, prove that all of the following quantities are equal:
    \begin{enumerate}[(a)]
        \item $\sup_{\norm{v}_2 = 1} \norm{Mv}_{2}$, the \emph{operator norm} of $M$, which is commonly denoted by $\norm{M}$;
        \item $\sup_{v \neq 0} \frac{\norm{Mv}_{2}}{\norm{v}_2}$;
        \item $\sup_{\norm{u}_2 = \norm{v}_2 = 1} u^\top M v$;
        \item $\sigma_1(M)$, the largest singular value of $M$;
        \item $\sqrt{\lambda_1(M M^\top)}$, the square root of the largest eigenvalue of $M M^\top$;
        \item $\sqrt{\lambda_1(M^\top M)}$, the square root of the largest eigenvalue of $M^\top M$.
    \end{enumerate}
\end{myexercise}

\begin{myexercise}[\level\sep Maximal entry bound]\label{prob:max_entry}
    Given a matrix $X \in \R^{n \times m}$, show that for any $i \in [n]$ and $j \in [m]$ we have 
    \begin{equation*}
        - \norm{X} \leq \abs{X_{ij}} \leq \norm{X}.
    \end{equation*}
\end{myexercise}

\begin{myexercise}[\level\sep Symmetrization of matrices]\label{prob:symmetrization}
    We are going to explore three ways in which an $m \times n$ (with $m \leq n$) real-valued matrix $M$ can be symmetrized.
    \begin{enumerate}[(a)]
        \item Let $A$ be an $m \times m$ matrix defined by
        \begin{equation*}
            A \coloneqq M M^\top.
        \end{equation*}
        Check that $A$ is symmetric, and show that its $m$ eigenvalues are given by: $\sigma_1(M)^2, \sigma_2(M)^2, \ldots, \sigma_m(M)^2$.

        \item Show that $A$ and $B \coloneqq M^\top M$ have the same non-zero eigenvalues, up to multiplicities.
        
        \item Let $C$ be an $(m+n) \times (m+n)$ matrix defined by
        \begin{equation*}
            C \coloneqq \begin{pmatrix} 0_{m \times m} & M \\ M^\top & 0_{n \times n} \end{pmatrix},
        \end{equation*}
        where $0_{r \times r}$ is an $r \times r$ all-zeros matrix. Check that $C$ is symmetric, and show that its $m+n$ eigenvalues are given by:
        $$\text{a) }\, \sigma_1(M), \ldots, \sigma_m(M); \, \text{b) }
        -\sigma_1(M), \ldots, -\sigma_m(M); \, 
        \text{c) $n-m$ of them are 0}.$$    
    \end{enumerate}
\end{myexercise}

\begin{myexercise}[\level\level\sep Gershgorin circle theorem]\label{prob:gershgorin}
    Let $A \in \R^{n \times n}$ be a symmetric matrix with entries $(a_{ij})_{i,j\in[n]}$. For $i\in [n]$ let $R_{i}$ be the sum of the absolute values of the non-diagonal entries in the $i$-th row:
    \begin{equation*}
        R_{i}=\sum _{j\neq {i}}\left|a_{ij}\right|.
    \end{equation*}
    Prove that every eigenvalue of $A$ lies within at least one of the Gershgorin discs $D(a_{ii},R_{i})$, i.e. for any eigenvalue $\lambda$ of $A$ we can find $i \in [n]$ such that $\abs{\lambda - a_{ii}} \leq R_i$.
\end{myexercise}

\begin{myexercise}[\level\level\sep Quadratic form optimization]\label{prob:quadratic_form}
    Let $A \in \R^{n\times n}$ be a symmetric matrix with eigenvalues $\lambda_1\geq  \ldots \geq \lambda_n$. Given $r \in \brac{1, 2, \ldots, n}$, consider the following optimization problem:
    \begin{equation*}
        \max_{v_1,\ldots,v_r \in \R^n} \sum_{i=1}^r v_i^\top A v_i \quad \text{s.t.}\quad v_i^\top v_j = \delta_{ij} \text { for } 1 \leq i,j \leq r,
    \end{equation*}
    where $\delta_{ij}$ is the Kronecker delta defined as
    \begin{equation*}
        \delta_{ij} = \begin{cases}
            1 & i = j, \\
            0 & i \neq j.
        \end{cases}
    \end{equation*}
    \begin{enumerate}[(a)]
        \item Show that $\trace(A)$ is the solution of the problem when $r = n$.
        \item Determine the solution of the problem in terms of the eigenvalues of $A$ when $r < n$.
    \end{enumerate}
\end{myexercise}

\begin{myexercise}[\level\level\sep Courant-Fischer minimax formula for symmetric matrices]\label{prob:courant_fischer_formula}
    Let $A \in \R^{n \times n}_{\text{sym}}$ ($n \times n$ real symmetric matrix). Prove that for any $k \in [n]$:
    \begin{equation*}
        \lambda_k(A) = \max_{\substack{V \subseteq \R^n\\ \dim(V) = k}} \min_{\substack{v \in V\\ \norm{v} = 1}} v^\top A v
        \qquad\text{and}\qquad
        \lambda_k(A) = \min_{\substack{V \subseteq \R^n\\ \dim(V) = n-k+1}} \max_{\substack{v \in V\\ \norm{v} = 1}} v^\top A v.
    \end{equation*}
\end{myexercise}
\begin{hint}
    Any $V$ with $\dim(V) = n-k+1$ intersects with the subspace spanned by the top $k$ eigenvectors of $A$.
\end{hint}

\begin{myexercise}[\level\level\sep Consequences of Courant-Fischer]\label{prob:courant_fischer_consequences}
    Let $X, Y \in \R^{m \times n}$ ($m \times n$ real matrices), with $m \leq n$.
    \begin{enumerate}[(a)]
        \item Using Problem~\ref{prob:symmetrization}(c) extend the Courant-Fischer formula to hold for asymmetric matrices: for any $k \in [m]$
        \begin{equation*}
            \sigma_k(X) = \max_{\substack{V \subseteq \R^n\\ \dim(V) = k}} \min_{\substack{v \in V\\ \norm{v} = 1}} \norm{X v}
            \quad\text{and}\quad            
            \sigma_k(X) = \min_{\substack{V \subseteq \R^n\\ \dim(V) = n-k+1}} \max_{\substack{v \in V\\ \norm{v} = 1}} \norm{X v}.
        \end{equation*}
        
        \item Show the Eckart–Young–Mirsky Theorem for Spectral norm: given any rank-$r$ matrix $B$ with $r < m$,
        \begin{equation*}
            \norm{X-B} \geq \sigma_{r+1}(X).
        \end{equation*}

        \item Show Weyl's inequality for singular values: for all $1 \leq i, j \leq m$ satisfying $i+j-1 \leq m$ we have
        \begin{equation*}
            \sigma_{i+j-1}(X+Y) \leq \sigma_i(X) + \sigma_j(Y).
        \end{equation*}
    \end{enumerate}
\end{myexercise}

\begin{myexercise}[\level\sep Inner product between matrices]\label{prob:matrix_inner_product}
    For any two matrices $A, B \in \R^{n \times n}$, consider the map
    \begin{equation*}
        \bran{A, B} \coloneqq \trace\brap{A B^\top}.
    \end{equation*}
    \begin{enumerate}[(a)]
        \item Prove that $\bran{\cdot, \cdot}$ is an inner product on the space of $n \times n$ matrices.

        \item Show that $\norm{A}_F^2 = \bran{A, A}$.

        \item Deduce the \textit{matrix Cauchy-Schwarz inequality}: $\bran{A, B} \leq \norm{A}_F \norm{B}_F$.
    \end{enumerate}
\end{myexercise}

\begin{myexercise}[\level\level\sep Rotation minimization]\label{prob:rotation_min}
    Let $A, B \in \R^{m\times n}$ be two arbitrary matrices. Find the solution, in terms of $A$ and $B$, or their SVD decompositions, of the following optimization problem:
    \begin{equation*}
        \argmin_{\Omega \in O(m)} \norm{\Omega A-B}_{F}.
    \end{equation*}
    Here $O(m)$ denotes the set of all $m\times m$ orthogonal matrices.
\end{myexercise}

\begin{myexercise}[\level\sep Polar decomposition]\label{prob:polar_decomposition}
    Let $A \in \R^{n\times n}$. Prove that there exists a positive semi-definite matrix $P$ and an orthogonal matrix $Q$ such that $A = PQ$.
\end{myexercise}

\begin{myexercise}[\level\level\sep Power method]\label{prob:power_method}
    Let $A \in \R^{n\times n}$ be a symmetric positive semi-definite matrix with eigenvalues $\lambda_1 \geq \ldots \geq \lambda_n \geq 0$ and associated eigenvectors $v_1, \ldots, v_n \in \R^n$ (that form an orthonormal basis). In this exercise, the goal is to show that the power method converges exponentially fast.
    
    \begin{enumerate}[(a)]
        \item Let $y_0 \in \R^n$ be an initial vector that satisfies $A y_0 \neq 0$. Define the power method iteration for $k \geq 0$: 
        \begin{equation*}
            y_{k+1} = \frac{A y_k}{\norm{A y_k}}.
        \end{equation*}
        Prove that these iterations are well-definied, i.e. that $\norm{A y_k} \neq 0$ for any $k \geq 1$.
        
        \item Define the Rayleigh quotient as $\xi_k = y_k^\top A y_k$ and its relative error by 
        \begin{equation*}
            \err(\xi_k) = \frac{ \lambda_1 - \xi_k}{\lambda_1}.
        \end{equation*}
        If $A$ is diagonal, i.e., $A = \diag(\lambda_1, \dots, \lambda_n)$ and $\lambda_1 = 1$, show that we can represent the error as
        \begin{equation*}
            \err(\xi_k) = \frac{\sum_{i=2}^n w_i^2 \lambda_i^{2k} (1 - \lambda_i)}{w_1^2 + \sum_{i=2}^n w_i^2 \lambda_i^{2k}},
        \end{equation*}
        where $w_i = \bran{y_0, v_i}$ for all $i \in [n]$.

        \item If $A$ is diagonal with $1 = \lambda_1 > \lambda_2 > \lambda_3$, and we start from $y_0$ such that $w_1 \neq 0$ and $w_2 \neq 0$, show that
        \begin{equation*}
            \frac{\err(\xi_{k+1})}{\err(\xi_k)} \to \brap{\frac{\lambda_2}{\lambda_1}}^{2} \quad \text{as $k\to \infty$}.
        \end{equation*}
        
        \item Generalize the result in (c) for an arbitrary matrix (not necessarily diagonal) having $\lambda_1 > \lambda_2 > \lambda_3$.
    \end{enumerate}
\end{myexercise}

\begin{myexercise}\label{ex:mse_unbiased}
This exercise relates to Theorem~\ref{th:mle}. Show that
an unbiased estimator of $\sigma^2$ is given by
$\hat{\sigma}^2_{\text{MLE}} = \frac{1}{n-p}\|y - X\hat{\beta}_{\text{LS}}\|^2$.
\end{myexercise}

\begin{myexercise}[\level\level\sep Moore-Penrose Pseudoinverse]\label{prob:pseudoinverse}
    \begin{enumerate}[(a)]
        \item Let $\Sigma$ be an $n\times m$ rectangular diagonal matrix ($\Sigma_{ij}=0$ for $i\neq j$) with non-negative entries. Find an $m\times n$ rectangular diagonal matrix $\Sigma^+$ that is a pseudoinverse of $\Sigma$.
        
        \item Given a general $n \times m$ matrix $A$, consider the singular value decomposition of $A = U\Sigma V^\top  $ with $U$ and $V$ being orthogonal matrices, and $\Sigma$ being (rectangular) diagonal. Prove that the matrix $A^+$, given by $A^{+} = V\Sigma^+U^\top  $, is a pseudoinverse of $A$.
        
        \item Prove that if $A$ is an invertible $n \times n$ matrix, then $A^{-1}$ is a pseudoinverse of $A$.
        
        \item Prove that if $A$ has full column rank (its columns are linearly independent) then its pseudoinverse is given by
        \begin{equation*}
            A^{+} = \brap{A^\top   A}^{-1} A^\top  .
        \end{equation*}
        
        \item Prove that the pseudoinverse is unique.
    \end{enumerate}
\end{myexercise}

\begin{myexercise}[\level\level\level\sep BBP for spiked Wigner model]\label{prob:spiked_wigner}
    In this chapter we analyzed the BBP transition for the Wishart model, i.e., when we observe $Y = \frac{1}{n} XX^\top$, where $X$ is an $p \times n$ matrix with columns drawn independently from $\NN(0, I_p + \beta u u^\top)$. We will explore the similar type of phase transition for another model.
    
    Let $W$ be an $n \times n$ Wigner matrix, $v$ be a unit-norm vector in $\R^n$ and $\xi \geq 0$. We define the \textit{spiked Wigner model} as observing $Y = \frac{1}{\sqrt n} W + \xi v v^\top$, with the aim of recovering the \textit{signal} $v$.
    This model exhibits the following phase transition (as $n \to \infty$) for
    \begin{enumerate}
        \item the largest eigenvalue $\lambda_{\max}$ of $Y$:
        \begin{equation*}
            \lambda_{\max} \rightarrow \begin{cases}
                2 &\text{if } \xi \leq 1,\\
                \xi+\frac{1}{\xi} &\text{if } \xi > 1;\\
            \end{cases}
        \end{equation*}
        
        \item the leading eigenvector $v_{\max}$ of $Y$:
        \begin{equation*}
            \abs{\bran{v_{\max}, v}}^2 \rightarrow \begin{cases}
                0 &\text{if } \xi \leq 1,\\
                1-\frac{1}{\xi^2} &\text{if } \xi > 1.\\
            \end{cases}
        \end{equation*}
    \end{enumerate}
    
    To measure the quality of the recovery procedure we define the \textit{mean squared error} of an estimate $w \in \R^n$ as
    \begin{equation*}
        \mse(w) =  \E\bras{\norm{w w^\top - v v^\top}^2}.
    \end{equation*}
    
    Find the asymptotic behavior of the mean squared error for the PCA estimator, i.e., the value of
    \begin{equation*}
        \lim_{n\to \infty} \mse(v_{\max})
    \end{equation*}
    as a function of $\xi$. 
\end{myexercise}
\begin{remark}
    The mean square error defined as above might appear unnatural for this problem since one can measure the difference between two vectors using $\ell_2$ norm (up to a sign). However, this type of MSE definition can be more useful when the perturbation is not rank-one and non-symmetric (e.g., in low-rank matrix estimation problems). Additionally, it addresses the issue of sign invariance in the model, where the observation remains the same whether the signal is $v$ or $-v$, and therefore,  we can recover the vector only up to a sign.
\end{remark}

\begin{myexercise}
This exercise is about PCA. We use the classical iris data set as an example (see \verb|https://en.wikipedia.org/wiki/Iris_flower_data_set|
for an explanation of that dataset). You can download a subset of the iris dataset from the UCI Machine Learning Repository).

The dataset contains fifty flowers and for each flower there are four measurements: sepal length, sepal width, petal length and petal width.
We want to understand via PCA
how these four measurements are related to each other. Are they essentially independent of each other or are some of them (weakly or strongly) correlated?

\noindent
(a) Assume you loaded the iris data in Matlab into a data matrix $X$ of size $4 \times 50$. You can examine the structure of the data using plotmatrix: \\
\verb| [~,ax]=plotmatrix(X');|  \\
\verb| ax(1,1).YLabel.String='Sepal L';| \\
\verb| ax(2,1).YLabel.String='Sepal W';| \\
\verb| ax(3,1).YLabel.String='Petal L';|\\
\verb| ax(4,1).YLabel.String='Petal W';|\\
\verb| ax(4,1).XLabel.String='Sepal L';|\\
\verb| ax(4,2).XLabel.String='Sepal W';|\\
\verb| ax(4,3).XLabel.String='Petal L';|\\
\verb| ax(4,4).XLabel.String='Petal W';|\\

Which measurements are related and which ones are not-related? Compare your findings based on this visual inspection with your findings based on computing the variance and covariance of the data.

\noindent
(b) Next carry out a principal component analysis by projecting the data onto the eigenvectors of the (centered) covariance matrix of the data.
How much of the variation in the data is captured by the individual components? 
\end{myexercise}

\begin{myexercise}
\textbf{(PCA for Dimensionality Reduction).} In this exercise we work with the MNIST dataset.\\
(a) Standardize the data.\\
(b) Apply PCA and retain enough components to explain 95\% variance.\\
(c) Visualize the first two principal components for a sample of images.
\end{myexercise}

\begin{myexercise}\textbf{(Image Compression with Truncated SVD).} A grayscale image is represented by a matrix $A \in \R^{m \times n}$. Take a gray-scale image of your choice (not too small, at least of size $256\times 256$). \\
(a) Compute the rank-$k$ approximation using the SVD
for each $k=1,\dots,p$, where $p =\min(m,n)$.
Measure the relative Frobenius error and plot the reconstruction error as a function of $k$. For which range of $k$ do you see no visual distortion of the compressed image?
\end{myexercise}    

\chapter{Linear Regression and Regularization}
\label{c:linreg_ls}

\newcommand{\regpar}{\theta}

It is dawn at Mavericks, California. The swell forecast reads 10 meters at 18 seconds---a 
massive storm system has sent energy across thousands of miles of Pacific Ocean. 
For big wave surfers, these are exciting conditions! But they also come with all the dangers of a wipeout.

The {\em hold-down}---the time a surfer is trapped beneath the surface after a 
wipeout---is one of the most critical variables in a big-wave wipeout.
Professional big wave surfers can hold their breath for four to five minutes under controlled 
conditions, but after the physical exertion of paddling into a 12-meter wave, that 
window shrinks dramatically. Add the disorientation of being tumbled by thousands of tons of water, and you have maybe roughly 20-30 seconds before panic sets in, and perhaps 60-90 seconds before blackout.
This motivates a practical and (for some readers) urgent question: Given ocean conditions and wave characteristics, can we predict how long is a surfer likely to be held underwater?

Imagine we have collected data from 50 big wave sessions at various locations 
(Mavericks, Nazaré, Jaws, Teahupo'o,...) using underwater cameras and GPS trackers on  professional surfers. For each wipeout, we have recorded hold-down time (seconds) $y$,  wave height (meter, measured face) $x_1$,
wave period (seconds between waves) $x_2$,
 water depth at impact zone (meter) $x_3$, and
 wind speed (knots) $x_4$.
Tomorrow, a massive northwest swell is forecast with 
wave height 15m, period 20s, depth 12m, and wind 18 kts.
How long should we expect the hold-down to be? 

While the true physics is nonlinear (involving fluid dynamics, buoyancy, and 
turbulent flow), a linear approximation may capture the first-order effects within the  range of observed conditions. We thus postulate that hold-down time is approximately a linear function of the 
wave characteristics
\begin{equation*}
  y = \regpar_0 + \regpar_1 x_1 + \regpar_2 x_2 + \regpar_3 x_3 + \regpar_4 x_4 + \epsilon,
\end{equation*}
where $\regpar_0$ is the baseline hold-down time,
$\regpar_1$ is the increase in hold-down per meter of wave height, $\regpar_2$ is the effect of wave period, $\regpar_3$ is the effect of water depth (deeper water means shorter hold-downs), $\regpar_4$ is the effect of wind, and $\epsilon$ represents random variation (wave shape, surfer position, etc.)

With our 50 observations, we can write the model in matrix form
$$y = X \regpar + \eps$$
where $y \in \R^{50}$ is the vector of observed hold-down times, 
$X \in \R^{50 \times 5}$ contains the wave characteristics (with a column of 1's 
for the intercept), and $\regpar \in \R^5$ are the unknown coefficients we wish to estimate. Once we have found $\hat{\regpar}$, we can predict tomorrow's expected hold-down.

But how do we find the ``best'' estimate $\hat{\regpar}$? Is our estimator optimal in some sense? How accurate are our predictions?  Should we include all variables, or is simpler better? Linear regression and least squares will provide answers to these questions.

\medskip
This chapter studies linear regression and least squares from four complementary viewpoints. 
After introducing the basic idea of linear regression, we first show that least squares is an orthogonal projection problem. Second, we show that it arises naturally from Gaussian noise and conditional-mean prediction. Third, we analyze its bias, variance, and prediction risk. Finally, we show how ridge regression stabilizes least squares by shrinking low-information directions, and we reinterpret this shrinkage as Wiener filtering.

\section{Linear regression}

Linear regression is the foundational {\em supervised} learning model in data science, statistics, and applied mathematics~\cite{hastieelements,LVanderberghe_SBoyd_book,Golub_MatrixComputations}.
While it is often motivated by statistical inference, its mathematical core is a geometric problem: finding the ``best'' projection of a target vector onto the subspace spanned by the predictors.

Let us formulate the linear regression problem in greater generality.
Consider a dataset consisting of $n$ observations and $p$ features $x_1,\dots,x_p$, represented by the matrix $X \in \mathbb{R}^{n \times p}$ (often called the design matrix), and a vector of responses $y \in \mathbb{R}^n$. Now we introduce a {\em rather strong assumption} on the underlying data model: we {\em assume a linear relationship}
$y = X\regpar + \epsilon$,
where, as in the previous example, $\regpar \in \mathbb{R}^p$ is the vector of unknown coefficients and $\epsilon$ represents noise. The goal is now to find the best estimate $\hat{\regpar}$ of $\regpar$.

\smallskip
Linear regression serves two primary purposes:
\begin{enumerate}
\item Prediction: Given a new predictor $x_{\text{new}}$, we wish to predict 
the corresponding response $y_{\text{new}}$. Our postulated linear model suggests the prediction
$\hat{y}_{\text{new}} = x_{\text{new}}^\top \hat{\regpar}$
where $\hat{\regpar}$ is an estimate of $\regpar$.
\item Inference: We wish to understand the relationship between predictors and 
response, quantified by $\regpar$, such as: Which predictors are important?  How does changing $x_j$ affect $y$? 
What is the uncertainty in our estimates? 
\end{enumerate}

But how can we find an estimate $\hat{\regpar}$ in the first place? 
Because of noise, we do not expect to recover $\regpar$ exactly from finite data, nor do we expect $y$to lie exactly in the column space of $X$.
So, which error function (or {\em loss function} in the terminology of data science, statistics, and machine learning) shall we use to quantify a good (or even optimal) fit? 

We propose to compute the $\hat{\regpar}$ that minimizes the sum of squared residuals
\begin{equation}\label{eq:l2reg}
f(\regpar) = \|y - X\regpar\|^2.   
\end{equation}

This proposal immediately raises the question, why did we use the least-squares error in~\eqref{eq:l2reg} and not some other measure of fit?
After all, we could consider many alternatives, such as for instance the sum of absolute errors $\sum_{i=1}^n |y_i -x_i^\top\regpar|$, or the maximum absolute error $\max_{i=1,\ldots,n} |y_i - x_i^\top\regpar|$. 

The choice of the squared loss \eqref{eq:l2reg} is not arbitrary. It emerges from multiple 
complementary perspectives: geometric interpretation, 
computational tractability, and statistical optimality. We examine each justification in turn\footnote{The dominance of the least squares approach is also deeply rooted in the history of astronomy and the development of the normal distribution.
While Adrien-Marie Legendre was the first to publish the method in 1805 (as a purely algebraic way to fit curves to planetary data), Carl Friedrich Gauss published his version in 1809 and famously claimed he had been using it since 1795~\cite{stigler1981gauss}. Gauss provided the first probabilistic justification for the method, effectively inventing the field of statistical inference as we know it today~\cite{stigler1990history}.}.

\begin{figure}
\centering
\includegraphics[width=0.55\textwidth]{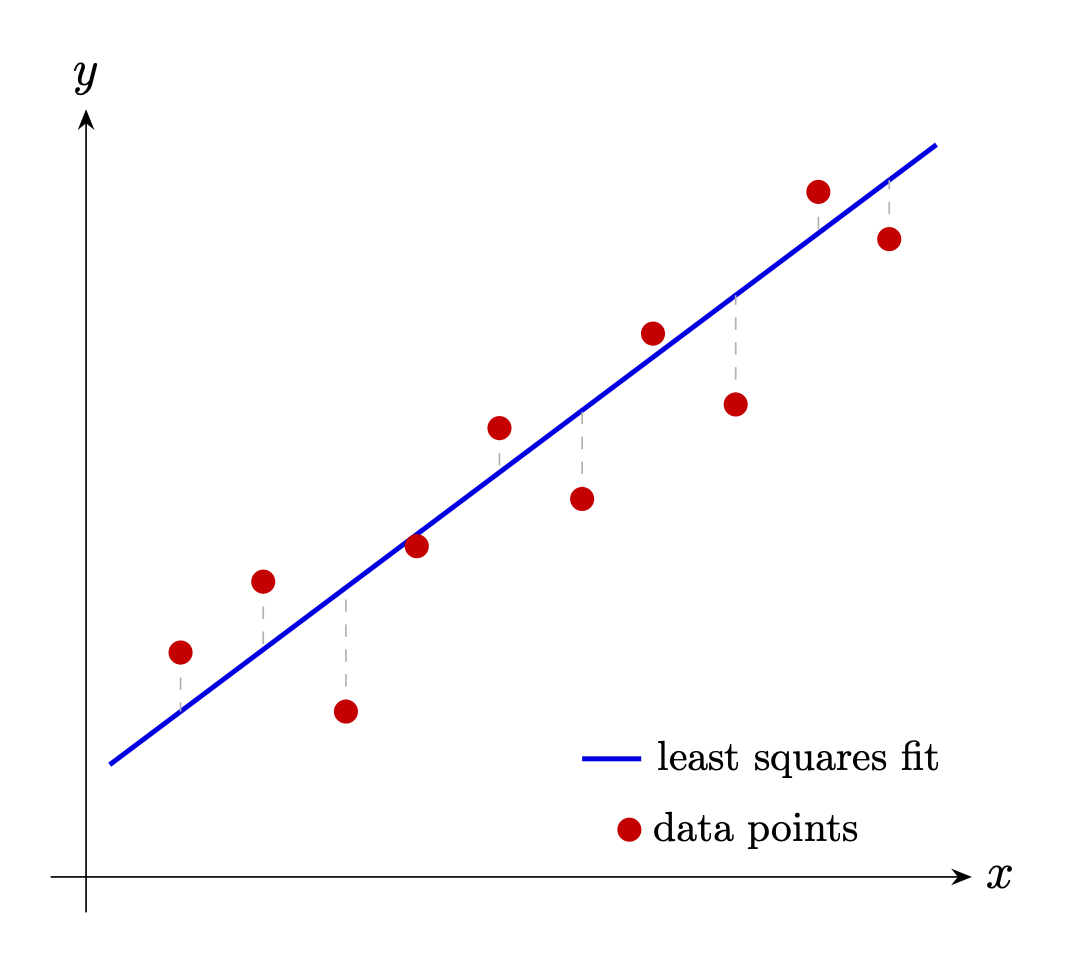}
\caption{Least squares regression line fitted to a set of observations. Dashed lines indicate the residuals $r_i =  y_i - \hat{y}_i$. We seek the linear
function that minimizes the sum of squared residuals.}
\label{fig:linearregression}
\end{figure}

\medskip
It is helpful to keep in mind the fundamental differences between linear regression and PCA (discussed in Chapter~\ref{s:PCA}).
While linear regression and PCA both involve fitting a linear structure to data, they have fundamentally different goals, and assumptions. Linear regression is a {\em supervised method}. The goal is to model the relationship between a set of predictor variables 
$X$ and a response variable $y$. It assumes the $X$ values are known perfectly\footnote{{\em Total Least Squares}, a generalization of least squares, is a regression technique that assumes errors exist in {\em both} $X$ and $y$, see e.g.~\cite{Golub_MatrixComputations}. The total least squares solution is equivalent to finding the first principal component of the combined matrix $[X \, y]$.} and all noise is in $y$. In contrast, PCA is an {\em unsupervised method} that does not involve a response variable. Instead, it seeks a low-dimensional linear structure that captures as much variation in the given data $X$ as possible. In a nutshell,
regression asks: ``How well can $X$ explain $y$?'' while PCA asks: ``What is the most efficient way to represent $X$?''

\section{Geometry, existence, and uniqueness of least squares}

The least squares loss has some appealing geometric interpretations that are not shared by other loss functions, which we collect below (and the verification of which is left as exercises).

Let $S \subset \mathbb{R}^n$ be a linear subspace. For every $y \in \mathbb{R}^n$, there exists a unique decomposition
\begin{equation}
  \label{th:orthoproj}
y = s + r, \quad
\text{such that
$s \in S$, and $r \in S^\perp$.}
\end{equation}
Furthemore, $s$ is the unique minimizer of
$$
\min_{u \in S} \|y - u\|^2.
$$

Let $\Col(X)$ denote the column space of $X$ and
$\Pcol$ the orthogonal projection onto $\Col(X)$.
Applying the subspace decomposition above to the case  $S = \Col(X)$ implies that
for any $y \in \mathbb{R}^n$, there exists $\hat \regpar \in \mathbb{R}^p$ such that
$$
X\hat \regpar = \Pcol y.
$$
Moreover, we have
$$
\hat \regpar \in \arg\min_{\regpar \in \mathbb{R}^p} \|y - X\regpar\|^2.
$$
In the hold-down example, the fitted vector $X \hat{\regpar}$
is the orthogonal projection of the observed hold-down times onto the subspace spanned by the measured wave features.

Let $X \in \mathbb{R}^{n \times p}$ and $y \in \mathbb{R}^n$. 
A vector $\hat \regpar \in \R^p$ minimizes
$$\|y - X\regpar\|^2$$
if and only if
$$
X^\top (y - X\hat \regpar) = 0.
$$
Equivalently,
\begin{equation}
\label{eq:normalequations}    
X^\top X \hat \regpar = X^\top y,
\end{equation}
which are also referred to as the {\em normal equations}.

Furthermore, if $X$ has full column rank ($\mathrm{rank}(X)=p$), then $X^T X$ is invertible and the normal equations have as unique solution
the least squares estimator given by 
\begin{equation}\label{eq:ols_standard}
  \hat{\regpar}_{\text{LS}} = (X^\top X)^{-1}X^\top y.
\end{equation}

However, this approach has both practical and theoretical limitations: (i)~Computing $X^\top X$ squares the condition 
number, amplifying rounding errors. This is not a big deal when the condition number of $X$ is small (say, $\kappa(X) = \OOO(1)$), but becomes a major concern when $X$ is not well conditioned. (ii)~When $\rank(X) < p$, the matrix $X^\top X$ is singular and cannot be inverted; (iii)~When $p > n$, the system is underdetermined.

The SVD provides an elegant solution that avoids forming 
$X^\top X$ explicitly and handles all cases (full rank, rank deficient, underdetermined) in a unified framework\footnote{The SVD may be computationally too expensive for large-scale problems. Conjugate gradient based iterative solvers are often the algorithms of choice in that case~\cite{Golub_MatrixComputations,hanke2017conjugate}. See also the randomized SVD in Chapter~\ref{s:RSVD} as a way to reduce the computational complexity of the SVD at the cost of reduced accuracy.}.

\begin{theorem} \label{th:ls_svd}
Let $X \in \R^{n \times p}$ have the SVD $X = U\Sigma V^\top$ with $\rank X = r$.
Then the minimum 2-norm least squares solution to $X\theta = y$ is
\begin{equation}\label{eq:ols_svd_solution}
\hat{\regpar} = V\Sigma^+U^\top y = \sum_{i=1}^r \frac{1}{\sigma_i}(u_i^\top y)v_i
\end{equation}
where $\Sigma^+$ is the Moore-Penrose pseudoinverse of $\Sigma$:
\begin{equation}\label{eq:pseudoinverse}
  \Sigma^+ = \begin{bmatrix}
    \diag(1/\sigma_1, \ldots, 1/\sigma_r) & 0 \\
    0 & 0
  \end{bmatrix} \in \R^{p \times n}.
\end{equation}
\end{theorem}

\begin{proof}
We seek to minimize $\|y - X\regpar\|^2$. We substitute the SVD $X = U\Sigma V^\top$ and obtain
$$\|y - X\regpar\|^2 = \|y - U\Sigma V^\top\regpar\|^2.$$
Since $U$ is orthogonal, $\|U^\top z\| = \|z\|$ for any vector $z$. 
Thus
$$\|y - U\Sigma V^\top\regpar\|^2 = \|U^\top y - \Sigma V^\top\regpar\|^2.$$
Define $\tilde{y} = U^\top y \in \R^n$ and $\alpha = V^\top\regpar \in \R^p$. 
The problem becomes
\begin{equation}\label{eq:pinv_alpha}
\min_{\alpha} \|\tilde{y} - \Sigma\alpha\|^2.
\end{equation}

We partition $\Sigma$, $\tilde{y}$, and $\alpha$ as follows:
$$\Sigma = \begin{bmatrix} \Sigma_r & 0 \\ 0 & 0 \end{bmatrix}, \quad
\alpha = \begin{bmatrix} \alpha_r \\ \alpha_{\perp} \end{bmatrix}, \quad
\tilde{y} = \begin{bmatrix} \tilde{y}_r \\ \tilde{y}_{\perp} \end{bmatrix},$$
where $\Sigma_r = \diag(\sigma_1, \ldots, \sigma_r) \in \R^{r \times r}$.

Then
\begin{align*}
  \|\tilde{y} - \Sigma\alpha\|^2
  &= \left\|\begin{bmatrix} \tilde{y}_r - \Sigma_r\alpha_r \\ \tilde{y}_{\perp} \end{bmatrix}\right\|^2\\
  &= \|\tilde{y}_r - \Sigma_r\alpha_r\|^2 + \|\tilde{y}_{\perp}\|^2.
\end{align*}

The second term is independent of $\alpha$. To minimize the first term,
$$\Sigma_r\alpha_r = \tilde{y}_r \quad \Rightarrow \quad 
\alpha_r = \Sigma_r^{-1}\tilde{y}_r 
= \begin{bmatrix} \tilde{y}_1/\sigma_1 \\ \vdots \\ \tilde{y}_r/\sigma_r \end{bmatrix}.$$

For now the component $\alpha_{\perp}$ is undetermined. We note that setting 
$\alpha_{\perp} = 0$ gives the minimum norm solution to~\eqref{eq:pinv_alpha}.

Transforming back gives
$$
 \regpar = V\alpha = \begin{bmatrix} V_r & V_{\perp} \end{bmatrix} \begin{bmatrix} \Sigma_r^{-1}\tilde{y}_r \\ \alpha_{\perp}  \end{bmatrix}  =  V_r \Sigma_r^{-1}\tilde{y}_r + V_{\perp} \alpha_{\perp}.
$$  
Since $V$ is orthogonal, by Pythagoras the 2-norm of $\regpar$ is minimized
if $V_{\perp} \alpha_{\perp}$ is zero, i.e., if $\alpha_{\perp} =0$.
With this choice we obtain
$$
 \regpar =  V_r \Sigma_r^{-1}\tilde{y}_r  
  = V\Sigma^+U^\top y
  = \sum_{i=1}^r v_i \cdot \frac{1}{\sigma_i} \cdot (u_i^\top y)
  = \sum_{i=1}^r \frac{1}{\sigma_i}(u_i^\top y)v_i.
  $$

\end{proof}

The SVD solution has an appealing geometric interpretation: We project $y$ onto each left singular vector $u_i^\top y$, then scale by $1/\sigma_i$, weight the result by the corresponding right singular vector $v_i$, and sum over all positive singular values. This explicitly constructs the solution as a linear combination of the right singular  vectors, with coefficients determined by the singular values and projections of $y$.

\begin{remark}[\textbf{Minimum 2-norm solution}]
When $\rank(X) = r < p$, the matrix $X^\top X$ is singular and the normal equations have infinitely many solutions. We have seen in Theorem~\ref{th:ls_svd}  that the SVD handles this gracefully. Indeed,  when $\rank(X) = r < p$, the set of least squares solutions is
$$\mathcal{S} = \left\{\regpar \in \R^p : \|y - X\regpar\|^2 = \min_{\regpar} 
\|y - X\regpar\|^2\right\}.$$
An inspection of the proof of Theorem~\ref{th:ls_svd} reveals (namely, setting
$\alpha_{\perp} = 0$) that among all $\regpar \in \mathcal{S}$, 
the SVD solution $\hat{\regpar} = V\Sigma^+U^\top y$ 
has the smallest Euclidean norm, i.e.,
$$\|\hat{\regpar}\| = \min_{\regpar \in \mathcal{S}} \|\regpar\|.$$ 
The minimum-norm solution is not just a mathematical convenience it is often the state that nature itself selects. This reflects a fundamental tendency in physics: among all configurations consistent with the governing constraints, systems gravitate toward states of minimum energy. In this light, the pseudoinverse is not merely a matrix operator, but a mathematical expression of physical equilibrium.
\end{remark}

For dense full-rank least-squares problems, QR factorization is usually the standard method because it is numerically stable and significantly cheaper than the SVD~\cite{Golub_MatrixComputations}. The SVD is more robust, especially when the matrix is ill-conditioned or rank-deficient, because it reveals the numerical rank and naturally produces minimum-norm solutions.

\section{Statistical properties of least squares}

Before exploring the benefits of using the least squares error function from a statistical viewpoint, we will first impose the following {\em standard assumptions} on our linear model $y= X\theta+\eps$, which we will make throughout this section unless otherwise noted:

\begin{enumerate}
\item[] (A1) Exogeneity: $\E[\eps | X] = 0$.
\item[] (A2) Homoscedasticity: $\Var(\epsilon | X) = \sigma^2 I$.
\item[] (A3) Full rank: $\rank(X) = p$ (assumes $n \geq p$).
\end{enumerate}

It is instructive to explain the role of each of these assumptions.

Exogeneity is the most important statistical assumption. 
It stipulates that the errors are uncorrelated with the predictors.
Exogeneity would mean, for instance, that the unmodeled part of the hold-down time in our surfing example is not systematically correlated with the recorded wave height, period, depth, or wind speed.
It also means there is no ``omitted variable bias'', i.e., we have not left out important confounders. Thus, exogeneity is required for unbiasedness and consistency of the least squares solution, as well as for a causal interpretation of coefficients\footnote{Here, a causal interpretation 
means if we were to intervene and change $x_j$ 
while holding other relevant factors fixed, $y$
would change by $\regpar_j$ (on average).}.

Homoscedasticity means the errors have constant variance, i.e.,  the  ``variability'' of the errors around  the regression line is the same for all values of the predictors\footnote{This is clearly a very idealized assumption. A more realistic assumption is heteroscedasticity. For example, homoscedasticity would imply that the variance of income is the same for people with 4 years of education as for people with 16 years. In contrast, heteroscedasticity would indicate that the income variance increases with education level, which is a much more realistic viewpoint since in general highly educated people have more diverse career outcomes.}.
Without homoscedasticity, the least squares estimator may still be unbiased but can be inefficient and produce misleading standard errors.

We already encountered the full rank assumption in the previous section, and have seen that it ensures invertibility of $X^T X$ and uniqueness of the solution. While non-invertibility is not a ``deal-breaker'', enforcing uniqueness of the solution would require additional assumptions and regularization (for instance we can replace the inverse with the Moore-Penrose pseudo-inverse, as previously shown), which we will discuss in more detail in Chapter~\ref{ss:predictionrisk}.

\medskip
The most compelling statistical justification of the least squares error function~\eqref{eq:l2reg} arguably comes from {\em maximum likelihood theory}. 

Let us recall the concept of the likelihood function.
Assume we have a set of i.i.d.\ observations $\mathcal{D} = \{x_1, x_2, \dots, x_n\}$ drawn from a probability density (or mass) function $f(x | \regpar)$. The joint probability of the entire dataset is the product of the individual probabilities
$\P(\mathcal{D} | \regpar) = \prod_{i=1}^n f(x_i | \regpar).$
When we treat the data $\mathcal{D}$ as fixed and the parameter $\regpar$ as a variable, this function is called the {\em likelihood function}, denoted as $\mathcal{L}(\regpar)$,
$$\mathcal{L}(\regpar; x_1, \dots, x_n) = \prod_{i=1}^n f(x_i | \regpar).$$
The Maximum Likelihood Estimator (MLE) $\hat{\regpar}_{MLE}$ is defined as the argument that maximizes the likelihood function, i.e., 
$$\hat{\regpar}_{MLE} := \argmax_{\regpar \in \Theta} \mathcal{L}(\regpar),$$
where $\Theta$ denotes a parameter space.

Under suitable regularity conditions (identifiability, correct model specification, and mild technical assumptions), one can show (using e.g.\ the Uniform Law of Large Numbers and the Argmax Theorem) that the maximum likelihood estimator is {\em consistent}~\cite{wald1949note,van2000asymptotic}, i.e.
$$
\hat{\theta}_n \xrightarrow{P} \theta_0 \quad \text{as } n \to \infty,
$$
where $\theta_0$ is the true parameter and 
$\hat{\theta}_n$ is the  MLE based on $n$ samples.

\smallskip
Notice that to actually compute the MLE numerically, one usually considers the
{\em log-likelihood function} $\ell(\regpar) = \log \mathcal{L}(\regpar)$.
Since the natural logarithm is a monotonically increasing function, the value of $\regpar$ that maximizes $\mathcal{L}(\regpar)$ also maximizes the log-likelihood
$$\ell(\regpar) = \sum_{i=1}^n \log f(x_i | \regpar),$$
which has numerically much more stable properties.
The {\em score} is the gradient of the log-likelihood function with respect to $\regpar$.
To find the MLE, we typically solve the {\em score equation} by setting the gradient of the log-likelihood to zero,
$$\nabla_\regpar \ell(\regpar) = 0.$$

Except for a few cases (like least squares, see also Theorem~\ref{th:mle} below) the MLE typically lacks a closed-form solution. While many standard models (like logistic regression, see Chapter~\ref{s:logisticregression}) remain computationally tractable due to convexity, more complex models result in non-convex optimization problems where finding the global MLE is often computationally intractable.

When we add a distributional assumption on the errors, we can show that
the least squares estimator is actually the maximum likelihood estimator.

\begin{theorem}[Least squares as Maximum Likelihood Estimator]\label{th:mle}
Suppose the errors are i.i.d.\ Gaussian, i.e,
$\epsilon_i \sim \mathcal{N}(0, \sigma^2)$, for $i = 1, \ldots, n$. Under the standard assumptions (A1)--(A3), the maximum likelihood 
estimator (MLE) of $\regpar$ is
$$\hat{\regpar}_{\text{MLE}} = \hat{\regpar}_{\text{LS}} = (X^\top X)^{-1}X^\top y,$$
and the MLE of $\sigma^2$ is
$$\hat{\sigma}^2_{\text{MLE}} = \frac{1}{n}\|y - X\hat{\regpar}_{\text{LS}}\|^2.$$
\end{theorem}

\begin{proof}
The likelihood function is:
\begin{align*}
  L(\regpar, \sigma^2 | y, X)
  &= \prod_{i=1}^n \frac{1}{\sqrt{2\pi\sigma^2}}\exp\left(-\frac{(y_i - x_i^\top\regpar)^2}{2\sigma^2}\right)\\
  &= (2\pi\sigma^2)^{-n/2}\exp\left(-\frac{1}{2\sigma^2}\|y - X\regpar\|^2\right).
\end{align*}

The log-likelihood is:
$$\ell(\regpar, \sigma^2) = -\frac{n}{2}\log(2\pi) - \frac{n}{2}\log(\sigma^2) - \frac{1}{2\sigma^2}\|y - X\regpar\|^2.$$

For fixed $\sigma^2$, maximizing $\ell$ with respect to $\regpar$ is equivalent to 
minimizing $\|y - X\regpar\|^2$, which yields $\hat{\regpar}_{\text{LS}}$.

Substituting $\hat{\regpar}_{\text{LS}}$ and maximizing with respect to $\sigma^2$:
$$\frac{\partial\ell}{\partial\sigma^2} = -\frac{n}{2\sigma^2} + \frac{1}{2(\sigma^2)^2}\|y - X\hat{\regpar}_{\text{LS}}\|^2 = 0$$
gives
\begin{equation}\label{eq:biasedMSE}
\hat{\sigma}^2_{\text{MLE}} = \frac{1}{n}\|y - X\hat{\regpar}_{\text{LS}}\|^2,
\end{equation}
and the proof is complete.

\end{proof}

Note that the MLE~\eqref{eq:biasedMSE} is biased; the unbiased estimator uses $n-p$ instead of $n$ in the denominator (see also Exercise~\ref{ex:mse_unbiased}).

\begin{remark}
The Gaussian assumption is often reasonable for errors 
that arise from the sum of many small independent sources (by the Central Limit Theorem).
But what if the error $\eps$ is not Gaussian? If the error follows a different distribution, the MLE leads to a different loss function.
For example for Laplace errors ($\epsilon_i \sim \text{Laplace}(0, b)$) this results in minimizing $\sum_i |y_i - x_i^\top\regpar|$, known as $\ell_1$ regression. And for uniform errors ($\epsilon_i \sim \text{Uniform}(-a, a)$), we minimize $\max_i |y_i - x_i^\top\regpar|$, also referred to as Chebyshev regression.
\end{remark}

\begin{theorem}\label{th:squared_optimal}
Suppose $Y$ is a random variable and we wish to predict $Y$ with a constant $c \in \R$. 
Among all possible choices of $c$, the minimizer of the expected squared error is:
$$c^* = \argmin_{c \in \R} \E[(Y - c)^2] = \E[Y].$$

More generally, for conditional prediction given $x$, the function that minimizes 
expected squared error is:
$$f^*(x) = \argmin_{f} \E[(Y - f(x))^2 | x] = \E[Y | x].$$
\end{theorem}

\begin{proof}
For the unconditional case we compute
\begin{align*}
  \E[(Y - c)^2]
  &= \E[(Y - \E[Y] + \E[Y] - c)^2]\\
  &= \E[(Y - \E[Y])^2] + 2\E[(Y - \E[Y])](\E[Y] - c) + (\E[Y] - c)^2\\
  &= \Var(Y) + 0 + (\E[Y] - c)^2.
\end{align*}
The last expression is minimized when $c = \E[Y]$.

For the conditional case, the proof is identical but conditioning on $x$ throughout.

\end{proof}

Theorem~\ref{th:squared_optimal} tells us that if we believe $\E[Y | x] = x^\top\regpar$, 
then minimizing the squared error is the natural way to estimate $\regpar$; we are finding the 
linear function that best approximates the conditional mean. We leave it to the reader to verify that using instead the sum of absolute errors estimates the conditional median.

\begin{theorem}\label{th:ols_properties}
Under the standard assumptions (A1)--(A3), the least squares estimator $\hat{\regpar}_{\text{LS}}$ satisfies
\begin{enumerate}
  \item[(i)] {\em Unbiasedness:} $\E[\hat{\regpar}_{\text{LS}} | X] = \regpar$.
  \item[(ii)] {\em Covariance:} $\Var(\hat{\regpar}_{\text{LS}} | X) = \sigma^2(X^\top X)^{-1}$.
\end{enumerate}
\end{theorem}

\begin{proof}
To prove the unbiasedness condition, we compute
\begin{align*}
  \E[\hat{\regpar}_{\text{LS}} | X]
  &= \E[(X^\top X)^{-1}X^\top y | X]\\
  &= (X^\top X)^{-1}X^\top\E[y | X]\\
  &= (X^\top X)^{-1}X^\top X\regpar\\
  &= \regpar.
\end{align*}

The covariance property follows from the following calculation:
\begin{align*}
  \Var(\hat{\regpar}_{\text{LS}} | X)
  &= \Var((X^\top X)^{-1}X^\top y | X)\\
  &= (X^\top X)^{-1}X^\top \Var(y | X) X(X^\top X)^{-1}\\
  &= (X^\top X)^{-1}X^\top (\sigma^2I_n) X(X^\top X)^{-1}\\
  &= \sigma^2(X^\top X)^{-1}X^\top X(X^\top X)^{-1}\\
  &= \sigma^2(X^\top X)^{-1}.
\end{align*}

\end{proof}

A natural question is whether the least squares estimator is optimal among all \emph{unbiased} linear estimators?

\begin{definition}[Linear Estimator]\label{def:linear_estimator}
An estimator $\tilde{\regpar}$ is \emph{linear} if it can be written as $\tilde{\regpar} = B y$
for some matrix $B \in \R^{p \times n}$ that may depend on $X$ but not on $y$.
\end{definition}

\begin{theorem}[Gauss-Markov Theorem]\label{th:gauss_markov}
Under the standard assumptions (A1)--(A3), the least squares estimator $\hat{\regpar}_{\text{LS}}$ is the 
{\em Best Linear Unbiased Estimator} (BLUE). That is, among all linear unbiased 
estimators $\tilde{\regpar} = By$, the least squares estimator has the smallest 
covariance matrix
$$\Var(\tilde{\regpar} | X ) - \Var(\hat{\regpar}_{\text{LS}} | X ) \succeq 0$$
for any linear unbiased estimator $\tilde{\regpar}$.

Equivalently, for any linear combination $a^\top\regpar$, the least squares-based estimator 
$a^\top\hat{\regpar}_{\text{LS}}$ has the smallest variance among all linear unbiased 
estimators of $a^\top\regpar$.
\end{theorem}

\begin{proof}
Let $\tilde{\regpar} = By$ be any linear unbiased estimator. Unbiasedness requires:
$$\E[\tilde{\regpar} | X ] = B\E[y | X ] = BX \regpar = \regpar$$
for all $\regpar \in \R^p$. This implies $BX  = I_p$.

The covariance of $\tilde{\regpar}$ is
\begin{align*}
  \Var(\tilde{\regpar} | X )
  &= B\Var(y | X )B^\top\\
  &= \sigma^2BB^\top.
\end{align*}

Decompose $B$ as
$$B = (X^\top X )^{-1}X^\top + C$$
where $C = B - (X^\top X)^{-1} X ^\top$.

The constraint $B X  = I_p$ implies:
$$C X  = B X  - ( X^\top X )^{-1} X^\top X = I_p - I_p = 0.$$

Now compute
\begin{align*}
  \Var(\tilde{\regpar} | X)
  &= \sigma^2[(X^\top X)^{-1}X^\top + C][(X^\top X)^{-1}X^\top + C]^\top\\
  &= \sigma^2[(X^\top X)^{-1} 
     + (X^\top X)^{-1}X^\top C^\top 
     + CX(X^\top X)^{-1} 
     + CC^\top]\\
  &= \sigma^2[(X^\top X)^{-1} + CC^\top]\\
  &= \Var(\hat{\regpar}_{\text{LS}} | X) + \sigma^2 CC^\top.
\end{align*}

Since $CC^\top \succeq 0$ (positive semi-definite), we have:
$$\Var(\tilde{\regpar} | X) \succeq \Var(\hat{\regpar}_{\text{LS}} | X)$$
with equality if and only if $C = 0$, i.e., $\tilde{\regpar} = \hat{\regpar}_{\text{LS}}$.

\end{proof}

\section{Bias, variance, and risk}
\label{ss:predictionrisk}

A fundamental tension in statistical estimation is the bias-variance tradeoff: 
reducing variance often increases bias, and vice versa.
In fact, the bias–variance tradeoff is a fundamental identity of squared-error prediction and randomness in estimation, and not specific to linear models. It arises from basic properties of expectations and variance.

\begin{definition}[Mean Squared Error]\label{def:mse}
For an estimator $\hat{\regpar}$ of $\regpar$, the \emph{mean squared error (MSE)} is defined as
$$\MSE(\hat{\regpar}) = \E[\|\hat{\regpar} - \regpar\|^2].$$
\end{definition}

Also recall that 
$\Bias(\hat{\theta}) = \E[\hat{\theta}] - \theta$ and 
$\Cov(\hat{\theta}) = \E\left[ (\hat{\theta} - \E[\hat{\theta}])(\hat{\theta} - \E[\hat{\theta}])^T \right]$.

\smallskip

\begin{theorem}[Bias-variance decomposition for parameter estimation]\label{th:bias_variance}
Let $\theta \in \mathbb{R}^p$ be a fixed, unknown parameter vector. Let $\hat{\theta}$ be an estimator of $\theta$, which is a random variable depending on the training data $\mathcal{D}$. Assume that 
$\E [ \|\hat{\theta}\|^2 ] < \infty$. Then
\begin{equation}\label{eq:bias_variance_param}
\MSE(\hat{\theta}) = \|\Bias(\hat{\theta})\|_2^2 + \tr(\Cov(\hat{\theta})).
\end{equation}

\end{theorem}

\begin{proof}
Let $\bar{\theta} = \E[\hat{\theta}]$ denote the expected value of the estimator. We compute
\begin{align*}
\MSE(\hat{\theta}) & = \E \left[ \| (\hat{\theta} - \bar{\theta}) + (\bar{\theta} - \theta) \|_2^2 \right] \\ & = \E \left[ \|\hat{\theta} - \bar{\theta}\|_2^2 + \|\bar{\theta} - \theta\|_2^2 + 2(\hat{\theta} - \bar{\theta})^T (\bar{\theta} - \theta) \right]\\
 & = \E [ \|\hat{\theta} - \bar{\theta}\|_2^2 ] + \E [ \|\bar{\theta} - \theta\|_2^2 ] + 2\E [ (\hat{\theta} - \bar{\theta})^T (\bar{\theta} - \theta) ].
\end{align*}
We first consider the variance term $\E [ \|\hat{\theta} - \bar{\theta}\|_2^2 ]$.
By the property of the trace and the definition of the covariance matrix we get
$$\E [ \|\hat{\theta} - \bar{\theta}\|_2^2 ] = \E [ \tr((\hat{\theta} - \bar{\theta})(\hat{\theta} - \bar{\theta})^T) ] = \tr(\E [ (\hat{\theta} - \bar{\theta})(\hat{\theta} - \bar{\theta})^T ]) = \tr(\Cov(\hat{\theta})).$$

We now proceed to the bias term $\E [ \|\bar{\theta} - \theta\|_2^2 ]$.
Since both $\bar{\theta}$ and $\theta$ are deterministic, the expectation is simply the value itself, i.e.,
$$E [ \|\bar{\theta} - \theta\|_2^2 ] = \|\bar{\theta} - \theta\|_2^2 = \|\Bias(\hat{\theta})\|_2^2.$$

Regarding the cross term $2\E [ (\hat{\theta} - \bar{\theta})^T (\bar{\theta} - \theta) ]$ we note that since $(\bar{\theta} - \theta)$ is a constant vector, we can pull it out of the expectation and get
$$2\E [ \hat{\theta} - \bar{\theta} ]^T (\bar{\theta} - \theta) = 2(\E[\hat{\theta}] - \bar{\theta})^T (\bar{\theta} - \theta).$$
Since $\E[\hat{\theta}] = \bar{\theta}$, the term $\E[\hat{\theta}] - \bar{\theta} = 0$. Thus, the cross-term vanishes.
Summing the remaining terms yields the desired decomposition
$$\MSE(\hat{\theta}) = \|\Bias(\hat{\theta})\|_2^2 + \tr(\Cov(\hat{\theta})).$$

To complete the proof we note that our assumption ensures that $\hat{\regpar}$ has finite first and second moments, 
so all expectations are well-defined and finite.
\end{proof}

We point out that Theorem~\ref{th:bias_variance} does allow for {\em heteroscedasticity} and for {\em correlated errors}, Moreover, it does not impose 
any distributional assumptions beyond zero-mean and finite second moments. Thus, the decomposition holds for \emph{any} estimator under very mild conditions. This 
generality makes it a fundamental result in statistical learning theory.

While Theorem~\ref{th:bias_variance} concerns estimating the parameter vector $\regpar$, 
in practice we often care about \emph{predicting} new observations. The bias-variance 
decomposition for prediction includes an additional irreducible error term.

\begin{corollary}[Bias-variance decomposition for prediction]\label{cor:bias_variance_prediction}
Assume we are given a training set $\mathcal{D} = \{(x_i, y_i)\}_{i=1}^n$ where $y_i = f(x_i) + \epsilon$ is the target variable, $f$ is the true (but unknown) deterministic function, and $\epsilon$ is random noise  with $\E[\epsilon] = 0$ and $\Var(\epsilon) = \sigma^2$. Assume that $\epsilon$ is independent of the training data.
Let $\hat{f}(x; \mathcal{D})$ be an estimator (a model) trained on $\mathcal{D}$. Then,
expected squared error of the estimator at a fixed point $x$, taken over the distribution of all possible datasets $\mathcal{D}$, is given by
\begin{equation}\label{eq:prediction_mse}
\E_\mathcal{D} \left[ (y - \hat{f}(x; \mathcal{D}))^2 \right] = \Bias[\hat{f}(x)]^2 + \Var[\hat{f}(x)] + \sigma^2.
\end{equation}

\end{corollary}

\begin{proof}
To simplify notation, let $y = f + \epsilon$, $\hat{f} = \hat{f}(x; \mathcal{D})$, and $\bar{f} = \E_\mathcal{D}[\hat{f}]$. Note that $f$ is deterministic, so $\E_\mathcal{D}[f] = f$.

The objective is $\E_\mathcal{D}[(y - \hat{f})^2]$. Substituting $y = f + \epsilon$ we obtain

\begin{align*}
\E_\mathcal{D}[(f + \epsilon - \hat{f})^2] & = \E_\mathcal{D}[((f - \hat{f}) + \epsilon)^2] \\
& = \E_\mathcal{D}[(f - \hat{f})^2 + 2\epsilon(f - \hat{f}) + \epsilon^2] \\
& = \E_\mathcal{D}[(f - \hat{f})^2] + 2\E_\mathcal{D}[\epsilon(f - \hat{f})] + \E_\mathcal{D}[\epsilon^2].
\end{align*}

Since the noise $\epsilon$ is independent of the training data $\mathcal{D}$ it is also independent of $\hat{f}$). Hence,
$$\E_\mathcal{D}[\epsilon(f - \hat{f})] = \E[\epsilon] \cdot \E_\mathcal{D}[f - \hat{f}] = 0 \cdot (f - \bar{f}) = 0$$
Also, $\E_\mathcal{D}[\epsilon^2] = \Var(\epsilon) = \sigma^2$. Thus, our expression simplifies to
\begin{equation}\label{eq:biasvar5}
\E_\mathcal{D}[(y - \hat{f})^2] = \E_\mathcal{D}[(f - \hat{f})^2] + \sigma^2.
\end{equation}

Now we focus on $\E_\mathcal{D}[(f - \hat{f})^2]$. We have 

$$\E_\mathcal{D}[(f - \hat{f})^2] = \E_\mathcal{D}[(f - \bar{f} + \bar{f} - \hat{f})^2]= \E_\mathcal{D}[(f - \bar{f})^2 + 2(f - \bar{f})(\bar{f} - \hat{f}) + (\bar{f} - \hat{f})^2].$$

Note that 
$$\E_\mathcal{D}[(f - \bar{f})^2] = (f - \bar{f})^2 = \Bias^2.$$
Also, 
$$\E_\mathcal{D}[2(f - \bar{f})(\bar{f} - \hat{f})] = 2(f - \bar{f}) \E_\mathcal{D}[\bar{f} - \hat{f}] = 0,$$ 
since $\E_\mathcal{D}[\bar{f} - \hat{f}] = \bar{f} - \E_\mathcal{D}[\hat{f}] = \bar{f} - \bar{f} = 0$.

Finally, 
$$\E_\mathcal{D}[(\bar{f} - \hat{f})^2] = \Var [\hat{f}(x)]$$.

Substituting these back into~\eqref{eq:biasvar5} yields
$$\E_\mathcal{D}[(y - \hat{f})^2] = (f - \bar{f})^2 + \E_\mathcal{D}[(\bar{f} - \hat{f})^2] + \sigma^2,$$
which proves our claim.
\end{proof}

The decomposition~\eqref{eq:prediction_mse} in Corollary~\ref{cor:bias_variance_prediction} reveals three sources of prediction error:
\begin{enumerate}
  \item Bias:
    Systematic error from using a biased estimator or misspecified model. For least squares, 
    this is zero since $\E[\hat{\regpar}_{\text{LS}}] = \regpar$. For ridge regression (discussed below), this 
    is nonzero but can be outweighed by variance reduction.
   \item Variance:
    Random error from the particular training sample. Predictions vary across 
    different training datasets. This term is larger when
  $x_{\text{new}}$ is far from the training data (extrapolation), the design matrix $X$ is ill-conditioned, or the sample size $n$ is small relative to dimension $p$.
    \item Irreducible error:
    Fundamental noise in the response variable that cannot be eliminated by any 
    estimator. This represents the intrinsic randomness in $y$ that cannot be 
    explained by $x$, even with infinite data.
\end{enumerate}

The bias-variance tradeoff refers to the tension between the first two terms. This brings up the question whether we can accept some bias to achieve 
substantial variance reduction to achieve lower total prediction error? This is indeed possible and can be accomplished via proper {\em regularization}.

For example, for the case of linear regression, when the matrix $X$ is ill-conditioned, the least squares solution remains unbiased but can have extremely high variance.  Historically, the focus was on unbiasedness, because datasets often had rather small values of $p$ compared to $n$, making ill-conditionedness of $X$ a much less common scenario. The ill-conditionedness of $X$
happens typically more often when the number of variables $p$ is comparable to the sample size $n$ (here, we still assume $p \le n$)\footnote{For example if $X$ is a Gaussian random matrix, we will see in Chapter~\ref{c:probability-gaussiananalysis} that the condition number of $X$ grows quickly as $p \to n$. 
Indeed, let $\gamma = p/n$, then we have that $\kappa(X) \approx \frac{\sqrt{n}+\sqrt{p}}{\sqrt{n}-\sqrt{p}}$. See~Theorem\ref{thm:Gordon:afterGconcentration} for a precise statement.}. In this regime, minimizing prediction error rather than preserving unbiasedness becomes central, and regularization, such as {\em ridge regression}, improves performance by trading bias for variance reduction.
We will explore this topic in Chapter~\ref{ss:ridge}. 

\medskip

Corollary~\ref{cor:bias_variance_prediction} also forms the basis for constructing prediction intervals, see e.g.~\cite{seber2003linear,hastieelements} for  details. Specializing to the case of a linear model $y = X\theta + \eps$, if the model is correctly specified and the estimator is {\em unbiased}, then
under Gaussian errors a $(1-\alpha)$ prediction interval for $y_{\text{new}}$ is
\begin{equation}\label{eq:predictioninterval}   \hat{y}_{\text{new}} \pm t_{1-\alpha/2, n-p} \cdot \sqrt{\hat{\sigma}^2 + \hat{\sigma}^2 \, x_{\text{new}}^\top(X^\top X)^{-1}x_{\text{new}}},
\end{equation}
where $\hat{\sigma}^2 = \frac{1}{n-p}\|y - X\hat{\regpar}\|^2$ and $t_{1-\alpha/2, n-p}$ is the critical value of the $t$-distribution.
Note that the first term under the square root
in~\eqref{eq:predictioninterval} accounts for
the irreducible error and the second term for parameter uncertainty.
We emphasize that the probability is in the
{\em procedure}, not in a {\em single realization}. Namely, if we repeatedly collect a training dataset, fit a model (with the same method), construct a prediction interval, and
observe a new data point, then on average, 
$(1-\alpha) \times 100\%$ of these intervals will contain the new observation. 
We emphasize that for a biased estimator (such as ridge regression or lasso) constructing prediction intervals is more involved~\cite{efron1994introduction,hastieelements}.

\medskip
\noindent
\textbf{Generalization gap.}
In statistical learning theory, the bias-variance tradeoff tells us how the components of {\em risk} behave as model complexity changes. Let us recall the notion of  risk, which represents the average cost incurred by a predictive model when applied to the entire population of possible data points. Formally, we define risk as follows:
\begin{definition}\label{de:risk}
Let $\mathcal{X}$ be the input space and $\mathcal{Y}$ the output space. Let $P$ be a joint probability distribution over $\mathcal{X} \times \mathcal{Y}$. Given a decision function (or hypothesis) $f: \mathcal{X} \to \mathcal{Y}$ and a measurable loss function $L: \mathcal{Y} \times \mathcal{Y} \to \mathbb{R}_{\geq 0}$, the Risk $R(f)$ is defined as the expectation of the loss,
$$R(f) = \mathbb{E}_{(X, Y) \sim P} [L(Y, f(X))].$$
\end{definition}

Using the definition of expectation, the risk can be written as the integral over the joint distribution
$$R(f) = \int_{\mathcal{X} \times \mathcal{Y}} L(y, f(x)) \, dP(x, y).$$

In data science, the distribution $P$ is almost always unknown. This leads to the distinction that is central to learning theory:
\begin{itemize}
\item Theoretical Risk ($R(f)$): This is the risk we {\em want to minimize} but cannot calculate in practice (also known as {\em True Risk} or {\em Population Risk}). 
\item Empirical Risk ($\hat{R}_n(f)$): This is what we {\em actually minimize} using a training set of size $n$; it is given by
$$\hat{R}_n(f) = \frac{1}{n} \sum_{i=1}^n L(y_i, f(x_i)).$$
\end{itemize}

The {\em generalization gap} is the difference between these two, $|R(f) - \hat{R}_n(f)|$. An important part of theory in data science and machine learning is dedicated to bounding this gap to ensure that a model that performs well on training data will also perform well ``in the wild''.

Sometimes it is stated in the literature that the generalization gap is the difference between test error and training error. However, the test error  is only an estimator of the true risk. Hence, while the test error minus training error is a standard proxy, the true generalization gap is defined against the population risk, and conflating the two can obscure the effects of test set noise, data reuse, and distribution shift.

\medskip
\noindent
\textbf{Model mismatch, underfitting, overfitting:}
In reality, the prediction error decomposition~\eqref{eq:prediction_mse} is almost always wrong, since the true data-generating process does rarely perfectly fit the chosen model. For instance in case of linear regression,  
when the underlying data is generated by a process that is more complex than a linear model (as is  the case with the surfing/hold-down time example used in the beginning of this chapter), we encounter {\em model mismatch}, also known as {\em specification error} or {\em structural bias}.

Model mismatch is closely related to the issue of underfitting/overfitting, but they are not identical.
Model mismatch is a {\em design-time issue}: the chosen model class does not contain a function capable of representing the true relationship in the data. Even with infinite data, the best possible predictor within the model class cannot recover the true signal.

Underfitting, in contrast, is a {\em run-time phenomenon}: the fitted model fails to capture the structure of the data. This may occur because the model class is too restrictive (model mismatch), for example fitting a line to a quadratic relationship. But it can also arise when the estimation procedure overly constrains the parameters, for example through strong regularization or insufficient training. Underfitting typically corresponds to large approximation bias.
Overfitting occurs when the model class is too flexible, and thus the estimator fits noise in finite samples. Examples include using a high-degree polynomial or  linear regression when $p > n$ without regularization (see Chapter~\ref{ss:ridge}).
Overfitting manifests as large variance.

\section{Regularization and ridge regression}\label{ss:ridge}

Assume we want to predict the optimal drug dosage for a patient using the expression levels of thousands of genes. In such problems the number of features (genes) can be far larger than the number of patients in the study, leading to a high-dimensional regression problem with $p \gg n$.

This problem falls into the area of  {\em high-dimensional linear regression} (assuming one considers the underlying model to be linear). More generally, high-dimensional linear regression refers broadly to regimes where $p$ is comparable to or larger than $n$. This includes the cases $p > n$, $p \gg n$, and $p/n \to \gamma > 0$.

Standard least squares works well when the number of predictors $p$ is  small relative to the sample size $n$ and the design matrix $X$ is well-conditioned. 
However, when predictors are highly correlated or $p$ approaches $n$, least squares becomes 
unstable, since small changes in the data can lead to large changes in coefficient estimates. 
In the  case $p > n$, 
$X^\top X$ is singular, and the minimizer is not unique.

\smallskip

{\em Ridge regression}\footnote{The term was introduced by Hoerl and Kennard in 1970, who used the word ``ridge'' specifically to refer to this flat elongated trough in the least squares objective surface that arises under multi-collinearity. They called their technique ridge regression because it was designed to deal with that ridge.} addresses these problems by adding a penalty term that shrinks coefficients toward zero, trading a small amount of bias for a potentially substantial variance 
reduction~\cite{hoerl1970ridge,hastieelements}. In numerical analysis and optimization ridge regression is also known as {\em Tikhonov regularization} (named for Andrey Tikhonov~\cite{tikhonov1963solution}).

\begin{definition}[Ridge Regression]\label{def:ridge}
Given observations $\{(x_i, y_i)\}_{i=1}^n$, the \emph{ridge regression estimator} 
with regularization parameter $\lambda \geq 0$ is
\begin{equation}\label{eq:ridge_def}
  \hat{\regpar}_{\text{ridge}}(\lambda) = \argmin_{\regpar \in \R^p} 
  \left\{\|y - X\regpar\|^2 + \lambda\|\regpar\|^2\right\}.
\end{equation}
\end{definition}

The ridge objective balances two competing goals:
\begin{itemize}
  \item {\em Fit the data:} Minimize $\|y - X\regpar\|^2$,
  \item {\em Keep coefficients small:} Minimize $\|\regpar\|^2$ (regularization).
\end{itemize}
The {\em regularization parameter} $\lambda$ controls the trade-off between these two goals.

Unlike many regularized methods, ridge regression has an explicit solution.

\begin{theorem}[Ridge Regression Solution]\label{thm:ridge_solution}
The ridge regression estimator has the closed-form solution:
\begin{equation}\label{eq:ridge_normal}
  \hat{\regpar}_{\text{ridge}}(\lambda) = (X^\top X + \lambda I_p)^{-1}X^\top y.
\end{equation}
Furthermore, $X^\top X + \lambda I_p$ is invertible for any $\lambda > 0$, 
even when $X^\top X$ is singular (i.e., when $p > n$ or when $X$ has 
collinear columns).
\end{theorem}

\begin{proof}
The objective function is
$$f(\regpar) = \|y - X\regpar\|^2 + \lambda\|\regpar\|^2 
= (y - X\regpar)^\top(y - X\regpar) + \lambda\regpar^\top\regpar.$$

Expanding this expression gives
$$f(\regpar) = y^\top y - 2\regpar^\top X^\top y + \regpar^\top X^\top X\regpar + \lambda\regpar^\top\regpar.$$
We take the gradient 
$$\nabla_{\regpar} f(\regpar) = -2X^\top y + 2X^\top X\regpar + 2\lambda\regpar 
= 2(X^\top X + \lambda I_p)\regpar - 2X^\top y,$$
and set it to zero,
$$(X^\top X + \lambda I_p)\regpar = X^\top y.$$
For $\lambda > 0$, the matrix $X^\top X + \lambda I_p$ is positive definite, i.e.,
$$\langle \regpar, (X^\top X + \lambda I_p)\regpar \rangle = \|X\regpar\|^2 + \lambda\|\regpar\|^2 > 0$$
for all $\regpar \neq 0$. Hence, we can compute 
$$\hat{\regpar}_{\text{ridge}}(\lambda) = (X^\top X + \lambda I_p)^{-1}X^\top y.$$
The Hessian $\nabla^2 f(\regpar) = 2(X^\top X + \lambda I_p) \succ 0$ , which confirms that $\hat{\regpar}_{\text{ridge}}(\lambda)$ is indeed a minimum.
\end{proof}

When $X$ has full column rank, the least squares solution is given by
$\hat{\regpar}_{\text{LS}} = (X^\top X)^{-1}X^\top y$.
Ridge regression modifies this by adding $\lambda I_p$ to $X^\top X$, which stabilizes the inversion when eigenvalues of $X^\top X$ are small and ensures invertibility even when $X^\top X$ is singular.

The SVD provides deep insight into how ridge regression works.

\begin{theorem}[Ridge Regression via SVD]\label{thm:ridge_svd}
Let $X = U\Sigma V^\top$ be the SVD of $X \in \R^{n \times p}$ 
with singular values $\sigma_1 \geq \sigma_2 \geq \cdots \geq \sigma_r > 0$ where 
$r = \rank(X)$. Write $U = [u_1, \ldots, u_n]$ and 
$V = [v_1, \ldots, v_p]$.
Then,
\begin{equation}\label{eq:ridge_svd}
  \hat{\regpar}_{\text{ridge}}(\lambda) = \sum_{i=1}^r \frac{\sigma_i}{\sigma_i^2 + \lambda}
  (u_i^\top y)v_i
  = VD_\lambda U^\top y,
\end{equation}
where $D_\lambda = \diag\left(\frac{\sigma_1}{\sigma_1^2 + \lambda}, \ldots, 
\frac{\sigma_r}{\sigma_r^2 + \lambda}\right)$.
\end{theorem}

\begin{proof}
Starting from the ridge solution $\hat{\regpar}_{\text{ridge}} = (X^\top X + \lambda I_p)^{-1}X^\top y$ we have
\begin{align}
  \hat{\regpar}_{\text{ridge}}(\lambda)
  &= (V\Sigma^\top\Sigma V^\top + \lambda I_p)^{-1}V\Sigma^\top U^\top y \notag\\
  &= V(\Sigma^\top\Sigma + \lambda I_p)^{-1}\Sigma^\top U^\top y.\label{eq:ridgeSVD1}
\end{align}

Since $\Sigma^\top\Sigma = \diag(\sigma_1^2, \ldots, \sigma_r^2, 0, \ldots, 0)$, it follows that
$$\Sigma^\top\Sigma + \lambda I_p = 
\diag(\sigma_1^2 + \lambda, \ldots, \sigma_r^2 + \lambda, \lambda, \ldots, \lambda).$$

Therefore,
$$(\Sigma^\top\Sigma + \lambda I_p)^{-1}\Sigma^\top
= \diag\left(\frac{\sigma_1}{\sigma_1^2 + \lambda}, \ldots, 
\frac{\sigma_r}{\sigma_r^2 + \lambda}, 0, \ldots, 0\right).$$

Substituting this expression into~\eqref{eq:ridgeSVD1} gives
\begin{equation}\label{eq:ridgeSVD2}
\hat{\regpar}_{\text{ridge}}(\lambda) = VD_\lambda U^\top y
= \sum_{i=1}^r \frac{\sigma_i}{\sigma_i^2 + \lambda}(u_i^\top y)v_i.
\end{equation}

\end{proof}

It is instructive to compare~\eqref{eq:ridgeSVD2} with the least squares solution (when $r = p$):
\begin{equation}\label{eq:LS_svd}
  \hat{\regpar}_{\text{LS}} = \sum_{i=1}^p \frac{1}{\sigma_i}(u_i^\top y)v_i.
\end{equation}
 In the surfing exmaple, if wave height and wave period are highly correlated in the dataset, ridge regression prevents the fitted coefficients from becoming unstable by shrinking poorly determined combinations of features.

So, how does ridge regression fare with respect to the bias-variance tradeoff? The following result is straightforward (the proof is left to the reader as Exercise~\ref{ex:ridge_bias}).

\begin{proposition}\label{prop:ridge_bias}
Under the assumptions (A1) and (A2), the ridge regression estimator $\hat{\regpar}_{\text{ridge}}(\lambda)$ has the following properties:
\begin{align*}
\Bias(\hat{\regpar}_{\text{ridge}}(\lambda)) &  = \left[(X^\top X + \lambda I)^{-1}X^\top X - I\right]\theta, \\
\tr(\Cov(\hat{\regpar}_{\text{ridge}}(\lambda))) & = \tr\big((X^\top X + \lambda I)^{-1}X^\top \cdot \sigma^2 I \cdot X(X^\top X + \lambda I)^{-1}\big).
\end{align*}
Furthermore, let the SVD of $X$ be $X=U \Sigma V^\top$ with $\rank X = r$, and write  $\tilde\theta = V^\top\theta$, then the MSE of the estimator $\hat{\regpar}_{\text{ridge}}(\lambda)$ is given by 
$$\text{MSE}(\hat{\regpar}_{\text{ridge}}(\lambda)) = \underbrace{\lambda^2\sum_{i=1}^r \frac{\tilde\theta_i^2}{(\sigma_i^2+\lambda)^2}}_{\text{bias}^2} + \underbrace{\sigma^2\sum_{i=1}^r \frac{\sigma_i^2}{(\sigma_i^2+\lambda)^2}}_{\text{variance}}.$$

\end{proposition}

Unlike in Theorem~\ref{th:ols_properties}, we do not require that $X$ has full rank (assumption (A3)), since being able to handle the rank-deficient case is exactly a key motivation for introducing ridge regression in the first place.

\begin{figure}
\centering
\includegraphics[width=0.85\textwidth]{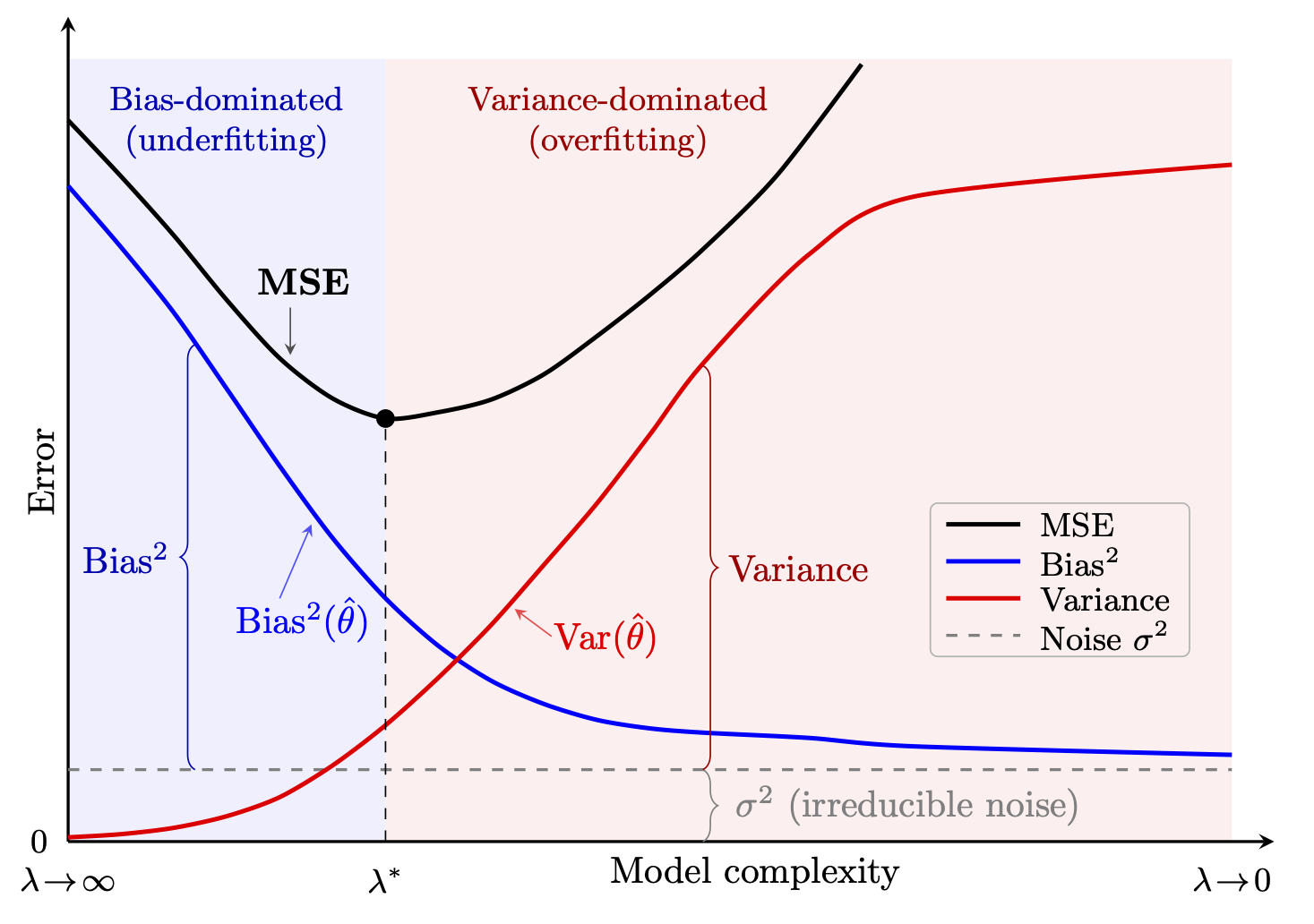}
\caption{ The bias-variance tradeoff of the prediction error as a function of model complexity
(equivalently, decreasing regularization strength $\lambda^{-1}$).
The MSE (black) decomposes as $\mathrm{MSE} = \mathrm{Bias}^2(\hat\theta)
                + \mathrm{Var}(\hat\theta) + \sigma^2$,
where $\sigma^2$ is the irreducible noise floor (dashed).
The squared bias (blue) decreases monotonically with complexity;
the variance (red) increases monotonically. The optimal regularization $\lambda^*$  is located at the \emph{minimum of the MSE curve}, which occurs where $-d\,\mathrm{Bias}^2/d\lambda = d\,\mathrm{Var}/d\lambda$ (the slopes cancel).}
\label{fig:biasvariance}
\end{figure}

\medskip
\noindent
\textbf{Choosing the regularization parameter:}
Proposition~\ref{prop:ridge_bias} suggests that by 
choosing an appropriate value for the regularization parameter $\lambda$ we can achieve a good balance between overfitting and underfitting of the solution.
See Figure~\ref{fig:biasvariance} for an illustration of this bias-variance tradeoff for the prediction error and the interplay between regularization and over/underfitting.
The optimal choice for $\lambda$ via the minimum of the MSE, indicated in Figure~\ref{fig:biasvariance}, is unfortunately not computable in practice, since it would require access to the true solution $\theta$. This begs the question 
how to find $\lambda$ in practice.

{\em Cross-validation}, a widely used technique in data science and machine learning for hyperparameter selection, is a standard way to determine a good choice for $\lambda$. We describe the cross-validation approach in the context of ridge regression, but it should be understood that this technique is more broadly applicable.

\begin{definition}[$K$-fold cross-validation for ridge regression]\label{def:cv_ridge}
To select $\lambda$:
\begin{enumerate}
  \item Partition the data into $K$ folds $\mathcal{F}_1, \ldots, \mathcal{F}_K$ 
        (typically $K = 5$ or $10$) by generating a random permutation of all indices, and then dividing the permuted indices into $K$ contiguous, disjoint blocks. Each block becomes one fold.
  \item For each candidate value $\lambda \in \{\lambda_1, \ldots, \lambda_m\}$ 
        and each fold $k=1,\dots,K$:
    \begin{itemize}
      \item[(i)] Train ridge regression on all data except fold $k$
      \item[(ii)] Compute the prediction error on fold $k$: 
            $$\text{MSE}_k(\lambda) = \frac{1}{|\mathcal{F}_k|}\sum_{i \in \mathcal{F}_k}(y_i - \hat{y}_i(\lambda))^2.$$
    \end{itemize}
  \item Average over folds: $\text{CV}(\lambda) = \frac{1}{K}\sum_{k=1}^K \text{MSE}_k(\lambda)$
  \item Choose: $\lambda^* = \argmin_{\lambda} \text{CV}(\lambda)$
\end{enumerate}
\end{definition}

\medskip

If we know that the noise is Gaussian and we have a good estimate for the noise variance $\sigma^2$, instead of the rather expensive cross validation we can use  Stein’s unbiased risk estimate (SURE). SURE  provides an analytic estimate of prediction risk under Gaussian noise~\cite{stein1981estimation}
and can therefore be used to select hyperparameters such as the ridge penalty or the lasso regularization parameter (see~\eqref{eq:lasso1}) by minimizing the SURE value. 
While SURE often has lower variance than cross-validation and uses the full dataset, it relies on stronger assumptions---besides assuming Gaussian noise and knowledge of the noise variance, it imposes  a certain smoothness condition (weak differentiability) on the estimator.  Cross-validation, though computationally heavier, is thus more broadly applicable.

The work of Donoho and Johnstone~\cite{donoho1995adapting} was 
instrumental in moving SURE from a theoretical result in statistics into a cornerstone of signal and image processing, in particular for data denoising (see also~\cite{blu2007sure}).

Stein's SURE has its roots in his earlier work~\cite{stein1956inadmissibility,james1961estimation}. In fact, while the Gauss-Markov theorem states that the least squares estimator is the best linear unbiased estimator, Charles Stein proved  in a seminal paper~\cite{stein1956inadmissibility} the startling result that in dimensions $p\ge 3$, the least squares estimator is ``inadmissible''. This result, also known as {\em Stein's Paradox}, means there exists a biased ``shrinkage'' estimator---the James-Stein estimator~\cite{james1961estimation}---that achieves a lower MSE for any true parameter value by pulling individual estimates toward a common average or the origin~\cite{gruber2017improving}.

\medskip
\noindent
\textbf{Linear regression and sparsity:}
In many applications in high-dimensional settings---such as in financial econometrics and risk management, natural language processing, recommender systems, neuroimaging, or cheminformatics---it is often believed that only a small subset of variables truly influence the process we try to model. Consequently, it is desirable to construct estimators that set many coefficients exactly to zero. Such sparse models can improve predictive performance by reducing variance, and they also lead to models that are much easier to interpret. 

This motivation leads to the lasso (least absolute shrinkage and selection operator)~\cite{Tib96,hastieelements}, which replaces the $\ell_2$-penalty with an $\ell_1$-penalty, namely we consider the regularized regression problem
\begin{equation}\label{eq:lasso1}
\min_{\regpar} \, \, \|y - X\regpar\|^2 + \lambda \|\regpar\|_1,
\end{equation}
for some $\lambda > 0$.
Unlike ridge regression, the $\ell_1$-penalty encourages coefficients to be exactly zero, thereby performing variable selection as part of the estimation procedure.
We will discuss the reason why the $\ell_1$-penalty promotes sparsity, associated theoretical guarantees, and numerical considerations extensively in Chapter~\ref{c:cs} in connection with {\em compressive sensing}.

\medskip
\noindent
\textbf{Outliers:}
An important consideration in linear regression (or in any data science task for that matter) is how different error functions handle {\em outliers}, i.e., data points that deviate significantly from the pattern exhibited by the majority of observations.
The least squares solution is sensitive to outliers, since a single observation with large error  $(y_i - x_i^\top\regpar)$ contributes quadratically to the objective, potentially  dominating the fit. In contrast, the solution of the $\ell_1$ objective (also called $\ell_1$-regression or least absolute deviations regression) given by
$$
\min_{\regpar} \, \, \|y - X\regpar\|_1,
$$
is more robust, since outliers contribute only linearly. However, one downside of the $\ell_1$ loss function is its lack of smoothness, since the function $f(x)=|x|$ is not differentiable at the origin. Also, $\ell_1$ estimators are typically less statistically efficient compared to $\ell_2$ estimators when the noise is Gaussian.

\begin{figure}[h]
\begin{center}
 \subcaptionbox{Linear regression with outliers}{\includegraphics[width = 52mm,height=38.6mm]{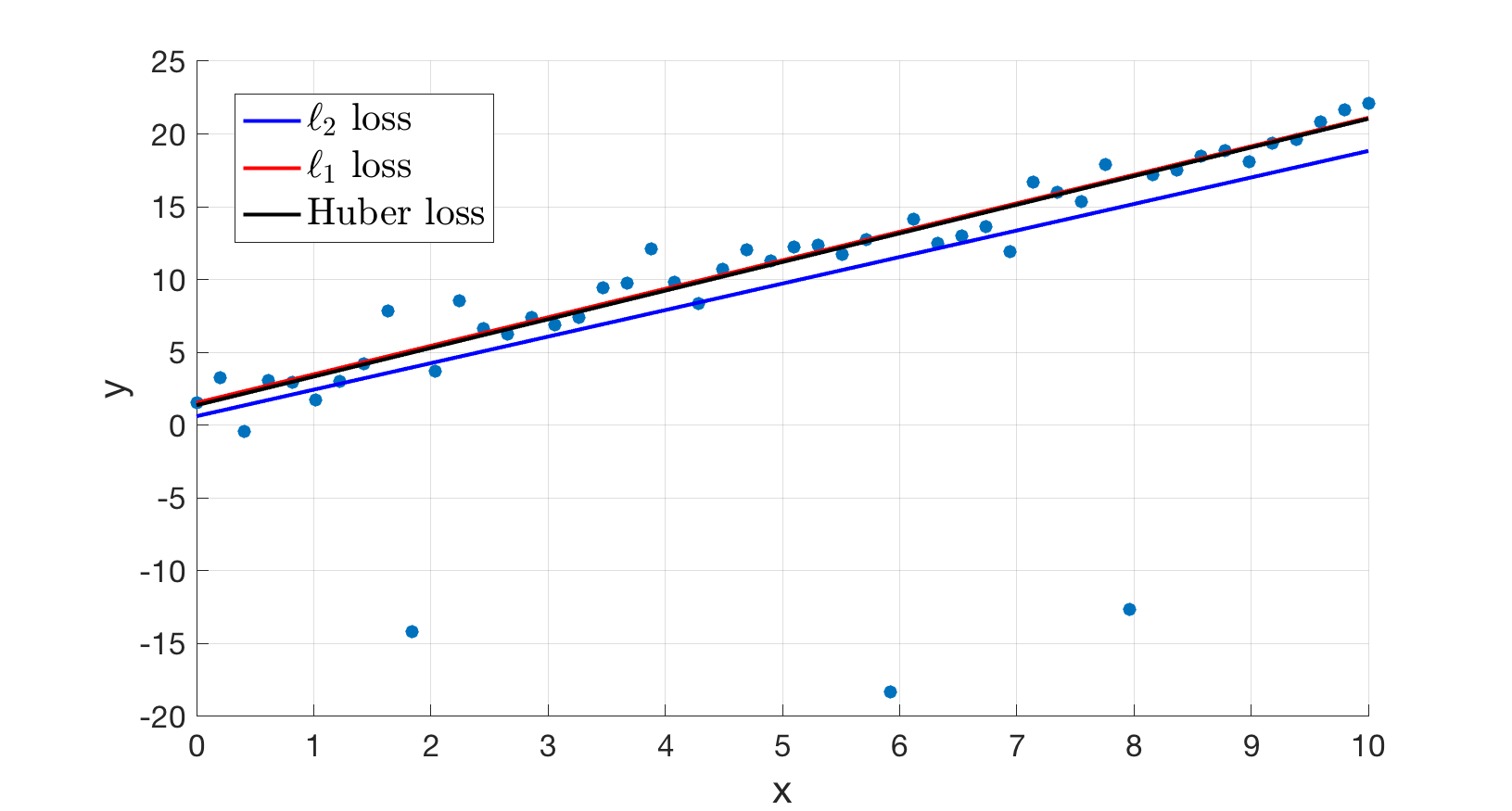}}
 \,
 \subcaptionbox{Loss function landscape}{
\includegraphics[width = 60mm,height=38.6mm]{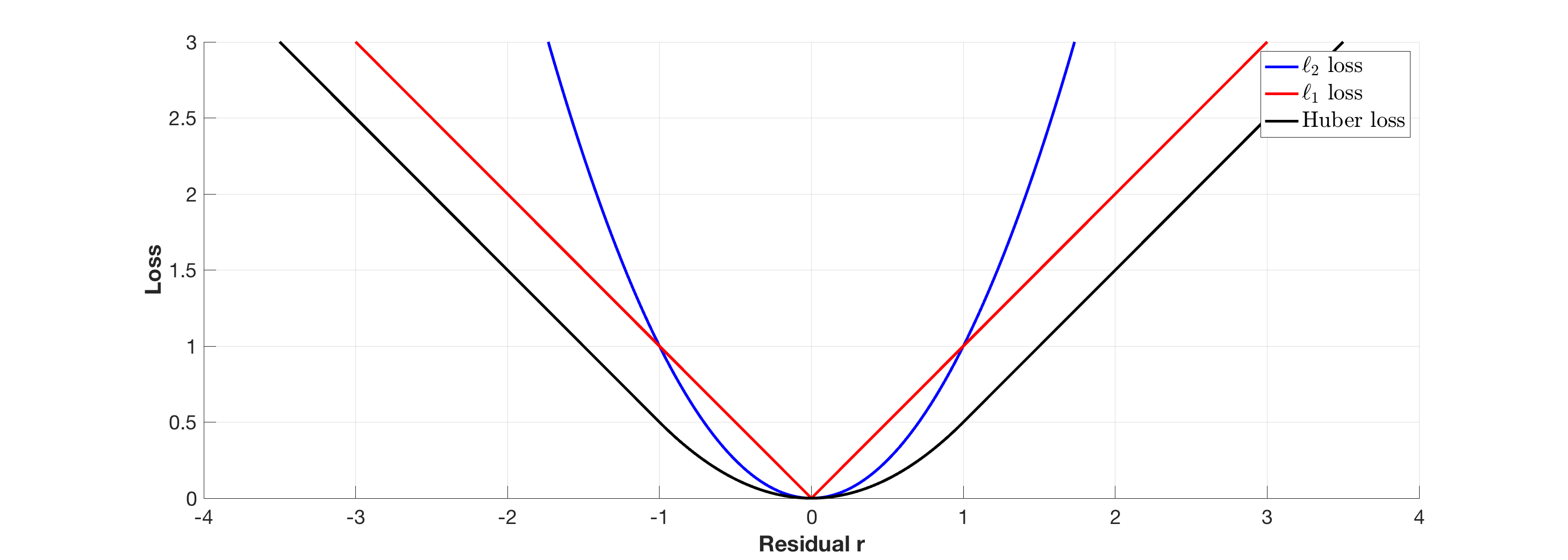}}
\caption{The performance of linear regression using different loss functions. The solution computed via the $\ell_2$-loss is pulled toward the outliers.
The solutions associated with $\ell_1$-loss and with Huber loss, respectively, are almost identical.
The $\ell_1$-loss is robust in the presence of outliers, but not smooth, while Huber loss combines robustness of the $\ell_1$-loss with the smoothness of the $\ell_2$-loss.}
\label{fig:outliers}
\end{center}
\end{figure}

A good compromise is provided by the 
 {\em Huber method}~\cite{huber1992robust}. The Huber loss is defined as
$$L_\delta(r_i) = \begin{cases}
\frac{1}{2}r_i^2 & \text{if } |r_i| \leq \delta, \\
\delta |r_i| - \frac{1}{2}\delta^2 & \text{if } |r_i| > \delta,
\end{cases}$$
where $r_i = y_i - x_i^\top \regpar$ is the $i$-th residual and 
$\delta > 0$ is a tuning parameter (see also Figure~\ref{fig:outliers}(b)).
Because the derivative is continuous everywhere (including the transition points at $\pm \delta$), the Huber loss is $C^1$-continuous. This ensures that gradient-based optimization algorithms (like gradient descent, see Chapter~\ref{s:grad}) converge reliably.

Figure~\ref{fig:outliers} shows a comparison of the $\ell_2$-loss, the $\ell_1$-loss, and the Huber loss for a linear regression example with a few outliers (left panel), together with the associated loss functions (right panel). It is evident that the solution of the $\ell_2$ objective is affected most by the few
outliers. The solutions associated with $\ell_1$-loss and with Huber loss, respectively (which are almost identical in this case) are only marginally affected by the outliers.

\section{Wiener filtering}\label{s:wiener}

In classical linear regression, we often treat the data as a collection of fixed points to be ``fit.'' However, in the presence of noise, a more powerful perspective is to treat the measurement as a corrupted signal. This leads us to the {\em Wiener Filter}, the mathematically optimal way to extract a signal from noise via a linear estimator by balancing experimental evidence against prior knowledge~\cite{wiener1949extrapolation,kailath2000linear}.

In this section we will show that ridge regression is simultaneously a
maximum a posteriori (MAP) estimator under a Gaussian prior, the solution to a
properly normalized loss function whose two terms have direct statistical
interpretations, and a componentwise Wiener filter applied in the SVD basis of the design matrix.  The unifying object is the
signal-to-noise ratio (SNR), which controls how strongly the estimator trusts
the data relative to the prior.
We will see that ridge regression acts as a component-wise Wiener filter in the SVD basis.


\medskip

The Wiener filter addresses a fundamental question: Given a noisy measurement $y = s + n$, how much should we trust $y$ to estimate the true signal $s$?
Let us first gain some intuition for the scalar case.
If we assume the signal $s$ and noise $n$ are independent, zero-mean scalar random variables with finite variances $\sigma_s^2$ and $\sigma_n^2$, the optimal linear estimate $\hat{s} = Gy$ is found by minimizing the MSE
\begin{equation}\label{eq:wiener1}
\min_{G} \, \E[(s - Gy)^2].
\end{equation}

To find the minimum mean squared error (MMSE) estimator, we expand the cost function~\eqref{eq:wiener1}, use the fact the signal and the noise are uncorrelated, and set the
derivative of~\eqref{eq:wiener1} with respect to $G$ to zero. We arrive at the solution, also called {\em Wiener Gain},
$$G = \frac{\sigma_s^2}{\sigma_s^2 + \sigma_n^2} = \frac{1}{1 + \frac{1}{\text{SNR}}},$$
where $\text{SNR} = \sigma_s^2 / \sigma_n^2$
represents the {\em signal-to-noise ratio}\footnote{Because many signals have a very wide dynamic range, signals are often expressed using the logarithmic decibel scale. Thus, in practice the SNR is often expressed in decibels as
$\text{SNR}_{\text{dB}} =10\log_{10}(\text{SNR})$.}.
For high SNR, i.e., if the signal is much stronger than the noise, $G \approx 1$, we trust the measurement. For low SNR, i.e., if the noise dominates, $G \approx 0$, we ignore the measurement to avoid a high-variance error, relying instead on our prior knowledge (that the signal is centered at zero).

We can now generalize this logic to the multivariate case.
Consider the linear model
\begin{equation}
  \label{eq:linear_model}
  y = X\theta + \varepsilon,
  \qquad
  \varepsilon \sim \mathcal{N}(0, \sigma^2 I_n),
\end{equation}
where $y \in \mathbb{R}^n$ is the measurement vector, $X \in \mathbb{R}^{n \times p}$
is the design matrix, $\theta \in \mathbb{R}^p$ is the unknown parameter vector,
and $\varepsilon$ is Gaussian noise with known variance $\sigma^2 > 0$.  The
likelihood of $\theta$ given the observations is therefore
\begin{equation}
  \label{eq:wiener2}
  p(y \mid \theta)
  = (2\pi\sigma^2)^{-n/2}
    \exp \Big(-\frac{1}{2\sigma^2}\|y - X\theta\|^2\Big).
\end{equation}
We place a zero-mean isotropic Gaussian prior on $\theta$:
\begin{equation}
  \label{eq:prior}
  \theta \sim \mathcal{N}(0, \tau^2 I_p),
\end{equation}
where $\tau^2 > 0$ is the prior variance. 

By Bayes' theorem the posterior is
$p(\theta | y) \propto p(y | \theta)\,p(\theta)$.
Taking the negative log-posterior and dropping terms that do not depend on
$\theta$, the MAP estimate is the minimizer of
\begin{equation} \label{eq:normalised_loss} 
\mathcal{L}(\theta)
  = \frac{1}{2\sigma^2}\|y - X\theta\|^2
  + \frac{1}{2\tau^2}\|\theta\|^2.
\end{equation}
This is the \emph{normalized} loss function.  The first term is the
\emph{data-fidelity term}: it measures how well $\theta$ explains the
observations, weighted by the inverse noise variance $1/\sigma^2$.  The second
term is the \emph{regularization term}: it penalizes large coefficients,
weighted by the inverse prior variance $1/\tau^2$.  Both terms now have
identical units (nats of log-probability), making the trade-off between them
dimensionally consistent.

\begin{remark}[Connection to standard ridge regression]
Multiplying $\mathcal{L}(\theta)$ by $2\sigma^2$ gives the standard
ridge objective $\|y - X\theta\|^2 + \lambda\|\theta\|^2$ with regularization
parameter
\begin{equation}
  \label{eq:lambda_sigma_tau}
  \lambda = \frac{\sigma^2}{\tau^2}.
\end{equation}
The standard formulation obscures the individual roles of $\sigma^2$ and
$\tau^2$ by collapsing them into the single scalar $\lambda$.  The normalized
form~\eqref{eq:normalised_loss} keeps them explicit and makes the probabilistic
interpretation transparent.
\end{remark}

Consider~\eqref{eq:normalised_loss} and
set $\nabla_\theta \mathcal{L}(\theta) = 0$; this gives
\begin{equation}
  \label{eq:gradient}
  -\frac{1}{\sigma^2}X^\top(y - X\theta) + \frac{1}{\tau^2}\theta = 0.
\end{equation}
Rearranging terms and multiplying  by $\sigma^2$ yields
\begin{equation}
  \label{eq:ridge_normal1}
  \left(X^\top X + \lambda I_p\right)\hat\theta = X^\top y,
  \qquad \lambda = \frac{\sigma^2}{\tau^2}.
\end{equation}
Since $X^\top X \succeq 0$ and $\lambda > 0$, the matrix
$X^\top X + \lambda I_p \succ 0$ is strictly positive definite for every
$\lambda > 0$, regardless of the rank of $X$.  The unique MAP (and ridge)
estimate is therefore
\begin{equation}
  \label{eq:ridge_estimate}
  \hat\theta_{\mathrm{MAP}}
  = \hat\theta_{\mathrm{ridge}}
  = \left(X^\top X + \lambda I_p\right)^{-1}X^\top y.
\end{equation}
We already confirmed earlier that this gives a strict global minimum.

Because both the likelihood~\eqref{eq:wiener2} and the
prior~\eqref{eq:prior} are Gaussian, the posterior is also Gaussian:
\begin{equation}
  \label{eq:posterior}
  \theta \mid y
  \;\sim\;
  \mathcal{N}\!\left(
    \hat\theta_{\mathrm{MAP}},\;
    \sigma^2\left(X^\top X + \lambda I_p\right)^{-1}
  \right).
\end{equation}

\medskip

It is instructive to revisit the componentwise structure of the ridge estimator by using the
SVD of the design matrix, $X=U \Sigma V^\top$, which we derived in~\eqref{eq:ridgeSVD2}. The ridge estimate takes the form
\begin{equation}
  \label{eq:ridge_svd1}
  \hat\theta_{\mathrm{ridge}}
  = V(\Sigma^2 + \lambda I_p)^{-1}\Sigma U^\top y
  = \sum_{i=1}^p
    \frac{\sigma_i}{\sigma_i^2 + \lambda}
    (u_i^\top y)\, v_i.
\end{equation}
Now we apply the rotations
\begin{equation}
  \label{eq:rotated_coords}
  \tilde\theta = V^\top\theta \in \mathbb{R}^p,
  \qquad
  \tilde y = U^\top y \in \mathbb{R}^p.
\end{equation}
In these coordinates the model~\eqref{eq:linear_model} becomes
$\tilde y_i = \sigma_i \tilde\theta_i + \tilde\varepsilon_i$ with
$\tilde\varepsilon_i \sim \mathcal{N}(0, \sigma^2)$ i.i.d., and the ridge
estimate~\eqref{eq:ridge_svd1} decouples into $p$ independent scalar problems:
\begin{equation}
  \label{eq:ridge_component}
  \hat{\tilde\theta}_i
  = \frac{\sigma_i}{\sigma_i^2 + \lambda}\,\tilde y_i
  = \frac{1}{\sigma_i}\cdot\frac{\sigma_i^2}{\sigma_i^2 + \lambda}\,\tilde y_i,
  \qquad i = 1, \ldots, p.
\end{equation}

We now introduce the \emph{per-component signal-to-noise ratio}.  Under the
prior \eqref{eq:prior}, the $i$-th rotated coefficient satisfies
$\tilde\theta_i \sim \mathcal{N}(0, \tau^2)$, so the signal component
$\sigma_i\tilde\theta_i$ has variance $\sigma_i^2\tau^2$.  The corresponding
observation $\tilde y_i = \sigma_i\tilde\theta_i + \tilde\varepsilon_i$ has
noise variance $\sigma^2$.  The SNR in the $i$-th  direction is
therefore
\begin{equation}
  \label{eq:snr_def}
  \mathrm{SNR}_i
  = \frac{\sigma_i^2\,\tau^2}{\sigma^2}.
\end{equation}
Substituting $\lambda = \sigma^2/\tau^2$, note that
\begin{equation}
  \label{eq:lambda_snr}
  \frac{\lambda}{\sigma_i^2}
  = \frac{\sigma^2/\tau^2}{\sigma_i^2}
  = \frac{1}{\mathrm{SNR}_i},
\end{equation}
so the shrinkage factor in~\eqref{eq:ridge_component} can be written as
\begin{equation}
  \label{eq:shrinkage_snr}
  \frac{\sigma_i^2}{\sigma_i^2 + \lambda}
  = \frac{1}{1 + \lambda/\sigma_i^2}
  = \frac{1}{1 + 1/\mathrm{SNR}_i}
    = \frac{\mathrm{SNR}_i}{1 + \mathrm{SNR}_i}.
\end{equation}
The ridge estimate in the principal-component basis therefore takes the form
\begin{equation} \label{eq:wiener_form}
 \hat{\tilde\theta}_i
  = \underbrace{\frac{1}{\sigma_i}}_{\text{LS inversion}}
\cdot  \underbrace{\frac{1}{1 + 1/\mathrm{SNR}_i}}_{\text{Wiener filter gain}}
    \cdot\;
    \tilde y_i,
    \qquad i = 1, \ldots, p.
\end{equation}

The connection to Wiener filtering deserves elaboration.  Consider the scalar
model $\tilde y_i = \sigma_i\tilde\theta_i + \tilde\varepsilon_i$ with
$\tilde\theta_i \sim \mathcal{N}(0,\tau^2)$ and
$\tilde\varepsilon_i \sim \mathcal{N}(0,\sigma^2)$ independent.  This is
exactly the signal-plus-noise model underlying the Wiener filter that we introduced at the beginning of this section: we observe the signal $\sigma_i\tilde\theta_i$
corrupted by noise $\tilde\varepsilon_i$, and we wish to estimate
$\tilde\theta_i$.

The linear MMSE estimator of $\tilde\theta_i$ given $\tilde y_i$ is:
\begin{equation}
  \label{eq:mmse_component}
  \mathbb{E}[\tilde\theta_i \mid \tilde y_i]
  = \frac{\mathrm{Cov}(\tilde\theta_i,\tilde y_i)}{\mathrm{Var}(\tilde y_i)}
    \,\tilde y_i
  = \frac{\sigma_i\tau^2}{\sigma_i^2\tau^2 + \sigma^2}
    \,\tilde y_i
  = \frac{1}{\sigma_i}\cdot
    \frac{\sigma_i^2\tau^2}{\sigma_i^2\tau^2 + \sigma^2}
    \,\tilde y_i.
\end{equation}
Dividing numerator and denominator of the second term on the right-hand side of~\eqref{eq:mmse_component} by $\sigma^2$ gives
\begin{equation}
  \label{eq:mmse_snr}
  \mathbb{E}[\tilde\theta_i \mid \tilde y_i]
  = \frac{1}{\sigma_i}\cdot\frac{\mathrm{SNR}_i}{1 + \mathrm{SNR}_i}
    \,\tilde y_i
  = \frac{1}{\sigma_i}\cdot\frac{1}{1 + 1/\mathrm{SNR}_i}
    \,\tilde y_i,
\end{equation}
which is identical to~\eqref{eq:wiener_form}.  Since the model is jointly
Gaussian, the linear MMSE estimator coincides with the MAP estimator, confirming that
ridge regression achieves the minimum MSE in each principal direction
simultaneously.

\medskip

The Wiener filter gain $W_i = \mathrm{SNR}_i/(1+\mathrm{SNR}_i)$ interpolates
between two extreme regimes of high SNR and low SNR by continuously interpolating between the data and the prior. The fraction of the least squares estimate that is retained is exactly $\mathrm{SNR}_i/(1 + \mathrm{SNR}_i) \in (0,1)$.

\bigskip
Reflecting on the derivations above, we have shown that the following three problems have identical solutions:
\begin{enumerate}

  \item \textbf{Regularized least squares (ridge regression).}
  Minimize the standard ridge objective:
  \[
    \hat\theta_{\mathrm{ridge}}
    = \argmin_\theta\; \|y - X\theta\|^2 + \lambda\|\theta\|^2.
  \]

  \item \textbf{MAP estimation under a Gaussian prior.}
  Minimize the normalized probabilistic loss:
  \[
    \hat\theta_{\mathrm{MAP}}
    = \argmin_\theta\;
      \frac{1}{2\sigma^2}\|y - X\theta\|^2 + \frac{1}{2\tau^2}\|\theta\|^2,
    \qquad \lambda = \frac{\sigma^2}{\tau^2}.
  \]

  \item \textbf{Componentwise Wiener filtering in the SVD basis.}
  Apply the Wiener filter gain to each singular value component:
  \[
    \hat{\tilde\theta}_i
    = \frac{1}{\sigma_i}\cdot\frac{1}{1 + 1/\mathrm{SNR}_i}\cdot\tilde y_i,
    \qquad
    \mathrm{SNR}_i = \frac{\sigma_i^2\tau^2}{\sigma^2} = \frac{\sigma_i^2}{\lambda}.
  \]

\end{enumerate}

\section*{Exercises}
\addcontentsline{toc}{section}{Exercises}

\begin{myexercise}
 Show that if $x_1,x_2,\ldots, x_n$ are $n$ independent observations of a $p$-dimensional random variable
$X\in \mathbb{R}^p$, then
$$\Sigma_n = \frac{1}{n-1}\sum_{i=1}^n (x_i-\mu_n)(x_i-\mu_n)^T,$$
is an unbiased estimate of $\Sigma = \mathbb{E}(X-\mu)(X-\mu)^T$, where $\mu = \mathbb{E}X$ is the true mean and $\mu_n = \frac{1}{n} \sum_{i=1}^n x_i$ is the sample mean. That is, show $\mathbb{E}(\Sigma_n) = \Sigma$.
    
\end{myexercise}

\begin{myexercise}\label{ex:ols_bias_variance}
For the  least squares estimator $\hat{\theta}_{\text{LS}} = (X^\top X)^{-1}X^\top y$ show that the
bias-variance expression~\eqref{eq:bias_variance_param}  takes the form
$$\MSE(\hat{\theta}_{\text{LS}}) = 0 + \tr(\sigma^2(X^\top X)^{-1}) 
= \sigma^2\tr((X^\top X)^{-1}).$$
\end{myexercise}

\begin{myexercise}\label{ex:mse_unbiased}
This exercise relates to Theorem~\ref{th:mle}. Show that
an unbiased estimator of $\sigma^2$ is given by
$\hat{\sigma}^2_{\text{MLE}} = \frac{1}{n-p}\|y - X\hat{\theta}_{\text{LS}}\|^2$.
\end{myexercise}

\begin{myexercise}\label{ex:ridge_bias}
Prove Proposition~\ref{prop:ridge_bias}.
\end{myexercise}

\begin{myexercise}
Consider the linear model
$$y = X\theta + \epsilon$$
where $X \in \mathbb{R}^{n \times p}$ is a random matrix with entries $X_{ij} \sim \mathcal{N}(0, 1/n)$ and $\epsilon \sim \mathcal{N}(0, \sigma^2 I_n)$. Assume $n > p$ so that $X$ is full rank almost surely.
\begin{enumerate}
\item 
Write the log-likelihood function $L(\theta; y, X)$ and show that the MLE $\hat{\theta}_{MLE}$ is identical to the  Least Squares  estimator.
\item
Show that conditioned on $X$, the estimator follows a Gaussian distribution
$$\hat{\theta}_{MLE} | X \sim \mathcal{N}(\theta, \sigma^2 (X^T X)^{-1}).$$
\item Large sample limit: As $n, p \to \infty$ with $p/n = \gamma \in (0, 1)$, the matrix $X^T X$ converges to the Identity matrix $I_p$ (by the Law of Large Numbers). What is the expected MSE $E[\|\hat{\theta}_{MLE} - \theta\|^2]$ in this limit?
\end{enumerate}

\end{myexercise}

\begin{myexercise}
Use the same setup as in the previous exercise. Implement a simulation to verify the theoretical MSE.
Fix $p = 50$ and vary $n$ from $60$ to $500$.
For each $n$, generate 100 trials of $X$ and $\epsilon$.
Compute the average $\| \hat{\theta}_{MLE} - \theta \|^2$ and plot it against the theoretical prediction $\sigma^2 \Tr((X^T X)^{-1})$.
   
\end{myexercise}

\begin{myexercise}
Generate a synthetic dataset with $n=100$ observations and $p=10$ features where only the first 3 features are truly informative (how do you translate this condition 
``only the first 3 features are truly informative'' into your experimental setup?). Now add Gaussian noise $\epsilon \sim \mathcal{N}(0, 0.5)$ to the observations.
\begin{enumerate}
\item  Compute the least squares solution.
\item Implement Ridge Regression for a range of $\lambda \in [10^{-3}, 10^3]$.
\item Plot the MSE on a held-out test set as a function of $\lambda$. Identify the ``elbow'' where the bias-variance tradeoff is optimized.
\end{enumerate}  
\end{myexercise}

\begin{myexercise}
Suppose $s$ is a Bernoulli signal taking values $\pm 1$ with equal probability (so $\E[s] = 0$, $\sigma_s^2 = 1$), and $n \sim \mathcal{N}(0, \sigma_n^2)$ independent of $s$. The observation is $y = s + n$. 
\begin{enumerate}
\item Compute the best {\em linear} estimator via
the Wiener filter. 
\item Show that the true MMSE estimator is the conditional mean
$$\E[s | y] = \tanh (y/\sigma_n^2).$$
\end{enumerate}
This exercise tells us the Wiener filter is the best linear approximation to $\tanh (y/\sigma_n^2)$, but the two (approximately) coincide only at low SNR (where both are approximately linear) and the gap grows as the SNR increases.

\end{myexercise}

\begin{myexercise}
Suppose the true relationship between a feature $x$ and a response $y$ is governed by a quadratic function with additive noise
$$y = x^2 + \epsilon, \quad \epsilon \sim \mathcal{N}(0, 0.1^2)$$
We are given a small training set $S$ of $n=5$ points sampled uniformly from $x \in [-1, 1]$.
\begin{enumerate}
\item  Underfitting (structural bias):
Assume a data analyst attempts to model this data using a linear hypothesis class $\mathcal{H}_1 = \{f(x) = w_1x + w_0\}$.
Explain why, even with an infinite number of training points, the expected prediction error will remain high.
In the language of risk, does this error stem from the estimation error or the approximation error?
\item Overfitting (high variance):
Her colleague decides to ensure the training error is exactly zero. They use a 4th-degree polynomial hypothesis class $\mathcal{H}_4 = \{f(x) = \sum_{i=0}^4 w_ix^i\}$.
Since we have $n=5$ distinct data points, a 4th-degree polynomial can interpolate the data perfectly (zero empirical risk). Sketch what this fitted curve might look like compared to the true parabola $y = x^2$.
If we draw a new test point $x_{test}$ from the same distribution, why is the squared error likely to be much larger for the 4th-degree model than for a simple quadratic model?
\item The generalization gap:
Let $\hat{R}_S(f)$ be the training error and $R(f)$ be the population risk.
For which model ($\mathcal{H}_1$ or $\mathcal{H}_4$) do you expect the generalization gap $G(f) = R(f) - \hat{R}_S(f)$ to be larger?
Provide a brief geometric argument involving the "sensitivity" of the coefficients to small perturbations in the training labels $y$.
\end{enumerate}   
\end{myexercise}

\begin{myexercise}
\textbf{Total Least Squares.} Suppose we wish to fit a linear model $y = X\theta$ where
$X \in \mathbb{R}^{n \times p}$ is a matrix of predictors, $y \in \mathbb{R}^n$ is a vector of observations $\theta \in \mathbb{R}^p$ is an unknown parameter vector. In standard least squares we assume that the predictor matrix $X$ is known exactly and that only $y$ contains noise.

However, suppose {\em both} $X$ and $y$ are contaminated by measurement errors. Let the true relationship be
$$
(y + \delta y) = (X + \Delta X)\theta,
$$
where $\Delta X$ and $\delta y$ are unknown perturbations.

\begin{enumerate}
\item  Explain why ordinary least squares may give biased parameter estimates when the predictor variables are measured with error.
\item
Total least squares seeks corrections $\Delta X$ and $\delta y$ that solve
$$
(y + \delta y) = (X + \Delta X)\theta,
$$
while minimizing the total squared perturbation
$$
\|\Delta X\|_F^2 + \|\delta y\|^2.
$$

Show that this problem can be written as
$$
\min_{\Delta A} \|\Delta A\|_F
\quad 
\text{subject to} \quad (A + \Delta A)
\begin{bmatrix}
\theta \\
-1
\end{bmatrix}
= 0,
$$
where $A = [X | y]$ is the augmented data matrix.

\item
Let the SVD of the augmented matrix be $A = U \Sigma V^T.$
Show that the TLS solution is obtained from the right singular vector corresponding to the smallest singular value.

\item 
If $v =[v_1, \dots, v_p, v_{p+1}]^\top$
is this singular vector, show that
$$
\theta_{TLS} = -\frac{1}{v_{p+1}}
[v_1,\dots, v_p]^\top.
$$
\end{enumerate}
\end{myexercise}

\begin{myexercise} 
Consider a random variable $X$ with $\E[X] = \mu$ and $\Var(X) = \sigma^2$. Suppose you have $n$ i.i.d.\ observations $x_1, \dots, x_n$. You are considering two estimators for $\mu$:
\begin{enumerate}
\item[]  Estimator A (Sample Mean): $\hat{\mu}_A = \frac{1}{n}\sum x_i$
\item[] Estimator B (Constant): $\hat{\mu}_B = c$, where $c$ is a fixed constant (e.g., $c=0$).
\end{enumerate}

\begin{enumerate}
\item  Calculate the bias and variance for both estimators.
\item  Give the MSE for both.
\item Under what specific condition (in terms of $n, \sigma^2, \mu,$ and $c$) is the ``dumb'' constant estimator actually better (lower MSE) than the sample mean?
\end{enumerate}
\end{myexercise}

\begin{myexercise} 
This exercise is about ridge regression and cross-validation.
Use the following Matlab-type code to generate a synthetic high-dimensional dataset where only a few features are truly informative:
\begin{Verbatim}[fontsize=\small, frame=single]
% Set seed for reproducibility
rng(100);
n = 500; p = 1000; % High-dimensional setting (p > n)
X = randn(n, p);
% Only first 10 features matter
theta_true = [ones(10, 1); zeros(p-10, 1)]; 
y = X * theta_true + 0.5 * randn(n, 1); % Additive Gaussian noise
\end{Verbatim}

\begin{enumerate}
\item 
Recall that the ridge estimator is given by  $\hat{\theta}_{\lambda} = (X^T X + \lambda I)^{-1} X^T y$.
Write a function \texttt{myridge(X, y, lambda)} that returns the regression coefficients. 
Efficiency Hint: Since $p > n$, it is computationally cheaper to use the identity $(X^T X + \lambda I)^{-1} X^T = X^T (XX^T + \lambda I)^{-1}$. Prove this identity in your write-up.

\item 
Implement a 5-fold cross-validation routine to select the optimal $\lambda$ from an equally spaced grid: \texttt{lambdas = linspace(0.1, 5, 50)}.
Partition the $n=500$ observations into five  folds of equal size.
For each $\lambda$ in the grid:
Apply the ridge estimator to the four folds and calculate the Mean Squared Error (MSE) on the remaining fold.
Average the MSE across all five folds to get the Cross-Validation Error, $CV(\lambda)$.
Plot the $CV(\lambda)$ curve against $\log(\lambda)$. Mark the $\lambda^*$ that minimizes the error.
\item
The ``One-Standard-Error'' Rule: Often, the $\lambda$ that strictly minimizes $CV(\lambda)$ is too small. Identify the largest $\lambda$ whose error is within one standard error of the minimum. Why might a researcher prefer this ``simpler'' model?
\item 
Compute the SVD of $X = U\Sigma V^T$. Explain how $\lambda$ modifies the singular values $\sigma_i$ in the solution. Specifically, what happens to the components of the solution corresponding to very small $\sigma_i$ as $\lambda$ increases?  
\end{enumerate}
\end{myexercise}

\chapter{Graphs, Networks, and Clustering}
\label{c:graphs}

A crucial part of data science consists of the study of networks. Network science, or graph theory, unifies the study of diverse types of networks, such as social networks, protein-protein interaction networks, gene-regulation networks, and the internet. In this chapter we introduce graph theory and treat the problem of clustering, to identify similar data points, or vertices, in (network) data.

\section{PageRank}

Before we introduce the formalism of graph theory, we describe the celebrated PageRank algorithm. This algorithm is a principal component\footnote{It is difficult to resist using this pun.} behind web search algorithms, in particular in Google. The goal of PageRank is to quantitatively rate the importance of each page on the web, allowing the search algorithm to rank the pages and thereby present to the user the more important pages first.
Search engines such as Google have to carry out three basic steps:\footnote{Another important component of modern search engines is personalization, which we do not discuss here.}
\begin{itemize}
\setlength{\parsep}{-0.3ex}
\setlength{\itemsep}{-0.3ex}
\item Crawl the web and locate all, or as many as possible, accessible webpages.
\item Index the data of the webpages from step 1, so that they can be searched efficiently for relevant
key words or phrases.
\item Rate the importance of each page in the database, so that when a user does
a search and the subset of pages in the database with the desired information
has been found, the more important pages can be presented first.
\end{itemize}
Here, we will focus on the third step. We follow mainly the derivation in~\cite{Bryan2006}. We aim to develop a score of importance for each webpage. A score will be a
non-negative number. A key idea in assigning a score to any given webpage is that the page's score is derived from the
links made to that page from other webpages --- ``A person is important not if it knows a lot of people, but if a lot of
people know that person''. 

\begin{figure}[h]
\begin{center}
\includegraphics[width=40mm]{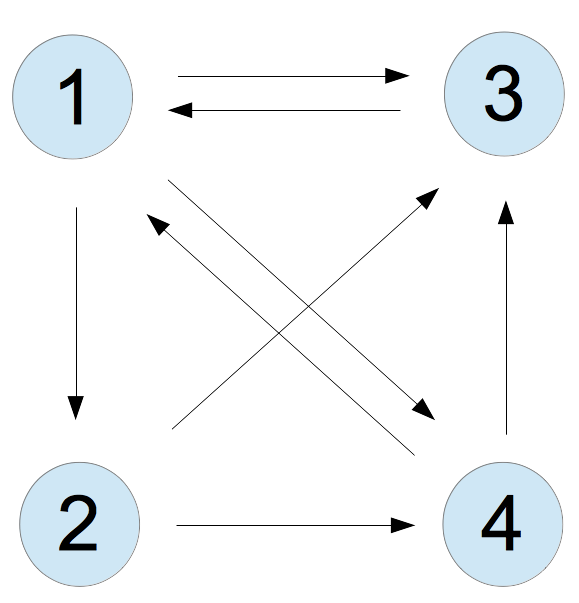}
\end{center}
\caption{A toy example of the Internet}
\label{fig:web}
\end{figure}

Suppose the web of interest contains $n$ pages, each page indexed by an integer $k$,
$1 \le k \le n$. A typical example is illustrated in Figure~\ref{fig:web}, in which an arrow from page
$k$ to page $j$ indicates a link from page $k$ to page $j$. Such a web is an example of a
directed graph. The links to a given page are called
the backlinks for that page.
We will use $x_k$ to denote the importance score of page $k$ in the web. $x_k$
is nonnegative and $x_j > x_k$ indicates that page $j$ is more important than page $k$.

A very simple approach is to take $x_k$ as the number of backlinks for page $k$. In
the example in Figure~\ref{fig:web}, we have $x_1 = 2, x_2 = 1, x_3 = 3$, and $x_4 = 2$, so that page 3
is the most important, pages 1 and 4 tie for second, and page 2 is least important. A
link to page $k$ becomes a vote for page $k$'s importance.
This approach ignores an important feature one would expect a ranking algorithm
to have, namely, that a link to page $k$ from an important page should boost page $k$'s
importance score more than a link from an unimportant page.
In the web of Figure~\ref{fig:web}, pages 1 and 4 both have two backlinks: each links to the other, but the second backlink from page
1 is from the seemingly important page 3, while the second
backlink for page 4 is from the relatively unimportant page 2. As such, perhaps the algorithm should rate
the importance of page 1 higher than that of page 4.

As a first attempt at incorporating this idea, let us compute the score of page $j$ as
the sum of the scores of all pages linking to page $j$. For example, consider the web
in our toy example. The score of page 1 would be determined by the relation $x_1 = x_3 + x_4$.
However, since $x_3$ and $x_4$ will depend on $x_1$, this seems like a circular definition, since it is
self-referential (it is exactly this self-referential property that will establish a connection to eigenvector
problems!).

We also seek a scheme in which a webpage does not gain extra influence
simply by linking to lots of other pages. We can do this by reducing the impact of each link, as more and more outgoing 
links are added to a webpage. If page $j$ contains $n_j$ links, one of which
links to page $k$, then we will boost page $k$'s score by $x_j/n_j$, rather than by $x_j$. In
this scheme, each webpage gets a total of one vote, weighted by that web page's score,
that is evenly divided up among all of its outgoing links. To quantify this for a web
of $n$ pages, let $L_k \subset \{1,2,\dots, n\}$ denote the set of pages with a link to page $k$, that
is, $L_k$ is the set of page $k$'s backlinks. For each $k$ we require
$$x_k = \sum_{j \in L_k} \frac{x_j}{n_j},$$
where $n_j$ is the number of outgoing links from page $j$. 

If we apply these scheme to the toy example in Figure~\ref{fig:web}, then for page 1 we have
$x_1 = x_3/1 + x_4/2$,  since pages 3 and 4 are backlinks for page 1 and page 3 contains
only one link, while page 4 contains two links (splitting its vote in half). Similarly,
$x_2 = x_1/3, x_3 = x_1/3 + x_2/2 + x_4/2$, and $x_4 = x_1/3 + x_2/2$. These conditions can be expressed as linear
system of equations $Ax =x$, where $x=[x_1,x_2,x_3,x_4]^T$ and 
$$A = 
\begin{bmatrix}
0 & 0 & 1 & \frac{1}{2} \\
\frac{1}{3} & 0 & 0 & 0 \\
\frac{1}{3} & \frac{1}{2} & 0 & \frac{1}{2} \\
\frac{1}{3} & \frac{1}{2} & 0 & 0 
\end{bmatrix}
$$
Thus, we end up with an eigenvalue/eigenvector problem: Find the eigenvector $x$ of the matrix $A$, associated with the
eigenvalue 1. We note that $A$ is a column-stochastic matrix, since it is a square matrix for which all of its
entries are nonnegative and the entries in each column sum to 1. If we build a random walk on the internet where each link is clicked with equal probability then $A_{ij}$ is the probability that a random walked in page $j$ goes to page $i$. Later, in Chapter~\ref{c:diffusion}, we will use random walks on (undirected) graphs in order to embed their nodes in euclidean space, we note that, in that context, we will mostly work with $M=A^T$ the matrix for which  $M_{ij}$ corresponds to the probability of taking step from $i$ to $j$. Stochastic matrices arise in the study of Markov chains and in a variety of modeling problems in economics and
operations research.  See e.g.~\cite{HJ90} for more details on stochastic matrices. The fact that 1 is an eigenvalue of $A$ is not just coincidence in this example, but holds true in general for stochastic matrices. 

\begin{proposition}\label{prop:columnstochastichaseig1}
A column-stochastic matrix $A$ has an eigenvalue equal to 1 and 1 is also its largest eigenvalue.
\end{proposition}

\begin{proof}
Let $A$ be an $n\times n$ column-stochastic matrix.
We first note that $A$ and $A^T$ have the same eigenvalues (their eigenvector will usually be different though).
Let $\mathbf{1} = [1,1,\dots,1]^T$ be the vector of length $n$ which has all ones as entries. Since $A$ is
column-stochastic, we have $A^T \mathbf{1} = \mathbf{1}$ (since all columns of $A$ sum up to 1). Hence $\mathbf{1}$ is an eigenvector of
$A^T$ (but not of $A$) with eigenvalue 1. Thus 1 is also an eigenvalue of $A$.

To show that 1 is the largest eigenvalue of $A$ we apply the Gershgorin Circle Theorem~\cite{HJ90} to $A^T$. Consider
row $k$ of $A^T$. Let us call the diagonal element $a_{k,k}$ and the radius will be
$\sum_{i \neq k} |a_{k,i}| = \sum_{i \neq k} a_{k,i}$ since $a_{k,i} \ge 0$. This is a circle with its center
at $a_{k,k} \in [0,1]$ and with radius $\sum_{i \neq k} a_{k,i} = 1 - a_{k,k}$. Hence, this circle has 1 on its
perimeter. This holds for all Gershgorin circles for this matrix.  Thus, since all eigenvalues lie in the
union of the Gershgorin circles, all eigenvalues $\lambda_i$ satisfy $|\lambda_i| \le 1$.

\end{proof}

In our example, we obtain as eigenvector $x$ of $A$ associated with eigenvalue 1 the vector
$x = [x_1,x_2,x_3,x_4]^T$ with entries
 $x_1 = \frac{12}{31}, x_2 = \frac{4}{31}, x_3 = \frac{9}{31}$, and $x_4 =
\frac{6}{31}$. Hence, perhaps somewhat surprisingly, page 3 is no longer the most important one, but page 1.
This can be explained by the fact, that the in principle quite important page 3 (which has three webpages linking to
it) has only one outgoing link, which gets all its ``voting power'', and that link points to page 1.

In reality, $A$ can easily be of size a billion times a billion. Fortunately, we do not need compute all eigenvectors
of $A$, only the eigenvector associated with the eigenvalue 1, which, as we know, is also the largest eigenvalue of $A$.
This in turn means we can resort to standard {\em power iteration} to compute $x$ fairly efficiently (and we can also make
use of the fact that $A$ will be a sparse matrix, i.e., many of its entries will be zero).
The actual PageRank algorithms adds some minor modifications,\footnote{One such modification is the addition of very small weight edges between every pair of pages, corresponding to a ``random teleport'' of the random walker (or web crawler) that happens with very low probability. One advantage this brings is that the networks becomes irreducible. A reducible directed graph is one for which there are nodes $u$ and $v$ for which there is no directed path from $u$ to $v$, if a networks is not reducible, it is called irreducible.} but the essential idea is as described above.

The idea to use eigenvectors for ranking dates back to the late 1800s to the work of Edmund Landau in the context
of ranking in chess tournaments, it was actually Landau's first mathematical paper when he was 18 years old~\cite{Landau1895Turnierresultate} (with a follow-up later~\cite{Landau1914Preisverteilung}). You
can read about this nice story in~\cite{sinn2022landau}.

\section{Graphs and clustering}

\begin{figure}[h]
\begin{center}
\includegraphics[width=0.4\textwidth]{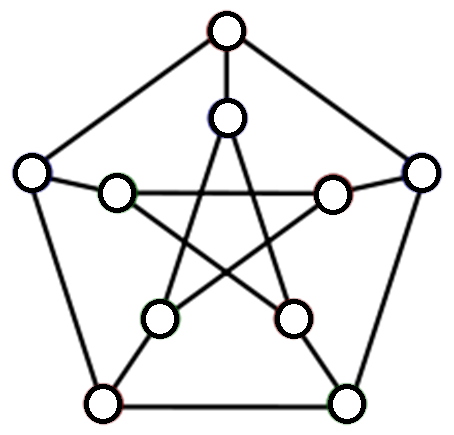}
\caption{The Petersen graph}
\label{fig:Petersen}
\end{center}
 
\end{figure}

We now introduce the formalism for undirected\footnote{The previous Chapter featured directed graphs, in which edges (links) have a meaningful direction. In what follows we will focus in undirected graphs in which an edge represents a connection, without meaningful direction.} graphs, one of the main objects of study in what follows. A graph $G = (V,E)$ contains a set of nodes $V=\left\{v_1,\dots,v_n\right\}$ and edges $E\subseteq {V \choose 2 }$. An edge $(i,j)\in E$ if $v_i$ and $v_j$ are connected. Figure \ref{fig:Petersen} depicts one of the graph theorists' favorite examples, the Petersen graph\footnote{The Petersen graph is often used as a counter-example in graph theory.}.

Let us recall some concepts about graphs that we will need.
\begin{itemize}
\item A graph is {\em connected} if, for all pairs of vertices, there is a path between these vertices in the graph. The number of connected components is simply the size of the smallest partition of the nodes into connected subgraphs. The Petersen graph is connected (and thus it has only $1$ connected component).

\item A {\em clique} of a graph $G$ is a subset $S$ of its nodes such that the subgraph corresponding to it is complete. In other words $S$ is a clique if all pairs of vertices in $S$ share an edge. The clique number $c(G)$ of $G$ is the size of the largest clique of $G$. The Petersen graph has a clique number of $2$.

\item An {\em independent set} of a graph $G$ is a subset $S$ of its nodes such that no two nodes in $S$ share an edge. Equivalently it is a clique of the complement graph $G^c \defeq \left(V,E^c\right)$. The independence number of $G$ is simply the clique number of $S^c$. The Petersen graph has an independence number of $4$.

\end{itemize}

A particularly useful way to represent a graph is through its adjacency matrix. Given a graph $G=(V,E)$ on $n$ nodes ($|V|=n$), we define its adjacency matrix $A \in \RR^{n\times n}$ as the symmetric matrix with entries
\begin{equation}
 \label{adjacencymatrix}   
A_{ij} = \left\{  \begin{array}{ll} 1 & \text{ if } (i,j)\in E, \\ 0 & \text{otherwise}.   \end{array}   \right.
\end{equation}

Sometimes, we will consider weighted graphs $G=(V,E,W)$, where edges may have weights $w_{ij}$ that are non-negative $w_{ij}\geq 0$ and symmetric $w_{ij} = w_{ji}$.

Much of the sequel will deal with graphs. Chapter~\ref{c:diffusion} will treat (network) data visualization, dimension reduction, and embeddings of graphs in Euclidean space. Chapter~\ref{c:community} will introduce and study important random graph models. The rest of this Chapter will be devoted to spectral graph theory, clustering, and graph metrics. 

Clustering is one of the central tasks in machine learning. Given a set of data points, or nodes of a graph, the purpose of clustering is to partition the data into a set of clusters where data points assigned to the same cluster correspond to similar data points (depending on the context, it could be for example having small distance to each other if the points are in Euclidean space, or having high connectivity if on a graph). We will start with an example of clustering points in Euclidean space, and later move back to graphs.

\begin{figure}[h]
\begin{center}
\includegraphics[width=0.8\textwidth]{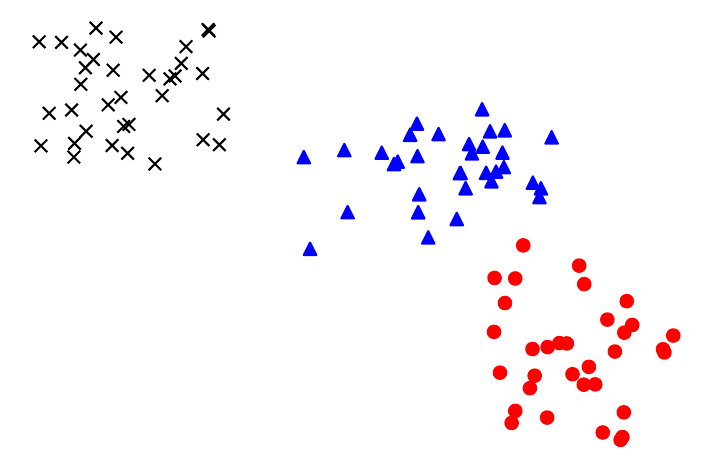}
\caption{Example of points separated into clusters. By using different symbols and colors, the three clusters appear more distinct than they actually are. }
\label{figure:3:clustering_00}
\end{center}
\end{figure}

\section{$k$-means clustering}

One the most popular methods used for clustering is $k$-means clustering. Given $x_1,\dots,x_n\in\RR^p$ the $k$-means clustering partitions the data points into clusters $S_1\cup \dots \cup S_k$ with centers $\mu_1,\dots,\mu_k\in\RR^p$ as the solution to:
\begin{equation}\label{eq:3:kmeans_obj}
\min_{\text{partition}\ \substack{S_1,\dots,S_k \\ \mu_1,\dots,\mu_k}}\sum_{l=1}^k \sum_{i\in S_l} \left\|  x_i - \mu_l\right\|^2.
\end{equation}
Note that, given the partition, the optimal centers are given by
\[
\mu_l = \frac1{\left| S_l \right|} \sum_{i\in S_l}x_i.
\]

Lloyd's algorithm~\cite{Lloyd:kmeans} (also sometimes known as the $k$-means algorithm), is an iterative algorithm that alternates between
\begin{itemize}
\item Given centers $\mu_1,\dots,\mu_k$, assign each point $x_i$ to the cluster $S_l$ with nearest center 
\[
l=\argmin_{l=1,\dots, k}\left\|  x_i - \mu_l \right\|.
\]
\item Update the centers $\mu_l =  \frac1{\left| S_l \right|} \sum_{i\in S_l}x_i$.
\end{itemize}

Unfortunately, Lloyd's algorithm is not guaranteed to converge to the solution of~\eqref{eq:3:kmeans_obj}. Indeed, Lloyd's algorithm oftentimes gets stuck in local optima of~\eqref{eq:3:kmeans_obj}. In the following we will discuss convex relaxations for clustering, which can be used as an alternative algorithmic approach to Lloyd's algorithm, but since optimizing~\eqref{eq:3:kmeans_obj} is $NP$-hard there is no polynomial time algorithm that works in the worst-case (assuming the widely believed conjecture $P\neq NP$, see also Chapter~\ref{maxcutapprox})

\begin{figure}[h]
\begin{center}
\includegraphics[width=0.7\textwidth]{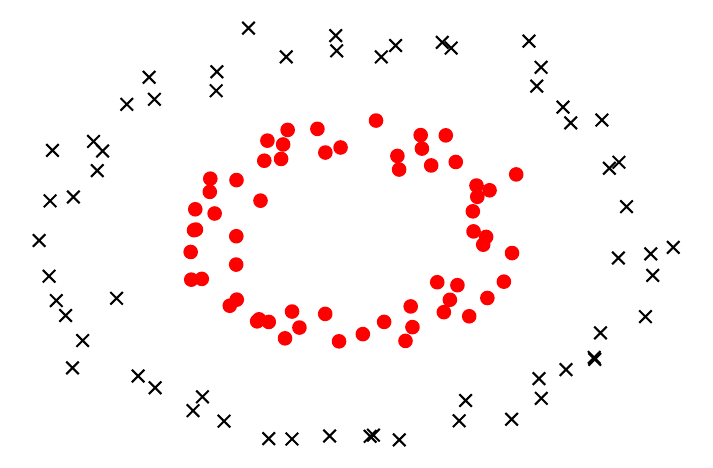}
\caption{Because the solutions of $k$-means are always convex clusters, it is not able to handle some cluster structures. This is one of the several limitations of $k$-means.}
\label{figure:3:cluster_kmeans_circles}
\end{center}
\end{figure}

While popular, $k$-means clustering has some potential issues:
\begin{itemize}
\item One needs to set the number of clusters a priori (a typical way to overcome this issue is by trying the algorithm for different number of clusters).

\item The way~\eqref{eq:3:kmeans_obj} is defined it needs the points to be defined in an Euclidean space, oftentimes we are interested in clustering data for which we only have some measure of affinity between different data points, but not necessarily an embedding in $\RR^p$ (this issue can be overcome by reformulating~\eqref{eq:3:kmeans_obj} in terms of distances only).

\item The formulation is computationally hard, so algorithms may produce suboptimal instances.

\item The solutions of $k$-means are always convex clusters. This means that $k$-means may have difficulty in finding cluster such as in Figure~\ref{figure:3:cluster_kmeans_circles}.

\end{itemize}

\section{Spectral clustering}

A natural way to try to overcome the issues of $k$-means depicted in Figure~\ref{figure:3:cluster_kmeans_circles} is by transforming the data into a graph and cluster the graph: Given the data points we can construct a weighted graph $G = (V,E,W)$ using a similarity kernel $K_{\epsilon}$, such as $K_{\epsilon}(u) = \exp\left(-\frac1{2\epsilon}u^2\right)$, by associating each point to a vertex and, for which pair of nodes, set the edge weight as
\[
w_{ij} = K_{\epsilon}\left( \|x_i-x_j\| \right).
\]
Another popular procedure to transform data into a graph is by constructing a graph where data points are connected if they correspond to the nearest neighbors. We note that this procedure only needs a measure of distance, or similarity, of data points and not necessarily that they lie in an Euclidean space. Given this motivation, and the prevalence of network data, we will now address the problem of clustering the nodes of a graph.

\subsubsection*{Normalized Cut}

Given a graph $G=(V,E,W)$, the goal is to partition the graph into clusters in a way that keeps as many of the edges, or connections, within the clusters and has as few edges as possible across clusters. We will focus on the case of two clusters, and briefly address extensions at the end of this chapter. A natural way to measure a vertex partition $\left(S,S^c\right)$ is
\[
\cut(S) = \sum_{i\in S}\sum_{j\in S^c} w_{ij}.
\]

If we represent the partition by a vector $y\in\{\pm 1\}^n$ where $y_i=1$ if $i\in S$, and $y_i=-1$ otherwise, then the cut is a quadratic form on the Graph Laplacian.

\begin{definition}[Graph Laplacian and Degree Matrix]
\label{def:graphLaplacian} 
Let $G=(V,E,W)$ be a graph and $W$ the matrix of weights (or adjacency matrix if the graph is unweighted). The degree matrix $D$ is a diagonal matrix with diagonal entries
\begin{equation}
    \label{degreematrix}
D_{ii} = \deg(i) = \sum_{j} w_{ij}.
\end{equation}

The graph Laplacian of $G$ is given by

\begin{equation}
\label{graphlalpacian}
L_G = D - W.
\end{equation}

Equivalently
\[
 L_G \defeq  \sum_{i<j}w_{ij} \left( e_i - e_j \right)\left( e_i - e_j \right)^T,
\]
recall that $e_i$ denotes the $i$-th canonical basis element with entries given by $(e_i)_j=\delta_{ij}$.
\end{definition}
Note that the entries of $L_G$ are given by
\[
 \left(L_G\right)_{ij} = \left\{   \begin{array}{ll}  -w_{ij} & \text{ if  } i\neq j \\ \deg(i) & \text{ if  } i=j.      \end{array} \right.
\]

If $S\subset V$ and $y\in\{\pm 1\}^n$ such that $y_i=1$ if $i\in S$, and $y_i=-1$ otherwise, then it is easy to see that
\[
\cut(S)  =  \frac14 \sum_{i<j}w_{ij}(y_i-y_j)^2.
\]
The following proposition establishes
\begin{equation}\label{eq:cut_Laplacian}
\cut(S) = \frac14 y^TL_Gy,
\end{equation}
for $y\in\{\pm 1\}^n$ such that $y_i=1$ if and only if $i\in S$. 

\begin{proposition}\label{prop:cut_Laplacian}
Let $G=(V,E,W)$ be a graph and $L_G$ its graph Laplacian. For any $x\in\RR^n$ 
\[
x^TL_Gx = \sum_{i<j}w_{ij}(x_i-x_j)^2.
\]
\end{proposition}
\begin{proof}
\begin{eqnarray*}
\sum_{i<j}w_{ij}\left( x_i - x_j \right)^2 & = & \sum_{i<j}w_{ij}\left[ x^T \left( e_i - e_j \right) \right]\left[ \left( e_i - e_j \right)^T x \right] \\
 & = & \sum_{i<j}w_{ij} x^T\left( e_i - e_j \right)\left( e_i - e_j \right)^Tx \\
 & = & x^T\left[ \sum_{i<j}w_{ij} \left( e_i - e_j \right)\left( e_i - e_j \right)^T \right]x.
\end{eqnarray*}
\end{proof}
Note that Proposition \ref{prop:cut_Laplacian} implies that the graph Laplacian $L_G$ is a positive semidefinite (PSD) matrix, since $x^T L x \geq 0$ for all $x$ (assuming non-negative weights $w_{ij} \geq 0$). This property can also be deduced directly from Definition \ref{def:graphLaplacian} of $L_G$ as the (non-negative) weighted sum of rank-1 PSD matrices.

While $\cut(S)$ is a good way of measuring the fit of a partition, it has a major drawback: the minimum cut is achieved for $S=\emptyset$ (since $\cut(\emptyset) = 0$) which is a rather meaningless choice of partition. Constraining the partition to be non-trivial would not overcome this drawback, because very unbalanced partitions would still be favored (e.g., partitions with $|S|=1$ containing just a single node). Below we discuss how to promote more balanced partitions.

\begin{remark}\label{remark:balancedLG}
One simple way to address this is to simply ask for an exactly balanced partition, $|S| = |S^c|$ (let us assume the number of vertices $n=|V|$ is even). We can then identify a partition with a label vector $y\in\{\pm 1\}^n$ where $y_i=1$ if $i\in S$, and $y_i=-1$ otherwise. Also, the balanced condition can be written as $\sum_{i=1}^n y_i = 0$. This means that we can write the minimum balanced cut as
\begin{equation}\label{eq:minbis-andspectrum}
\min_{\substack{S\subset V \\ |S|=|S^c|}}\cut(S) = 
\frac14 \min_{\substack{y\in\{-1,1\}^n \\ \1^Ty = 0}}  y^TL_Gy,
\end{equation}
which is suggestive of the connection between clustering and spectral properties of $L_G$. This connection will be made precise below.\footnote{The attentive reader might notice that, because any point in the hypercube has $\ell_2$-norm $\sqrt{n}$, \eqref{eq:minbis-andspectrum} readily implies $\mathrm{minbis(G)}\geq \frac{n}{4}\lambda_2(L_G)$, where $\mathrm{minbis(G)}$ denotes the minimum bisection of $G$ (also known as the minimum balanced cut). In the sequel we will establish a more useful version of such a spectral inequality.}
\end{remark}

Asking for the partition to be exactly balanced is too restrictive in many cases. There are several ways to evaluate a partition that are variations of $\cut(S)$ that take into account the intuition that one wants both $S$ and $S^c$ not to be too small (although not necessarily equal to $|V|/2$). A prime example is Cheeger's cut.

\begin{definition}[Cheeger's cut]
Given a graph and a vertex partition $(S,S^c)$, the Cheeger cut (also known as conductance, or expansion) of $S$ is given by
\[
h(S) = \frac{\cut(S)}{\min\{\vol(S), \vol(S^c) \}},
\]
where $\vol(S) = \sum_{i\in S}\deg(i)$.

Also, the Cheeger's constant of $G$ is given by
\[
h_G = \min_{S \subset V}  h(S).
\]
\end{definition}

A similar object is the Normalized Cut, $\Ncut$, which is given by
\[
\Ncut(S) = \frac{\cut(S)}{\vol(S)} + \frac{\cut(S^c)}{\vol(S^c)}.
\]

Note that $\Ncut(S)$ and $h(S)$ are tightly related, in fact it is easy to see that:
\[
h(S) \leq \Ncut(S) \leq 2h(S).
\]

\subsubsection{Normalized Cut as a spectral relaxation}

Below we will show that $\Ncut$ can be written in terms of a minimization of a quadratic form involving the graph Laplacian $L_G$, analogously to the balanced partition as described in Remark~\ref{remark:balancedLG}.

Recall that the cut value of a balanced partition can be written as
\[
\frac14 \min_{\substack{y\in\{-1,1\}^n \\ \1^Ty = 0}}  y^TL_Gy.
\]
An intuitive way to relax the balanced condition is to allow the labels $y$ to take values in two different real values $a$ and $b$ (e.g. $y_i = a$ if $i\in S$ and $y_j = b$ if $i\notin S$) but not necessarily $\pm1$. We can then use the notion of volume of a set to ensure a less restrictive notion of balanced by asking that
\begin{equation}\label{eq:1TDy0}
a\vol\left(S\right) + b\vol\left(S^c\right) = 0,
\end{equation}
where
\begin{equation}\label{eq:vol}
\vol(S) = \sum_{i\in S} \deg(i).
\end{equation}
Thus \eqref{eq:1TDy0} corresponds to $\1^TDy = 0$.

We also need to fix a scale for $a$ and $b$:
\[
a^2\vol\left(S\right) + b^2\vol\left(S^c\right) = 1,
\]
which corresponds to $y^TDy = 1$.

This suggests considering
\begin{equation*}
\min_{\substack{y\in\{a,b\}^n \\ \1^TDy = 0,\, y^TDy = 1}}  y^TL_Gy.
\end{equation*}

As we will see below, this corresponds precisely to $\Ncut$.

\begin{proposition}
For $a$ and $b$ to satisfy $a\vol\left(S\right) + b\vol\left(S^c\right) = 0$ and $a^2\vol\left(S\right) + b^2\vol\left(S^c\right) = 1$ it must be that
\[
a = \left(  \frac{\vol(S^c)}{\vol(S)\vol(G)}\right)^{\frac12} \quad \text{ and }\quad b =  -  \left(  \frac{\vol(S)}{\vol(S^c)\vol(G)}\right)^{\frac12}  ,
\]
corresponding to
\[
y_i = \left\{ \begin{array}{rl}  \left(  \frac{\vol(S^c)}{\vol(S)\vol(G)}\right)^{\frac12} & \text{ if } i\in S \\
  -  \left(  \frac{\vol(S)}{\vol(S^c)\vol(G)}\right)^{\frac12}  & \text{ if } i\in S^c.     \end{array}   \right.
\]
Note that $\vol$ is defined in \eqref{eq:vol}.
\end{proposition}

\begin{proof}
The proof involves only doing simple algebraic manipulations together with noticing that $\vol(S)+\vol(S^c) = \vol(G)$.
\end{proof}

\begin{proposition}
\[
\Ncut(S) = y^TL_Gy,
\]
where $y$ is given by
\[
y_i = \left\{ \begin{array}{rl}  \left(  \frac{\vol(S^c)}{\vol(S)\vol(G)}\right)^{\frac12} & \text{ if } i\in S \\
  -  \left(  \frac{\vol(S)}{\vol(S^c)\vol(G)}\right)^{\frac12}  & \text{ if } i\in S^c.     \end{array}   \right.
\]
\end{proposition}

\begin{proof}
\begin{eqnarray*}
y^TL_Gy & = & \frac12 \sum_{i,j} w_{ij} (y_i - y_j)^2 \\
& = & \sum_{i\in S} \sum_{j\in S^c} w_{ij} (y_i - y_j)^2 \\
& = & \sum_{i\in S} \sum_{j\in S^c} w_{ij} \left[   \left(  \frac{\vol(S^c)}{\vol(S)\vol(G)}\right)^{\frac12}  + \left(  \frac{\vol(S)}{\vol(S^c)\vol(G)}\right)^{\frac12}    \right]^2 \\
& = & \sum_{i\in S} \sum_{j\in S^c} w_{ij}\frac{1}{\vol(G)} \left[ \frac{\vol(S^c)}{\vol(S)} + \frac{\vol(S)}{\vol(S^c)} + 2   \right] \\
& = & \sum_{i\in S} \sum_{j\in S^c}w_{ij} \frac{1}{\vol(G)} \left[ \frac{\vol(S^c)}{\vol(S)} + \frac{\vol(S)}{\vol(S^c)} +  \frac{\vol(S)}{\vol(S)} + \frac{\vol(S^c)}{\vol(S^c)}  \right] \\
& = & \sum_{i\in S} \sum_{j\in S^c}w_{ij}  \left[ \frac{1}{\vol(S)} + \frac{1}{\vol(S^c)}  \right] \\
& = & \cut(S)  \left[ \frac{1}{\vol(S)} + \frac{1}{\vol(S^c)}  \right]\\
& = & \Ncut(S).
\end{eqnarray*}
\end{proof}

This means that finding the minimum $\Ncut$ corresponds to solving

\begin{equation}\label{eq:3:Ncut:norelax}
\begin{array}{rl}
\min & y^TL_Gy \\
\text{s. t.} & y\in \{a,b\}^n \text{ for some } a \text{ and } b \\
 & y^TDy = 1 \\
 & y^TD\1 = 0.
\end{array}
\end{equation}

Since solving~\eqref{eq:3:Ncut:norelax} is, in general, NP-hard, we consider a similar problem where the constraint that $y$ can only take two values is removed:

\begin{equation}\label{eq:3:Ncut:relaxed}
\begin{array}{rl}
\min & y^TL_Gy \\
\text{s. t.} & y\in \RR^n \\
& y^TDy = 1 \\
 & y^TD\1 = 0.
\end{array}
\end{equation}

Given a solution of~\eqref{eq:3:Ncut:relaxed} we can \emph{round} it to a partition by setting a threshold $\tau$ and taking $S = \left\{ i\in V:\, y_i \leq \tau  \right\}$. We will see below that $\eqref{eq:3:Ncut:relaxed}$ is an eigenvector problem (for this reason we call~\eqref{eq:3:Ncut:relaxed} a spectral relaxation).

In order to better see that~\eqref{eq:3:Ncut:relaxed} is an eigenvector problem (and thus computationally tractable), set $z = D^{\frac12}y$ and
\begin{equation}\label{eq:LLL_G}
\LLL_G = D^{-\frac12} L_G D^{-\frac12},
\end{equation}
then~\eqref{eq:3:Ncut:relaxed} is equivalent

\begin{equation}\label{eq:3:Ncut:relaxed_LLL}
\begin{array}{rl}
\min & z^T\LLL_Gz \\
\text{s. t.} & z\in \RR^n \\
& \|z\|^2 = 1 \\
 & \left(D^{\frac12}\1\right)^Tz = 0.
\end{array}
\end{equation}

Note that $\LLL_G = I - D^{-\frac12}WD^{-\frac12}$. We order its eigenvalues in increasing order $0=\lambda_1\left( \LLL_G\right) \leq \lambda_2\left( \LLL_G\right) \leq \cdots \leq  \lambda_n\left( \LLL_G\right)$. The eigenvector associated with the smallest eigenvalue is given by $D^{\frac12}\1$. By the variational interpretation of the eigenvalues, it follows that the solution of \eqref{eq:3:Ncut:relaxed_LLL} is $\lambda_2\left( \LLL_G\right)$ and the minimizer is given by the second smallest eigenvector of $\LLL_G = I - D^{-\frac12}WD^{-\frac12}$, which we call $v_2$. Note that this corresponds also to the second largest eigenvector of $D^{-\frac12}WD^{-\frac12}$. This means that the optimal $y$ in~\eqref{eq:3:Ncut:relaxed} is given by $\varphi_2 = D^{-\frac12}v_2$. This motivates Algorithm~\ref{algorithm:3:spectralclustering2clusters}, which is often referred to as {\em Spectral Clustering}:
\begin{algorithm}[h]
Given a graph $G=(V,E,W)$, let $v_2$ be the eigenvector corresponding to the second smallest eigenvalue of the normalized Laplacian $\LLL_G$, as defined in~\eqref{eq:LLL_G}. Let $\varphi_2 = D^{-\frac12}v_2$.
Given a threshold $\tau$ (one can try all different possibilities, or run $k$-means for $k=2$), set
\[
S = \{ i\in V:\, \varphi_2(i)\leq \tau \}.
\]
\caption{Spectral Clustering}
\label{algorithm:3:spectralclustering2clusters}
\end{algorithm}

\section{Cheeger's Inequality} 
Because the relaxation~\eqref{eq:3:Ncut:relaxed} is obtained from~\eqref{eq:3:Ncut:norelax} by removing a constraint we immediately have that
\[
\lambda_2\left(\LLL_G\right) \leq \min_{S\subset V} \Ncut(S).
\]
This means that
\[
\frac12\lambda_2\left(\LLL_G\right) \leq h_G.
\]

In what follows we will show a performance guarantee for Algorithm~\ref{algorithm:3:spectralclustering2clusters}.

\begin{lemma}\label{lemma:3:neededforCheeger}
There is a threshold $\tau$ producing a partition $S$ such that
\[
h(S) \leq \sqrt{2\lambda_2\left(\LLL_G\right)}.
\]
\end{lemma}

This implies in particular that
\[
h(S) \leq \sqrt{4h_G},
\]
meaning that Algorithm~\ref{algorithm:3:spectralclustering2clusters} is suboptimal at most by a square-root factor. Note that this also directly implies the famous Cheeger's Inequality, stated next. 

\begin{theorem}[Cheeger's Inequality]\label{thm:cheegerinequality}
Recall the definitions above. The following holds:
\[
\frac12\lambda_2\left(\LLL_G\right) \leq h_G \leq \sqrt{2\lambda_2\left(\LLL_G\right)}.
\]
\end{theorem}

Cheeger's inequality was first established for manifolds by Jeff Cheeger in 1970~\cite{JCheeger_1970}, and the graph version is due to Noga Alon and Vitaly Milman~\cite{NAlon_1986,NAlon_VMilman_1986} in the mid 80s. 
Cheeger's inequality for a closed manifold $M$ gives a bound on the area of a hypersurface $E$, that partitions $M$ into two disjoint parts.
\begin{theorem}[Cheeger's Inequality (Cheeger 1970)]
Let $h_M$ be the Cheeger isoperimetric constant of $M$, defined as
\begin{eqnarray*}
h_M = \inf_{E}	\frac	{\text{area of $E$}}	{\min\{\text{vol}(A),\text{vol}(B)\}}
\end{eqnarray*}
where $E$ is a smooth submanifold of $M$ of $n-1$ dimensions that divides it into two disjoint submanifolds $A$ and $B$.
Let $\lambda_M$ be the smallest positive eigenvalue of the Laplacian, or Laplace-Bertrami operator, on $M$.
Then
\begin{eqnarray*}
\lambda_M \geq \frac{h_M^2}{4}
\end{eqnarray*}
\end{theorem}

\subsubsection{Bounding the diameter of a graph using $\lambda_2(\LLL_G)$}
Before proving Cheeger's inequality for graphs, we first discuss another eigenvalue inequality, that gives a flavor of the connection between the second eigenvalue of the normalized graph Laplacian and the geometry of the graph (see also \cite{FanChung_SpectralGraphTheory}). 

Associate a cost $c(u,v)$ for each pair of nodes inversely proportional to the weight between them:
\begin{equation}
c(u,v) = \left\{\begin{array}{cc}
                  \frac{1}{w_{uv}} & \text{for } (u,v)\in E, \\ & \\
                  +\infty & \text{for } (u,v)\notin E.
                \end{array}
 \right.
\end{equation}
A {\em path} of length $k$ connecting $u$ and $v$ is of the form
$$P = (u=v_1,v_2,\ldots,v_k=v).$$
The cost associated with this path is
$$c(P) = \sum_{i=1}^{k-1} c(v_i,v_{i+1}).$$
The {\em geodesic distance} or {\em shortest path} between $u$ and $v$ is the minimum of the cost over all possible paths connecting them:
$$d_g(u,v) = \min \{c(P) \,| \, P(1)=u,\, P(k)=v,\; |P|=k\}$$
The {\em diameter} of the graph $G$ is the largest geodesic distance
$$\text{diam}(G) = \max_{u,v} d_g(u,v).$$

We demonstrate that for $\lambda_2(\LLL_G)$, the second smallest eigenvalue of
the normalized graph Laplacian $\LLL_G = D^{-1/2}L_G D^{-1/2}$,
\begin{equation}
\label{eq:diam-lambda}
\text{diam}(G) \geq \frac{
1}{\text{vol}(G)\lambda_2(\LLL_G)}.
\end{equation}
That is, a small second eigenvalue implies that the diameter of the graph must be large. We already know that $\lambda_2(\LLL_G) = 0$ implies that the graph is disconnected and the diameter is infinite. The bound (\ref{eq:diam-lambda}) is more quantitative from that standpoint.

\smallskip
\begin{proof}[of~\eqref{eq:diam-lambda}] Taking $f$ to be an eigenfunction achieving the Rayleigh quotient
characterization $$\lambda_2=\inf_{f^T D\mathbf{1}=0}
\frac{\frac{1}{2}\sum_{u,v} w_{uv}(f(v)-f(u))^2}{\sum_{v} f(v)^2 \deg(v)}.$$
Choose $v_0, u_0$ such that $$f(v_0)=\max_v |f(v)|,$$ and $$f(u_0)<0,$$
which exist since
$$\sum_{v} f(v)\deg(v)=0.$$ Now, suppose $$P=(u_0=v_1,v_2,\ldots,v_k=v_0)$$ is the shortest path
from $u_0$ to $v_0$. We have
\begin{eqnarray*}
\lambda_2 &=& \frac{\frac{1}{2}\sum_{u,v} w_{uv}(f(v)-f(u))^2}{\sum_{v} f(v)^2 \deg(v)} \\
&\geq&
 \frac{\sum_{i=1}^{k-1}
w_{v_i,v_{i+1}}(f(v_i)-f(v_{i+1}))^2}{f(v_0)^2\sum_{v} d_v} \\
& \geq & \frac{\sum_{i=1}^{k-1}
w_{v_i,v_{i+1}}(f(v_i)-f(v_{i+1}))^2}{f(v_0)^2\text{vol}(G)} \frac{d_g(u_0,v_0)}{\text{diam}(G)} \\
&=& \frac{\sum_{i=1}^{k-1}
w_{v_i,v_{i+1}}(f(v_i)-f(v_{i+1}))^2}{f(v_0)^2\text{vol}(G)} \frac{\sum_{i=1}^{k-1} w_{v_i,v_{i+1}}^{-1}}{\text{diam}(G)} \\
&\geq& \frac{\left(\sum_{i=1}^{k-1}
f(v_i)-f(v_{i+1})\right)^2}{f(v_0)^2\text{vol}(G)\text{diam}(G)} \\
&=& \frac{(
f(v_0)-f(u_0))^2}{f(v_0)^2\text{vol}(G)\text{diam}(G)} \\
&\geq& \frac{
1}{\text{vol}(G)\text{diam}(G)},
\end{eqnarray*}
there the second-to-last inequality is an application of Cauchy-Schwarz inequality in dimension $k-1$ (and relies on the fact that the weights $w_{v_i,v_{i+1}}$ are positive).
\end{proof}

\subsubsection{The proof of Cheeger's Inequality}

The upper bound in Cheeger's inequality (corresponding to Lemma~\ref{lemma:3:neededforCheeger}) is more interesting but more difficult to prove, and it is often referred to as the ``the difficult part'' of Cheeger's inequality. We will prove this Lemma in what follows. There are several proofs of this inequality (see~\cite{Chung_CheegersIneq} for four different proofs!). The proof that follows is an adaptation of the proof in this blog post~\cite{trevisan:blog:cheegerproof} for the case of weighted graphs.

\bigskip

\begin{proof}[of Lemma~\ref{lemma:3:neededforCheeger}]

We will show that given $y\in\RR^n$ satisfying
\[
\RRR(y) \defeq \frac{y^TL_Gy}{y^TDy} \leq \delta,
\]
and $y^TD\1 = 0$, there is a ``rounding of it'', meaning a threshold $\tau$ and a corresponding choice of partition
\[
S = \{  i\in V:\, y_i \leq \tau \}
\]
such that
\[
h(S) \leq \sqrt{2\delta}.
\]
Since $y=\varphi_2$ satisfies the conditions with $\delta = \lambda_2\left(\LLL_G \right)$, this proves the Lemma.

We will pick this threshold at random and use the probabilistic method to show that at least one of the thresholds works.

First we can, without loss of generality, assume that $y_1\leq \cdots \leq y_n$ (we can simply relabel the vertices). Also, note that scaling of $y$ does not change the value of $\RRR(y)$. Also, if $y^TD\1=0$ adding a multiple of $\1$ to $y$ can only decrease the value of $\RRR(y)$: the numerator does not change and the denominator $(y+c\1)^TD(y+c\1) = y^TDy + c^2 \1^TD\1 \geq y^TDy$.

This means that we can construct (from $y$ by adding a multiple of $\1$ and scaling) a vector $x$ such that
\[
x_1\leq \cdots \leq x_n, \
x_m=0, \ \text{ and } x_1^2 + x_n^2 = 1,
\]
and
\[
\frac{x^TL_Gx}{x^TDx} \leq \delta,
\]
where $m$ be the index for which $\vol(\{1,\dots,m-1\}) \leq \vol(\{m,\dots,n\})$ but $\vol(\{1,\dots,m\}) > \vol(\{m+1,\dots,n\})$.

We consider a random construction of $S$ with the following distribution. $S = \{  i\in V:\, x_i \leq \tau \}$ where $\tau \in [x_1,x_n]$ is drawn at random with the distribution
\[
\Prob\left\{  \tau \in [a,b]   \right\} = \int_{a}^{b} 2|\tau | d\tau,
\]
where $x_1 \leq a\leq b \leq x_n$.

It is not difficult to check that
\[
\Prob\left\{  \tau \in [a,b]   \right\} = \left\{   \begin{array}{rl}     \left|  b^2-a^2 \right|   & \text{ if } a \text{ and } b \text{ have the same sign} \\
       a^2+b^2  & \text{ if } a \text{ and } b \text{ have different signs}    \end{array}  \right.
\]

Let us start by estimating $\EE \cut(S)$.
\begin{eqnarray*}
\EE \cut(S) & = & \EE \frac12 \sum_{i\in V}\sum_{j\in V} w_{ij}\1_{(S,S^c) \text{ cuts the edge } (i,j)} \\
 & = & \frac12 \sum_{i\in V}\sum_{j\in V} w_{ij}\Prob\{  (S,S^c) \text{ cuts the edge } (i,j)  \}
\end{eqnarray*}

Note that $ \Prob\{  (S,S^c) \text{ cuts the edge } (i,j)  \} $ is  $\left| x_i^2 - x_j^2 \right| $ if $x_i$ and $x_j$ have the same sign and $x_i^2 + x_j^2$ otherwise. Both cases can be conveniently upper bounded by
$ \left| x_i - x_j\right| \left( | x_i | + |x_j | \right)$. This means that
\begin{eqnarray*}
\EE \cut(S) & \leq  & \frac12 \sum_{i , j }w_{ij}  \left| x_i - x_j\right| \left( | x_i | + |x_j | \right) \\
 & \leq & \frac12\sqrt{  \sum_{ij}  w_{ij} (x_i-x_j)^2      } \sqrt{  \sum_{ij}  w_{ij} (|x_i|+|x_j|)^2      },
\end{eqnarray*}
where the second inequality follows from the Cauchy-Schwarz inequality.

From the construction of $x$ we know that
\[
 \sum_{ij}  w_{ij} (x_i-x_j)^2  = 2x^T L_G x \leq 2\delta x^TDx.
\]
Also,
\[
 \sum_{ij}  w_{ij} (|x_i|+|x_j|)^2 \leq  \sum_{ij}  w_{ij} (2x_i^2+2x_j^2)  
 = 2 \left( \sum_i \deg(i)2x_i^2\right)
 = 4x^TDx.
\]
This means that
\[
\EE \cut(S)  \leq \frac12 \sqrt{2\delta x^TDx} \sqrt{ 4 x^TDx} = \sqrt{2\delta }\, x^TDx.
\]

On the other hand,
\[
\EE \min\{ \vol{S},\vol{S^c} \} = \sum_{i=1}^n \deg(i) \Prob\{ x_i \text{ is in the smallest set (in terms of volume)} \},
\]
to break ties, if $\vol(S) = \vol(S^c)$ we take the ``smallest'' set to be the one with the first indices.

Note that $m$ is always in the largest set. Any vertex $j<m$ is in the smallest set if $ x_j \leq  \tau \leq x_m = 0$ and any $j>m$ is in the smallest set if $0 = x_m \leq \tau \leq x_j$. This means that,
\[
\Prob\{ x_j \text{ is in the smallest set (in terms of volume)}\}  = x_j^2.
\]
Which means that
\[
\EE \min\{ \vol{S},\vol{S^c} \} = \sum_{i=1}^n \deg(i)x_i^2 = x^TDx.
\]

Hence,
\[
\frac{\EE \cut(S)  }{\EE \min\{ \vol{S},\vol{S^c} \}} \leq \sqrt{2\delta}.
\]

Note, however, that $\frac{\EE \cut(S)  }{\EE \min\{ \vol{S},\vol{S^c} \}}$ and  $\EE\frac{\cut(S)  }{\min\{ \vol{S},\vol{S^c} \}}$ do not necessarily coincide. Therefore, we do not necessarily have
\[
\EE\frac{\cut(S)  }{\min\{ \vol{S},\vol{S^c} \}} \leq \sqrt{2\delta}.
\]
However, since both random variables are positive,
\[
\EE \cut(S) \leq \EE \min\{ \vol{S},\vol{S^c}\} \sqrt{2\delta},
\]
or equivalently
\[
\EE \left[ \cut(S) -  \min\{ \vol{S},\vol{S^c} \}\sqrt{2\delta} \right] \leq 0,
\]
which guarantees, by the probabilistic method, the existence of $S$ such that
\[
\cut(S) \leq \min\{ \vol{S},\vol{S^c} \} \sqrt{2\delta},
\]
which is equivalent to
\[
h(S) = \frac{\cut(S) }{ \min\{ \vol{S},\vol{S^c} \}} \leq  \sqrt{2\delta},
\]
which concludes the proof of the Lemma.
\end{proof}

\begin{remark}
Spectral clustering can also be viewed from the perspective of random walks. In fact, Proposition~\ref{prop:3:NcutandRandomWalks} shows that spectral clustering can be viewed as attempting to find clusters so as to minimize the probability of the random walker to jump from one cluster to the other.
\end{remark}

\subsubsection*{Multiple Clusters}

Much of the above can be easily adapted to multiple clusters. Algorithm~\ref{algorithm:3:spectralclustering} is a natural extension of spectral clustering to multiple clusters.\footnote{We will see in Chapter~\ref{c:diffusion} that the map $\phi:V\to\RR^{k-1}$ defined in Algorithm~\ref{algorithm:3:spectralclustering} can also be used for data visualization, not just clustering.} 

\begin{algorithm}[h]
Given a graph $G=(V,E,W)$, let $v_2,\dots,v_k$ be the eigenvectors corresponding to the second through $k$'th eigenvalues of the normalized Laplacian $\LLL_G$, as defined in~\eqref{eq:LLL_G}. Let $\varphi_m = D^{-\frac12}v_m$. Consider the map $\phi:V\to\RR^{k-1}$ defined as
\[
\phi(v_i) =\left[  \begin{array}{c}   \varphi_2(i) \\ \vdots \\  \varphi_{k}(i) \end{array} \right].
\]
Cluster the $n$ points in $k-1$ dimensions into $k$ clusters using $k$-means.
\caption{Spectral Clustering}
\label{algorithm:3:spectralclustering}
\end{algorithm}

There is also an analogue of Cheeger's inequality. A natural way of evaluating $k$-way clustering is via the $k$-way expansion constant (see~\cite{JRLee_SOGharan_LTrevisan_2011}):
\[
 \rho_G(k) = \min_{S_1,\dots,S_k} \max_{l=1,\dots,k} \left\{  \frac{\cut(S_l)}{\vol(S_l)}   \right\},
\]
where the maximum is over all choice of $k$ disjoint subsets of $V$ (but not necessarily forming a partition).

Another natural definition is
\[
 \varphi_G(k) = \min_{S: \vol{S}\leq \frac1k \vol(G)} \frac{\cut(S)}{\vol(S)}.
\]
It is easy to see that
\[
 \varphi_G(k) \leq \rho_G(k).
\]
The following are analogues of Cheeger's inequality for multiple clusters.

\begin{theorem}[\cite{JRLee_SOGharan_LTrevisan_2011}]
Let $G=(V,E,W)$ be a graph and $k$ a positive integer
\begin{equation}\label{eq:3:multywaycheeger:k2}
 \rho_G(k) \leq \OOO\left(k^2\right) \sqrt{\lambda_k}.
\end{equation}
Also,
\[
 \rho_G(k) \leq \OOO\left(\sqrt{\lambda_{2k} \log k} \right).
\]
\end{theorem}

\section*{Exercises}
\addcontentsline{toc}{section}{Exercises}

\begin{definition}[Irreducible matrix]
    A matrix $A \in \mathbb{R}^{n \times n}$ is called irreducible if there is no permutation matrix $P$ such that
    \begin{equation*}
        P^\top A P = \begin{pmatrix} A_{11} & A_{12} \\ 0 & A_{22} \end{pmatrix},
    \end{equation*}
    where $A_{11}$ and $A_{22}$ are square matrices (not necessarily of same dimensions). In other words, an irreducible matrix cannot be transformed into block upper-triangular matrix by simultaneous row/column permutations.
\end{definition}

\begin{myexercise}[\level\level\sep Irreducibility and graphs]\label{prob:irreducible_matrix}
    Let $A \in \R^{n\times n}$ be a matrix with non-negative entries. We define $G(A)$, a directed graph associated to $A$, in the following way: there is a link from $i$ to $j$ if and only if $A_{ij} > 0$.
    \begin{enumerate}[(a)]
        \item Prove that if $A$ is irreducible, and $x$ is its eigenvector with non-negative entries, then $x$ has only positive entries.
        \item Show that the statement above is false if we drop the assumption that $A$ is irreducible.
        \item Show that $A$ is irreducible if and only if the associated graph $G(A)$ is strongly connected, which means that for every ordered pair of nodes $(i,j)$ there is a path from $i$ to $j$ (of any length).
    \end{enumerate}
\end{myexercise}

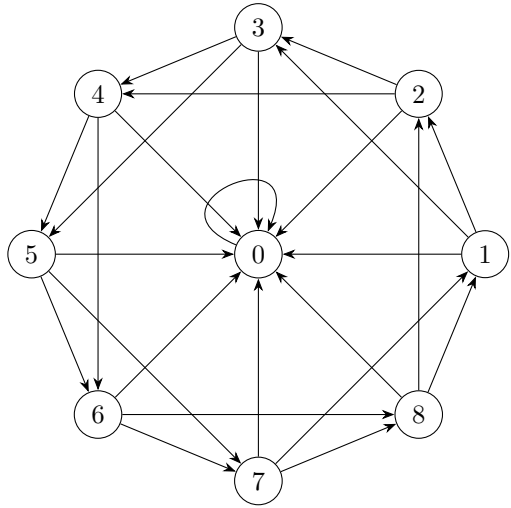
\begin{figure}[h!]
    \centering
    \begin{tikzpicture}[>=Stealth, node distance=2cm,
        every node/.style={circle, draw, minimum size=6mm}]

        \node (0) at (0,0) {0};
        \node (1) at (0:3) {1};
        \node (2) at (45:3) {2};
        \node (3) at (90:3) {3};
        \node (4) at (135:3) {4};
        \node (5) at (180:3) {5};
        \node (6) at (225:3) {6};
        \node (7) at (270:3) {7};
        \node (8) at (315:3) {8};

        \foreach \i in {1,2,3,4,5,6,7,8}
            \draw[->] (\i) -- (0);

        \draw[->, loop above, looseness=8, in=67.5, out=157.5] (0) to (0);

        \foreach \i/\j in {1/2,2/3,3/4,4/5,5/6,6/7,7/8,8/1}
            \draw[->] (\i) -- (\j);

        \foreach \i/\j in {1/3,2/4,3/5,4/6,5/7,6/8,7/1,8/2}
            \draw[->] (\i) -- (\j);

    \end{tikzpicture}
    \caption{Example of the directed graph from Problem~\ref{prob:pagerank} with $n=8$ and $k=2$.}
    \label{fig:digraph}
\end{figure}

\begin{myexercise}[\level\level\level\sep PageRank and Random Teleports]\label{prob:pagerank}
    In the lectures, we considered the PageRank algorithm designed for ranking pages based on their importance by analysing their ingoing and outgoing links. However, there exist graphs such that PageRank fails to predict meaningful scores. We will consider a simple fix for this problem which is often referred to as Random Teleports.

    Let $n > k > 1$. Consider a directed graph on $n + 1$ vertices, labelled $0, 1, \ldots, n$, with the following links. Vertex $0$ links only to itself, and any other vertex $j \in [n]$ has $k$ outgoing edges to its next $k$ vertices: $j + 1, \ldots, j + k \mod n$, and an edge to vertex $0$. See Figure~\ref{fig:digraph} for an example of such a graph.
    \begin{enumerate}[(a)]
        \item Compute the rank of vertices according to the PageRank scheme described in Section 4.1 of Lecture Notes.
        \item We define PageRank with Random Teleports as follows: with probability $\beta$ a random walker follows a link at random, and with probability $ 1 - \beta$, jumps to a random vertex (link or vertex is chosen uniformly at random). We form a new random walk matrix $M \in \mathbb{R}^{(n+1)\times (n+1)}$ whose entries $m_{ij}$ equal to the probability of going from vertex $j$ to vertex $i$. The ranking is then defined as the leading eigenvector of the constructed matrix $M$. For $k=1$ and fixed $0 < \beta < 1$ compute the PageRank scores for nodes with the teleport probability $1 - \beta$.
    \end{enumerate}
\end{myexercise}
\begin{remark}
    Have you recognized the connection with the notion of irreducibility in this exercise?
\end{remark}

\begin{myexercise}[\level\sep Lloyd's algorithm - monotonicity]\label{prob:lloyd_monotonicity}
    Recall the problem of $k$-means clustering: given $k \in \N$, we wish to separate $n$ points $x_1, \ldots, x_n \in \R^p$ (with $n \geq k$) into $k$ clusters using the following objective function
    \begin{equation}\label{eq:kmeansobjective}
        \text{cost}_2(S_1, \ldots, S_k; \mu_1, \ldots, \mu_k) \coloneqq \sum_{l=1}^k \sum_{i \in S_l} \norm{x_i - \mu_l}_2^2.
    \end{equation}
    This function depends on clusters $S_1, \ldots, S_k$ (that partition $[n]$) and their centers $\mu_1, \ldots, \mu_k \in \R^p$. Denote the minimum value by
    \begin{equation*}
        \text{opt}_2 \coloneqq \min_{\substack{\text{partition }S_1,\ldots,S_k\\\text{centers }\mu_1, \ldots, \mu_k}} \text{cost}_2(S_1, \ldots, S_k; \mu_1, \ldots, \mu_k)
    \end{equation*}
    Prove the following three properties.
    \begin{enumerate}[(a)]
        \item Given a choice for the partition $S_1, \ldots, S_k$ (of non-empty sets), the centers that minimize (\ref{eq:kmeansobjective}) are given by
        \begin{equation*}
            \mu_l = \frac{1}{\abs{S_l}}\sum_{i \in S_l}x_i.
        \end{equation*}
        
        \item Given the centers $\mu_1, \ldots, \mu_k \in \R^p$ the partition that minimizes (\ref{eq:kmeansobjective}) assigns each point $x_i$ to the cluster
        \begin{equation*}
            l = \argmin_{l=1, \ldots, k}\norm{x_i-\mu_l}_2.
        \end{equation*}

        \item Show that the optimization problem above can be rephrased in terms of the simplified objective function, that only depends on pairwise distances (double summation is taken only over unordered pairs):
        \begin{equation}\label{eq:kmeans_alternative}
            \text{opt}_2 = \min_{\text{partition }S_1,\ldots,S_k}
            \sum_{l=1}^k \frac{1}{|S_l|}\sum_{i,j \in S_l}\|x_i-x_j\|_2^2.
        \end{equation}
    \end{enumerate}
\end{myexercise}
\begin{hint}
    (c): Expand the square in \eqref{eq:kmeans_alternative} at the optimal choice of centers $\mu_l = \frac{1}{|S_l|}\sum_{i\in S_l} x_i$.
\end{hint}

\begin{myexercise}[\level\sep Lloyd's algorithm - convergence]\label{prob:lloyd_convergence}
    Given any set of $n$ points in $\R^p$, prove that Lloyd's algorithm stops after a finite number of iterations, in other words, that the objective eventually stops decreasing.
\end{myexercise}
\begin{hint}
    There are finitely many partitions of $n$ points.
\end{hint}

\begin{myexercise}[\level\level\sep Lloyd's algorithm - different objective]\label{prob:lloyd_different_objective}
    Let $n \geq k \geq 2$, and $x_1, \ldots, x_n \in \R^p$. Instead of minimizing sum-of-squares of $\ell_2$ norms \eqref{eq:kmeansobjective}, suppose we want to minimize an objective function with $\ell_1$ norms:
    \begin{equation}\label{eq:kmeansobjective1}
        \text{cost}_1(S_1, \ldots, S_k; \mu_1, \ldots, \mu_k) \coloneqq \sum_{l=1}^k \sum_{i \in S_l} \norm{x_i - \mu_l}_1.
    \end{equation}
    Denote the minimum value by
    \begin{equation*}
        \text{opt}_1 \coloneqq \min_{\substack{\text{partition }S_1,\ldots,S_k\\\text{centers }\mu_1, \ldots, \mu_k}} \text{cost}_1(S_1, \ldots, S_k; \mu_1, \ldots, \mu_k).
    \end{equation*}
    \begin{enumerate}[(a)]
        \item Given a choice for the partition $S_1, \ldots, S_k$ (of non-empty sets), which centers do minimize the alternative objective function \eqref{eq:kmeansobjective1}? A proof of minimality needs to be provided.
        \item Develop an algorithm analogous to Lloyd's algorithm using the alternative objective function \eqref{eq:kmeansobjective1}.
        \item Prove that it is always the case that $\text{opt}_2 \leq \text{opt}_1^2$.
    \end{enumerate}
\end{myexercise}

\begin{myexercise}[\level\sep Properties of the adjacency matrix]\label{prob:graph_adjacency_matrix}
    Let $G = (V,E)$ be a graph and $A$ its adjacency matrix.
    \begin{enumerate}[(a)]
        \item Show that for any graph $G$ the adjacency matrix $A$ satisfies $\norm{A} \geq d_{\text{ave}} \coloneqq \frac{1}{n}\sum_{i \in [n]}\deg(i)$.
        
        \item Show that for any graph $G$ the adjacency matrix $A$ satisfies $\norm{A} \leq d_{\text{max}} \coloneqq \max_{i \in [n]}\deg(i)$.
        
        \item Conclude if $G$ is $d$-regular then $\norm{A} = d$.
    \end{enumerate} 
\end{myexercise}

\begin{myexercise}[\level\sep Properties of the Laplacian matrix]\label{prob:graph_laplacian_matrix}
    Let $G = (V,E)$ be a graph and $L$ its Laplacian matrix. Recall the convention: eigenvalues of $L$ are ordered in non-decreasing order, i.e. $\lambda_1(L) \leq \lambda_2(L) \leq \ldots \leq \lambda_n(L)$.
    \begin{enumerate}[(a)]
        \item Show that we can express the Laplacian matrix $L$ as
        \begin{equation*}
            L = \sum_{(i,j) \in E} (e_i - e_j)(e_i - e_j)^\top
        \end{equation*}
        
        \item Deduce that $L$ is a positive semi-definite matrix.

        \item Let $S$ be a component (maximal connected subgraph) in $G$, and $\1_S$ a vector given by
        \begin{equation*}
            (\1_S)_i = \begin{cases}
                1 &\text{if } i\in S,\\
                0 &\text{otherwise}.
            \end{cases}
        \end{equation*}
        Show that $\1_S$ is in the kernel of $L$.

        \item Deduce that
        \begin{itemize}
            \item the smallest eigenvalue $\lambda_1(L)$ is always zero;
            \item the second smallest eigenvalue $\lambda_2(L)$ is zero if and only if $G$ is disconnected;
            \item the nullity of $L$ (i.e. dimension of the kernel $\ker(L)$) equals the number of components of $G$.
        \end{itemize}
    \end{enumerate} 
\end{myexercise}

\begin{myexercise}[\level\sep Normalized Laplacian]\label{prob:normalized_laplacian}
    Given an undirected weighted graph $G = (V, E, W)$, we define the normalized Laplacian matrix $\LL_G = D^{-1/2} L_G D^{-1/2}$, where $D$ is the degree matrix and $L_G$ is the graph Laplacian.
    \begin{enumerate}[(a)]
        \item Show that $\LL_G$ is symmetric and PSD (positive semi-definite).
        
        \item Show that all the eigenvalues of $\LL_G$ are real numbers, between $0$ and $2$.
    \end{enumerate}
\end{myexercise}

\begin{myexercise}
A contraction of a graph $G$ is formed by identifying two distinct vertices, say
$u$ and $v$, into a single vertex $v^*$. The weights of edges incident to $v^*$ are defined as
follows:
\begin{eqnarray*}
w(x,v^*) &=& w(x,u) + w(x,v), \\
w(v^*,v^*) &=& w(u,u) + w(v,v) + 2w(u,v).
\end{eqnarray*}
Show that if $H$ is formed by contractions from a graph $G$, then the second eigenvalue of the normalized Laplacians satisfy
$$\lambda_2(\LL_{G}) \leq \lambda_2(\LL_{H})$$    
\end{myexercise}

\begin{myexercise}[\level\level\sep Tightness of the upper bound in Cheeger's inequality]\label{prob:cheeger_upper_bound}
    Let $n \geq 4$ be an even number, and $C_n$ a cycle graph on $n$ vertices, labelled $1$ to $n$. We choose uniform weights $w_{ij} = \1\bras{\{i,j\} \in E}$, so that $W = A$.
    \begin{enumerate}[(a)]
        \item Prove that for every cut $S$, with $\emptyset \subsetneq S \subsetneq [n]$, its Cheeger's cut is lower bounded as
        \begin{equation*}
            h(S) \geq \frac{2}{n}.
        \end{equation*}
        
        \item Denote by $\lambda_2(C_n)$ the second smallest eigenvalue of the Laplacian of the graph $C_n$. Prove that 
        \begin{equation*}
            \lambda_2(C_n) \leq \frac{c}{n^2},
        \end{equation*}
        where $c > 0$ is an absolute constant.
        
        \item Conclude that the upper bound in Cheeger's inequality is tight up to an absolute constant.
    \end{enumerate}
\end{myexercise}
\begin{hint}    
    (b): Consider the quadratic form $x^\top L_{C_n} x$ for the vector $x \in \R^n$ given by $x_i = \abs{i-\frac{n}{2}}-\frac{n}{4}$.
\end{hint}

\begin{myexercise}[\level\level\sep Tightness of the lower bound in Cheeger's inequality]\label{prob:cheeger_lower_bound}
    Let $d \geq 2$ be an integer, $G = (V,E)$ be the $d$-dimensional hypercube, and $\LL_G$ its normalized Laplacian. We index the $n=2^d$ vertices of the hypercube by $d$-dimensional $\brac{0, 1}$-vectors, i.e. $V = \brac{0,1}^d$, and given any $x, y \in V$, we have $\brac{x,y} \in E$ if and only if $x$ and $y$ differ in exactly one coordinate.
    The example for $d = 3$ is given in Figure~\ref{fig:hypercube}.
    
    \begin{figure}[h!]
        \centering
        \begin{tikzpicture}[scale=3]
            \coordinate (A) at (0,0,0);
            \coordinate (B) at (1,0,0);
            \coordinate (C) at (1,1,0);
            \coordinate (D) at (0,1,0);
            \coordinate (E) at (0,0,1);
            \coordinate (F) at (1,0,1);
            \coordinate (G) at (1,1,1);
            \coordinate (H) at (0,1,1);
        
            \draw[thick] (A) -- (B) -- (C) -- (D) -- cycle;
        
            \draw[thick] (E) -- (F) -- (G) -- (H) -- cycle;
        
            \draw[thick] (A) -- (E);
            \draw[thick] (B) -- (F);
            \draw[thick] (C) -- (G);
            \draw[thick] (D) -- (H);

            \foreach \point in {(A), (B), (C), (D), (E), (F), (G), (H)}
                \fill \point circle (0.5pt);
        
            \node[ left, font=\footnotesize] at (A) {(0,0,0)};
            \node[ right, font=\footnotesize] at (B) {(1,0,0)};
            \node[ right, font=\footnotesize] at (C) {(1,1,0)};
            \node[ left, font=\footnotesize] at (D) {(0,1,0)};
            \node[ left, font=\footnotesize] at (E) {(0,0,1)};
            \node[ right, font=\footnotesize] at (F) {(1,0,1)};
            \node[ right, font=\footnotesize] at (G) {(1,1,1)};
            \node[ left, font=\footnotesize] at (H) {(0,1,1)};
        
        \end{tikzpicture}
        \caption{3-dimensional hypercube.}
        \label{fig:hypercube}
    \end{figure}
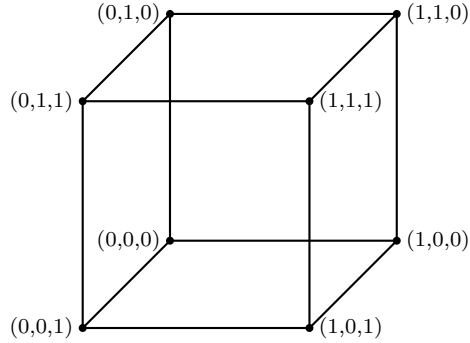

    Given a (possibly empty) subset $T \subseteq [d]$ let $v_T \in \R^n$ be a vector, whose coordinates are indexed by $n=2^d$ vertices of the hypercube and defined by
    \begin{equation*}
        v_T(x) = (-1)^{\sum_{i \in T} x_i},
    \end{equation*}
    where $x_i$ is the $i$-th coordinate of the vertex $x\in\{0,1\}^d$.
    Also, let $S_T \subseteq V$ be the subset of vertices given by
    \begin{equation*}
        S_T = \brac{x \in V \colon v_T(x) = 1}.
    \end{equation*}
    \textit{[When $T = \emptyset$, we interpret the empty sum as zero, i.e. $\sum_{i \in \emptyset}x_i = 0$.]}
    
    \begin{enumerate}[(a)]
        \item Compute $h(S_{\brac{1}})$, the Cheeger's cut of the subset $S_{\brac{1}}$.

        \item Show that every vertex in $S_T$ has exactly $d-\abs{T}$ neighbours in $S_T$.
        
        \item Show that for any $T \subseteq [d]$, $v_T$ is an eigenvector of $\LL_G$ with eigenvalue $\frac{2\abs{T}}{d}$.
        
        \item Show that if $T' \subseteq [d]$ is distinct from $T$, then $v_T$ and $v_{T'}$ are orthogonal.

        \item Conclude that for any $0 \leq k \leq d$, the eigenspace corresponding to the eigenvalue $\frac{2k}{d}$ has dimension ${d \choose k}$.

        \item Compute $h_G$, the Cheeger’s constant of $G$.
    \end{enumerate}
\end{myexercise}

\begin{myexercise}[\level\level\level\sep Characterization of bipartite graphs]\label{prob:bipartite_graphs}
    Let $G = (V,E)$ be a graph.
    \begin{enumerate}[(a)]
        \item Prove that $G$ is bipartite if and only if there is no cycle of odd length.
        \item Prove that for any $k \geq 1$, $A^k$ is a matrix whose entry $(i, j)$ contains the number of paths of length $k$ between $i$ and $j$.
        \item Prove that $G$ is bipartite if and only if the spectrum of $A$ is symmetric, i.e. $\lambda_i(A) = - \lambda_{n+1-i}(A)$ for all $i \in [n]$.
        \item If in addition $G$ is connected and $d$-regular, prove that $G$ is bipartite if and only $\lambda_1(A) = - \lambda_n(A)$.
    \end{enumerate} 
\end{myexercise}
\begin{hint}
    (d): Try similar ideas as in the proof of the Gershgorin circle theorem, using $\lambda_n(A) = -d$.
\end{hint}

\begin{myexercise}
We will try various clustering methods on MNIST.  \\
(a) Implement your own version of Lloyd's k-means algorithm (do not use an existing k-means function; programming  it yourself will make you better aware of potential pitfalls or other aspects you need to pay attention to). Test it on MNIST. How well does k-means cluster the data? You need to come up with a performance metric to quantify what ``how well'' means (in case of MNIST we know the ground truth.). Also note that k-means may give you different results, based on different random initializations - how do you handle that?

\noindent
(b) Implement  spectral clustering and apply it to MNIST. There are parameters that you need to choose (a good value for $\sigma$, how may eigenvectors do you keep,...).
Recall that spectral clustering has a second step (the ``rounding step''), in which you use your k-means algorithm from (a) to cluster the spectrally-embedded data. As in (a), evaluate the performance of your clustering algorithm.
\end{myexercise}

\chapter{Nonlinear Dimension Reduction and Diffusion Maps}
\label{c:diffusion}

In Chapter~\ref{c:svd} we discussed dimension reduction via Principal Component Analysis. Many datasets however have low dimensional structure that is not linear. In this chapter we will discuss nonlinear dimension reduction techniques. Just as with spectral clustering in Chapter~\ref{c:graphs} we will focus on graph data while noting that most types of data can be transformed into a weighted graph by means of a similarity kernel (Section~\ref{subs:diffusionmapskernel}). The goal of this chapter is to embed the nodes of a graph in Euclidean space in a way that best preserves the intrinsic geometry of the graph (or the data that gave rise to the graph).

\section{Diffusion maps}

Diffusion maps \cite{RRCoifman_SLafon_2006} will allow us to represent (weighted) graphs $G=(V,E,W)$ in $\RR^d$, i.e. associating, to each node, a point in $\RR^d$. 
Before presenting diffusion maps, we will introduce a few important notions. The reader may notice the similarities with the objects described in the context of PageRank in Chapter~\ref{c:graphs}, the main difference is that here the connections between graphs have no direction, meaning that the weight matrix $W$ is symmetric\footnote{The goal in the context of PageRank was to do ranking, in which case a directed network is usually best. In this Chapter we will be interested in understanding similarities between data points, in which case it is more common to make use of undirected graphs.}; this will be crucial in the derivations below (as we will see in Section~\ref{sec:MarkovChains} this symmetry can be viewed as corresponding to a certain Markov Chain being \emph{reversible}).

Given $G=(V,E,W)$ we consider a random walk (with independent steps) on the vertices of $V$ with transition probabilities:
\[
 \Prob\left\{ X(t+1) = j | X(t) = i  \right\} = \frac{w_{ij}}{\deg(i)},
\]
where $\deg(i) = \sum_j w_{ij}$. Let $M$ be the matrix of these probabilities,
\begin{equation}\label{eq:def:M}
 M_{ij} = \frac{w_{ij}}{\deg(i)}.
\end{equation}
It is easy to see that $M_{ij}\geq 0$ and $M\1 = \1$ (indeed, $M$ is a transition probability matrix). Recalling that $D$ is the diagonal matrix with diagonal entries $D_{ii} = \deg(i)$ we have
\[
 M = D^{-1}W.
\]

If we start a random walker at node $i$ ($X(0)=1$) then the probability that, at step $t$, is at node $j$ is given by
\[
 \Prob\left\{ X(t) = j | X(0) = i   \right\} = \left(M^t\right)_{ij}.
\]
In other words, the probability cloud of the random walker at point $t$, given that it started at node $i$ is given by the row vector
\[
 \Prob\left\{ X(t) | X(0) = i   \right\} = e_i^TM^t = M^t[i,:].
\]

\begin{remark}\label{remark:2:DMwithoutPhi}
 A natural representation of the graph would be to associate each vertex to the probability cloud above, meaning
 \[
  i \to M^t[i,:].
 \]
This would place nodes $i_1$ and $i_2$ for which the random walkers starting at $i_1$ and $i_2$ have, after $t$ steps, very similar distribution of locations. However, this would require $d=n$. In what follows we will construct a similar mapping but for considerably smaller $d$.
\end{remark}

$M$ is not symmetric, but a matrix similar to M, $S = D^{\frac12} M D^{-\frac12}$ is, indeed $S = D^{-\frac12}WD^{-\frac12}$. We consider the spectral decomposition of $S$
\begin{equation}\label{eq:SisVlambdaVT}
S = V\Lambda V^T,
\end{equation}
where $V= \left[ v_1,\dots,v_n  \right]$ satisfies $V^TV = I_{n\times n}$ and $\Lambda$ is diagonal with diagonal elements $\Lambda_{kk} = \lambda_k$ (and we organize them as $\lambda_1\geq \lambda_2 \geq \cdots \geq \lambda_n$). Note that $Sv_k = \lambda_k v_k$. Also,
\[
M  = D^{-\frac12} S D^{\frac12} = D^{-\frac12} V\Lambda V^T D^{\frac12} = \left( D^{-\frac12} V\right) \Lambda \left( D^{\frac12}  V \right)^T.
\]
We define $\Phi = D^{-\frac12} V$ with columns $\Phi = \left[ \varphi_1,\dots,\varphi_n\right]$  and $\Psi = D^{\frac12} V$ with columns $\Psi = \left[ \psi_1,\dots,\psi_n\right]$. Then
\begin{equation}\label{eq:MisPhiLambdaPsiT}
M = \Phi \Lambda \Psi^T,
\end{equation}
and $\Phi,\, \Psi$ form a biorthogonal system in the sense that $\Phi^T\Psi = I_{n\times n}$ or, equivalently, $\varphi_j^T\psi_k = \delta_{jk}$. Note that $\varphi_k$ and $\psi_k$ are, respectively right and left eigenvectors of $M$, indeed, for all $1\leq k\leq n$:
\[
M\varphi_k = \lambda_k \varphi_k \quad \text{ and } \psi_k^T M = \lambda_k \psi_k^T.
\]
Also, we can rewrite this decomposition as
\[
M = \sum_{k=1}^n \lambda_k \varphi_k \psi_k^T.
\]
and it is easy to see that
\begin{equation}\label{eq:Masrank1withphiandpsi}
M^t = \sum_{k=1}^n \lambda_k^t \varphi_k \psi_k^T.
\end{equation}

Let us revisit the embedding suggested on Remark~\ref{remark:2:DMwithoutPhi}. It would correspond to
\[
i \to M^t[i,:] = \sum_{k=1}^n \lambda_k^t \varphi_k(i) \psi_k^T,
\]
it is written in terms of the basis $\psi_k$. The diffusion map will essentially consist of the representing a node $i$ by the coefficients of the above map
\begin{equation}\label{eq:2:DM:withcoordinate1}
i \to 
 \left[ \begin{array}{c} \lambda_1^t \varphi_1(i) \\ \lambda_2^t \varphi_2(i) \\ \vdots \\ \lambda_n^t \varphi_n(i) \end{array} \right],
\end{equation}
Note that $M\1 = \1$, meaning that one of the right eigenvectors $\varphi_k$ is simply a multiple of $\1$ and so it does not distinguish the different nodes of the graph. We will show that this indeed corresponds to the the first eigenvalue.

\begin{proposition}
All eigenvalues $\lambda_k$ of $M$ satisfy $\left|\lambda_k\right| \leq 1$.
\end{proposition}

\begin{proof}

Let $\varphi_k$ be a right eigenvector associated with $\lambda_k$ whose largest entry in magnitude is positive $\varphi_k\left(i_{\max}\right)$. Then,
\[
\lambda_k \varphi_k\left(i_{\max}\right) = M \varphi_k\left(i_{\max}\right) =  \sum_{j=1}^n M_{i_{\max}, j} \varphi_k\left(j\right).
\]
This means, by triangular inequality that, that
\[
\left| \lambda_k \right| = \sum_{j=1}^n \left| M_{i_{\max}, j} \right| \frac{\left|\varphi_k\left(j\right)\right|}{\left|\varphi_k\left(i_{\max}\right)\right|} \leq \sum_{j=1}^n \left| M_{i_{\max}, j} \right| = 1.
\]

\end{proof}

\begin{remark}\label{remark:lazyDMaps}
It is possible that there are other eigenvalues with magnitude $1$ but only if $G$ is disconnected or if $G$ is bipartite. Provided that $G$ is a connected graph, a natural way to remove potential periodicity issues (like the graph being bipartite) is to make the walk lazy, i.e. to add a certain probability of the walker to stay in the current node. This can be conveniently achieved by taking, e.g.,
\[
M' = \frac12 M + \frac12 I.
\]
\end{remark}

By the proposition above we can take $\varphi_1 = \1$, meaning that the first coordinate of~\eqref{eq:2:DM:withcoordinate1} does not help differentiate points on the graph. This suggests removing that coordinate:

\begin{definition}[Diffusion Map]
Given a graph $G=(V,E,W)$ construct $M$ and its decomposition $M = \Phi \Lambda \Psi^T$ as described above. The diffusion map is a map $\Dphi_t: V\to \RR^{n-1}$ given by
\[
\Dphi_t\left(v_i\right) =  \left[ \begin{array}{c} \lambda_2^t \varphi_2(i) \\ \lambda_3^t \varphi_3(i) \\ \vdots \\ \lambda_n^t \varphi_n(i) \end{array} \right].
\]
\end{definition}

This map is still a map to $n-1$ dimensions. But note now that each coordinate has a factor of $\lambda_k^t$ which, if $\lambda_k$ is small will be rather small for moderate values of $t$. This motivates truncating the diffusion map by taking only the first $d$ coefficients.

\begin{definition}[Truncated Diffusion Map]
Given a graph $G=(V,E,W)$ and dimension $d$, construct $M$ and its decomposition $M = \Phi \Lambda \Psi^T$ as described above. The diffusion map truncated to $d$ dimensions is a map $\Dphi_t: V\to \RR^{d}$ given by
\[
\Dphi_t^{(d)}\left(v_i\right) =  \left[ \begin{array}{c} \lambda_2^t \varphi_2(i) \\ \lambda_3^t \varphi_3(i) \\ \vdots \\ \lambda_{d+1}^t \varphi_{d+1}(i) \end{array} \right].
\]
\end{definition}

We note that the truncated diffusion map without scaling the coordinates by $\lambda^t$ (or equivalently, taking $t=0$) is known as {\em Laplacian Eigenmap}, introduced by Belkin and Niyogi \cite{belkin2001laplacian,belkin2003laplacian}. 

In the following theorem we show that the Euclidean distance in the diffusion map coordinates (called diffusion distance) meaningfully measures distance between the probability clouds after $t$ iterations.

\begin{theorem}
For any pair of nodes $v_{i_1}$, $v_{i_2}$ we have
\begin{align*}    
& \left\| \Dphi_t\left(v_{i_1}\right) - \Dphi_t\left(v_{i_2}\right) \right\|^2 =  \\
& \qquad \qquad
= \sum_{j=1}^n \frac{1}{\deg(j)}\left[  \Prob\left\{ X(t)=j | X(0) = {i_1}  \right\} - \Prob\left\{ X(t)=j | X(0) = {i_2}  \right\} \right]^2.
\end{align*}
\end{theorem}

\begin{proof}
Note that 
$$\sum_{j=1}^n \frac{1}{\deg(j)}\left[  \Prob\left\{ X(t)=j | X(0) = {i_1}  \right\} - \Prob\left\{ X(t)=j | X(0) = {i_2}  \right\} \right]^2$$ 
can be rewritten as
\begin{align*} 
& \sum_{j=1}^n \frac{1}{\deg(j)}\left[  \sum_{k=1}^n \lambda_k^t \varphi_k(i_1) \psi_k(j) -  \sum_{k=1}^n \lambda_k^t \varphi_k(i_2) \psi_k(j) \right]^2  =  \\ & \qquad \qquad = \sum_{j=1}^n \frac{1}{\deg(j)}\left[  \sum_{k=1}^n \lambda_k^t\left( \varphi_k(i_1) - \varphi_k(i_2) \right) \psi_k(j) \right]^2
\end{align*}
and
\begin{align*}
& \sum_{j=1}^n \frac{1}{\deg(j)}\left[  \sum_{k=1}^n \lambda_k^t\left( \varphi_k(i_1) - \varphi_k(i_2) \right) \psi_k(j) \right]^2  \\    & = \sum_{j=1}^n \left[  \sum_{k=1}^n \lambda_k^t\left( \varphi_k(i_1) - \varphi_k(i_2) \right) \frac{\psi_k(j)}{\sqrt{\deg(j)}} \right]^2 \\
  & = \left\|  \sum_{k=1}^n \lambda_k^t\left( \varphi_k(i_1) - \varphi_k(i_2) \right) D^{-\frac12} \psi_k \right\|^2.
\end{align*}

Note that $D^{-\frac12}\psi_k = v_k$ which forms an orthonormal basis, meaning that
\begin{eqnarray*}
\left\|  \sum_{k=1}^n \lambda_k^t\left( \varphi_k(i_1) - \varphi_k(i_2) \right) D^{-\frac12} \psi_k \right\|^2 & = &\sum_{k=1}^n \left( \lambda_k^t\left( \varphi_k(i_1) - \varphi_k(i_2) \right)    \right)^2 \\
& = &\sum_{k=2}^n \left( \lambda_k^t \varphi_k(i_1) - \lambda_k^t\varphi_k(i_2)   \right)^2,
\end{eqnarray*}
where the last equality follows from the fact that $\varphi_1=\1$. The proof of the theorem is complete.

\end{proof}

\begin{figure}[h]
\begin{center}
\includegraphics[width=0.4\textwidth]{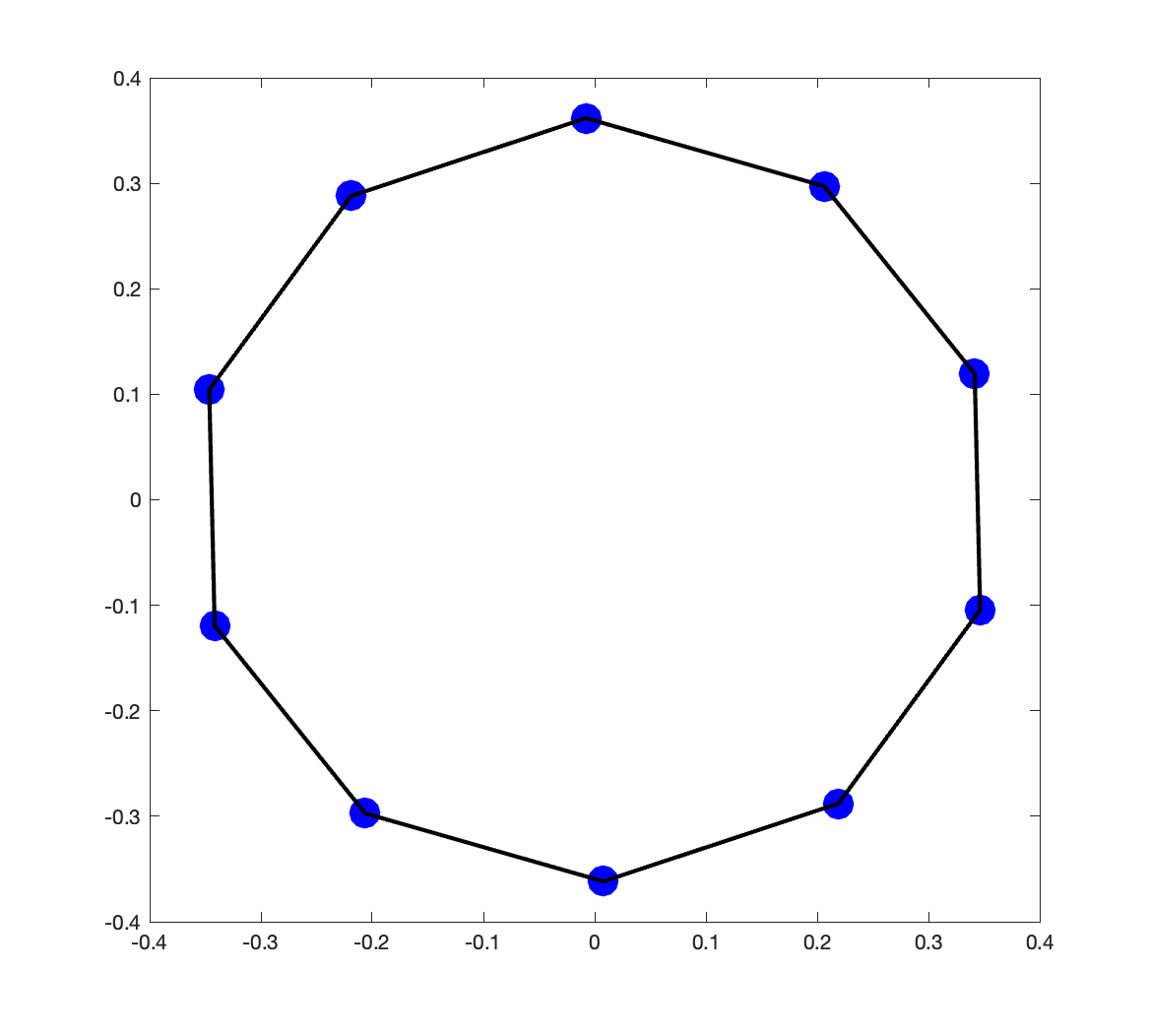}
\caption{The diffusion map, truncated to two dimensions,  of the cycle graph $C_{10}$.}
\label{figure:DM:ring}
\end{center}
\end{figure}

As simple example, let us consider the cycle graph  (or ring graph) $C_n$, which is a graph on $n$ nodes $\{1,\dots,n\}$ such that node $k$ is connected to $k-1$ and $k+1$ and $1$ is connected to $n$. The diffusion map of $C_{10}$ truncated to two dimensions, shown in Figure~\ref{figure:DM:ring}, gives a very natural way of displaying the graph (indeed, if one is asked to draw the ring graph, this is probably the drawing that most people would do). It is not difficult to analytically compute the diffusion map of $C_n$ and confirm that it displays the points in a circle.

\subsection{Diffusion maps of point clouds}\label{subs:diffusionmapskernel}

Very often we are interested in embedding in $\RR^d$ a point cloud of points $x_1,\dots,x_n \in \RR^p$ and not necessarily a graph. One option is to use Principal Component Analysis (PCA), but PCA is only designed to find linear structure of the data and the low dimensionality of the dataset may be non-linear. For example, let us say that our dataset is images of the face of a person taken from different angles and lighting conditions, for example, the dimensionality of this dataset is limited by the amount of muscles in the head and neck and by the degrees of freedom of the lighting conditions 
but it is not clear that this low dimensional structure is linearly apparent on the pixel values of the images.

\begin{figure}[h]
\begin{center}
\includegraphics[width=0.4\textwidth]{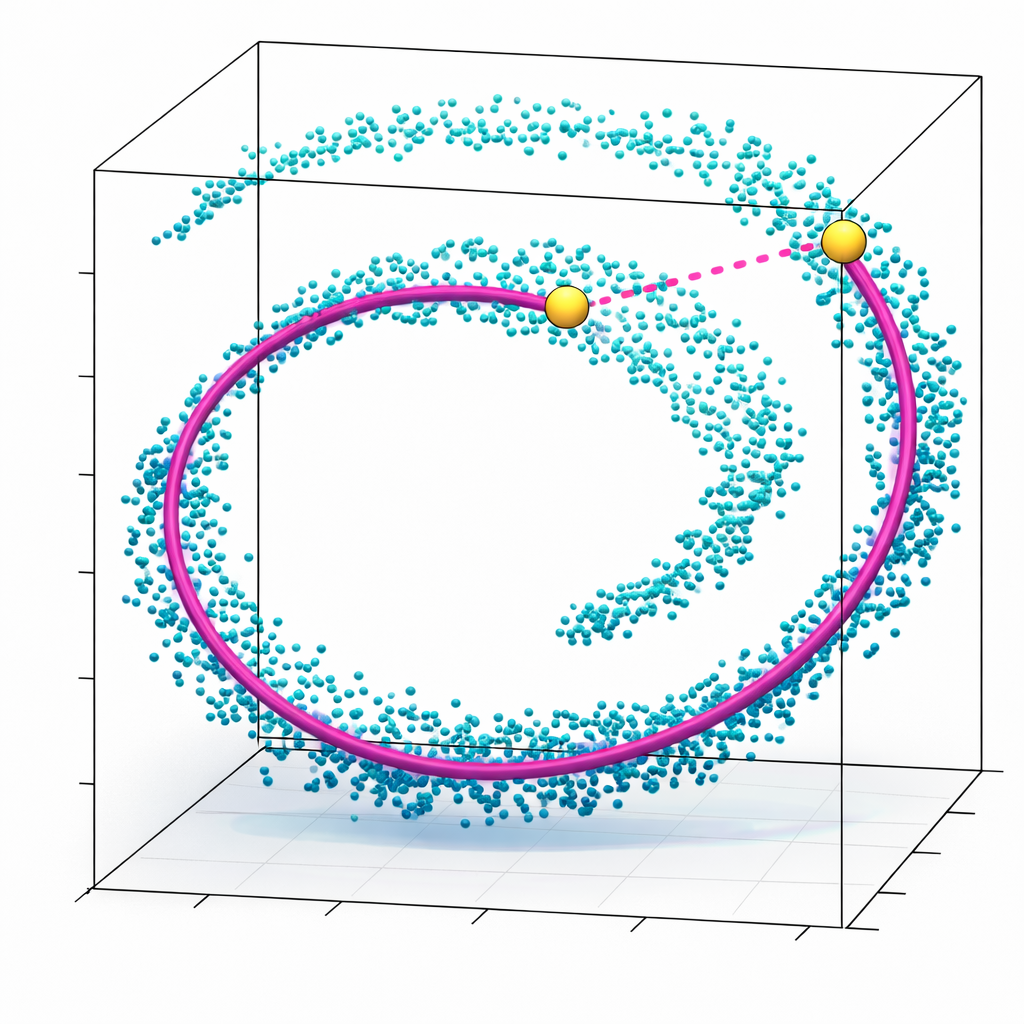}
\caption{A swiss roll point cloud (see, for example,~\cite{ISOMAP_paper_2000}). The points are sampled from a two dimensional manifold curved in $\RR^3$ and then a graph is constructed where nodes correspond to points.}
\label{figure:DM:SwissRoll}
\end{center}
\end{figure}

Let us consider a point cloud that is sampled from a two dimensional swiss roll embedded in three dimension (see Figure~\ref{figure:DM:SwissRoll}). In order to learn the two dimensional structure of this object we need to differentiate points that are near each other because they are close by in the manifold and not simply because the manifold is curved and the points appear nearby even when they really are distant in the manifold (see Figure~\ref{figure:DM:SwissRoll} for an example). We will achieve this by creating a graph from the data points.

Our goal is for the graph to capture the structure of the manifold. To each data point we will associate a node. For this we should only connect points that are close in the manifold and not points that maybe appear close in Euclidean space simply because of the curvature of the manifold. This is achieved by picking a small scale and linking nodes if they correspond to points whose distance is smaller than that scale. This is usually done smoothly via a kernel\footnote{A detailed discussion of kernel based methods will be given in Chapter~\ref{s:kernellearning}. For now it suffices to think of a kernel as a function.} $K_\eps$, and to each edge $(i,j)$ associating a weight
\[
w_{ij} = K_\eps\left( \|x_i - x_j\|_2 \right),
\]
a common example of a kernel is $K_\eps(u) = \exp\left( -\frac1{2\eps} u^2 \right)$, that gives essentially zero weight to edges corresponding to pairs of nodes for which $\|x_i-x_j\|_2 \gg \sqrt{\eps}$. We can then take the the diffusion maps of the resulting graph.

\subsection{An illustrative simple example and noisy data}

A simple and illustrative example is to take images of a blob on a background in different positions (imagine a white square on a black background and each data point corresponds to the same white square in different positions). This dataset is clearly intrinsically two dimensional, as each image can be described by the (two-dimensional) position of the square. However, we do not expect this two-dimensional structure to be directly apparent from the vectors of pixel values of each image; in particular we do not expect vectors of images with this type of structure to lie in a two dimensional affine subspace!\footnote{It is worth noting that in a very stylized example where the only possibilities are the blobs not intersecting, or intersecting by a fixed amount, it is possible that, since this effect is captured by just three values, it can still be captured linearly, and that in such simple examples PCA could still capture such low-dimensional structure.}

\begin{figure}[h]
\begin{center}
\includegraphics[width=0.4\textwidth]{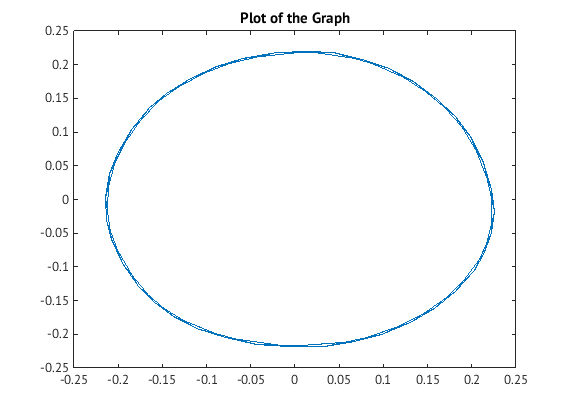}
\caption{The two-dimensional diffusion map of the dataset  where each data point is an image with the same vertical strip in different positions in the x-axis, the circular structure is apparent.}
\label{figure:DM:blob1d}
\end{center}
\end{figure}

Let us start by experimenting with the above example for one dimension. In that case the blob is a vertical stripe and simply moves left and right. We think of our space as the one in many arcade games, if the square or stripe moves to the right all the way to the end of the screen, it shows up on the left side (and same for up-down in the two-dimensional case). Not only should this point cloud have a one dimensional structure but it should also exhibit a circular structure. Remarkably, this structure is completely apparent when taking the two-dimensional diffusion map of this dataset, see Figure~\ref{figure:DM:blob1d}.

\begin{figure}[h]
\begin{center}
\includegraphics[width=45mm,height=45mm]{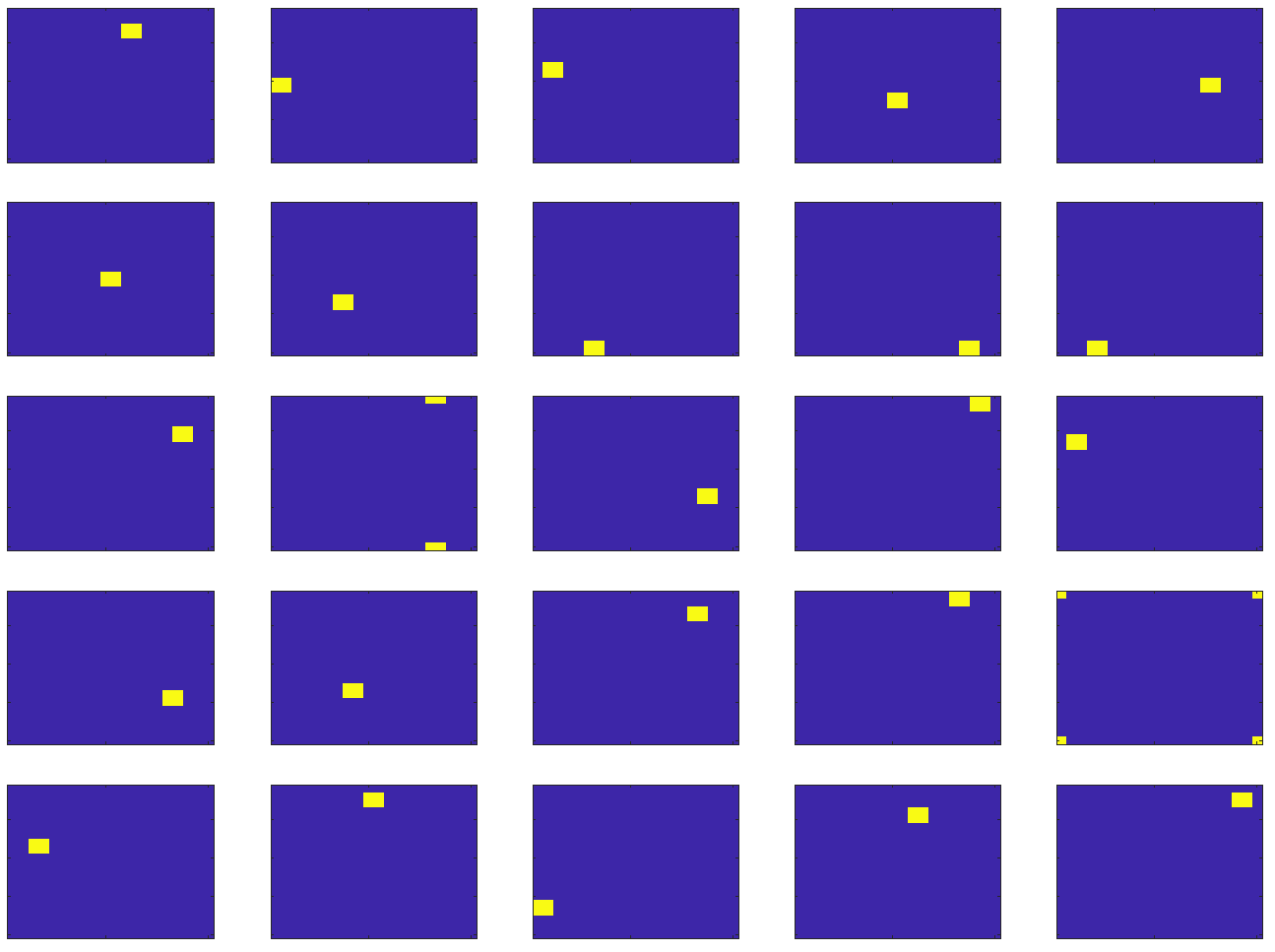}\qquad
\includegraphics[height=45mm]{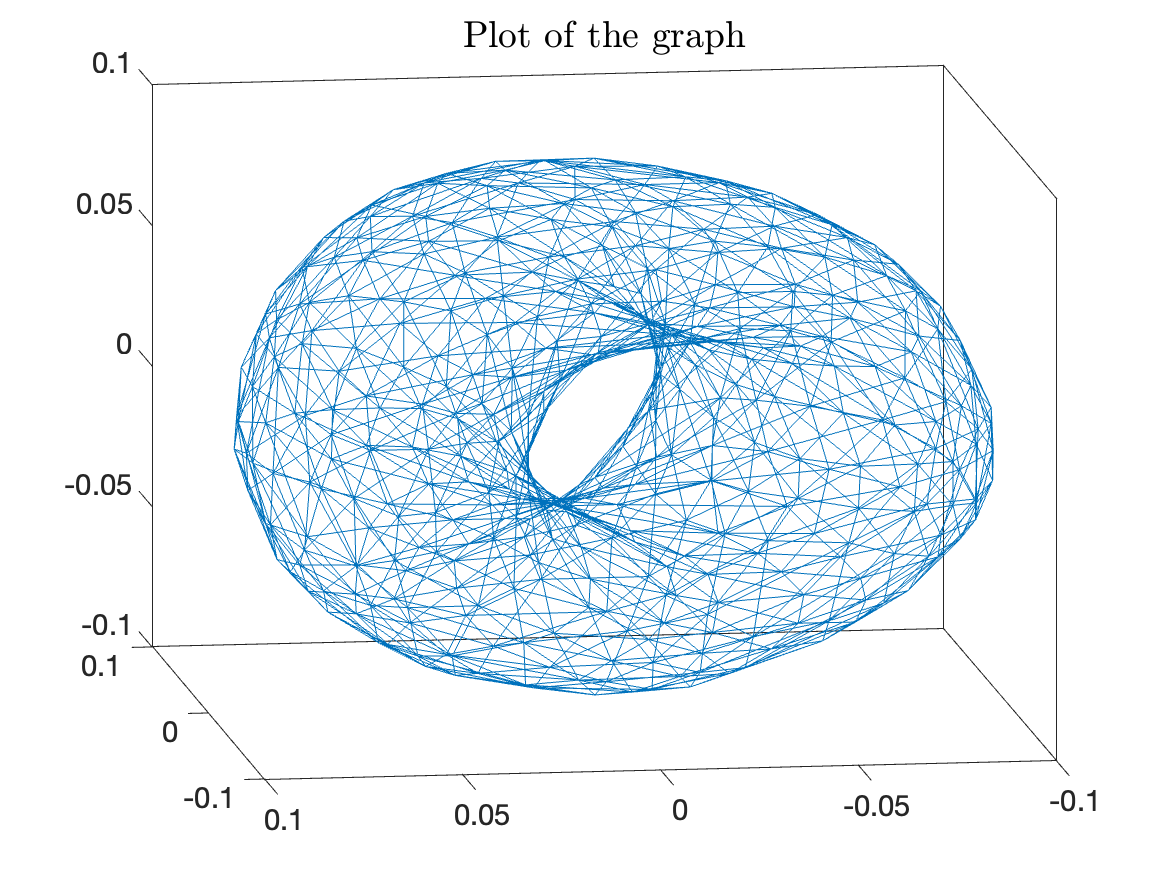}
\caption{On the left a subset of the data set considered and on the right its three dimensional diffusion map. The fact that the manifold is a torus is remarkably captured by the embedding.}
\label{figure:DM:blob2d}
\end{center}
\end{figure}

For the two dimensional example, we expect the structure of the underlying manifold to be a two-dimensional torus. Indeed, Figure~\ref{figure:DM:blob2d} shows that the three-dimensional diffusion map captures the toroidal structure of the data.

In many applications, the data are noisy and the Euclidean distances $\|x_i - x_j\|$ would be affected by the measurement noise. It is therefore common to apply a denoising algorithm to the data prior to the application of diffusion maps. A popular choice is to first denoise and reduce dimensionality using PCA, followed by diffusion maps in the reduced space of principal components. The dimension of PCA is chosen large enough to retain the non-linear structure of the data and the signal variability, but small enough to effectively diminish the noise. See~\cite{singer2013two,gilles2025cryo} for the application of PCA followed by diffusion maps in biomedical imaging. Fitting a manifold in the presence of noise is a topic of ongoing research~\cite{aizenbud2025estimation,fefferman2025fitting}. 

\subsection{Diffusion maps, images, and data manifolds}

To illustrate the ability of the diffusion map to recover physically meaningful
latent structure from image data, we experimentally study its performance on two standard image datasets, MNIST and the Yale Face Database.

In both experiments, the affinity between two images 
$x_i, x_j$ is defined by the Gaussian kernel
$$
W_{ij} = \exp\!\left(-\frac{\|x_i - x_j\|^2}{2\varepsilon}\right),
$$
where $\varepsilon > 0$ is the bandwidth parameter. Here we choose $\varepsilon$ proportional to the median of the pairwise squared distances: $\varepsilon = \frac{1}{2}\mathrm{median}\bigl\{\|x_i - x_j\|^2 : i \neq j\bigr\}$.
The median heuristic (as well as the mean heuristic) is a standard data-adaptive choice that scales $\varepsilon$ to the typical squared distance between points in the dataset, avoiding the need to tune the bandwidth by hand.

\subsubsection{Example: Diffusion maps on MNIST handwritten digits}

To exhibit the geometry recovered by the diffusion map, we apply it to a subset of the MNIST database. Each image is represented as a vector in $\mathbb{R}^{784}$ by flattening its pixel intensities, so the data matrix $X \in \mathbb{R}^{784 \times 600}$
has images as columns and pixel intensities as rows.

\begin{figure}[h]
\begin{center}
\includegraphics[width=0.66\textwidth]{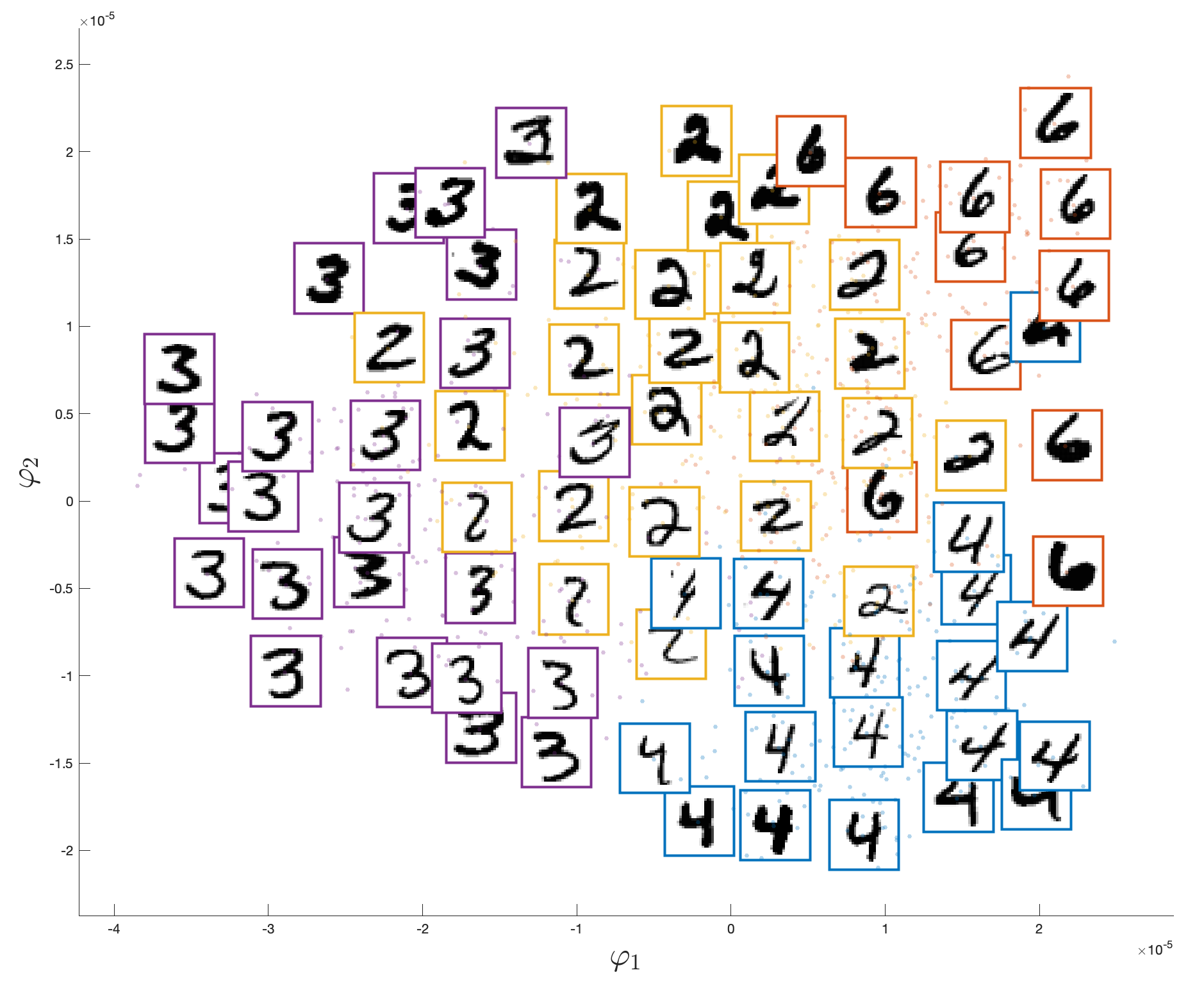}
\caption{Diffusion map embedding of
600 MNIST images (150 each of
digits 2,3,4 6) in the $(\varphi_1, \varphi_2)$
plane. A subset of the images is
displayed as a thumbnail of the corresponding digit image.}
\label{fig:mnist_diffusion}
\end{center}
\end{figure}

The resulting embeddings are shown in Figure~\ref{fig:mnist_diffusion} and respectively.
Figure~\ref{fig:mnist_diffusion} depicts the diffusion map 
 embedding obtained from the first two diffusion coordinates $\varphi_1, \varphi_2$ for all 
600 images, with a subset of the images displayed as a small thumbnail at its embedding location and colored borders indicating digit class. 

The four digit classes separate quite well in the $(\varphi_1, \varphi_2)$ plane without any use of class labels during training: the diffusion map discovers the class structure from geometry alone. This gives support to the assumption that images of the same digit class lie on a lower-dimensional manifold within $\mathbb{R}^{784}$, corresponding to the natural variability in handwriting style, stroke width, and slant within that class. Within each digit class the images are organized continuously: adjacent points in the embedding correspond to images that look similar, and one can trace smooth trajectories through each cluster along which the appearance changes gradually.

\begin{figure}[h]
\begin{center}
\includegraphics[width=0.66\textwidth]{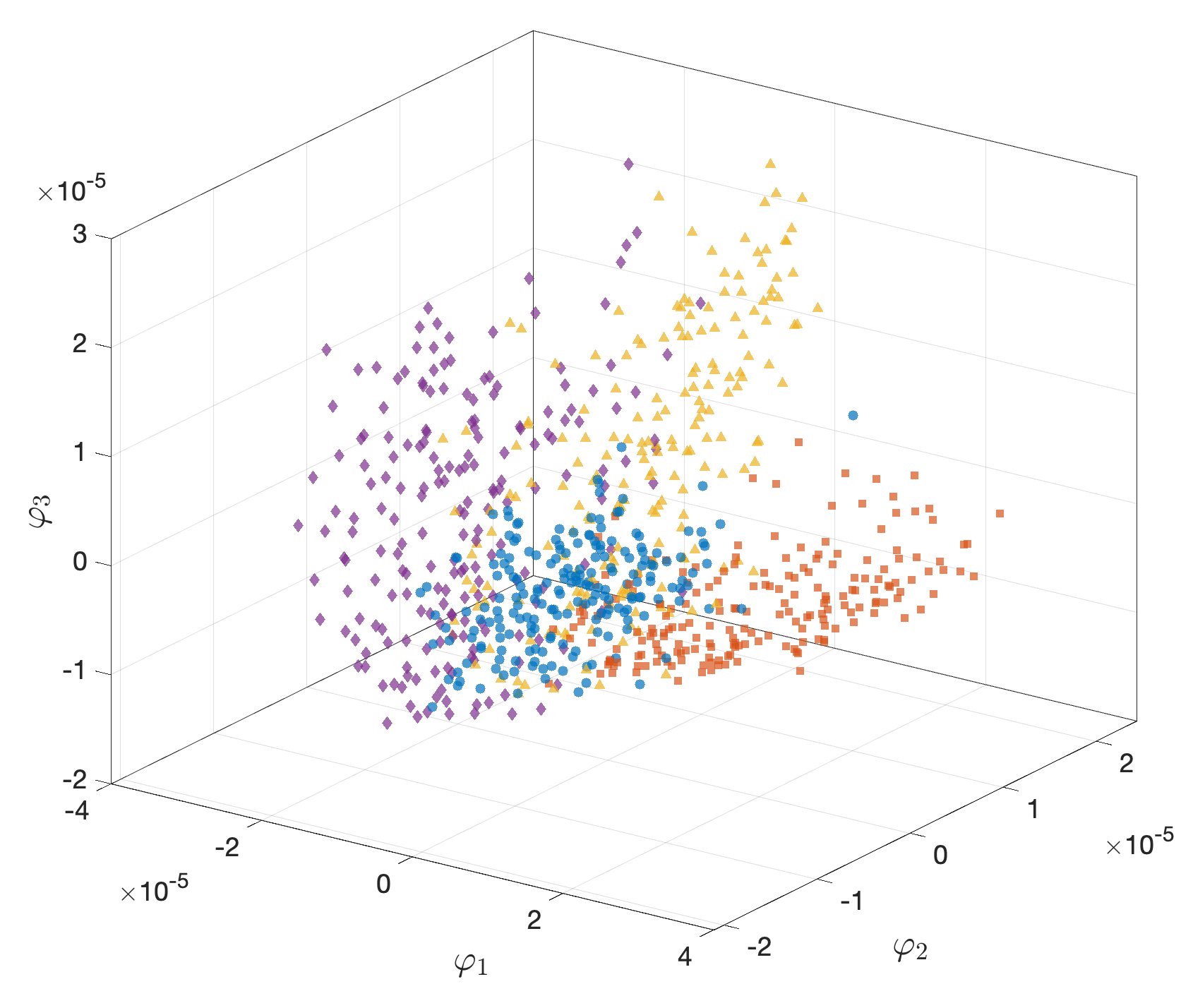}
\caption{Diffusion map embedding of
600 MNIST images (150 each of
digits 2,3,4 6) in the $(\varphi_1, \varphi_2,\varphi_3)$
space (2 = yellow, 3 = purple, 4 = blue, 6 = red).}
\label{fig:mnist_diffusion3d}
\end{center}
\end{figure}

Figure~\ref{fig:mnist_diffusion3d} shows the  the diffusion map embedding corresponding to the first three diffusion coordinates $\varphi_1, \varphi_2, \varphi_3$, which reveals
a good (but certainly not perfect) unsupervised clustering of the data points according to their digit labels (here, 2 = yellow, 3 = purple, 4 = blue, 6 = red).
It is important to emphasize that two and three diffusion coordinates were chosen for visualization purposes, but depending on the task at hand more diffusion dimensions may be needed to accomplish the desired goal. See also the discussion in Chapter~\ref{s:tsne} about using diffusion maps as a preprocessing step.

\subsubsection{Example: Diffusion maps on the Extended Yale B Face Database}

In this experiment we apply the diffusion map to the
Extended Yale~B database \cite{georghiades2001few}, a controlled
face image dataset consisting of $38$ subjects photographed under
$64$ distinct illumination conditions. Each image has been
cropped and normalized to $192 \times 168$ pixels. For this
experiment we select a single subject and work with all $64$ of
their frontal-pose images, representing each as a vector in
$\mathbb{R}^{896}$ obtained by resizing to $32 \times 28$ pixels
and flattening. 

The illumination conditions are parameterized by two angles: the
azimuth $\phi \in [-130^\circ, +130^\circ]$ and the elevation
$\theta \in [-40^\circ, +90^\circ]$ of the light source relative
to the camera axis. As the light source moves around the face, the
image intensity pattern changes dramatically: under extreme lateral
illumination large portions of the face fall into shadow, while
under frontal illumination the face appears nearly uniform. Despite
this large photometric variation, all $64$ images depict the same
individual with the same geometry, so they lie on a smooth
two-dimensional manifold in $\mathbb{R}^{896}$ parameterized by
$(\phi, \theta)$.

\begin{figure}[h]
\begin{center}
\includegraphics[width=0.9\textwidth]{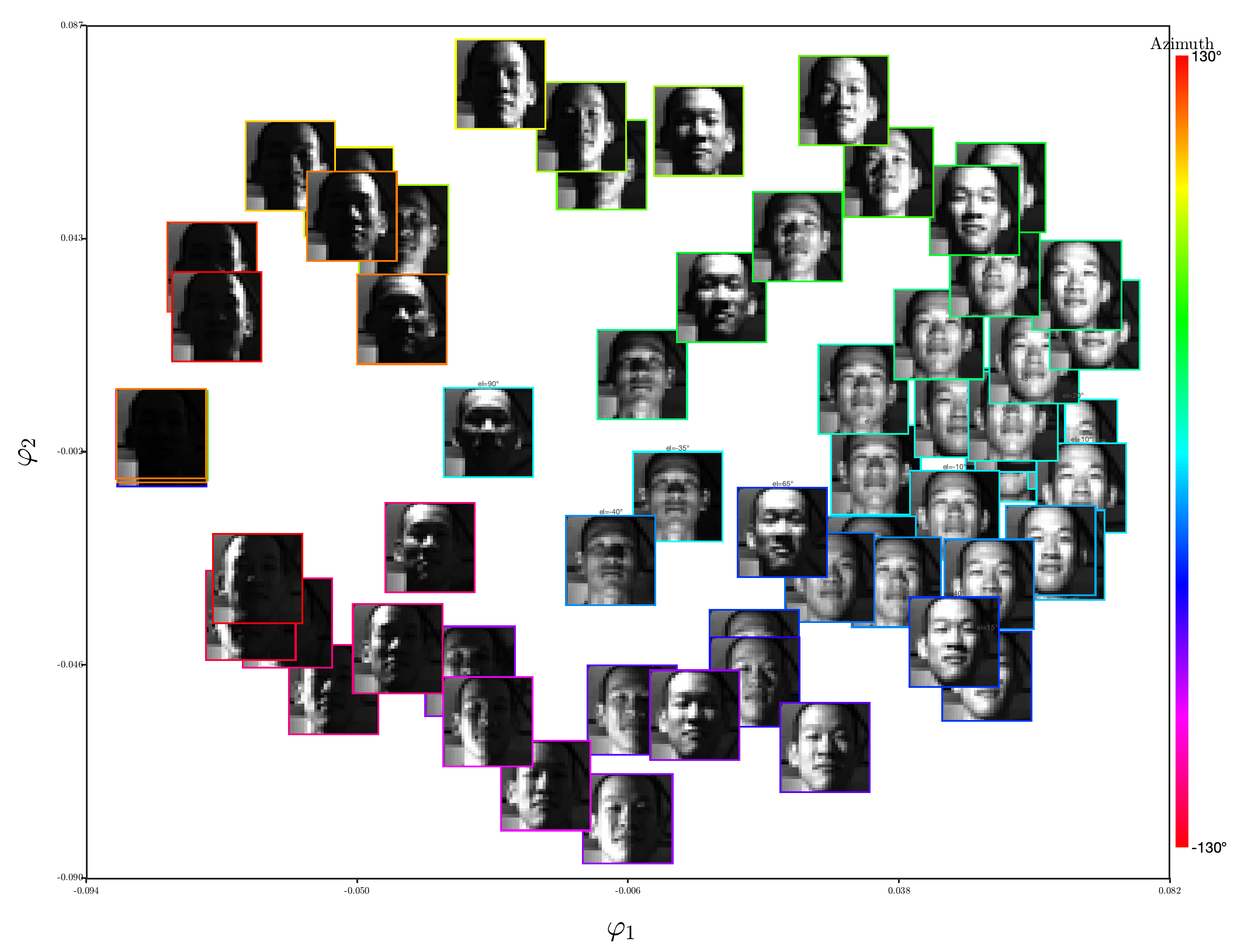}
\caption{Diffusion map embedding of 64 frontal-pose images of a
single subject from the Extended Yale B database, varying only in
illumination condition. Each point is displayed as a thumbnail of
the corresponding face image, with border color indicating the
azimuth of the light source (color scale shown at right). The
embedding appears to organize the images by illumination geometry: $\varphi_1$
correlates with azimuth (left--right position of the light source)
and $\varphi_2$ with elevation (above--below position). Images with
similar shadow patterns are nearby in the embedding; images with
qualitatively different lighting are far apart.}
\label{fig:faces_diffusion}
\end{center}
\end{figure}

The embedding provided by the diffusion map reveals the intrinsic two-dimensional structure of
the illumination manifold, see Figure~\ref{fig:faces_diffusion}. The first diffusion coordinate $\varphi_1$
correlates strongly with the azimuth of the light source: moving
along $\varphi_1$ corresponds to sweeping the light from left to right
across the face, with the shadow pattern shifting accordingly.
The second diffusion coordinate $\varphi_2$ captures the elevation of
the light source: moving along $\varphi_2$ corresponds to raising or
lowering the light from below to above the face.

\section{Connections between diffusion maps and spectral clustering}

Diffusion maps are tightly connected to spectral clustering (described in Chapter~\ref{c:graphs}). In fact, spectral clustering can be understood as simply performing $k$-means on the embedding given by diffusion maps truncated to $k-1$ dimensions.

A natural way to try to overcome the issues of $k$-means depicted in Figure~\ref{figure:3:cluster_kmeans_circles} is by using diffusion maps: Given the data points we construct a weighted graph $G = (V,E,W)$ using a kernel $K_{\epsilon}$, such as $K_{\epsilon}(u) = \exp\left(-\frac1{2\epsilon}u^2\right)$, by associating each point to a vertex and, for which pair of nodes, set the edge weight as
\[
w_{ij} = K_{\epsilon}\left( \|x_i-x_j\| \right).
\]

Recall the construction of a matrix $M = D^{-1}W$ as the transition matrix of a random walk
\[
\Prob\left\{ X(t+1) = j | X(t) = i \right\} = \frac{w_{ij}}{\deg(i)} = M_{ij},
\]
where $D$ is the diagonal with $D_{ii} = \deg(i)$. The $d$-dimensional diffusion map is given by
\[
\Dphi_t^{(d)}(i) = \left[  \begin{array}{c}  \lambda_2^t \varphi_2(i) \\ \vdots \\  \lambda_{d+1}^t \varphi_{d+1}(i) \end{array} \right],
\]
where $M  = \Phi \Lambda \Psi^T$ where $\Lambda$ is the diagonal matrix with the eigenvalues of $M$ and $\Phi$ and $\Psi$ are, respectively, the right and left eigenvectors of $M$ (note that they form a bi-orthogonal system, $\Phi^T\Psi = I$).

If we want to cluster the vertices of the graph in $k$ clusters, then it is natural to truncate the diffusion map to have $k-1$ dimensions (since in $k-1$ dimensions we can have $k$ linearly separable sets). If indeed the clusters were linearly separable after embedding then one could attempt to use $k$-means on the embedding to find the clusters, this is precisely the motivation for spectral clustering.

\begin{algorithm}
Spectral Clustering: Given a graph $G=(V,E,W)$ and a number of clusters $k$ (and $t$), spectral clustering consists in taking a $(k-1)$ dimensional diffusion map
\[
\Dphi_{t}^{(k-1)}(i) =  \left[  \begin{array}{c}  \lambda_2^t \varphi_2(i) \\ \vdots \\  \lambda_{k}^t \varphi_{k}(i) \end{array} \right]
\]
and clustering the points $\Dphi_{t}^{(k-1)}(1),\Dphi_{t}^{(k-1)}(2),\dots,\Dphi_{t}^{(k-1)}(n)\in\RR^{k-1}$ using, for example, $k$-means clustering. Usually, the scaling of $\lambda_m^t$ is ignored (corresponding to $t=0$).
\caption{Spectral clustering described using diffusion maps.}
\label{DMAPS:algorithm:3:spectralclustering}
\end{algorithm}

In order to show that this indeed coincides with Algorithm~\ref{algorithm:3:spectralclustering}, it is enough to show that $\varphi_m = D^{-\frac{1}{2}}v_m$ where $v_m$ is the eigenvector associated with the $m$-th smallest eigenvalue of $\LLL_G$. This follows from the fact that $S=D^{-\frac12}WD^{-\frac12}$ as defined in~\eqref{eq:SisVlambdaVT} is related to $\LLL_G$ by
\[
\LLL_G = I - S,
\]
and $\Phi = D^{-1/2}V$.

\begin{figure}[h]
\begin{center}
\includegraphics[width=0.7\textwidth]{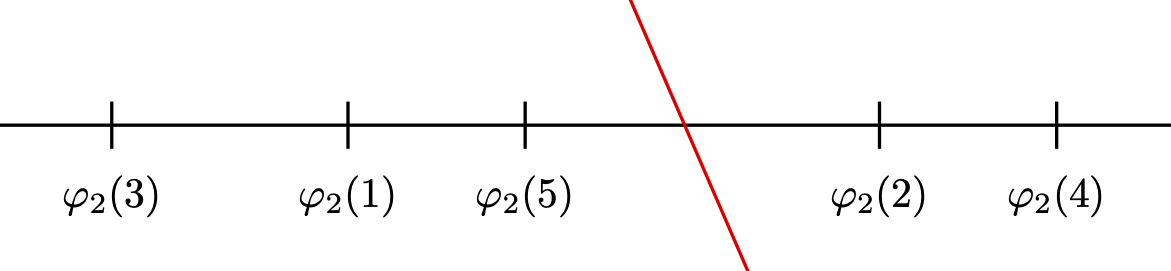}
\caption{For two clusters, spectral clustering consists in assigning to each vertex $i$ a real number $\varphi_2(i)$, then setting a threshold $\tau$ and taking $S = \{ i\in V:\, \varphi_2(i)\leq \tau \}$. This real number can both be interpreted through the spectrum of $\LLL_G$ as in Algorithm~\ref{algorithm:3:spectralclustering2clusters} or as the diffusion map embedding as in Algorithm~\ref{DMAPS:algorithm:3:spectralclustering}.}
\label{figure:3:cluster_2means_threshold}
\end{center}
\end{figure}

Proposition~\ref{prop:3:NcutandRandomWalks} below establishes a connection between $\Ncut$ (as described in Chapter~\ref{c:graphs}) and the random walks introduced above. Let $M$ as defined in~\eqref{eq:def:M} denote the matrix of transition probabilities. 
Recall that $M\1 = \1$, corresponding to $M\varphi_1 = \varphi_1$, which means that $\psi_1^TM =\psi_1^T$, where 
\[
\psi_1 = D^{\frac12}v_1 = D\varphi_1 = \left[\deg(i) \right]_{1\leq i\leq n}.
\]
This means that $ \left[\frac{ \deg(i)} { \vol(G)} \right]_{1\leq i\leq n}$ is the stationary distribution of this random walk. Indeed it is easy to check that, if $X(t)$ has a certain distribution $p_t$ then $X(t+1)$ has a distribution $p_{t+1}$ given by $p_{t+1}^T = p_{t}^TM$

\begin{proposition}\label{prop:3:NcutandRandomWalks}
Given a graph $G=(V,E,W)$ and a partition $(S,S^c)$ of $V$, $\Ncut(S)$ corresponds to the probability, in the random walk associated with $G$, that a random walker in the stationary distribution goes to $S^c$ conditioned on being in $S$ plus the probability of going to $S$ condition on being in $S^c$, more explicitly:
\[
\Ncut(S) = \Prob\left\{ X(t+1) \in S^c | X(t) \in S  \right\} + \Prob\left\{ X(t+1) \in S | X(t) \in S^c  \right\},
\]
where $\Prob\{X(t) = i\} =\frac{  \deg(i)} { \vol(G)} $.
\end{proposition}

\begin{proof}
Without loss of generality we can take $t=0$. Also, the second term in the sum corresponds to the first with $S$ replaced by $S^c$ and vice-versa, so we'll focus on the first one. We have:

\begin{eqnarray*}
\Prob\left\{ X(1) \in S^c | X(0) \in S  \right\} & = & \frac{\Prob\left\{ X(1) \in S^c \cap X(0) \in S  \right\}}{\Prob\left\{ X(0) \in S  \right\}} \\
& = & \frac{\sum_{i\in S}\sum_{j\in S^c} \Prob\left\{ X(1) \in j \cap X(0) \in i  \right\}}{\sum_{i\in S} \Prob\left\{ X(0) = i  \right\}} \\
& = & \frac{\sum_{i\in S}\sum_{j\in S^c} \frac{\deg(i)}{\vol(G)} \frac{w_{ij}}{\deg(i)}}{\sum_{i\in S} \frac{\deg(i)}{\vol(G)}} \\
& = & \frac{\sum_{i\in S}\sum_{j\in S^c} w_{ij}}{ \sum_{i\in S} \deg(i)}  =  \frac{\cut(S) }{ \vol(S)}.
\end{eqnarray*}

Analogously,
\[
\Prob\left\{ X(t+1) \in S | X(t) \in S^c  \right\} = \frac{\cut(S) }{ \vol(S^c)},
\]
which concludes the proof.
\end{proof}

\section{Commute time distance}\label{s:commute}

We saw that the diffusion distance is a way to equip the graph with a metric that has a nice probabilistic interpretation.  
A natural question is if there are other meaningful and useful metrics for graphs. Perhaps the first metric that comes to mind is the {\em shortest path metric}, or geodesic distance. A cost $c_{ij}$ is associated with each edge, e.g., $c_{ij} = w_{ij}^{-1}$. The cost of a path is the sum of costs along the path edges. The geodesic distance between $i$ and $j$ is the smallest possible cost over all possible paths connecting $i$ and $j$. It can be verified that the geodesic distance is indeed a metric (i.e., it satisfies all properties required from a metric), but it is not necessarily an $\ell^2$ metric. Not every metric can be embedded in a Euclidean (or in general, Hilbert) space. For example, the $\ell^1$ distance (Manhattan distance) cannot be embedded in a Euclidean space (it does not satisfy the parallelogram identity). Also, the geodesic distance has other shortcomings that will be discussed later in the course, such as its sensitivity to noise. We remark that the geodesic distance is the basis for a remarkable non-linear dimensionality reduction method called Isomap \cite{ISOMAP_paper_2000}, which will be revisited later on.

Are there any other interesting/natural graph metrics? Given the random walk process over the graph, it is natural to consider the average time it takes to get from node $i$ to node $j$. It can be easily seen to satisfy the triangle inequality. However, it is not symmetric: the average time to get from $i$ to $j$ may not equal the time from $j$ to $i$. So, we can symmetrize it. Clearly, the time to get from a node to itself is 0, so overall we get a metric. It is called the {\em commute time distance}. Surprisingly, a kind of miracle happens: The commute time distance turns out to be a Euclidean distance corresponding to an embedding that is also given in terms of eigenvectors and eigenvalues of the graph Laplacian. Our goal is to prove this by starting from the probabilistic interpretation (see also \cite{doyle1984random,lovasz1993random}).

Consider an irreducible random walk  $\{X(t)\}_{t=0}^\infty$ with jump probabilities $M_{ij} = \Prob\left\{ X(t+1) = j | X(t) = i   \right\} = w_{ij}/\deg(i)$ where $w_{ij}$ are the weights of a undirected graph; irreducibility is then equivalent to the graph being connected. We define the {\em first passage time} and {\em mean first passage time} as follows:

\paragraph*{First Passage Time}

\begin{equation}
\tau_{ij}=\left\{\begin{array}{cc}
                   \inf\left\{ t>0 \,|\, X(t)={x}_{j}, \; X(0)={x}_{i}\right\} & \mbox{for }
i\neq j, \\
&\\
                   0 & \mbox{for } i=j.
                 \end{array}
\right.
\end{equation}

\paragraph*{Mean First Passage Time}
\begin{equation}
t_{ij}=\mathbb{E}[\tau_{ij}].
\end{equation}
The {\em expected commute time} between ${x}_{i}$ and ${x}_{j}$
is defined as
\begin{equation}
t_{ij}+t_{ji}.
\end{equation}
Next, we show that the expected commute
time is a metric. Let
\begin{equation}
\tau'_{ij}=\inf\left\{ t>0 \,|\, X(t)={x}_{j},\,X(0)={x}_{i}\right\}.
\end{equation}
Notice that $\tau_{ii} \neq \tau'_{ii}$ because $\tau'_{ii} > 0.$ Let
\begin{equation}
t'_{ij}=\mathbb{E}[\tau'_{ij}].
\end{equation}
The random walker can get from $x_i$ to $x_j$ either directly in one time step with probability $M_{ij}=w_{ij}/\deg(i)$, or it first goes from $x_i$ to $x_k$ ($k\neq j$) with probability $M_{ik}$ and the expected time to get to $x_j$ is $1+t'_{kj}$. Let $T$ and $T'$ denote the matrices with entries respectively  $t_{ij}$ and $t'_{ij}$. We have,\footnote{This step needs irreducibility as otherwise some of the terms would be infinite and the cancellation in the last equality wouldn't necessarily hold}
\begin{eqnarray}
T'_{ij}&=&M_{ij}\cdot1+\sum_{k\neq j}M_{ik}(1+T'_{kj}) \nonumber \\
&=& \sum_{k}M_{ik} + \sum_{k\neq j} M_{ik}T'_{kj} \nonumber \\
&=& 1 + \sum_{k} M_{ik}T'_{kj} - M_{ij}T'_{jj}
\end{eqnarray}
that is,
\begin{equation}
T'=\1\1^T+M(T'-\ddiag(T')),\label{eq:cmt1}
\end{equation}
 where $\1\1^T$ denotes the all-ones matrix and $\ddiag(T')$ is the diagonal matrix obtained by setting to zero all non-diagonal entries of $T'$.

Let us try to determine $\ddiag(T')$ first. To that end,
let ${\psi}_{1}^T=[\deg(i)]_{1\leq i\leq n}$ be
be the left eigenvector of $M$
\begin{equation}
{\psi}_{1}^{T}M={\psi}_{1}^{T}.
\end{equation}
Multiply (\ref{eq:cmt1}) by ${\psi}_{1}^{T}$ on both sides to obtain
\begin{eqnarray}
{\psi}_{1}^{T}T' & = & {\psi}_{1}^{T}\1\1^T+{\psi}_{1}^{T}M(T'-\ddiag(T'))\\
 & = & {\psi}_{1}^{T}\1\1^T+{\psi}_{1}^{T}(T'-\ddiag(T')).\nonumber
\end{eqnarray}
Thus, we have
\begin{equation}
{\psi}_{1}^{T}\ddiag(T')={\psi}_{1}^{T}\1\1^T={\psi}_{1}^{T}\1\1^T=\left(\sum_{k=1}^{n}\deg(k)\right)\1^T,
\label{eq:cmt2}
\end{equation}
or equivalently,
\begin{equation}
T'_{ii}=\frac{1}{\deg(i)}\sum_{k=1}^{n}\deg(k) = \frac{1}{\deg(i)} \vol(G),
\end{equation}
recall that te {\em volume} of $G$ is the sum of the node degrees. 
$T'_{ii}$ is the mean return (or recurrence) time from node $i$ to itself. We see that it is proportional to $\vol(G)$: The larger the graph, the more time the random walker wanders (on average) around before returning back. $T'_{ii}$ is inversely proportional to $\deg(i)$: The larger $\deg(i)$, the more possibilities the random walker has to return to node $i$. Remarkably, $T'_{ii}$ does not see much about the structure of the graph, besides the total edge weights and the specific node degree.

Now we will show uniqueness of the solution $T'$ to (\ref{eq:cmt1}). Indeed,
assume that $T_2'$ is another solution such that
\begin{equation}
T_2'=\1\1^T+M(T_2'-\ddiag(T')),
\end{equation}
then
\begin{equation}
T'-T_2'=M(T'-T_2').
\end{equation}
We see that the columns on $T'-T_2'$ are right eigenvectors of $M$ with eigenvalue 1. But if $M$ is irreducible, the only eigenvector with eigenvalue is 1 is the all-ones vector. Therefore,
\begin{equation}
T'-T_2'={\textbf 1}{u}^{T}
\end{equation}
 for some $u$; the fact that $T'$ and $T'_2$ need to agree on the diagonal means that $u$ must be zero and we must have $T'_2=T'$.

Now that we established that $T'$ is unique let us solve for it. Notice that that the mean first passage time matrix $$T = [t_{ij}]_{1\leq i,j \leq n}$$ can be written as
\begin{equation}
T = T' - \ddiag(T').
\end{equation}
Therefore, (\ref{eq:cmt1}) is equivalent to $T+\ddiag(T')=\1\1^T+MT$ which we can rewrite as
\begin{equation}
(I-M)T=\1\1^T-\ddiag(T').\label{eq:cmt3}
\end{equation}
Unfortunately, $(I-M)$ is not invertible, so we cannot obtain $T$ by simply inverting $(I-M)$.

Recall that the graph Laplacian is defined
as
\begin{equation}
L=D-W=D(I-M).\label{eq:graph_laplacian}
\end{equation}
Multiplying (\ref{eq:cmt3}) by $D$ gives
\begin{equation}
LT=D\1\1^T-D\ddiag(T').\label{eq:cmt4}
\end{equation}
We also recall that $L\1=0$ and $L\succeq 0$.

The eigen-decomposition of $L$ can be written as
\begin{equation}
L=\sum_{l=1}^{n}\mu_{l}{\phiLap}_{l}{\phiLap}_{l}^{T} = \sum_{l=2}^{n}\mu_{l}{\phiLap}_{l}{\phiLap}_{l}^{T}
\end{equation}
with $0=\mu_{1}\leq\mu_{2}\leq\cdots\leq\mu_{n}$, and $\phiLap_1 = \frac{1}{\sqrt{n}}{\textbf 1}$. Note that since $G$ is connected then $\mu_{2}>0$. 
The pseudo-inverse of the graph Laplacian is given by
\begin{equation}
L^\dagger=\sum_{l=2}^{n}\frac{1}{\mu_{l}}{\phiLap}_{l}{\phiLap}_{l}^{T}.
\end{equation}
Then,  $L^\dagger$ has the  properties listed in~\eqref{eq:pinvproperties}. Moreover,
\begin{equation}
L^\dagger L=\sum_{l=2}^{n}{\phiLap}_{l}{\phiLap}_{l}^{T}=I-{\phiLap}_{1}{\phiLap}_{1}^{T} = I - \frac{1}{n}{\textbf 1}{\textbf 1}^T.\label{eq:cmt5}
\end{equation}

Multiplying (\ref{eq:cmt4}) by $L^\dagger$ gives
\begin{equation}
\left(L^\dagger L\right)T=L^\dagger D\1\1^T -L^\dagger DM_{d}.\label{eq:cmt6}
\end{equation}
Combining (\ref{eq:cmt5}) and (\ref{eq:cmt6}) for each column of $T$, we have
\begin{equation}
T=L^\dagger D\1\1^T-L^\dagger D\ddiag(T')+{\textbf 1}{ u}^{T}
\end{equation}
for some vector $u$, that is,
\begin{equation}
t_{ij}=\sum_{k=1}^{n}L_{ik}^\dagger \deg(k)-\vol(G)L_{ij}^\dagger +u_{j}\label{eq:cmt7}
\end{equation}
(here we used that $D\,\ddiag(T')= \vol(G) I$).
From $t_{ii}=0$ we find
\begin{equation}
u_{i}=-\sum_{k=1}^{n}L_{ik}^\dagger \deg(k)+\text{vol}(G)L_{ii}^\dagger .\label{eq:cmt8}
\end{equation}
From (\ref{eq:cmt7}) and (\ref{eq:cmt8}), for $i\neq j$, the average
commute time between ${x}_{i}$ and ${x}_{j}$ is
\begin{eqnarray}
t_{ij}+t_{ji} & = & \sum_{k=1}^{n}L_{ik}^\dagger \deg(k)-\text{vol}(G)L_{ij}^\dagger -\sum_{k=1}^{n}L_{jk}^\dagger \deg(k)+\text{vol}(G)L_{jj}^\dagger \nonumber \\
 &  & +\sum_{k=1}^{n}L_{jk}^\dagger \deg(k)-\text{vol}(G)L_{ji}^\dagger -\sum_{k=1}^{n}L_{ik}^\dagger \deg(k)+\text{vol}(G)L_{ii}^\dagger \nonumber \\
 & = & \text{vol}(G)\left(L_{ii}^\dagger +L_{jj}^\dagger -2L_{ij}^\dagger \right),\label{eq:cmt9}
\end{eqnarray}
where
\begin{eqnarray}
L_{ij}^\dagger  & = & \sum_{l=2}^{n}\frac{1}{\mu_{l}}{\phiLap}_{l}(i){\phiLap}_{l}(j).\label{eq:cmt10}
\end{eqnarray}

Define the {\em commute time mapping} $\Psi : V \mapsto \mathbb{R}^{n-1}$ as
\begin{equation}
\Psi(x_{i}) = \left(\frac{1}{\sqrt{\mu_{2}}}{\phiLap}_{2}(i),\,\frac{1}{\sqrt{\mu_{3}}}{\phiLap}_{3}(i),\,\cdots,\,\frac{1}{\sqrt{\mu_{n}}}{\phiLap}_{n}(i)\right)^{T}.\label{eq:commute_time_embedding}\end{equation}
We see that the commute time between $x_i$ and $x_j$ is equivalent to the Euclidean distance between $\Psi(x_i)$ and $\Psi(x_j)$:
\begin{equation}
t_{ij}+t_{ji}=\text{vol}(G)\left\Vert \Psi({x}_{i})-\Psi({x}_{j})\right\Vert ^{2}.
\end{equation}
Also,
\begin{equation}
t_{ij}+t_{ji}=\text{vol}(G) (e_i - e_j)^T L^\dagger (e_i - e_j),
\end{equation}
where $e_i$ is the $i$'th standard unit vector.

\subsection{Comparison between the diffusion distance and the commute time distance}

Let us assume that the graph is regular, that is, $\deg(i)=d$ for all $i=1,2,\ldots,n$, or equivalently $D=dI$.
Then,
\begin{equation}
L = D - W = D(I-M) = d(I-M),
\end{equation}
so $M$ and $L$ share the same eigenvectors
\begin{equation}
\phi_l = \phiLap_l,
\end{equation}
and the eigenvalues are related by
\begin{equation}
\mu_l = d(1-\lambda_l).
\end{equation}

\begin{figure}[h]
\begin{center}
\includegraphics[width=100mm,height=65mm]{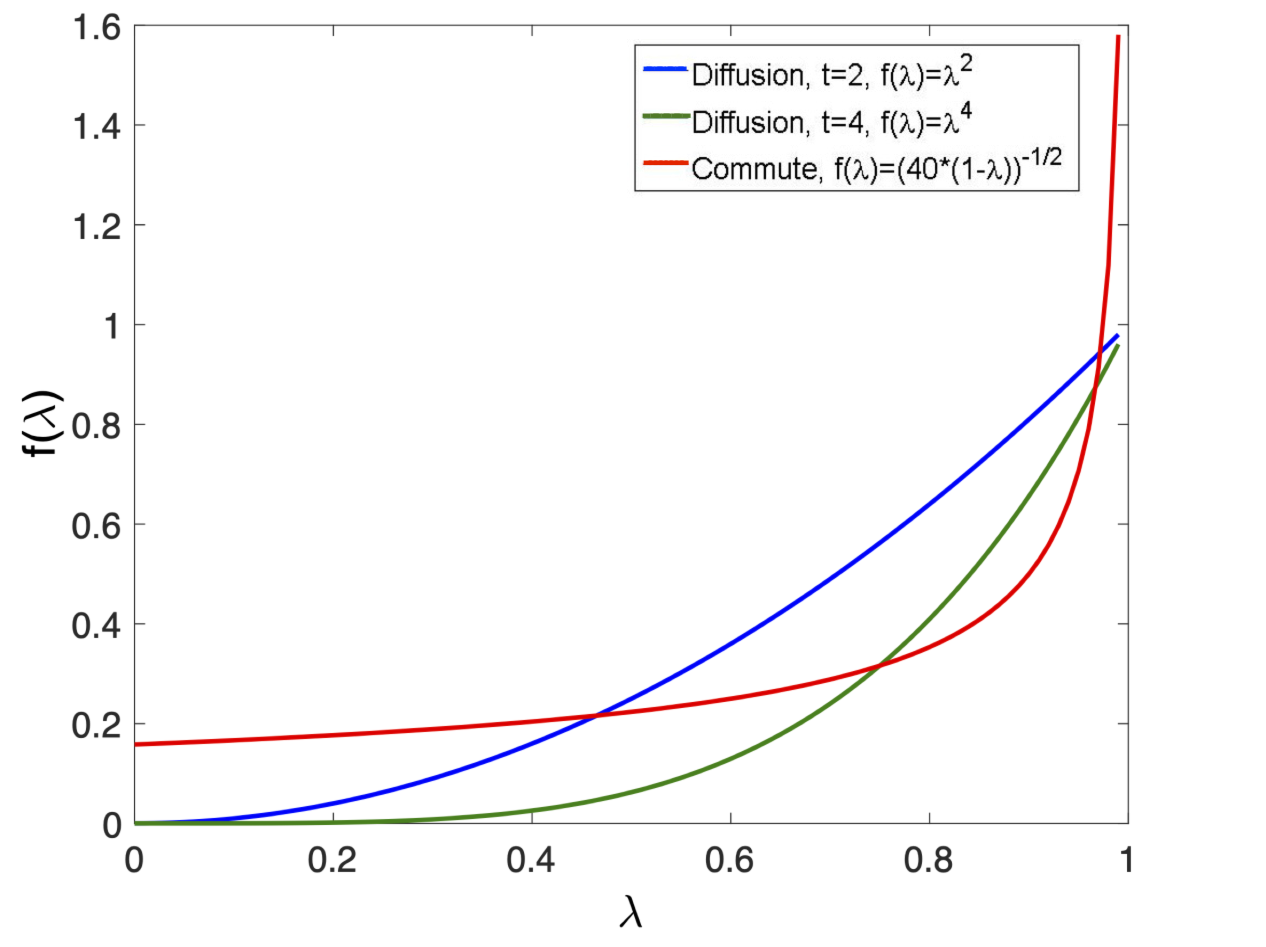}
\end{center}
\caption{Filter transfer function $f(\lambda)$ for diffusion map ($t=2$ and $t=4$) and the commute time map.}
\end{figure}

The commute time distance can become meaningless for some families of graphs. We see that the filter corresponding to the commute time distance associates a similar non-zero weight to all small eigenvalues. If all eigenvalues are assigned the same weight (i.e, constant filter), then the resulting distance between nodes $i$ and $j$ is simply $\frac{1}{d_i} + \frac{1}{d_j}$. This is quite boring, as it tells us nothing about the connectivity between the two nodes. It turns out that in certain cases the commute time distance has exactly this boring behavior \cite{von2014hitting}.  

We comment that many other filter/transfer functions are possible in practice, and it all depends on the problem at hand. 

\subsection{Effective resistance}
The commute time distance is also proportional to the {\em resistance distance}, where the graph is viewed as an electrical network of resistors, with the resistance of each edge given by $$r_{ij} = w_{ij}^{-1}$$ Using Ohm's Law and Kirckhoff's Law it is possible to compute the effective resistance between any two nodes in the graph. The effective resistance turns out to be exactly the same as the commute time distance just without the factor of $\text{vol}(G)$. This provides an alternative physical interpretation of the commute time distance.

To see this, suppose that $\mb{v} \in \mathbb{R}^n$ is the vector of electric potentials at the vertices. Recall Ohm's Law that relates the potential with the resistance $R$ and the electric current $I$
$$V = RI$$
Let $\mb{i} \in \mathbb{R}^{2|E|}$ be the vector of signed edge currents, with 
$$\mb{i}_{i \to j} = \frac{\mb{v}_i - \mb{v}_j}{r_{ij}} = w_{ij}(\mb{v}_i - \mb{v}_j)$$  
We see that
\begin{equation}
\mb{i} = \Delta U\mb{v},
\end{equation}
where $U$ is an $|E|\times n$ signed edge adjacency matrix of $G$, meaning that the row of $U$ corresponding to $(i,j)\in E$ is given by $e_i-e_j$. The signing is arbitrary but needs to be set. $\Delta=\diag(\vecc[w_{ij}]_{(i,j)\in E})$ is the $|E|\times |E|$ matrix whose diagonal elements are the edge weights.

Also, Kirckhoff's Law says that a resistor network cannot hold current: all 
flow entering vertex $i$ from
edges in the graph must exit $i$ to an external source. Let $\mb{i}_{\text{ext}} \in \mathbb{R}^n$ denote the vector of external currents,
where $\mb{i}_{\text{ext}}(i)$ is the amount of current entering the graph through node $i$. We then have
\begin{equation}
\mb{i}_{\text{ext}}(i) = \sum_{j : (i,j)\in E} \mb{i}_{i\to j}.
\end{equation}
Equivalently, in matrix form
\begin{equation}
\mb{i}_{\text{ext}} = U^T \mb{i} = U^T \Delta U\mb{v} = L\mb{v}
\end{equation}
Thus, by applying the Laplacian on the vector of potentials we get the external source current. In particular, we distinguish between internal nodes $i$ for which $\mb{i}_{\text{ext}}(i)=0$, and boundary nodes with $\mb{i}_{\text{ext}}(i) \neq 0$. Notice that for internal nodes
$$L \mb{v}(i)=0$$
is equivalent to
$$\mb{v}_i = \frac{1}{\deg(i)} \sum_{j:(i,j)\in E} w_{ij} \mb{v}_j$$  
that is, the potential at the node is the weighted average of the potential at its neighbors.

Now, in order to calculate the effective resistance between nodes $i$ and $j$ we need to solve
\begin{equation}
L\mb{v} = \mb{i}_{\text{ext}}  
\end{equation} 
with nodes $i$ and $j$ being boundary nodes and all other nodes being internal nodes. That is, we let
$$\mb{i}_{i,j,\text{ext}}(a) = I (e_i - e_j) = \left\{\begin{array}{rc}
                                    I & a=i \\
                                    -I & a=j \\
                                    0 & \text{otherwise} 
                                  \end{array}
 \right.$$
Notice that $\mb{1}^T \mb{i}_{i,j,\text{ext}}=0$, therefore if the graph is connected (the eigenvalue 0 is simple), the solution for the potential vector $\mb{v}$ is given by
\begin{equation}
\mb{v} = L^\dagger \mb{i}_{i,j,\text{ext}} = I L^\dagger(e_i - e_j)
\end{equation} 
The potential drop between nodes $i$ and $j$ is
\begin{equation}
\mb{v}_i - \mb{v}_j =  I (e_i - e_j)^T L^\dagger (e_i - e_j)
\end{equation}
Which shows that the effective resistance is 
\begin{equation}
R^{\text{eff}}(i,j) = \frac{\mb{v}_i - \mb{v}_j}{I} = (e_i - e_j)^T L^\dagger(e_i - e_j)
\end{equation}
Therefore, the effective resistance and the commute time distance are equivalent (up to the factor of $\text{vol}(G)$). 

\subsection{Time synchronization on graphs}
Here we demonstrate another application of the pseudo-inverse of the Laplacian and the commute time distance / effective resistance.
Consider the time-synchronization problem, in which $n$ sensing nodes are part of a multi-hop
communication network. Each node has a local clock,
but each pair of clocks differs by a constant offset. However,
nodes that communicate directly can estimate the difference
between their local clocks by exchanging
messages that are time stamped with local clock times. For
example, suppose that nodes $i$ and $j$ can communicate
directly with each other and have clock offsets $t_i$ and $t_j$
with respect to a reference clock. By passing messages
back and forth, the nodes can measure the relative clock
offset $t_i-t_j$ with the real-valued measurement
\begin{equation}
\label{offsets}
s_{ij} = t_i - t_j + \xi_{ij},\qquad (i,j)\in E
\end{equation}
where $\xi_{ij}$ is the measurement noise. The task is to estimate the clock offsets $t_1,\ldots, t_n$ (up to a global offset) from such pair-wise measurements.
It is also an instance of a ranking problem, in which $t_i$ may represent the ``strength" of the $i$'th player, and $s_{ij}$ is the outcome or score of a match between players $i$ and $j$.
Clearly, we need that the graph of measurements is connected. Also, suppose that the random variables $(\xi_{ij})_{(i,j)\in E}$ are independent zero-mean Gaussians, i.e., $$\xi_{ij} \sim \mathcal{N}(0,\sigma_{ij}^2),\qquad (i,j)\in E$$ Clearly, using the offset measurements we can only hope to estimate $t_1,\ldots,t_n$ up to a global shift. We may therefore assume that they are centered, i.e., $\sum_{i=1}^n t_i = 0$, or
$${\mathbf 1}^T t = 0$$
We seek to find a solution to the over-determined system of linear equations (\ref{offsets}) which in vector notation can be written as 
$$s = Ut + \xi$$
and $\xi \sim \mathcal{N}(0,\diag(\sigma)^2)$, where $\diag(\sigma)$ is the diagonal $|E|\times |E|$ matrix with $\sigma_{ij}$ in the diagonal. We pre-whiten the noise by dividing each equation with the standard deviation $\sigma_{ij}$ to get
$$\diag(\sigma)^{-1} s = \diag(\sigma)^{-1} Ut + z$$
where $$z \sim \mathcal{N}(0,I)$$
The least squares estimator $\hat{t}$ satisfies 
$$U^T \diag(\sigma)^{-2} U\hat{t} = U^T \diag(\sigma)^{-2} s$$
We therefore define weights 
$$w_{ij} = \sigma_{ij}^{-2}$$
for which the least squares estimator satisfies
$$L \hat{t} = U^T \Delta s$$,
where $\Delta = \diag(\vecc[w_{ij}]_{(i,j)\in E})$ is the $|E|\times |E|$ matrix whose diagonal elements are the edge-weights. The solution (assuming $\mb{1}^T t = 0$) is then given by
\begin{equation}
\hat{t} = L^\dagger U^T \Delta s
\end{equation} 
Notice that the least squares estimator is unbiased: 
\begin{equation}
\mathbb{E}[\hat{t}] = L^\dagger U^T \Delta \mathbb{E}[s] = L^\dagger U^T \Delta Ut = L^\dagger L t = t
\end{equation}
The error is given by
\begin{equation}
\hat{t} - t = L^\dagger U^T \Delta (s - \mathbb{E}[s]) = L^\dagger U^T \Delta \xi 
\end{equation}
Hence, the covariance of the estimation error is given by
\begin{equation}
\mathbb{E}[(\hat{t} - t)(\hat{t}-t)^T] = L^\dagger U^T \Delta \Delta^{-1} \Delta U L^\dagger = L^\dagger L L^\dagger = L^\dagger
\end{equation}
In particular, the mean squared error (MSE) of the $i$'th clock is the $i$'th diagonal term of $L^\dagger$
\begin{equation}
\mathbb{E}[(\hat{t}_i - t_i)^2] = L^\dagger_{ii},
\end{equation}
and the MSE of all $n$ estimators is
\begin{equation}
\mathbb{E}[\|\hat{t} -  t\|^2] = \sum_{i=1}^n L^\dagger_{ii} = \operatorname{Tr}(L^\dagger).
\end{equation}
Finally, the offset $t_i-t_j$ between nodes $i$ and $j$ is estimated as $\hat{t}_i - \hat{t}_j$. The MSE of the offset is
\begin{eqnarray}
\mathbb{E}[((\hat{t}_i - \hat{t}_j) - (t_i - t_j))^2] &=& \mathbb{E}[((\hat{t}_i - t_i) - (\hat{t}_j - t_j))^2] \nonumber \\
&=& \mathbb{E}[(\hat{t}_i - t_i)^2] + \mathbb{E}[(\hat{t}_j - t_j)^2] - 2\mathbb{E}[(\hat{t}_i - t_i)(\hat{t}_j - t_j)] \nonumber \\
&=& L^\dagger_{ii} + L^\dagger_{jj} - 2L^\dagger_{ij},
\end{eqnarray}
which is the effective resistance distance. More accurate offset estimates are reflected by shorter commute time distances.

\section{Other non-linear dimensional reduction techniques}

There are several other similar non-linear dimensional reduction methods. A particularly popular one is Isomap~\cite{ISOMAP_paper_2000}. The idea is to find an embedding in $\RR^d$ for which Euclidean distances in the embedding correspond as much as possible to geodesic distances in the graph. This can be achieved by, between pairs of nodes $v_i,\,v_j$ finding their geodesic distance and then using, for example, Multidimensional Scaling to find points $y_i\in\RR^d$ that minimize (for example)
\[
\min_{y_1,\dots,y_n \in \RR^d}\sum_{i,j} \left(   \|y_i-y_j\|^2 - \delta_{ij}^2   \right)^2,
\]
which can be done with spectral methods (it is a good exercise to compute the optimal solution to the above optimization problem). 

Locally linear embedding (LLE) \cite{roweis2000nonlinear} and Isomap are often considered landmarks in the area of nonlinear dimension reduction.  Beyond diffusion maps \cite{RRCoifman_SLafon_2006} and Laplacian eigenmaps \cite{belkin2001laplacian,belkin2003laplacian}, there are several other nonlinear dimension reduction methods worth mentioning here, including Hessian LLE \cite{donoho2003hessian}, local tangent space alignment \cite{zhang2004principal}, and UMAP \cite{mcinnes2018umap}. There are other variants of diffusion maps such as anisotropic diffusion maps based on the Mahalanobis distance \cite{singer2008non} and alternating diffusion \cite{lederman2018learning} for learning the geometry of common latent variables. Here, we choose to present in more detail t-Distributed Stochastic Neighbor Embedding (t-SNE) \cite{maaten2008visualizing}, a popular method for visualization of high-dimensional data, and later in Chapter~\ref{s:vectordiffusionmap} vector diffusion maps \cite{ASinger_HTWu_2011_VDM}  that uses the graph connection Laplacian and generalizes diffusion maps from scalar functions to vector fields.

\subsection{t-Distributed Stochastic Neighbor Embedding}\label{s:tsne}

t-SNE is a nonlinear dimension reduction method designed primarily for the visualization of high-dimensional data~\cite{maaten2008visualizing}.
Unlike spectral methods such as diffusion maps, which are based on eigenvalue problems and have a clear operator-theoretic interpretation, t-SNE is formulated as a stochastic optimization problem whose objective is to preserve local neighborhood relations in a probabilistic sense.

The central idea of t-SNE is that points that are close in high-dimensional space should remain close in the embedding, while distant points are modeled with much weaker constraints. To that end we will define two distances, one suitable for the original high-dimensional data space, and one for the low-dimensional embedding space.

Given a set of points $\{x_1, \dots, x_n\} \subset \mathbb{R}^d$, we define the similarity of point $x_j$ to $x_i$ as the conditional probability $p_{i|j}$ given by
$$p_{i|j} = \frac{\exp(-\|x_i - x_j\|^2 / 2\sigma_i^2)}{\sum_{k \neq i} \exp(-\|x_i - x_k\|^2 / 2\sigma_i^2)} \quad \text{and} \quad
p_{ij} = \frac{p_{j|i} + p_{i|j}}{2n}.
$$
In contrast to graph Laplacians and diffusion maps,
the bandwidth $\sigma_i$ is not global. Instead it is determined such that it is adapting the scale to the local density of the data.

Let $\{y_1, \dots, y_n\} \subset \mathbb{R}^m$ (where $m \ll d$, typically $m=2$ or $3$) denote the embedded points. In the low-dimensional space, we use a heavy-tailed Student t-distribution with one degree of freedom (which is a Cauchy distribution) to model similarities
$$q_{ij} = \frac{(1 + \|y_i - y_j\|^2)^{-1}}{\sum_{k \neq l} (1 + \|y_k - y_l\|^2)^{-1}}.$$
The choice of the t-distribution in the embedding space alleviates the so-called crowding problem, which arises when high-dimensional neighborhoods are projected into low-dimensional spaces.
The heavy tails of the t-distribution allow distant points in $\mathbb{R}^d$ to be modeled by even larger distances in $\mathbb{R}^m$, preventing the ``crowding'' at the center of the map often seen in standard SNE~\cite{hinton2002stochastic}.

The embedding is determined by minimizing the Kullback-Leibler (KL) divergence between the joint probability distribution $P$ (in $\mathbb{R}^d$) and $Q$ (in $\mathbb{R}^m$)
\begin{equation}
    \label{KL-tsne}
C = \text{KL}(P \| Q) = \sum_{i \neq j} p_{ij} \log \frac{p_{ij}}{q_{ij}}
\end{equation}
with respect to the variables $y_1,\dots,y_n$.
This cost function is non-convex and is typically optimized via gradient descent\footnote{The method of gradient descent is discussed in detail in Chapter~\ref{c:optimization}.}. The gradient has a particularly intuitive ``spring-force'' interpretation
$$\frac{\partial C}{\partial y_i} = 4 \sum_j (p_{ij} - q_{ij})\frac{y_i - y_j}{1 + \|y_i - y_j\|^2}$$
with two opposing forces:
\begin{itemize}
\item[$\bullet$] 
Attraction: The term $p_{ij}$ pulls $y_i$ toward $y_j$ if they are close in the original space. 
\item[$\bullet$] 
Repulsion: The term $q_{ij}$ pushes $y_i$ away from $y_j$ if they are close in the embedding space but not in the original space.
\end{itemize}

We note that  the embedding provided by t-SNE is invariant under global rotations and translations but not under scaling, 
and global distances and large-scale geometry are generally distorted. Thus, t-SNE may therefore better be viewed as a local neighborhood-preserving visualization method rather than a general-purpose dimension reduction technique.

\medskip

\begin{figure}[h!]
\begin{center}
\includegraphics[width=27mm,height=55mm]{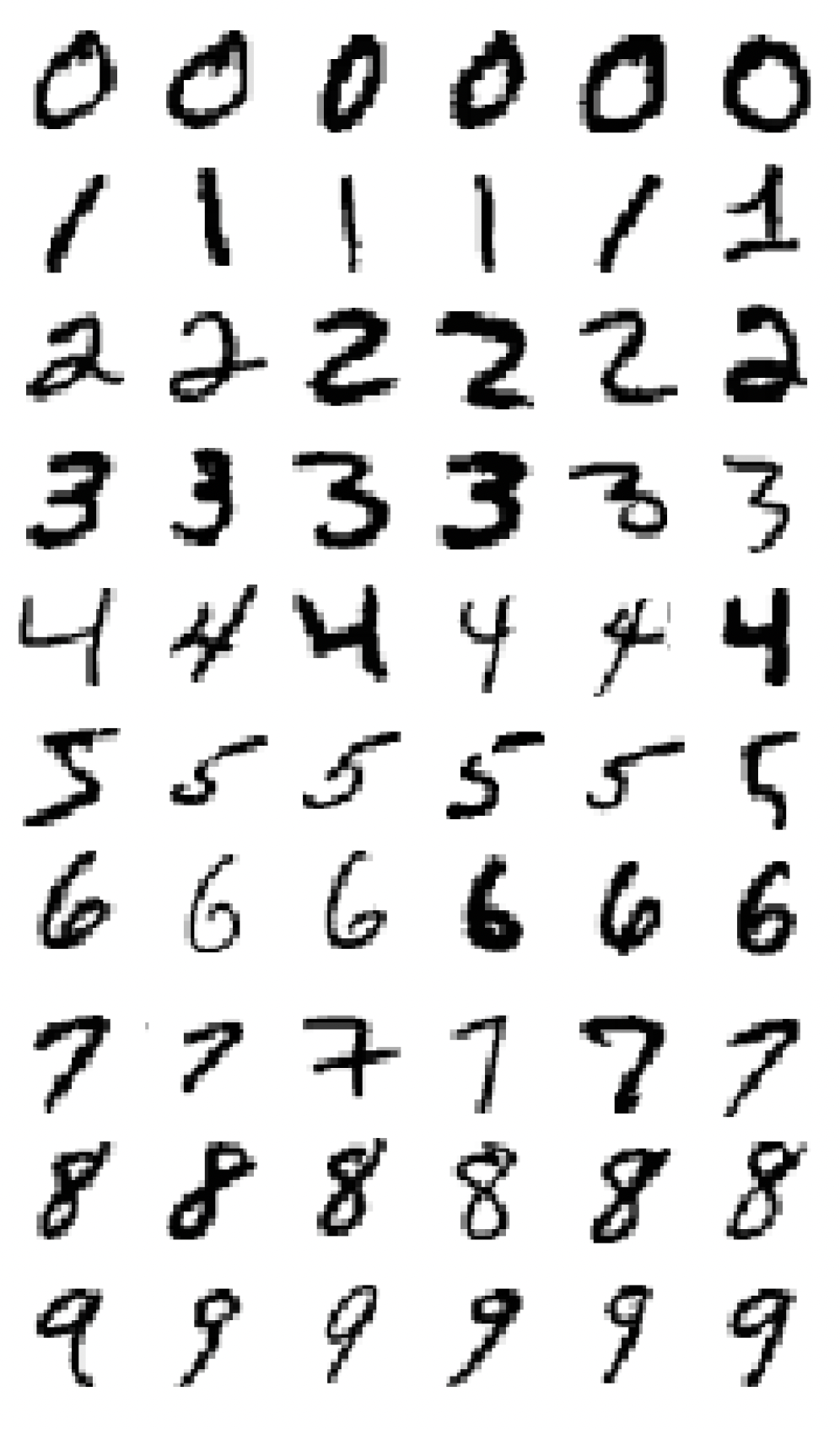}\qquad\quad
\includegraphics[width=80mm,height=55mm]{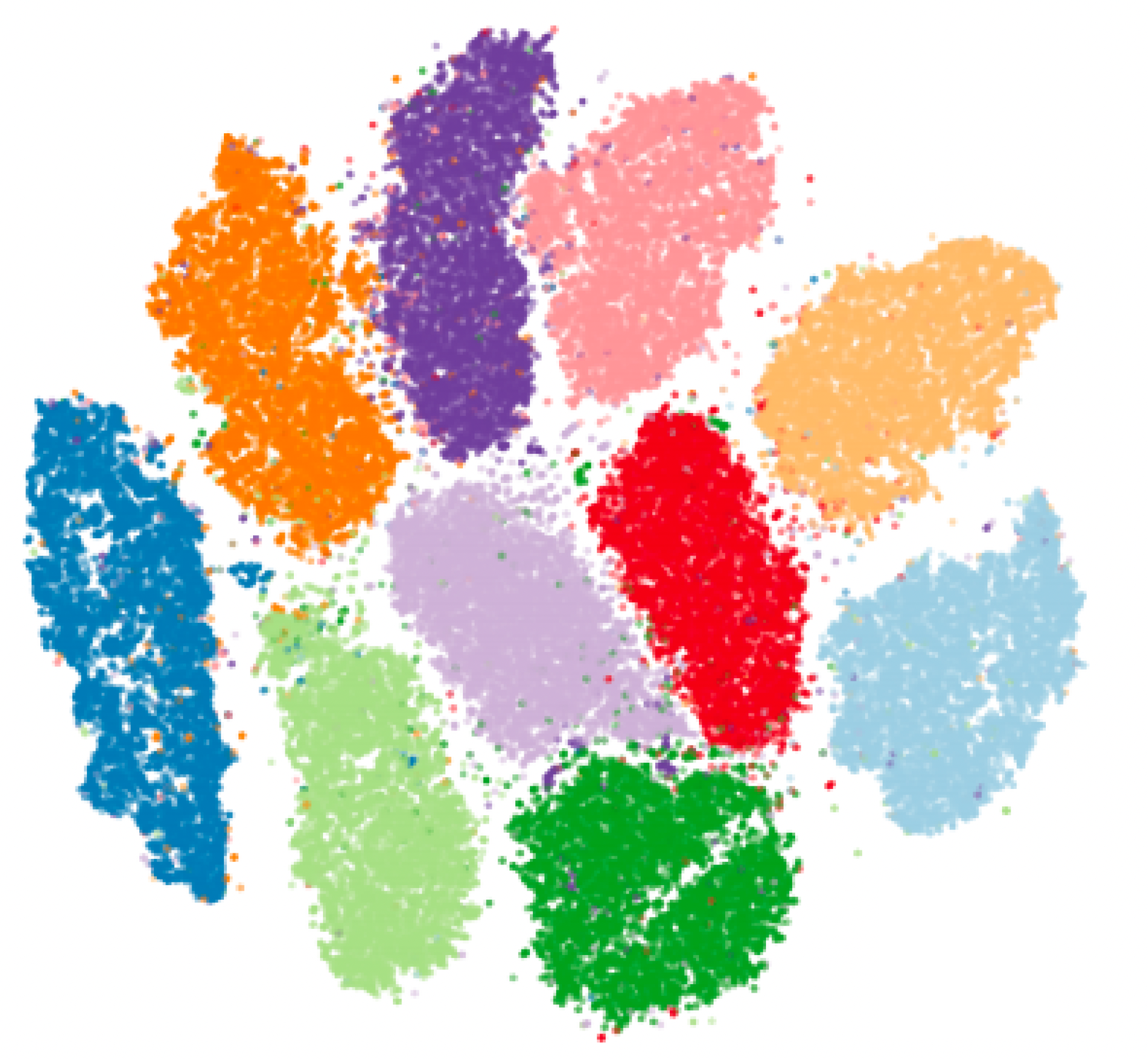}
\caption{Embedding of MNIST via t-SNE using the implementation proposed in~\cite{linderman2019clustering}. The left panel shows some example images from MNIST. The right panel shows the t-SNE embedding. Different colors correspond to different digits. (Image courtesy of Stefan Steinerberger.)}
\label{fig:mnist_tsne}
\end{center}
\end{figure}

We demonstrate the effectiveness of t-SNE by applying it to the MNIST (Modified National Institute of Standards and Technology)  database~\cite{lecun1998mnist} of handwritten digits. 
Each datapoint (image) $x_i$ is an $28 \times 28$ black-and-white image containing a handwritten digit. See the left panel of Figure~\ref{fig:mnist_tsne} for some example images.
The corresponding 2-D embedded points $y_i$ are computed via t-SNE.  Since the t-SNE objective is non-convex, the final embedding is sensitive to initialization and hyperparameters. Here, we use the implementation proposed in~\cite{linderman2019clustering}.
 The resulting embedding is shown in the right panel of Figure~\ref{fig:mnist_tsne}.
It is evident that t-SNE is quite effective in separating the different digits.

There are some interesting connections between t-SNE and spectral clustering, as uncovered in~\cite{linderman2019clustering} and~\cite{cai2022theoretical}, 
which provides some of the first rigorous mathematical justifications for why t-SNE behaves so effectively as a clustering tool. First note that the attractive term resembles a graph Laplacian $(L_P y)_i = \sum_j p_{ij}(y_i - y_j)$,
where $P$ acts as a weighted adjacency matrix. Also, the gradient descent flow during the initial phase behaves similarly to power iteration on a graph Laplacian~\cite{cai2022theoretical}. Specifically, the early iterations essentially perform a form of spectral clustering.
The clusters are formed based on the dominant eigenvectors of the transition matrix, much like the initial stages of a diffusion process.
It is furthermore shown in~\cite{linderman2019clustering} that if the data contains well-separated clusters (quantified by the spectral gap of the Laplacian), t-SNE is provably guaranteed to keep these clusters separated in the low-dimensional embedding. These connections should not be interpreted as placing t-SNE within the framework of spectral dimension reduction methods; rather, they provide insight into the local dynamics of its optimization landscape.

Diffusion maps are sometimes used as a preprocessing step  for t-SNE~\cite{wolf2018scanpy,moon2019visualizing,haghverdi2015diffusion}. The standard pipeline, common in computational biology \cite{wolf2018scanpy}, is to first reduce the data to a moderate number of diffusion coordinates which removes noise and compresses the intrinsic geometric structure into a tractable representation, and then to apply t-SNE (or the closely related UMAP algorithm~\cite{mcinnes2018umap}) to this 
$k$-dimensional intermediate representation to produce the final 2D visualization. This two-stage approach aims to combine the strengths of both methods: diffusion maps provide a geometrically meaningful and noise-robust intermediate representation, while t-SNE  optimizes that representation specifically for visual interpretability in two dimensions.

\section{Kernel learning}\label{s:kernellearning}

Linear methods are computationally efficient and well-understood, but they fail when the relationship between variables is inherently non-linear. The kernel approach allows us to lift our data into a high-dimensional (potentially infinite-dimensional) feature space where linear boundaries can capture complex structures, all without ever explicitly computing the coordinates in that space, a procedure referred to as the {\em kernel trick}. 

One of the primary advantages of the kernel approach is its ability to bypass the often difficult task of explicit feature engineering. In many domains---network analysis being a prime example---practitioners may struggle to define a concrete feature vector for an object, yet they can intuitively provide a measure of affinity or similarity between pairs of objects.

From a mathematical perspective, this ``user-defined'' similarity measure can be accompanied with some rigorous guarantees, as we will demonstrate in this section: If the practitioner's affinity function qualifies as a positive semi-definite  kernel (a property verifiable via Bochner’s Theorem, Theorem~\ref{th:bochner}), then Mercer’s Theorem (Theorem~\ref{th:mercer}) guarantees that this measure corresponds to a valid inner product in some high-dimensional feature space. Consequently, the kernel trick allows us to perform sophisticated non-linear learning via a similarity metric, rather than requiring the explicit construction of a feature map.

\subsection{Positive semidefinite kernels and reproducing kernels}

In practice, choosing a kernel is a combination of principled heuristics and empirical validation.
The linear kernel $K(x,x') = x^\top x'$ assumes the data is already linearly separable (or nearly so) in its original coordinates.
The polynomial kernel $K(x,x') = 
(\langle x, x' \rangle + r)^d$,
is useful in image processing where pixel interactions matter. The Gaussian kernel $K(x,x') = \exp (-\|x-x'\|^2/2\eps^2)$, which is sometimes also called RBF kernel (where RBF stands for Radial Basis Function) serves as a powerful default for most general-purpose applications, provided its hyperparameter is tuned carefully.
The Laplacian kernel $K(x,x') = \exp\bigl(-\gamma\|x-x'\|\bigr)$ for $\gamma>0$ is well suited for modeling physical processes or signals that have sharp changes or disjointed behavior (e.g., ``edges'' in natural images) that a Gaussian kernel might over-smooth.

One property that turns out to be very useful for a kernel is positive-definiteness.

\begin{definition}[Positive semi-definite kernel]\label{def:psd_kernel}
Let $\calX$ be a nonempty set.  A function $K\colon\calX\times\calX\to\R$ is called
a \emph{positive semi-definite (PSD) kernel}  if for all
$n\in\N$, all choices of points $x_1,\ldots,x_n\in\calX$, and all
$\alpha_1,\ldots,\alpha_n\in\R$,
\begin{equation}\label{eq:psd_condition}
  \sum_{i=1}^n\sum_{j=1}^n \alpha_i\alpha_j K(x_i,x_j) \;\ge\; 0.
\end{equation}
Equivalently, the \emph{Gram matrix} $\gramK\in\R^{n\times n}$ defined by
$(\gramK)_{ij} = K(x_i,x_j)$ for $i,j =1,\dots,n$ is positive semi-definite for all such choices.
\end{definition}

Condition \eqref{eq:psd_condition} is purely algebraic and requires no topological
structure on $\calX$.  The set $\calX$ can be discrete, a manifold, a function space,
or any abstract set.

Reproducing Kernel Hilbert Spaces (RKHS) provide a convenient and rigorous functional analysis framework
for kernel learning.

\begin{definition}\label{def:rkhs}
Let $\Hilb$ be a Hilbert space of real-valued functions on $\calX$ with inner product
$\langle \cdot, \cdot \rangle_{\Hilb}$.  We say $\Hilb$ is a \emph{reproducing kernel Hilbert space}
(RKHS) if there exists a function $K\colon\calX\times\calX\to\R$ such that:
\begin{enumerate}
\item For each $x\in\calX$, the function $K_x(\cdot) := K(x,\cdot)$ belongs to $\Hilb$;
\item For all $f\in\Hilb$ and all $x\in\calX$,
\begin{equation}\label{eq:reproducing}
  f(x) = \langle f, K_x\rangle_{\Hilb} = \langle f, K(x,\cdot)\rangle_{\Hilb}.
\end{equation}
\end{enumerate}
The function $K$ is called the \emph{reproducing kernel} of $\Hilb$.
\end{definition}

The following proposition (whose proof is left to the reader) summarizes basic properties of the reproducing kernel.

\begin{proposition}\label{prop:rk_properties}
Let $\Hilb$ be an RKHS with reproducing kernel $K$.  Then:
\begin{enumerate}
\item $K$ is symmetric: $K(x,x') = K(x',x)$ for all $x,x'\in\calX$.
\item $K$ is positive semi-definite in the sense of Definition~\ref{def:psd_kernel}.
\item For all $x,x'\in\calX$, $K(x,x') = \langle K_x, K_{x'} \rangle_{\Hilb}$.
\item The reproducing kernel is unique.
\end{enumerate}
\end{proposition}

A kernel $K$ can often be understood via an explicit {\em feature map} $\phi: \mathcal{X} \to \mathcal{F}$, where $\mathcal{F}$ is some (possibly high-dimensional) feature space, such that
\begin{equation}\label{eq:kerneltrick}
    K(x, x') = \langle \phi(x),\, \phi(x') \rangle_{\mathcal{F}}.
\end{equation}
This means the kernel computes the inner product between the feature representations of $x$ and $x'$, without needing to construct $\phi$ explicitly. This is the {\em kernel trick}: algorithms that depend only on inner products between inputs can be kernelized by replacing $\langle x, x' \rangle$ with $K(x, x')$, implicitly working in $\mathcal{F}$.

As a simple example, consider $x \in \mathbb{R}^2$ and the polynomial kernel given by $K(x, x') = (x^\top x')^2$. Expanding this out reveals it equals $\langle \phi(x), \phi(x') \rangle$ for $\phi(x) = (x^2,\, (x')^2,\, \sqrt{2} x x')^\top$, a three-dimensional feature map computed via a scalar operation.

\subsection{Mercer's Theorem}

While the kernel trick allows us to compute similarities without knowing the coordinates of the data in the feature space, Mercer’s Theorem provides the explicit mathematical construction of those coordinates. It provides a constructive answer to the question {\em ``which functions $K$ are valid kernels, i.e., correspond to an inner product in some feature space?''}

\begin{theorem}[Mercer's Theorem]\label{th:mercer}
Let $\mathcal{X} \subseteq \mathbb{R}^d$ be closed, 
 let $\mu$ be a strictly positive Borel measure on ${\mathcal X}$, and let $K: \mathcal{X} \times \mathcal{X} \to \mathbb{R}$ be a continuous, symmetric function. If $K$ is positive semi-definite, then there exist non-negative eigenvalues $\{\lambda_j\}_{j=1}^\infty$ and an orthonormal sequence of continuous eigenfunctions $\{\psi_j\}_{j=1}^\infty$ in $L^2(\mathcal{X})$ such that
\begin{equation}
    K(x, x') = \sum_{j=1}^{\infty} \lambda_j\, \psi_j(x)\, \psi_j(x'),
\end{equation}
where the series converges absolutely for each pair $(x,x') \in \mathcal{X} \times \mathcal{X}$
and uniformly on every compact subset of $\mathcal{X} \times \mathcal{X}$.
\end{theorem}

\begin{proof}
Define the linear operator $T_k: L^2(X, \mu) \to L^2(X, \mu)$ by
$$(T_k f)(x) = \int_X K(x, t) f(t) d\mu(t).$$
Because $K$ is continuous and symmetric, $T_k$ is a compact, self-adjoint operator. By the Spectral Theorem for compact self-adjoint operators, there exists an orthonormal basis of $L^2(X, \mu)$ consisting of eigenfunctions $\{\phi_j\}_{j=1}^\infty$ with corresponding eigenvalues $\lambda_j$.
Due to positive definiteness, $\lambda_j \ge 0$.
We order them such that $\lambda_1 \ge \lambda_2 \ge \dots \to 0$.
The eigenfunctions corresponding to $\lambda_j > 0$ are continuous because $\phi_j(x) = \frac{1}{\lambda_j} \int_X K_j(x, t) \phi_j(t) d\mu(t)$.

To establish pointwise convergence we 
consider the truncated kernel expansion $K_n(x, t) = \sum_{j=1}^n \lambda_j \phi_j(x) \phi_j(t)$. We want to show that as $n \to \infty$, $K_n \to K$.
Define the remainder kernel $R_n(x, t) := K(x, t) - \sum_{j=1}^n \lambda_j \phi_j(x) \phi_j(t)$.
$R_n$ is also a continuous, symmetric, positive definite kernel. For any positive definite kernel, the diagonal elements must be non-negative, i.e.,
$$R_n(x, x) = K(x, x) - \sum_{j=1}^n \lambda_j \phi_j(x)^2 \ge 0.$$
This implies that for all $n$ we have
$$\sum_{j=1}^n \lambda_j \phi_j(x)^2 \le K(x, x).$$
Since the partial sums are monotonic and bounded, the series $\sum_{j=1}^\infty \lambda_j \phi_j(x)^2$ converges pointwise. By the Cauchy-Schwarz inequality
\begin{equation}\label{eq:mercercauchyschwarz}
\Big| \sum_{j=m}^n \lambda_j \phi_j(x) \phi_j(t) \Big| \le \sqrt{\sum_{j=m}^n \lambda_j \phi_j(x)^2} \sqrt{\sum_{j=m}^n \lambda_j \phi_j(t)^2}.
\end{equation}
This proves the absolute convergence of the series for each pair $(x, t)$.

To prove uniform convergence on a compact subset $S \subset X$, we invoke Dini's Theorem.
Let $g_n(x) = \sum_{j=1}^n \lambda_j \phi_j(x)^2$. We know $g_n(x)$ is a sequence of continuous functions (sums of squares of continuous eigenfunctions).
$g_n(x)$ is monotonically increasing ($g_{n+1} \ge g_n$ since $\lambda_j \ge 0$).
The limit $g(x) = \sum_{j=1}^\infty \lambda_j \phi_j(x)^2$ is equal to $K(x, x)$ (which is continuous).
By Dini's Theorem, if a monotonic sequence of continuous functions converges pointwise to a continuous function on a compact set, the convergence is uniform. Thus, $\sum \lambda_j \phi_j(x)^2$ converges uniformly to $K(x, x)$ on $S$. Utilizing once more the bound~\eqref{eq:mercercauchyschwarz}, the uniform convergence of the diagonal terms implies the uniform convergence of the cross-terms $K(x, t)$.

Finally, we must show the limit is indeed $K(x, t)$. To that end, we define $R(x, t) := K(x, t) - \sum_{j=1}^\infty \lambda_j \phi_j(x) \phi_j(t)$.
Since $T_R$ is a positive operator with all eigenvalues equal to zero (as we subtracted the entire spectral decomposition), the spectral radius of $T_R$ is 0. For a self-adjoint compact operator, this implies $T_R = 0$. Since $R(x, t)$ is continuous and $\mu$ is strictly positive, $T_R = 0$ implies $R(x, t) = 0$ for all $(x, t)$, hence
$$K(x, t) = \sum_{j=1}^{\infty} \lambda_j \phi_j(x) \phi_j(t),$$
which completes the proof.

\end{proof}

The closely related but more general Moore-Aronszajn Theorem~\cite{aronszajn1950theory} states that 
every positive definite kernel defines a unique RKHS (existence and structure). In comparison, Mercer's Theorem establishes under additional analytic assumptions, that the kernel admits an eigenfunction expansion which comes with an explicit feature map given by
 $$\phi(x) = \bigl(\sqrt{\lambda_1}\psi_1(x),\,\sqrt{\lambda_2}\psi_2(x),\,\ldots\bigr).$$
While this concrete construction of the feature map is useful for theoretical purposes, in practice we avoid computing it by using the kernel trick.

\subsection{Bochner's Theorem}\label{s:bochner}

While Mercer's Theorem provides a link between a PSD kernel and a feature map in $L_2$, {\em Bochner's Theorem} provides a powerful characterization of PSD kernels for the case when the kernel is {\em translation-invariant}, i.e., a kernel that can be expressed in the form $K(x, x') = \kappa(x - x')$ for some function $\kappa$.

Consider a finite Borel measure $\mu$ of $\mathbb{R}^p$. That is, $\mu(\mathbb{R}^p) < \infty$ and $\mu(A) = \int_{A \subset \mathbb{R}^p} \,d\mu$. For example, if $\mu$ is absolutely continuous with respect to the Lebesgue measure $dx$ then $d\mu = f(x)\,dx$ where $f$ is the probability density function. The Fourier transform of $\mu$ is
\begin{equation}
\hat{\mu}(\xi) = \int_{\mathbb{R}^p} e^{-2\pi \imath \langle \xi, x \rangle}\,d\mu(x).
\end{equation}
Observe that
\begin{equation}
|\hat{\mu}(\xi)| \leq \mu(\mathbb{R}^p) < \infty.
\end{equation}
That is, $\hat{\mu}$ is bounded. Also, $\hat{\mu}$ is a continuous function. Moreover, for every $n=1,2,\ldots$ we can associate with $\hat{\mu}$ a PSD quadratic form:
\begin{eqnarray}
\sum_{i,j=1}^n \hat{\mu}(\xi_i-\xi_j) \alpha_i \bar{\alpha}_j &=& \sum_{i,j=1}^n \alpha_i \bar{\alpha}_j \int_{\mathbb{R}^p} e^{-2\pi \imath \langle \xi_i - \xi_j, x\rangle}\,d\mu(x) \\
&=& \int_{\mathbb{R}^p} \left| \sum_{i=1}^n \alpha_i e^{-2\pi \imath \langle \xi_i, x \rangle }\right|^2 \,d\mu(x) \geq 0.
\end{eqnarray}
We conclude that $\hat{\mu}$ is bounded, continuous, and the matrix $(\hat{\mu}(\xi_i-\xi_j))_{i,j=1}^n$ is PSD for every $n=1,2,\ldots$ and for all $\xi_1,\ldots,\xi_n \in \mathbb{R}^p$.

But the converse is also true: If $f$ is bounded, continuous, and the matrix $(f(x_i-x_j))_{i,j=1}^n$ is PSD for all choices of $n=1,2,\ldots$ and $x_1,\ldots,x_n\in \mathbb{R}^p$, then $f$ is the Fourier transform of a finite Borel measure.

From the convolution theorem and Parseval's theorem we get
\begin{equation}
\langle \hat{K}  \hat{\alpha}, \hat{\alpha}  \rangle \geq 0.
\end{equation}
It follows that
\begin{equation}
\int_{\mathbb{R}^p} |\hat{\alpha}|^2(\xi) \hat{K}(\xi)\,d\xi \geq 0.
\end{equation}
Since we can choose $\alpha(x)$ (or $\hat{\alpha}(\xi)$), we must have $\hat{K}(\xi) \geq 0$. 

We thus have proved  Bochner's theorem~\cite{bochner1932vorlesungen}:
\begin{theorem}\label{th:bochner}
A (complex-valued) function $K \in L^{\infty}(\mathbb{R}^p) \cap C(\mathbb{R}^p)$ is positive definite on $\mathbb{R}^p$
if and only if it is the (inverse) Fourier transform of a finite non-negative
Borel measure $\mu$ on $\mathbb{R}^p$, i.e.,
\begin{equation}
K(x) = \int_{\mathbb{R}^p} e^{2\pi \imath \langle \xi, x \rangle}\,d\mu(\xi).
\end{equation}
\end{theorem}

\medskip
When $K$ is normalized so that $K(0) = 1$, the measure $\mu(\xi)$ becomes a probability distribution over frequencies $\xi \in \mathbb{R}^d$. For the Gaussian kernel
$$    K(x, x') = \exp\!\left(-\frac{\|x - x'\|^2}{2\eps^2}\right),
$$
the spectral measure $\mu(\xi)$ is a Gaussian distribution with variance $1/\eps^2$ (because the inverse Fourier transform of the Gaussian is also a Gaussian, see e.g.~\cite[Lemma 1.5.1]{Gro01}), which immediately confirms that the Gaussian kernel is PSD.

This  has immediate consequences for diffusion maps. Let $x_1,x_2,\ldots, x_n \in \mathbb{R}^p$ and consider the $n\times n$ weight matrix $W$ with entries
$
W_{ij} = e^{-\frac{\|x_i - x_j \|^2}{2\epsilon^2}}$.
Then $W$ is PSD for all values of $n$, $p$, and $\epsilon$. More generally,  Bochner's theorem implies that for positive kernels the eigenvalues of $W$ and the adjacency matrix $A$ are non-negative and we can use the diffusion map with any positive real $t>0$.

\subsection{The Representer Theorem}

The following theorem, known as {\em Representer Theorem}, is fundamental in kernel methods because it guarantees that the optimal solution to a regularized cost functional lives in a finite-dimensional subspace, regardless of how large the underlying Hilbert space is.

\begin{theorem}\label{th:representer}
Let $\mathbb{H}$ be an RKHS with reproducing kernel $K$. 
Given a set of training data $\{(x_i, y_i)\}_{i=1}^n \subset \mathcal{X} \times \mathbb{R}$, consider a cost function of the form
$$J(f) = L(f(x_1), \dots, f(x_n)) + \Psi(\|f\|_{\mathbb{H}}^2)$$
where $L: \mathbb{R}^n \to \mathbb{R}$ is an arbitrary loss function and $\Psi: [0, \infty) \to \mathbb{R}$ is a strictly monotonically increasing regularization function.

Then, any minimizer $f^* \in \mathbb{H}$ of $J(f)$ admits a representation of the form
$$f^*(x) = \sum_{i=1}^n \alpha_i K(x, x_i)$$
for some coefficients $\alpha_i \in \mathbb{R}$.
\end{theorem}

\begin{proof}
Let $\mathbb{H}_S = \text{span}\{K(\cdot, x_1), \dots, K(\cdot, x_n)\}$ be the subspace of $\mathbb{H}$ spanned by the kernel functions at the training points. Any function $f \in \mathbb{H}$ can be decomposed into its projection onto $\mathbb{H}_S$ and an orthogonal component $f_\perp \in \mathbb{H}_S^\perp$:
$$f = f_S + f_\perp, \quad \text{where } f_S = \sum_{i=1}^n \alpha_i K(\cdot, x_i).$$
By the reproducing property, for any $j \in \{1, \dots, n\}$, the evaluation of $f$ at $x_j$ is
$$f(x_j) = \langle f, K(\cdot, x_j) \rangle_{\mathbb{H}} = \langle f_S + f_\perp, K(\cdot, x_j) \rangle_{\mathbb{H}}.$$
Since $f_\perp$ is orthogonal to the span of the kernels, $\langle f_\perp, K(\cdot, x_j) \rangle_{\mathbb{H}} = 0$, thus
$$f(x_j) = \langle f_S, K(\cdot, x_j) \rangle_{\mathbb{H}} = f_S(x_j).$$
This shows that the loss term $L$ depends only on $f_S$. Now consider the regularization term. By the Pythagorean theorem we have
$$\|f\|_{\mathbb{H}}^2 = \|f_S + f_\perp\|_{\mathbb{H}}^2 = \|f_S\|_{\mathbb{H}}^2 + \|f_\perp\|_{\mathbb{H}}^2.$$
Since $\Psi$ is strictly monotonically increasing, $\|f_\perp\|_{\mathbb{H}}^2$ must be zero for $f$ to minimize $J(f)$. If $\|f_\perp\|_{\mathbb{H}} > 0$, then $f_S$ would achieve a strictly lower cost than $f$. Thus, the minimizer $f^*$ must lie entirely in $\mathbb{H}_S$. 

\end{proof}

The representer theorem acts as a kind of dimensionality reduction principle for optimization: although learning occurs in a potentially infinite-dimensional RKHS, the optimal solution lies in the finite-dimensional span of kernel evaluations at the training data. Consequently, the optimization problem depends on the number of training observations rather than the dimensionality of the feature space.

\subsection{Applications of kernel learning}

\noindent
\textbf{Kernel ridge regression:} 
The kernelized version of ridge regression replaces  $\min \|X\theta - y\|^2 + \lambda\|\theta\|^2$ with a minimization over functions $f$ in a RKHS $\mathbb{H}$, i.e., we consider
\begin{equation}
    \min_{f \in \mathbb{H}} \,\frac{1}{m}\sum_{i=1}^m (f(x_i) - y_i)^2 + \lambda \|f\|^2_{\mathbb{H}}.
\end{equation}
By Theorem~\ref{th:representer}, the solution takes the form $f^*(x) = \sum_{i=1}^m \alpha_i K(x_i, x)$, and substituting this into the objective reduces the problem to
\begin{equation}
    \min_{\alpha \in \mathbb{R}^m} \,\frac{1}{m}\|K\ \alpha - y\|^2 + \lambda\, \alpha^\top K \alpha,
\end{equation}
where $G_K$ is the $m \times m$ Gram matrix with entries $(G_K)_{ij} = K(x_i, x_j)$. Setting the gradient to zero yields the closed-form solution:
\begin{equation}
    \alpha = (K + \lambda m I)^{-1}y.
\end{equation}
Predictions at a new point $x$ are then given by $f^*(x) = K(x)^\top \alpha$, where $K(x) = (K(x_1, x), \dots, K(x_m, x))^\top$. The key insight is that the algorithm never requires explicit access to the feature map $\phi(x)$; all computations are performed through the kernel matrix $G_K$.

\medskip
\noindent
\textbf{Kernel regression:}
While kernel ridge regression is an optimization-based approach, kernel regression is a localized density estimation approach. In particular, it is a non-parametric regression method that estimates the relationship between an input variable $x$ and an output $y$ by taking a weighted average of observed responses, where the weights depend on the similarity between inputs measured by a kernel function. 

Suppose we observe data $(x_1,y_1),\dots,(x_n,y_n)$, to predict the value of $y$ at a new point $x$, kernel regression assigns larger weights to observations whose inputs are close to $x$.
The resulting prediction is a weighted average of the form
$$
\hat f(x) = \frac{\sum_{i=1}^{n} K(x,x_i) y_i}
{\sum_{i=1}^{n} K(x,x_i)}.
$$
This formula is known as the {\em Nadaraya–Watson kernel regression estimator}~\cite{nadaraya1964estimating,watson1964smooth}.

See~\cite{gilles2025cryo} for an application of kernel regression in cryogenic electron microscopy (cryo-EM).

\medskip
\noindent
\textbf{Kernel PCA:}
Standard PCA finds directions of maximum variance by computing the eigenvectors of the sample covariance matrix. Kernel PCA~\cite{scholkopf1998nonlinear} extends this to nonlinear settings by implicitly performing PCA in the feature space $\mathcal{F}$. 
By exploiting the kernel trick, kernel PCA does not need to set up a feature map explicitly.
The (centered) covariance operator in $\mathcal{F}$ is:
\begin{equation}
    C = \frac{1}{m}\sum_{i=1}^m \langle \phi(x_i), \phi(x_i)\rangle,
\end{equation}
and one seeks eigenvectors $v \in \mathcal{F}$ satisfying $Cv = \mu v$. Since any such eigenvector must lie in the span of $\{\phi(x_i)\}$, we can write $v = \sum_{i=1}^m \alpha_i \phi(x_i)$, and the eigenvalue problem reduces to:
\begin{equation}
    \tilde{G}\alpha = m\mu\, \alpha,
\end{equation}
where $\tilde{G}$ is the {\em centered} Gram matrix
\begin{equation}
    \tilde{G} = G_K - \mathbf{1}_m G_K - G_K\mathbf{1}_m + \mathbf{1}_m G_K \mathbf{1}_m,
\end{equation}
with $\mathbf{1}_m = \frac{1}{m}\mathbf{1}\mathbf{1}^\top$ the $m\times m$ matrix with all entries $1/m$. The projection of a new point $x$ onto the $r$-th kernel principal component is:
\begin{equation}
    z_r(x) = \sum_{i=1}^m \alpha_i^{(r)}\, K(x_i, x),
\end{equation}
where $\alpha^{(r)}$ is the $r$-th eigenvector of $\tilde{G}$. Kernel PCA is particularly effective for dimensionality reduction and feature extraction when the data lies on a nonlinear manifold, as the feature space $\mathcal{F}$ may ``unfold'' structure that is invisible to linear PCA in the input space.

\smallskip
Both kernel PCA and diffusion maps are spectral manifold learning techniques that embed high-dimensional data into a lower-dimensional space by solving an eigenvalue problem derived from a kernel matrix $K(x,x')$. But kernel PCA operates on a centered version of $K$. It seeks the directions of maximum variance in a high-dimensional feature space, hence it is concerned with certain geometric projections in Hilbert space. It is also highly sensitive to data density. In contrast, diffusion maps operate on a normalized version of $K$, so it can be made more invariant to density via renormalization. It views $K$ as a transition matrix for a Markov chain on the data and emphasizes intrinsic geometry and connectivity, rather than variance in feature space.

\medskip
\noindent
\textbf{Random Fourier Features:} 
Assume we are given a  dataset $\{x_1, \dots , x_n\} \in \R^d$.
Since the measure $\mu(\xi)$ arising in Bochner's Theorem is a probability distribution (for a properly normalized kernel $K$), we can write
\begin{equation*}
    K(x, x') = \mathbb{E}_{\xi \sim \mu}\!\left[e^{2\pi \imath \xi^\top(x - x')}\right]
    = \mathbb{E}_{\omega \sim \mu}\!\left[\cos\!\left(\omega^\top(x - x')\right)\right],
\end{equation*}
where the second equality holds when $\mu$ is symmetric. This means we can approximate $K(x, x')$ by drawing random frequencies $\xi_1, \dots, \xi_r \sim \mu$ with $r \ll n$ and constructing the $r$-dimensional random Fourier feature map
\begin{equation*}
    \phi(x) = \sqrt{\frac{2}{d}}\Bigl[\cos(\xi_1^\top x + b_1),\; \dots,\; \cos(\xi_r^\top x + b_r)\Bigr]^\top,
\end{equation*}
where $b_i \sim \text{Uniform}[0, 2\pi]$, so that $K(x, x') \approx \langle \phi(x), \phi(x') \rangle$. This approximation (due to Rahimi \& Recht~\cite{rahimi2007random}) allows kernel methods to scale to large datasets by replacing expensive $\mathcal{O}(n^2)$ kernel matrix computations with cheaper inner products in a $r$-dimensional feature space.

The Random Fourier Features method can be used for instance to create a low-rank approximation of the kernel matrix for kernel PCA. This technique makes kernel PCA scalable to large datasets. Instead of working with the kernel matrix $K \in \R^{n \times n}$, we work 
with the \emph{explicit feature matrix} $\Phi \in \R^{n \times D}$:
\begin{equation}\label{eq:feature_matrix}
  \Phi = \begin{bmatrix}
    \phi(x_1)^\top\\
    \phi(x_2)^\top\\
    \vdots\\
    \phi(x_n)^\top
  \end{bmatrix} \in \R^{n \times D}
\end{equation}
Then one can show that with a certain probability $K \approx \Phi\Phi^\top$, see~\cite{rahimi2007random}. This is a form of matrix sketching, see also Chapter~\ref{s:sketching}.

\smallskip
\noindent
\textbf{Kernel SVM:}
Another natural use case of the kernel approach arises in connection with Support Vector Machines, a classification method which will be discussed in detail (together with its kernel variant) in Chapter~\ref{s:svm}.

\section{Semi-supervised learning}\label{s:semisupervised}

Classification is a central task in machine learning which we will explore in detail in Chapters~\ref{c:classification} and~\ref{c:deeplearning}. In this section, we discuss an application of graph Laplacians and diffusion maps to classification in the {\em semi-supervised} setting.

In a supervised learning setting we are given many labeled examples and want to use them to infer the label of a new, unlabeled example. For simplicity, let us focus on the case of two labels, $\{-1,+1\}$. 

\begin{figure}[h]
\begin{center}
\includegraphics[width=0.22\textwidth]{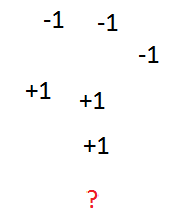}
\caption{Given a few labeled points, the task is to label an unlabeled point.}
\label{figure:2:SupervisedLearning}
\end{center}
\end{figure}

Let us consider the task of labeling the point ``?'' in Figure~\ref{figure:2:SupervisedLearning} given the labeled points. The natural label to give to the unlabeled point would be $1$.

\begin{figure}[h]
\begin{center}
\includegraphics[width=0.4\textwidth]{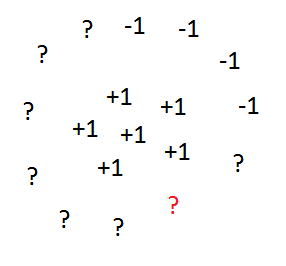}
\caption{In this example we are given many unlabeled points, the unlabeled points help us learn the geometry of the data.}
\label{figure:2:SemiSupervisedLearning}
\end{center}
\end{figure}

However, if we are given not just one unlabeled point, but many, as in Figure~\ref{figure:2:SemiSupervisedLearning}; then it starts being apparent that $-1$ is a more reasonable guess.

Intuitively, the unlabeled data points allowed us to better learn the intrinsic geometry of the dataset. That is the idea behind {\em Semi-Supervised Learning} (SSL). While supervised learning relies exclusively on labeled examples, semi-supervised learning combines labeled and unlabeled data to leverage the underlying structure of the input space when labeled data are limited.

So, let us explore how we can exploit the structure of the input data to improve classification when labels are scarce.
Just as above, we will use the data points to construct (via a kernel $K_\eps$) a graph $G=(V,E,W)$ where nodes correspond to points. More precisely, let $l$ denote the number of labeled points with labels $f_1,\dots,f_l$, and $u$ the number of unlabeled points (with $n=l+u$), the first $l$ nodes $v_1,\dots,v_l$ correspond to labeled points and the rest $v_{l+1},\dots,v_n$ are unlabaled. We want to find a function $f:V\to \{-1,1\}$ that agrees on labeled points: $f(i)=f_i$ for $i=1,\dots,l$ and that is ``as smooth as possible'' on the graph. A way to pose this is the following \cite{zhu2003semi} 
\[
 \min_{f:V\to \{-1,1\}:\, f(i)=f_i\, i=1,\dots,l} \sum_{i<j}w_{ij}\left( f(i) - f(j) \right)^2.
\]
Instead of restricting ourselves to giving $\{-1,1\}$ we allow ourselves to give real valued labels, with the intuition that we can ``round'' later by, e.g., assigning the sign of $f(i)$ to node $i$.

We thus are interested in solving
\[
 \min_{f:V\to \RR:\, f(i)=f_i\, i=1,\dots,l} \sum_{i<j}w_{ij}\left( f(i) - f(j) \right)^2.
\]

If we denote by $f$ the vector (in $\RR^n$ with the function values) then, recalling Proposition~\ref{prop:cut_Laplacian}, we are can rewrite the problem as
\begin{eqnarray*}
 \sum_{i<j}w_{ij}\left( f(i) - f(j) \right)^2 
= f^TL_Gf.
\end{eqnarray*}

\begin{remark}
 Consider an analogous example on the real line, where one would want to minimize
 \[
  \int f'(x)^2 dx.
 \]
Integrating by parts
\[
 \int f'(x)^2 dx = \mathrm{Boundary\ Terms} - \int f(x)f''(x)dx.
\]
Analogously, in $\RR^d$:
\[
 \int \left\|\nabla f(x)\right\|^2 dx =  \int \sum_{k=1}^d \left( \frac{\partial f}{\partial x_k}(x) \right)^2 dx = \mathrm{B.\ T.} - \int f(x) \sum_{k=1}^d \frac{\partial^2 f}{\partial x_k^2}(x)dx =
 \]
 $\displaystyle{= \mathrm{B.\ T.} - \int f(x) \Delta f(x) dx}$,

which helps motivate the use of the term graph Laplacian for $L_G$.
\end{remark}

Let us consider our problem
\[
 \min_{f:V\to \RR:\, f(i)=f_i\, i=1,\dots,l} f^TL_G f.
\]
We can write
\[
 D = \left[ \begin{array}{cc} D_L & 0  \\  0  &  D_U  \end{array}  \right], \quad W = \left[ \begin{array}{cc} W_{LL} & W_{LU}  \\  W_{UL}  &  W_{UU}  \end{array}  \right], \quad L_G = \left[ \begin{array}{cc} D_L - W_{LL} & -W_{LU}  \\  -W_{UL}  & D_U - W_{UU}  \end{array}  \right], 
 \]
and $f =  \left[ \begin{array}{c} f_L   \\  f_U  \end{array}  \right]$, where $L$ and $U$ correspond to the set of indices of  labeled and unlabeled nodes, respectively (not to be confused with the graph Laplacian $L_G$).

Then we want to find (recall that $W_{UL}=W_{LU}^T$)
\[
\min_{f_U\in \RR^{u}} f_L^T\left[ D_L - W_{LL}\right] f_L - 2f_U^TW_{UL}f_L + f_U^T\left[ D_U - W_{UU}\right] f_U.
\]
By first-order optimality conditions\footnote{Readers who want to refresh their memory about first-order optimality conditions may take a peak at Chapter~\ref{s:optimality}.}, it is easy to see that the optimal satisfies
\[
 \left( D_U - W_{UU}\right) f_U = W_{UL}f_L.
\]
If $D_U-W_{UU}$ is invertible\footnote{It is not difficult to see that unless the problem is in some form degenerate, such as the unlabeled part of the graph being disconnected from the labeled one, then this matrix will indeed be invertible.}, then
\[
 f_U^\ast = \left( D_U - W_{UU}\right)^{-1}W_{UL}f_L.
\]

\begin{remark}
 The function $f$ we constructed is called a harmonic extension \cite{zhu2003semi}. Indeed, it shares properties with harmonic functions in Euclidean space such as the mean value property and maximum principles; if $v_i$ is an unlabeled point then
 \[
  f(i) = \left[ D_U^{-1} \left( W_{UL}f_l + W_{UU}f_u \right) \right]_i = \frac1{\deg(i)}\sum_{j=1}^n w_{ij}f(j),
 \]
which immediately implies that the maximum and minimum value of $f$ needs to be attained at a labeled point.
\end{remark}

\subsubsection{An interesting experiment and the Sobolev Embedding Theorem}

Let us try a simple experiment. Let us say we have a grid on $[-1,1]^d$ dimensions (with say $m^d$ points for some large $m$) and we label the center as $+1$ and every node that is at distance larger or equal to $1$ to the center, as $-1$. We are interested in understanding how the above algorithm will label the remaining points, hoping that it will assign small numbers to points far away from the center (and close to the boundary of the labeled points) and large numbers to points close to the center.

\begin{figure}[h]
\begin{center}
\includegraphics[width=0.45\textwidth]{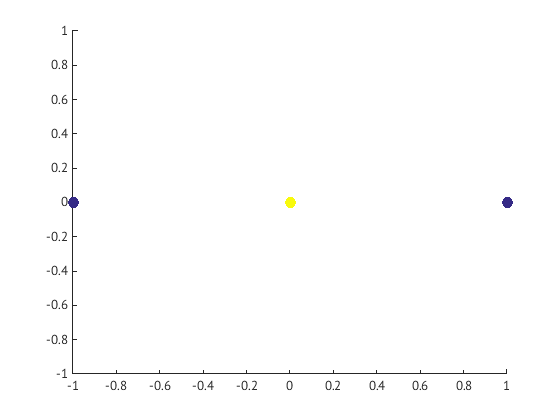}
\includegraphics[width=0.45\textwidth]{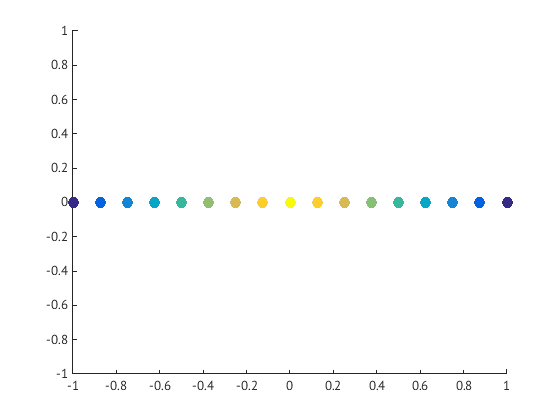}
\caption{The $d=1$ example of the use of this method to the example described above, the value of the nodes is given by color coding. For $d=1$ it appears to smoothly interpolate between the labeled points.}
\label{figure:2:SSL_1d_labels}
\end{center}
\end{figure}

\begin{figure}[h]
\begin{center}
\includegraphics[width=0.45\textwidth]{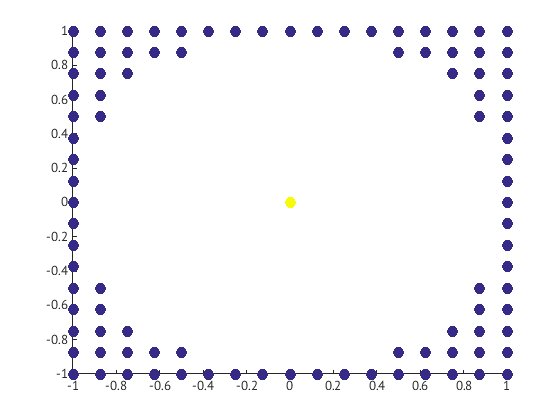}
\includegraphics[width=0.45\textwidth]{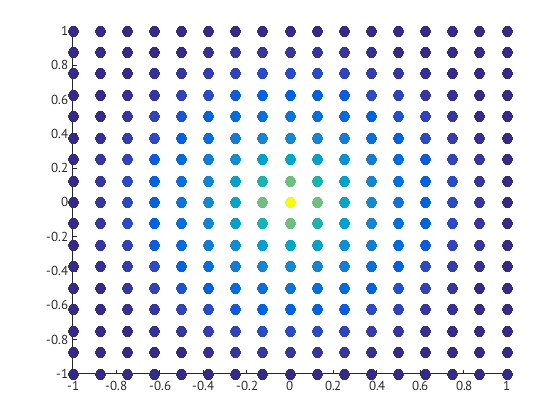}
\caption{The $d=2$ example of the use of this method to the example described above, the value of the nodes is given by color coding. For $d=2$ it appears to smoothly interpolate between the labeled points.}
\label{figure:2:SSL_2d_labels}
\end{center}
\end{figure}

\begin{figure}[h]
\begin{center}
\includegraphics[width=0.45\textwidth]{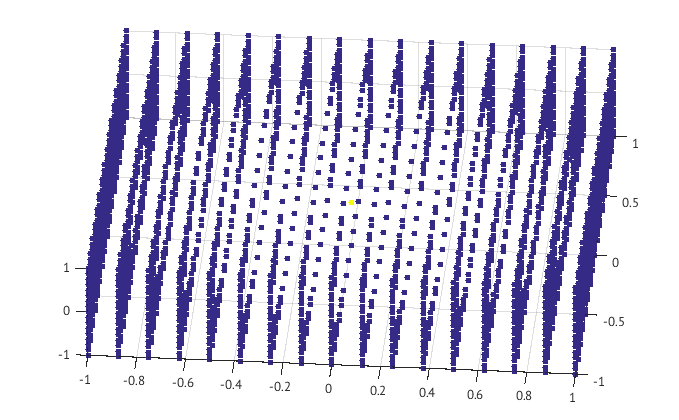}
\includegraphics[width=0.45\textwidth]{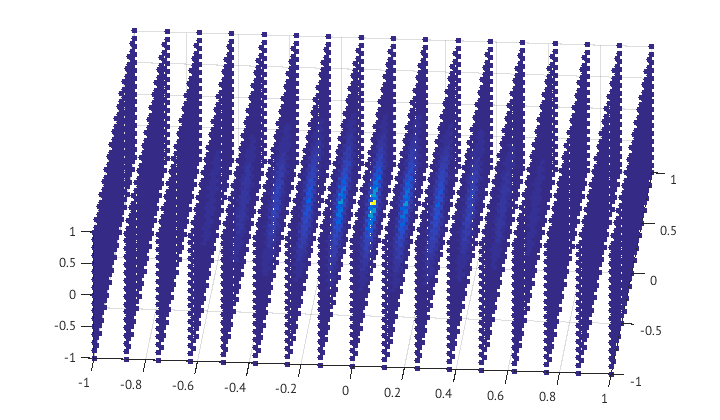}
\caption{The $d=3$ example of the use of this method to the example described above, the value of the nodes is given by color coding. For $d=3$ the solution appears to only learn the label $-1$.}
\label{figure:2:SSL_3d_labels}
\end{center}
\end{figure}

See the results for $d=1$ in Figure~\ref{figure:2:SSL_1d_labels}, $d=2$ in Figure~\ref{figure:2:SSL_2d_labels}, and $d=3$ in Figure~\ref{figure:2:SSL_3d_labels}. While for $d\leq 2$ it appears to be smoothly interpolating between the labels, for $d=3$ it seems that the method simply learns essentially $-1$ on all points, thus not being very meaningful. Let us turn to $\RR^d$ for intuition:

Let us say that we want to find a function in $\RR^d$ that takes the value $1$ at zero and $-1$ at the unit sphere, that minimizes $\int_{B_{0}(1)}\|\nabla f(x)\|^2 dx$. Let us consider the following function on $B_0(1)$ (the ball centered at $0$ with unit radius)
\[
 f_\eps(x) = \left\{  \begin{array}{cc}   1-2\frac{|x|}{\eps} & \text{if} |x|\leq \eps \\  -1 & \text{otherwise.}   \end{array} \right.
\]
A quick calculation suggest that
\[
 \int_{B_{0}(1)}\|\nabla f_\eps(x)\|^2 dx = \int_{B_{0}(\eps)}\frac1{\eps^2}dx = \vol(B_0(\eps)) \frac1{\eps^2}dx \approx \eps^{d-2},
\]
meaning that, if $d>2$, the performance of this function is improving as $\eps \to 0$, explaining the results in Figure~\ref{figure:2:SSL_3d_labels}.

\begin{figure}[h]
\begin{center}
\includegraphics[width=0.45\textwidth]{figures/Chapter5-SSL_3d_labels.png}
\includegraphics[width=0.45\textwidth]{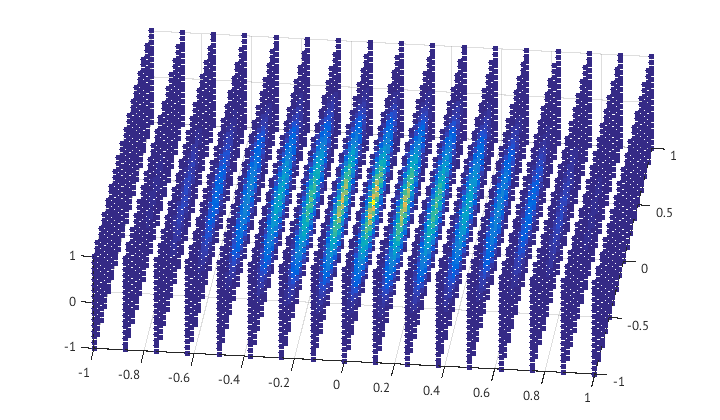}
\caption{The $d=3$ example of the use of this method with the extra regularization $f^TL^2f$ to the example described above, the value of the nodes is given by color coding. The extra regularization seems to fix the issue of discontinuities.}
\label{figure:2:SSL_3d_labelsm2}
\end{center}
\end{figure}

One way of thinking about what is going on is through the Sobolev Embedding Theorem. $H^m\left( \RR^d\right)$ is the space of function whose derivatives up to order $m$ are square-integrable in $\RR^d$, Sobolev Embedding Theorem says that if $m > \frac{d}2$ then, if $f\in H^m\left( \RR^d\right)$ then $f$ must be continuous and bounded (i.e., $f \in C_0(\RR^d)$), which would rule out the behavior observed in Figure~\ref{figure:2:SSL_3d_labels}. It also suggests that if we are able to control also second derivatives of $f$ then this phenomenon should disappear (since $2>\frac32$). While we will not describe it here in detail, there is, in fact, a way of doing this by minimizing not $f^TLf$ but $f^TL^2f$ instead, Figure~\ref{figure:2:SSL_3d_labelsm2} shows the outcome of the same experiment with the $f^TLf$ replaced by $f^TL^2f$ and confirms our intuition that the discontinuity issue should disappear (see, e.g.,~\cite{Nadler_SSL_sobolev_09} for more on this phenomenon).

\section{A brief primer on reversible Markov chains}\label{sec:MarkovChains}

With the material already described in this chapter one is able to very easily describe the basics of the theory of reversible Markov chains, so we will take the opportunity to do this (while pointing the interested reader to~\cite{levinperesmarkovchains2017} for a thorough treatment). We will focus here on Markov chains on finite state spaces\footnote{The theory relatively easily extends to discrete-step Markov chains on infinite state spaces, but also (with more effort) to continuous-time Markov processes, see Remark~\ref{remark:infiniteMarkovChains}.}. 

A Markov chain on $N$ states is a random walk on the state space, identified with $[N]$, whose transition probabilities depend only on the current state and not on the past (the so-called \emph{Markovian property}). Let us describe the transition probabilities by
\begin{equation}\label{eq:MarkovChainTransitionMatrix}
    M_{ij} = \PP\left( X(t+1)=j | X(t) = i\right).
\end{equation}

Let $\pi\in[0,1]^{N}$ (with $\pi^\top\1=1$) be a stationary measure of the Markov chain, meaning that $\pi^\top M = \pi^\top$.\footnote{An argument analogous to Proposition~\ref{prop:columnstochastichaseig1}, together with the Peron-Frobenius can show the existence of a stationary measure.} A Markov chain is said to be \emph{irreducible} if there is no proper subset of states where the probability mass can be fully trapped, in other words: A Markov chain is irreducible if for all $i,j\in[N]$, $\exists_{t>0}$ such that $\PP\left( X(t)=j | X(0) = i\right)>0$. One can show that irreducible Markov chains have unique stationary measures (and the stationary measure needs to be strictly positive).\footnote{There are a few different proofs of this fact, it also follows from Peron-Frobenius. Note that irreducibility alone does not imply that the Markov chain needs to converge to the stationary measure, for that aperiodicity is needed~\cite{levinperesmarkovchains2017}.}

In the rest of this section we will focus on Markov chains with a unique stationary $\pi$ that furthermore satisfies $\pi>0$ (we will identify the Markov chain with the transition matrix $M$ given in~\eqref{eq:MarkovChainTransitionMatrix}). Since $\pi$ is a stationary measure of $M$ we must have that, for each state $i$, the probability inflow at the: $\sum_{j\in[N]}\pi_jM_{ji}$ is the same as the probability outflow: $\sum_{j\in[N]}\pi_iM_{ij}$. Note that this is indeed equivalent to stationarity as $\sum_{j\in[N]}\pi_jM_{ji}=\left(\pi^\top M\right)_i$ and $\sum_{j\in[N]}\pi_iM_{ij} = \pi_i\sum_{j\in[N]}M_{ij}=\pi_i$. A natural stronger condition is to request that the inflow and outflow are the same per edge, not just total per state: for $\pi>0$ the stationary measure and for all $i,j\in[N]$,
\begin{equation}\label{eq:reversibleMC}
 \pi_jM_{ji} = \pi_i M_{ij}.
\end{equation}
Condition~\eqref{eq:reversibleMC} is known as the detailed balance condition. Markov chains satisfying detailed balance~\eqref{eq:reversibleMC} are called reversible Markov chains, they form a very important class of Markov chains. A reversible Markov chain can be thought of as a Markov chain that, at stationarity (or equilibrium), has the same dynamics when viewed running forward or backwards in time.

Reversibility (condition~\eqref{eq:reversibleMC}) can be viewed as $M$ being self-adjoint in the inner product $\langle \cdot, \cdot \rangle_\pi$ (given by $\langle x, y \rangle_\pi = \sum_{i=1}^N \pi_i x_i y_i$). Namely 
$$\langle x, My \rangle_\pi = \sum_{i=1}^N\sum_{j=1}^N \pi_i x_i M_{ij}y_j = \sum_{i=1}^N\sum_{j=1}^N \pi_j x_i M_{ji}y_j = \langle M^\top x, y \rangle_\pi.$$

Given $M$ a reversible Markov chain with a stationary measure $\pi>0$, let us define 
\begin{equation}\label{eq:WisdiagpiM}
    W = \diag(\pi)M,
\end{equation}
where $\diag(\pi)$ is the diagonal matrix with diagonal entries given by $\pi$. Reversibility is equivalent to $W$ being a symmetric matrix, since $W\geq 0$ it can be viewed as the weight matrix of a graph on $N$ nodes. This means that any reversible Markov chain with a strictly positive stationary measure can be viewed as the diffusion map from some weighted graph $G=([N],E,W)$ and thus the theory developed in this chapter (including connections with spectral clustering and Cheeger's Inequality) can be directly used to understand reversible Markov chains.

A very important concept in the study of Markov Chains is that of mixing time: how many iterations of the Markov Chain are needed, when starting from an initial distribution $\mu_0$, to reach a distribution that is at most $\eps$ away from $\pi$ in Total Variation (TV) distance. The TV distance between two probability distributions $p$ and $q$ (defined on the same probability space) is given by $\sup_E\|p(E)-q(E)\|$ where the supremum is taken over all measurable sets in the probability space. Since in our case the distributions are finite (they are distributions over [N]) we have
\[
\|\mu-\nu\|_{TV} =\frac12\sum_{i\in [N]}\left| \mu_i - \nu_i \right| = \frac12\|\mu-\nu\|_{1},
\]
where $\mu_i$ and $\nu_i$ are the probabilty of $i$ under, respectively, $\mu$ and $\nu$, and $\|\cdot\|_1$ denotes the $\ell_1$-norm in $\RR^N$.

\begin{definition}[Mixing time]
Given a Markov Chain with transition matrix $M\in\RR^{N\times N}$ and unique stationary measure $\pi\in\RR^N$ and $\eps>0$, we define the mixing time from a starting distribution $\mu^0$ as
\[
T_{mix}(M,\mu_0,\eps) \defeq \inf\left\{t\geq 0:\,\|\mu_0^\top M^t - \pi\|_{TV}\leq \eps\right\}.
\]
The worst-case mixing time is given by the mixing time when the Markov Chain is started from the worst state
\[
T_{mix}(M,\eps) \defeq \sup_{i\in[N]} T_{mix}(M,\delta_i,\eps),
\]
where $\delta_i$ is the delta distribution at state $i$; when viewed as a vector $\delta_i=e_i$.
\end{definition}

A core idea in the study of Markov Chains is to bound mixing times via functional inequalities that involve the spectrum of the transition matrix. A central object in this context is the {\em Dirichlet Form} which measures how a function varies under the transitions of the Markov Chain: given a function $f:[N]\to\RR$ (here viewed as a vector $f\in\RR^N$) the Dirichlet Form of $M$ is given by
\begin{equation}
\mathcal{E}_M(f) \defeq \frac12 \sum_{i,j=1}^N \pi_i f_i (I-M)_{ij}f_j.     
\end{equation}
The Markov Chain is said to satisfy a {\em Poincar\'{e} inequality} with constant $C_M$ if, for all $f\in\RR^{N}$,
\begin{equation}\label{eq:poincare-for-MC}
    \Var_{\pi}(f) \leq C_M \mathcal{E}_M(f),
\end{equation}
where $\Var_{\pi}(f)$ is the variance of $f$ under the measure $\pi$. In other words: $ \Var_{\pi}(f) = \sum_{i=1}^n \pi_i\left( f_i - \pi^\top f \right)^2$. Note that $\pi^\top f=\EE_\pi f$ corresponds to the mean of $f$ under the distribution $\pi$.

The best constant in \eqref{eq:poincare-for-MC} is known as the {\em spectral gap} of the Markov Chain:
\begin{equation}\label{eq:Gap-for-MC}
    \mathrm{gap}_M \defeq \inf_{\substack{f\in\RR^N \\ \Var_{\pi}(f)\neq 0  }}\frac{\mathcal{E}_M(f)}{\Var_{\pi}(f)}.
\end{equation}

Note that we have
\begin{equation}\label{eq:DirichletAndLaplacian}
    \mathcal{E}_M(f) = \frac12  f^\top \diag(\pi) (I-M) f = \frac12  f^\top (D_W-W) f = \frac12  f^\top L_W f,
\end{equation}
where $W$ is the weight matrix of a graph, as in~\eqref{eq:WisdiagpiM}, and $D_W=\diag(\pi)$ and $L_W=D_W-W$ are the associated degree and graph Laplacian matrices, respectively. Thus, $\Var_{\pi}(f)=0\Rightarrow \mathcal{E}_M(f)=0$ and $\mathrm{gap}_M$ in~\eqref{eq:Gap-for-MC} indeed corresponds to the best Poincar\'e constant in~\eqref{eq:poincare-for-MC}. Similarly, noting that $\pi^\top f = \1^\top D_W f$,
\begin{eqnarray}
    \Var_\pi(f) & = &  \sum_{i=1}^n \pi_i\left( f_i - \pi^\top f \right)^2 = \EE_\pi f^2 - \left( \EE_\pi f \right)^2  \nonumber\\
    & = & f^\top D_W f - \left(\1^\top D_W f\right)^2. \label{eq:VarAndDW} 
\end{eqnarray}

Recall that, by construction $\tr D_W = \sum_{i=1}^N \pi_i = 1$ and that we assume in this section that $\pi_i>0$ for all $i\in[N]$.

Recall from Chapter~\ref{c:graphs} that the normalized graph Laplacian is given by $\LLL_W = I - D_W^{-\frac12} W  D_W^{-\frac12}$ and its second smallest eigenvalue is given (as we saw in~\eqref{eq:3:Ncut:relaxed} and~\eqref{eq:3:Ncut:relaxed_LLL}) by
 $$\lambda_2(\LLL_W) = \inf_{\substack{y\in\RR^N\setminus\{0\} \\ \1^\top D_W y = 0}}   \frac{y^\top L_W y}{y^\top D_W y}.$$

Note that adding a multiple of $\1$ to $f\in\RR^N$ does not affect either $\mathcal{E}_M(f)$ or $\Var_\pi(f)$ (see~\eqref{eq:DirichletAndLaplacian}). This means we can write
\[
\mathrm{gap}_M = \inf_{\substack{f\in\RR^N\setminus \{0\}\\ \1^\top D_W f=0  }}\frac{\mathcal{E}_M(f)}{\Var_{\pi}(f)} = \inf_{\substack{f\in\RR^N\setminus \{0\}\\ \1^\top D_W f=0  }}\frac{f^\top L_W f}{f^\top D_W f} = \lambda_2(\LLL_W) = 1-\lambda_2(M),
\]
where $\lambda_2(M)$ is the second largest eigenvalue of $M$. The second inequality used~\eqref{eq:DirichletAndLaplacian} and~\eqref{eq:VarAndDW}. The fourth inequality follows from the fact that $M=\diag(\pi)^{-1}W$ (recall~\eqref{eq:WisdiagpiM}) is similar to $\diag(\pi)^{-\frac12}W\diag(\pi)^{-\frac12}$ and $\LLL_W = I - \diag(\pi)^{-\frac12}W\diag(\pi)^{-\frac12}$.

Similarly to the situation in Remark~\ref{remark:lazyDMaps} it is simpler to focus on lazy walks. We say a Markov Chain $M$ is lazy if $M_{ii}\geq \frac12$ for all $i\in[N]$ (this ensures that all eigenvalues of $M$ are non-negative). Given a transition matrix $M$ we can consider a lazy version of it by taking $M_{\text{lazy}}=\frac12M+\frac12I$. This does not affect reversibility and reduces the spectral gap by half.

We now show a central result in this area: a control on the spectral gap (or, equivalently on $\lambda_2(\LLL_W)$) provides a bound on the mixing time of a Markov Chain.

\begin{theorem}[Spectrum and mixing, see Theorem 12.4 in~\cite{levinperesmarkovchains2017}]\label{thm:spectralgapmixingtime}
 Let $M$ be a reversible Markov Chain with unique stationary measure $\pi$ (satisfying $\pi>0$). Let $\mathrm{gap}_M$ denote the spectral gap of $M$ (meaning that $M$ satisfies a Poincar\'e inequality with constant $\mathrm{gap}_M$). Let $M_{\text{lazy}}=\frac12+\frac12M$ be the lazy version of $M$. We have:
 \[
T_{mix}(M_{\text{lazy}},\eps) \leq \frac2{\mathrm{gap}_M}
\left(
\log\left(\frac1{\pi_{\min}}\right)+\log\left(\frac1{\eps}\right)
\right).
\]
If $M$ itself is a lazy Markov Chain then
 \[
T_{mix}(M,\eps) \leq \frac1{\mathrm{gap}_M}
\left(
\log\left(\frac1{\pi_{\min}}\right)+\log\left(\frac1{\eps}\right)
\right).
\]
\end{theorem}

\begin{proof}
Let $\varphi_1,\dots,\varphi_N$ be the eigenvectors of $M$ built in~\eqref{eq:MisPhiLambdaPsiT} (associated with eigenvalues $1=\lambda_1(M)>\cdots\geq\lambda_N(M)$). Recall that $\varphi_1=\1$ and that $\varphi_k^\top D_W\varphi_\ell=\delta_{k\ell}$ (in other words, they are orthonormal in the inner product $\langle\cdot,\cdot\rangle_\pi$). We can write (see~\eqref{eq:MisPhiLambdaPsiT}) $M = \Phi\Lambda\Psi^\top$ where $\Phi=[\varphi_1,\dots,\varphi_N]$, $\Lambda$ is a diagonal matrix with entries $\Lambda_ii= \lambda_i(M)$, and $\Psi = D_W\Phi$. Furthermore, for $t>0$ an integer, $M^t = \Phi\Lambda^t\Psi^\top=\sum_{k=1}^N\lambda_k^t(M)\varphi_k (D_W\varphi_k)^\top$ (see~\eqref{eq:Masrank1withphiandpsi} and recall that $D_W=\diag(\pi)$). 
Given a starting state $i\in[N]$ and respective starting measure $\delta_i$ we have
\begin{eqnarray*}
\delta_i^\top M^t &=& \delta_i^\top\sum_{k=1}^N\lambda_k^t(M)\varphi_k D_W\varphi_k^\top =  \delta_i^\top\left( \1\1^\top D_W + \sum_{k=2}^N\lambda_k^t(M)\varphi_k \varphi_k^\top D_W\right) \\
    &=& \pi^\top + \sum_{k=2}^N\lambda_k^t(M)\varphi_k(i) \varphi_k^\top D_W,
\end{eqnarray*}
where $\varphi_k(i)$ denotes the $i$-th entry of $\varphi_k$.

Thus, for every $j\in[N]$,
\begin{eqnarray}
    \left| \frac{\left(\delta_i^\top M^t\right)_j}{\pi_j} - 1 \right| & = & \left| \sum_{k=2}^N\lambda_k^t(M)\varphi_k(i) \varphi_k(j) \right| \leq \lambda_\ast^t   \sum_{k=2}^N\left|\varphi_k(i) \varphi_k(j) \right| \nonumber\\
    &\leq & \lambda_\ast^t \left( \sum_{k=2}^N  \varphi_k(i)^2\right)^{\frac12}
    \left( \sum_{k=2}^N  \varphi_k(j)^2\right)^{\frac12}, \label{eq:MCequation2}
\end{eqnarray}
where $\lambda_\ast = \max\left\{\lambda_2(M),|\lambda_n(M)|\right\}$ and the second inequality is an application of Cauchy-Schwarz.

Since $D_W^{\frac12}\Phi$ is an orthogonal matrix, it holds that
\(
I = D_W^{\frac12}\Phi\left(D_W^{\frac12}\Phi \right)^\top, 
\)
and thus, for every $i\in[N]$,
\begin{equation}\label{eq:MCequation3}
1 = I_{ii} = \sqrt{\pi_i} \left( \sum_{k=1}^N  \varphi_k(j)^2\right) \sqrt{\pi_i} = \pi_i \left( 1 + \sum_{k=2}^N  \varphi_k(j)^2 \right).
\end{equation}

Thus, combining~\eqref{eq:MCequation2} and~\eqref{eq:MCequation3}
\begin{equation}\label{eq:MCequation4}
\left| \frac{\left(\delta_i^\top M^t\right)_j}{\pi_j} - 1 \right| \leq \lambda_\ast^t \sqrt{\pi_i^{-1}-1}\sqrt{\pi_j^{-1}-1}\leq \frac{\lambda_\ast^t}{\pi_{\min}}.
\end{equation}

This means that
\begin{eqnarray*}\label{eq:MCequation5}
 \|\delta_iM^t-\pi\|_{TV} 
 &=& \frac12 \sum_{j=1}^N \left| \left(\delta_i^\top M^t\right)_j - \pi_j \right| 
 = \frac12 \sum_{j=1}^N \pi_j\left|  \frac{\left(\delta_i^\top M^t\right)_j}{\pi_j} - 1\right| \\
 &\leq & \frac12\max_j \left|  \frac{\left(\delta_i^\top M^t\right)_j}{\pi_j} - 1\right| \leq \frac12 \frac{\lambda_\ast^t}{\pi_{\min}}.
\end{eqnarray*}

For $M$ a lazy Markov Chain (meaning that $M_{ii}\geq \frac12$ for all $i\in[N]$) we have $\lambda_n\geq 0$ and so $\mathrm{gap}_M = 1-\lambda_\ast$. Thus
\begin{equation}
\|\delta_iM^t-\pi\|_{TV} \leq \frac12 \frac{(1-\mathrm{gap}_M)^t}{\pi_{\min}}\leq \frac12 \frac{\exp(-\mathrm{gap}_Mt)}{\pi_{\min}}.
\end{equation}
This gives
\[
T_{mix}(M,\eps) \leq \frac1{\mathrm{gap}_M}\log\left(\frac1{2\eps\pi_{\min}}\right)\leq \frac1{\mathrm{gap}_M}
\left(
\log\left(\frac1{\pi_{\min}}\right)+\log\left(\frac1{\eps}\right)
\right),
\]
the result for $M_{\text{lazy}}$ can be obtained by noting that $M_{\text{lazy}}$ is always a lazy Markov Chain and that $\mathrm{gap}_{M_{\text{lazy}}}=\frac12\mathrm{gap}_M$.
\end{proof}

Theorem~\ref{thm:spectralgapmixingtime} shows that mixing times are governed by a Fiedler eigenvalue $\lambda(\LLL_W)$ of the underlying weighted graph. Cheeger's inequality. Theorem~\ref{thm:cheegerinequality}, further shows that these are directly related to bottlenecks (small cuts) in the graph. These results together can be interpreted as saying that if there are no bottlenecks in the graph, the associated Markov Chain necessarily mixes fast.

It is generally possible to improve the dependency on $\pi_{\min}$ on the control of mixing times. A (now) standard way is through so-called Log-Sobolev inequalities. The idea is to develop analogues to Poincar\'e inequalities that control the entropy of a test function, rather than the variance. This tends to allow one to replace the $\log(\pi_{\min}^{-1})$ term by a $\log\log(\pi_{\min}^{-1})$ term. See, e.g.,~\cite{Diaconisetal-LogSobolev96}.

\begin{remark}\label{remark:infiniteMarkovChains}
The theory above adapts relatively easily to Markov Chains on infinite state spaces by considering the spectrum of an associated Markov transition kernel. Furthermore, one can also analogously develop a theory of continuous time Markov Processes (see, e.g.,~\cite{FukushimaOshimaTakeda2011}). In fact, the Gaussian Poincar\'e inequality (Proposition~\ref{prop:GaussianPoincare}) plays the same role as~\eqref{eq:poincare-for-MC} for a certain Markov Process (see Remark~\ref{remark:Ornstein-Uhlenbeck}, and~\cite{vanHandel_LectureNotesProb_14} (Section 2)).
\end{remark}

As we will see below, many Markov Chains of interest have exponential state spaces (such as $\{\pm1\}^n$, meaning that $N=2^n$) and the goal is to have polynomial in $n$ mixing times.

\subsection{A word about Markov Chain Monte Carlo methods}

An important use of Markov Chains is to sample from high dimensional distributions that are easy to access up to normalization. This strategy usually goes by the name of Markov Chain Monte Carlo~\cite{hastings1970monte,robert1999monte}. These are central objects in both Statistical Physics~\cite{krauth2006statistical} and high dimensional Bayesian Statistics~\cite{gelman1995bayesian}.

Let us focus on a (rich) class of examples for the sake of exposition: The goal is to sample from a probability distribution over the hypercube $\{\pm1\}^n$ (meaning that $N=2^n$), which $\pm1$ entry is usually called \emph{a spin}. Let $J\in\RR^{n\times n}$ be a symmetric matrix. Given $\beta\geq 0$ we define the following probability distribution over the hypercube, known as a Gibbs measure\footnote{In Statistical Physics $\beta$ corresponds to the inverse temperature ($\frac{1}{T}$) of the system, governing a trade-off between energy and entropy.}:
\begin{equation}\label{eq:gibbsmeasurexTJx}
\pi_x = \frac{1}{Z_n}\exp{\left(\beta x^TJx\right)},
\end{equation}
for $x\in\{\pm1\}^n$, where $Z_n = \sum_{x\in\{\pm1\}^n}\exp{\left(\beta x^TJx\right)}$ is known as a the normalization constant, or partition function.

Distributions of the form of~\eqref{eq:gibbsmeasurexTJx} are central in both Statistical Physics and Bayesian Statistics. In Statistical Physics, by taking different choices of $J$, it includes very important models such as Curie-Weiss, Ising, Sherrington-Kirkpatrick, and Edward-Anderson models. It is also central in Statistics. In fact the posterior distribution of the node memberships given the graph edges in the Stochastic Block Model (in two communities) of Chapter~\ref{c:community} is naturally written in the form~\eqref{eq:gibbsmeasurexTJx}, see~\cite{Abbe_SBM_survey}.

The goal will be to construct a Markov Chain whose unique stationary measure is the target distribution. A key property of these measures is that while they are hard to compute (because computing $Z_n$ requires summing over exponentially many summands) they are easy to access up to normalization, meaning that it is easy to compute relative probabilities, such as $\pi_x/\pi_y$ for two states $x$ and $y$.

Motivated by the example above let us show the construction of a reversible Markov Chain to sample from a (positive) probability distribution $\pi$ in $[N]$. We start by building a graph on $[N]$, and for the sake of exposition, let us say this is a simple unweighted $d$-regular graph with adjacency $W$. Each step of the Markov Chain proceeds as follows: From state $x\in[N]$, the chain picks a neighbor uniformly at random, let us call that neighbor $y\in[N]$; then the Markov Chain moves to $y\in[N]$ with probability $\frac{\pi_y}{\pi_x+\pi_y}$ and stays in $x$ with probability $\frac{\pi_x}{\pi_x+\pi_y}$ (note that while computing $\pi_x$ is difficult for, e.g.~\eqref{eq:gibbsmeasurexTJx}, computing these update probabilities is easy).  The reader can check that this Markov Chain is reversible and that $\pi$ is indeed a stationary measure. This essentially corresponds to the celebrated {\em Metropolis algorithm} (see, e.g.~\cite{Newman-Barkema-MCMC99}).\footnote{The choice of neighboring node to potentially move to does need to be uniformly random, in which case the probability of \emph{accepting} or \emph{rejecting} the move depends on those probabilities, as to make the chain reversible on $\pi$ (and so it is no longer given by $\frac{\pi_y}{\pi_x+\pi_y}$ and $\frac{\pi_y}{\pi_x+\pi_y}$).}

Returning to the hypercube and the target distribution~\eqref{eq:gibbsmeasurexTJx}, a classical Markov Chain is obtained by taking the contribution above for the graph whose nodes are elements of $\{\pm1\}^n$ and two nodes are connected if they differ in only one entry (often called a spin). In this case $N=2^n$ and $d=n$. This gives rise to the celebrated Glauber Dynamics (see~\cite{levinperesmarkovchains2017}).\footnote{This can be viewed as what is called in Physics a ``Heat Bath'' where the system is resampled conditioned on all but one spin.} In many situations this process allows one to sample distributions of the form of~\eqref{eq:gibbsmeasurexTJx} in time that is polynomial in $n$ (and so logarithmic in $N$). There has been exciting recent progress on understanding mixing times of these (and other related) chains~\cite{AnariLiuOveisGharanFOCS2020,ChenEldan2025LocalizationSchemes}.

\section*{Exercises}
\addcontentsline{toc}{section}{Exercises}

\begin{myexercise}[\level\sep Random walks: eigenvalues]\label{prob:random_walks_evals}
    Let $G = (V, E, W)$ be an undirected weighted graph.
    \begin{enumerate}[(a)]
        \item Suppose that $A, B \in \R^{n\times n}$ are similar matrices, which means that there exists an invertible matrix $P\in \R^{n\times n}$ such that $B = P^{-1} A P$. Prove that they have the same eigenvalues, with equal geometric multiplicities.
        
        \item Show that the transition probability matrix $M \coloneqq D^{-1}W$ is similar to the matrix $S \coloneqq D^{-1/2}W D^{-1/2}$.
        
        \item Deduce that all eigenvalues of $M$ are real.

        \item Prove that every eigenvalue of $M$ belongs to the interval $[-1,1]$, and show that $1$ is an eigenvalue.
    \end{enumerate}
\end{myexercise}

\begin{myexercise}[\level\level\sep Random walks: connectivity]\label{prob:random_walks_connectivity}
    Let $G = (V, E, W)$ be an undirected weighted graph. Show that the largest eigenvalue of $M = D^{-1}W$ has multiplicity one if and only if the graph is connected. Here, two vertices are connected if and only if there exists a path from one to another along which all edges have positive weights.
\end{myexercise}

\begin{myexercise}[\level\level\sep Equilibrium distribution]\label{prob:equilibrium_distribution}
    Let $G = (V, E, W)$ be an undirected weighted graph, and $X$ a random walk (with independent set) associated to it, the one having transition probabilities
    \begin{equation*}
        \P\brap{X(t+1) = j \mid X(t) = i} = \frac{w_{ij}}{\deg(i)}.
    \end{equation*}
    Suppose that there is an equilibrium distribution $\pi$ on $V$, which means that starting from any $i \in V$,
    \begin{equation*}
        \P\brap{X(t) = j \mid X(0) = i} \to \pi_j \text{ as }t \to \infty
    \end{equation*}
    holds for any $j \in V$. Prove that $\lambda_2(M) < 1$.
\end{myexercise}
\begin{hint}
   This means that for any $i_1, i_2, j \in V$, we have $\P\brap{X(t) = j \mid X(0) = i_1} - \P\brap{X(t) = j \mid X(0) = i_2} \to 0$.
\end{hint}

\begin{myexercise}[\level\level\sep Truncated diffusion map of a cycle graph]\label{prob:truncated_diffusion_map_cycle}
    Let $n \geq 2$ and $\omega = e^{-2\pi i / n} $ be the primitive $n$-th root of unity. In this problem, we index rows and columns of $n \times n$ matrices from $0$ to $n-1$, rather than $1$ to $n$.
    \begin{enumerate}[(a)]
        \item The \emph{Discrete Fourier Transform} (DFT) matrix   
        is an $n \times n$ matrix $F = (F_{j,k})_{j,k=0}^{n-1}$ with
        \begin{equation*}
            F_{j,k} \coloneqq \frac{1}{\sqrt{n}}\omega^{jk}.
        \end{equation*}
        Prove that $F$ is a unitary matrix, meaning that $F^*F = FF^* = I_n$. 
        
        \item A \emph{Circulant} matrix is an $n \times n$ matrix $C = (C_{j,k})_{j,k=0}^{n-1}$ with $C_{j,k} = c_{j-k \text{ mod }n}$, where $c_0, \ldots, c_{n-1}$ are $n$ real numbers:
        \begin{equation*}
            C = 
            \begin{pmatrix}
                c_0 & c_{n-1} & c_{n-2} & \cdots & c_{1}\\
                c_{1} & c_0 & c_{n-1} & \cdots & c_{2}\\
                c_{2} & c_{1} & c_0 & \cdots & c_{3}\\
                \vdots & \vdots & \vdots & \ddots & \vdots \\
                c_{n-1} & c_{n-2} & c_{n-3} & \cdots & c_0
            \end{pmatrix},
        \end{equation*}

        Prove that $F$ diagonalizes $C$, meaning that $FCF^*$ is a diagonal matrix.
        
        \item Deduce that the eigenvalues of $C$ are given by
        \begin{equation*}
            \lambda_j = \sum_{k=0}^{n-1} c_k \omega^{kj} = c_0 + c_1 \omega^{j} + \dots + c_{n-1} \omega^{(n-1)j}, \quad j = 0, 1, \ldots, n-1.
        \end{equation*}
        Note that the eigenvalues $\lambda_0, \ldots, \lambda_{n-1}$ are not necessarily ordered.

        \item Let $C_n$ be a cycle graph of length $n$, where $n \geq 3$. Find the diffusion map truncated to $2$ dimensions.
    \end{enumerate}
\end{myexercise}

\begin{myexercise}[\level\level\sep Diffusion map of a complete graph]\label{prob:diffusion_map_complete_graph}
    Let $K_n$ be complete graph on $n$ nodes, where $n \geq 3$. Find the diffusion map. (Since there are more bases of real-valued eigenvectors, it is sufficient to pick one.)
\end{myexercise}

\begin{myexercise}[\level\level\sep Diffusion map of a lazy walk]\label{prob:diffusion_map_lazy_walk}
    Let $G = (V, E, W)$ be an undirected weighted graph with the associated transition probability matrix $M$. Let $\varphi_t \colon V \to \R^{n-1}$ be the diffusion map made from $M$. Now let $M' = \frac{1}{2}(M + I)$ be the transition probability matrix of the associated lazy walk, and $\varphi'_t \colon V \to \R^{n-1}$ be the diffusion map made from $M'$. Prove that for any $t \geq 1$:
    \begin{equation*}
        \varphi'_t = 2^{-t}\sum_{u=0}^t {t \choose u} \varphi_u.
    \end{equation*}
\end{myexercise}

\begin{myexercise}[\level\level\level\sep Aperiodicity of a lazy walk]\label{prob:aperiodicity_lazy_walk}
    Let $G = (V, E, W)$ be an undirected weighted graph.
    \begin{enumerate}[(a)]
        \item Let $M' = \frac{1}{2}(M + I)$ be the transition probability matrix of the associated lazy walk. Show that $M'$ is not necessarily symmetric, but it has $n$ non-negative eigenvalues.

        \item Prove that the lazy random walk is aperiodic, which means that for any vertex $i \in V$ there is no (period) integer $k > 1$ such that for any (time) $t \geq 1$:
        \begin{equation*}
            \brap{(M')^t}_{ii} > 0 \implies k \mid t.
        \end{equation*}
        
        \item Suppose that $W$ is an irreducible matrix. Prove that there exists $T \in \N$ such that for every $t \geq T$, all entries of $(M')^t$ are positive. (This means that the associated Markov Chain is regular.)
    \end{enumerate}
\end{myexercise}

\begin{myexercise}
Show that the mean first passage time $t_{ij} = \mathbb{E}[\tau_{ij}]$ satisfies that triangle inequality.
\end{myexercise} 

\begin{myexercise}
Let $L = D - W$ be the Laplacian of a weighted undirected connected graph. Show that the pseudo-inverse $L^\dagger$ is given by
$$L^\dagger = (L + \frac{1}{n}{\textbf 1}{\textbf 1}^T)^{-1} - \frac{1}{n} {\textbf 1}{\textbf 1}^T.$$
\end{myexercise} 

\begin{myexercise}
Is it necessarily the case that all entries of $L^\dagger$ are non-negative, i.e., that $L^\dagger_{ij} \geq 0$ for $i,j=1,\ldots,n$? 
\end{myexercise}

\begin{myexercise}[\level\level\sep Hitting times and semi-supervised learning]\label{prob:hitting_times_semi_supervised}
    Let $G = (V,E,W)$ be an undirected, connected graph with non-negative weights $w_{ij}$. The vertex set is partitiond as $V = V_{+} \cup V_{\scriptstyle -} \cup V^{\star}$, where $V_{+}$ are labeled as 1, $V_{\scriptstyle -}$ are labeled as 0 and $V^{\star}$ are unlabeled. Suppose that every unlabelled vertex ($V_{\star}$) is connected to at least one labelled vertex ($V_{+} \cup V_{\scriptstyle -}$) by an edge.
    
    To predict the label of the unlabeled vertices, you wish to find a function $f^{\star}: V \to \mathbb{R}$ which agrees on the labeled vertices and predicts the values of the unlabeled vertices as values in $\mathbb{R}$ as smoothly as possible:
    \begin{equation}\label{eq:optproblem}
        f^\star := \arg\,\min_{\substack{f: V \to \mathbb{R}: \\ f(i) = 1, i \in V^+ \\ f(i) = 0, i \in V^-}} \sum_{i < j} w_{ij} (f(i) - f(j))^2.
    \end{equation}
    
    Now, consider a random walk $X$ on $V$ given by the following transition probabilities:
    \begin{equation*}
        \P\brap{X(t+1)=j\mid X(t)=i} = \frac{w_{ij}}{\deg(i)}.
    \end{equation*}
    Given a node $i \in V$, let $g(i)$ be the probability that a random walker starting at $i$ reaches a node in $V_{+}$ before reaching one in $V_{\scriptstyle -}$. I.e. if $T_{+} = \inf\brac{t \geq 0 \colon X(t) \in V_{+}}$ and $T_{\scriptstyle -} = \inf\brac{t \geq 0 \colon X(t) \in V_{\scriptstyle -}}$, then
    \begin{equation*}
        g(i) \coloneqq \P\brap{T_{\scriptstyle +} < T_{\scriptstyle -} \mid X(0)=i}.
    \end{equation*}
    \begin{enumerate}[(a)]
        \item Show that for any $i \in V$ and $t \geq 0$, $\sum_{j \in V} \P\brap{X(t+1)=j\mid X(t)=i} = 1$, so that $X$ is a random walk.
        \item Show that $g$ satisfies constraints of the optimization problem: $g(i) = 1$ for $i \in V_{+}$, and $g(i) = 0$ for $i \in V_{\scriptstyle -}$.
        \item Prove that $g$ satisfies the following equality for any unlabelled $i \in V^\star$:
        \begin{equation}\label{eq:gconditions}
            g(i) = \frac{1}{\deg(i)} \sum_{j\in V_{+}} w_{ij} +  \frac{1}{\deg(i)} \sum_{j\in V^{\star}} w_{ij} g(j).
        \end{equation}
        \item By analyzing first-order optimality conditions (it is enough to state the formula from the notes, without proving) of the optimization problem \eqref{eq:optproblem}, show that $f^\star$ also satisfies \eqref{eq:gconditions}, and conclude that $f^\star = g$.
    \end{enumerate}
\end{myexercise}

\begin{myexercise}
Suppose we want to denoise the function $f \in L^2(\Omega)$ where $\Omega \subset \mathbb{R}^p$ by solving the following regularization problem:
$$\min_g \|f-g\|^2_{L^2(\Omega)} + \mu\int_\Omega g(x) \Delta^k g(x) \,dx$$
where $\Delta^k$ is the $k$'th iterated Laplacian. Find $g$ and determine to what Sobolev space it belongs. Consider the similar graph regularization problem
$$\min_{g:V\to \mathbb{R}} (f-g)^T D (f-g) + \mu g^T L^k g,$$
where $L^k$ is the $k$'th iterated combinatorial Laplacian $L^k=(D-W)^k$.\\
\begin{hint}
Express $g$ in terms of the eigenfunctions/eigenvectors of a suitable operator.    
\end{hint}
\end{myexercise}
 
\begin{myexercise}
Show that $f \in H^1(\mathbb{R}^2)$ does not imply $f\in C_0(\mathbb{R}^2)$ by constructing a sequence of functions $\{f_n\}_{n=1}^\infty$ in $H^1(\mathbb{R}^2)$ satisfying $$\lim_{n\to \infty} \|f_n\|_{H^1(\mathbb{R}^2)}=0,$$ but $\lim_{n\to \infty} f_n \notin C_0(\mathbb{R}^2)$
\begin{hint}
The $\log$ function.    
\end{hint}    
\end{myexercise}

\begin{myexercise}
Derive the solution to the extension problem with $l$ labeled points and $u=n-l$ unlabeled points using the discrete Sobolev norm $H^m(V)$ (with $m\geq 1$) as the cost function. That is, find $f$ that solves the minimization problem
\begin{eqnarray*}
\min_{f:V \to \mathbb{R}} f^T (D+L^m) f,\\
\mbox{s.t. } f(i)=f_i,\quad \mbox{for } i=1,\ldots,l.
\end{eqnarray*}
\end{myexercise}

\begin{myexercise}
Derive the solution to the `noisy' extension problem, where the $l$ labeled points may be contaminated by measurement errors. Specifically, find the solution to the minimization problem
\begin{eqnarray*}
\min_{g:V \to \mathbb{R}} \sum_{i=1}^l d_i (f(i)-g(i))^2 + \mu g^T (D+L^m) g\\
\end{eqnarray*}
\end{myexercise}

\begin{myexercise}
Use a computer simulation (e.g., using Matlab or Python) to sample independently $n$ unlabeled points $x_1,x_2,\ldots,x_n$ in $\mathbb{R}^2$ from a Gaussian mixture model with two isotropic independent Gaussians $X_1$ and $X_2$ centered at $\mu_1=(-1,-1)$ and $\mu_2=(1,1)$. That is, let $$X_1\sim \mathcal{N}(\mu_1,\sigma^2I)\quad \mbox{and } \quad X_2 \sim \mathcal{N}(\mu_2,\sigma^2I),$$ where $I = \left(
                                                                                                                 \begin{array}{cc}
                                                                                                                   1 & 0 \\
                                                                                                                   0 & 1 \\
                                                                                                                 \end{array}
                                                                                                               \right)$ and define the random vector $X$ from which $x_1,\ldots,x_n$ are drawn as $$X=ZX_1+(1-Z)X_2,$$ where $Z$ is a Bernoulli random variable independent of $X_1$ and $X_2$:
    $$Z=\left\{\begin{array}{cc}
         1 & \mbox{with probability } \frac{1}{2}, \\
         &\\
         0 & \mbox{with probability } \frac{1}{2}.
       \end{array}\right.
    $$
Add the two centers as labeled points. That is, define the label of $\mu_1$ as ``RED'' and the label of $\mu_2$ as ``BLUE".
\begin{enumerate}
\item Use the harmonic extension algorithm to label the $n$ unlabeled points (use $w_{ij}=e^{-\|x_i-x_j\|^2_{\mathbb{R}^2}/\varepsilon}$ as your similarity measure). Try different values of $n$, $\sigma$ and $\varepsilon$, especially notice the behavior of the algorithm for large $n$.
\item Replace the cost function $f^T L f$ by $f^T(D+L^2)f$ corresponding to the Sobolev norm $\|f\|^2_{H^2(\mathbb{R}^2)}$. Compare the classification errors of the two methods. Explain your results in light of the Sobolev embedding theorem.
\end{enumerate}

\end{myexercise}


\chapter{Linear Dimension Reduction via Random Projections}
\label{c:johnson}

In Chapters~\ref{c:svd} and~\ref{c:diffusion} we saw both linear and non-linear methods for dimension reduction.
In this chapter we will see one of the most fascinating consequences of the phenomenon of concentration of measure in high dimensions, one of the \emph{blessings of high dimensions} described in Chapter~\ref{c:surprises}. When given a data set in high dimensions, we will see  that the projection to a lower dimensional space, taken at random, can preserve (under conditions made precise in the next section) certain geometric features of the data set. The remarkable aspect here is that this ``lower'' dimension can be strikingly lower. This gives rise to significant computational savings in many data processing tasks by including a random projection as a pre-processing step. Similar ideas are also the driving force behind many algorithms in {\em randomized linear algebra}, as we will demonstrate in this chapter via the {\em randomized SVD}.
There is however another less obvious implication of this phenomenon with important practical implications: since the projection is agnostic of the data, it can be leveraged even when the data set is not explicit, such as the set of all \emph{natural images} or the set of all ``possible'' \emph{brain scans}; this is at the heart of \emph{Compressive Sensing}, which we will explore in Chapter~\ref{c:cs}

\section{The Johnson-Lindenstrauss Lemma}\label{s:jl}

Suppose one has $n$ points, $X = \{x_1,\dots,x_n\}$, in $\RR^p$ (with $p$ large). If $p>n$, the points actually lie in a subspace of dimension $n$, so the projection $f:\RR^p\to \RR^n$ of the points to that subspace acts without distorting the geometry of $X$. In particular, for every $x_i$ and $x_j$, $\|f(x_i)-f(x_j)\|^2 = \|x_i-x_j\|^2$, meaning that $f$ is an isometry in $X$.
Suppose instead we allow a bit of distortion, and look for a map $f:\RR^p\to \RR^d$ that is an $\epsilon-$isometry\footnote{Another possible convention is to ask for $(1-\epsilon) \|x_i-x_j\| \leq \|f(x_i)-f(x_j)\| \leq (1+\epsilon) \|x_i-x_j\|$; this is equivalent up to a multiplicative constant on $\epsilon$ since for $0\leq\epsilon\leq 1$ we have $1+\epsilon\leq (1+\epsilon)^2\leq 1+3\epsilon$ and $1-2\epsilon\leq (1-\epsilon)^2\leq 1-\epsilon$.}, meaning that
\begin{equation}\label{def_epsilonisometry}
(1-\epsilon) \|x_i-x_j\|^2 \leq \|f(x_i)-f(x_j)\|^2 \leq (1+\epsilon) \|x_i-x_j\|^2.
\end{equation}
Can we do better than $d=n$?

In 1984, Johnson and Lindenstrauss~\cite{Johnson_Lindenstrauss_84} showed a remarkable lemma that answers this question affirmatively.

\begin{theorem}[Johnson-Lindenstrauss Lemma~\cite{Johnson_Lindenstrauss_84}]\label{JL_random_0}
 For any $0<\epsilon<1$ and for any integer $n$, let $d$ be such that
 \begin{equation}\label{kdim}
 d \geq 4\frac1{\epsilon^2/2 - \epsilon^3/3}\log n.
\end{equation}
Then, for any set $X$ of $n$ points in $\RR^p$, there is a linear map $f:\RR^p\to\RR^d$ that is an $\epsilon-$isometry for $X$ (see (\ref{def_epsilonisometry})).
This map can be found in randomized polynomial time\footnote{Randomized polynomial time refers to the complexity class of decision problems for which a randomized algorithm can solve the problem in polynomial time with a high probability of correctness.}.
\end{theorem}

\begin{proof}
We follow~\cite{Dasgupta_Gupta_JLsimpleproof} for an elementary proof for Theorem~\ref{JL_random_0}. 
We will start by showing that, given a pair $x_i,x_j$ a projection onto a random subspace of dimension $d$ will satisfy (after appropriate scaling) property (\ref{def_epsilonisometry}) with high probability. Without loss of generality we can assume that $u = x_i - x_j$ has unit norm. Understanding what is the norm of the projection of $u$ on a random subspace of dimension $d$ is the same as understanding the norm of the projection of a (uniformly) random point on $S^{p-1}$ the unit sphere in $\RR^p$ on a specific $d$-dimensional subspace---let us say the one generated by the first $d$ canonical basis vectors.

This means that we are interested in the distribution of the norm of the first $d$ entries of a random vector drawn from the uniform distribution over $S^{p-1}$ -- this distribution is the same as taking a standard Gaussian vector in $\RR^p$ and normalizing it to the unit sphere.

Let $g:\RR^p\to\RR^d$ be the projection on a random $d-$dimensional subspace and let $f:\RR^p\to\RR^d$ defined as $f = \sqrt{\frac{p}d}g$. Then (by the above discussion), given a pair of distinct points $x_i$ and $x_j$, $\frac{\|f(x_i)-f(x_j)\|^2}{\|x_i-x_j\|^2}$ has the same distribution as $\frac{p}dL$, as defined in Lemma~\ref{lemma_COM}. Using Lemma~\ref{lemma_COM}, we have, given a pair $x_i,x_j$,

\[
 \Pr\left[ \frac{\|f(x_i)-f(x_j)\|^2}{\|x_i-x_j\|^2} \leq (1-\epsilon) \right] \leq \exp\left( \frac{d}2(1-(1-\epsilon) +\log(1-\epsilon)) \right),
\]
since for $\epsilon\geq 0$, $\log(1-\epsilon) \leq -\epsilon -\epsilon^2/2$, using~\eqref{kdim} we have
\begin{eqnarray*}
 \Pr\left[ \frac{\|f(x_i)-f(x_j)\|^2}{\|x_i-x_j\|^2} \leq (1-\epsilon) \right] & \leq & \exp\left( -\frac{d\epsilon^2}4 \right) \\
      & \leq & \exp\left( -2\log n \right) = \frac1{n^2}.
\end{eqnarray*}
On the other hand,
\[
 \Pr\left[ \frac{\|f(x_i)-f(x_j)\|^2}{\|x_i-x_j\|^2} \geq (1+\epsilon) \right] \leq \exp\left( \frac{d}2(1-(1+\epsilon) +\log(1+\epsilon)) \right).
\]
since for $\epsilon\geq 0$, $\log(1+\epsilon) \leq \epsilon -\epsilon^2/2 + \epsilon^3/3$, using~\eqref{kdim} we have
\begin{eqnarray*}
 \Prob\left[ \frac{\|f(x_i)-f(x_j)\|^2}{\|x_i-x_j\|^2} \geq (1+\epsilon) \right] & \leq & \exp\left( -\frac{d\left(\epsilon^2 - 2\epsilon^3/3\right)}4 \right) \\
      & \leq & \exp\left( -2\log n \right) = \frac1{n^2}.
\end{eqnarray*}
By the union bound it follows that 
\[
 \Pr\left[ \frac{\|f(x_i)-f(x_j)\|^2}{\|x_i-x_j\|^2} \notin [1-\epsilon,1+\epsilon] \right] \leq \frac2{n^2}.
\]
Since there exist ${n \choose 2}$ such pairs, again, a simple union bound gives
\[
 \Pr\left[ \exists_{i,j}:\frac{\|f(x_i)-f(x_j)\|^2}{\|x_i-x_j\|^2} \notin [1-\epsilon,1+\epsilon] \right] \leq \frac2{n^2} \frac{n(n-1)}{2} = 1-\frac1n.
\]
Therefore, choosing $f$ as a properly scaled projection onto a random $d$-dimensional subspace gives an $\epsilon$-isometry on $X$ (see (\ref{def_epsilonisometry})) with probability at least $\frac1n$. We can achieve any desirable constant probability of success by trying $\OOO(n)$ such random projections, meaning we can find an $\epsilon-$isometry in randomized polynomial time.

\end{proof}

Note that by considering $d$ slightly larger one can get a good projection on the first random attempt with  high confidence. In fact, it is trivial to adapt the proof above to obtain the following lemma:

\begin{proposition}\label{JL_random}
 For any $0<\epsilon<1$, $\tau>0$, and for any integer $n$, let $d$ be such that
\[
 d \geq \frac{2(2+\tau)}{\epsilon^2/2 - \epsilon^3/3}\log n.
\]
Then, for any set $X$ of $n$ points in $\RR^p$, take $f:\RR^p\to\RR^d$ to be a suitably scaled projection on a random subspace of dimension $d$, then $f$ is an $\epsilon-$isometry for $X$ (see (\ref{def_epsilonisometry})) with probability at least $1-\frac1{n^{\tau}}$.
\end{proposition}

Proposition \ref{JL_random} is quite remarkable. Consider the situation where we are given a high-dimensional data set in a streaming fashion -- meaning that we get each data point at a time, consecutively. To run a dimension-reduction technique like PCA or Diffusion maps we would need to wait until we received the last data point and then compute the dimension reduction map (both PCA and Diffusion Maps are, in some sense, data adaptive). Using Lemma \ref{JL_random} one can just choose a projection at random in the beginning of the process (all one needs to know is an estimate of the logarithm of the size of the data set) and just map each point using this projection matrix which  can be done online -- we do not need to see the next point to compute the projection of the current data point. Proposition~\ref{JL_random} ensures that this (seemingly na\"ive) procedure will, with high probably, not distort the data by more than $\epsilon$.

One might wonder if such low-dimensional embeddings as provided by the Johson-Lindenstrauss Lemma also extend to other norms besides the Euclidean norm.  For the $\ell_1$-norm there exist lower bounds which prevent such a dramatic dimension reduction (see~\cite{lee2004embedding}), and for the $\ell_{\infty}$-norm one can easily construct examples that demonstrate the impossibility of dimension reduction. Hence, the Johnson-Lindenstrauss Lemma seems to be a subtle result about the Euclidean norm.

\subsection{The Fast Johnson-Lindenstrauss transform}

Let us continue thinking about our example of high-dimensional streaming data. After we draw the random projection matrix\footnote{An orthogonal projection $P$ must satisfy $P=P^{\ast}$ and $P^2=P$. Here, it is not $M$ that represents a projection, but $M^{\ast} M$, yet for our purposes of approximate norm-preserving dimension reduction it suffices to apply $M$ instead of $M^{\ast} M$.  However, with a slight abuse of terminology, we still refer to $M$ as projection.}, say $M$, for each data point $x$, we still have to compute $Mx$ which has a computational cost of $\OOO(\epsilon^{-2}\log(n)p)$  since $M$ has $\OOO(\epsilon^{-2}\log(n)p)$ entries (since $M$ is a random matrix, generically it will be a dense matrix).  In some applications this might be too expensive, raising the natural question of whether one can do better. 
Moreover, storing a large-scale dense matrix $M$  is not very desirable either.
There is no hope of significantly reducing the number of rows in general, as it is known that  the Johnson-Lindenstrauss Lemma is orderwise optimal~\cite{Alon_JL_lowerbound_2003,Larsen_Nelson_JL}.

We might hope to replace the dense random matrix $M$ by a sparse matrix $S$ to speed up the matrix-vector multiplication and to  reduce the storage requirements.  This method was proposed and analyzed in~\cite{Ailon_Chazelle_fast_JL}. Here we discuss a slightly simplified version,  see also~\cite{cohen2015optimal}.

We let $S$ be a very sparse $d \times p$ matrix, where each row of $S$ has just one single non-zero entry of value
$\sqrt{p/d}$ at a location drawn uniformly at random. Then, for any vector $x \in \R^p$ 
\begin{equation*}
  \underset{i}{\E}
   [(Sx)_i^2]=\sum\limits_{j=1}^p \P(S_{ij}\neq 0) \cdot \frac{p}{d} \cdot x^2_j = \frac{1}{d} \|x\|^2,
\end{equation*}
hence $\E [\|Sx\|^2] = \E [\sum\limits^d_{i = 1} (Sx_i)^2]=\|x\|^2$.
This result is satisfactory with respect to expectation (even for $d=1$), but not with respect to the variance of $\|Sx\|^2$. For instance,
if $x$ has only one non-zero entry  we need $d \sim \OOO(p)$ to ensure that $\|Sx\|^2 \neq 0$ with non-negligible probability. More generally, if one coordinate of $x$ is much larger (in absolute value) than all its other coordinates, then we will need a rather large value for $d$ to guarantee that $\|Sx\| \approx \|x\|$.

A natural way to quantify the ``peakiness'' of a vector $x$ is via the {\em peak-to-average ratio}\footnote{This quantity also plays an important role in wireless communications. There, one tries to avoid transmitting signals with a large peak-to-average ratio, since such signals would suffer from nonlinear distortions when they are passing through  the power amplifiers that are usually installed in cell phones. The potentially large 
peak-to-average ratio of OFDM signals is one of the alleged reasons why CDMA was dominant over OFDM for such a long time.}
measured by the quantity $\|x\|_\infty/\|x\|$.  It is easy to see that we have (assuming $x$ is not the zero-vector)
$$ \frac{1}{\sqrt{p}} \le \frac{\|x\|_\infty}{\|x\|} \le 1.$$
The upper bounds is achieved by vectors with only one non-zero entry, while the lower bound is met by constant-modulus vectors.
Thus, if  
\begin{equation}\label{approxpar}
\frac{\|x\|_\infty}{\|x\|} \approx \frac{1}{\sqrt{p}},
\end{equation}
we can hope that sparse subsampling of $x$ will still preserve its Euclidean norm.

Thus, this suggests to include a preprocessing step by applying a rotation so that sparse vectors become non-sparse in the new basis, thereby reducing their $\infty$-norm (while their 2-norm remains invariant under rotation). Two natural choices for such a rotation are  the Discrete Fourier transform (which maps unit-vectors into constant modulus vectors) and its $\Z_2$-cousin, the Walsh-Hadamard matrix\footnote{Hadamard matrices do not exist for all dimension $d$. But we can always pad $x$ with zeroes to achieve the desired length.}.
But since the Fast Johnson-Lindenstrauss Transform (FJLT) has to work for all vectors, we  need to avoid that this rotation maps dense vectors into sparse vectors. We can address this issue by ``randomizing'' the rotation, thereby ensuring with overwhelming probability that dense vectors are not mapped into sparse vectors. This can be accomplished in a numerically efficient manner (thus maintaining our overall goal of numerical efficiency) by first randomizing the signs of $x$ before applying the rotation. Putting these steps together we arrive at the following map.
\begin{definition}\label{FastJL}
The {\em Fast Johnson-Lindenstrauss Transform} is  the map $\Psi: \C^p \to \C^d$, defined by $\Psi:= S F D$, where 
$S$ and $D$ are random matrices and $F$ is a deterministic matrix. In particular,
\begin{itemize}
\setlength{\itemsep}{-0.5ex}
\setlength{\parsep}{-0.5ex}
\item $S$ is a $d \times p$ matrix, where each row of $S$ has just one single non-zero entry of value $\sqrt{p/d}$ at a location drawn uniformly at random.
\item $F$ is either the $p \times p$ DFT matrix or the $p \times p$ Hadamard matrix (if it exists), in each case normalized by $1/\sqrt{p}$ to obtain a unitary matrix.
\item $D$ is a $p\times p$ diagonal matrix whose entries are drawn independently  from $\{-1,+1\}$ with probability $1/2$.
\end{itemize}
\end{definition}

We can carry out the matrix-vector multiplication by the DFT matrix via the Fast Fourier Transform (FFT) in $\OOO(p \log p)$ operations; a similar algorithm exists for the Walsh-Hadamard matrix.
The FJLT allows for a dimension reduction that is competitive with the Johnson-Lindenstrauss Lemma as manifested by the following theorem.

\begin{theorem}[Fast Johnson-Lindenstrauss Transform]\label{th:fjl}
For $0<\eps<1$ and $0<\delta<1$, there is a random matrix $\Psi$ of size $d \times p$ with $d = \OOO \big( \log(p/\delta) \log(1/\delta) /\eps^2\big)$ such that, for
each $x \in \C^p$,
$$\| \Psi x \| \in [1-\eps,1+\eps] \cdot \|x\|$$
holds with probability at least $1-\delta$. Matrix-vector multiplication with $\Psi$ takes $\OOO(p\log p +d)$ operations.
\end{theorem}

For convenience we will prove Theorem~\ref{th:fjl} for the case when $F$ is the Walsh-Hadamard matrix so that the random variables are real-valued and the exposition is lighter. But it is straighforward to adapt the argument to the case when $F$ is the DFT matrix. Keeping this minor simplification in mind, the proof of Theorem~\ref{th:fjl} follows from the two lemmas below.
We first show that with high probability the random rotation $FD$ produces vectors with a sufficiently low peak-to-average ratio.

\begin{lemma}\label{le:par2}
Let $y = FDx$, where $F$ and $D$ are as in Definition~\ref{FastJL}. Then
\begin{equation}\label{jlpar}
\underset{D}{\P} \left( \frac{\| y \|_\infty}{\|y\|} \ge  \sqrt{\frac{2 \log (4p/\delta)}{p}}\right) \le \frac{\delta}{2}.
\end{equation}
\end{lemma}

\begin{proof}
Since $FD$ is unitary, the quantity $\|FDx\|_\infty/\|FDx\|$ is invariant under rescaling of $x$ and therefore we can assume $\|x\|=1$.

Let $\xi_i = \pm 1$ be the $i$-th diagonal entry of $D$. We have $y_i  =  \sum_{j=1}^p \xi_j F_{ij} x_j$ and note that the terms of this sum are i.i.d.\ bounded random variables. We thus can apply Hoeffding's inequality. In the notation of Theorem~\ref{thm:hoeffding1},
let $X_j =  \xi_j F_{ij} x_j$. We note that $X_j = \pm  F_{ij} x_j$,  hence $\E[X_j] =0$ and 
$|X_j  | \le a_j$, where $a_j = \left|F_{ij} x_j\right|$. It holds that
$$\sum_{j=1}^p a_j^2 = \sum_{j=1}^p |F_{ij}|^2 x_j^2  = \sum_{j=1}^p \frac{1}{p} x_j^2 = \frac{\|x\|^2}{p} = \frac{1}{p}.$$

We can now use Theorem~\ref{thm:hoeffding1} with $t = \sqrt{2 \log(4p/\delta)/p}$ and obtain
$$\P \left( |y_i| >  \sqrt{ \frac{2\log(4p/\delta)}{p}} \right) \le 2\exp\left(-\displaystyle{\frac{2 \log(4p/\delta)/p}{2/p}}\right) =   \frac{\delta}{2p}.$$

Applying the union bound finishes the proof. 
\end{proof}

\begin{lemma} \label{le:par3}
Let $0<\eps<1$ and $y\in \RR^p$ satisfy  $\|y \|_\infty^2 <  \frac{2\log (4p/\delta)}{p}$ and $\|y\|=1$, then it possible to pick $d \sim 1/\eps^2 \log(p/\delta) \log(1/\delta)$ such that
$$\P\big( \|Sy\|^2 - 1 | \le \eps \big) \le  1- \frac{\delta}{2}.$$
\end{lemma}

\begin{proof}

The idea is to use Bernstein's Inequality (Theorem~\ref{thm:Bernstein}).
We write $S_{ji} = \sqrt{p/d} \delta_{ji}$, where $\delta_{ji} \in \{0,1\}$ is our random sample of the columns for row $j$. Hence for all $j$,
$\sum_{i=1}^p \delta_{ji} = 1$. We write $z:= Sy$ and compute
\begin{align*}
q_j & := z_j^2 \\
      & = \frac{p}{d} \left( \sum_{i=1}^p \delta_{ji} y_i \right)^2 \\
      & = \frac{p}{d} \left( \sum_{i} \delta_{ji} y_i^2 +   \sum_{i \neq \ell} \delta_{ji} \delta_{j\ell} y_i y_{\ell} \right) \\
       & = \frac{p}{d} \sum_{i} \delta_{ji}y_i^2.
\end{align*}
We want to bound the random variable $\|z\|^2 -1= \sum_j \left(q_j-\frac1d\right)$.  Since the $q_j$'s are independent, and $\EE\left[q_j-\frac1d\right]=0$, we can apply Berstein's Inequality (Theorem~\ref{thm:Bernstein}, provided we bound $\sigma^2$ and $a$. Firstly,
$$a \le \frac{p}{d}\|y\|_\infty^2 -\frac1d < \frac{p}{d}\|y\|_\infty^2 \leq  \frac{2\log (4p/\delta)}{d}.$$
and secondly,
\begin{align*}
d\sigma^2 &  \le d \E[ q_1^2]  - 1 \le d \E[ q_1^2]  = d \E \left[ \frac{p^2}{d^2} \sum_i \delta_{ji} y_i^4 \right] \\
& = \frac{p^2}{d} \frac1p\sum_i y_i^4 \\
& \leq \|y\|_\infty^2 \frac{p}{d} \sum_i y_i^2 \\
& = \frac{p}{d} \|y\|_\infty^2  \|y\|^2 \\
& = \frac{p}{d} \|y\|_\infty^2 \\
& \leq  \frac{2\log (4p/\delta)}{d}.
\end{align*}
Plugging these terms into Bernstein's Inequality (Theorem~\ref{thm:Bernstein}) gives
\begin{align*} \P \big( | \|Sy\|^2 - 1 | > \eps \big)  & = \P \Big( \Big| \sum_{j=1}^d q_j - 1 \Big| > \eps \Big) \\
& \leq 2 \exp\left(  \frac{-\eps^2}{2d\left(\frac{2\log (4p/\delta)}{d^2}\right) + \frac23\left(\frac{2\log (4p/\delta)}{d}\right)\eps  } \right)\\
& \leq 2 \exp\left(  \frac{-\eps^2}{\frac83\left(\frac{2\log (4p/\delta)}{d}\right)  } \right),\\
& = 2 \exp\left(  \frac{-3\eps^2d}{16\log (4p/\delta)  } \right),
\end{align*}
where the last inequality uses $\eps<1$.
Picking $d = \frac{16}{3\eps^2} \log(4p/\delta) \log(4/\delta)$ gives the desired $\delta/2$-bound.

\end{proof}

Combining Lemmas~\ref{le:par2} and~\ref{le:par3}, union bounding over the two events, and recalling that if $\|x\|=1$ then $\|y\|=1$ since both $D$ and $F$ are unitary, establishes Theorem~\ref{th:fjl}.

Besides the potential speedup, another advantage of the FJLT  is that it requires significantly less memory  compared to storing an unstructured random projection matrix as is the case for the standard Johnson-Lindenstrauss approach.

There are many applications where one wants to perform dimension reduction while preserving the geometry of certain sets with infinite points, such as subspaces. Via an $\epsilon$-net argument it is possible to show that random matrices that are good as Johnson-Lindenstrauss maps for $N$ points, also approximately preserve the geometry of a ($\log N$)-dimensional subspace. These go by the name of Oblivious Subspace Embeddings (see, e.g., Section 5.2.2. in~\cite{KireevaTropp} where the fast JL construction we introduced above is shown to be an Oblivious Subspace Embedding). A similar idea will be explored below (Section~\ref{s:RSVD}) where a random projection will allow us to still approximate the subspace spanned by the singular vectors corresponding to the largest singular values, resulting in a randomized SVD algorithm that is workhorse of large scale linear algebra. In Chapter~\ref{c:probability-gaussiananalysis} we will develop analogues of the Johnson-Lindenstrauss Lemma for more general sets,  through the celebrated Gordon's espace through a mesh theorem.

\section{Randomized SVD}\label{s:RSVD}

Randomized linear algebra has gained significant relevance in recent years due to its ability to efficiently handle large-scale data problems that traditional methods struggle with.  
The Johnson-Lindenstrauss dimension reduction discussed in the previous section is one prominent example. The techniques encountered in the Johnson-Lindenstrauss transform naturally lend themselves as key ingredient of the \emph{randomized SVD},  a method to compute an approximation of the SVD of a matrix. It is especially useful when the matrix is large and dense and can be well approximated by a low-rank matrix. 

In this section we will give a detailed description of the randomized SVD.
It is based on using random projections to reduce the dimensionality of the matrix before performing a more traditional SVD on the reduced matrix.

Recall from Chapter~\ref{c:svd} that  the SVD of  a matrix $A\in\R^{m\times n}$ is given by
\begin{equation}\label{eq:SVDrand}
 A = U \Sigma V^{\ast},
\end{equation}
where $U \in \mathbb{R}^{m \times m}$ is a unitary matrix containing the left singular vectors,
$\Sigma \in \R^{m \times n}$ is a diagonal matrix with singular values (in decreasing order) on the diagonal,
and $V \in \mathbb{R}^{n \times n}$ is a unitary matrix  containing the right singular vectors.

In practice $A$ is often large, and we are interested in only a small number $s$ of the largest singular values and corresponding singular vectors, giving a rank-$k$ approximation of $ A $:
$$
A \approx U_k \Sigma_k V_k^{\ast}
$$
where $ U_k \in \mathbb{R}^{m \times k} $, $ \Sigma_k \in \R^{k \times k} $, and $ V_k \in \mathbb{R}^{n \times k} $.

We define the projection $P_k$ onto the the span of the leading $k$ left singular vectors of $A$ via
$$
P_k = \sum_{i=1}^k u_i u_i^{\ast}.
$$
We know from Theorem~\ref{th:rank_k_Frob} that the best rank-$k$ approximation of $A$ with respect to the Frobenius norm is given by
$$\min_{\operatorname{rank}(B) \leq k} \|A - B \|_F = \|A - P_k A \|_F =  \sqrt{\sum_{i > k} \sigma_i^2}.$$
Thus, we can compute the best approximation by projecting the target matrix onto the $k$-dimensional
subspace that captures ``most of the action'' of the matrix $A$.

In many applications it suffices to compute the best approximation with moderate accuracy. Hence, following~\cite{HMT11}, we let $s=k+p$ for small $p$, and seek a rank-$s$ approximation $\hat{A}_s$ that is comparable to the best rank-$k$ approximation, i.e.,
$$\|A - \hat{A}_s\|_F^2  \le (1+\eps) \|A - P_k A \|_F^2 = (1+\eps)\sum_{i > k} \sigma_i^2 $$
for some chosen tolerance $\eps > 0$.

The aim of the Randomized SVD (RSVD) is essentially to generate an efficient approximation for the best rank-$k$ approximation of the matrix $P_k A$. 
We  proceed in two steps. In the first step we compute a low-dimensional subspace that captures the range of $A$. In the second step we project $A$ onto the subspace and compute the (standard SVD) of the resulting smaller matrix.
Regarding the first step, recall that $P_k$ is the orthogonal projection  onto the $k$
leading left singular vectors of $A$. 
But since we do not know the subspace spanned by $P_k$, a sensible choice, as suggested by the  Johnson-Lindenstrauss Lemma,  is to draw some {\em probing} random vectors from a rotationally invariant distribution to estimate the directions of the vectors $u_i$ that make up $P_k$. We collect these probing random vectors in a matrix $\Psi$.

Hence, let $ \Psi \in \R^{n \times (k + p)} $ be a  random matrix, where $ k $ is the desired number of singular values/vectors,
and $ p $ is an oversampling parameter (typically $ p $ is small, like 5 or 10).  A promising choice for $ \Psi $ is to select
each entry of $ \Psi $ as a sample from a standard normal distribution $ \mathcal{N}(0, 1) $.
We then form the matrix $ Y \in \R^{m \times (k + p)} $ as
$$
Y = A \Psi.
$$
This matrix $ Y $ captures the action of $ A $ on a random subspace of dimension $ k + p $. We will see (in Theorem~\ref{th:randSVD}) that, with high probability, the subspace spanned by the columns of $ Y $ contains a good approximation of the subspace spanned by the top $ k $ left singular vectors of $ A $.\footnote{In many applications the matrix $A$ is implicit and one has access to it via matrix-vector products, if this is the case computing $Y$ can be done with $k+p$ matrix-vector products.}
Next, we find an orthonormal basis for the range of $ Y $. Using the standard QR decomposition, we compute
$$
Y = Q R,
$$
where $ Q \in \mathbb{R}^{m \times (k + p)} $ has orthonormal columns, and $ R \in \R^{(k + p) \times (k + p)} $ is an upper triangular matrix.
Now, we project $ A $ onto the lower-dimensional subspace spanned by the columns of $ Q $ by computing
$$
B = Q^{\ast} A.
$$
Here, $ B \in \R^{(k + p) \times n} $ is a much smaller matrix that approximates $ A $ (in the sense that $QB$ approximates $A$).

Next, we compute the exact SVD of the smaller matrix $ B $
$$
B = U_B \Sigma_B V_B^{\ast}.
$$
where $U_B \in \mathbb{R}^{(k + p) \times (k + p)} $, $ \Sigma_B \in \R^{(k + p) \times n} $, and $ V_B \in \R^{n \times  (k + p)} $.
Finally, the approximate SVD of $ A $ is given by
$$
A \approx \tilde{ U} \tilde{ \Sigma} \tilde{V}^{\ast}
$$
where $ \tilde{U} = Q U_B \in \mathbb{R}^{m \times (k + p)} $, $ \tilde{\Sigma} = \Sigma_B \in \R^{ (k + p) \times  (k + p)} $, and $ \tilde{V} = V_B \in \R^{n \times  (k + p)} $.

We summarize these steps in Algorithm~\ref{alg:RandSVD}.

\begin{algorithm}
\caption{Randomized SVD}
\label{alg:RandSVD}
\textbf{Input:} Matrix  $A \in\R^{m\times n}$, target rank $k>0$, oversampling factor $p>0$; set $s=k+p$.
\begin{enumerate}
\item Generate a random matrix $\Psi \in \mathbb{R}^{N \times (k+p)}$ 
\item  Compute the product $Y = A \Psi$ 
\item Compute the economy-sized QR factorization $Y = Q R$, with $Q \in\R^{m\times (k+p)}$ and  $R \in\R^{(k+p)\times (k+p)}$ 
\item Compute the product  $B = Q^{\ast} A$ 
\item Compute the economy-sized SVD  $B =  U_B \Sigma_B V_B^{\ast}$, with $U_B \in\R^{m\times (k+p)}, \Sigma_B \in\R^{(k+p)\times (k+p)}$ and  $V_B \in\R^{n \times (k+p)}$ 
\item Set $\tilde{U} = Q U_B, \tilde{\Sigma} = \Sigma_B, \tilde{V} = V_B.$ 
\end{enumerate}
\textbf{Output:} Orthogonal $\tilde{U} \in \R^{m \times (k+p)}$, orthogonal $\tilde{V} \in \R^{n \times (k+p)}$, and diagonal $\tilde{\Sigma} \in  \R^{(k+p) \times (k+p)}$ 
such that $A \sim \hat{A}_s:= \tilde{U} \tilde{\Sigma} \tilde{V}^{\ast}$ 

\end{algorithm}

The random matrix $ \Psi $ used in the random projection step ensures that with high probability, the subspace spanned by the columns of $ Y = A \Psi $ contains a good approximation to the dominant singular subspace of $ A $.

\begin{theorem}[\cite{HMT11}]\label{th:randSVD}
Consider a matrix $A \in \R^{m \times n}$ with singular values $\sigma_1 \ge \sigma_2 \ge \sigma_3 \ge \cdots$. Fix the target rank  $k \le \min(m,n)$. When $s :=k+p \ge k+2$, the randomized SVD method  produces a random rank-$s$
approximation $\hat{A}_s$ that satisfies
\begin{equation}
\label{randSVDbound}
\E[ \|A - \hat{A}_s \|_F ] \le \left(1 + \frac{k}{p-1}\right)^\frac{1}{2} \Big( \sum_{j > k}\sigma_{j}^2\Big)^\frac{1}{2}.
\end{equation}
\end{theorem}

The error on the right hand side consists of the best rank-$k$ approximation error and an additional factor due to using a randomized projection instead of the optimal projection $P_k$.

For the proof of Theorem~\ref{th:randSVD} we need some preparation. In what follows it will be convenient to rewrite the SVD of $A$, cf.~\eqref{eq:SVDrand}, in block matrix form:
\begin{equation}
A = \begin{bmatrix}  U_k & U_\perp \end{bmatrix} \begin{bmatrix}  \Sigma_k & 0 \\ 0 & \Sigma_\perp \end{bmatrix} \begin{bmatrix}  V_k^{\ast} \\ V_\perp^{\ast} \end{bmatrix},
\end{equation}
where $ U_\perp \in \R^{m \times m-k}$ denotes the orthogonal complement of $U_k \in \R^{m \times k}$ with respect to $U \in \R^{m \times m}$.
Furthermore, we set 
$$\Psi_k : = V_k^\ast \Psi \qquad \text{and} \qquad \Psi_\perp : = V_\perp^\ast \Psi .$$
Here, $\Psi_k$ captures the alignment of $\Psi$ with the leading right singular vectors of $A$, while  $\Psi_\perp$ captures
the alignment with the trailing right singular vectors of $A$. 
The matrix $P_Y$ denotes the orthogonal projection onto the range of $Y$. Since this projection is unique, we have that $P_Y = Q Q^{\ast}$.

The following deterministic error bound (the bound does not
require $\Psi$ to be Gaussian) will be instrumental:

\begin{lemma}\label{le:SVDerrorbound}
We use the same notation as in Theorem~\ref{th:randSVD}. Let $\Psi \in \R^{n \times (k+p)}$ be an arbitrary matrix with 
 $\rank(\Psi)=k$. Then the rank-$s$ approximation $\hat{A}_s$ computed by Algorithm~\ref{alg:RandSVD} satisfies
\begin{equation}\label{RSVDdeterministic}
\| A - \hat{A}_s\|^2_F \le \|\Sigma_\perp\|_F^2 + \|\Sigma_k \Psi_k \Psi_\perp^{\dagger}\|_F^2.
\end{equation}
\end{lemma}

\begin{proof}
We first fix the target rank $k \le \min{\{m,n\}}$ and the oversampling factor $p$ and set $s=k+p$. 
Denoting $\tilde{A} = U^{\ast} A$ and $\tilde{Y}= \tilde{A}\Psi$, we obtain
\begin{equation}\label{RSVDdeterministic1}
\| A - \hat{A}_s\|^2_F = \| (I - P_Y)A \|^2_F = \| U^{\ast} (I - P_Y) U\tilde{A} \|^2_F = \| ( I - P_{\tilde{Y}} ) \tilde{A} \|^2_F.
\end{equation}
Here, the first equality uses the fact that the Frobenius norm is unitarily invariant, whereas in the second equality we have substituted $P_{\tilde{Y}} = U^{\ast} P_Y U$.
Next, we set $Z=\tilde{Y} \Psi_k^{\dagger} \Sigma_k^{-1} =: [I \,\,\, F]^{\ast}$ where $F = \Sigma_\perp \Psi_\perp \Psi_k^{\dagger} \Sigma_k^{-1}$.
Note that $\range(Z) \subseteq \range(\tilde{Y})$. This in turn yields
\begin{equation}\label{RSVDdeterministic2}
\| (I-P_{\tilde{Y}} )\tilde{A}\|_F^2 \le \| (I - P_Z)\tilde{A} \|_F^2 = \tr \big( \tilde{A}^{\ast} (I - P_Z)\tilde{A} \big) = \tr \big(\Sigma (I - P_Z) \Sigma \big).
\end{equation}
Note that $P_Z$ has the expansion
$$
P_Z =  \begin{bmatrix}  I \\ F  \end{bmatrix}  \begin{bmatrix}  I + F^{\ast} F  \end{bmatrix}^{-1}  \begin{bmatrix}  I \\ F  \end{bmatrix}^{\ast}.
$$
Since $P_Z$ is a projection, we have that $I - P_Z \succ 0$. After some linear algebra calculations (left to the reader), we obtain
$$ 0 \prec I - P_X \preceq \begin{bmatrix}  F^{\ast}F & B \\ B^{\ast} &  I \end{bmatrix},$$
where $B = -(I+F^{\ast}F)^{-1} F^{\ast}$. Multiplying both sides by $\Sigma$ gives
\begin{equation}\label{RSVDdeterministic4}
\Sigma(I-P_Z)\Sigma  \preceq \begin{bmatrix}  \Sigma_k F^{\ast}F \Sigma_k & \Sigma_k B \Sigma_\perp \\ \Sigma_\perp B^{\ast} \Sigma_k & \Sigma_\perp \Sigma_\perp \end{bmatrix}.
\end{equation}
Combining~\eqref{RSVDdeterministic1},  \eqref{RSVDdeterministic2} and ~\eqref{RSVDdeterministic4} gives
\begin{flalign*}
\| A- \hat{A}_s \|_F^2 & = \|(I-P_{\tilde{Y}}) \tilde{A} \|_F^2  \\
                                  & \le \tr \big( \Sigma (I-P_Z) \Sigma \big) \\
                                  & \le \tr \big(\Sigma_k F^{\ast} F \Sigma_\perp \big) + \tr \big( \Sigma_\perp \big) \\
                                  & \le \|\Sigma_\perp \Psi_\perp \Psi_k^{\dagger} \|_F^2 + \|\Sigma_\perp\|_F^2,
\end{flalign*}
where in the penultimate inequality we have used the fact that the trace is monotone on the cone of positive semidefinite matrices.

\end{proof}

Lemma~\ref{le:SVDerrorbound} holds for {\em any} rank-$k$ matrix $\Psi$.  As pointed out in~\cite{KireevaTropp}, the first term on
the right-hand side of~\eqref{RSVDdeterministic}  captures the trailing singular values of the matrix, which is independent of the
specific choice of $\Psi$, since this loss occurs for every rank-$k$ approximation. The second term is the error caused by the
dimension reducing matrix $\Psi$.
We want the component $\Psi_k$ in the leading directions to be well conditioned (ideally, an
identity matrix). We want the component $\Psi_\perp$ in the trailing directions to be as small as possible.

The bound~\eqref{RSVDdeterministic} suggests that
in order to establish Theorem~\ref{th:randSVD} we need to get a handle on $ \Psi_k \Psi_\perp^{\dagger}$ when $\Psi_{ij}
\overset{\text{i.i.d.}}{\sim} {\mathcal N}(0,1).$ 
To that end, the following lemma will be helpful.

\begin{lemma}\label{pr:Gaussnorm2}
Let  $G \in \R^{m \times n}$ be a standard Gaussian random matrix. 
 Assume $m \ge 2$ and $n-m \ge 2$. Then
\begin{equation}\label{eq:gaussnorm3}
\E [ \| G^{\dagger} \|_F^2 ] = \frac{m}{n-m-1}.
\end{equation}
\end{lemma}

\begin{proof}
We follow the proof of the corresponding result in the appendix of~\cite{HMT11}.
 First, note that $(G G^{\ast})$ is invertible almost surely. We compute
$$  \| G^{\dagger} \|_F^2 = \trace [(G^{\dagger})^{\ast} G^{\dagger}] = \trace [(G G^{\ast})^{-1}].$$
The random matrix $(G G^{\ast})^{-1}$ follows the inverted Wishart distribution. Hence, we can use formula (12) on page 97 of~\cite{muirhead2009aspects} for the expected trace and obtain~\eqref{eq:gaussnorm3}.

\end{proof}

With these lemmata in place, we are now ready to prove Theorem~\ref{th:randSVD}.
\begin{proof}[of Theorem~\ref{th:randSVD}]
Since $\Psi$ is chosen to be a standard Gaussian matrix and the Gaussian distribution is invariant under unitary transformations, it follows that $\Psi_k=V^{\ast}_k \Psi$ and $\Psi_\perp =V^{\ast}_\perp \Psi$ are also standard Gaussian. 
Moreover, note that matrix $\Psi_\perp$ has full rank with probability one, since the rows of a (fat) Gaussian matrix are almost surely in general position.

We consider the bound established in Lemma~\ref{le:SVDerrorbound}, apply expectation to~\eqref{RSVDdeterministic} and use H\"{o}lder's inequality to obtain
\begin{equation}\label{RSVDbound4}
\E [ \| A - \hat{A}_s\|_F]  
\le \big(\E [ \| (I-P_Y)A\|_F^2] \big)^{1/2}
\le \Big( \|\Sigma_\perp\|_F^2 + \E [\|\Sigma_k \Psi_k \Psi_\perp^{\dagger}\|_F^2 ]  \Big)^{1/2}.
\end{equation}
Using the law of total expectation as well as the independence of
$\Psi_k$ and $\Psi_\perp$ we estimate
\begin{align}
\E [ \|\Sigma_k \Psi_k \Psi_\perp^{\dagger}\|_F^2 ]  & = \E_{\Psi_k} \big[ \E [ \| \Sigma_\perp \Psi_\perp \Psi_k^{\dagger} \|_F^2 | \Psi_k ] \big] \notag \\ 
& \le  \E \big [ \| \Sigma_\perp \|_F^2 \,  \|\Psi_k^{\dagger} \|_F^2 \big]  \notag \\
& \le  \| \Sigma_\perp \|_F^2 \, \E \big[ \|\Psi_k^{\dagger} \|_F^2 \big]  \notag \\
& \le  \frac{k}{p-1}  \sum_{j > k} \sigma_j^2, 
\label{RSVDbound6}
\end{align}
where the second  inequality follows from Lemma~\ref{pr:Gaussnorm2}.

Combining now~\eqref{RSVDbound4} with~\eqref{RSVDbound6} establishes the bound~\eqref{randSVDbound}.

\end{proof}

\subsection{Computational complexity of the Randomized SVD}

What is the computational complexity of the RSVD? The dominant costs for the RSVD are:
\begin{enumerate}
\item ${\mathcal O}(m n (k + p)) $ for the matrix-matrix multiplication $ A \Psi $, assuming that $\Psi$ is an unstructured, non-sparse matrix (like a Gaussian matrix).
\item $ {\mathcal O}(m (k + p)^2) $ for the QR decomposition of $ Y $.
\item ${\mathcal O}(mn (k + p))$  for computing $Q^{\ast} A$.
\item  ${\mathcal O}((k + p)^2 n) $ for computing the  SVD of the small matrix $ B $.
\end{enumerate}
Thus, the overall complexity of the randomized SVD is ${\OOO}(m n (k + p)) $, which is significantly smaller than
the $ {\OOO}(m n^2) $ cost of the classical SVD when $ k $ is much smaller than $ n $.

We can further reduce the cost of the matrix-matrix product  $ A \Psi $ by introducing structure or sparsity into $\Psi$. We already encountered one such choice in Definition~\ref{FastJL} in the form of the Fast Johnson-Lindenstrauss transform. The improved computational efficiency usually comes at the cost of additional log-factors in terms of required samples to guarantee the same accuracy.

When the singular values of $A$ decay slowly, Algorithm~\ref{alg:RandSVD} is not very accurate in the regime of interest, namely when $k \ll n$. The reason is that to achieve a small approximation error the matrix $Q$ must align well with the first $k + p$ left singular vectors of $A$. As a consequence, the product $A\Psi$ must reveal
the singular vectors, which in turn would require the singular vectors associated with small singular values not to interfere with the calculation. 

How can we remedy this issue? We can take inspiration from the power iteration. 
By raising our matrix to a power $q$ via $(AA^{\ast} )^q A$ prior to applying it to $\Psi$ we can enhance the decay of the 
spectrum of our matrix without effecting the direction of the eigenvectors. Hence, we replace in Algorithm~\ref{alg:RandSVD} the step $Y = A \Psi$ by $Y = (AA^{\ast} )^q A \Psi$.
A small power of $q$ (such as $q=1$ or 2) is often sufficient. In this modified RSVD  it is also advisable to increase the oversampling factor $p$. We refer to~\cite[Corollary 10.10]{HMT11} for an approximation error bound for this modified RSVD as well as to~\cite{tropp2023randomized} for a more detailed discussion.

The key ideas behind the RSVD also form the main ingredients for some other algorithms, such as randomized least squares and randomized Choleksy factorization, see~\cite{rokhlin2008fast,HMT11,KireevaTropp}.

\section{Matrix sketching}\label{s:sketching}

The methods we encountered in this chapter are closely intertwined with the concept of {\em matrix sketching}~\cite{liberty2013simple,mahoney2011randomized,rudelson2007sampling,HMT11}, a family of randomized techniques for constructing a low-dimensional representation of a large matrix while approximately preserving its essential linear-algebraic structure. 

Given a data matrix $A \in \mathbb{R}^{n \times d}$, the goal is to replace $A$ by a much smaller matrix—called a {\em sketch}---that enables accurate approximation of quantities such as norms, inner products, or dominant singular subspaces. 

Two complementary formulations are common. In the {\em projection-based} viewpoint, one forms a sketch $SA$, where $S \in \mathbb{R}^{m \times n}$ is a random embedding (e.g., Gaussian or subsampled randomized Hadamard), generalizing the Johnson–Lindenstrauss lemma from vectors to entire subspaces~\cite{woodruff2014sketching,tropp2017practical}. 

In the {\em sampling-based} viewpoint, one constructs a sketch by randomly selecting and rescaling columns (or rows) of $A$, often using leverage-score sampling to retain the most informative directions~\cite{frieze2004fast,mahoney2009cur,drineas2006subspace}. Both approaches can be understood as approximately preserving the geometry of the column space of $A$. 

Matrix sketching does not only underlie the randomized SVD, it provides a principled way to trade controlled approximation error for potentially substantial reductions in computational and memory costs. As such, it enables distributed linear algebra~\cite{woodruff2014sketching} and it is the fundamental mathematical too that makes {\em streaming} possible for high-dimensional data~\cite{harvey2008sketching,mahoney2011randomized}. Here, streaming refers to a computational model in which data are processed sequentially, in one (or very few) passes, using limited memory, without storing the entire dataset~\cite{muthukrishnan2005data,woodruff2014sketching}. Streaming is essential when
the dataset is too large to store in memory,
data arrive continuously (logs, sensor data, social networks), or computation must be done online.

Let us explore some exemplary results in sketching beyond the Johnson-Lindenstrauss transform and the randomized SVD.

\begin{definition}
A $(1\pm \eps)$ $\ell_2$-subspace embedding for the column space of an $n \times d$ matrix $A$ is a matrix $S$ for which for all $x \in \R^d$
$$
(1-\varepsilon)\|Ax\|^2 \le \|SAx\|^2 \le (1+\varepsilon)\|Ax\|^2
\quad \text{for all } v \in \mathbb{R}^d.
$$
\end{definition}

We will occasionally abuse notation and say that $S$ is an $\ell_2$-subspace embedding for $A$ itself, although it should be clear from the definition that this property is independent from the choice of a particular basis for the representation
of the column space of $A$.

\begin{theorem}[$\ell_2$-subspace embedding]\label{th:subspaceembedding} 
Let $U \subset \mathbb{R}^n$ be a fixed subspace of dimension $r$.
Let $S \in \mathbb{R}^{m \times n}$ be a random matrix such that for any fixed vector $x \in \mathbb{R}^n$,
$$
\mathbb{P}\left( \left| \|Sx\|^2 - \|x\|^2 \right| > \varepsilon \|x\|^2 \right)
\le 2e^{-c\varepsilon^2 m}
$$
If
$$
m \ge C \frac{r + \log(1/\delta)}{\varepsilon^2},
$$
then with probability at least $1-\delta$,
$$
(1-\varepsilon)\|x\|^2 \le \|Sx|^2 \le (1+\varepsilon)\|x\|^2
\quad \forall x \in U.
$$
\end{theorem}

\begin{proof}
By homogeneity, it suffices to prove the inequality for all
$$
x \in \mathbb{S}_U := \{x \in U : \|x\| = 1\}.
$$
There exists an $\eta$-net $\mathcal{N} \subset \mathbb{S}_U$ with
$
|\mathcal{N}| \le \left(1+\frac{2}{\eta}\right)^r.
$
Choose $\eta = \frac{\varepsilon}{4}$, then
$
|\mathcal{N}| \le \left(\frac{8}{\varepsilon}\right)^r
 \le
e^{C r}.
$
For each fixed $x \in \mathcal{N}$,
$$
\P \left(
\bigl| \|Sx\|^2 - 1 \bigr| > \frac{\varepsilon}{2}
\right)
 \le
2 e^{-c \varepsilon^2 m}.
$$
By the union bound,
$$
\P \left(
\exists x \in \mathcal{N} :
\bigl| \|Sx\|^2 - 1 \bigr| > \frac{\varepsilon}{2}
\right)
\le
|\mathcal{N}| \cdot 2 e^{-c \varepsilon^2 m}.
$$
Using the bound on the size of the net, $|\mathcal{N}| \le e^{C r}$,
we obtain
$$
\P \left(\exists x \in \mathcal{N} :
\bigl| \|Sx\|^2 - 1 \bigr| > \frac{\varepsilon}{2} \right)
\le 2 e^{ C r - c \varepsilon^2 m }.
$$
If
$$
m \ge C \frac{r + \log(1/\delta)}{\varepsilon^2},
$$
then
$$
Cr - c \varepsilon^2 m \le -\log(2/\delta).
$$

Hence
$$
\P \left(
\forall x \in \mathcal{N} :
\bigl| \|Sx\|^2 - 1 \bigr| \le \frac{\varepsilon}{2}
\right)
\ge 1-\delta.
$$
From now on, we condition on this event.

Let $y \in \mathbb{S}_U$ and choose $x \in \mathcal{N}$ such that
$$
\|y - x\| \le \eta = \frac{\varepsilon}{4}.
$$
We write
$$
\|Sy\|
\le
\|Sx\| + \|S(y-x)\|.
$$

Using concentration again for the fixed vector $y-x$,
$$
\|S(y-x)\|^2
\le
(1+\varepsilon)\|y-x\|^2
\le
(1+\varepsilon)\frac{\varepsilon^2}{16}.
$$

Combining with $\|Sx\|^2 \le 1+\varepsilon/2$, one obtains
$\|Sy\|^2 \le 1+\varepsilon$.
A symmetric argument gives the lower bound
$\|Sy\|^2 \ge 1-\varepsilon$.

Hence, with probability at least $1-\delta$,
$$
(1-\varepsilon)\|x\|^2
\le
\|Sx\|^2
\le
(1+\varepsilon)\|x\|^2
\quad \forall x \in U.
$$
 
\end{proof}

It is easy to see from our previously derived concentration bounds that a Gaussian random matrix and a Subsampled Randomized Hadamard (or Fourier) Transform as introduced in Definition~\ref{FastJL}
satisfy the conditions of Theorem~\ref{th:subspaceembedding}.

Another interesting and useful construction of a sketching matrix is the so-called {\em CountSketch}~\cite{charikar2002finding,thorup2012tabulation}. CountSketch is extremely sparse, making it an appealing choice for processing massive datasets or high-speed data streams. It is often used to estimate frequencies of elements in data streams.

A CountSketch matrix $S \in \mathbb{R}^{m \times n}$ is defined by two independent {\em hash functions}:
\begin{enumerate}
\item $h: \{1, \dots, n\} \to \{1, \dots, m\}$: This {\em bucket hash function} maps each column index to one of the $m$ rows.
\item $g: \{1, \dots, n\} \to \{+1, -1\}$: This {\em sign hash function} assigns a random sign to each column.
\end{enumerate}

The entries of $S$ are constructed as follows:
$$S_{i,j} = \begin{cases} g(j) & \text{if } h(j) = i, \\ 0, & \text{otherwise}. \end{cases}$$

The CountSketch construction enjoys the following useful properties: (i)~Sparsity: Each column of $S$ has exactly one non-zero entry. (ii)~Efficiency: To compute $Sx$, we do not perform a standard matrix-vector product. Instead, for each entry $x_j$, we simply update one row of the sketch: $sketch[h(j)] \leftarrow sketch[h(j)] + g(j)x_j$. (iii)~Time Complexity: Computing $SA$ for $A \in \mathbb{R}^{n \times d}$ takes only $\OOO(\text{nnz}(A))$ (the number of non-zero entries in $A$) operations.

A typical realization of $S$ might look like this:
$$S = \begin{bmatrix} 0 & 0 & 0 & 0 & -1 & 0 \\ +1 & 0 & 0 & 0 & 0 & 0 \\ 0 & 0 & 0 & +1 & 0 & 0 \\ 0 & -1 & 0 & 0 & 0 & +1 \\ 0 & 0 & -1 & 0 & 0 & 0 \\ 0 & 0 & 0 & 0 & 0 & 0 \end{bmatrix}.$$
In this specific example, the two underlying hash functions performed the following actions:
(i)~The bucket hash ($h$):
$h(1)=2, h(2)=4, h(3)=5, h(4)=3, h(5)=1, h(6)=4$.
Note that row 4 received two assignments (from columns 2 and 6, respectively), while row 6 is empty. This ``collision'' is expected in hashing. (ii)~The sign hash ($g$):
$g(1)=+1, g(2)=-1, g(3)=-1, g(4)=+1, g(5)=-1, g(6)=+1$.

Before proving subspace embeddings, we establish that CountSketch preserves the inner product.

\begin{lemma}\label{le:countsketchinner}
For any two vectors $x, y \in \mathbb{R}^n$, the inner product of their sketches via the CountSketch matrix $S$ is an unbiased estimator of their true inner product:
$$\E[\langle Sx, Sy \rangle] = \langle x, y \rangle.$$
\end{lemma}

The proof is left to the reader in Exercise~\ref{ex:countsketch}.

\begin{theorem}\label{thm:sketch-countsketch}
Let $U \subset \mathbb{R}^n$ be a subspace of dimension $d$. If the number of rows $m$ satisfies
$$m \ge \frac{2d^2}{\delta\epsilon^2},$$
then with probability $1-\delta$
$$(1-\epsilon)\|x\|^2 \le \|Sx\|^2 \le (1+\epsilon)\|x\|^2 \quad \forall x \in U.$$
\end{theorem}

\begin{proof}
A matrix $S$ is an $\epsilon$-subspace embedding for the column space of $U$ if and only if the eigenvalues of $(SU)^T (SU)$ are within $[1-\epsilon, 1+\epsilon]$. This is equivalent to
$$\| (SU)^T (SU) - I_d \| \le \epsilon.$$
Since the spectral norm is upper-bounded by the Frobenius norm, it suffices to show that 
$\mathbb{P}(\| (SU)^T (SU) - I_d \|_F^2 > \epsilon^2) \le \delta$.

Let $M = (SU)^T (SU) - I_d$. The entries of the product $(SU)^T (SU)$ are
$$((SU)^T (SU))_{jk} = \langle Su_j, Su_k \rangle$$
where $u_j$ and $u_k$ are the $j$-th and $k$-th columns of $U$.
From Lemma~\ref{le:countsketchinner} we know $\E[\langle Su_j, Su_k \rangle] = \langle u_j, u_k \rangle$.
Since $U$ has orthonormal columns, $\langle u_j, u_k \rangle = \delta_{jk}$. Thus, $\E[M] = 0$.

The squared Frobenius norm is the sum of the squares of the entries
$$\E[\|M\|_F^2] = \sum_{j=1}^d \sum_{k=1}^d \E[((SU)^T (SU) - I_d)_{jk}^2].$$
We expand the expectation for a single pair $(j, k)$. Let $X_{jk} = \langle Su_j, Su_k \rangle$. We seek $\Var(X_{jk})$ because $\E[(X_{jk} - \delta_{jk})^2] = \Var(X_{jk})$.
Using the definition of the CountSketch matrix, we can expand
$X_{jk}$ as follows:
$$X_{jk} = \sum_{a=1}^n \sum_{b=1}^n g(a)g(b) \mathbf{1}_{h(a)=h(b)} U_{aj} U_{bk}$$
Using the property of random signs $g$, the expectation of $X_{jk}^2$ survives only when indices match in pairs. After some algebraic simplification using the independence of $h$ and $g$ we obtain
$$\Var(\langle Su_j, Su_k \rangle) = \frac{1}{m} \left( \|u_j\|^2 \|u_k\|^2 + \langle u_j, u_k \rangle^2 \right) - \frac{2}{m} \sum_{a=1}^n u_{aj}^2 u_{ak}^2$$
Summing over all $j, k \in \{1, \dots, d\}$ gives
$$\E[\|M\|_F^2] = \sum_{j,k} \Var(X_{jk}) \le \frac{1}{m} \sum_{j,k} (\|u_j\|^2 \|u_k\|^2 + \langle u_j, u_k \rangle^2)$$
Since $\|u_j\|^2 = 1$ and $\sum_{j,k} \langle u_j, u_k \rangle^2 = \|U^T U\|_F^2 = \|I_d\|_F^2 = d$, we get
$$\E[\|M\|_F^2] \le \frac{d^2 + d}{m} \le \frac{2d^2}{m}.$$

Applying Chebyshev's inequality gives
$$\mathbb{P}(\|M\|_F^2 > \epsilon^2) \le \frac{\E[\|M\|_F^2]}{\epsilon^2} \le \frac{2d^2}{m \epsilon^2}$$
To ensure this probability is at most $\delta$, we set
$$\frac{2d^2}{m \epsilon^2} = \delta \implies m = \frac{2d^2}{\delta \epsilon^2}$$
This completes the proof. 
\end{proof}

The attentive reader will notice a couple of places where this argument is rather suboptimal: the tail bound is obtained with Chebyshev's inequality, and the quantity controlled is the Frobenius norm rather than the spectral norm (indeed, many $d\times d$ random matrices tend to have Frobeniues norm that is about $\sim\sqrt{d}$ times larger than the spectral norm). Using random matrix tools similar to the ones we will develop in Chapter~\ref{c:probability-matrixconcentration}, it is possible to significantly improve Theorem~\ref{thm:sketch-countsketch} (see e.g.~\cite{nelson2013osnap}). 
In any case, while CountSketch is efficient, it requires a larger sketch size than Gaussian matrices to achieve the same $\ell_2$-subspace embedding guarantee. But CountSketch is the preferred choice when $n$ and $d$ are so large that even a single dense matrix multiplication is prohibitive.

\bigskip

The following theorem illustrates the use of sketching to speed up the solution of large-scale least squares problems, as they arise for example in high-dimensional linear regression.

\begin{theorem}[Approximate regression via sketching]
Let $A \in \mathbb{R}^{n\times d}$, $y \in \mathbb{R}^n$, and let $S$ be an $\ell_2$-subspace embedding for $\mathrm{Range}([A; y])$ (where $\mathrm{Range}([A; y])$ denotes the subspace spanned by the columns of $A$ and the vector $y$).
Let
$$
\tilde{z} = \underset{z}{\argmin} \, \|SAx - Sy\|.
$$
Then
$$
\|A\tilde{x} - y\| \le (1+\varepsilon)\min_x \|Ax - y\|.
$$
\end{theorem}

\begin{proof}
Let
$$
r(x) := Ax - y.
$$
The least-squares solution satisfies
$$
x^* = \arg\min_x \|r(x)\|.
$$

Since $r(x) \in \range ([A; y])$, the $\ell_2$-subspace embedding implies
$$
(1-\varepsilon)\|r(x)\|
\le \|Sr(x)\|
\le (1+\varepsilon)\|r(x)\|
\quad \forall x.
$$

By definition of $\tilde{x}$,
$$
\|S r(\tilde{x})\| \le \|S r(x^*)\|.
$$

Applying the embedding bounds yields
$$
(1-\varepsilon)\|r(\tilde{x})\|
\le (1+\varepsilon)\|r(x^*)\|.
$$

Dividing by $1-\varepsilon$,
$$
\|A\tilde{x} - y\|
\le \frac{1+\varepsilon}{1-\varepsilon} \|Ax^* - y\|.
$$

For $0 < \varepsilon \le 1/3$,
$$
\frac{1+\varepsilon}{1-\varepsilon} \le 1 + 3\varepsilon,
$$
which yields the desired bound (up to rescaling $\varepsilon$).
     
\end{proof}

\subsection{Fast approximate matrix multiplication via sketching?}

Computing the product of two matrices is one of the most common and basic procedures in numerical linear algebra.
Let $A,\,B$ be two arbitrary $n\times n$ matrices. The cost for the classical way (the standard inner product method) to compute the product $AB$  is ${\mathcal O}(n^3)$.

In 1969 Strassen took the linear algebra community by surprise when he showed that the standard matrix
multiplication algorithm is not optimal from a computational viewpoint by presenting an algorithm that uses  only
${\OOO}(n^{2.808})$ operations~\cite{Strassen1969}.
Strassen's algorithm proceeds as follows.
Let us decompose the product of $A$ and $B$ into blocks
in the following way
$$
\begin{bmatrix} A_{11}&A_{12} \\ A_{21}&A_{22} \end{bmatrix}
\begin{bmatrix} B_{11}&B_{12} \\ B_{21}&B_{22} \end{bmatrix} =
\begin{bmatrix}  C_{11}&C_{12} \\ C_{21}&C_{22} \end{bmatrix},
$$
where  each block is of size $\frac{n}{2}\times\frac{n}{2}$. Following the standard
way of multiplication, this requires eight $\left(\frac{n}{2}\times\frac{n}{2}\right)$ matrix multiplications
and four additions of $\frac{n}{2}\times\frac{n}{2}$ matrices.
Strassen devised a formula that computes this block matrix multiplication using only {\em seven}
$\frac{n}{2}\times\frac{n}{2}$ matrix multiplication and eighteen $\frac{n}{2}\times\frac{n}{2}$ additions.
The cost for multiplying two $\frac{n}{2}\times\frac{n}{2}$ matrices is ${\mathcal O}\left(\frac{n}{2}\right)^3$ and the cost
for adding two $\frac{n}{2}\times\frac{n}{2}$ matrices is ${\mathcal O}\left(\frac{n}{2}\right)^2$. Since matrix addition is cheaper than matrix multiplication, there is a net gain in computational cost.
Proceeding recursively, with blocks of size $\frac{n}{2}$, $\frac{n}{4}$, \ldots, (up to blocks of size 16)
we end up with {\em Strassen's algorithm}. The total cost now is $\mathcal{O}(n^{\log_2 7})\sim \mathcal{O}(n^{2.8074})$.

Coppersmith and Winograd improved Strassen's algorithm further, currently sitting at around ${\mathcal O}(n^{2.37})$ operations at the cost of an increased constant.

An important open question is whether we can we construct a fast, practical algorithm\footnote{Although fast matrix multiplication algorithms such as Strassen’s method and the Coppersmith-Winograd algorithm improve the asymptotic complexity of matrix multiplication, they are rarely used in practical numerical linear algebra. Their large constant factors, increased numerical instability, higher memory requirements,  poor cache behavior and high communication overhead in distributed systems typically outweigh the theoretical savings. Consequently, most practical libraries rely on highly optimized implementations of the classical algorithm.} that can multiply two arbitrary $n\times n$ matrices
in ${\mathcal O}(n^{2+\epsilon})$ (or, say, ${\mathcal O}(n^{2}\log n)$) operations.

Yet, in certain cases one may be content with computing $AB$ approximately instead of exactly. By sacrificing precision, this should allow for a reduced complexity algorithm. Randomized algorithms are a natural candidate to realize this goal.

Let $A \in \mathbb{R}^{m \times n}$ and $B \in \mathbb{R}^{n \times p}$. An intuitive approach, popularized by Drineas, Kannan, and Mahoney~\cite{drineas2006fast}, treats the product $AB$ as a sum of $n$ rank-one outer products
$$AB = \sum_{i=1}^n A^{(i)} B_{(i)}.$$
Instead of summing all $n$ terms, we pick $c$ indices $\{i_1, i_2, \dots, i_c\}$ from $\{1, \dots, n\}$ with replacement, based on a probability distribution $\{p_i\}$. We then form the approximation
$$\widetilde{C} = \sum_{t=1}^c \frac{1}{c p_{i_t}} A^{(i_t)} B_{(i_t)}$$
This is equivalent to $AB \approx (AS)(S^T B)$, where $S$ is a sparse sampling and scaling matrix.

The quality of the approximation depends heavily on the sampling probabilities $\{p_i\}$.
If we use  probabilities that are proportional to the Euclidean norms of the columns and rows, 
$$p_i = \frac{\|A^{(i)}\| \|B_{(i)}\|}{\sum_{j=1}^n \|A^{(j)}\| \|B_{(j)}\|},$$
the expected error is bounded by
$$E[\|AB - \widetilde{C}\|_F] \leq \frac{1}{\sqrt{c}} \|A\|_F \|B\|_F.$$
While $\|A\|_F \|B\|_F$ can be large, this bound may be sufficient for applications where the matrices have a  are already noisy or when the dimension $n$ is in the order of millions and $c$ is in the order of 10000, say. However, the cost for computing all the probabilities $p_i$ must be factored into the total cost of matrix-matrix multiplication.

Random projections (like the subsampled Randomized Hadamard Transform or Gaussian sketches) lead to bounds of the form
$$\|AB - (AS)(S^T B)\| \leq \|AB - (AB)_k\| + \epsilon \|A\|_F \|B\|_F,$$
where $(AB)_k$ is the best rank-$k$ approximation of the product, see e.g.~\cite{HMT11,cohen2015optimal}. This indicates that the randomized product effectively captures the most significant singular values of the true product.

\smallskip
In conclusion, randomized sketching can approximate large matrix products efficiently when the matrices share a very large dimension. However, the benefits are limited when the output size is large, when strong spectral guarantees are required, or when the matrices have large stable rank. Consequently, sketching is most effective when the matrix product serves as an intermediate step in a larger algorithm, such as regression, low-rank approximation, or graph computations.

\section*{Exercises}
\addcontentsline{toc}{section}{Exercises}

\begin{myexercise}[\level\level\sep Optimality of the Johnson-Lindenstrauss Lemma]
    \label{prob:optimality_jl}
    Recall that the Johnson-Lindenstrauss lemma states that for any set of $n$ points in $\R^p$ there is a linear map $f\colon \R^p \to \R^d$, which acts almost as an $\eps$-isometry, for some $d \asymp \log n$. We will show that this dependency is indeed optimal.
    
    Fix some $d \in \N$ and $\eps > 0$. A subset $S \subseteq \R^d$ is said to be $\eps$-\textit{separated} if there are no distinct $s, t \in S$ such that $\norm{s-t}_2\leq\eps$. A \textit{packing number} of a subset $T \subseteq \R^d$ is defined as the size of its largest subset that is $\eps$-separated, i.e.
    \begin{equation*}
        \DD(T, \eps) = \max_{\substack{S \subseteq T \\ S \text{ is } \eps-\text{separated}}} \abs{S}.
    \end{equation*}
    \begin{enumerate}[(a)]
        \item Suppose $S$ is an $\eps$-separated subset of $B^d(1)$, the unit ball in $\R^d$. Using a volume argument, prove that
        \begin{equation*}
            \DD\brap{B^d(1), \eps} \leq \brap{\frac{2+\eps}{\eps}}^d.
        \end{equation*}

        \item Consider the set $X = \brac{x_1,\ldots,x_n}$ of $n \geq d$ points in $\R^n$, given by
        \begin{equation*}
            x_i = \begin{cases}
                e_i &\text{if }i < n,\\
                0 &\text{otherwise};
            \end{cases}
        \end{equation*}
        where $e_i$ is $i$-th coordinate vector in $\R^n$. Suppose that $f\colon\R^n\mapsto\R^d$ is an $\eps$-isometry for $X$. Show that 
        \begin{equation*}
            f(X) = \brac{f(x_1), \ldots, f(x_n)}
        \end{equation*}
        is an $(1-\eps)$ separated subset of some translated copy of $B_d(1+\eps)$.

        \item Conclude that
        \begin{equation*}
            n \leq \brap{\frac{3+\eps}{1-\eps}}^d.
        \end{equation*}
    \end{enumerate}
\end{myexercise}

\begin{myexercise}[\level\level\sep Johnson-Lindenstrauss Lemma: Alternative Version]
    \label{prob:jl_lemma}
    The goal of this exercise is to prove the random projection lemma and then to use this result to prove another version of the Johnson-Lindenstrauss lemma.
    \begin{enumerate}[(a)]
        \item Let $P \in \R^{n\times m}$ be the coordinate projection, which maps a vector in $\R^n$ onto its first $m$ coordinates in $\R^m$. Let $z\in S^{n-1}$ be a random vector sampled uniformly on the sphere $S^{n-1}$. Show that 
        \begin{equation*}
            \E\norm{Pz}_2^2 = \frac{m}{n} \norm{z}_2^2.
        \end{equation*}
        
        \item Prove the following statement using the result in Problem~\ref{prob:lipschitz_concentration}: There exists an absolute constant $c > 0$, such that for any $\eps > 0$, the following inequality holds with probability at least $1-2\exp(-c\eps^2 m)$
        \begin{equation}\label{eq:random_proj_lm}
            (1-\eps)\sqrt{\frac{m}{n}}\norm{z}_2\le \norm{Pz}_2\le (1+\eps)\sqrt{\frac{m}{n}}\norm{z}_2.
        \end{equation}
        
        \item Note that the result in (b) is stated for a random vector, while in dimension reduction we wish to find a randomized projection such that it preserves geometry for a fixed set of points. However, it can be shown that the same result \eqref{eq:random_proj_lm} holds when $z\in\R^n$ is a fixed vector, and $P$ is a orthogonal projection onto an $m$-dimensional subspace chosen uniformly at random from all $m$-dimensional subspaces in $\R^n$. In fact, these two models are equivalent.
        
        You can use the mentioned fact \textbf{without proof}. Using (b), show the following result: Let $\mathcal{X}$ be a set of $n$ points in $\R^n$ and let $\eps > 0$. Suppose that 
        \begin{equation*}
            m \geq \frac{C}{\eps^2} \log n.
        \end{equation*}
        
        Consider a random subspace $E$ of dimension $m$ chosen uniformly from all $m$-dimensional subspaces in $\R^n$, and let $P$ be an orthogonal projection on this set. Then with probability at least $1 - 2 \exp(-c \eps^2 m)$, the scaled projection $Q = \sqrt{\tfrac{n}{m}} P$ is an $\eps$-approximate isometry for $\mathcal{X}$, i.e., for all $x, y \in \mathcal{X}$,
        \begin{equation*}
            (1 - \eps) \| x - y \| \le \| Q x - Q y \| \le (1 + \eps) \| x - y\|.
        \end{equation*}
        Here $C,c > 0$ are universal constants.
    \end{enumerate}
\end{myexercise}

\begin{myexercise}[\level\level \sep Random projections flatten vectors]\label{prob:projections_flatten_vectors}
     We saw that random projections are good at preserving euclidean distances. In this exercise we'll study their interaction with the $\ell_\infty$-norm. A vector is considered "flat" if its $\ell_\infty$-norm is small compared to its euclidean norm. We will show that random projections flatten vectors. 
    \begin{enumerate}[(a)]
        \item We are given a deterministic matrix $A \in \R^{m \times n}$, a fixed vector $v \in \R^n$ with $\norm{v} =1$ and a random diagonal matrix $D \in \R^{n \times n}$ that has independent Rademacher entries on its diagonal (so $\mathbb{P}(D_{i,i} = 1) = \mathbb{P}(D_{i,i} = -1) = \frac12$). Prove
        \begin{equation*}
            \mathbb{P}(\norm{ADv}_\infty \geq \norm{A}_\infty t) \leq 2m e^{-t^2/2},
        \end{equation*}
        where $\norm{A}_\infty = \max_{i,j} |A_{i,j}|$.
        \item Let $P \in \R^{m \times n}$ be a uniform random projection matrix. Prove that for some universal constant $c>0$
        $$\mathbb{P} \left( \norm{P}_\infty \geq  \frac{t}{\sqrt{n}} \right) \leq 2mn e^{-ct^2}.$$
        You can use without proof that the distribution of any row of $P$ is a uniform random vector on the euclidean sphere $S^{n-1}$.
        \item Conclude that for $\mathcal{X} \subseteq S^{n-1}$ being a set of unit vectors of cardinality $|\mathcal{X}| = k$, we have that with probability at least $1- \frac Cn$ for all $x \in \mathcal{X}$
        $$ \sqrt{\frac nm }\norm{Px }_\infty \leq C'\sqrt{\frac{\log(nmk)}{m}},$$
        where $C,C'>0$ are sufficiently large universal constants.
    \end{enumerate}
\end{myexercise}

\begin{myexercise}
Implement the Randomized SVD algorithm to approximate the rank-$k$ SVD of a $1000 \times 1000$ Hilbert matrix (a matrix known for rapid singular value decay).\\
(a) Compute the true SVD and the Randomized SVD with k=10 and p=5. \\
(b) Plot the singular values obtained from both methods on the same graph. \\
(c) Report the error.
\end{myexercise}

\begin{myexercise}
Let $A \in \mathbb{R}^{m \times n}$ be a matrix of rank $r$, constructed as
$A = U \Sigma  V^T$, where $U \in \mathbb{R}^{m\times r}$ and $V \in \mathbb{R}^{n\times r}$ have orthonormal columns and $\Sigma =\operatorname{diag}(\sigma_1,\dots,\sigma_r)$. Consider the following two singular value decay regimes:
$$\text{(i) Exponential decay:} \qquad
   \sigma_j = e^{-\alpha j}, \quad \alpha > 0.
$$
$$\text{(ii) Polynomial decay:} \qquad
   \sigma_j = j^{-p}, \quad p > 1.$$
For each regime: \\
(a) Generate a test matrix $A$ of size $1000\times 1000$. \\
(b) Compute a rank-$k$ approximation using randomized SVD without power iterations and
compute the relative Frobenius norm error
$\frac{\|A - A_k^{\mathrm{rSVD}}\|_F}{\|A\|_F}$.
\\
(d) Repeat the experiment with power iterations.\\
(e) Compare the results with the optimal truncated SVD.
\end{myexercise}

\begin{myexercise}
Show that if $S$ is a $(1\pm \eps)$ $\ell_2$-subspace embedding for $A$, then it is also a $(1\pm \eps)$ $\ell_2$-subspace embedding for $U$, where $U$ is an orthonormal basis for the column space of $A$.    
\end{myexercise}

\begin{myexercise}\label{ex:countsketch}
Prove Lemma~\ref{le:countsketchinner}.    
\end{myexercise}

\begin{myexercise}(Comparing sketching matrices in least squares)
Let $A \in \mathbb{R}^{n \times d}$ with $n \gg d$ and let $b \in \mathbb{R}^n$. Consider the overdetermined least-squares problem
$$
x^\star = \arg\min_{x \in \mathbb{R}^d} \|Ax - b\|.
$$

Compare the performance of different sketching matrices $S \in \mathbb{R}^{m \times n}$ for approximating $x^\star$ via the sketch-and-solve approach.
\begin{enumerate}
\item Generate a matrix $A \in \mathbb{R}^{2000 \times 50}$ with i.i.d.\ standard Gaussian entries. Generate a ground-truth vector $x_0 \in \mathbb{R}^{50}$ with i.i.d.\ standard Gaussian entries. Set $b = Ax_0 + \eta$, where $\eta \sim \mathcal{N}(0, 0.01 I)$.
Compute the exact least-squares solution
$x^\star = \arg\min_x \|Ax - b\|$
using a standard solver (QR or SVD).
Record $\|Ax^\star - b\|$ and the runtime.

\item 
For each sketch size $m \in \{100, 200, 400, 800\}$, repeat the following experiment 10 times for each sketching matrix:
\begin{enumerate}
\item Gaussian sketch: $S_{ij} \sim \mathcal{N}(0, 1/m)$.
\item SRHT sketch:
$S = \sqrt{\frac{n}{m}} R H D$, where $D$ is a random sign matrix, $H$ is the Hadamard transform, and $R$ samples $m$ rows uniformly.
\item CountSketch: $S$ has one nonzero per column, with random hash function and random signs.
\end{enumerate}

For each sketch $S$, compute the sketched solution
$\tilde{x} = \arg\min_x \|SAx - Sb\|$.

Record (i) the relative solution error: $\|\tilde{x} - x^\star\| / \|x^\star\|$, (ii) the relative residual error: $\|A\tilde{x} - b\| / \|Ax^\star - b\|)$, and (iii) the runtime for forming $SA$ and solving the reduced problem.
\item 
Plot the average relative residual error versus sketch size $m$ for each sketching method. Plot the average runtime versus sketch size $m$.
\end{enumerate}

Comment on which sketch reaches a given accuracy with the smallest $m$, which sketch is fastest overall, and 
how does sparsity of the sketch affect runtime.

\begin{myexercise}
Repeat the previous exercise with a sparse random matrix $A$, where each row has only 10 nonzero entries (the locations and the entries of the 10 non-zero entries in each row are chosen at random).  How does the performance of CountSketch change relative to Gaussian and SRHT?
\end{myexercise}

\end{myexercise}

\begin{myexercise}
Let $L$ be the graph Laplacian introduced in Definition~\ref{def:graphLaplacian}. 
Let $S \in \mathbb{R}^{m \times n}$ be a sketching matrix (Gaussian, SRHT, or CountSketch).
\begin{itemize}
\item[(a)] 
Define the sketched quadratic form
$$
q_S(x) := \|S L^{1/2} \|^2
$$
and suppose $S$ is an $\ell_2$-subspace embedding for the column space of $L^{1/2} U$.
Show that, with high probability,
$$
   (1-\varepsilon) x^T L x \le q_S(x) \le (1+\varepsilon) x^T L x
   \quad \forall x \in U.
$$

\item[(b)] Generate a random geometric graph with $n = 2000$ nodes  as follows: Sample $n$ points $x_1,\dots,x_n$ independently and uniformly at random in the unit square $[0,1]^2$. Fix a radius $r>0$ and create an undirected edge between nodes $i$ and $j$ if $\|x_i - x_j\| \le r$.
Now, form the graph Laplacian $L$ and
compute the first five nontrivial eigenvectors of $L$ using (i) the full Laplacian, (ii) a sketched Laplacian using a Gaussian sketch, (iii) a sketched Laplacian using CountSketch. For each method, compare runtime, approximation error in eigenvalues,
and subspace error (principal angles between eigenspaces). 
\item[(c)] Why does the presence of the nullspace of $L$ require care when applying sketching?
\end{itemize}

\end{myexercise}


\chapter{Optimization for Data Science}
\label{c:optimization}

\newcommand{\dom}{{\mathcal D}}
\newcommand{\feas}{{\mathcal F}}
\newcommand{\finf}{{f_{\inf}}}
\newcommand{\loss}{{\mathcal C}}
\newcommand{\cost}{{\mathcal C}}
\newcommand{\PL}{{Polyak-\L ojasiewicz}}
\newcommand{\PLC}{{Polyak-\L ojasiewicz\ condition}}

Optimization lies at the heart of modern data science. Whether fitting a linear regression model, clustering data, finding a sparse solution to an underdetermined system of equations, or training a deep neural network, the underlying task is almost always the same: {\em find the parameters that minimize (or maximize) a well-defined objective function}. 

This chapter provides a brief, rigorous and practical introduction to optimization from a data science perspective. We begin by formalizing what an optimization problem is: a function to be minimized, variables to be chosen, and constraints that restrict which solutions are allowed. From there, we discuss convexity,   
introduce the necessary tools for analyzing optimality, including gradients, Hessians, the Karush–Kuhn–Tucker conditions, duality, and rate of convergence. Finally, we study gradient descent and popular variants, which play a central role in large-scale machine learning. For a comprehensive study of optimization we recommend the sources~\cite{LVanderberghe_SBoyd_book,calafiore2014optimization,nesterov2013introductory,nocedal2006numerical}.

A mathematical optimization problem can be expressed as\footnote{The setup~\eqref{eq:optim} also includes the case of maximizing an objective $f(x)$ by simply considering the minimization of $-f(x)$.}
\begin{align}\label{eq:optim}
\begin{split}
\underset{x \in \mathcal{D}}{\operatorname{minimize}} & \qquad   f(x)  \\
 \text{subject to} & \qquad g_i(x) \le b_i, \quad \, i = 1, \dots, m, \\
 & \qquad h_j(x) = c_j, \quad j=1,\dots,r.
\end{split}
\end{align}
Here, the vector $x \in \dom$ is the {\em optimization variable} of the problem and $\dom$ is the search space or domain of $x$; often we set $\dom = \R^n$. The
function $f: \dom \to \R$ is the {\em objective function}; in data science and machine learning $f$ is commonly referred to as the {\em cost function} or {\em loss function} when minimization is sought, and the {\em utility function}  when maximization is sought.
The functions
$g_i: \dom \to \R, i=1,\dots,m$,  $h_j: \dom \to \R, j=1,\dots,r$ are the equality and inequality constraint functions, and the constants $b_i, c_j$
are the limits, or bounds, for the constraints, respectively.\footnote{The attentive reader might notice that it would have been equivalent to simply embedded the constraints on the search space $\dom$, however there are many benefits to keeping $\dom$ as a ``canonical set'' such as $\RR^n$ or $\{\pm1\}^n$ and separating the instance specific constraints; this will be more apparent when we discuss constraints optimization below.} If there are no constraints in~\eqref{eq:optim}, the problem is called {\em unconstrained}, otherwise it falls in the realm of {\em constrained optimization}. 

A vector $x^*$ is called {\em optimal}, or a
{\em solution} of the problem~\eqref{eq:optim}, if it has the smallest objective value among all vectors that satisfy the constraints; that is, for any $z\in\dom$ with $g_i(z) \le b_i, i=1,\dots,m$, $h_j(z) = c_j, j=1,\dots,r$, we have $f(z) \ge f(x^*)$. 
The set of all points satisfying the constraints is known as the {\em feasible region}, which we denote by $\feas$.

Many foundational data science models naturally fit into this structure.
For example, linear regression minimizes the mean squared error over all parameter vectors, usually without constraints. Logistic regression minimizes cross-entropy loss, again as an unconstrained problem unless we explicitly add regularization or parameter bounds. Data clustering and community detection can be expressed in the form~\eqref{eq:optim} where the constraints are of discrete type (cf.~Chapter~\ref{c:graphs} and Chapter~\ref{c:community}, respectively).
Understanding this basic formulation gives the language and framework needed for all the optimization techniques that follow.

The structure of an optimization problem determines both the difficulty of finding a solution and the type of algorithms we can use. We have already seen several optimization problems in the preceding chapter, and will see many more in chapters that follow. For example, the PCA objective~\eqref{eq:PCA:goodnessoffit0} is an optimization problem on continuous variables, the $k$-means objective~\eqref{eq:3:kmeans_obj} was an optimization problem with both discrete variables (the assignments to clusters) and continuous variables (the cluster centers), and the Minimimum bisection problem~\eqref{eq:minbis-andspectrum} is an optimization problem on discrete variables. Below we will focus on continuous optimization. Indeed, while this book covers several combinatorial optimization problems the approach we take (as in Chapter~\ref{c:graphs}) is often to relax to a suitable continuous optimization problems. We point the reader to~\cite{wolsey1999integer,schrijver2003combinatorial} for more on combinatorial optimization.

We have already seen (and will see more) optimization problems for which there are efficient algorithms (such as the PCA example) and optimization problems for which no such algorithm is believed to exist (such as minimum bisection). While the reader might be tempted to conjecture that discrete optimization problems are harder than continuous ones, the situation is more subtle: it is true that combinatorial optimization problems are often formulated over exponential sized sets (such as $\{\pm1\}^n$) but that on itself does not make them computationally hard (just take $\min_{x\in\{\pm1\}^n} \1^Tx$ as an easy example). Also, note that while the Max-Cut problem~\eqref{MAXCUTproblem} is an NP-hard combinatorial problem, it can be equivalently formulated as a continuous one by replacing $x\in\{\pm1\}^n$ by $-1\leq x_i\leq 1$ (see Exercise~\ref{ex:maxcutcontinuousoptimizationproblem}).

\section{Convexity in Optimization}

Convexity is one of the most mathematically important concepts in continuous optimization as it tends to directly impact the complexity and solvability of the problem.

\begin{definition}\label{convexfun}
A function $f : \mathbb{R}^d \to \mathbb{R}$ is {\em convex} if for all $x,y \in \mathbb{R}^d$ and all $\lambda \in [0,1]$,
\begin{equation}
 \label{def:convex}
f(\lambda x + (1-\lambda)y) \le \lambda f(x) + (1-\lambda) f(y).
\end{equation}
\end{definition}
Geometrically, this means that the line segment connecting any two points on the graph of the function lies entirely on or above the graph\footnote{A good intuitive way to think about convexity it is that if an algorithm sits at a point $y$ and there is a point $x$ with a lower objective, the line between $y$ and $x$ must have lower objective than $y$, meaning that there is no barrier for local optimization algorithms (this will be made precise below).}.
We call $f$ {\em strictly convex} if the inequality in~\eqref{def:convex} becomes a strict inequality.

\begin{figure}[h]
\begin{center}
    \includegraphics[width=.7\textwidth]{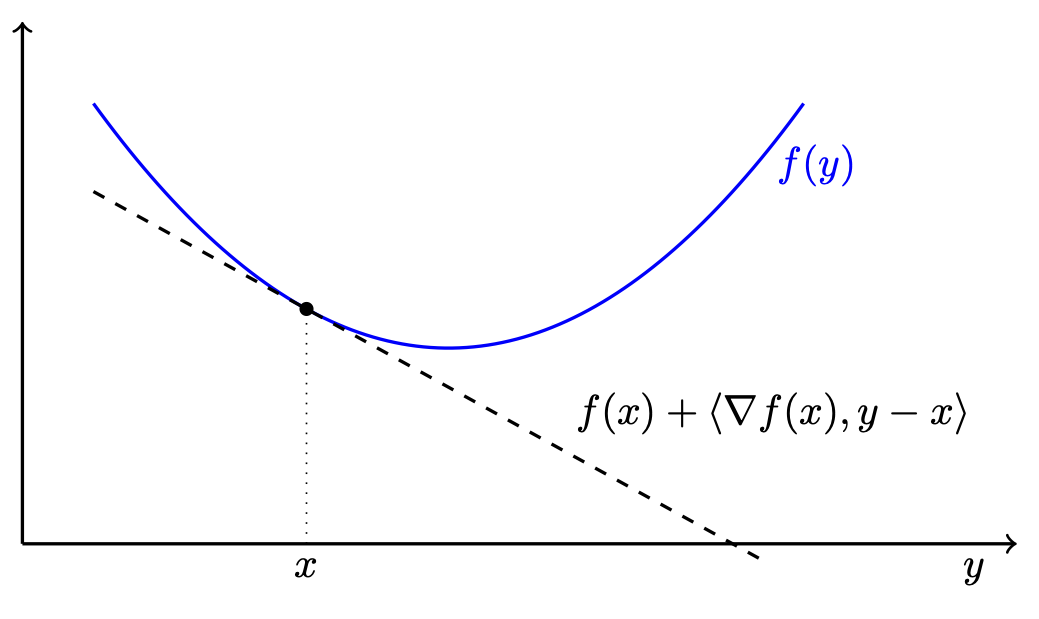}
    \caption{A convex function lies above its tangent at every point $x$.}
    \label{fig:plotconvex}
\end{center}
\end{figure}

If in addition $f$ is continuously differentiable, convexity is equivalent to the {\em first-order condition} (we leave the proof of this equivalence as an exercise):
\begin{equation}
 \label{def:geomconvex}
f(y) \ge f(x) +  \langle \nabla f(x), y - x\rangle, \quad \forall x,y \in \mathbb{R}^d.
\end{equation}
This means that the function $f$ is above its tangent at $x$, as illustrated in Figure~\ref{fig:plotconvex}.

Strengthening convexity by requiring  an additional
positive quadratic term on the right-hand side of~\eqref{def:geomconvex} leads to {\em strong convexity}.

\begin{definition}
    A differentiable function $f : \mathbb{R}^d \to \mathbb{R}$ is $\mu$-strongly convex ($\mu > 0$) if
\begin{equation}
 \label{def:stronglyconvex}
f(y) \ge f(x) + \langle \nabla f(x),  y - x \rangle + \frac{\mu}{2}\|y - x\|^2, \quad \forall x,y \in \mathbb{R}^d.
\end{equation}
\end{definition}
This assumption will later allow us to establish exponential convergence rates.

\begin{lemma}\label{le:minima}
(i)~Let $f: \R^d \to \R$ be a convex function. Then the set of minimizers of $f$ is convex and has cardinality $0, 1$, or $\infty$. (ii)~If in addition $f$ is (differentiable) $\mu$-strongly convex, then $f$
has exactly one minimizer.
\end{lemma}

\begin{proof}
Let $f$ be convex, and assume that $x^{*}$ and $x^{**}$ are two minimizers of $f$. By~\eqref{def:convex}, every convex combination is also a minimizer, which establishes the first part of the lemma. Now, assume that $f$ is $\mu$-strongly convex. Then~\eqref{def:stronglyconvex} implies $f$ that is lower bounded by a (strictly) convex
quadratic function. Hence there exists at least one minimizer $x^*$, which means $\nabla f(x^*)=0$.  It follows from~\eqref{def:stronglyconvex} that $f(x) > f(x^*)$ for every $x \neq x^*$. 
\end{proof}

\begin{figure}[h]
\begin{center}
    \includegraphics[width=.7\textwidth]{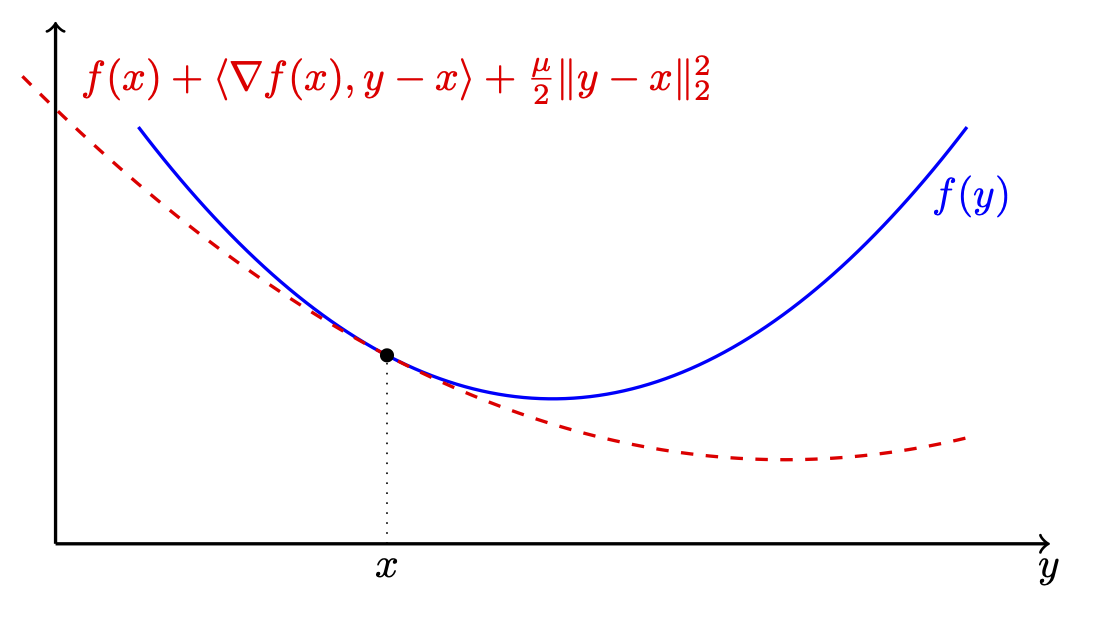}
    \caption{A $\mu$-strongly convex function}
    \label{fig:plotstronglyconvex}
\end{center}
\end{figure}

\subsection{Convex Optimization Problems}

Recall that a set $\mathcal{F}$ is a convex set if, for any two points $x, y \in \feas$ and any scalar $\lambda \in [0, 1]$, the convex combination $\lambda x + (1-\lambda) y$ also belongs to $\feas$.

\begin{definition}[Convex Optimization Problem]
A problem is a {\em convex optimization problem} if it satisfies two conditions:
\begin{enumerate}
\item[(i)] The feasible region $\feas$ is a convex set.
\item[(ii)] The objective function $f(x)$ must be convex\footnote{As stated in~\eqref{eq:optim}, throughout this chapter, we always consider the minimization of a function unless stated otherwise. If the problem is a maximization problem, the objective function $f(x)$ must be concave in order for the optimization problem to be convex (in Chapter~\ref{c:community} the problems will be written as maximization problems).}.
\end{enumerate}
\end{definition}

\begin{theorem}[Fundamental Theorem of Convex Optimization.]\label{th:convopt}
Let $f : \mathbb{R}^n \to \R$ be a convex function on a convex set $C \subseteq \R^n$.
If $x^* \in C$ is a local minimum of $f$ over $C$, then $x^*$ is also a global minimum.
If $f$ is furthermore strictly convex
, the global minimum is unique.
\end{theorem}

\begin{proof}
Assume $x^*$ is {\em not} a global minimum
Then there exists a point $y \in C$ such that
$$
f(y) < f(x^*).
$$
Since $C$ is convex, for any $\lambda \in [0,1]$, the point
$x_\lambda := (1-\lambda)x^* + \lambda y
$ belongs to $C$.
By convexity of $f$,
$$
f(x_\lambda)
\le
(1-\lambda)f(x^*) + \lambda f(y).
$$
Since $f(y) < f(x^*)$, we have
$$
(1-\lambda)f(x^*) + \lambda f(y)
<
(1-\lambda)f(x^*) + \lambda f(x^*) = 
f(x^*).
$$
Thus,
$$
f(x_\lambda) < f(x^*).
$$
As $\lambda \to 0^+$,
$$
x_\lambda \to x^*.
$$
Hence, for sufficiently small $\lambda > 0$, the point $x_\lambda$ lies inside any neighborhood $B_\varepsilon(x^*)$.
But we have shown
$$
f(x_\lambda) < f(x^*),
$$
which contradicts the assumption that $x^*$ is a local minimum.
The assumption that $x^*$ is not a global minimum must be false. Therefore,
$$
f(x^*) \le f(x) \quad \forall x \in C.
$$
Hence, $x^*$ is a global minimum.

The proof of the claim of uniqueness is left as an exercise (see Exercise~\ref{eq:ex_uniqueness}).

\end{proof}

Theorem~\ref{th:convopt} is quite useful, since it guarantees that any optimization algorithm that converges to a stationary point\footnote{The function $f$ has a stationary point at $x$ if the gradient of $f$ has norm zero at $x$.} is guaranteed to have found the best possible solution.

Convex problems are highly desirable: every local minimum is a global minimum, and efficient algorithms exist with strong guarantees.\footnote{It is worth mentioning that an optimization problem being convex is not sufficient to guarantee that it is efficient to solve: sometimes it is hard to check if a candidate solution is in the feasibility set $C$, e.g. imagine a convex feasibility set $C$ that is defined with exponentially many constraints.}
Many classical models in data science and machine learning  such as linear regression, compressive sensing, logistic regression, and support vector machines are convex optimization problems.

\subsection{Some important classes of convex optimization problems}

We briefly discuss several noteworthy special cases of convex optimization problems, which play an important role in practice and which we will encounter later in this book.

A convex optimization problem is often stated in the form
\begin{align}\label{eq:convexproblem}
\begin{split}
 \underset{x\in\R^n}{\text{minimize}} & \qquad  f(x)  \\
 \text{subject to} & \qquad g_i(x) \le 0, \quad \, i = 1, \dots, m, \\
 & \qquad Ax = b,
\end{split}
\end{align}
where the functions $f$, $g_i, i=1,\dots, m$ are convex.

The optimization problem~\eqref{eq:convexproblem} is called a {\em linear program} if the objective and constraint functions  are linear (affine).
A linear program is thus typically written in the form
\begin{align}\label{eq:linprog}
\begin{split}
 \underset{x\in\R^n}{\text{minimize}} & \qquad   \langle c,x\rangle  \\
 \text{subject to} & \qquad \langle g_i, x\rangle  \le c_i, \quad \, i = 1, \dots, m, \\
 & \qquad Ax = b, 
\end{split}
\end{align}
where $A \in \R^{r \times n}$, $c = \{c_i\}_{i=1}^m \in\RR^m$,  $g_1,\dots,g_m\in\R^n$, $b \in \R^r$. 

A {\em quadratic program} is also a special case of~\eqref{eq:convexproblem}, where the objective and the inequality
constraints are quadratic functions, i.e., they are polynomials in the $x$
variable of degree at most two. Linear programs and quadratic programs appear in Chapter~\ref{c:cs} (note that not all quadratics are convex, and so quadratic programs can also be non-convex).

In a {\em semidefinite program} (SDP) we minimize a linear function subject to the constraint that an affine combination of symmetric matrices is positive semidefinite. Such a constraint is nonlinear but convex, so semidefinite programs are convex optimization problems. Furthermore, efficient algorithms exist for SDPs~\cite{LVanderberghe_SBoyd_1996}. 
Denote by $\symS_n$  the space of all $n \times n$ real symmetric matrices.
The space is equipped with the Hilbert-Schmidt inner product 
$\langle A,B \rangle = \trace(A^T B)$.
An SDP can be stated in standard form as
\begin{align}\label{eq:sdp}
\begin{split}
 \underset{X\in\symS_n}{\text{minimize}} & \qquad   \langle C,X\rangle  \\
 \text{subject to} & \qquad \langle A_i, X\rangle = b_i, \quad \, i = 1, \dots, m, \\
 & \qquad X \succeq 0, 
\end{split}
\end{align}
where $C, A_1,\dots,A_m \in \symS_n$. 
Semidefinite programs will feature prominently in Chapter~\ref{c:community}. It is worth mentioning that~\eqref{eq:sdp} is not naturally written in form~\eqref{eq:convexproblem} but rather as an optimization problem over a convex cone $K$, in this case the cone of positive semidefinite matrices (a cone in $\RR^n$ is a set that is closed under non-negative scaling). While  Section~\ref{s:duality} will focus on problems of the form~\eqref{eq:convexproblem}, the arguments and derivations can be adapted to the setting of optimization over convex cones (such as the PSD cone). For example, dual optimization problems tend to involve the dual cone $K^\ast = \{y\in\RR^n: \langle y,x\rangle \geq 0, \forall x\in K\}$. Note that both the PSD cone (of PSD symmetric matrices), and the cone of non-negative vectors in $\RR^n$, are self-dual cones. In Chapter~\ref{c:community} we derive the dual optimization problem for a Semidefinite Program.

\subsection{Non-convex optimization}

\begin{figure}[h]
\begin{center}
    \includegraphics[width=.7\textwidth]{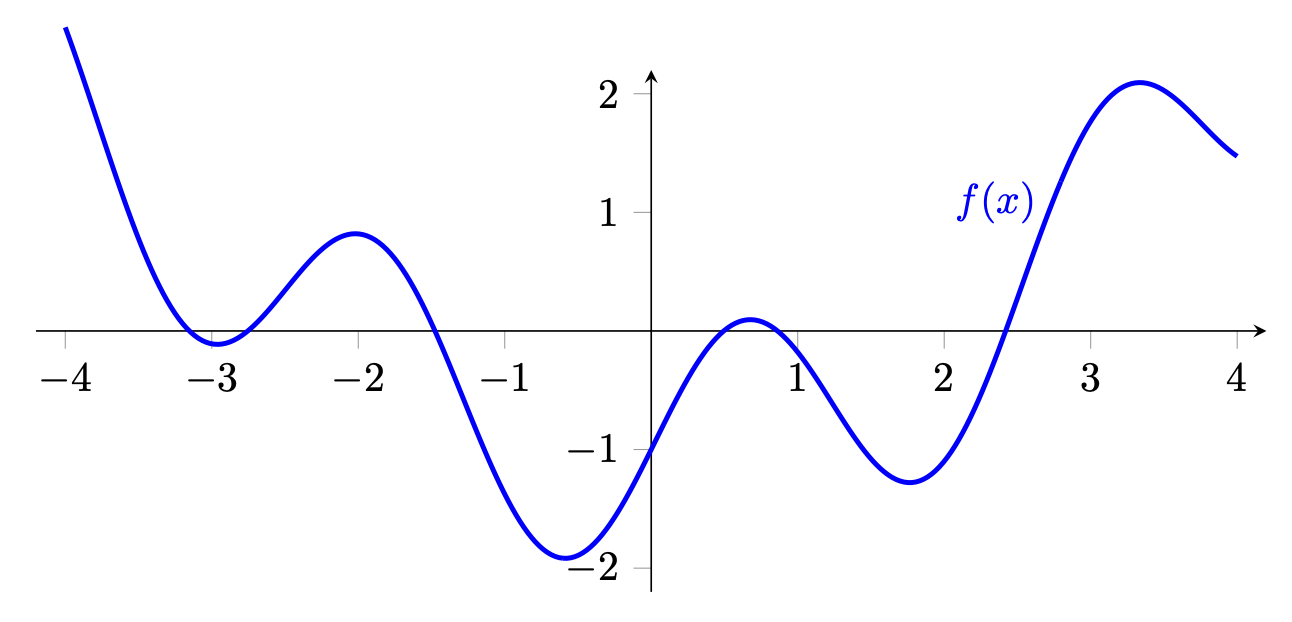}
    \caption{A nonconvex function. Nonconvex functions can have many local minima, even in one dimension.}
    \label{fig:plotnonconvex}
\end{center}
\end{figure}

An optimization problem is non-convex if either the objective function $f(x)$ is not convex, or the feasible region $\mathcal{F}$ is not convex (or both). Non-convex problems, common in data science and deep learning, may contain many local minima and saddle points, see Figure~\ref{fig:plotnonconvex} for an illustration.
These problems often require heuristic or iterative methods and seldom offer guarantees of global optimality. See also the convergence results for gradient descent in Section~\ref{s:gd:nonconvex}.

\begin{remark}[Smooth vs.\ non-smooth optimization]

A smooth optimization problem is characterized by having a differentiable objective function (and constraints). Gradient-based algorithms, which will be discussed in detail in Section~\ref{s:grad}, are typically effective for such problems.
Non-smooth problems involve objectives or constraints that are not differentiable everywhere. Examples include the $\ell_1$-penalty used in compressive sensing (see Chapter~\ref{c:cs}) or hinge loss in support vector machines (see Chapter~\ref{c:classification}). These require specialized methods such as subgradient algorithms or proximal techniques. 

\end{remark}

\section{Optimality conditions in unconstrained optimization}\label{s:optimality}
 
Optimality conditions provide mathematical criteria for determining whether a candidate solution of an optimization problem is locally or globally optimal. These conditions form the theoretical foundation for most optimization algorithms, as they describe the necessary (and sometimes sufficient) properties that optimal solutions must satisfy.

We start with unconstrained optimization problems.

\subsection{First-order necessary conditions (unconstrained)}

As the name suggests, first-order conditions rely on gradient information of the optimization function $f$.
The gradient of $f$ with respect to $x$ is a vector of partial derivatives
\begin{equation}
\nabla_x f(x) = \left( \frac{\partial f(x)}{\partial x_1}, \frac{\partial f(x)}{\partial x_2}, \dots, \frac{\partial f(x)}{\partial x_d} \right).
\label{gradient}
\end{equation}
If it is clear from the context we simply write $\nabla f(x)$ instead of $\nabla_x f(x)$.

\begin{theorem}
 \label{th:firstorder}
Let $f:\R^n\to\R$ be continuously differentiable. If $x^\star$ is a  local minimum of $f$, then
$$
\nabla f(x^\star)=0.
$$
\end{theorem}

\begin{proof}
Let $d\in\R^n$ be any direction and define
$$
\varphi(t)=f(x^\star+td).
$$
Since $x^\star$ is a local minimum, $t=0$ is a local minimizer of $\varphi$, hence
$$
\varphi'(0)=0.
$$
By the chain rule,
$$
\varphi'(0) = \nabla f(x^\star)^T d.
$$

Since this must hold for {\em all directions} $d$, the only vector satisfying
$$
\nabla f(x^\star)^T d =0 \quad \forall d
$$
is
$$
\nabla f(x^\star)=0.
$$
\end{proof}

The first-condition in Theorem~\ref{th:firstorder} provides a necessary, but not a sufficient criterion to characterize whether a function has a minimum at a specific point. If $\nabla f(x)=0$ we may or may not have a local extremum at $x$. We therefore call any point $x$ for which $\nabla f(x)=0$ a {\em critical point}.

\subsection{Second-order conditions (unconstrained)}

To find out whether a critical point is indeed a local extremum, we will utilize second-order conditions. 

\begin{figure}
\begin{center}
    \includegraphics[width=.5\textwidth]{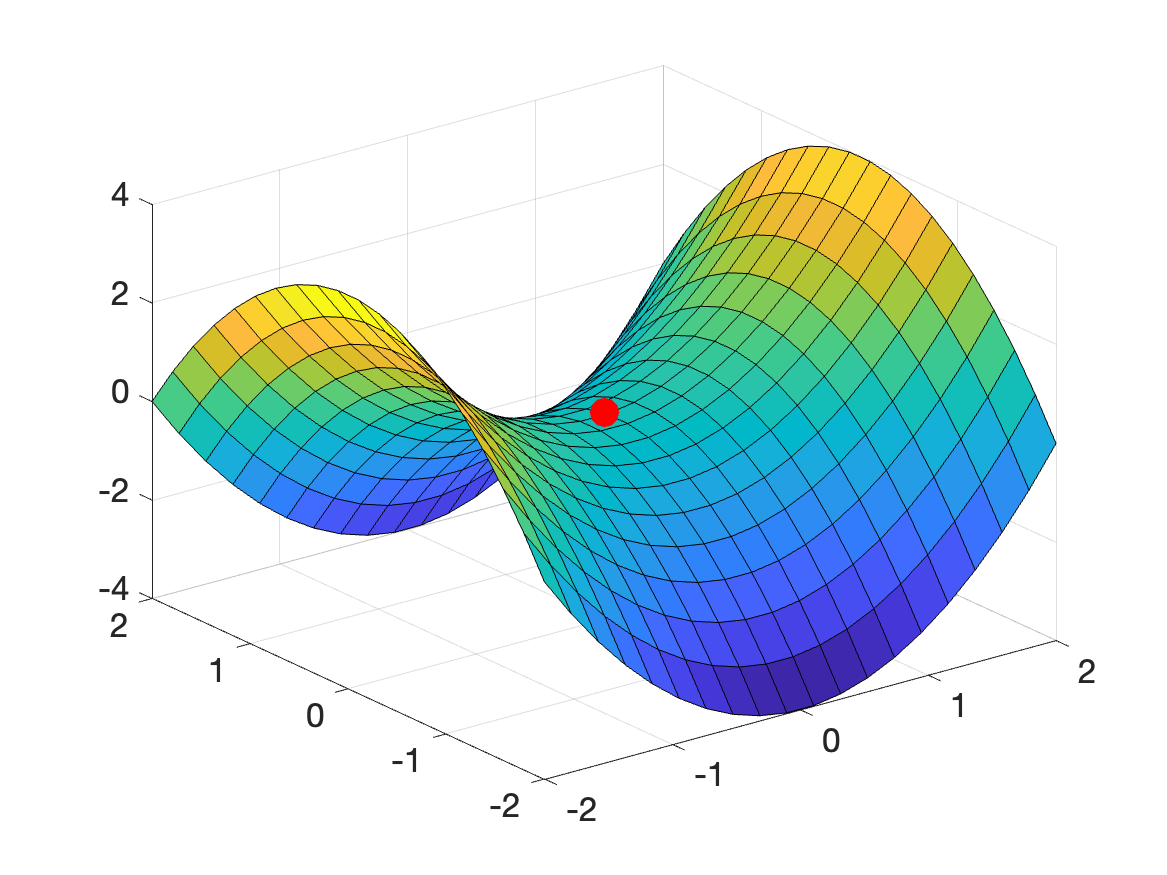}
    \caption{A saddle point (shown in red) on the graph of the function $f(x,y)=x^2-y^2$'}
    \label{fig:saddle}
\end{center}
\end{figure}

\begin{definition}
 \label{def:saddle}   
Let $f\in C^2$. A point $x^\star$ is a {\em saddle point} of $f$ if
$\nabla f(x^\star)=0$ but the Hessian $\nabla^2 f(x^\star)$ is indefinite.
\end{definition}
Thus, a saddle point is a critical point, but it is not a local extremum of the function, see also Figure~\ref{fig:saddle}. 

\begin{theorem}\label{th:secondorder}
If $f\in C^2$ and $x^\star$ is a local minimum, then
 $\nabla^2 f(x^\star)\succeq 0$ (i.e., the Hessian is positive semidefinite).
\end{theorem}

\begin{proof}
Consider the Taylor expansion
$$
f(x^\star + d) = f(x^\star) + \nabla f(x^\star)^T d +
 \frac{1}{2} d^T \nabla^2 f(x^\star) d + o(\|d\|^2).
$$
At a local minimum, for sufficiently small $d$,
$$
f(x^\star+d)-f(x^\star) \ge 0.
$$
Using $\nabla f(x^\star)=0$, we obtain
$$
\tfrac12 d^T \nabla^2 f(x^\star) d + o(\|d\|^2) \ge 0.
$$
Dividing by $\|d\|^2$ and letting $d\to 0$ yields
$$
d^T \nabla^2 f(x^\star) d \ge 0
$$
for all $d$. Thus the Hessian is positive semidefinite.

\end{proof}

A straightforward variation of Theorem~\ref{th:secondorder} is that if $x^*$ is a local maximum, then $\nabla^2 f(x^\star)\preceq 0$, i.e., the Hessian is negative semidefinite.

\begin{theorem}\label{th:hessian}
Let $f\in C^2$. If $\nabla f(x^\star)=0$ and $\nabla^2 f(x^\star)\succ 0$, then $x^\star$ is a {\em strict local minimum}.
\end{theorem}

\begin{proof}
Positive definiteness implies
$$
d^T \nabla^2 f(x^\star) d \ge c\|d\|^2
$$
for some $c>0$. Using the Taylor expansion again:
$$
f(x^\star+d)-f(x^\star)
= \frac{1}{2} d^T \nabla^2 f(x^\star) d + o(\|d\|^2)
\ge \frac{c}{2}\|d\|^2 + o(\|d\|^2).
$$
For sufficiently small $d$, the right-hand side is strictly positive. Thus $x^\star$ is a strict local minimizer. 
\end{proof}

Notice that if  $f\in C^2$ and the Hessian $\nabla^2 f$ is positive definite, then $f$ is strictly convex, but the converse is not necessarily true. Consider for example the function $f(x)=x^4$. Then it is easy to check that $f$ is strictly convex, but at $x=0$ we have $f''(x) =0$, which implies that the (in this case scalar-valued) Hessian is not positive definite.

\medskip
\noindent
\textbf{Example: Least squares.} In Chapter~\ref{c:linreg_ls} we analyzed the least squares problem from a statistical perspective. Now we will investigate it through an optimization lens and put the aforementioned optimality conditions to the test. We consider the least squares problem
\begin{equation}\label{optim:lsq}
    \min_x \frac{1}{2} \|Ax - b\|^2,
\end{equation}
where $A \in \R^{m \times n}$ with $m \ge n$.
A standard way to compute its solution
is via orthogonal projections, as is usually done in numerical linear algebra. Here, we follow instead the path of optimization by setting the gradient of $f(x)=\frac{1}{2} \|Ax - b\|^2$ to zero, as suggested by Theorem~\ref{th:firstorder}. 

\begin{figure}[h!]
\begin{center}
    \includegraphics[width=.65\textwidth]{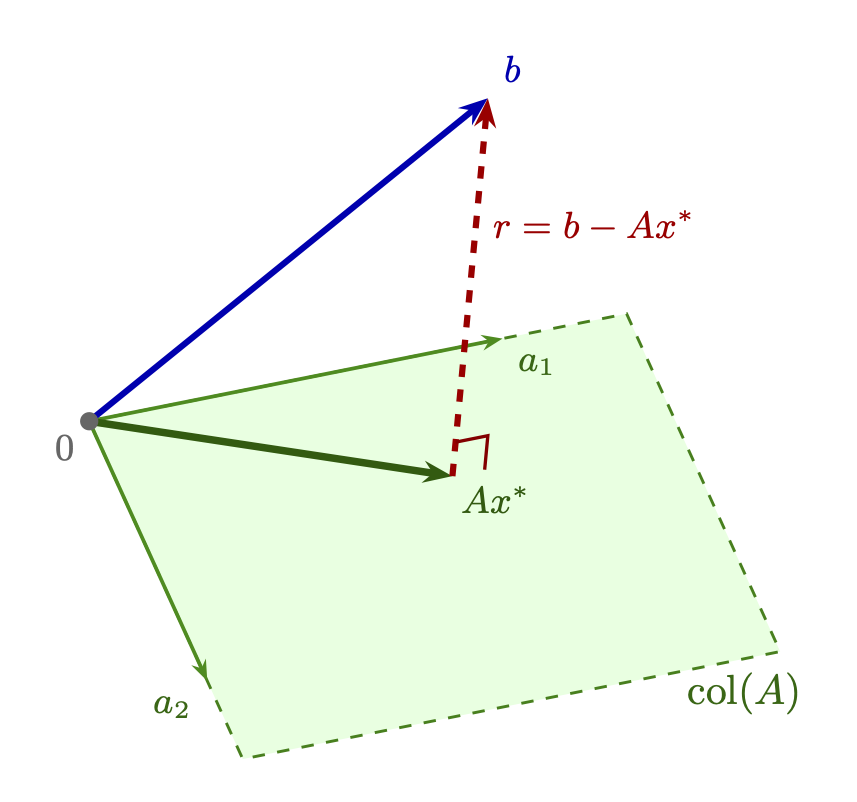}
    \caption{The vector $b$ is decomposed into its orthogonal projection $Ax^*$ onto $\operatorname{col}(A)$ (the column space of $A$), and the residual $r = b - Ax^*$ which is perpendicular to $\operatorname{col}(A)$. The least-squares solution $x^*$ minimizes $\|r\|$ and satisfies the normal equations $A^\top r = 0$.}
    \label{fig:lsqproj}
\end{center}
\end{figure}

To that end we first expand~\eqref{optim:lsq} into
$$  f(x) =  \frac{1}{2}\Big( x^T A^T A x - x^T A^T b - b^T Ax + b^T b\Big).$$
Next we compute the gradient of $f$ with respect to $x$. Standard vector calculus yields
$$\nabla f = A^T A x - A^T b,$$
Setting the gradient to zero gives
\begin{equation}\label{lsqnormal}
A^T A x - A^T b = 0 \quad \Longleftrightarrow \quad  A^T A x = A^T b.
\end{equation}
The equations on the right hand side of~\eqref{lsqnormal} are known as the {\em normal equations}. Let the residual be $r(x) = b - Ax$. The normal equations are equivalent to the orthogonality condition
$$
A^T r(x)=0 \quad\Longleftrightarrow\quad r(x)\perp \operatorname{col}(A).
$$
Thus, the columns of $A$ span a subspace $\operatorname{col}(A)\subset\R^n$, and the least-squares minimizer $x^*$ makes $A x^{*}$ the orthogonal projection of $b$ onto $\operatorname{col}(A)$, as illustrated in Figure~\ref{fig:lsqproj}

Equivalently,
$$
Ax^\star = P_{\operatorname{col}(A)} b,
$$
and the projection matrix (when $A$ has full column rank) is
$$
P_{\operatorname{col}(A)} = A(A^T A)^{-1}A^T.
$$
The matrix $A^T A$ is symmetric and if $A$ has full rank $n$, then $A^T A$
is  positive definite, and thus invertible. In this case
the solution to~\eqref{optim:lsq} is given by
\begin{equation}\label{sollsq}
    x^* = (A^T A)^{-1} A^T b.
\end{equation}
This $x^*$ minimizes $f$ because the Hessian $\nabla^2 f = A^T A$ is positive definite, so $f$ is strictly convex. Thus in this case, by Theorem~\ref{th:convopt}, $x^*$ is a {\em unique} global minimum.

\smallskip
\noindent
\textbf{Least squares least norm solutions:}
If $A$ does not have full column rank ($\rank(A) < n$), then 
$A^T A$ is symmetric positive {\em semidefinite} and singular. The normal equations $A^T Ax = A^T b$
may have many solutions (note that there must exists at least one solution since the column space of $A^T b$ is always contained in the column space of $A^T A$, which means that the system $A^\top Ax=A^\top b$ is always consistent).

However, for the least-squares problem there is always at least one canonical minimizer: The set of all solutions of the normal equations (i.e. all minimizers) is
\begin{equation}\label{eq:normal_kernel}
\{x_0 + z \,:\,z\in\ker(A)\},
\end{equation}
where $x_0$ is any particular solution satisfying $A^T A x_0 = A^T b$. Among all solutions of the form~\eqref{eq:normal_kernel} there is a unique solution of particular interest, namely the solution that has minimal 2-norm. This unique solution can be computed via  the Moore–Penrose pseudoinverse $A^\dagger$ which we already encountered in Section~\ref{ss:pinv} (recall that  $A^\dagger = (A^T A)^{-1}A^T$ when $\rank A = n$). More precisely,
$$
x^\dagger: = A^\dagger b
$$
is the solution of the optimization problem
\begin{align}\label{eq:optimpinv}
\begin{split}
 \text{minimize} & \qquad   \|x\|  \\
 \text{subject to} & \qquad A^T A x = A^T b.
\end{split}
\end{align}
If the system is consistent, the constraint $A^T A x = A^T b$ reduces to $Ax = b$.

The minimal 2-norm solution $x^\dagger$ satisfies
$$ x^\dagger \perp {\mathcal N}(A).$$
Thus, it avoids extra nullspace components that do not affect the output.

The minimum-2-norm solution corresponds to the least energy solution in physics, and to the
maximum likelihood estimate when the solution is assumed to have i.i.d.\ Gaussian priors.

\section{Constrained Optimization: Duality and KKT conditions} 
\label{s:duality}

We consider the optimization problem~\eqref{eq:optim}, which we restate here without loss of generality and with a slight abuse of notation in a  more convenient form regarding the constraints:
\begin{align}\label{eq:optim4}
\begin{split}
 \text{minimize} & \qquad   f(x)  \\
 \text{subject to} & \qquad g_i(x) \le 0, \quad \, i = 1, \dots, m, \\
 & \qquad h_j(x) = 0, \quad j=1,\dots,r. 
\end{split}
\end{align}
Its optimal value is denoted by $p^*$, and
we do not (yet) assume that the problem is convex.

\subsection{The Lagrangian and the Lagrangian dual}

The key idea in Lagrangian duality is to incorporate the constraints in~\eqref{eq:optim4} into the objective function by augmenting it with a weighted sum of the constraint functions.

To that end, we introduce the {\em Lagrangian} ${\mathcal L}: \R^n \times \R^m\times \R^r \to \R$ associated with~\eqref{eq:optim} as
$$
\mathcal{L}(x,\lambda,\nu)
= f(x) + \sum_{i=1}^m \lambda_i g_i(x)
+ \sum_{j=1}^r \nu_j h_j(x),
$$
where $\lambda = \{\lambda_i\}_{i=1}^m$ and $\nu = \{\nu_j\}_{j=1}^r$.
We refer to $\lambda_i\ge 0$, as {\em Lagrange multiplier} associated with the inequality constraint $g_i(x)\le 0$, and to $\nu_j$ as the Lagrange
multiplier associated with the  equality constraint $h_j(x)=0$. With a slight abuse of notation we write 
$\lambda \ge 0$ when this inequality holds entrywise, i.e., when $\lambda_i \ge 0, \forall i=1,\dots,m$.

We define the {\em Lagrange dual function} as
$$
\theta(\lambda,\nu) = \inf_{x\in \dom} \mathcal{L}(x,\lambda,\nu).
$$
The vectors $\lambda$ and $\nu$ are also
called the {\em dual variables} associated with the problem~\eqref{eq:optim}.
When the Lagrangian is unbounded below in $x$, the dual function takes on the
value $-\infty$, otherwise, $\theta(\lambda,\nu)$ is a
finite value. 

The following properties of the dual function will be very useful.

\begin{lemma}\label{le:duallowerbound}
The dual function $\theta(\lambda,\nu)$is jointly concave in $(\lambda,\nu)$. Moreover, for any $\lambda \ge 0$  and any $\nu$ we have
\begin{equation}\label{eq:duallowerbound}
    \theta(\lambda,\nu) \le p^*.
\end{equation}
\end{lemma}
\begin{proof} 
Note that for each fixed $x$, the function $\mathcal{L}(x,\lambda,\nu)$ is affine in $(\lambda,\nu)$, hence it is also concave in $(\lambda,\nu)$, because linear and affine functions are both convex and concave.
Since the dual function $\theta(\lambda,\nu)$ is the pointwise infimum of a family of affine
functions of $(\lambda,\nu)$, it is concave. Note that this conclusion holds even when the problem~\eqref{eq:optim} is not convex.

To prove the claim  about the lower bound, let $\lambda \ge 0$ and
assume that $\bar{x}$ is a feasible point for~\eqref{eq:optim4}, i.e., $g_i(\bar{x})  \le 0$, and $h_j(\bar{x})  = 0$. 

Now consider the Lagrangian for this feasible $\bar{x}$, assuming $\lambda \ge 0$:
$${\mathcal L}(\bar{x}, \lambda, \nu) = f(\bar{x}) + \sum_{i=1}^m \underbrace{ \lambda_i g_i(\bar{x})}_{\le 0} \, +  \, \sum_{j=1}^r \underbrace{ \nu_j h_j(\bar{x})}_{= 0},$$
where each term $\lambda_i g_i(\bar{x})$ is non-positive since $\lambda_i \ge 0$ and $g_i(\bar{x}) \le 0$.
Because the summation terms are either zero or non-positive, we have
\begin{equation}\label{eq:dual5}
{\mathcal L}(\bar{x}, \lambda, \nu) \le f(\bar{x}).
\end{equation}

The dual function is defined as the infimum of the Lagrangian over all $x \in \mathcal{D}$, which includes the subset of feasible points:
$$\theta(\lambda, \nu) = \inf_{x' \in \mathcal{D}} {\mathcal L}(x', \lambda, \nu) \le {\mathcal L}(x, \lambda, \nu).$$
Combining this with inequality~\eqref{eq:dual5} gives
$$\theta(\lambda, \nu) \le f(x).$$
Since $\theta(\lambda, \nu) \le f(x)$ holds for every feasible $x$, it must also be less than or equal to the minimum value attainable by $f(x)$ on the feasible set:
$$\theta(\lambda, \nu) \le \inf \{ f(x) \mid x \text{ is feasible} \} = p^*$$
Therefore, $\theta(\lambda, \nu) \le p^*$ for all $\lambda \ge 0$ and any $\nu$.

\end{proof}

Let us consider some examples for which we will derive the Lagrange dual function analytically.

\noindent
\textbf{Example: Underdetermined linear equations.}
We consider the problem~\eqref{eq:optimpinv} for consistent systems, i.e.,
\begin{align}\label{eq:optimpinv2}
\begin{split}
 \text{minimize} & \qquad   x^T x  \\
 \text{subject to} & \qquad Ax = b,
\end{split}
\end{align}
where $A \in \R^{r \times n}$. This problem has $r$ equality constraints (and no inequality constraints).
The Lagrangian is 
$$ \mathcal{L}(x,\nu) = x^T x + \nu^T (Ax-b),$$ 
and the dual function is given by $\theta(\nu) = \inf_{x \in \dom} \mathcal{L}(x,\nu)$. $\mathcal{L}$ is strictly convex as the reader may readily convince herself. Hence, we can find the unique minimizer via the first-order optimality condition 
$$\nabla_x \mathcal{L}(x,\nu) = 2x + A^T \nu = 0,$$
which gives $x = - \frac{1}{2} A^T \nu$.  The associated dual function is 
\begin{eqnarray*}\theta(\nu) =  \mathcal{L}\big(- \frac{1}{2} A^T \nu,\nu\big) &=& 
 \frac{1}{4} \nu^T A A^T \nu  + \nu^T ( -\frac{1}{2}A  A^T\nu - b)\\
&=&
-\frac{1}{4} \nu^T A A^T \nu  - \nu^T b.
\end{eqnarray*}
Lemma~\ref{le:duallowerbound} implies that for any $\nu \in \R^r$ we have the lower bound
$$ -\frac{1}{4} \nu^T A A^T \nu  - \nu^T b \le 
\inf_x \big\{\|x\|^2: Ax = b\big\}.$$

\medskip
\noindent
\textbf{Example: Dual of a linear program.}
Consider the linear program in standard form
\begin{align}\label{LPstandard}
\begin{split}
\text{minimize} \quad & c^T x \\
\text{subject to} \quad & Ax = b, \\
& x \ge 0,
\end{split}
\end{align}
where $x \in \mathbb{R}^n$, $A \in \mathbb{R}^{m \times n}$,
$b \in \mathbb{R}^m$, and $c \in \mathbb{R}^n$.
To compute the Lagrangian we introduce dual variables $\nu \in \mathbb{R}^m$ for the equality constraint $Ax=b$ (which we can rewrite as $Ax-b=0$), and
dual variables $\lambda \in \mathbb{R}^n$ for the inequality constraint $x \ge 0$ (which we can rewrite as $-x \le 0$).

The Lagrangian is
$$
\Lag(x,\nu,\lambda)
= c^T x + \nu^T (Ax - b) - \lambda^T x 
= (c + A^T \nu - \lambda)^T x - b^T \nu.
$$
Here, $\lambda \ge 0$ is required by dual feasibility.

The Lagrange dual function is defined as
$$
\theta(\lambda, \nu,)
= \inf_{x \in \mathbb{R}^n} \Lag(x,\lambda,\nu).
$$
We now compute this infimum.

Since the Lagrangian is affine in $x$, we distinguish two cases:

\noindent
Case 1: $c + A^T \nu - \lambda = 0$:
Then $L(x,\nu,\lambda) = - b^T \nu$ for all $x$, thus
$\theta(\lambda,\nu) = -b^T \nu$.

\noindent
Case 2: $c + A^T \nu - \lambda \neq 0$:
Then the affine function $(c + A^T \nu - \lambda)^T x - b^T \nu$ is unbounded below as a function of $x$, and therefore
$\theta(\lambda,\nu) = -\infty$.

Combining the two cases gives
\begin{equation}\label{LPdualfunction}
    \theta(\lambda,\nu)
=
\begin{cases}
-b^T \nu, & \text{if } c + A^T \nu - \lambda = 0, \\
-\infty, & \text{otherwise}.
\end{cases}
\end{equation}
The lower bound property~\eqref{eq:duallowerbound} is nontrivial only when $\lambda$ and $\nu$ satisfy $\lambda \ge 0$ and
$c + A^T \nu - \lambda = 0$. When this happens, then 
$-b^T \nu$ is a lower bound on the optimal
value of the linear program~\eqref{LPstandard}.

\subsection{The Lagrangian dual problem}

Lemma~\ref{le:duallowerbound} tells us that for each pair $(\lambda,\nu)$ the Lagrange dual function gives us a lower bound on the optimal value $p^*$
of the optimization problem~\eqref{eq:optim}. This raises the question, what the best lower bound is that we can derive  from the Lagrange dual function.

In other words, we are concerned with the so-called {\em Lagrange dual problem} associated with problem~\eqref{eq:optim} 
\begin{align}\label{eq:dualproblem}
\begin{split}
 \text{maximize} & \qquad   \theta(\lambda,\nu)  \\
 \text{subject to} & \qquad \lambda \ge 0,
\end{split}
\end{align}
where the mazimization is sought with respect to the variables $\lambda$ and $\nu$.
The problem~\eqref{eq:optim} is also referred to as the {\em primal problem}.
We refer to the pair $(\lambda^*,\nu^*)$ as {\em dual optimal} or {\em optimal Lagrange multipliers}
if they are the maximizer of~\eqref{eq:dualproblem}.

An important point is that the Lagrange dual problem~\eqref{eq:dualproblem} is {\em convex}, since the objective to be maximized is concave and the constraint is convex, and this is true
whether or not the primal problem~\eqref{eq:optim} is convex.

\medskip
To put this concept to test, let us compute the Lagrange dual problem for a linear program. We already computed the Lagrangian dual function for a linear program, resulting in~\eqref{LPdualfunction}. The Lagrange dual problem of the linear program~\eqref{LPstandard} is to maximize this dual function $\theta$ subject to $\lambda \ge 0$. That means we need to solve
\begin{align}\label{eq:dualproblem1}
\begin{split}
 \text{maximize} & \quad \theta(\lambda,\nu) = 
 \begin{cases}
-b^T \nu, & \text{if } A^T \nu - \lambda +c = 0, \\
-\infty, & \text{otherwise},
\end{cases}\\
 \text{subject to} & \quad \lambda \ge 0.
\end{split}
\end{align}
Since $\theta$ is finite only when $A^T \nu - \lambda +c = 0$, we can make the equality constraints explicit via the following {\em equivalent} problem
\begin{align}\label{LPdual2}
\begin{split}
\text{maximize} \quad & -b^T \nu \\
\text{subject to} \quad & A^T \nu -\lambda + c = 0, \\
& \lambda \ge 0,
\end{split}
\end{align}
which in turn can be written as the following linear program in inequality form:
\begin{align}\label{LPdual3}
\begin{split}
\text{maximize} \quad & -b^T \nu \\
\text{subject to} \quad & A^T \nu + c \ge 0. 
\end{split}
\end{align}

\begin{remark}[Duality in Semidefinite Programming]
    One can construct the dual of a semidefinite program with a similar construction. The main point to note is that the variables $\lambda$ in~\eqref{LPdual2}, dual variables to the constraints $x\geq 0$ can be viewed as being in a dual cone: Let $Q_+$ denote the cone of non-negative vectors in $\RR^n$, then the constraint $x\geq 0$ corresponds to $x\in Q_+$; the dual variables are then in the dual cone $Q_+^\ast$ (in this case $Q_+^\ast=Q_+$). In a semidefinite program a matrix variable $X$ is in $K$ the cone of symmetric PSD matrices. It is also self-dual so this variable will have a dual variable $Q\in K$. See Chapter~\ref{c:community} for an actual derivation of the dual of an SDP.
\end{remark}

\subsubsection{Weak duality and strong duality}\label{ss:weakstrong}
We denote the optimal value of the Lagrange dual problem by $d^*$. By definition, $d^*$ is the best lower bound for $p^*$ that we can compute from the Lagrange dual
function. Moreover, we have the straightforward inequality
\begin{equation}
 \label{weakduality}   
\text{(dual optimum) } d^* \le p^* \text{ (primal optimum)}.
\end{equation}
And by virtue of Lemma~\ref{le:duallowerbound} this bound 
holds even if the original problem is not convex. This property is also called {\em weak duality}.

The weak duality inequality~\eqref{weakduality} also holds when $p^*$ and $d^*$ are infinite: if the primal problem is unbounded below, i.e., $p^* = -\infty$, this implies $d^* = -\infty$, which means  the Lagrange dual problem is infeasible;  if the
dual problem is unbounded above, i.e., $d^* = \infty$, this implies $p^* = \infty$, which in turn says that the
primal problem is infeasible.

\begin{definition}\label{dualitygap}
The difference $p^* - d^*$ is called the 
{\em optimal duality gap} of the original
problem.
\end{definition}
Thus, the optimal duality gap (often with a slight abuse of terminology simply called duality gap), which is always non-negative, gives the gap between the optimal value of the primal problem
and the best lower bound on it that can be obtained from the
Lagrange dual function.

We say that {\em strong duality} holds if the duality gap is zero, that is if we have the equality
\begin{equation}
 \label{strongduality}   
d^* = p^*.
\end{equation}

Unlike weak duality, strong duality does not hold in general.  Consider for example the problem
\begin{align*}
\text{minimize}  & \quad  x \\
\text{subject to} & \quad x^2 \ge 1.
\end{align*}
The constraint implies
$
x \in (-\infty,-1] \cup [1,\infty)$.
The objective is minimized at
$x^* = -1$ with optimal value $p^* = -1$.

To compute the Lagrangian dual, we rewrite the constraint as
$$
g(x) = 1 - x^2 \le 0.
$$

The Lagrangian is
$$
\mathcal{L}(x,\lambda) = x + \lambda(1 - x^2),
\qquad \lambda \ge 0.
$$

The dual function is
$$
\theta(\lambda) = \inf_x \big[ x + \lambda(1 - x^2) \big].
$$

If $\lambda > 0$, the term $-\lambda x^2$ dominates and
  $\theta(\lambda) = -\infty$.
If $\lambda = 0$. then $\theta(0) = \inf_{x} x = -\infty$.
Thus, $\theta(\lambda) = -\infty$ for all $\lambda \ge 0$.
The dual problem is
$$
d^* = \sup_{\lambda \ge 0} \theta(\lambda) = -\infty.
$$
Hence, the primal optimal value is $p^* = -1$, while
the dual optimal value is $d^* = -\infty$,
so there is a  strict (and rather large) duality gap.

\medskip

Sometimes the dual problem can be easier to solve compared with the primal problem and the primal solution can be constructed from the dual solution.
Thus, we are interested in understanding which conditions can guarantee that 
strong duality holds. A significant body of work in the optimization literature has been devoted to exploring 
different possibilities for these conditions. Among the most prominent conditions are {\em Slater's Condition} and the 
{\em Karush–Kuhn–Tucker Conditions}, which we will explore in the following sections.

\subsubsection{Max-min characterization of weak and strong duality}\label{ss:minmax}

In essence, at the core of duality is the question of whether minimizing after maximizing gives the same result as maximizing after minimizing. We will make this slogan now more precise.

For convenience we consider the primal problem without equality constraints (it is straightforward to extend our derivations to include equality constraints):
\begin{align}\label{eq:primalnoequality}
  p^* = \min_{x \in \dom} \quad & f(x) \notag \\
  \text{subject to} \quad & g_i(x) \leq 0, \quad i = 1, \dots, m.
\end{align}
Notice that for any $x \in \dom$ the Lagrangian $\Lag(x,\lambda)$ satisfies
$$
  \sup_{\lambda \geq 0} \Lag(x, \lambda)
  = \sup_{\lambda \geq 0} \left[ f(x) + \sum_{i=1}^{m} \lambda_i g_i(x) \right]
  =
  \begin{cases}
    f(x)   & \text{if } g_i(x) \leq 0 \text{ for all } i, \\
    +\infty & \text{otherwise.}
  \end{cases}
$$
Indeed, if any $g_i(x) > 0$ we may send $\lambda_i \to +\infty$ to make the expression
unbounded; if $x$ is feasible all terms $\lambda_i g_i(x) \leq 0$, and the supremum
is achieved at $\lambda = 0$.

Therefore the primal problem can be reformulated as
\begin{equation}\label{eq:primal_minmax}
 p^* = \min_{x \in \dom} \sup_{\lambda \geq 0} \Lag(x, \lambda).
\end{equation}
By definition of the dual function and dual optimal value we have that
\begin{equation}\label{eq:dual_maxmin}
  d^* = \sup_{\lambda \geq 0} \inf_{x \in \dom} \Lag(x, \lambda).
\end{equation}
Hence, we can express weak duality as
\begin{equation}\label{eq:weakminmax}   
\sup_{\lambda \geq 0} \inf_{x \in \dom} L(x, \lambda)\le \inf_{x \in \dom} \sup_{\lambda \geq 0}  L(x, \lambda),
\end{equation}
and strong duality as the equality
$$
\sup_{\lambda \geq 0} \inf_{x \in \dom} L(x, \lambda) = \inf_{x \in \dom} \sup_{\lambda \geq 0}  L(x, \lambda).
$$

Weak duality $d^* \leq p^*$ is actually an immediate consequence of the following universal
inequality, which requires \emph{no assumptions} on $\Lag$ or the sets involved.

\begin{theorem}[Max-Min Inequality]
  For any function $\phi : {\mathcal X} \times {\mathcal Y} \to \R$ and any sets ${\mathcal X}, {\mathcal Y}$:
  $$
    \sup_{y \in {\mathcal Y}} \inf_{x \in {\mathcal X}} \phi(x, y)
    \leq
    \inf_{x \in {\mathcal X}} \sup_{y \in {\mathcal Y}} \phi(x, y).
  $$
\end{theorem}

\begin{proof}
  Fix any $x_0 \in {\mathcal X}$ and $y_0 \in {\mathcal Y}$. Then
  $$
    \inf_{x \in {\mathcal X}} \phi(x, y_0) \leq \phi(x_0, y_0) \leq \sup_{y \in {\mathcal Y}} \phi(x_0, y).
  $$
  Taking the supremum over $y_0$ on the left-hand side and the infimum over $x_0$
  on the right-hand side gives the result.
\end{proof}

Strong duality, in turn, is equivalent to the existence of a \emph{saddle point}.

\begin{theorem}\label{th:saddleploint}
  The following are equivalent:
  \begin{enumerate}
    \item Strong duality holds and both the primal and dual optima are attained.
    \item There exists a saddle point $(x^*, \lambda^*)$ of $\Lag$, i.e.,
    $$
      \Lag(x^*, \lambda) \leq \Lag(x^*, \lambda^*) \leq \Lag(x, \lambda^*)
      \quad \forall\, x \in \dom,\; \lambda \geq 0.
    $$
  \end{enumerate}
  Moreover, at any saddle point, $\Lag(x^*, \lambda^*) = p^* = d^*$.
\end{theorem}

We leave the proof of this theorem as an exercise. The saddle point characterization of strong duality shows that the solution to the optimization problem is an equilibrium point where the ``force'' of the constraints (the dual variables) perfectly balances the ``pull'' of the objective function.

\subsubsection{Strong duality and Slater's condition}

In convex optimization, strong duality is not guaranteed by default, though it holds for the vast majority of problems satisfying basic regularity conditions. The conditions (beyond convexity), under
which strong duality holds are called constraint qualifications. One such constraint qualification is  introduced in the next
theorem~\cite{rockafellar2015convex,LVanderberghe_SBoyd_book}.

\begin{theorem}[Slater's Theorem]\label{th:slater}
Consider the convex optimization problem
\begin{align} \label{eq:slaterproblem}
\underset{x \in \mathcal{D}}{\operatorname{minimize}} \quad & f(x) \notag \\ \text{subject to} 
    \quad & g_i(x) \leq 0, \quad i = 1, \dots, m \\ & Ax = b, \quad A \in \R^{p \times n,}
\end{align}
where the minimization is over $x \in \dom$.
Assume the problem is convex.
If there exists a point $\tilde{x} \in  \operatorname {relint} (\dom)$, the relative interior\footnote{The relative interior of a set $\dom$ (usually denoted by $\operatorname {relint} (\dom)$) is defined as its interior within the affine hull of $\dom$. For example, if $\dom$ is a  line segment in a two-dimensional ambient space, then $\operatorname {int} (\dom)$ is empty, but $\operatorname {relint} (\dom)$  is the line segment without its endpoints.} of $\mathcal{D}$, such that
\begin{equation}\label{Slaterscondition}
    g_i(\tilde{x}) < 0, \quad i = 1, \dots, m \quad \text{and} \quad A\tilde{x} = b,
\end{equation}
then strong duality holds ($p^* = d^*$). Furthermore, if $p^* > -\infty$, there exists a dual optimal point $(\lambda^*, \nu^*)$ that attains the dual supremum.
\end{theorem}

\begin{figure}[h]
\begin{center}
\includegraphics[width=70mm]{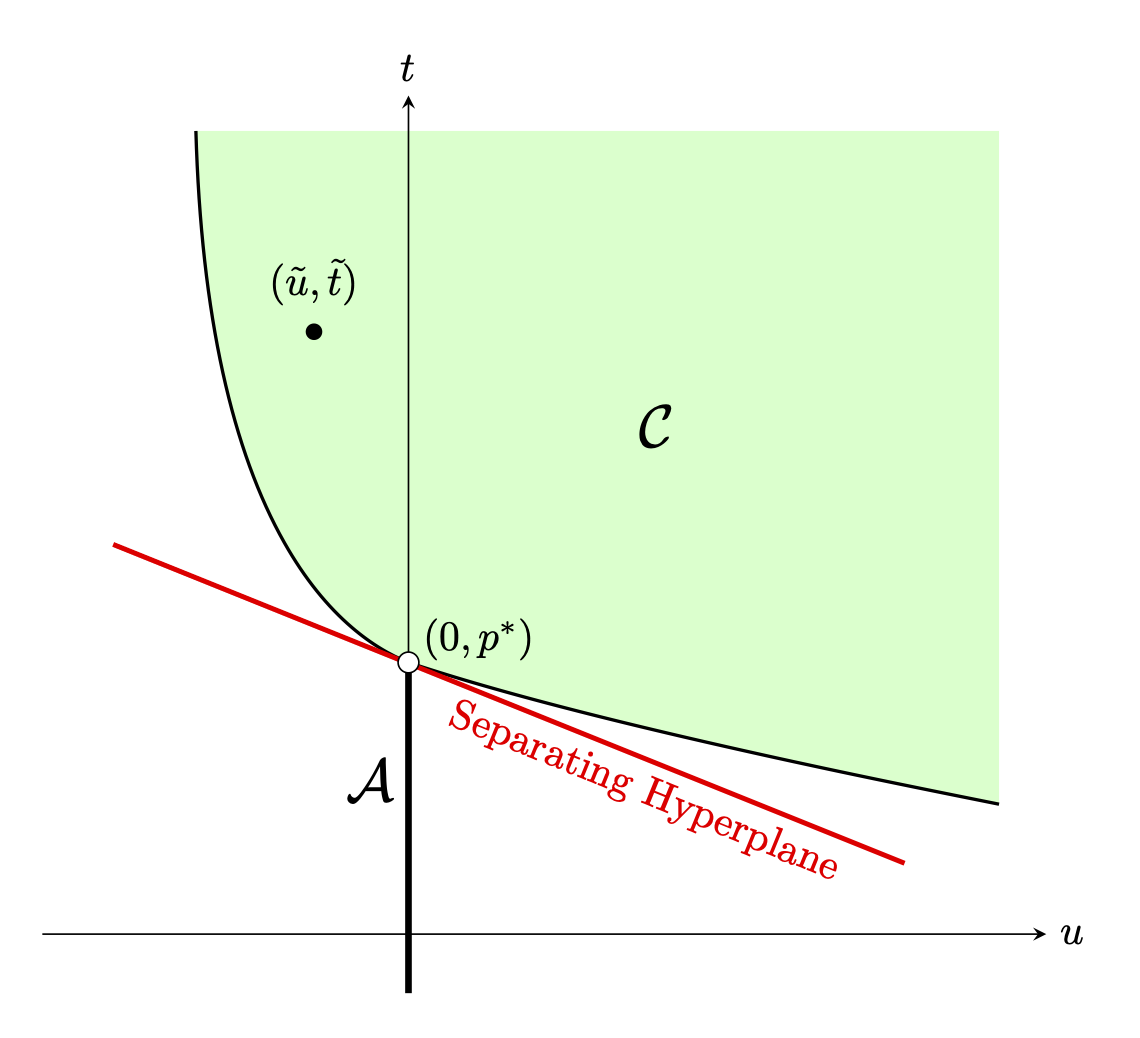}
\caption{The idea of the proof of Slaters' Theorem for a problem with one inequality constraint.  The two sets ${\mathcal C}$ and ${\mathcal A}$ are convex and non-intersecting, hence there exists a separating hyperplane. Slater's condition ensures
that any separating hyperplane must be non-vertical, since it must pass
to the left of the point $(\tilde{u},\tilde{t}) = (g_1(\tilde{x}),f(\tilde{x}))$, where $\tilde{x}$ is strictly feasible. If the hyperplane is not vertical, one can divide by its slope to get the Lagrange multipliers.}
\label{fig:slater}
\end{center}
\end{figure}

The conditions~\eqref{Slaterscondition} are known as {\em Slater's condition}. 
Geometrically, Slater's condition means that
the feasible set has nonempty interior relative to its affine hull. This ``thickness'' is exactly what allows separating hyperplanes to exist at the boundary without degeneracy, as the proof will establish rigorously, see also Figure~\ref{fig:slater}.

\begin{proof}
Define the set
$$
\mathcal{C}
:= \left\{ (u,t) \in \mathbb{R}^m \times \mathbb{R}\,\, \middle| 
\,\,\exists x \in \dom \text{ s.t. }\,\,
\begin{aligned}
& g_i(x) \le u_i,\ i=1,\dots,m, \\
& Ax=b, \\
& f(x) \le t
\end{aligned}
\right\}.
$$
Since $f$ and $g_i, i=1,\dots,m$ are convex, $\mathcal{C}$ is a convex set.
We emphasize that every point $(u,t) \in \mathcal{C}$
arises from an $x$ that satisfies all equality constraints exactly. Hence, we are working in an affine subspace determined by $Ax=b$.

Note that
$p^* = \inf \{ t: (0,t) \in \mathcal{C} \}$ and that
the point $(0,p^*)$ is not in the {\em interior} of $\mathcal C$. Indeed, for any $\epsilon > 0$, the point $(0, p^* - \epsilon)$ is not in $\mathcal{C}$ by definition of the infimum.
Now define the set
$$
\mathcal{A} := \{ (0,t) \mid t < p^* \}.
$$
Then $\mathcal{A}$ is a convex set, $\mathcal{A} \cap \mathcal{C} = \emptyset$ (a feasible point achieving $t < p^*$ while satisfying all constraints does not exist), and $\mathcal{C}$ has nonempty relative interior due to Slater’s condition.
Indeed, since $\tilde{x} \in \operatorname{relint} \dom$ and satisfies strict inequalities $g_i(\tilde{x}) < 0$, it follows that
$$
(g_1(\tilde{x}),\dots,g_m(\tilde{x}),f(\tilde{x})) \in \operatorname{relint} \mathcal{C}.
$$

By the Separating Hyperplane Theorem~(see e.g.~\cite[Chapter 2]{LVanderberghe_SBoyd_book}), there exists nonzero vector $(\lambda,\mu) \in \mathbb{R}^m \times \mathbb{R}$,
$\lambda \ge 0, \mu > 0$,
and a scalar $\alpha$ such that
\begin{align}    
\lambda^T u + \mu t \ge \alpha
\quad & \text{for all } (u,t) \in \mathcal{C},  \label{slatereq1}
\\
\lambda^T u + \mu t \le \alpha
\quad & \text{for all } (u,t) \in \mathcal{A}.\label{slatereq2}
\end{align}
Since $u = 0$ for all $(u,t) \in \mathcal{A}$,
 condition~\eqref{slatereq2} reduces to
\begin{equation}
\label{slartereq3}
\mu t \le \alpha \quad \forall t < p^*.
\end{equation}

We now use Slater’s condition  to prove that the hyperplane cannot be ``vertical'' (i.e., we must have $\mu \neq 0$).
If $\mu < 0$, then letting $t\to -\infty$ in~\eqref{slartereq3} implies $\mu t \to +\infty$, contradicting finiteness of $\alpha$. Hence $\mu \ge 0$. To show that $\mu >0$ we assume for
 contradiction that $\mu=0$. Then~\eqref{slartereq3} gives $\alpha \ge 0$, and~\eqref{slatereq1} becomes
$$
\lambda^T u \ge \alpha \ge 0
\quad \forall (u,t)\in\mathcal C.
$$
By Slater’s condition, there exists $\tilde{x}$ with
$$
u_i = g_i(\tilde x) < 0,\quad A\tilde x=b.
$$
Hence $(\tilde u,\tilde t)\in\mathcal C $ with $\tilde u \in \mathrm{int}(\mathbb{R}^m_-)$. Then $\lambda^T \tilde u \ge 0$,
but since $\lambda \ge 0$ and $\tilde u < 0$, this implies
$\lambda = 0$.
Thus $(\lambda,\mu)=(0,0)$, contradicting nontrivial separation.
Therefore, $\mu > 0$.

Equation~\eqref{slartereq3} implies that $\alpha \ge \mu p^*$.
By definition of $p^*$, there exists a sequence $t_k \downarrow p^*$ such that $(0,t_k)\in\mathcal C$.
Applying~\eqref{slatereq1} yields $\mu t_k \ge \alpha$
and taking limits gives $\mu p^* \ge \alpha$.
Hence $\alpha = \mu p^*$.

Now we construct a dual optimal point. For any feasible $x$ (i.e. $Ax=b$),
$$
\sum_{i=1}^m \lambda_i g_i(x) + \mu f(x) \ge \mu p^*.
$$
We divide by $\mu>0$ and define
$$
\lambda_i^* = \frac{\lambda_i}{\mu} \ge 0.
$$
Then
$$
f(x) + \sum_{i=1}^m \lambda_i^* g_i(x) \ge p^*
\quad \forall x \text{ with } Ax=b.
$$
Adding $\nu^T (Ax-b)=0$, we obtain
$$
L(x,\lambda^*,\nu) \ge p^*
\quad \forall x.
$$
Taking the infimum over $x$,
$$
\theta(\lambda^*,\nu) \ge p^*.
$$
But weak duality always gives
$$
\theta(\lambda,\nu) \le p^* \quad \forall (\lambda,\nu),
$$
so we conclude
$$
\theta(\lambda^*,\nu^*) = p^*.
$$
Thus
$d^* = p^*$ (strong duality), and
the dual optimum is attained at $(\lambda^*,\nu^*)$.

\end{proof}

Slater's condition  plays a {\em dual role} in the proof: 
{\em Structurally}, it guarantees $\operatorname{relint}(\mathcal{C}) \neq \emptyset$, making the supporting hyperplane theorem applicable. 
{\em Analytically}, the Slater point $\tilde{x}$ is used a second time to rule out $\mu = 0$, forcing the separating hyperplane to be non-degenerate in the objective direction.

Without Slater's condition, $\mathcal{C}$ might still have nonempty relative interior, but we would have no guarantee. And more critically, we would lose the second use of $\tilde{x}$ that forces $\mu > 0$. Both uses are essential, and both stem from the same strict feasibility assumption.

\subsection{Karush–Kuhn–Tucker (KKT) Conditions}
\label{ss:KKT}

For unconstrained optimization, optimality is at least in part characterized by the first-order condition $\nabla f(x^*) = 0$.
For constrained problems, the KKT conditions play an analogous role, since they describe when no feasible descent direction exists and they encode how constraints interact with the objective. In this sense, KKT is a natural extension of first-order optimality to constrained optimization\footnote{A useful example to keep in mind to build intuition is one where a point $x$ is a local optimum because the direction of steepest descent is perpendicular to the constraints; thus to improve one would need to leave the feasible set. The KKT conditions provide formal rigorous versions of this intuition.}.

Before deriving the KKT conditions, we  introduce some helpful tools.

\smallskip
\noindent
\textbf{Complementary slackness:}
We say that an inequality constraint is {\em slack} when it is satisfied with strict inequality at the optimum, and the constraint is {\em tight}  when it is satisfied with equality at the optimum. If the constraint has ``slack,'' it cannot exert any ``force'' on the optimal solution\footnote{We will encounter this concept again in Section~\ref{s:svm} in the context of Support Vector Machines where the decision boundary used for classification is determined solely by the points known as support vectors. Data points that are far from this boundary have ``slack'' and do not affect the orientation of the hyperplane that defines the boundary.}.

{\em Complementary slackness} is the condition that at optimality, for each inequality constraint $i$ we have 
$$\lambda_i^* g_i(x^*) = 0.$$
Since $\lambda_i^* \geq 0$ and $g_i(x^*) \leq 0$, this says that at least one of the two must be zero for every $i$. There are exactly two cases: \\
Case 1: The constraint is active (tight): $g_i(x^*) = 0$. The constraint is binding at the optimum, and $\lambda_i^*$ is free to be positive -- the dual variable is ``switched on'' and is actively penalizing that constraint.\\
Case 2: The constraint is inactive (slack): $g_i(x^*) < 0$. The constraint has room to spare, and so $\lambda_i^* = 0$ -- the dual variable is ``switched off'' and contributes nothing to the Lagrangian.

\medskip

Assume now that strong duality holds, i.e., the duality gap is zero. Let $x^*$ be a primal optimal and $(\lambda^*,\nu^*)$ be a dual optimal point. We compute
\begin{align*}
f(x^*) & = \theta(\lambda^*,\nu^*) \\
& = \inf_x \Big( f(x) + \sum_{i=1}^m \lambda^{*}_i g_i(x)
+ \sum_{j=1}^r \nu^{*}_j h_j(x)\Big) \\
& \le f(x^*) + \sum_{i=1}^m \lambda^{*}_i g_i(x^*)
+ \sum_{j=1}^r \nu^{*}_j h_j(x^*) \\
& \le  f(x^*).  
\end{align*}
Here, the equality in the first line is due to the assumption that the duality gap is zero, the second line is just the definition of the Lagrange dual function, the third line follows from the fact the the infimum of the
Lagrangian over $x$ is less than or equal to its value at $x = x^*$, and the fourth line follows from $\lambda^{*}_i \ge 0$, $g_i(x^*) \le 0, i = 1,\dots,m$, and $h_j(x^*) = 0, j=1,\dots,r$. Therefore, the two inequalities above must hold with equality.

A take-away of the fact that the inequality in the third line is actually an equality is that $x^*$ minimizes $\mathcal{L}(x,\lambda,\nu)$ over $x$. Note that $\mathcal{L}(x,\lambda,\nu)$ could have other minimizers as well.

The fact that the inequality in the fourth line is actually an equality means that $\sum_{i=1}^m \lambda^{*}_i g_i(x^*)=0$. Since each term in this sum is non-positive, we draw the conclusion that
\begin{equation}\label{eq:slackness}
    \lambda^{*}_i g_i(x^*) = 0, \quad i=1,\dots,m,
\end{equation}
i.e., we have complementary slackness.

If we had a slack constraint with $\lambda_i > 0$ for some $i$, we could slightly improve the objective function by moving in the corresponding direction of $-\nabla f$ without violating that constraint. This would contradict the assumption that $x^*$ was already optimal. Whereas if $\lambda_i >0$, the constraint must be active (i.e., $g_i(x^*) = 0$). Consequently, examining the optimal dual variables allows us to identify which primal constraints are active at the solution.

\medskip
We assume that the functions $f$, $g_1,\dots,g_m$, $h_1, . . . , h_r$ are continuously differentiable, but not necessarily  convex (we will address the convex case later). Furthermore, let $x^*$ and $(\lambda^*, \nu^*)$ 
be any primal and dual optimal points with zero duality gap. Since $x^*$ minimizes $\mathcal{L}(x,\lambda^*,\nu^*)$ over $x$, it follows that the gradient  $\nabla \mathcal{L}(x,\lambda^*,\nu^*)$ (with respect to $x$) vanishes at $x^*$:
$$ \nabla f(x^*) + \sum_{i=1}^m \lambda^*_i \nabla g_i(x^*)
+ \sum_{j=1}^r \nu^*_j \nabla h_j(x^*) = 0.$$
This provides us with the set of conditions
\begin{align}\label{eq:kkt}
\begin{split}
g_i(x^*) & \le 0, \quad i=1,\dots,m  \\
h_j(x^*) & = 0, \quad j=1,\dots,r  \\
\lambda^{*}_i & \ge 0, \quad i=1,\dots,m \\
\lambda^{*}_i g_i(x^*) & = 0, \quad i=1,\dots,m  \\
\nabla f(x^*) + \sum_{i=1}^m \lambda^{*}_i \nabla g_i(x^*)  + \sum_{j=1}^r \nu^{*}_j \nabla h_j(x^*) & = 0, 
\end{split}
\end{align}
called the {\em Karush-Kuhn-Tucker (KKT)} conditions.
Here, the first two conditions ensure {\em primal feasibility} and the third condition {\em dual feasibility}. The fourth condition is the {\em complementary slackness} condition we just discussed and the fifth condition represents
{\em Lagrangian stationarity}.

We can summarize this insight provided by the KKT conditions in the following proposition.

\begin{proposition}\label{prop:kkt1}
For any optimization problem with continuously differentiable objective and constraint functions for which strong duality holds, any pair of primal and dual
optimal points must satisfy the KKT conditions~\eqref{eq:kkt}.
\end{proposition}

We next show that when the primal problem is convex, the KKT conditions are also sufficient.

\begin{theorem}\label{th:kkt2}
Let the functions $f$, $\{g_i\}_{i=1}^m, \{h_j\}_{j=1}^r$ be continuously differentiable. If in addition $f$ and $\{g_i\}_{i=1}^m$ are convex and $\{h_j\}_{j=1}^r$ are
affine, then the KKT conditions~\eqref{eq:kkt} are necessary and sufficient for the points $x^*, \lambda^*, \nu^*$ to be primal and dual optimal with zero optimality gap.
\end{theorem}

\begin{proof}
The claim about necessity follows from Proposition~\ref{prop:kkt1}. Thus we only need to prove the claim about sufficiency.

Note that the first three KKT
conditions imply that $x^*$ is primal feasible, and $\lambda^*$ is dual feasible.
Furthermore, since ${\mathcal L}(x,\lambda^*,\nu^*)$  is convex in $x$, the fifth KKT condition states
that $x^*$ is a global minimizer of
${\mathcal L}(x,\lambda^*,\nu^*)$, and therefore
\begin{align*}
 \theta(\lambda^*,\nu^*) & = \inf_{x\in \dom}  {\mathcal L}(x,\lambda^*,\nu^*)  = {\mathcal L}(x^*,\lambda^*,\nu^*) \\
 & = f(x^*) + \sum_{i=1}^m \lambda^{*}_i g_i(x^*)
+ \sum_{j=1}^r \nu^{*}_j h_j(x^*) \\
& = f(x^*),
\end{align*}
where the last equality follows from the fact that $h_i(x^*) = 0$ (from primal feasibility), and that $\lambda^{*}_i g_i(x^*)=0$ (from complementary slackness).
This implies that the primal-dual feasible points $x^*,(\lambda^*,\nu^*)$
are optimal, since the corresponding duality gap is zero.

\end{proof}

\subsection{The dual certificate}\label{ss:dualcertificate}

In the context of Lagrangian duality and KKT conditions, the 
concept of a {\em dual certificate}\footnote{The terminology ``certificate'' is borrowed from complexity theory and highlights an asymmetry in computational complexity: Finding the solution often requires an algorithm which may be computationally expensive; while certifying optimality of the solution usually involves simple matrix-vector multiplications or checking eigenvalues. In this context, the dual certificate is a {\em witness} of optimality.} turns out to be very useful.
Given a convex optimization problem
\begin{align*}
\underset{x \in \mathcal{D}}{\operatorname{minimize}} & \qquad   f(x)  \\
 \text{subject to} & \qquad g_i(x) \le b_i, \quad \, i = 1, \dots, m, \\
 & \qquad h_j(x) = c_j, \quad j=1,\dots,r,
\end{align*}
a dual certificate for a candidate point $x^*$ is a set of dual variables $(\lambda^*, \nu^*)$ that satisfy the KKT conditions~\eqref{eq:kkt}. If the problem is convex and satisfies a constraint qualification (like Slater's condition), the existence of these variables is both necessary and sufficient for global optimality.

The most critical part of the certificate is the stationarity condition
$$\nabla f(x^*) + \sum_{i=1}^m \lambda_i^* \nabla g_i(x^*) + \sum_{j=1}^p \nu_j^* \nabla h_j(x^*) = 0$$
Geometrically, this means the gradient of the objective function at $x^*$ must lie in the span of the gradients of the active constraints (with the additional constraint that $\lambda_i \ge 0 \enspace \forall i$). The dual variables $\lambda_i^*$ and $\nu_j^*$ act as the ``weights''  that prove the objective cannot be decreased further without violating a constraint.

At a feasible point $x^*$, the KKT conditions can be understood
geometrically through the normal cone of the feasible set $\mathcal{F}$ at $x^*$,
defined as
\begin{equation}
    N_{\mathcal{F}}(x^*) = \left\{ d \in \mathbb{R}^n : d^\top (x - x^*) \leq 0 
    \;\; \forall x \in \mathcal{F} \right\},
\end{equation}
that is, the set of all directions that make a non-positive inner product with every
feasible direction from $x^*$. The stationarity condition requires that $-\nabla
f(x^*) \in N_{\mathcal{F}}(x^*)$, meaning the negative gradient of the objective must
point in a direction that is normal to the feasible set at $x^*$, so that no feasible
descent direction exists. When $\mathcal{F}$ is described by smooth inequality
constraints $g_i(x) \leq 0$ and equality constraints $h_j(x) = 0$, and an appropriate
constraint qualification holds, the normal cone at $x^*$ admits an explicit
representation,
\begin{equation}\label{eq:normalcone}
    N_{\mathcal{F}}(x^*) = \Big\{ \sum_{i \in \mathcal{A}} \lambda_i \nabla g_i(x^*) 
    + \sum_j \mu_j \nabla h_j(x^*) \;:\; \lambda_i \geq 0 \Big\},
\end{equation}
where $\mathcal{A}$ denotes the set of active inequality constraints at $x^*$. The dual
multipliers $\lambda_i^*$ and $\mu_j^*$ are then precisely the coefficients that
witness membership of $-\nabla f(x^*)$ in this cone, and together they form the dual
certificate that verifies optimality. For convex $f$ and convex $\mathcal{F}$, this
condition is both necessary and sufficient for global optimality; in the non-convex
setting it remains a necessary condition that any local minimum must satisfy.

We illustrate the above observation via the following linear program, which for ease of visualization has no equality constraints:
\begin{align}\label{eq:kkt_example}
\begin{split}
  \max_x & \quad 3x_1 + 2x_2 \\  
  \text{s.t} & \quad g_1(x):=x_1 \le 4,\,\, g_2(x):=x_2 \le 4  \\
  & \quad g_3(x):=x_1+x_2 \le 6 \\
  & \quad g_4(x):=x_1 \ge 0, \,\, g_5(x):=x_2 \ge 0.
\end{split}    
\end{align}

\begin{figure}[h]
\begin{center}
\includegraphics[width=80mm]{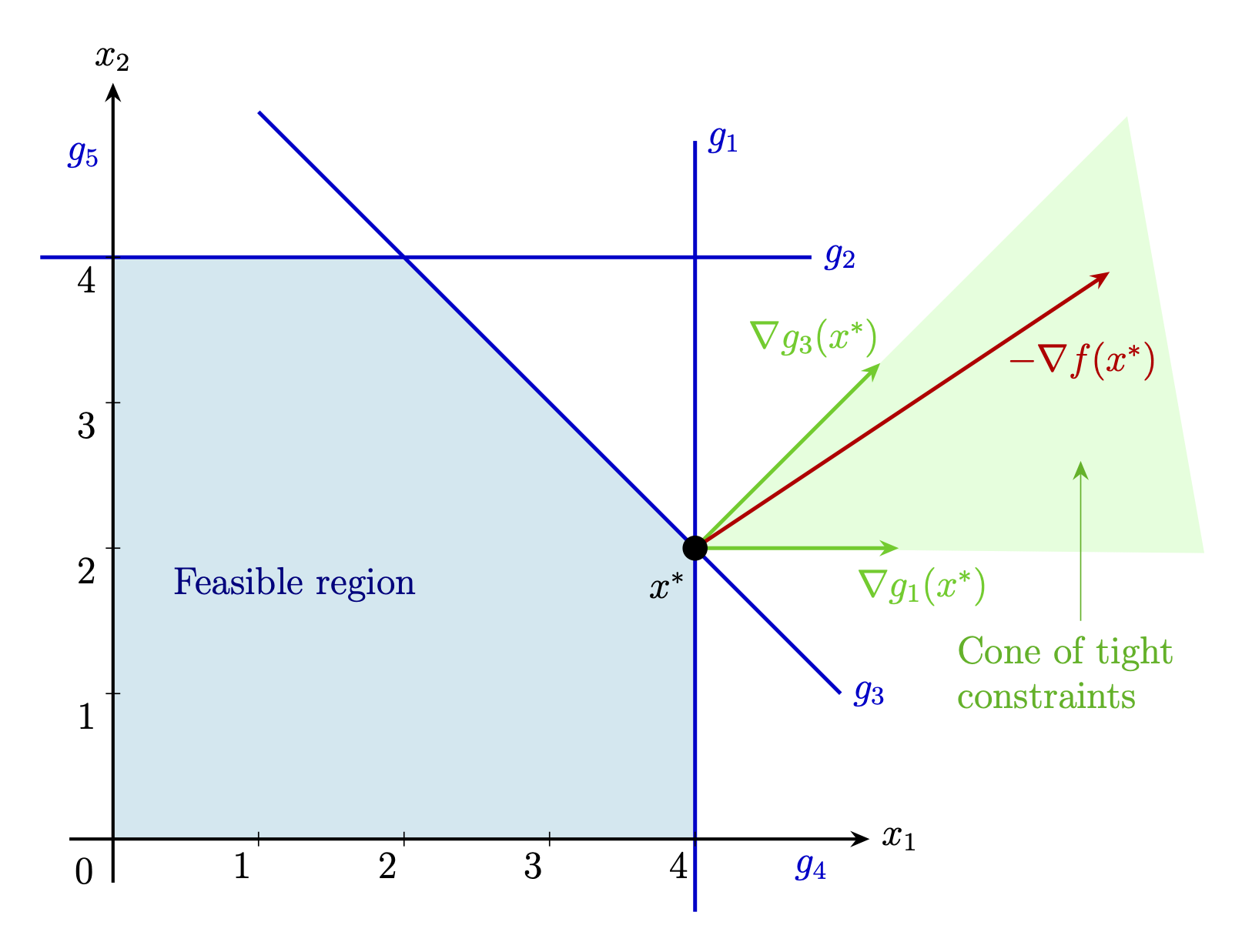}
\caption{KKT conditions and the normal cone. The tight
constraints for $x$ are $g_1(x)= x_1 \le 4$ and $g_3(x)= x_1+x_2 \le 6$. In this example, $-\nabla f(x)$ is inside the cone formed by $\nabla g_1(x^*
)$ and $\nabla g_3(x^*)$, thus the feasible solution $x^*=(4,2)$ is optimal in this case.}
\label{fig:kkt}
\end{center}
\end{figure}

Figure~\ref{fig:kkt} is a geometric depiction of optimality for the LP in~\eqref{eq:kkt_example}. The optimal solution is  $x^* = (4,2)$, the tight
constraints  are $g_1(x):= x_1 \le 4$ and $g_3(x):= x_1+x_2 \le 6$. 
As the condition~\eqref{eq:normalcone} stipulates,
$-\nabla f(x)$ is inside the cone formed by $\nabla g_1(x^*
)$ and $\nabla g_3(x^*)$, as required for the solution $x^*$ to be optimal (in the figure the cone of tight
constraints is depicted with its origin shifted to the point $x^*$).

\medskip

\medskip

Assume that the conditions of Theorem~\ref{th:kkt2} hold and that we have computed the dual solution $(\lambda^*,\nu^*)$. We can now find the solution $x^*$ to the primal problem using the stationarity condition (the fifth condition in~\eqref{eq:kkt}):
$$\nabla_x \mathcal{L}(x^*,\lambda^*,\nu^*) = 0.$$
Given the optimal dual variables $(\lambda^*,\nu^*)$, this is an equation in $x^*$. Solving it  gives the primal optimizer.
The candidate $x^*$ obtained from stationarity must satisfy
$$
g_i(x^*) \le 0, \quad h_j(x^*)=0.
$$
If multiple solutions satisfy stationarity, feasibility selects the correct one.

In some cases the KKT conditions can be solved analytically.
Many algorithms for convex optimization are
designed  (or can be interpreted as) as methods for solving the KKT conditions.

\medskip

\noindent
\textbf{Example: Quadratically constrained quadratic program
 (QCQP).} Consider
\begin{align*}
  & \text{minimize} \qquad f(x): = x^2+1 \\
  & \, \text{subject to} \quad \,\, g(x): = (x-3)(x-4) \le 0.
\end{align*}
Note that the constraint means
$$(x-3)(x-4) \le 0 \qquad \Longleftrightarrow \qquad x\in [3,4].
$$

We now compute the Lagrangian and the associated dual function:
\begin{align*}
\mathcal{L}(x,\lambda) & = x^2 +1 +\lambda(x^2-7x+12), \\
\theta(\lambda) & = \inf _x \mathcal{L}(x,\lambda) = \inf_x \, \big((\lambda+1)x^2 -7\lambda x + 12 \lambda+1\big).
\end{align*}
Since $\mathcal{L}(x,\lambda)$ is a quadratic function of $x$, we minimize it by setting the derivative $\nabla_x \mathcal{L}(x,\lambda)$ to zero, yielding
$$
2(\lambda+1)x - 7\lambda = 0 \Longrightarrow x = \frac{7\lambda}{2(\lambda+1)}.
$$
Substituting back into $\mathcal{L}(x,\lambda)$, simplifying the resulting $\theta(\lambda)$ and using the fact that we always have $\lambda \ge 0$, the dual problem becomes 
$$\theta(\lambda)
= \mathcal{L}(x^*(\lambda),\lambda)
= 12\lambda + 1 - \frac{49\lambda^2}{4(1+\lambda)}.
$$
This expression is finite only if the quadratic is convex in $x$, i.e.
$$
1+\lambda > 0.
$$
Since $\lambda \ge 0$, this condition is automatically satisfied.
Hence, the dual problem is
\begin{align*}
  & \max_\lambda \,\,\, 12\lambda + 1 - \frac{49\lambda^2}{4(1+\lambda)} \\
  & \, \text{s.t.} \,\,\,\,\,\,\,\, \lambda \ge 0.
\end{align*}
We can readily confirm that $\theta(\lambda)$ is concave (negative second derivative), which is to be expected because dual functions are always concave.
A quick calculation yields that the unique optimal point is at $\lambda^* = 6$. At this value $\theta(6)=10 = d^*$.

We can compute the optimal value $x^*$ by solving
$\nabla_x \mathcal{L}(x^*,\lambda^*,\nu^*) = 0$, which we solved already previously. Namely, we have $x = \frac{7\lambda}{2(\lambda+1)}$, and inserting $\lambda=\lambda^*=6$ gives
$x^*=3$ and $f(3)=10=p^*$. Thus $d^{*}=p^{*}$, confirming optimality and strong duality in this problem.

\medskip
While convex quadratically constrained quadratic programs (QCQPs), as in the example above, can be solved efficiently, their nonconvex counterparts constitute a class of particularly challenging optimization problems.
This fact should not come unexpected, since non-convex QCQPs represent a bridge
between continuous and discrete optimization~\cite{calafiore2014optimization} (e.g. $x^2=1$ is equivalent to $x\in\{\pm1\}$).

\medskip

Dual certificates also play an important role in mathematically understanding the optimal solution of a convex optimization problem, they will be key in Chapter~\ref{c:community} where it will be shown that the solution of a convex optimization is precisely one is attempt to estimate, in that case the community structure of a network. Moreover, we will make frequent use of the dual certificate in Chapter~\ref{c:cs} when we investigate compressive sensing and sparsity and Chapter~\ref{c:lowrank} when we analyze low-rank matrix recovery.

\medskip
\noindent
\textbf{The dual of the dual:}
What happens if we take the dual of the dual optimization problem? Do we get back the primal problem? The answer depends on the nature of the primal problem.

For convex optimization problems satisfying a constraint qualification (e.g. Slater's condition), strong duality holds, meaning the duality gap is zero. In this case taking the dual twice does indeed returns us to the primal problem.

If the primal problem is non-convex, the dual of the dual does not necessarily return the original problem. Instead, it returns a {\em convex relaxation} of the primal problem, also called the {\em convex biconjugate} or {\em biconjugate relaxation}. Specifically, the objective of the ``dual of the dual'' will be the convex envelope (the largest lower-bounding convex function) of the original primal objective and the associated feasible set  will be the convex hull of the original primal feasible set. 

An interesting aspect here is that in non-convex optimization adding redundant constraints to the primal can introduce additional dual variables, thereby strengthening the dual problem and yielding a tighter convex relaxation when taking the dual of the dual. Thus in non-convex optimization, 
an algebraically redundant constraint (one that does not change the primal feasible set) is not necessarily computationally redundant. 

This is also a key idea behind the Lasserre Hierarchy (and the dual sum-of-squares perspective, which we explore in Chapter~\ref{c:community}). 
The Lasserre hierarchy does not just add a few random redundant constraints; it systematically 
considers all sum-of-squares weighted combinations of the original constraints up to a given degree. By ``lifting'' the problem into a higher-dimensional semidefinite program, it provides a sequence of convex relaxations that, under mild conditions, is guaranteed to converge to the global optimum.

\subsection{Subdifferential and subgradient}
\label{ss:subgradient}

We will encounter important situations in which the 
objective function $f$ is convex but not differentiable at the solution. This happens for example for $f(x)=\|x\|_1$.
In this situation the {\em subdifferential} takes the role of the derivative.

Let us recall the definition of a subdifferential.

\begin{definition}
Let $f: \mathbb{R}^n \to \mathbb{R} \cup \{+\infty\}$ be a convex function. The {\em subdifferential} of $f$ at a point $x$, denoted by $\partial f(x)$, is the set of all vectors $g \in \mathbb{R}^n$ such that for all $y \in \mathbb{R}^n$:
$$f(y) \geq f(x) + \langle g,y - x\rangle.$$
Each vector $g$ in $\partial f(x)$ is called a {\em subgradient}.
\end{definition}

\begin{figure}[h]
\begin{center}
\includegraphics[width=90mm]{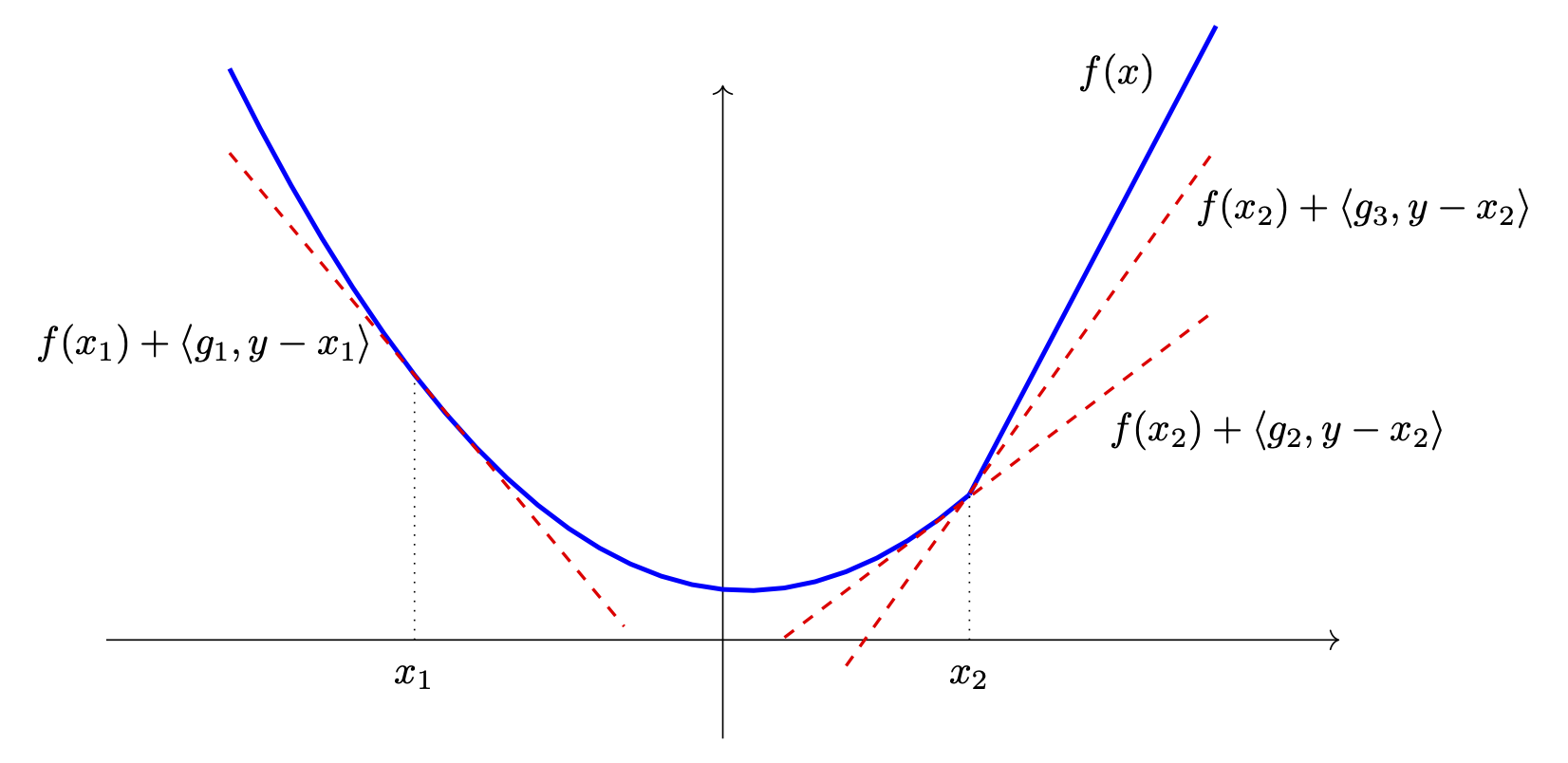}
\caption{At the point $x_1$, the convex function $f$ is differentiable, and $g_1$ is the unique subgradient (which is also the derivative) of $f$ at $x_1$. At the point $x_2$, $f$ is not differentiable. At this point, $f$ has (infinitely) many subgradients: two subgradients, $g_2$ and $g_3$, are shown.}
\label{fig:subgradient}
\end{center}
\end{figure}

As shown in Figure~\ref{fig:subgradient}, there can be more than one subgradient of a function $f$ at a point $x$.

Thus, when the constraint functions $g_i$ (inequalities) and $h_j$ (equalities) are differentiable but the objective function $f$ is not, the stationarity condition becomes a {\em set-inclusion} problem. Instead of searching for a point where the gradient of the Lagrangian is zero, we search for a point where the {\em zero vector} is contained within the set of all possible ``slopes'' generated by the subdifferential of the objective and the gradients of the constraints.
Therefore, in this case the stationarity condition of the KKT conditions~\eqref{eq:kkt} essentially states that there must exist a specific subgradient $s \in \partial f(x^*)$ such that
$$
s + \sum_{i=1}^m \lambda^{*}_i \nabla g_i(x^*)  + \sum_{j=1}^r \nu^{*}_j \nabla h_j(x^*)  = 0. 
$$

\medskip
\noindent
\textbf{Example: Subdifferential for the $\ell_1$-norm.}
Let us compute the subdifferential for the function $f(x) = \|x\|_1$. Because the $\ell_1$-norm is a separable function, its subdifferential is the Cartesian product of the subdifferentials of each individual absolute value component. So, we start with the scalar case:  $f(x) = |x|$.
For a single variable, the subdifferential $\partial |x|$ is defined by the set of slopes $g$ that satisfy the subgradient inequality: $|y| \geq |x| + g(y - x)$ for all $y$.
If $x > 0$, the function is differentiable with slope $1$. Thus, $\partial |x| = \{1\}$.
If $x < 0$, the function is differentiable with slope $-1$. Thus, $\partial |x| = \{-1\}$.
If $x = 0$, we need $|y| \geq 0 + g(y - 0)$, which simplifies to $|y| \geq gy$. This holds for all $y$ if and only if $g \in [-1, 1]$.
We conclude that the subdifferential of the absolute value is the sign function
$$
\partial |x| = \text{sign}(x) = 
\begin{cases} \{1\} & x > 0 \\ \{-1\} & x < 0 \\ 
[-1, 1] & x = 0 \end{cases}
$$
Therefore, for the vector case, $f(x)=\|x\|_1$, the subdifferential is the set of vectors $g$ such that:
$$g_i = \begin{cases} \text{sign}(x_i) & \text{if } x_i \neq 0 \\ \in [-1, 1] & \text{if } x_i = 0 \end{cases}$$

\medskip
\noindent
\textbf{Example: Subdifferential for the nuclear norm.} 
Recall that the nuclear norm of a matrix $X\in \R^{n_1 \times n_2}$ is defined as the sum of its singular values, i.e.,
$$\|X\|_* = \sum_{i=1}^{\min(n_1,n_2)}\sigma_i(X),$$
where $X=U\Sigma V$ is the SVD of $X$ and the $\sigma_i$ are the diagonal entries of $\Sigma$.
The nuclear norm plays an important role in optimization, in particular as a convex surrogate in rank minimization problems, see also Chapter~\ref{c:lowrank}. We want to compute the subdifferential of $f(X) = \|X\|_*$.

We  need to identify the set of matrices $Z$ that satisfy the subgradient inequality
$$\|Y\|_* \geq \|X\|_* + \langle Z, Y - X \rangle \qquad \forall \, Y \in \mathbb{R}^{n_1 \times n_2}.$$

Recall that the nuclear norm and the spectral norm are dual to each other. By the definition of the dual norm
$$\|X\|_* = \sup_{\|Y\|\le 1} \langle Y, X \rangle.$$
From the definition of a subgradient $Z \in \partial f(X)$, it follows that $X$ must satisfy
\begin{enumerate}
\item $\|Z\| \leq 1$ \quad (Membership in the dual unit ball)
\item $\langle Z, X \rangle = \|X\|_*$ \quad (Equality in the dual characterization)
\end{enumerate}

Let the SVD of $X$ be $X = U \Sigma V^T$, where $U \in \mathbb{R}^{n_1 \times r}$, $V \in \mathbb{R}^{n_2 \times r}$, and $\Sigma$ is an $r \times r$ diagonal matrix with strictly positive singular values.
We define the tangent space $\Tn$ at $X$ relative to the manifold of rank-$r$ matrices as
$$\Tn = \{ U A^T + B V^T : A \in \mathbb{R}^{n \times r}, B \in \mathbb{R}^{m \times r} \}.$$
The orthogonal complement $\Tn^{\perp}$ consists of matrices whose rows are orthogonal to $V$ and columns are orthogonal to $U$
$$\Tn^{\perp} = \{ W : U^T W = 0, W V = 0 \}.$$
Any matrix $Z$ can be decomposed as
$$Z = P_\Tn(Z) + P_{\Tn^\perp}(Z).$$

We require $\langle Z, X \rangle = \|X\|_*$. Substituting the SVD
$$\langle Z, U \Sigma V^T \rangle = \text{tr}(Z^T U \Sigma V^T) = \text{tr}(V^T Z^T U \Sigma) =  \sum_{i=1}^r (U^T Z V)_{ii} \sigma_i.$$
For this sum to equal $\sum \sigma_i$ (given $\|Z\| \leq 1$), we must have $(U^T Z V)_{ii} = 1$ for all $i$. In fact, the only way to satisfy $\|Z\| \leq 1$ and achieve this inner product is if
$$P_T(Z) = UV^T.$$

Now consider the component $W = P_{\Tn^\perp}(Z)$. For $Z$ to be a valid subgradient, the operator norm $\|Z\|$ must not exceed 1. Since $UV^T$ and $W$ have orthogonal row and column spaces (by the definition of $\Tn^\perp$):
$$\|Z\| = \|UV^T + W\| = \max(\|UV^T\|, \|W\|)$$
Because $\|UV^T\| = 1$, we must ensure that $\|W\| \leq 1$.
Furthermore, the orthogonality conditions for $W \in \Tn^\perp$ are
$$U^T W = 0 \quad \text{and} \quad WV = 0$$
Combining these requirements yields that the subdifferential has the form
\begin{equation}\label{eq:nuclearsubdifferential}
\partial \|X\|_* = \{ UV^T + W : U^T W = 0, WV = 0, \|W\| \leq 1 \}.
\end{equation}

\section{Gradient-based methods and convergence} 
\label{s:grad}

{\em Gradient descent} is an iterative optimization algorithm used to minimize a differentiable function by moving in the direction of the {\em steepest descent}\,\footnote{Therefore gradient descent is also called the method of steepest descent in numerical analysis.}, as defined by the negative of the gradient. The gradient of a function at a point gives the direction of the steepest ascent. Therefore, to minimize the function, the gradient descent algorithm updates the parameters iteratively by moving in the opposite direction of the gradient
\begin{equation}
x_{k+1} = x_{k} - \eta \nabla_{\! x} \, f(x_k),  \qquad k=0,1,,\dots,
\label{gradientdescent}
\end{equation}
where $x_{0}$ is some initial guess and $\eta $ is the {\em step size} or {\em learning rate}.

Typical choices for the initial guess are the zero vector or some random initialization.
The step size may be chosen constant, decaying, or adaptively. A step size that is too large might overshoot the minimum, while a step size that is too small may result
in poor convergence.

In this section we will discuss gradient descent methods and their convergence, focusing on unconstrained optimization problems. We will start with guarantees of converge of gradients for $L$-smooth functions in Section~\ref{s:gd:Lsmoothfunctions}; in Section~\ref{s:gd:convexLsmooth} we upgrade the conditions to include convexity of the function to be optimized and show convergence of function values to the minimum value; we then introduce the  \PLC to show converge of the iterates themselves to the minimizer in Section~\ref{s:gd:PolyakLojasiewicz}. 


\subsection{Gradient descent for $L$-smooth functions}\label{s:gd:Lsmoothfunctions}

We recall a useful notion of regularity that quantifies how much a function changes, such that its gradient 
provides a reliable local linear approximation, and which in turn will enable us to quantify the rate of convergence of gradient descent.

\begin{definition}\label{Lipschitz}
    The gradient of a function $f : \mathbb{R}^d \to \mathbb{R}$ is {\em $L$-Lipschitz continuous}  if there exists a constant $L > 0$ such that
\begin{equation}
    \label{def:Lipschitz}
\|\nabla f(x) - \nabla f(y)\| \le L \|x - y\|, \quad \forall x,y \in \mathbb{R}^d.
\end{equation}
\end{definition}

It is not difficult to see (and left as an exercise) that the $L$-Lipschitz condition is equivalent to the notion of $L$-smoothness often used in the literature in connection with gradient descent.

\begin{figure}[h]
\begin{center}
    \includegraphics[width=.7\textwidth]{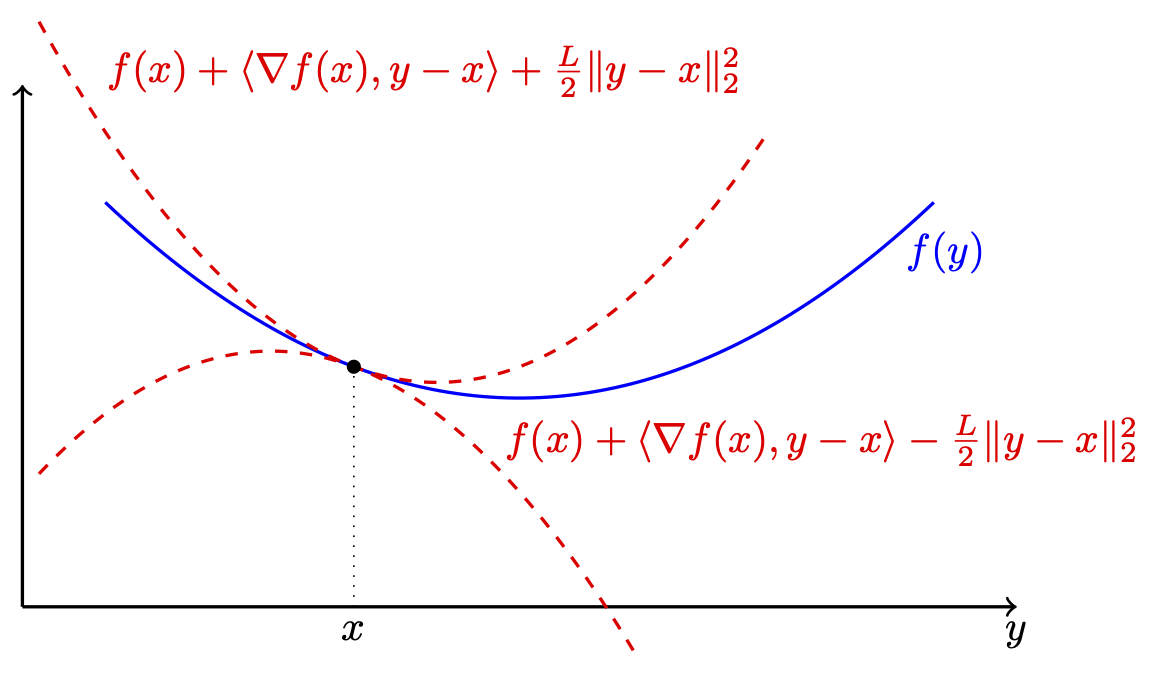}
    \caption{A convex and $L$-smooth function}
    \label{fig:plotsmooth}
\end{center}
\end{figure}

\begin{definition}\label{def:Lsmooth}
 A function $f : \mathbb{R}^d \to \mathbb{R}$  is called {\em $L$-smooth} if there exists a constant $L > 0$ such that
\begin{align} \label{quadraticbound}
\begin{split}
f(y) & \le f(x) + \langle \nabla f(x) ,y - x \rangle  + \frac{L}{2} \|x - y\|^2 \quad \forall x,y \in \mathbb{R}^d, \\
f(y) & \ge f(x) + \langle \nabla f(x) ,y - x \rangle  - \frac{L}{2} \|x - y\|^2 \quad \forall x,y \in \mathbb{R}^d.
\end{split}
\end{align}
\end{definition}
In other words, $f$ is bounded above and below by a quadratic function that touches its graph at each point $x$, see also Figure~\ref{fig:plotsmooth}.

We can use $L$-smoothness to quantify the improvement of one
step of the gradient descent algorithm. The resulting 
statement is often referred to as the {\em gradient descent lemma}.

\begin{lemma}[Gradient Descent Lemma]\label{le:descentlemma}
 Asssume that $f: \R^d \to \R$ is $L$-smooth. Then for any stepsize $\eta$ with $0 < \eta \le \frac{1}{L} $ each step of gradient descent~\eqref{gradientdescent} satisfies
\begin{equation}\label{eq:descentlemma}
f(x_{k+1}) \le f(x_k) - \frac{\eta}{2}\|\nabla f(x_k)\|^2.
\end{equation}
\end{lemma}

\begin{proof}
We apply~\eqref{quadraticbound} with $x=x_k$ and $y=x_{k+1}$:
\begin{equation}
   \label{descent1} 
f(x_{k+1}) \le f(x_k) + \langle \nabla f(x_k),  x_{k+1} - x_k \rangle + \frac{L}{2}\|x_{k+1} - x_k\|^2.
 \end{equation}
 Substituting the gradient descent formula $x^{(k+1)} = x^{(k)} - \eta \nabla f(x^{(k)})$ into~\eqref{descent1}
 gives
 \begin{align*}
     f(x_{k+1}) & \le f(x_k)  - \frac{\eta}{2}\|\nabla f(x_k)\|^2 + \frac{L}{2}\|x_{k+1} - x_k\|^2 \\
     & = f(x_k)  - \eta \big( 1 - \frac{L}{2} \eta \big) \|\nabla f(x_k)\|^2.
 \end{align*}
 The assumption  $\eta \le \frac{1}{L}$ implies that $1 - \frac{L}{2} \eta \ge \frac{1}{2}$,
 and the proof is complete.
 
\end{proof}
This suggests that, for $L$-smooth functions, a natural choice for the stepsize is $\eta = \frac{1}{L}.$

If $f$ is lower bounded, the gradient descent lemma also implies that the gradients
of the iterates $x_k$ must eventually become small: indeed, if $\frac{\eta}{2}\|\nabla f(x_k)\|^2$ stayed non-negligible over many iterates, then~\eqref{eq:descentlemma} would force the function of the values of the iterates to be arbitrarily small, which cannot happen for functions that are bounded below. This will be discussed in Section~\ref{s:gd:nonconvex} (see Theorem~\ref{th:stationary}). Note however that we have not yet assumed convexity of $f$ and, without
assuming convexity (or other suitable regularity conditions), guarantees of small gradients may be of limited use, because they do not necessarily imply that the associated function value is close to optimal.  To see this, consider for example the function $f(x) = \eps \log(1+e^x)$. The gradient of $f$ satisfies $\nabla f \le \eps$ for all $x \in \R^d$. Hence, $x$ could be arbitrarily far away from the optimum, while the associated gradient is nevertheless small. 

\subsection{Gradient descent for convex $L$-smooth functions}\label{s:gd:convexLsmooth}

With convexity we will be able to guarantee convergence of function values. To that end we first introduce a useful tool, the {\em Euclidean mirror descent lemma}.

\begin{lemma}[Euclidean Mirror Descent Lemma]\label{euclideandescentlemma}
Let $f: \R^d \to \R$ be a differentiable and convex function. Then, for any choice of stepsize $\eta$ and any two consecutive iterates $x_k, x_{k+1}$ produced by gradient descent~\eqref{gradientdescent} it holds that
\begin{equation}\label{eq:mirror1}
f(x_k) \le f(y) + \frac{1}{2\eta} \Big( \|y-x_k\|^2 - \|y-x_{k+1}\|^2  + \|x_{k+1} - x_k\|^2 \Big), \qquad \forall y \in \R^d.
\end{equation}
\end{lemma}

\begin{proof}
An easy calculation shows that
\begin{equation}
    \label{eq:mirror2}
\langle x_k - x_{k+1}, y - x_k \rangle = - \frac{1}{2}\Big( \|y-x_k\|^2 - \|y-x_{k+1}\|^2  + \|x_{k+1} - x_k\|^2 \Big).
\end{equation}
Since $f$ is convex, we can employ the bound~\eqref{def:geomconvex} and obtain that for any $y \in \R^d$,
\begin{equation}
    \label{eq:mirror3}
f(y) \ge f(x_k) + \langle \nabla f(x_k) , y - x_k \rangle.
\end{equation}
Substituting the gradient descent iteration rule $x_{k+1} = x_k - \eta \nabla f(x_k)$ in~\eqref{eq:mirror3} we obtain
\begin{equation}
    \label{eq:mirror4}
f(y) \ge f(x_k) + \frac{1}{\eta} \langle x_k - x_{k+1} , y - x_k \rangle.
\end{equation}
Using~\eqref{eq:mirror2}, and rearranging terms gives
$$
f(x_k) \le f(y) + \frac{1}{2\eta} \Big( \|y-x_k\|^2 - \|y-x_{k+1}\|^2  + \|x_{k+1} - x_k\|^2 \Big),
$$
which is what we wanted to prove. 
\end{proof}
We will now use the Euclidean mirror descent lemma  to obtain a convergence rate
in terms of the objective function value.

\begin{theorem}\label{th:gradconv1}
Let $f : \mathbb{R}^d \to \mathbb{R}$ be convex with $L$-Lipschitz continuous gradient, and let $x^* \in \arg\min_x f(x)$. 
Assume we apply gradient descent with constant step size $ 0 < \eta\le  \frac{1}{L}$. 
Then, for all $k \ge 1$,
$$
f(x_k) - f(x^*) \le \frac{\|x_0 - x^*\|^2}{2k\eta}.
$$
Hence, the error in function value decreases at rate ${\mathcal O}(1/k)$.
\end{theorem}

\begin{proof}
Since $x_{k+1} - x_k = \eta \nabla f(x_k)$ by definition of the gradient descent algorithm, we can rewrite the bound~\eqref{eq:mirror1} as
\begin{equation}\label{eq:mirror5}
f(x_k) \le f(y) + \frac{1}{2\eta} \Big( \|y-x_k\|^2 - \|y-x_{k+1}\|^2  + \|\eta \nabla f(x_k)\|^2 \Big), \qquad \forall y \in \R^d.
\end{equation}
We apply Lemma~\ref{le:descentlemma} and get
\begin{equation}
    \label{eq:mirror6}
\frac{\eta}{2} \|\nabla f(x_k)\|^2 \le 
f(x_{k}) - f(x_{k+1}).
\end{equation}
Plugging~\eqref{eq:mirror6} into~\eqref{eq:mirror5} yields
\begin{equation}\label{eq:mirror7}
f(x_{k+1}) \le f(y) + \frac{1}{2\eta} \Big( \|y-x_k\|^2 - \|y-x_{k+1}\|^2 \Big), \qquad \forall y \in \R^d.
\end{equation}
In particular, setting $y=x^{*}$ gives
\begin{equation}\label{eq:mirror7}
f(x_{k+1}) \le f(x^*) + \frac{1}{2\eta} \Big( \|x^*-x_k\|^2 - \|x^*-x_{k+1}\|^2 \Big),
\end{equation}
whenever $\eta \le \frac{1}{L}$. By summing over $k=0,\dots,K-1$ and telescoping over the iterations, we obtain
\begin{equation}\label{eq:mirror8}
\sum_{k=0}^{K-1} f(x_{k+1}) \le K f(x^*) + \frac{1}{2\eta}  \|x^*-x_0\|^2.
\end{equation}
Lemma~\ref{le:descentlemma} implies that $f(x_k)$ is non-increasing in $k$, which implies that $\sum_{k=0}^{K-1} f(x_{k+1}) \le K f(x^K)$. Therefore,
$
K f(x_K) \le K f(x^*) + \frac{1}{2\eta}  \|x^*-x_0\|^2,
$
from which we conclude
$
f(x_K) - f(x^*) \le \frac{1}{2K\eta}  \|x^*-x_0\|^2$. 

\end{proof}

\subsection{Iterate convergence of gradient descent for strongly convex functions}\label{s:gd:PolyakLojasiewicz}

Can we improve Theorem~\ref{th:gradconv1} to obtain a statement
 at the level of the iterates, and not just at the level of the function values? I.e., can we obtain a result stating that $\|x_k - x^*\|$
 decreases at rate ${\OOO}(1/k)$?  Alas, without further assumptions this is not possible. As Nesterov has shown (see Theorem 2.1.7 in~\cite{nesterov2018lectures}), there exist convex $L$-smooth functions for which convergence via gradient descent to the optimal point may be arbitrarily slow.

One standard way to obtain the desired convergence rate bound at the level of the iterates $x_k$ is by assuming  in addition that $f$ is strongly convex. Here, we take a slightly different approach by using a weaker assumption, namely the so-called {\em \PL condition}, introduced independently by Polyak~\cite{Pol63} and \L ojasiewicz~\cite{Loj63}.

\begin{definition}[\PLC]\label{PL condition}
A lower-bounded function $f: \R^d \to \R$ satisfies the \PL\ condition if there exists $\mu > 0$ such that
$$
\frac{1}{2} \|\nabla f(x) \|^2 \ge \mu (f(x) - \finf), \qquad \forall x \in \R^d.
$$
\end{definition}

We leave it as an exercise to show that strong convexity implies the \PL\ condition. However, the reverse is in general not true. Take for example the constant function $f(x)=c$ for some constant $c \in \R$, which trivially satisfies the \PL\ condition, but is clearly not strongly convex. In fact, the \PL\ condition does not even imply convexity. As case in point, consider the function $f(x) = x^2 + 3\sin^2(x)$, which is obviously not convex, but can be shown to satisfy the \PL\ condition (with $\mu = \frac{1}{32}$), see~\cite{karimi2016linear}.

We need the following lemma in order to establish the sought-after convergence rate bound on the iterates.

\begin{lemma}
   \label{le:PL} 
Let $f: \R^d \to \R$ be an $L$-smooth function which is lower-bounded by $\finf$ and which satisfies the \PL\ condition for some constant $\mu > 0$. Let the stepsize be given by $\eta = \frac{1}{L}$. Then, the $k$-th iterate $x_k$ of gradient descent satisfies
\begin{equation}
f(x_k) - f(x^*) \le \Big(1 - \frac{\mu}{L} \Big)^k (f(x_0) - \finf).
\end{equation}
\end{lemma}

\begin{proof}
Using the gradient descent lemma (Lemma~\ref{le:descentlemma}) (for $\eta = \frac{1}{L}$) and the \PL\  condition, we have
\begin{align*}
f(x_{k+1}) & \le f(x_k) - \frac{1}{2L} \| \nabla f(x_k)\|^2 \\
& \le f(x_k) - \frac{\mu}{2L} (f(x_k) - \finf) \\
& \le \Big(1-\frac{\mu}{L}\Big) f(x_k) + \frac{\mu}{L} \finf.
\end{align*}
Subtracting $\finf$ on both sides gives
$$
f(x_{k+1}) -  \finf \le \Big(1-\frac{\mu}{L}\Big) \big( f(x_k) - \finf \big).
$$
Solving the recurrence completes the proof.
    
\end{proof}

We are now ready to show that  a function that satisfies the \PLC\ must attain a minimum and
that the iterates produced by gradient descent converge at a rate exponential in $k$.

\begin{theorem}\label{th:PLCbound}
Let $f: \R^d \to \R$ be an $L$-smooth function which is lower-bounded by $\finf$ and which satisfies the \PLC\ for some constant $\mu > 0$. Let the stepsize $\eta$ satisfy $0 < \eta \le \frac{1}{L}$. Then, the sequence of iterates $x_k$ produced by gradient descent converges to some point $x^* \in \R^d$, with
\begin{equation}\label{eq:mirror7}
\|x_{k+1} - x^*\|^2 \le \frac{8 \eta L^2}{\mu^2}  
\Big(1-\frac{\mu}{L}\Big)^k \big( f(x_0) - \finf \big).
\end{equation}
Moreover, 
$\lim_{k \to \infty} f(x_k) = f(x^*) = \finf$.
\end{theorem}

\begin{proof}
For any $k=0,1,2,\dots$, we have
\begin{align*}
\|x_{k+1} - x_k \|^2 & = \eta^2 \|\nabla f(x_k)\|^2 \\
& \le 2 \eta \big(f(x_k) - f(x_{k+1})\big) \\
& \le 2 \eta \big(f(x_k) - f(x^*)\big) \\
& \le 2 \eta \Big(1-\frac{\mu}{L}\Big)^k \big( f(x_0) - \finf \big),
\end{align*}
where the second line follows from Lemma~\ref{le:descentlemma}
and the fourth line from Lemma~\ref{le:PL}. 

For convenience, we denote $q:=\Big(1-\frac{\mu}{L}\Big)$.
For any $m < n$,
\begin{align*}
\|x_m - x_n\| & \le \sum_{k=m}^{n-1}\|x_k - x_{k+1}\| \\
              & \le \sqrt{2\eta(f(x_0) - \finf)} \cdot \sum_{k=m}^{n-1} \sqrt{q}^k \\
              & \le \sqrt{2\eta(f(x_0) - \finf)} \cdot \frac{\sqrt{q}^m}{1-\sqrt{q}},
\end{align*}
where the use of the geometric series formula is justified since  $0 \le q < 1$.
Therefore, $\|x_m - x_n\| \to 0$ as $m,n\to \infty$, which establishes that $\{x_k\}$ is a Cauchy sequence. Since $\R^d$ is complete, 
$x_k$ converges to some $x^* \in \R^d$. Now, we set $m=k$, let $n \to \infty$ and obtain
$$
\|x_k - x^*\| \le \sqrt{2\eta(f(x_0) - \finf)} \cdot \frac{\sqrt{q}^m}{1-\sqrt{q}}.
$$
Squaring both sides gives
\begin{align}
\|x_k - x^*\|^2 & \le 2\eta(f(x_0) - \finf) \cdot \frac{q^m}{(1-\sqrt{q})^2} \notag \\
& \le 8\eta(f(x_0) - \finf) \cdot \frac{q^m}{q(1-q)^2}, \label{eq:mirror8}
\end{align}
where the last line follows from the fact that 
$\frac{1}{(1-\sqrt{q})^2} \le \frac{4}{q(1-q)^2}$.
Substituting back the definition of $q$ in~\eqref{eq:mirror8} yields the claimed bound in~\eqref{eq:mirror8}.

\end{proof}

Note that Theorem~\ref{th:PLCbound} does not guarantee that the solution $x^*$ is the {\em unique} minimizer, since
the \PLC \ is not strong enough for that.
Now, if $f$ is instead $\mu$-strongly convex (in addition to being $L$-smooth), then $x^*$ is indeed the {\em unique} minimizer and we can  tighten the convergence bound in~\eqref{eq:mirror7} (see e.g.~\cite{nesterov2018lectures}) as follows:
\begin{equation}\label{eq:linearconvergenceGDstronglyconvex}
\|x_{k} - x^*\|^2 \le   
\Big(1-\frac{\mu}{L}\Big)^k \|x_{0} - x^*\|^2.
\end{equation}

\begin{remark}
    Sequences of iterates $\{x_k\}_{k=0}^\infty$ that converge satisfying
    $$
\frac{\|x_{k+1} - x^*\|}{{\|x_{k} - x^*\|}} \le C
    $$
 for some constant $C$ are said to {\em converge linearly}, since the number of digits of precision increases linearly with iterations\footnote{This rate is sometimes also referred to as exponential convergence, because of the exponential decay with respect to the iteration count.}. Note that a convergence satisfying~\eqref{eq:linearconvergenceGDstronglyconvex} corresponds to linear convergence with $C = \sqrt{1-\frac{\mu}{L}}$.

 Some second-order methods, such as Newton's method that involves Hessians, often converge quadratically in the sense
 that the error at each iteration is approximately proportional to the square of the error at the previous iteration, 
 which in practice means that once the algorithm is close to the solution, the number of correct digits of precision roughly doubles with each iteration. The caveat is that these second-order methods come with much higher computational cost per iteration due to forming and inverting the Hessian.
\end{remark}

\subsection{Gradient descent in non-convex optimization}\label{s:gd:nonconvex}

We now turn to the behavior of gradient descent when the objective function $f:\mathbb{R}^d \to \mathbb{R}$ is {\em not convex}. In this regime, the global convergence guarantees of the convex case no longer hold, since a nonconvex function may have multiple local extrema and saddle points. 
This is typically the case for cost functions employed in deep learning. 

In principle, we did already briefly dive into the non-convex setting in the previous section when exploring the \PLC. However, this assumption is still quite strong and often not fulfilled for optimization problems arising in data science.
So, what are minimal assumptions we can build on in the non-convex setting that still give meaningful convergence statements about the stationarity of the iterates?

When a problem fails to be convex, some of our ideas, theorems, and methods may still apply ``locally''. One way to do this is by focusing on ``critical points''.
Recall that a stationary point is a point where the gradient of a function vanishes. A first-order stationary point satisfies $\nabla f(x^\ast) = 0$, meaning the function is ``locally flat'', although it may correspond to a minimum, maximum, or saddle point.  

We again consider the gradient descent iteration
$$
x_{k+1} = x_k - \eta \nabla f(x_k),
$$
for a differentiable (possibly non-convex) function $f$.
We make the following two standard assumptions: 

\smallskip
\noindent
(A1) $f$ has an $L$-Lipschitz continuous gradient (is $L$-smooth). 

\smallskip
\noindent
(A2) $f$ is bounded below:
$f(x) \ge f_{\inf} > -\infty, \, \forall x\in\mathbb{R}^d$, where $f_{\inf}:= \inf_{x} f(x)$.

\smallskip
These assumptions are mild and are (locally) satisfied by many neural network loss functions in practice.

By the gradient descent Lemma (Lemma~\ref{le:descentlemma}) we have for any step size $0 < \eta \le 1/L$:
\begin{equation}
\label{descentlemma2}    
f(x_{k+1}) \le f(x_k) - \frac{\eta}{2}\|\nabla f(x_k)\|^2.
\end{equation}
This inequality holds regardless of convexity.
It shows that, as long as the step size is not too large, each iteration monotonically decreases the objective (unless $\nabla f(x_k)=0)$.

We can now establish convergence to stationary points.

\begin{theorem}\label{th:stationary}
Suppose that $f$ satisfies assumptions (A1) and (A2), and that gradient descent uses a constant step size $0 < \eta \le 1/L$.
Then:   
\begin{itemize}
\item[(a)]  The sequence $\{f(x_k)\}_{k=1}^{\infty}$ is nonincreasing and convergent. 
\item[(b)] The gradients vanish asymptotically:
   $$
   \lim_{k \to \infty} \|\nabla f(x_k)\| = 0.
   $$
\item[(c)] Every limit point of $(x_k)$ is a first-order stationary point of $f$.
\end{itemize}

\end{theorem}

\begin{proof}
Using assumption (A1), we can apply~\eqref{descentlemma2} and obtain
$$
f(x_k) - f(x_{k+1}) \ge \frac{\eta}{2}\|\nabla f(x_k)\|^2.
$$
Summing over $k=0,\dots,K-1$ and employing assumption (A2) gives
$$
\frac{\eta}{2}\sum_{k=0}^{K-1}\|\nabla f(x_k)\|^2
\le f(x_0) - f(x_K)
\le f(x_0) - f_{\inf}.
$$
Thus the sum of gradient norms squared is finite:
$$
\sum_{k=0}^{\infty} \|\nabla f(x_k)\|^2 < \infty.
$$
In particular, $\|\nabla f(x_k)\| \to 0$ as $k \to \infty$.
Monotone bounded sequences $\{f(x_k)\}$ converge; and by continuity of $\nabla f$, any accumulation point of $\{x_k\}$ must satisfy $\nabla f(x_\infty)=0.$  
\end{proof}

The above theorem guarantees convergence to a point where $\nabla f(x)=0$, but not necessarily to a local minimum/maximum.
However, for a number of high-dimensional problems (notably in deep learning), the set of strict saddle points is of measure zero under certain conditions~\cite{lee2016gradient,panageas2017,sun2018geometric}.
Thus, in practice, gradient descent (especially with stochastic perturbations and careful parameter tuning) frequently avoids saddle points and converges to local minima.
This seems to play a role in the practical success of gradient-based optimization in deep learning.

\medskip
The proper choice of the stepsize  plays a key role for the performance of gradient descent. We have seen that in the convex setting, a fixed stepsize (which must be sufficiently small to respect the Lipschitz constant), suffices. Although recent work has shown that improvements can be gained in the convex setting from variable stepsizes, see e.g.~\cite{altschuler2024acceleration}.
Conversely, in non-convex landscapes, the step size functions as a ``tool for exploration''. While a small step size is necessary for fine-tuning near a solution, a larger or more dynamic step size is often required to provide the ``numerical momentum'' needed to escape shallow local minima and navigate past stagnant saddle points.

\subsection{Gradient descent with momentum and Nesterov acceleration}\label{s:gd:nesterovaccelation}

We previously mentioned that vanilla gradient descent can converge slowly, particularly in ill-conditioned problems where the objective function exhibits steep curvature in some directions and shallow curvature in others. In such cases, the algorithm may oscillate across narrow valleys while making little progress toward the optimum.

Momentum methods address this limitation by introducing a velocity variable that accumulates gradient information across iterations~\cite{polyak1964some}. Rather than moving solely in the direction of the current gradient, the update incorporates a running average of past gradients, thereby smoothing the optimization trajectory and accelerating convergence along consistent descent directions.

The classical momentum method updates the velocity $v_k$ and parameters $x_k$ as
\begin{align*}
v_{k+1} &= \mu v_k - \eta \nabla f(x_k), \\
x_{k+1} &= x_k + v_{k+1},
\end{align*}
where $\mu \in [0,1)$ is the momentum coefficient. Intuitively, the velocity acts as an exponentially decaying memory of past gradients. When successive gradients point in similar directions, their contributions reinforce one another, leading to larger effective steps. Conversely, oscillations in directions of high curvature are damped, improving stability.

Nesterov Accelerated Gradient (NAG)~\cite{nesterov1983method}---often referred to as Nesterov momentum---is a ``look-ahead'' modification designed to correct this behavior. Standard momentum is known in mathematical optimization as the ``Heavy Ball'' method: Think of a heavy ball rolling down a hill. It gains speed based on the current slope and its previous velocity. However, it does not know it should slow down until it has already started climbing the opposite side of the valley.

Nesterov momentum refines the momentum approach by computing the gradient at a {\em look-ahead} position rather than at the current parameters. Instead of evaluating $\nabla f(x_k)$, Nesterov momentum evaluates the gradient at the anticipated future position $x_k + \mu v_k$.

The update equations become
\begin{align*}
v_{k+1} &= \mu v_k - \eta \nabla f(x_k + \mu v_k), \\
x_{k+1} &= x_k + v_{k+1}.
\end{align*}

\begin{figure}
\begin{center}
    \includegraphics[width=.8\textwidth]{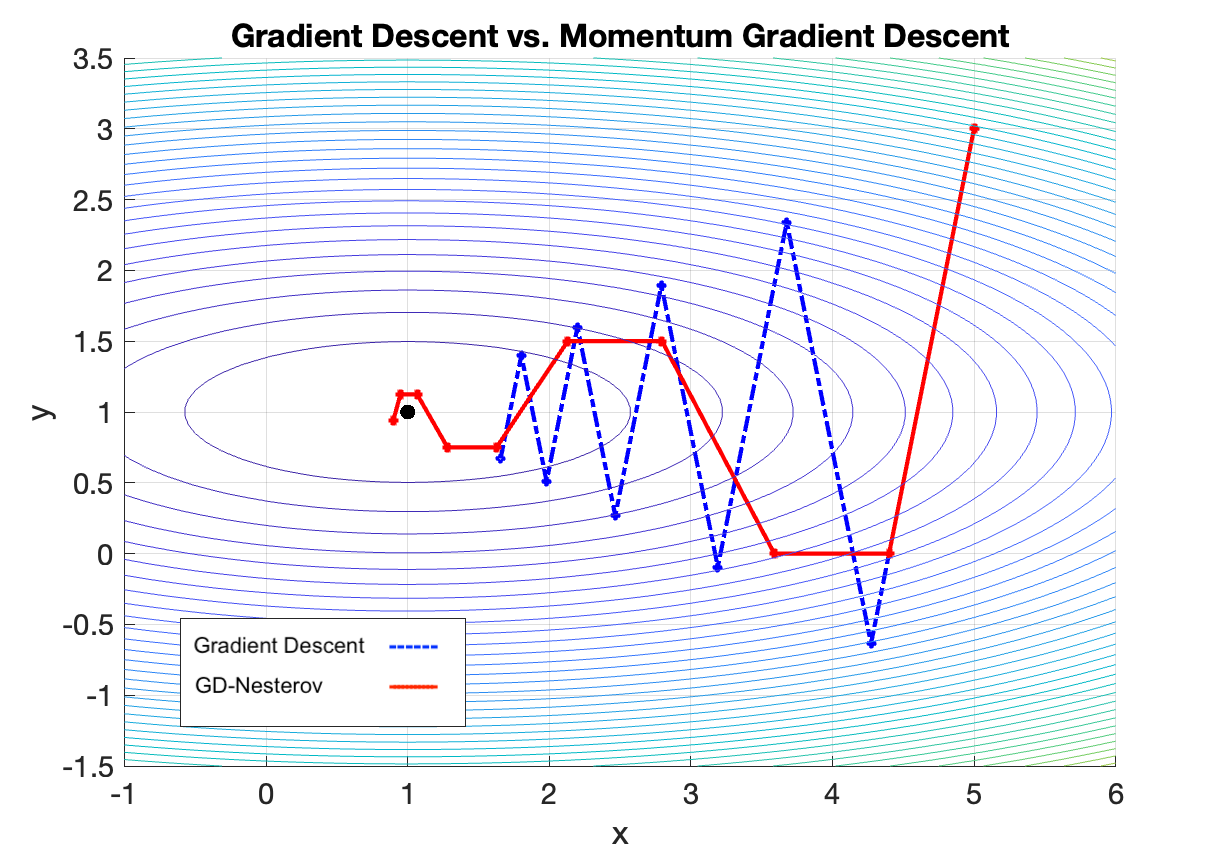}
    \caption{Convergence behavior of standard gradient descent and gradient descent with Nestorov momentum.  The convergence path of gradient descent is shows as blue dashed line, while the one of gradient descent with Nestorov momentum as red solid line. The level sets illustrate the contour of the optimization landscape. The solution at point (1,1) is depicted as black dot.}
    \label{fig:nestorov}
\end{center}
\end{figure}

As a result, Nesterov momentum tends to reduce overshooting and often reaches the global minimum in fewer iterations (and the cost per iteration increases only slightly) compared to vanilla gradient descent, as illustrated in Figure~\ref{fig:nestorov}.

From a theoretical perspective, Nesterov’s method achieves an improved convergence rate for convex optimization problems. We have seen that for smooth convex functions, standard gradient descent converges (in function values) at a rate of ${\mathcal O}(1/k)$ (see Theorem~\ref{th:gradconv1}). In comparison,  Nesterov momentum achieves a rate of ${\mathcal O}(1/k^2)$ under appropriate conditions. This improvement is significant and has made Nesterov momentum an  cornerstone of modern optimization, particularly in large-scale data science and machine learning problems.

In practice, momentum coefficients are commonly set in the range $\mu \in [0.9, 0.99]$, while the learning rate $\eta$ must still be tuned to the problem at hand. Nesterov momentum often yields faster and more stable convergence than classical momentum with minimal additional computational cost, as it requires only one gradient evaluation per iteration.

Both momentum and Nesterov acceleration are foundational ideas that underpin more advanced adaptive optimization algorithms,  such as Adam~\cite{kingma2014adam} or AdaGrad~\cite{duchi2011adaptive} which are often used in  high-dimensional, non-convex settings.

\subsection{Subgradient descent}\label{s:gd:subgradient}

Subgradient descent is the most straightforward extension of gradient descent to non-smooth functions, such as the $\ell_1$-penalty used in compressive sensing (see Chapter~\ref{c:cs}) or the nuclear norm in low-rank matrix recovery (see Chapter~\ref{c:lowrank}).

The idea is straightforward: Replace the gradient $\nabla f(x)$ with any subgradient $g \in \partial f(x)$.
But because the subgradient does not necessarily point in a descent direction, the objective value can fluctuate  during iterations. This often results in rather slow convergence. For a general convex function, the converges  rate is $\OOO(1/\sqrt{k})$, which means one needs $\OOO(1/\epsilon^2)$ iterations to reach an accuracy of $\epsilon$.

For example, when used in connection with the sparsity-promoting $\ell_1$-penalty, subgradient descent rarely produces exactly sparse iterates during the process; sparsity emerges only asymptotically. That is, coefficients will be very small but not exactly zero. Therefore, other methods such as proximal methods or Riemannian manifold optimization (see~Section~\ref{s:nonconvexmodels} for both) are usually preferable in practice.

\subsection{Stochastic gradient descent}\label{s:gd:stochasticGD}

Stochastic Gradient Descent (SGD) is a variant of gradient descent that updates the parameters using a randomly selected subset of data points (often just a single data point) instead of the entire dataset. This makes SGD more computationally efficient and allows it to escape local minima more easily. Note, however, that even for very small $\eta$, the update~\eqref{gradientdescent} is no longer guaranteed to reduce the value of the loss function. After all, we have traded
the mean for one sample. Hence, as Higham and Higham~\cite{higham2019deep} point out, the term ``stochastic gradient descent'' is a bit of a misnomer, and it would be more correct to call it just ``stochastic gradient''. Alas, we will be more sloppy than~\cite{higham2019deep} and will keep using the term stochastic gradient descent.

In the standard gradient descent algorithm, the gradient is computed as
$$
\nabla_x f(x) = \frac{1}{n} \sum_{i=1}^{n} \nabla_x f_i(x),
$$
where $f_i(x)$ represents the error at the $ i $-th data point.

In contrast, SGD updates the parameters using the gradient computed from a single randomly selected data point $ (x_i, y_i) $:
$$
x^{(k+1)} = x^{(k)} - \eta \nabla_x f_i(x^{(k)}).
$$
Here, the gradient is an unbiased estimate of the true gradient, which introduces noise into the optimization process. 
The noise introduced by stochastic sampling in SGD can have both positive and negative effects. On the one hand, noise can help the algorithm escape local minima by introducing random perturbations to the parameters. On the other hand, excessive noise can hinder convergence and lead to suboptimal solutions.

\subsubsection{Stochastic gradient descent and the Randomized Kaczmarz method}\label{ss:kaczmarz}

We can gain some valuable insight into the rate of convergence of SGD in the linear setting by exploiting a useful connection to the randomized Kaczmarz method~\cite{SV06}. Indeed, for linear systems, SGD is essentially equivalent to  randomized Kaczmarz, as we will demonstrate below. 

Consider a (possibly overdetermined) system of linear equations 
\begin{equation}\label{kaczmarzsystem}
    Ax=b,
\end{equation}
where $A \in \mathbb{R}^{m \times n}$ and $b \in \mathbb{R}^m$. We write the rows of $A$ as $a_1^T,\dots,a_m^T$, so the system is
$$
a_i^T x = b_i, \quad i=1,\dots,m.
$$

The Kaczmarz method~\cite{karczmarz1937} iteratively projects the current iterate onto the solution hyperplane of one equation
$$
x^{k+1}
= x^k + \frac{b_i - \langle a_i, x^k \rangle}{\|a_i\|^2} a_i,
$$
where the row index $i$ is chosen cyclically or deterministically.
Geometrically, each step is an orthogonal projection onto the hyperplane
$H_i = \{x : \langle a_i, x\rangle = b_i\}$.

The randomized Kaczmarz method selects the row index $i$ at random, with probability
\begin{equation}
\label{eq:importancesampling}   
\mathbb{P}(i = j) = \frac{\|a_j\|^2}{\|A\|_F^2},
\end{equation}
where $\|A\|_F^2 = \sum_{j=1}^m \|a_j\|^2$.

The update is
$$
x^{k+1} = x^k + \frac{b_i - \langle a_i, x^k\rangle}{\|a_i\|^2} a_i.
$$

Now, let us consider the same problem as a least-squares minimization task. We want to minimize the objective function
$$f(x) = \frac{1}{2} \|Ax - b\|^2 = \sum_{i=1}^m \frac{1}{2} (\langle a_i,x\rangle - b_i)^2.$$
In SGD, instead of computing the gradient of the entire sum, we pick one component $f_i(x) = \frac{1}{2} (\langle a_i, x\rangle - b_i)^2$ at random and update $x$ using its gradient
$$\nabla f_i(x) = (\langle a_i, x\rangle - b_i) a_i.$$
The SGD update rule with learning rate $\eta_k$ is
\begin{align*}
x_{k+1} & = x_k - \eta_k \nabla f_i(x_k) \\
        & = x_k - \eta_k (\langle a_i, x_k\rangle - b_i) a_i \\
        & = x_k + \eta_k (b_i - \langle a_i, x_k \rangle) a_i. 
\end{align*}
By comparing the two update formulas, the equivalence becomes clear: 
\begin{align*}
\text{SGD Update:} & \quad x_{k+1} = x_k + \eta_k (b_i - \langle a_i, x_k \rangle) a_i, \\
\text{RK Update:} & \quad x_{k+1} = x_k + \frac{1}{\|a_i\|^2} (b_i - \langle a_i, x_k \rangle) a_i.
\end{align*}
If we choose a specific step size for SGD such that
$\eta_k = \frac{1}{\|a_i\|^2}$,
the two methods become identical, cf.\ also~\cite{needell2014stochastic}.

The convergence rate of classical (deterministic) Kaczmarz method is highly sensitive to the ordering of the rows.
If the rows are highly correlated, the projections move the estimate in very small increments, significantly slowing down the convergence rate. Randomization comes to our rescue. Indeed, the randomization in picking the rows during the iterations of the Kaczmarz method can dramatically improve convergence guarantees as the following theorem shows, see also~\cite{SV06}.

\begin{theorem}
Let $Ax = b$ be consistent, and let $x^*$ be the unique solution in $\mathrm{range}(A^T)$.
Then the randomized Kaczmarz method converges in expectation, with the average error
$$
\mathbb{E}\big[\|x^k - x^*\|^2\big]
\le
\left(1 - \frac{\sigma_{\min}^2(A)}{\|A\|_F^2}\right)^k
\|x^0 - x^*\|^2,
$$
where $\sigma_{\min}(A)$ is the smallest nonzero singular value of $A$.
\end{theorem}

Thus, the convergence rate is linear (exponential) in expectation. We note that the expression $\|A\|_F/\sigma_{\min}(A) = \|A\|_F \|A^{-1}\|:= \tilde{\kappa}(A)$ is also known as the scaled condition number, introduced by Demmel in~\cite{demmel1988probability}.

\begin{proof}
We have
\begin{equation}
\sum_{j=1}^m |\langle z, a_j \rangle |^2 \ge  \frac{\|z\|^2}{\|A^{-1}\|^2}
 \qquad \text{for all $z \in \C^n$.}        \label{rows of A}
\end{equation}
Using the fact that $\|A\|_F^2=\sum_{j=1}^m \|a_j\|^2$ we can
write~\eqref{rows of A} as
\begin{equation} 
\label{normalizedrows}
\sum_{j=1}^m \frac{\|a_j\|^2}{\|A\|_F^2} \;
  \Bigl| \Big\langle z, \frac{a_j}{\|a_j\|} \Big\rangle \Big|^2
\ge \tilde{\kappa}(A)^{-2} \|z\|^2
 \ \ \ \text{for all $z \in \C^n$}.
\end{equation}
The main point of the proof is to view the left hand side
in~\eqref{normalizedrows} as an expectation of some random variable.
Namely, recall that the solution space of the $i$-th equation of \eqref{kaczmarzsystem}
is the hyperplane $\{y : \langle y, a_i \rangle = b_i \}$, whose normal is
$\frac{a_i}{\|a_i\|}$.
Define a random vector $Z$ whose values are the normals to all
the equations of \eqref{kaczmarzsystem}, with probabilities as in our algorithm:
\begin{equation}                    \label{Z}
Z  =  \frac{a_j}{\|a_j\|}
\ \ \ \text{with probability} \ \ \
\frac{\|a_j\|^2}{\|A\|_F^2}, \ \ \
j = 1, \ldots, m.
\end{equation}
Then \eqref{normalizedrows} says that
\begin{equation}                    \label{expectation}
\E | \langle z, Z\rangle |^2  \ge \tilde{\kappa}(A)^{-2} \|z\|^2
 \ \ \ \text{for all $z \in \C^n$.}
\end{equation}
The orthogonal projection $P$ onto the solution space of a random
equation of \eqref{kaczmarzsystem} is given by
$P z = z - \langle z - x, Z \rangle \, Z$.

Now we are ready to analyze our algorithm. We want to show that
the error $\|x_k - x\|^2$ reduces at each step in average
(conditioned on the previous steps) by at least the factor
of $(1 - \tilde{\kappa}(A)^{-2})$.
The next approximation $x_k$ is computed from $x_{k-1}$ as
$x_k = P_k x_{k-1}$, where $P_1, P_2, \ldots$ are independent realizations
of the random projection $P$.
The vector $x_{k-1} - x_k$ is in the kernel of $P_k$. It is
orthogonal to the solution space of the equation onto which $P_k$
projects, which contains the vector $x_k - x$ (recall that $x$ is the
solution to all equations). The orthogonality of these two vectors then yields
$$
\|x_k - x\|^2 = \|x_{k-1} - x\|^2 - \|x_{k-1} - x_k\|^2.
$$
To complete the proof, we have to bound $\|x_{k-1} - x_k\|^2$ from below.
By the definition of $x_k$, we have
$$
\|x_{k-1} - x_k\|  =  \langle x_{k-1} - x, Z_k \rangle
$$
where $Z_1, Z_2, \ldots$ are independent realizations of the random vector $Z$.
Thus
$$
\|x_k - x\|^2
\le \Big( 1 - \Big| \Big\langle \frac{x_{k-1} - x}{\|x_{k-1} - x\|},
                  Z_k
                  \Big\rangle \Big|^2 \Big)
  \;\|x_{k-1} - x\|^2.
$$
Now we take the expectation of both sides conditional upon
the choice of the random vectors $Z_1, \ldots, Z_{k-1}$
(hence we fix the choice of the random projections $P_1, \ldots, P_{k-1}$ and
thus the random vectors $x_1, \ldots, x_{k-1}$, and we average over the
random vector $Z_k$). Then
$$
\E_{\{Z_1,\ldots,Z_{k-1}\}}  \|x_k - x\|^2
\le \Big( 1 - \E_{\{Z_1,\ldots,Z_{k-1}\}}
                     \Big| \Big\langle \frac{x_{k-1} - x}{\|x_{k-1} - x\|},
                     Z_k
                     \Big\rangle \Big|^2 \Big)
  \; \|x_{k-1} - x\|^2.
$$
By \eqref{expectation} and the independence,
$$
\E_{\{Z_1,\ldots,Z_{k-1}\}}  \|x_k - x\|^2
\le   \big( 1 - \tilde{\kappa}(A)^{-2} \big) \; \|x_{k-1} - x\|^2.
$$
Taking the full expectation of both sides, we conclude that
$$
\E \|x_k - x\|^2
\le   \big( 1 - \tilde{\kappa}(A)^{-2} \big) \; \E \|x_{k-1} - x\|^2.$$
We complete the proof by induction.

\end{proof}

The randomized Kaczmarz method can also be
cast as an iterative sketching method, see~\cite{gower2015randomized}.

\begin{figure}[h]
\begin{center}
\includegraphics[width=90mm]{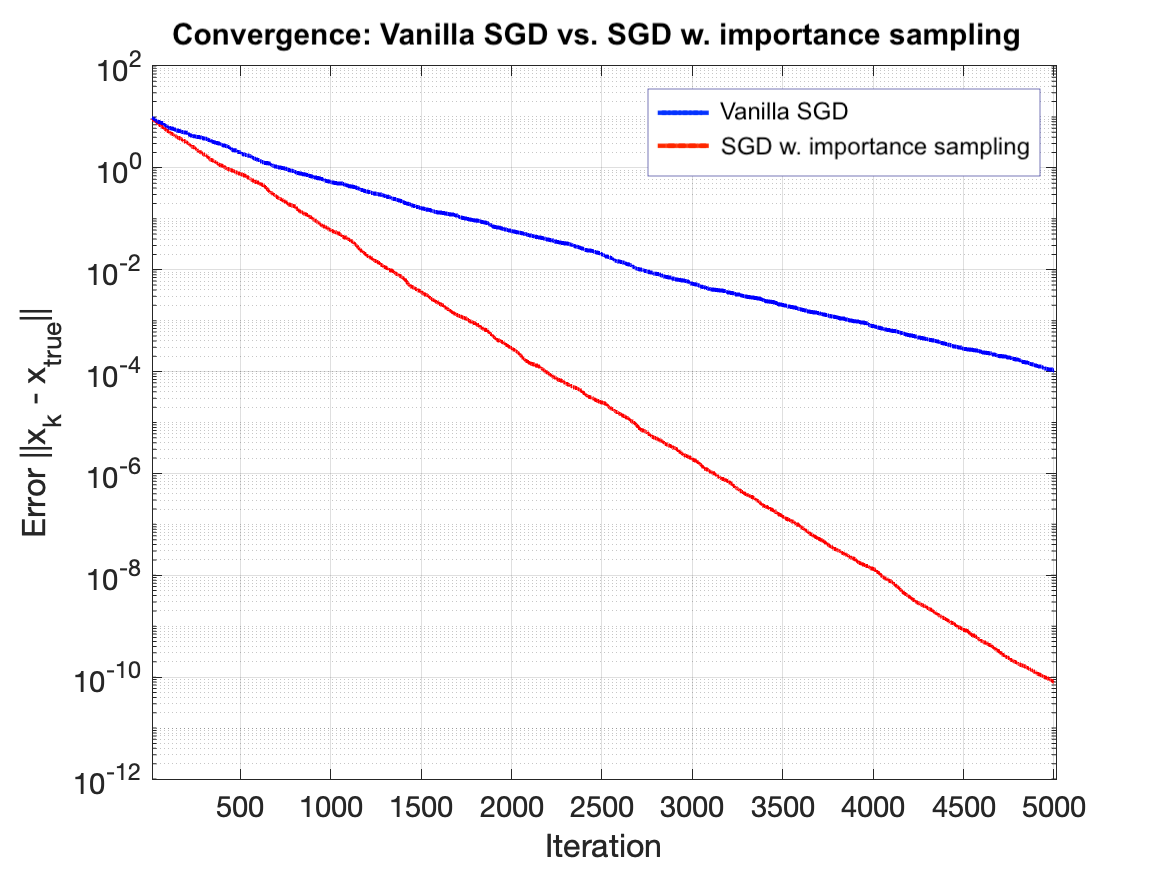}
\caption{Effect of importance sampling for Stochastic Gradient Descent.}
\label{fig:SGD}
\end{center}
\end{figure}

\bigskip
\noindent
\textbf{Stochastic gradient descent with importance sampling:}
We have seen that in the randomized Kaczmarz algorithm, rows are typically sampled with probability proportional to $\|a_i\|^2$. In the realm of SGD, this corresponds to {\em importance sampling}, which helps to reduce the variance of the stochastic gradient estimate~\cite{needell2014stochastic}. See also Figure~\ref{fig:SGD}, which 
demonstrates how importance sampling---choosing training examples based on their``influence''---can accelerate convergence (training of a linear model) compared to standard uniform sampling.
We consider a dataset with $m = 1000$ samples and $n = 100$ features. To simulate a ``real-world'' scenario, such as in sensor networks, where some data points are more informative than others, we scale the feature vectors (rows of $A$) such that their squared norms $\|a_i\|^2$ vary across two orders of magnitude. We then generate a consistent target vector $b = Ax_{\text{true}}$.
For vanilla SGD (i.e., vanilla Kaczmarz) at each iteration, a single sample $(a_i, b_i)$ is chosen with uniform probability $1/m$. This represents the standard approach in most deep learning frameworks. For SGD with importance sampling (i.e., randomized Kaczmarz with~\eqref{eq:importancesampling}), samples are selected with probability $P(i) = \|a_i\|^2 / \|A\|_F^2$. This strategy prioritizes samples that have a larger gradient magnitude, effectively spending more ``computational effort'' on the most impactful data points. The convergence plot, cf.~Figure~\ref{fig:SGD}, illustrates the significant speedup enabled by importance sampling. The point of this experiment is to illustrate the potential gain in convergence speed. A full comparison would also need to include the computational cost of calculating the norms of all the rows $a_i$ for SGD with importance sampling.

\section*{Exercises}
\addcontentsline{toc}{section}{Exercises}

\begin{myexercise}\label{ex:maxcutcontinuousoptimizationproblem}
Show that optimization problem~\eqref{eq:minbis-andspectrum} remains mathematically unchanged if the constraints $x\in\{\pm1\}^n$ are replaced by $-1\leq x_i\leq 1$ (for all $1\leq i\leq n$).
\end{myexercise}

\begin{myexercise}
Prove that if $f$ is continuously differentiable, convexity is equivalent to the {\em first-order condition} 
\begin{equation*}
 \label{def:geomconvex}
f(y) \ge f(x) +  \langle \nabla f(x), y - x\rangle, \quad \forall x,y \in \mathbb{R}^d.
\end{equation*}    
\end{myexercise}

\begin{myexercise}
Show that strong convexity implies the \PLC. 
\end{myexercise}

\begin{myexercise} \label{eq:ex_uniqueness}
This concerns the second claim in Theorem~\ref{th:convopt}.
Prove that if the function $f$ is strictly convex, then the global minimum is unique.    
\end{myexercise}

\begin{myexercise}
Prove that if  $f\in C^2$ and the Hessian $\nabla^2 f$ is positive definite, then $f$ is strictly convex.
\end{myexercise}

\begin{myexercise}
Show that the $L$-Lipschitz condition is equivalent to the notion of $L$-smoothness.    
\end{myexercise}

\begin{myexercise}
Show that the general linear program~\eqref{eq:linprog} is equivalent to the linear program (sometimes called linear program in standard form):
\begin{align*}
\begin{split}
\underset{x\in\R^n}{\text{minimize}} & \qquad  \langle d, z \rangle  \\
 \text{subject to} & \qquad \tilde{A}z = \tilde{b}, \\
 & \qquad z \ge 0,
\end{split}
\end{align*}
where $\tilde{A}$ and $\tilde{b}$ are properly augmented versions of $A$ and $b$, respectively.\\
(Hint: decompose $x$ into $x=x^{+} - x^{-}$, where $x^{+}, x^{-} \ge 0$, and introduce slack variables.)

\end{myexercise}

\begin{myexercise}
Prove Theorem~\ref{th:saddleploint}
\end{myexercise}

\begin{myexercise}
Consider the problem
\begin{align*}
    \min_x &  \quad x \\
    \text{s.t.} & \quad x^2 \le 0.
\end{align*}
Show that Slater's condition fails in this case.
\end{myexercise}

\begin{myexercise}
    
Consider the problem
\begin{align*}
    \min_{x,y} &  \quad -x \\
    \text{s.t} & \quad y - (1-x)^3  \le 0 \\
               & \quad x \ge 0 \\
               & \quad y \ge 0
\end{align*}
Show that the KKT conditions fail in this case.
\end{myexercise}

\begin{myexercise}
 Consider the problem
\begin{align*}
\min_x & \quad x^T Ax + 2b^T x \\
\text{s.t.} & \quad x^T x \le 1,
\end{align*}
where the matrix $A \in \R^{n\times n}$ is symmetric, but not positive definite, and $b\in \R^n$. Thus this is not a convex problem. Show that strong duality nevertheless holds in this case.
\end{myexercise}

\begin{myexercise}
Consider the absolute value function $f(x) = |x|$ for  $x \in \mathbb{R}$.
If we perform subgradient descent with a constant step size $\eta$ on $f(x)$, starting at $x^{(0)} = 0.5\eta$, show that the iterate will ``oscillate'' across the origin rather than staying at the sparse solution $x=0$.
\end{myexercise}

\begin{myexercise}
Find the dual problem for the {\em entropy maximization problem}:
\begin{align*}
 \min_x & \quad  \sum_{i=1}^n x_i \log x_i \\
 \text{s.t.} & \quad Ax \le b \\
                   & \quad {\mathbf 1}^\top x = 1
\end{align*}
with domain $\dom = \R^n_+$   
\end{myexercise}

\begin{myexercise}
Consider the optimization problem
\begin{align*}
 \min_x & \quad  e^{-x} \\
 \text{s.t.} & \quad \frac{x^2}{y} \le 0
\end{align*}
with variables $x,y$ and domain $\dom = \{(x,y) \, | \, y >0\}$.\\
(a) Verify that this is a convex optimization problem. Find the optimal value. \\
(b) Give the Lagrange dual problem, and find the optimal solution $\lambda^*$ and optimal value $d^*$ of the dual problem. What is the optimal duality gap? \\
(c) Does Slater’s condition hold for this problem?    
\end{myexercise}

\begin{myexercise}
Let
$$
f(x) = \frac{1}{2} x^T A x - b^T x,
$$
where $A \in \mathbb{R}^{d \times d}$ is symmetric positive definite. \\
(a) Show that gradient descent converges linearly if
$$
0 < \eta < \frac{2}{\lambda_{\max}(A)}.
$$
(b) Express the convergence rate in terms of the condition number
$$
\kappa(A) = \frac{\lambda_{\max}(A)}{\lambda_{\min}(A)}.
$$    
\end{myexercise}

\begin{myexercise}
(Euclidean projection onto the probability simplex.) 
For some probabilistic data science applications, we require that each of our data points is a discrete probability distribution (i.e., the components are positive and sum to 1). Namely, each point $x$ should belongs to the {\em probability simplex}
$$
    \Delta^{n-1}\;=\;\Bigl\{x\in\mathbb{R}^{n}\;\Big|\;
                    \sum_{i=1}^{n} x_i = 1,\;x_i\ge 0 \,\,\,\forall i\Bigr\}.
$$
However, when a data point is not in such a probability simplex we may need to project the data onto this space.
For a given vector $z\in\mathbb{R}^{n}$, the projection onto the probability simplex is the solution of
\begin{equation}\label{eq:projsimplex} 
    \min_{x\in\mathbb{R}^{n}}\;
         \tfrac12\|x - z\|_{2}^{2}
    \quad\text{s.t.}\quad
         \mathbf{1}^{\top}x = 1,\;x\ge 0.
\end{equation}

\begin{enumerate}
\item  Derive the KKT conditions for~\eqref{eq:projsimplex} using a scalar multiplier  $\nu$ for the equality constraint and a vector of multipliers  $\lambda\ge 0$ for the inequality constraints.
\item  Show from the stationarity condition that
$$x^\star_i \;=\;\max\!\{\,z_i - \nu,\;0\,\},
                 \qquad i=1,\dots,n.$$
Explain how complementary slackness implies that $\lambda_i = \max\{\,\nu-z_i,\,0\}$.
\end{enumerate}
\end{myexercise}

\begin{myexercise}
Consider the function $f(x,y)=10x^2+y^2$. This is a ``narrow valley'' potential (or a ridge, from which the name ``ridge regression'' is derived).  We want to solve $\min f(x,y)$ via gradient descent (pretending we do not know that the solution is obviously $(0,0)$). \\
(a) Compare the convergence path and the number of iterations needed to reach $\|\nabla f\| \le 10^{-4}$ for $\eta = 0.01$, $\eta = 0.05$, and $\eta = 0.1$. \\
(b) Apply Nesterov momentum with $\eta = 0.01$, $\eta = 0.05$, and $\eta = 0.1$, and compare your results to those in (a).
\end{myexercise}

\begin{myexercise} (Farkas’ Lemma) Let $A \in \mathbb{R}^{m \times n}$ and $b \in \mathbb{R}^m$. Prove that exactly one of the following two statements holds:
\begin{enumerate}
\item There exists $x \in \mathbb{R}^n$ such that
$$ Ax = b$$
\item There exists $y \in \mathbb{R}^m$ such that
$$A^T y = 0,\quad y \ge 0,\quad b^T y < 0.$$
\end{enumerate}
\begin{hint}
    Use a separating hyperplane argument.
\end{hint}
\end{myexercise}

\begin{myexercise} (Self-dual linear programs).
Consider the linear program
\begin{align*}
\max_x & \quad c^T x \\
\text{s.t.} & \quad Ax \le b \\
& \quad x \ge 0,
\end{align*}
where $A \in \mathbb{R}^{n \times n}$ is skew-symmetric $(A = -A^T)$ and $b = -c$. \\
(a) Derive the dual linear program and show that it is equivalent to the primal. \\
(b) Show that if the problem is feasible, then the primal and dual optimal values exist and are equal to zero. \\
(c) Prove that any primal feasible solution $x$ and dual feasible solution $y$ satisfy
$$
c^T x + c^T y = 0.
$$
(d) Let $x^*$ be an optimal primal solution. By Parts (a) and (b), $x^*$ is also dual feasible (with $y = x^*$).
Show that complementary slackness implies $x^T A x = 0$. 
\end{myexercise}

\begin{myexercise}
Find the subdifferential of $f(x) =\|x\|_p$  for
$p=2$ and $p=\infty$.
\end{myexercise}

\begin{myexercise}
The {\em Geometric Median} (also known as the Fermat point) of a set of points $\{x_1, x_2, \dots, x_n\} \subset \mathbb{R}^d$ is defined as the point $x$ that minimizes the sum of Euclidean distances
$$f(x) = \sum_{i=1}^{n} \|x - x_i\|_2.$$
(a) Prove that the function $g(x) = \|x - a\|_2$ is a convex function for any fixed $a \in \mathbb{R}^d$. \\
(b) Using the properties of convex functions, prove that the objective function $f(x)$ for the geometric median is convex.\\
(c) Is $f(x)$ strictly convex? Consider the case where all points $x_i$ lie on a single line.\\
(d) Compute the gradient $\nabla f(x)$ for $x \notin \{x_1, \dots, x_n\}$. Set the gradient to zero to find the optimality condition. Express the optimal solution $x^*$ as a weighted average of the points $x_i$.\\
(e) Explain why this expression is not a ``closed-form'' solution in the same sense as the arithmetic mean $\bar{x} = \frac{1}{n}\sum x_i$. (Hint: Look at where $x^*$ appears in your equation).
\end{myexercise}

\begin{myexercise}
(Optimal power allocation and the {\em water-filling algorithm}.)
In communication systems, we often want to allocate a fixed amount of power $P$ across $n$ independent channels to maximize the channel capacity. If the noise level in channel $i$ is represented by $\alpha_i$, and we allocate power $x_i$, the capacity is proportional to $\log(1 + x_i/\alpha_i)$. Maximizing this is equivalent to the following convex problem
\begin{align*}
\text{minimize} & \quad -\sum_{i=1}^n \log(\alpha_i + x_i) \\
\text{subject to} & \quad \sum_{i=1}^n x_i = P \\
& \quad x_i \geq 0, \quad i=1, \dots, n
\end{align*}
For this exercise, we normalize the total power to $P=1$.

\begin{enumerate}
\item  Write down the Lagrangian $L(x, \lambda, \nu)$ for this problem, using $\nu$ for the equality constraint and $\lambda_i$ for the inequality constraints.
\item  Derive the KKT conditions for this problem.
\item Show that the optimal power allocation $x_i^*$ must satisfy:
   $$x_i^* = \max(0, \eta - \alpha_i)$$
   where $\eta = 1/\nu$ (often called the ``water level'').
\item Explain why this is called ``water-filling.'' (Hint: Imagine $\alpha_i$ as the floor of a vessel. As you pour in a unit of ``water'' (power), which channels get filled first?)
\item Assume you have $n = 6$ channels with the noise profile 
$$\alpha = [0.1, 0.2, 0.5, 1.0, 2.0, 4.0].$$
Solve the optimization problem using CVX (in Matlab) or CVXOPT (Python). Plot the results using a bar chart: (i)~Plot the noise levels $\alpha_i$ as the base of the bars; (ii)~Plot the allocated power $x_i$ on top of the noise levels; (iii)~Draw a horizontal line representing the resulting ``water level'' $\eta$.
\end{enumerate}

\end{myexercise}

\chapter{Classification} \label{c:classification}

Classification is a fundamental task in data science and machine learning which arises in a wide range of applications, including medical diagnosis, finance,  computer vision, text analysis, and scientific data analysis.
Patient data (symptoms, lab results, imaging features) are classified to predict the presence or absence of a disease; tumors are classified as benign or malignant based on imaging or genomic features. Loan applicants are classified as low-risk or high-risk based on financial history and demographic features. Images are classified according to the object they contain (e.g., bicycle vs.\ car). News articles are classified into categories such as politics, sports, technology, or finance. In the area of protein folding, classification models identify the functional category of a protein based on its amino acid sequence, which is essential for drug discovery.
In each case, the objective is to assign observations to a finite set of categories based on measured features.

While clustering and classification share the fundamental objective of partitioning the feature space, they have essential differences:  In clustering, which falls in the realm of unsupervised learning, the goal is to discover structure (groups) in unlabeled data. In classification, which falls in the realm of supervised learning, the goal is to assign data points to known groups using labeled examples. Clustering is primarily descriptive: it uses optimization to tell us about the structure of the current dataset. Classification is predictive: it uses optimization to create a rule that generalizes to unseen data.

In this chapter we study two central approaches to  classification:
\begin{enumerate}
  \item {\em Logistic regression}, motivated by probabilistic modeling and likelihood maximization~\cite{cox1958regression};
  \item {\em Support vector machines}, motivated by geometric margin maximization~\cite{cortes1995support}.
\end{enumerate}
Both methods lead to convex optimization problems and play a central role in modern data science.  

Deep learning based classification, which has emerged as the dominant technique for large scale datasets, will be discussed in Chapter~\ref{c:deeplearning}.

\section{Binary classification and linear classifiers}

We begin our exploration of classification (or discrimination) with the special but very important and instructive case of {\em binary classification}:
Given labeled data
$\{(x_i,y_i)\}_{i=1}^n$ where the data $x_i \in \mathbb{R}^d$ and the binary labels $y_i$ are usually represented by 
$\{-1,+1\}$
, the goal is to learn a rule that predicts the label $y$ of a new observation $x$.

In a nutshell, a linear classifier is one whose decision boundaries are linear. Formally, a linear classifier is defined by parameters $(w,b) \in \mathbb{R}^d \times \mathbb{R}$ and the decision rule
$$
f(x) = \operatorname{sign}(w^T x + b).
$$
The decision boundary
$$
\{x \in \mathbb{R}^d : w^T x + b = 0\}
$$
is a hyperplane that separates the input space into two half-spaces.
Learning a classifier thus amounts to choosing $w$ and $b$ so that points with different labels lie on different sides of this hyperplane. 

\begin{figure}
\begin{center}
    \includegraphics[width=.5\textwidth]{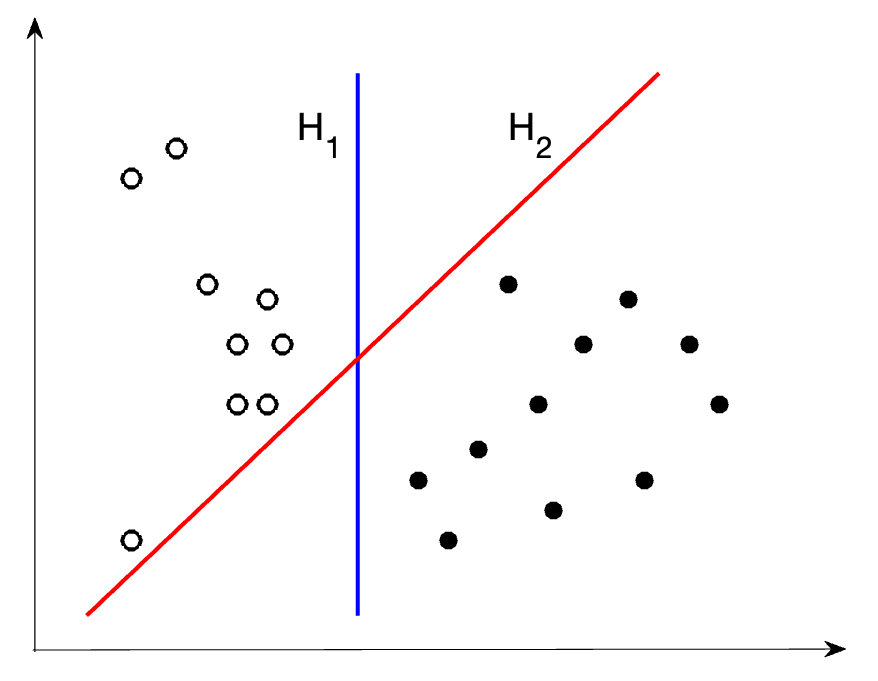}
    \end{center}
    \caption{The open and filled circles represent data from two different classes. 
    These two sets can be linearly separated by a hyperplane. The depicted hyperplanes $H_1$ and $H_2$ are two out of infinitely many possible choices for such a separating hyperplane. Which hyperplane shall we choose a classifier?}
    \label{fig:linearclass}
\end{figure}

Figure~\ref{fig:linearclass} shows a simple example of two sets of points and two  possible choices of linear classifiers (hyperplanes) that classify the two sets.
Which hyperplane shall we choose among the potentially infinitely many candidates? Is there an optimal choice? If so, how can we find it?

\section{Logistic regression}\label{s:logisticregression}

Logistic Regression, which has some of its roots\footnote{See~\cite{Cramer2002} for a detailed account on the history of logistic regression.} in the works of  Verhulst~\cite{verhulst1844recherches}, Berkson~\cite{berkson1944application} and Cox~\cite{cox1958regression}, is one of the most popular algorithms for (binary) classification problems.
 While the name contains ``regression,'' it is fundamentally a tool for estimating the probability that an input belongs to a specific category.

We start by considering a random variable (the binary label) $y\in \{0,1\}$, with $\P(y=1) = p$ and $\P(y=0) = 1-p$,
where $p \in [0,1]$. The probability $p$ is assumed to depend on one or more explanatory variables, represented by the vector $x \in \R^n$. For instance, let us say we want to predict whether a patient has a certain disease ($y=1$) or not ($y = 0$). The probability of having this disease is $q$, which is modeled as being based on explanatory variables such as age, gender, weight,  blood pressure, temperature, collected in the vector $x.$

In linear regression, we model the output as a linear combination of inputs 
$$z = w^T x + b.$$ 
However, $z$ can range from $-\infty$ to $+\infty$, making it unsuitable for representing a probability, which must be constrained to the interval $[0, 1]$.

To map the linear predictor to a probability space, we use the logistic sigmoid function $\sigma(z)
= (1 + e^{-z})^{-1}$.
This gives
$$\P(y=1 | x; w,b) = \sigma(w^T x + b) = \frac{1}{1 + \exp(-(w^T x + b))}.$$

Logistic regression can be viewed as a linear model for the log-odds (logit) of the probability $p$. If $p = \P(y=1|x)$, the odds are defined as $p/(1-p)$. An easy calculation shows that taking the natural logarithm gives
$$\log\Big( \frac{p}{1-p} \Big) = w^T x + b.$$

This reveals that the decision boundary (where $p=0.5$) is the hyperplane defined by $w^T x + b = 0$.

Unlike linear regression, which uses ordinary least squares, the parameters $w$ and $b$ in logistic regression are typically estimated using Maximum Likelihood Estimation. Assuming the observations are independent and identically distributed (i.i.d.), the likelihood function for $n$ observations is\footnote{Recall that the probability mass function of a Bernoulli variable is $\P(Y=y) = p^y (1-p)^{1-y}\,\, y \in \{0,1\}$.}
$$L(w,b) = \prod_{i=1}^n [\sigma(w^T x_i + b)]^{y_i} [1 - \sigma(w^T x_i + b)]^{1 - y_i}.$$

Maximizing the likelihood is equivalent to minimizing the {\em negative log-likelihood}, 
$$
J(w,b) = - \frac{1}{n} \sum_{i=1}^n
[y_i \log \sigma( w^T x_i + b) + (1-y_i)\log\bigl(1-\sigma( w^T x_i + b)\bigr)],
$$
which is also known as the \emph{logistic loss} or \emph{cross-entropy loss}.

It is not difficult to see that $J(w)$ is a convex function.
As a result, logistic regression admits a unique global minimizer (up to degeneracies in the data) and can be solved efficiently using first-order or second-order optimization methods, such as gradient descent or Newton’s method.

The gradient of the loss  is remarkably elegant
$$
\nabla_{w,b} J(w,b) =
\frac{1}{n}\sum_{i=1}^n
\bigl(\sigma(w^T x_i + b) - y_i\bigr)x_i,
$$
This gradient resembles the one found in linear regression, though the underlying hypothesis is non-linear.

In the basic logistic regression model,  if the data is linearly separable, the optimization algorithm will try to drive the weights $w$ toward infinity.
This is so, since a larger $\|w\|$ makes the sigmoid function $\sigma(w^T x + b)$ steeper, pushing the predicted probabilities closer to  $1$ or $0$. This leads to overfitting and the model becomes  sensitive to small fluctuations in the training data.
It also can lead to numerical instability and to 
``exploding'' gradients.

There are several pathways to address this problem.
To derive the regularized version, we can for instance treat $w$ not as a fixed parameter to be found, but as a random variable with a prior distribution. This is known as Maximum A Posteriori (MAP) estimation.

Here, we assume a Gaussian prior on the weights $w$ centered at zero, namely $w \sim {\mathcal N}(0,\eta^2 I)$. However, we do not place any prior on $b$. We can express this ``flat'' prior on $b$ as
$b \sim {\mathcal N}(0,\gamma^2)$, where $\gamma \to \infty$ (this is sometimes written as
$b \sim {\mathcal N}(0,\infty)$).

According to Bayes’ Theorem~\cite[Chapter 7.2.3]{casella2024statistical}, the posterior probability of the weights $w$ given the data $x$ is
\begin{equation}\label{eq:bayes}
\pi(w,b | x, y)  \propto  \pi(w)\pi(b) \pi(y | x, w,b) ,
\end{equation}
where $\pi(y | x, w,b)$ is the standard likelihood function used in basic logistic regression.
Colloquially,~\eqref{eq:bayes} is 
$$\quad \,\, \text{posterior}  \propto \text{prior} \times \text{likelihood.}$$    
To find the MAP estimate, we maximize the log-posterior
$$\log \pi(w,b | x, y) =  \log \pi(w)+\log \pi(b) + \log \pi(y | x, w,b)  + C,$$
where $C$ is a constant independent of $w$.

Substituting the ``flat'' prior for $b$, we have
$\log \pi(b) = - \frac{b}{2\gamma^2} + C$, which  gives $\log \pi(b) \to 0$ as $\gamma \to \infty$.

We now substitute the Gaussian prior for $w$ and obtain
$$\log \pi(w) = - \frac{1}{2} w^T (\eta^2 I)^{-1} w +C = -\frac{1}{2\eta^2} \|w\| +C.$$
Thus, by flipping the sign to convert this into a minimization problem, we arrive at the $\ell_2$-regularized objective
$$J(w, b) =-\frac{1}{n} \sum_{i=1}^n \left[ y_i \log \sigma(w^T x_i +b) + (1-y_i) \log(1-\sigma(w^T x_i +b)) \right] + \lambda \|w\|^2,$$
where the regularization parameter $\lambda > 0$ controls the trade-off between fitting the data and keeping the weights small.

The $\ell_2$-term $\|w\|^2$ is strictly convex. Adding a strictly convex function to a convex function results in a strictly convex objective. Hence, for $\lambda>0$, the objective is strictly convex and thus has a unique minimizer, even in the presence of data degeneracies. It also makes the objective strongly convex, which greatly improves convergence of gradient descent (see Chapter~\ref{c:optimization}).

Note that logistic regression outputs probabilities and not binary decisions. Thus, strictly speaking, logistic regression by itself is not a classifier.  
Logistic regression produces a {\em score}
$$
s(x) := \sigma(w^T x + b) \in (0,1),
$$
which is interpreted as an estimated posterior probability.
Classification is then typically performed by applying the decision rule
$$
\hat y_\tau =
\begin{cases}
1, & s(x) \ge \tau,\\
0, & s(x) < \tau,
\end{cases}
$$
for a particular choice of $\tau \in [0,1]$
to the output of logistic regression.
This is equivalent to a ($\tau$-dependent) {\em linear decision boundary}
$$
w(\tau)^T x + b(\tau) = 0,
$$
which defines a hyperplane separating the two classes. Thus, logistic regression is a linear classifier, although it differs from margin-based methods such as support vector machines in its objective and interpretation.

But how shall we choose the threshold $\tau$? While it seems natural and tempting to set $\tau = \frac{1}{2}$, it is not clear that this would be the best choice. Lowering the threshold typically reduces false negatives while increasing false positives, and raising it has the opposite effect. So how shall we go about choosing $\tau$? This question deserves some consideration, and is directly related to a bigger question, namely how should we evaluate a supervised learning classifier? We take a brief interlude and explore this question in the next section.

\section{Evaluation of classifiers, confusion matrix, and ROC}

The confusion matrix provides a simple way to shed some light  on the performance of an algorithm’s predictions against the actual ground truth for binary classification.
It breaks down the predictions into four categories:
\begin{itemize}
\item True Positive (TP): The model correctly predicted a positive outcome i.e the actual outcome was positive.
\item True Negative (TN): The model correctly predicted a negative outcome i.e the actual outcome was negative.
\item False Positive (FP): The model incorrectly predicted a positive outcome i.e the actual outcome was negative. It is also known as a Type I error.
\item False Negative (FN): The model incorrectly predicted a negative outcome i.e the actual outcome was positive. It is also known as a Type II error.
\end{itemize}

\begin{table}[h!]
\centering
\renewcommand{\arraystretch}{1.3}
\setlength{\tabcolsep}{10pt}
\begin{tabular}{cc>{\centering\arraybackslash}p{2.8cm}>{\centering\arraybackslash}p{2.8cm}}

  \multicolumn{2}{c}{} 
    & \multicolumn{2}{c}{\textbf{Predicted}} \\[0.4ex]

  \multicolumn{2}{c}{}
    & \textbf{Positive}
    & \textbf{Negative} \\[0.3ex]

  \cline{3-4}

  \multirow{2}{*}{\rotatebox[origin=c]{90}{\textbf{Actual}}}
    & \textbf{Positive}
    & \multicolumn{1}{|>{\centering\arraybackslash}p{2.8cm}|}{\makecell{True Positive \\ (TP)}}
    & \multicolumn{1}{>{\centering\arraybackslash}p{2.8cm}|}{\makecell{False Negative \\ (FN)}} \\

  \cline{3-4}

    & \textbf{Negative}
    & \multicolumn{1}{|>{\centering\arraybackslash}p{2.8cm}|}{\makecell{False Positive \\ (FP)}}
    & \multicolumn{1}{>{\centering\arraybackslash}p{2.8cm}|}{\makecell{True Negative \\ (TN)}} \\

  \cline{3-4}

\end{tabular}
\caption{Confusion matrix for binary classification.}
\end{table}

Let us take this opportunity to introduce a few more terms that are often used in practice when evaluating the performance of a classifier.
{\em Sensitivity} (true positive rate) is the probability of a positive test result, conditioned on the individual truly being positive.
{\em Specificity} (true negative rate) is the probability of a negative test result, conditioned on the individual truly being negative.
False positives, i.e., negative instances incorrectly predicted as positive are also referred to as {\em Type I error}, while false negatives, i.e., positive instances incorrectly predicted as negative are called {\em Type II error}.

From the counts appearing in the confusion matrix, several frequently used metrics can be derived:
\begin{align}\label{def:recall}
\begin{split}
\text{Accuracy} = \frac{\text{TP}+\text{TN}}{\text{TP}+\text{FP}+\text{TN}+\text{FN}},& \quad
\text{Precision} = \frac{\text{TP}}{\text{TP}+\text{FP}},\\
\text{F1} = 2 \cdot \frac{\text{Precision}\cdot \text{Recall}}{\text{Precision + Recall}}, & \quad
\text{Recall (TPR)} = \frac{\text{TP}}{\text{TP}+\text{FN}}.
\end{split}
\end{align}

\begin{figure}[h!]
\begin{center}
\includegraphics[width=0.5\textwidth]{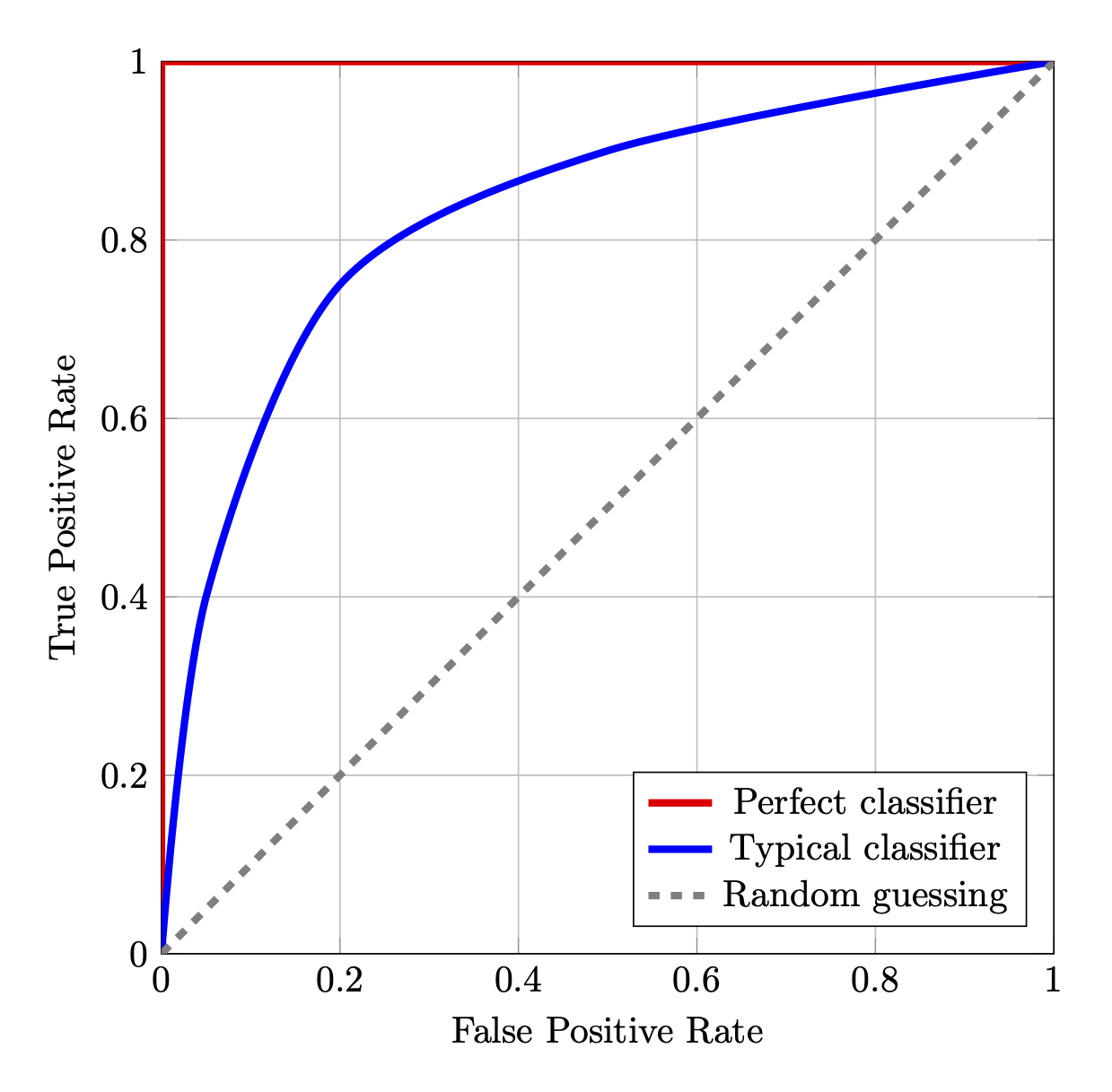}
\caption{An ROC (Receiver Operating Characteristic) curve represents the performance of a classification model by plotting the True Positive Rate against the False Positive Rate across all possible classification thresholds.}
\label{fig:ROC}
\end{center}
\end{figure}

To visualize the tradeoff between True Positive Rate (TPR) and the False Positive Rate (FPR) as a function of the threshold, one commonly uses the Receiver Operating Characteristic (ROC) curve\footnote{The Receiver Operating Characteristic curve was originally developed in the context of radar signal detection, hence the name.}, see also Figure~\ref{fig:ROC}. A random classifier produces a diagonal line (gray line),  while a useful classifier lies above this line (blue curve), and a perfect classifier reaches the top-left corner $(0,1)$ (red curve). 

The Area Under the Curve (AUC) summarizes the ROC curve as a single number. If AUC=1, we have perfect class separation, while AUC=0.5 corresponds to random guessing. 
The AUC can be interpreted as the probability that the classifier assigns a higher score to a randomly chosen positive sample than to a randomly chosen negative sample.

In situations with class imbalance, the ROC curve can be misleading. The Precision–Recall (PR) curve may be preferable in that case. It plots Precision versus Recall (both are defined in~\eqref{def:recall}) across thresholds. As such, it highlights performance on the positive class, which is critical in rare-event detection (e.g., fraud, disease diagnosis).

\medskip
So, where does that leave us regarding how to choose the threshold $\tau$ for logistic regression, the question that triggered this excursion into evaluation of classifiers? The ROC curve does not determine a unique decision threshold $\tau$. The default choice $\tau = 0.5$ is appropriate when class priors are approximately equal and false positives and false negatives have about equal cost. It is often used as a baseline, but rarely optimal in practice.  Maximizing the F1 score is a pragmatic way to select a threshold when the impact of different error types is unknown~\cite{van2004geometry}.
Another frequently used choice is to pick the $\tau$ that minimizes the distance to the top-left corner $(0,1)$ (the optimal point of a perfect classifier), i.e.,
$$\tau^* = \argmin_{\tau} \sqrt{(1-\text{TPR}(\tau))^2 +
\text{FPR}(\tau)^2},$$
which treats false positives and false negatives symmetrically.

Ultimately, the threshold is a decision parameter, not a model parameter. As such,
the ``right'' choice of $\tau$ depends on the application-specific trade-off between false positives and false negatives.

\section{Support Vector Machines}\label{s:svm}

A natural idea to construct a linear classifier would be to find an affine function that approximately classifies the points such that one minimizes the number of points misclassified. However, this is in general a computationally infeasible combinatorial optimization problem.  One heuristic to circumvent the computational hardness is to instead try to maximize a ``gap'' between classes, relying only on a critical subset of points called Support Vectors. 
Support Vector Machines (SVMs)~\cite{cortes1995support} thus represent a shift from the probabilistic perspective of Logistic Regression to a geometric perspective. An important property of support vector machines is that the
determination of the model parameters corresponds to a convex optimization problem (as was also the case for logistic regression).

\subsection{Hard-margin SVM}

Consider a linearly separable dataset $\{(x_i, y_i)\}_{i=1}^n$ where $y_i \in \{-1, 1\}$. We seek a hyperplane defined by $w^T x + b = 0$ that separates the two classes.

Let us first determine the distance from a point $x_i$ to this hyperplane.
We can decompose $x_i$ as
$$x_i = x_i^* + r \frac{w}{\|w\|},$$
where $x_i^*$ is the orthogonal projection of $x_i$ onto the hyperplane (so $w^T x_i^* + b = 0$), and $r$ is the signed  distance -- positive if $x_i$ is on the same side as $w$, negative on the other side. We want to find $r$. Since $x_i^* = x_i - r\frac{w}{\|w\|}$ lies on the hyperplane
$$w^T\!\left(x_i - r\frac{w}{\|w\|}\right) + b = 0,$$
which, after some basic linear algebra, can be written as 
$$r = \frac{w^T x_i + b}{\|w\|}.$$
Hence, the signed distance from $x_i$ to the hyperplane, measured along $w/\|w\|$, is $r$. Its magnitude is the geometric (unsigned) distance.

Now let us incorporate the label $y_i$ to get a signed margin.
The expression $r = (w^T x_i + b)/\|w\|$ is positive when $x_i$ is on the $w$-side of the hyperplane and negative on the other side.  If $y_i = +1$, then $x_i$ should be on the positive side, so $w^T x_i + b > 0$, and $r > 0$. If $y_i = -1$, then $x_i$ should be on the negative side, so $w^T x_i + b < 0$, and $r < 0$.

In both cases, $y_i \cdot r > 0$. Multiplying by $y_i$ therefore flips the sign for negative-class points, making every correctly classified distance positive, giving
$$\gamma_i = y_i \cdot r = \frac{y_i(w^T x_i + b)}{\|w\|}.$$

\begin{figure}
\begin{center}
    \includegraphics[width=.6\textwidth]{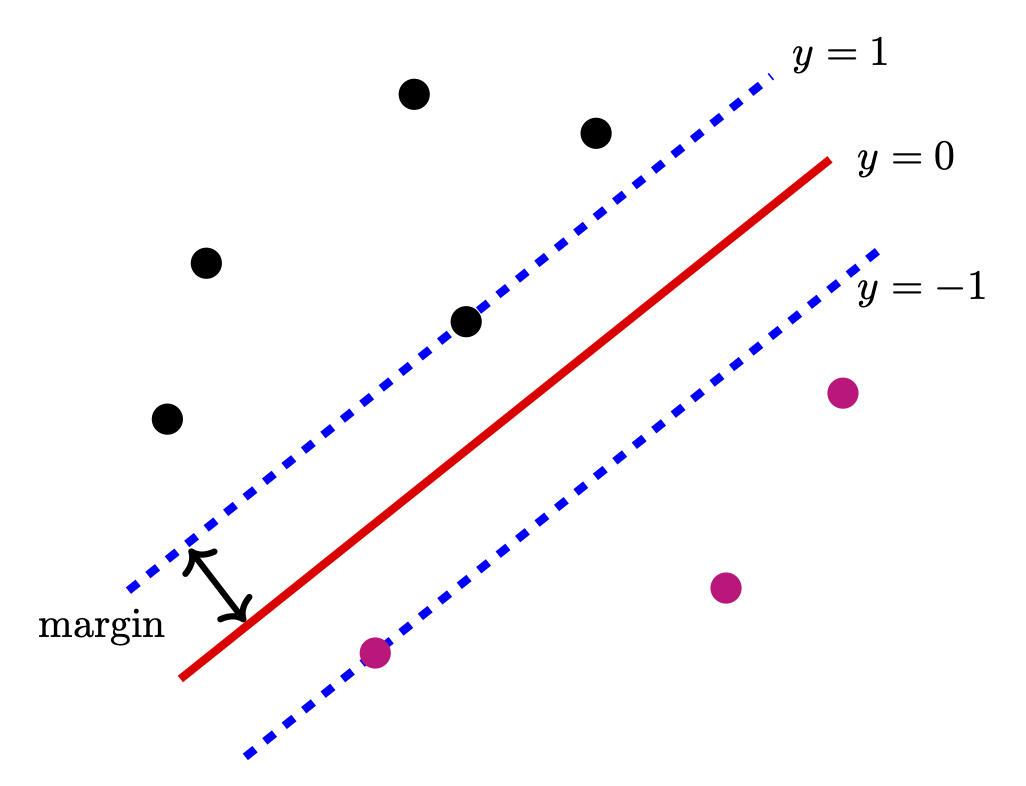}
    \end{center}
    \caption{The margin is defined as the distance between the decision boundary (the red line) and the closest data point.}
    \label{fig:margin}
\end{figure}

The goal of the SVM is to maximize the margin $\gamma$, which is the distance to the nearest point,
see also Figure~\ref{fig:margin}.

For convenience, we scale $w$ and $b$ such that the nearest points satisfy $y_i(w^T x_i + b) = 1$. This is called canonical scaling in the SVM literature and the expression $y_i(w^T x_i + b)$ is also called functional margin. Without loss of generality we can set this quantity to unity, since the direction of the hyperplane is unchanged when we scale both $w$ and $b$ by the same factor. The margin then becomes $1/\|w\|$.

Maximizing $1/\|w\|$ is equivalent to minimizing $\frac{1}{2}\|w\|^2$:
\begin{align}\label{svm}
\begin{split}
\min_{w, b} & \quad \frac{1}{2}\|w\|^2 \\
\text{s.t.} & \quad y_i(w^T x_i + b) \geq 1, \quad \forall i.
\end{split}
\end{align}

To solve the constrained optimization, we use the method of Lagrange multipliers, as described in Chapter~\ref{s:duality}. The Lagrangian becomes
$$\mathcal{L}(w, b, \alpha) = \frac{1}{2}\|w\|^2 - \sum_{i=1}^n \alpha_i [y_i(w^T x_i + b) - 1],$$
where $\alpha_i \geq 0$ are the Lagrange multipliers.

Setting the partial derivatives of $\mathcal{L}$ with respect to $w$ and $b$ to zero gives
\begin{align*}
& \nabla_{w} \mathcal{L} = w - \sum \alpha_i y_i x_i = 0 \implies w = \sum_{i=1}^n \alpha_i y_i x_i, \\
& \frac{\partial \mathcal{L}}{\partial b} = \sum \alpha_i y_i = 0.
\end{align*}
Substituting these back into the Lagrangian yields the dual objective
\begin{align}\label{SVMdual}
\begin{split}
 \max_{\alpha} & \quad \sum_{i=1}^n \alpha_i - \frac{1}{2} \sum_{i,j=1}^n \alpha_i \alpha_j y_i y_j \langle x_i, x_j\rangle \\
\text{s.t.}      & \quad \sum \alpha_i y_i = 0,  \\
& \quad \,\, \alpha_i \geq 0.
\end{split}
\end{align}

By the KKT Complementary Slackness condition~\eqref{eq:slackness}, 
$$\alpha_i [y_i(w^T x_i + b) - 1] = 0.$$ 

This  equation dictates exactly which points ``exist'' in the final model. To see this, note that based on this KKT condition, all training points $x_i$ fall into two mutually exclusive categories:
\begin{enumerate}
\item  \textbf{Non-Binding Constraints (the ``disappearing'' points):}
If a point $x_i$ lies strictly outside the margin, then its margin is greater than 1, i.e., 
$$y_i(w^T x_i + b) > 1.$$
For the product in the KKT condition to be zero, the Lagrange multiplier must be
$$\alpha_i = 0.$$
Looking at the formula for the weight vector, 
\begin{equation}\label{eq:svmweight}
w = \sum_{i=1}^n \alpha_i y_i x_i,
\end{equation}
we see that these points make no contribution to the sum. If one were to remove these points from the dataset and retrain the model, the resulting hyperplane would remain identical. In the context of complementary slackness (cf.~Chapter~\ref{ss:KKT}), we also refer to such data points as having ``slack''.

\item \textbf{Binding Constraints (the support vectors):}
If a point $x_i$ lies exactly on the margin, then
$$y_i(w^T x_i + b) = 1.$$
In this case, the constraint is active, and $\alpha_i$ can be non-zero ($\alpha_i > 0$). These are the {\em support vectors}: they are the only points that ``support'' the hyperplane.
\end{enumerate}
Let $S = \{i : y_i(w^T x_i + b) = 1\}$ be the set of indices of the support vectors and denote the optimal weight vector by $w^*$. We can rewrite the summation~\eqref{eq:svmweight} as
$$w^* = \sum_{i \in S} \alpha_i y_i x_i.$$
Thus, $w^*$ is a linear combination of only the support vectors. 

We summarize this insight in the following theorem:
\begin{theorem}[Sparsity of the SVM]
The optimal weight vector $w^*$ of a Support Vector Machine is determined solely by the subset of training points for which the canonical functional margin is exactly $1$.
\end{theorem}

\begin{figure}
\begin{center}
    \includegraphics[width=0.6\textwidth]{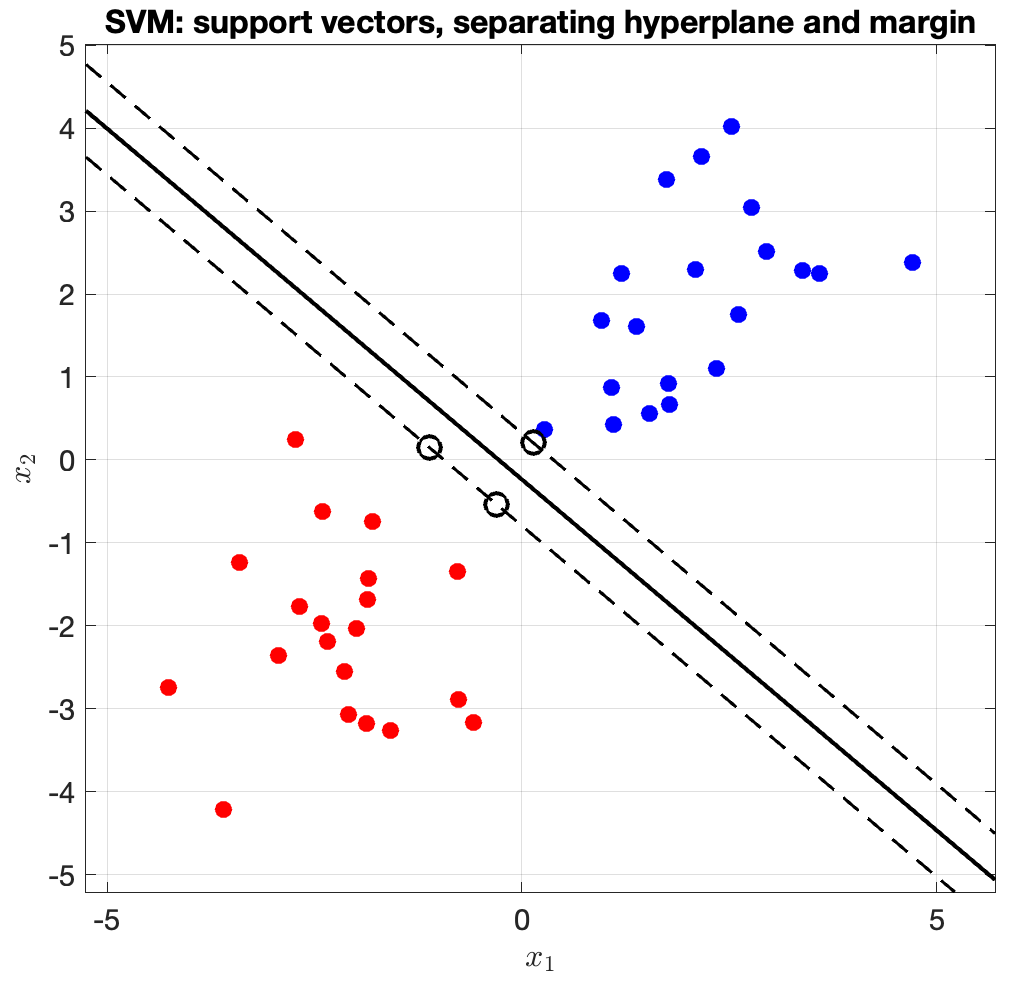}
    \end{center}
    \caption{By solving the optimization problem~\eqref{svm} we find an affine function that gives the largest gap  between the two sets. Geometrically, we are finding the thickest slab that separates the two sets of points. In the plot, open circles represent the support vectors.}
    \label{fig:svm}
\end{figure}

Figure~\ref{fig:svm} illustrates SVM in action. Depicted are a linearly separable
data set, the separating hyperplane computed by solving~\eqref{svm}, the associated maximal margin and the support vectors.

\subsection{Soft-margin SVM}

In real-world data, classes often overlap. To accommodate 
a trade-off between the margin (the distance between the
two critical hyperplanes) and the classification error on the training set, we introduce slack variables $\xi_i \geq 0$ that allow points to be on the ``wrong side'' of the margin.
Under this {\em soft-margin} constraint, the primal problem takes the form 
\begin{align*}
\min_{w, b, \xi} & \quad \frac{1}{2}\|w\|^2 + \mu \sum_{i=1}^n \xi_i, \\
\text{s.t.} & \quad y_i(w^T x_i + b) \geq 1 - \xi_i, \\
& \quad \xi_i \ge 0.
\end{align*}
This is a convex quadratic program  with linear constraints.
Here, $\mu$ is a hyperparameter that controls the trade-off between maximizing the margin and minimizing classification errors. In the limit $\mu \to \infty$, we will recover the earlier hard-margin support vector machine.

The goal of the soft-margin SVM is to find a classifier that balances margin maximization and training errors, yielding robustness to noise and better generalization. Soft-margin SVM is also more robust to outliers, since it prevents a single anomalous data point from dictating the entire boundary.

\begin{figure}
\begin{center}
    \includegraphics[width=.62\textwidth]{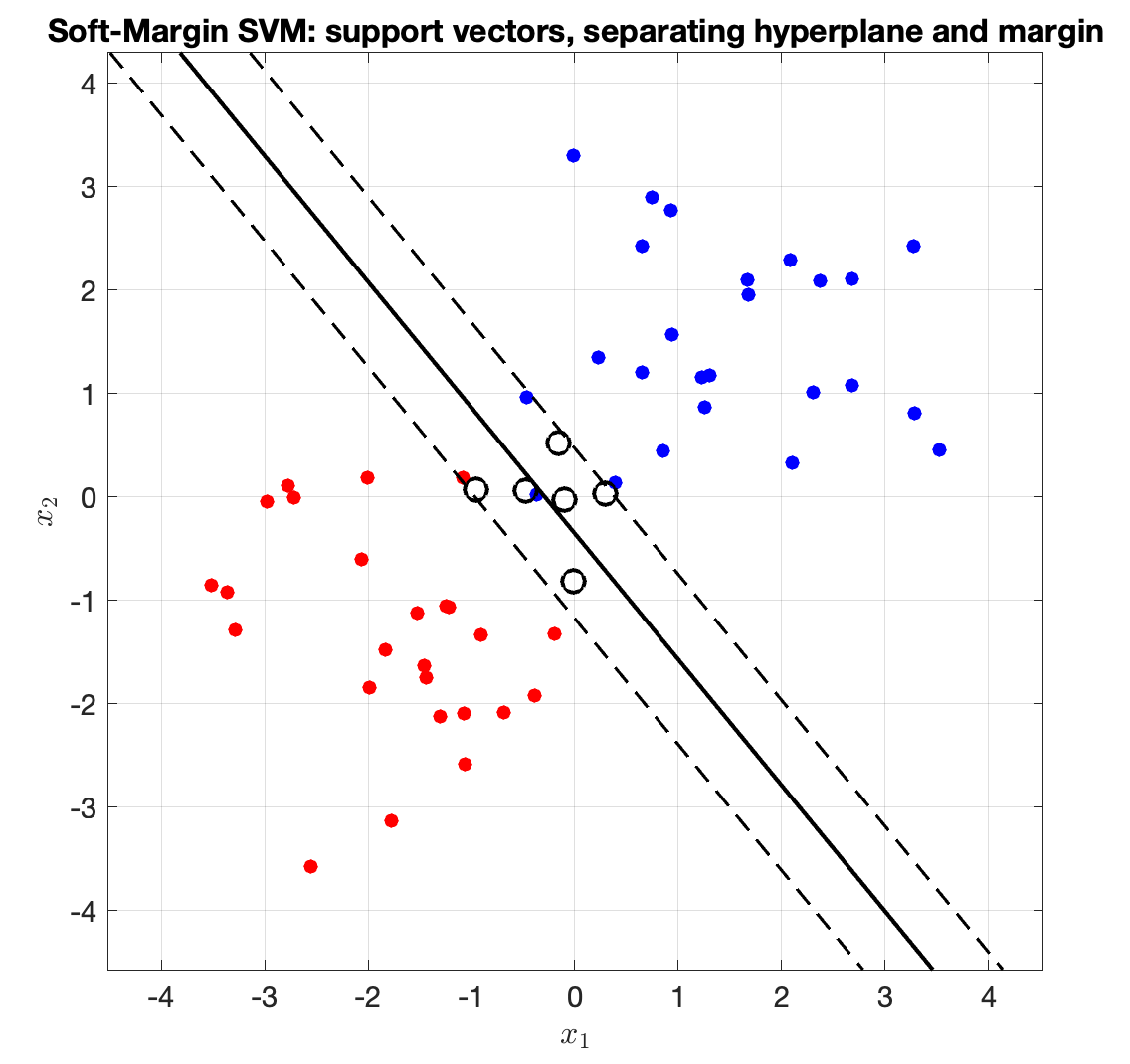}
    \end{center}
    \caption{Classification via soft-margin SVM. 
    Open circles represent the support vectors. The larger margin leads to misclassification of one (blue) point in this example.  Soft-margin SVM 
    improves generalization by allowing for a trade-off between maximizing the margin and minimizing classification errors.}
    \label{fig:softsvm}
\end{figure}

It is instructive to rewrite the soft-margin SVM as an unconstrained optimization problem.
To that end, we first express $\xi_i$ solely in terms of $w, b$, and the data. From the constraints, we see that
if the point is correctly classified and outside the margin ($y_i(w^T x_i + b) \geq 1$), the smallest possible $\xi_i$ that satisfies the constraints is $0$. On the other hand,
if the point violates the margin ($y_i(w^T x_i + b) < 1$), the smallest $\xi_i$ that satisfies the first constraint is $1 - y_i(w^T x_i + b)$.
This can be written concisely as
$$\xi_i = \max(0, 1 - y_i(w^T x_i + b)).$$
By substituting the expression for $\xi_i$ back into the primal objective, we get
$$\min_{w, b} \,\,\,  \sum_{i=1}^n \max(0, 1 - y_i(w^T x_i + b))  + \lambda \|w\|^2.$$

The term $\max(0, 1 - z)$ is known as the {\em hinge loss}.
It is called ``hinge'' because of its shape: it is flat (zero) for values $\ge 1$ and increases linearly for values $<1$.
As the hinge loss is an upper bound on the 0-1 indicator loss (misclassification count), minimizing the SVM objective provides a convex surrogate for minimizing classification error.

\medskip

Logistic regression can also be considered as a variant of the SVM where the hinge function $\max(0,1 - z)$ is replaced with the (smooth and convex) logistic
function $\ln(1 + e^{-z})$.
Because the hinge loss is exactly zero for points far from the boundary, the gradient is also zero for those points. This is why SVMs are sparse (only support vectors matter).
But unlike the logistic loss, the hinge loss is not differentiable at the ``elbow'' (where $y_i(w^T x_i +b)=1$). In practice, we use sub-gradient descent to solve this.

\begin{figure}
\begin{center}
    \includegraphics[width=.9\textwidth]{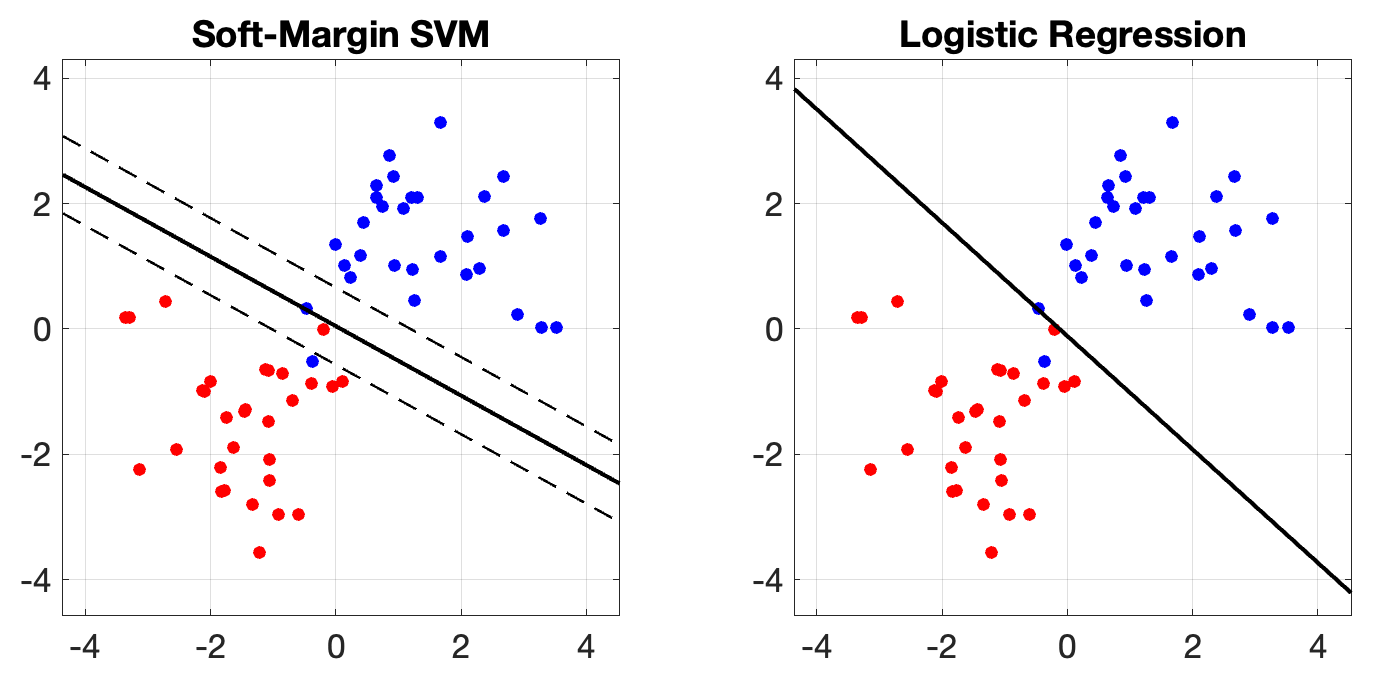}
    \end{center}
    \caption{A comparison between logistic regression and soft-margin SVM.
    In logistic regression even a point far away from the decision boundary exerts a tiny ``pull'' on the solution. In contrast, in SVM, the hinge loss has a flat region. Once a point is correctly classified and beyond the margin, its gradient becomes exactly zero. It ``disappears'' from the optimization.
    }
    \label{fig:compsvmlogreg}
\end{figure}

Figure~\ref{fig:compsvmlogreg} shows a comparison between logistic regression and soft-margin SVM. As we can see, the methods produce somewhat different decision boundaries. In logistic regression every point has a non-zero influence on the gradient because the sigmoid derivative is never exactly zero. In contrast, in SVM, once a point is correctly classified and beyond the margin, its gradient becomes exactly zero. 

\subsection{Kernel SVM}

The SVMs we discussed so far are limited to datasets that are (approximately) linearly separable. However, for many real-world datasets this may not be the case. We can address this by mapping the input features $x \in \mathbb{R}^d$ into a higher-dimensional feature space  via a non-linear mapping function $\Phi(x)$ with the aim that in this higher-dimensional space, a linear separation may become possible.

However, if we perform the mapping $\Phi(x)$ explicitly, we face a major computational challenge.
Indeed, if we want to work in a high-dimensional space, the SVM dual objective requires us to compute the inner product of these transformed vectors. Thus, the kernel-analog of~\eqref{SVMdual}  takes the form
\begin{align}\label{SVMdualkernel}
\begin{split}
     \max_{\alpha} & \quad 
 \sum \alpha_i - \frac{1}{2} \sum \alpha_i \alpha_j y_i y_j \langle \Phi(x_i), \Phi(x_j) \rangle \\
\text{s.t.} & \quad \alpha_i \geq 0 \\
            & \quad \sum \alpha_i y_i = 0. 
 \end{split}           
\end{align}
In high dimensions calculating $\Phi(x_i)$ and then the inner product $\langle \Phi(x_i), \Phi(x_j)\rangle$ for every pair of points is computationally impossible.

The {\em kernel trick} introduced in Chapter~\ref{s:kernellearning}, comes to our rescue: 
We can simply replace the inner product $\langle \Phi(x_i), \Phi(x_j) \rangle$  in~\eqref{SVMdualkernel} with the kernel function $K(x_i, x_j)$. We are now effectively computing the boundary in a high-dimensional space while only performing simple arithmetic in the original low-dimensional space.
This modification of SVM is referred to as {\em kernel SVM}~\cite{boser1992training,scholkopf2002learning}.
The Gaussian kernel $K(x,x') = \exp (-\|x-x'\|^2/2\eps)$ is a common default choice for kernel SVM.

\begin{figure}
\begin{center}
\includegraphics[width=0.45\textwidth,height=45.5mm]{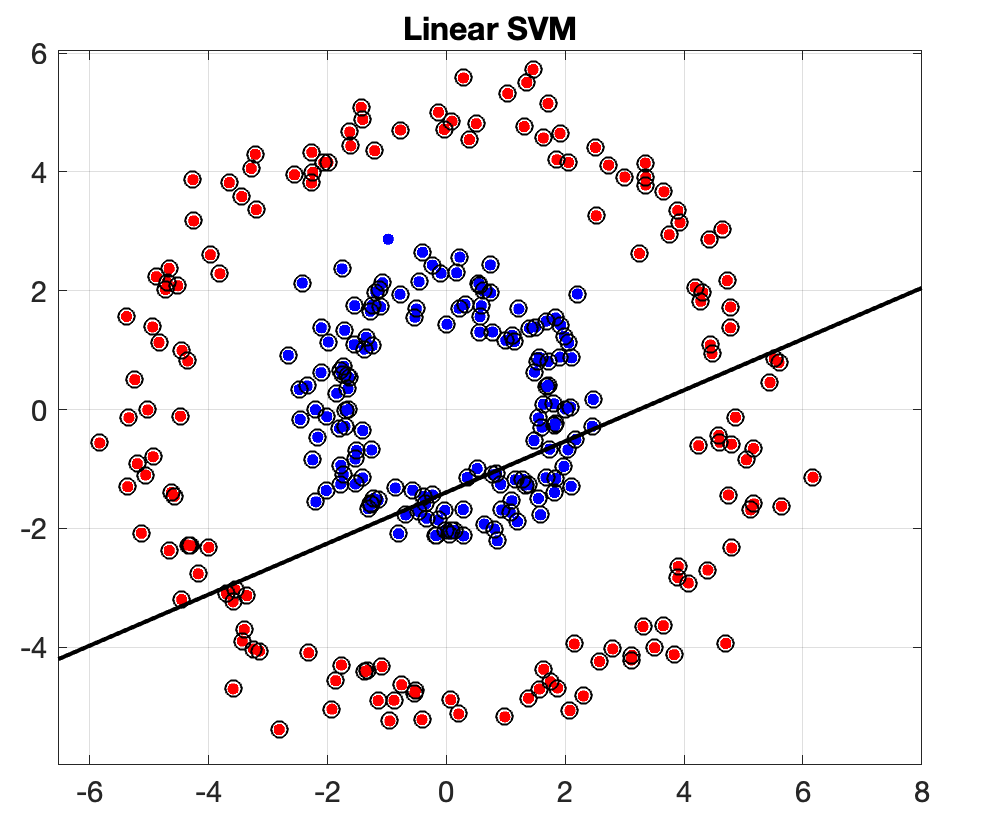}
 \qquad
\includegraphics[width=0.45\textwidth,height=45mm]{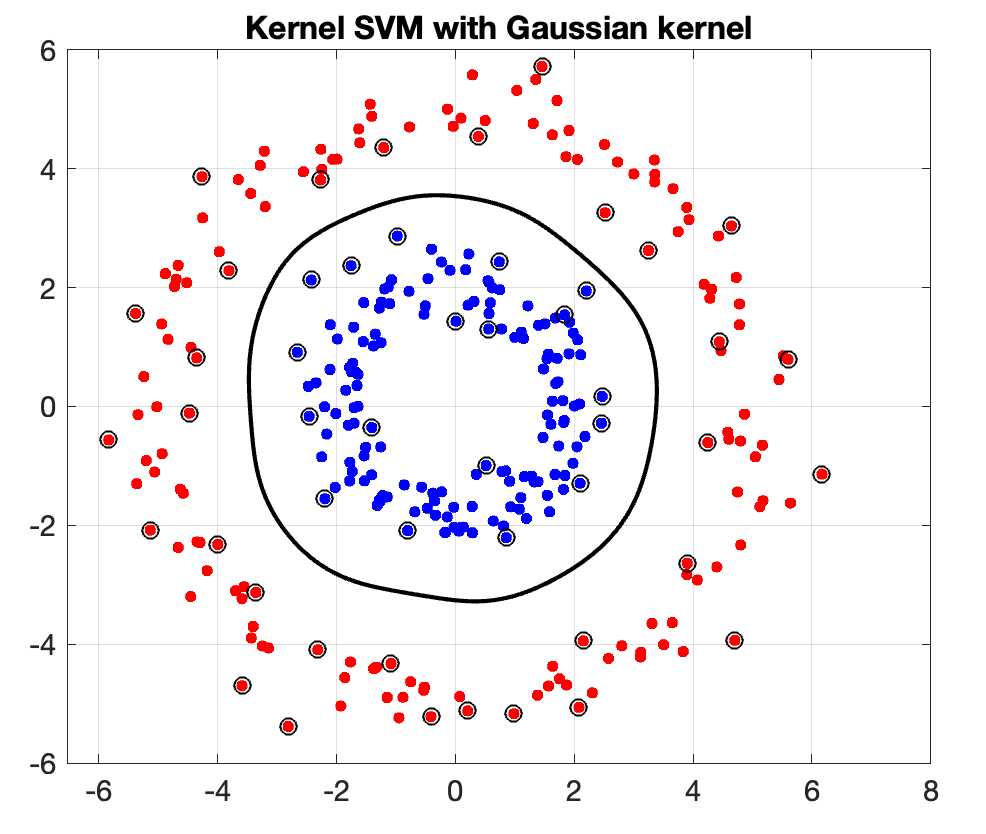}
\end{center}
    \caption{Linear SVM (left) and kernel SVM (right) with a Gaussian kernel $K(x,x') = \exp (-\|x-x'\|^2/2\eps)$ with $\eps =1$, applied to 
    a non-linearly separable dataset with two classes. Class membership is illustrated by red and blue colors. The decision boundary is plotted in black. The support vectors are indicated by open circles. While linear SVM obviously must fail to separate the two clusters, kernel SVM (with properly chosen hyperparameter) succeeds in disentangling them.}
    \label{fig:kernelsvm}
\end{figure}

\section{Beyond binary classification}

Binary classification addresses problems with only two possible outcomes. However, many practical and scientific applications naturally involve multiple classes, making multiclass classification essential.
Indeed, in many domains, categories are inherently multi-valued, such as 
medical diagnosis with multiple disease categories, object recognition (car, pedestrian, bicycle,...), or document topic classification (politics, sports, technology,...).

\subsection{Multiclass Logistic Regression (Softmax Regression)}

While binary logistic regression uses the sigmoid function to map a score to a probability, multiclass logistic regression (also known as Softmax Regression) generalizes this by using the Softmax function.

For $K$ classes, we maintain $K$ separate weight vectors $\{w_1, w_2, \dots, w_K\}$. For a given input $x$, the probability that it belongs to class $k$ is
\begin{equation}    \label{multiclass}
\P(y=k \mid x) = \frac{\exp(w_k^T x+b_k)}{\sum_{j=1}^K \exp(w_j^T x + b_j)},
\end{equation}
where the function on the right-hand side in~\eqref{multiclass} is also known as softmax function.

The denominator ensures that $\sum_{k=1}^K \P(y=k \mid x) = 1$, creating a valid probability distribution.
The optimization objective generalizes to the categorical cross-entropy loss
$$J(W) = -\sum_{i=1}^n \sum_{k=1}^K \mathbf{1}\{y_i = k\} \log(\P(y_i=k \mid x_i)).$$
Here, the indicator $\mathbf{1}\{y_i = k\}$ is the ``multiclass version'' of the $y$ and $(1-y)$ used in binary logistic regression, which equals 1 if the true class of the 
$i$-th sample is $k$, and it equals 0 otherwise.  
$W$ is a matrix where each column (or row) corresponds to the weight vector $w_k$ for class $k$.
The decision rule consists of predicting the class with the highest probability: $\hat{y} = \arg\max_k \P(y=k \mid x)$.

\subsection{Multiclass Support Vector Machines}

Unlike logistic regression, the SVM objective does not naturally produce a probability distribution across multiple classes. Instead, SVMs use meta-strategies to decompose the multiclass problem into multiple binary problems~\cite{hsu2002comparison}.

One such strategy is called ``one-vs-all,'' this is the most common approach. The strategy consists of training $K$ separate binary SVMs. For each class $k$, we train a classifier where class $k$ is the positive label ($+1$) and all other classes are the negative label ($-1$). For the prediction step we calculate the decision function $f_k(x) = w_k^T x  + b_k$ for all $K$ classifiers. The input is assigned to the class with the highest (most positive) score.
However, this approach comes with the drawback that
the individual classifiers are trained on imbalanced datasets (one class vs.\ all other classes). 

This issue can be mitigated by considering a
``one-vs.-one'' strategy. There, one trains a binary SVM for every possible pair of classes. This results in $K(K-1)/2$ classifiers. The prediction is done via a ``majority vote'': Each classifier performs a ``vote.'' If the SVM for class 1 vs.\ class 2 chooses class 1, then class 1 gets a point. The class with the most votes wins. This strategy  has the benefit that each binary problem is smaller and more balanced, and it has the obvious downside that it is computationally more intense.

\section{Rademacher complexity and generalization bounds}\label{ss:rademacher}

In this section, we develop a systematic connection between geometric measures of complexity and generalization guarantees for SVMs.
We show how {\em Rademacher complexity} provides a natural measure that leads directly to margin-based risk bounds~\cite{bartlett2002rademacher,koltchinskii2000rademacher,shalev2014understanding}. In a nutshell, Rademacher complexity is a measure of the richness of a class of (binary or real-valued) functions with respect to a probability distribution.

Let us start with a thought experiment\footnote{This thought experiment is inspired by lectures notes of Aditya Bhaskara on the {\em Theory of Machine Learning}, \url{https://utah.instructure.com/courses/1023142}.}. 
Suppose there is an unknown distribution ${\mathcal D}$ from which we draw $m$ i.i.d.\ samples 
$S = \{(x_1, y_1), \dots, (x_n, y_n)\}$. Let
$\ell$ be a given loss function and
let $\loss_S(f)=\frac{1}{n} \sum_{i=1}^n \ell(f(x_i), y_i)$ denote the associated empirical loss of a hypothesis $f \in {\mathcal F}$, where ${\mathcal F}$ denotes a (possibly infinite) hypothesis class\footnote{In statistical learning theory, a hypothesis class is the set of all prediction rules that a learning algorithm is allowed to choose from. For example, for linear classifiers,  $\mathcal{F} = \{ x \mapsto \operatorname{sign}(w^\top x + b) : w \in \mathbb{R}^d, b \in \mathbb{R} \}$.}, and $\loss_{\mathcal D}(f) = \mathbb{E}_{(x,y) \sim {\mathcal D}} [\ell(f(x), y)]$ its true (population) loss. We can compute $\loss_S(f)$, but
since ${\mathcal D}$ is usually unknown, we cannot calculate $\loss_{\mathcal D}(f)$ directly.
In order to establish a generalization bound, we aim to show that the discrepancy between these two quantities is small uniformly over all $f \in {\mathcal F}$. 

To build intuition, consider the following experiment. Suppose we are given a fixed hypothesis $f$ together with the sample $S$, and we wish to estimate the quantity $\loss_S(f) - \loss_{\mathcal D}(f)$. A natural approach is to employ a form of cross-validation: partition the sample $S$ into two disjoint subsets, $S_1$ (training) and $S_2$ (test), where for simplicity we assume that $|S_1| = |S_2| = |S|/2$. We aim to approximate $\loss_S(f) - \loss_{\mathcal D}(f)$ by
$$
\loss_S(f) - \loss_{\mathcal D}(f) \approx \loss_{S_1}(f) - \loss_{S_2}(f).
$$
We now expand this expression more explicitly. Let ${\textbf 1}_{(f, x_i)}$ be the indicator function that equals 1 if the hypothesis $f$ makes an error on the example $x_i$, and equals $0$ otherwise.
The expression above can therefore be rewritten as
\begin{equation}\label{eq:rademacher1}
\loss_S(f) - \loss_{\mathcal D}(f)
\approx
\frac{1}{|S_1|}\sum_{x_i \in S_1} {\textbf 1}_{(f, x_i)} 
- \frac{1}{|S_2|}\sum_{x_i \in S_2} {\textbf 1}_{(f, x_i)}.
\end{equation}
 
Fixing the sample $S$, each loss term ${\textbf 1}_{(f, x_i)}$ appears with a coefficient of $+1$ if $x_i \in S_1$ and with a coefficient of $-1$ if $x_i \in S_2$. Let us denote this sign by $\sigma_i$. With this notation, we may write
$$
\loss_{S_1}(f) - \loss_{S_2}(f) = \frac{2}{n} \sum_i \sigma_i 
{\textbf 1}_{(f, x_i)}.$$

Now suppose the sample $S$ is fixed, and we form the partition into $S_1$ and $S_2$ uniformly at random. This is equivalent to choosing the signs $\sigma_i$ independently and uniformly from ${\pm 1}$, in which case the sum above can be viewed as a randomized estimate of the generalization gap. 
The resulting quantity is commonly referred to as a Rademacher average. Informally, Rademacher complexity,  is the supremum of the quantity above, over $f \in {\mathcal F}$.

We will make this concept now rigorous. We distinguish between the empirical version (calculated on a specific set) and the population version (an average over all possible sets) of Rademacher complexity.

\begin{definition}\label{de:empiricalrademacher}
Let $\mathcal{F}$ be a class of real-valued functions $f : \mathcal{X} \to \mathbb{R}$, and let
$S = (x_1, x_2, \ldots, x_n)$ be a fixed sample from $\mathcal{X}$. Let $\sigma_1, \ldots, \sigma_n$ be independent Rademacher random variables, i.e.
$$
\mathbb{P}(\sigma_i = +1) = \mathbb{P}(\sigma_i = -1) = \tfrac12.
$$
The {\em empirical Rademacher complexity} of $\mathcal{F}$ with respect to $S$ is defined as
$$
\widehat{\Rade}_S(\mathcal{F})
\coloneqq 
\mathbb{E}_{\sigma} \Big[
\sup_{f \in \mathcal{F}}
\frac{1}{n}
\sum_{i=1}^n \sigma_i  f(x_i)
\Big].
$$
\end{definition}
Moving from this data-dependent estimate to its expected, distribution-level counterpart gives the population Rademacher complexity.

\begin{definition}\label{de:empiricalrademacher}
Let ${\mathcal D}$ be a distribution over $\mathcal{X}$. The (population) Rademacher complexity of $\mathcal{F}$ with respect to $n$ samples from ${\mathcal D}$ is
$$
\Rade_n(\mathcal{F})
\coloneqq
\mathbb{E}_{S \sim {\mathcal D}^n}
\left[\widehat{\Rade}_S(\mathcal{F})
\right] =
\mathbb{E}_{S \sim {\mathcal D}^n} \mathbb{E}_{\sigma}
\left[\sup_{f \in \mathcal{F}} \frac{1}{n}
\sum_{i=1}^n \sigma_i f(x_i) \right].
$$
\end{definition}
If $\mathcal{F} \subseteq \{ f : \mathcal{X} \to \{0,1\} \}$, the same definitions apply verbatim, with $f(x_i) \in \{0,1\}$.

Rademacher complexity is closely related to the {\em Gaussian width}, a concept arising in high-dimensional geometry and statistical learning theory
which we will explore in depth in Chapter~\ref{s:gordon}. For the current purposes, it suffices to know that for a closed set $T \subset \mathbb{R}^n$, the Gaussian width $w(T)$ is defined as
$$w(T) = \mathbb{E}_{g \sim \mathcal{N}(0, I_n)} \left[ \sup_{t \in T} \,\langle g, t \rangle \right].$$
Thus, for Rademacher complexity we are essentially replacing Gaussian random variables with Rademacher random variables.
The close connection between these two concepts is further manifested  by the following relationship between the Rademacher complexity and Gaussian width (see Exercise~\ref{ex:rademacher}):
$$\sqrt{\frac{2}{\pi}}\,w(\mathcal{F}) \leq  \Rade_n(\mathcal{F})\leq \sqrt{2 \ln(2n)} \,w(\mathcal{F}).$$

\medskip
Rademacher complexity is one of the central tools 
in classical statistical learning theory.
 If a hypothesis class has low Rademacher complexity (or Gaussian width), it will generalize well, as established by the following theoretical analysis. 

\begin{definition}\label{de:risk}
Given a hypothesis $f \in \mathcal{F}$, a loss function $\loss$, and a training sample $S = \{(x_1, y_1), \dots, (x_n, y_n)\}$, the empirical risk $\hat{R}_S(f)$ is defined as 
$$\hat{R}_S(f) = \frac{1}{n} \sum_{i=1}^n \loss(f(x_i), y_i).$$
\end{definition}

While the ``true risk'' (or ``expected risk'') is the error we expect to see on all future, unseen data, the empirical risk is the error we can actually calculate using the data we have (the training set).

The following theorem connects the true risk to the empirical risk  using the population Rademacher complexity.
Intuitively, if a hypothesis class cannot fit random noise well, then empirical performance reliably predicts true performance.

\begin{theorem}[Fundamental Generalization Bound]\label{th:rademachergeneralization}
Let $\mathcal{F}$ be a hypothesis class of functions mapping to $[0, 1]$. For any $\delta > 0$, with probability at least $1 - \delta$ over a sample $S$ of size $n$, every $h \in \mathcal{F}$ satisfies
\begin{equation}\label{eq:radebound}
R(f) \leq \hat{R}_S(f) + 2\Rade_n(\mathcal{F}) + \sqrt{\frac{\ln(1/\delta)}{2n}},
\end{equation}
where $R(f) = \mathbb{E}[\loss(f(x), y)]$ is the expected risk, $\hat{R}_S(f)$ is the empirical risk on the sample $S$, and  $\Rade_n(\mathcal{F})$ is the population Rademacher complexity.
\end{theorem}

\begin{proof}
The proof has three main components: (i)~concentration, (ii)~symmetrization, and (iii)~reduction to Rademacher complexity.

(i)~Concentration via McDiarmid’s Inequality:
Define a function $\Phi(S)$ that represents the maximum gap between true and empirical risk for our class
$$\Phi(S) = \sup_{f \in \mathcal{F}} (R(f) - \hat{R}_S(f))$$
We want to show that $\Phi(S)$ does not vary much if we change one element in $S$. Let $S$ and $S'$ differ only in the $i$-th element ($x_i$ vs $x'_i$).
We compute
$$|\Phi(S) - \Phi(S')| \leq \sup_{f \in \mathcal{F}} |\hat{R}_{S'}(f) - \hat{R}_S(f)| = \sup_{f \in \mathcal{F}} \frac{1}{n} |f(x'_i) - f(x_i)| \leq \frac{1}{n},$$
where we have used that the range of $f$ is $[0, 1]$. By McDiarmid’s Inequality~\cite{mcdiarmid1989method}, with probability $1 - \delta$ it holds that
\begin{equation}\label{eq:rade4}
\Phi(S) \leq \mathbb{E}_S[\Phi(S)] + \sqrt{\frac{\ln(1/\delta)}{2n}}.
\end{equation}

(ii) Symmetrization:
Now we bound the expectation 
$$\mathbb{E}_S[\Phi(S)] = \mathbb{E}_S \left[ \sup_{f \in \mathcal{F}} (R(f) - \hat{R}_S(f)) \right].$$
To that end, we introduce a ``ghost sample'' $S' = \{x'_1, \dots, x'_n\}$ drawn from the same distribution. Note that $R(f) = \mathbb{E}_{S'}[\hat{R}_{S'}(f)]$.
We have
\begin{align*}
&\mathbb{E}_S[\Phi(S)]  =  \mathbb{E}_S \left[ \sup_{f \in \mathcal{F}} (\mathbb{E}_{S'}[\hat{R}_{S'}(f)] - \hat{R}_S(f)) \right] \\
& \quad \quad \leq \mathbb{E}_{S, S'} \left[ \sup_{f \in \mathcal{F}} (\hat{R}_{S'}(f) - \hat{R}_S(f)) \right] = \mathbb{E}_{S, S'} \left[ \sup_{f \in \mathcal{F}} \frac{1}{n} \sum_{i=1}^n (f(x'_i) - f(x_i)) \right],
\end{align*}
where we have used Jensen’s inequality in the last step.

(iii) Introducing Rademacher variables:
Since $x_i$ and $x'_i$ are i.i.d., the distribution of $(f(x'_i) - f(x_i))$ is symmetric. We can multiply by Rademacher variables $\sigma_i \in \{-1, 1\}$ without changing the expectation and obtain
\begin{align}\label{eq:rade5}
\begin{split}
& \mathbb{E}_{S, S', \sigma} \left[ \sup_{f \in \mathcal{F}} \frac{1}{n} \sum_{i=1}^n \sigma_i (f(x'_i) - f(x_i)) \right]
\le  \\
& \qquad \leq \mathbb{E}_{S', \sigma} \left[ \sup_{f \in \mathcal{F}} \frac{1}{n} \sum_{i=1}^n \sigma_i f(x'_i) \right] + \mathbb{E}_{S, \sigma} \left[ \sup_{f \in \mathcal{F}} \frac{1}{n} \sum_{i=1}^n -\sigma_i f(x_i) \right],
\end{split}
\end{align}
where we have used the property of suprema ($\sup(A-B) \leq \sup A + \sup(-B)$).
Both terms are identical and equal to the population Rademacher complexity $\Rade_n(\mathcal{F})$.
Therefore, $\mathbb{E}_S[\Phi(S)] \leq 2\Rade_n(\mathcal{F})$.

Substituting the result of~\eqref{eq:rade5} back into the McDiarmid bound from~\eqref{eq:rade4} gives
$$\Phi(S) \leq 2\Rade_n(\mathcal{F}) + \sqrt{\frac{\ln(1/\delta)}{2n}}$$
Since $R(f) - \hat{R}_S(f) \leq \Phi(S)$ for all $f \in \mathcal{F}$, the bound is proved. 

\end{proof}

If we wish to use the empirical Rademacher complexity $\hat{\Rade}_S(\mathcal{F})$ (which we can actually compute from data), the bound becomes slightly looser to account for the variance of the sample, namely
\begin{equation}\label{eq:radebound2}
R(f) \leq \hat{R}_S(f) + 2\hat{\Rade}_S(\mathcal{F}) + 3\sqrt{\frac{\ln(2/\delta)}{2n}}.
\end{equation}
The proof of this bound is left to the reader (see Exercise~\ref{ex:rademacher2}).

\medskip

We are now ready to state the Rademacher Complexity  generalization bound for linear classes.

\begin{theorem}\label{th:rademacher_linear}
Denote $\mathcal{F}_B := \{x \mapsto \langle w, x \rangle : \|w\|\leq B\}$ and let $S = \{x_1, \dots, x_n\}$ be a sample such that $\|x_i\| \leq r$ for all $i$. The empirical Rademacher complexity of $\mathcal{F}_B$ satisfies
$$\hat{\Rade}_S(\mathcal{F}_B) \leq \frac{Br}{\sqrt{n}}.$$
\end{theorem}

\begin{proof}
We have
$$\hat{\Rade}_S(\mathcal{F}_B) = \mathbb{E}_{\sigma} \left[ \sup_{\|w\| \leq B} \frac{1}{n} \sum_{i=1}^n \sigma_i \langle w, x_i \rangle \right]  = \frac{1}{n} \mathbb{E}_{\sigma} \Big[ \sup_{\|w\| \leq B} \Big\langle w, \sum_{i=1}^n \sigma_i x_i \Big\rangle \Big].$$
To maximize the inner product $\langle w, \sum \sigma_i x_i \rangle$ subject to the constraint $\|w\| \leq B$, we choose $w$ to be in the same direction as the vector $\sum \sigma_i x_i$.
Specifically, the optimal $w^*$ is $B \frac{\sum \sigma_i x_i}{\|\sum \sigma_i x_i\|}$. Substituting this in gives
\begin{align}\label{eq:rade6}
\begin{split}
\hat{\Rade}_S(\mathcal{F}_B) & = \frac{1}{n} \mathbb{E}_{\sigma} \Big[ B \Big\| \sum_{i=1}^n \sigma_i x_i \Big\| \Big] = \frac{B}{n} \mathbb{E}_{\sigma} \Big[ \Big( \Big\| \sum_{i=1}^n \sigma_i x_i \Big\|^2 \Big)^{\frac{1}{2}} \Big] \\
& \leq \frac{B}{n} \sqrt{ \mathbb{E}_{\sigma} \Big[ \Big\| \sum_{i=1}^n \sigma_i x_i \Big\|^2 \Big] },
\end{split}
\end{align}
where we have used the concavity of the square root function and Jensen's Inequality.

Furthermore, 
\begin{equation}
\mathbb{E}_{\sigma} \Big\| \sum_{i=1}^n \sigma_i x_i \Big\|^2 =
\mathbb{E}_{\sigma} \Big[ \sum_{i,j} \sigma_i \sigma_j \langle x_i, x_j \rangle \Big] = \sum_{i,j} \mathbb{E}_{\sigma}[\sigma_i \sigma_j] \langle x_i, x_j \rangle.
\end{equation}

Since the $\sigma_i$ are independent and have mean zero we have that if $i \neq j$, then $\mathbb{E}[\sigma_i \sigma_j] = \mathbb{E}[\sigma_i]\mathbb{E}[\sigma_j] = 0$, and  if $i = j$, then $\mathbb{E}[\sigma_i^2] = 1$. Hence,  the double sum collapses to
$\sum_{i=1}^n \mathbb{E}[\sigma_i^2] \langle x_i, x_i \rangle = \sum_{i=1}^n \|x_i\|^2$.

Given our assumption that $\|x_i\| \leq R$, we have $\sum_{i=1}^n \|x_i\|^2 \leq nR^2$. Substituting this back into the inequality~\eqref{eq:rade6} yields
$$\hat{\Rade}_S(\mathcal{F}_B) \leq \frac{B}{n} \sqrt{nR^2} =  \frac{BR}{\sqrt{n}}.$$

\end{proof}

The following result, known as Talagrand's Contraction Lemma, 
is the key tool for passing from linear predictors to hinge, logistic, or other Lipschitz losses in SVM analysis. It
allows us to relate the complexity of the ``composed'' function class (loss $\circ$ hypothesis) back to the original hypothesis class.

\begin{theorem}[Talagrand’s Contraction Lemma~\cite{LT91}]\label{th:talagrandcontraction}
Let $\mathcal{F}$ be a class of functions mapping $\mathcal{X} \to \mathbb{R}$. Let $\phi_1, \dots, \phi_n$ be functions such that each $\phi_i: \mathbb{R} \to \mathbb{R}$ is $L$-Lipschitz. Then for any sample $S = \{x_1, \dots, x_n\}$
$$\mathbb{E}_{\sigma} \left[ \sup_{f \in \mathcal{F}} \sum_{i=1}^n \sigma_i \phi_i(f(x_i)) \right] \leq L \cdot \mathbb{E}_{\sigma} \left[ \sup_{f \in \mathcal{F}} \sum_{i=1}^n \sigma_i f(x_i) \right].$$
\end{theorem}

\begin{proof}
We assume $L=1$ without loss of generality. We proceed by induction on $n$, the number of samples.

Let $n=1$:
We want to show that
$$\mathbb{E}_{\sigma_1} [\sup_{f \in \mathcal{F}} \sigma_1 \phi_1(f(x_1))] \leq \mathbb{E}_{\sigma_1} [\sup_{f \in \mathcal{F}} \sigma_1 f(x_1)].$$
Expanding the expectation gives
$$\frac{1}{2} \Big( \sup_{f \in \mathcal{F}} \phi_1(f(x_1)) + \sup_{g \in \mathcal{F}} [-\phi_1(g(x_1))] \Big) = \frac{1}{2} \sup_{f, g \in \mathcal{F}} [\phi_1(f(x_1)) - \phi_1(g(x_1))]$$
By the Lipschitz property we have $\phi_1(f(x_1)) - \phi_1(g(x_1)) \leq |f(x_1) - g(x_1)|$.
The right-hand side of the case $n=1$ is $\frac{1}{2} \sup_{f, g \in \mathcal{F}} [f(x_1) - g(x_1)]$.
We note that $\sup_{f,g} |f-g| = \sup_{f,g} (f-g)$ in our case, since $f$ and $g$ are interchangeable in the supremum, hence the inequality holds.

Assume now the theorem holds for $n-1$. Let $U_{n-1}(f) := \sum_{i=1}^{n-1} \sigma_i \phi_i(f(x_i))$. We analyze
\begin{equation}\label{eq:radeE}
E := \mathbb{E}_{\sigma_{1:n}} \left[ \sup_{f \in \mathcal{F}} U_{n-1}(f) + \sigma_n \phi_n(f(x_n)) \right],
\end{equation}
(where we have used the shorthand $\sigma_{1:n} = (\sigma_1,\dots,\sigma_n)$) and isolate the expectation over $\sigma_n$. We get
$$E = \mathbb{E}_{\sigma_{1:n-1}} \left[ \frac{1}{2} \sup_{f \in \mathcal{F}} \left( U_{n-1}(f) + \phi_n(f(x_n)) \right) + \frac{1}{2} \sup_{g \in \mathcal{F}} \left( U_{n-1}(g) - \phi_n(g(x_n)) \right) \right].$$
For any fixed values of $\sigma_{1:n-1}$, let $f^*$ and $g^*$ be the functions that achieve the two suprema above. Then the expression inside the outer expectation is
\begin{equation}\label{eq:radeU}
\frac{1}{2} \left[ U_{n-1}(f^*) + \phi_n(f^*(x_n)) + U_{n-1}(g^*) - \phi_n(g^*(x_n)) \right].
\end{equation}
Using the Lipschitz property, $\phi_n(f^*(x_n)) - \phi_n(g^*(x_n)) \leq |f^*(x_n) - g^*(x_n)|$.
We now consider the sign of $(f^*(x_n) - g^*(x_n))$:
If $f^*(x_n) \geq g^*(x_n)$,~\eqref{eq:radeU} becomes $\frac{1}{2} [ U_{n-1}(f^*) + U_{n-1}(g^*) + f^*(x_n) - g^*(x_n) ]$.
If $f^*(x_n) < g^*(x_n)$,~\eqref{eq:radeU} becomes $\frac{1}{2} [ U_{n-1}(f^*) + U_{n-1}(g^*) - f^*(x_n) + g^*(x_n) ]$.

In both cases, notice that this matches the expansion of the supremum of the untransformed $n$-th variable
$$\frac{1}{2} \left[ \sup_{f \in \mathcal{F}} (U_{n-1}(f) + f(x_n)) + \sup_{g \in \mathcal{F}} (U_{n-1}(g) - g(x_n)) \right].$$
Substituting this back into~\eqref{eq:radeE} gives
$$E \leq \mathbb{E}_{\sigma_{1:n-1}} \mathbb{E}_{\sigma_n} \left[ \sup_{f \in \mathcal{F}} \sum_{i=1}^{n-1} \sigma_i \phi_i(f(x_i)) + \sigma_n f(x_n) \right].$$
By repeating this argument for $\sigma_{n-1}, \sigma_{n-2}, \dots, \sigma_1$, we peel away each $\phi_i$ one by one. After $n$ iterations, we obtain
$$\mathbb{E}_{\sigma} \left[ \sup_{f \in \mathcal{F}} \sum_{i=1}^n \sigma_i \phi_i(f(x_i)) \right] \leq \mathbb{E}_{\sigma} \left[ \sup_{f \in \mathcal{F}} \sum_{i=1}^n \sigma_i f(x_i) \right].$$
This completes the proof.

\end{proof}

It is easy to see that the hinge loss and the logistic loss are 1-Lipschitz.
Therefore, by virtue of Talagrand's Contraction Lemma, generalization bounds for linear predictors automatically extend to SVM and logistic regression, as manifested by the Theorem~\ref{th:marginbound} below.

We define the margin loss (or ramp loss) $\Phi_\gamma: \mathbb{R} \to [0, 1]$  by
$$\Phi_\gamma(u) = \min\Big(1, \max\big(0, 1 - \frac{u}{\gamma}\big)\Big) = \begin{cases} 1 & \text{if } u \leq 0, \\ 1 - u/\gamma & \text{if } 0 < u \leq \gamma, \\ 0 & \text{if } u > \gamma. \end{cases}$$

\begin{theorem}\label{th:marginbound}
Let $\mathcal{D}$ be a distribution over $\mathcal{X} \times \{-1, +1\}$, where $\mathcal{X}$ is an input space such that $\|x\| \leq R$ for all $x \in \mathcal{X}$. Let $\mathcal{F}_B$ denote the hypothesis class defined by
$$\mathcal{F}_B = \{ x \mapsto \langle w, x \rangle : \|w\| \leq B \}.$$
For any margin $\gamma > 0$, and for any  $\delta \in (0, 1)$, with probability at least $1 - \delta$ over the choice of a sample $S = \{(x_1, y_1), \dots, (x_n, y_n)\}$ drawn i.i.d. from $\mathcal{D}$, every $f \in \mathcal{F}_B$ satisfies
$$\P_{(x,y) \sim \mathcal{D}}(y f(x) \leq 0) \leq \frac{1}{n} \sum_{i=1}^n \Phi_\gamma(y_i f(x_i)) + \frac{2BR}{\gamma \sqrt{n}} + 3 \sqrt{\frac{\ln(2/\delta)}{2n}}.$$

\end{theorem}

\begin{proof}
We want to bound the true risk $R(f) = P_{(x,y) \sim \mathcal{D}}(yf(x) \leq 0)$. Define the indicator loss $\loss_{0/1}(u) = \mathbb{1}_{u \leq 0}$.
The margin loss function $\Phi_\gamma(u)$ is defined such that for all $u \in \mathbb{R}$
$$\mathbf{1}_{u \leq 0} \leq \Phi_\gamma(u)$$
Therefore, the true risk is dominated by the expected margin loss
$$R(f) = \mathbb{E}[\mathbb{1}_{yh(x) \leq 0}] \leq \mathbb{E}[\Phi_\gamma(yh(x))].$$
Let $\mathcal{G}$ be the family of functions $\mathcal{G} = \{ (x, y) \mapsto \Phi_\gamma(y f(x)) : f \in \mathcal{F}_B \}$. Since the range of $\Phi_\gamma$ is $[0, 1]$, we apply the bound~\eqref{eq:radebound2}. For any $\delta > 0$, with probability at least $1 - \delta$
\begin{equation}\label{eq:rade7}
\mathbb{E}[\Phi_\gamma(yf(x))] \leq \frac{1}{n} \sum_{i=1}^n \Phi_\gamma(y_i f(x_i)) + 2\hat{\mathcal{R}}_S(\mathcal{G}) + 3\sqrt{\frac{\ln(2/\delta)}{2n}}.\end{equation}

Note that $\Phi_\gamma(u)$ is $L$-Lipschitz with $L = \frac{1}{\gamma}$. By Talagrand’s Contraction Lemma,
\begin{equation}\label{eq:radem8}
\hat{\mathcal{R}}_S(\mathcal{G}) = \mathbb{E}_\sigma \left[ \sup_{f \in \mathcal{F}_B} \frac{1}{n} \sum_{i=1}^n \sigma_i \Phi_\gamma(y_i h(x_i)) \right] \leq \frac{1}{\gamma} \mathbb{E}_\sigma \left[ \sup_{f \in \mathcal{F}_B} \frac{1}{n} \sum_{i=1}^n \sigma_i y_i h(x_i) \right].
\end{equation}

Consider the term $\mathbb{E}_\sigma [ \sup \frac{1}{n} \sum \sigma_i y_i f(x_i) ]$. Since $y_i \in \{-1, 1\}$ are fixed constants in the empirical complexity and $\sigma_i$ are independent Rademacher variables, the product $\sigma_i y_i$ is also a Rademacher variable with the same distribution as $\sigma_i$. Thus,
$$\mathbb{E}_\sigma \left[ \sup_{f \in \mathcal{F}_B} \frac{1}{n} \sum_{i=1}^n \sigma_i y_i f(x_i) \right] = \mathbb{E}_\sigma \left[ \sup_{f \in \mathcal{F}_B} \frac{1}{n} \sum_{i=1}^n \sigma_i f(x_i) \right] = \hat{\mathcal{R}}_S(\mathcal{F}_B) \leq \frac{BR}{\sqrt{n}},$$
where we have used Theorem~\ref{th:rademacher_linear} in the last step.
We combine this bound with the inequality~\eqref{eq:radem8} and obtain
\begin{equation}\label{eq:rade9}
\hat{\mathcal{R}}_S(\mathcal{G}) \leq \frac{1}{\gamma} \cdot \frac{BR}{\sqrt{n}}.
\end{equation}

Substitute the  bound from~\eqref{eq:rade9} into the generalization inequality~\eqref{eq:rade7}
gives
$$\mathbb{E}[\Phi_\gamma(yh(x))] \leq \frac{1}{n} \sum_{i=1}^n \Phi_\gamma(y_i h(x_i)) + \frac{2BR}{\gamma \sqrt{n}} + 3\sqrt{\frac{\ln(2/\delta)}{2n}}.$$
We complete the proof by noting that
$R(h) \leq \mathbb{E}[\Phi_\gamma(yh(x))]$.

\end{proof}

When applying standard SVM, we assume the data is (essentially) linearly separable. Assuming the linear separability holds strictly,  then in hard-margin SVM the optimization ensures that for all training points, $y_i f(x_i) \geq 1$. If we set our theoretical margin parameter $\gamma = 1$, then the hard-margin SVM ensures $y_i f(x_i) \geq 1$, which means $\Phi_\gamma(y_i f(x_i)) = 0$ for every single training point. Therefore, the empirical risk term in Theorem~\ref{th:marginbound} becomes zero. The generalization bound then tells us that the true error is entirely bounded by the complexity term $\frac{2BR}{\sqrt{n}}$ and the concentration term.

In a soft-margin SVM, we allow some points to violate the margin using slack variables $\xi_i$. Recall that the hinge loss, $\loss_{\text{Hinge}}(u) = \max(0, 1-u)$, is an upper bound on the ramp loss $\Phi_\gamma(u)$. While the ramp loss is what we use in the theorem to get a tight bound between 0 and 1,  the soft-margin SVM actually minimizes the hinge loss because it is convex and easier to solve via quadratic programming.
Because $\loss_{\text{Hinge}} \geq \Phi_\gamma$,  minimizing the hinge loss is a principled approach to trying to minimize the empirical term in the margin generalization bound.

\medskip
The generalization bounds are also applicable to kernel-SVM. It is not difficult to establish the following result (see also Exercise~\ref{ex:kernelSVMbound}).

\begin{theorem}\label{th:marginkernelSVM}
Let $K: \mathcal{X} \times \mathcal{X} \to \mathbb{R}$ be a symmetric, positive-definite kernel, and let $\mathbb{H}_K$ be its associated Reproducing Kernel Hilbert Space with norm $\|\cdot\|_K$. Assume there exists a constant $R > 0$ such that $K(x, x) \leq R^2$ for all $x \in \mathcal{X}$.
Let 
$$\mathcal{F}_B = \{ f \in \mathbb{H}_K : \|f\|_K \leq B \}.$$
Then with probability at least $1-\delta$
$$R(f) \leq \hat{R}_{S, \gamma}(f) + \frac{2B \sqrt{\tr(K)/n}}{\gamma \sqrt{n}} + 3\sqrt{\frac{\ln(2/\delta)}{2n}}.$$
\end{theorem}

\medskip
In this RKHS setting, the empirical Rademacher complexity is strictly bounded by the trace of the kernel matrix
$$\hat{\mathcal{R}}_S(\mathcal{F}_B) \leq \frac{B}{n} \sqrt{\sum_{i=1}^n K(x_i, x_i)} \leq \frac{BR}{\sqrt{n}}.$$

Thus, Theorem~\ref{th:marginkernelSVM} shows that even if $\mathbb{H}_K$ is infinite-dimensional (like with a Gaussian kernel), the complexity of the learning task is governed only by the norm of the function (the parameter $B$) and the kernel's behavior on the data (the parameter $R$), not the dimensionality of the space.

\section{Alternative classification methods}

The choice of classification method depends strongly on factors like the inherent structure of the data (such as its dimensionality, sparsity, and linearity), the practical setting (for instance, a high-frequency trading platform might prioritize speed, while in medical diagnostics accuracy or need for interpretability may have higher priority), the available computational resources, as well as the amount of labeled data available and the signal-to-noise ratio.

Thus, the  optimal classifier is not necessarily the most complex one, but rather the model that best balances accuracy, interpretability, robustness, and feasibility within the constraints of the intended application. With this in mind, we briefly mention a variety of alternative classification methods beyond linear models and margin-based classifiers which are widely used in practice.

$k$-Nearest Neighbor (k-NN) methods~\cite{cover1967nearest} classify data points based on local proximity in feature space and require minimal modeling assumptions, although they scale poorly with dimension. 

Random Forests and related tree-based ensembles~\cite{ho1998random,breiman2001random} construct classifiers by aggregating many randomized decision trees, offering strong empirical performance and robustness at the cost of reduced interpretability. 

Boosting methods, such as AdaBoost and gradient boosting~\cite{freund1997decision,friedman2001greedy,chen2016xgboost}, combine weak learners into a powerful classifier through iterative reweighting, with strong theoretical guarantees under suitable conditions.

Neural networks are currently the state-of-the-art approach for multiclass classification, particularly when the relationship between the input features and the classes is non-linear or highly complex (like in images, audio, or natural language) and when a large number of training data is available. 

In fact, the multiclass logistic regression model can be interpreted as a ``single-layer'' neural network. Deep Learning expands this by adding hidden layers between the input and the final softmax output. We will explore neural networks in more detail in Chapter~\ref{c:deeplearning}.

\section*{Exercises}
\addcontentsline{toc}{section}{Exercises}

\begin{myexercise}\hfill

\noindent
(a) Write down the KKT conditions for the soft-margin SVM.\\
(b) Interpret each KKT condition geometrically. \\
(c) Explain how complementary slackness characterizes support vectors.
\end{myexercise}

\begin{myexercise}
Generate a binary classification dataset in $\mathbb{R}^2$ via: 
\begin{itemize}
\item[] \quad $\bullet$ \, Class $+1$: 100 samples from $\mathcal N((2,2), I)$,
\item[] \quad $\bullet$ \, Class $-1$: 100 samples from $\mathcal N((-2,-2), I)$.
\end{itemize}
(a) Fit logistic regression with no regularization;\\
(b) Fit logistic regression with $\ell_2$-regularization $(\lambda = 0.01, 0.1, 1)$.
\\
Compare norm of the weight vector, classification accuracy, and margin width. Visualize decision boundaries.
    
\end{myexercise}

\begin{myexercise}\label{ex:rademacher}
Show that the following relationship between the Rademacher complexity and Gaussian width holds:
$$\sqrt{\frac{2}{\pi}}\,w(\mathcal{F}) \leq  \Rade_n(\mathcal{F})\leq \sqrt{2 \ln(2n)} \,w(\mathcal{F}).$$
\end{myexercise}

\begin{myexercise}\label{ex:rademacher2}
Prove the following data-dependent generalization bound: \\
For a hypothesis class $\mathcal{F}$ mapping to $[0, 1]$, with probability at least $1 - \delta$ over a sample $S$ of size $n$, every $f \in \mathcal{F}$ satisfies
$$R(f) \leq \hat{R}_S(f) + 2\hat{\Rade}_S(\mathcal{F}) + 3\sqrt{\frac{\ln(2/\delta)}{2n}}.$$
\end{myexercise}

\begin{myexercise}\label{ex:kernelSVMbound}
Prove Theorem~\ref{th:marginkernelSVM}.
\end{myexercise}

\begin{myexercise}
Consider a soft-margin SVM where each training example is a triplet $(x_i,y_i,p_i)$ with
$0 < p \le 1$ denoting the importance of example $i$. 
\begin{itemize}
\item[(a)] Modify the primal optimization problem so that the penalty for violating the margin constraint on example $i$ is scaled by $p_i$.
\item[(b)] Derive the corresponding dual optimization problem and describe how the importance weights affect the dual variables.
\item[(c)] Briefly interpret the effect of $p_i$ on the solution. 
\end{itemize}

\end{myexercise}

\begin{myexercise}
Suppose we have a binary classification problem in one dimension. The two classes, $\mathcal{C}_0$ and $\mathcal{C}_1$, are distributed according to Gaussian probability density functions: 
\begin{itemize}
\item[(i)] 
Class 0 (Negative): $p(x | \mathcal{C}_0) \sim \mathcal{N}(\mu_0, \sigma^2)$,
\item[(ii)] Class 1 (Positive): $p(x | \mathcal{C}_1) \sim \mathcal{N}(\mu_1, \sigma^2)$.
\end{itemize}
Assume $\mu_1 > \mu_0$ and that the classes have equal prior probabilities, $\P(\mathcal{C}_0) = \P(\mathcal{C}_1) = 0.5$. We define a classifier with a decision threshold (cut-off) $\tau$, such that we predict class $\mathcal{C}_1$ if $x > \tau$, and class $\mathcal{C}_0$ otherwise.

Derive analytical expressions for the following metrics as a function of the threshold $\tau$, the means $\mu_0, \mu_1$, and the variance $\sigma^2$. Use the standard normal cumulative distribution function $\Phi(z)$ or the complementary error function $\text{erfc}(z)$ in your answers.
\begin{enumerate}
\item  True Positive Rate (Sensitivity): $\text{TPR}(\tau)$.
\item
True Negative Rate (Specificity): $\text{TNR}(\tau)$.
\item
False Positive Rate: $\text{FPR}(\tau)$.
\item
The Confusion Matrix: Express the expected values of the matrix entries for a dataset of size $N$.
\item
ROC Curve Equation: Derive the functional relationship between $\text{TPR}$ and $\text{FPR}$ by eliminating $\tau$.
\item
Area Under the Curve: Show that the AUC for this 1-D Gaussian case is given by $\Phi\big(\frac{\mu_1 - \mu_0}{\sqrt{2\sigma^2}}\big)$.
\end{enumerate}
\end{myexercise}

\begin{myexercise}
The MNIST dataset consists of grayscale images of handwritten digits $0,1,\dots,9$, each represented as a vector in $\mathbb{R}^{784}$. 
In this problem, we consider the binary classification task of distinguishing between the digits 3 and 8.
Let
$$
{(x_i,y_i)}_{i=1}^n \subset \mathbb{R}^{784} \times \{-1,+1\}
$$
denote the dataset, where $y_i = +1$ corresponds to the digit 8 and $y_i = -1$ corresponds to the digit 3.
Extract all images of digits 3 and 8 from the MNIST dataset, and normalize each image so that its pixel values lie in $[0,1]$. Split the data into a training set $80\%$ and a test set $20\%$. \\
(a) Compare logistic regression and soft-margin SVM in terms of classification accuracy, convergence behavior, and robustness to misclassified samples. \\
(b) Visualize several test images that are misclassified by each method.
\end{myexercise}

\begin{myexercise}(The Breast Cancer Wisconsin Diagnostic Dataset).
Download the {\em Breast Cancer Wisconsin (Diagnostic) dataset} (e.g.\ from the UCI Machine Learning Repository). The goal in this exercise is to predict whether a tumor is malignant or benign based on 30 features (radius, texture, smoothness, etc.). \\
(a) Train  SVM on the raw data. Then, standardize the features (subtract mean, divide by standard deviation) and retrain. Why does the SVM performance change so drastically with scaling, while Logistic Regression is relatively more robust? \\
(b) Extract the weights $w$ from the Logistic Regression model. Which physical feature (e.g., ``area'' vs. ``concavity'') is the strongest predictor of malignancy? 
\end{myexercise}

\begin{myexercise}
The Spambase dataset (UCI Machine Learning Repository) consists of email messages represented as feature vectors of 57 attributes. Each email is labeled as either {\em spam} or {\em non-spam}. Let
$$
{(x_i, y_i)}_{i=1}^n \subset \mathbb{R}^{57} \times \{-1,+1\}
$$
denote the dataset, where $y_i = +1$ corresponds to spam and $y_i = -1$ corresponds to non-spam. \\
(a) Train logistic regression models for $\ell_1$-regularization, $\ell_2$-regularization, and 
no regularization. \\
(b) Train a soft-margin SVM. \\
Compare logistic regression and soft-margin SVM in terms of classification accuracy, sparsity of learned parameters, and robustness to regularization parameter choice.
\end{myexercise}

\begin{myexercise}
The goal of this exercise is to compare the performance of Support Vector Machines  and Feed-Forward Neural Networks as a function of training set size. This project consists of four parts.

\smallskip
\noindent
Load the MNIST Dataset: Filter the dataset to include only the two digits 4 and 9.
This pair is chosen because these two digits are visually (somewhat) similar and provide a meaningful challenge.
Flatten the $28 \times 28$ images into 784-dimensional vectors. Scale the pixel values to the range $[0, 1]$. Reserve a fixed test set of 1,000 images for evaluation.

You will train two models:
\begin{itemize}
\item Model A (SVM): Use a Gaussian kernel with $C=10$ and $\gamma=0.01$.
\item Model B (Neural Network): A simple Multi-Layer Perceptron (MLP) with one hidden layer of 100 neurons, using ReLU activation and a Sigmoid output layer. Use the Adam optimizer with a learning rate of $0.001$.
\end{itemize}

Train both models on subsets of increasing size $N$ of the training data: for each subset size $N \in \{20, 50, 100, 500, 1000, 5000\}$, record the test accuracy.
(To ensure statistical stability, you should repeat each training session five times with different random subsets and report the average accuracy.)

\begin{itemize}
\item[(a)] At what value of $N$ (roughly) does the neural network begin to outperform the SVM? Explain why the SVM performs better at very low values of $N$.
\item[(b)] 
Overfitting: Compare the training accuracy vs.\ test accuracy for the neural network at $N=50$. What does this gap tell you about the model's complexity relative to the data size?
\item[(c)] 
Support Vectors: For the $N=500$ SVM model, calculate the number of support vectors. How does this number compare to the total number of parameters in your neural network?
\end{itemize}

\end{myexercise}

\begin{myexercise} This exercise deals with a highly imbalanced dataset, a problem we often encounter in practice.

From the MNIST dataset select two classes: Pick digit ``0'' (Majority) and digit ``8'' (Minority). Create an imbalanced dataset by taking 5,000 samples of the digit ``0'', and taking only 50 samples of the digit ``8''.
Combine them into a training set where the minority class represents only 1\% of the data.
But use a balanced test set (500 of each) to see how the model generalizes.
Train a baseline by training a standard Logistic Regression or SVM on this imbalanced set. Then calculate accuracy: You will likely achieve ~99\% accuracy (why?)
Calculate the Precision, Recall, and F1-Score. If the model simply predicts ``0'' for every single image, what would the accuracy be? Is this model actually useful?

We will explore different intervention strategies to address this imbalance problem.

\vspace*{-2mm}
\begin{enumerate}
\item Class weighting:
Modify the SVM or Logistic Regression loss function to penalize mistakes on the minority class more heavily.

\item Precision-recall thresholding:
Instead of using the default 0.5 probability threshold, move the decision boundary.
Plot a Precision-Recall Curve. Find the threshold that yields a Recall of at least 0.80 and report the resulting Precision.
The ROC Curve and the Precision-Recall Curve are both used to evaluate classifiers. In this imbalanced experiment, which curve was more ``honest'' about the model's poor performance on the minority class? Why?

\item Data augmentation: Come up with your own strategy to create additional (synthetic) versions of the digit ``8''. But you can only use the current 50 digits of ``8'' to do this, you cannot simply add other digits ``8'' from the MNIST dataset or some other existing dataset.

\item Subsampling: Create a balanced dataset by downsampling the majority class to 50 samples. 
\end{enumerate}

Compare the results of the three different strategies to the baseline in terms of accuracy, precision for digit ``8'', precision for digit ``0'', and F1-score. Comment on the advantages and disadvantages of each strategy.
Which intervention strategy seems most promising?
    
\end{myexercise}

\chapter{A Mathematical Introduction  to Deep Learning}
\label{c:deeplearning}

\newcommand{\bfa}{{\mathbf a}}
\newcommand{\bfb}{{\mathbf b}}
\newcommand{\bfh}{{\mathbf h}}

\newcommand{\bfu}{{\mathbf u}}
\newcommand{\bfv}{{\mathbf v}}
\newcommand{\bfw}{{\mathbf w}}
\newcommand{\bfW}{{\mathbf W}}
\newcommand{\bfx}{{\mathbf x}}
\newcommand{\bfy}{{\mathbf y}}
\newcommand{\bfz}{{\mathbf z}}

Deep learning has emerged as one of the most intriguing developments in modern data science, offering a powerful framework for learning complex patterns directly from data. Unlike traditional machine learning methods that rely heavily on feature engineering, deep neural networks automatically extract hierarchical representations that capture both low-level and high-level structures in data. This capability makes deep learning particularly effective for classification tasks, where the goal is to assign inputs---such as images, sounds, or text---to discrete categories. By composing multiple layers of nonlinear transformations, deep learning models can approximate highly intricate decision boundaries, often achieving impressive performance across a wide range of domains.

The 2024 Nobel Prize in Physics and the 2024 Nobel Prize in Chemistry marked a significant moment of recognition for the field of deep learning and its far-reaching scientific impact. Together, these awards signal how deep-learning architectures have moved from niche algorithmic innovation to being recognized as powerful scientific tools.

Despite this phenomenal empirical success and scientific recognition, the field's underlying mathematical theory is still nascent; many fundamental questions regarding optimization, generalization bounds, and the precise reasons for the efficacy of overparameterized models remain open problems at the intersection of mathematics, statistics, and computer science.

At its core, deep learning leverages artificial neural networks with many layers---hence the term ``deep''---to model complex relationships in data. This chapter provides a mathematical perspective on deep learning, focusing on the foundational concepts and structures that underpin the field. 
Deep learning has developed its own jargon, often coming up with new names for already well established procedures in signal and image processing, optimization, or statistics. We follow in part the very accessible article by Higham and Higham~\cite{higham2019deep} when bridging deep learning and applied mathematics. We also recommend~\cite{petersen2024mathematical} which covers several topics that we do not discuss in this chapter,  such as the generalization properties of deep neural networks.

\section{Neural networks}

Neural networks are a cornerstone of modern artificial intelligence, capable of learning patterns from data. 
Neural networks are used in a variety of applications, including image recognition, natural language processing, and even playing complex games like Go and Chess. 
Here we focus on using neural networks for classification.

\begin{figure}
\begin{center}
    \includegraphics[width=.35\textwidth]{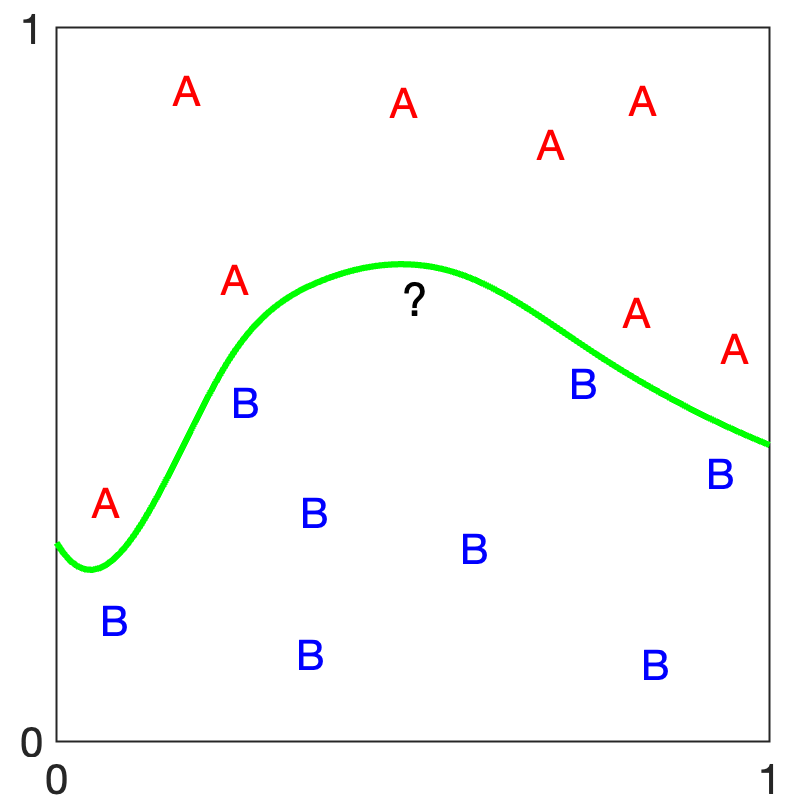}
    \caption{Labeled and unlabeled data points in $\R^2$ and a possible decision boundary.}
    \label{fig:classification}
\end{center}
\end{figure}

Suppose we have a training dataset of $n$ points in $\R^d$ consisting 
of two classes $A$ and $B$, drawn from some underlying distribution of data with the 
same labels. Based on this training set, we are looking to design a function 
$F: \R^d \rightarrow \{\textrm{A,B}\}$ that will generate labels for new 
data, for example, the black question mark in Figure~\ref{fig:classification}
If the green line represents the decision boundary of $F$, then the question mark will be
classified as a $B$. Artificial neural networks (or neural networks for short) are a way of generating this function $F$.

\subsection{Perceptrons}

A neural network is  composed of layers of {\em neurons}, where each neuron (or node) is a computational unit that processes input and produces an output.
The perceptron, introduced by Frank Rosenblatt in 1958~\cite{Rosenblatt}, is one of the simplest models of a neuron. Mathematically, a perceptron can be described as:
$$
y = \sigma \left(\sum_{i=1}^n w_i x_i + b\right),
$$
where  $x_i$ are the input features, $w_i$ are the weights associated with the inputs, $b$ is the so-called bias term, $\sigma$ is an activation function (typically a Heaviside step function, but more generally it can be a nonlinear function like the sigmoid).

A multilayer perceptron (MLP) extends the perceptron by stacking multiple layers of nodes. An MLP consists of an input layer, one or more hidden layers, and an output layer. Each layer $ \ell $ in an MLP is described by the following equation:

$$
a^{(\ell)} = \sigma\left(W^{(\ell)} a ^{(\ell-1)} + b^{(\ell)}\right),
$$
where $ a^{(\ell-1)}$ is the output of the previous layer (or the input data for the first layer),
$W^{(\ell)}$ is the weight matrix for layer $ \ell $,
$b^{(\ell)}$ is the bias vector for layer $\ell$,
and $ \sigma$ is an element-wise activation function. 
An MLP is a specific type of neural network, more specifically a subclass of feedforward neural networks. More generally, {\em neural network} is an umbrella term that includes various architectures beyond just feedforward networks.

To introduce some further standard terminology, the first layer of a neural network is called the input layer, which usually just corresponds to the data. A hidden layer in a neural network is any layer of nodes that is situated between the first layer and the last layer. The term ``hidden'' refers to the fact that these layers are not directly exposed to the input data or the final output; they are ``hidden'' in the sense that they exist only within the network's architecture.
The last layer is the output layer, typically associated with a loss function that measures the discrepancy between the predicted output and the true output.
 The output layer shapes the data into a shape more appropriate for the downstream task. For example the output layer may transform the data from a hidden shape of, say, 1024, to a shape of 2 for a binary prediction.

\subsection{General setup}

We assume the following general setup: The network has $L$ layers, with layers $1$ and $L$ being the input and output layers, respectively.
We assume that the number of nodes in the $\ell$-th layer is $n_\ell$, for $\ell = 1, 2, 3, \ldots, L$. Thus the input dimension is $n_1$ and the output dimension is $n_L$. We denote the matrix of weights at layer $\ell$ by $W^{(\ell)} \in \R^{n_\ell \times n_{\ell-1}}$, where
$w^{(\ell)}_{jk}$ is the weight that node $j$ at layer $\ell$ applies to the output from node $k$ at layer $\ell-1$. Similarly, $b^{(\ell)} \in \R^{n_l}$ is the vector of biases for layer $\ell$. Hence, the bias $b^{(\ell)}_j$ is associated with node $j$ at layer $\ell$.

\begin{figure}
\begin{center}
 \includegraphics[width=0.8\textwidth]{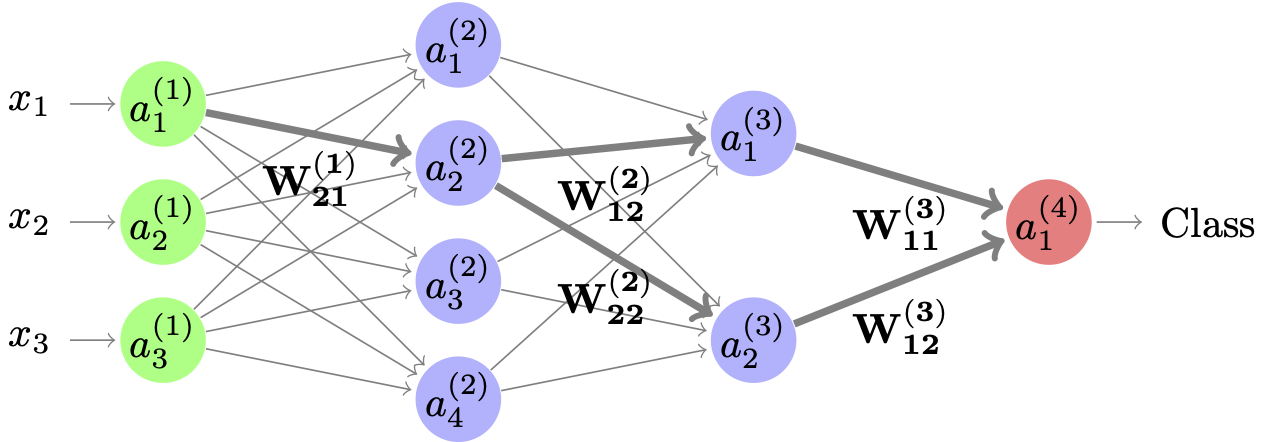}
\caption{A neural network with two hidden layers.}
\label{fig:NN}
\end{center}
\end{figure}

A neural network with four layers is illustrated in Figure~\ref{fig:NN} (for simplicity the bias terms are omitted in that figure).
$W_{kj}^{(\ell)}$ represents the connection between unit $j$ in layer $\ell$ and
unit $k$ in layer $\ell+1$, so the rows of $W^{(\ell)}$ correspond to
the weights going into node $k$. In this spirit, we define $W_i^{
(\ell)}$ to be the $i^{\textrm{th}}$ row vector of $W^{(\ell)}$. In the
example shown in Figure~\ref{fig:NN}, $W^{(1)}\in\R^{4\times 3}$,  $W^{(2)}\in\R^{2\times 4}$, and $W^{(3)}\in\R^{1\times 2}$.
Also, $a_k^{(\ell)}$ denotes the output, or activation, of unit $k$ in layer $\ell$,
while $z_k^{(\ell)}=\sum_{j}W_{kj}^{(\ell)}x_J$ is the sum into node $k$.
Hence, in the above example:
\begin{align*}
  \ell = 1 ~&~ a_k^{(1)}  := x_k\\
  \ell = 2 ~&~ a_k^{(2)} := \sigma_1 \big(W_k^{(1)}x \big) = \sigma_1 \big(W_
  {k1}^{(1)}x_1 + W_{k2}^{(1)}x_2  + W_{k3}^{(1)} x_3\big)\\
  \ell = 3 ~&~ a_k^{(3)} := \sigma_2 \big(W_k^{(2)} a^{(2)}\big) = \sigma_2 \big(W_
  {k1}^{(2)}x_1 + W_{k2}^{(2)}x_2  + W_{k3}^{(1)} x_3 + W_
  {k4}^{(1)} x_3\big)\\
  \ell = 4 ~&~ a_k^{(4)} := \sigma_3 \big(W_k^{(3)} a^{(3)} \big) = \sigma_3 \big(
  W_{k1}^{(3)} x_1 + W_{k2}^{(3)} x_2 \big)
\end{align*}

We can now represent the action of a  deep neural network with $L$ layers in the following form:
Given an input $x \in \R^{d_1}$, let $a^{(\ell)}_j$ 
denote the output, or activation, from neuron $j$ at layer $\ell$. We compute:
\begin{equation}
\label{neuralnet}
\begin{split}
    a^{(1)} & = x \in \R^{d_1}, \\
    a^{(\ell)} & = \sigma_k \Big( W^{(\ell)} a^{(\ell-1)} + b^{(\ell)}\Big) \in \R^{d_\ell}, \qquad \text{for $\ell = 2,\dots, L$,}
\end{split}
\end{equation}
where, the activation functions $\sigma_\ell$ and the number of nodes $d_\ell$ may differ from layer to layer. 

Suppose now that we are given a set of $N$ data points $x_1,\dots, x_N \in \R^{d_1}$ (our {\em training set}), for which we have associated target outputs $y_1,\dots, y_N$. 
The $x_i$ might be images, each showing either a car or a bicycle, and the corresponding $y_i$ might be binary labels indicating class membership, i.e., if the depicted object in image $x_i$ is in fact a car or a bicycle.
The task of training a neural net is to minimize
a user-specified cost function or loss function ${\mathcal C}$ which takes as input $\{x_i\}_{i=1}^N,\{y_i\}_{i=1}^N$ and which will depend on all the weights $\{W^{(\ell)}\}_{\ell=1}^L$ and biases $\{b^{(\ell)}\}_{\ell=1}^L\big)$. One example of such a cost function is
$${\mathcal C}\big(\{x_i\}_{i=1}^N,\{y_i\}_{i=1}^N \big) = \frac{1}{N}  \sum_{i=1}^N \| y(x_i) - a^{(L)} (x_i)  \|_2^2. $$
We will discuss other choices of loss functions  and how such a network is typically trained (i.e, how to find the weights and biases that minimize the loss function) in Section~\ref{nn:training}.

\subsection{Activation functions}

The activation function $ \sigma $ introduces non-linearity into the model, with the intention that it should enable the network  (assuming sufficiently many nodes and layers)  to learn complex patterns. We briefly discuss some common activation functions. {\em Sigmoidal} type functions satisfy the properties $\lim_{x \to \infty } \sigma(x) = 1$   and  $\lim_{x \to -\infty } \sigma(x) = 0$. The {\em sigmoid} or {\em logistic} function
  $$
  \sigma(x) = \frac{1}{1 + e^{-x}},
  $$
an example of a sigmoid function, maps the input $ x $ to a value between 0 and 1, making it useful for binary classification problems, as we already have seen in Chapter~\ref{c:classification}.
The {\em hyperbolic tangent} 
  $$
  \tanh(x) = \frac{e^x - e^{-x}}{e^x + e^{-x}}
  $$
maps the input to a value between -1 and 1, and it is often used in hidden layers of a neural network.
The {\em Rectified Linear Unit} (ReLU)
  $$
  \text{ReLU}(x) = \max(0, x)
  $$
 is the most popular activation function in deep learning. It is computationally efficient and helps mitigate the vanishing gradient problem~\cite{hochreiter1991untersuchungen,bengio1994learning}. However, it is not differentiable at the origin.
The {\em softmax} function:
  \begin{equation}
   \label{softmax}   
  \text{Softmax}(x_i) = \frac{e^{x_i}}{\sum_{j=1}^n e^{x_j}}
    \end{equation}
  is typically used in the output layer of a network for multi-class classification problems. It converts the raw output scores into probabilities.

\begin{figure}
\begin{center}
    \includegraphics[width=.3\textwidth]{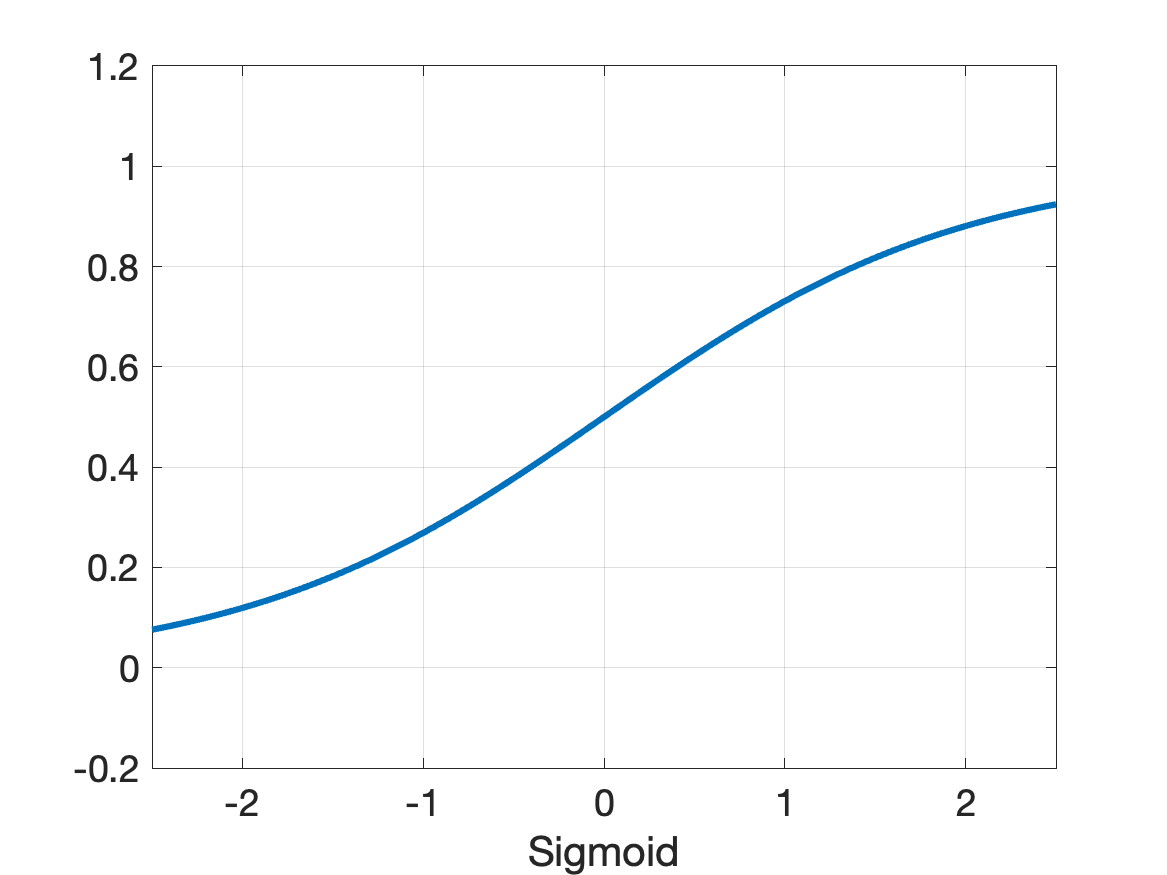}
    \includegraphics[width=.3\textwidth]{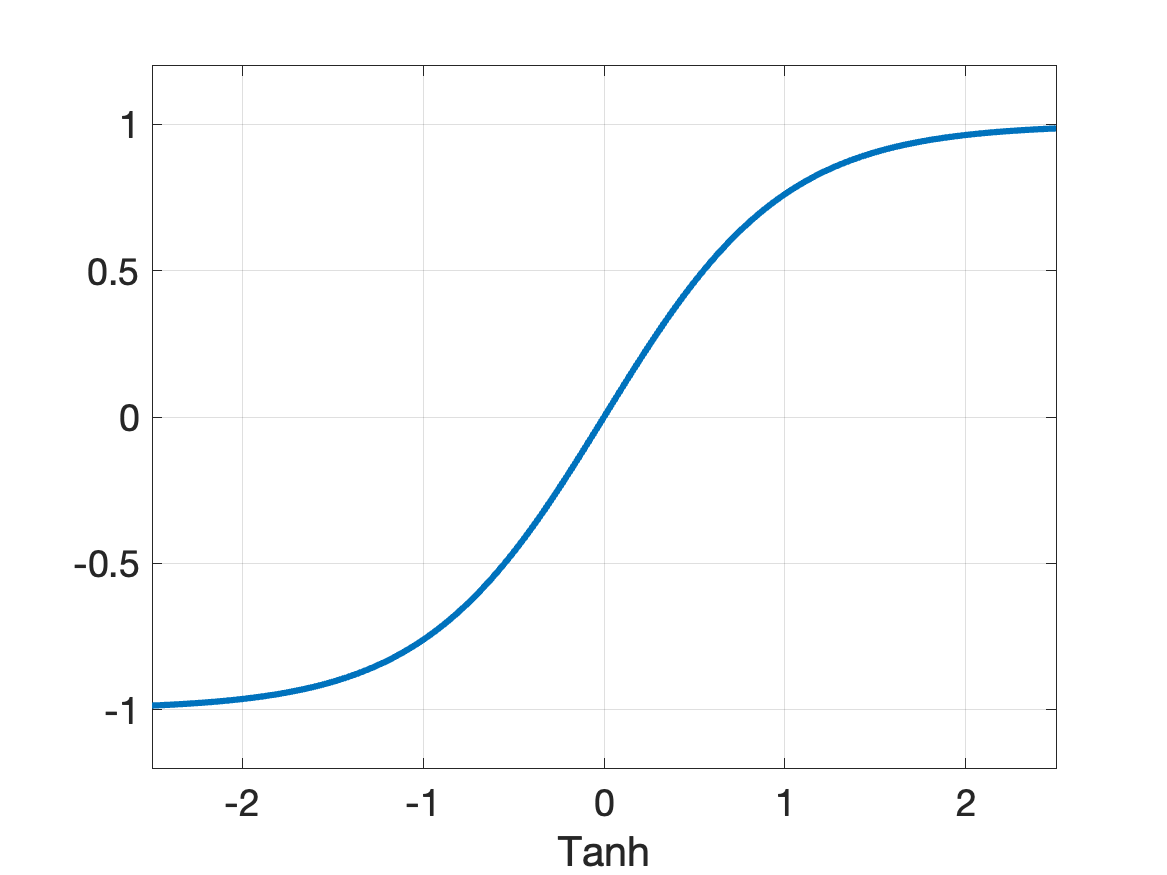}
    \includegraphics[width=.3\textwidth]{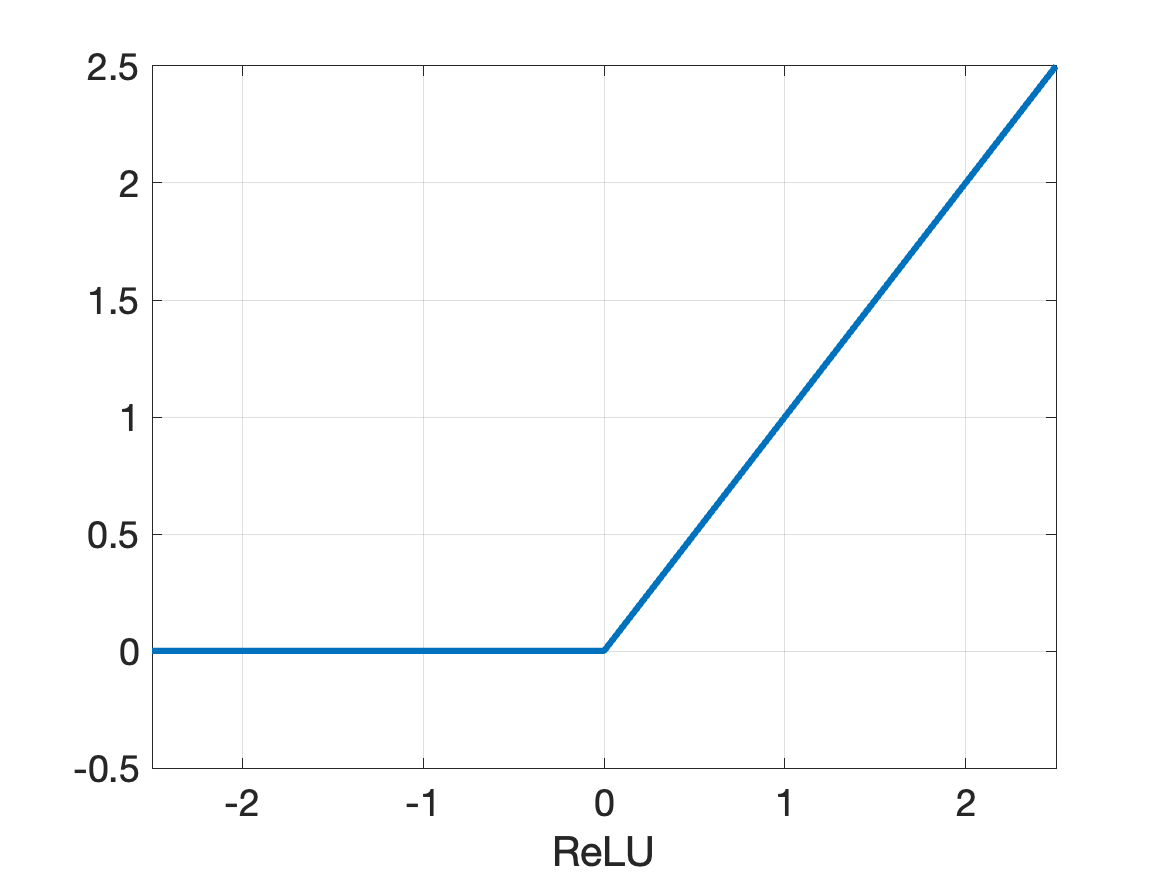}
    \caption{Some activation functions commonly used in deep learning: the logistic (sigmoid) function, the hyperbolic tangent (tanh), and the  Rectified Linear Unit (ReLU).}
    \label{fig:activ}
\end{center}
\end{figure}

Which activation function shall we use in which layer? 
Except for the last layer, where the activation function is in part determined by the task at hand (e.g.\ a classification task will suggest something like a softmax activation function while a ReLU would not help in that case), choosing an activation function is done mainly via trial and error.

While trial and error is often good enough for most application, one might ask if, instead of such an ad hoc approach, can we take a data driven approach and {\em learn} the activation functions like we do for the weights and biases?
One approach to do this in a principled manner is given by Kolmogorov-Arnold Networks which have learnable activation functions. Indeed, Kolmogorov-Arnold Networks have no linear weights at all, since every weight parameter is replaced by a univariate function parametrized as a spline.
We refer to interested reader to~\cite{liu2024kan} for details.

\section{The Universal Approximation Theorem}

Although many facets of deep learning still lack a thorough mathematical explanation, the approximation properties of (shallow) neural networks are relatively well understood.
The Universal Approximation Theorem~\cite{hornik1990universal,hornik1991approximation,cybenko1989approximation,pinkus1999approximation,barron2002universal}
is a fundamental result in neural network theory. It essentially states that a feedforward neural network with a single hidden layer containing a finite number of nodes can approximate any continuous function on compact subsets of $\R^n$, given appropriate parameters (weights and biases).

\subsection{A ``shallow'' theorem}

One of the foundational versions of the Universal Approximation Theorem (UAT) for shallow neural networks is due to George Cybenko~\cite{cybenko1989approximation}. It applies to single-hidden-layer feedforward networks with sigmoidal activation functions. We need some notation before we can present the rigorous statement.

Let \( I_n = [0,1]^n \) be the unit hypercube in \( \mathbb{R}^n \), and let \( C(I_n) \) be the space of continuous real-valued functions on \( I_n \), equipped with the supremum norm \( \|f\| = \sup_{x \in I_n} |f(x)| \).
We recall that a function \( \sigma: \mathbb{R} \to \mathbb{R} \) is called {\em sigmoidal} if \( \lim_{t \to +\infty} \sigma(t) = 1 \) and \( \lim_{t \to -\infty} \sigma(t) = 0 \).

\begin{theorem}[Cybenko, 1989]: \label{th:UAT} Let \( \sigma \) be any continuous sigmoidal function. Then, the set of all finite sums of the form
\[
G(x) = \sum_{j=1}^N \alpha_j \sigma(\langle y_j, x\rangle + \theta_j),
\]
where \( N \in \mathbb{N} \), \( \alpha_j, \theta_j \in \mathbb{R} \), and \( y_j \in \mathbb{R}^n \), is dense in \( C(I_n) \). In other words, for any \( f \in C(I_n) \) and any \( \varepsilon > 0 \), there exists such a \( G \) satisfying
\[
|G(x) - f(x)| < \varepsilon \quad \forall x \in I_n.
\]
\end{theorem}

This theorem implies that a single-hidden-layer neural network with a continuous sigmoidal activation function (e.g., the logistic sigmoid \( \sigma(t) = \frac{1}{1 + e^{-t}} \)) can approximate any continuous function on a compact set to arbitrary accuracy, provided sufficiently many hidden neurons are used. Note that the result extends naturally to any compact subset of \( \mathbb{R}^n \) by rescaling and shifting, and to multi-output functions component-wise.

The proof relies on concepts from functional analysis, including the Hahn-Banach theorem and the Riesz representation theorem. It first establishes a more general result for ``discriminatory'' functions defined below, and then shows that continuous sigmoidal functions are discriminatory.

\begin{definition}
Let \( M(I_n) \) denote the space of finite signed regular Borel measures on \( I_n \).
A function \( \sigma: \mathbb{R} \to \mathbb{R} \) is called {\em discriminatory} if, for any \( \mu \in M(I_n) \),
  \[
  \int_{I_n} \sigma(\langle y, x\rangle + \theta) \, d\mu(x) = 0 \quad \forall y \in \mathbb{R}^n, \, \forall \theta \in \mathbb{R}
  \]
  implies \( \mu = 0 \).
 \end{definition}

\begin{theorem} \label{th:UATgen} Let \( \sigma \) be any continuous discriminatory function. Then, finite sums of the form
\[
G(x) = \sum_{j=1}^N \alpha_j \sigma(\langle y_j, x\rangle + \theta_j)
\]
are dense in \( C(I_n) \).
\end{theorem}

\begin{proof}
Let \( S \) be the subspace of \( C(I_n) \) consisting of all finite linear combinations of functions of the form \( x \mapsto \sigma(\langle y, x \rangle + \theta) \) for \( y \in \mathbb{R}^n \) and \( \theta \in \mathbb{R} \) (i.e., the span of such functions). Clearly, \( S \) is a linear subspace.

Assume, for contradiction, that the closure \( \overline{S} \) (in the supremum norm) is a proper closed subspace of \( C(I_n) \). By the Hahn-Banach theorem, there exists a nonzero bounded linear functional \( L: C(I_n) \to \mathbb{R} \) such that \( L(g) = 0 \) for all \( g \in \overline{S} \) (and hence for all \( g \in S \)).

By the Riesz representation theorem, since \( L \) is a bounded linear functional on \( C(I_n) \), there exists a unique \( \mu \in M(I_n) \) such that
\[
L(h) = \int_{I_n} h(x) \, d\mu(x) \quad \forall h \in C(I_n).
\]
In particular, for each fixed \( y \in \mathbb{R}^n \) and \( \theta \in \mathbb{R} \), the function \( h(x) = \sigma(\langle y, x\rangle + \theta) \) belongs to \( S \) (and thus to \( \overline{S} \)), so
\[
L(\sigma(\langle y, \cdot \rangle + \theta)) = \int_{I_n} \sigma(\langle y, x \rangle+ \theta) \, d\mu(x) = 0 \quad \forall y \in \mathbb{R}^n, \, \forall \theta \in \mathbb{R}.
\]
Since \( \sigma \) is discriminatory, this implies \( \mu = 0 \). But \( \mu = 0 \) means \( L = 0 \), contradicting the assumption that \( L \neq 0 \).

Therefore, \( \overline{S} = C(I_n) \), so \( S \) is dense in \( C(I_n) \). 
\end{proof}

To apply Theorem~\ref{th:UATgen}  to sigmoidal functions, we need the following result:

\begin{lemma}  \label{le:UAT} Any continuous sigmoidal function \( \sigma \) is discriminatory.
\end{lemma}

\begin{proof}
Assume
$$
\int_{I_n} \sigma(\langle y, x \rangle+ \theta) \, d\mu(x) = 0 \quad \forall y \in \mathbb{R}^n, \, \forall \theta \in \mathbb{R}.
$$
We aim to show that $\mu = 0$.

Fix \( y \in \mathbb{R}^n \), \( \theta \in \mathbb{R} \), and consider scaling the argument. For any \( \lambda > 0 \), replace \( y \) with \( \lambda y \) and \( \theta \) with \( \lambda \theta \):
\[
\int_{I_n} \sigma(\lambda (y^T x + \theta)) \, d\mu(x) = 0.
\]
As \( \lambda \to +\infty \), the integrand \( \sigma(\lambda (\langle y,x\rangle  + \theta)) \) converges pointwise to the step function
\[
\psi(x) = 
\begin{cases} 
1 & \text{if } \langle y, x\rangle + \theta > 0, \\
0 & \text{if } \langle y, x\rangle + \theta < 0, \\
\sigma(\theta) & \text{if } \langle y, x \rangle + \theta = 0 
\end{cases}
\]
(since \( \sigma \) is sigmoidal with limits 1 and 0). Moreover, since \( \sigma \) is continuous and thus bounded on \( \mathbb{R} \), the convergence is bounded.

By the Lebesgue dominated convergence theorem (applicable because \( I_n \) has finite Lebesgue measure and the functions are bounded),
\[
\lim_{\lambda \to +\infty} \int_{I_n} \sigma(\lambda (\langle y, x \rangle + \theta)) \, d\mu(x) = \int_{I_n} \psi(x) \, d\mu(x) = \mu(H_{y,\theta}) + \sigma(0) \cdot \mu(\Pi_{y,\theta}) = 0,
\]
where \( H_{y,\theta} = \{ x \in I_n | \langle y, x\rangle + \theta > 0 \} \) is the open half-space and \( \Pi_{y,\theta} = \{ x \in I_n | \langle y, x \rangle + \theta = 0 \} \) is the hyperplane.

To isolate terms, note that by varying \( \theta \) slightly or using different scalings, we can show \( \mu(\Pi_{y,\theta}) = 0 \) for almost all hyperplanes (since hyperplanes have measure zero in general position). More rigorously, consider the pushforward measure along the line in direction \( y \).

Define the linear functional \( F: C(\mathbb{R}) \to \mathbb{R} \) by
$$
F(g) = \int_{I_n} g(\langle y, x \rangle) \, d\mu(x),
$$
which is well-defined and bounded (since $\mu $ is finite). The assumption implies \( F(\sigma(\cdot + \theta)) = 0 \) for all \( \theta \), and by the scaling argument above, \( F \) vanishes on the indicators of half-lines \( (a, +\infty) \) for any \( a \in \mathbb{R} \) (by choosing appropriate \( \theta \)).

Since indicators of intervals generate the Borel algebra, and simple functions (finite linear combinations of indicators) are dense in $ L^1(\mathbb{R}, \nu) $ for the induced measure $\nu$ (the pushforward of $\mu$ under the map $x \mapsto \langle y, x\rangle $), $ F$ vanishes on a dense set in the continuous functions.

In fact, since $F(\exp(2\pi i t \cdot)) = 0$ for the Fourier basis (by approximating exponentials via sigmoidal superpositions or directly from the vanishing on steps), the characteristic function of the pushforward measure is zero everywhere, implying the pushforward is zero. 
Thus,
$$
\int_{I_n} e^{2\pi i m \langle y, x\rangle} \, d\mu(x) = 0 \quad \forall m \in \mathbb{Z},
$$
implying the Fourier transform of $\mu$ vanishes, so $\mu = 0$ by uniqueness of Fourier transforms for measures. 
\end{proof}

\begin{proof}[of Theorem~\ref{th:UAT}]
The proof of  Theorem~\ref{th:UAT} follows now immediately by combining Theorem~\ref{th:UATgen} and Lemma~\ref{le:UAT}. 
\end{proof}

This proof demonstrates the validity of Theorem~\ref{th:UAT} without constructing explicit approximations, relying instead on separation theorems in functional analysis. Extensions (e.g.,~\cite{hornik1990universal,hornik1991approximation}) provide alternative proofs for broader classes of activations, see also~\cite{petersen2024mathematical}.

However, since the proof is nonconstructive, it ``only'' guarantees the existence of an approximating neural network, but it does not provide a method to find the weights $ \alpha_j, y_j, \theta_j $ or the number of neurons $ N $.

There exists extensive theory  establishing convergence rates for the approximation of a function $f$ by a neural network, see e.g.~\cite{mhaskar1993approximation,pinkus1999approximation,devore1998nonlinear}. For classical smoothness spaces it has been shown that to achieve an accuracy of $\eps$, the size of the network needs to increase exponentially in $d$, and this exponential dependence on $d$ is unavoidable~\cite{novak2009approximation,devore1998nonlinear}. 

Thus, additional assumptions on the function $f$ are needed if one wants to avoid this curse of dimensionality. One such regularity assumption is captured by the so-called {\em Barron class}.
Barron, in~\cite{barron2002universal}, introduces a Banach space of regular functions (now known as Barron class), defined by the semi-norm
$\|f\|_B:= \int |\hat{f}(\omega) | |\omega| \,d\omega$. Barron shows that, for any $n$, there exists a network with $n$
neurons such that the approximation error is of the order $1/\sqrt{n}$, when measured in the $L^2$-norm.

\subsection{When is deep  better than shallow?}

Adding more layers (depth) does not increase the class of functions that can be approximated in principle, since we just saw that even one hidden layer already suffices for density. 
But it dramatically changes the {\em efficiency} and {\em representational complexity} of approximation.

It is known (see e.g.~\cite{poggio2017and,yarotsky2017error,telgarsky2016benefits})
that some functions can be approximated exponentially more efficiently by deep networks than by shallow ones. For example
let $f(x) = x_1 x_2 \cdots x_d$ be a high-order monomial. In this case it has been shown that a shallow network needs exponentially many neurons in $d$ to approximate $f(x)$, while a deep network can approximate $f(x)$ with ${\mathcal O}(d)$ neurons using compositional structure~\cite{poggio2017and}. 

However, while it is not too difficult to construct a neural network (deep or shallow) that enables some form of universal approximation power, it is important to emphasize that this sheds little light on the success of deep learning. How come (carefully fine-tuned) neural networks with a huge number of parameters (weights) are able to generalize astonishingly well?  Traditional approaches fail to explain why large neural networks generalize well in practice~\cite{zhang2016understanding,zhang2021understanding}. 

Can Rademacher complexity help?
Recall that Rademacher complexity bounds measure the worst-case capacity of an entire hypothesis class, and for large neural networks these bounds are often {\em huge}, leading to vacuous generalization bounds.
However, the effective complexity seems to be much smaller than the worst-case class complexity. Understanding precisely when and why overparameterized networks generalize remains one of the central open questions in modern learning theory.

The ``double descent phenomenon'' is a noteworthy theoretical development in this context, which helps to shed some light on why deep neural networks generalize so well despite being massively overparameterized~\cite{belkin2019reconciling}.
It suggests that in the highly over-parameterized regime, deep networks achieve generalization by finding the most mathematically parsimonious way to interpolate the training data. Thus, the double descent curve refines our understanding of the bias-variance tradeoff in the overparametrized regime. Generalization depends not just on model size, but how the model is trained and which solutions are preferred.

\subsection{Infinite depth: Neural differential equations}

When we let the depth of a neural network approach infinity, we arrive at so-called 
neural differential equations~\cite{chen2018neural}. To see this, let us consider a standard feedforward neural network with $L$ layers, where each layer applies a transformation of the form
$$
a^{(\ell+1)} = a^{(\ell)} + f(a^{(\ell)}, \theta_{\ell}), \quad \ell = 0, 1, \dots, L-1,
$$
with $a^{(0)} = x$ denoting the input and 
$a^{(L)}$ the final hidden representation or output. This formulation corresponds to a residual network (ResNet) architecture, in which the update rule explicitly adds a residual term to the identity mapping.

To interpret this recursion as a discretization scheme, introduce a continuous ``depth'' variable $t \in [0, T]$ and let each layer correspond to a discrete time point $t_\ell = \ell,\Delta t$ with step size $\Delta t = T/L$. Then the layer-wise update can be rewritten as
$$
a^{(\ell+1)}(t_{\ell+1}) = a^{(\ell)}(t_\ell) + \Delta t  f\big(a^{(\ell)}(t_\ell), t_\ell, \theta(t_\ell)\big).
$$
This is precisely the forward (explicit) Euler method applied to the initial value problem
$$
\frac{da(t)}{dt} = f\big(a(t), t, \theta(t)\big), \quad a(0) = x.
$$
Hence, in the limit as the number of layers $L \to \infty$ (and thus $\Delta t \to 0)$, the discrete-depth neural network converges formally to a {\em neural differential equation}, or more specifically, a {\em neural ordinary differential equation}. The dynamics of the hidden representation are governed by
$$
\frac{da(t)}{dt} = f\big(a(t), t, \theta(t)\big),
$$
where $f$ is a neural network parameterizing the instantaneous rate of change of $a(t)$.

This continuous-depth perspective allows one to treat neural network architectures as parameterized dynamical systems. The terminal state $a(T)$ is obtained by integrating the differential equation from $t=0$ to $t=T$,
$$
a(T) = a(0) + \int_0^T f\big(a(t), t, \theta(t)\big) dt,
$$
and the parameters $\theta(t)$ are optimized to minimize a loss functional defined over the final state.

From an analytical viewpoint, this interpretation reveals that the expressiveness of deep networks can be understood through the lens of continuous-time dynamical systems theory, with residual connections corresponding to first-order numerical integration schemes.

\section{Training neural networks}\label{nn:training}

Training a neural network involves finding the optimal weights $W^{(\ell)}$ and biases $b^{(\ell)}$ that minimize a cost function or loss function $\cost$. 
Due to the typically large amount of data, higher-order optimization methods are usually prohibitively expensive. The work horse for this process typically comprises gradient-based optimization algorithms, regardless of the specific network architecture. 

\subsection{Loss functions}\label{s:loss}

Loss functions (called cost functions in optimization or error functions in numerical analysis) are critical in guiding the learning process by quantifying how well the model's predictions align with the true targets. The choice of the loss function depends  on the specific task (e.g., regression, classification) and the characteristics of the data. For example, the mean squared error is often employed for regression tasks:
$$
\cost(y, \hat{y}) = \frac{1}{n} \sum_{i=1}^n ( y_i - \hat{y}_i )^2,
$$
The  mean absolute error is given by
$$
\cost(y, \hat{y}) =  \frac{1}{n} \sum_{i=1}^{n} |y_i - \hat{y}_i|.
$$
It is also used for regression tasks, particularly when robustness to outliers is desired, since it is less sensitive to outliers than the mean squared error.

The cross-entropy loss is a standard choice for classification tasks:
$$
\cost(y, \hat{y})  = - \sum_{i=1}^n y_i \log(\hat{y}_i).
$$
The multi-class cross-entropy loss generalizes binary cross-entropy to multiple classes. In detail,
for multi-class classification, where the true target is a one-hot encoded vector $ \mathbf{y}_i $ and the predicted probabilities for each class are given by $ \hat{\mathbf{y}}_i $, the cross-entropy loss is:
$$
\cost(y, \hat{y})  = -\frac{1}{n} \sum_{i=1}^{n} \sum_{c=1}^{C} y_{ic} \log(\hat{y}_{ic}),
$$
where $ C $ is the number of classes, $ y_{ic} $ is the true label (1 or 0) for class $ c $, and $ \hat{y}_{ic} $ is the predicted probability for class $ c $.

The Hinge Loss is primarily used in Support Vector Machines  but  also finds use in  neural networks.
For binary classification, where $ y_i \in \{-1, 1\} $ and $ \hat{y}_i $ is the predicted score (not a probability), the hinge loss is defined as:
$$
\cost(y,\hat{y}) = \frac{1}{n} \sum_{i=1}^{n} \max(0, 1 - y_i \hat{y}_i).
$$
The hinge loss is used in contexts where the predictions are scores rather than probabilities, and it aims to ensure that the true class is not only predicted correctly but with a margin of at least 1. If $ y_i \hat{y}_i \geq 1 $, the loss is 0, meaning the prediction is correct with a sufficient margin.
 If $ y_i \hat{y}_i < 1 $, the loss increases linearly, penalizing the model for incorrect or insufficiently confident predictions.

The Kullback-Leibler (KL) Divergence measures how one probability distribution diverges from a second, expected probability distribution. It is typically used in variational autoencoders and reinforcement learning.
For two probability distributions $ P $ (true distribution) and $ Q $ (predicted distribution), the KL divergence is defined as:
$$
\text{KL}(P \parallel Q) = \sum_{i=1}^{n} P(i) \log\left(\frac{P(i)}{Q(i)}\right).
$$
It is not symmetric, meaning $ \text{KL}(P \parallel Q) \neq \text{KL}(Q \parallel P) $.
In deep learning, the KL divergence is often used in settings where the model needs to match its output distribution $ Q $ to a target distribution $ P $.

The Huber loss attempts to combine the best properties of the mean squared error and the mean absolute error. For a prediction $ \hat{y}_i $ and target $ y_i $, it is defined as:
$$
\cost(y,\hat{y}) = \frac{1}{n} \sum_{i=1}^{n} C_{\delta}(y_i - \hat{y}_i ),
$$
where
$$
C_{\delta}(r) = \begin{cases}
\frac{1}{2} r^2 & \text{for } |r| \leq \delta, \\
\delta (|r| - \frac{1}{2} \delta) & \text{for } |r| > \delta.
\end{cases}
$$
The Huber loss is typically used for regression tasks, particularly when dealing with outliers. It is less sensitive to outliers than the mean squared error.

\subsection{Stochastic gradient descent}\label{ss:SGDsubsection}

Training a neural network corresponds to choosing its parameters, i.e., the weights and biases, that
minimize the cost function. 
Given the high dimensionality and non-linearity of neural networks, finding the optimal parameters becomes a challenging task. So, how shall we do that?

Enter {\em gradient descent}\,!
The use of gradient descent is usually discouraged in classical numerical optimization and often for good reason: Gradient descent generally can have a slow rate of convergence, is sensible to the choice of step size and initialization, and can easily get stuck in saddle points and local extrema~\cite{nocedal2006numerical}. However, in many applications in data science and machine learning it suffices to compute a solution with relatively low accuracy and often one is content with finding local minima. Also, since gradient descent uses only first-order information (namely, the gradient), it can better cope with limited computing resources vis-{\`a}-vis very large datasets compared to second-order methods. Due to these considerations gradient descent has become the workhorse in training neural networks.

Specializing the results in Chapter~\ref{s:grad} to the setting to neural networks, we interpret $f(x)$ as the network's cost function $\cost(\theta)$, with its parameters $\theta$ (the weights and biases) collected in the vector $x$.

Deep learning uses stochastic gradient descent (SGD), which was discussed in Chapter~\ref{s:grad}, instead of standard (full-batch) gradient descent because of computational, statistical, and optimization-landscape reasons that become decisive at the scale and structure of modern neural networks.

Recall that in the standard gradient descent algorithm, the gradient is computed as
$$
\nabla_\theta \loss(\theta) = \frac{1}{n} \sum_{i=1}^{n} \nabla_\theta \loss_i(\theta),
$$
where $\loss_i(\theta)$ represents the loss for the $ i $-th data point.

In contrast, SGD updates the parameters using the gradient computed from a single randomly selected data point $ (x_i, y_i) $:
$$
\theta^{(k+1)} = \theta^{(k)} - \eta \nabla_\theta \loss_i(\theta^{(k)}).
$$

A common variant of SGD in deep learning is {\em mini-batch gradient descent}, where the gradient is computed over a small batch of data points instead of a single data point
$$
\theta^{(k+1)} = \theta^{(k)} - \eta \frac{1}{|B_k|} \sum_{i \in B_k} \nabla_\theta \loss_i(\theta^{(k)}),
$$
where $ B_k $ is the set of indices in the $ k $-th mini-batch, and $ |B_k| $ is the batch size. Mini-batch gradient descent  attempts to strike a balance between the computational efficiency of SGD and the stability of full-batch gradient descent. It is widely used in practice due to its ability to leverage parallelism and its effectiveness in noisy environments.
The convergence of SGD can be analyzed in terms of the expected value of the loss function, see also Section~\ref{ss:kaczmarz} for the special case of applying SGD to a linear system of equations.

The step size of gradient descent, in this context called the \emph{learning rate}, is an important design parameter. If the learning rate is too large, the noise can cause the algorithm to diverge. Conversely, if the learning rate is very small, the algorithm may converge rather slowly.
If the learning rate $ \eta $ is sufficiently small, SGD converges to a neighborhood around a local minimum. The exact nature of this neighborhood depends on the noise characteristics and the curvature of the loss function.

One way to enhance stability is to use a decaying learning rate
$$
\eta_k = \frac{\eta_0}{1 + \alpha k},
$$
where $ \eta_0 $ is the initial learning rate, and $ \alpha $ is a decay factor. 

\medskip

Adaptive methods such as Adam~\cite{kingma2014adam} or AdaGrad~\cite{duchi2011adaptive} adjust step sizes dynamically and this frequently leads to empirical improvements but complicates the theoretical guarantees~\cite{ward2020adagrad}.
While Adam is often the method of choice for training speed, it has been observed in simulations that switching to SGD with momentum (see Chapter~\ref{s:grad} for a discussion of gradient descent with momentum) for the final stages of training can achieve better generalization on test data for some architectures~\cite{wilson2017marginal}.

\subsection{Backpropagation}

Backpropagation is the fundamental algorithm used for training neural networks by computing the gradients of the loss function with respect to each weight in the network. It leverages the chain rule from calculus to efficiently compute the gradient of the loss function with respect to every weight in the network.

It will be convenient in the following to denote a neural network by a function $f(x,W)$, where $x$ are the input data and $W$ represents all the weight and bias parameters. The goal of training is to find the set of parameters $W^*$ that minimizes a loss function $\loss$. It will be helpful in our derivation of the backpropagation algorithm to make the dependence of the loss function on the
the network parameters explicit by writing $\loss\big(f(x_i, W), y_i\big)$.
We  are concerned with solving:
$$W^* = \underset{W}{\arg\min} \left\{ \frac{1}{N} \sum_{i=1}^N \loss(f(x_i, W), y_i) \right\},$$
where, as above, $y_i$ is the true output for input ${x}_i$. Backpropagation computes the gradient $\nabla_{W} L$ so that the weights can be updated using (stochastic) gradient descent:
$$W \leftarrow W - \eta \nabla_{W} L,$$
where $\eta$ is the learning rate.

The goal is to compute $\frac{\partial \loss}{\partial w_{jk}^{(\ell)}}$ and $\frac{\partial \loss}{\partial b_j^{(\ell)}}$ for all weights $w_{jk}^{(\ell)}$ and biases $b_j^{(\ell)}$.
The input  $z_j^{(\ell)}$ to neuron $j$ in layer $\ell$ is a weighted sum of the outputs from the previous layer, $a_k^{(\ell-1)}$, i.e., 
    $$z_j^{(\ell)} = \sum_{k} w_{jk}^{(\ell)} a_k^{(\ell-1)} + b_j^{(\ell)}.$$
Applying the chain rule, the gradient of the loss $\loss$ with respect to a weight $w_{jk}^{(\ell)}$ in layer $\ell$ is:
$$\frac{\partial \loss}{\partial w_{jk}^{(\ell)}} = \frac{\partial \loss}{\partial z_j^{(\ell)}} \frac{\partial z_j^{(\ell)}}{\partial w_{jk}^{(\ell)}}.$$
From the definition of $z_j^{(\ell)}$, we have
$$\frac{\partial z_j^{(\ell)}}{\partial w_{jk}^{(\ell)}} = a_k^{(\ell-1)}.$$
Substituting this back, and defining the term $\delta_j^{(\ell)}$
$$\delta_j^{(\ell)} \equiv \frac{\partial \loss}{\partial z_j^{(\ell)}},$$
the weight gradient becomes
\begin{equation}
    \label{BP-1}
    \frac{\partial \loss}{\partial w_{jk}^{(\ell)}} = \delta_j^{(\ell)} a_k^{(\ell-1)}.
\end{equation}
Similarly, the bias gradient is
\begin{equation}
    \label{BP-2}
\frac{\partial \loss}{\partial b_j^{(\ell)}} = \frac{\partial \loss}{\partial z_j^{(\ell)}} \frac{\partial z_j^{(\ell)}}{\partial b_j^{(\ell)}} = \delta_j^{(\ell)} \cdot 1 = \delta_j^{(\ell)}.
\end{equation}

The key challenge is to compute the error term $\delta_j^{(\ell)}$ for all layers.
The loss $\loss$ depends directly on the output activations $a_j^{(L)}$ and the activations $a_j^{(L)}$ depend on the net inputs $z_j^{(L)}$. We have
$$\delta_j^{(L)} = \frac{\partial \loss}{\partial z_j^{(L)}} = \frac{\partial \loss}{\partial a_j^{(L)}} \frac{\partial a_j^{(L)}}{\partial z_j^{(L)}}.$$

Since $a_j^{(L)} = \sigma(z_j^{(L)})$, the second term is the derivative of the activation function
$$\frac{\partial a_j^{(L)}}{\partial z_j^{(L)}} = \sigma'(z_j^{(L)}).$$
Hence, the output layer error is
\begin{equation}
    \label{BP-3}
\delta_j^{(L)} = \frac{\partial \loss}{\partial a_j^{(L)}} \sigma'(z_j^{(L)}).
\end{equation}

The error term $\delta_j^{(\ell)}$ for a hidden layer $\ell$ is found by relating it to the errors in the next layer, $\delta_k^{(\ell+1)}$. This is where the name {\em backpropagation} comes from.

Applying the chain rule again gives
\begin{equation}
\label{delta0}    
\delta_j^{(\ell)} = \frac{\partial \loss}{\partial z_j^{(\ell)}} = \sum_{k} \frac{\partial \loss}{\partial z_k^{(\ell+1)}} \frac{\partial z_k^{(\ell+1)}}{\partial z_j^{(\ell)}},
\end{equation}
where the sum is over all neurons $k$ in layer $\ell+1$.

Substituting the definition $\delta_k^{(\ell+1)} = \frac{\partial \loss}{\partial z_k^{(\ell+1)}}$ into~\eqref{delta0} yields
\begin{equation}
    \label{deltaeq}
\delta_j^{(\ell)} = \sum_{k} \delta_k^{(\ell+1)} \frac{\partial z_k^{(\ell+1)}}{\partial z_j^{(\ell)}}.
\end{equation}

Now we compute the second term. From $z_k^{(\ell+1)} = \sum_{j'} w_{kj'}^{(\ell+1)} a_{j'}^{(\ell)} + b_k^{(\ell+1)}$ and $a_j^{(\ell)} = \sigma(z_j^{(\ell)})$ we get
$$\frac{\partial z_k^{(\ell+1)}}{\partial z_j^{(\ell)}} = \frac{\partial}{\partial z_j^{(\ell)}} \left( \sum_{j'} w_{kj'}^{(\ell+1)} a_{j'}^{(\ell)} + b_k^{(\ell+1)} \right) = w_{kj}^{(\ell+1)} \frac{\partial a_j^{(\ell)}}{\partial z_j^{(\ell)}} = w_{kj}^{(\ell+1)} \sigma'(z_j^{(\ell)}).$$

Plugging this back into equation~\eqref{deltaeq} gives
$$\delta_j^{(\ell)} = \sum_{k} \delta_k^{(\ell+1)} w_{kj}^{(\ell+1)} \sigma'(z_j^{(\ell)}).$$

We now factor out the term $\sigma'(z_j^{(\ell)})$ which is independent of the summation and arrive at
\begin{equation}
    \label{BP-4}
\delta_j^{(\ell)} = \Big( \sum_{k} w_{kj}^{(\ell+1)} \delta_k^{(\ell+1)} \Big) \sigma'(z_j^{(\ell)}).
\end{equation}

This is the ``Backpropagation Error Equation'': the error at layer $\ell$ is calculated by taking a weighted sum of the errors from the next layer $\ell+1$ and multiplying it by the derivative of the activation function at layer $\ell$.

Backpropagation has two phases, a {\em forward pass}  and a {\em backward pass}.
The forward pass consists of the following steps:
\begin{enumerate}[(i)]
\setlength{\itemsep}{-0.3ex}
\item Initialize all weights $W$ and biases ${b}$.\\
\item Propagate the input $x$ through the network, layer by layer, computing all net inputs ${z}^{(\ell)}$ and activations ${a}^{(\ell)}$ for all $\ell$.
    $${a}^{(\ell)} = \sigma\big(W^{(\ell)} {a}^{(l-1)} + {b}^{(\ell)}\big).$$
\item Compute the loss $\loss$ at the output layer ${a}^{(L)}$.
\end{enumerate}
The backward pass consists of four steps which comprise the gradient calculation and update:
\begin{enumerate}[(i)]
\setlength{\itemsep}{-0.3ex}
\item Output error ($\delta^{(L)}$): Compute the error term for the output layer $L$ using equation~\eqref{BP-3}:
    $$\delta_j^{(L)} = \frac{\partial \loss}{\partial a_j^{(L)}} \sigma'(z_j^{(L)}).$$
\item Backpropagate error ($\delta^{(\ell)}$): Compute the error term for each preceding layer $\ell=L-1, L-2, \dots, 1$ using the errors from the layer above (equation~\eqref{BP-4}):
    $$\delta_j^{(\ell)} = \Big( \sum_{k} w_{kj}^{(\ell+1)} \delta_k^{(\ell+1)} \Big) \sigma'(z_j^{(\ell)}).$$
\item Calculate gradients ($\nabla_{W} \loss$ and $\nabla_{{b}} \loss$): Use the computed error terms $\delta^{(\ell)}$ to find the gradients for all weights and biases, (equations~\eqref{BP-1} and~\eqref{BP-2}):
\begin{align*}
    \frac{\partial \loss}{\partial w_{jk}^{(\ell)}} & = \delta_j^{(\ell)} a_k^{(\ell-1)}, \\
    \frac{\partial \loss}{\partial b_j^{(\ell)}} & = \delta_j^{(\ell)}.
\end{align*}
\item Update parameters: Perform a gradient descent step to update the weights and biases
\begin{align*}
    W^{(\ell)} & \leftarrow W^{(\ell)} - \eta \frac{\partial \loss}{\partial W^{(\ell)}}, \\
    {b}^{(\ell)} & \leftarrow {b}^{(\ell)} - \eta \frac{\partial \loss}{\partial {b}^{(\ell)}}.
\end{align*}
\end{enumerate}

Backpropagation is computationally efficient because it reuses intermediate computations from the forward pass. Specifically, it computes gradients layer by layer in reverse, leveraging the fact that the gradient of the loss with respect to earlier layers can be expressed recursively in terms of gradients from later layers. This efficiency is crucial for training deep neural networks, where the number of parameters can be large, and the computation of gradients must be optimized.

\subsection{Regularization}

Training deep networks presents several challenges, which necessitate the use of some form of regularization. These challenges include:

\noindent
{\em Vanishing and exploding gradients:} During backpropagation, gradients can become very small or very large, making learning difficult. Techniques such as careful initialization, batch normalization, and gradient clipping as well as using ReLU as activation function are used to mitigate these issues.

\noindent
{\em Overfitting:} Deep networks with many parameters are prone to overfitting the training data. To prevent overfitting, regularization techniques are often employed. Common methods include   adding a penalty term such as $ \lambda \|W\|_2^2 $ to the loss function and a technique called \emph{dropout}: to randomly drop units from the neural network during training to prevent co-adaptation of neurons. 

\noindent
{\em Early stopping} is a particularly effective strategy in which training is halted once performance on the validation set ceases to improve. This implicitly regularizes the model and reduces the risk of fitting noise in the training data.

\noindent
{\em Data augmentation} is another technique used to avoid overfitting by increasing the diversity of the training data  without actually collecting new data. It involves applying various transformations to the existing data to create new, slightly modified versions of it. For example, in image classification, an object should be recognizable regardless of its position, orientation, or lighting conditions. By augmenting the data to include controlled variations in the image training data, such as shifts, rotations, flips, scaling, and changes in brightness or contrast,  we can help the model to generalize better to unseen data.

\subsection{Practical setup for training neural networks}

Training a neural network is not only a matter of choosing an architecture and an optimization algorithm; it also requires a carefully designed experimental protocol to ensure that the resulting model generalizes beyond the data it was trained on. We briefly describe best practices for setting up a reliable neural network training pipeline.
These principles actually apply to any supervised data science learning task.

A fundamental principle of {\em supervised learning} is the separation of data into disjoint subsets with distinct roles. Typically, the available dataset is divided into a {training set}, a {\em validation set}, and a {\em test set}. A common heuristic is to split the data into 70\% training, 10\% validation, and 20\% test data\footnote{While an 70-10-20 split is a typical starting point for medium-sized datasets, in the era of ``Big Data'' , it is not unusual to see 98-1-1 splits, as even 1\% of data can provide a statistically significant evaluation. Nevertheless, a more conservative split provides a better defense}.

The training set is used to fit the model parameters via optimization algorithms such as stochastic gradient descent. The validation set is used to tune hyperparameters and make modeling decisions, while the test set is held out and used only once to obtain an unbiased estimate of the final model’s generalization performance. 
Cross-validation (introduced in connection with ridge regression in  Definition~\ref{def:cv_ridge}) is typically used for parameter tuning.
Crucially, the test set must not influence any aspect of model design or training.

Neural networks involve numerous hyperparameters, including learning rates, batch sizes, network depth and width, regularization coefficients, and optimization settings. 
Hyperparameter tuning is typically performed by training multiple models on the training set and evaluating their performance on the validation set. Techniques range from manual tuning and grid search to more sophisticated approaches such as random search or Bayesian optimization. Importantly, all hyperparameter choices must be finalized before evaluating performance on the test set.

Data leakage occurs when information from outside the training set is inadvertently used during training or model selection, leading to overly optimistic performance estimates. Common sources of leakage include preprocessing steps (such as normalization or feature selection) performed using the full dataset, or repeated evaluation on the test set during model development.

To avoid data leakage, all data-dependent preprocessing operations should be fit exclusively on the training data and then applied unchanged to the validation and test sets. Likewise, the test set should remain completely isolated until the very end of the training pipeline.

Once the training and tuning process is complete, the final model is evaluated on the test set to estimate its generalization performance. This evaluation should be performed exactly once to prevent data leakage.

\subsection{Computational complexity} 
Training large neural networks requires significant computational resources. The development of specialized hardware (e.g., GPUs, TPUs) and distributed training techniques has helped to alleviate this problem. 

However, computational complexity remains a critical concern, in particular for Large Language Models. Training these models demands immense computational resources, with cost and time scaling rapidly with model size, dataset volume, and sequence length. Architectures like Transformers involve operations that scale quadratically with input length, making efficiency a key bottleneck.

\section{Deep learning architectures}

Various architectures have been developed to handle different types of data and tasks. In the previous sections we already have discussed  (fully connected) feedforward networks, also known as multilayer perceptrons. In this section we briefly survey  other widely used architectures.

\subsection{Convolutional neural networks}

Convolutional Neural Networks~\cite{fukushima1980neocognitron,lecun2002gradient} (CNNs) are the architecture that fundamentally enabled the modern breakthrough in computer vision. Indeed,
CNNs are particularly effective for processing grid-like data such as images. They use convolutional layers to extract local patterns through learned filters. 

Mathematically, a CNN is a neural network where the weight vectors in a given layer are formed by discrete convolution
with a finite number of kernels. For one-dimensional data, this is equivalent to saying that the weight
matrix $W^{(\ell)}$ is a Toeplitz matrix, i.e., 
\begin{equation}\label{layertoeplitz}
  W^{(\ell)} = 
  \begin{pmatrix}
    t_1       & t_2       & \ldots & t_{n}\\
    t_2^{\ast}  & t_1       & \ddots & \vdots\\
    \vdots & \ddots & \ddots  & t_2\\
    t_{n}^{\ast}      & \ldots & t_2^{\ast}   &  t_1
  \end{pmatrix},
\end{equation}
where for convenience we have omitted the layer dependence in the individual matrix entries in~\eqref{layertoeplitz}.
The Toeplitz matrix can be equivalently represented by a {\em filter} or {\em kernel} $g$ defined via $g(j) = t_j$. 
When the input is two-dimensional (such as images), the weight matrix of a CNN has the structure of a block-Toeplitz matrix with Toeplitz blocks. In either case, one can take advantage of the Toeplitz-type structure, which gives rise to fast matrix-vector multiplication via the (1D or 2D) Fast Fourier transform.

CNNs typically use multiple kernels in their convolutional layers. This is motivated by the hope  that each filter will learn to detect a different kind of feature in the input. Having multiple filters allows the layer to build up a set of complementary feature maps that together describe the input from different perspectives.

There are two main advantages to a CNN compared to a general neural network:
\begin{enumerate}
  \item The convolutional structure reduces the number of unknown weights; instead of having $nk$
  weights, where $n$ is the input size and $k$ is the number of units in the
  layer, there are only $n$ weights for $n$ units.
  \item  Convolution is translation-invariant, since the convolution operator commutes with the translation operator. So if the item to be classified is also translation invariant, such as objects in images, or sounds in
  audio signals, then the resulting classifier will also be translation
  invariant.
\end{enumerate}
Usually this convolution is combined with subsampling, which is intended to create insensitivity
against scaling. In the CNN literature, this is known as pooling, and there are
several ways this is implemented, such as max pooling or average pooling. Consider a ``pooling window'' of size $k \times k$ centered or anchored at position $(i,j)$ (in the initial layer position $(i,j)$ corresponds to the pixel in the $i$-th row and $j$-th column of the input image). Max pooling selects the maximum value within the window. It acts as a non-linear feature detector, identifying the presence of a specific pattern (like an edge or texture) regardless of its exact location within the window. Average pooling computes the arithmetic mean of all values in that window. It treats the feature map as a signal and acts as a low-pass filter, smoothing out the activations

\begin{figure}
\begin{center}
\includegraphics[width=\textwidth]{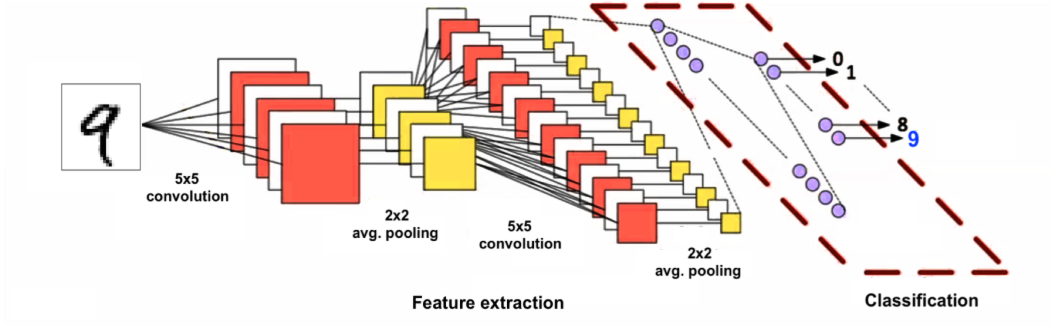}
\caption{A common architecture for a Convolutional Neural Network (CNN) used for image classification, specifically for recognizing handwritten digits. The network takes an image of a handwritten digit. The feature extraction phase involves alternating convolutional layers (in this example using a $5\times 5$ kernel) and pooling  layers (in this example using $2\times 2$ average pooling) to automatically extract features from the image. In the classification step the extracted features are flattened and passed to a fully connected network, which then classifies the input image into one of the possible output classes (digits 0 through 9).}
\label{fig:cnn}
\end{center}
\end{figure}

Figure~\ref{fig:cnn} illustrates a common architecture for a Convolutional Neural Network (CNN) used for image classification.

\begin{figure}
\begin{center}
\includegraphics[width=.3\textwidth,height=0.21\textwidth]{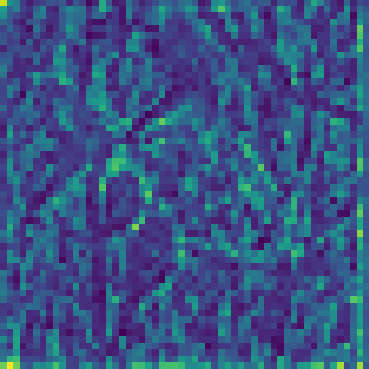}
\qquad \qquad
\includegraphics[width=.3\textwidth,height=0.21\textwidth]{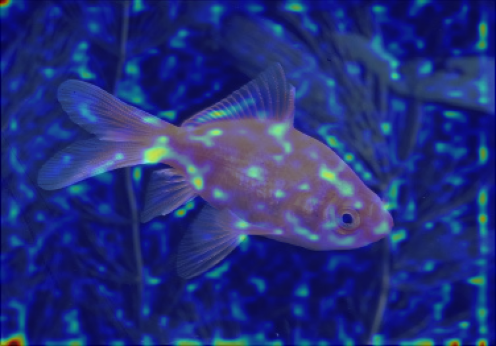}

\bigskip

\includegraphics[width=.3\textwidth,height=0.21\textwidth]{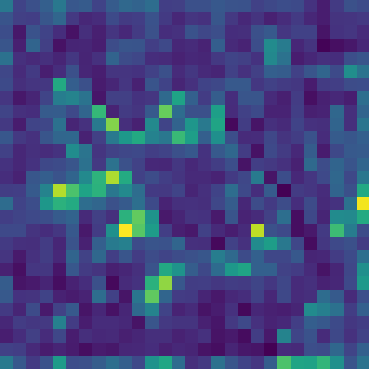}
\qquad \qquad
\includegraphics[width=.3\textwidth,height=0.21\textwidth]{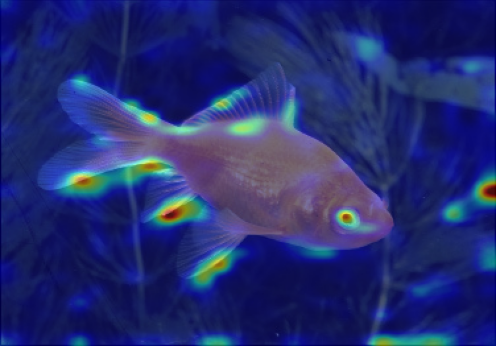}
\bigskip

\includegraphics[width=.3\textwidth,height=0.21\textwidth]{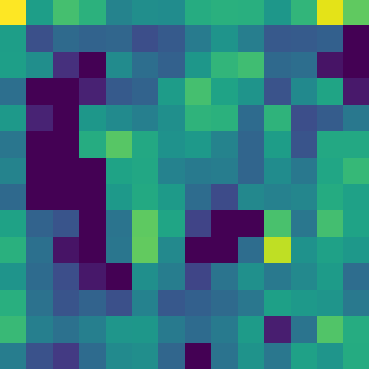}
\qquad \qquad
\includegraphics[width=.3\textwidth,height=0.21\textwidth]{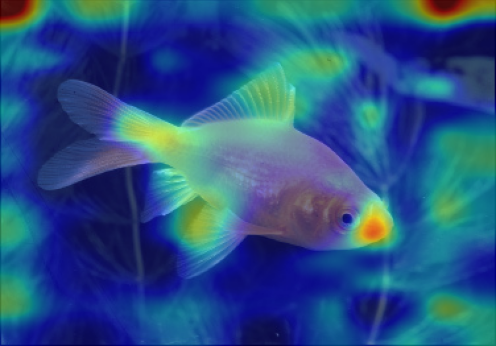}

\bigskip

\includegraphics[width=.3\textwidth,height=0.21\textwidth]{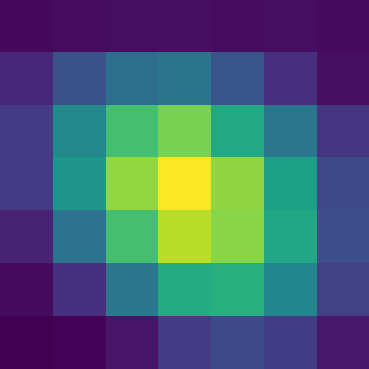}
\qquad \qquad 
\includegraphics[width=.3\textwidth,height=0.21\textwidth]{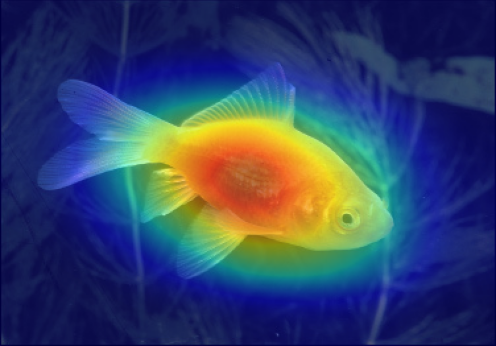} 
\caption{The left column shows the Class Activation Map (CAM) for layers 1 (top) 2,3, and 4 (bottom). The right columns show the CAM overlaid with the original image.
It is evident how the CNN picks up localized information in the initial layers and more global features in the latter layers.}
\label{fig:CAM}
\end{center}
\end{figure}

There is ample empirical evidence that 
in early layers of the CNN, filters might detect simple local patterns (edges, corners, color gradients), while
in deeper layers, filters respond to more complex ``global'' patterns (textures, shapes, or object parts).
See Figure~\ref{fig:CAM} for an illustration of this observation.  Here we use a CNN with four convolutional layers. The input image to the CNN is an image of a goldfish.
The figure displays a so-called Class Activation Map (CAM) for each layer. The CAM is a technique used for CNN's in connection with image classification to visualize which regions of an input image are most responsible for a model’s decision for a particular class~\cite{zhou2016learning}. It is intended to provide an interpretable {\em heatmap} highlighting the spatial regions that contribute most strongly to the predicted class score. A CAM is essentially a weighted sum of all the feature maps of a specific layer.  The left column in Figure~\ref{fig:CAM} shows the CAM for layers 1 to 4 (from top to bottom). The right columns depicts the CAM overlaid with the original image\footnote{The plots were generated using the software package in~\cite{torcham2020}}. It is evident how the CNN picks up localized information in the initial layers and more global features in the latter layers.

CNNs are widely used in computer vision tasks, such as
image classification~\cite{lecun2002gradient,krizhevsky2012imagenet} (assigning a label to an entire image), object detection~\cite{redmon2016you} (identifying and locating objects within an image), and image segmentation~\cite{long2015fully,ronneberger2015u} (classifying each pixel in an image to a particular class), which is useful in tasks like medical image analysis.

\subsection{Graph Neural Networks}

A Graph Neural Network (GNN) is a machine learning model designed to perform inference on data structured as graphs~\cite{scarselli2008graph,Kipf2017}. A GNN learns representations by iteratively propagating and transforming information along the edges of a graph, combining local connectivity via message passing with learnable nonlinear transformations

Given a graph $G=(V,E)$ with nodes $v \in V$ and edges $e \in E$,
let $h_{v}^{(k)}$ denote the representation of node $v$ at layer $k$.
A generic message-passing update is:
\begin{equation}
    \label{GNN}
h_{v}^{(k+1)} = \sigma \Big( W  h_v^{(k)} + \underbrace{\sum_{u \in \mathcal{N}(v)} \phi(h_v^{(k)}, h_u^{(k)}, e_{uv})}_{m_v^{(k)}} \Big),
\end{equation}
where $\mathcal{N}(v)$ denotes neighbors of $v$, $m_v^{(k)}$ is the updated message, and $\phi$ is a learnable message function possibly depending on edge features $e_{uv}$. This preserves  permutation invariance of node orderings. 

At a high level, a graph neural network layer consists of two conceptually distinct operations, repeated across layers:
\begin{enumerate}

\item {\em Message passing (neighborhood aggregation):}
Each node $v$ receives ``messages'' from its neighbors $u \in \mathcal{N}(v)$
based on their current features and possibly the connecting edge features.
These messages are then aggregated (e.g. by summing, averaging, or taking a weighted combination). This corresponds to the term
$m_v^{(k)}=\sum_{u \in \mathcal{N}(v)} \phi(h_v^{(k)}, h_u^{(k)}, e_{uv})$ in equation~\eqref{GNN}.
\item {\em Feature update (node feature transformation):}
The aggregated message is combined with the node’s own current feature vector, passed through a learnable transformation (often followed by a nonlinearity), yielding updated node 
embeddings. This corresponds to the 
term $\sigma(W  h_v^{(k)} + m_v^{(k)})$ in equation~\eqref{GNN}.
\end{enumerate}
By stacking multiple GNN layers, a node's final representation $h_v^{(L)}$ can capture information from its $L$-hop neighborhood (nodes up to $L$ steps away). The final representations are then used for downstream tasks like node classification, link prediction, or graph classification.

A popular GNN variant, is the Graph Convolutional Network (GCN)~\cite{Kipf2017}. The layer-wise propagation rule of a GCN can be expressed as:
$$
H^{(k)} = \sigma \left( \tilde{D}^{-\frac{1}{2}} \tilde{A} \tilde{D}^{-\frac{1}{2}} H^{(k-1)} W^{(k)} \right),
$$
where $H^{(k-1)} \in \mathbb{R}^{|V| \times d_{k-1}}$ is the matrix of node representations from layer $k-1$, $W^{(k)}$ is a learnable weight matrix (the model's parameters), $\sigma$ is a non-linear activation function,
$\tilde{A} = A + I$ is the adjacency matrix with self-loops (allowing a node to include its own previous state in the aggregation), where $A$ is as defined in~\eqref{adjacencymatrix}, and
$\tilde{D}$ is the degree matrix (as defined in~\eqref{degreematrix}) of $\tilde{A}$.

The reader may wonder where the concept of a convolution enters in graph convolutional networks, which would justify their name. Spectral graph theory provides the answer. We can define the concept of a convolution on a graph in terms of the graph Laplacian
$L = D - A$, as defined in~\eqref{graphlalpacian}.
The graph Laplacian has an orthonormal eigendecomposition $L = U \Lambda U^T$,
where $U$ contains the eigenvectors (graph Fourier modes), and $\Lambda$
the eigenvalues (graph frequencies) of $L$.
We can now  define convolution on graphs by analogy to the standard convolution theorem where $U$ plays the role of the Fourier transform on a graph. One can then show that a GCN is a  first-order approximation of  convolution on graphs~\cite{defferrard2016convolutional}.

Graph neural networks find applications in particle physics~\cite{kipf2018neural},  molecular and chemical modeling to predict molecular properties~\cite{kearnes2016molecular},  drug discovery~\cite{bongini2021molecular}, and protein structure analysis~\cite{fout2017protein}, as well as in traffic network modeling and in recommendation systems~\cite{zhou2020graph}.

\subsection{ Recurrent Neural Networks }

For sequential data, Recurrent Neural Networks (RNNs)~\cite{cho2014learning} introduce {\em recurrence} by updating a hidden state $h_t$ using input $x_t$ via
$$
h_t = \sigma(W_h h_{t-1} + W_x x_t + b), \quad y_t = W_o h_t,
$$
where we emphasize that $h_t$ depends on $h_{t-1}$.
The hidden state carries memory across time steps. Variants such as Long Short-Term Memory (LSTM)~\cite{Hochreiter1997} modify the update rule by introducing {\em gating mechanisms} to mitigate vanishing and exploding gradients, enabling modeling of longer dependencies.

RNNs have been applied to natural language modeling (before transformers took over), in
time-series forecasting (financial markets, weather prediction, physiological monitoring in healthcare), and 
to sequential control (robotics and reinforcement learning environments with temporal structure).

In RNNs, the influence of a past input $x_t$ on later states $h_{t+k}$ propagates through repeated multiplication by the recurrent Jacobian
$$
\frac{\partial h_{t+k}}{\partial h_t} = \prod_{i=1}^k \frac{\partial h_{t+i}}{\partial h_{t+i-1}}.
$$

This leads to vanishing (or sometimes exploding) gradients, making it difficult to learn dependencies over long distances in the sequence.
While standard feed forward neural networks also can suffer from this problem as they grow (deeper for neural networks, longer for RNNs), the mathematical mechanism in RNNs makes the decay exponential relative to the sequence length. RNN multiplies the same recurrent matrix many times. This makes the vanishing/exploding behavior potentially much more systematic and severe.

Although LSTMs were explicitly designed to solve the vanishing gradient problem that plagued simple RNNs, they still struggle to effectively maintain information and capture dependencies over very long sequences in practice.

The fundamental design of RNNs and LSTMs requires them to process sequence data (like words in a sentence) one step at a time. The calculation for the current word's hidden state depends on the hidden state of the previous word,  meaning one cannot compute the representation for the entire sequence simultaneously. This sequential dependency makes it impossible to fully parallelize the computation across modern hardware (like GPUs or TPUs). Training RNNs and LSTMs, therefore, becomes computationally expensive and time-consuming as datasets and sequences become very large,
because the network must be ``unrolled'' over the entire sequence and trained via backpropagation through time, leading to linear growth in computation and memory with sequence length. Additionally, the inherently sequential structure prevents parallelization across time steps, and LSTMs add further per-step computational overhead.

\subsection{Transformer networks}

The creation of Transformers was driven primarily by the two aforementioned major limitations of RNNs and their variants like LSTM networks: the inability to efficiently capture long-range dependencies and the lack of parallelization. The transformer architecture~\cite{vaswani2017attention} addresses these limitations by replacing recurrence with so-called attention mechanisms, creating a direct link between every word and every other word in the sequence, regardless of distance and allowing global context to be modeled in parallel. Transformers  have become the state-of-the-art architecture for natural language processing tasks and as such are the core architecture behind modern Large Language Models (LLMs).

We give a brief mathematical derivation of transformers, following in part the approach taken in~\cite{turner2023introduction}. A {\em token} is the fundamental unit into which input data of a transformer is divided before being processed by the model.
For text, a token usually corresponds to a word, subword, or even a single character, depending on the chosen tokenization scheme. Similarly,  an image can be broken up into a set of patches which are then used as tokens.
Each token is mapped to a {\em vector embedding} before entering the transformer. The embeddings can be fixed or they can be learned together with the other parameters of the model.
Each token is mapped to a vector embedding (a technique pioneered by methods like {\em word2vec}~\cite{mikolov2013efficient}) before entering the transformer.

Let a sequence of $n$ $d$-dimensional tokens be given as
$$
x_j^{(0)}, \quad j=1,\dots, n, \quad x_j^{(0)} \in \mathbb{R}^d,
$$
where each $x_j^{(0)}$ is a vector embedding of the $j$-th token.

Since the core attention mechanism is invariant to the order of the input, and since the order of inputs matters for applications like NLP, 
the network needs to be explicitly supplied with information about the position of each token. This is achieved by adding a positional encoding vector $p_j \in \R^d$ to the input embedding $x_j$:
\begin{equation}\label{positionencoding}
    z_j^{(0)} = x_j^{(0)} + p_j, \qquad j=1,\dots,n.
\end{equation}
The original Transformer uses fixed sinusoidal functions for the positional encoding. An alternative, often used in modern LLMs, is to make the positional encodings learnable parameters.

The collection of inputs $z_j$ forms a matrix
$$
Z^{(0)} = \begin{bmatrix} z_1^{(0)}, \dots , z_j^{(0)} \end{bmatrix} \in \mathbb{R}^{d \times n}.
$$
The transformer will take as input the position-encoded token sequence matrix $Z^{(0)}$ and return a representation of the sequence in terms of another $d \times n$ matrix $Z^{(L)}$, where $L$ denotes the number of layers in the
transformer.

The output $Z^{(L)}$ is calculated by iteratively applying
a transformer block
$$
Z^{(\ell)} = \text{transformer-block}\,(Z^{(\ell-1)}), \qquad \ell=1,\dots,L.$$

The transformer block consists of two stages: one stage that operates along the sequence dimension and another that operates across the feature dimension.
The first stage contains the key innovation of transformers, the so-called self-attention mechanism. 
In this first stage, each feature is refined based on dependencies among tokens across the sequence---for instance, how a word at position $j$ depends on a previous word at positions $j'$, or how two image patches correspond to one another. This stage acts vertically across the columns of $Z^{(\ell-1)}$. In the second stage the features associated with each token are refined. This stage acts horizontally along the rows of $Z^{(\ell-1)}$. Through repeated application of the transformer block, the representation of token $j$ and feature $i$ becomes influenced by information from token $j'$ and feature $i'$. As Turner points out in~\cite{turner2023introduction}, the concept of interleaving processing across the sequence and across features is common among many machine learning architectures. For example, in
graph neural networks one interleaves processing
across nodes and across features.

\subsubsection{Stage 1: Self-attention across the sequence}

We will describe the first stage of the tranformer-block, namely the multi-head self-attention in more detail.

\smallskip
\noindent
\textbf{Attention.}
The output of the first stage of the transformer block in the $\ell$-iteration is a $d \times n$ matrix $Y^{(\ell)}$. The output is produced by aggregating information across the sequence
independently for each feature using an operation called {\em attention}. 
This is the fundamental operation of transformers, which proceeds as follows. The output vector at location $j$, denoted by $y^{(\ell)}_j$ is calculated
by a weighted average of the input features $z^{(\ell-1)}$ at location $j' = 1,\dots,n$ via
\begin{equation}\label{eq:trans1}
y_j^{(\ell)} = \sum_{j'=1}^n  z_{j'}^{(\ell-1)} A_{j',j}^{(\ell)}  
\end{equation}
Here the weighting is done by a so-called attention matrix $A_{j,j'}^{(\ell)}$ of size $n \times n$. The attention matrix is normalized such that $\sum_{j'=1}^n  A_{j',j}^{(\ell)} = 1$ for all $j=1,\dots,n$. 

The idea of the attention matrix is that $A_{j,j'}^{(\ell)}$ will take a large value for locations in the sequence $j'$ that are of high relevance for location $j$, and a small value in case there is little relevance between the two.

We can conveniently write this correspondence
as matrix multiplication
$$Y^{(\ell)} = Z^{(\ell-1)}  A^{(\ell)}.$$

\noindent
\textbf{Self-attention.}
But we have not yet explained how we decide which values to assign to the entries of the attention matrix $A_{j,j'}^{(m)}$. One of the key aspects of the transformer is that the entries of the attention matrix are determined via the input data themselves, i.e., attention is computed via {\em self-attention}.
We could do this by measuring the similarity between two locations by the inner product between the
features at those two locations, i.e., by considering $\langle z_{j'}^{(\ell)}, z_{j}^{(\ell)} \rangle$. And we could turn these inner products into probabilities by applying the softmax function defined in equation~\eqref{softmax}. 
The softmax operation ensures that attention weights are nonnegative and sum to one across each row.
So, shall we compute the attention matrix as
$$
A_{j',j}^{(\ell)} = \frac{\exp\big(\langle z_{j}^{(\ell)}, z_{j'}^{(\ell)} \rangle\big)}{\sum_{j''=1}^n \exp\big(\langle z_{j'}^{(\ell)}, z_{j''}^{(\ell)} \rangle\big)} \,\,\, ?
$$
Alas, this simple approach is doomed to fail, since it entangles information about the similarity between
locations in the sequence with the content of the sequence itself.

One possibility to
curtail this undesirable entanglement is to  apply  a linear transformation $W$ to the sequence $z_j^{(\ell)}$. That is, shall we consider
$$
A_{j',j}^{(\ell)} = \frac{\exp\big(\langle W z_{j}^{(\ell)}, W z_{j'}^{(\ell)} \rangle\big)}{\sum_{j''=1}^n \exp\big(\langle W z_{j'}^{(\ell)}, W z_{j''}^{(\ell)} \rangle\big)}\,\,\, ?
$$
Yet, this choice is still not quite suitable, at least not for natural language processing tasks. Indeed, the inherent symmetry of the expression $\exp\big(\langle W z_{n}^{(\ell)} , W z_{n'}^{(\ell)} \rangle\big)$ contrasts with the lack of symmetry we typically encounter in the relationship between words. For instance, the word {\em marsupial} is strongly related to the word {\em mammal}. In contrast, the word {\em mammal} is only weakly associated with {\em marsupial}, since there are many other mammals and many among those  (e.g., {\em lion}, {\em elephant}, {\em whale},...) are much more frequently associated with mammals.

It is easy to break this symmetry by introducing for instance two different linear transforms, let us call them $W_{k}$ and $W_q$. For convenience, we choose both matrices to be of dimension $d_k \times d$. Usually one chooses $d_k < d$ such that $W_k$ and $W_q$ project onto lower-dimensional subspaces, with the aspiration that only the most important features of $z_j^{(\ell)}$ prevail in the similarity computations. 
Moreover, for improved numerical stability we divide the matrices in the exponents by the square-root of the dimensionality of the projected vectors. 
With this asymmetric notion of similarity and normalization in place, we arrive at the desired formula for our attention matrix, namely
\begin{equation}
\label{attentionmatrix}   
A_{j',j}^{(\ell)} = \frac{\exp\big(\langle W_k z_{j}^{(\ell)}, W_q z_{j'}^{(\ell)} \rangle /\sqrt{d_k}\big)}{\sum_{j''=1}^n \exp\big(\langle W_k z_{j'}^{(\ell)}, W_q z_{j''}^{(\ell)} \rangle /\sqrt{d_k}\big)}.
\end{equation}
Equations~\eqref{eq:trans1} and~\eqref{attentionmatrix}
taken together represent what is referred to as the {\em self-attention mechanism}.  Owing to the fact that the entries of $A_{j',j}^{(\ell)}$ are non-negative and each column of $A$ sums up to 1 by construction, 
each output vector $y_j^{\ell}$ is formed by a convex combination of the input features $z_j^{\ell-1}$.

The two quantities $q_j = W_q z_j^{(\ell)}$ and $k_j = W_k z_j^{(\ell)}$ arising in the inner products of~\eqref{attentionmatrix} are typically known as the {\em queries} and the {\em keys}, respectively (where we have adopted the convention of suppressing the subscripts in the queries and the keys).

The reader who is used to seeing the attention mechanism  in the form
\begin{equation}
    \label{attention2}
\operatorname{Att}(Q, K, V) = \operatorname{softmax} \left( \frac{QK^\top}{\sqrt{d_k}} \right)V
\end{equation}
may rest assured that  we can easily recover expression~\eqref{attention2} from~\eqref{attentionmatrix}
and~\eqref{eq:trans1}.
To that end we first collect the queries $q_j$ and keys $k_j$ in matrices $Q = [W_q Z^{(\ell)}]^T \in \mathbb{R}^{n \times d_k}$ and $K =  [W_k Z^{(\ell)}]^T \in \mathbb{R}^{n \times d_k}$, respectively.
This already establishes that $\operatorname{softmax} \big( QK^\top/\sqrt{d_k} \big) = A^{(\ell)}$.

Why is the value-matrix $V$ (or $W_v$ for that matter) not present in~\eqref{eq:trans1} or in~\eqref{attentionmatrix}? The reason we skipped $V$ until now is that the original transformer architecture introduces another projection in the so-called multi-head self-attention (discussed in subsequent paragraphs), which as noted in~\cite{turner2023introduction}, creates unnecessary redundancy that we are avoiding here.

\smallskip

The transformer's self-attention can be viewed as a kernel smoother: after learned linear projections 
$Q$,$K$,$V$ the output at query $q$
is a normalized, similarity-weighted average of value vectors $z_j$.  This is mathematically analogous to  kernel regression estimation (cf.~Chapter~\ref{s:kernellearning} and also~\cite{nadaraya1964estimating,watson1964smooth}) with the attention score 
$\exp (\langle q, k \rangle)$ acting as the  kernel.
The difference is that in classical kernel regression, the kernel function, often chosen as $K(z,z') = \exp(\frac{-\|z-z'\|^2}{2\sigma^2})$, is fixed (up to hyperparameters like the kernel bandwidth $\sigma$). It defines a static notion of similarity between data points. By contrast, in a transformer’s attention, the (non-symmetric) kernel $K(z_i,z_j)=\exp(\langle q_i, k_j \rangle)=\exp(\langle W_q z_i , W_k z_j\rangle)$
is learned and adaptive. The similarity metric itself depends on trainable parameters $W_q, W_k$.
As training progresses, these parameters change to reflect task-specific or context-dependent notions of similarity.

\medskip
\noindent
\textbf{Multi-head self-attention.}
A single attention head must use one set of attention weights per token to encode all the different features and relationships. This forces it to average or blend various types of information, which limits its expressive power. This can be addressed by using multiple heads.

The {\em multi-head attention mechanism} is a refinement that allows the model to collectively attend to information from different representation subspaces at different positions. Instead of performing a single attention function, the query, key, and value matrices are linearly projected several times with different, learned parameter matrices.

Instead of a single attention operation, the transformer block employs attention in $H$ parallel 
sets, called {\em heads}, and then
linearly projects the output onto $d \times n$ arrays. This enables the model to capture dependencies across multiple representational subspaces simultaneously. The use of multiple heads in transformers is akin to the use of multiple channels in CNNs.
In detail, the multi-head self-attention operation is given by
\begin{equation}\label{multihead}
Y^{(\ell)} = \sum_{h=1}^H V_h^{(\ell)} Z^{(\ell)}A_h^{(\ell)} ,
\end{equation}
where the $H$  matrices $V_h^{(\ell)}$ are of dimension $d \times d$.
The standard multi-head presentation can be recovered
by simply setting $V_h$ to be a low-rank matrix
$V_h = W_h W_{v,h}$ where $W_h \in \R^{d \times d_v}$ and $W_{v,h} \in \R^{d_v \times d}$.
In this case, the matrices $V_h^{(\ell)}$ act like (non-orthogonal) projections onto a $d_v$-dimensional linear subspace.
The matrices $W_k^{(\ell)}, W_q^{(\ell)}$, $W_{v,h}^{(\ell)}$, and $W_v^{(\ell)}$ are  unknown parameters that need to be learned during the training.

\subsubsection{Stage 2: Multi-layer perceptron across features}

In the second stage of the tranformer block a {\em pointwise feed-forward network} is applied row-wise across features. This simple two-layer fully connected network  takes the form
   $$
   f_{\text{FFN}}(z) = \sigma(zW_1 + b_1) W_2 + b_2, \quad W_1 \in \mathbb{R}^{d \times d_{\text{ff}}},  W_2 \in \mathbb{R}^{d_{\text{ff}} \times d}.
   $$

\subsubsection{The transformer block}
A transformer layer combines multi-head attention with feed-forward transformations (together with two standard tricks of the trade, namely residual connections and layer normalization, to stabilize training). Stacking $L$ such layers yields the encoder (or decoder)  of a transformer. 

An encoder processes input sequence embeddings into contextual representations. A decoder generates outputs, using both self-attention and cross-attention, where queries attend over encoder outputs. More specifically, in cross-attention the Queries (Q) come from the target sequence (e.g., the sentence being generated in the Decoder), while the Keys (K) and Values (V) come from the source sequence (e.g., the sentence being translated in the encoder).
In self-attention the model asks, ``How do the words in this sentence relate to each other?''. 
In cross-attention the model asks, ``Which part of the input sentence is relevant to the word I am currently writing?''

\bigskip
The transition from high-dimensional internal hidden states to a discrete output (like words or tokens) occurs in the final output head. This stage is structured to map a continuous vector representation back into a probability distribution over the model's vocabulary.

After the final decoder block, each token position is represented by a hidden vector $h \in \mathbb{R}^d$, where $d$ is the model dimension (e.g., 512 or 768). To map this to the vocabulary, the model applies a linear transformation
$$u = h W_U + b,$$
where $W_U \in \mathbb{R}^{d \times V}$is the so-called unembedding matrix, $V$ is the vocabulary size (often 30,000 to 100,000+), and the 
$u \in \mathbb{R}^V$ are the {\em logits}. Each element $u_i$ represents the raw ``score'' for the $i$-th word in the dictionary. The logits are unconstrained real numbers. To interpret them as probabilities, the softmax function is applied
$$P(y = i | \text{context}) = \frac{\exp(u_i / \tau)}{\sum_{j=1}^{V} \exp(u_j / \tau)},$$
where the hyperparameter $\tau$ is called the temperature\footnote{The term is borrowed from thermodynamics, similarly to how it was used in Section~\ref{sec:MarkovChains}.
}, which controls the ``sharpness'' of the resulting probability distribution across every single unique token in the model's vocabulary. A low $\tau$ ($\tau < 1$) makes the model more deterministic, while a high $\tau$ ($\tau > 1$) increases randomness.

\bigskip
The transformer architecture underlies a number of well known models such as BERT~\cite{devlin2019bert} (encoder-only), GPT~\cite{radford2018improving}, Gemini~\cite{team2023gemini}, and Claude~\cite{anthropic2024claude3} (generally understood to be decoder-only) and of course the original Transformer~\cite{vaswani2017attention} (encoder–decoder). Regarding expressivity, it has been shown that transformers are universal approximators of sequence-to-sequence functions under mild assumptions~\cite{yun2019transformers}.

\medskip
While attention-based architectures have become ubiquitous in machine learning, our mathematical understanding of the reasons for their effectiveness remains limited.
There is no comprehensive theory yet describing how architectural elements such as depth, width, attention heads, and positional encodings jointly determine expressivity or enable emergent algorithmic structure.
Also, the quadratic scaling of the self-attention mechanism creates a significant computational bottleneck for long-sequence data.
Progress will likely require new tools from functional analysis, dynamical systems, information theory, and statistical learning theory, as well as closer links between theory and empirical behavior.

\subsection{Learning low-dimensional structure with autoencoders}

In previous chapters we encountered different approaches 
to dimension reduction: PCA finds 
the optimal linear subspace by minimizing the $\ell_2$-reconstruction error, the  Johnson-Lindenstrauss lemma guarantees nearly isometric embeddings in logarithmically  small dimensions, and diffusion maps reveal the intrinsic geometry of a data manifold  through the eigenfunctions of a Markov operator. Each of these methods imposes  structure on the embedding---linearity, isometry, or smoothness with respect to a 
diffusion process---and derives guarantees from that structure.

Autoencoders offer a complementary perspective based on neural networks. Rather than prescribing the form of the embedding analytically, they  \emph{learn} an encoder--decoder pair from data by minimizing a reconstruction  objective. The latent representation that emerges occupies the same conceptual space as the principal components or diffusion coordinates: a 
low-dimensional summary of high-dimensional data~\cite{hinton2006reducing,bengio2013representation}. What changes is the hypothesis class. By using nonlinear, multi-layer mappings, autoencoders can capture structure 
that lies beyond the reach of PCA.

\begin{definition}[Autoencoder]
An \emph{autoencoder} consists of an \emph{encoder} $f_\phi : \mathbb{R}^d \to 
\mathbb{R}^k$, where $k \ll d$, and a \emph{decoder} $g_\psi : \mathbb{R}^k \to \mathbb{R}^d$, each 
parameterized by weights $\phi$ and $\psi$ drawn from some neural network 
architecture. Given a dataset $\{x_1, \ldots, x_n\} \subset \mathbb{R}^d$, the 
parameters are found by minimizing the \emph{reconstruction loss}
\begin{equation}
    \mathcal{L}(\phi, \psi) 
    = \frac{1}{n} \sum_{i=1}^{n} \| x_i - g_\psi(f_\phi(x_i)) \|^2.
    \label{eq:ae_loss}
\end{equation}
The composition $\Pi = g_\psi \circ f_\phi : \mathbb{R}^d \to \mathbb{R}^d$ is 
called the \emph{reconstruction map}, and the image $f_\phi(x_i) \in \mathbb{R}^k$ 
is the \emph{latent code} or \emph{latent representation} of $x_i$.
\end{definition}

\begin{figure}[h]
\begin{center}
\includegraphics[width=0.6\textwidth]{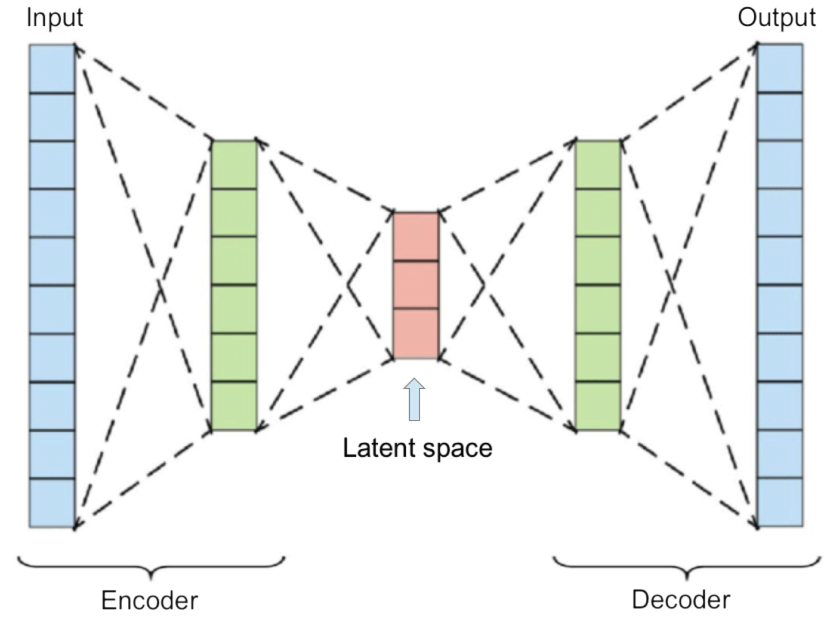}
\caption{Schematic depiction of the architecture of an autoencoder.}
\label{fig:autoencoder}
\end{center}
\end{figure}

The architecture of the encoder and decoder is flexible.  Typically, both are deep  networks with multiple hidden layers and pointwise nonlinearities. The bottleneck layer of dimension $k$ 
forces the network to compress information, producing a low-dimensional 
representation as a byproduct of learning to reconstruct, as illustrated in Figure~\ref{fig:autoencoder}.

It is instructive to consider {\em linear
autoencoders}. A \emph{linear autoencoder} constrains both $f_\phi$ and $g_\psi$ to be linear maps:
$$
    f_\phi(x) = Wx, \quad g_\psi(z) = Vz,
$$
where $W \in \mathbb{R}^{k \times d}$ and $V \in \mathbb{R}^{d \times k}$. The 
reconstruction loss becomes
\begin{equation}
    \mathcal{L}(W, V) = \frac{1}{n}\sum_{i=1}^n \|x_i - VWx_i\|^2 
    = \|X - XW^\top V^\top\|_F^2 / n,
    \label{eq:linear_ae_loss}
\end{equation}
where $X \in \mathbb{R}^{n \times d}$ is the data matrix with rows $x_i^\top$.
A straightforward calculations shows that the  global minimizers of this objective are  precisely the PCA subspaces~\cite{baldi1989neural}.

However, the power of autoencoders lies in their nonlinear generalizations. When $f_\phi$ and 
$g_\psi$ are nonlinear maps---realized as multi-layer networks---the autoencoder can 
in principle learn to exploit nonlinear structure in the data manifold.

The Universal Approximation Theorem  implies that 
sufficiently wide or deep encoders and decoders can approximate any continuous 
function on a compact set. This expressivity comes at a cost: unlike PCA, nonlinear 
autoencoders have non-convex loss landscapes, and gradient-based training finds only  local minima in general.

Vanilla autoencoders as described so far are purely discriminative: they learn a 
deterministic encoder--decoder pair without placing any structure on the latent 
space. Indeed, without regularization, the learned latent space may not have any useful  geometric structure: nearby points in $\mathbb{R}^d$ need not map to nearby codes in $\mathbb{R}^k$. Several structured variants address these issues. Arguably the most successful approach to date is the variational autonecoder that we will describe next.

\subsubsection{Variational autoencoders}

In a standard autoencoder, the encoder maps each data point 
to a single code, and there is no mechanism preventing the latent codes from  collapsing into isolated clusters with large empty regions between them. Decoding 
a point from such an empty region produces meaningless output. A so-called {\em variational autoencoder} (VAE) tries to prevent this by imposing a probabilistic model on 
the latent space, and thereby aims to enable both principled inference and generation of new samples~\cite{kingma2013auto,diederik2019introduction}.

A \emph{latent variable model} posits that each observed data point $x \in 
\mathbb{R}^d$ is generated by first sampling an unobserved \emph{latent variable} 
$z \in \mathbb{R}^k$ from a prior distribution $p(z)$, and then sampling $x$ 
from a conditional distribution $p_\psi(x | z)$ parameterized by $\psi$. 
The joint distribution is
$$
p_\psi(x, z) = p_\psi(x | z) \, p(z),
$$
and the marginal likelihood of an observation is
\begin{equation}
    p_\psi(x) = \int_{\mathbb{R}^k} p_\psi(x | z) \, p(z) \, dz.
    \label{eq:marginal_likelihood}
\end{equation}
The canonical choice is a standard Gaussian prior $p(z) = \mathcal{N}(0, I_k)$
and a Gaussian or Bernoulli likelihood $p_\psi(x | z)$ parameterized by a
neural network. The goal of learning is to find $\psi$ maximizing the
log-likelihood $\sum_{i=1}^n \log p_\psi(x_i)$.

The fundamental difficulty is that the integral \eqref{eq:marginal_likelihood}
is intractable for nonlinear $p_\psi(x | z)$; one cannot evaluate it in
closed form, and naive Monte Carlo estimation has prohibitively high variance
in high dimensions. The VAE resolves this by introducing a tractable
\emph{approximate posterior}.

We proceed as follows. For any distribution $q(z | x)$, Jensen's inequality applied to the concave
logarithm gives
\begin{align}
    \log p_\psi(x) 
    &= \log \int p_\psi(x | z) \, p(z) \, dz \notag \\
    &= \log \int \frac{p_\psi(x | z) \, p(z)}{q(z | x)} \, q(z | x) \, dz \notag \\
    &= \log \, \mathbb{E}_{q(z | x)}\!\left[\frac{p_\psi(x | z) \, p(z)}{q(z | x)}\right] \notag \\
    &\geq \mathbb{E}_{q(z | x)}\!\left[\log \frac{p_\psi(x | z) \, p(z)}{q(z | x)}\right] \notag \\
    &= \mathbb{E}_{q(z | x)}\bigl[\log p_\psi(x | z)\bigr] 
       - \text{KL}\bigl(q(z | x) \,\|\, p(z)\bigr).
    \label{eq:elbo_derivation}
\end{align}
The right-hand side of \eqref{eq:elbo_derivation} is called the \emph{evidence lower 
bound} (ELBO), also written $\mathcal{L}_{\mathrm{ELBO}}(x; \phi, \psi)$. 
The gap between the log-likelihood and the ELBO is exactly the KL divergence 
between the approximate and true posterior:
\begin{equation}
    \log p_\psi(x) - \mathrm{ELBO}(\phi, \psi; x) 
    = \text{KL}\bigl(q_\phi(z | x) \,\|\, p_\psi(z | x)\bigr) \geq 0.
    \label{eq:elbo_gap}
\end{equation}
This identity has two 
important consequences. First, the ELBO is indeed a lower bound on the 
log-likelihood for every choice of $q$. Second, the bound is tight (i.e., 
$\mathrm{ELBO} = \log p_\psi(x)$) if and only if $q_\phi(z | x) = 
p_\psi(z | x)$ almost surely, that is, when the approximate posterior 
equals the true posterior. Maximizing the ELBO therefore simultaneously pushes 
toward higher likelihood and toward a better approximation of the posterior. 

This is the key for the variational autoencoder, which parameterizes the approximate posterior as a
Gaussian with a diagonal covariance:
\begin{equation}
    q_\phi(z | x) = \mathcal{N}\bigl(\mu_\phi(x),\, 
    \mathrm{diag}(\sigma_\phi^2(x))\bigr),
    \label{eq:vae_encoder}
\end{equation}
where $\mu_\phi : \mathbb{R}^d \to \mathbb{R}^k$ and $\sigma_\phi^2 : 
\mathbb{R}^d \to \mathbb{R}^k_{>0}$ are neural networks with shared weights 
up to the final layer (the \emph{encoder}). The likelihood is parameterized as
\begin{equation}
    p_\psi(x | z) = \mathcal{N}\bigl(g_\psi(z),\, \tau^2 I_d\bigr),
    \label{eq:vae_decoder}
\end{equation}
where $g_\psi : \mathbb{R}^k \to \mathbb{R}^d$ is a neural network 
(the \emph{decoder}) and $\tau^2 > 0$ is a fixed or learned variance. The 
parameters $(\phi, \psi)$ are trained to maximize the ELBO summed over the 
dataset, i.e,
\begin{equation}
    \max_{\phi,\psi} \,\, \sum_{i=1}^n \underbrace{\Bigl[
        \mathbb{E}_{q_\phi(z | x_i)}\bigl[\log p_\psi(x_i | z)\bigr]
        - \text{KL}\bigl(q_\phi(z | x_i) \,\|\, p(z)\bigr)
    \Bigr]}_{\mathrm{ELBO}(\phi, \psi; x_i)}.
    \label{eq:vae_objective}
\end{equation}

The ELBO term in~\eqref{eq:vae_objective} decomposes into two competing terms,
each with a distinct interpretation:

\begin{enumerate}
    \item Reconstruction term: 
    $\mathbb{E}_{q_\phi(z | x_i)}[\log p_\psi(x_i | z)]$. 
    With the Gaussian likelihood \eqref{eq:vae_decoder}, this equals 
    $-\frac{1}{2\tau^2}\mathbb{E}[\|x_i - g_\psi(z)\|^2] + \mathrm{const}$, 
    so maximizing it encourages the decoder to reconstruct $x_i$ accurately 
    from codes $z$ sampled from the encoder distribution. This is the 
    probabilistic analog of the deterministic autoencoder loss.

    \item KL regularization term: 
    $-\text{KL}(q_\phi(z | x_i) \,\|\, p(z))$. 
    This penalizes the encoder for producing a posterior $q_\phi(z | x_i)$ 
    that deviates too far from the prior $\mathcal{N}(0, I_k)$. It prevents 
    the encoder from collapsing to a deterministic map (which would reduce 
    the VAE to a standard autoencoder) and regularizes the latent space to 
    be well-covered by the prior.
\end{enumerate}

Without the KL regularization term, the model could simply assign each input to a very distant narrow point in latent space to minimize reconstruction error, leading to a ``fractured''latent space. The KL term forces the clusters to overlap and stay centered, with the goal that
points close to each other in the latent space decode to visually or structurally similar outputs.

\bigskip

Autoencoders have found broad application across scientific and engineering domains, owing to their ability to learn compact, task-relevant representations without labeled data~\cite{balle2016end,vincent2010stacked,an2015variational,gomez2018automatic}.
The latent codes learned by an autoencoder can serve as features for downstream supervised tasks, especially in settings where labeled data is scarce~\cite{way2018extracting,broecker2025multimodal}.
Autoencoders now frequently appear as components of larger systems, rather than standalone dimension-reduction tools. Examples include privacy-preserving and federated learning pipelines, anomaly detection systems, and generative models. 

While in generative AI autoencoders have been outperformed by diffusion models in terms of quality of the generated data~\cite{song2020score,ho2020denoising}, 
latent diffusion models often rely on high-quality autoencoder-style compression as a preprocessing step, underscoring the continued relevance of learned low-dimensional representations even when the final objective is generative rather than descriptive~\cite{preechakul2022diffusion,rombach2022high}.

\section*{Exercises}
\addcontentsline{toc}{section}{Exercises}

\begin{myexercise}(Linear classification with the perceptron). The perceptron is a simple linear classifier for binary classification. Given a dataset
$$
{(x_i, y_i)}_{i=1}^n \subset \mathbb{R}^d \times {-1, +1},
$$
the perceptron algorithm iteratively updates a weight vector $w \in \mathbb{R}^d$ and bias $b \in \mathbb{R}$ as follows:
\begin{enumerate}
\item Initialize $w = 0, b = 0$.
\item For each misclassified point $x_i$ (i.e., $y_i(w^T x_i + b) \le 0)$ update:
$$   w \gets w + y_i x_i, \quad b \gets b + y_i $$
\end{enumerate}
(a) Prove that if the data are {\em linearly separable}, there exists a finite number of updates after which the perceptron algorithm finds a separating hyperplane.\\
(b) Let $R = \max_i |x_i|$ and $\gamma$ the margin of separation. Show that the number of updates is at most $(R/\gamma)^2$.
\end{myexercise}

\begin{myexercise}
Let $d \in \mathbb{N}$, and let
$$
f(x) = \sum_{i=1}^\infty c_i ,\sigma_{\mathrm{ReLU}}(\langle a_i, x\rangle + b_i),
\quad x \in \mathbb{R}^d,
$$
where
$$
\|a_i\| = 1, \quad |b_i| \le 1, \quad \text{and} \quad \|c\|_1 : < \infty.
$$
Show that for every $N \in \mathbb{N}$, there exists a one-hidden-layer ReLU neural network $f_N$ with $N$ neurons such that
$$
\|f - f_N\|_{L^2(B_1^d)} \le \frac{C \|c\|_1}{\sqrt{N}},
$$
where $C>0$ is an absolute constant. Conclude that infinite ReLU networks with $\ell^1$-summable coefficients admit dimension-independent approximation rates $O(N^{-1/2})$.
\end{myexercise}

\begin{myexercise} (Why the sigmoidal assumption in the UAT is important).
Let $\sigma:\mathbb{R}\to\mathbb{R}$ be a polynomial. 
\begin{enumerate}[label=(\alph*)]
\item Show that any function in 
$$\mathcal{N}_\sigma =
   \left\{
   x\mapsto \sum_{j=1}^m a_j \sigma(w_j\cdot x + b_j)
   \right\}
   $$
   is a polynomial in $x$.

\item Conclude that $\mathcal{N}_\sigma$ is not dense in $C([0,1]^n)$.

\item Explain why this shows that the nonpolynomial assumption in the classical UAT is not merely technical.
\end{enumerate}
\end{myexercise}

\begin{myexercise}
Assume $\sigma$ is a sigmoidal activation.
\begin{enumerate}[label=(\alph*)]
\item Suppose all weight vectors $w_j$ are constrained to satisfy $\|w_j\| = 1$. Is $\mathcal{N}_\sigma$ still dense in $C([0,1]^n)$?
\item Suppose all biases $b_j$ are set to zero. Is universality preserved?
\item Discuss which constraints destroy universality and which merely affect efficiency.
\end{enumerate}

\end{myexercise}

\begin{myexercise}(Universal approximation in other norms).
Let $1 \le p < \infty$. 
\begin{enumerate}[label=(\alph*)]
\item Show that if $\mathcal{N}_\sigma$ is dense in $C([0,1]^n)$, then it is dense in $L^p([0,1]^n)$.
\item Give an example of a function in $L^p$ that is not continuous but can still be approximated in $L^p$ by neural networks.
\end{enumerate}

\end{myexercise}

\begin{myexercise}
Backpropagation and the ``vanishing gradient'':
Consider a simple $L$-layer neural network where each layer has a single neuron. The output is $y = a_L$, where
$a_l = \sigma(z_l)$
$z_l = w_l a_{l-1} + b_l$
$\sigma(z) = \frac{1}{1 + e^{-z}}$ is the sigmoid activation.\\
(a) Using the chain rule, derive the expression for the gradient of the loss $\mathcal{L}$ with respect to the first weight $w_1$, i.e., $\frac{\partial \mathcal{L}}{\partial w_1}$.\\
(b) The derivative of the sigmoid function $\sigma'(z)$ is maximized at $z=0$. What is this maximum value? \\
(c) Explain why, as the number of layers $L$ increases, the gradient $\frac{\partial \mathcal{L}}{\partial w_1}$ tends toward zero if the weights are initialized near 1.
\end{myexercise}

\begin{myexercise}
Batch normalization transforms a mini-batch of activations $X = \{x_1, \dots, x_m\}$ to a new batch $\hat{X}$ by
$$\hat{x}_i = \frac{x_i - \mu_B}{\sqrt{\sigma_B^2 + \epsilon}}$$
(a) Prove that for any input batch $X$, the transformed batch $\hat{X}$ has a mean of 0 and a variance of 1.\\
(b) If we add a learnable scale $\gamma$ and shift $\beta$ such that $y_i = \gamma \hat{x}_i + \beta$, what happens to the representation if the network learns $\gamma = \sqrt{\sigma_B^2 + \epsilon}$ and $\beta = \mu_B$?
\end{myexercise}

\begin{myexercise}(The efficiency of depth).
This exercise is about an empirical comparison of the number of parameters required by a shallow network versus a deep network to approximate a high-frequency repeating pattern.

We want to approximate a nested sawtooth function $f(x)$ on the interval $[0, 1]$. A sawtooth function $S(x)$ is defined as
$$S(x) = \begin{cases} 2x & 0 \le x < 0.5 \\ 2(1-x) & 0.5 \le x \le 1 \end{cases}.$$

We are interested in the $k$-th composition, $f(x) = S(S(S(\dots S(x) \dots )))$, which creates $2^{k-1}$ ``teeth'' or oscillations. For this problem, we will target $k=4$, which results in 8 oscillations.

\begin{itemize}
\item  Shallow approximation:
According to the UAT, a single hidden layer can approximate this. Create a neural network with one hidden layer and ReLU activation. Find the minimum number of neurons $N$ required to achieve a Mean Squared Error (MSE) below $0.01$.
Calculate the total number of trainable parameters (weights + biases) for this network. Note: For a single hidden layer with $N$ neurons: total parameters $= (1 \times N) + N + (N \times 1) + 1 = 3N + 1$.
\item Deep approximation:
A remarkable property of ReLU units is that they can ``mirror'' the sawtooth shape perfectly through nesting.
Create a deep network where each hidden layer has only 2 neurons, but you have $L$ layers ($L$ should correspond to the number of compositions $k$).
Train this network on the same sawtooth data.
Calculate the total number of trainable parameters. Note: For $L$ layers of width 2: Total Parameters $\approx L \times (2^2 + 2) \approx 6L$.
\end{itemize}

Compare the mean squared error for these different network architectures and parameters.
\end{myexercise}

\begin{myexercise}(Another ``depth vs width'' exploration).
Consider the function
$$
f(x) = \sin(8\pi x), \qquad x \in [0,1].
$$

Approximate $f$ using two neural network architectures with ReLU activation:
\begin{enumerate}
\item[(a)] A shallow network: one hidden layer with $m$ neurons
\item[(b)] A deep network: $L$ hidden layers, each with a small fixed width.
\end{enumerate}

\smallskip
\noindent
Part (a): Shallow network approximation:

\vspace*{-2mm}
\begin{enumerate}
\item Construct a fully connected neural network with
one input neuron, one hidden layer of width $m$, 
ReLU activation, and one linear output neuron.

\item Train the network to approximate $f$ using mean squared error loss on a uniform grid of training points in $[0,1]$.

\item For $m \in \{5, 10, 20, 40, 80\}$, train the network until convergence, record the final training error, and plot the approximation $x \mapsto \hat f_m(x)$.
\end{enumerate}

\smallskip
\noindent
Part (b): Deep network approximation:

\vspace*{-2mm}
\begin{enumerate}
\item Construct a fully connected neural network with
one input neuron, $L$ hidden layers, each of width 4, ReLU activation, and one linear output neuron.

\item Train the network as above.

\item For $L \in \{2, 4, 8, 16\}$ train the network, record the final training error and plot the approximation $x \mapsto \hat f_L(x)$.
\end{enumerate}

For each trained model, compute the total number of trainable parameters.
Create a plot of approximation error vs number of parameters, with shallow and deep networks on the same axes.
Identify which architecture achieves a given target error (e.g. $10^{-3})$ with fewer parameters.

\end{myexercise}

\begin{myexercise}
Implement and compare the performance for a  network with three hidden layers using different activation functions: sigmoid, tanh, ReLU, and Leaky ReLU on MNIST digit classification. \\
(a) Plot the gradient norms at each layer during training for each activation function. Which exhibits the worst vanishing gradient problem? \\
(b) Compare final test accuracy and training time for each activation function.    
\end{myexercise}

\chapter{Large Sample Limit of Graph Laplacians}
\label{c:convergence}

We want to analyze the behavior of diffusion maps and other non-linear dimensionality reduction methods in the case where the data points $$x_1,x_2,\ldots,x_n \in \mathbb{R}^p$$ are i.i.d samples from a probability distribution over $\mathcal{M}$, a $d$-dimensional Riemannian manifold embedded in $\mathbb{R}^p$, i.e., $\operatorname{dim}M = d \ll p$. We call $\mathcal{M}$ the {\em intrinsic manifold} and $\mathbb{R}^p$ the {\em ambient space}. The weights are defined via a kernel function $K$ with a local scale $\sqrt{\varepsilon}$ as follows $$w_{ij} = K_{\varepsilon}(\|x_i - x_j \|) = K\left(\frac{\|x_i - x_j \|}{\sqrt{\varepsilon}}\right).$$  For example, weights for the Gaussian kernel are given by  $$w_{ij} = \exp{\left(-\frac{\|x_i - x_j\|^2}{2\varepsilon}\right)}.$$ We have previously discussed the random walk matrix $A = D^{-1}W$, the random walk Laplacian $L_{rw} = I - A$, the graph Laplacian $L = D - W$, and the normalized graph Laplacian $\LL = I - D^{-1}W$. We would like to understand the possible limits
\begin{eqnarray*}
D - W &\to& ? \\
I - D^{-1}W &\to & ?
\end{eqnarray*}
as $n\to \infty$ and $\varepsilon \to 0$, and perhaps we also need to specify the dependency of $\eps$ on $n$ (i.e., $\varepsilon = \varepsilon_n$) for the limit to exist and in what sense. We expect that the discrete random walk would converge to a continuous diffusion process over the manifold, but we have to be careful when taking limits. Specifically, we need to be mindful of how to choose $\varepsilon$ as a function of $n$ to guaranty convergence, understand the convergence rate, and different types of convergence. Intuitively, one might hope that the discrete graph Laplacian converges to the Laplace-Beltrami operator over the manifold. It turns out that the limiting operator also depends on the distribution of sampled point clouds over the manifold, and different normalizations of the graph Laplacian can yield different limiting operators. Understanding the graph Laplacian at the large sample limit is a powerful tool for gaining insight into the behavior of diffusion maps and related algorithms for nonlinear dimension reduction and the main topic of this chapter.

\section{Non-parametric density estimation} 
The problem of density estimation from a point cloud turns out to be intimately related to that of understanding large sample limits of graph Laplacians, and we therefore discuss it first.  

\subsection{Non-parametric density estimation in 1-D}

We first consider the problem of density estimation on the real line.
Let $p(x)$ be the probability density function (pdf) of the random variable $X$, and let $x_1,x_2,\ldots,x_n$ be i.i.d observations of $X$.
We employ the {\em Parzen Window} \cite{parzen1962estimation} for density estimation of the pdf $p(x)$ from the samples $\{x_i\}_{i=1}^n$.
We assume that $p(x)$ is uniformly continuous.
To build our estimator, consider a sum of bump functions for estimating $p(x)$, $p_{n,\varepsilon}(x)$:
\[
p_{n,\varepsilon}(x) = \frac{1}{n} \sum_{i=1}^n \frac{1}{\sqrt{\varepsilon}} K\left(\frac{x-x_i}{\sqrt{\varepsilon}} \right)
\]
It makes sense to choose $K$ that satisfies the following conditions:
\begin{itemize}
\item $K \geq 0$
\item $\int_\mathbb{R} K(x)\,dx = 1$
\item $K(x) = K(-x)$
\item $\lim_{|x|\to \infty} K(x)=0$
\item $K$ is monotonic decreasing for $x>0$ (optional but makes sense)
\item Bounded variation: $\int |K'(x)|\,dx < \infty$. If $K$ is monotonic, then $\int |K'(x)|\,dx = 2K(0)$.
\item $K$ is continuous (because $p(x)$ is continuous).
\end{itemize}
The estimator $p_{n,\eps}$ is nothing but a smoothed histogram.

In the limit $n\rightarrow \infty$, the Law of Large Numbers gives:
\begin{equation}
p_{n,\varepsilon}(x)
\rightarrow \mathbb{E}\left[\frac{1}{\sqrt{\eps}}K\left(\frac{x-X}{\sqrt{\eps}}\right)  \right]
= \frac{1}{\sqrt{\eps}} \int_\mathbb{R} K\left(\frac{x-y}{\sqrt{\eps}}\right)p(y)\,dy
= (K_\eps \ast p)(x),
\label{parzen}
\end{equation}
where $$K_\eps(x) = \displaystyle \frac{1}{\sqrt{\eps}} K \left(\displaystyle \frac{x}{\sqrt{\eps}}\right)$$
is a scaled kernel whose width is $\sqrt{\eps}$ and height is $1/\sqrt{\eps}$.
The estimator $p_{n,\eps}$ is a convolution of the kernel $K_\eps$ with the true density $p$ (i.e. blurring due to $\eps$). In order to reduce the blurring it is tempting to take
$\eps \rightarrow 0$. This corresponds to narrower bumps about the data (like a delta function), and we see with the above expression that our estimator satisfies $$\lim_{\eps \to 0}\lim_{n\to \infty} p_{n,\eps}(x) \rightarrow p(x).$$
In practice, we do not have an infinite number of samples, so taking $\eps$ too small can lead to significant errors. We move on to analyze the finite sample case.

For small $\eps>0$ we use a change of variables $z = (y-x)/\sqrt{\eps}$ on the integral in (\ref{parzen}), and then a Taylor expansion to get:

\begin{align}
& \frac{1}{\sqrt{\eps}}\int_\mathbb{R} K\left(\frac{x-y}{\sqrt{\eps}}\right)p(y)dy = \int_\mathbb{R} K(z)p(x+z\sqrt{\eps})\,dz\nonumber\\
& \qquad \qquad = \int_\mathbb{R} K(z)\left(p(x)+\sqrt{\eps}p^\prime(x) z + \frac{\eps}{2}p^{\prime \prime}(x) z^2 + O(\eps^{3/2})\right)\,dz\nonumber\\
& \qquad \qquad = p(x) + \eps \frac{m_2}{2} p^{\prime\prime} (x) + O(\eps^2), \label{eq:2}
\end{align}
where $$m_2 = \int K(z)z^2\,dz.$$
In (\ref{eq:2}) the $\sqrt{\eps}p^\prime(x) z$ term drops out because $z$ is odd while $K$ is an even function of $z$. Similarly, all integrals involving odd powers of $z$ vanish.
Thus, for $\eps>0$ our estimator does not converge to $p(x)$ as $n\to \infty$. The estimator is biased, with 
$$\text{Bias} = p(x) - \mathbb{E}[p_{n,\eps}(x)] = O(\eps).$$

Recall the bias-variance decomposition of the mean squared error derived in Chapter~\ref{ss:predictionrisk}.
Now we turn our attention to the variance error of $p_{n,\eps}$ for finite $n$:
\begin{eqnarray}
\text{Var }[p_{n,\eps}(x)] &=& \frac{1}{n} \text{Var}\left[\frac{1}{\sqrt{\eps}} K\left( \frac{x-y}{\sqrt{\eps}}\right) \right] \nonumber\\
&=& \frac{1}{n} \left[ \frac{1}{\eps} \int_\mathbb{R} K^2\left(\frac{x-y}{\sqrt{\eps}}\right)p(y)\,dy - \left(\frac{1}{\sqrt{\eps}} \int_\mathbb{R} K\left(\frac{x-y}{\sqrt{\eps}}\right)p(y)\,dy\right)^2\right] \nonumber \\
&=& \frac{1}{n} \left[ \frac{1}{\eps} \int_\mathbb{R} K^2\left(\frac{x-y}{\sqrt{\eps}}\right)p(y)\,dy - O(1) \right] \label{var2}\\
&=& \frac{1}{n} \left[ \frac{1}{\sqrt{\eps}}\int_\mathbb{R} K^2(z)p(x+\sqrt{\eps}z)\,dz + O(1) \right] \nonumber\\
&=& \frac{1}{n} \left[ \frac{1}{\sqrt{\eps}}\int_\mathbb{R} K^2(z)\,dz \;p(x) + O(1) \right] \nonumber\\
&=:& \frac{1}{n\sqrt{\eps}} \left[M_2 p(x) + O(\sqrt{\eps})\right]\label{var5}
\end{eqnarray}
where for (\ref{var2}), as $\eps \rightarrow 0$, we showed that it approaches $p(x)$, so it is $O(1)$.
In (\ref{var5}) we define
$$M_2 = \int_\mathbb{R} K^2(z)\,dz.$$
Overall,
\begin{equation}
\label{eq:density_estimation_MSE}
p_{n,\eps}(x) = p(x) + \eps \frac{m_2}{2} p^{\prime\prime} (x) + \xi_{n,\eps} + O(\eps^2),
\end{equation}
where
$$
\mathbb{E}[\xi_{n,\eps}] = 0 \,\,\text{ and }\,\, \text{Var}[\xi_{n,\eps}] = \frac{M_2 p(x)}{n\sqrt{\eps}}\left[1 + O(\sqrt{\eps})\right].
$$

We see as we try to take $\eps \rightarrow 0$, that the standard deviation $\sigma(\xi_{n,\eps}) = \sqrt{\text{Var}[\xi_{n,\eps}]} \rightarrow \infty$, so we must pick some optimal choice of $\eps = \eps_n$ to balance things out. The MMSE criterion
$$\min_{\eps} \text{Bias}^2 + \text{Var} = \min_{\eps} O(\eps^2) + O(\frac{1}{n\eps^{1/2}})$$
gives
$$\eps_n = n^{-2/5}.$$
The decay of $\eps_n$ to 0 is rather slow: the bandwidth is $\sqrt{\eps_n} = n^{-1/5}$. For this choice, the MSE decays as $n^{-4/5}$ instead of the usual $n^{-1}$ decay in parametric estimation. The decay of the MSE to 0 means that we get pointwise convergence: for every $x$, as $n\to \infty$, $p_{n,\eps}(x)\to p(x)$ with probability 1.

\subsection{Non-parametric density estimation in higher dimensions}
Suppose $x_1,\ldots,x_n$ are i.i.d samples of a random variable $X$ in $\mathbb{R}^d$.
We define the density estimator as
$$p_{n,\eps}(x) = \frac{1}{n} \sum_{i=1}^n \frac{1}{\eps^{d/2}}K\left(\frac{x-x_i}{\sqrt{\eps}}\right)$$
If
$$\int_{\mathbb{R}^d} K(x)\,dx = 1$$
then
$$K_{\eps}(x) = \frac{1}{\eps^{d/2}}K(\frac{x}{\sqrt{\eps}})$$
satisfies
$$\int_{\mathbb{R}^d} K_{\eps}(x)\,dx = 1$$
We also assume that the kernel function is radially symmetric, that is, it is only a function of the distance:
$$K(x) = k(\|x\|)$$

\subsection{Bias term and the Laplacian}
As before, we calculate the expectation using the change of variables $z = \frac{x-y}{\sqrt{\eps}}$
\begin{eqnarray*}
\mathbb{E}[p_{n,\eps}(x)] &=& \frac{1}{\eps^{d/2}}\int_{\mathbb{R}^d} K\left(\frac{x-y}{\sqrt{\eps}}\right)p(y)\,dy \\
&=& \int_{\mathbb{R}^d} K(z)p(x+\sqrt{\eps}z)\,dz \\
&=& \int_{\mathbb{R}^d} K(z)\left[p(x) + \sqrt{\eps}\sum_{i=1}^n \frac{\partial p(x)}{\partial x_i} z_i + \frac{\eps}{2}\sum_{i,j=1}^n \frac{\partial^2 p(x)}{\partial x_i \partial x_j} z_i z_j + \cdots \right]\,dz \\
&=& p(x) + \eps  \frac{m_2}{2} \sum_{i=1}^n \frac{\partial^2 p}{\partial x_i^2}(x) + O(\eps^2) \\
&=& p(x) + \eps  \frac{m_2}{2} \Delta p(x) + O(\eps^2)
\end{eqnarray*}
where
$$m_2 = \int_{\mathbb{R}^d} K(z)z_i^2\,dz = \frac{1}{d} \int_{\mathbb{R}^d} K(z)\|z\|^2\,dz.$$
Notice that integrals involving mixed terms $z_iz_j$ for $i\neq j$ vanish due to the radial symmetry of $K$. We see that the Laplacian emerges in a natural way. Let us explain the emergence of the Laplacian from a different viewpoint. Suppose $K$ is the Gaussian kernel
$$K(x) = \frac{1}{(2\pi)^{d/2}}e^{-\frac{\|x\|^2}{2}}.$$
For $\eps = 2t$ we have
$$K_{2t}(x) = \frac{1}{(4\pi t)^{d/2}}e^{-\frac{\|x\|^2}{4t}}$$
which is the heat kernel in $\mathbb{R}^d$. The convolution
$$K_{2t}\ast p(x)$$
is simply the solution of the heat equation
$$u_t(t,x) = \Delta u(t,x),\quad x\in \mathbb{R}^d, \quad t > 0,$$
with the initial condition $$u(0,x) = p(x).$$
In other words, the Gaussian kernel $K_{2t}(x)$ is the Green function for the heat equation. 
Formally, the solution is
$$K_{2t}\ast p = u(t,x) = e^{t\Delta}p = (I + t\Delta + \cdots)p(x) = p(x) + t\Delta p(x) + O(t^2).$$
This shows that for the Gaussian kernel
$$K_{\eps}\ast p = p(x) + \frac{\eps}{2}\Delta p(x) + O(\eps^2),$$
so that $m_2 = 1$ (which we already know, since the variance of the standard Gaussian is 1). The Laplacian appears in the bias for all kernels: the only difference is the coefficient $m_2$.
\subsection{Variance term and optimal choice of $\eps_n$}
Now, let us calculate the variance. The calculation is similar to the one dimensional case: 
\begin{eqnarray*}
\text{Var }[p_{n,\eps}(x)] &=& \frac{1}{n} \text{Var}\left[\frac{1}{\eps^{d/2}} K\left( \frac{x-y}{\sqrt{\eps}}\right) \right] \nonumber\\
&=& \frac{1}{n} \left[ \frac{1}{\eps^d} \int_{\mathbb{R}^d} K^2\left(\frac{x-y}{\sqrt{\eps}}\right)p(y)\,dy - \left(\frac{1}{\eps^{d/2}} \int_{\mathbb{R}^d} K\left(\frac{x-y}{\sqrt{\eps}}\right)p(y)\,dy\right)^2\right] \nonumber \\
&=& \frac{1}{n} \left[ \frac{1}{\eps^d} \int_{\mathbb{R}^d} K^2\left(\frac{x-y}{\sqrt{\eps}}\right)p(y)\,dy - O(1) \right] \\
&=& \frac{1}{n} \left[ \frac{1}{\eps^{d/2}}\int_{\mathbb{R}^d} K^2(z)p(x+\sqrt{\eps}z)\,dz + O(1) \right] \nonumber\\
&=& \frac{1}{n} \left[ \frac{1}{\eps^{d/2}}\int_{\mathbb{R}^d} K^2(z)\,dz \;p(x) + O(1) \right] \nonumber\\
&=& \frac{1}{n\eps^{d/2}} \left[M_2 p(x) + O(\eps)\right]
\end{eqnarray*}
where
$$M_2 = \int_{\mathbb{R}^d} K^2(z)\,dz.$$
Minimizing the mean squared error
$$\text{MMSE} = \min_{\eps} \eps^2 + \frac{1}{n\eps^{d/2}}$$
gives
$$\eps = \frac{C_d}{n\eps^{d/2+1}}$$
or
$$\eps^{d/2 + 2} = C_d n^{-1}$$
so
$$\eps_n = O\left(n^{-\frac{1}{\frac{d}{2}+2}}\right).$$
The MSE is
$$\text{MSE} = O\left(n^{-\frac{1}{\frac{d}{4}+1}}\right).$$
This is another form of the curse of dimensionality. The number of sample points needed for a given approximation error (MSE) increases exponentially with the dimension $d$:
$$n = O\left(\frac{1}{\text{MSE}^{d/4+1}}\right).$$

\section{Pointwise convergence of graph Laplacians}
We now turn our attention back from density estimation to the problem of convergence of graph Laplacians in the large sample limit. Note that there are different notions of convergence, and here we choose to focus on pointwise convergence, which although weaker than spectral convergence of eigenvalues and eigenvectors, it does not require too advanced analysis yet highlights the main concepts. The convergence of graph Laplacians is covered in numerous papers and remains a topic of active research \cite{belkin2005towards,hein2005graphs,singer2006graph,belkin2006convergence,gine2006empirical,singer2017spectral,trillos2018variational,garcia2020error,calder2022improved}.

\subsection{Laplacians in $\mathbb{R}^d$}

Suppose $x_1,x_2,\ldots,x_n$ are i.i.d samples from a distribution over $\mathbb{R}^d$ with pdf $p(x)$. We construct an $n\times n$ weight matrix $W$ with
$$w_{ij} = \frac{1}{n}K_\eps(x_i-x_j)$$
The graph Laplacian is the $n\times n$ matrix $$L = D-W$$ and the random walk Laplacian is
$$L_{rw} = I - D^{-1}W$$
We view all matrices as operators from $\mathbb{R}^{|V|} \to \mathbb{R}^{|V|}$. That is, we can apply the matrices to vectors of length $n=|V|$ that are functions over the data. In other words, we regard the vector $$f=(f_1,f_2,\ldots,f_n)\in \mathbb{R}^{|V|}$$ as $$f = (f(x_1),f(x_2),\ldots,f(x_n))$$ with $f_i = f(x_i)$. We abuse notation and think of the vector $f$ as $n$ samples of the function $f : \mathbb{R}^d \to \mathbb{R}$, where the exact usage should be clear from the context. We regard the function $f$ as being deterministic. That is, given $x$ the value $f(x)$ is completely known. Furthermore, we assume that $f$ is a nice function in the sense that $f\in C^4(\mathbb{R}^d)$.

\subsection{Limiting operator of $L = D-W$}
The entries of the diagonal matrix $D$ are given by
$$D_{ii} = \sum_{j=1}^n w_{ij} = \frac{1}{n}\sum_{j=1}^n K_{\eps}(x_i - x_j)$$
The previous discussion about kernel density estimators shows that for a fixed value of $\eps$,
$$\mathbb{E}[D_{ii}] = p(x_i) + \eps\frac{m_2}{2}\Delta p(x_i) + O(\eps^2)$$
Since $f$ is deterministic, we also have
$$\mathbb{E}[(Df)(i)] = \mathbb{E}[D_{ii}f(x_i)] = f(x_i)p(x_i) + \eps\frac{m_2}{2}f(x_i)\Delta p(x_i) + O(\eps^2)$$
We remark that the term corresponding to $w_{ii}$ needs to be excluded from the sum, since it is deterministic. However, its contribution is only $\frac{K(0)}{n}$ and tends to 0 as $n\to \infty$.
Now, let us consider $Wf$:
$$(Wf)(i) = \frac{1}{n}\sum_{j=1}^n K_{\eps}(x_i-x_j)f(x_j)$$
and its expected value is given by
\begin{eqnarray*}
\mathbb{E}[(Wf)(i)] &=& \int_{\mathbb{R}^d} K_{\eps}(x_i-y)f(y)p(y)\,dy \\
&=& f(x_i)p(x_i) + \eps\frac{m_2}{2}\Delta(fp)(x_i) + O(\eps^2).
\end{eqnarray*}
It follows that
\begin{eqnarray*}
\mathbb{E}[(Lf)(i)] &=& \mathbb{E}[(D-W)f(i)] \\
&=& [p(x_i) + \eps\frac{m_2}{2}\Delta p(x_i)]f(x_i) - [f(x_i)p(x_i) + \eps\frac{m_2}{2}\Delta(fp)(x_i)] + O(\eps^2) \\
&=& \eps\frac{m_2}{2} [f\Delta p - \Delta(fp)](x_i) + O(\eps^2) \\
&=& \eps\frac{m_2}{2} [f\Delta p - f\Delta p - 2\nabla f \cdot \nabla p - p\Delta f](x_i) + O(\eps^2)\\
&=& -\eps\frac{m_2}{2} [p\Delta f + 2\nabla p \cdot \nabla f](x_i) + O(\eps^2)
\end{eqnarray*}
Therefore,
\begin{eqnarray*}
-\frac{1}{\eps}(Lf)(i) &\to& \frac{m_2}{2} [p\Delta f + 2\nabla p \cdot \nabla f](x_i) + O(\eps) \\
&=& \frac{m_2 p(x_i)}{2} [\Delta f + 2\frac{\nabla p}{p} \cdot \nabla f](x_i) + O(\eps)
\end{eqnarray*}
almost surely as $n\to \infty$. For the uniform distribution, $p(x)=const$ the gradient vanishes $\nabla p = 0$ and we end up with the Laplacian. However, for non-uniform distributions, we get a different operator with diffusion constant and drift that both depend on $p$. We conclude that
$$\lim_{\eps\to 0}\lim_{n\to \infty} -\frac{1}{\eps}(Lf)(i) = \frac{m_2 p(x_i)}{2} [\Delta f + 2\frac{\nabla p}{p} \cdot \nabla f](x_i)$$

\subsection{Limiting operator of $L_{rw}=I - D^{-1}W$}
Let us consider the limiting operator for the random walk Laplacian $$L_{rw} = I - A = I - D^{-1}W$$ We have
$$(L_{rw}f)(i) = f(x_i) - \frac{\sum_{j=1}^n w_{ij}f(x_j)}{\sum_{j=1}^n w_{ij}}$$
We cannot simply conclude that the expected value $\mathbb{E}[(L_{rw}f)(i)]$ is the ratio of expectations of the numerator and of the denominator since they are not independent. Instead, we use the Law of Large Numbers to conclude that as $n\to \infty$
\begin{align*}
& (L_{rw}f)(i) \to f(x_i) - \frac{f(x_i)p(x_i) + \eps\frac{m_2}{2}\Delta(fp)(x_i) + O(\eps^2)}{p(x_i) + \eps\frac{m_2}{2}\Delta p(x_i) + O(\eps^2)} \\
& \quad = f(x_i) - \frac{f(x_i)p(x_i) + \eps\frac{m_2}{2}f(x_i)\Delta p(x_i) + \eps\frac{m_2}{2}\left(p \Delta f + 2\nabla p \cdot \nabla f \right) + O(\eps^2)}{p(x_i) + \eps\frac{m_2}{2}\Delta p(x_i) + O(\eps^2)} \\
& \quad = -\eps\frac{m_2}{2}\left(\Delta f + 2\frac{\nabla p}{p}\cdot \nabla f \right)(x_i) + O(\eps^2).
\end{align*}
For the uniform distribution ($p=const$) we get that the random walk Laplacian converges to the Laplacian. In the non-uniform case, we get convergence to the backward Fokker-Planck operator \cite{nadler2006diffusion}: Define a potential function $$U(x) = -2\log p(x),$$ then $$\nabla U = -2\frac{\nabla p}{p}$$ and we have
$$-\frac{1}{\eps}(L_{rw}f)(i) \to \frac{m_2}{2}\left(\Delta f - \nabla U \cdot \nabla f\right)(x_i) + O(\eps)$$
For example, if the samples come from a Gaussian distribution, i.e., $$p(x) = \frac{1}{(2\pi)^{d/2} (\det \Sigma)^{1/2}} e^{-\frac{1}{2}x^T \Sigma^{-1} x}$$ then $$U(x)=x^T\Sigma^{-1}x + const$$ is a parabolic potential, and the diffusive particle sees a linear drift towards the origin with a slope depending on the location.

\subsection{The effect of curvature}
Suppose that the data points are sampled from a regular $C^3$ curve embedded in $\mathbb{R}^p$ parameterized as $\gamma(s)$, where $s$ is the natural arc-length parameterization, satisfying
$$\|\gamma'(s)\|=1$$
Differentiating $\|\gamma'(s)\|^2=1$ gives
$$\gamma'(s)\cdot \gamma''(s) = 0$$
that is, $\gamma''(s)$ is perpendicular to $\gamma'(s)$. Taylor expansion near $s=0$ gives
$$\gamma(s) = \gamma(0) + \gamma'(0)s + \frac{1}{2}\gamma''(0)s^2 + O(s^3)$$
We can set a local coordinate system with the origin at $\gamma(0)$, the $x$-axis in the direction of $\gamma'(0)$, and the $y$-axis in the direction of $\gamma''(0)$ (if $\gamma''(0)=0$ then just pick the $y$-axis to be in any direction orthogonal to the $x$-axis). In the local coordinate system
$$\gamma(s) = (s, \frac{1}{2}as^2, 0, 0, \ldots, 0) + O(s^3)$$
where $a = \|\gamma''(0)\|$ is the curvature. This just means that locally every regular curve looks like a planar parabola. We also need information about the third derivative. Differentiating $\gamma'(s)\cdot \gamma''(s) = 0$ gives
$$\|\gamma''\|^2 + \gamma' \cdot \gamma''' = 0$$
so
$$\gamma'(0) \cdot \gamma'''(0) = -a^2$$
This gives a slightly refined Taylor expansion
$$\gamma(s) = (s - \frac{1}{6}a^2 s^3, \frac{1}{2}as^2, 0, 0, \ldots, 0) + (o(s^3),O(s^3),O(s^3),\ldots,O(s^3))$$

In practice, however, we can only measure Euclidean distances in $\mathbb{R}^p$. We have no knowledge of the curve and we cannot measure geodesic distances.
Luckily, the Euclidean distance is a very good approximation of the geodesic distance for short distances. Specifically, for the chordal distance:
$$\|\gamma(s)\|^2 = s^2 - \frac{1}{3}a^2 s^4 + \frac{1}{4}a^2 s^4 + o(s^4) = s^2(1 - \frac{1}{12}a^2 s^2 + o(s^2))$$
hence
\begin{eqnarray}
\|\gamma(s)\| &=& s\sqrt{1 - \frac{1}{12}a^2 s^2 + o(s^2)} = s(1 - \frac{1}{24}a^2 s^2 + o(s^2)) \nonumber \\
&=& s-\frac{1}{24}a^2 s^3+o(s^3) \label{eq:u}
\end{eqnarray}
so we see that the chordal and geodesic distances are the same up to both first and second order.
Let $u=\|\gamma(s)\|$ be the chordal distance, then
$$u = s - \frac{1}{24}a^2 s^3+o(s^3)$$
and
$$s = u + \frac{1}{24}a^2 u^3+o(u^3)$$
so
$$\frac{ds}{du} = 1 + \frac{1}{8}a^2 u^2 + o(u^2).$$

Suppose that the distribution of samples along our path has density $p(s)$, and $x_i=0$. The expectation of $D_{ii} = \sum_{j=1}^n w_{ij}$ is (we neglect the contribution of the term $j=i$ which is $O(n^{-1})$) 
\begin{eqnarray*}
\mathbb{E}\left[D_{ii}\right] &=& \mathbb{E}\left[\sum_{j=1}^n w_{ij}\right] \\
&=& \mathbb{E}\left[K_\eps(\|x_i-X\|)\right] \\
&=& \int \frac{1}{\sqrt{\eps}} K\left(\frac{\|\gamma(s)\|}{\sqrt{\eps}} \right)p(s)\,ds\\
 &=& \int \frac{1}{\sqrt{\eps}} K\left(\frac{u}{\sqrt{\eps}}\right)p(u+O(u^3))(1+\frac{a^2u^2}{8} + o(u^2))\,du  \\
&=& \int K(z) p(\sqrt{\eps}z + O(\eps^{3/2}))(1+\frac{\eps}{8}a^2z^2+o(\eps))\, dz\\
&=& \int K(z) ( p(0)+\sqrt{\eps}zp^\prime (0) + \frac{\eps}{2}p^{\prime \prime}(0)z^2 + O(\eps^{3/2})   )(1+\frac{\eps}{8}a^2z^2+o(\eps))\, dz \nonumber \\
&=& p(0) + \eps \frac{m_2}{2}\left[p^{\prime \prime}(0) + \frac{a^2}{4}p(0) \right] + o(\eps)
\end{eqnarray*}
In the calculation of the integral we substituted $u =\|\gamma(s)\|$, and used the change of coordinates $\frac{ds}{du} = 1 + \frac{1}{8}a^2u^2+ o(u^2)$, and then
substituted $z = u/\sqrt{\eps}$, rewrote $m_2 := \int K(z)z^2\,dz$, and used the fact that $\int K(z)dz=1$, and its symmetry $K(z)=K(-z)$. If the curve is $C^4$ then the $o(\eps)$ can be replaced by $O(\eps^2)$.

It follows that
$$\mathbb{E}[Df] = fp + \eps \frac{m_2}{2}\left[fp'' + \frac{a^2}{4}fp \right] + o(\eps)$$
and
$$\mathbb{E}[Wf] = fp + \eps \frac{m_2}{2}\left[(fp)'' + \frac{a^2}{4}fp \right] + o(\eps)$$
Therefore
\begin{eqnarray*}
\mathbb{E}[Lf] &=& \mathbb{E}[(D-W)f] \\
&=&  \eps \frac{m_2}{2}\left[fp^{\prime \prime} - (fp)''  \right] + o(\eps) \\
&=& -\eps \frac{m_2}{2}\left[f''p + 2f'p'   \right] + o(\eps)
\end{eqnarray*}
While curvature affects both $D$ and $W$, luckily the effect cancels out for the Laplacian! This means that it does not matter how the curve is embedded in $\mathbb{R}^p$, you can bend and twist it, yet $L$ remains the same (to leading order). In that sense, $L$ depends only on the intrinsic geometry of the data.

A similar cancellation of the curvature effect also applies to the random walk Laplacian:
\begin{eqnarray*}
L_{rw}f &\to& f - \frac{fp + \eps \frac{m_2}{2}\left[(fp)'' + \frac{a^2}{4}fp \right] + o(\eps)}{p + \eps \frac{m_2}{2}\left[p'' + \frac{a^2}{4}p \right] + o(\eps)} \\
&=& f - \frac{fp + \eps \frac{m_2}{2}\left[fp'' + \frac{a^2}{4}fp + pf'' + 2p'f'  \right] + o(\eps)}{p + \eps \frac{m_2}{2}\left[p'' + \frac{a^2}{4}p \right] + o(\eps)} \\
&=& -\eps\frac{m_2}{2}\left[f'' + 2\frac{p'}{p}f' \right].
\end{eqnarray*}
Thus, even though we have curvature, our normalization brings us back to the Fokker-Planck operator.

\subsection{General manifold}
What happens on a general manifold?
In 1-d we saw that the effect of curvature can be summarized as $$p'' \to p'' + \frac{a^2}{4} p.$$
In general, through any point on a $d$-dimensional manifold we can find an orthogonal system of geodesics.
Suppose we can take through a point $x$ a direction $x_1$, and an orthogonal direction $x_2$, and another $x_3$, etc.
We have a system $s_1, s_2, \ldots, s_d$.

The Laplace Beltrami operator takes the form
$$
\Delta_M f = \sum_{i=1}^d \frac{\partial^2 f}{ \partial s_i^2}.
$$
In the case $\operatorname{dim}(\mathcal{M}) = d$, every direction may have different curvature $a_1(x), a_2(x), \ldots, a_d(x)$, and thus there will be a correction term in the Laplacian.
In 1D, the correction was $\frac{a^2}{4}p$.
Similarly, it can be shown that in $d$ dimensions the correction term $E(x)$ in $$\Delta_M p(x) + \frac{1}{4}E(x)p(x)$$ is
$$
E(x) = \sum_{i=1}^da_i^2(x) - \sum_{i=1}^d\sum_{j\neq i}a_i(x)a_j(x).
$$

Still, the curvature term $E(x)$ drops in both $L$ and $L_{rw}$ and we have
$$\lim_{\eps\to 0} \lim_{n\to \infty} -\frac{1}{\eps}Lf = \frac{m_2}{2} \left[p \Delta_M f + 2\nabla p \cdot \nabla f \right]$$
and
$$\lim_{\eps\to 0} \lim_{n\to \infty} -\frac{1}{\eps}L_{rw}f = \frac{m_2}{2} \left[\Delta_M f + 2\frac{\nabla p}{p} \cdot \nabla f \right].$$

\subsection{Backward and Forward Fokker-Planck operators and the Neumann boundary condition}

Again, if we define a potential $U(x) = -2\log p(x)$ then the random walk Laplacian converges to the backward Fokker-Planck operator
$$\mathcal{L}f = \Delta_M f - \nabla U \cdot \nabla f.$$
The backward Fokker-Planck operator plays a significant role in the analysis of stochastic ordinary differential equations (SDEs) and It\^o diffusion processes. Similar to the Laplacian that appears in the heat/diffusion equation describing the evolution of the probability density function of the Brownian motion, the time evolution of the probability density function of a Brownian motion in a potential well $U(x)$ (i.e., Brownian motion + drift)  is governed by the forward Fokker-Planck equation.

The Laplace-Beltrami operator is a self-adjoint operator. Indeed, if we integrate over a manifold with no boundary, we see that
\[
\int_\mathcal{M} f \Delta_M g = \int_\mathcal{M} g \Delta_M f
\]
i.e. $\langle f, \Delta g \rangle = \langle \Delta f, g \rangle$.
This fact that the Laplacian is self adjoint is left as an exercise.\\

The adjoint $\mathcal{L}^*$ of the backward Fokker-Planck operator, known as the forward Fokker-Plank operator, is given by
\[
\mathcal{L}^*g = \Delta_M g + \nabla ( g \nabla U ).
\]

The forward operator can be written as divergence of flux:
$$\mathcal{L}^* g = \nabla \cdot \left[\nabla g + g\nabla U \right] = -\nabla \cdot J$$
where
$$-J = \nabla g + g\nabla U.$$
Here, the first term is flux due to concentration gradient and the second term is due to the drift $F = -\nabla U$.

Intuitively, the discrete random walk on the graph of data points cannot get absorbed and the total probability is conserved. As the random walk reaches a boundary point it has no choice but to be reflected back into another data point on the manifold. We therefore have a reflection boundary condition, or no-flux boundary condition given by $J=0$. Let us derive the corresponding boundary condition for $f$ and show that $\mathcal{L}^*$ is indeed the adjoint of $\mathcal{L}$.
\begin{eqnarray*}
\int_{\mathcal{M}} f \mathcal{L}^* g \,dv &=& -\int_{\mathcal{M}} f \nabla \cdot J \,dv\\
&=& -\int_{\mathcal{M}} \nabla \cdot [fJ] \,dv + \int_{\mathcal{M}} J \cdot \nabla f \,dv \\
&=& -\int_{\partial \mathcal{M}} fJ\,dS + \int_{\mathcal{M}} J\cdot \nabla f \,dv ~~~~~~~~~\text{no flux b.c.:}\, J=0\\
&=& -\int_{\mathcal{M}} \left[\nabla g + g \nabla U \right] \cdot \nabla f \,dv \\
&=& -\int_{\mathcal{M}} g\nabla U \cdot \nabla f \,dv - \int_{\mathcal{M}}\nabla g \cdot \nabla f \, dv \\
&=& -\int_{\mathcal{M}} g\nabla U \cdot \nabla f \,dv - \int_{\mathcal{M}}\nabla\cdot(g\nabla f) + \int_{\mathcal{M}} g\Delta f\,dv \\
&=& \int_{\mathcal{M}} g\left[\Delta f - \nabla U \cdot \nabla f  \right]\,dv - \int_{\partial \mathcal{M}} g \frac{\partial f}{\partial \nu}\,dS ~~~~~~~\text{adjoint b.c.: } \frac{\partial f}{\partial \nu}=0 \\
&=& \int_{\mathcal{M}} g \mathcal{L} f \,dv
\end{eqnarray*}
This shows that $\mathcal{L}^*$ is the formal adjoint of $\mathcal{L}$ if the boundary condition for $f$ is the homogeneous Neumann
$$\frac{\partial f}{\partial \nu} = 0$$

The eigenvectors of $L_{rw}$ are discrete approximations of the eigenfunctions of the backward Fokker-Plank operator with homogeneous Neumann boundary condition in case $\partial \mathcal{M} \neq \emptyset$:

\begin{eqnarray*}
\mathcal{L}\phi &=& \lambda \phi,\quad x\in \mathcal{M}\\
\frac{\partial \phi}{\partial \nu} &=& 0,\quad x\in \partial \mathcal{M}
\end{eqnarray*}

\subsection{Density invariant graph Laplacians}
We now turn our attention to a different normalization of the similarities $w_{ij}$ that leads to different limiting operators.
Recall that we began with the symmetric similarity matrix $$w_{ij} = w_{ji} = \frac{1}{n}K_{\eps}(x_i-x_j)$$ but now we introduce a new similarity:
\[
\tilde{w}_{ij} = \frac{w_{ij}}{d_i d_j} = \frac{1}{n}\frac{K_{\eps}(x_i-x_j)}{d_i d_j}
\]
where $$d_i = \sum_{j=1}^n w_{ij}$$ are the row sums.
This yields the symmetric matrix
\[
\tilde{{W}} = {D}^{-1} {W} {D}^{-1}
\]
which can be used to define a new diagonal matrix $\tilde{D}$ with
\[
\tilde{D}_{ii} = \sum_{j=1}^n \tilde{w}_{ij}
\]
and a new Laplacian and a random walk Laplacian
$$\tilde{L} = \tilde{D} - \tilde{W}$$
and
$$\tilde{L}_{rw} = I - \tilde{D}^{-1}\tilde{W}.$$
Then, $\tilde{L}_{rw}$ converges to the Laplace-Beltrami operator independent of the density $p$, while $\tilde{L}$ still has some dependency on $p$:
$$\lim_{\eps\to 0}\lim_{n\to \infty} -\frac{1}{\eps}\tilde{L}f = \frac{m_2}{2} \frac{1}{p(x_i)}\Delta_M f$$
and
$$\lim_{\eps\to 0}\lim_{n\to \infty} -\frac{1}{\eps}\tilde{L}_{rw}f = \frac{m_2}{2} \Delta_M f.$$
This means that using this re-normalization the random walk Laplacian $\tilde{L}_{rw}$ approaches the Laplace-Beltrami operator, even if the sampling distribution is non-uniform.
We don't even need to know the distribution of sampling (i.e. no density estimation required)!
However, $d_i$ in some sense does give you the density.
Indeed, $d_i \propto p(x_i)$ since we previously showed that:
\begin{eqnarray}
d_i &\to& p(x_i) + \eps \frac{m_2}{2}\left(\Delta_M p(x_i) + \frac{1}{4}E(x_i)p(x_i)\right) + O(\eps^2)
\end{eqnarray}
as $n\to \infty$. We have
$$\frac{1}{d} \to \frac{1}{p} - \eps\frac{m_2}{2p}\left(\frac{\Delta p}{p} + \frac{1}{4}E \right) + O(\eps^2).$$
Therefore,
$$\tilde{w}_{ij} \to w_{ij}\left[\frac{1}{p(x_i)p(x_j)} - \eps\frac{m_2}{2p(x_i)p(x_j)}\left(\frac{\Delta p(x_i)}{p(x_i)} + \frac{1}{4}E(x_i) + \frac{\Delta p(x_j)}{p(x_j)} + \frac{1}{4}E(x_j)\right) + O(\eps^2) \right].$$
Hence,
\begin{eqnarray*}
&& \tilde{D}_{ii} = \sum_{j=1}^n \tilde{w}_{ij}  \\
&=& \frac{1}{n} \sum_{j=1}^n K_{\eps}(x_i-x_j)\left[\frac{1}{p(x_i)p(x_j)} - \eps\frac{m_2}{2p(x_i)p(x_j)}\left(\frac{\Delta p(x_i)}{p(x_i)} + \frac{1}{4}E(x_i) + \frac{\Delta p(x_j)}{p(x_j)} + \frac{1}{4}E(x_j)\right) + O(\eps^2) \right] \\
&\to& \frac{1}{p(x_i)} + \eps \frac{m_2}{2} \frac{1}{p(x_i)}\left[\Delta 1 + \frac{1}{4}E(x_i) \right] - \eps\frac{m_2}{2p(x_i)}\left(\frac{\Delta p(x_i)}{p(x_i)} + \frac{1}{4}E(x_i) + \frac{\Delta p(x_i)}{p(x_i)} + \frac{1}{4}E(x_i)\right) + O(\eps^2) \\
&=& \frac{1}{p(x_i)} + \eps \frac{m_2}{2} \frac{1}{p(x_i)}\left[\frac{1}{4}E(x_i) \right] - \eps\frac{m_2}{p(x_i)}\left(\frac{\Delta p(x_i)}{p(x_i)} + \frac{1}{4}E(x_i) \right) + O(\eps^2) \\
&=& \frac{1}{p(x_i)}\left[1 - \eps\frac{m_2}{2}\left(2\frac{\Delta p(x_i)}{p(x_i)} +\frac{1}{4}E(x_i)\right) + O(\eps^2) \right]
\end{eqnarray*}
And similarly,
\begin{eqnarray*}
&& (\tilde{W}f)(x_i) = \sum_{j=1}^n \tilde{w}_{ij}f(x_j) \\
&=& \frac{1}{n} \sum_{j=1}^n K_{\eps}(x_i-x_j)\left[\frac{f(x_j)}{p(x_i)p(x_j)} - \eps\frac{m_2 f(x_j)}{2p(x_i)p(x_j)}\left(\frac{\Delta p(x_i)}{p(x_i)} + \frac{1}{4}E(x_i) + \frac{\Delta p(x_j)}{p(x_j)} + \frac{1}{4}E(x_j)\right) + O(\eps^2) \right] \\
&\to& \frac{f(x_i)}{p(x_i)} + \eps \frac{m_2}{2} \frac{1}{p(x_i)}\left[\Delta f(x_i)+\frac{1}{4}E(x_i)f(x_i) \right] - \eps\frac{m_2f(x_i)}{p(x_i)}\left(\frac{\Delta p(x_i)}{p(x_i)} + \frac{1}{4}E(x_i) \right) + O(\eps^2) \\
&=& \frac{f(x_i)}{p(x_i)}\left[1 - \eps\frac{m_2}{2}\left(-\frac{\Delta f(x_i)}{f(x_i)} + 2\frac{\Delta p(x_i)}{p(x_i)} +\frac{1}{4}E(x_i)\right) + O(\eps^2) \right]
\end{eqnarray*}
Altogether,
\begin{eqnarray*}
(\tilde{L}f)(x_i) &=& (\tilde{D}-\tilde{W})f(x_i) \\
&\to& \frac{f(x_i)}{p(x_i)}\left[1 - \eps\frac{m_2}{2}\left(2\frac{\Delta p(x_i)}{p(x_i)} +\frac{1}{4}E(x_i)\right) + O(\eps^2) \right] \\
&& -\frac{f(x_i)}{p(x_i)}\left[1 - \eps\frac{m_2}{2}\left(-\frac{\Delta f(x_i)}{f(x_i)} + 2\frac{\Delta p(x_i)}{p(x_i)} +\frac{1}{4}E(x_i)\right) + O(\eps^2) \right] \\
&=& -\eps \frac{m_2}{2}\frac{1}{p(x_i)}\Delta_M f(x_i) + O(\eps^2)
\end{eqnarray*}
For the random walk Laplacian,
\begin{eqnarray*}
(\tilde{L}_{rw}f)(x_i) &=& (I-\tilde{D}^{-1}\tilde{W})f(x_i) \\
&=& -\eps \frac{m_2}{2}\Delta_M f(x_i) + O(\eps^2)
\end{eqnarray*}
Thus, the density invariant normalized random walk Laplacian converges to the Laplace-Beltrami operator, regardless of the density. The normalization separates the density from the geometry.

We remark that other normalizations and variations of the graph Laplacian have been proposed, analyzed, and applied in practice. For example, the bi-stochastic normalization that can be computed efficiently by Sinkhorn–Knopp (SK) iterations is robust to noise heteroskedasticity \cite{landa2021doubly,cheng2024bi}. Motivated by the success of self-tuning spectral clustering \cite{zelnik2004self}, variable bandwidth diffusion kernels and anisotropic diffusion were analyzed in \cite{singer2008non,berry2016variable}. Diffusion maps with affinity kernels based on non-Euclidean distances such as optimal transportation Wasserstein distances and others were considered in \cite{kileel2021manifold,Xu2026}.        

\subsection{Variance error}
For simplicity of exposition, we assume that the points are sampled in $\mathbb{R}^d$. The discussion can be extended to the $d$-dimensional manifold case without much difficulty.
We previously derived the limiting operators for the graph Laplacian and the random walk Laplacian as the number of samples goes to infinity. In practice, however, the number of samples is finite and we need to choose $\eps = \eps_n$ in a suitable manner in order to ensure that the variance error is not too large. A similar consideration was made earlier for non-parametric density estimation. 

We begin by considering the graph Laplacian
$$Lf(i) = (D - W)f(i) = \frac{1}{n}\sum_{j=1}^n K_{\eps}(x_i-x_j) (f(x_i)-f(x_j))$$
The variance is
\begin{eqnarray*}
\text{Var}(Lf(i)) &=& \frac{1}{n}\left[\int_{\mathbb{R}^d} K_{\eps}^2(x_i-y)(f(x_i)-f(y))^2p(y)\,dy + O(\eps^2)  \right] ~~~~~ (\mathbb{E}[Lf]=O(\eps))\\
&=& \frac{1}{n\eps^{d/2}}\left[M_2 (f(x_i)-f(x))^2p(x)|_{x=x_i} + \eps \frac{m_{2,2}}{2}\Delta ((f(x_i)-f(x))^2p(x))|_{x=x_i} + O(\eps^2) \right] \\
&=& \frac{1}{n\eps^{d/2-1}} \left[\frac{m_{2,2}}{2}\Delta ((f(x_i)-f(x))^2p(x))|_{x=x_i} + O(\eps) \right]
\end{eqnarray*}
where
$$m_{2,2} = \frac{1}{d}\int K(x)^2 \|x\|^2\,dx$$
Typically, the leading order term is of size $\frac{1}{n\eps^{d/2}}$, but since the function vanishes at $x_i$ ($f(x_i) - f(x)|_{x=x_i} = 0$), the leading order term vanishes and we get an asymptotically smaller variance.
Recall
$$\Delta (g^2) = 2g\Delta g + 2\|\nabla g\|^2.$$
Therefore, for the uniform distribution $p(x)=const$ and $g(x)=f(x_i)-f(x)$ we have
$$\Delta(f(x_i)-f(x))^2|_{x=x_i} = 2\|\nabla f(x_i)\|^2$$
Hence, for the uniform distribution the variance error is $O(\frac{1}{n\eps^{d/2-2}})$ whenever $\nabla f = 0$. For example, if the manifold is the sphere in $\mathbb{R}^3$, then $f(x,y,z)=z$ is extremal at the north and south poles, and the variance error is asymptotically smaller there.

\subsection{Stochastic fluctuation for the random walk Laplacian}
The random walk Laplacian is given by
$$L_{rw}f(i) = (I - D^{-1}W)f(i) = \frac{\frac{1}{n}\sum_{j=1}^n K_{\eps}(x_i-x_j) (f(x_i)-f(x_j))}{\frac{1}{n}\sum_{j=1}^n K_{\eps}(x_i-x_j)}$$
It is a ratio of two dependent random variables, therefore the variance cannot be simply computed. We can write
$$L_{rw}f(i) = \frac{\sum_{j=1}^n F_j}{\sum_{j=1}^n G_j}$$
where
\begin{eqnarray*}
F_j &=& K_{\eps}(x_i-x_j) (f(x_i)-f(x_j))\\
G_j &=& K_{\eps}(x_i-x_j)
\end{eqnarray*}
We want to show that
$$L_{rw}f(i) = \frac{\sum_{j=1}^n F_j}{\sum_{j=1}^n G_j} \approx \frac{\mathbb{E}[F_j]}{\mathbb{E}[G_j]}$$
and to control the size of the fluctuation as a function of $n$ and $\eps$.
Recall
\begin{eqnarray*}
\mathbb{E}[F_j] &=& \int_{\mathbb{R}^d} K_{\eps}(x_i-y) (f(x_i)-f(y))p(y)\,dy \\ &=& \eps \frac{m_2}{2}\Delta ((f(x_i)-f(y))p(y))|_{y=x_i} + O(\eps^2) \\
\mathbb{E}[G_j] &=& \int_{\mathbb{R}^d} K_{\eps}(x_i-y) p(y)\,dy \\ &=& p(x_i) + O(\eps)
\end{eqnarray*}
and
\begin{eqnarray*}
\mathbb{E}[F_j^2] &=& \int_{\mathbb{R}^d} K_{\eps}^2(x_i-y) (f(x_i)-f(y))^2 p(y)\,dy \\
&=& \frac{1}{\eps^{d/2-1}} \frac{m_{2,2}}{2}\Delta ((f(x_i)-f(y))^2p(y))|_{y=x_i} + O(\frac{1}{\eps^{d/2-2}}) \\
\mathbb{E}[G_j^2] &=& \int_{\mathbb{R}^d} K_{\eps}^2(x_i-y) p(y)\,dy \\
&=& \frac{M_2}{\eps^{d/2}}p(x_i) + O(\frac{1}{\eps^{d/2-1}}) \\
\mathbb{E}[F_j G_j] &=& \int_{\mathbb{R}^d} K_{\eps}^2(x_i-y) (f(x_i)-f(y))p(y)\,dy \\
&=& \frac{1}{\eps^{d/2-1}} \frac{m_{2,2}}{2}\Delta ((f(x_i)-f(y))p(y))|_{y=x_i} + O(\frac{1}{\eps^{d/2-2}})
\end{eqnarray*}
We conclude that
\begin{eqnarray*}
\text{Var}(F_j) &=& \frac{1}{\eps^{d/2-1}} \frac{m_{2,2}}{2}\Delta ((f(x_i)-f(y))^2p(y))|_{y=x_i} + O(\frac{1}{\eps^{d/2-2}}) \\
\text{Var}(G_j) &=& \frac{M_2}{\eps^{d/2}}p(x_i) + O(1,\frac{1}{\eps^{d/2-1}}) \\
\text{Cov}(F_j, G_j) &=& \mathbb{E}[F_j G_j] - \mathbb{E}[F_j]\mathbb{E}[G_j] \\
&=& \frac{1}{\eps^{d/2-1}} \frac{m_{2,2}}{2}\Delta ((f(x_i)-f(y))p(y))|_{y=x_i} + O(\eps,\frac{1}{\eps^{d/2-2}})
\end{eqnarray*}
The correlation between $F_j$ and $G_j$ is
\begin{eqnarray*}
\rho (F_j,G_j) &=& \frac{\text{Cov}(F_j,G_j)}{\sqrt{\text{Var}(F_j)}\sqrt{\text{Var}(G_j)}} \\
&=& O\left(\frac{1}{\eps^{d/2-1}} \sqrt{\eps^{d/2 + d/2 - 1}}\right) \\
&=& O(\sqrt{\eps})
\end{eqnarray*}
Altogether,
\begin{eqnarray*}
L_{rw}f(i) &=& \frac{\frac{1}{n}\sum_{j=1}^n F_j}{\frac{1}{n}\sum_{j=1}^n G_j} \\
&=& \frac{\mathbb{E}[F_j] + \xi_{n,\eps}^f}{\mathbb{E}[G_j] + \xi_{n,\eps}^g} \\
&=& \frac{\eps \frac{m_2}{2}\Delta ((f(x_i)-f(y))p(y))|_{y=x_i} + O(\eps^2) + \xi_{n,\eps}^f}{p(x_i) + O(\eps) + \xi_{n,\eps}^g} \\
&=& \eps\frac{\frac{m_2}{2}\Delta ((f(x_i)-f(y))p(y))|_{y=x_i} + O(\eps) + O_P(\frac{1}{n^{1/2}\eps^{d/4+1/2}})}{p(x_i) + O(\eps) + O_P(\frac{1}{n^{1/2}\eps^{d/4}})}
\end{eqnarray*}
Therefore,
$$\frac{1}{\eps}L_{rw}f(i) = \frac{m_2}{2p(x_i)}\Delta ((f(x_i)-f(y))p(y))|_{y=x_i} + O(\eps) + O_P(\frac{1}{n^{1/2}\eps^{d/4+1/2}})$$
By balancing the variance and squared bias we get
$$\eps^2 = \frac{1}{n \eps^{d/2+1}}$$
or
$$\eps^{d/2+3} = \frac{1}{n}$$
so the optimal kernel bandwidth is
$$\eps_n = O\left(\frac{1}{n^{1/(d/2+3)}}\right).$$

\subsection{Large deviation bounds}
We can also get bounds on large deviation with high probability. The random variables $F_j$ are uniformly bounded by $$c=O(\eps^{-d/2})$$ and their variance is $$\sigma^2 = O(\eps^{-(d/2-1)})$$ We see that $$\sigma^2 \ll c$$ so Bernstein's inequality could in principle provide a large deviation bound that is tighter than that provided by Hoeffding's inequality. Recall Bernstein's inequality
$$\Pr \left\{\frac{1}{n}\sum_{j=1}^n (F_j - \mathbb{E}[F_j]) > \alpha \right\} \leq e^{-\frac{n\alpha^2}{2\sigma^2 + \frac{2}{3}c\alpha}}$$
Since we estimate a quantity of size $O(\eps)$ (the prefactor of the Laplacian), we need to take $\alpha \ll \eps$. Let us take $\alpha=o(\eps)$. The exponent in Bernstein takes the form
$$\frac{n\alpha^2}{2\sigma^2 + \frac{2}{3}c\alpha} = \frac{n \alpha^2}{O(\eps^{-(d/2-1)}) + o(\eps^{-d/2}\eps)} = O(n\alpha^2\eps^{d/2-1})$$
Choosing
$$n\alpha^2\eps^{d/2-1} = \log n $$
means that the deviation happens with probability that goes to 0 as $n\to \infty$ (even at all points by union bound).
This means that the deviation is of size
$$\alpha = \frac{\sqrt{\log n}}{n^{1/2}\eps^{d/4-1/2}}.$$
We see that the only price to pay in order to get a large deviation bound is a factor of $\sqrt{\log n}$. Taking a series $\eps_n\to 0$ with $\frac{\sqrt{\log n}}{n^{1/2}\eps_n^{d/4-1/2}}\to 0$ guarantees point-wise convergence, and also uniform convergence.

In the case of data points sampled from a manifold, it is desirable to show that the discrete eigenvectors of the Laplacian operators converge to the eigenfunctions of the differential operators. The pointwise convergence and even the uniform convergence are unfortunately not sufficient to demonstrate this type of convergence. The precise treatment of such spectral convergence is beyond the scope of this textbook.

\subsection{Sturm-Liouville}
Consider the eigenvalue problem
\[
\Delta f - \nabla U \cdot \nabla f = -\lambda f
\]
which in 1-D is
\[
f^{\prime \prime} - U^\prime f^\prime = -\lambda f
\]
and if we multiply by the integrating factor $\exp(-U)$ we get 
\[
(e^{-U} f^\prime)^\prime + \lambda e^{-U} f = 0,
\]
which is a Sturm-Liouville equation.
Thus we know that it has a countable number of eigenvalues, its eigenfunctions form an orthonormal basis for the suitable weighted $L^2$ space, and everything else that comes with the Sturm-Liouville theory. In particular, Sturm-Liouville theory implies that the first non-trivial eigenfunction sorts points on an open curve according to its arclength, and the embedding of a closed curve in the plane encircles the origin exactly once (see rexercises at the end of this chapter).  

\section{Can one hear the shape of the data?}
Cheeger's inequality provides a geometric meaning for the second eigenvalue of the Laplacian as an isoperimtric inequality. It holds for both the discrete (graph) and the continuous (manifold) Laplacians. What other geometric information is encoded by the eigenvalues of the Laplacian? The continuous case has been widely explored, and it turns out that the eigenvalues encode some shape information, as was put famously by Mark Kac in his paper in the American Mathematical Monthly in 1966 \cite{kac1966}.

\subsection{Wave equation}
The Laplacian appears in many of the important PDEs describing physical phenomenon such as wave propagation, diffusion and heat equation, quantum mechanics (Schr\"odinger equation), and more. This is perhaps due to the fact that the Laplacian commutes with rotations and translations and therefore appears in the description of physical phenomenon with such symmetries.

Consider a ``drum" or vibrating membrane $D \subset \mathbb{R}^2$, whose configuration over time $u(t,x)$ is described by the {\em{wave equation}}
\begin{eqnarray*}
u_{tt}(t,x) &=&c^2\Delta u (t,x), \quad x\in D, \; t>0,
\end{eqnarray*}
where $\Delta u = u_{x_1x_1} + u_{x_2x_2}$, and $c$ is the wave velocity. Fixing the boundary accounts for the Dirichlet boundary condition
$$u(t,x) = 0, \quad x \in \partial D,\; t \geq 0.$$

We can derive the solution using the usual separation of variables technique.  Suppose $u(t,x) = T(t) X(x)$, then
\begin{eqnarray*}
T''X&=&c^2T\Delta X\\
&\Updownarrow&\\
\frac{T''}{T} &=& c^2\frac{\Delta X}{X} = -\lambda\\
\end{eqnarray*}
resulting in  $T''(t)=-\lambda T \Rightarrow T(t) = e^{\pm \imath \sqrt{\lambda}t}$ and

\begin{displaymath}
\left\{
     \begin{array}{lr}
       c^2\Delta X = -\lambda X  &\;\;\text{($X$ is a Laplacian eigenfunction)}\\
       X(x)=0 , x\in\partial D &\;\;\text{(Dirichlet boundary condition)}\\
     \end{array}
   \right.
\end{displaymath}

Therefore, we wish to find the eigenfunctions $\lbrace\varphi_n\rbrace_{n=1}^\infty$ of the Laplacian $\Delta$ with Dirichlet boundary condition:
\begin{eqnarray*}
       \Delta \varphi_n &=& -\lambda_n \varphi_n,\quad x\in D,\\
       \varphi_n(x)&=&0, \quad \quad x\in\partial D,
\end{eqnarray*}
with eigenvalues $0 < \lambda_1 \leq \lambda_2 \leq \cdots \leq \lambda_n \leq \cdots$. We see that the following relationship holds: $$\lambda = c^2\lambda_n.$$
Because $\varphi_1, \varphi_2, \cdots$ form an orthonormal basis to $L^2(D)$, we can write
\begin{equation*}
u(t,x)=\sum_{n=1}^\infty \left( a_n e^{\imath c\sqrt{\lambda_n} t} + b_n e^{-\imath c \sqrt{\lambda_n} t}\right) \varphi_n(x)\\
\end{equation*}
To find $a_n$ and $b_n$, we need initial conditions $$u(0,x)=f(x),$$ and $$u_t(0,x)=g(x).$$

\subsection{Diffusion Equation}
In 1-D, we have $u_t=\Delta u$ over the internal $x\in[0,L]$. With Neumann boundary conditions $u_x(t,0)=0$, $u_x(t,L)=0$, the eigenfunctions $$\phi_n''=-\lambda_n \phi_n$$ are given by
$$\phi_n = \cos(\sqrt{\lambda_n} x)$$
with
$$\sin{(\sqrt{\lambda_n} L)} = 0 \Rightarrow \sqrt{\lambda_n} L=n\pi,$$
$$\lambda_n=\frac{n^2 \pi^2}{L^2}, \; n=0,1,2,\ldots.$$
Therefore,
\begin{eqnarray*}
u(t,x)&=&\sum_{n=0}^\infty a_n e^{-\lambda_n t} \cos{(\sqrt{\lambda_n} x)}\\
&=&\sum_{n=0}^\infty a_n e^{-\frac{n^2 \pi^2}{L^2} t} \cos\left(\frac{n\pi x}{L}\right).\\
\end{eqnarray*}

In 2-D, the Laplacian is $\Delta u=u_{xx} + u_{yy}$. For a rectangle of size $L_x\times L_y$ we construct the eigenfunction $\phi_{n m}(x,y)$ by taking the tensor products of eigenfunctions of the respective 1-D equations, e.g., $$\phi_{n m}(x,y) = X_n(x)Y_n(y).$$ The eigenfunction problem then becomes
\begin{equation*}
-\Delta\phi_{nm} = \lambda_nX_nY_m + \lambda_mX_nY_m = (\lambda_n+\lambda_m) \phi_{nm}\\
\end{equation*}
Then $$\lambda_{nm} = \lambda_n+\lambda_m = \left(\frac{n\pi}{L_x}\right)^2+\left(\frac{m\pi}{L_y}\right)^2, \quad n,m=0,1,2,\ldots.$$ Notice that the growth of eigenvalues for 2-D is different than the 1-D case.

We count the number of eigenvalues: $$N(\lambda) = \#\{\lambda_n \le \lambda\}.$$
For the 1-D case: $\left(\frac{n\pi}{L}\right)^2 \le \lambda$, so $N(\lambda)=\sqrt{\frac{\lambda L^2}{\pi^2}}=\frac{L}{\pi}\lambda^{1/2}$\\
For the 2-D case: $\left(\frac{n\pi}{L_x}\right)^2 + \left(\frac{m\pi}{L_y}\right)^2\le \lambda$, so $\frac{n^2}{\left(\frac{\sqrt{\lambda}L_x}{\pi}\right)^2} + \frac{m^2}{\left(\frac{\sqrt{\lambda} L_y}{\pi}\right)^2} \le 1$\\.

In the 2-D, $N(\lambda)$ is the number of regular lattice points inside the first quadrant of the ellipse with semi axes $\frac{\sqrt{\lambda}L_x}{\pi}$ and $\frac{\sqrt{\lambda}L_y}{\pi}$. Therefore, $N(\lambda)$ is approximately $\frac{1}{4}$ the area of the ellipse
\begin{eqnarray*}
N(\lambda) & \approx & \frac{1}{4} \pi a b = \frac{1}{4}\pi \left(\frac{\sqrt{\lambda} L_x}{\pi}\right) \left(\frac{\sqrt{\lambda} L_y}{\pi}\right)\\
&=& \frac{\lambda L_x L_y}{4\pi}.
\end{eqnarray*}

We see that in 1-D, $N(\lambda) = \frac{L}{\pi}\lambda^{1/2}$, and in 2-D, $N(\lambda) \sim \frac{\lambda L_x L_y}{4\pi}$. Similar considerations can be made for higher dimensional boxes.

From this, we can guess that $N(\lambda) \sim c \lambda^{d/2}$, $c = c(d) |D|$, where $c(d)$ is some constant that depends on $d$ and $|D|$ is the volume of the domain.

\subsection{Green Function/Heat Kernel}
The Green's function $G_t(x,y)$ enables us to write the solution to the diffusion equation for any initial distribution $u(0,y)=f(y)$ as
\begin{equation*}
u(t,x)=\int\limits_D G_t(x,y)f(y)dy
\end{equation*}
We see that $G_t(x,y)$ needs to satisfy some conditions. First, from $u_t = \Delta u$ we must have
\begin{equation*}
\int\limits_D \frac{\partial G}{\partial t} f(y) dy = \int\limits_D \Delta_x G_t(x,y)f(y)dy.
\end{equation*}
Since we want it to hold for any $f$, then $G$ should satisfy the diffusion equation by itself
\begin{equation*}
\frac{\partial}{\partial t}G_t(x,y)=\Delta_x G_t(x,y).
\end{equation*}
Also, $G_t(x,y)$ must satisfy the initial and boundary conditions, e.g., $$G_t(x,y) \xrightarrow{t\rightarrow 0}\delta(x,y)$$ and
$$G_t(x,y) = 0, \quad x\in \partial D,\; y\in D,$$
for Dirichlet boundary condition, or
$$\frac{\partial G_t(x,y)}{\partial n_x} = 0, \quad x\in \partial D,\; y\in D,$$
for Neumann boundary condition.

The heat kernel $G_t(x,y)$ has the following probabilistic interpretation: it is the probability density function of a Brownian motion $X(t)$ that starts at $X(0)=y$ at time 0 to be at $X(t)=x$ at time $t$. It can therefore be regarded as the propagator of the probability density. 

Since $G_t(\cdot,y)$ satisfies the diffusion equation with the boundary condition, we can write $$G_t(x,y) = \sum_{n=1}^\infty a_n(y) e^{-\lambda_n t} \phi_n(x).$$ Also, for $t=0$ we formally have
\begin{equation*}
\delta(x,y) = G_0(x,y) = \sum_{n=1}^\infty a_n(y)\phi_n(x),
\end{equation*}
and upon taking inner product with $\phi_m$ we get
\begin{equation*}
a_m(y) = \phi_m(y),
\end{equation*}
where we used the orthogonality of Laplacian eigenfunctions. Therefore,
$$G_t(x,y) = \sum_{n=1}^\infty e^{-\lambda_n t} \phi_n(x)\phi_n(y).$$
Just like with spectral decomposition of finite matrices, the last equation gives the spectral decomposition of the heat kernel $G_t$ in terms of its eigenvalues and eigenfunctions.
We can also understand the spectral representation of the heat kernel from the fact that the solution to the diffusion equation
$$u_t = \Delta u$$
is
$$u = e^{t \Delta } f$$
in the same way that the solution to a system of ODEs
$$\frac{du}{dt} = Au$$
with a constant coefficient matrix $A$ is
$$u = e^{tA}u_0.$$

The {\em{trace}} of $G$ is defined as
\begin{eqnarray*}
\text{Tr}(G_t) &=& \int_D G_t(x,x)\,dx\\
&=&\sum_{n=1}^\infty e^{-\lambda_n t} \int_D  \phi_n^2(x)\, dx\\
&=&\sum_{n=1}^\infty e^{-\lambda_n t}.
\end{eqnarray*}

If we know the spectrum $\{\lambda_n\}_{n=1}^\infty$ then we can construct the trace, and vice versa, from the trace we can get the eigenvalues (e.g., $$\lambda_1 = -\lim_{t\to \infty} \frac{\log \text{Tr}(G_t)}{t},$$ and we can peel-off the eigenvalues one by one).

\subsection{Weyl's Law}
Motivated by the eigenvalues of the Laplacian in a box, we conjectured that
\begin{equation*}
N(\lambda) \sim \lambda^{d/2},
\end{equation*}
or equivalently,
\begin{equation*}
\lambda_n \sim n^{2/d},
\end{equation*}
which is known as Weyl's law, that we now derive.

First, we observe that $$\text{Tr}(G_t) = \sum_{n=1}^\infty e^{-\lambda_n t}$$ diverges as $t\to 0$ because it becomes an infinite sum of ones. At what rate does it diverge?
We show that for a region $\Omega \subset \mathbb{R}^d$ we have
\begin{equation*}
\text{Tr}(G_t)=\int_\Omega G_t(x,x)\, dx = \sum_{n=1}^\infty e^{-\lambda_n t} \sim \frac{|\Omega|}{(4\pi t)^{d/2}},\quad \text{as } t\to 0.
\end{equation*}

\paragraph{Diffusion over the real line}
We start with diffusion over the entire real line (no boundary conditions). In this case, the Green function is the Gaussian kernel. Indeed,
the Green function satisfies $$\frac{\partial}{\partial t}G_t(x,y)=\frac{\partial^2}{\partial x^2}G_t(x,y),$$ with the initial condition $$\lim_{t\to 0}G_t(x,y)=\delta(x,y).$$  We ask then: what does the heat kernel $G_t(x,y)$ look like? The answer is $$G_t(x,y) = \frac{1}{\sqrt{4\pi t}} e^{-(x-y)^2/{4t}},$$
so that the solution to the heat equation is obtained by convolving the initial distribution with $G_t$:
\begin{equation*}
u(x,t)=\int_{-\infty}^\infty G_t(x,y)f(y)\,dy = \frac{1}{\sqrt{4\pi t}}\int_{-\infty}^\infty e^{-\frac{(x-y)^2}{4t}}f(y)\,dy.
\end{equation*}
This result can be proven by taking the Fourier transform of the diffusion equation:
\begin{equation*}
\frac{\partial}{\partial t}\hat{G}_t(\xi,y)= -4\pi^2\xi^2\hat{G}_t(\xi,y),
\end{equation*}
which is a linear first order ODE for each $\xi$ and $y$ with solution
\begin{equation*}
\hat{G}_t(\xi, y) = A(\xi,y) e^{-4\pi^2 \xi^2 t}.
\end{equation*}
Taking the Fourier transform of the the initial condition
\begin{equation*}
G_0(x,y) = \delta(x,y),
\end{equation*}
we get
\begin{equation*}
\hat{G}_0(\xi,y) = \int_{-\infty}^\infty e^{-2\pi \imath \xi x} \delta(x,y)\,dx = e^{-2\pi \imath \xi y}.
\end{equation*}
Therefore,
\begin{equation*}
\hat{G}_0(\xi,y) = A(\xi,y) = e^{-2\pi \imath \xi y},
\end{equation*}
and we have
\begin{equation*}
\hat{G}_t(\xi, y) = e^{-2\pi \imath \xi y} e^{-4\pi^2 \xi^2 t}.
\end{equation*}
This is a shifted Gaussian in $\xi$. The (inverse) Fourier transform of a Gaussian is also a Gaussian. It can be easily verified that 
\begin{equation*}
G_t(x,y) = \int_{-\infty}^\infty e^{2\pi \imath \xi x}e^{-2\pi \imath \xi y} e^{-4\pi^2 \xi^2 t}\,d\xi = \frac{1}{\sqrt{4\pi t}} e^{-(x-y)^2/{4t}}.
\end{equation*}

\paragraph{Diffusion over the semi-infinite line}

In considering $\mathbb{R}_+$ for $u_t=\Delta u$ with Neumann boundary condition $u_x(0)=0$, we use the {\em{method of images}} to derive the heat kernel. This is done by placing an imaginary diffusive particle at $-y$, so its trajectory enters the positive line whenever the original particle leaves it, thus preserving the no-flux boundary condition. The corresponding Green's function is therefore
\begin{equation*}
G_t(x,y)=\frac{1}{\sqrt{4\pi t}}e^{\frac{-(x-y)^2}{4t}}+\frac{1}{\sqrt{4\pi t}}e^{\frac{-(x+y)^2}{4t}}\\
\end{equation*}
Note that for the Dirichlet boundary condition at the origin, the Green's function is the difference between the two terms.

\paragraph{Diffusion over a finite interval}
It is a homework exercise to derive the heat kernel for diffusion in the finite interval $[0,L]$. This can be done using the method of images, but now an infinite number of images is required. Still, the short time asymptotic behavior is that of the heat kernel over the entire real line: at very short times our diffusive particle does not ``see" the boundary.

{\textbf{Short-time asymptotics of the heat kernel}}: for small times ($t\rightarrow 0$), we obtain $G_t(x,y)\sim\frac{1}{\sqrt{4\pi t}} e^{\frac{-(x-y)^2}{4t}}$, again as $t\rightarrow0$.

\subsection{Any domain $\Omega\subset\mathbb{R}^d$}
Again, using the Fourier transform we can derive the heat kernel for diffusion in $\mathbb{R}^d$ and get
\begin{equation*}
G_t(x,y) = \frac{1}{(4\pi t)^{d/2}} e^{-\frac{\|x-y\|^2}{4t}}.
\end{equation*}
Alternatively, the heat kernel in $\mathbb{R}^d$ is the product of $d$ one-dimensional heat kernels, corresponding to independent Brownian motions in each dimension.
This is also the short time asymptotic behavior of the heat kernel for any domain $\Omega \subset \mathbb{R}^d$, by arguing that at very short times the particle does not see the boundary.
More formally, we can consider an asymptotic expansion of the form
\begin{equation*}
G_t(x,y)\sim\sum\limits_k \frac{1}{(4\pi t)^{d/2}}e^{-\frac{S_k^2(x,y)}{4t}}\sum_n Z_{k,n}(x,y)t^n,\quad \text{as } t\to 0.
\end{equation*}

Plugging this into the heat equation, we get the Eikonal equation $|\nabla S_k|=1$ for the $S_k$'s, and transport equations for the $Z_{k,n}$'s.  The $S_k$'s correspond to length of geodesics, or billiard trajectories from $y$ to $x$ that can be reflected from the boundary. The expansion of diagonal terms is $G_t(x,x)\sim\frac{1}{(4\pi t)^{d/2}}$. Therefore, the trace becomes
\begin{equation*}
\text{Tr}(G_t)=\int\limits_\Omega G_t(x,x)\,dx\sim\frac{|\Omega|}{(4\pi t)^{d/2}}
\end{equation*}
as $t\rightarrow0$.

\subsection{Weyl's Law}
To examine $N(\lambda)=\#\{\lambda_n<\lambda\}$, we look at
\begin{equation*}
\sum_{n=1}^\infty e^{-\lambda_n t} = \int_0^\infty e^{-\lambda t}\, dN(\lambda)\sim \frac{|\Omega|}{(4\pi t)^{d/2}}
\end{equation*}
Assuming $N(\lambda)\sim c \lambda^\alpha$ and $N'(\lambda)\sim\alpha c \lambda^{\alpha-1}$ we get
\begin{eqnarray*}
\frac{|\Omega|}{(4\pi t)^{d/2}}\sim\int\limits_0^\infty e^{-\lambda t}\alpha c \lambda^{\alpha-1}\,d\lambda &\underbrace{=}_\text{$s=\lambda t$}& \alpha c\int\limits_0^\infty e^{-s}\left(\frac{s}{t}\right)^{\alpha-1}\frac{ds}{t}\\
&=&\frac{\alpha c}{t^\alpha}\underbrace{\int\limits_0^\infty e^{-s}s^{\alpha-1}ds}_\text{$\Gamma(\alpha)$}
\end{eqnarray*}
Hence,
$$\alpha=\frac{d}{2}$$
Therefore,
\begin{equation*}
\frac{|\Omega|}{(4\pi)^{d/2}} = \frac{d}{2}c \Gamma(\frac{d}{2}) \Rightarrow c=\frac{|\Omega|}{(4\pi)^{d/2}\frac{d}{2}\Gamma(\frac{d}{2})}\\
\end{equation*}
and so we get the asymptotic behavior (Weyl's Law):
\begin{equation}
N(\lambda)\sim\frac{|\Omega|}{(4\pi)^{d/2}\frac{d}{2}\Gamma(\frac{d}{2})}\lambda^{d/2}
\end{equation}
For $d=1$, we have
\begin{eqnarray*}
N(\lambda)&\sim&\frac{L}{\sqrt{4\pi}\frac{1}{2}\underbrace{\sqrt{\pi}}_\text{$\Gamma(\frac{1}{2})$}} \lambda^{1/2}\\
&\sim&\frac{L}{\pi}\lambda^{1/2} \text{\hspace{1cm} (same as previous)}\\
\end{eqnarray*}
For $d=2$, we have
\begin{eqnarray*}
N(\lambda)&\sim&\frac{|\Omega|}{4\pi}\lambda, \hspace{0.5cm}|\Omega|=L_xL_y \text{\hspace{1cm} (same as previous)}\\
\end{eqnarray*}

From the spectrum we can therefore infer the dimension of the region and also its volume by considering the growth rate of the eigenvalues (e.g., by looking at the eigenvalues on a log-plot).

For $d=2$, Kac found additional higher order terms in the asymptotic expansion:
\begin{equation*}
\sum_{n=1}^\infty e^{-\lambda_n t} \sim \frac{|\Omega|}{4\pi t}-\frac{|\partial\Omega|}{8(\pi t)^{1/2}}+\frac{1}{6}(1-h)+O(\sqrt{t})
\end{equation*}
basically showing that besides the area of the domain $|\Omega|$, we can hear the length of its boundary $|\partial \Omega|$ and the number of holes $h$ inside the domain. In fact, all these terms are just integrals over the boundary of curvature (Gauss' formula gives the area, length of boundary is clear, Euler's characteristic is given by the Gauss-Bonnet theorem, etc.)

However, by just looking at the eigenvalues, it is typically impossible to {\textit{exactly}} determine the manifold/domain. There are domains which are different but have the same spectrum. These are called isospectral domains. So, no, one cannot hear the shape of the drum exactly. Still, the eigenvalues provide a spectral signature, acting as a hash function of the domain.\\

\subsection{Diffusion map revisited}
Define the embedding of the domain/manifold into $\ell^2$:
\begin{equation*}
\Phi_t(x) = \left(e^{-\lambda_1 t/2}\phi_1(x),e^{-\lambda_2 t/2}\phi_2(x), \ldots \right)
\end{equation*}
where
\begin{equation}
\Delta \phi_l = -\lambda_l \phi_l,
\end{equation}
are the eigenfunctions of the Laplacian over the manifold with Neumann boundary conditions (if there is a boundary) $0 = \lambda_1 < \lambda_2 \leq \lambda_3 \leq \cdots$.
Indeed,
\begin{equation*}
\Phi_t(x)\in\ell^2,
\end{equation*}
because
$$\|\Phi_t(x)\|^2=\sum_{n=1}^\infty e^{-\lambda_n t}\phi_n(x)^2 = G_t(x,x),$$ that is, the diagonal of the heat kernel is the squared norm, and in general, the inner products are given by $$\left\langle\Phi_t(x),\Phi_t(y)\right\rangle = \sum_{n=1}^\infty e^{-\lambda_n t}\phi_n(x)\phi_n(y) = G_t(x,y).$$
(again, notice that $G_t$ is PSD.)

From the short time asymptotic behavior $G_t(x,x)\sim\frac{1}{(4\pi t)^{d/2}}$ and $$\|\Phi_t(x)\|^2=G_t(x,x) \sim \frac{1}{(4\pi t)^{d/2}},$$ we conclude that all $\Phi_t(x)$ have roughly the same norm as $t\rightarrow 0$ for all $x$. In other words, $\Phi_t$ maps the domain/manifold to some subset of the sphere of radius $\frac{1}{(4\pi t)^{d/4}}$. Also, from the probabilistic interpretation of $G_t(x,y)$ it follows that $$G_t(x,y) \geq 0$$ for all $x$, $y$, and $t$. The angle between any two vectors $\Phi_t(x)$ and $\Phi_t(y)$ is acute. As a result $\Phi_t(\Omega)$ occupies only a small portion of the sphere.

The diffusion distance
\begin{eqnarray*}
D_t^2(x,y) &=& \left\langle\Phi_t(x),\Phi_t(x)\right\rangle + \left\langle\Phi_t(y),\Phi_t(y)\right\rangle -2\left\langle\Phi_t(x),\Phi_t(y)\right\rangle\\
&=& G_t(x,x)+G_t(y,y)-2 G_t(x,y)\\
&\sim& \frac{1}{(4\pi t)^{d/2}} \left[1+1-2 e^{-\|x-y|^2/{4t}}\right]\\
&=&\frac{2}{(4\pi t)^{d/2}} \left(1-e^{-\frac{\|x-y\|^2}{4t}}\right)\\
&\sim&\frac{\|x-y\|^2}{(4\pi t)^{d/2} 2t}
\end{eqnarray*}
whenever $\|x-y\| \ll \sqrt{t}$. We see that the diffusion distance behaves locally like the geodesic distance. However, the global behavior is very different:
\begin{equation*}
D_t^2(x,y) \sim \frac{2}{(4\pi t)^{d/2}},
\end{equation*}
for $\|x-y\| \gg \sqrt{t}$ as $t\to 0$. That is, the distances between points of large geodesic distance are roughly the same, that is, the distance saturates.

The idea of embedding Riemannian manifolds by their heat kernel goes back to B\'erard, Besson, and Gallot in 1994 \cite{berard1994embedding}. It took another decade until the idea was rediscovered by the applied math community.

\subsection{Commute time distance}
Let us turn our attention to the commute time distance introduced in Chapter~\ref{s:commute}.There, we defined the commute time embedding as
\begin{equation*}
\Psi(x) = \left(\frac{1}{\sqrt{\lambda_2}}\phi_2(x),\frac{1}{\sqrt{\lambda_3}}\phi_3(x),\ldots\right).
\end{equation*}
We claim that for $d\geq 2$, the embedding cannot be in $\ell^2$. Indeed,
$$\|\Psi(x)\|^2 = \sum_{n=2}^\infty \frac{1}{\lambda_n} \phi_n^2(x).$$ Assume to the contrary that there is a constant  $M>0$ such that $$\|\Psi(x)\|^2 \leq M$$ for all $x\in \Omega$. Integrating over $\Omega$ gives
$$\sum_{n=2}^\infty \frac{1}{\lambda_n} = \int_{\Omega} \|\Psi(x)\|^2\,dx \leq M |\Omega|.$$ But this is a contradiction to Weyl's Law, because
$$\lambda_n\sim n^{2/d}$$
and the sum
$$\sum_{n=1}^\infty \frac{1}{n^{2/d}}$$
diverges for $d \geq 2$. Only for $d=1$ the sum converges. Thus, the commute time embedding is in $\ell^2$ only for $d=1$.

We can understand this in terms of the behavior of the Green's function for Poisson's equation in $\mathbb{R}^d$. The Green's function in that case is the fundamental solution for the electric/gravitational potential due to a point charge/mass. It is singular for dimension $d\geq 2$ (log singularity $\log(r)$ for $d=2$ and algebraic singularity $1/r^{d-2}$ for $d\geq 3$) and is regular only for $d=1$. Recall that Poisson's equation is
$$\Delta u = f$$
Again, we can write the solution in terms of the fundamental solution
$$\Delta_x G(x,y) = \delta(x,y)$$
with Neumann boundary conditions, as
$$u(x) = \int G(x,y)f(y)\,dy.$$
And similar to the previous derivation
$$G(x,y) = \sum_{n=2}^\infty \frac{1}{\lambda_n}\phi_n(x)\phi_n(y)$$

\subsection{The biharmonic distance}
The operator $\Delta^2$ which is the second iterate of the Laplacian is called the {\textit{biharmonic operator}}. It appears in linear elasticity and fluid flow. The Green's function to
$$\Delta^2 u = f$$
with appropriate boundary conditions is
$$G(x,y) = \sum_{n=2}^\infty \frac{1}{\lambda_n^2}\phi_n(x)\phi_n(y).$$
By Weyl's Law
$$\sum_{n=2}^\infty \frac{1}{\lambda_n^2} \sim \sum_{n=2}^\infty \frac{1}{n^{4/d}}$$
which converges for $d < 4$, for which the fundamental solution is regular. The biaharmonic distance is popular in computer graphics in order to measure distances on surfaces: unlike the commute time distance it does not diverge as the mesh is refined, and unlike the diffusion distance it is global, i.e., the distance does not saturate. Up to a logarithmic factor, the biharmonic distance behaves similar to the geodesic distance.  In general, given the dimension $d$ of space, one can define an embedding using any suitable iterate of the Laplacian $\Delta^m$.

\section{Vector diffusion maps and the graph connection Laplacian}\label{s:vectordiffusionmap}

We conclude this chapter with vector diffusion maps and the graph connection Laplacian that extend diffusion maps and the graph Laplacian mathematical from scalar functions to vector fields \cite{ASinger_HTWu_2011_VDM}. 

\subsection{Quick review of diffusion maps}

To set the stage for vector diffusion maps, it is convenient to briefly recall the construction and some basic properties of standard diffusion maps. Recall that in this setting we have data points $x_1, x_2, \ldots, x_n$ that are related by affinities $w_{ij}$ in a weighted graph. We build the weight matrix $W$ from the affinities $w_{ij}$ and consider the random walk matrix $A = D^{-1}W$, where $D$ is the diagonal matrix of degrees $d_i = \sum \limits_{j} w_{ij}$. $A$ is similar to the symmetric matrix $D^{-1/2}WD^{-1/2}$. By the spectral theorem, this can be written as $V \Lambda V^T$, where $\Lambda$ is the diagonal matrix of eigenvalues $\lambda_i$ in decreasing order of magnitude, and $V$ is the matrix of eigenvectors. Then the eigenvectors of $A$ are $\phi_l = D^{-1/2}v_l$.

Then recall that the \emph{diffusion map} is the function from the data to $\mathbb{R}^n$ given by:

\[ \Phi_t(i) : x_i \to \left(\lambda_l^t\phi_l(i)\right)_{l=1}^n. \]
The inner product between the embeddings of $x_i$ and $x_j$ is:
\[ \langle \Phi_t(i), \Phi_t(j) \rangle = \sum \limits_{l=1}^n \lambda_l^{2t}\phi_l(i)\phi_l(j). \]
This inner product can also be expressed, up to a normalizing factor, via the entries of $A^{2t}$. Indeed from the spectral decomposition of $A$:
\[ A^{2t}(i,j) = \sum_{l=1}^n \lambda_l^{2t}\phi_l(i)\psi_l(j), \]
where $\psi_l$ satisfy $\psi_l(j) = d_j \phi_l(j)$, therefore $$A^{2t}(i,j)/d_j = \langle \Phi_t(i), \Phi_t(j) \rangle.$$ Recall that $A^{2t}(i,j)$ gives the probability that starting at $i$, the random walk according to $A$ will end up at $j$ in $2t$ time steps.

\subsection{Vector diffusion maps}

The basic setup of vector diffusion maps is analogous to that of diffusion maps. We have data points $x_i$ and scalar affinities $w_{ij}$ between them, and we think of the data as forming a weighted graph. In addition, however, we also have orthogonal matrices $O_{ij}$ of size $d\times d$ (for some $d$) that give additional information about the similarity between the data points. For instance, if the data points are two-dimensional images, then $O_{ij}$ may be the in-plane rotation that aligns them best, assuming that the images have been centered.

\begin{figure}[h]
\begin{center}
\includegraphics[width=0.4\textwidth]{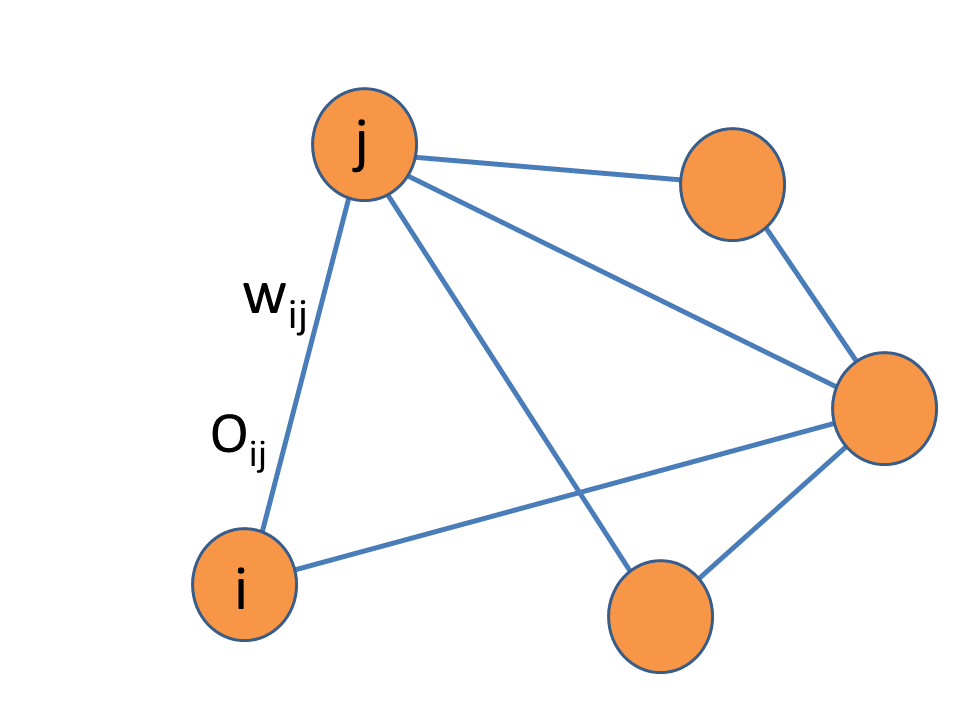}
\caption{In vector diffusion maps, the relations between data points are encoded by weights $w_{ij}$ and orthogonal transformations $O_{ij}$.}
\end{center}
\end{figure}

\begin{figure}[h]
\begin{center}
\subcaptionbox{$I_i$}{
\includegraphics[width=0.2\textwidth]{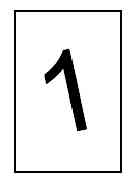}
}
\qquad
\subcaptionbox{$I_j$}{
\includegraphics[width=0.2\textwidth]{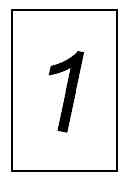}
}
\qquad
\subcaptionbox{$I_k$}{
\includegraphics[width=0.204\textwidth]{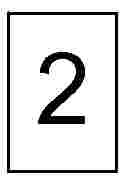}
}
\end{center}
\caption{An example of a weighted graph with orthogonal transformations: $I_i$ and $I_j$ are two different images of the digit one, corresponding to nodes $i$ and $j$ in the graph. $O_{ij}$ is the $2\times 2$ rotation matrix that rotationally aligns $I_j$ with $I_i$ and $w_{ij}$ is some measure for the affinity between the two images when they are optimally aligned. The affinity $w_{ij}$ is large, because the images $I_i$ and $O_{ij}I_j$ are actually the same. On the other hand, $I_k$ is an image of the digit two, and the discrepancy between $I_k$ and $I_i$ is large even when these images are optimally aligned. As a result, the affinity $w_{ik}$ would be small, perhaps so small that there is no edge in the graph connecting nodes $i$ and $k$. The matrix $O_{ik}$ is clearly not as meaningful as $O_{ij}$. If there is no edge between $i$ and $k$, then $O_{ik}$ is not represented in the weighted graph.}
\label{fig:digits}
\end{figure}

In general, we can think of an abstract group $G$ acting on the set $X$ where the data comes from. If $g_{ij}$ denotes the group element that aligns the data points best, and $G$ has an orthogonal representation $\rho$, then $O_{ij}$ may correspond to $\rho(g_{ij})$.

With this additional information we want to say that two data points are similar if they are connected by many of paths in the graph, but also if the transformations one obtains by traversing those paths and composing the $O_{ij}$ along the way are similar.

We make use of the transformations by constructing the following block matrix $W_1$:

\[W_1(i,j) = w_{ij}O_{ij}, \quad 1\leq i,j \leq n.\]

Note that $W_1(i,j)$ is a $d \times d$ matrix, thus $W_1$ is of size $nd \times nd$. We think of $W_1$ as having $n \times n$ blocks of size $d \times d$. The diagonal blocks of $W_1$ are set to $w_{ii}I$. Since $O_{ij} = O_{ji}^T$ (e.g., the matrix that aligns $i$ to $j$ is the inverse of the matrix that aligns $j$ to $i$), we see that $W_1$ is a symmetric matrix. We can make $W_1$ sparse by setting some blocks $W_1(i,j)$ equal to zero, for instance those with small affinity $w_{ij}$.

In analogy with diffusion maps, we also define the diagonal matrix $D_1$ of degrees of size $nd \times nd$:

\[D_1(i,i) = d_i I_{d \times d},\]
where $d_i = \sum_j w_{ij}$ is the node degree. We define the {\em graph connection Laplacian} as
$$L_1 = D_1 - W_1$$ \vfill

$$L_1(i,j) = \left\{\begin{array}{cc}
                    -w_{ij}O_{ij} & i\neq j, \\
                    \sum_{k\neq i} w_{ik}I_{d\times d} & i=j.
                  \end{array}\right. $$ 

The graph connection Laplacian $L_1$ can be applied to vectors $v$ of length $nd$, viewed as $n$ vectors in $\mathbb{R}^d$, i.e., $v = [v(1),\ldots,v(n)]^T$ with $v(i) \in \mathbb{R}^d$. Moreover, as a quadratic form $L_1$ satisfies 
$$v^T L_1 v = \frac{1}{2} \sum_{i,j=1}^n w_{ij} \|v(i) - O_{ij} v(j)\|^2$$
(the proof is left as an exercise). It follows that $L_1 \succeq 0$, and $L_1$ has a non-trivial null space iff there exists a non-zero $v$ such that $v(i) = O_{ij}v(j)$ for all $i\sim j$. We refer to $v^T L_1 v$ as the {\em frustration of $v$}, and can also define the {\em frustration of the connection graph} as the minimum over all $v$ with $\|v(i)\|=1$ for all $i$. It is possible to bound this frustration using the eigenvalues of graph connection Laplacian, which can be regarded as a generalization of Cheeger's inequality to the graph connection Laplacian \cite{Bandeira_Singer_Spielman_OdCheeger}.

The matrix $D_1^{-1}W_1$ is similar to the symmetric matrix $D_1^{-1/2}W_1D_1^{-1/2}$ via
\[D_1^{-1}W_1 = D_1^{-1/2}D_1^{-1/2}W_1D_1^{-1/2}D_1^{1/2}.\]
Therefore, as in diffusion maps, $D_1^{-1}W_1$ has a complete basis of eigenvectors, but now there are $nd$ of them. These vectors are of length $nd$, and we think about them in blocks of $d$. 

 Let us consider the spectral decomposition of the matrix $\tilde{W}_1=D_1^{-1/2}W_1D^{-1/2}$:

\[\tilde{W}_1(i,j) = \sum \limits_{l=1}^{nd} \lambda_lv_l(i)v_l(j)^T. \]

We emphasize that here the eigenvalues $\lambda_l$ are different from those of the random walk matrix $A = D^{-1}W$, in particular there are now $nd$ of them. From the above decomposition, we obtain as in the case of diffusion maps:

\[\left( D_1^{-1}W_1\right)(i,j) = \sum \limits_{l=1}^{nd} \lambda_lw_l(i)u_l(j)^T = \sum \limits_{l=1}^{nd} \lambda_lw_l(i)w_l(j)^Td_j, \]
where the vectors $w_l, u_l$ are defined by $w_l(i) = d_i^{-1/2}v_l(i)$ and $u_l(i) = d_i^{1/2}v_l(i)$. Then $D_1^{-1}W_1$ raised to the power $2t$ can be expressed as

\[\left( D_1^{-1}W_1\right)^{2t}(i,j) = d_j\sum \limits_{l=1}^{nd} \lambda_l^{2t} w_l(i)w_l(j)^T.\]

The matrix $\left( D_1^{-1}W_1\right)^{2t}$ plays a critical role in vector diffusion maps. Specifically, $\left( D_1^{-1}W_1\right)^{2t}(i,j)$ is obtained by considering all paths of length $2t$ between $i$ and $j$, and for each path summing the probability of traversing that path multiplied by the composition of transformations along the path. We want to define an affinity between our vertices so it depends on the `size' or `magnitude' of this matrix. If the transformations along several paths lead to similar transformations, then the overall magnitude will be large, while if the transformations cancel out, then the magnitude will be smaller. It turns out that the Frobenius (or Hilbert-Schmidt) norm

\[\left\| \frac{\left( D_1^{-1}W_1\right)^{2t}(i,j)}{d_j}\right\|_{\text{F}}^2\]
will be convenient for measuring the magnitude of that matrix. To understand this quantity, we use the formula derived above in terms of the eigenvalues and eigenvectors of $D_1^{-1/2}W_1D_1^{-1/2}$:

\[\left\|\sum \limits_{l=1}^{nd} \lambda_l^{2t} w_l(i)w_l(j)^T\right\|_{\text{F}}^2 =
\text{Tr}\left(\sum \limits_{l,r=1}^{nd} \lambda_l^{2t} \lambda_r^{2t}w_l(i)w_l(j)^T w_r(j)w_r(i)^T \right)\]

Now we try to simplify the terms in the sum. Note that $w_l(j)^T w_r(j) = \langle w_l(j), w_r(j) \rangle$ is a scalar. In addition, the trace function is linear, so we can bring it inside the summation, and for each term factor out the scalar. Finally, we also have $\text{Tr}(AB)=\text{Tr}(BA)$ for any two matrices, and we can rewrite the above as:
\[\left\|\sum \limits_{l=1}^{nd} \lambda_l^{2t} w_l(i)w_l(j)^T\right\|_{\text{F}}^2 = \sum \limits_{l,r=1}^{nd} (\lambda_l\lambda_r)^{2t} \langle w_l(i), w_r(i) \rangle \langle w_l(j), w_r(j) \rangle. \]
Therefore, we define the \emph{vector diffusion map} as
\[ V_t(i) : i \to \left((\lambda_l\lambda_r)^t\langle w_l(i), w_r(i) \rangle \right)_{l,r=1}^{nd}, \]
and we have the identity
\[\left\| \frac{\left( D_1^{-1}W_1\right)^{2t}(i,j)}{d_j}\right\|_{\text{F}}^2 = \langle V_t(i), V_t(j) \rangle. \]

The vector diffusion map is an embedding of the data in a real Euclidean space, just like the ordinary diffusion map. However, we note some differences. The dimension of the space into which the data is embedded is now $(nd)^2$, as opposed to $n$ for diffusion map. Also, the coordinates of the embedding are given by inner products of eigenvectors, not just their evaluations. In fact, one way to think about the embedding is that it takes all eigenvectors $w_1,\ldots,w_{nd}$, looks at their $i$-th block, and evaluates all possible inner products between them. There is some redundancy in the current definition, since the coordinates $(l,r)$ and $(r,l)$ always agree. This way we can get a reduction by a factor of almost two in the dimension of the embedded space.

From the interpretation of $\left( D_1^{-1}W_1\right)^{2t}(i,j)$ we see that $\langle V_t(i), V_t(j) \rangle$ gives us a notion of affinity between the vertices. We mention that we can also define vector diffusion distances $d_{\text{VDM},t}^2$ from the inner product $\langle V_t(i), V_t(j) \rangle$ as
$$d_{\text{VDM},t}^2 = \langle V_t(i), V_t(i) \rangle + \langle V_t(j), V_t(j) \rangle - 2 \langle V_t(i), V_t(j) \rangle$$

One apparently inconvenient feature of the vector diffusion map is its high dimensionality. However, the eigenvalues of $D_1^{-1}W_1$ help with this issue - indeed, all eigenvalues are at most one in absolute value (the proof is left as an exercise). Therefore, we can proceed as in diffusion maps, and truncate the embedding to keep only the largest eigenvalues. Note also that unlike diffusion maps, here it can happen that all eigenvalues have norm strictly less than one. Similar to diffusion maps, it is possible to use different normalizations of the graph connection Laplacian to define other normalized vector diffusion maps. For example, the mapping $i \to \left((\lambda_l\lambda_r)^t\langle v_l(i), v_r(i) \rangle \right)_{l,r=1}^{nd},$ (that uses $v$'s instead of $w$'s) corresponds to the unnormalized graph connection Laplacian. 

\subsection{Vector diffusion maps in manifold learning}

Next, we consider the manifold learning setup, where the data points are assumed to lie on a $d$-dimensional manifold $\mathcal{M}$ embedded in $\mathbb{R}^p$, where $d \ll p$. In this case, the transformations $O_{ij}$ can be obtained from the data, and below we outline a procedure for this task.

For simplicity, we assume that the dimension $d$ of the manifold is known in advance. If that is not the case, it is possible to estimate it from the local variability of the data set computed at different scales using the SVD at each point \cite{AVLittle_YMJung_MMaggioni_2009_ManDim}. 

\medskip
\noindent
{\textbf Obtaining the transformations $O_{ij}$:}
If the dimension of the manifold is $d$, we construct $d$-dimensional orthogonal transformations as follows. The first step is known as \emph{local PCA}. We choose a parameter $\epsilon_{PCA}$ and perform PCA for each point $x_i$ on its set of neighbors closer than $\sqrt{\epsilon_{PCA}}$ in Euclidean distance, i.e., using $x_j$'s for which $\|x_j - x_i\| \leq \sqrt{\epsilon_{PCA}}$. The first $d$ singular vectors are a numerical approximation of the tangent plane to the manifold at $x_i$. An asymptotic order of magnitude of $\epsilon_{PCA}$ for which this procedure works well can be derived in terms of $n$ and $d$ \cite{ASinger_HTWu_2011_VDM}. 

\begin{figure}
\begin{center}
\subcaptionbox{}{
\includegraphics[width=0.45\textwidth]{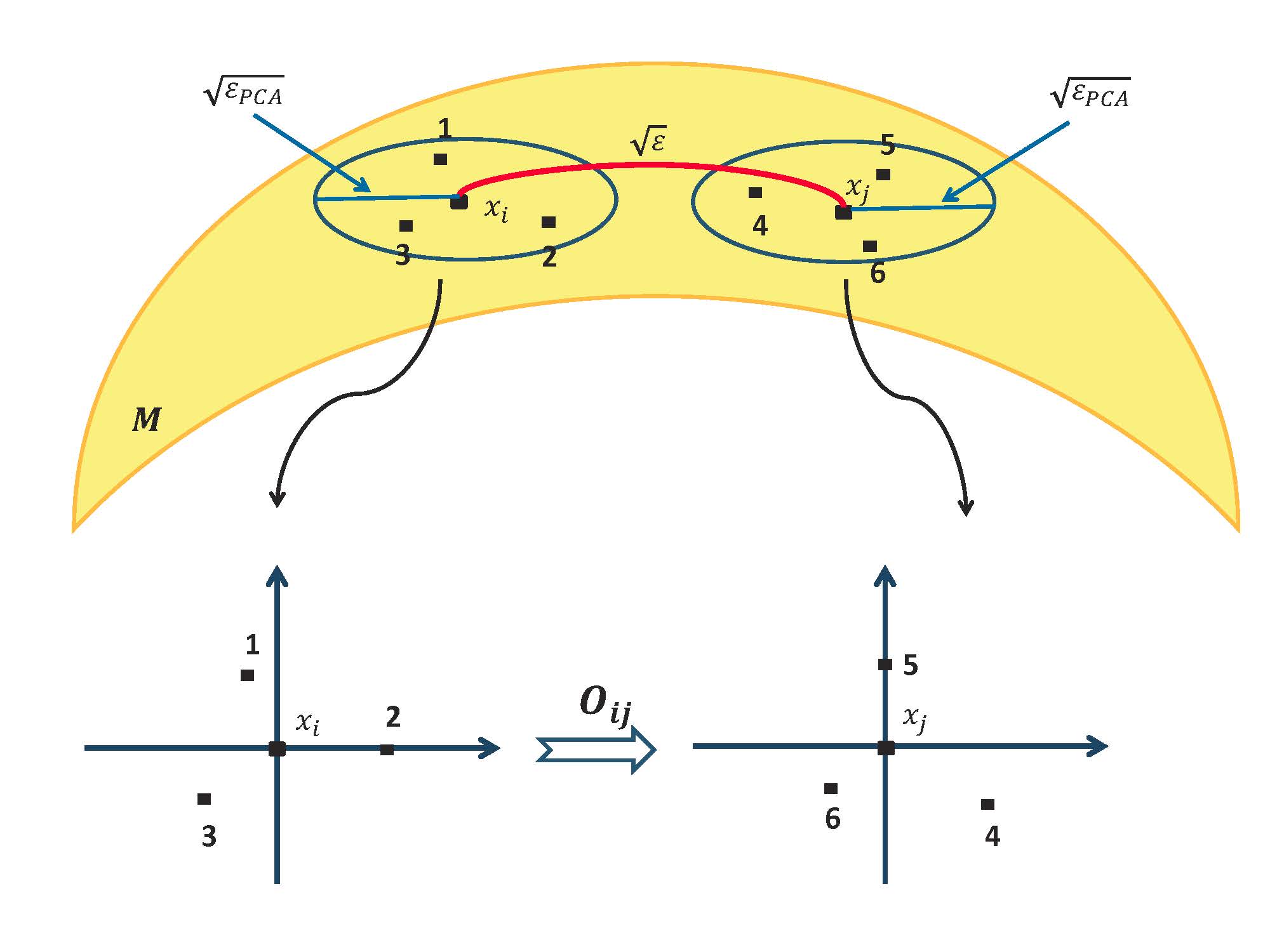}
}
\subcaptionbox{}{
\includegraphics[width=0.45\textwidth]{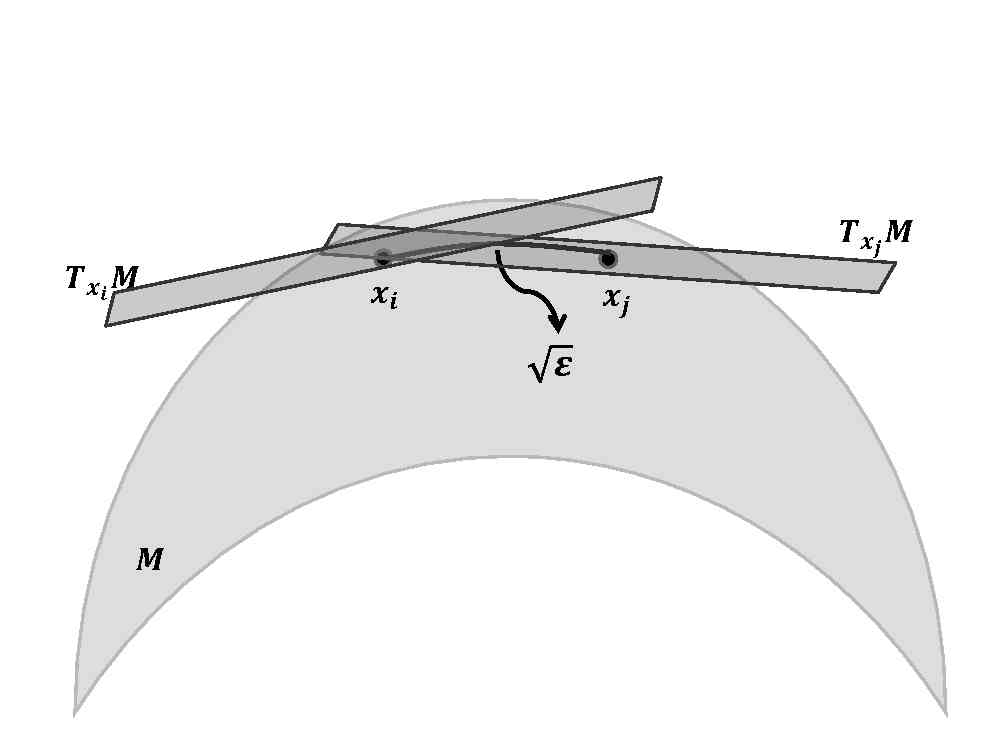}
}
\end{center}
\caption{The orthonormal basis of the tangent plane $T_{x_i}\MMM$ is determined by local PCA using data points
inside a Euclidean ball of radius $\sqrt{\epsilon_\text{PCA}}$ centered at $x_i$. The bases for $T_{x_i}\MMM$ and
$T_{x_j}\MMM$ are optimally aligned by an orthogonal transformation $O_{ij}$ that can be viewed as a mapping from
$T_{x_j}\MMM$ to $T_{x_i}\MMM$.}
\end{figure}

Let us denote by $O_i$ the $p \times d$ matrix obtained by local PCA at $x_i$. Then the columns of $O_i$ are orthonormal: $O_i^TO_i = I_{d\times d}$. Note that we could take any linear combination of the columns and still have the same approximation to the local tangent plane - however, the vector diffusion maps only depend on inner products of vectors, and these inner products are invariant to the choice of the basis.

Now we come to the second step, known as \emph{alignment}. Here we use the $O_i$ and $O_j$ for neighboring $x_i$ and $x_j$ to obtain the $d\times d$ orthogonal transformation $O_{ij}$. We consider another scale parameter $\epsilon$, much larger than $\epsilon_{PCA}$. To make our final matrix $W_1$ sparse, we can set to zero all entries $W_1(i,j)$ where $\|x_i-x_j\|>\sqrt{\epsilon}$. We consider only the remaining pairs $(i,j)$.

To see how the matrices $O_{ij}$ come about, we can look at the simple example when the manifold is $\mathbb{R}^2$ (i.e., flat manifold).  In that case, the matrices $O_i^TO_j$ are orthogonal, and show how to change basis from the local tangent plane at $x_j$ to the local tangent plane at $x_i$. In this case we can simply take $O_{ij} = O_i^TO_j$.

However, for manifolds with non-vanishing curvature, $O_i^TO_j$ will not be orthogonal in general. Still, we can define $O_{ij}$ as the closest orthogonal transformation to $O_i^T O_j$:

\[O_{ij} = \arg\min \limits_{O\in O(d)} \|O_i^TO_j - O\|_{\text{HS}}^2. \]
This optimization problem can be solved efficiently using the SVD of $O_i^TO_j$. Namely, if the SVD is $O_i^TO_j = U\Sigma V^T$, then $O_{ij} = UV^T$. Thus, we construct $O_{ij}$ in two steps, using local PCA and alignment. Using this construction of $W_1$, we can proceed to compute the vector diffusion maps as before.

\subsection{Parallel transport and the connection Laplacian}

Now we want to give some interpretations of the above construction in the limit as the number of points $n$ tends to infinity, and as $\epsilon \to 0$. It turns out that in this limit $O_{ij}$ approximates the \emph{parallel transport} operator from differential geometry.

Here we will give an intuitive explanation of the parallel transport operator. Consider two points $x_i$ and $x_j$ in the manifold $\mathcal{M}$. The parallel transport operator $P_{x_ix_j}$ is a linear mapping from the tangent space at $x_j$, denoted by $T_{x_j}$, to the tangent space $T_{x_i}$. Consider a geodesic between $x_i$ and $x_j$. If we take a vector in $T_{x_j}$, and slide it parallel to the geodesic to $T_{x_i}$, we get the image of the vector under parallel transport.

It is possible to show that $O_{ij}$ constructed above gives a good approximation to the parallel transport operator for nearby points. We mention that obtaining the best orthogonal estimate as opposed to simply using $O_i^TO_j$ as an approximation is crucial for deriving convergence theorems \cite{ASinger_HTWu_2011_VDM}. 

The operator $D_1^{-1}W_1$ can be viewed as an averaging operator over vector fields. Indeed,

\[\left(D_1^{-1}W_1 v\right)(i) = \frac{1}{d_i}\sum_{j\sim i} w_{ij}O_{ij}v(j).\]

Thus, each vector $v(j)$ is brought from the tangent plane $T_{x_j}$ to the tangent plane $T_{x_i}$ via parallel transport, and then weighted averaged by $w_{ij}$.

We previously obtained the large sample limits for diffusion maps. Under uniform sampling, the discrete graph Laplacian converges to a differential operator known as the Laplace-Beltrami operator on the manifold in the limit of infinite sample. In that setup, the operator was acting on scalar functions on the manifold. As mentioned above, we now think of our vectors as having $n$ entries of dimension $d$. Thus, in this case, the limiting operator will act on tangent vector fields (or one-forms) on the manifold. We mention that it is possible to devise similar constructions for higher order forms as well by taking tensor powers of $O_{ij}$.

For diffusion maps, we have the convergence result 
\[\left(D^{-1}W - I\right)f  \to c\Delta f\]
for sufficiently smooth functions $f$. In vector diffusion maps, we have an analogous statement:
\[\left(D_1^{-1}W_1 - I\right)X  \to c\nabla^2 X\]
where $X$ is a vector field and $\nabla^2$ denotes the \emph{connection Laplacian} over the manifold. 

\subsection{Orientability and double cover} 

Vector diffusion maps can be used to obtain topological information about the underlying manifold. For instance, orientability can be detected by forming the signed matrix of $\det(O_{ij})$ and considering its eigenvectors and eigenvalues \cite{ASinger_HTWu_2011_OrDM}. Specifically, we encode the information about reflections in a symmetric $n \times n$ matrix $Z$ with entries
\begin{equation*}
Z_{ij} = \left\{\begin{array}{ccl} \det O_{ij} & & (i,j)\in E, \\ 0 & & (i,j)\notin E.
\end{array} \right.
\end{equation*}
That is, $Z_{ij}=1$ if no reflection is needed, $Z_{ij}=-1$ if a reflection is needed, and $Z_{ij}=0$ if the points are not nearby.
We then normalize $Z$ by the node degrees, i.e., consider the eigenvalues of $D^{-1}Z$. 

If the manifold is orientable, the matrix can be ``synchronized" in the sense that we can find a vector of signs $\{z_i\}_{i=1}^n$ such that
$$\det(O_{ij}) = z_i z_j$$ 
(equivalently, using our previous notation, the graph connection Laplacian has no frustration). This recovers the orientations of the local bases. Otherwise, it can be used to construct an embedding of the orientable double cover in Euclidean space. Thus, orientability can be detected with vector diffusion maps, which look only at local information about the data set.

\begin{figure}[h]
\begin{center}
\subcaptionbox{$S^2$}{
\includegraphics[width=0.3\textwidth]{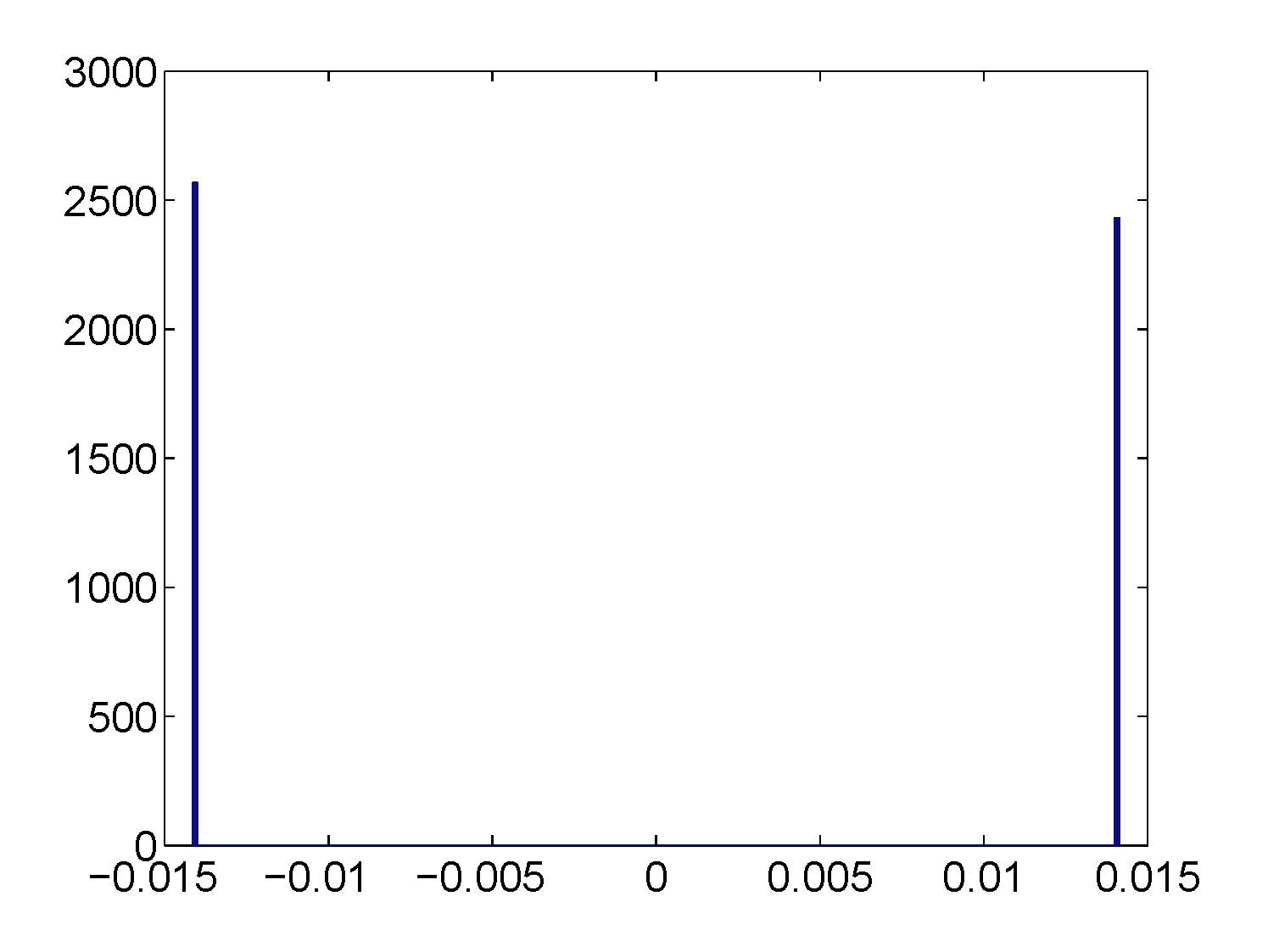}
}
\subcaptionbox{Klein bottle}{
\includegraphics[width=0.3\textwidth]{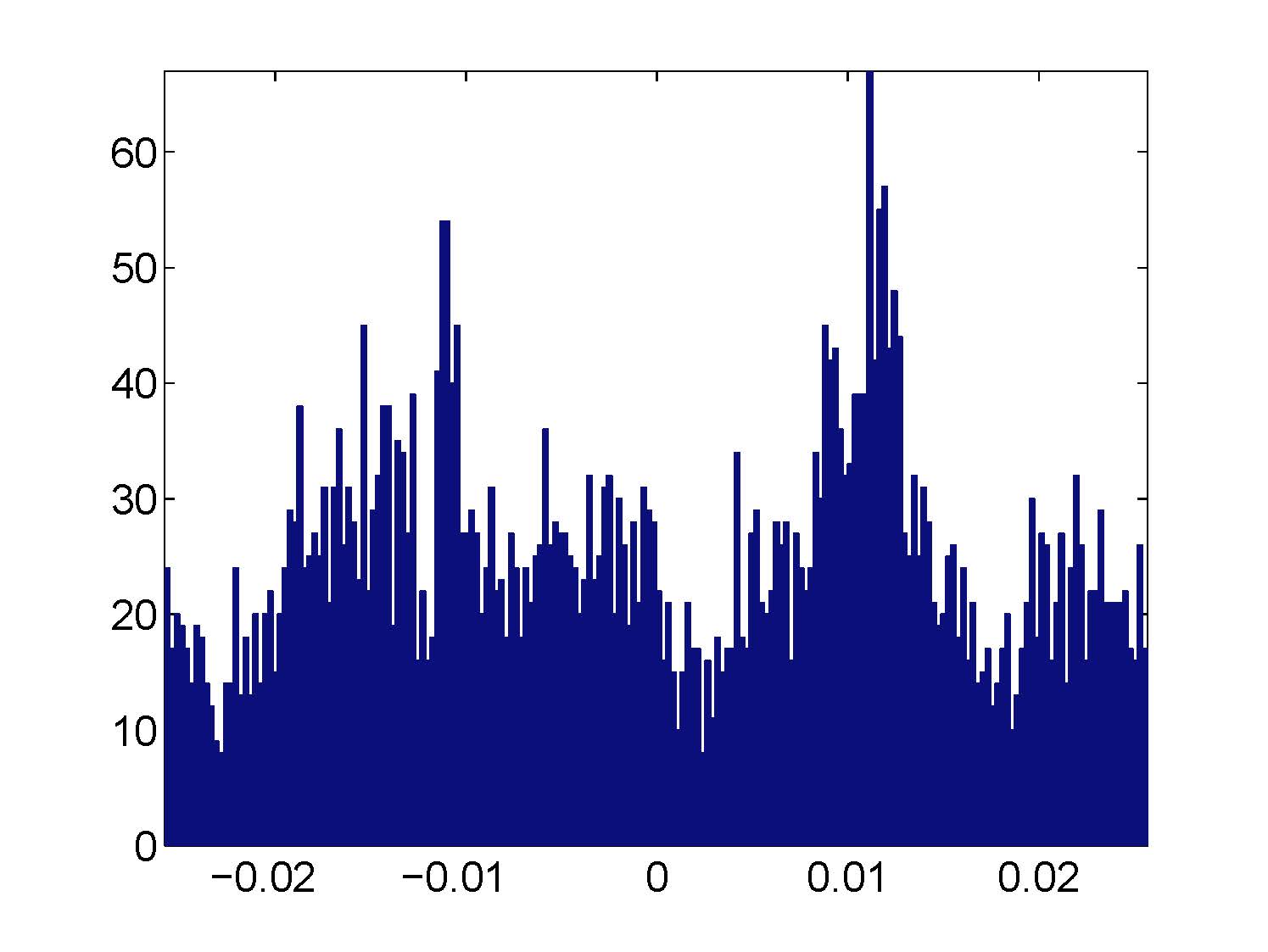}
}
\subcaptionbox{$\mathbb{R}P^2$}{
\includegraphics[width=0.3\textwidth]{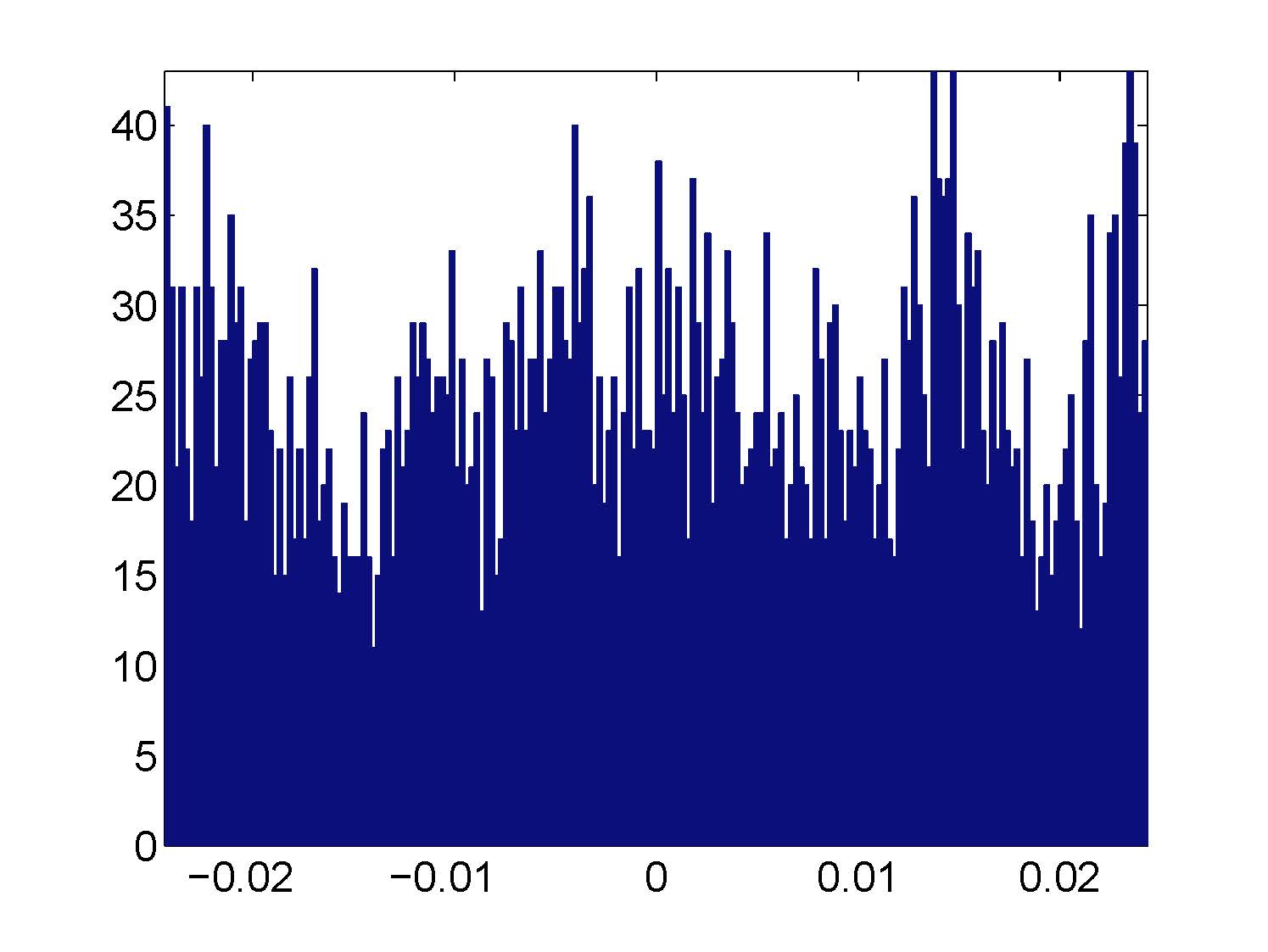}
}
\caption{Histogram of the values of the top eigenvector of $D^{-1}Z$.}
\label{fig:vectordiffusion1}
\end{center}
\end{figure}

Embedding obtained using the eigenvectors of the (normalized) matrix gives the double cover of non-orientable manifolds. 
\begin{equation*}
\left[
\begin{array}{rr}
Z & -Z\\
-Z & Z
\end{array}
\right] = \left(
            \begin{array}{rr}
              1 & -1 \\
              -1 & 1 \\
            \end{array}
          \right) \otimes Z,
\end{equation*}

\begin{figure}[h]
\begin{centering}
\subcaptionbox{}{
\includegraphics[width=0.3\textwidth]{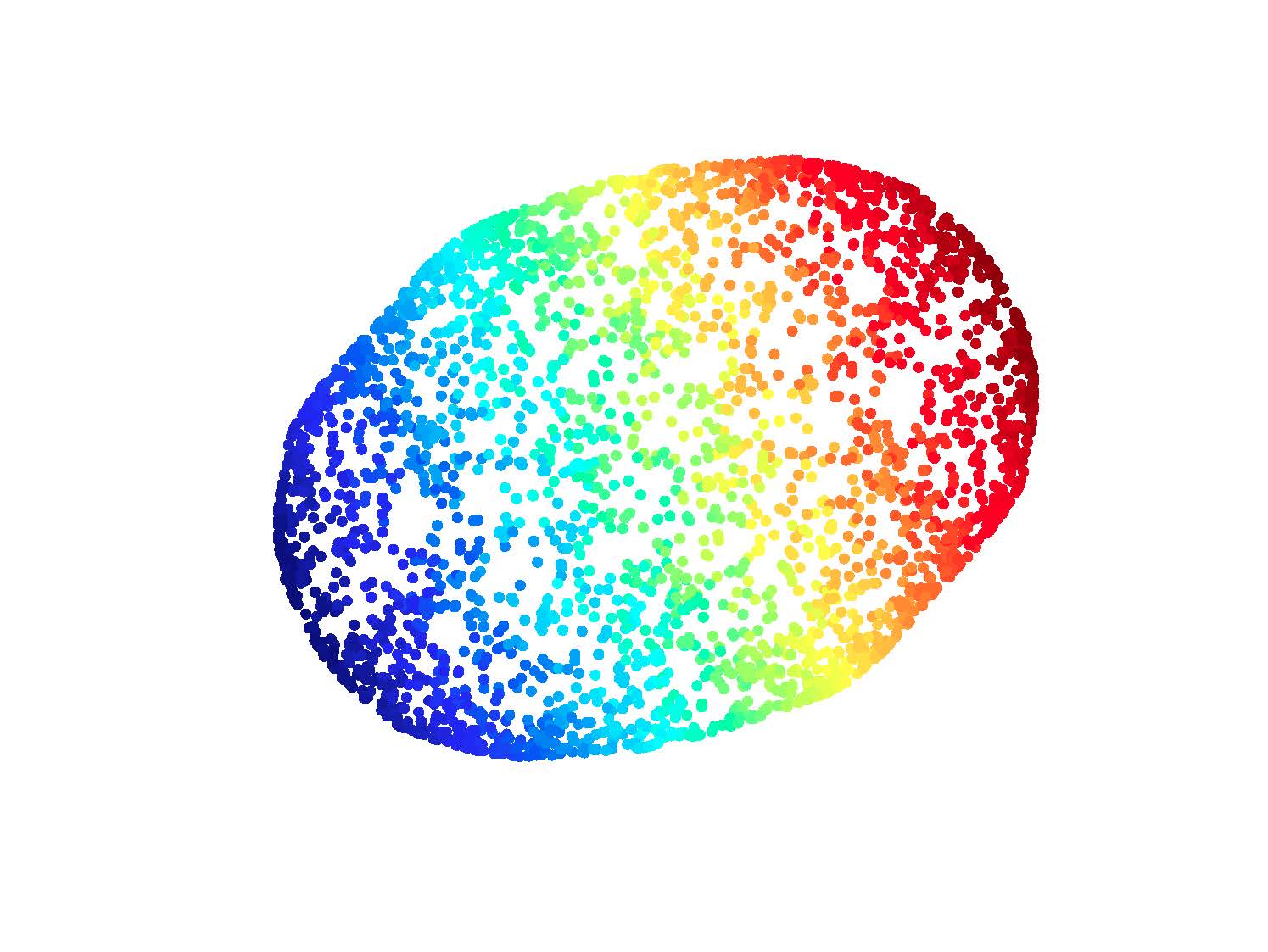}
}
\subcaptionbox{}{
\includegraphics[width=0.3\textwidth]{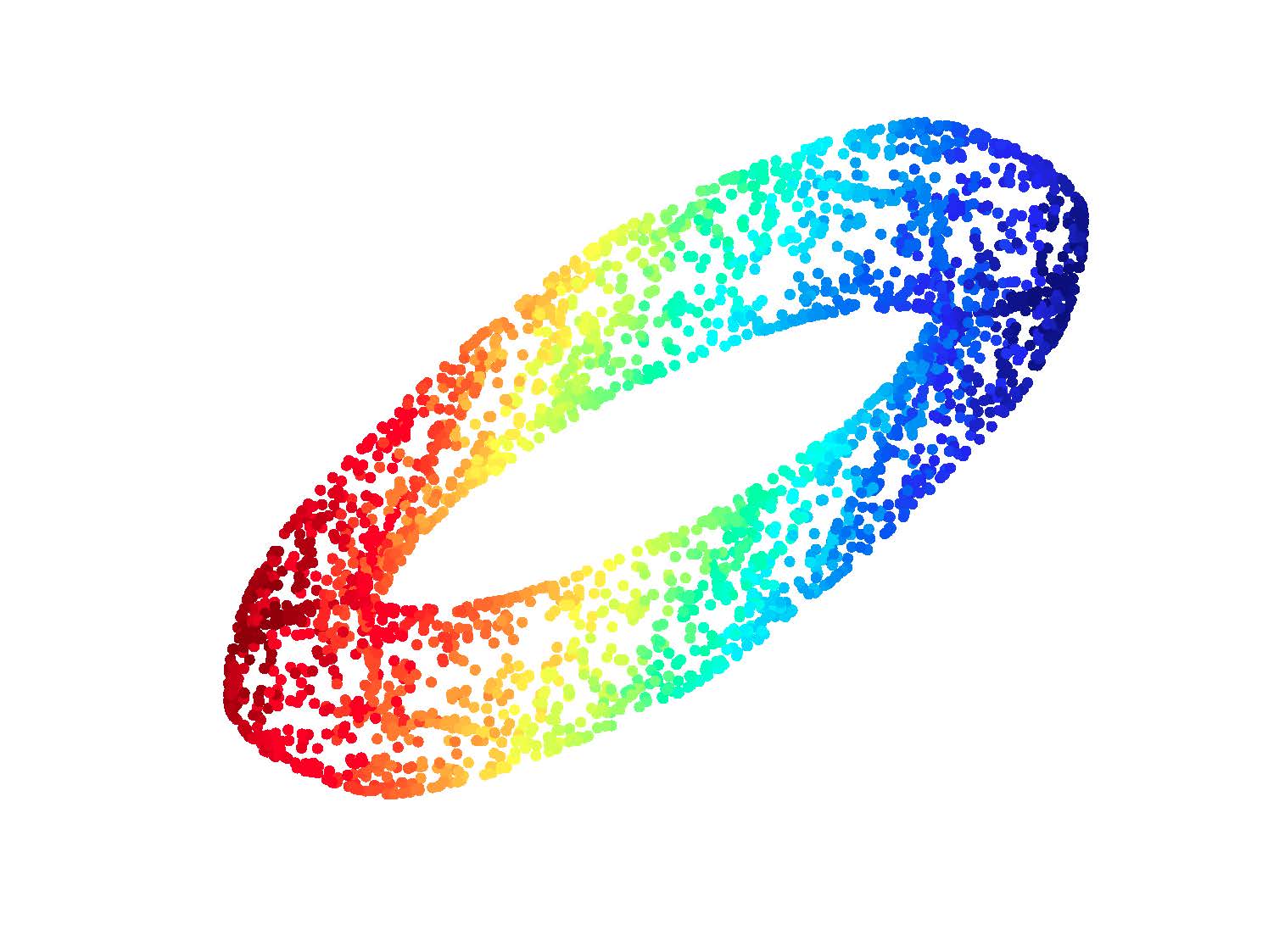}
}
\subcaptionbox{}{
\includegraphics[width=0.3\textwidth]{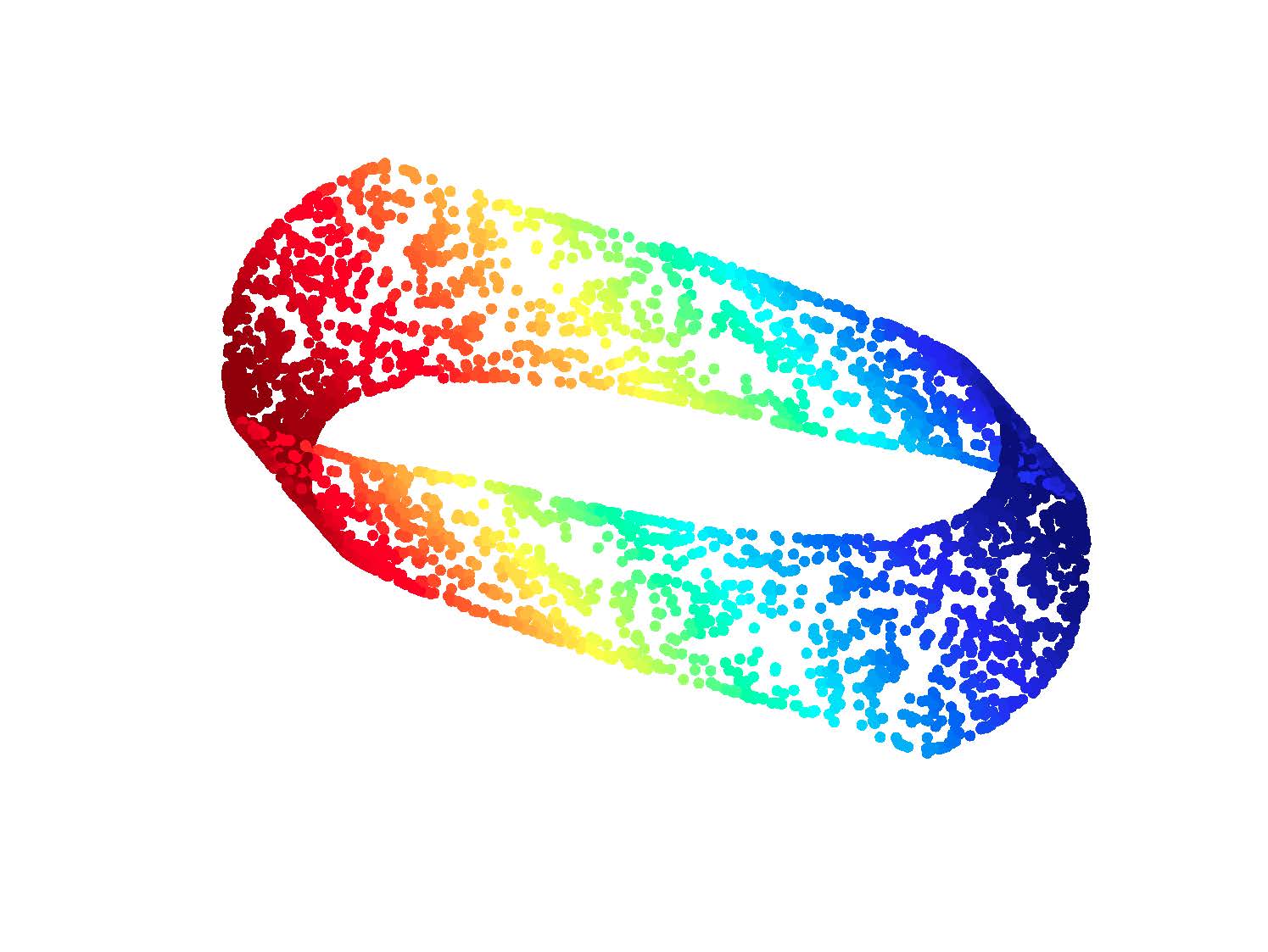}
}
\end{centering}
\caption{Left: the orientable double covering of $\mathbb{R} P(2)$, which is $S^2$; Middle: the orientable double covering of the Klein bottle, which is $T^2$; Right: the orientable double covering of the M\"obius strip, which is a cylinder.}
\label{double_rp2_klein}
\end{figure}

We remark that learning the topology of a manifold from random samples \cite{niyogi2008finding} is still an active area of research. Beyond the graph Laplacian and the connection Laplacian, other Laplacian operators, namely, the Hodge Laplacians reveal topological information such as the Betti numbers (number of connected components, number of holes, etc.) through their harmonic forms. Readers interested in topological data analysis are referred to \cite{carlsson2009topology} as a starting point for this topic.

\section*{Exercises}
\addcontentsline{toc}{section}{Exercises}

\begin{myexercise}
Find an `optimal kernel' $K$ that minimizes the variance error term for a given bias error of the density estimator (\ref{eq:density_estimation_MSE}). That is, find $K$ that minimizes
$$
\min_{K} \int_\mathbb{R} K^2(z)\,dz
$$
such that $K \geq 0$, $\int K(z)\,dz = 1$, $K(z) = K(-z)$, and $\int_{\mathbb{R}}K(z)z^2\,dz = m_2$.    
\end{myexercise}

\begin{myexercise}
In the uniform sampling case, show that if $L_{rw}\phi = \lambda \phi$, as $n\rightarrow \infty$ followed by $\varepsilon \to 0$, then $\phi$ satisfies the boundary condition $\frac{\partial \phi}{\partial \nu} = 0$ $\forall x \in \partial \mathcal{M}$ where $\nu$ is the normal vector to the boundary $\partial \mathcal{M}$ at $x$.\\
\begin{hint}
Start with the 1D case where $\mathcal{M}$ is the interval $[0,L]$ and observe that at the endpoints (i.e., the boundary) the integral of the kernel function $K_{\eps}$ cannot be approximated any longer as a Dirac delta function plus an $O(\varepsilon)$ Laplacian (second derivative) correction term as we saw for interior points. At the boundary only half of the kernel function gets to be integrated, which can be viewed as half a Dirac delta function, and we also get an $O(\sqrt{\varepsilon})$ correction term involving the {\em first derivative}, because the integral involving the linear term does not vanish as it did before. Argue that the $O(\sqrt{\varepsilon})$ term must vanish which leads to the no-flux boundary condition. A rigorous treatment requires an asymptotic analysis in powers of $\sqrt{\varepsilon}$ and matching the outer solution to the boundary layer.   
\end{hint}    
\end{myexercise}

\begin{myexercise}
Suppose the normalization of $w_{ij}$ is
\[
w_{ij}^{(\alpha)} = \frac{w_{ij}}{d_i^\alpha d_j^\alpha}
\]
for $\alpha \in [0,1]$. That is,
$$W^{(\alpha)} = D^{-\alpha} W D^{-\alpha}$$
and define the corresponding diagonal matrix
$$D^{(\alpha)}_{ii} = \sum_{j=1}^n w^{(\alpha)}_{ij}$$
What are the limiting operators for $L^{(\alpha)}$ and $L_{rw}^{(\alpha)}$?
For the random walk Laplacian we know that for $\alpha=0$ we get the backward Fokker-Planck operator, and for $\alpha=1$ we get the Laplace-Beltrami operator. What is the limiting operator for $\alpha$ between 0 and 1, and specifically for $\alpha=\frac{1}{2}$?  
\end{myexercise}

\begin{myexercise}[\level\level\sep Spherical harmonics]
Find the eigenfunctions and eigenvalues of the Laplacian on the sphere $\mathbb{S}^2 \subset \mathbb{R}^3$.
 What are the multiplicities of the eigenvalues? (Hint: use the chain rule to get the Laplacian in spherical coordinates $(r,\theta,\phi)$ and find its form for functions that are independent of $r$).    
\end{myexercise}

\begin{myexercise}[\level\level\sep Hermite polynomials]
The density of a one-dimensional Gaussian is $$p(x) = \frac{1}{\sqrt{2\pi\sigma^2}}e^{-\frac{x^2}{2\sigma^2}},$$ and the corresponding potential function is $$U(x)=-2\log p(x) = \frac{x^2}{\sigma^2} + const.$$  Find the eigenfunctions $\phi_n(x)$ and eigevalues $\lambda_n$ of the backward Fokker-Planck operator
    $$\phi_n'' - \frac{2x}{\sigma^2}\phi_n' = -\lambda_n \phi_n.$$
\end{myexercise}

\begin{myexercise}[\level\level\sep Line segment + noise]
Consider $n$ data points $z_1,z_2,\ldots,z_n$ in $\mathbb{R}^2$ sampled independently from the probability density function $p(x,y)$ (to avoid confusion, we denote the first coordinate of $z\in \mathbb{R}^2$ by $x$ and the second coordinate by $y$, that is, $z=(x,y)$):
$$p(x,y) = p_1(x)p_2(y),$$
where
$$p_1(x) = \left\{\begin{array}{cc}
\frac{1}{L} & 0\leq x \leq L \\
0 & otherwise \\
\end{array}\right.$$
and
$$p_2(y) = \frac{1}{\sqrt{2\pi\sigma^2}}e^{-\frac{y^2}{2\sigma^2}}.$$
In other words, the $x$-coordinate is uniformly distributed on the interval $[0,L]$ and the $y$-coordinate has a normal distribution with zero mean and standard deviation $\sigma$.

\begin{enumerate}
\item Find the eigenfunctions and eigenvalues of the backward Fokker-Planck operator
$$\Delta f - \nabla U \cdot \nabla f = -\lambda f,$$
where $U(x,y) = -2\log (p_1(x)p_2(y))$ and $\Delta = \frac{\partial^2}{\partial x^2} + \frac{\partial^2}{\partial y^2}$.
\begin{hint}Separation of variables. 
\end{hint}

\item Find the condition (in terms of $\sigma$ and $L$) for which the first non-trivial eigenfunction is a function of $x$ rather than a function of $y$. In other words, find a condition that ensures that the first coordinate of diffusion maps (in the limit of $n\to \infty$) would coincide with the $x$ coordinate (``the intrinsic coordinate"), rather than the $y$ coordinate (``the direction of noise").

\item Find the $2 \times 2$ true covariance matrix $\Sigma = \mathbb{E}(Z-\mathbb{E}Z)(Z-\mathbb{E}Z)^T$ that corresponds to the density function $p(x,y)$. Find the condition (in terms of $\sigma$ and $L$) for which the first principal component coincides with the $x$ axis and the second principal component coincides with the $y$ axis. Compare your result to part (2) and try to explain why one of the methods succeeds in finding the intrinsic axis before the other.

\end{enumerate}

\end{myexercise}
\begin{myexercise}
Show that the first nontrivial eigenfunction of the second order ordinary differential equation $$f^{\prime \prime} - U^\prime f^\prime = -\lambda f$$ 
is monotonic. Conclude that the first nontrivial eigenfunction basically orders the data points sampled from an open curve according to its arclength, i.e. $x_i \rightarrow \phi(x_i)$ is an embedding of the high dimensional data points in 1-D.
 
\begin{hint}
Use the Pr\"ufer transformation (see \cite{coddington_levinson}). 
\end{hint}
\end{myexercise}

\begin{myexercise}
For the Fokker-Planck operator in 1-D with periodic boundary conditions, show that we get an embedding that encircles the origin exactly once.
Also show that the angle that the two eigenfunctions make for the different points on the curve is monotonic.
\begin{hint}
Again, the key to this proof is the Pr\"ufer transformation.   
\end{hint}
\end{myexercise}

\begin{myexercise}
Find the heat kernel for diffusion in the finite interval $[0,L]$ with Neumann boundary conditions at the endpoints. 
\begin{hint}
The method of images with an infinite number of images. 
\end{hint}     
\end{myexercise}

\begin{myexercise}
The Riemann $\zeta$-function is defined as
$$\zeta(s) = \sum_{n=1}^\infty \frac{1}{n^s}.$$
Define the $\zeta$-function of a domain (or, in general, a manifold) as
$$\zeta(x,y,s) = \sum_{n=1}^\infty \frac{1}{\lambda_n^s}\phi_n(x)\phi_n(y).$$
That is, $\zeta(x,y,s)$ is the Green function for $(-\Delta)^s$ with Dirichlet boundary conditions.
Show that $\zeta(x,y,s)$ is given by the Mellin transform of the heat kernel, that is, show:
$$\Gamma(s) \zeta(x,y,s) = \int_0^\infty G_t(x,y) t^{s-1}\,dt,$$
where $\Gamma(s)$ is the Gamma function. Can you give a probabilistic interpretation of the right-hand-side integral of the heat kernel in terms of moments of first passage/hitting times?    
\end{myexercise}

\begin{myexercise}
Consider the commute-time embedding of a domain $\Omega \subset \mathbb{R}^d$:
$$x \mapsto (\frac{1}{\sqrt{\lambda_2}}\phi_2(x),\frac{1}{\sqrt{\lambda_3}}\phi_3(x),\ldots ),$$
where $\Delta \phi_n = -\lambda_n \phi_n$.
Explain the failure of this embedding for $d\geq 2$ in light of the Green functions $G(r)$ for Poisson's equation from electrostatics $\Delta u = f$ (i.e., $G(r)=\frac{1}{2\pi}\log r$ for $d=2$ and $G(r)=-\frac{1}{4\pi r}$ for $d=3$). Find the distance $D^2(x,y) = G(x,x)+G(y,y)-2G(x,y)$, where $G(x,y)$ is the Green's function for the Poisson equation on the interval $[0,L]$ and observe its relation to the geodesic distance.
\end{myexercise}

\begin{myexercise}[\level\level\sep Biharmonic distance]
In computer graphics $2$-dimensional surfaces in $\mathbb{R}^3$ are very popular and one often wants to have distances that resemble the geodesic distances. It was suggested in \cite{Lipman:2010:BD} to use the embedding
$$x \mapsto (\frac{1}{\lambda_2}\phi_2(x),\frac{1}{\lambda_3}\phi_3(x),\ldots ),$$
that corresponds to the Green function of the biharmonic equation $$\Delta^2 u = f.$$
Find the Green function for the biharomic equation in $\mathbb{R}^2$  and derive the squared distance $D^2(x,y)$ $(x,y\in \mathbb{R}^2)$. To what extent does this distance resembles the geodesic distance $\|x-y\|$?    
\end{myexercise}

\begin{myexercise}
Show that the graph connection Laplacian $L_1 = D_1 - W_1$ satisfies 
$$v^T L_1 v = \frac{1}{2} \sum_{i,j=1}^n w_{ij} \|v(i) - O_{ij} v(j)\|^2$$
for any $v \in \RR^{nd}$ viewed as $n$ vectors $v(1),\ldots, v(n) \in \RR^d$.
\end{myexercise}

\begin{myexercise}
Show that if $\lambda$ is an eigenvalue of $\tilde{W}_1=D_1^{-1/2}W_1D^{-1/2}$ then $-1 \leq \lambda \leq 1$.     
\end{myexercise}

\begin{myexercise}[\level\level\sep Vector spherical harmonics]
Use a computer simulation to generate $n$ points on the unit sphere $S^2 \subset \mathbb{R}^3$ and compute the eigenvalues of $D_1^{-1}W_1$ (use $\epsilon_{\text{PCA}}$ and $\epsilon$ as you see fit). What are the multiplicities of the eigenvalues? The corresponding eigen-vector-fields are known as the ``vector spherical harmonics'' and are routinely used in geophysics and other applications. Try to derive the vector spherical harmonics from the scalar spherical harmonics by taking their gradient and projecting to the tangent plane to the sphere.    
\end{myexercise}

\begin{myexercise}[\level\level\sep $\mathbb{R}P^2$, orientability and double-cover]
The real projective plane $\mathbb{R}P^2$ is the space of lines in $\mathbb{R}^3$ passing through the origin. $\mathbb{R}P^2$ can be regarded as the quotient of the two-sphere $$S^2 = \{(x,y,z) \in \mathbb R^3 : x^2+y^2+z^2 = 1\}$$ by the antipodal relation $$(x,y,z)\sim (-x,-y,-z).$$ It is a two-dimensional non-orientable manifold that cannot be embedded in $\mathbb{R}^3$ without intersecting itself, but it can be embedded in $\mathbb{R}^4$ and higher-dimensional Euclidean spaces.
\begin{enumerate}
\item Suppose we sample $n$ points from the uniform distribution on $S^2 \subset \mathbb{R}^3$ denoted $x_1,x_2,\ldots,x_n$ and define affinities $w_{ij}$ between them as
$$w_{ij} = K(|\langle x_i , x_j \rangle |),$$
where $K$ is a positive function, e.g., $$w_{ij} = |\langle x_i , x_j \rangle |^2, \quad \mbox{or} \quad w_{ij} = \exp\left\{\frac{1-|\langle x_i , x_j \rangle |^2}{\epsilon}\right\},$$ for some $\epsilon>0$. Determine the eigenfunctions and eigenvalues of the corresponding diffusion map matrix in the limit $n\to \infty$. 
\begin{hint}Distinguish between odd and even functions on the sphere, and observe that the limiting operator commutes with reflection.
\end{hint}

\item Explain how to obtain an embedding $\Phi$ of $\mathbb{R}P^2$ in $\mathbb{R}^5$ using the diffusion map eigenfunctions from above.

\item Use a computer simulation to embed $\mathbb{R}P^2$ in $\mathbb{R}^5$ using the above construction. Then, perform local PCA and alignment to construct $O_{ij}$ for vector diffusion map on $\mathbb{R}P^2$ (choose $\epsilon_{\text{PCA}}$ and $\epsilon$ as you see fit). Determine the multiplicities of the eigenvalues of the vector diffusion map matrix $D_1^{-1}W_1$. Can you explain your numerical findings?

\item Use a computer simulation to construct the $n\times n$ symmetric matrix $Z$ with
$$Z_{ij} = \det O_{ij}, \quad \mbox{for } \|\Phi(x_i)-\Phi(x_j)\| < \epsilon,$$
and
$$Z_{ij} = 0, \quad \mbox{for } \|\Phi(x_i)-\Phi(x_j)\| \geq \epsilon,$$
where $\Phi$ is the diffusion map embedding in $\mathbb{R}^5$. That is, the elements of $Z$ are either $1$, $-1$ or $0$.
Determine that $\mathbb{R}^2$ is non-orientable by inspecting the top eigenvalue and eigenvector of $D^{-1}Z$ (where $D$ is a diagonal $n\times n$ matrix with the weighted degrees on its diagonal).

\item Use a computer simulation to construct the $2n \times 2n$ symmetric matrix $\tilde{Z}$ given by $$\tilde{Z} = Z \otimes \left(
                                                                               \begin{array}{rr}
                                                                                 1 & -1 \\
                                                                                 -1 & 1 \\
                                                                               \end{array}
                                                                             \right).$$
Observe the numerical multiplicities of $\tilde{D}^{-1}\tilde{Z}$ (where $\tilde{D}$ is a $2n\times 2n$ diagonal matrix with the weighted degrees).

\item $S^2$ is the orientable double-cover of $\mathbb{R}P^2$. Use the eigenfunctions of $\tilde{D}^{-1}\tilde{Z}$ to embed $S^2$ in $\mathbb{R}^3$. Notice that we started with $n$ points sampled from $\mathbb{R}P^2$ embedded in $\mathbb{R}^5$ and ended up with $2n$ points from its double cover $S^2$ in $\mathbb{R}^3$. Explain how this procedure can be generalized to find the embedding of the double covering of any non-orientable manifold.

\end{enumerate}

\end{myexercise}


\chapter{Community \kern-1.2pt Detection and the \kern-1.5pt   Power \kern-1.2pt of Convex \kern-1.2pt Relaxation}
\label{c:community}

This chapter presents the idea of a convex relaxation, illustrated via problems on graphs. We start with the introduction of the Goemans-Williamson semidefinite relaxation for the Max-Cut problem, and move on to the problem of community detection in the Stochastic Block Model.

\section{Max-Cut, lifting, and approximation algorithms}
\label{c:maxcut}\label{maxcutapprox}

Many data analysis tasks include in them a step consisting of solving a computational problem, oftentimes in the form of finding a hidden parameter that best explains the data, or model specifications that provide best-fits. Many such problems, including examples in previous chapters, are computationally intractable. In complexity theory this is often captured by $NP$-hardness. Unless the widely believed $P\neq NP$ conjecture is false, there is no polynomial algorithm that can solve all instances of an NP-hard problem. Thus, when faced with an NP-hard problem (such as the \emph{Max-Cut} problem discussed below) one has three natural options: to use an exponential type algorithm that solves exactly the problem in all instances, to design polynomial time algorithms that only work for some of the instances (hopefully relevant ones), or to design polynomial algorithms that, in all instances, produce guaranteed approximate solutions. 

This section is about the third option, another example of this approach is the earlier discussion on Spectral Clustering and Cheeger's inequality. The second option, of designing algorithms that work in many, rather than all, instances is discussed later in this chapter, notably these goals are often achieved by the same algorithms.

The \emph{Max-Cut} problem is defined as follows: Given a graph $G=(V,E,W)$ with non-negative weights $w_{ij}$ on the edges, find a set $S\subset V$ for which $\text{cut}(S)$ is maximal.\footnote{The problems of maximizing and minimizing the cut are related if one considers the complement graph. As we saw in Chapter~\ref{c:graphs}, if we want to cluster the nodes of a graph (by minimizing the cut) we have to add a mechanism that balances the size of clusters. The Max-Cut problem does not suffer from this issue, as the partitions that maximize the cut tend to be well balanced. In fact, this makes the Max-Cut problem conceptionally closer to clustering (and community detection) than the Min-Cut problem, making it an excellent theoretical testbed for algorithms and analysis. Furthermore, it also enjoys several direct applications.}
Goemans and Williamson~\cite{MXGoemans_DPWilliamson_1995} introduced an approximation algorithm that runs in polynomial time, has a randomized component in it, and is able to obtain a cut whose expected value is guaranteed to be no smaller than a particular constant $\alpha_{GW}$ times the optimum cut. The constant $\alpha_{GW}$ is referred to as the approximation ratio.

Let $V=\{1,\ldots,n\}$. One can restate \texttt{Max-Cut} as
\begin{equation}\label{MAXCUTproblem}
\begin{array}{l}
\max \quad  \frac12 \sum_{i<j}w_{ij}(1-y_iy_j) \\
s.t. \quad \quad  |y_i|=1
\end{array}
\end{equation}
The $y_i$'s are binary variables that indicate set membership, i.e., $y_i=1$ if $i\in S$ and $y_i=-1$ otherwise.

Consider the following relaxation of this problem:
\begin{equation}\label{MAXCUTproblem_Relaxed}
\begin{array}{l}
\max \quad  \frac12 \sum_{i<j}w_{ij}(1-u_i^Tu_j) \\
s.t. \quad \quad  u_i\in \mathbb{R}^n, \|u_i\|=1.
\end{array}
\end{equation}
This is a relaxation because if we restrict $u_i$ to be a multiple of $e_1$, the first element of the canonical basis, then (\ref{MAXCUTproblem_Relaxed}) is equivalent to \eqref{MAXCUTproblem}. For this to be a useful approach, the following two properties should hold:
 \begin{enumerate}[{(a)}]

 \item \label{it-1} Problem \eqref{MAXCUTproblem_Relaxed} is easy to solve.
 \item \label{it-2} The solution of \eqref{MAXCUTproblem_Relaxed} is, in some way, related to the solution of \eqref{MAXCUTproblem}.
 \end{enumerate}
 
 \begin{definition}\label{def_maxcut_rmaxcut}
 Given a graph $G$, we define $\mathrm{MaxCut}(G)$ as the optimal value of~\eqref{MAXCUTproblem} and $\mathcal{R}\mathrm{MaxCut}(G)$ as the optimal value of~\eqref{MAXCUTproblem_Relaxed}.
 \end{definition}
 
We start with property (\ref{it-1}). Set $X$ to be the Gram matrix of $u_1,\ldots,u_n$, that is, $X=U^TU$ where the $i$'th column of $U$ is $u_i$. We can rewrite the objective function of the relaxed problem as $$\frac12\sum_{i<j}w_{ij}(1-X_{ij})$$ One can exploit the fact that $X$ having a decomposition of the form $X=Y^TY$ is equivalent to being positive semidefinite, denoted $X\succeq 0$. The set of PSD matrices is a convex set. Also, the constraint $\|u_i\|=1$ can be expressed as $X_{ii}=1$. This means that the relaxed problem is equivalent to the following semidefinite program (SDP):

\begin{equation}\label{MAXCUTproblem_SDP}
\begin{array}{l}
\max \quad  \frac12 \sum_{i<j}w_{ij}(1-X_{ij}) \\
s.t. \quad \quad X\succeq 0 \text{ and } X_{ii}=1,\; i=1,\ldots,n.
\end{array}
\end{equation}
This SDP can be solved (up to $\epsilon$ accuracy) in time polynomial on the input size and $\log(\epsilon^{-1})$~\cite{LVanderberghe_SBoyd_1996} (It is a convex optimization problem, which were discussed in Chapter~\ref{c:optimization}; note however that in this chapter optimization problems are more often formulated as maximization problems).

There is an alternative way of viewing \eqref{MAXCUTproblem_SDP} as a relaxation of \eqref{MAXCUTproblem}. By taking $X = yy^T$ one can formulate a problem equivalent to \eqref{MAXCUTproblem}

\begin{equation}\label{MAXCUTproblem_lifted}
\begin{array}{l}
\max \quad  \frac12 \sum_{i<j}w_{ij}(1-X_{ij}) \\
s.t. \quad \quad X\succeq 0 \text{ , } X_{ii}=1,\; i=1,\ldots,n, \text{ and } \rank(X) = 1.
\end{array}
\end{equation}
The SDP \eqref{MAXCUTproblem_SDP} can be regarded as a relaxation of \eqref{MAXCUTproblem_lifted} obtained by removing the non-convex rank constraint. In fact, this is how we will later formulate a similar relaxation for the minimum bisection problem, in Section~\ref{s:community}.

We now turn to property (\ref{it-2}), and consider the problem of forming a solution to \eqref{MAXCUTproblem} from a solution to \eqref{MAXCUTproblem_SDP}. From the solution $\{u_i\}_{i=1,\ldots,n}$ of the relaxed problem \eqref{MAXCUTproblem_SDP}, we produce a cut subset $S'$ by randomly picking a vector $r \in \mathbb{R}^n$ from the uniform distribution on the unit sphere and setting $$S'=\{i|r^Tu_i\geq0\}$$ In other words, we separate the vectors $u_1,\ldots,u_n$ by a random hyperplane (perpendicular to $r$). We will show that the cut given by the set $S'$ is comparable to the optimal one.

\begin{figure}[h]
\begin{center}
\includegraphics[width=0.4\textwidth]{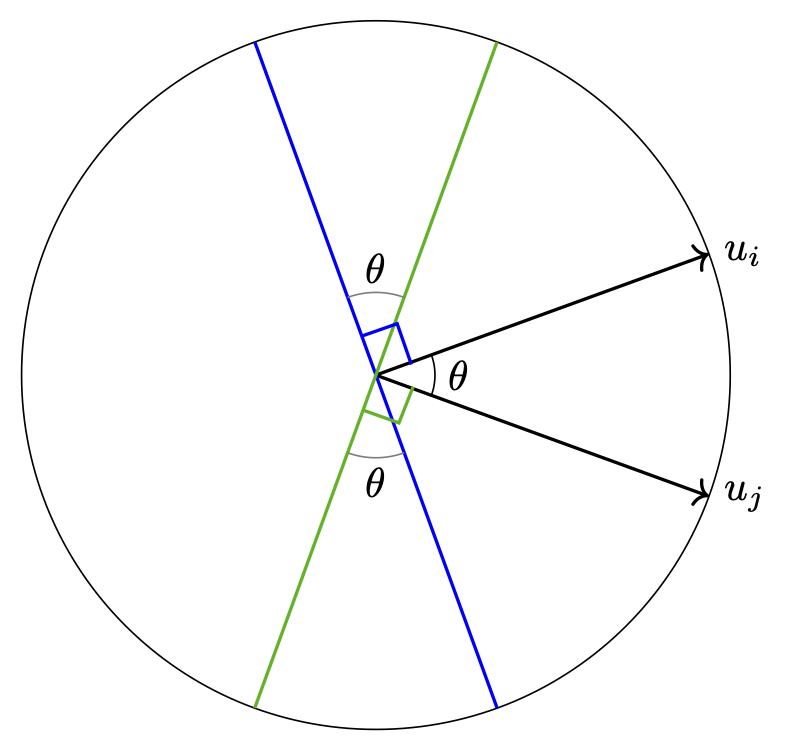}
\caption{Illustration of the relationship between the angle between vectors and their inner product, $\theta = \arccos(u_i^Tu_j)$}
\label{figure:8:MaxCut}
\end{center}
\end{figure}

Let $W$ be the value of the cut produced by the procedure described above. Note that $W$ is a random variable, whose expectation is easily seen (see Figure~\ref{figure:8:MaxCut}) to be given by

\begin{eqnarray*}
\mathbb{E}[W] & = & \sum_{i<j}w_{ij} \Pr\left\{\text{sign}(r^Tu_i)\neq \text{sign}(r^Tu_j)\right\} \\
     & = & \sum_{i<j}w_{ij}\frac1\pi \arccos(u_i^Tu_j). \\
\end{eqnarray*}
If we define $\alpha_{GW}$ as
\[
 \alpha_{GW} = \min_{-1\leq x \leq 1} \frac{\frac1\pi \arccos(x)}{\frac12(1-x)},
\]
it can be shown that $\alpha_{GW}>0.87$ (see, for example~\cite{MXGoemans_DPWilliamson_1995}).

By linearity of expectation
\begin{equation}
\mathbb{E}[W] = \sum_{i<j}w_{ij}\frac1\pi \arccos(u_i^Tu_j) \geq \alpha_{GW} \frac12 \sum_{i<j}w_{ij}(1-u_i^Tu_j).
\end{equation}

Let $\mathrm{MaxCut}(G)$ be the maximum cut of $G$, meaning the maximum of the original problem \eqref{MAXCUTproblem}. Naturally, the optimal value of \eqref{MAXCUTproblem_Relaxed} is larger or equal than $\mathrm{MaxCut}(G)$. Hence, an algorithm that solves \eqref{MAXCUTproblem_Relaxed} and uses the random rounding procedure described above produces a cut $W$ satisfying
\begin{equation}
\mathrm{MaxCut}(G)\geq \mathbb{E}[W] \geq \alpha_{GW} \frac12 \sum_{i<j}w_{ij}(1-u_i^Tu_j) \geq \alpha_{GW} \mathrm{MaxCut}(G),
\end{equation}
thus having an approximation ratio (in expectation) of $\alpha_{GW}$. Note that one can run the randomized rounding procedure several times and select the best cut.\footnote{It is worth noting that one is only guaranteed to solve \eqref{MAXCUTproblem_Relaxed} up to an approximation of $\eps$ from its optimum value. However, since this $\eps$ can be made arbitrarily small, one can get the approximation ratio arbitrarily close to $\alpha_{GW}$.}
We thus have
\[
\mathrm{MaxCut}(G) \geq  \mathbb{E}[W] \geq  \alpha_{GW}\mathcal{R}\mathrm{MaxCut}(G)\geq \alpha_{GW} \mathrm{MaxCut}(G)
\]

\subsection*{Can $\alpha_{GW}$ be improved?}

A natural question is to ask whether there exists a polynomial time algorithm that has an approximation ratio better than $\alpha_{GW}$. 

\begin{figure}[h]
\begin{center}
\includegraphics[width=0.34\textwidth]{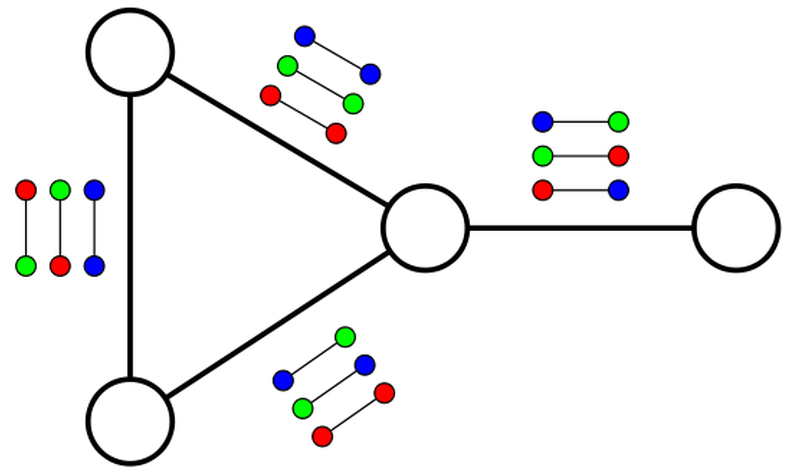}
\quad \quad
\includegraphics[width=0.34\textwidth]{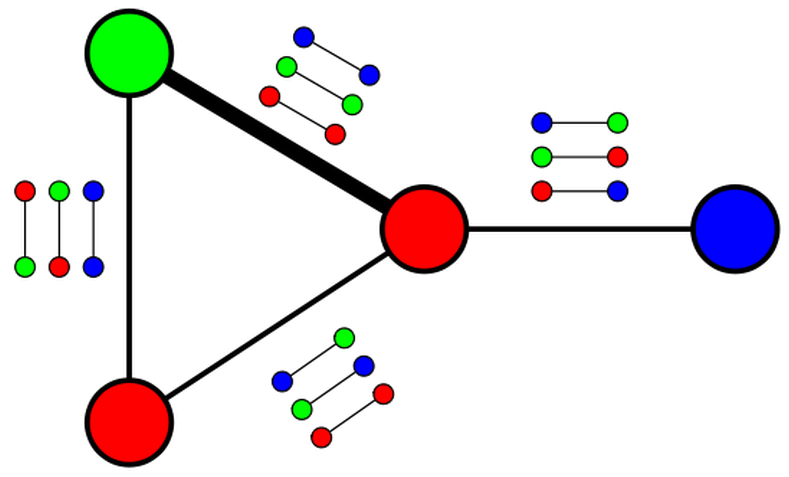}
\caption{Illustration of the Unique Games Problem}
\label{figure:8:UGP}
\end{center}
\end{figure}

The unique games problem (as depicted in Figure~\ref{figure:8:UGP}) is the following: Given a graph and a set of $k$ colors, and, for each edge, a matching between the colors, the goal in the unique games problem is to color the vertices as to agree with as high of a fraction of the edge matchings as possible. For example, in Figure~\ref{figure:8:UGP} the coloring agrees with $\frac34$ of the edge constraints, and it is easy to see that one cannot do better.

The Unique Games Conjecture of Khot~\cite{SKhot_2002}, has played a major role in hardness of approximation results.

\begin{conjecture}
For any $\epsilon>0$, the problem of distinguishing whether an instance of the Unique Games Problem is such that it is possible to agree with a $\geq 1-\epsilon$ fraction of the constraints or it is not possible to even agree with a $\epsilon$ fraction of them, is NP-hard.
\end{conjecture}

There is a sub-exponential time algorithm capable of distinguishing such instances of the unique games problem~\cite{Arora_Barak_Steurer_2010}, however no polynomial time algorithm has been found so far. At the moment one of the strongest candidates to break the Unique Games Conjecture is a relaxation based on the Sum-of-squares hierarchy that we will discuss below.

Remarkably, approximating~\texttt{Max-Cut} with an approximation ratio better than $\alpha_{GW}$ is as hard as refuting the Unique Games Conjecture (UG-hard)~\cite{Khot_Kindler_Mossel_ODonnel_2005}. More generality, if the Unique Games Conjecture is true, the semidefinite programming approach described above produces optimal approximation ratios for a large class of problems~\cite{Raghavendra_2008_optimalitySDP_UG}.

Not depending on the Unique Games Conjecture, there is a NP-hardness of approximation of $\frac{16}{17}$ for \texttt{Max-Cut}~\cite{Hastad_inapproximability}.

\begin{remark}
Note that a simple greedy method that assigns membership to each vertex as to maximize the number of edges cut involving vertices already assigned achieves an approximation ratio of $\frac12$ (even of $\frac12$ of the total number of edges, not just of the optimal cut).
\end{remark}

\subsection{A Sums-of-Squares interpretation}

We now give a different interpretation to the approximation ratio obtained above. Let us first slightly reformulate the problem (recall that $w_{ii}=0$).

Recall from Proposition~\ref{prop:cut_Laplacian} that a cut can be rewritten as a quadratic form involving the graph Laplacian. We can rewrite~\eqref{MAXCUTproblem} as
\begin{equation}\label{MAXCUTproblem_LG}
\begin{array}{cl}
\max &\frac14 y^T L_G y \\
 & y_{i}=\pm1,\; i=1,\ldots,n.
\end{array}
\end{equation}

Similarly,~\eqref{MAXCUTproblem_SDP} can be written (by taking $X$ as the PSD matrix variable that is a surrogate for $X=yy^T$) as

\begin{equation}\label{MAXCUTproblem_SDP_LG} 
\begin{array}{cl}
\max &\frac14 \tr\left( L_G X\right) \\
s.t. & X\succeq 0 \\
 & X_{ii}=1,\; i=1,\ldots,n.
\end{array}
\end{equation}

In Section~\ref{s:community} we will derive the the dual program to~\eqref{MAXCUTproblem_SDP_LG} in the context of recovery in the Stochastic Block Model. Here we will simply state it, and show weak duality as it will be important for the argument that follows.

\begin{equation}\label{MAXCUTproblem_SDP_LG_Dual} 
\begin{array}{cl}
\min & \tr\left( D \right)\\
s.t. & D \text{ is a diagonal matrix}\\
 & D - \frac14 L_G \succeq 0.
\end{array}
\end{equation}

Indeed, if $X$ is a feasible solution to~\eqref{MAXCUTproblem_SDP_LG} and $D$ a feasible solution to~\eqref{MAXCUTproblem_SDP_LG_Dual} then, since $X$ and $D - \frac14 L_G$ are both positive semidefinite $\tr\left[ X\left( D - \frac14 L_G \right)  \right] \geq 0$ which gives
\[
0 \leq \tr\left[ X\left( D - \frac14 L_G \right)  \right] = \tr(XD) - \frac14\tr\left( L_G X \right) = \tr(D) - \frac14\tr\left( L_G X \right),
\]
since $D$ is diagonal and $X_{ii}=1$. This shows weak duality, the fact that the value of~\eqref{MAXCUTproblem_SDP_LG_Dual} is larger than the one of~\eqref{MAXCUTproblem_SDP_LG}.\footnote{Note how in this chapter optimization problems are formulated as maximization problems, while in Chapter~\ref{c:optimization} they are formulated as minimization problems}

If the Slater condition~\eqref{Slaterscondition} is satisfied then strong duality holds, i.e., the optimal values of both programs coincide. In this case, the Slater condition asks whether there is a matrix strictly positive definite that is feasible for~\eqref{MAXCUTproblem_SDP_LG}, and the identity is such a matrix. This means that there exists $D^{\natural}$ feasible for~\eqref{MAXCUTproblem_SDP_LG_Dual} such that
\[
\tr(D^{\natural}) = \mathcal{R}\mathrm{MaxCut}.
\]

Since, for any $y\in\RR^n$, we have $y^TD^\natural y=\sum_{i=1}^n D^\natural_{ii}y_i$, we can write the following equality that holds for any $y\in\RR^n$:
\begin{equation}\label{eq:8:SOScertificateMaxCut}
\frac14 y^TL_Gy =  \mathcal{R}\mathrm{MaxCut} - y^T\left( D^{\natural} - \frac14L_G\right)y + \sum_{i=1}^n D_{ii}^{\natural}\left( y_i^2 - 1 \right).
\end{equation}

Since $D^{\natural} - \frac14L_G\succeq 0$, there exists $V$ such that $D^{\natural} - \frac14L_G = VV^T$ with the columns of $V$ denoted by $v_1,\dots,v_n$, meaning that $y^T\left( D^{\natural} - \frac14L_G\right)^Ty = \left\| V^T y\right\|^2 = \sum_{k=1}^n (v_k^Ty)^2$. Hence, for any $y\in\RR^n$,
\begin{equation}\label{eq:8.5:SOScertificateMaxCut}
\mathcal{R}\mathrm{MaxCut} - \frac14 y^TL_Gy   = \sum_{k=1}^n (v_k^Ty)^2+ \sum_{i=1}^n D_{ii}^{\natural}\left( y_i^2 - 1 \right).
\end{equation}

There is a particularly fruitful interpretation of the equality~\eqref{eq:8.5:SOScertificateMaxCut}: It provides a certificate that no cut of $G$ is larger than $\mathcal{R}\mathrm{MaxCut}$. Indeed, if $y\in\{\pm1\}^n$ then $y_i^2=1$ and so $\sum_{i=1}^n D_{ii}^{\natural}\left( y_i^2 - 1 \right)=0$ while $\sum_{k=1}^n (v_k^Ty)^2\geq 0$ since it is a sum of squares (of degree 2). Furthermore, while it certifies a bound on $2^n$ many possible assignments $y\in\{\pm1\}^n$,~\eqref{eq:8.5:SOScertificateMaxCut} can be verified efficiently: because it needs to hold for all $y\in\RR^n$, it amounts to checking that the coefficients of the quadratic polynomials on the left- and right-hand sides match.

This is known as a sum-of-squares certificate~\cite{Barak_Steurer_surveyICM,Barak_SOS_LectureNotes,Parrilo_thesis_SOS,Lassere_01_SOS,Shor_87_SOS,Nesterov_00_SOS}; by following this argument in reverse one can show that $\mathcal{R}\mathrm{MaxCut}$ is precisely the smallest real number for which a certificate of the type of~\eqref{eq:8.5:SOScertificateMaxCut} exists (while $\mathrm{MaxCut}$ is the smallest real number for which $\mathrm{MaxCut} - \frac14 y^TL_Gy$ is non-negative in the hypercube. This gap shrinks if one allows sum-of-square certificates of higher degree\footnote{This is related to Hilbert's 17th problem~\cite{Schmudgen_Hilbert17} and Stengle's Positivstellensatz~\cite{Stengle_Positivstellensatz} }).

The remarkable fact is that sum-of-squares certificates of at most a specified degree can be found using Semidefinite programming~\cite{Parrilo_thesis_SOS,Lassere_01_SOS} (in fact, the SDP~\eqref{MAXCUTproblem_SDP_LG_Dual} is finding the smallest real number $\Lambda$ for which a certificate such as~\eqref{eq:8.5:SOScertificateMaxCut} (of degree 2) exists.  
On the other hand, the primal is in some sense constraining the degree $2$ moments of $y$, $X_{ij}=y_iy_j$. Many natural questions remain open towards a precise understanding of the power of SDPs corresponding to higher degree sum-of-squares certificates. There are several very nice lecture notes, surveys, and courses on Sum-of-Squares, see for example~\cite{Barak_SOS_LectureNotes,Fleming-etal-proofsandalgorithms,Steurer-etal-ICM2018}\footnote{There are also very nice courses and notes on the topic, such as courses by Boaz Barak and David Steurer (\url{https://www.sumofsquares.org/public/index.html}), by Tselil Schramm (\url{https://tselilschramm.org/sos-paradigm/winter21.html}), by Sam Hopkins (\url{http://www.samuelbhopkins.com/teaching/sos-fall-24/sos-fall-24.html}), Tim Kunisky (\url{http://www.kunisky.com/teaching/2022spring-sos/}), and Aaron Potechin (\url{https://canvas.uchicago.edu/courses/17604}).}

\begin{remark}[triangular inequalities and Sum of squares level 4]\label{remark:9:SOS4MaxCut}
A natural follow-up question is whether the relaxation of degree $4$ is actually strictly tighter than the one of degree $2$ for Max-Cut (in the sense of forcing extra constraints). What follows is an interesting set of inequalities that degree $4$ enforces and that degree $2$ does  not, known as triangular inequalities. This example helps illustrate the differences between Sum-of-Squares certificates of different degree.

Since $y_i\in \{\pm 1\}$ we naturally have that, for all $i,j,k$
\[
y_iy_j + y_jy_k + y_ky_i \geq -1,
\]
this would mean that, for $X_{ij} = y_iy_j$ we would have,
\[
X_{ij} + X_{jk} + X_{ik} \geq -1,
\]
however it is not difficult to see that the SDP~\eqref{MAXCUTproblem_SDP_LG} of degree 2 is only able to constraint
\[
X_{ij} + X_{jk} + X_{ik} \geq -\frac32,
\]
which is considerably weaker. There are a few different ways of thinking about this, one is that the three vector $u_i,u_j,u_k$ in the relaxation may be at an angle of 120 degrees with each other. Another way of thinking about this is that
the inequality $y_iy_j + y_jy_k + y_ky_i \geq -\frac32$ can be proven using sum-of-squares proof with degree $2$:
\[
(y_i+y_j+y_k)^2 \geq 0 \quad  \Rightarrow \quad y_iy_j + y_jy_k + y_ky_i \geq -\frac32
\]
However, the stronger constraint cannot.

On the other hand, if degree $4$ monomials are involved, (let us say $X_S = \prod_{s\in S} y_s$, note that $X_{\emptyset} =1$ and $X_{ij}X_{ik} = X_{jk}$) then the constraint
\[
\left[\begin{array}{c}
X_{\emptyset} \\
X_{ij} \\
X_{jk} \\
X_{ki}
\end{array}\right]
\left[\begin{array}{c}
X_{\emptyset} \\
X_{ij} \\
X_{jk} \\
X_{ki}
\end{array}\right]^T
= 
\left[\begin{array}{cccc}
1       & X_{ij} & X_{jk} & X_{ki} \\
X_{ij} & 1      & X_{ik} & X_{jk} \\
X_{jk} & X_{ik}& 1       & X_{ij} \\
X_{ki} & X_{jk} & X_{ij} & 1
\end{array}\right]
\succeq 0
\]
implies $X_{ij} + X_{jk} + X_{ik} \geq -1$ just by taking
\[
\1^T  \left[\begin{array}{cccc}
1       & X_{ij} & X_{jk} & X_{ki} \\
X_{ij} & 1      & X_{ik} & X_{jk} \\
X_{jk} & X_{ik}& 1       & X_{ij} \\
X_{ki} & X_{jk} & X_{ij} & 1
\end{array}\right]   \1 \geq 0.
\]
Also, note that the inequality $y_iy_j + y_jy_k + y_ky_i \geq -1$ can indeed be proven using sum-of-squares proof with degree $4$ (recall that $y_i^2=1$):
\[
(1+y_iy_j+y_jy_k+y_ky_i)^2 \geq 0 \quad  \Rightarrow \quad y_iy_j + y_jy_k + y_ky_i \geq -1.
\]
Interestingly, it is known~\cite{Khot_Vishnoi_2013} that these extra inequalities alone will not increase the approximation power (in the worst case) of~\eqref{MAXCUTproblem_SDP}.
\end{remark}

\section{Community detection}\label{s:community}

The problem of detecting communities in network data is a central problem in data science, examples of interest include social networks, the internet, or biological and ecological networks. In Chapter~\ref{c:graphs} we discussed clustering in the context of graphs, and described performance guarantees for spectral clustering (based on Cheeger's Inequality) that made no assumptions on the underlying graph. While these guarantees are remarkable, they are worst-case and hence pessimistic in nature. In an effort to understand the performance of some of these approaches on more realistic models of data, we will now analyze a generative model for graphs with community structure, the stochastic block model. On the methodology side, we will focus on convex relaxations, based on semidefinite programming (as in Chapter~\ref{maxcutapprox}), and will show that this approach achieves exact recovery of the communities on graphs drawn from this model. The techniques developed to prove these guarantees mirror the ones used to prove analogous guarantees for a variety of other problems where convex relaxations yield exact recovery (such as some of the problems covered in Chapter~\ref{c:lowrank}).

\subsection{The Stochastic Block Model}

The Stochastic Block Model is a random graph model that produces graphs with a community structure. While, as with any model, we do not expect it to capture all properties of a real world network (examples include network hubs, power-law degree distributions, and other structures) the goal is to study a simple graph model that produces community structure, as a test bed for understanding fundamental limits of community detection and analyzing the performance of recovery algorithms. 

\begin{definition}[Stochastic Block Model]
Let $n$ and $k$ be positive integers representing respectively the number of nodes and communities, $c\in [k]^n$ be the vector of community memberships for the different nodes, and $P\in [0,1]^{k\times k}$ a symmetric matrix of connectivity probabilities. A graph $G$ is said to be drawn from the Stochastic Block Model on $n$ nodes, when for each pair of nodes $(i,j)$ the probability that $(i,j)\in E$ is independent from all other edges and given by $P_{c_i,c_j}$.
\end{definition}

\begin{figure}[h]
\begin{center}
\subcaptionbox{}{\includegraphics[width=0.3\textwidth,height=0.28\textwidth]{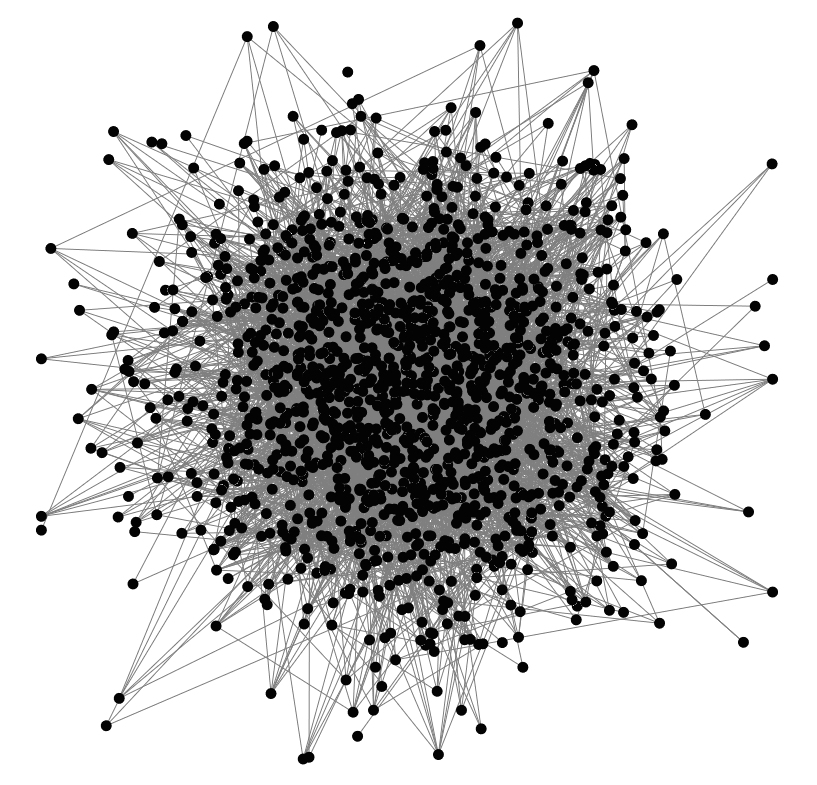}}\qquad
\subcaptionbox{}{\includegraphics[width=0.6\textwidth,height=0.28\textwidth]{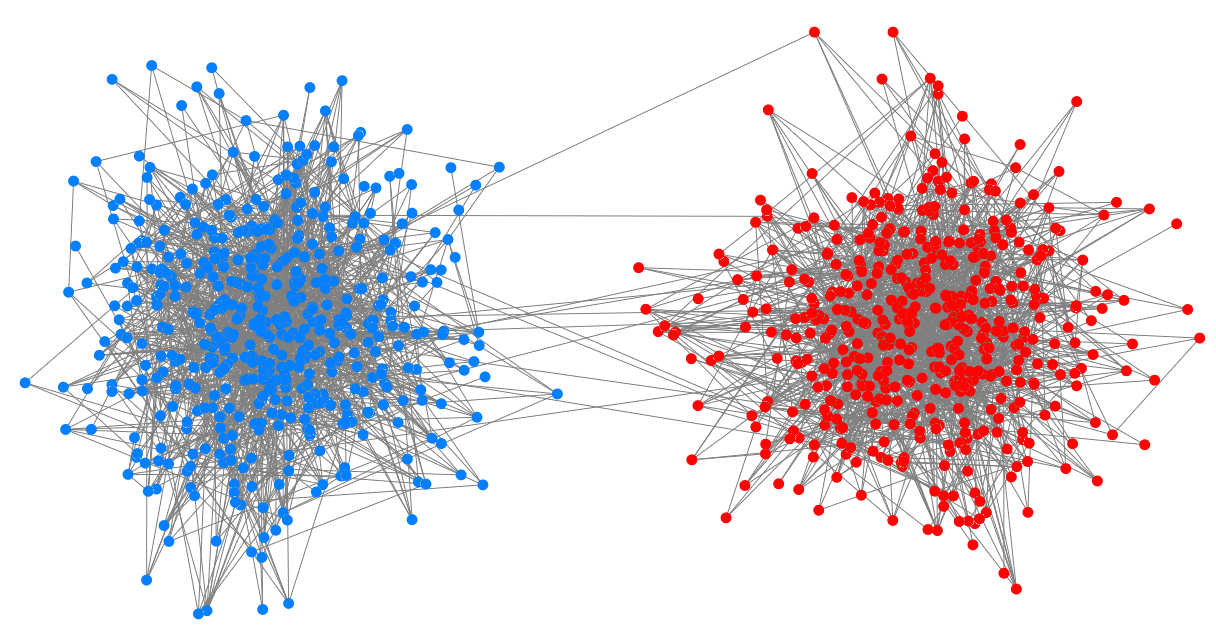}}
\caption{A graph generated form the stochastic block model with 600 nodes and 2 communities, scrambled in Fig.~\ref{fig:sbm}(a),  clustered and color-coded in Fig.~\ref{fig:sbm}(b). 
Nodes in this graph connect with probability $p = 6/600$ within communities and $q = 0.1/600$ across communities. (Image courtesy of Emmanuel Abbe.)}
\label{fig:sbm}
\end{center}
\end{figure}

We will focus on the special case of the two communities ($k=2$) balanced symmetric block model where $n$ is even, both communities are of the same size, and
\[
P = \left[\begin{array}{cc} p & q \\ q & p \end{array}\right],
\]
where $p,q\in[0,1]$ are constants, cf.~Figure~\ref{fig:sbm}. Furthermore, we will focus on the associative case ($p>q$), while noting that all that follows can be easily adapted to the disassociate case ($q>p$). 
 
We note also that when $p=q$ this model reduces to the classical Erd\H{o}s-Reny\'{i} model described in Chapter~\ref{c:graphs}. Since there are only two communities we will identify their membership labels with $+1$ and $-1$.

Many fascinating questions can be asked in the context of this model. Natural questions include to characterize statistics of the model, such as number of triangles or larger cliques. In this chapter, motivated by the problem of community detection, we are interested in understanding when is it possible to reconstruct, or estimate, the community memberships from an observation of the graph, and what efficient algorithms succeed at this inference task.

Before proceeding we note that the difficulty of this problem should certainly depend on the value of $p$ and $q$. As illustrative examples, this problem is trivial when $p=1$ and $q=0$ and hopeless when $p=q$ (notice that even in the easy case the actual membership can only be determined up to a re-labeling of the communities). As $p>q$, we will attempt to recover the original partition by trying to compute the minimum bisection of the graph; while related to the Max-Cut problem described in Section~\ref{c:maxcut}, notice how the objective here is to produce the minimum balanced cut.

\subsection{Spike model prediction}

A natural approach is to draw motivation from Chapter~\ref{c:graphs} and to use a form of spectral clustering to attempt to partition the graph.

Let $A$ be the adjacency matrix of $G$,
\begin{align}
A_{ij} = \left\{ \begin{array}{cc} 1 & \text{ if } (i,j)\in E(G) \\ 0 & \text{ otherwise.} \end{array}  \right.
\end{align}
Note that in our model, $A$ is a random matrix.  We would like to solve
\begin{align}
\max\ & \sum_{i,j}A_{ij}x_ix_j \nonumber\\
\text{s.t.}\ &x_i=\pm1, \forall i \label{eq:10:MinBisection}\\
 & \sum_j x_j = 0, \nonumber
\end{align}
The optimal solution $x$ of \eqref{eq:10:MinBisection} takes the value $+1$ on one side of a partition and $-1$ on the other side, where the partition is balanced and achieves the minimum cut between the resulting clusters.

Relaxing the condition $x_i=\pm1,\ \forall_i$ to $\|x\|_2^2=n$ would yield a spectral method
\begin{align}
\max\ & \sum_{i,j}A_{ij}x_ix_j \nonumber\\
\text{s.t.}\ & \|x\|_2 = \sqrt{n} \label{MLE_A_spectralmethod}\\
 & \1^T x = 0 \nonumber
\end{align}
The solution of~\eqref{MLE_A_spectralmethod} corresponds to the leading eigenvector of the matrix obtained by projecting $A$ on the orthogonal complement of the all-ones vector $\1$.

The matrix $A$ is a random matrix whose expectation is given\footnote{For simplicity we assume that self-loops also have probability $p$. This does not affect any of the conclusions, as it does not give information about the community memberships.} by
\[
 \EE[A_{ij}] =  \left\{ \begin{array}{cl} p & \text{ if $i$ and $j$ are in the same community } \\ q & \text{ otherwise.} \end{array}  \right.
\]
Let $g$ denote the vector corresponding to the true community memberships, with entries $+1$ and $-1$; note that this is the vector we want to recover.\footnote{We want to recover either $g$ or $-g$, as they correspond to different labelings of the same community structure.} We can write
\[
 \EE[A] = \frac{p+q}2 \1\1^T +  \frac{p-q}2 gg^T,
\]
and
\[
 A = \big(A-\EE[A]\big) +  \frac{p+q}2 \1\1^T +  \frac{p-q}2 gg^T.
\]
In order to remove the term $\frac{p+q}2 \1\1^T$ we consider the random matrix
\[
 \AAAA = A - \frac{p+q}2 \1\1^T.
\]
It is easy to see that
\[
 \AAAA = \big(\AAAA-\EE[\AAAA]\big) +  \frac{p-q}2 gg^T.
\]
This means that $\AAAA$ is the sum of a random matrix whose expected value is zero and a rank-1 matrix, i.e. $$\AAAA = W + \lambda vv^T$$ where $W = \big(\AAAA-\EE[\AAAA]\big)$ and $\lambda vv^T  = \frac{p-q}2n \left(\frac{g}{\sqrt{n}}\right)\left(\frac{g}{\sqrt{n}}\right)^T$. In Chapter~\ref{c:svd} we saw that for a large enough rank-1 additive perturbation to a Wigner matrix, there is an eigenvalue associated with the perturbation that pops outside of the distribution of eigenvalues of a Wigner Gaussian matrix $W$. Moreover, whenever this happens, we saw that the leading eigenvector has a non-trivial correlation with $g$. 

Since $\AAAA$ is simply $A$ minus a multiple of $\1\1^T$, problem~\eqref{MLE_A_spectralmethod} is equivalent to
\begin{align}
\max\ & \sum_{i,j}\AAAA_{ij}x_ix_j \nonumber\\
\text{s.t.}&\ \|x\|_2 = \sqrt{n} \label{MLE_A_spectralmethod_2}\\
 & \1^T x = 0 \nonumber
\end{align}

Since we have subtracted a multiple of $\1\1^T$, we will drop the constraint $\1^T x =0$. Notice how a deviation from $\1^T x =0$ would be penalized in the new objective, the fact that the multiple we subtracted is sufficient for us to drop the constraint will be confirmed by the success of the new optimization problem, now given by
\begin{align}
\max\ & \sum_{i,j}\AAAA_{ij}x_ix_j \nonumber\\
\text{s.t.}&\ \|x\|_2 = \sqrt{n} \label{MLE_A_spectralmethod_22},
\end{align}
whose solution corresponds to the leading eigenvector of $\AAAA$.

Recall that if $\AAAA-\EE[\AAAA]$ is a Wigner matrix with i.i.d. entries with zero mean and variance $\sigma^2$ then its empirical spectral density follows the semicircle law and it is essentially supported in $[-2\sigma\sqrt{n},2\sigma\sqrt{n}]$. We would then expect the top eigenvector of $\AAAA$ to correlate with $g$ as long as
\begin{equation}\label{conditionspike}
 \frac{p-q}2n  > \frac{2\sigma\sqrt{n}}2.
\end{equation}

Unfortunately $\AAAA-\EE[\AAAA]$ is not a Wigner matrix in general. In fact, half of its entries have variance $p(1-p)$ while the variance of the other half is $q(1-q)$.

Putting rigor aside for a second, if we were to take $\sigma^2 = \frac{p(1-p)+q(1-q)}2$ then \eqref{conditionspike} would suggest that the leading eigenvector of $\AAAA$ correlates with the true partition vector $g$ as long as
\begin{equation}\label{condition_notjustified}
 \frac{p-q}2  > \frac1{\sqrt{n}}\sqrt{\frac{p(1-p)+q(1-q)}2},
\end{equation}
This argument is of course not valid, because the matrix in question is not a Wigner matrix. The special case $q = 1-p$ can be easily salvaged, since all entries of $\AAAA-\EE[\AAAA]$ have the same variance and they can be made to be identically distributed by conjugating with $\mathrm{diag}(g)$. This is still an impressive result, it says that if $p=1-q$ then $p-q$ needs only to be around $\frac1{\sqrt{n}}$ to be able to make an estimate that correlates with the original partitioning!

An interesting regime (motivated, for example, by friendship networks in social sciences) is when the average degree of each node is constant. This can be achieved by taking $p = \frac{a}n$ and $q = \frac{b}n$ for constants $a$ and $b$. While the argument presented to justify condition \eqref{condition_notjustified} is not valid in this setting, it nevertheless suggests that the condition on $a$ and $b$ needed to be able to make an estimate that correlates with the original partition, often referred to as partial recovery, is
\begin{equation}\label{condition_notjustified_ab}
 (a-b)^2  > 2(a+b).
\end{equation}

Remarkably this was posed as a conjecture by Decelle et al.~\cite{Decelle_SBM} and proved in a series of works by Mossel et al.~\cite{Mossel_SBM1,Mossel_SBM2} and Massoulie~\cite{Massoulie_SBM}. While describing the proof of this conjecture is outside the scope of this book, we note that the conjectures were obtained by studying fixed points of a certain linearization of belief propagation using techniques from statistical physics. The lower bound can be proven by showing contiguity between the two models below the phase transition, and the upper bound is obtained by analyzing an algorithm that is an adaptation of belief propagation and studying the so-called non-backtracking operator. We refer the reader to the excellent survey of Abbe~\cite{Abbe_SBM_survey} and references therein for further reading.

\begin{remark}[More than three communities]
The balanced symmetric stochastic block model with $k>3$ communities is conjectured to have a fascinating statistical-to-computational gap. In the sparse regime of inner probability $p=\frac{a}{n}$ and outer probability $q = \frac{b}{n}$ it is believed that, for $k>3$ there is a regime of the parameters $a$ and $b$ such that the problem of partially recovering the community memberships is statistically, or information-theoretically, possible but that there does not exist a polynomial-time algorithm to perform this task. These conjectures are based on insight obtained with tools from statistical physics. We refer the reader to~\cite{Decelle_SBM,Zhang_Moore_2014,Ghasemian_Moore_SBM_2015,Abbe_Sandon_KSbound} for further reading.
\end{remark}

\subsection{Exact recovery}

We now turn our attention to the problem of recovering the cluster membership of every single node correctly, not simply having an estimate that correlates with the true labels. We will keep our focus on the balanced, symmetric, two communities setting. This will serve as an illustration for two phenomena: (i) the fact that convex relaxations often find optimal solutions, not just approximations, when the input are ``typical instances''; (ii) convex relaxations such as semidefinite programs over random instances can often be readily understood via matrix concentration inequalities (such as the ones in Chapter~\ref{c:probability-matrixconcentration}). If the inner-probability is $p = \frac{a}n$ then it is not hard to show that each cluster will (with high probability) have isolated nodes, making it impossible to recover the membership of every possible node correctly. In fact this is the case whenever $p \leq  \frac{(2-\eps)\log n}n$, for some $\eps>0$. For that reason we focus on the regime
\begin{equation}
p = \pp \text{ and } q = \qq,
\end{equation}
for some constants $\alpha>\beta$.

A natural algorithm would be to compute the minimum bisection~\eqref{eq:10:MinBisection} which corresponds to the Maximum Likelihood Estimator, and also the Maximum a Posteriori Estimator when the community memberships are drawn uniformly at random. In fact, it is known (see~\cite{Abbe_SBMExact} for a proof) that if
\begin{equation}\label{condition_MLE}
\sqrt{\alpha} - \sqrt{\beta}  > \sqrt{2},
\end{equation}
then, with high probability, (\ref{eq:10:MinBisection}) recovers the true partition. Moreover, if
\[
\sqrt{\alpha} - \sqrt{\beta}  < \sqrt{2},
\]
no algorithm can, with high probability, recover the true partition.

In this section we will analyze a semidefinite programming relaxation, analogous to the ones described in Section~\ref{c:maxcut} for Max-Cut. By making use of convex duality, we will derive conditions for exact recovery with this particular algorithm, reducing the problem to a problem in random matrix theory. We will present a solution to the resulting random matrix question, using the matrix concentration tools developed in Chapter~\ref{c:probability-matrixconcentration}. While not providing the strongest known guarantee, this approach is extremely adaptable and can be used to solve a large number of similar questions.\footnote{A tight guarantee can be obtained by adapting the random matrix theory part of the argument, the convex geometry part can remain the same; this is briefly discussed below.}

\subsection{A semidefinite relaxation}

Let $x\in\RR^n$ with $x_i =\pm 1$ represent a partition of the nodes (recall that there is an ambiguity in the sense that $x$ and $-x$ represent the same partition).
Note that if we remove the constraint $\sum_j x_j = 0$ in (\ref{eq:10:MinBisection}) then the optimal solution becomes $x=\1$. Let us define $B = 2A - (\1\1^T -I)$, meaning that

\begin{align}
B_{ij} = \left\{ \begin{array}{rl} 0 & \text{ if } i=j \\ 1 & \text{ if } (i,j)\in E(G) \\ -1 & \text{ otherwise} \end{array}  \right.
\end{align}
It is clear that the problem

\begin{align}
\max\ & \sum_{i,j}B_{ij}x_ix_j \nonumber\\
\text{s.t.}\ &x_i=\pm1, \forall_i \label{MLE_B_old}\\
 & \sum_j x_j = 0 \nonumber
\end{align}
has the same solution as (\ref{eq:10:MinBisection}). However, when the constraint is dropped,

\begin{align}
\max\ & \sum_{i,j}B_{ij}x_ix_j \nonumber\\
\text{s.t.}\ &x_i=\pm1, \forall_i \label{MLE_B},
\end{align}
$x= \1$ is no longer an optimal solution. As with~\eqref{MLE_A_spectralmethod_22} above, the penalty created by subtracting a large multiple of $\1\1^T$ will be enough to discourage unbalanced partitions (the reader may notice the connection with Lagrangian duality). In fact, (\ref{MLE_B}) is the problem we will set ourselves to solve.

Unfortunately (\ref{MLE_B}) is in general NP-hard (one can encode, for example, \emph{Max-Cut} by picking an appropriate $B$). We will relax it to an easier problem by the same technique used to approximate the  Max-Cut problem in the previous section (this technique is often known as \emph{matrix lifting}). If we write $X=xx^T$ then we can formulate the objective of (\ref{MLE_B}) as
\[
\sum_{i,j}B_{ij}x_ix_j = x^TBx = \tr(x^TBx) = \tr(Bxx^T) = \tr(BX)
\]
Also, the condition $x_i = \pm1$ implies $X_{ii} = x_i^2 = 1$. This means that (\ref{MLE_B}) is equivalent to

\begin{align}
\max\ \qquad & \tr(BX)\nonumber\\
\text{s.t.}\ \qquad &X_{ii}=1, \forall_i \label{eq:relaxation_SBM}\\
 & X=xx^T  \text{ for some } x\in\RR^n.\nonumber
\end{align}

The fact that $X=xx^T$ for some $x\in\RR^n$ is equivalent to $\rank(X) = 1$ and $X\succeq 0$. This means that (\ref{MLE_B}) is equivalent to
\begin{align}
\max\ \qquad& \tr(BX)\nonumber\\
\text{s.t.}\ \qquad &X_{ii}=1, \forall_i \\
 & X \succeq 0 \nonumber\\
 & \rank(X) = 1.\nonumber
\end{align}
We now relax the problem by removing the non-convex rank constraint
\begin{align}
\max\ \qquad & \tr(BX)\nonumber\\
\text{s.t.}\ \qquad & X_{ii}=1, \forall_i \label{MLE_SDP}\\
 & X \succeq 0 \nonumber.
 \end{align}
This is an SDP that can be solved (up to arbitrary precision) in polynomial time~\cite{LVanderberghe_SBoyd_1996} (note that it is the same SDP as~\eqref{MAXCUTproblem_SDP_LG}, with a different coefficient matrix).

Since we removed the rank constraint, the solution to (\ref{MLE_SDP}) is no longer guaranteed to be rank-1. We will take a different approach from the one we used in Section~\ref{c:maxcut} to obtain an approximation ratio for \texttt{Max-Cut}, which was a worst-case approximation ratio guarantee. In this section we will show that, for some values of $\alpha$ and $\beta$, with high probability, the solution to (\ref{MLE_SDP}) not only satisfies the rank constraint but it coincides with $X = gg^T$ where $g$ corresponds to the true partition. From $X$ one can compute $g$ by simply calculating its leading eigenvector. Besides being a good approximation algorithm, SDP relaxations such as~\eqref{MAXCUTproblem_SDP_LG} provide optimal solutions in many instances (but not always).

\subsection{Convex duality}

A standard technique to show that a candidate solution is the optimal one for a convex problem is to use convex duality, which we discussed in Section~\ref{ss:KKT}. Note that unlike in Chapter~\ref{c:optimization}, problems here are formulated as maximization problems.

We will now describe duality with a game theoretical intuition in mind. The idea will be to rewrite \eqref{MLE_SDP} without imposing constraints on $X$ but rather have the constraints be implicitly enforced. Consider the following optimization problem.

\begin{equation}\label{MLE_SDP_Lagrangian}
\max_{X} \min_{\substack{Z,\ Q \\ Z \text{ is diagonal } \\ Q\succeq 0}} \tr(BX) + \tr(QX) + \tr\left(Z\left(I_{n\times n}-X\right)\right).
\end{equation}

Let us provide a game theoretical interpretation for~\eqref{MLE_SDP_Lagrangian}. Suppose that there is a {\em primal player} (picking $X$) whose objective is to maximize the objective and a {\em dual player} (picking $Z$ and $Q$ after seeing $X$) trying to make the objective as small as possible. If the primal player does not pick $X$ satisfying the constraints of \eqref{MLE_SDP} then we claim that the dual player is capable of driving the objective to $-\infty$. Indeed, if there is an $i$ for which $X_{ii} \neq 1$ then the dual player can simply pick $Z_{ii} = -c\frac1{1-X_{ii}}$ and make the objective as small as desired by taking a large enough $c$. Similarly, if $X$ is not positive semidefinite, then the dual player can take $Q=c vv^T$ where $v$ is such that $v^TXv <0$. If, on the other hand, $X$ satisfies the constraints of \eqref{MLE_SDP} then
\[
 \tr(BX) \leq \min_{\substack{Z,\ Q \\ Z \text{ is diagonal } \\ Q\succeq 0}} \tr(BX) + \tr(QX) + \tr\left(Z\left(I_{n\times n}-X\right)\right).
\]
Since equality can be achieved if for example the dual player picks $Q = 0_{n\times n}$, then it is evident that the values of \eqref{MLE_SDP} and \eqref{MLE_SDP_Lagrangian} are the same:
\[
 \max_{\substack{X,\\ X_{ii}\ \forall_i \\ X\succeq 0}} \tr(BX) = \max_{X} \min_{\substack{Z,\ Q \\ Z \text{ is diagonal } \\ Q\succeq 0}} \tr(BX) + \tr(QX) + \tr\left(Z\left(I_{n\times n}-X\right)\right)
\]
With this game theoretical intuition in mind, it is clear that if we change the ``rules of the game'' and have the dual player decide their variables before the primal player (meaning that the primal player can pick $X$ knowing the values of $Z$ and $Q$) then it is clear that the objective can only increase, which means that:
\[
 \max_{\substack{X,\\ X_{ii}\ \forall_i \\ X\succeq 0}} \tr(BX) \leq \min_{\substack{Z,\ Q \\ Z \text{ is diagonal } \\ Q\succeq 0}} \max_{X}  \tr(BX) + \tr(QX) + \tr\left(Z\left(I_{n\times n}-X\right)\right).
\]
Note that we can rewrite
\[
  \tr(BX) + \tr(QX) + \tr\left(Z\left(I_{n\times n}-X\right)\right) = \tr\left(\left( B + Q - Z \right)X\right) + \tr(Z).
\]
When playing:
\[
  \min_{\substack{Z,\ Q \\ Z \text{ is diagonal } \\ Q\succeq 0}} \max_{X}  \tr\left(\left( B + Q - Z \right)X\right) + \tr(Z),
\]
if the dual player does not set $B+Q-Z=0_{n\times n}$ then the primal player can drive the objective value to $+\infty$, this means that the dual player is forced to chose $Q = Z-B$ and so we can write
\[
  \min_{\substack{Z,\ Q \\ Z \text{ is diagonal } \\ Q\succeq 0}} \max_{X}  \tr\left(\left( B + Q - Z \right)X\right) + \tr(Z) = \min_{\substack{Z,\\ Z \text{ is diagonal } \\ Z-B\succeq 0}} \max_{X}  \tr(Z),
\]
which clearly does not depend on the choices of the primal player. This means that
\[
  \max_{\substack{X,\\ X_{ii}\ \forall_i \\ X\succeq 0}} \tr(BX) \leq \min_{\substack{Z,\\ Z \text{ is diagonal } \\ Z-B\succeq 0}}  \tr(Z).
\]
This is known as weak duality (strong duality says that, under some conditions the two optimal values actually match, see for example~\cite{LVanderberghe_SBoyd_1996}, recall that we used strong duality when giving a sum-of-squares interpretation to the Max-Cut approximation ratio, a similar interpretation can be given in this problem, see~\cite{Bandeira_PCC}).

Also, the problem
\begin{align}
\min\ \qquad & \tr(Z)\nonumber\\
\text{s.t.}\ \qquad & Z \text{ is diagonal} \label{MLE_SDP_DUAL}\\
 & Z - B  \succeq 0 \nonumber
 \end{align}
is called the dual problem of (\ref{MLE_SDP}).

The derivation above explains why (in maximization optimization problems) the objective value of the dual problem is always greater or equal to the primal problem (for minimization problems the opposite is true; in fact,~\eqref{MLE_SDP} is the dual to~\eqref{MLE_SDP_DUAL}). Nevertheless, there is a much simpler proof (although not as enlightening): let $X,Z$ be  a feasible point of (\ref{MLE_SDP}) and (\ref{MLE_SDP_DUAL}), respectively.
Since $Z$ is diagonal and $X_{ii}=1$, it follows that $\tr(ZX)=\tr(Z)$. Also, $Z-B\succeq 0$ and $X\succeq 0$, therefore $\tr[(Z-B)X] \geq 0$. Altogether,
\[
\tr(Z) - \tr(BX) = \tr[(Z-B)X] \geq 0,
\]
as stated.

Recall that we want to show that $gg^T$ is the optimal solution of (\ref{MLE_SDP}). Then, if we find $Z$ diagonal, such that $Z-B \succeq 0$ and
\[
\tr[(Z-B)gg^T] = 0,  \quad \text{(this condition is known as complementary slackness)}
\]
then $X=gg^T$ must be an optimal solution of (\ref{MLE_SDP}). To ensure that $gg^T$ is the unique solution we just have to ensure that the nullspace of $Z-B$ only has dimension 1 (which corresponds to multiples of $g$). Essentially, if this is the case, then for any other possible solution $X$ one could not satisfy complementary slackness.

This means that if we can find $Z$ with the following properties:
\begin{enumerate}[leftmargin=10mm,itemsep=0.2pt,topsep=2pt]
\item[(1)] $Z$ is diagonal
\item[(2)] $\tr[(Z-B)gg^T] = 0$
\item[(3)] $Z-B \succeq 0$
\item[(4)] $\lambda_2(Z-B)> 0$,
\end{enumerate}
 then $gg^T$ is the unique optimum of (\ref{MLE_SDP}) and so recovery of the true partition is possible (with an efficient algorithm).
 $Z$ is known as the dual certificate, or dual witness (note that this essentially corresponds to establishing the KKT conditions discussed in Section~\ref{ss:KKT}).

 \subsection{Building the dual certificate}

The idea to build $Z$ is to construct it to satisfy properties (1) and (2) and try to show that it satisfies (3) and (4) using concentration. In fact, since $Z$ is diagonal this design problem has $n$ free variables. If $Z-B \succeq 0$ then condition (2) is equivalent to $(Z-B)g=0$ which provides $n$ equations, as the resulting linear system is non-singular, the candidate arising from using conditions (1) and (2) will be unique.
 
 A couple of preliminary definitions will be useful before writing out the candidate $Z$. Recall that the degree matrix $D$ of a graph $G$ is a diagonal matrix where each diagonal coefficient $D_{ii}$ corresponds to the number of neighbors of vertex $i$ and that $\lambda_2(M)$ is the second smallest eigenvalue of a symmetric matrix $M$.
\begin{definition}\label{Def:L_SBM}
Let $\Gp$ (resp. $\Gm$) be the subgraph of $G$ that includes the edges that link two nodes in the same community (resp.\ in different communities) and $A$ the adjacency matrix of $G$. We denote by $\Dp$ (resp.\ $\Dm$) the degree matrix of $\Gp$ (resp.\ $\Gm$) and define the Stochastic Block Model Laplacian to be
\begin{eqnarray*}
L_{SBM}= \Dp-\Dm-A.
\end{eqnarray*}
\end{definition}
Note that the inclusion of  self loops does not change $L_{SBM}$. Also, we point out that $L_{SBM}$ is not in general positive-semidefinite.

Now we are ready to construct the candidate $Z$. Condition (2) implies that we need $Z_{ii} = \frac1{g_i}B[i,:]g$. Since $B = 2A - (\1\1^T -I)$ we have
\[
Z_{ii} = \frac1{g_i}(2A - (\1\1^T -I))[i,:]g = 2\frac1{g_i}(Ag)_i + 1,
\]
meaning that
\[
Z = 2(\Dp-\Dm) + I.
\]
This is our candidate dual witness.
As a result
\[
Z - B = 2(\Dp-\Dm) + I - \left[ 2A - (\1\1^T -I) \right] = 2 L_{SBM} +11^T.
\]
It trivially follows (by construction) that
\[
(Z-B)g = 0.
\]
This immediately gives the following lemma.

\begin{lemma} \label{lemma:lemma1SDP}
Let $L_{\text{SBM}}$ denote the Stochastic Block Model Laplacian as defined in Definition~\ref{Def:L_SBM}. If
\begin{equation}
\lambda_2(2 L_{\text{SBM}} +11^T) >0 \label{SDP:condition},
\end{equation}
then the relaxation~\eqref{MLE_SDP} recovers the true partition.
\end{lemma}
Note that $2 L_{\text{SBM}} +11^T$ is a random matrix and so this reduces to ``an exercise" in random matrix theory.

\subsection{Matrix concentration and Matrix Bernstein Inequality}\label{sc:matrixBernstein}

We will dedicate Chapter~\ref{c:probability-matrixconcentration} to random matrix theory and matrix concentration inequalities that aim to control the largest eigenvalue or spectral norm of random matrices. 
In this section we present a general use concentration inequality for sums of independent random matrices, while noting that, as with scalars, many random variables can be written as sums of independent random variables even when it's not trivially apparent (and we will see below that this is the case for the matrix appearing in Lemma~\ref{lemma:lemma1SDP}).

Let us recall Bernstein's inequality (Theorem~\ref{thm:Bernstein}) copied here with slightly different notation, and with only one of the tails: If $X_1,X_2,\ldots,X_n$ are independent centered random variables satisfying $|X_i| \leq r$ and $\mathbb{E}[X_i^2]=\frac1n\nu^2$. Then,
\begin{equation}\label{prob:Bernstein:scalar}
\Prob \left\{ \sum_{i=1}^n X_i > t \right\} \leq \exp\left(-\frac{t^2}{2\nu^2 + \frac{2}{3}rt}\right).
\end{equation}

A very useful generalization of this inequality exists for bounding the largest eigenvalue of the sum of independent random matrices
\begin{theorem}[Matrix Bernstein Inequality, Theorem~1.4 in~\cite{Tropp:TailBoundsRM}]\label{thm:4:MatrixBernstein}
 Let $\{X_k\}_{k=1}^n$ be a sequence of independent random symmetric $d\times d$ matrices. Assume that each $X_k$ satisfies:
 \[
  \mathbb{E} X_k = 0 \text{ and } \lambda_{\max}\left( X_k \right) \leq R \text{ almost surely.}
 \]
Then, for all $t\geq 0$,
\[
 \Prob\left\{ \lambda_{\max}\left( \sum_{k=1}^n X_k \right) \geq t \right\} \leq d \cdot \exp\left( \frac{-t^2}{2\sigma^2 + \frac23Rt} \right) \text{ where } \sigma^2 = \left\| \sum_{k=1}^n\mathbb{E}\left(X_k^2\right) \right\|.
\]
\end{theorem}
Recall that $\|A\|$ denotes the spectral norm of $A$. Comparing with \eqref{prob:Bernstein:scalar} the attentive reader will notice an extra dimensional factor of $d$; a simple change of variables shows that this corresponds to a poly-logarithmic factor on the random variable, a factor that will be discussed later in this chapter where we will also include an improved inequality (Theorem~\ref{thm:improvedBernstein}).
Theorem~\ref{thm:4:MatrixBernstein} can be easily extended to rectangular matrices, see e.g.~Theorem~1.6 in~\cite{Tropp:TailBoundsRM}.

In Chapter~\ref{c:probability-matrixconcentration} we will discuss matrix concentration inequalities in detail and prove inequalities similar to Theorem~\ref{thm:4:MatrixBernstein} (for the proof of the precise bound in Theorem~\ref{thm:4:MatrixBernstein}, which is based on a different toolkit than what we will explore in Chapter~\ref{c:probability-matrixconcentration}, we point the reader to~\cite{Tropp:TailBoundsRM}). For now we will use this inequality to show that the SDP approach presented above achieves exact recovery of communities in the stochastic block model.

\subsection{Using matrix concentration}
In this section we show how the resulting question amounts to controlling the largest eigenvalue of a random matrix, which can be done with the matrix concentration tools described above.

Let us start by noting that
\[
\EE\left[ 2 L_{\text{SBM}} +11^T \right] = 2\EE L_{\text{SBM}} +11^T = 2\EE \Dp- 2\EE \Dm- 2\EE A  +11^T,
\]
and
$\EE \Dp = \frac{n}2 \pp I$, $\EE\Dm = \frac{n}2 \qq I$. Moreover, $\EE A$ is a matrix with four $\frac{n}2\times \frac{n}2$ blocks where the diagonal blocks have entries $\pp$ and the off-diagonal blocks have entries $\qq$.\footnote{For simplicity we assume the possibility of self-loops; notice however that the matrix in question does not depend on this, only its decomposition in the degree matrices and $A$.} In other words 
$$\EE A = \frac12\left( \pp + \qq \right)11^T +  \frac12\left( \pp - \qq \right)   gg^T.$$
This means that
\[
\EE\left[ 2 L_{\text{SBM}} +11^T \right] = \left( (\alpha-\beta) \log n\right)I  + \left( 1 - (\alpha+\beta)\frac{\log n}n \right)11^T - (\alpha-\beta)\frac{\log n}n gg^T.
\]

Since $L_{\text{SBM}}g=0$ and $11^Tg = 0$, we can safely ignore what happens in the span of $g$, and it is not hard to see that
\[
\lambda_2 \left( \EE\left[ 2 L_{\text{SBM}} +11^T \right] \right) = (\alpha-\beta) \log n.
\]

Thus, it is enough to show that
\begin{equation}\label{largedeviations_matrix}
\left\|  L_\text{SBM}  - \EE\left[ L_\text{SBM} \right] \right\| < \frac{\alpha-\beta}2 \log n,
\end{equation}
which is a large deviation inequality; recall that $\|\cdot\|$ denotes operator norm.

The idea is to write $L_\text{SBM}  - \EE\left[ L_\text{SBM} \right]$ as a sum of  independent random matrices and use the Matrix Bernstein Inequality (Theorem~\ref{thm:4:MatrixBernstein}). This gives an illustrative example of the applicability of matrix concentration tools, as many random matrices of interest can be rewritten as sums of independent matrices.

Let us start by defining, for $i$ and $j$ in the same community (i.e. $g_i=g_j$), 
\begin{align*}
\gamma^+_{ij}=\begin{cases} 1 & \text{ if } (i,j)\in E \\ 0 & \text{otherwise,}\end{cases}
\end{align*}
and
\begin{align*}
\Delta^+_{ij}&= (e_i-e_j)(e_i-e_j)^T,
\end{align*}
where $e_i$ (resp. $e_j$) is the vector of all zeros except the $i^{th}$ (resp. $j^{th}$) coefficient which is 1.

For $i$ and $j$ in different communities (i.e. $g_i \neq g_j$), define 
\begin{align*}
\gamma^-_{ij}=\begin{cases} 1 & \text{ if } (i,j)\in E \\ 0 & \text{otherwise,} \end{cases}
\end{align*}
and
\begin{align*}
\Delta^-_{ij}&= -(e_i+e_j)(e_i+e_j)^T.
\end{align*}
 We have
\[
L_\text{SBM} = \sum_{i<j: g_i=g_j} \gamma^+_{ij} \Delta^+_{ij}+\sum_{i<j: g_i\neq g_j} \gamma^-_{ij} \Delta^-_{ij}.
\]
We note how $(\gamma_{ij}^+)_{i,j}$ and  $(\gamma_{ij}^-)_{i,j}$ are jointly independent random variables with expectations $\EE (\gamma_{ij}^+) = \frac{\alpha\log n}{n}$ and $\EE (\gamma_{ij}^-)= \frac{\beta\log n}{n}$. $ \Delta^+_{ij}$ and $\Delta^-_{ij}$ are deterministic matrices. This means that
\[
L_\text{SBM} - \EE L_\text{SBM} = \sum_{\substack{i<j:\\ g_i=g_j}} \left(\gamma^+_{ij} - \frac{\alpha\log n}{n}\right) \Delta^+_{ij}+\sum_{\substack{i<j:\ g_i\neq g_j}} \left(\gamma^-_{ij} - \frac{\beta\log n}{n} \right) \Delta^-_{ij}.
\]

We can then use Theorem~\ref{thm:4:MatrixBernstein} by setting
\begin{equation}\label{eq:matrix_inside_sigma2}
\sigma^2 = \Big\|  \mathrm{Var}\left[\gamma^+\right]\sum_{i<j:\ g_i=g_j}\left(\Delta^+_{ij}\right)^2+ \mathrm{Var}\left[\gamma^-\right]\sum_{i<j:\ g_i\neq g_j}\left(\Delta^-_{ij}\right)^2  \Big\|,
\end{equation}
and $R=2$, since $\left\|\Delta^+_{ij}\right\|=\left\|\Delta^-_{ij}\right\|=2$ and both $(\gamma_{ij}^+)_{i,j}$ and $(\gamma_{ij}^-)_{i,j}$ take values in $[-1,1]$. Note how this bound is for the spectral norm of the summands, not just the largest eigenvalue, as our goal is to bound the spectral norm of the random matrix.
In order to compute $\sigma^2$, we write
$$ \sum_{i<j:\ g_i=g_j}\left(\Delta^+_{ij}\right)^2 = nI - \left(\1\1^T + gg^T\right),$$
and
$$ \sum_{i<j:\ g_i\neq g_j}\left(\Delta^-_{ij}\right)^2 = nI + \left(\1\1^T - gg^T\right)$$.

Since $\mathrm{Var}\left[\gamma^+\right] \leq \frac{\alpha\log n}{n}$, $\mathrm{Var}\left[\gamma^-\right] \leq \frac{\beta\log n}{n}$, and all the summands are positive semidefinite we have
\[
\sigma^2 \leq \left\| \frac{(\alpha+\beta)\log n}{n}\left(nI-gg^T\right) - \frac{(\alpha-\beta)\log n}{n}\1\1^T \right\| = (\alpha+\beta)\log n.
\]

Using Theorem~\ref{thm:4:MatrixBernstein} for $t= \frac{\alpha-\beta}2\log n$ on both the largest and smallest eigenvalue gives
\begin{align*}
& \Prob\left\{ \left\|  L_\text{SBM}  - \EE\left[ L_\text{SBM} \right] \right\| \geq \frac{\alpha-\beta}2  \log n \right\}    \leq{}  \\
 &  \hspace*{47mm} \le    2n \cdot \exp\left(  \frac{-\left( \frac{\alpha-\beta}2 \log n \right)^2}{2\left(\alpha+\beta\right) \log n  + \frac43\left( \frac{\alpha-\beta}2 \log n \right)} \right) \hspace*{40mm} \\
 & \hspace*{47mm} =  2 \cdot \exp\left( -\frac{(\alpha-\beta)^2 \log n }{8\left(\alpha+\beta\right)  + \frac{8}3\left( \alpha-\beta  \right)} + \log n \right)\\
 & \hspace*{47mm} =  2 n^{-\left(\frac{(\alpha-\beta)^2 }{8\left(\alpha+\beta\right)  + \frac{8}3\left( \alpha-\beta  \right)} -1 \right)}.
\end{align*}

Together with Lemma~\ref{lemma:lemma1SDP}, this implies that as long as
\begin{equation}\label{SBM_exact_suboptimalcondition}
(\alpha-\beta)^2 > 8\left(\alpha+\beta\right) + \frac{8}3\left( \alpha-\beta  \right),
\end{equation}
the semidefinite programming relaxation~\eqref{MLE_SDP} recovers the true partition, with probability tending to $1$ as $n$ increases.

While it is possible to obtain a stronger guarantee for this relaxation, the derivation above illustrates the matrix concentration technique in a simple, yet powerful, instance. In fact, the analysis in~\cite{Abbe_SBMExact} uses the same technique. 
 In order to obtain a sharp guarantee (Theorem~\ref{thm:10:SBM_extact_SDP} below) one needs more specialized tools. We refer the interested reader to~\cite{Bandeira_Laplacian} or~\cite{Hajek_et_al_SBM_SDP} for a discussion and proof of Theorem~\ref{thm:10:SBM_extact_SDP}; the main idea is to separate the diagonal from the non-diagonal part of $L_\text{SBM}  - \EE\left[ L_\text{SBM} \right]$, treating the former with scalar concentration inequalities, and the latter with specialized matrix concentration inequalities such as the ones in~\cite{Bandeira_NARandomMatrixBound}.

\begin{theorem}\label{thm:10:SBM_extact_SDP}
Let $G$ be a random graph with $n$ nodes drawn according to the stochastic block model on two communities with edge probabilities $p$ and $q$. Let $p = \frac{\alpha \log n}n$ and $q = \frac{\beta \log n}n$, where $\alpha> \beta$ are constants. Then, as long as
\begin{equation}\label{eq:conditionsalphabetaSBMcor_2}
 \sqrt{\alpha} - \sqrt{\beta} > \sqrt{2},
\end{equation}
 the semidefinite program considered above coincides with the true partition with high probability. \end{theorem}

Note that, if 
\begin{equation}\label{eq:10:exactrecoveryfails}
 \sqrt{\alpha} - \sqrt{\beta} < \sqrt{2},
\end{equation}
then exact recovery of the communities is impossible, meaning that the SDP algorithm is optimal. Furthermore, in this regime~\eqref{eq:10:exactrecoveryfails}, one can show that there will be a node on each community that is more connected to the other community than to its own, meaning that a partition that swaps them would have more likelihood. The fact that the SDP will start working essentially when this starts happening appears naturally in the analysis in~\cite{Bandeira_Laplacian}. Later, it was proven that the spectral method~\eqref{MLE_A_spectralmethod}, followed by a simple thresholding step, also gives exact recovery of the communities~\cite{Abbe_et_al_spectralexact}. An analogous analysis has recently been obtained for the (normalized or unnormalized) graph Laplacian in place of the adjacency matrix, see~\cite{denglingstrohmer}. However, the proof techniques for the graph Laplacian are different and a bit more involved, since---unlike the adjacency matrix---the graph Laplacian does not  exhibit row/column-wise independence.

\begin{remark}
An important advantage of semidefinite relaxations is that they are often able to produce certificates of optimality. Indeed, if the solution of the relaxation~\eqref{MLE_SDP} is rank 1 then the user is sure that it must be the solution of~\eqref{MLE_B}. These advantages, and ways of producing such certificates while bypassing the need to solve the semidefinite program are discussed in~\cite{Bandeira_PCC}.  Semidefinite relaxation has also been effectively used to show rigorously that spectral clustering can indeed outperform clustering via $k$-means~\cite{ling2020certifying,boedihardjo2021performance,deng2021strong}
\end{remark}

\bigskip

In fact, the convex relaxation of optimization problems involving low-rank constraints plays an important role in many areas of science and engineering. There one often one seeks to minimize the rank of a matrix subject to linear or convex constraints. Rank minimization is intrinsically non-convex and combinatorial, making it computationally intractable in general. A standard convex relaxation replaces the rank function with the {\em nuclear norm} (or Schatten-1 norm), defined as the sum of the singular values of a matrix. 

This relaxation leads to tractable convex programs—often semidefinite programs or efficiently solvable via first-order methods—that promote low-rank solutions while retaining strong theoretical guarantees. Amazingly, under suitable conditions, nuclear norm minimization is provably {\em exact}, recovering the original minimum-rank solution. As a result, convex nuclear norm minimization provides a principled and computationally efficient surrogate for rank minimization in a wide range of applications, including matrix completion, system identification, and robust principal component analysis. We will discuss this idea and several applications in detail in Chapter~\ref{c:lowrank}.

\section*{Exercises}
\addcontentsline{toc}{section}{Exercises}

\begin{myexercise}
[\level \level \sep Max-Cut Bound]
\label{prob:max-cut-bound}
Show that every undirected graph with $|E|$ edges contains a cut that separates at least $|E|/2$ edges.
\begin{hint}
Consider the expected cut value of a random partition $(S,S^c)$ where each vertex has probability $1/2$ to be in $S$.   
\end{hint}
\end{myexercise}

\begin{myexercise}[\level \level \sep Random MaxCut and Boosting]
    \label{prob:random_max_cut}
    We consider the following naive (but surprisingly effective) procedure to find a large cut in a graph $G$ with an even number of vertices: choose a set $S$ of $n/2$ vertices uniformly at random in $G$. We want to show that the partition $(S, S^c)$ cuts a large number of edges with some (small but positive) probability. Then we boost the procedure to increase the probability of finding a large cut. 

    \begin{enumerate}[(a)]
        \item Show that for any fixed $\eps \in (0, \frac{1}{2})$
        $$
        \Prob \left( \cut(S) > \left(\frac{1}{2} - \eps\right) |E|  \right) \ge \eps,
        $$
        where $|E|$ is the number of edges in graph $G$. 
        \begin{hint}
        For (a) it may be easier to study the set of edges that are NOT being cut.
        \end{hint}
        \item The result of the previous subproblem is rather unsatisfying, since if we want to find a cut with $0.49 |E|$ edges, the probability of success may be as low as $0.01$. For this reason we will modify our procedure, namely to improve our accuracy we sample $S$ several times.

        Suppose we run the procedure $k$ times and get sets $S_1, \dots, S_k$. We want to construct a set $S^*$ from these outputs such that for any $\delta \in (0,1)$,
        $$
        \Prob\left(\cut(S^*) > \left(\frac{1}{2} - \eps\right) |E|  \right) \ge 1 - \delta
        $$
        
        Find such cut and give an estimate on the required number of trials $k(\delta, \epsilon)$ depending on probability parameter~$\delta$ and approximation parameter $\epsilon$.
    \end{enumerate}

\emph{Remark: observe that the same technique can be applied to other randomized algorithms as well. }

\end{myexercise}

\begin{myexercise}[\level \level \sep Dual SDP]
    \label{prob:dual_sdp}
       To find the solution to the community detection problem in SBM, in the course we introduce convex relaxation of the problem and subsequently use convex duality to certify the optimality. In this problem, we will find the dual problem using the Lagrangian function. 
    
    Recall the definition of a semidefinite program (SDP).

    \begin{definition}
    A semidefinite program (SDP) is an optimization problem of the following type:
    \begin{equation}\label{eq:SDP_def}
        \max_{X\in \R^{n\times n}} \langle A, X \rangle \quad \text{subject to } X \succeq 0, \langle B_i, X \rangle = b_i,\quad i = 1, \dots, m,
    \end{equation}

    where $A, B_1, \dots, B_m \in \mathbb{R}^{n \times n}$ and $b_i \in \mathbb{R}$ are given.
    \end{definition}

    In SDPs, one of the constraints is positive semidefiniteness of a matrix. This constraint can be incorporated in the Lagrangian function as follows:
    $$
    \mathcal L(X, \nu, Y) =  \langle A, X \rangle + \sum_{i=1}^m \nu_i ( b_i - \langle B_i, X \rangle) + \langle Y, X \rangle,
    $$
    where $Y \in \R^{n\times n}$ is positive semidefinite matrix, and $\nu \in \R^m$. Using this Lagrangian, we can easily check that 
    $$
    p^* = \max_X \min_{\substack{\nu, Y\\Y \succeq 0}}\mathcal L(X, \nu, Y)
    $$
    coincides with the optimal value of the original SDP \eqref{eq:SDP_def}.
    \begin{enumerate}[(a)]
        \item Using the expression for the Lagrangian function, find the dual function 
        $$
        g(Y, \nu) =  \max_{X\in \R^{n\times n}} \mathcal L(X, \nu, Y)
        $$
        defined for PSD matrices $Y\in \R^{n\times n}$ and $\nu \in R^m$ (note that the dual function may be infinite for certain values of $Y$). Then write the dual program of the SDP \eqref{eq:SDP_def} (the dual program just minimizes the dual function and contains the constraints that prevent the dual function from being infinite). 

        \item Using (a), find the dual of the following semidefinite program: 
        \begin{align*}
        \max\quad &\tr(BX) \\
        \text{s.t.}\quad &X_{ii} = 1 \quad\text{ for each } i\\
        &X\succeq 0.
        \end{align*}

    \end{enumerate}
    
\end{myexercise}

\begin{myexercise}[\level\level\level\sep Connectedness of the Erd\H{o}s-R\'enyi graph]
    \label{prob:erdos_renyi_connectedness}

We define the Erd\H{o}s-R\'enyi graph as a random graph $G \sim \mathcal{G}(n,p)$ with $n$ vertices generated by placing each possible edge independently at random with probability $p$. The Erd\H{o}s-R\'enyi graph is a popular model to study the performance of several optimization algorithms on graphs. Many of these algorithms rely on the graphs being connected and in this problem we study when this is the case for the Erd\H{o}s-R\'enyi model.

We define $p:=\frac{\lambda\log n}{n}$ for some constant $\lambda >0$.

\begin{enumerate}[(a)]
    
    \item Prove that if $\lambda \le 1-c$, where $c>0$ is an absolute constant, then the graph $G$ has an isolated vertex with probability $1-o(1)$. (We use the standard asymptotic notation,  $f(n) = o(1)$ if $\lim_{n \rightarrow \infty}f(n) = 0$.)

    \begin{hint}
       For (a) consider the random variable that counts the number of isolated vertices.
    \end{hint}

    \begin{hint}
        Use Problem~\ref{prob:paley_zygmund}.
    \end{hint}
    
    \item Now observe the following: A graph is disconnected if and only if there exits a set of $k$ nodes such that $k\le \floor{\frac{n}{2}}$ and there is no edge connecting the set of $k$ nodes with the complement set of $n-k$ nodes. Use this fact to prove that if $\lambda \ge 1+c$ for an absolute constant $c>0$, then the graph is connected with probability $1-o(1)$.
    
\end{enumerate}

\end{myexercise}

\begin{myexercise}[\level\sep Sum of Squares Proof]
    \label{prob:sos_proof}
        Let $x,y$ be real numbers, prove
        $$ x^4 + y^4 + 4xy + 2 \geq 0.$$
        \begin{hint}
        Add and subtract an appropriate monomial.
    \end{hint}
\end{myexercise}

\begin{myexercise}[\level\sep Smallest Eigenvalue Program]
    \label{prob:smallest_eigenvalue}

Let $A \in \R^{d \times d}$ be a symmetric matrix. Prove that the following optimization problem has the smallest eigenvalue of $A$ as optimal value:
\begin{align*}
        \min \quad &\Tr(AX) \\
        \text{s.t.}\quad &\Tr(X) = 1 \\
        &X\succeq 0.
        \end{align*}
(Recall that $X \succeq 0$ means that $X$ is PSD.)
\end{myexercise}

\begin{myexercise}[\level\level\sep Discrepancy Relaxation]
    \label{prob:discrepancy_relaxation}
    
    Let $A \in \R^{d \times m}$ be a matrix. We define its discrepancy as the optimal value of the following minimization problem:
    $$ \operatorname{disc}(A) = \min_{\eps \in \{-1,1\}^m} \norm{A \eps}_{\infty}.$$
    The vector discrepancy of $A$ is the minimal value of the problem
    $$ \operatorname{vecdisc}(A) = \min_{u_1, \ldots, u_m \in \mathbb{S}^{m-1}} \max_{1 \leq i \leq d} \norm{ \sum_{j=1}^m A_{i,j}u_j}_2,$$
    where $\mathbb{S}^{m-1} \subset \R^m$ denotes the euclidean unit sphere, so $\norm{u_i}_2 = 1$ for all $1\leq i \leq m$. The goal of this exercise is to show that vector discrepancy is a convex relaxation of discrepancy, which can be solved using a semidefinite program.
    \begin{enumerate}[(a)]
        \item Prove the inequality
        $$ \operatorname{vecdisc}(A)^2 \leq \operatorname{disc}(A)^2.$$
        \item Prove that the quantity $\operatorname{vecdisc}(A)^2$ is the optimal value of the following semidefinite program:
        \begin{align*}
        \min \quad &D \in \R\\
        \text{s.t.}\quad &(AXA^\top)_{i,i} \leq D \quad \forall 1 \leq i \leq d  \\
        \text{and} \quad &X_{i,i} = 1 \quad \forall 1 \leq i \leq m \\
        &X\succeq 0 \in \R^{m \times m}.
        \end{align*}
    \begin{hint}
        Use the square-root of the matrix $X$ to construct your unit vectors.
    \end{hint}
    \end{enumerate}
    
\end{myexercise}

\begin{myexercise}[\level\level\level\sep Minimum Bisection and Community Detection]
    \label{prob:min_bisection_community_detection}
    
    The goal of this exercise is to relate the minimum bisection problem with exact recovery in the community detection problem. Assume that $n$ is even and consider the graph with $n$ vertices drawn from the stochastic block model with two balanced communities, i.e, each community has size $n/2$, moreover, the two communities are chosen uniformly at random. Let $p$ be the probability that an edge is placed inside the communities and $q$ across the communities with $p>q$. 

Our goal is to estimate the partition $\Omega$ induced by the communities with an estimator $\hat{\Omega}(G)$ that depends only on one sample of the random graph $G$. Prove that the estimator that minimizes the probability of error is equivalent to solve the minimum bisection of the observed graph $G$ (the minimum bisection is a partition into two equally-sized subsets, such that the number of edges being cut by such a partition is minimal). The probability of error $P_e$ is given by
\begin{equation*}
P_e:=\mathbb{P}(\hat{\Omega}\neq \Omega) = \sum_{g}\mathbb{P}(\hat{\Omega}(G)\neq \Omega|G=g)\mathbb{P}(G=g).
\end{equation*}
Here the sum is taken over all possible realizations of the random graph $G$.

\begin{hint}
    Use Bayes' rule to simplify and note that you can ignore terms that do not depend on $\hat{\Omega}$.
\end{hint}

\end{myexercise}

\begin{myexercise}[\level\sep PSD Set Convexity]
    \label{prob:psd_convexity}
    Show that the set $S_n^+ = \{A \in \R^{n\times n}: A \succeq 0 \}$ is convex and that it is invariant under multiplication with a positive scalar.
\end{myexercise}

\begin{myexercise}[\level\level\sep Spectral Algorithm for Planted Clique]
    \label{prob:spectral_planted_clique}

    We want to analyze parts of a spectral algorithm, which is used to find the largest clique in a graph $G$ on $n$ vertices. This algorithm is often analyzed for the planted clique model, where $G \sim \mathcal{G}(n,1/2)$ is random Erd\H{o}s-R\'enyi graph and then $k$ vertices of $G$ are randomly uniformly selected and then edges will be added to $G$ until these $k$ vertices become a clique (fully connected amongst each other). We call the graph we get after this procedure $\tilde{G}$. The goal is to find this planted clique with the so called "AKS spectral algorithm", which relies on computing the top eigenvector of the matrix $M \coloneqq A - \frac12 \mathbf{1}_n\mathbf{1}_n^\top$, where $A$ is the adjacency matrix of $\tilde{G}$ and $\mathbf{1}_n \in \R^n$ is the all-ones vector. The idea behind this algorithm is that the matrix $M$ is typically close to the matrix $\frac 12 \mathbf{1}_S \mathbf{1}_S^T$, where $\mathbf{1}_S$ is the indicator vector of the planted clique $S$, and if the matrices are close, then their top eigenvectors should in some sense also be close. This is the part that you will prove in this exercise:
    \begin{enumerate}[(a)]
        \item Let $0<  \eps < 1$, and suppose there exists a symmetric matrix $M \in \R^{n \times n}$ and a subset $S \subset [n]$ with the property $\abs{S} > 2(1+ \eps^{-1})\norm{M- \frac12 \mathbf{1}_S \mathbf{1}_S^\top}$, where $\mathbf{1}_S \in \R^n$ is the indicator vector of $S$ (so $1$ whenever the coordinate is in $S$ and $0$ otherwise). If $v$ is an eigenvector corresponding to the largest eigenvalue of $M$ with norm $\norm{v}_2^2 = \abs{S}$, prove that the following inequality holds:
        $$ \min \{ \norm{v - \mathbf{1}_S}_2^2 , \norm{-v - \mathbf{1}_S}_2^2 \} \leq 2\abs{S} \eps^2.$$
        You may use the following theorem without proof:
        \begin{theorem}
            Let $M \in \R^{n \times n}$ be a symmetric matrix and let $v$ be an eigenvector corresponding to the largest eigenvalue of $M$. Let $y \in \R^n$ be any vector and let $\theta$ be the angle between $y$ and $v$, then
            $$ \abs{\sin(\theta)} \leq \frac{\norm{M-  yy^\top}}{|\norm{yy^\top} - \norm{M-  yy^\top}|}$$
        \end{theorem}

    \begin{hint}
        Use the fact that the dot product of $\mathbf{1}_S$ and $v$ depends on the angle between these vectors.
    \end{hint}
    \end{enumerate}
\end{myexercise}

\begin{myexercise}[\level\level\sep "Baby" Davis-Kahan Theorem]
    \label{prob:baby_davis_kahan}
    The aim of this problem is to prove the matrix peturbation theorem you were given in Problem~\ref{prob:spectral_planted_clique}. Let $M \in \R^{n \times n}$ be a symmetric matrix and let $v$ be a unit-length eigenvector corresponding to the largest eigenvalue of $M$ (in magnitude). Let $y \in \R^n$ be any vector and let $\theta$ be the angle between $y$ and $v$, then
    $$ \abs{\sin(\theta)} \leq \frac{\norm{M-  yy^\top}}{|\norm{yy^\top} - \norm{M-  yy^\top}|}$$
     \begin{enumerate}[(a)]
        \item Let $P_{y^\perp}$ be the $n \times n$ orthogonal projection matrix that projects a vector onto the orthogonal complement of $y$ and let $\lambda$ be the eigenvalue of $M$ corresponding to $v$. Prove
        $$ |\lambda| \norm{P_{y^\perp}v } \leq \norm{M-yy^\top}. $$
        \item Prove $|\lambda| \geq |\norm{yy^\top} - \norm{M-  yy^\top}|$.
        \item Show $\abs{\sin(\theta)} =  \norm{P_{y^\perp}v } $ and finish the proof of the theorem.
    \end{enumerate}
    
\end{myexercise}

\begin{myexercise}[\level\sep Sign Rounding]
    \label{prob:sign_rounding}
    As seen in the lecture, often we would like to round an eigenvector we computed in a spectral algorithm to a sign vector. A very naive (but sometimes effective) method is to simply take the sign of every entry of the vector. So given $x \in \R^d$ we define $\operatorname{sgn}(x) \in \R^d$ by $\operatorname{sgn}(x)_i = 1$ if $x_i \geq 0$ and $\operatorname{sgn}(x)_i = -1$ otherwise. This method works well when the eigenvector is already close to a sign vector as you will show.
     \begin{enumerate}[(a)]
        \item Let $x \in \R^d$ be a vector and let $y \in \{-1,+1\}^d$ be a sign vector. Prove that if 
        $$ \norm{x-\frac{1}{\sqrt{d}}y}_2 \leq \varepsilon,$$
        then the number of indices $i$ with $\operatorname{sgn}(x)_i \neq y_i$ is at most $\eps^2 d$.
        \item Conclude $ \frac{1}{\sqrt{d}}\norm{\operatorname{sgn}(x)-y}_2 \leq 2\varepsilon.$
    \end{enumerate}
    
\end{myexercise}

\begin{myexercise}[\level\level\sep Little Grothendieck Problem]
    \label{prob:little_grothendieck}
    Let $C\succeq 0$ ($C \in \R^{n \times n}$ is positive semidefinite). In this problem you will show an approximation ratio of $\frac{2}{\pi}$ to the problem
\[
 \max_{x_i=\pm 1}\sum_{i,j=1}^n C_{ij}x_ix_j.
\]

Similarly to \texttt{Max-Cut}, we consider
\[
 \max_{\substack{v_i\in \R^n \\ \|v_i\|^2 = 1}}\sum_{i,j=1}^n C_{ij}v_i^Tv_j.
\]

The goal is to show that, for $ g \sim\mathcal{N}\left(0,I_{n\times n}\right)$, taking $x_i= \operatorname{sign}(v_i^T g)$ a randomized rounding,
\begin{equation}\label{eq:lilGrothendieck}
 \E \left[ \sum_{i,j=1}^n C_{ij}x_i x_j  \right] \geq \frac{2}{\pi} \sum_{i,j=1}^n C_{ij}v_i^Tv_j
\end{equation}
The difficulty lies in the fact that $\E[x_i x_j]$ is not easy to compute, which is why we divide this exercise into two parts.

\begin{enumerate}[(a)]
    \item Compute the quantity $\E[\operatorname{sign}(v_i^T g) \langle v_j, g \rangle]$.
    \begin{hint}
        This number should only depend on the inner product between $v_j$ and $v_i$.
    \end{hint}
    \item Define the matrix $S \in \R^{n \times n}$ with entries 
    $$S_{i,j} = (\langle v_i, g \rangle - \sqrt{\pi/2}\operatorname{sign}(v_i^T g))(\langle v_j, g \rangle - \sqrt{\pi/2}\operatorname{sign}(v_j^T g)).$$ Show that 
    $$ \Tr(CS) \geq 0$$
    holds, and use this fact to prove the inequality~\eqref{eq:lilGrothendieck}.
\end{enumerate}
\end{myexercise}

\begin{myexercise}[\level\level\sep Minimum Bisection Relaxation]
    \label{prob:min_bisection_relaxation}
    
    In Problem~\ref{prob:min_bisection_community_detection} we saw that the problem of community detection is equivalent to solving the minimum bisection problem in a graph $G$. Unfortunately, solving the minimum bisection problem is NP-hard, so we would like to solve a semidefinite relaxation of it. For a graph on $n$ vertices with adjacency matrix $A$ the semidefinite relaxation of the minimum bisection problem is given by 
    \begin{align*}
        \max \quad & \sum_{i,j=1}^n A_{i,j} \langle u_i, u_j \rangle  \in \R\\
        \text{s.t.}\quad &\langle u_i, u_i \rangle =1 \quad \forall 1 \leq i \leq n  \\
        \text{and} \quad &\sum_{i=1}^n u_{i} = 0 \\
        &u_i \in \R^n \quad \forall 1 \leq i \leq n.
    \end{align*}
    Find matrices $B_1, \ldots, B_k$ and real numbers $b_1, \ldots, b_k$ for some integer $k$, such that the problem can be written as
    \begin{equation}\label{eq:SDP_bisection}
        \max_{X\in \R^{n\times n}} \langle A, X \rangle \quad \text{subject to } X \succeq 0, \langle B_i, X \rangle = b_i,\quad i = 1, \dots, k.
    \end{equation}
\end{myexercise}


\chapter{Concentration of Measure and Gaussian Analysis}
\label{c:probability-gaussiananalysis}

In this chapter we significantly expand on the concepts presented in Chapter~\ref{c:surprises}, showcasing several other instances of the \emph{Concentration of Measure} phenomena, and focusing on Gaussian Analysis to develop a toolkit that is used throughout the book, including the celebrated Gordon's Escape through a Mesh Theorem. Our treatment draws inspiration from the excellent texts~\cite{vanHandel_LectureNotesProb_14,Ramon-StFlour-2022}.

\emph{Notation}: In this chapter, we try to reserve $g$ for an isotropic Gaussian vector and use $X$ and $Y$ to denote Gaussian vectors with other covariances.

\section{Gaussian integration by parts and Wick's formula}

We start by covering some basics in Gaussian analysis, starting with a very useful fact about Gaussian random variables: Gaussian integration by parts.

\begin{lemma}[Gaussian integration by parts]\label{lemma:GIbP}
Let $g$ be an $n$ dimensional vector of iid $\NNN(0,1)$ random variables ($g\sim\NNN(0,I_{n\times n})$). Let $\psi:\RR^n\to\RR$ be any differentiable function with at most exponential growth,\footnote{We will not make an effort to describe the minimal regularity assumptions as to not deviate from the main goal of this book. When we say differentiable with at most exponential growth we mean that the derivatives also are of at most exponential growth. The interested reader will realize that this is tightly connected to Sobolev spaces on the Gaussian measure.} then for any $i\in[n]$,
\[
\EE[g_i \psi(g)] = \EE\left[ \frac{\partial \psi}{\partial x_i}(g) \right].
\]
\end{lemma}

\begin{proof}
For $\phi$ a univariate function, integration by parts gives
\begin{eqnarray*}
\EE [ g_i \phi(g_i) ] & = & \int_{-\infty}^\infty x\phi(x) \frac{\exp\left(-\frac{x^2}{2}\right)}{\sqrt{2\pi}}dx \\
& = & - \int_{-\infty}^\infty \phi(x) \frac{d}{dx}\left(\frac{\exp\left(-\frac{x^2}{2}\right)}{\sqrt{2\pi}}\right)dx  \\
& = & \int_{-\infty}^\infty \frac{d}{dx} \phi(x) \frac{\exp\left(-\frac{x^2}{2}\right)}{\sqrt{2\pi}} dx = \EE \left[  \phi'(g_i) \right],
\end{eqnarray*}
as long as $\phi$ is differential and has at most exponential growth. The proof follows by taking $\phi(x)=F(g_1,\dots,g_{i-1},x,g_{i+1},\dots,g_n)$ and conditioning on $g_1,\dots,g_{i-1},g_{i+1},\dots,g_n$.
\end{proof}

It is straightforward to extend this formula to non-isotropic Gaussian vectors, and we record this here as a corollary.
\begin{corollary}\label{cor:GIbP}
Let $X$ be a centered $n$-dimensional Gaussian vector with covariance $\Sigma$, i.e. $X\sim\NNN(0,\Sigma)$. Let $\varphi\in\RR^n\to\RR$ be any differentiable function with at most exponential growth,
then for any $j\in[n]$,
\[
\EE[X_j \varphi(X)] = \sum_{k=1}^n  \Sigma_{jk}  \EE\left[\frac{\partial \varphi}{\partial x_k}(X)\right].
\]
\end{corollary}
\begin{proof}
We can write $X=\Sigma^{\frac12}g$ where $g$ has iid $\NNN(0,1)$ entries (note that $\Sigma \succeq 0$
). Defining $\psi(g) = \varphi(\Sigma^{\frac12}g) 
$ and using Lemma~\ref{lemma:GIbP} yields
\begin{eqnarray*}
\EE[X_j \varphi(X)] & = & \EE\left[\left(\Sigma^{\frac12}g\right)_{j} \varphi(\Sigma^{\frac12}g)\right] = \EE\left[\left(\Sigma^{\frac12}g\right)_{j} \psi(g)\right] \\
& = & \EE\left[ \left(\sum_{i=1}^n \Sigma^{\frac12}_{ji}g_i \right) \psi(g) \right]
=
 \sum_{i=1}^n \Sigma^{\frac12}_{ji} \EE\left[ g_i   \psi(g) \right] \\
 & = & 
 \sum_{i=1}^n \Sigma^{\frac12}_{ji} \EE\left[   \frac{\partial \psi}{\partial x_i}(g) \right] = \sum_{i=1}^n \Sigma^{\frac12}_{ji} \EE\left[ \sum_{k=1}^n \frac{\partial \varphi}{\partial x_k}(X) \Sigma^{\frac12}_{ki} \right] \\
 & = & \sum_{k=1}^n \left( \sum_{i=1}^n \Sigma^{\frac12}_{ji} \Sigma^{\frac12}_{ki} \right) \EE\left[\frac{\partial \varphi}{\partial x_k}(X) \right] = \sum_{k=1}^n  \Sigma_{jk}  \EE\left[\frac{\partial \varphi}{\partial x_k}(X) \right],
\end{eqnarray*}
where in the first equality in the third line we used Lemma~\ref{lemma:GIbP}. The last equality follows from matrix multiplication and the fact that $\Sigma^{\frac12}$ is a symmetric matrix.
\end{proof}

\begin{lemma}[Gaussian moments]\label{lemma:gaussianmoments}
Let $g$ be a standard univariate Gaussian. We have
\[
\EE g^{2p} = (2p-1)!!,
\]
where $(2p-1)!! = (2p-1)(2p-3)\cdots3\cdot1$.
Also, 
\[
\left[\EE g^{2p}\right]^{\frac1{2p}} \leq \sqrt{2p}.
\]
\end{lemma}

\begin{proof} This is a straightforward application of Gaussian integration by parts (Lemma~\ref{lemma:GIbP}). Let $\phi(g)=g^{2p-1}$ then $\EE[g\cdot g^{2p-1}] = \EE[(2p-1) g^{2p-2}]$. Iterating and using $\EE g^2=1$ gives $\EE g^{2p} = (2p-1)!!$. While it is easy to see that $(2p-1)!!\leq (2p)^{p}$, which proves the intended inequality, it is instructive to see a different, direct, proof:

Since $\EE[g^{2p}] = \EE[(2p-1) g^{2p-2}]$ and, by Jensen, $\EE[g^{2p-2}]\leq \left(\EE[g^{2p}]\right)^{\frac{2p-2}{2p}}$, we have 
\[
\EE[g^{2p}] \leq (2p-1) \left(\EE[g^{2p}]\right)^{\frac{2p-2}{2p}},
\]
which can be rewritten as $\left(\EE[g^{2p}]\right)^{\frac{1}{p}}\leq (2p-1)$.
\end{proof}

We will be interested in computing trace moments of Gaussian matrices, which will involve computing expected values of polynomials of Gaussians. It is clear that the expected value of any polynomial of i.i.d.\ standard Gaussians can be computed using Gaussian integration by parts for each monomial (just as in Lemma~\ref{lemma:gaussianmoments}). Wick's formula\footnote{Wick's formula is also known as Isserlis' Theorem~\cite{isserlis1918formula} in statistics and probability and dates back to 1918. It was rediscovered in the context of quantum field theory by Wick in 1950~\cite{wick1950evaluation}.} is a very useful way of organizing this calculation. Before stating Wick's formula, we need to define the notion of pair partition.

\begin{definition}[Pair Partition]\label{def:pairpartition}
Given $k$ a positive integer, we define $\PP_2[k]$ as the set of partitions of $[k]$ into subsets of size 2 each. If $k$ is odd then $\PP_2[k]$ is empty. Given a function $u$ on $[k]$ and a pair partition $\nu\in\PP_2[k]$ we say that $u$ is compatible with $\nu$, and write $u\sim\nu$ if for all sets $\{i,j\}\in\nu$ we have $u(i)=u(j)$.
\end{definition}

\begin{lemma}[Wick's formula]\label{lemma:Wicksformula}
Let $g_1,\dots,g_n$ be iid standard Gaussian random variables and let $u:[p]\to[n]$ then
\[
\EE[g_{u(1)}\cdots g_{u(p)}] = \sum_{\nu\in \PP_2[p]} 1_{u \sim \nu},
\]
where $\PP_2[p]$ denotes a set of pair partitions and $u \sim \nu$ means that the function $u$ is compatible with $\nu$ (see Definition~\ref{def:pairpartition})
\end{lemma}

\begin{proof}
Since both sides are zero when $p$ is odd we can focus on the case when $p$ is even. We can then prove the identity by induction. It is trivially true for $p=2$, let $p\geq 4$ an even number. Using Gaussian integration by parts
\begin{eqnarray*}
\EE[g_{u(1)}\cdot g_{u(2)} \cdots g_{u(p)}] & = & \EE\left[\frac{\partial}{\partial g_{u(1)}} g_{u(2)} \cdots g_{u(p)}\right] \\
&=& \EE\left[\sum_{i=2}^p g_{u(2)} \cdots g_{u(i-1)} 1_{u(1)=u(i)} g_{u(i+1)}\cdots g_{u(p)}\right] \\
&=&\sum_{i=2}^p 1_{u(1)=u(i)} \EE\left[ g_{u(2)} \cdots g_{u(i-1)} g_{u(i+1)}\cdots g_{u(p)}\right]. \\
\end{eqnarray*}
Using the induction hypothesis for $\EE\left[ g_{u(2)} \cdots g_{u(i-1)} g_{u(i+1)}\cdots g_{u(p)}\right]$ (which is a product of $p-2$ factors) we get
\begin{eqnarray*}
\EE[g_{u(1)}\cdot g_{u(2)} \cdots g_{u(p)}]  & = & \sum_{i=2}^p 1_{u(1)=u(i)} \sum_{\nu'\in\PP_2([p]\setminus\{1,i\})} 1_{u_{|[p]\setminus\{1,i\}}\sim\nu'} \\
& = & \sum_{i=2}^p \sum_{\nu'\in\PP_2([p]\setminus\{1,i\} )} 1_{u\sim(\nu'\cup\{1,i\}) } = \sum_{\nu\in \PP_2[p]} 1_{u \sim \nu},
\end{eqnarray*}
where $u_{|[p]\setminus\{1,i\} }$ denotes the restriction of $u$ to the set $[p]\setminus\{1,i\}$.
\end{proof}

\section{Gaussian interpolation, Poincar\'{e}, and concentration}

In this section, we start with a classical tool in Gaussian analysis that we will use several times, namely the idea of Gaussian Interpolation.

\begin{lemma}[Gaussian Interpolation]\label{lemma:gaussianinterpolation}
Let $X$ and $Y$ be two centered independent $n$ dimensional Gaussian vectors with covariances, respectively $\Sigma^X$ and $\Sigma^Y$. Define, for $t\in[0,1]$,
\begin{equation}\label{eq:interpolationGI}
Z_t = \sqrt{t}X + \sqrt{1-t}Y,
\end{equation}
with $X$ and $Y$ independent from each other, and let $f:\RR^n\to\RR$ be a twice differentiable function with at most exponential growth. Then 
\begin{equation}
\frac{d}{dt} \EE \left[ f(Z_t) \right] = \frac12 \sum_{i,j=1}^n \left( \Sigma^X_{ij} - \Sigma^Y_{ij} \right) \EE \left[ \frac{\partial^2 f}{\partial x_i\partial x_j}(Z_t)\right].
\end{equation}
\end{lemma}

Before proving this lemma, we note that the parameterization of the path, although perhaps peculiar at first, is the one that in the case that $X$ and $Y$ are identically distributed, renders $Z_t$ identically distributed to them for any $t$; it is also the parameterization for which $\Cov(Z_t) = t\Cov(X)+(1-t)\Cov(Y)$, where $\Cov(X)=\EE XX^T$ (since $\EE X=0$).

\begin{proof} The idea will be to use Gaussian integration by parts (Corollary~\ref{cor:GIbP}). By the chain rule (and swapping the order of differentiating and taking the expectation) 
\begin{eqnarray}
\frac{d}{dt} \EE \left[ f(Z_t) \right]& = &  \EE \left[ \sum_{i=1}^n\frac{\partial f}{\partial x_i}(Z_t)\frac{d}{dt}(Z_t)_i\right] \nonumber \\ &= &  \EE \left[ \sum_{i=1}^n\frac{\partial f}{\partial x_i}(Z_t)\left(\frac1{2\sqrt{t}}X_i-\frac1{2\sqrt{1-t}}Y_i\right)\right]  \\ &= &  \EE \left[ \sum_{i=1}^n\frac{\partial f}{\partial x_i}(Z_t)\frac1{2\sqrt{t}}X_i\right]-\EE \left[ \sum_{i=1}^n\frac{\partial f}{\partial x_i}(Z_t)\frac1{2\sqrt{1-t}}Y_i\right]. 
\end{eqnarray}

For the first term, conditioning on $Y$ and using Corollary~\ref{cor:GIbP} on the random variable $\sqrt{t}X$ gives
\begin{eqnarray*}
\EE \left[ \sum_{i=1}^n\frac{\partial f}{\partial x_i}(Z_t)\frac1{2\sqrt{t}}X_i\right] & = & \frac1{2t}\EE \left[ \sum_{i=1}^n\frac{\partial f}{\partial x_i}(\sqrt{t}X + \sqrt{1-t}Y)\sqrt{t}X_i\right] \\ & = & \frac1{2t} \sum_{j=1}^n \left( t\Sigma^{X}\right)_{ij} \EE \left[ \sum_{i=1}^n\frac{\partial^2 f}{\partial x_j \partial x_i }(\sqrt{t}X + \sqrt{1-t}Y)\right]\\ & = & \frac1{2} \sum_{i,j=1}^n \Sigma^{X}_{ij} \EE \left[ \frac{\partial^2 f}{\partial x_j \partial x_i }(Z_t)\right].
\end{eqnarray*}
An analogous calculation for the second terms completes the proof.
\end{proof}

Gaussian interpolation will play (at least) two important roles in the following: (i)~together with the fundamental theorem of calculus it will allows us to compare $\EE f(X)$ with $\EE f(Y)$ for various functions $f$ and Gaussian random variables $X$ and $Y$, this is the key tool behind the comparison inequalities of Section~\ref{sec:gaussiancomparisonineqs}; (ii)~another important role of Gaussian interpolation is to provide concentration inequalities, roughly speaking by interpolating between $(X,X)$ containing two exact copies of a random variable and $(X,X')$ containing two independent copies, allows to quantitatively say how much two independent copies of a random variable behave similarly, resulting in concentration of measure. The first instance of this idea is in the so-called Gaussian covariance identity, but we will see it again in the Gaussian Poincar\'{e} inequality and in Gaussian concentration.

\begin{lemma}[Gaussian covariance identity]\label{lemma:gaussiancovarianceidentity}
Let $\phi$ and $\psi$ be two differentiable functions (from $\RR^n$ to $\RR$) of at most exponential growth. Let $g$ and $g'$ be two iid $n$ dimensional $\NNN(0,I)$ vectors. Define
\begin{equation}\label{eq:interpolationCI}
g_{(t)} = tg + \sqrt{1-t^2}g'.
\end{equation}
We have
\[
\cov(\phi(g),\psi(g)) = \int_{0}^1 \EE \langle \nabla\phi(g),\nabla\psi(g_{(t)}) \rangle dt.
\]
\end{lemma}
Before proving this lemma, let us remark on how the interpolation in~\eqref{eq:interpolationCI} is related to the one in~\eqref{eq:interpolationGI}. The interpolation used in~\eqref{eq:interpolationGI} is an interpolation between two distributions of random variables linearly parameterized on the covariance of these random variables. The interpolation in the covariance identity~\eqref{eq:interpolationCI} creates a path between a random variable and an i.i.d. copy of it, linearly parameterized by the covariance between $g$ and $g_{(t)}$. As we will see in the proof below, we can study~\eqref{eq:interpolationCI} by taking $Z_t = \sqrt{t}X+\sqrt{1-t}Y$ where $X$ and $Y$ are two $2n$-dimensional Gaussian random variables with covariances
\begin{equation}\label{eq:2diffinterpolations}
X \sim \NNN\left( 0, \left[\begin{array}{cc} I_{n\times n} & 0_{n\times n} \\  0_{n\times n} & I_{n\times n} \end{array}\right]\right) \text{ and } Y \sim \NNN\left( 0, \left[\begin{array}{cc} I_{n\times n} & I_{n\times n} \\  I_{n\times n} & I_{n\times n} \end{array}\right]\right),
\end{equation}
since, for each $t\in[0,1]$, $Z_t \sim \left[\begin{array}{c} g \\ g_{(t)}\end{array}\right]$ (in the since of having the same probability law).

\begin{proof}
Let us start by defining $\ell(t) = \EE \left[\phi(g)\psi(g_{(t)})\right]$. One has
\[
\ell(0) = \left(\EE \phi(g)\right)\left(\EE \psi(g)\right) \text{ and } \ell(1) = \EE \left[\phi(g)\psi(g)\right],
\]
thus
\[
\cov(\phi(g),\psi(g)) = \ell(1)-\ell(0) = \int_0^1 
\ell'(t)dt.
\]
The idea now is to compute $\ell'(t)$ using Lemma~\ref{lemma:gaussianinterpolation}. We leverage the construction above and consider $Z_t=\sqrt{t}X+\sqrt{1-t}Y$ where $X$ and $Y$ are independent Gaussian vectors distributed accordingly to~\eqref{eq:2diffinterpolations}. Since, for each $t\in[0,1]$, $Z_t$ has the same law as $\left[\begin{array}{c} g \\ g_{(t)}\end{array}\right]$ we have that $f(Z_t)$ has the same law as $\phi(g)\psi(g_{(t)})$ for $f:\RR^{2n}\to\RR$ given by $f\left(\left[\begin{array}{c} y \\ z\end{array}\right]\right) = \phi(y)\psi(z)$ for $y,z\in\RR^n$, and so $\ell(t) = \EE \left[ f(Z_t) \right]$

The rest of the proof is a straightforward computation using Lemma~\ref{lemma:gaussianinterpolation}:
\begin{eqnarray*}
\ell'(t) & = &  \frac{d}{dt}\EE \left[ f(Z_t) \right] =  \frac12 \sum_{a,b=1}^{2n} \left( \Sigma^{Z_1}_{ab} - \Sigma^{Z_0}_{ab} \right) \EE \left[ \frac{\partial^2 f}{\partial x_a\partial x_b}(Z_t)\right] \\
& = &\frac12 \left(2\sum_{i=1}^n \EE \frac{\partial}{\partial x_i}\phi(g) \frac{\partial}{\partial x_i}\psi(g_{(t)}) \right) \\
& = & \EE \langle \nabla\phi(g),\nabla\psi(g_{(t)}) \rangle.
\end{eqnarray*}
\end{proof}

An elegant consequence of this lemma is the celebrated Gaussian Poincar\'{e} inequality, which bounds the variance of a function of a Gaussian vector by its gradient yielding a form concentration of measure: ``If a function $f$ has small gradient, $f(X)$ has small variance''.

\begin{proposition}[Gaussian Poincar\'{e} Inequality]\label{prop:GaussianPoincare}
Let $f:\RR^n\to\RR$ be a differentiable function with at most exponential growth. Let $g\sim\NNN\left(0,I_{n\times n}\right)$. Then
\[
\Var\left( f(g) \right) \leq \EE \| \nabla f(g)\|^2.
\]
\end{proposition}
We note that this inequality (as the others in this chapter) can be effortlessly extended to Lipchitz functions by a standard approximation argument.

\begin{proof}
By the Gaussian covariance identity (Lemma~\ref{lemma:gaussiancovarianceidentity}) we have
\begin{eqnarray*}
\Var\left( f(g) \right) = \cov(f(g),f(g)) &=& \int_{0}^1 \EE \langle \nabla f(g),\nabla f(g_{(t)}) \rangle dt \\
& \leq  & \int_{0}^1 \left(\EE\|\nabla f(g)\|^2\right)^{\frac12}\left(\EE\|\nabla f(g_{(t)})\|^2\right)^{\frac12}  dt \\
& = & \int_{0}^1 \EE\|\nabla f(g)\|^2dt = \EE\|\nabla f(g)\|^2, 
\end{eqnarray*}
where the inequality follows from Cauchy-Schwarz on both the inner-product and the expectation and the penultimate equality follows from the fact that $g\sim g_{(t)}$.
\end{proof}

We will now prove one of the most important results in concentration of measure, Gaussian concentration. Although being a concentration result specific for normally distributed random variables, it will be very useful throughout this book. Much like Poincar\'{e}'s inequality above, it intuitively it says that if $F:\RR^n\to\RR$ is a function that is stable in terms of its input then $F(X)$ is well concentrated around its mean, where $X\in\NNN(0,I)$. Unlike Pointcar\'{e}'s inequality above, Gaussian concentration provides Gaussian tail bounds. More precisely:

\begin{theorem}[Gaussian Concentration]\label{theorem:4:gaussianconcentration}
 Let $g\sim\NNN(0,I_{n\times n})$ be a vector with i.i.d.\ standard Gaussian entries and $f:\RR^n\to\RR$ a $\sigma$-Lipchitz function (i.e.: $|f(x)-f(y)|\leq \sigma \|x-y\|$, for all $x,y\in\RR^n$). Then, for every $t\geq 0$
 \[
  \Prob\left\{ \left| f(g) - \EE f(g)\right| \geq t   \right\} \leq 2 \exp\left( -\frac{t^2}{2\sigma^2} \right).
 \]
\end{theorem}

\begin{proof}
We will assume that the function $f$ is smooth as a limiting argument can generalize the result from smooth functions to general Lipschitz functions. The existence of the integrals we use follows from the constraints that the Lipschitz property forces on growth of $f$. We also assume $\EE f(g)=0$ for ease of notation, as we can simply consider $\phi(g)=f(g)-\EE f(g)$.

We consider the moment generating function (as in~\eqref{eq:MomentGeneratingFunction})
\[
M_f(\lambda) = \EE\left[ \exp\left( \lambda f(g) \right)  \right],
\]
and note that
\[
M_f'(\lambda) = \EE\left[ f(g)\exp\left( \lambda f(g) \right) \right] = \cov\left(f(g),\exp\left( \lambda f(g) \right) \right),
\]
since $\EE f(g)=0$.

The Gaussian covariance identity (Lemma~\ref{lemma:gaussiancovarianceidentity}) then gives
\begin{eqnarray*}
M_f'(\lambda) &=& \int_{0}^{1}\EE \langle \nabla f(g),\nabla \exp\left( \lambda f(g_{(t)}) \right) \rangle dt \\ &=& \int_{0}^{1}\EE \left[ \langle \nabla f(g),\nabla f(g_{(t)}) \rangle \lambda   \exp\left( \lambda f(g_{(t)}) \right) \right] dt.
\end{eqnarray*}
The Lipchitz condition implies that $|\langle \nabla f(g),\nabla f(g_{(t)}) \rangle|\leq \sigma^2$ which together with the fact that $M_f(\lambda)\geq 0$ gives
\[
\left|M_f'(\lambda)\right| \leq \int_{0}^{1}\sigma^2\EE \left| \lambda   \exp\left( \lambda f(g_{(t)}) \right) \right|  dt = \sigma^2|\lambda| M_f(\lambda).
\]
Since $M_f(0)=1$ this implies (one can see this, e.g., by noting that $\log M_f(0)=0$ and $\left|\frac{d}{d\lambda}\log M_f(\lambda) \right| \leq |\lambda| \sigma^2$) that
\[
M_f(\lambda)\leq \exp\left(\frac{\lambda^2 \sigma^2}{2}\right).
\]
The rest of the proof follows from the standard Chernoff trick and Markov Inequality (the same we did in Proposition~\ref{prop:gaussian}), Formally, we showed that $f(g)$ is a sub-Gaussian random variable (in the sense of Definition~\ref{def:subgaussian}) and the proof can be finished by using Proposition~\ref{prop:subgaussian}.

\end{proof}

There is a direct argument for a weaker version of this inequality that does not require the covariance identity. We refer the interested reader to  Theorem~2.1.12 in~\cite{Tao_topicsRMT} (the original argument is due to Maurey and Pisier).

\subsection{Talagrand's concentration inequality}

A remarkable result by Talagrand~\cite{Talagrand_concentration_1995}, Talagrand's concentration inequality, provides an analogue of Gaussian concentration for bounded random variables.

\begin{theorem}[Talagrand concentration, Thm.~2.1.13~\cite{Tao_topicsRMT}]
Let $K>0$, and let $X_1,\dots,X_n$ be independent bounded random variables with $|X_i|\leq K$ for all $1\leq i \leq n$. Let $F:\RR^n \to \RR$ be a $\sigma$-Lipschitz and convex function. Then, for any $t\geq 0$,
\[
 \Prob\left\{ \left|F(X) - \EE\left[F(X)\right] \right| \geq t K\right\} \leq c_1 \exp\left( -c_2 \frac{t^2}{\sigma^2} \right), 
\]
for positive constants $c_1$, and $c_2$.
\end{theorem}

Other useful similar inequalities (with explicit constants) are available in~\cite{Massart_constants}.

\section{Gaussian comparison principles}\label{sec:gaussiancomparisonineqs}

We will start this section with the celebrated Slepian's Comparison Lemma, also known as the Sudakov-Fernique inequality,\footnote{They are essentially the same statement, on being a slight generalization of the other, we state here the more general form: Theorem~\ref{thm:Slepian_SF_ineq}.} are crucial tools to compare Gaussian processes. A Gaussian process is a family of Gaussian random variables indexed by some set $T$, $\left\{X_t\right\}_{t\in T}$ (if $T$ is finite this is simply a Gaussian vector). Given a Gaussian process $X_t$, a particular quantity of interest is $\EE\left[\sup_{t\in T}X_t\right]$.

We will focus on Gaussian Processes that are whose maximum is well approximated by maximum on finite sets, meaning Gaussian Processes for which
    \begin{equation}\label{eq:separablegaussianprocess}
    \EE\left[\sup_{t\in T}X_t\right] = \sup_{\substack{T' \subseteq T \\ |T'| < \infty}} \EE\left[\max_{t\in T'}X_t\right].
    \end{equation}

To not significantly deviate from the main message of this process we won't discuss regularity and measurability issue of Gaussian processes in detail, but note that the standard class of Gaussian processes that satisfy~\eqref{eq:separablegaussianprocess} are so-called \emph{separable Gaussian processes} (you can see a definition in Definition~5.22 of~\cite{vanHandel_LectureNotesProb_14}, it essentially requires that there exists a countable subset $T_0$ of the domain for which, for any $t\in T$, $X_t$ lies in the limit points of $X_s$ for $s\in T_0$ and $s\to t$). In any case, the Gaussian processes we will consider are either finite, which case they are trivially separable (and~\eqref{eq:separablegaussianprocess} holds), or they are defined over a separable set $T$ and have path continuity on the process metric $d(t,u) = \sqrt{\EE(X_t-X_u)^2}$ which also ensures~\eqref{eq:separablegaussianprocess}. In most cases, in fact, one can simply think of the approximating subsets as $\epsilon$-nets of $T$.

We will be particularly interested in comparing two different Gaussian processes, usually one of interest with one that is simpler to understand. 
Intuitively, if we have two Gaussian processes $X_t$ and $Y_t$ with mean zero $\EE\left[X_t\right] = \EE\left[Y_t\right] = 0$, for all $t\in T$, and the same variance, then the process that has the ``least correlations'' should have a larger maximum (think the maximum entry of vector with i.i.d. Gaussian entries versus one always with the same Gaussian entry). Theorem~\ref{thm:Slepian_SF_ineq} makes this intuition precise and extends it to processes with different variances.\footnote{Although intuitive in some sense, this turns out to be a delicate statement about Gaussian random variables, as it does not hold in general for other distributions.}

\begin{theorem}[Slepian/Sudakov-Fernique inequality]\label{thm:Slepian_SF_ineq}

Let $\left\{X_u\right\}_{u\in U}$ and $\left\{Y_u\right\}_{u\in U}$ be two (almost surely bounded) separable centered Gaussian processes indexed by the same (compact) set $U$ (
satisfying~\eqref{eq:separablegaussianprocess}
). If, for every $u,v\in U$:
\begin{equation}
\EE\left[X_{u}-X_{v}\right]^2 \leq  \EE\left[Y_{u}-Y_{v}\right]^2,
\end{equation}
then
\[
 \EE\left[\max_{u\in U}X_u\right] \leq \EE\left[\max_{u\in U}Y_u\right].
\]
\end{theorem}

\begin{proof}
Since $X$ and $Y$ are separable processes, we can reduce to the case in which $U$ is finite, by showing the inequality for all finite subsets of the indexing set (see~\eqref{eq:separablegaussianprocess}).

The idea is to use Gaussian Interpolation (Lemma~\ref{lemma:gaussianinterpolation}) on a soft-max. We take
\[
Z(t) = \sqrt{t}X + \sqrt{1-t^2}Y.
\]
For $\beta>0$ we define
\[
F_\beta(Z(t)) = \frac{1}{\beta} \log\left(\sum_{u\in U} e^{\beta Z_u(t)} \right). 
\]
Gaussian Interpolation (Lemma~\ref{lemma:gaussianinterpolation}) implies that
\begin{eqnarray*}
\frac{d}{dt}\EE F_\beta(Z(t)) &=&  \frac12 \sum_{u,v\in U}^n \left( \Sigma^X_{uv} - \Sigma^Y_{uv} \right) \EE \left[ \frac{\partial^2 F}{\partial x_u\partial x_v}(Z(t))\right].
\end{eqnarray*}
Setting $p_u(Z) = e^{\beta Z_u(t)} / \sum_{v\in U} e^{\beta Z_v(t)}$ (we dropped the $t$ from $p_u(Z(t))$ to ease notation), we have
\begin{eqnarray*}
\frac{\partial F}{\partial x_u}(Z(t)) & = & \frac{1}{\beta}  \frac{\beta e^{\beta Z_u(t)} }{\sum_{v\in U} e^{\beta Z_v(t)}} = p_u(Z), \\ \frac{\partial^2 F}{\partial x_u\partial x_v}(Z(t)) & = & \delta_{uv} \frac{\beta e^{\beta Z_u(t)} }{\sum_{w\in U} e^{\beta Z_w(t)}} - e^{\beta Z_u(t)}\frac{\beta e^{\beta Z_v(t)}  }{\left(\sum_{w\in U} e^{\beta Z_w(t)}\right)^2} \\ & = & \beta\left( \delta_{uv}p_u(Z) - p_u(Z)p_v(Z)  \right).
\end{eqnarray*}

Our goal is to show that $\frac{d}{dt}\EE F_\beta(Z(t))$ is non-positive, implying that $\EE F_\beta(X) \leq \EE F_\beta(Y)$ which by taking $\beta\to\infty$, and recalling that $$\max_{u\in U}Z_u(t) \leq F_\beta(Z(t)) \leq \frac{\log n}{\beta} + \max_{u\in U}Z_u(t),$$ finishes the proof.

Showing that $\frac{d}{dt}\EE F_\beta(Z(t))$ is non-positive is mechanical computation (using $\sum_{u\in U}p_u(Z)=1$ and $\EE\left[X_{u}-X_{v}\right]^2 = \Sigma^X_{uu}-2\Sigma^X_{uv}+\Sigma^X_{vv}$). Indeed,
\begin{eqnarray}
\frac{d}{dt}\EE F_\beta(Z(t)) &=& \frac{\beta}{2}\sum_{u\in U} \left( \Sigma^X_{uu} - \Sigma^Y_{uu} \right)\left( p_u(Z) - p_u(Z)p_u(Z)  \right)
 \\ & &  - \frac{\beta}{2}\sum_{u\neq v\in U} \left( \Sigma^X_{uv} - \Sigma^Y_{uv} \right) p_u(Z)p_v(Z),
\end{eqnarray}

and
\begin{eqnarray*}
0 & \geq & \frac12 \sum_{u\neq v\in U} \left( \EE\left[X_{u}-X_{v}\right]^2 - \EE\left[Y_{u}-Y_{v}\right]^2 \right) p_u(Z)p_v(Z) \\
 & = & \frac12 \sum_{u\neq v\in U} \left( \Sigma^X_{uu} - 2\Sigma^X_{uv} + \Sigma^X_{vv} - \Sigma^Y_{uu} + 2\Sigma^Y_{uv} -\Sigma^Y_{vv} \right) p_u(Z)p_v(Z)  \\
 & = &  \sum_{u\neq v\in U} \left( \Sigma^X_{uu} -  \Sigma^Y_{uu}\right) p_u(Z)p_v(Z) -   \sum_{u\neq v\in U} \left( \Sigma^X_{uv} -  \Sigma^Y_{uv}\right) p_u(Z)p_v(Z) \\ 
 & = &  \sum_{u\in U} \left( \Sigma^X_{uu} -  \Sigma^Y_{uu}\right) p_u(Z)\left(\sum_{v\in U\setminus \{u\}}p_v(Z)\right) -   \sum_{u\neq v\in U} \left( \Sigma^X_{uv} -  \Sigma^Y_{uv}\right) p_u(Z)p_v(Z) \\
 & = & \frac2{\beta}\frac{d}{dt}\EE F_\beta(Z(t)),
\end{eqnarray*}
where the last equality follows from $\sum_{v\in U\setminus \{u\}}p_v(Z) = 1-p_u(Z)$.
\end{proof}

We will also use in the following an extension due to Gordon~\cite{Gordon_comparison_85,Gordon_EscapeMesh}. While we omit the proof here, we note that it can be shown by a similar argument as above, by taking a soft min-max, see, e.g.,~\cite{Gordon_EscapeMesh} or Problem 6.2. in~\cite{vanHandel_LectureNotesProb_14}).

\begin{theorem}\label{Thm:SlepiansIneq_extension_Gordon}[Theorem A in~\cite{Gordon_EscapeMesh}]
Let $\left\{X_{t,u}\right\}_{(t,u)\in T\times U}$ and $\left\{Y_{t,u}\right\}_{(t,u)\in T\times U}$ be two (almost surely bounded) separable centered Gaussian processes\footnote{Here we need the min-max to be well approximated by finite subsets of $T$ and $U$ in the sense of~\eqref{eq:separablegaussianprocess}, as before the processes on which we use this result will be well approximated by $\epsilon$-nets.} indexed by the same (compact) sets $T$ and $U$.

If, for every $t_1,t_2\in T$ and $u_1,u_2 \in U$:
\begin{equation}
\EE\left[X_{t_1,u_1}-X_{t_1,u_2}\right]^2 \leq  \EE\left[Y_{t_1,u_1}-Y_{t_1,u_2}\right]^2,
\end{equation}
and, for $t_1\neq t_2$,
\begin{equation}
\EE\left[X_{t_1,u_1}-X_{t_2,u_2}\right]^2 \geq  \EE\left[Y_{t_1,u_1}-Y_{t_2,u_2}\right]^2,
\end{equation}
then
\[
 \EE\left[\min_{t\in T}\max_{u\in U}X_{t,u}\right] \leq \EE\left[\min_{t\in T}\max_{u\in U}Y_{t,u}\right].
\]
\end{theorem}

Note that Theorem~\ref{thm:Slepian_SF_ineq} easily follows by setting $|T|=1$.

\subsection{Spectral norm of a Wigner Matrix}

Before proceeding with Gordon's Escape Through a Mesh Theorem we take a brief detour to showcase the usefulness of what we already developed in this chapter.

Let $W\in\RR^{n\times n}$ be a standard Gaussian Wigner matrix, a symmetric matrix with (otherwise) independent Gaussian entries, the off-diagonal entries have unit variance and the diagonal entries have variance $2$ (recall~\eqref{eq:WignerSC}). $\lambda_{\max}(W)$ depends on $\frac{n(n+1)}2$ independent (standard) Gaussian random variables and it is easy to see that it is a $\sqrt{2}$-Lipschitz function of these variables, since
\[
\left|\lambda_{\max}(W^{(1)}) - \lambda_{\max}(W^{(2)})  \right| \leq \lambda_{\max}\left( W^{(1)} - W^{(2)} \right) \leq \left\| W^{(1)} - W^{(2)} \right\|_F.
\]
The symmetry of the matrix and the variance $2$ of the diagonal entries are responsible for an extra factor of $\sqrt{2}$.

Using Gaussian Concentration (Theorem~\ref{theorem:4:gaussianconcentration}) we immediately get
\[
\Prob\left\{ \lambda_{\max}(W) \geq \EE \lambda_{\max}(W) + t  \right\} \leq 2\exp\left( -\frac{t^2}{4} \right).
\]

On the other hand, one can prove $\EE \lambda_{\max}(W) \leq 2\sqrt{n}$ using Slepian's inequality (Theorem~\ref{thm:Slepian_SF_ineq}) by comparing the Gaussian process $X_u = u^Wu$ with the Gaussian process $Y_u = \sqrt{2}g^Tu$ for $g\in\NNN\left(0,I_{n\times n}\right)$, this is an excellent exercises that we leave to the reader. Combining these we get the following.

\begin{proposition}\label{proposition:4:WignerSpectralNormTail}
Let $W\in\RR^{n\times n}$ be a standard Gaussian Wigner matrix, a symmetric matrix with (otherwise) independent Gaussian entries, the off-diagonal entries have unit variance and the diagonal entries have variance $2$. Then,
\[
\Prob\left\{ \lambda_{\max}(W) \geq 2\sqrt{n} + t  \right\} \leq 2\exp\left( -\frac{t^2}{4} \right).
\]
\end{proposition}

Note that this gives a rather precise control of the fluctuations of $\lambda_{\max}(W)$.\footnote{It turns out that the fluctuations of $\lambda_{\max}(W)$ are even smaller, corresponding to the so-called Tracy-Widom distribution~\cite{Anderson_Guionnet_Zeitouni_IntroRandomMatrices}}
In fact, for $t = 2\sqrt{\log n}$ this gives
\[
\Prob\left\{ \lambda_{\max}(W) \geq 2\sqrt{n} + 2\sqrt{\log n}  \right\} \leq 2\exp\left( -\frac{4\log n}{4} \right) = \frac2n.
\]
This illustrated the level of concentration we can expect from the spectral norm, or largest eigenvalues, of random matrices. As we will see in Chapter~\ref{c:probability-matrixconcentration} the main challenge in many cases is in controlling the expected value of the spectral norm, or largest eigenvalue.

\section{Gordon's Theorem}\label{s:gordon}

In Section~\ref{s:jl} we showed that in order to approximately preserve the distances (up to $1\pm \eps$) between $n$ points, it suffices to randomly project them to $\Theta\left(\epsilon^{-2} \log n \right)$ dimensions. The key argument was that a random projection approximately preserves the norm of every point in a set $S$, in this case the set of differences between pairs of $n$ points. What we showed is that in order to approximately preserve the norm of every point in $S$, it is enough to project to $\Theta\left(\epsilon^{-2} \log |S| \right)$ dimensions. The question this section is meant to answer is: can this be improved if $S$ has a special structure? Given a set $S$, what is the measure of complexity of $S$ that explains how many dimensions one needs to project on to still approximately preserve the norms of points in $S$. We shall see below that this will be captured---via Gordon's Theorem---by the so called {\em Gaussian width} of $S$.

\begin{definition}[Gaussian width]\label{def:gaussianwidth}
Given a closed set $S\subset \RR^{p}$, its \emph{Gaussian width} $\omega(S)$ is defined as:
\[
\omega(S) = \EE \max_{x\in S} \left[ g_p^Tx  \right],
\]
where $g_p\sim \NNN\left(0,I_{p\times p}\right)$.  
\end{definition}

Similarly to the proof of Theorem~\ref{JL_random_0} we will restrict our attention to sets $S$ of unit norm vectors, meaning that $S\subseteq \SSS^{p-1}$ (which in particular implies it is compact).

Also, we will focus our attention not in random projections but in the similar model of random linear maps $G:\RR^{p}\to\RR^{d}$ that are given by matrices with i.i.d.\ Gaussian entries. For this reason the following proposition will be useful:

\begin{proposition}
Let $g_d\sim  \NNN\left(0,I_{d\times d}\right)$, and define
\[
a_d\defeq \EE\|g_d\|.
\]
Then $\sqrt{d-1}\leq\sqrt{\frac{d}{d+1}}\sqrt{d}\leq a_d\leq \sqrt{d}$.
\end{proposition}
We will prove here the weaker lower bound $\sqrt{d-1}\leq a_d\leq \sqrt{d}$ which is sufficient for our purposes and has a very elegant proof.\footnote{There is a very nice discussion of this lower bound in this blog post~\cite{Epperly_blog_gaussiannorm}.} 
\begin{proof}
The upper bound is a straightforward consequence of Jensen's inequality $$a_d^2=\left(\EE\|g_d\|\right)^2 \leq \EE\|g_d\|^2 =d.$$
The lower bound $\sqrt{d-1}\leq a_d$ follows from the Gaussian Poincar\'e Inequality (Proposition~\ref{prop:GaussianPoincare}):
let $f(g_d)=\|g_d\|$ (and recall that since $f$ is Lipschitz it has a gradient a.e. and we can apply Proposition~\ref{prop:GaussianPoincare}. Thus $\Var(f(g_d)) \leq \EE \| \nabla f(g_d) \|^2$ which gives: $$d - \left(\EE\|g\|\right)^2 =  \EE\|g\|^2 - \left(\EE\|g\|\right)^2 = \Var(f(g)) \leq \EE \| \nabla f(g) \|^2 = \EE \| g/\|g\| \|^2 = 1$$
\end{proof}

We are now ready to present Gordon's Theorem.

\begin{theorem}[Gordon's Escape Through a Mesh~\cite{Gordon_EscapeMesh}]\label{GordonsTheorem}
Let $G\in\RR^{d\times p}$ be a random matrix with independent $\NNN(0,1)$ entries and $S\subset \SSS^{p-1}$ be a closed subset of the unit sphere in $p$ dimensions. Then
\begin{equation}\label{eq:gordon1}   
\EE \max_{x\in S} \left\|\frac1{a_d}Gx\right\| \leq 1 + \frac{\omega(S)}{a_d},  
\end{equation}

and
\begin{equation}   \label{eq:gordon2}  
\EE \min_{x\in S} \left\|\frac1{a_d}Gx\right\| \geq 1 - \frac{\omega(S)}{a_d},
\end{equation}

where $a_d = \EE\|g_d\|$ and $\omega(S)$ is the Gaussian width of $S$. Recall that $\sqrt{\frac{d}{d+1}}\sqrt{d}\leq a_d\leq \sqrt{d}$.
\end{theorem}

Before proving Gordon's Theorem we will note some of its direct implications. The theorem suggests that $\frac1{a_d}G$ preserves the norm of the points in $S$ up to $1\pm \frac{\omega(S)}{a_d}$, indeed we can make this precise with Gaussian concentration (Theorem~\ref{theorem:4:gaussianconcentration}).

Note that the function $F(G) = \max_{x\in S} \left\|Gx\right\|$ is 1-Lipschitz. Indeed
\begin{eqnarray*}
\left| \max_{x_1\in S}\left\| G_1x_1\right\|  - \max_{x_2\in S}\left\| G_2x_2\right\| \right| & \leq & \max_{x\in S} \left| \left\| G_1x\right\|  - \left\| G_2x\right\| \right| \leq \max_{x\in S} \left\| \left( G_1 - G_2 \right)x\right\| \\
& \leq & \left\|  G_1 - G_2\right\| \leq \left\|  G_1 - G_2\right\|_F.
\end{eqnarray*}

Similarly, one can show that $F(G) = \min_{x\in S} \left\| Gx\right\|$ is 1-Lipschitz.
Thus, one can use Gaussian concentration to get
\begin{equation}\label{eq:5:usingGaussianconcentration_1}
\Prob\left\{ \max_{x\in S} \|Gx\| \geq a_d + \omega(S) + t  \right\} \leq \exp\left( -\frac{t^2}2 \right),
\end{equation}
and
\begin{equation}\label{eq:5:usingGaussianconcentration_2}
\Prob\left\{ \min_{x\in S} \|Gx\| \leq a_d - \omega(S) - t  \right\} \leq \exp\left( -\frac{t^2}2 \right).
\end{equation}

This gives us the following theorem.

\begin{theorem}\label{thm:Gordon:afterGconcentration}
 Let $G\in\RR^{d\times p}$ a random matrix with independent $\NNN(0,1)$ entries and $S\subset \SSS^{p-1}$ be a closed subset of the unit sphere in $p$ dimensions. Then, for $\eps>\frac{\omega(S)}{a_d}$, with probability $\geq 1-2\exp\left[ -\frac{a_d^2}{2}\left( \eps - \frac{\omega(S)}{a_d}  \right)^2\right]$:
\[
 (1-\eps)\|x\| \leq \left\| \frac1{a_d} Gx  \right\| \leq (1+\eps)\|x\|,
\]
for all $x\in S$.

Recall that $d\frac{d}{d+1} \leq a_d^2 \leq d$.
 \end{theorem}

\begin{proof}
This is readily obtained by taking $\eps = \frac{\omega(S)+t}{a_d}$, using~\eqref{eq:5:usingGaussianconcentration_1} and~\eqref{eq:5:usingGaussianconcentration_2}.
\end{proof}

\begin{remark}
 Note that a simple use of a union bound\footnote{This follows from the fact that the maximum of $n$ standard Gaussian random variables is $\lesssim \sqrt{2\log n}$.} shows that $\omega(S)\lesssim \sqrt{2\log|S|}$, which means that taking $d$ to be of the order of $\log|S|$ suffices to ensure that $\frac1{a_d} G$ to have the Johnson Lindenstrauss property. This observation shows that Theorem~\ref{thm:Gordon:afterGconcentration} essentially directly implies Theorem~\ref{JL_random_0} (although not exactly, since $\frac1{a_d} G$ is not a projection).
\end{remark}

\subsection{Gordon's Escape Through a Mesh Theorem}
Theorem~\ref{thm:Gordon:afterGconcentration} suggests that, if $\omega(S)\leq a_d$, a uniformly chosen random subspace of $\RR^n$ of dimension $(n-d)$ (which can be seen as the nullspace of $G$) avoids a set $S$ with high probability. This is indeed the case and is known as Gordon's Escape Through a Mesh Theorem (Corollary 3.4. in Gordon's original paper~\cite{Gordon_EscapeMesh}). See also~\cite{Dustin_blog_gordon} for a description of the proof. We include the Theorem below for the sake of completeness.

\begin{theorem}[Corollary 3.4. in~\cite{Gordon_EscapeMesh}]\label{thm:GordonEscapeMesh}
 Let $S\subset \SSS^{p-1}$ be a closed subset of the unit sphere in $p$ dimensions. If $\omega(S)<a_d$, then for a $(p-d)$-dimensional subspace $\Lambda$ drawn uniformly from the Grassmanian manifold we have
\[
\Prob\left\{ \Lambda \cap S \neq \emptyset   \right\}  \leq \frac72 \exp \left( -\frac1{18} \left( a_d - \omega(S) \right)^2 \right),
\]
where $\omega(S)$ is the Gaussian width of $S$ and $a_d = \EE\|g_d\|$ where $g_d\sim \NNN(0,I_{d\times d})$.
\end{theorem}

\subsection{Proof of Gordon's Theorem}

We are now ready to prove Gordon's Theorem.

\begin{proof}[of Theorem~\ref{GordonsTheorem}]

Let $G\in\RR^{d\times p}$ with i.i.d. $\NNN(0,1)$ entries. We define two Gaussian processes: For $v\in S\subset \SSS^{p-1}$ and $u\in \SSS^{d-1}$ let $g\sim\NNN\left(0,I_{d\times d}\right)$ and $h\sim\NNN\left(0,I_{p\times p}\right)$ and define the following processes:
\[
 A_{u,v} = g^Tu + h^Tv, 
\]
and
\[
 B_{u,v} = u^T Gv. 
\]

For all $v,v'\in S \subset \SSS^{p-1}$ and $u,u'\in \SSS^{d-1}$,
\begin{gather*}
 \EE \left| A_{v,u} - A_{v',u'} \right|^2 - \EE \left| B_{v,u} - B_{v',u'} \right|^2  =  4 - 2\left(u^Tu' + v^Tv' \right) - \sum_{ij}\left( v_iu_j - v_i'u_j' \right)^2 \\
  =  4 - 2\left(u^Tu' + v^Tv' \right) - \left[2-  2\left(v^Tv'\right) \left(u^Tu'\right)  \right] \\
  =  2 - 2\left( u^Tu' + v^Tv' - u^Tu'v^Tv' \right) \\
  =  2\left(1-u^Tu'\right)\left(1-v^Tv'\right).
\end{gather*}
This means that $\EE \left| A_{v,u} - A_{v',u'} \right|^2 - \EE \left| B_{v,u} - B_{v',u'} \right|^2\geq 0$ and $\EE \left| A_{v,u} - A_{v',u'} \right|^2 - \EE \left| B_{v,u} - B_{v',u'} \right|^2 = 0$ if $v=v'$.
This implies that we can use Theorem~\ref{Thm:SlepiansIneq_extension_Gordon} with $X=A$ and $Y=B$ (these processes are well approximated by $\epsilon$-nets and separable, see~\eqref{eq:separablegaussianprocess}), to get
\[
 \EE \min_{v\in S}\max_{u\in \SSS^{d-1}} A_{v,u} \leq \EE \min_{v\in S}\max_{u\in \SSS^{d-1}} B_{v,u}.
\]
Noting that
\[
\EE \min_{v\in S}\max_{u\in \SSS^{d-1}} B_{v,u} = \EE\min_{v\in S}\max_{u\in \SSS^{d-1}}u^TGv =  \EE\min_{v\in S}\left\|Gv\right\|,
\]
and
\begin{align*}
\EE \left[\min_{v\in S}\max_{u\in \SSS^{d-1}} A_{v,u} \right] & = \EE\max_{u\in \SSS^{d-1}}g^Tu +  \EE\min_{v\in S}h^Tv \\
& =   \EE\max_{u\in \SSS^{d-1}}g^Tu -\EE\max_{v\in S}(-h^Tv) = a_d - \omega(S),
\end{align*}
gives the second part of the theorem.

On the other hand, since $\EE \left| A_{v,u} - A_{v',u'} \right|^2 - \EE \left| B_{v,u} - B_{v',u'} \right|^2\geq 0$ then we can similarly use Theorem~\ref{thm:Slepian_SF_ineq} with $X = B$ and $Y = A$, to get
\[
\EE \max_{v\in S}\max_{u\in \SSS^{d-1}} A_{v,u} \geq \EE \max_{v\in S}\max_{u\in \SSS^{d-1}} B_{v,u}.
\]
Noting that
\[
\EE \max_{v\in S}\max_{u\in \SSS^{d-1}} B_{v,u} = \EE\max_{v\in S}\max_{u\in \SSS^{d-1}}u^TGv =  \EE\max_{v\in S}\left\|Gv\right\|,
\]
and
\[
\EE \left[\max_{v\in S}\max_{u\in \SSS^{d-1}} A_{v,u} \right] = \EE\max_{u\in \SSS^{d-1}}g^Tu +  \EE\max_{v\in S}h^Tv = a_d + \omega(S),
\]
concludes the proof of the theorem.

\end{proof}

A remarkable application of Gordon's Theorem is that one can use it for abstract sets $S$ such as the set of all \emph{natural images} or the set of all plausible \emph{user-product} ranking matrices. In these cases Gordon's Theorem suggests that a measurements corresponding just to a random projection may be enough to keep geometric properties of the data set in question, in particular, it may allow for reconstruction of the data point from just the projection. These phenomenon and the sensing savings that arises from it is at the heart of compressive sensing and several recommendation system algorithms, among many other data processing techniques. Motivated by these two applications we will focus in this section on understanding which projections are expected to preserve the norm of sparse vectors and low-rank matrices. 
Both compressed sensing and low-rank matrix modeling will be discussed in length in Chapters~\ref{c:cs} and~\ref{c:lowrank}, respectively.

\begin{remark}[Ornstein-Uhlenbeck process]\label{remark:Ornstein-Uhlenbeck}
There is an important side of Gaussian Analysis that we do not cover, related to the theory of Markov semigroups, and in particular, the Ornstein-Uhlenbeck process. In fact, this approach is arguably the most natural way of proving Poincar\'{e}'s inequality (this is intimately connected to the material in Section~\ref{sec:MarkovChains}), and also yields Log-Sobolev and hypercontractivity inequalities. We point the reader to Chapter 2 of~\cite{vanHandel_LectureNotesProb_14} for an excellent pedagogical introduction to these ideas. 
\end{remark}

\begin{remark}[Hermite Polynomials]\label{remark:HermitePolynomials}
Another central set of ideas in Gaussian analysis that we do not cover in this book is the theory of Hermite polynomials, which are a basis of orthogonal polynomials in the Gaussian measure, forming an analogue of Fourier analysis in the Gaussian setting. Writing a function $f\in L^2(\NNN(0,1))$, or in $L^2(\NNN(0,I))$, in its Hermite polynomial expansion is a classical idea, often known as a Wiener Chaos expansion. Due to Gaussian integration by parts (Lemma~\ref{lemma:GIbP}) taking the gradient of such a function yields simple transformations on the expansion coefficients. In fact, this yields a very simple proof of Poincar\'{e}'s inequality (try it!).\footnote{Hermite polynomials also play a crucial role on the development of low degree method for computational hardness of hypothesis testing (see, e.g., Appendix B of~\cite{KuniskyWeinBandeira2022LowDegreeSurvey}).} The Hermite polynomials are the eigenfunctions of the Ornstein-Uhlenbeck Operator (mentioned in Remark~\ref{remark:Ornstein-Uhlenbeck}). We point the reader to the classical reference~\cite{Szego1975OrthogonalPolynomials} for more on Hermite polynomials and an introduction to the beautiful theory of orthogonal polynomials.
\end{remark}

\section*{Exercises}
\addcontentsline{toc}{section}{Exercises}

\begin{myexercise}[\level\sep Moments of Gaussians]
    \label{prob:gaussian_moments}
    Let $Z$ be a standard gaussian random variable. Recall that Gaussian integration by parts states the following:
    given any differentiable function $f \colon \R \to \R$ whose derivative is absolutely integrable with respect to the standard normal measure, we have
    \begin{equation*}
        \E\bras{Zf(Z)} = \E\bras{f'(Z)}.
    \end{equation*}
    Let $p \geq 1$ be an integer and $Z$ be a standard gaussian random variable. Show that
    \begin{equation*}
        \E\bras{Z^p} = \begin{cases}
            (p-1)!! &\text{if } p \text{ is even};\\
            0 &\text{if } p \text{ is odd}.
        \end{cases}
    \end{equation*}
    Here, $!!$ denotes the double factorial, defined as $n!! = n \cdot (n-2) \cdots 3 \cdot 1$ for an odd natural number $n$.
    (There are $(p-1)!!$ possible pairings of $p$ elements (when $p$ even), and this is not a coincidence!)
\end{myexercise}

\begin{myexercise}[\level\level\sep Maximum of Gaussians]
    \label{prob:maximum_gaussians}
     Let $g_1, \ldots, g_d$ be a collection of (not necessarily independent) Gaussian random variables with zero mean and variance $\sigma^2$.
    \begin{enumerate}[(a)]
        \item Prove that the following bound holds
        \begin{equation*}
            \mathbb{E}\max_{i=1,\ldots,d}g_i \le \sigma \sqrt{2\log d}.
        \end{equation*}
        \emph{(If you do not manage to prove the inequality with sharp constant 2 in the square root, you can replace it by an absolute constant $C>0$.)}
        
        \item Prove that the bound in the previous item is sharp up to an absolute constant if we assume that all the Gaussian random variables are independent, i.e. there exists a universal constant $c > 0$ such that
        \begin{equation*}
            \mathbb{E}\max_{i=1,\ldots,d}g_i \geq c\, \sigma \sqrt{\log d}.
        \end{equation*}
        
        \item Show that the conclusion of (b) is false if we drop the assumption that $g_1, \ldots, g_d$ are independent.
    \end{enumerate}
\end{myexercise}

\begin{myexercise}[\level\level\level\sep Application of Slepian's lemma]
    \label{prob:slepian_lemma}
    Let $W$ be a $d \times d$ Gaussian Wigner matrix.
    \begin{enumerate}[(a)]
        \item Prove that
        \begin{equation*}
            \mathbb{E}\sup_{v\in S^{d-1}}\langle g,v\rangle = \mathbb{E}\|g\|_2 \le \sqrt{d},
        \end{equation*}
        where $g \in \mathbb{R}^d$ is a standard Gaussian random vector.
        
        \item Apply Slepian's lemma to prove that
        \begin{equation*}
            \mathbb{E}\lambda_{\max}(W) \le 2\sqrt{d}.
        \end{equation*}
        
        \item Show that the upper bound above is tight up to an absolute constant.
    \end{enumerate}
\end{myexercise}
\begin{hint}
   For (b), consider the stochastic process $Y_{v}:=2\langle g,v\rangle$. For (c), first show $\E\norm{g}_2 \leq c\sqrt{d}$ by integration.
\end{hint}

\begin{lemma}[Gamma Function Bound]
    Let $x \geq \frac12$, we have 
    $$ \Gamma(x) = \int_0^\infty t^{x-1}e^{-t} \, dt \leq 3x^x.$$
\end{lemma}

\begin{myexercise}[\level\sep Moments of Subgaussian Variables]
    \label{prob:subgaussian_moments}
    Let $Y$ be a $\sigma^2$-subgaussian random variable, so for all $t \geq 0$ it holds
    $$ \Prob(\abs{Y} \geq \sigma t) \leq 2e^{-\frac{t^2}{2}}.$$
    Prove that for all $p\geq 1$ we have
    $$ \E[\abs{Y}^p]^{\frac1p} \leq C\sigma\sqrt{p},$$
    where $C>0$ is a universal constant.
    \begin{hint}
        Use Problem~\ref{prob:integral_identity}(a) and bounds on the gamma function.
    \end{hint}
\end{myexercise}

\begin{theorem}[Gaussian Lipschitz concentration]
    Let $g_1, \ldots, g_n$ be i.i.d standard Gaussian random variables. Let $f:\mathbb{R}^n\rightarrow \mathbb{R}$ be a $L$-Lipschitz function with respect to the Euclidean norm. Then, for all $t>0$,
    \begin{equation}\label{conc_gauss}
    \mathbb{P}(|f(g_1,\ldots,g_n)-\mathbb{E}f(g_1,\ldots,g_n)|\ge L t)\le 2e^{-t^2/2}.
    \end{equation}
\end{theorem}

\begin{myexercise}[\level \level \sep Variance of a Gaussian Process]
    \label{prob:gaussprocessvariance}
    Let $g \in \R^d$ be a standard random gaussian vector and let $v_1, \ldots, v_n \in \R^d$ be fixed vectors. Define $f(g) = \sup_{1 \leq i\leq n} \langle v_i,g \rangle$. We aim to show that the variance of $f$ is (up to a constant factor) at most the maximal variance of the variables we take the supremum over.
    \begin{enumerate}[(a)]
    \item Let $x,y \in \R^d$ be arbitrary. Prove $\abs{f(x)-f(y)}\leq \sup_{1\leq i \leq n} \norm{v_i}_2 \norm{x-y}_2$.
    \item Use tail integration and gaussian concentration to prove that $\Var(f(g)) \leq 4 \sup_{1\leq i \leq n} \norm{v_i}_2^2$
\end{enumerate}
    
\end{myexercise}

\begin{myexercise}[\level \level \sep Sudakov's Inequality]
    \label{prob:sudakovinequality}
    Let $g \in \R^d$ be a standard random gaussian vector and let $ I=  \{v_1, \ldots, v_n\} \subseteq \R^d$ be a finite set of fixed vectors. Define $f(g) = \sup_{v \in I } \langle v,g \rangle$. The goal of this problem is to show the link between packing numbers and gaussian processes. We say a set $S$ is $\eps$-separated, if for all $v,w \in S$ with $v \neq w$ we have $\norm{v-w}_2 > \eps$.
    \begin{enumerate}[(a)]
    \item Let $ \eps >0$ and suppose $I$ has an $\eps$-separated subset $S$. Combine part (b) of Problem~\ref{prob:maximum_gaussians} with Slepian's Lemma to prove
    $$ \E[f(g)] \geq c \eps \sqrt{\log(\abs{S})}$$
    for some universal constant $c>0$.
    \item Let $\mathcal{D}(\eps)$ be the maximal cardinality of an $\eps$-separated subset of $I$. Conclude
    $$ \E[f(g)] \geq c \sup_{\eps >0} \eps \sqrt{\log(\mathcal{D}(\eps))}.$$
\end{enumerate}
    
\end{myexercise}

\begin{myexercise}[\level  \sep Absolute Values Only Make Small Differences]
    \label{prob:absvaluessmalldifference}
    Let $X$ be a real scalar random variable. In many scenarios we would like to estimate its expected absolute value with an optimal leading constant. For technical reasons this is however less tractable than just estimating its expectation in many cases, but when the variance of $X$ is small, the quantities bound each other reasonably well. This also turns out to be the case when we look at maxima of random variables under certain assumptions.
    \begin{enumerate}[(a)]
        \item Prove $ \abs{\E[X]} \leq \E [\abs{X}] \leq \abs{\E[X]} + \sqrt{\Var[X]}$.
        \item Let $ Y_1, \ldots, Y_n$ be a collection of scalar random variables (not necessarily i.i.d.), such that for all $i$ both $Y_i$ and $-Y_i$ are identically distributed and consider $Z = \max_{1 \leq i \leq n} Y_i$. Prove using a union bound and Chebyshev's inequality that
        $$ \P \left( \max_{1 \leq i \leq n} |Y_i| \geq \E[Z] + t \right) \leq \frac{2\Var[Z]}{t^2}.$$
        \item Use tail integration to conclude 
        $$ \E \left[ \max_{1 \leq i \leq n} |Y_i|  \right]  \leq \E[Z] + (1+ \sqrt{2})\sqrt{\Var[Z]}.$$
        
    \end{enumerate}

\end{myexercise}

\begin{theorem}[Lipschitz Concentration on the Sphere]
    Assume that $f:\sqrt{n}S^{n-1} \rightarrow \mathbb{R}$ is a $L$-Lipschitz function and let $X$ be a random vector uniformly distributed on the sphere $\sqrt{n}S^{n-1}$. Then, for all $t>0$,
    \begin{equation}\label{conc_sphere}
        \mathbb{P}(|f(X) - \mathbb{E}f(X)| \ge L t)\le 2e^{-ct^2}.
    \end{equation}
    Here $c>0$ is an absolute constant.
\end{theorem}

\begin{myexercise}[\level\level\sep Lipschitz Concentration around $L_p$ norms]
    \label{prob:lipschitz_concentration}

    In this problem we will extend the Lipschitz concentration on the sphere to concentration around $L_p$ norms for $p \ge 1$, i.e., we will prove that under the same assumptions as in~\eqref{conc_sphere} and additionally assuming that $f$ is non-negative,
    \begin{equation}\label{eq:lpsphereconc}
    \mathbb{P}(|f(X) - \|f(X)\|_{L_p}| \ge L t)\le 2e^{-c_pt^2},
    \end{equation}
where $c_p>0$ is a constant only depending on $p$ and $\| Z \|_{L_p} := (\E |Z|^p)^{1/p}$ for a random variable $Z$ such that its $p$-th absolute moment is well-defined. 

We will split the proof into several steps. 
\begin{enumerate}[(a)]
    \item Let $Y$ be an $L^2$-subgaussian random variable in the sense of Problem~\ref{prob:subgaussian_moments}. Prove that for any $A \geq 0$ there exists a constant $c_A$ only depending on $A$, such that
    $$ \Prob(\abs{Y-LA} \geq Lt ) \leq 2e^{-c_At^2}$$
    holds for some constant $c_A >0$ only depending on $A$.
    \item Show that for any non-negative random variable $Z$,
    $$
    | \E Z - (\E Z^p)^{1/p}| \le (\E \abs{Z - \E Z}^p)^{1/p}.
    $$
    You can assume that all the moments are well-defined. Use Problem~\ref{prob:subgaussian_moments} and~\eqref{conc_sphere} to conclude
    $$ | \E[f(X)] - \E [f(X)^p]^{1/p}|  \leq CL \sqrt{p}$$
    for some universal constant $C>0$.
    \item Now we have all the ingredients to prove the theorem. Using (a), (b), and~\eqref{conc_sphere}, complete the proof of inequality~\eqref{eq:lpsphereconc}.
\end{enumerate}

\end{myexercise}

\begin{myexercise}[\level\level\level \sep MGF bounds via the Poincaré inequality]
    \label{prob:poincaremgf}
    Suppose we have a random vector $X \in \R^d$ satisfies the following Poincaré inequality: For any bounded differentiable function $f$ on $\R^d$ we have 
    $$ \Var(f(X)) \leq C_P \E[\norm{\nabla f(X)}_2^2].$$
    The goal of this exercise is to show that the moment generating functions with small gradient can in this case be bounded well.
    \begin{enumerate}[(a)]
        \item Prove for any differentiable bounded function $g$ on $\R^d$ that $\nabla(e^{g(x)}) = e^{g(x)} \nabla g(x)$ and combine this with the Poincaré inequality to conclude that if $\norm{\nabla g(X)}_2^2 \leq 1$ holds almost surely, then 
        $$ \E[e^{\lambda g(x)}]- \E[e^{\lambda g(x)/2}]^2 \leq \frac{C_p\lambda^2}{4} \E[e^{\lambda g(x)}]$$
        holds for all $\lambda \geq 0 $
        \item Assume further that $\frac{C_p\lambda^2}{4} <(1/2)$ and that $\E[e^{\lambda g(x)/n}]^n$ converges to some number $\alpha$ as $n$ goes to infinity. Use (a) recursively to show
        $$ \E[e^{\lambda g(x)}] \leq e^{K}\alpha$$
        for some universal constant $K>0$. You can use without proof that $\frac{1}{1-x}\leq e^{2x}$ holds for $0\leq x \leq 1/2$.
        \item Prove that for bounded functions $g$ and any $\lambda$ we have 
        $$ \lim_{n \to \infty } \E[e^{\lambda g(x)/n}]^n = \alpha = e^{\lambda \E[g(x)]}. $$
    \end{enumerate}
\end{myexercise}

\begin{myexercise}[\level\level\level\sep Duality and Covering Numbers]
    \label{prob:duality_covering_numbers}
    We will present the principle of duality in a completely different context. Suppose you have a set $T \subseteq \R^n$ and some number $\eps>0$. We call a subset $S \subseteq T$ an $\eps$-covering of $T$, if for every $t \in T$ there exists an $s \in S$, such that $\norm{s-t}_2 \leq \eps$. We call a subset $S \subseteq T$ an $\eps$-separated set, if for every $s \neq s' \in S$ we have $\norm{s-t}_2 > \eps$. We define the following optimal values:
    $$ \mathcal{N}(T, \eps) \coloneqq \hspace{-3mm} \min_{\substack{S \subseteq T \\ S \, \eps-\mbox{covering of} \, T}} \hspace{-8mm} \abs{S}  \qquad \qquad \mathcal{D}(T, \eps) \coloneqq \hspace{-3mm} \max_{\substack{S \subseteq T \\ S \, \eps-\mbox{separated}}} \hspace{-5mm} \abs{S} $$
    \begin{enumerate}[(a)]
        \item Show that these two optimization problems are duals in the following sense:
        $$ \mathcal{N}(T, \eps) \leq \mathcal{D}(T, \eps)  \leq \mathcal{N}(T, \eps/2)  $$
        \item Use part (a) to show that for the euclidean ball in $d$-dimensions one has the following covering number estimates for every $0< \eps < 1$:
        $$ \left( \frac{1}{\eps} \right)^d  \leq  \mathcal{N}(\mathbb{B}_2^d, \eps) \leq \left( \frac{3}{\eps} \right)^d$$
    \end{enumerate}
\end{myexercise}

\begin{myexercise}[\level\sep Norm of a Gaussian Vector]
    \label{prob:norm_gaussian_vector}
    
    Given a standard Gaussian vector $g \in \mathbb{R}^d$, it is an easy consequence of Jensen's inequality that $\mathbb{E}\|g\|_2 \le \sqrt{d}$. The goal of this exercise is to give a simple proof that this is sharp using the Gaussian Poincaré inequality. 
\begin{enumerate}[(a)]
    \item Prove that the variance of $\|g\|_2$ is at most an absolute constant.
    \item Show that $\mathbb{E}\|g\|_2/\sqrt{d}$ converges to one as $d$ goes to infinity.
\end{enumerate}

\end{myexercise}

\begin{myexercise}[\level\sep Gaussian Width of the Simplex]
    \label{prob:gaussian_width_simplex}
    The gaussian width of a set $S \subseteq \R^d$ is defined as
    $$ \omega(S) \coloneqq \E \left[ \sup_{s \in S} \, \langle s, g  \rangle \right],$$
    where $g \in \R^d$ is a standard gaussian vector, so all coordinates of $g$ are independent and $\mathcal{N}(0,1)$-distributed. Consider the $d-1$-dimensional simplex
    $$ S_d \coloneqq \left \{ x \in \R^d \, \vert \, 0\leq x_i \leq 1, \, \sum_{i=1}^d x_i = 1 \right \}$$
    Our goal is to show that there exist universal constants $c,C >0$, such that
    $$ c \sqrt{\log(d)} \leq \omega(S_d) \leq C \sqrt{\log(d)}.$$
    \begin{enumerate}
        \item[(a)] For any subset $T \subseteq \R^d$ we define its convex hull $\operatorname{conv}(T)$ as follows: 
        $$ \operatorname{conv}(T) = \left \{ \sum_{i=1}^k x_i t_i \, \vert \, k\in \mathbb{N}_{>0}, \, x_i \in S_k, \, t_i \in T \right \}$$
        Prove the equality
        $$ \omega(\operatorname{conv}(T)) = \omega(T).$$
        \item[(b)] Find a finite set $T \subseteq \R^d$, such that $S_d = \operatorname{conv}(T)$. Use the result of another problem in this section to deduce the desired inequalities.
    \end{enumerate}
\end{myexercise}

\begin{myexercise}
Consider the following quantity:\\
\textbf{Definition: (Gaussian Complexity)} {\em The Gaussian complexity of a subset $T \in \R^n$ is defined as
$$
\gamma(T) := \E \left[ \max_{x\in T} \big|\langle g, x\rangle \big|\right],
$$
where $g \in {\mathcal N}(0, I_n)$.}

What can you say about the relationship between Gaussian complexity and Gaussian width of a set $T$? Under what circumstances on $T$ does equality hold? Using this, what can you say
about the set $T-T$? Give an example where these quantities are certainly different.  
\end{myexercise}


\chapter{Matrix Concentration Inequalities}
\label{c:probability-matrixconcentration}

There are many applications where one needs to control the spectrum of random matrices. Depending on the context, these matrices may represent the noise whose effect in a spectral algorithm is controlled by its spectral norm, or the size of a dual variable that needs to be controlled to show the exactness of a convex relaxation (such as in Chapter~\ref{c:community}). While some of the tools we developed in Chapter~\ref{c:surprises} could be used to control the size of the entries of random matrices, which could translate to spectral bounds, this would likely introduce many suboptimal dimensional factors. 

In what follows we will state and prove various matrix concentration results, somewhat similar to Theorem~\ref{thm:4:MatrixBernstein} (in Section~\ref{sc:matrixBernstein}). We will focus on understanding, and bounding, the typical value of the spectral norm of random matrices by upper bounding $\EE\|X\|$, as these tend to be high dimensional objects themselves they often have enough concentration that tail bounds are then easy to obtain (as it was illustrated in Chapter~\ref{c:probability-gaussiananalysis} with~Proposition~\ref{proposition:4:WignerSpectralNormTail}). Our presentation will rely on Gaussian analysis (developed in Chapter~\ref{c:probability-gaussiananalysis}). Our treatment of the Non-commutative Khintchine inequality, and the Matrix Concentration inequality we will prove follows parts of~\cite{Ramon-StFlour-2022} and~\cite{Tropp_MatrixConcentrationElementary}.\footnote{For an approach to matrix concentration that includes a direct proof of Theorem~\ref{thm:4:MatrixBernstein}  we recommend Tropp's excellent monograph~\cite{Tropp:TailBoundsRM_Monograph}.}

\section{Non-commutative Khintchine inequality}

We start with a particularly important inequality involving the expected value of a random matrix. It is intimately related to the non-commutative Khintchine inequality~\cite{Pis03}, and for that reason we will often refer to it as Non-commutative Khintchine (see, for example,~(4.9) in~\cite{Tropp:TailBoundsRM}). We will prove this inequality in Section~\ref{subsec:proofNCK}.

\begin{theorem}[Non-commutative Khintchine (NCK)]\label{thm:4:Gaussianseries}
Let $A_1,\dots,A_n\in \RR^{d\times d}$ be symmetric matrices and $g_1,\dots,g_n\sim\NNN(0,1)$ i.i.d., then: 
\begin{equation}\label{logterm}
\EE \Big\| \sum_{k=1}^{n} g_k A_k \Big\| \leq \Big(  2\lceil \log d \rceil+1  \Big)^{\frac12} \sigma,
\end{equation}
where
\begin{equation}\label{eq:4:sigma:forGaussianseries}
\sigma^2 = \Big\| \sum_{k=1}^n A_k^2  \Big\|. 
\end{equation}
\end{theorem}

Note that, akin to Proposition~\ref{proposition:4:WignerSpectralNormTail}, we can also use Gaussian Concentration to get a tail bound on $\left\| \sum_{k=1}^{n} g_k A_k \right\|$. We consider the function
\[
F:\RR^n \to \Big\| \sum_{k=1}^n g_kA_k \Big\|.
\]
We now estimate its Lipschitz constant; let $g,h\in\RR^n$ then
\begin{eqnarray*}
 \left| \Big\| \sum_{k=1}^n g_kA_k \Big\| -  \Big\| \sum_{k=1}^n h_kA_k \Big\| \right| & \leq & \Big\| \Big(\sum_{k=1}^n g_kA_k\Big) -  \Big( \sum_{k=1}^n h_kA_k \Big) \Big\| \\
 & = & \Big\| \sum_{k=1}^n (g_k-h_k)A_k \Big\| \\
 & = & \max_{v:\,\|v\|=1 } \Big|v^T\Big( \sum_{k=1}^n (g_k-h_k)A_k \Big) v \Big|\\
 & = & \max_{v:\,\|v\|=1 } \Big| \sum_{k=1}^n (g_k-h_k)\Big(v^TA_kv\Big) \Big| \\
 & \leq & \max_{v:\,\|v\|=1 } \sqrt{\sum_{k=1}^n (g_k-h_k)^2} \sqrt{\sum_{k=1}^n \Big(v^TA_kv\Big)^2}\\
 & = & \sqrt{ \max_{v:\,\|v\|=1 } \sum_{k=1}^n \Big(v^TA_kv\Big)^2}\,  \|g-h\|_2,
\end{eqnarray*}
where in the first inequality we made use of the triangular inequality and in the last one we used Cauchy-Schwarz.

This motivates us to define a new parameter, the weak variance $\sigma_\ast$.

\begin{definition}[Weak Variance (see, for example,~\cite{Tropp:TailBoundsRM_Monograph})]
 Given symmetric matrices $A_1,\dots,A_n\in\RR^{d\times d}$ . We define the weak variance parameter as
 \[
  \sigma_\ast^2 = \max_{v:\,\|v\|=1 } \sum_{k=1}^n \left(v^TA_kv\right)^2.
 \]
\end{definition}

This means that, using Gaussian concentration (Theorem~\ref{theorem:4:gaussianconcentration}), and setting $t = u\sigma_\ast$, we have
\begin{equation}
\Prob\left\{ \Big\| \sum_{k=1}^{n} g_k A_k \Big\| \geq \Big(  2\lceil \log d \rceil+1 \Big)^{\frac12} \sigma + u\sigma_\ast \right\} \leq \exp\Big( - \frac12u^2 \Big).
\end{equation}

Thus, although the expected value of $\left\| \sum_{k=1}^{n} g_k A_k \right\|$ is controlled by the parameter $\sigma$, its fluctuations seem to be controlled by $\sigma_\ast$. We compare the two quantities in the following proposition.

\begin{proposition}
 Given $A_1,\dots,A_n\in\RR^{d\times d}$ symmetric matrices, recall that
 \[
  \sigma = \sqrt{\Big\| \sum_{k=1}^n A_k^2  \Big\|} \text{ and } \sigma_\ast = \sqrt{ \max_{v:\,\|v\|=1 } \sum_{k=1}^n \left(v^TA_kv\right)^2  }.
  \]
  We have
 \[
  \sigma_\ast \leq \sigma.
 \]
\end{proposition}

\begin{proof}
Using the Cauchy-Schwarz inequality,
\begin{eqnarray*}
 \sigma_\ast^2 & = & \max_{v:\,\|v\|=1 } \sum_{k=1}^n \left(v^TA_kv\right)^2  =  \max_{v:\,\|v\|=1 } \sum_{k=1}^n  \left(v^T \left[A_kv\right]\right)^2 \\
 & \leq & \max_{v:\,\|v\|=1 } \sum_{k=1}^n  \left( \|v\| \|A_kv\|\right)^2  =  \max_{v:\,\|v\|=1 } \sum_{k=1}^n  \|A_kv\|^2 \\
 & = & \max_{v:\,\|v\|=1 } \sum_{k=1}^n  v^TA_k^2v  =  \Big\| \sum_{k=1}^n  A_k^2 \Big\|  =  \sigma^2.
\end{eqnarray*}
\end{proof}

\subsection{How tight is the Non-commutative Khintchine inequality?}\label{sec:tightnessNCK}

The following simple calculation is suggestive that the parameter $\sigma$ in Theorem~\ref{thm:4:Gaussianseries} is indeed the correct parameter to understand $\EE \left\| \sum_{k=1}^{n} g_k A_k \right\|$.
\begin{eqnarray}
 \EE \Big\| \sum_{k=1}^{n} g_k A_k \Big\|^2 & = & \EE \Big\| \Big( \sum_{k=1}^{n} g_k A_k \Big)^2 \Big\| = \EE \max_{v:\ \|v\|=1} v^T\Big( \sum_{k=1}^{n} g_k A_k \Big)^2v \nonumber\\
 & \geq & \max_{v:\ \|v\|=1} \EE  v^T\Big( \sum_{k=1}^{n} g_k A_k \Big)^2v = \max_{v:\ \|v\|=1}  v^T\Big( \sum_{k=1}^{n} A_k^2 \Big)v = \sigma^2. \nonumber
\end{eqnarray}

But a natural question is whether the logarithmic term in~\eqref{logterm} is needed. Motivated by this question we will explore a couple of examples.

\begin{example}\label{example:WignerMatrix}
 We can write a $d\times d$ Wigner matrix $W$ (recall Section~\ref{subsection:WignerMatrices}) as a Gaussian series, by taking $A_{ij}$ for $i\leq j$ defined as
 \[
  A_{ij} = e_ie_j^T + e_je_i^T,
 \]
if $i\neq j$, and
\[
 A_{ii} = \sqrt{2} e_ie_i^T.
\]
It is not difficult to see that, in this case, $\sum_{i\leq j}A_{ij}^2 = (d+1)I_{d\times d}$, meaning that $\sigma = \sqrt{d+1}$. This implies that Theorem~\ref{thm:4:Gaussianseries} gives us
\[
 \EE\|W\| \lesssim \sqrt{d\log d},
\]
however, we know that $ \EE\|W\| \asymp \sqrt{d}$, meaning that the bound given by NCK (Theorem~\ref{thm:4:Gaussianseries}) is, in this case, suboptimal by a logarithmic factor.\footnote{By $a\asymp b$ we mean $a\lesssim b$ and $a \gtrsim b$.}
\end{example}

The next example will show that the logarithmic factor is in fact needed in some examples

\begin{example}
 Consider $A_k = e_ke_k^T\in\RR^{d\times d}$ for $k=1,\dots,d$. The matrix $ \sum_{k=1}^{n} g_k A_k$ corresponds to a diagonal matrix with independent standard Gaussian random variables as diagonal entries, and so its spectral norm is given by $\max_k |g_k|$. It is known that $\max_{1\leq k \leq d} |g_k| \asymp \sqrt{\log d}$. On the other hand, a direct calculation shows that $\sigma = 1$. This shows that the logarithmic factor cannot, in general, be removed.
\end{example}

This motivates the question of trying to understand when is it that the extra dimensional factor is needed. For both these examples, the resulting matrix $X = \sum_{k=1}^{n} g_k A_k$ has independent entries (except for the fact that it is symmetric). In the case of independent entries it is possible to write a stronger inequality using $\sigma_\ast$~\cite{Bandeira_NARandomMatrixBound}:\footnote{There are notable improvements of Theorem~\ref{thm:4:matrixconcentrationindependententries}, including a dimension free bounds~\cite{latala2018dimension}, bounds that often capture the right constant on the leading order term~\cite{Bandeiraetal_FreeProbability}, and bounds that capture Tracy-Widom level tails~\cite{vanHandel-Brailovskaya-indentries2024}.}

\begin{theorem}[\cite{Bandeira_NARandomMatrixBound}]\label{thm:4:matrixconcentrationindependententries}
 If $X$ is a $d\times d$ random symmetric matrix with Gaussian independent entries (except for the symmetry constraint) whose entry $i,j$ has variance $b_{ij}^2$ then
 \[
  \EE \| X\| \lesssim \sqrt{ \max_{1\leq i\leq d} \sum_{j=1}^d b_{ij}^2 } + \max_{ij}\left| b_{ij} \right| \sqrt{\log d}.
 \]
\end{theorem}

\begin{remark}
 $X$ in the theorem above can be written in terms of a Gaussian series by taking
 \[
  A_{ij} = b_{ij} \left( e_ie_j^T + e_je_i^T  \right),
 \]
for $i< j$ and $A_{ii} = b_{ii}e_ie_i^T$. One can then compute $\sigma$ and $\sigma_\ast$:
\[
 \sigma^2 = \max_{1\leq i\leq d} \sum_{j=1}^d b_{ij}^2 \text{ and } \sigma_\ast^2 \asymp \max_{i,j}b_{ij}^2.
\]
This means that, when the random matrix in NCK (Theorem~\ref{thm:4:Gaussianseries}) has independent entries (modulo symmetry) then
\begin{equation}\label{eq:4:sigmasigmastar}
 \EE\|X\| \lesssim \sigma + \sqrt{\log d}\,\sigma_\ast.
\end{equation}
\end{remark}

Recently an improvement, using ideas from Free Probability to remove the dimensional factor in some situations, was obtained in~
\cite{Bandeiraetal_FreeProbability} (see~\cite{Tatianaetal_Universality} for non-Gaussian extensions). Interestingly, the same work shows that \eqref{eq:4:sigmasigmastar} does not hold in general, disproving a conjecture that was included in an earlier version of this manuscript. See Section~\ref{subsec:freeprobimprovement} for a more thorough discussion on these improvements.

\subsection{Proof of the Non-commutative Khintchine inequality}\label{subsec:proofNCK}

We are now ready to prove Theorem~\ref{thm:4:Gaussianseries}, the Non-commutative Khintchine inquality (NCK). See Section 7.2. in \cite[Section 7.2]{Tropp:TailBoundsSecondOrder} and \cite{Ramon-StFlour-2022} for presentations of the same argument.

\begin{proof}[of Theorem~\ref{thm:4:Gaussianseries}]

Let $p$ be a positive integer. Let $X =  \sum_{k=1}^{n} g_k A_k$, we have
\[
\left(\EE\left\| X\right\|\right)^{2p} \leq \EE\left\| X \right\|^{2p} =  \EE\left\| X^{2p} \right\| \leq \EE \tr X^{2p} ,
\]
where the first inequality follows from Jensen and the second from the fact that the trace of a positive semidefinite matrix dominates its spectral norm.

Using Gaussian integration by parts (Lemma~\ref{lemma:GIbP}) we get
\begin{eqnarray*}
\EE \tr X^{2p} &=& \EE \tr\sum_{k=1}^ng_kA_k X^{2p-1} =  \EE \tr\sum_{k=1}^n\sum_{q=0}^{2p-2}A_kX^qA_kX^{2p-2-q} \\ 
&=&  \EE\sum_{k=1}^n\sum_{q=0}^{2p-2} \tr\left(A_kX^qA_kX^{2p-2-q}\right).
\end{eqnarray*}

If the matrices $A_1,\dots,A_n$ were commutative, then we would have that $\tr\left(A_kX^qA_kX^{2p-2}\right) = \tr\left(A_k^2X^{2p-2}\right)$ for all $q$. It turns out that this is the worst possible situation and the commutative upper bound always dominates. We state and prove in Lemma~\ref{lemma:commutativeisworstGIbP} below that, for all $0\leq q\leq 2p-2$, $\tr\left(A_kX^qA_kX^{2p-2}\right) \leq \tr\left(A_k^2X^{2p-2}\right)$. This implies that
\begin{eqnarray*}
\EE \tr X^{2p} &\leq & \EE\sum_{k=1}^n(2p-1) \tr\left(A_k^2X^{2p-2}\right) = (2p-1) \EE\tr\left(\left(\sum_{k=1}^nA_k^2\right)X^{2p-2}\right).
\end{eqnarray*}
H\"older's inequality on Schatten norms then gives
\begin{eqnarray*}
\EE \tr X^{2p} &\leq & (2p-1) \left\| \sum_{k=1}^nA_k^2 \right\| \EE\sum_{k=1}^n \tr\left(X^{2p-2}\right) = (2p-1)\sigma^2 \EE\sum_{k=1}^n \tr\left(X^{2p-2}\right).
\end{eqnarray*}
Iterating this procedure gives
\begin{equation*}
\EE \tr X^{2p} \leq (2p-1)!! \sigma^{2p} d,
\end{equation*}
which implies
\begin{equation}
\EE\|X\| \leq \left( \tr X^{2p} \right)^{\frac1{2p}} \leq  \left( (2p-1)!!\, \sigma^{2p} d \right)^{\frac1{2p}} \leq [(2p-1)!!]^{\frac1{2p}} \sigma d^{\frac1{2p}}.
\end{equation}
Taking $p = \lceil \log d \rceil$ and using the fact that $(2p-1)!! \leq \left( \frac{2p+1}{e} \right)^p$ (see~\cite{Tropp_MatrixConcentrationElementary} for an elementary proof consisting essentially of taking logarithms and comparing the sum with an integral, note also that Lemma~\ref{lemma:gaussianmoments} would yield the same bound up to a constant of $e$) we get
\[
\EE \|X\| \leq \left( \frac{2\lceil \log d \rceil+1}{e} \right)^{\frac12} \sigma d^{\frac1{2 \lceil \log d \rceil }} \leq \left( 2\lceil \log d \rceil+1 \right)^{\frac12} \sigma.
\]
\end{proof}

\begin{lemma}[Commutative is the worst-case I]\label{lemma:commutativeisworstGIbP}
Let $A$ and $X$ be symmetric matrices. For any $p,q$ non-negative integers satisfying $0\leq q \leq 2p-2$ we have
\[
\tr\left(AX^qAX^{2p-2-q}\right) \leq \tr\left(A^2X^{2p-2}\right).
\] 
\end{lemma}

\begin{proof}
Consider the spectral decomposition $X = \sum_{i=1}^d \lambda_i v_iv_i^T$, then
\begin{eqnarray*}
\tr\left(AX^qAX^{2p-2-q}\right) & = & \sum_{i,j=1}^d\lambda_i^q\lambda_j^{2p-2-q}\tr\left(Av_iv_i^TAv_jv_j^T\right) \\
& = & \sum_{i,j=1}^d\lambda_i^q\lambda_j^{2p-2-q}\left(v_i^TAv_j\right)^2 \\
& \leq & \left( \sum_{i,j=1}^d|\lambda_i|^{2p-2}\left(v_i^TAv_j\right)^2 \right)^{\frac{q}{2p-2}} \left( \sum_{i,j=1}^d|\lambda_j|^{2p-2}\left(v_i^TAv_j\right)^2 \right)^{\frac{2p-2-q}{2p-2}}\\
& = & \sum_{i,j=1}^d|\lambda_i|^{2p-2}\left(v_i^TAv_j\right)^2 =  \sum_{i,j=1}^d\lambda_i^{2p-2}\left(v_j^TAv_i\right)\left(v_i^TAv_j\right) \\
& = & \sum_{i,j=1}^d\tr\left(\lambda_i^{2p-2}v_j^TAv_iv_i^TAv_j\right) = \tr\left(A^2X^{2p-2}\right),
\end{eqnarray*}
where the inequality follows from H\"older's inequality (Proposition~\ref{prop:holderineq}, in particular~\eqref{eq:HolderRS}).
\end{proof}

\begin{remark}
Theorem~\ref{thm:4:Gaussianseries} is an upper bound on a Gaussian process, and thus an orderwise optimal bound can (in theory) be obtained by Talagrand's generic chaining, and up to a logarithmic factor by covering number arguments via a Dudley's (see~\cite{Talagrand14}). Interestingly, there is no known proof of a bound that is optimal up to logarithmic factors based on these techniques (in other words, without using operator theoretical arguments). Such an argument would likely extend well beyond the setting of spectral norm in matrices, see~\cite{Lucca-et-al-TensorNCK} for a discussion on this and related problems.
\end{remark}

\section{Matrix concentration inequalities}

In what follows, we closely follow~\cite{Tropp_MatrixConcentrationElementary} and present an elementary proof of a few useful matrix concentration inequalities using Theorem~\ref{thm:4:Gaussianseries}. We start with a Rademacher version of Theorem~\ref{thm:4:Gaussianseries}.

\begin{theorem}\label{thm:4:RademacherSeries}
 Let $H_1,\dots,H_n\in \RR^{d\times d}$ be symmetric matrices and $\eps_1,\dots,\eps_n$ i.i.d. Rademacher random variables (meaning $=+1$ with probability $1/2$ and $=-1$ with probability $1/2$), then: 
\[
\EE \left\| \sum_{k=1}^{n} \eps_k H_k \right\| \leq \Big(  \frac\pi2 + \pi\lceil\log(d)\rceil \Big)^{\frac12} \sigma,
\]
where
\begin{equation}\label{eq:4:sigma:forrademacherseries}
\sigma^2 = \left\| \sum_{k=1}^n H_k^2  \right\|.
\end{equation}
\end{theorem}

\begin{proof}
Let $g_1,\dots,g_n$ by iid standard Gaussians independent from the Rademacher random variables, since the sign and absolute value of $g_k$ are independent and $\EE|g_k| = \sqrt{\frac2\pi}$, Jensen implies
\begin{eqnarray*}
\EE \left\| \sum_{k=1}^{n} \eps_k H_k \right\| & = &\frac{1}{\EE|g_1|} \EE \left\| \sum_{k=1}^{n} \left(\EE|g_k|\right)\eps_k H_k \right\| \\ & \leq & \sqrt{\frac{\pi}2} \EE \left\| \sum_{k=1}^{n} |g_k|\eps_k H_k \right\| = \sqrt{\frac{\pi}2} \EE \left\| \sum_{k=1}^{n} g_k H_k \right\|. 
\end{eqnarray*}
The conclusion then follows by applying Theorem~\ref{thm:4:Gaussianseries}.
\end{proof}

We note that Theorem~\ref{thm:4:RademacherSeries} holds without the extra $\frac{\pi}{2}$ factor (see~\cite{Tropp_MatrixConcentrationElementary} for a proof that boils down to the same as argument as the one above to prove its Gaussian counterpart, Theorem~\ref{thm:4:Gaussianseries}).

Using Theorem~\ref{thm:4:RademacherSeries}, we will prove the following theorem.

\begin{theorem}\label{thm:4:matrixconcentration:PSDcase}
Let $T_1,\dots,T_n\in\RR^{d\times d}$ be random independent symmetric positive semidefinite matrices, then
\[
\EE \left\| \sum_{i=1}^n T_i  \right\| \leq \left[  \left\| \sum_{i=1}^n \EE T_i \right\|^{\frac12}  + \sqrt{C(d)} \left(   \EE \max_i \| T_i\|   \right)^{\frac12}    \right]^2,
\]
where
\begin{equation}\label{eq:4:Cd_constant}
C(d) \defeq 2\pi + 4\pi\lceil \log d \rceil.
\end{equation}
\end{theorem}

A key step in the proof of Theorem~\ref{thm:4:matrixconcentration:PSDcase} is an idea that is extremely useful in Probability, the trick of symmetrization. For this reason we isolate it in a lemma. 
\begin{lemma}[Symmetrization]\label{lemma:4:symmetrization}
Let $T_1,\dots,T_n$ be independent random matrices (note that they do not necessarily need to be positive semidefinite, for the sake of this lemma) and $\eps_1,\dots,\eps_n$ random i.i.d. Rademacher random variables (independent also from the matrices). Then
\[
\EE \Big\| \sum_{i=1}^n T_i  \Big\| \leq \Big\| \sum_{i=1}^n \EE T_i \Big\| + 2 \EE  \Big\| \sum_{i=1}^n \eps_i T_i  \Big\|
\]
\end{lemma}
\begin{proof}
The triangular inequality gives 
\[
\EE \Big\| \sum_{i=1}^n T_i  \Big\| \leq \Big\| \sum_{i=1}^n \EE T_i \Big\| + \EE  \Big\| \sum_{i=1}^n  \Big(T_i - \EE T_i\Big)  \Big\|.
\]
Let us now introduce, for each $i$, a random matrix $T_i'$ identically distributed to $T_i$ and independent (all $2n$ matrices are independent). Then
\begin{eqnarray*}
\EE  \Big\| \sum_{i=1}^n  \Big(T_i - \EE T_i\Big)  \Big\| & = & \EE_T  \Big\| \sum_{i=1}^n  \Big(T_i -  \EE T_i - \EE_{T_i'}\Big[ T_i' - \EE_{T_i'}T_i' \Big]\Big)  \Big\|  \\ & = & \EE_T  \Big\| \EE_{T'} \sum_{i=1}^n  \Big(T_i -  T_i' \Big)  \Big\| \leq \EE  \Big\| \sum_{i=1}^n  \Big(T_i -  T_i' \Big)  \Big\|,
\end{eqnarray*}
where we use the notation $\EE_a$ to mean that the expectation is taken with respect to the variable $a$ and the last step follows from Jensen's inequality with respect to $\EE_{T'}$.

Since $T_i-T_i'$ is a symmetric random variable,\footnote{Note that we use the notation ``symmetric random variable'' to mean $X\sim -X$ and ``symmetric matrix'' to mean $X^T=X$} it is identically distributed to $\eps_i \left( T_i - T_i' \right)$, which gives
\begin{align*}    
&\EE  \left\| \sum_{i=1}^n  \left(T_i -  T_i' \right)  \right\|  = \EE  \left\| \sum_{i=1}^n  \eps_i\left(T_i -  T_i' \right)  \right\|  \\
 \leq \,\, & \EE  \left\| \sum_{i=1}^n  \eps_i T_i  \right\| + \EE  \left\| \sum_{i=1}^n  \eps_i T_i'  \right\| = 2 \EE  \left\| \sum_{i=1}^n  \eps_i T_i  \right\|,
\end{align*}
concluding the proof.
\end{proof}

\begin{proof}[Proof of Theorem~\ref{thm:4:matrixconcentration:PSDcase}]

Using Lemma~\ref{lemma:4:symmetrization} and Theorem~\ref{thm:4:RademacherSeries} we get
\[
\EE \left\| \sum_{i=1}^n T_i  \right\| \leq \left\| \sum_{i=1}^n \EE T_i \right\| + \sqrt{C(d)} \EE \left\| \sum_{i=1}^n T_i^2  \right\|^{\frac12}
\]

The trick now is to make a term like the one in the LHS appear in the RHS. For that we start by noting (you can see Fact 2.3 in~\cite{Tropp_MatrixConcentrationElementary} for an elementary proof) that, since $T_i\succeq 0$,
\[
\left\| \sum_{i=1}^n T_i^2  \right\| \leq \max_i \| T_i\|   \left\| \sum_{i=1}^n T_i  \right\|.
\]
This means that
\[
\EE \left\| \sum_{i=1}^n T_i  \right\| \leq \left\| \sum_{i=1}^n \EE T_i \right\| + \sqrt{C(d)} \EE \left[ \left(  \max_i \| T_i\| \right)^{\frac12} \left\| \sum_{i=1}^n T_i  \right\|^{\frac12} \right].
\]
Furthermore, applying the Cauchy-Schwarz inequality for $\EE$ gives,
\[
\EE \left\| \sum_{i=1}^n T_i  \right\| \leq \left\| \sum_{i=1}^n \EE T_i \right\| + \sqrt{C(d)}  \left( \EE \max_i \| T_i\| \right)^{\frac12} \left( \EE \left\| \sum_{i=1}^n T_i  \right\|\right)^{\frac12},
\]
Now that the term $\EE \left\| \sum_{i=1}^n T_i  \right\|$ appears in the RHS, the proof can be finished with a simple application of the quadratic formula (see Section 6.1. in~\cite{Tropp_MatrixConcentrationElementary} for details).

\end{proof}

We now show an inequality for general symmetric matrices

\begin{theorem}\label{thm:4:matrixconsymmetric}
Let $Y_1,\dots,Y_n\in\RR^{d\times d}$ be random independent symmetric matrices satisfying $\EE Y_i=0$, then
\[
\EE \left\| \sum_{i=1}^n Y_i  \right\| \leq \sqrt{C(d)}\sigma  + C(d) L,
\]
where,
\begin{equation}
\sigma^2 = \left\|  \sum_{i=1}^n \EE Y_i^2   \right\| \text{ and } L^2 = \EE \max_i \| Y_i\|^2
\end{equation}
and, as in~\eqref{eq:4:Cd_constant},
\begin{equation*}
C(d) \defeq 2\pi + 4\pi\lceil \log d \rceil.
\end{equation*}
\end{theorem}

\begin{proof}

Using Symmetrization (Lemma~\ref{lemma:4:symmetrization}) and Theorem~\ref{thm:4:RademacherSeries}, we get
\[
\EE \left\| \sum_{i=1}^n Y_i  \right\| \leq 2 \EE_Y \left[  \EE_\eps \left\|   \sum_{i=1}^n \eps_i Y_i \right\| \right] \leq \sqrt{C(d)} \EE \left\| \sum_{i=1}^n Y_i^2  \right\|^{\frac12}.
\]
Jensen's inequality gives
\[
\EE \left\| \sum_{i=1}^n Y_i^2  \right\|^{\frac12} \leq \left( \EE \left\| \sum_{i=1}^n Y_i^2  \right\| \right)^{\frac12},
\]
and the proof can be concluded by noting that $Y_i^2\succeq 0$ and using Theorem~\ref{thm:4:matrixconcentration:PSDcase}.
\end{proof}

\begin{remark}[The rectangular case]
One can extend Theorem~\ref{thm:4:matrixconsymmetric} to general rectangular matrices $S_1,\dots,S_n\in\RR^{d_1\times d_2}$ by setting
\[
Y_i = \left[ \begin{array}{cc}  0 & S_i \\ S_i^T & 0   \end{array} \right],
\]
the so-called Hermitian dilation, and noting that
\[
\left\| Y_i^2 \right\| = \left\| \left[ \begin{array}{cc}  0 & S_i \\ S_i^T & 0   \end{array} \right]^2   \right\| 
= \left\| \left[ \begin{array}{cc}  S_iS_i^T & 0 \\ 0 & S_i^TS_i    \end{array} \right]   \right\| = \max\left\{ \left\| S_i^TS_i \right\|, \left\| S_iS_i^T \right\|  \right\}.
\]
For details we refer to~\cite{Tropp_MatrixConcentrationElementary}.
\end{remark}

\section{Improvements leveraging intrinsic freeness}\label{subsec:freeprobimprovement}

In Section~\ref{sec:tightnessNCK} we saw that the logarithmic factor in the Non-commutative Khintchine (Theorem~\ref{thm:4:Gaussianseries}) inequality is required in the worst-case but superfluous in some instances. If we pay close attention to the proof of Theorem~\ref{thm:4:Gaussianseries}, we notice that there is a potential loss in using Lemma~\ref{lemma:commutativeisworstGIbP} when $A_1,\dots,A_n$ are non-commutative, and that perhaps the $(2p-1)$ factor coming from bounding every summand with the $q=0$ one, could be (in some cases) replaced by a constant, which would remove the logarithmic factor in the final bound.\footnote{It is worth noting that both the term $q=0$ and $q=2p-2$ are equal to the upper bound in Lemma~\ref{lemma:commutativeisworstGIbP}, however while it is tempting to hope to bound all other other terms by a smaller order parameter, this would yield a final bound of $\sqrt{2}\sigma$, which we know does not hold for Wigner matrices (which would need $2\sigma$). The notion of crossing and non-crossing partitions (See Definition~\ref{partitions:crossing}) is an elegant way of singling out which terms are substantial.}
 Tropp~\cite{Tropp:TailBoundsSecondOrder} made an important early connection between improvements in the Non-commutative Khintchine inequality and ideas in Free Probability~\cite{NicaSpeicher-FreeProbability} showing an improvement of Theorem~\ref{thm:4:Gaussianseries} in~\cite{Tropp:TailBoundsSecondOrder}. This is done by using Gaussian Integration by parts twice and controlling, with a (sometimes) smaller parameter $\omega$ -- called the matrix alignment parameter -- the summands where the factors from different applications of Gaussian Integration by parts ``cross''. This allows for a more efficient iteration argument and replaces the $\theta\big((\log n)^{\frac12}\big)$ factor multiplying $\sigma$ by a $\theta\big((\log n)^{\frac14}\big)$ factor.

We briefly describe below an improvement to Theorem~\ref{thm:4:Gaussianseries} in~\cite{Bandeiraetal_FreeProbability} that often yields sharp bounds. To describe it,  it is worth showing a slightly different proof of Theorem~\ref{thm:4:Gaussianseries}. Let $X=\sum_{k=1}^ng_kA_k$ and $p\geq 1$ as in Theorem~\ref{thm:4:Gaussianseries}. Wick's formula (Lemma~\ref{lemma:Wicksformula}) gives
\begin{eqnarray*}
\EE X^{2p} &=& \sum_{u:[2p]\to[n]} \EE[g_{u(1)}\cdots g_{u(2p)}] \tr\left(A_{u(1)}\cdots A_{u(2p)}\right) \\
&=& \sum_{u:[2p]\to[n]} \sum_{\nu\in \PP_2[2p]} 1_{u \sim \nu}\tr\left(A_{u(1)}\cdots A_{u(2p)}\right) \\
&=& \sum_{\nu\in \PP_2[2p]}\sum_{\substack{u:[2p]\to[n] \\ u\sim\nu} } \tr\left(A_{u(1)}\cdots A_{u(2p)}\right).
\end{eqnarray*}
If the matrices $A_k$ were commutative then the summands 
$$\sum_{\substack{u:[2p]\to[n] \\ u\sim\nu} } \tr\left(A_{u(1)}\cdots A_{u(2p)}\right)$$ would coincide for all pair partitions, and in particular with the pair partition $\nu_0 \defeq \{(1,2),(3,4),\dots,(2p-1,2p)\}$ on which the summand is given by
\[
\sum_{\substack{u:[2p]\to[n] \\ u\sim\nu_0} } \tr\left(A_{u(1)}\cdots A_{u(2p)}\right) = \sum_{\substack{u:[p]\to[n] } } \tr\left(A_{u(1)}^2\cdots A_{u(p)}^2\right)=\tr\left( \sum_{k=1}^n A_k^2\right)^p.
\]
The following Lemma is analogous to Lemma~\ref{lemma:commutativeisworstGIbP} and due to Buchholz~\cite{Buchholz01}.

\begin{lemma}[Commutative is the worst-case II~\cite{Buchholz01}]\label{lemma:commutativeisworstWicks}
For any $\nu\in\PP[2p]$ and $A_1\dots,A_{n}$ symmetric matrices
\[
\sum_{\substack{u:[2p]\to[n] \\ u\sim\nu} } \tr\left(A_{u(1)}\cdots A_{u(2p)}\right) \leq \tr\left( \sum_{k=1}^n A_k^2\right)^p.
\]
\end{lemma}

If $\sum_{k=1}^n A_k^2$ is a multiple of the identity matrix, there are many pair partitions that match the upper bound above: whenever there is an adjacent pair, it can be ``peeled-off'' in the sum and potentially make adjacent pairs that were not adjacent before; the partitions that can be fully ``peeled-off'' this way are precisely the so-called \emph{non-crossing} partitions and they indeed match the upper bound above (see Lemma~\ref{prop:GUENONcrossingpairings}).

\begin{definition}[Crossing and Non-crossing Partitions]\label{partitions:crossing}
We say $\nu\in\PP[2p]$ is a crossing partition when it has pairs $(i_1,i_2)\in\nu$ and $(j_1,j_2)\in\nu$ such that $i_1< j_1 < i_2 < j_2$. Otherwise we say $\nu$ is \emph{non-crossing}. The set of non-crossing partition is denoted by $\NC[2p]\subset \nu\in\PP[2p]$.
\end{definition}

Bandeira, Boedihardjo, and van Handel~\cite{Bandeiraetal_FreeProbability} showed that, in many settings, the random matrix $X$ exhibits intrinsic freeness in the sense that only the summand corresponding to non-crossing partitions are non-negligible.

This is done by interpolating (using Gaussian Interpolation, Lemma~\ref{lemma:gaussianinterpolation}) the Gaussian model $X$ with a Free Probability model $X_{\mathrm{free}}$ where the Gaussians are replaced by a \emph{semi-circular family} -- these can be viewed, in a sense, as ``non-commutative random variables'' for which Wick's formula (Lemma~\ref{lemma:Wicksformula}) holds when summing only over $\mathrm{NC}[2p]$ (this can be viewed as the limiting behavior of Lemma~\ref{lemma:WicksformulaGUE} below, and indeed the argument can be viewed as replacing Gaussians by ever larger Wigner matrices). While the matrix alignment parameter of Tropp is a key component in the argument, intrinsic freeness is controlled by $v^2=\left\| \mathrm{Cov} X \right\|$, the spectral norm of the covariance matrix of the entries of $X$ (this covariance matrix is a $d^2\times d^2$ matrix). When $\frac{\sigma}{v} \gg\polylog(d) $ the improvement yields $\EE\|X\| \leq (2+o(1))\sigma$. 

Furthermore, Brailovskaya and van Handel~\cite{Tatianaetal_Universality}  developed a universality principle that roughly states that for any sum of independent random matrices  $Y = \sum_{i=1}^n Y_i$ (such as in Theorem~\ref{thm:4:matrixconsymmetric}), as long as the summands are sufficiently small, the matrix $Y$ behaves like a Gaussian analogue $Y_G$ where the entries of $Y$ are replaced by Gaussian random variables with the same mean and covariance, and for which the results in~\cite{Bandeiraetal_FreeProbability} can be used (notice that this is different than the argument, based on symmetrization, done to prove Theorem~\ref{thm:4:matrixconsymmetric} from Theorem~\ref{thm:4:Gaussianseries}). Combining these two tools, one obtains an improvement over the matrix Bernstein inequality.

\begin{theorem}[\cite{Bandeiraetal_FreeProbability,Tatianaetal_Universality}] \label{thm:improvedBernstein}
Let $Y_1,\dots,Y_n\in\RR^{d\times d}$ be random independent symmetric matrices satisfying $\EE Y_i=0$, and such that $\|Y_i\|\leq R$, for all $i\in[n]$, almost surely. Then
\[
\EE \left\| \sum_{i=1}^n Y_i  \right\| \leq 2\sigma + C \Big( v^{\frac12}\sigma^{\frac12}(\log d)^\frac34 + R^{\frac13}\sigma^{\frac23}(\log d)^\frac23 + R \log d \Big),
\]
and
\[
\PP\Big[  \left\| \sum_{i=1}^n Y_i  \right\| \geq 2\sigma + C \Big( v^{\frac12}\sigma^{\frac12}(\log d)^\frac34 + \sigma_\ast t^{\frac12}+ R^{\frac13}\sigma^{\frac23}t^\frac23 + R t \Big)    \Big] \leq de^{-t}
\]
where, $C$ is a universal constant, 
\begin{equation}
\sigma^2 = \left\|  \sum_{i=1}^n \EE Y_i^2   \right\| \text{, } v^2 = \left\| \mathrm{Cov}(Y) \right\| \text{ and } \sigma_\ast^2 = \sup_{\|u\|=\|w\|=1}\EE \left| u^TYw \right|^2.
\end{equation}
\end{theorem}

Note that if $v,\sigma_\ast,R\ll \frac{1}{\polylog(d)}\sigma$, which happens often in applications (see~\cite{Bandeiraetal_FreeProbability,Tatianaetal_Universality}), then all terms multiplying the universal constant $C$ are negligible and, in the tail bound, the tail parameter $t$ only appears in low-order terms.

\subsection{Crossings and Wick's formula for the GUE}

An instructive pursuit is to compute an analogue of Wick's formula for a complex valued analogue of the Wigner matrix that has appeared in Section~\ref{subsection:WignerMatrices} and Example~\ref{subsection:WignerMatrices}). This is an instance in which the phenomenon of cancellations arising from crossing partitions is particularly transparent (and it is tightly connected to the seminal work of Voiculescu~\cite{Voiculescu1991} and Haagerup-Thorbj{\o}rnsen~\cite{HaagerupThorbjornsen2005}, and to the methods in~\cite{Bandeiraetal_FreeProbability} that give rise to the inequalities in Theorem~\ref{thm:improvedBernstein}).

\begin{definition}[Gaussian Unitary Ensemble (GUE)]
A random $D
\times D$ GUE matrix $W$ is a random Hermitian (self-adjoint) matrix whose upper off-diagonal entries are iid $\CNNN(0,\frac1D)$ and whose diagonal entries are iid $\NNN(0,\frac1D)$ (also independent from the off-diagonal entries). $W_{ij}\sim \CNNN(0,\frac1D)$ means that $\Repart\left(W_{ij}\right) \sim ~\NNN(0,\frac1D)$, $\Impart\left(W_{ij}\right) \sim ~\NNN(0,\frac1D)$, and that both are independent.
\end{definition}

The goal of this subsection is to show that, for large-dimensional GUEs, there is Wick's formula whose non-negligible terms only involve non-crossing partitions.

\begin{lemma}[Wick's formula for GUEs (cf. Lemma~\ref{lemma:Wicksformula})]\label{lemma:WicksformulaGUE}
Let $W_1,\dots,W_n$ be iid $D\times D$ GUE matrices and let $u:[2p]\to[n]$ then
\[
\frac1D\EE\left[\tr W_{u(1)}\cdots W_{u(2p)}\right] = \sum_{\nu\in \NC_2[2p]} 1_{u \sim \nu} + \mathcal{O}\left(\frac1{D^2}\right).
\]
where $\NC_2[2p]$ denotes de set of non-crossing pair partitions, $u \sim \nu$ means the function $u$ is compatible with $\nu$, and the constant in $\mathcal{O}$ may depend on $n,p,$ and $u$ (see Definitions~\ref{def:pairpartition} and~\ref{partitions:crossing}).
\end{lemma}

We start by showing that non-crossing terms do not have cancellations.

\begin{definition}
 Let $\nu\in\PP[2p]$ (see Definition~\ref{def:pairpartition}). We define $u_{\nu}:[2p]\to[p]$ as first surjective assignment $u:[2p]\to[p]$, in the lexicographic order, that satisfies $u\sim \nu$.\footnote{For all our uses here, we could have picked any surjective assignment $u:[2p]\to[p]$, the one that is lexicographically first is an arbitrary choice.}
\end{definition}

\begin{proposition}\label{prop:GUENONcrossingpairings}
Let $\nu\in\NC[2p]$, and let $W_1,\dots,W_{p}$ be iid $D\times D$ GUE matrices, then
\begin{equation}\label{eq:GUENONcrossingpairings}
    \EE\left[\tr W_{u_\nu(1)}\cdots W_{u_\nu(2p)}\right] = \tr(I)
\end{equation}
\end{proposition}
\begin{proof}
Since $\nu$ is non-crossing there must exist $i\in[2p-1]$ such that $u_\nu(i)=u_\nu(i+1)$. For such as $i$ we have that the LHS of~\eqref{eq:GUENONcrossingpairings} is equal to
\[
\EE\left[\tr W_{u_\nu(1)}\cdots W_{u_{\nu(i-1)}} \left(\EE_{W_{u_\nu(i)}}  W_{u_\nu(i)}^2 \right) W_{u_\nu(i+2)}\cdots W_{u_\nu(2p)}\right],
\]
where the outside expectation is over the other random matrices. Furthermore, since $\left(\EE_{W_{u_\nu(i)}}  W_{u_\nu(i)}^2 \right)=I$, such pairs can be ``peeled-off'' one by one to finish the proof (formally, one can do induction by noting that $\nu\setminus\{i,i+1\}$ is also non-crossing).
\end{proof}

We are now ready to examine crossing partitions.

\begin{proposition}\label{prop:GUEcrossingFormula}
Let $B_1,B_2,B_3,B_4$ be $D\times D$ (complex) matrices and let $W,V$ be two independent $D\times D$ GUE matrices. We have
\begin{equation}\label{eq:crossingformula}
\EE\left[ \tr W B_1 V B_2 W B_3 V B_4 \right] = \frac{1}{D^2} \tr\left( B_3B_2B_1B_4 \right).
\end{equation}
\end{proposition}
\begin{proof}
This follows from a computation, and noting that for $W\sim\mathrm{GUE}$ we have $
\EE W_{i_1j_1}W_{i_2j_2} = \frac1{D}\delta_{i_1j_2}\delta_{i_2j_1}$ (where $\delta_{ij}$ is the indicator of $i=j$).\footnote{An analogous  calculation can be done for real value Wigner matrices (Example~\ref{example:WignerMatrix}), also called GOE, but there are more choices of $i_1,j_1,i_2,j_2$ that yield non-zero moments, making the calculation less transparent).} Indeed, expanding the LHS of~\eqref{eq:crossingformula} we see it is equal to
\begin{eqnarray*}
\mathrm{LHS}\text{ of }\eqref{eq:crossingformula} & = & \sum_{i_1,\dots,i_8=1}^D \EE\left[W_{i_1i_2} (B_1)_{i_2i_3} V_{i_3i_4} (B_2)_{i_4i_5} W_{i_5i_6} (B_3)_{i_6i_7} V_{i_7i_8} (B_4)_{i_8i_1} \right] \\ 
& = & \sum_{i_1,\dots,i_8=1}^D \frac1{D^2}\delta_{i_1i_6}\delta_{i_2i_5}\delta_{i_3i_8}\delta_{i_4i_7}  (B_1)_{i_2i_3}  (B_2)_{i_4i_5}  (B_3)_{i_6i_7}  (B_4)_{i_8i_1} \\
& = & \frac1{D^2}\sum_{i_1,\dots,i_4=1}^D  (B_1)_{i_2i_3}  (B_2)_{i_4i_2}  (B_3)_{i_1i_4}  (B_4)_{i_3i_1} \\
& = & \frac1{D^2}\tr\left[ B_3 B_2 B_1 B_4\right].
\end{eqnarray*}
\end{proof}

\begin{proposition}\label{prop:GUEcrossingpairings}
 Let $\nu\in\PP[2p]\setminus\NC[2p]$ be a crossing partition, and let $W_1,\dots,W_{p}$ be iid $D\times D$ GUE matrices, then
\begin{equation}
    0\leq \EE\left[\tr W_{u_\nu(1)}\cdots W_{u_\nu(2p)}\right] \leq \frac1{D^2}\tr(I)
\end{equation}
\end{proposition}
\begin{proof}
This follows by taking a crossing $\{i,j,i',j'\}$ such that $i<j<i'<j'$,  $u_\nu(i)=u_\nu(i')$, and $u_\nu(j)=u_\nu(j')$, applying Proposition~\ref{prop:GUEcrossingFormula} on $W_{u_\nu(i)}$ and $W_{u_\nu(j)}$ and using induction on $p$.
\end{proof}

\begin{proof}[of Lemma~\ref{lemma:WicksformulaGUE}]

With all the ingredients collected, this follows directly from the useful equality
\begin{equation}\label{eq:WicksMixedFormula?}
    \EE\left[\tr W_{u(1)}\cdots W_{u(2p)}\right] = \sum_{\nu\in\PP[2p]} 1_{u \sim \nu} \EE\left[\tr W_{u_{\nu}(1)}\cdots W_{u_{\nu}(2p)}\right],
\end{equation}
and Propositions~\ref{prop:GUEcrossingpairings} and~\ref{prop:GUENONcrossingpairings}. 

The equality~\eqref{eq:WicksMixedFormula?} is a consequence of Lemma~\ref{lemma:Wicksformula}, a formal proof follows by noting that $W_{\ell}$ are Gaussian matrices and thus we can write $W_{\ell} = \sum_{j=1}^m g_{\ell;j} A_j$ for some $A_j$ (complex valued). This means we have
\begin{eqnarray*}
    \EE\left[\tr W_{u(1)}\cdots W_{u(2p)}\right] 
    &=& \sum_{j_1,\dots,j_{2p}=1}^{m} \tr\left(A_{j_1}
    \cdots A_{j_{2p}}\right) \EE\left[ g_{u(1);j_1}
    \cdots g_{u(2p);j_{2p}} \right]\\
    &=&  \sum_{j_1,\dots,j_{2p}=1}^{m} \tr\left(A_{j_1}
    \cdots A_{j_{2p}}\right) \sum_{\nu\in\PP[2p]} 1_{u \sim \nu} 1_{j \sim \nu} \\
    &=& \sum_{\nu\in\PP[2p]} 1_{u \sim \nu} \left( \sum_{j_1,\dots,j_{2p}=1}^{m} \tr\left(A_{j_1}
    \cdots A_{j_{2p}}\right)   1_{j \sim \nu} \right),
\end{eqnarray*}
and on the other hand,
\begin{eqnarray*}
    \EE\left[\tr W_{u_{\nu}(1)}\cdots W_{u_{\nu}(2p)}\right] 
    &=& \sum_{j_1,\dots,j_{2p}=1}^{m} \tr\left(A_{j_1}
    \cdots A_{j_{2p}}\right) \EE\left[ g_{u_{\nu}(1);j_1}
    \cdots g_{u_{\nu}(2p);j_{2p}} \right]\\
    &=&  \sum_{j_1,\dots,j_{2p}=1}^{m} \tr\left(A_{j_1}
    \cdots A_{j_{2p}}\right) 1_{j \sim \nu},
\end{eqnarray*}
where the second equality uses the fact that $\nu$ is the only pairing compatible with $u_\nu$.
\end{proof}

\section*{Exercises}
\addcontentsline{toc}{section}{Exercises}

\begin{definition}[Gaussian Wigner matrix]
    Let $W$ be a $d \times d$ matrix, which is a gaussian series by taking $W = \sum_{1 \leq k \leq l \leq d}g_{kl}A_{kl}$, where
    \begin{equation*}
        A_{kl} = \begin{cases}
            e_ke_l^\top + e_le_k^\top &\text{for }k<l,\\
            \sqrt{2} e_ke_k^\top &\text{for }k=l.
        \end{cases}
    \end{equation*}
\end{definition}

\begin{myexercise}[\level\sep Constructing Wigner]\label{prob:constructing_wigner}
    Let $Z$ be a $d \times d$ random matrix whose entries are all independent standard gaussians (in total $d^2$
of them). Show that $\frac{1}{\sqrt{2}}\brap{Z + Z^\top}$ is a Gaussian Wigner matrix.
\end{myexercise}

\begin{definition}[Diagonal matrix with Gaussian entries]
    Let $D$ be a $d \times d$ diagonal matrix, which is a gaussian series by taking $D = \sum_{1 \leq k \leq d}g_{k}A_{k}$, where
    \begin{equation*}
        A_{k} = e_ke_k^\top.
    \end{equation*}
\end{definition}

\begin{myexercise}[\level\sep Computing the $\sigma$ parameter]\label{prob:computing_sigma}
    Let $W$ be a $d \times d$ Gaussian Wigner matrix and let $D$ be a $d \times d$ diagonal matrix with independent standard gaussians on the diagonal. Show that $\norm{\mathbb{E}W^2}^{\frac12} = \sigma(W) = \sqrt{d+1}$ and $\norm{\mathbb{E}D^2}^{\frac12} = \sigma(D) = 1$, and upper bound $\E\norm{W}$ and $\E\norm{D}$ using the Non-commutative Khintchine inequality.
\end{myexercise}

\begin{myexercise}[\level\level\sep Intrinsically free Non-commutative Khintchine Inequality]\label{prob:strong_NCK}
    In fact, a stronger version of the Non-commutative Khintchine inequality is known. Let $A_1, \ldots, A_n \in \R^{d \times d}$ by symmetric matrices and $g_1, \ldots, g_n \in \NN(0,1)$ i.i.d. The gaussian series $X = \sum_{t=1}^n g_t A_t$ satisfies
    \begin{equation*}
        \E\norm{X} \leq 2 \sigma + C\,v\,(\log d)^{\frac{3}{2}},
    \end{equation*}
    where $C > 0$ is an absolute constant, $\sigma^2 = \norm{ \sum_{i=1}^n A_i^2}$, and $v$ is given by
    \begin{equation*}
        v^2 = \norm{\Cov(X)}.
    \end{equation*}
    Here, matrix covariance $\Cov(X)$ is a $d^2 \times d^2$ matrix, whose row and column coordinates are indexed by pairs of indices, and entries are given by
    \begin{equation*}
        \Cov(X)_{ij,kl} = \E\bras{X_{ij}X_{kl}} \quad\text{for }i,j,k,l\in[d].
    \end{equation*}
    Compute $\Cov(W)$ and $\Cov(D)$, where $W$ and $D$ are as in Problem~\ref{prob:computing_sigma}, and deduce using the intrinsically free Non-commutative Khintchine Inequality that there is an absolute constant $C' > 0$, such that
    \begin{equation*}
        \E\norm{W} \leq C'\,\sqrt{d}.
    \end{equation*}
\end{myexercise}
\begin{hint}
    As a start, show that $\Cov(X)_{ij,kl} = \sum_{t=1}^n \bras{A_t}_{ij} \bras{A_t}_{kl}$.
\end{hint}

\begin{myexercise}[\level\sep Hermitian dilation]\label{prob:hermitian_dilation}
    In order to extend the matrix Bernstein inequality from symmetric to general rectangular matrices, we will use Hermitian dilation. For a matrix $S \in \R^{d_1 \times d_2}$, the Hermitian dilation $\mathcal{H}(S)$ is defined as
    \begin{equation*}
        \mathcal{H}(S) \coloneqq \begin{pmatrix}
            0_{d_1 \times d_1} & S \\
            S^\top & 0_{d_2 \times d_2}
        \end{pmatrix}\in\R^{(d_1 + d_2)\times (d_1 + d_2)}.
    \end{equation*}
    In Problem~\ref{prob:symmetrization}(c) we showed that $\mathcal{H}(S)$ is symmetric and that $\norm{\mathcal{H}(S)} = \norm{S}$.
    
    Let $S_1, \dots, S_n \in \R^{d_1 \times d_2}$ be random rectangular matrices satisfying $\E\bras{S_i} = 0$ for every $i \in [n]$. Show that 
    \begin{equation*}
        \E \norm{ \sum_{i=1}^n S_i } \le \sqrt{C(d)}\, \sigma + C(d)\, L,
    \end{equation*}
    where $d = d_1 + d_2$, $C(d)  = 4 + 8 \ceil{\log d}$ and
    \begin{equation*}
        \sigma^2 = \max \brac{\norm{ \sum_{i=1}^n \E\bras{S_i S_i^\top}},  \norm{ \sum_{i=1}^n \E\bras{S_i^\top S_i}}},\qquad L^2 = \E \max_i \norm{S_i}^2.
    \end{equation*}
\end{myexercise}

\begin{myexercise}[\level\level\level\sep Bernstein's inequality - expectation bound]\label{prob:bernstein_expectation_bound}
    Let $\brac{X_k}_{k=1}^n$ be a sequence of independent random symmetric $d \times d$ matrices. Assume that each $X_k$ satisfies:
    \begin{equation*}
        \E X_k=0 \text { and } \|X_k\|\leq R \text { almost surely. }
    \end{equation*}
    In this exercise, the goal is to show that
    \begin{equation}\label{eq:bernstein_expectation}
        \E \norm{\sum_{k=1}^n X_k} \le C \brap{ \sigma \sqrt{\log{(d+1)}} + R\log(d+1)}.
    \end{equation}
    To get a bound on expectation from the tail bound in Theorem~\ref{thm:4:MatrixBernstein}, we will use an integral identity for the expectation (see Problem~\ref{prob:integral_identity}(a)).
    \begin{enumerate}[(a)]
        \item Show that for any $a, b > 0$
        \begin{equation*}
            \exp\brap{-\frac{2}{a+b}} \le \max\brac{\exp\brap{-\frac{1}{a}}, \exp\brap{-\frac{1}{b}}} \le \exp\brap{-\frac{1}{a}} + \exp\brap{-\frac{1}{b}}.
        \end{equation*}
        Apply it to the exponent in the right-hand side of matrix Bernstein's inequality so that you get integrals that are easier to compute. 
        \item For very small $t$, the tail bound is loose since the exponent is close to $1$. Think how we can isolate the case of $t$ close to $0$.
        \item Once you split the integral, it remains only to compute the individual parts and choose the right constants in your argument. To find it, you can closely examine the final bound \eqref{eq:bernstein_expectation}.  
    \end{enumerate}
\end{myexercise}

\begin{myexercise}[\level\level\sep Randomized matrix multiplication]\label{prob:randomized_matrix_mult}
    Let $A \in \R^{n\times m}$ be a real-valued matrix with unit spectral norm $\norm{A} = 1$. The cost of computing $AA^\top$, using the standard matrix multiplication method, is of the order of $n^2 m$, which can be prohibitive when $n$ and $m$ are very large. In some cases, it is sufficient to obtain only an approximate solution which allows us to reduce the costs significantly. In this problem you will show that by using randomness we can get an approximation of the product more efficiently.

    Denote $a_1, \dots, a_m \in \R^n$ columns of matrix $A$. Define a random matrix $X$ such that $\P\brap{X = m \cdot a_k a_k^\top} = \frac{1}{m}$. 
    \begin{enumerate}[(a)]
        \item Suppose we draw $s$ independent copies of $X$, denoted by $X_1, \dots, X_s$, and then average them $\hat{X}_s = \frac{1}{s} \sum_{k=1}^s X_k$. Prove that $\hat{X}_s$ in an unbiased estimator of $A A^\top$, meaning that $\E \hat{X}_s = A A^\top$.
        
        \item Define the coherence statistic as $\mu(A) = m \cdot \displaystyle{\max_{k = 1, \dots, m} \|a_k\|^2} $. Show that:
        \begin{equation*}
            \max_{k = 1, \dots, s} \norm{\E \bras{(X_k - \E X_k)^2}}  \le 2 \mu(A)
        \end{equation*}
        and 
        \begin{equation*}
            \max_{k = 1, \dots, s} \|X_k - \E X_k \| \le 2\mu(A), \text{ almost surely}.
        \end{equation*}
        
        \item Use Problem~\ref{prob:bernstein_expectation_bound} (proof not needed) to show that, for an absolute constant $C$, if the number of samples $s$ satisfies
        \begin{equation*}
            s \ge C \max \brac{\frac{1}{\eps}, \frac{1}{\eps^2} } \mu \log n
        \end{equation*}
        then the procedure achieves $\eps$-accuracy, i.e., $\E\norm{\hat{X}_s - AA^\top} \le \eps$.
    \end{enumerate}
\end{myexercise}

\begin{myexercise}[\level\sep Commuting vs. simultaneously diagonalizable]\label{prob:commuting_vs_sim_diag}
    Let $A, B \in \R^{d\times d}$ be two symmetric matrices.
    \begin{enumerate}[(a)]
        \item If $A$ and $B$ are simultaneously diagonalizable, show that they commute.
        
        \item If $A$ and $B$ commute, and $B$ has all eigenvalues distinct, show that they are simultaneously diagonalizable.

        \item Generalize (a) and (b) for $n$ symmetric matrices $A_1,\ldots, A_n \in \R^{d\times d}$.
    \end{enumerate}
\end{myexercise}
\begin{hint}
    For (b) show that if $B v = \lambda v$ for some $v \in \R^n$ and $\lambda \in \R$ then $B (A v) = \lambda (A v)$.
\end{hint}

\begin{myexercise}[\level\level\sep NCK for commuting matrices]\label{prob:NCK_commuting}
    In the book we discussed the role of commutativity of the matrices for the upper bound of the expected value of the spectral norm of a random matrix. Recall that for $X \coloneqq \sum_{i=1}^n g_i A_i$, where $A_1, \dots, A_n \in \R^{d\times d}$ are symmetric matrices and $g_1, \dots, g_n \overset{\text{iid}}{\sim} \NN(0,1)$, it holds
    \begin{equation*}
        \sigma \lesssim \E \norm{X} \lesssim  \sigma \sqrt{\log d},
    \end{equation*}
    where $\sigma^2 \coloneqq \norm{ \sum_{i=1}^n A_i^2}$.
    
    \begin{enumerate}[(a)]
        \item Suppose that $A_1, \dots, A_n \in \R^{d\times d}$ are symmetric commuting matrices. This means that they are simultaneously diagonalizable (well known fact, no proof needed), so there is an orthogonal matrix $Q \in \R^{d \times d}$ so that $D_i \coloneqq Q A_i Q^{-1}$ is a diagonal matrix for any $i \in [n]$. Let $\lambda^{(i)}_{1}, \dots, \lambda^{(i)}_{d}$ be the entries that appear, in that order, on the diagonal of $D_i$. Show that
        \begin{equation*}
            \E\norm{ X } = \E \max_{k = 1, \ldots, d} \abs{\sum_{i=1}^n g_i \lambda^{(i)}_{k}}.
        \end{equation*}
        
        \item Deduce using Problem~\ref{prob:maximum_gaussians}(a) that  
        \begin{equation*}
            \E \norm{X} \lesssim \sigma \sqrt{\log d}.
        \end{equation*}
        
        \item Find an example of commuting matrices $A_1, \dots, A_n$ such that Problem~\ref{prob:maximum_gaussians}(b) implies
        \begin{equation*}
            \E \norm{X} \gtrsim \sigma \sqrt{\log d}.
        \end{equation*}
    \end{enumerate}
\end{myexercise}

\begin{myexercise}[\level\level\sep Trace Commutativity Inequality]\label{prob:trace_commutativity_ineq}
    The goal of this exercise is to show the following key inequality that proves why commuting matrices perform worse than non-commuting matrices in trace moment estimations. One can actually follow a slightly different approach than the one shown in the book if one follows the hint in (b).
    \\
    Let $X,A \in \R^{d \times d}$ be symmetric matrices and let $k,l$ be nonnegative integers with $k+l$ being even, then 
    $$ \tr( AX^{k}AX^{l}) \leq \tr(A^2X^{k+l}).$$
    \begin{enumerate}[(a)]
        \item Let $X = \sum_{i=1}^d \lambda_i u_i u_i^\top$ be the the eigenvalue decomposition of $X$, prove
        $$ \tr( AX^{k}AX^{l}) \leq \sum_{i,j=1}^d \abs{\lambda_i}^k \abs{\lambda_j}^l (u_i^\top A u_j)^2.$$
        \item Finish the proof by showing
        $$ \sum_{i,j=1}^d \abs{\lambda_i}^k \abs{\lambda_j}^l (u_i^\top A u_j)^2 \leq \tr(A^2X^{k+l}). $$
        \begin{hint}
        Use Young's Inequality on the product of eigenvalues.
        \end{hint}
    \end{enumerate}
\end{myexercise}


\chapter{Compressive Sensing and Sparsity}
\label{c:cs}

Most of us have noticed how saving an image in JPEG dramatically reduces the space it occupies on our hard drives (as opposed to file types that save the value of each pixel in the image). The idea behind these compression methods is to exploit known structure in the images; although our cameras will record the  value (even three values in RGB) for each pixel, it is clear that most collections of pixel values will not correspond to pictures that we would expect to see. Natural images do not correspond to arbitrary arrays of pixel values, but have some specific structure to them. It is this  special structure one aims to exploit by choosing a proper representation of the image. Indeed, natural images are known to be approximately sparse in certain bases (such as the wavelet bases) and this is the core idea behind JPEG (actually, JPEG2000; JPEG uses a different basis).

\section{Sparse recovery}\label{s:sparse}

Let us think of $x\in \CC^p$ as the signal corresponding to the image already represented in the basis in which it is sparse. The modeling assumption is that $x$ is $s$-sparse, or $\|x\|_0\leq s$, meaning that $x$ has at most $s$ non-zero components and, usually, $s\ll p$. The $\ell_0$-norm\footnote{We recall that the $\ell_0$ norm is not actually a norm, as it does not necessarily rescale linearly with a rescaling of $x$.} $\|x\|_0$ of a vector $x$ is the number of non-zero entries of $x$. 
This means that when we take a picture, our camera makes $p$ measurements (each corresponding to a pixel) but then, after an appropriate change of basis, it keeps only $s\ll p$ non-zero coefficients and drops the others. This seems a rather wasteful procedure and thus motivates the question: ``If only a few degrees of freedom are kept after compression, why not in the first place measure in a more efficient way and take considerably less than $p$ measurements?''. 

The question whether we can carry out  data acquisition and compression simultaneously is at the heart of {\em Compressive Sensing}~\cite{Candes_CS1,Candes_CS2,Candes_CS3,Candes_CS4,Donoho_CS,FoucartRauhut_CSbook}. It is particularly important in MRI imaging \cite{lustig2007sparse,feng2017compressed}, as fewer measurements potentially means shorter data acquisition time. Indeed, current MRI technology based on concepts from compressive sensing can, in certain cases, reduce the time needed to collect the data by a factor of six or more~\cite{lustig2007sparse}, which has been reported to have significant benefits especially in pediatric MR imaging~\cite{vasanawala2010improved}.
We recommend the book~\cite{FoucartRauhut_CSbook} as a great in-depth reference about compressive sensing.

In mathematical terms, the acquired measurements $y \in \CC^m$ are connected to the signal  of interest $x \in \CC^p$, with $m \ll p$, via 
\begin{equation}\label{cs}
\left[\begin{array}{c} \\ y \\ \  \end{array}\right] = \left[\begin{array}{cccccccccccccccccc} \\  & & & & & &  & A & & & & & & & & & & \\ \  \end{array}\right] \left[\begin{array}{c}  \\ \\ \\ x \\ \\ \\ \   \end{array}\right].
\end{equation}
The matrix $A \in\CC^{m \times p}$ models the linear measurement (information) process. Classical linear algebra tells us that if $m < p$, then the linear system~\eqref{cs} is underdetermined and that there are infinitely many solutions (assuming that there exists at least one solution). In other words, without additional information, it is impossible to
recover $x$ from $y$ in the case $m < p$.

In this chapter, we assume that $x$ is $s$-sparse with $s<m\ll p$. The goal is to recover $x$ from this underdetermined system and do this in a computationally efficient manner.
We emphasize that we {\em do not  know} the location of the non-zero coefficients of $x$ a priori\footnote{And therein lies the challenge, since $s$-sparse signals  do not form a linear subspace of $\RR^p$ (the sum of two $s$-sparse signals is in general no longer $s$-sparse but $2s$-sparse).}, otherwise the task would be trivial.

\subsection{Gaussian width of $s$-sparse vectors}\label{ss:gausswidth}

Before discussing algorithms, let us discuss well-posedness of the problem at hand: In order to be able to reconstruct $x$ from $y$ we need at the very least that $A$ is injective on sparse vectors.
Furthermore, in order for reconstruction to be stable, one should ask not only that $A$ is injective with respect to $s$-sparse vectors, but actually that it is almost an isometry, meaning that the $\ell_2$ distance between $Ax_1$ and $Ax_2$ should be comparable to the distances between $x_1$ and $x_2$, if they are $s$-sparse. Since the difference between two $s$-sparse vectors is in general a $2s$-sparse vector, we can alternatively ask for $A$ to approximately preserve the norm of $2s$-sparse vectors. Gordon's Theorem (Theorem~\ref{GordonsTheorem}) suggests that we can take $A\in\RR^{m\times p}$ to have i.i.d.\ Gaussian entries and to take $m\approx \omega^2\left(\mathcal{S}_{2s}\right)$, where $\mathcal{S}_{2s} = \left\{x:\,x\in\SSS^{p-1},\,\|x\|_0\leq 2s\right\}$ is the set of $2s$-sparse vectors, and $\omega\left(\mathcal{S}_{2s}\right)$ denotes the Gaussian width of $\mathcal{S}_{2s}$ (see Definition~\ref{def:gaussianwidth}).

\begin{proposition}\label{proposition:ssparse:gaussianwidth}
 If $s\leq p$, the Gaussian width $\omega\left(\mathcal{S}_{s}\right)$ of $\mathcal{S}_s$, the set of unit-norm vectors that are at most $s$-sparse, satisfies
 \[
  \omega\left(\mathcal{S}_{s}\right)^2 \lesssim s \log\left( \frac{p}s \right).
 \]
\end{proposition}

\proofb{
 \[
  \omega\left(\mathcal{S}_{s}\right) = \EE \max_{v\in\SSS^{p-1},\, \|v\|_0\leq s} g^Tv, 
 \]
 where $g\sim \NNN(0,I_{p\times p})$. We have
 \[
  \omega\left(\mathcal{S}_{s}\right) = \EE \max_{\Gamma\subset [p],\, |\Gamma|=s} \|g_{\Gamma}\|,
 \]
 where $g_{\Gamma}$ is the restriction of $g$ to the set of indices $\Gamma$.

Given a set $\Gamma$, Theorem~\ref{thm:chisquare_concentration}  yields
\[
 \Prob\left\{  \|g_{\Gamma}\|^2 \geq s + 2\sqrt{s}\sqrt{t} + 2t \right\} \leq \exp(-t).
\]
Union bounding over all $\Gamma\subset [p],\, |\Gamma|=s$ gives
\[
 \Prob\left\{\max_{\Gamma\subset [p],\, |\Gamma|=s} \|g_{\Gamma}\|^2 \geq s + 2\sqrt{s}\sqrt{t} + 2t \right\} \leq {p \choose s} \exp(-t).
\]

Taking $u$ such that $t=su$, gives
\begin{equation}\label{5:equation:boundsparsegaussianwidth}
 \Prob\left\{\max_{\Gamma\subset [p],\, |\Gamma|=s} \|g_{\Gamma}\|^2 \geq s\left(1+2\sqrt{u} + 2u\right)  \right\} \leq \exp\left[-su + s\log\left(e\frac{p}s\right)\right].
\end{equation}
Taking $u> \log\left(e\frac{p}s\right)$ reveals that the typical size of $\max_{\Gamma\subset [p],\, |\Gamma|=s} \|g_{\Gamma}\|^2$ is $\lesssim s\log\left(\frac{p}s\right)$. The proof can be completed by integrating~\eqref{5:equation:boundsparsegaussianwidth} in order to get a bound of the expectation of $\sqrt{\max_{\Gamma\subset [p],\, |\Gamma|=s} \|g_{\Gamma}\|^2}$.

}

This suggests that $\approx 2s\log\left(\frac{p}{2s}\right)$ measurements suffice to stably identify a $2s$-sparse vector. Indeed, we will see below that at this order of number of measurements it will actually be possible to efficiently recover an  $s$-sparse vector.

\subsection{Sparse recovery and $\ell_1$ optimization}

\begin{figure}[h]
\subcaptionbox{$p=0$}{ \includegraphics[height=18mm]{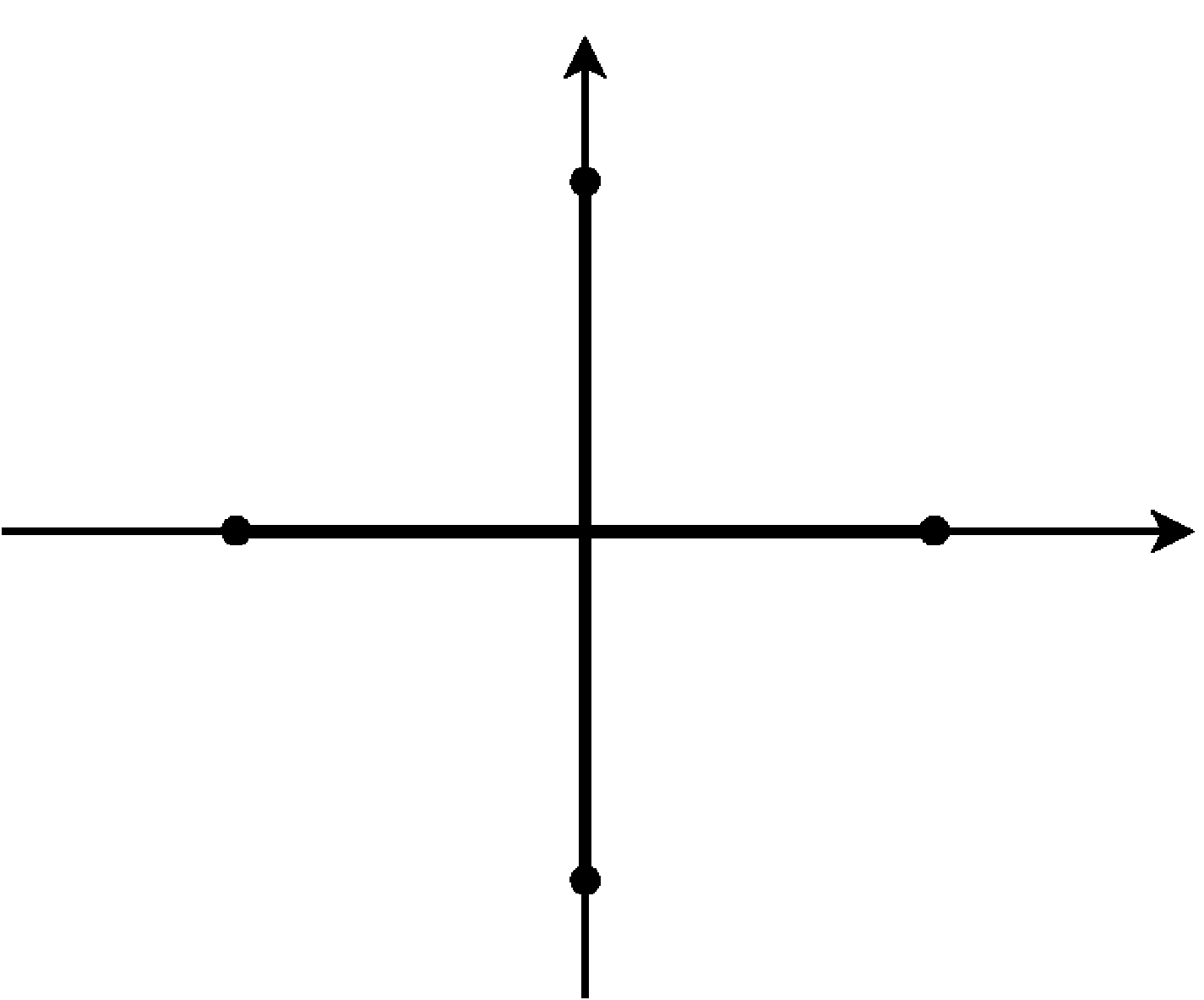}}
\subcaptionbox{$p=\frac{1}{2}$}{\includegraphics[height=18mm]{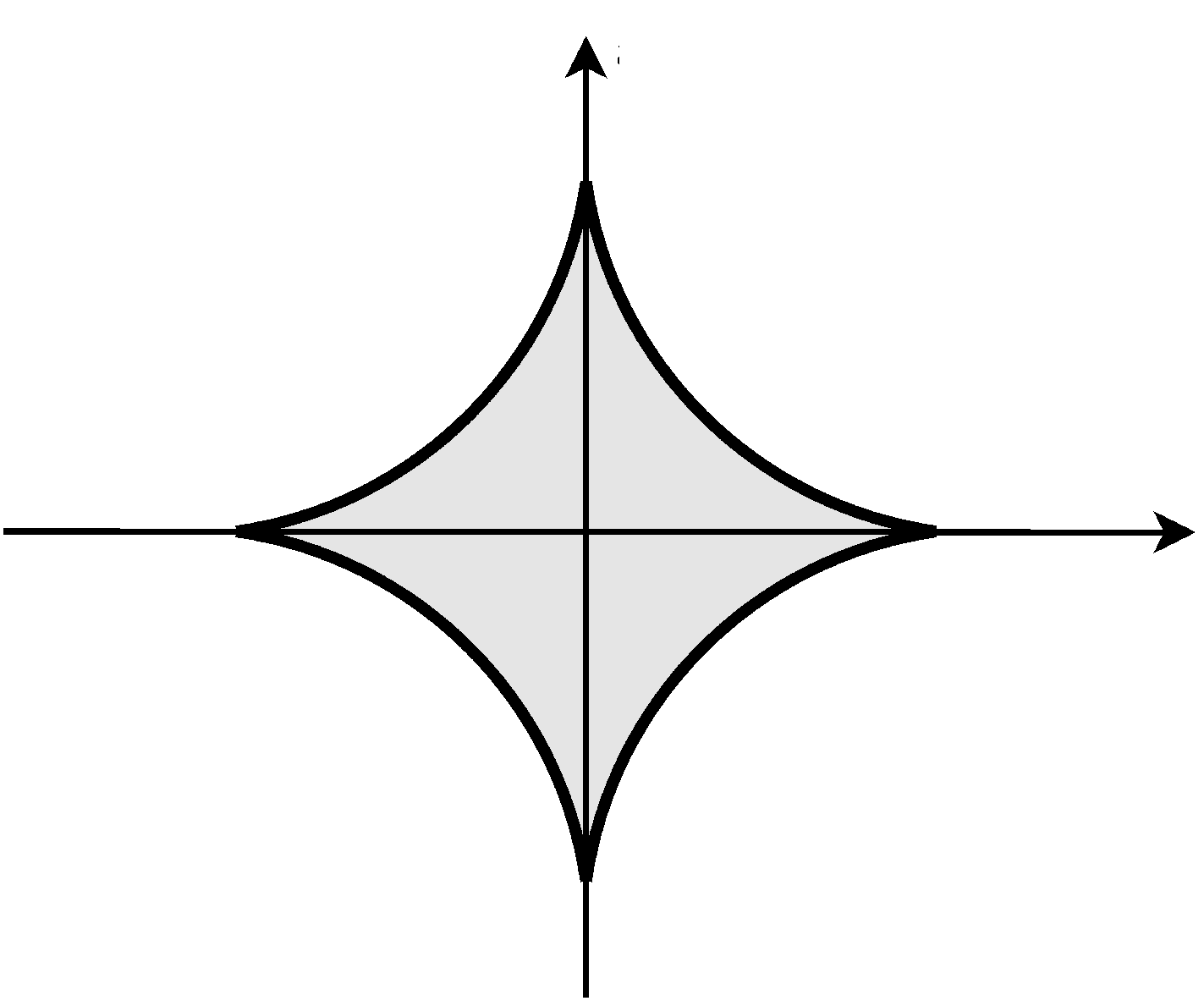}}
\subcaptionbox{$p=1$}{\includegraphics[height=18mm]{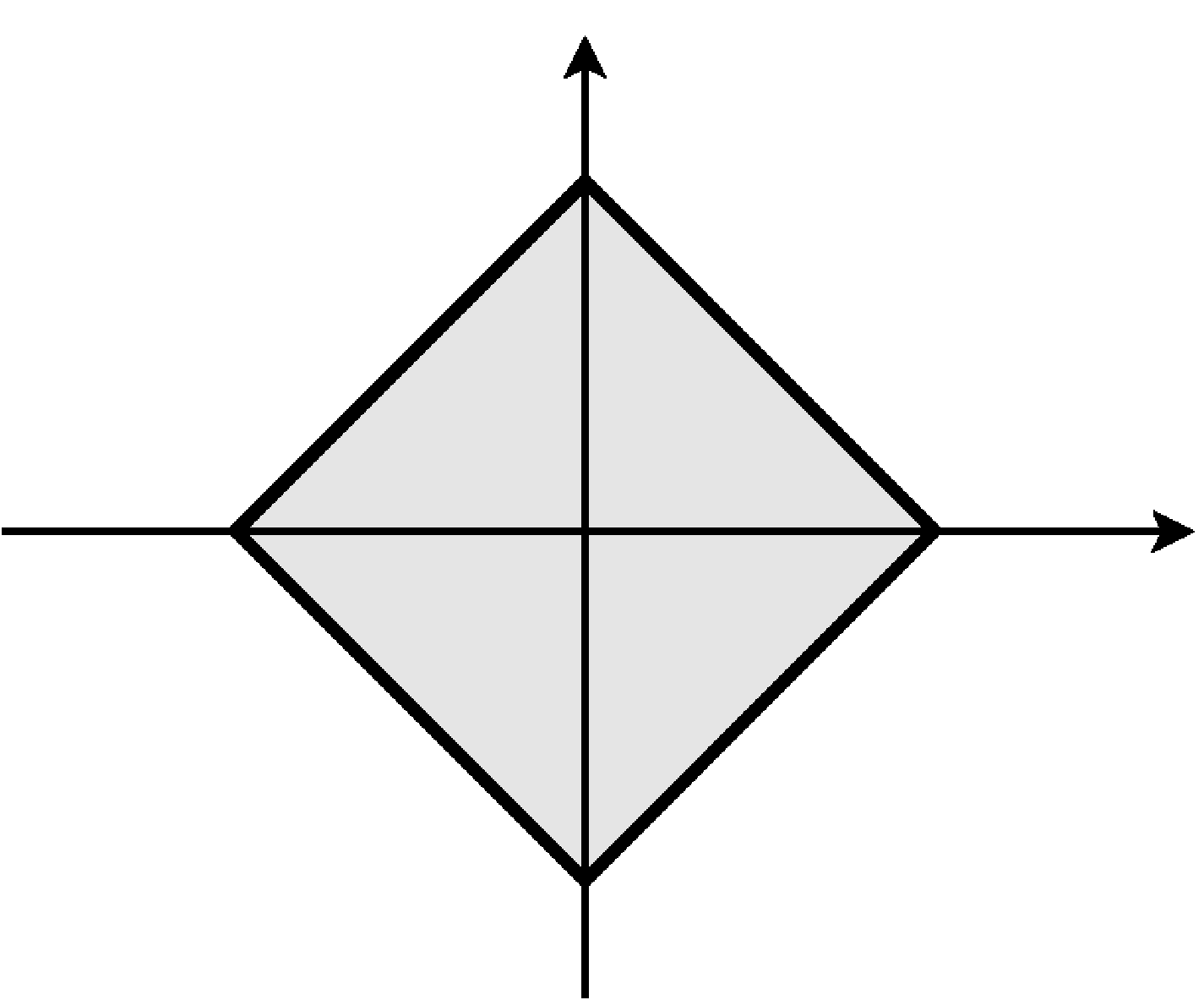}}
\subcaptionbox{$p=2$}{\includegraphics[height=18mm]{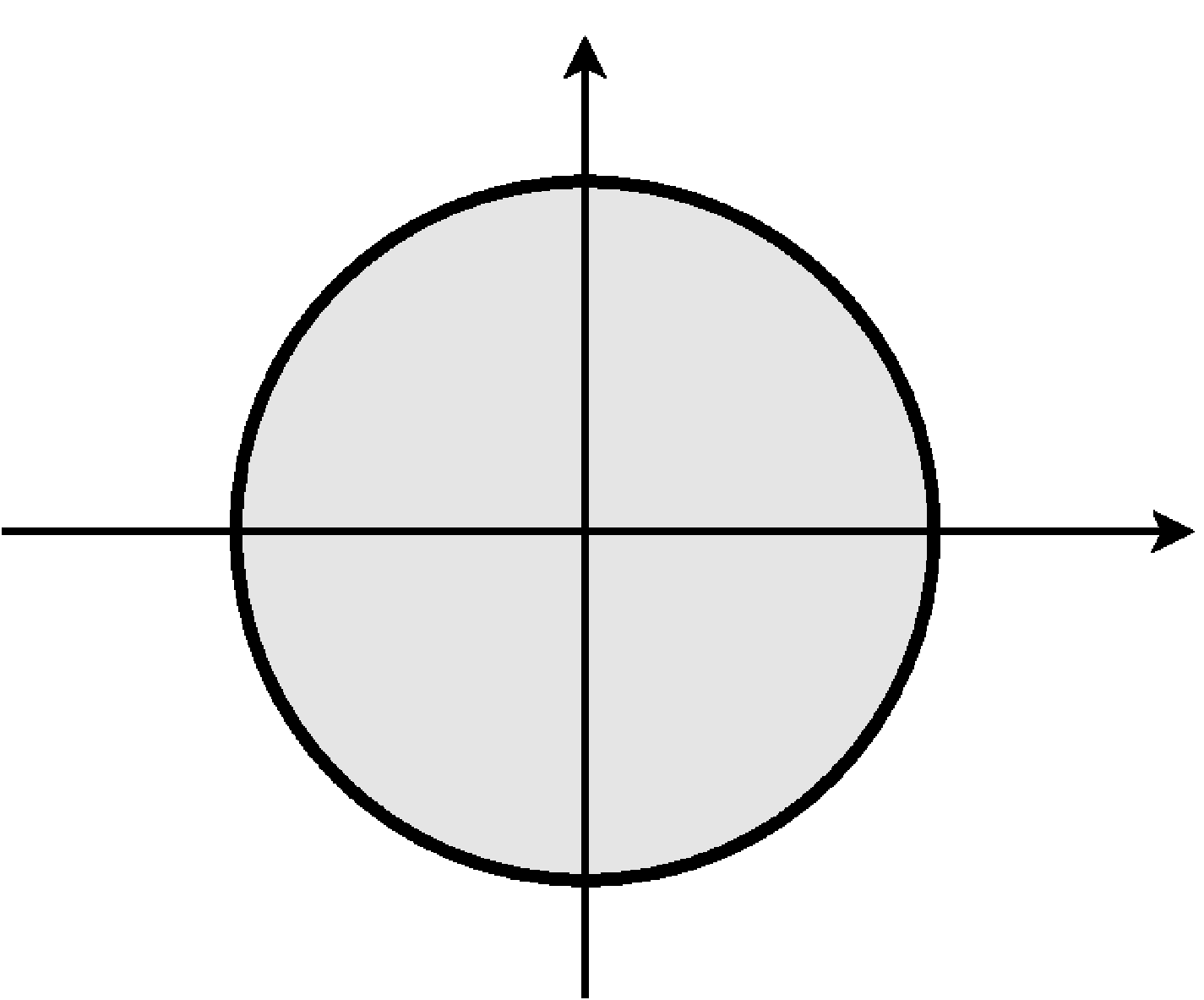}}
\subcaptionbox{$p=\infty$}{\includegraphics[height=18mm]{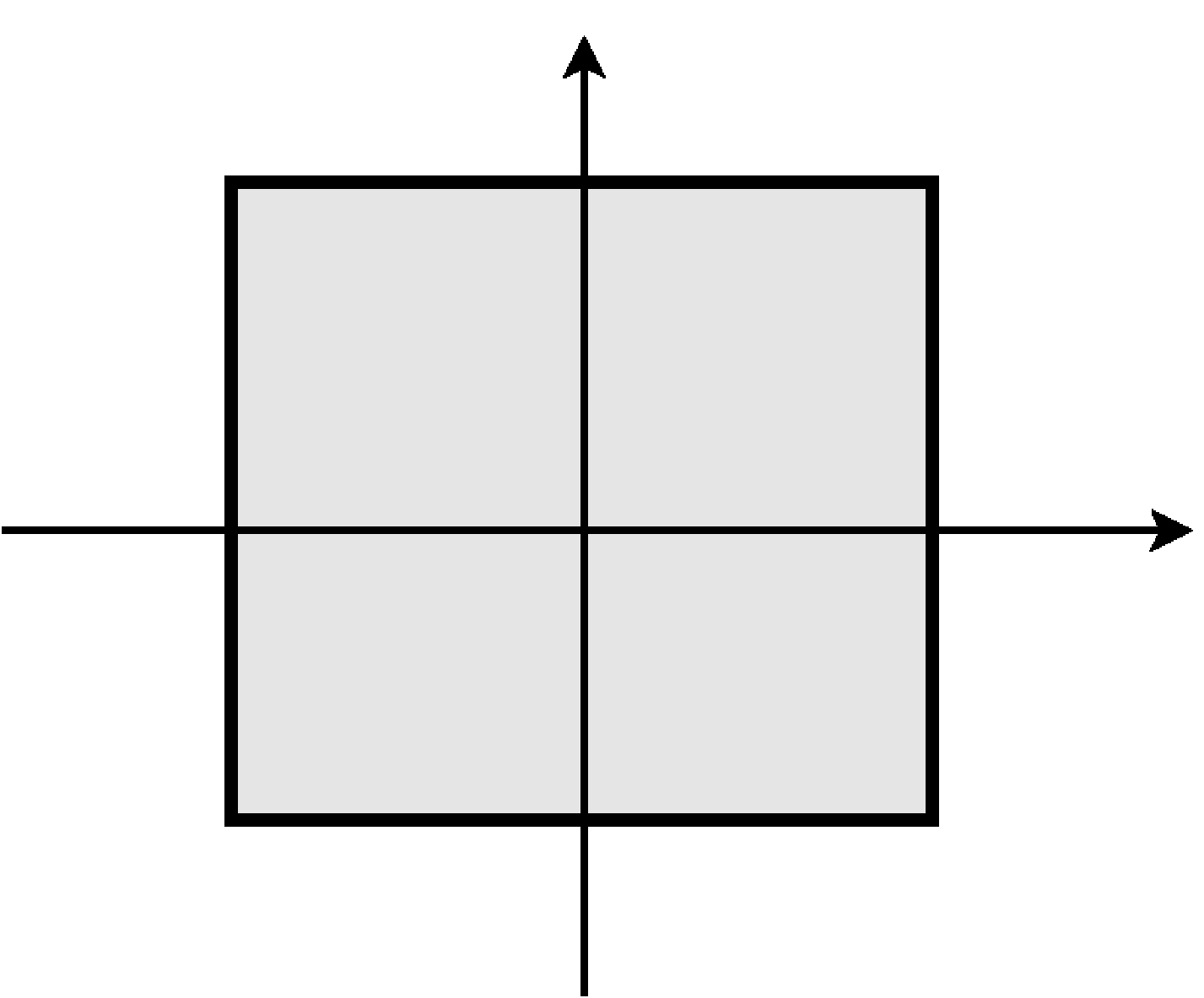}}
\caption{$\ell_p$ norm unit balls with different values for $p$}
\label{fig:lp-norm}
\end{figure}

Since the system~\eqref{cs} is underdetermined and we know that $x$ is sparse, the natural approach to try  to recover $x$ is to solve
\begin{equation}\label{eq:6:L0normmin}
\begin{array}{cl}
\min \,\,& \|z\|_0 \\
\text{s.t.} & Az = y,
\end{array}
\end{equation}
and hope that the optimal solution $z$ corresponds to the signal in question $x$. However the optimization problem~\eqref{eq:6:L0normmin} is  NP-hard  in general~\cite{natarajan1995sparse,FoucartRauhut_CSbook}. Instead, the approach usually taken in sparse recovery is to consider a convex surrogate of the $\ell_0$ norm, namely the $\ell_1$ norm: $\|z\|_1 = \sum_{i=1}^p |z_i|$. 
Figure~\ref{fig:lp-norm} depicts the $\ell_p$ balls and illustrates how the $\ell_1$ norm can be seen as a convex surrogate of the $\ell_0$ norm due to the \emph{pointiness} of the $\ell_1$ ball in the direction of the basis vectors, i.e. in ``sparse'' directions.  

\begin{figure}[h]
\subcaptionbox{$p=1$}{\includegraphics[height=40mm]{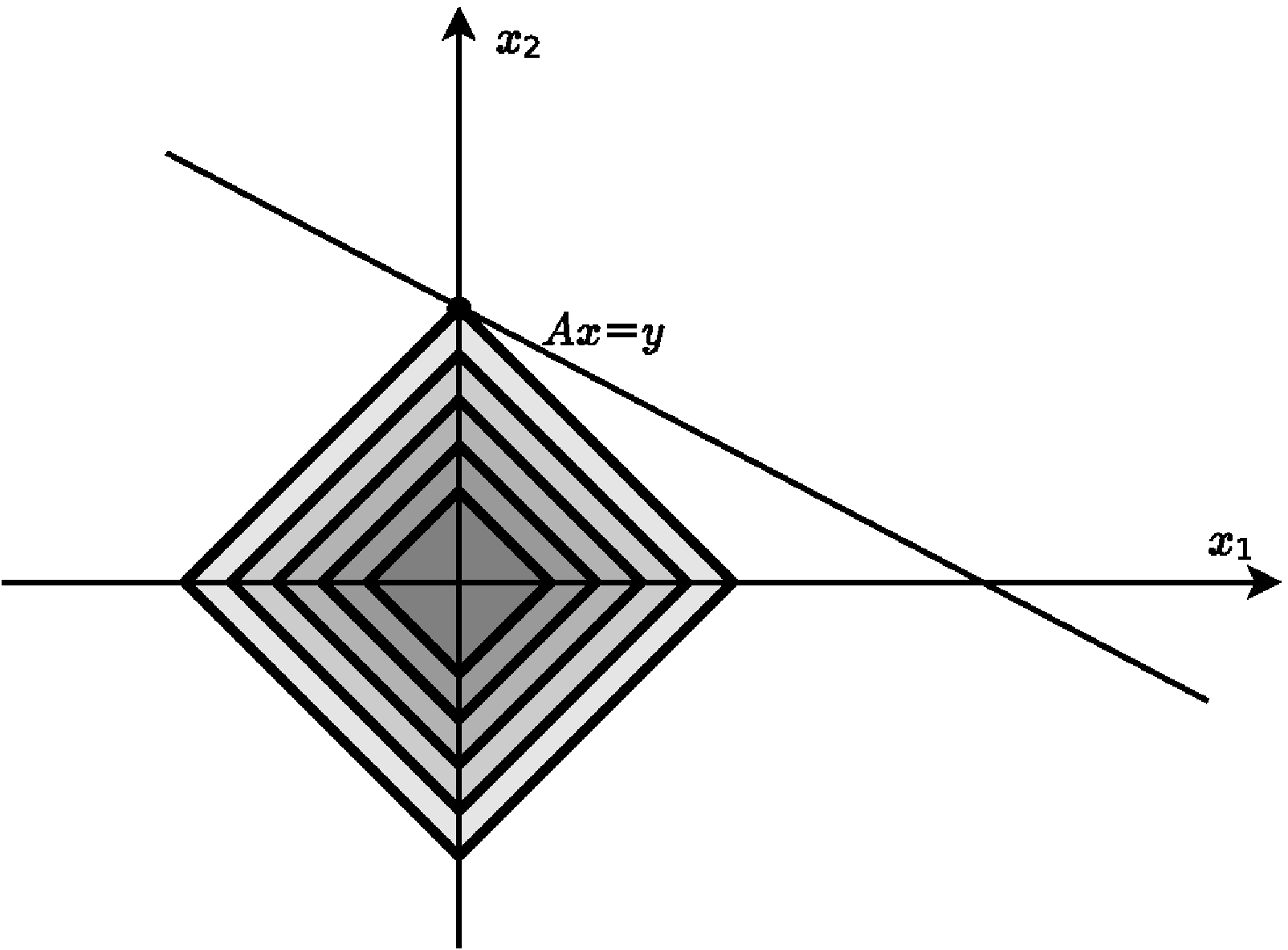}}\qquad
\subcaptionbox{$p=2$}{\includegraphics[height=40mm]{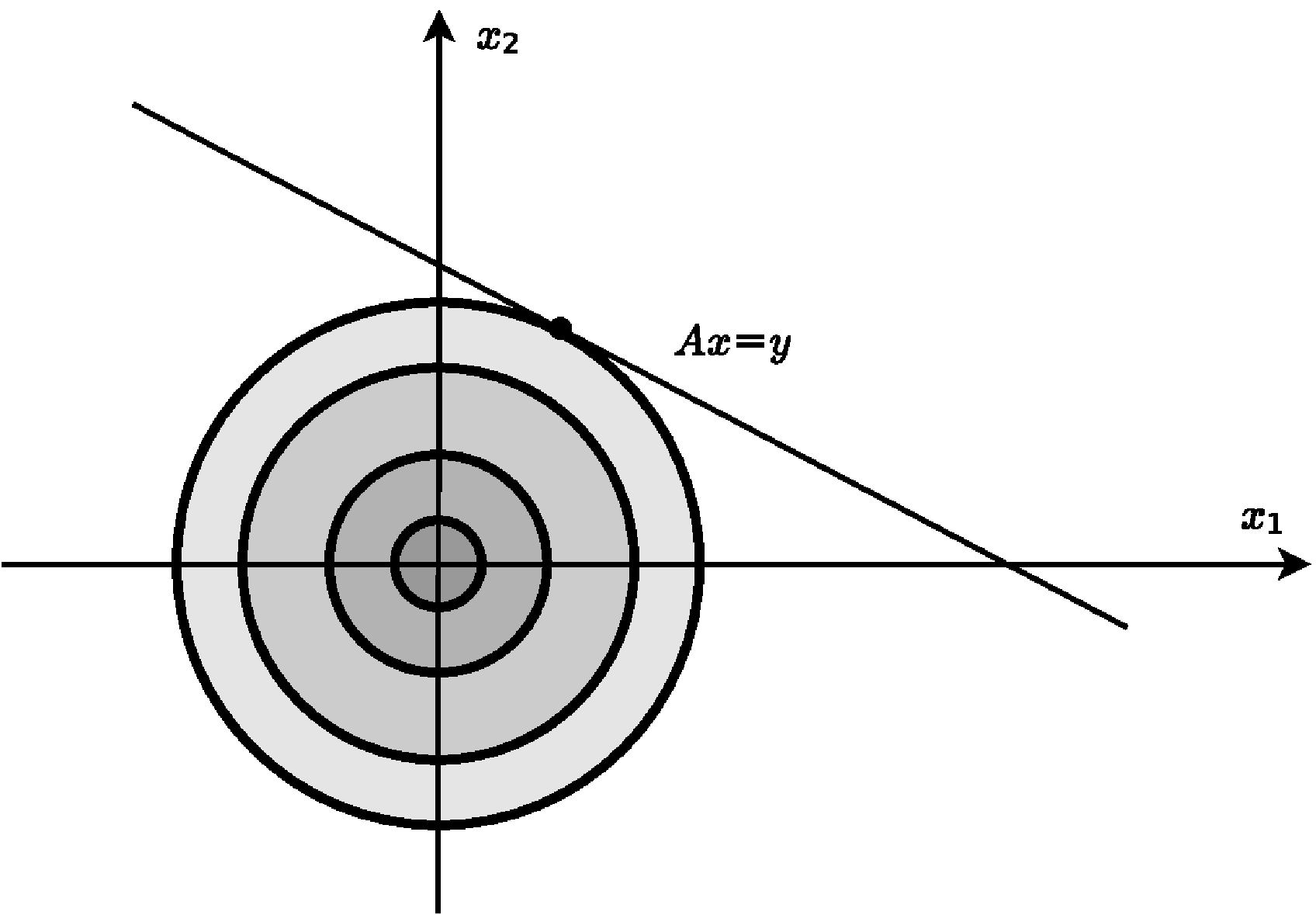}}
\caption{A two-dimensional depiction of $\ell_1$ and $\ell_2$ minimization. In $\ell_p$ minimization, one inflates the $\ell_p$ ball  until it hits the affine subspace of interest. This image conveys how the $\ell_1$ norm (left) promotes sparsity due to the ``pointiness'' of the $\ell_1$ ball. In contract,  $\ell_2$ norm minimization (right) does not favor sparse solutions.}
\label{fig:l1l2}
\end{figure}

The process of $\ell_p$ minimization can be understood as inflating (or deflating) the $\ell_p$ ball until one hits the affine subspace of interest. Figure~\ref{fig:l1l2} illustrates how $\ell_1$ norm minimization promotes sparsity, while $\ell_2$  norm minimization does not favor sparse solutions. 
We have seen in Chapter~\ref{ss:hypercube} that the $\ell_1$ ball becomes ``increasingly pointy'' with increasing dimension. This behavior works in our favor in compressive sensing---another manifestation of the {\em blessings of dimensionality}.

This motivates us to consider the following optimization problem  (surrogate to~\eqref{eq:6:L0normmin}):
\begin{equation}\label{eq:6:L1normmin}
\begin{array}{cl}
\min \,\, & \|z\|_1 \\
\text{s.t.} & Az = y,
\end{array}
\end{equation}

In order for~\eqref{eq:6:L1normmin} to be a good procedure for sparse recovery we need two things: for the solution of it to be meaningful (hopefully to coincide with $x$); and for~\eqref{eq:6:L1normmin} to be efficiently solvable.

\begin{remark}
We will consider for the moment the real-valued case $x\in\RR^p, A\in\RR^{m\times p}$ and formulate~\eqref{eq:6:L1normmin} as a linear program\footnote{In the complex case, we are dealing with a quadratic program.} (and thus show that it is efficiently solvable). Let us think of $\omega^{+}$ as the positive part of $z$ and $\omega^{-}$ as the symmetric of the negative part of it, meaning that $z = \omega^{+} - \omega^{-}$ and, for each $i$, either $\omega^{-}_i$ or $\omega^{+}_i$ is zero. Note that, in that case,
\[
\|z\|_1 = \sum_{i=1}^p \omega^{+}_i + \omega^{-}_i = \1^T \left(  \omega^{+} + \omega^{-} \right).
\]
Motivated by this, we consider:

\begin{equation}\label{eq:6:L1normmin_LP}
\begin{array}{cl}
\min \,\, &  \1^T \left(  \omega^{+} + \omega^{-} \right) \\
\text{s.t.} & A\left( \omega^{+} - \omega^{-} \right) = y \\
 & \omega^{+} \geq 0 \\
 & \omega^{-} \geq 0,
\end{array}
\end{equation}
which is a linear program. It is not difficult to see that the optimal solution of~\eqref{eq:6:L1normmin_LP} will indeed satisfy that, for each $i$, either $\omega^{-}_i$ or $\omega^{+}_i$ is zero and it is indeed equivalent to~\eqref{eq:6:L1normmin}; if both $\omega^{-}_i$ and $\omega^{+}_i$ are non-zero, one can lower the objective while keep satisfying the constraints by reducing both variables. Since linear programs are efficiently solvable~\cite{LVanderberghe_SBoyd_book}, this implies that the $\ell_1$-optimization problem~\eqref{eq:6:L1normmin} is efficiently solvable. This optimization program is also efficiently solvable over $\CC$ despite not being a linear program, but a quadratic program.
\end{remark}

In what follows we will discuss under which circumstances one can guarantee that the solution of~\eqref{eq:6:L1normmin} coincides with the sparse signal of interest. We will discuss a couple of different strategies to show this, as different strategies generalize better to other problems of interest. Later in this chapter we also discuss strategies for constructing sensing matrices.

\section{Null space property and exact recovery}
\label{s:nsp}

Given a $s$-sparse vector $x$, our goal is to show that under certain conditions $x$ is the unique optimal solution to

\begin{equation}
\begin{array}{cl}
\min \,\, & \|z\|_1 \\
\text{s.t.} & Az = y,
\end{array}
\end{equation}

Let $S = \supp(x)$, with $|S|=s$.\footnote{If $x$ has support size strictly smaller than $s$, in what follows, we can simply take a superset of it with size $s$} If $x$ is not the unique optimal solution of the $\ell_1$ minimization problem, there exists $z\neq x$ as optimal solution. Let $v = z-x$, it satisfies
\[
\|v+x\|_1 \leq \|x\|_1 \quad \text{ and } \quad A( v+x ) = Ax,
\]
this means that $Av=0$. Also, 
$$\|x\|_S = \|x\|_1 \geq \|v+x\|_1 = \|\left(v+x\right)_S\|_1 + \|v_{S^c}\|_1 \geq \|x_S\|_1 - \|v_S\|_1 + \|v_{S^c}\|_1,$$ where the last inequality follows by applying the triangle inequality. This means that $\|v_S\|_1 \geq \|v_{S^c}\|_1$, but since $|S|\ll p$ it is unlikely for $A$ to have vectors in its nullspace that are so concentrated on such few entries. This motivates the following definition.

\begin{definition}[Null Space Property]\label{def:nullspace}
$A$ is said to satisfy the $s$-Null Space Property ($A\in \text{s-NSP}$) if, for all $v$ in $\ker(A)$ (the nullspace of $A$) and all $|S| = s$, we have
\[
\|v_S\|_1 < \|v_{S^c}\|_1.
\]
\end{definition}

From the argument above, it is clear that if $A$ satisfies the Null Space Property for $s$, then $x$ will indeed be the unique optimal solution to~\eqref{eq:6:L1normmin}. In fact, as the property is described in terms of any set $S$ of size $s$, it implies recovery for any $s$-sparse vector.

\begin{theorem}\label{thm:NSPimpliesRecovery}
Let $x$ be an $s$-sparse vector. If $A\in\text{s-NSP}$ then $x$ is the unique solution to the $\ell_1$ optimization problem~$\eqref{eq:6:L1normmin}$ with $y=Ax$.
\end{theorem}

The Null Space Property is a statement about certain vectors not belonging to the null space of $A$, thus we can again resort to Gordon's Theorem  (Theorem~\ref{GordonsTheorem}) to establish recovery guarantees for Gaussian sensing matrices. Let us define the intersection with the unit-sphere of the cone of such vectors
\begin{equation}\label{eq:def:Cs}
C_s \defeq \left\{v\in\SSS^{p-1}: \exists_{S\subset [p],\, |S|=s} \left\|v_S\right\|_1\geq \left\|v_{S^c}\right\|_1 \right\}.
\end{equation} 

\subsection{Gordon's Theorem and the Null Space Property}\label{ss:gordonnullspace}

Since for an $m\times p$ matrix $A$, $A\in\text{s-NSP}$ is equivalent to $\ker(A) \cap C_s = \emptyset $, Gordon's Theorem, or more specifically Gordon's Escape Through a Mesh Theorem (Theorem~\ref{thm:GordonEscapeMesh}), implies that there exists a universal $C>0$ such that if $A$ is drawn with iid Gaussian entries, it will satisfy the s-NSP with high probability provided that $m \geq C\, \omega^2\left( C_s \right)$, where $ \omega\left( C_s \right)$ is the Gaussian width of $C_s$ (as defined in~\eqref{eq:def:Cs}).

\begin{proposition}
Let $s\leq p$ and $C_s\subset \SSS^{p-1}$ defined in~\eqref{eq:def:Cs}. There exists a universal constant $C$ such that
\[
\omega\left( C_s \right) \leq C \sqrt{s \log\left( \frac{p}s \right)},
\]
where $ \omega\left( C_s \right)$ is the Gaussian width of $C_s$ (as defined in~\eqref{eq:def:Cs}).
\end{proposition}

\proofb{
The goal is to upper bound
\[
\omega\left( C_s \right) = \EE \max_{v\in C_s} v^Tg,
\]
for $g\sim\NNN(0,I)$. Note that $C_s$ is invariant under permutations of the indices. Thus, the maximizer $v\in C_s$ will have its largest entries (in absolute value) in the coordinates where $g$ has its largest entries (in absolute value). Let $S$ be the set of the $s$ coordinates with largest absolute value of $g$. We have
\[
\EE \max_{v\in C_s} v^Tg = \EE \max_{v: \left\|v_S\right\|_1\geq \left\|v_{S^c}\right\|_1,\, \|v\|=1} v_S^Tg_S + v_{S^c}^Tg_{S^c}.
\]
The key idea is to notice that the condition $\left\|v_S\right\|_1\geq \left\|v_{S^c}\right\|_1$ imposes a strong bound on the $\ell_1$ norm of $v_{S^c}$ via $ \left\|v_{S^c}\right\|_1 \leq \left\|v_S\right\|_1 \leq \sqrt{s} \left\|v_S\right\| \leq \sqrt{s}$. This can be leveraged by noticing that
\[
v_S^Tg_S + v_{S^c}^Tg_{S^c} \leq \left\|v_S\right\| \left\|g_S\right\| + \left\|v_{S^c}\right\|_1 \left\|g_{S^c}\right\|_\infty.
\]
This gives 
\[
\omega\left( C_s \right) \leq \EE \left\|g_S\right\| + \sqrt{s} \left\|g_{S^c}\right\|_\infty,
\]
where $S$ corresponds to the set of the $s$ coordinates with largest absolute value of $g$.

We saw in the proof of Proposition~\ref{proposition:ssparse:gaussianwidth}, in the context of computing the Gaussian width of the set of sparse vectors, that $\EE \left\|g_S\right\| \lesssim  \sqrt{s \log\left( \frac{p}s \right)}$. Since all entries of $g_{S^c}$ are smaller, in absolute value than any entry in $g_S$ we have that $\left\|g_{S^c}\right\|_\infty^2 \leq \frac{1}s  \left\|g_S\right\|^2$. This implies that $ \EE \left\|g_{S^c}\right\|_\infty \lesssim \sqrt{\log\left( \frac{p}s \right)}$, concluding the proof.
}

Together with Theorem~\ref{thm:NSPimpliesRecovery} this implies the following recovery guarantee, matching the order of number of measurements suggested by the Gaussian width of sparse vectors.

\begin{theorem}\label{thm:GaussiansSatisfyNSP}
There exists a universal constant $C\geq 0$ such that if $A$ is a $m\times p$ matrix with i.i.d.\ Gaussian entries, and $m\geq C s \log\left(\frac{p}{s}\right)$, the following holds with high probability:
For any $x$ an $s$-sparse vector, $x$ is the unique solution to the $\ell_1$-optimization problem~\eqref{eq:6:L1normmin} with $y=Ax$.
\end{theorem}

\section{The Restricted Isometry Property}
\label{s:rip}

An alternative (and more classical) approach to establishing exact recovery via $\ell_1$-minimization is through the Restricted Isometry Property (RIP), which corresponds precisely with the property of approximately preserving the length of sparse vectors.

\begin{definition}[Restricted Isometry Property (RIP)]\label{def:6:RIP_2}
An $m\times p$ matrix $A$ (with either real or complex valued entries) is said to satisfy the $(s,\delta)$-Restricted Isometry Property (RIP),
\[
 (1-\delta)\|x\|^2 \leq \left\| Ax\right\|^2 \leq (1+\delta)\|x\|^2,
\]
for all $s$-sparse $x$.
\end{definition}

If $A$ satisfies the RIP  for sparsity $2s$, it means that it approximately preserves distances between $s$-sparse vectors (hence the name {\em RIP}). This can be leveraged to show that $A$ satisfies the NSP.

\begin{theorem}[\cite{Candes-RIP}]
Let $y=Ax$ where $x$ is an $s$-sparse vector. Assume that $A$ satisfies the RIP property
with $\delta_{2s} < \frac{1}{3}$, then the solution $x_*$ to the $\ell_1$-minimization problem
\begin{equation}\label{l1min}
\min_z \|z\|_1, \qquad \text{subject to } Az = y = Ax
\end{equation}
becomes $x$ exactly, i.e., $x_* = x$
\label{thm:maincs}
\end{theorem}

To prove this theorem we need the following lemma.

\begin{lemma}[\cite{Candes-RIP}]\label{lemma:np}
We have
\[
| \langle Ax, Ax^\prime \rangle | \leq \delta_{s+s^{\prime}} \|x\| \|x^\prime\|
\]
for all $x, x^\prime$ supported 
on disjoint subsets $S, S^\prime \subseteq [1, \cdots, p]$, $x, x^\prime \in \R^p$,
and $|S| \leq s$, $|S^\prime| \leq s^\prime$
\end{lemma}

\proofb{

Without loss of generality, we can assume $\|x\| = \|x^{\prime}\| = 1$,
so that the right hand size of the inequality becomes just $\delta_{s+s^\prime}$.
Since $A$ satisfies the RIP property, we have
\[
(1- \delta_{s+s^\prime}) \|x \pm x^\prime \|^2
\leq \|A(x \pm x^\prime) \|^2
\leq (1 + \delta_{s+s^\prime}) \|x \pm x^\prime\|^2.
\]
Since $x$ and $x^\prime$ have disjoint support,
$\| x \pm x^\prime \|^2 = \|x\|^2 + \|x^\prime\|^2 = 2$; the RIP property then becomes
\[
2(1- \delta_{s+s^\prime}) \leq \|Ax \pm Ax^\prime \|^2
\leq 2(1 + \delta_{s+s^\prime})
\]
The polarization identity implies:
\begin{align*}
| \langle Ax, Ax^\prime \rangle | 
& = \frac{1}{4} \Big| \|Ax + Ax^\prime \|^2 - \|Ax - Ax^\prime\|^2 \Big| \\
& \leq \frac{1}{4} \Big| 2(1+\delta_{s+s^\prime}) - 2(1-\delta_{s+s^\prime}) \Big| \\
& = \delta_{s+s^{\prime}}.
\end{align*}

}

To prove Theorem~\ref{thm:maincs}, we simply need to show that the
Null Space Property holds for the given conditions.

\begin{proof}[of Theorem~\ref{thm:maincs}]
Take $h \in \ker(A) \setminus \boldmath{0}$. Let index set $S_0$ be the set of indices of $s$
largest entries (by modulus) of $h$. Let index sets $S_1, S_2, \cdots$ be index sets
corresponding to the next $s$ to $2s$, $2s$ to $3s$, $\cdots$ largest entries of $h$.

Since $A$ satisfies the RIP, we have
\begin{align}
\|h_{S_0}\|^2 & \leq \frac{1}{1-\delta_s} \|Ah_{S_0}\|^2 \\
& = \frac{1}{1-\delta_s} \sum_{j \geq 1} \langle Ah_{S_0}, A(-h_{S_j}) \rangle 
& \text{(because $h_{S_0} = \sum_{j \geq 1} (-h_{S_j})$)} \\
& \le  \frac{1}{1-\delta_s} \sum_{j \geq 1} \delta_{2s} \|h_{S_0}\| \|h_{S_j}\|
& \text{(by Lemma~\ref{lemma:np})} \\
& \leq \frac{\delta_{2s}}{1 - \delta_s} \|h_{S_0}\| \sum_{j \geq 1} \|h_{S_j}\| \\
\|h_{S_0}\| & \leq \frac{\delta_{2s}}{1 - \delta_s} \sum_{j \geq 1} \|h_{S_j}\| .
\label{eq:pf1}
\end{align}
Note that
\[
\|h_{S_j}\| \leq s^{\frac{1}{2}} \|h_{S_j}\|_\infty \leq s^{-\frac{1}{2}} \|h_{S_{j-1}}\|_1.
\]
We can rewrite (\ref{eq:pf1}) as 
\begin{align}
\|h_{S_0}\| & \leq \frac{\delta_{2s}}{1 - \delta_s} s^{-\frac{1}{2}} \sum_{j \geq 1} \|h_{S_{j-1}}\|_1 \\
& = \frac{\delta_{2s}}{1 - \delta_s} s^{-\frac{1}{2}} \|h\|_1.
\label{eq:lastpf}
\end{align}
Also, by the Cauchy-Schwarz inequality,
\begin{equation}
\|h_{S_0}\|_1 = \sum_{i \in S_0} 1 \times |h_i| \leq \sqrt{\sum_{i \in S_0} 1^2}\sqrt{\sum_{i \in S_0} h_i^2} 
= \sqrt{s} \|h_{S_0}\|.
\label{eq:hs0}
\end{equation}
We have $\delta_{2s} < \frac{1}{3}$ as a condition, so
\begin{equation}
\frac{\delta_{2s}}{1-\delta_s} < \frac{\delta_{2s}}{1-\delta_{2s}} < \frac{1}{2} \quad \text{for $\delta_{2s} < \frac{1}{3}$}.
\label{eq:delta}
\end{equation}
Combining (\ref{eq:lastpf}), (\ref{eq:hs0}), and  (\ref{eq:delta}), we get
\begin{equation}
\|h_{S_0}\|_1 < \frac{1}{2} \|h\|_1.
\label{eq:pfhs0h1}
\end{equation}

Now we show that \eqref{eq:pfhs0h1} is equivalent to $\|h_{S}\|_1 < \|h_{S^C}\|_1$:
\begin{align*}
& & \|h_S\|_1 & < \|h_{S^C}\|_1 \\
& \Leftrightarrow & 2\|h_S\|_1 & < \|h_{S^C}\|_1 + \|h_S\|_1 \\
& \Leftrightarrow & 2\|h_S\|_1 & < \|h\|_1 \\
& \Leftrightarrow & \|h_S\|_1 & < \frac{1}{2} \|h\|_1.
\end{align*}

Thus, we have shown that $\|h_{S_0}\|_1 < \|h_{S^C}\|_1$, which is the Null Space Property and by virtue of Theorem~\ref{thm:NSPimpliesRecovery} our proof is complete.

\end{proof}

Many results in compressive sensing (such as Theorem~\ref{thm:maincs}) can be extended will little extra effort to the case where $x$ is not exactly $s$-sparse, but only approximately $s$-sparse, a property that is sometimes referred to as {\em compressible}. See~\cite{CP08,FoucartRauhut_CSbook} for a detailed discussion.

\subsection{Random matrices and the Restricted Isometry Property}

Theorem~\ref{thm:GaussiansSatisfyNSP} (and its proof) showed conditions under which a random gaussian matrices satisfy the NSP. To show the same for RIP is straightforward with the mathematical machinery we have now developed. 
Indeed,
using Proposition~\ref{proposition:ssparse:gaussianwidth} and Theorem~\ref{thm:Gordon:afterGconcentration}, one can readily show\footnote{Note that the $1\pm \delta$ term in the RIP property corresponds to $(1\pm\eps)^2$ in Gordon's Theorem. Since the RIP is a stronger property when $\delta$ is smaller, one can simply use $\eps = \frac13 \delta$.} that matrices with Gaussian entries satisfy the RIP with $m \approx s \log\left( \frac{p}{s} \right)$.

\begin{theorem}\label{thm:RIPmatricesGaussianEntries}
 Let $A$ be an $m\times p$ matrix with i.i.d. standard Gaussian entries, there exists a constant $C$ such that, if
 \begin{equation}\label{condgauss}
  m \geq Cs\log\left(\frac{p}s\right),
 \end{equation}
 then $\frac1{\sqrt{m}}A$ satisfies the $\left(s,\frac14\right)$-RIP with high probability. 
\end{theorem}

We point out an important aspect in this context. Theorems~\ref{thm:GaussiansSatisfyNSP} and~\ref{thm:RIPmatricesGaussianEntries} combined with Theorem~\ref{thm:maincs} yield a {\em uniform recovery guarantee} for sparse vectors with Gaussian sensing matrices. Once a Gaussian matrix satisfies the RIP or NSP (which it will for certain parameters with high probability), then exact recovery via $\ell_1$-minimization holds uniformly for {\em all} sufficiently sparse vectors.

\begin{figure}[h]
\begin{center}
\includegraphics[height=60mm]{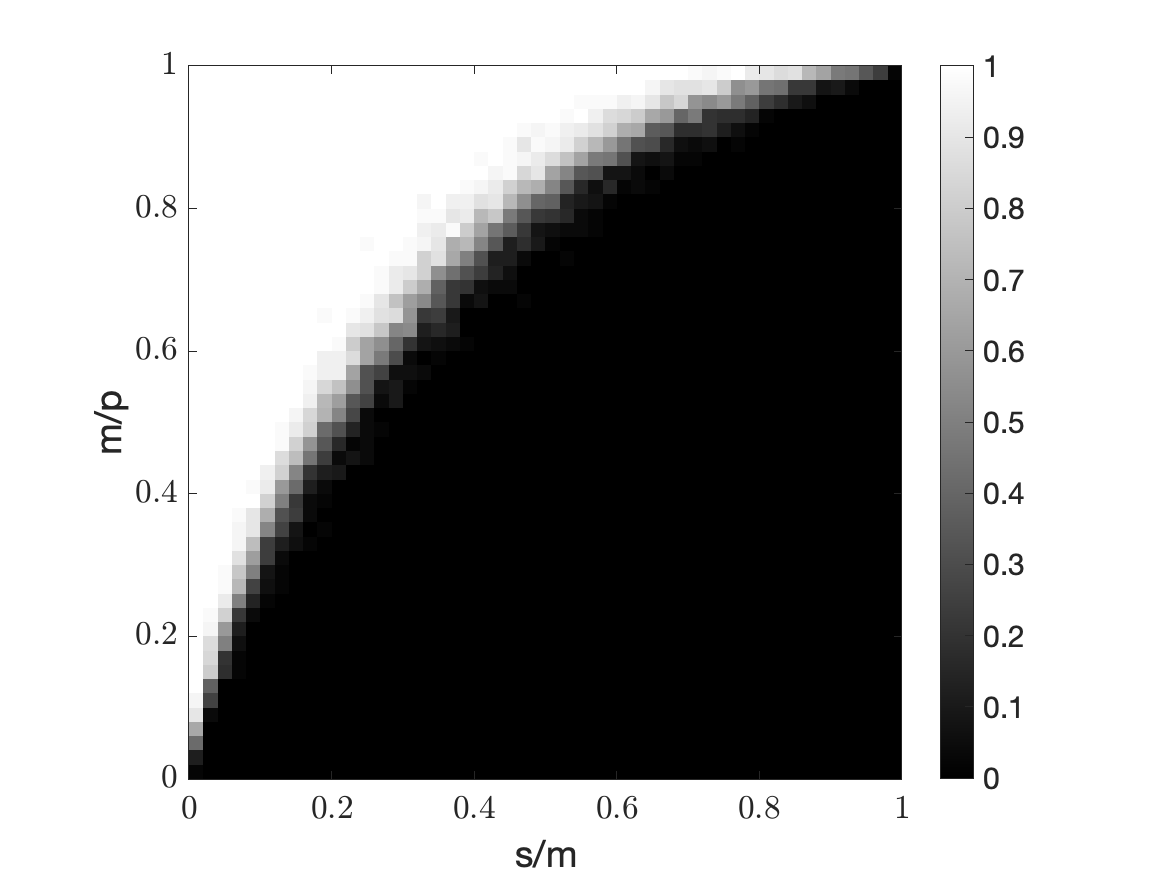}
\caption{The phase transition phenomenon in compressive sensing: if the number of measurements is above a certain threshold, $\ell_1$ minimization almost always succeeds in finding the sparsest solution, while it almost always fails if the number of  measurements is below that threshold. We plot the empirical
probability for different sparsity levels and different number of measurements at fixed ambient dimension, where  black represents 100\% failure and white represents 100\% success. Note the sharp transition between the two extremes. }
\label{fig:phaseplot}
\end{center}
\end{figure}

Figure~\ref{fig:phaseplot} illustrates  the phase transition phenomenon in compressive sensing: 
The plot shows that for a given sparsity level $s$, $\ell_1$ minimization almost always succeeds in finding the sparsest solution if the number of measurements is above a certain threshold, while it almost always fails if the number of  measurements is below that threshold. The very sharp transition between failure and success visible in the plot is supported by theoretical analysis~\cite{donoho2009observed,amelunxen2014living}.
The sensing matrix in this experiment is a Gaussian random matrix of dimension $m \times p$ where the  ambient dimension $p=100$ is fixed. The number of measurements, $m$ varies from 1 to 100, and the sparsity $s$ varies from 1 to $m$. For each choice of $s$ and $m$ we construct a sparse vector $x$ with non-zero random coefficients at  $s$ locations chosen uniformly at random from $1,\dots,100$. We compute $y=Ax$ and solve for $x$ via~\eqref{l1min}.
For  each choice of $s$ and $m$ this experiment  (randomly chosen $A$ and $x$) is repeated 50 times. We plot the empirical rate of success (here, success means that the relative reconstruction error is less than $10^{-5}$), where black means complete failure and white means complete success. A detailed analysis of this phase transition phenomenon can be found in~\cite{donoho2009observed,amelunxen2014living}.

While there are obvious similarities between  Johnson-Lindenstrauss projections and sensing matrices that satisfy the RIP, there are also important differences.\footnote{Interestingly, there are known relationships between the two objects~\cite{krahmer2011new}.}
We note that for JL dimension reduction to be applicable (an upper estimate of ) the number of vectors must be known a priori (and this number if finite).  JL projection preserves (up to $\epsilon$) pairwise distances between these vectors, but the vectors do not have to be sparse. As a consequence, JL  projections $P$ are, in general, a one-way street, as in general
 one cannot  recover $x$ from $y=Px$.
In contrast, a matrix that satisfies the RIP works for infinitely many vectors, however with the caveat that these vectors  must be sparse.  Moreover,  one can recover such sparse vectors $x$ from $y=Ax$ (and can do so numerically efficiently). 

As a consequence of these considerations, a matrix that satisfies the RIP does not necessarily have to satisfy the Johnson-Lindenstrauss Lemma.
While a Gaussian random matrix does indeed satisfy both, RIP and the Johnson-Lindenstrauss Lemma, other matrices do not satisfy both simultaneously. For example, take a randomly subsampled Fourier matrix $A$ of dimensions $m \times p$. 
In the notation of the definition of the Fast Johnson-Lindenstrauss transform, this matrix $A$ would correspond to
$A= SF$, but without the diagonal matrix $D$ that randomizes phases (or signs) of $x$. This matrix $A$ will not meet the Johnson-Lindenstrauss properties of Theorem~\ref{JL_random_0}. But the absence of the phase randomization matrix $D$ is not a hurdle for $A=SF$ to satisfy the RIP under appropriate conditions 
on the matrix dimensions.

Indeed, it is known~\cite{Candes_CS4} that if $m=\Omega_\delta(s\operatorname{polylog} p)$, then the partial Fourier matrix satisfies the RIP with high probability. The exact number of logarithmic factors needed is the object of much research with the best known upper bound due to Haviv and Regev~\cite{Haviv_Regev_2016}, giving an upper bound of $m=\Omega_\delta(s\log^2s\log p)$. On the side of lower bounds it is know that the asymptotics established for Gaussian matrices of $m=\Theta_\delta(s\log (p/s))$ are not achievable in general~\cite{Bandeira_UncertaintyPrinciple}.

Checking whether a matrix satisfies the RIP or not is in general NP-hard~\cite{Bandeira_etal_hardRIP,Tillmann_hardRIP}. While Theorem~\ref{thm:RIPmatricesGaussianEntries} suggests that RIP matrices are abundant for $s  \approx \frac{m}{\log\left( p \right)}$, it appears to be very difficult to deterministically construct matrices that satisfy RIP for $s\gg \sqrt{m}$, known as the square bottleneck~\cite{Tao_blog_deterministicRIP,Bandeira_RoadRIP,Bandeira_DerandomizingRIP,Bandeira_ConditionalPaley,JBourgain_etal_2011_RIP,Dustin_bookchapter_RIP}. The only known unconditional construction that is able to break this bottleneck is due to Bourgain et al.~\cite{JBourgain_etal_2011_RIP}; their construction achieves $s \approx m^{\frac12+\eps}$ for a small, but positive, $\eps$. There is also a conditional construction, based on the Paley Equiangular Tight Frame~\cite{Bandeira_RoadRIP,Bandeira_ConditionalPaley}.

In Section~\ref{s:coherence} we will consider more practical conditions for designing sensing matrices. These conditions, which are better suited for applications, are based on the concept of the  {\em coherence} of a matrix. Interestingly, the  phase randomization of $x$ that is notably absent in the partial Fourier matrix mentioned above, will reappear in this context in connection with {\em nonuniform recovery guarantees}.

\begin{remark}
If one is interested in understanding the probability of exact recovery of a specific sparse vector, and not a uniform guarantee on all sparse vectors simultaneously, then it is possible to do a more refined version of the arguments above that are able to predict the exact asymptotics of the number of measurements required; see~\cite{chandrasekaran2012convex} for an approach based on Gaussian widths and~\cite{amelunxen2014living} for an approach based on Integral Geometry.
\end{remark}

\section{Duality and exact recovery}

In this section we describe yet another approach to show exact recovery of sparse vectors via~\eqref{eq:6:L1normmin}. In this section we take an approach based on duality, the same strategy we took in Chapter~\ref{c:community} to show exact recovery in the Stochastic Block Model. The approach presented here is essentially the same as the one followed in~\cite{Fuchs_sparserepresentations} for the real case, and in~\cite{Tropp_shortcomplexlinearcombinations} for the complex case.

In Chapter~\ref{c:optimization} we derived the dual of a Linear Program using the Lagrangian formulation (we recommend the reader to work out the alternative derivation using the game theoretic approach used in Chapter~\ref{c:community} (there it was used to derived the dual of a Semidefinite Program). The dual to the Linear Program~\eqref{eq:6:L1normmin_LP} is given by:

\begin{equation}\label{eq:6:L1normmin_LP_Dual}
\begin{array}{cl}
\max_u & u^Ty \\
s.t. & -\1 \leq A^Tu   \leq  \1.
\end{array}
\end{equation}

Weak duality (the fact that $\eqref{eq:6:L1normmin_LP_Dual}\leq\eqref{eq:6:L1normmin_LP}$) is, as usual, easy to verify: 
If $\omega^{-}$ and $\omega^{+}$ are primal feasible and $u$ is dual feasible, then
\begin{eqnarray}\label{eq:6:weakduality}
0 & \leq & \left( \1^T - u^TA \right) \omega^{+} +  \left( \1^T + u^TA \right) \omega^{-} \label{eq:6:weakduality} \\  &= & 1^T\left( \omega^{+} + \omega^{-} \right) - u^T\left[ A\left( \omega^{+} - \omega^{-} \right)\right] = 1^T\left( \omega^{+} + \omega^{-} \right) - u^Ty, \nonumber
\end{eqnarray}
showing that $\eqref{eq:6:L1normmin_LP_Dual}\leq\eqref{eq:6:L1normmin_LP}$. 
Furthermore, we know from Chapter~\ref{s:duality} that strong duality holds, and it guarantees that the optimal values of the two programs actually match.

\subsection{Finding a dual certificate}

In order to show that $\omega^{+} - \omega^{-} = x$ is an optimal solution\footnote{For now we will focus on showing that it is an optimal solution, see Remark~\ref{remark:6:uniquenessofCSsol} for a brief discussion of how to strengthen the argument to show uniqueness.} to~\eqref{eq:6:L1normmin_LP}, we will find a dual feasible point $u$ for which the dual matches the value of $\omega^{+} - \omega^{-} = x$ in the primal, $u$ is known as a \emph{dual certificate} that we introduced in Chapter~\ref{ss:KKT}.

From~\eqref{eq:6:weakduality} it is clear that $u$ must satisfy 
$$\left( \1^T - u^TA \right) \omega^{+} =0 \quad \text{and} \quad \left( \1^T + u^TA \right) \omega^{-} = 0,$$ the \emph{complementary slackness} condition from Chapter~\ref{ss:KKT}. This means that we must take the entries of $A^Tu$ be $+1$ or $-1$ when $x$ is non-zero (and be $+1$ when it is positive and $-1$ when it is negative), in other words
\[
\left( A^Tu\right)_S = \sign\left(x_S\right),
\]
where $S = \supp(x)$, and $\left\| A^Tu\right\|_\infty \leq 1$ (in order to be dual feasible).

\begin{remark}\label{remark:6:uniquenessofCSsol}
It is not difficult to see that if we further ask $\left\| \left( A^Tu\right)_{S^c}\right\|_\infty < 1$, any optimal primal solution would have to have its support contained in the support of $x$. Indeed,  extending Theorem~\ref{th:kkt2} to the case when the objective $f$ is not differentiable but has a subdifferential (as is the case for $f(x) = \|x\|_1$), and then using the properties of the subdifferential for the $\ell_1$-norm that we computed in Chapter~\ref{ss:subgradient},
gives the following lemma.
\end{remark}

\begin{lemma}\label{le:csdual}
Consider the problem of sparse recovery discussed above. Let $S = \supp(x)$, if $A_S$ is injective and there exists $u\in\RR^{M}$ such that
\[
\left( A^Tu\right)_S = \sign\left(x_S\right),
\]
and
\[
\left\| \left( A^Tu\right)_{S^c}\right\|_\infty < 1,
\]
then $x$ is the unique optimal solution to the $\ell_1$-minimization problem~\eqref{eq:6:L1normmin}.
\end{lemma}

Since we know that $\left( A^Tu\right)_S = \sign\left(x_S\right)$ (and that $A_S$ is injective), we try to construct\footnote{Note how this differs from the situation in Chapter~\ref{c:community} where the linear inequalities were enough to determine a unique candidate for a dual certificate.} $u$ by least squares and hope that it satisfies $\left\| \left( A^Tu\right)_{S^c}\right\|_\infty < 1$. More precisely, we take
\[
u = \left( A^T_S \right)^\dagger \sign\left(x_S\right),
\]
where $\left( A^T_S \right)^\dagger = A_S \left( A_S^TA_S \right)^{-1}$ is the Moore-Penrose pseudo-inverse of $A^T_S$. This gives the following corollary.

\begin{corollary}\label{cor:6:dualcertificate}
Consider the problem of sparse recovery.  Let $S = \supp(x)$. If $A_S$ is injective and
\[
\left\| A_{S^c}^T A_S \left( A_S^TA_S \right)^{-1} \sign\left(x_S\right) \right\|_\infty < 1,
\]
then $x$ is the unique optimal solution to the $\ell_1$-minimization problem~\eqref{eq:6:L1normmin}.
\end{corollary}

Theorem~\ref{thm:RIPmatricesGaussianEntries} establishes that if $m \geq Cs\log\left(\frac{p}s\right)$, for a universal constant $C$, and $A \in \RR^{m\times p}$ is drawn with i.i.d. Gaussian entries $\NNN\left(0,\frac1m\right)$ then\footnote{Note that the normalization here is taken slightly differently: entries are normalized by $\frac1{\sqrt{m}}$, rather than $\frac{1}{a_m}$, but the difference is negligible for our purposes.} it will, with high probability, satisfy the $(s,1/3)$-RIP. Note that, if $A$ satisfies the $(s,1/3)$-RIP then, for any $|S|\leq s$ one has $\|A_S\| \leq \sqrt{1+\frac13}$ and $\|\left(A_S^TA_S\right)^{-1}\| \leq \left(1-\frac13\right)^{-1} = \frac32$, where $\|\cdot\|$ denotes the operator norm $\|B\| = \max_{\|x\|=1}\|Bx\|$.

This means that if we take $A$ random with i.i.d.\ $\NNN\left(0,\frac1m\right)$ entries then, for any $|S|\leq s$ we have that
\[
\| A_S \left( A_S^TA_S \right)^{-1} \sign\left(x_S\right) \| \leq \sqrt{1+\frac13} \frac32\sqrt{s} = \sqrt{3}\sqrt{s},
\]
and because of the independency among the entries of $A$, $A_{S^c}$ is independent of this vector and so for each $j\in S^c$ we have
\[
\Prob \left(  \left| A_{j}^T A_S \left( A_S^TA_S \right)^{-1} \sign\left(x_S\right) \right| \geq \frac1{\sqrt{M}}\sqrt{3}\sqrt{s} t \right)  \leq 2\exp\left(-\frac{t^2}2\right),
\]
where $A_j$ is the $j$-th column of $A$.

An application of the union bound gives
\[
\Prob \left(  \left\| A_{S^c}^T A_S \left( A_S^TA_S \right)^{-1} \sign\left(x_S\right) \right\|_\infty \geq \frac1{\sqrt{M}}\sqrt{3}\sqrt{s} t  \right)  \leq 2N\exp\left(-\frac{t^2}2\right),
\]
which implies
\begin{eqnarray*}
\Prob \left(  \left\| A_{S^c}^T A_S \left( A_S^TA_S \right)^{-1} \sign\left(x_S\right) \right\|_\infty \geq 1  \right) & \leq & 2p\exp\left(-\frac{\left( \frac{\sqrt{m}}{\sqrt{3s}} \right)^2}2\right) \\ &=& \exp\left(- \frac12\left[ \frac{m}{3s} - 2\log(2p) \right] \right),
\end{eqnarray*}
which means that we expect to exactly recover $x$ via $\ell_1$ minimization when $m \gg s\log(p)$. While this can be asymptotically worse then the bound of $m \gtrsim s\log\left(\frac{p}s\right)$, and this guarantee is not uniformly obtained for all sparse vectors, the technique in this section is generalizable to several circumstances and illustrates the flexibility of approaches based in construction of dual witnesses. We will discuss nonuniform guarantees in more detail in the next section.

\section{From theory to practice} \label{s:cs_practice0}

\subsection{Sensing matrices and incoherence}\label{s:coherence}

In applications, we usually cannot completely freely choose the sensing matrix to our liking. This means that Gaussian random matrices play an important role as benchmark, but from a practical viewpoint they play a marginal role. Clearly, randomness in the sensing matrix seems to be very beneficial for compressive sensing. However, in practice, there are numerous design constraints on the sensing matrix $A$, as in many applications one only has access to structured measurement systems. For example,  we may still have the freedom to choose, say the positions of the antennas in radar systems that employ multiple antennas~\cite{strohmer2014analysis,strohmer2013accurate}, the position of sensors in MRI~\cite{lustig2007sparse,Adcock-Hansen-book,block2007undersampled,ni2024auto}, or the sampling locations in digital signal acquisition~\cite{mishali2010theory}.
By choosing these randomly, we can still introduce randomness in our system. Or, we can transmit random waveforms in sonar and radar systems~\cite{herman2009high,potter2010sparsity}. Yet, in all these cases the overall structure of $A$ is still dictated by the physics of wave propagation. In other applications, it will be other physical constraints or design limitations that will dictate how much randomness we can introduce into the sensing matrix.

While establishing the RIP (with high probability) for Gaussian or Bernoulli random matrices is not too difficult, it is already significantly harder to do so for the partial Fourier matrix~\cite{Candes_CS4,rudelson2008sparse,Haviv_Regev_2016} and time-frequency matrices~\cite{dorsch2017refined}, and even harder for more specific sensing matrices.

A useful concept to overcome the practical limitations of the RIP is via the concept of the (in)coherence of a matrix. This concept has proven to be widely applicable in practice.
While we want to avoid the constraints of the RIP, we nevertheless take it as our point of departure.
Recall that the RIP (Definition~\ref{def:6:RIP_2}) asks that any $S\subset [p]$, $|S|\leq s$ satisfies:
\[
(1-\delta) \|x\|^2 \leq \left\| A_Sx\right\|^2 \leq (1+\delta) \|x\|^2,
\] 
for all $x\in \RR^{|S|}$. This is equivalent to
\[
\max_x \frac{x^T\left( A_S^TA_S - I \right)x}{x^Tx} \leq \delta,
\]
or equivalently
\[
\left\| A_S^TA_S - I \right\| \leq \delta.
\]

If the columns of $A$ are unit-norm vectors (in $\RR^m$), then the diagonal of $A_S^TA_S$ is all-ones, this means that $A_S^TA_S - I$ consists only of the non-diagonal elements of $A_S^TA_S$. If, moreover, for any two columns $a_i$, $a_j$, of $A$ we have $\left| a_i^Ta_j\right| \leq \mu$ for some $\mu$ then, Gershgorin's circle theorem tells us that $\left\| A_S^TA_S - I \right\| \leq \mu (s-1)$.

More precisely, given a symmetric matrix $B$, the Gershgorin's Circle Theorem~\cite{HJ90} states that all of the eigenvalues of $B$ are contained in the so called Gershgorin discs (for each $i$, the Gershgorin disc corresponds to 
$$\Big\{ \lambda : |\lambda - B_{ii}| \leq \sum_{j\neq i}\left|B_{ij}\right| \Big\}.$$ If $B$ has zero diagonal, then this reads: $\|B\| \leq \max_i \sum_{j\neq i} \left|B_{ij}\right|$.

Given a set of $p$ unit-norm vectors $a_1,\dots,a_p\in\RR^m$ we define its worst-case coherence $\mu$ as
\begin{equation}\label{coherence}
\mu(a_1,\dots,a_p) = \max_{i\neq j} \left| \langle a_i, a_j \rangle  \right|.
\end{equation}

Assume now the vectors $a_1,\dots,a_p\in\RR^m$ have worst-case coherence $\mu$. If we form a matrix with these vectors as columns, then it will satisfy the $\left( s , \mu(s-1) \right)$-RIP, implying that it will satisfy the $\left( s , \frac13 \right)$- RIP for $s  \leq \frac13\frac{1}\mu$.

This motivates the problem of designing sets of vectors $a_1,\dots,a_p\in\RR^m$ with smallest possible worst-case coherence. This is a central problem in Frame Theory~\cite{SH03,Chr03,Casazza_Kutyniok_2012,waldron2018introduction}.
Recall that in finite dimensions, a set of vectors $a_1,\dots,a_p \in\Hsp^m$ (where $\Hsp = \R$ or $\C$) is called a frame for $\Hsp^m$ if 
there exist constants (frame bounds) $0<A\le B < \infty$ such that\footnote{Since here we are dealing with a finite dimensional Hilbert space, the upper frame bound is always trivially fulfilled, and the lower bound is equivalent to the the frame forming a spanning set.}
\begin{equation}\label{framedef}
    A \|x\|^2 \le \sum_{k=1}^p | \langle x,a_k \rangle |^2 \le B \|x\|^2
\end{equation}
for every $x\in \Hsp^m$. 
The associated {\em frame operator} $S$ is defined by
\begin{equation} \label{frameop}
Sx = \sum_{i=1}^p \langle x,a_i \rangle a_i.
\end{equation}
The frame definition~\eqref{framedef} can be equivalently expressed as 
requiring that $A I \preccurlyeq S \preccurlyeq B I$ holds, where $\preccurlyeq$ denotes the positive-semidefinite order. 
A frame is called {\em tight} if $A=B$, in which case
$S=AI$. 
If $\|a_i\| = 1$ for all $i=1,\dots,p$ then $\{a_k\}_{i=1}^p$ is called a {\em unit norm frame}. We call a unit norm frame $\{a_k\}_{i=1}^p$ {\em equiangular} if
\begin{equation}
\label{defequi}
|\langle a_i,a_j \rangle| = c \quad 
\text{for all $i,j$ with $i \neq j,$}
\end{equation}
for some constant $c\ge 0$.
Obviously, any orthonormal basis is equiangular.

We can now provide an elucidating answer to the question of a lower bound on the coherence of a set of vectors $a_1,\dots,a_p$ via the following theorem from frame theory.
\begin{theorem}[\cite{SH03}]
\label{th:bound}
Let $\{a_k\}_{k=1}^p$ be a unit-norm frame for $\Hsp^m$, where $\Hsp = \R$ or $\C$ Then
\begin{equation}
\mu(a_1,\dots,a_p) \ge \sqrt{\frac{p-m}{m(p-1)}}.
\label{bound1}
\end{equation}
Equality holds in~\eqref{bound1} if and only if $\{a_k\}_{k=1}^p$ is an equiangular tight frame.

\end{theorem}

\begin{proof}
A simple calculation shows that 
\begin{equation*}
\trace (S) = \|S^{\frac{1}{2}}\|_F^2 =\sum_{k=1}^{p} \| a_k\|^2 = p, \,\, \text{and}\,
\trace (S^2) = \|S\|_F^2 =  \sum_{k,l=1}^{p}  |\langle a_k, a_l \rangle|^2,
\label{trace}
\end{equation*}
Let $\lambda_1 \ge   \dots \ge \lambda_m$ be the
eigenvalues of $S$. Let ${\textbf 1}$ denote the all-one vector.
By the Cauchy-Schwarz inequality
\begin{equation*}
\trace (S)^2 = \Big(\sum_{k=1}^{m} \lambda_k \Big)^2 =
\langle {\textbf 1} , \{\lambda_k\} \rangle
\le \| {\textbf 1}\|^2 \| \{\lambda_k\}\|^2 = 
m \sum_{k=1}^{m} \lambda_k^2 = m \trace S^2,
\label{trace1}
\end{equation*}
which can be stated equivalently as
\begin{equation}
  \label{frameforce}  
\sum_{k}^{p} \sum_{l=1}^{p} |\langle a_k, a_l \rangle|^2 \ge \frac{p^2}{m}.
\end{equation}
Equality holds in~\eqref{frameforce} if and only if $\lambda_1 = \dots = \lambda_p = p/m$, i.e., if and only if $\{a_k\}_{k=1}^p$ is a tight frame

We now consider
\begin{equation}
\label{framebound1}
\underset{k\neq l}\max  |\langle a_k, a_l \rangle|^2
\ge \frac{1}{p^2 - p}\sum_{k \neq l} |\langle a_k, a_l \rangle|^2 = 
\frac{1}{p^2 - p} \Big( \sum_{k =1}^p\sum_{l =1}^p |\langle a_k, a_l \rangle|^2 - p \Big),
\end{equation}
with equality if and only if $\{a_k\}_{k=1}^p$ is equiangular. By the bound~\eqref{frameforce} we have
\begin{equation}
\label{framebound2}
\frac{1}{p^2 - p} \Big( \sum_{k =1}^p\sum_{l =1}^p |\langle a_k, a_l \rangle|^2 - p \Big) \ge
\frac{1}{p^2 - p} \Big( \frac{p^2}{m} - p \Big)
= \frac{p-m}{m(p-1)},
\end{equation}
with equality if and only if $\{a_k\}_{k=1}^p$ is a tight frame. Inequalities~\eqref{framebound1} and~\eqref{framebound2} together give~\eqref{bound1} and the proof is complete.

\end{proof}

We note that if $\Hsp=\R$ equality in~\eqref{bound1} can only hold if  $p \le m(m+1)/2$, while if $\Hsp=\C$ then equality in~\eqref{bound1} can only hold if $p \le m^2$. This can be shown by a so-called tensoring trick (we invite the reader to try): by analyzing the gram matrix of the vectors $\{a_k\otimes \overline{a_k}\}_{k=1}^p$ (the $m(m+1)/2$ and $m^2$ bounds then appear by the dimension of the subspace in which these vectors lie), see Table II of~\cite{DGS75}.

The coherence bound~\eqref{bound1} of a set of $p$ unit-norm vectors in $m$ dimensions is also known as Welch bound~\cite{welch2003lower}. Due to this limitation implied by Theorem~\ref{th:bound}, deterministic constructions based on coherence cannot yield matrices that satisfy the RIP for $s\gg \sqrt{m}$, known as the square-root bottleneck~\cite{Bandeira_RoadRIP,Tao_blog_deterministicRIP,JBourgain_etal_2011_RIP,Bandeira_ConditionalPaley}.

However, as we will see in Section~\ref{s:cs_practice0}, if we are willing to accept slightly weaker recovery guarantees than provided by the RIP, matrices with low coherence do give rise to sensing matrices that come with appealing theoretical guarantees while also being useful in practice.

Theorem~\ref{th:bound} suggests that in order to design sensing vectors of minimal coherence, we should  look for equiangular tight frames (ETFs). The existence and construction of ETFs is a rather challenging problem that intersects with a wide range of areas such as harmonic analsyis, algebraic, combinatorics, finite geometry, group theory and representation theory, and coding theory~\cite{SH03,waldron2018introduction}. Applications include signal processing, wireless communications, numerical linear algebra, and quantum physics. Indeed, ETFs are deeply connected to Zauner's conjecture, particularly through their role in quantum information theory and the concept of SIC-POVMs (Symmetric Informationally Complete Positive Operator-Valued Measures)~\cite{zauner1999quantendesigns,Scott_Grassl_SICPOVM,appleby2019tight}.

Another interesting and very useful construction of sensing vectors with low coherence is given by so-called {\em mutually unbiased bases}~\cite{wootters1989optimal,durt2010mutually,bandyopadhyay2002new,calderbank1997,SH03}. These bases are widely used in quantum information theory as well as in radar and signal processing. Two orthonormal bases $\{ \phi_j \}_{j=0}^{m-1}, \{ \psi_j \}_{j=0}^{m-1} \in \C^m$ are called {\em mutually unbiased}
if and only if 
\begin{equation}\label{MUB}
\forall i,j \,\, | \langle \phi_i, \psi_j \rangle | = \frac{1}{\sqrt{m}}.
\end{equation}
The classical example of an MUB consists of the identity matrix and the Discrete Fourier transform matrix concatenated into a $m\times 2m$ matrix. But, perhaps surprisingly, for certain values of $m$, there exist MUBs in $\C^m$ consisting of $m+1$ orthonormal bases $\{ \phi^{(k)}_j \}_{j=0}^{m-1}, k=0,\dots,m$ such that condition~\eqref{MUB} is satisfied for any two orthonormal bases
$\{\phi^{(i)}_j\}_{j=0}^{m-1}, \{\phi^{(k)}_j\}_{j=0}^{m-1}$, 
see for example~\cite{wootters1989optimal,bandyopadhyay2002new,calderbank1997,SH03,alltop1980complex}.
Thus, MUBs give rise to structured sensing matrices of dimension $m \times m^2$ with coherence $\mu = 1/\sqrt{m}$, which thus almost achieve the Welch bound.
Here is a simple example for an MUB in $\C^m$: Let $m$ be a prime number $\ge 5$ and set $g(j) = \frac{1}{\sqrt{m}}e^{2\pi i j^3/m}$
for $j=0,\dots,m-1$. Then the vectors $\{g_{k,l}\}_{k,l=0}^{m-1}$ defined via
\begin{equation}
g_{k,l}(j) = g(j-k) e^{2\pi i lj/m}, \quad k,l=0,\dots,m-1,
\label{fingabor}
\end{equation}
satisfy $|\langle g_{k,l}, g_{k',l'} \rangle| \in \{0,1/\sqrt{m}\}$
for all $g_{k,l} \neq g_{k',l'}$, which follows from basic properties of quadratic Gauss sums (cf.~Theorem~2 in~\cite{alltop1980complex}).
We can add the $m \times m$ identity matrix and end up with $m^2+m$ vectors (split into $m+1$ ONBs) in $\C^m$ which form a MUB of maximal size.
For other, nonequivalent constructions, of MUBs (which, however, still revolve around translations and modulations in form of the finite Heisenberg group) 
see~\cite{calderbank1997}.
Remarkably, for $m$ not a power of a prime number it is not known how many MUB exist; this is an open problem even for $m=6$~\cite{RandomstrasseProblems2024}.
For $\R^m$ the construction and existence of MUBs faces more constraints. Even the existence of two bases is equivalent to the existence of so-called Hadamard matrices ($m\times m$ $\pm1$ matrices with orthogonal columns) for which the existence for dimensions that are a multiple of $4$ is still a (celebrated) open question (see, e.g.~\cite{RandomstrasseProblems2024}).

\subsection{Nonuniform recovery guarantees and coherence} \label{s:cs_practice0}

As mentioned before, it follows from Theorem~\ref{th:bound} that deterministic constructions based on coherence cannot yield matrices that satisfy the RIP for $s\gg \sqrt{M}$, known as the square-root bottleneck~\cite{Bandeira_RoadRIP,Tao_blog_deterministicRIP}.
To overcome this square root bottleneck it seems that something has to give. One fruitful direction is to sacrifice the {\em uniform recovery} granted by the RIP. Namely, once a matrix satisfies the RIP, it is guaranteed that the solution to the $\ell_1$-optimization problem is identical to the solution to the $\ell_0$-optimization problem (the sparsest solution) for {\em all} $s$-sparse vectors. In contrast, we can consider scenarios in which   $\ell_0$-optimization and $\ell_1$ optimization having the same solution holds ``only'' for {\em most} $s$-sparse vectors. This leads to {\em nonuniform recovery} results, which we will pursue below. The benefits however are often worth the sacrifice, since we end up with theoretical guarantees that are much more practical.

Recall that we consider a general linear system of equations $A x = y$, where $A \in \C^{m \times p}$, $x \in \C^p$
and $m \le p$. We introduce the following generic $s$-sparse model:
\begin{itemize}
\setlength{\parsep}{-0.3ex}
\item[(i)] The support $S \subset \{1,\dots,p\}$ of the $s$ nonzero
coefficients of $x$ is selected uniformly at random.
\item[(ii)] The non-zero entries of $\sign(x)$ form a Steinhaus sequence, i.e.,
$\sign(x_k):=x_k/|x_k|, k\in I_s,$ is a complex random variable that is
uniformly distributed on the unit circle, and independent from each other (and from $I_s$).
\end{itemize}

As an example for a theoretical nonuniform  guarantee using coherence-based sensing matrices  we state (without proof) the following theorem, and refer to~\cite{FoucartRauhut_CSbook} for a proof.

\begin{theorem}[Theorem 14.5 in~\cite{FoucartRauhut_CSbook}]\label{th:coherence}
Let $A \in \C^{m \times p}, m \le p$, be a matrix with $\ell_2$-normalized columns and coherence $\mu$. Let $S$ be a subset of $\{1,\dots,p\}$ selected at random according to the uniform model with $\card(S) = s$. Let $x \in \C^p$ be a vector supported on $S$ for which $\sign(x_S)$
is a Steinhaus sequence independent of $S$. Assume that, for $\eta,\eps \in (0,1)$
\begin{align}
  \eta & \le  \frac{c}{\ln(p/\eps)} \label{nonuniform_cond1}  \\
  \frac{s}{p} \|A\|^2 & \le  \frac{c}{\ln(p/\eps)}  \label{nonuniform_cond2}
\end{align}
for an appropriate constant $c > 0$. Then, with probability at least, the
vector $x$ is the unique minimizer of $\|z\|_1$  subject to  $Az = Ax$.
\end{theorem}

A simple calculation shows that for a unit-norm tight frame the
condition~\ref{nonuniform_cond1} is satisfied if $$m \ge Cs \ln(p/\eps),$$
which is in the ballpark of condition~\eqref{condgauss} for Gaussian sensing matrices.
Theorem~\ref{th:coherence} demonstrates that by allowing for nonuniform recovery guarantees one can successfully reconstruct sparse signals under
much milder coherence conditions than imposed by the aforementioned square root bottleneck. That being said, establishing that the support set of real-world signals can be modeled as random is, in general, difficult to justify rigorously. For instance, the wavelet coefficients of natural images are typically concentrated along edges, giving rise to structured, tree-like patterns in the locations of significant coefficients. This insight is utilized in practical compressive sensing systems as, for example, in MRI~\cite{lustig2007sparse,Adcock-Hansen-book}.
The fact that the support set of many natural signals does not conform to a uniform distribution highlights the relevance of the recovery results established in Sections~\ref{s:sparse}--\ref{s:rip}, which hold for {\em all} sufficiently sparse signals.
 Furthermore, as pointed out in~\cite{FoucartRauhut_CSbook}, the stability guarantees currently available for models based on random supports are notably weaker than those derived from the restricted isometry property.

Various other versions of {\em nonuniform recovery} results can be found e.g., in~\cite{Tropp_Dictionaries,CP08,FoucartRauhut_CSbook}. See~\cite{strohmer2014analysis,hugel2014remote,Adcock-Hansen-book} for some theoretical results geared towards applications.

\subsection{Further practical considerations} \label{s:cs_practice}

When moving the ideas of compressive sensing to move from theory into practice we encounter some challenges: (i) the choice of the sensing matrix is often limited by practical considerations and physical constraints (both usually rule out a Gaussian sensing matrix); (ii) signals are often not sparse in the measurement domain, but only with respect to some properly chosen transform (and even then, signals are often only approximately sparse);  (iii) for various reasons (speed being one of them) we may need alternative sparse solvers instead of vanilla $\ell_1$-minimization. We already have  discussed~(i) earlier in this section and will 
discuss items (ii) and (iii) succinctly below.  We recommend~\cite{Adcock-Hansen-book} for a detailed treatment of how to span the gap between theory and practice of compressive sensing.

As indicated above, signals are often not sparse in the canonical basis, but they are (approximately) sparse after a suitable transformation. For examples audio signals are sparse with respect to a localized Fourier transform such as a Gabor transform~\cite{daudet2006sparse} or some other local trigonometric transform~\cite{malvar1992signal,mallat_book}. 
JPEG exploits the sparsity of images in the discrete cosine basis or wavelet basis~\cite{mallat_book} while curvelet transforms are effective sparsifiers for images arising in astronomy~\cite{starck2003astronomical}. Radar signal are sparse when (properly) transformed into the time-frequency domain~\cite{herman2009high}. 
Some signals may even require a redundant dictionary (perhaps consisting of a combination of a wavelet basis, a curvelet basis, and a Fourier basis), see 
e.g.~\cite[Chapter 12]{mallat_book}. 
But how do such transform-sparse signals fit into our framework?

As before, let $x \in \C^p$ be the signal (or image) of interest and we observe $y=Ax$, where $A \in \C^{m \times p}$ is a sensing matrix, where $m < p$. 
We assume that $x$ is sparse with respect to some basis or frame $U \in \C^{p \times n}$, where $n \ge p$ (and $n=p$ if we restrict $U$ to be a basis).
Denote this sparse representation by $w \in \C^n$, i.e.,  $x= Uw$. We write $B:=AU$ and can now solve for $w$ via
\begin{equation} \label{transformsparse}
\min \,\, \|z\|_1 \,\,\, \text{s.t.} \,\,  Bz = y.
\end{equation}
Having solved this problem (assuming it does indeed yield $w$), we can recover $x$ from $w$ by simply computing $x=Uw$.

Thus, we can (try to) apply the theoretical results developed in the previous sections to the matrix $B=AU$ in place of $A$. In a nutshell, the sensing matrix $A$ has to be {\em incoherent} with respect to the transform in which the signal is sparse. An important example arises in MRI. There, the wavelet transform is often chosen as sparsifying transform and the sensing matrix is derived from randomly subsampling the two-dimensional Discrete Fourier Transform or the Radon transform, see~\cite{lustig2007sparse,Adcock-Hansen-book}.
In practice, it has been observed that even better results are obtained when the random subsampling of the Fourier coefficients is combined with a deterministic sampling of the  
Fourier coefficients corresponding the low frequencies, a strategy that has already been suggested in one of the original papers on compressive sensing~\cite{Donoho_CS}. We refer to~\cite{lustig2007sparse,Adcock-Hansen-book} for many more details and modifications. 

\begin{figure}[h]
\begin{center}
\includegraphics[width=\textwidth]{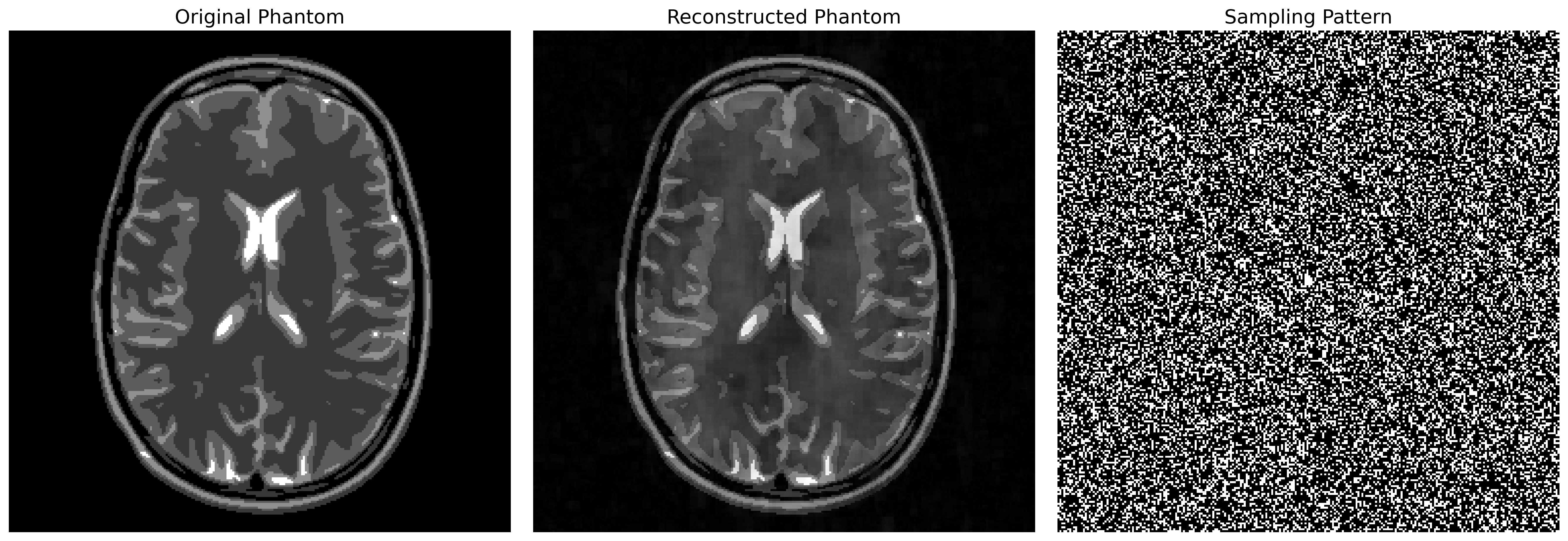}
\caption{The GLPU MRI phantom (left), its reconstruction (middle) via $\ell_1$ minimization from  30\% discrete randomly chosen Fourier measurements  using the wavelet transform as  sparsifying transform $U$. The sampling pattern is depicted on the right where white pixels correspond to sampling locations.}
\label{fig:CSMRI}
\end{center}
\end{figure}

Figure~\ref{fig:CSMRI} shows an example of compressive sensing applied to MRI.  The test image is the so-called GLPU phantom (introduced by Guerquin-Kern, Lejeune, Pruessmann and Unser in~\cite{guerquin2011realistic}). It is a continuous, piecewise constant MRI-type image with an analytic expression for its Fourier transform. 
 The reconstruction of the discrete GLPU phantom from 30\%  randomly chosen discrete Fourier measurements. The wavelet transform with a Haar wavelet has been used as sparsifying transform $U$. The image $x$ is recovered by first solving~\eqref{transformsparse} for $w$, followed by computing $x=Uw$.

\bigskip

There are various other efficient and rigorous methods to recover sparse vectors from underdetermined systems besides $\ell_1$-minimization and the lasso. For example, homotopy methods, greedy algorithms or methods based on approximate message passing. We refer to~\cite{FoucartRauhut_CSbook,Adcock-Hansen-book} for a comprehensive discussion of these techniques. 
Furthermore,~\cite{Adcock-Hansen-book} contains a detailed discussion of  some subtle potential numerical stability issues one should be aware of. 

\section*{Exercises}
\addcontentsline{toc}{section}{Exercises}

\begin{myexercise}[\level\level\sep Sparse vector approximation in $\ell^2$]\label{prob:sparse_vector_approx}
    Let $N \geq s \geq 1$ be integers and $x \in \C^N$ be a vector. Show that there exists an $s$-sparse vector $y \in \C^N$ such that
    \begin{equation*}
        \norm{x - y}_2 \leq \frac{1}{2\sqrt{s}}\norm{x}_1.
    \end{equation*}
\end{myexercise}

\begin{myexercise}[\level\sep $\ell_0$-Minimization Recovery]\label{prob:l0_minimization}
    
    Let $A \in \mathbb{C}^{m\times p}$ be a matrix. Suppose that every $s$-sparse vector $x$ can be uniquely recovered by $A$ via $\|.\|_0$ minimization, i.e, we choose $x^{\ast}$ that minimizes $\|z\|_{0}$ subject to the constraint $Ax=Az$ and $x$ is the unique minimum of this problem if $x$ has at most $s$ nonzero entries.
    \begin{enumerate}
        \item Prove that every $2s$ columns of $A$ are linearly independent.
        
        \item Prove that $m\ge 2s$.
        
        \item Prove that if a matrix $B\in \mathbb{C}^{m\times p}$ satisfies the condition that every $2s$ columns are linearly independent, then every $s$-sparse vector $x$ can be uniquely recovered by $B$ via $\|.\|_0$ minimization.

    \end{enumerate}

\end{myexercise}

\begin{myexercise}[\level\level\sep Stable nullspace property]\label{prob:stable_nullspace_property}
     A fundamental fact in compressed sensing is that in order to recover an $s$-sparse vector $x \in \mathbb{R}^N$ by minimizing the $\ell_1$ norm, the measurement matrix $\Phi \in \mathbb{R}^{d\times N}$ needs to satisfy the null space property. The goal of this exercise is to study the compressed sensing problem when $x$ is approximately sparse.

    We say that a matrix $\Phi \in \mathbb{R}^{d\times N}$ satisfies \emph{the $(s,\rho)$-stable null space property} if for every non-zero vector $v \in \ker(\Phi)$ and all sets $S$ such that $|S|\le s$, the following holds
    \begin{equation*}
        \|v_{S}\|_1 \le \rho \|v_{S^c}\|_1.
    \end{equation*}

    Prove the following facts
    \begin{enumerate}[(a)]
        \item Given a set $S\subset \{1,\ldots,N\}$ and vectors $x,z \in \mathbb{R}^N$,
        \begin{equation*}
            \|(x-z)_{S^c}\|_1 \le \|z\|_1 - \|x\|_1  + 2\|x_{S^c}\|_1 + \|(x-z)_{S}\|_1.
        \end{equation*}
        \item Prove that if $\Phi \in \mathbb{R}^{d\times N}$ satisfies the $(s,\rho)$-stable nullspace property with $\rho \in (0,1)$, then the solution of the optimization program 
        \begin{equation*}
            \hat{x}:= \mathrm{argmin}\, \|z\|_1 \quad \text{subject to} \quad \Phi z=\Phi x
        \end{equation*}
        satisfies 
        \begin{equation*}
            \|\hat{x}-x\|_1 \le 2\sigma_s(x)\frac{1+\rho}{1-\rho},
        \end{equation*}
        where $\sigma_s(x)$ is the $s$-best term approximation error of $x$, given by $\sigma_s(x) \coloneqq \inf_{z:\|z\|_0 \le s} \|x-z\|_1$.
        \item Show that the stable nullspace property with $\rho<1$ is sufficient for exact recovery when the vector $x$ is $s$-sparse.
    \end{enumerate}
\end{myexercise}

\begin{myexercise}
Let $i =\sqrt{-1}$ and set
$$
A=
\begin{bmatrix}
i & 0 & -i \\
0 & i & -i
\end{bmatrix}.
$$
Using the null space property, show that $\ell_1$-minimization will recover any
1-sparse vector $x$, given $Ax = y$.
\end{myexercise}

\begin{myexercise}
On the connection between (in)coherence parameter $\mu$  and restricted isometry constant $\delta_{s}$:
Show that $\delta_1 = 0, \delta_2 = \mu$, and $\delta_s \le (s-1)\mu$.
\end{myexercise}

\begin{myexercise}
Let $A \in \R^{k \times d}$ be a Gaussian random matrix. Give an estimate for the coherence $\mu$ of $A$.
\end{myexercise}

\begin{myexercise}
The proximal operator for the $\ell_1$ norm is defined as
$$\text{prox}_{\lambda \|\cdot\|_1}(y) = \arg\min_{x} \left( \lambda \|x\|_1 + \frac{1}{2}\|x-y\|^2 \right).$$
Derive the closed-form solution (Soft-Thresholding) for
$\text{prox}_{\lambda \|\cdot\|_1}(y)$.
\end{myexercise}

\begin{myexercise}
Consider $y=Ax$, where $A$ is a $100 \times 400$ Gaussian random matrix and $x$ is an $s$-sparse
vector of length 400. The locations of the non-zero entries of $x$ are chosen uniformly at random
and the non-zero coefficients of $x$ are normal-distributed. For $s=1,2,\dots,$ solve
$$\underset{z}{\min} \,\|z\|_1 \quad \text{subject to\,\,} Az=y,$$
(e.g.\ using the Matlab toolbox \texttt{cvx}, see \texttt {cvxr.com/cvx}, or its Python version \texttt{cvxopt}  or its R version \texttt{cvxr}.).
For each fixed $s$ repeat the experiment 10 times.
Create a graph plotting $s$ versus the relative reconstruction error (averaged over the ten experiments
for each $s$).
Starting with which value of $s$ (approximately)
does $\ell_1$-minimization fail to recover $x$?
\end{myexercise}

\begin{myexercise}
Same setup as in the previous exercise, but now the non-zero entries of $x$ are non-negative. Taking this information into
account, we now solve
$$\underset{z}{\min}\, \|z\|_1 \quad \text{subject to\,\,} Az=y \,\, \text{and\,\,} z \ge 0$$
(here, $z\ge 0$ is meant entrywise, i.e., for each $k: z_k \ge 0$).
(The positivity constraint is easy to include in {\texttt cvx}).
Repeat the simulations as described in the previous exercise.
Compare your findings to the results from your experiments of the previous exercise and try
to quantify the difference regarding the range for $s$ for which recovery is still possible in this case.    
\end{myexercise}

\begin{myexercise} 
Let $m$ be a prime number $\ge 5$ and set 
$$g(j) = \frac{1}{\sqrt{m}}e^{2\pi i j^3/m}$$
for $j=0,\dots,m-1$. Define the vectors $\{g_{k,l}\}_{k,l=0}^{m-1}$ 
\begin{equation*}
g_{k,l}(j) := g(j-k) e^{2\pi i lj/m}, \quad k,l=0,\dots,m-1.
\end{equation*}
Prove that these  $m^2$ vectors $\{g_{k,l}\}_{k,l=0}^{m-1}$ plus the $m$ unit vectors $\{e_k\}_{k=0}^{m-1}$ together form a mutually unbiased basis (MUB) of $m(m+1)$ vectors for $\C^m$.
\end{myexercise}

\begin{myexercise}
This exercise is a bit more involved. It asks you to prove {\em Zauner's Conjecture}, a
fascinating conjecture that pervades many areas of mathematics, including quantum information theory, harmonic analysis and frame theory, finite group theory and representation theory, algebraic number theory, algebraic geometry, discrete and combinatorial geometry, operator algebras, and matrix analysis.
(If you actually succeed in proving Zauner's Conjecture, T.S.\ will send you one of the famous {\em Zaunerstollen} (different Zauner) as reward; and you may also become a serious contender for significant mathematical awards). \\
\textbf{Zauner's Conjecture:}~\cite{zauner1999quantendesigns,renes2004symmetric} For every integer $d \ge 2$, there exists a unit vector $\psi \in \mathbb C^d$ (called a {\em fiducial vector} such that its orbit under the Weyl–Heisenberg group,
$$
\mathcal F = \{ T_x M_{\omega}\psi : (x,\omega)\in  \Z_d \times \Z_d \},
$$
forms an {\em equiangular tight frame} of $d^2$ vectors in $\mathbb C^d$. Here $T_x$ is the cyclic translation operator defined by $T_x f(t) = f(t-x)$ where $t= 0,\dots,d-1$ and $x= 0,\dots,d-1$ (thus the translation index $t-x$ is computed modulo $d$) and $M_{\omega}$ is the modulation operator defined by $M_{\omega} f(x) = f(x) e^{2\pi i \omega t/d}$ for $\omega =0,\dots,d-1.$.
\end{myexercise}


\chapter{Low-Rank Matrix Recovery}
\label{c:lowrank}

\newcommand{\PO}{P_{\Omega}}
\newcommand{\cA}{{\mathcal A}}
\newcommand{\Hn}{{\mathcal H}}
\newcommand{\Hnn}{{\mathcal H}_n}
\newcommand{\cT}{{\mathcal T}}
\newcommand{\Tp}{{\mathcal T}^{\perp}}
\newcommand{\cto}{{{\mathcal T}_{X_0}}}
\newcommand{\tpo}{{{\mathcal T}^{\perp}_{X_0}}}

Low-rank matrices arise throughout the sciences because many high-dimensional datasets are governed by a small number of latent factors, leading to strong correlations among variables~\cite{udell2019big}.
Low-rank matrix recovery is a fundamental problem in mathematics and computer science that deals with recovering a matrix
from {\em incomplete measurements}~\cite{WrightMa}. What does {\em incomplete} mean in this context? Consider for example a matrix
 $X \in \R^{n_1\times n_2}$. Then, $X$ is determined by $n_1 n_2$ parameters. Assume we are given  $k$ measurements of this matrix with $k \ll n_1n_2$. Without any further assumptions, the problem of recovering $X$ from these measurements is ill-posed in this case, admitting infinitely many solutions. 
But what if $X$ is a low-rank matrix? Can we recover a low-rank matrix from fewer measurements, and if so, can we do it in 
 a numerically robust and efficient manner? And what other conditions do the measurements have to satisfy to make such an endeavour possible? These are the questions we intend to answer in this chapter.

A prominent example of matrix completion and low-rank modeling is collaborative filtering, widely used in
recommender systems~\cite{koren2009matrix,agarwal2016statistical,udell2016generalized}.  Imagine a user-item rating matrix $X$, where $X_{i,j}$ represents the rating given by user
$i$ to item $j$.  Most entries in this matrix are typically missing, as users usually rate only a small fraction of the available items.  The goal of collaborative filtering is to predict the missing ratings, allowing the system to recommend relevant items to users.  By assuming that the rating matrix is approximately low-rank (reflecting latent factors like user preferences and item characteristics), we can use matrix completion techniques to infer the missing ratings and provide personalized recommendations.  A famous example is the ``Netflix Prize'' competition conducted between 2006 and 2009, see~\cite{netflix} for details.

An important application of low-rank matrix recovery arises in quantum tomography~\cite{gross2010quantum}.  In this context, the matrix $X$ represents the density matrix
of a quantum system.  Due to limitations in measurement devices, we often can only obtain incomplete information about
the density matrix.  The density matrix is a positive semi-definite matrix with trace 1, and often, it is low-rank,
especially for systems in a mixed state or when only a limited number of degrees of freedom are accessible.  Matrix
completion techniques can be used to reconstruct the full density matrix from the incomplete measurements. This
reconstruction is crucial for characterizing the state of the quantum system and performing quantum information
processing tasks. The low-rank assumption is justified by the physical nature of many quantum states, and
carefully exploiting it  allows for efficient and robust state reconstruction.  

Other applications of matrix recovery and low-rank modeling arise in link prediction in social networks~\cite{chiang2014prediction,menon2011link}, Euclidean distance matrix embedding~\cite{javanmard2013localization,dokmanic2015euclidean},  robust photometric stereo~\cite{WrightMa}, 
 phase retrieval~\cite{CESV2013}, blind deconvolution~\cite{ahmed2013blind}, and self-calibration~\cite{LS15}. We will discuss some of these applications in more detail later in this chapter.

As the reader may suspect, there are strong connections between compressive sensing and low-rank matrix recovery.
The connection lies in the fact that both rely on parsimonious structural assumptions (sparsity for compressive sensing and low-rank for matrix recovery) and can be tackled using convex relaxation. In both cases, the idea is to exploit these structures to recover the original data efficiently despite having far fewer observations than the ambient dimension of the signal of interest. Thus, it is not surprising that throughout this chapter we will see many parallels to compressive sensing. Nevertheless, matrix recovery poses some additional challenges, as the attentive reader will discover.

\section{Rank minimization and the nuclear norm}

We will refer to a matrix $X \in \R^{n_1 \times n_2}$ as {\em low-rank} if its rank $r$ is small compared to $\min\{n_1, n_2\}$. The choice that $X$ is real-valued is just for convenience, but all results can easily be adapted to the complex-valued setting.
A simple computation reveals that such an $X$ has $r(n_1+n_2-r)$ degrees of freedom, which corresponds to the 
dimension of the tangent space to the manifold of rank-$r$ matrices.

We write $\Hnn$ for the Hilbert space of all $n \times n$ Hermitian matrices equipped
with the Hilbert-Schmidt inner product $\langle X,Y \rangle := Tr(Y^\ast X)$
We will also deal with linear transformation which acts on the space $\R^{n_1\times n_2}$, we will use
calligraphic letters to denote these operators, as in ${\mathcal A}(X)$.

Now, consider a matrix  $X_0 \in \R^{n_1 \times n}$ with rank
$r \ll \min(n_1,n_2)$.
and assume we are given measurements of the form
\begin{equation}
\label{tracemeasures}
\langle A_j,X_0 \rangle  = y_j, \qquad j=1,\dots,m,
\end{equation}
where the $A_j$ are some matrices  in $\R^{n_1 \times n_2}$. For notational convenience, we define the measurement operator
\begin{equation}
\label{tracemeasureop}
\cA(X_0): \R^{n_1 \times n_2} \to \R^m, \quad 
\cA(X_0)_j = \langle A_j,X_0 \rangle, \qquad j=1,\dots,m.
\end{equation}
Thus, we can express the (possibly noisy) matrix measurement process in compact form via
$$y = {\mathcal A}(X_0) + \eps,$$
where $y=\{y_j\}_{j=1}^m$ contains all measurement outcomes and $\eps \in \R^m$ denotes additive noise.

For reference, the adjoint operator $\cA^*$ maps real-valued inputs
into Hermitian matrices, and is given by
\[
\begin{array}{lll}
  \R^{m} & \to & \mathcal{H}^{n \times n}\\
  z & \mapsto & \sum_i a_i \, a_k a_k^*.
\end{array}
\]

Let us first consider the noiseless case. A natural approach to recovering $X_0$ is to solve the rank minimization problem
\begin{align}\label{eq:rankmin}
\begin{split}
\min_{X \in \mathbb{R}^{n_1 \times n_2}} \quad & \operatorname{rank}(X) \\
\text{subject to} \quad & \mathcal{A}(X) = y.
\end{split}
\end{align}

This formulation directly encodes the assumption that the solution should have minimal rank among all matrices consistent with the measurements.

In a few cases, rank-minimization problems can be solved explicitly and efficiently. Consider for example the problem
\begin{align}\label{eq:ranksvd}
\begin{split}
\text{minimize} \quad & \operatorname{rank}(X) \\
\text{subject to} \quad & \|X - A\| \le \eps,
\end{split}
\end{align}
where $A \in \R^{n_1\times n_2}$ is a given matrix. Let $A= \sum_{i=1}^{\min(n_1,n_2)} \sigma_i u_i v_i^T$ denote the SVD of $A$. Utilizing our findings in Chapter~\ref{ss:lowranksvd}, it is easy to see that the solution is given by a properly truncated SVD of $A$, namely $X= \sum_{i=1}^{r} \sigma_i u_i v_i^T$, where $r$
is the smallest number such that $\sigma_{r+1} \le \eps$.

However, the rank function is non-convex, discontinuous, and combinatorial in nature. As a result, rank minimization problems are NP-hard in general, even in seemingly simple settings~\cite{LVanderberghe_SBoyd_1996,fazel2002matrix}. 
One way to see this is by considering the special case where $X$
is a diagonal matrix. In that case we have $\rank X = \|X\|_0$ and the rank minimization problem takes the form
\begin{align*}
\min \quad & \|X\|_0 \\
\text{subject to} \quad & \mathcal{A}(X) = y,
\end{align*}
which played a central role in Chapter~\ref{c:cs}, and we already know it to be 
NP-hard~\cite{natarajan1995sparse,FoucartRauhut_CSbook}.

This computational intractability motivates the search for tractable relaxations that retain the essential structure of the original problem.

\begin{figure}[h]
\begin{center}
\includegraphics[width=50mm,height=55mm]{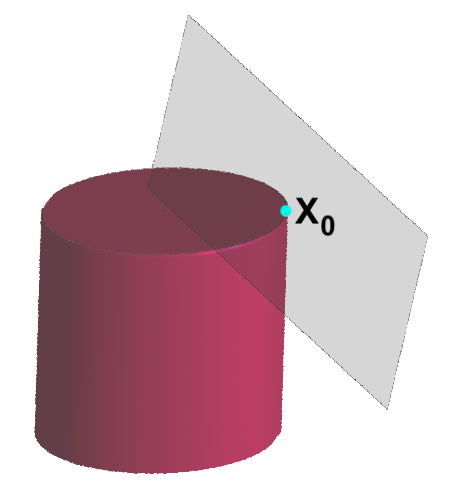}
\caption{The ``nuclear ball'' and the affine feasible set. 
The nuclear ball corresponds to $2\times 2$ symmetric matrices with nuclear norm at most equal to that of $X_0$.
The gray affine subspace represents the solution
space to the equation ${\mathcal A}(X) = {\mathcal A}(X_0)$.
When the feasible set is tangent to the ball at $X_0$, the
solution to~\eqref{eq:nuclearnormrelaxation} is exact.}
\label{fig:nucball}
\end{center}
\end{figure}

A standard convex surrogate for the rank function is the {\em nuclear norm}, defined as
$$
\|X\|_* := \sum_{i} \sigma_i(X),
$$
where ${\sigma_i(X)}$ are the singular values of $X$. The nuclear norm is convex and can be viewed as the matrix analogue of the $\ell_1$-norm for vectors, which is known to promote sparsity. In fact, the nuclear norm is the tightest convex lower bound of the rank function on the unit spectral-norm ball, see e.g. Theorem~1 in~\cite{fazel2001rank}. Compare also Figure~\ref{fig:nucball} to Figure~\ref{fig:l1l2}(a).

Replacing the rank objective in~\eqref{eq:rankmin} by the nuclear norm yields the convex optimization problem~\cite{fazel2002matrix,CR08,recht2010guaranteed}
\begin{align}\label{eq:nuclearnormrelaxation}
\begin{split}
\min_{X \in \mathbb{R}^{n_1 \times n_2}} \quad & \|X\|_* \\
\text{subject to} \quad & \mathcal{A}(X) = y.
\end{split}
\end{align}

This problem can be cast as a semidefinite program (see Exercise~\ref{ex:nuclear_SDP}). Consequently it can be solved efficiently using interior-point methods for moderate problem sizes or first-order methods for large-scale instances.\footnote{Generally, first-order methods have better time-complexity dependency on the problem size, but worse dependency on the desired distance to optimality.}  

In the presence of noise, i.e,, $y=\cA(X_0)+w$, where $w$ represents the noise term, one typically considers a relaxed formulation
\begin{align}\label{eq:nuclearnormrelaxationnoise}
\begin{split}
\text{minimize} \quad &  \|X\|_*  \\
\text{subject to} \quad & \|\mathcal{A}(X) - y\| \leq \tau,
\end{split}
\end{align}
where $\tau$ is an upper bound on the error, that is
$\|w\|\le \tau$.

Although nuclear norm minimization is only a relaxation of rank minimization, it has been shown to recover the true low-rank solution exactly under suitable conditions on the measurement operator $\mathcal{A}$. These conditions can be expressed in various ways, including restricted isometry properties, incoherence assumptions, or geometric criteria based on descent cones and Gaussian width.

The success of the convex relaxation is therefore not accidental: the nuclear norm captures the essential geometry of low-rank matrices in a way that allows both computational tractability and provable recovery guarantees, similarly to how the $\ell_1$ does for Sparsity (Chapter~\ref{c:cs}). As a result, nuclear norm minimization has become a cornerstone of modern low-rank matrix recovery theory. We will explore this topic in more detail in the next sections.

\section{Recovery guarantees via descent cone analysis}
\label{s:descentcone}

A central challenge in low-rank matrix recovery is to determine when a convex optimization procedure---typically nuclear norm minimization---can exactly recover a matrix of low rank from a set of linear measurements. A powerful and general framework to analyze this question arises from descent cone\footnote{The descent cone is sometimes equated to the tangent cone from convex geometry, although technically they do not usually coincide.} analysis and conic geometry, developed in~\cite{chandrasekaran2012convex,amelunxen2014living,tropp2015convex}
and refined in numerous followup papers, such as~\cite{kueng2017low,abbasi2019universality,fuchs2022proof}.
Our analysis will follow in part the perspicuous presentations in~\cite{tropp2015convex,fuchs2022proof}.

\begin{definition}\label{def:descentcone}
The descent cone $\mathcal{D}(f,X_0)$ of a proper convex function\footnote{A proper convex function is an extended real-valued convex function with a non-empty domain, that never takes on the value $-\infty$ and also is not identically equal to $+\infty$.} $f: \mathbb{C}^{n_1 \times n_2}\rightarrow \R$ at a point $X_0 \in \mathbb{C}^{n_1 \times n_2}$ is the conic hull of directions in which $f$ decreases near $X_0$ 
\begin{align*}
    \mathcal{D}(f,X_0) :=\{Z\in \mathbb{C}^{n_1 \times n_2}: f(X_0 + \epsilon Z) \le f(X_0) \text{ for some } \epsilon > 0\}.
\end{align*}
\end{definition}
The descent cone of a convex function is always a convex cone, but it may not be closed.

The descent cone analysis approach is widely applicable. In connection with low-rank matrix recovery, we are concerned with the case where $f(X) = \|X\|_*$.

Consider now the following two convex optimization problems:

\begin{multicols}{2}
\noindent
\begin{align}\label{fmin_nonoise}
\begin{split}
\text{min.} \quad  & f \left(X\right) \\
\text{s.t.} \quad & \mathcal{A} \left(X\right) = y.
\end{split}
\end{align}
\qquad
\begin{align}\label{fmin_noise}
\begin{split}
\text{min.} \quad  & f \left(X\right) \\
\text{s.t.} \quad & \|\mathcal{A} \left(X\right) - y\|_2\le \tau.
\end{split}
\end{align}
\end{multicols}

\begin{figure}[htbp]
\begin{center}
\subcaptionbox{Noiseless case}{
\includegraphics[width=45mm]{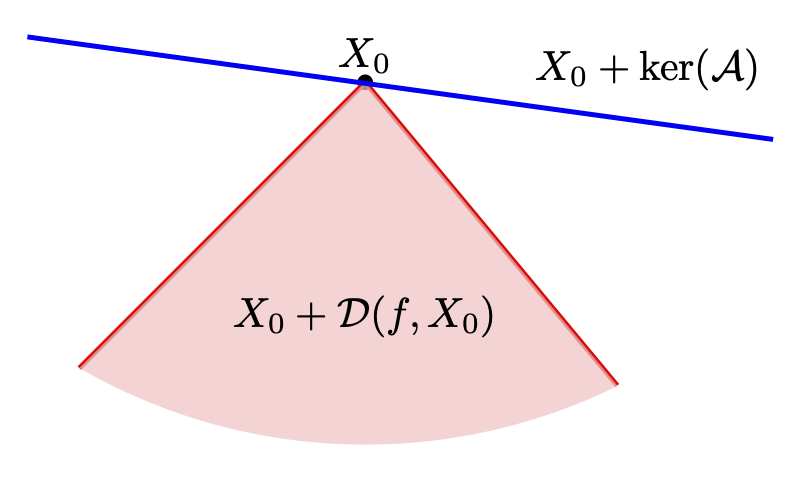}}
\qquad \qquad
\subcaptionbox{Noisy case}{
\includegraphics[width=45mm]{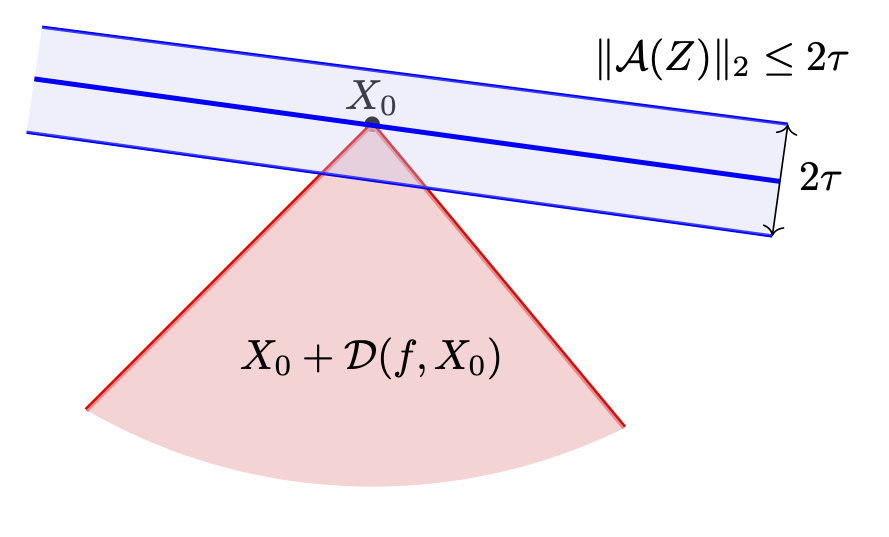}}
\caption{Descent cone analysis for the noiseless and the noisy case. Theoretical guarantees for low-rank matrix recovery can be obtained by analyzing the relative geometric orientation of the feasible space of the optimization problem with respect to the objective function’s descent cone (the reddish shaded areas) anchored at the matrix $X_0$.
For low-rank matrix recovery we take the function $f$ in the picture above to be $f(X)=\|X\|_*$.}
\label{fig:descentcone}
\end{center}
\end{figure}

We start with the noiseless case. If $X_0$ is the ground truth to $\mathcal{A} \left(X \right) = y$, then 
any minimizer $\hat{X}$ of~\eqref{fmin_nonoise} must satisfy $f(\hat{X}) \le f(X_0)$ and $\cA(\hat{X}) = y$.
This implies that $\hat{X}$ can be decomposed into $\hat{X} = X_0+Z$, where $Z \in \mathcal{D}(f,X_0) \cap \ker(\cA)$. If the intersection between the nullspace $\ker(\cA)$ and the descent cone $\mathcal{D}(f,X_0)$ only contains the zero element, then $X_0$ must be the 
unique optimal solution to~\eqref{fmin_nonoise}, as illustrated in Figure~\ref{fig:descentcone}(a).

In the noisy case, we cannot ask for exact recovery. But in light of~\eqref{fmin_noise} we hope that the error between a solution $\hat{X}$  and the ground truth $X_0$ is of the order of $\tau$. Let $\hat{X}=X_0+Z$ be a feasible solution of~\eqref{fmin_noise} and note that
the constraint $\|\mathcal{A} \left(X_0+Z\right) - y\|_2\le \tau$ implies $\|\mathcal{A} \left(Z\right)\|_2\le \tau$. Thus, to bound the error $\|\hat{X} - X_0\|_F = \|Z\|_F$, we need to focus our attention on the intersection of $\mathcal{D}(f,X_0)$ and
$\{Z: |\mathcal{A} \left(Z\right)\|_2\le \tau\}$. The smaller the size of the intersection, the smaller the error will be, cf.~also Figure~\ref{fig:descentcone}(b).

The size of $\mathcal{D}(f,X_0) \cap \{Z: |\mathcal{A} \left(Z\right)\|_2\le \tau\}$ can be controlled by a quantity called the {\em smallest conic singular value}\footnote{The terminology originates in the fact that $\sigma_{\min}\big(\cA,\R^{n_1\times n_2})$
coincides with the usual minimum singular value of $\cA$.}~\cite{tropp2015convex,chandrasekaran2012convex,fuchs2022proof}, which is defined by
\begin{equation}\label{conesv}    
\sigma_{\min}\big(\cA,\mathcal{D}(f,X_0)\big) :=   
\inf_{Z \in \mathcal{D}(f,X_0) \setminus \{0\}} \frac{\|\cA(Z)\|}{\|Z\|_F}.
\end{equation}
The following proposition shows that when $\sigma_{\min}\big(\cA,\mathcal{D}(f,X_0)\big)$ is large, the size of the intersection will be small, which in turn would yield a smaller error~\cite{tropp2015convex,chandrasekaran2012convex,fuchs2022proof}.
\begin{proposition} \label{prop:cone}
Let $\cA: \R^{n_1\times n_2} \to \R^m$ be a linear operator and assume
that $y = \cA(X_0)+w$ with $\|w\|\le \tau$.  Then, any minimizer $\hat{X}$ of the optimization problem~\eqref{fmin_noise} satisfies
\begin{equation}\label{cone_error1}
 \|\hat{X} - X_0\|_F \le \frac{2\tau}{\sigma_{\min}\big(\cA,\mathcal{D}(f,X_0)\big)}.  
\end{equation}
\end{proposition}

\begin{proof}
We have 
\begin{equation}\label{cone_error2}
\sigma_{\min}\big(\cA,\mathcal{D}(f,X_0)\big) \le \frac{\|\cA(Z)\|}{\|Z\|_F} \le 
\frac{\|\cA(Z)-w\|+\|w\|}{\|Z\|_F} \le 
\frac{2\tau}{\|Z\|_F},
\end{equation}
for any feasible $Z$. Here, the first inequality in~\eqref{cone_error2} follows from the definition of
$\sigma_{\min}\big(\cA,\mathcal{D}(f,X_0)\big)$, the second inequality from the triangle inequality, and the third one from $\|\mathcal{A} \left(Z\right)\|_2\le \tau$ and the assumption $\|w\|\le \tau$. Rearranging the terms in~\eqref{cone_error2} concludes the proof.

\end{proof}
As a consequence, if we can show that $\sigma_{\min}\big(\cA,\mathcal{D}(f,X_0)\big) > 0$, then this would imply exact recovery in the noiseless scenario. However, Proposition~\ref{prop:cone} is deceptive in its simplicity. In general it is actually rather difficult to compute or bound 
the smallest conic singular value of a matrix. The situation becomes more promising once we introduce  randomness in the measurement process, and thereby into $\cA$. We will explore such a scenario next.

\subsection{Low-rank matrix recovery from Gaussian measurements}\label{ss:matrixrecoverygaussian}

We consider the case of low-rank matrix recovery with measurements of the form
$$
\cA(X_0)_i = y_i = \langle A_i,X_0 \rangle, \qquad i=1,\dots,m,
$$
where the $A_i$ are $n_1 \times n_2$  Gaussian matrices.

The proof strategy to bound $\sigma_{\min}\big(\cA,\mathcal{D}(f,X_0)\big)$ for this Gaussian measurement operator is based on  Gordon's Theorem, the use of which was first proposed in the context of sparse recovery by Rudelson and Vershynin in~\cite[Section 4]{rudelson2008sparse}. It revolves around the concept of Gaussian width, which we introduced in Definition~\ref{def:gaussianwidth}.

Setting $S:= \{Z \in \mathcal{D}(\|\cdot \|_*,X_0):\|Z\|_F =1\}$, our goal becomes to bound $\inf_{Z \in S} \|\cA(Z)\|$ from below. 
Revisiting Gordon's Theorem,  an inspection of the proof of Theorem~\ref{GordonsTheorem} (in particular~\eqref{eq:5:usingGaussianconcentration_2}) yields the following tail bound\footnote{ The observant reader might have noticed that in~\eqref{gordontail} we have replaced the {\em minimum} in~\eqref{eq:gordon2} by the {\em infimum}. This switch (the verification of which is left to the reader) is due to the fact that the descent cone $S$ may not be closed.}
\begin{equation}\label{gordontail}
\inf_{x\in S} \| \cA(Z) \| \geq \sqrt{m-1} - \omega(S) - t, \qquad \text{for $t > 0$,}
\end{equation}
with probability at least $e^{-t^2/2}$,
where we have used the well known bound $a_d \ge \sqrt{d-1}$ in~\eqref{eq:gordon2} and $d=m$  to adapt the notation in Chapter~\ref{c:probability-gaussiananalysis} to the current setting ($m$ denotes the number of measurements in this chapter).
If $w(S)$ does not exceed $\sqrt{m-1}$, then we can obtain probabilistic recovery guarantees via Proposition~\ref{prop:cone}.

As pointed out in~\cite{fuchs2022proof}, Gordon's Theorem---and therefore also the tail bound in~\eqref{gordontail}---only requires $S$ to be a subset of the Frobenius unit sphere\footnote{The Frobenius unit sphere is ${\mathcal S}_F(\R^{n_1\times n_2}):= \{Z\in \R^{n_1\times n_2}: \|Z\|_F=1\}$.} ${\mathcal S}_F(\R^{n_1\times n_2})$. Therefore, one is not
restricted to a specific descent cone but one can instead choose the union over all possible descent
cones corresponding to rank-$r$ matrices in order to obtain uniform recovery guarantees:
\begin{align*} 
S_r = {\mathcal S}_F(\R^{n_1\times n_2}) \cap K_r \quad \text{where} \quad K_r = \bigcup_{{\small 
\begin{matrix}
    X \in \R^{n_1\times n_2} \\ \rank(X)=r
\end{matrix} }}
 \mathcal{D} \left( \| \cdot \|_*, X \right).
\end{align*}

\begin{proposition}[\cite{fuchs2022proof},  Corollary~2.1]\label{prop:matrixconebound}
    The Gaussian width of $S_r$, the union over all possible descent cones with an anchor point of rank-$r$ can be bounded by
    \begin{align*}
        w(S_r) 
        \lesssim \sqrt{r} \left( \sqrt{n_1} + \sqrt{n_2}\right).
    \end{align*}
    Furthermore, let $\mathcal{A}: \R^{n_1\times n_2}\rightarrow \R^m$ be a Gaussian measurement operator. Then, $\lambda_{\min} \left( \mathcal{A},  \mathcal{D}(f,X)\right)$ is bounded away from zero for any rank-$r$ matrix X w.h.p. if
    \begin{align*}
        m \gtrsim  r (n_1 + n_2).
    \end{align*}
\end{proposition}

The proof of Proposition~\ref{prop:matrixconebound} makes use of the following result, which we state without proof.
\begin{lemma}[\cite{fuchs2022proof},  Lemma~2.2]
\label{le:conerank}
Suppose that $Z \in \R^{n_1 \times n_2}$  is contained in the nuclear norm descent cone of a rank-$r$ matrix $X \in \R^{n_1 \times n_2}$. Then,
\begin{align*}
\|Z\|_* \leq  \left(1+\sqrt{2}\right) \sqrt{r}  \|Z\|_F.
\end{align*}
\end{lemma}

\begin{proof}[of Proposition~\ref{prop:matrixconebound}]
    Using H\"older's inequality and Lemma \ref{le:conerank}, the Gaussian width $w(S_r)$ can be bounded in terms of the expected operator norm of a standard Gaussian matrix:
\begin{align*}
    w(S_r) = \mathbb{E}\sup_{Z\in S_r} \langle A, Z\rangle & \leq \sup_{Z \in S_r} \|Z \|_* \,\mathbb{E} \|A  \| \\ & \leq\left(1+\sqrt{2}\right)
    \sqrt{r} \mathbb{E}\|A\| \\
    & \lesssim 
    \sqrt{r}(\sqrt{n_1}+\sqrt{n_2}),
\end{align*}
where in last inequality we have used the bound $\mathbb{E}\|A\| \lesssim \sqrt{n_1}+\sqrt{n_2}$ , which can be found in e.g.~\cite[Remark 4.4.4]{vershynin2010introduction}. Using the tail bound~\eqref{gordontail} derived from Gordon's Theorem, we obtain
\begin{align*}
    \inf_{X\in S_r}\|\mathcal{A}(Z)\|_2 = \inf_
    {{\small \begin{matrix}
    X \in \R^{n_1\times n_2} \\ \rank(X)=r
\end{matrix}} }
 \lambda_{\min} \left( \mathcal{A},  \mathcal{D}(\Vert \cdot \Vert_{\ast},X_0)  \right) \ge \sqrt{m-1} - w(S) - t
\end{align*}
 with probability at least $1-e^{-t^2/2}$. Therefore, if
$$
    m \gtrsim r (n_1+n_2) 
$$
we can pick $t>0$, such that $\inf_{X\in S_r}\|\mathcal{A}(Z)\|_2 $ is positive w.h.p. 
\end{proof}

\section{Low-rank matrix recovery and dual certificates}\label{s:matrixdual}

The descent cone approach, while geometrically appealing, faces arduous challenges\footnote{Some, but not all, of these challenges can be addressed by employing Mendelson’s small ball method, see~\cite{mendelson2015learning,koltchinskii2015bounding,tropp2015convex,fuchs2022proof}.}
when we deal with problems where additional structure, such as incoherence of the unknown signals, is critical to exclude exceptional instances where the reconstruction must necessarily fail (see~\cite{fuchs2022proof} for a more detailed discussion  of the issues that arise).
Thus, in applications like quantum state tomography, phase retrieval, and matrix completion, an approach based on dual certification introduced in Chapter~\ref{s:duality}, is often preferable.

The next lemma will pave the way to establishing tractable conditions that the nuclear norm relaxation~\eqref{eq:nuclearnormrelaxation}  attains its optimal value at a specific point is through the provision of a dual certificate (sometimes also called {\em dual witness}).

Recall from our computation of the subdifferential of the nuclear norm in Chapter~\ref{ss:subgradient} the definition the tangent space $\Tn$ at a rank$-r$ $n_1 \times n_2$ matrix $X$ with SVD $X = U \Sigma V^T$ relative to the manifold of rank-$r$ matrices:
$$\Tn_X = \{ U A^T + B V^T : A \in \mathbb{R}^{n_1 \times r}, B \in \mathbb{R}^{n_2 \times r} \}.$$

\begin{lemma}[Dual certificate,  \cite{candes2009exact}]\label{le:candesrechtlowrank}
Let $X_0 \in \R^{n_1 \times n_2}$ be such that $\mathcal{A}\left( X_0 \right) = y \in \R^m$.
Assume that the following two conditions hold:
\begin{enumerate}
 \item There exists a vector $\lambda \in \R^m$ such that $Y=\mathcal{A}^* \lambda$ satisfies
\begin{equation}\label{eq:nucleardualcond}
 \mathcal{P}_{\Tn_{X_0}}  Y = UV^* \quad \text{and} \quad \Vert \mathcal{P}_{\Tn^{\perp}_{X_0}} Y  \Vert < 1.
\end{equation}
\item The linear operator $ \mathcal{A}$ is injective when restricted to the tangent space $\Tn_{X_0}$.
\end{enumerate}
Then $X_0$ is the unique minimizer of~\eqref{eq:nuclearnormrelaxation}.
\end{lemma}

\begin{proof}
Suppose $X_0$ is feasible and has SVD
$X_0 = U \Sigma V^T$ with $\rank(X_0)=r$.
Let $P_{\cto}$ and $P_{\tpo}$  
denote the  orthogonal projections on $\Tn_{X_0}$ and
$\tpo$, respectively.

The Lagrangian is
$$
\mathcal L(X,\lambda)
= \|X\|_* + \langle \lambda,, \mathcal A(X)-b\rangle.
$$

The KKT conditions for optimality of $X_0$ are (note that we do not have inequality constraints):
(i)~Primal feasibility: $\mathcal A(X_0)=b$.
(ii)~Dual feasibility: $\mathcal A^*(\lambda) \in \partial\|X_0\|_*$.
(iii)~Complementary slackness: these are trivially fulfilled here since all constraints are equalities.
Thus, to certify optimality it suffices to construct a $\lambda$ such that
$$
Y := \mathcal A^*(\lambda) \in \partial\|X_0\|_*.
$$
We have seen in~\eqref{eq:nuclearsubdifferential}  that 
$$
\partial \|X_0\|_* =
\left\{UV^T + W : \mathcal P_{\mathcal T}(W)=0, \|W\|\le 1
\right\}.
$$
So any matrix of the form
$$
Y = UV^T + W,
\quad
W\in \tpo, \|W\|\le 1
$$
is a valid subgradient.

Suppose now there exists $Y=\mathcal A^*(\lambda)$ such that $\mathcal P_{\cto}(Y) = UV^T$ and 
$\|\mathcal P_{\tpo}(Y)\| < 1$.
We will show that this implies $Y\in\partial\|X_0\|_*$.

Write
$$Y = {\mathcal P}_{\cto}(Y) + {\mathcal P}_{\tpo}(Y)
= UV^T + W,
$$
where $W := \mathcal P_{\tpo}(Y)$.
By assumption, $W \in \tpo$ and $\|W\|< 1$, thus 
all conditions in~\eqref{eq:nuclearsubdifferential} are satisfied, and therefore
$Y \in \partial\|X_0\|_*$.
Since $X_0$ is feasible and
$Y=\mathcal A^*(\lambda)\in\partial\|X_0\|_*$,
the KKT conditions hold, so $X_0$ is optimal.

If additionally
$$
\|\mathcal P_{\tpo}(Y)\| < 1
\quad\text{and ${\mathcal A}$ is injective on $\cto$},
$$
then one can show that for any feasible $X\neq X_0$,
$$
\|X\|_* > \|X_0\|_*,
$$
hence $X_0$ is the unique minimizer.
This follows from the subgradient inequality
$$
\|X\|_* \ge \|X_0\|_* + \langle Y, X-X_0\rangle,
$$
with equality only if $X-X_0\in \cto$. 

\end{proof}

The hard part is now to verify that a given measurement operator $\cA$ actually  satisfies the conditions of Lemma~\ref{le:candesrechtlowrank}, in particular condition~\eqref{eq:nucleardualcond}.
One of the most successful and widely applicable approaches for this relies on the so-called {\em golfing scheme}, introduced by
David Gross~\cite{gross2011recovering}. 
In~\cite{gross2011recovering} Gross demonstrates that the conditions in Lemma~\ref{le:candesrechtlowrank} can be relaxed and that the
existence of an {\em approximate} dual certificate suffices to establish uniqueness
and this approximate dual certificate can be constructed via the golfing scheme in an iterative manner\footnote{In~\cite{fuchs2022proof} it is shown that a minor modification of the golfing scheme gives an explicit randomized construction for an exact dual certificate, which does not use more measurements than the original argument.}.

The concept of an {\em approximate dual certificate} allows for small errors on the tangent space $\mathcal{T}$ and a margin of safety on the orthogonal complement $\mathcal{T}^\perp$.

\begin{definition}\label{de:approxdual}
A matrix $Y \in \range(\mathcal{A}^*)$ is an $(\alpha, \beta)$-approximate dual certificate for $X_0 = U\Sigma V^\top$ if it satisfies the two conditions:
\begin{enumerate}
\item \begin{equation}\label{eq:dualalpha}
\text{Alignment on $\Tn$: $\|P_{\mathcal{T}}(Y) - UV^\top\|_F \le \alpha$,}
\end{equation}
\item 
\begin{equation}\label{eq:dualbeta}
\text{Constraint on $\Tn^\perp$: $\|P_{\mathcal{T}^\perp}(Y)\| \le \beta < 1$.}
\end{equation}
\end{enumerate}
\end{definition}

\medskip
The operator $\mathcal{P}_{\mathcal{T}} \mathcal{A}^* \mathcal{A} \mathcal{P}_{\mathcal{T}^\perp}$ will play an important role in our subsequent analysis. It dictates how much of a signal can hide in the kernel of the operator relative to its projection onto the tangent space $\Tn$.
The next lemma quantifies the ``spectral-to-Frobenius'' leakage. It ensures that any matrix in the kernel of the measurement operator cannot have its energy concentrated in the ``signal'' space $\mathcal{T}$ without having significantly more energy in the ``noise'' space $\mathcal{T}^\perp$.

\begin{lemma}[Leakage Lemma]\label{le:leakage}
Let $\mathcal{A}: \mathbb{C}^{d \times d} \to \mathbb{R}^m$ be a measurement operator. For any $H \in \ker(\mathcal{A})$, the following inequality holds
$$\|P_{\mathcal{T}}(H)\|_F \le \Cleak \|P_{\mathcal{T}^\perp}(H)\|,$$
where the {\em leakage constant} is defined as 
$$\Cleak := \frac{\|\mathcal{P}_{\mathcal{T}} \mathcal{A}^* \mathcal{A} \mathcal{P}_{\mathcal{T}^\perp}\|_{\text{op} \to \text{F}}}{\lambda_{\min}(\mathcal{P}_{\mathcal{T}} \mathcal{A}^* \mathcal{A} \mathcal{P}_{\mathcal{T}})},$$
and $\lambda_{\min}(\mathcal{P}_{\mathcal{T}} \mathcal{A}^* \mathcal{A} \mathcal{P}_{\mathcal{T}})$ denotes the smallest eigenvalue of $\mathcal{P}_{\mathcal{T}} \mathcal{A}^* \mathcal{A} \mathcal{P}_{\mathcal{T}}$.
\end{lemma}

\begin{proof}
 Since $H \in \ker(\mathcal{A})$, we have $\mathcal{A}(H) = 0$ and $\mathcal{A}(P_{\mathcal{T}}(H)) = -\mathcal{A}(P_{\mathcal{T}^\perp}(H))$.
Lower Bound (RIP on $\mathcal{T}$): Let $\mathcal{R}_{\mathcal{T}} = \mathcal{P}_{\mathcal{T}} \mathcal{A}^* \mathcal{A} \mathcal{P}_{\mathcal{T}}$. For any $Z \in \mathcal{T}$, we have $\|\mathcal{A}(Z)\|^2 = \langle Z, \mathcal{R}_{\mathcal{T}} Z \rangle \ge \lambda_{\min}(\mathcal{P}_{\mathcal{T}} \mathcal{A}^* \mathcal{A} \mathcal{P}_{\mathcal{T}}) \|Z\|_F^2$. Applying this to $P_{\mathcal{T}}(H)$ gives
$$\lambda_{\min}(\mathcal{P}_{\mathcal{T}} \mathcal{A}^* \mathcal{A} \mathcal{P}_{\mathcal{T}}) \|P_{\mathcal{T}}(H)\|_F^2 \le \|\mathcal{A}(P_{\mathcal{T}}(H))\|^2.$$
We have
\begin{align*}
\|\mathcal{A}(P_{\mathcal{T}}(H))\|^2 & = \langle \mathcal{A}(P_{\mathcal{T}}(H)), \mathcal{A}(P_{\mathcal{T}}(H)) \rangle = \langle \mathcal{A}(P_{\mathcal{T}}(H)), -\mathcal{A}(P_{\mathcal{T}^\perp}(H)) \rangle\\
& = -\langle P_{\mathcal{T}}(H), \mathcal{P}_{\mathcal{T}} \mathcal{A}^* \mathcal{A} \mathcal{P}_{\mathcal{T}^\perp} (P_{\mathcal{T}^\perp}(H)) \rangle.
\end{align*}
Applying Cauchy-Schwarz and induced norms gives
$$\lambda_{\min}(\mathcal{P}_{\mathcal{T}} \mathcal{A}^* \mathcal{A} \mathcal{P}_{\mathcal{T}}) \|P_{\mathcal{T}}(H)\|_F^2 \le \|P_{\mathcal{T}}(H)\|_F \cdot \|\mathcal{P}_{\mathcal{T}} \mathcal{A}^* \mathcal{A} \mathcal{P}_{\mathcal{T}^\perp}\|_{\text{op} \to \text{F}} \cdot \|P_{\mathcal{T}^\perp}(H)\|$$
Dividing both sides by $\lambda_{\min}(\mathcal{P}_{\mathcal{T}} \mathcal{A}^* \mathcal{A} \mathcal{P}_{\mathcal{T}}) \|P_{\mathcal{T}}(H)\|_F$ establishes the desired bound. 
\end{proof}

\begin{theorem}\label{th:dualnuclear}
Let $X_0$ be a rank-$r$ matrix with tangent space $\mathcal{T}$. Suppose there exists an $(\alpha, \beta)$-approximate  dual certificate $Y = \mathcal{A}^*(y)$.
 with leakage constant $\Cleak$ such that $\alpha < \frac{1-\beta}{\Cleak}$, then $X_0$ is the unique minimizer of the nuclear norm minimization problem~\eqref{eq:nuclearnormrelaxation}.
\end{theorem}

\begin{proof}
We must show that for any non-zero $H \in \ker(\mathcal{A})$, $\|X_0 + H\|_* > \|X_0\|_*$.

By the definition of the nuclear norm subgradient at $X_0$
$$\|X_0 + H\|_* \ge \|X_0\|_* + \langle UV^\top, H \rangle + \|P_{\mathcal{T}^\perp}(H)\|_*.$$
Since $H \in \ker(\mathcal{A})$ and $Y \in \text{range}(\mathcal{A}^*)$, we have $\langle Y, H \rangle = 0$. Thus
$$0 = \langle P_{\mathcal{T}}(Y), P_{\mathcal{T}}(H) \rangle + \langle P_{\mathcal{T}^\perp}(Y), P_{\mathcal{T}^\perp}(H) \rangle$$
$$\implies \langle UV^\top, P_{\mathcal{T}}(H) \rangle = \langle UV^\top - P_{\mathcal{T}}(Y), P_{\mathcal{T}}(H) \rangle - \langle P_{\mathcal{T}^\perp}(Y), P_{\mathcal{T}^\perp}(H) \rangle.$$
Substituting this identity into the subgradient inequality gives
$$\|X_0 + H\|_* - \|X_0\|_* \ge \|P_{\mathcal{T}^\perp}(H)\|_* - \langle P_{\mathcal{T}^\perp}(Y), P_{\mathcal{T}^\perp}(H) \rangle - \langle P_{\mathcal{T}}(Y) - UV^\top, P_{\mathcal{T}}(H) \rangle.$$
By H\"{o}lder's inequality, 
$$\langle P_{\mathcal{T}^\perp}(Y), H \rangle \le \|P_{\mathcal{T}^\perp}(Y)\| \|P_{\mathcal{T}^\perp}(H)\|_* \le \beta \|P_{\mathcal{T}^\perp}(H)\|_*.$$
Using Cauchy-Schwarz gives $\langle P_{\mathcal{T}}(Y) - UV^\top, P_{\mathcal{T}}(H) \rangle \le \alpha \|P_{\mathcal{T}}(H)\|_F$.
We now apply the Leakage Lemma and obtain
$\alpha \|P_{\mathcal{T}}(H)\|_F \le \alpha \Cleak \|P_{\mathcal{T}^\perp}(H)\|$.
Hence,
$$\|X_0 + H\|_* - \|X_0\|_* \ge (1-\beta) \|P_{\mathcal{T}^\perp}(H)\|_* - \alpha \Cleak \|P_{\mathcal{T}^\perp}(H)\|.$$
Since $\|P_{\mathcal{T}^\perp}(H)\|_* \ge \|P_{\mathcal{T}^\perp}(H)\|$ for any matrix we get
$$\|X_0 + H\|_* - \|X_0\|_* \ge (1 - \beta - \alpha \Cleak) \|P_{\mathcal{T}^\perp}(H)\|.$$
If $\alpha < \frac{1-\beta}{\Cleak}$, the term in the parentheses is strictly positive, which implies that $\|X_0 + H\|_* > \|X_0\|_*$.  
\end{proof}

\begin{remark}
The approach we have just outlined using the approximate dual certificate and the Leakage Lemma is essentially a dual-space perspective on the primal-space descent cone geometry that we described in Chapter~\ref{s:descentcone}.
In the descent cone approach we ask how much the operator $\mathcal{A}$ ``shinks'' vectors that are trying to decrease the nuclear norm. If the smallest conic singular value is large, the operator captures these directions, and they cannot be in the kernel. In the approximate dual certificate approach the leakage constant $\Cleak$ quantifies how much a kernel element $H$ can ``hide'' its energy in the tangent space. A large smallest conic singular value
$\sigma_{\min}\big(\cA,\mathcal{D}\big)$  corresponds to a small leakage constant $\Cleak$ (but not in a way that allows us to easily convert one into the other).
\end{remark}

The reader may wonder whether so far we just have been {\em `kicking the can down the road'}, since we moved from the primal problem to the  dual certificate to the approximate dual certificate to the leakage constant. After all, while Theorem~\ref{th:dualnuclear} proves that a certificate $Y$ allows for recovery, it still does not tell us how to find one. 

Fortunately\footnote{This is not a fortunate coincidence, but rather fortune that favors the prepared mathematician.}, there are now several techniques at our disposal to address this critical issue.
One such technique is the aforementioned {\em golfing scheme}. It provides an ingenious  algorithm to build $Y \in \range(\mathcal{A}^*)$ by successively projecting the ``residual error'' from the tangent space $\mathcal{T}$ back into the range of the measurement operator.
The concrete application of Theorem~\ref{th:dualnuclear} and the golfing scheme will depend on the specific nature and setting of the low-rank problem under consideration. In the following we will explore three compelling applications, matrix completion, quantum state tomography, and phase retrieval. 

While the overall proof architecture via approximate dual certificate and golfing scheme is shared across all three applications, the specific mathematical properties of the measurements affects
key steps in the proofs significantly. The difficulty in each proof lies in how the specific measurement operator $\cA$ interacts with the Tangent Space $\Tn$.

\section{Matrix completion}\label{s:matrixcompletion}

Matrix completion is the problem of recovering an unknown matrix from a subset of its entries, under the assumption that the matrix has low rank~\cite{CR08,candes2010power}. Cast in the form of~\eqref{tracemeasures}, this means that the measurement matrices take the form $e_{i} e_{j}^{\ast}$, whence  $\langle  e_i e_j^{\ast}, X \rangle = X_{i,j}$, i.e., the measurements reveal individual entries of the matrix.  We define the sampling operator $\mathcal{P}_\Omega = \sum_{(i,j) \in \Omega} \mathcal{P}_{ij}$, where $\mathcal{P}_{ij}(X) = \langle e_i e_j^\top, X \rangle e_i e_j^\top$, where $\Omega$ is some subset of the complete set of matrix entry indices $[n_1] \times [n_2]$. Thus, in the notation of~\eqref{tracemeasures}, ${\mathcal A}=P_{\Omega}$
and $({\mathcal A}(X))_{ij} = X_{ij}$.

Without further knowledge the problem of recovering $X$ from the entries $\mathcal{P}_\Omega(X)$ is ill-posed and has infinitely many solutions---just fill in the missing entries with arbitrary values. But assuming that  $X$ is a low-rank matrix, can we then fill in the missing entries of $X$ in a unique way and can we do so in a numerically feasible way? This problem arises for example in recommender systems, collaborative filtering (user–item ratings, preferences), crowdsourcing and survey data~\cite{koren2009matrix,WrightMa,agarwal2016statistical,udell2016generalized}.

By counting degrees of freedom of $X$, it is apparent that for low-rank matrix completion we need at least $r(n_1+n_2-r)$ many measurements. But this condition will in general not suffice. Take for example a rank-one matrix whose entries are all zeros, except for, say, its very first entry which we set equal to one. Since a priori we do not know the location of the non-zero entry, in the worst case we would need to take $n_1 n_2$ measurements (equal to the ambient dimension of the matrix) in order to successfully complete all missing matrix entries.

This should not come as a complete surprise, since we have encountered a similar situation in compressive sensing. There, we had to impose some notion of incoherence (e.g., via the RIP or directly via the concept of coherence) in order to develop a non-trivial, compelling theory (and numerics) of sparse recovery. We will need to do something similar for matrix completion.
In Chapter~\ref{s:coherence} (see equation~\eqref{coherence}) we introduced the concept of the (in)coherence of a set of vectors. For matrices we follow the notion of (in)coherence introduced in~\cite{CR08}. 

\begin{definition}\label{de:matrixcoherence}
    Let $X \in \mathbb{R}^{n_1 \times n_2}$ be a rank-$r$ matrix with  singular value decomposition $X = U \Sigma V^T$,
where $U \in \mathbb{R}^{n_1 \times r}$, $V \in \mathbb{R}^{n_2 \times r}$. We say that $X$ satisfies the incoherence condition with parameter $\mu$ if
$$\max_i \|U^\top e_i\|^2 \le \frac{\mu r}{n_1}, \quad \max_j \|V^\top e_j\|^2 \le \frac{\mu r}{n_2}, \,\text{and}\,\,\,
\|UV^\top\|_\infty \le \sqrt{\frac{\mu r}{n_1 n_2}}.$$
\end{definition}
Incoherence requires the energy of singular vectors to be spread across coordinates. This prevents ``spiky'' singular vectors, which would make recovery from few entries impossible.
The smallest coherence is 1, achieved if $U$ and $V$ are spanned by vectors whose entries all have magnitude 
$1/\sqrt{n_1}$ and $1/\sqrt{n_2}$, respectively.
The largest possible value for $\mu$ is $\max(n_1,n_2)/r$, 
which would correspond to any $U$ or $V$ that contains a standard basis element.

We will now apply Theorem~\ref{th:dualnuclear} and the golfing scheme to solve the matrix completion problem, see also~\cite{CR08,gross2011recovering}

\begin{theorem}\label{th:matrixcompletion}
Assume $X_0 \in \mathbb{R}^{n_1 \times n_2}$ is a rank-$r$ matrix and let $X_0$ be $\mu$-incoherent.  
Suppose we observe $m$ entries of $X_0$ with locations sampled uniformly with replacement. 
Then there exists a constant $C$ such that if
$$m \ge C \mu n r \log^2 n,$$
where $n = \max(n_1, n_2)$,
then  $X_0$ is the unique minimizer of the nuclear norm minimization problem
$$\min \|X\|_* \quad \text{subject to} \quad \mathcal{P}_\Omega(X) = \mathcal{P}_\Omega(M),$$
with probability at least $1-n^{-3}$.
\end{theorem}

In the proof of Theorem~\ref{th:matrixcompletion} we will make use of the following notation. We denote the
normalized sampling operator by $\mathcal{R} := \frac{1}{p} \mathcal{P}_\Omega$, where $p = m/(n_1 n_2)$, and the
associated subspace projections by
$\mathcal{R}_{\mathcal{T}}  := \mathcal{P}_{\mathcal{T}} \mathcal{R} \mathcal{P}_{\mathcal{T}}$ and 
$\mathcal{R}_{\mathcal{T}, \mathcal{T}^\perp}  := \mathcal{P}_{\mathcal{T}} \mathcal{R} \mathcal{P}_{\mathcal{T}^\perp}$. Note  that $\mathcal{P}_{\mathcal{T}}$  acts as identity on $\Tn$.

We will employ the following two lemmata. The first one establishes injectivity of the sampling operator on $\Tn$m the second one provides a bound on the leakage constant.

\begin{lemma}\label{le:matrixinjectivity} Given the assumptions of Theorem~\ref{th:matrixcompletion}, for $m \ge C \mu n r \log n$, we have
$$\|\mathcal{R}_{\mathcal{T}} - \mathcal{P}_{\mathcal{T}}\| \le \delta < 1,$$
 with probability at least $1 - 2n^{-3}$
\end{lemma}

\begin{proof}
We define the centered random operator $\mathcal{S} = \mathcal{R}_{\mathcal{T}} - \mathcal{P}_{\mathcal{T}}$. This can be written as a sum of $m$ independent, zero-mean operators $\mathcal{Z}_k$
$$\mathcal{S} = \sum_{k=1}^m \mathcal{Z}_k, \quad \text{where } \mathcal{Z}_k = \frac{1}{m} \left( \frac{1}{p} \mathcal{P}_{\mathcal{T}} \mathcal{P}_{ij} \mathcal{P}_{\mathcal{T}} - \mathcal{P}_{\mathcal{T}} \right)$$
We have 
\begin{align*}
\|\mathcal{Z}_k\|  &\le \|\mathcal{P}_{\mathcal{T}} \mathcal{P}_{ij} \mathcal{P}_{\mathcal{T}}\| + \|\mathcal{P}_{\mathcal{T}}\| 
=
\Big\| \sqrt{\frac{n^2}{m}} \mathcal{P}_{\mathcal{T}}(e_i e_j^\top) \Big\|_F^2 + \frac{1}{m} \\ &
= \frac{n^2}{m} \|\mathcal{P}_{\mathcal{T}}(e_i e_j^\top)\|_F^2 + \frac{1}{m} \le \frac{n^2}{m}\frac{2\mu r}{n} + \frac{1}{m}\le \frac{3\mu r n}{m},
\end{align*}
where we have used the incoherence assumption in the last step. For the variance $\sigma^2$ we have $\sigma^2 = \| \sum \mathbb{E}[\mathcal{Z}_k^* \mathcal{Z}_k] \| \le \frac{2\mu n r}{m}$.

The matrix Bernstein inequality of Theorem~\ref{thm:4:MatrixBernstein} reads in our setting
\begin{equation}\label{eq:Bernstein4}
\mathbb{P}(\|\mathcal{S}\| \ge \delta) \le 2nr \exp \left( \frac{-\delta^2}{2\sigma^2 + 2R\delta/3} \right).\end{equation}
Substituting the computed bounds for $\|\mathcal{Z}_k\|$ (the $R$ in~\eqref{eq:Bernstein4}) and $\sigma^2$ into the exponent of the right-hand side of~\eqref{eq:Bernstein4} gives for the exponent
$$\frac{-\delta^2}{\frac{4\mu n r}{m} + \frac{6\mu n r}{m} \frac{\delta}{3}} = \frac{-\delta^2  m}{2\mu n r (2 + \delta)}.$$
To ensure the probability is less than a small failure tolerance $\eta$ (in this lemma we use $\eta = n^{-3}$), we set
$$2nr \exp \left( \frac{-\delta^2 m}{2 \mu n r (2 + \delta)} \right) \le \eta$$
We take the logarithm, rearrange for $m$, and get
$$m \ge \frac{2 \mu n r (2 + \delta)}{\delta^2} \log\left(\frac{2nr}{\eta}\right).$$
If we treat $\delta$ as a fixed constant (such as e.g., $\delta = 1/2$), the term $\frac{2(2 + \delta)}{\delta^2}$ becomes a constant $C_1$. Since $\log(2nr/\eta) = \log(2r n^4) = C_2 \log n$, we get (by absorbing both constants into $C$)
\begin{equation}\label{eq:m_bound}
m \ge C \mu n r \log n.
\end{equation}
 
\end{proof}
We point out that the particular choice of $\delta \in (0,1)$ only affects the constant $C$ in~\eqref{eq:m_bound}.

\begin{lemma}\label{le:leakagecompletion}
Given the assumptions of Theorem~\ref{th:matrixcompletion}, 
suppose the number of measurements satisfies $m \ge C \mu n r \log n$.  Then with probability at least $1 - 2n^{-3}$ for every $H \in \ker(\mathcal{P}_\Omega)$ 
$$\|P_{\mathcal{T}}(H)\|_F \le \Cleak \|P_{\mathcal{T}^\perp}(H)\|$$
with leakage constant $\Cleak \le 4 \sqrt{\frac{nr \log n}{m}}$.
\end{lemma}

\begin{proof}
Let $H \in \ker(\mathcal{P}_\Omega)$. As meanwhile usual, we decompose $H$ into $H_{\mathcal{T}} + H_{\mathcal{T}^\perp}$ where $H_{\mathcal{T}} = P_{\mathcal{T}}(H)$.
A simple calculation gives 
\begin{equation}\label{eq:Hbound1}
\|H_{\mathcal{T}}\|_F \le \|(\mathcal{R}_{\mathcal{T}})^{-1}\| \cdot \|\mathcal{R}_{\mathcal{T}, \mathcal{T}^\perp}\|_{\text{op} \to \text{F}} \cdot \|H_{\mathcal{T}^\perp}\|.
\end{equation}

Define the event $\mathcal{E}_1 = \{ \|\mathcal{R}_{\mathcal{T}} - \mathcal{P}_{\mathcal{T}}\| \le 1/2 \}$.
We aim to apply the matrix Bernstein inequality to the sum of operators $\mathcal{Z}_k = \frac{1}{m} (\frac{1}{p} \mathcal{P}_{\mathcal{T}} \mathcal{P}_{ij} \mathcal{P}_{\mathcal{T}} - \mathcal{P}_{\mathcal{T}})$. We compute
$R = \|\mathcal{Z}_k\| \le \frac{n^2}{m} \frac{2\mu r}{n} = \frac{2\mu n r}{m}$. And $\sigma^2 \le \frac{2\mu n r}{m}$. Hence, by matrix Bernstein, if $m \ge C \mu n r \log n$, then $\mathbb{P}(\mathcal{E}_1^c) \le n^{-3}$.
On $\mathcal{E}_1$, the minimum eigenvalue $\lambda_{\min}(\mathcal{R}_{\mathcal{T}}) \ge 1 - 1/2 = 1/2$. Thus
$$\|(\mathcal{R}_{\mathcal{T}})^{-1}\| = \frac{1}{\lambda_{\min}(\mathcal{R}_{\mathcal{T}})} \le 2.$$

We define the event $\mathcal{E}_2 = \{ \|\mathcal{R}_{\mathcal{T}, \mathcal{T}^\perp}\|_{\text{op} \to \text{F}} \le \epsilon \}$.
Consider the operator $X \mapsto \mathcal{P}_{\mathcal{T}} \mathcal{R} \mathcal{P}_{\mathcal{T}^\perp}(X)$. This is a sum of $m$ independent operators $\mathcal{Y}_k(X) := \frac{n^2}{m} \langle e_i e_j^\top, X \rangle \mathcal{P}_{\mathcal{T}}(e_i e_j^\top)$, restricted to $X \in \mathcal{T}^\perp$.
Note that the induced norm $\|\cdot\|_{\text{op} \to \text{F}}$ of a rank-1 map $\langle A, \cdot \rangle B$ is $\|A\|_* \|B\|_F$. Hence
$$\|\mathcal{Y}_k\|_{\text{op} \to \text{F}} = \frac{n^2}{m} \underbrace{\|e_i e_j^\top\|_*}_{=1} \underbrace{\|\mathcal{P}_{\mathcal{T}}(e_i e_j^\top)\|_F}_{\le \sqrt{2\mu r/n}} \le \frac{n \sqrt{2\mu n r}}{m} =: R$$
For the variance $\sigma^2$ of $\mathcal{Y}_k$ we compute
$$\sigma^2 = m \cdot \mathbb{E} \left[ \left(\frac{n^2}{m}\right)^2 (1)^2 \|\mathcal{P}_{\mathcal{T}}(e_i e_j^\top)\|_F^2 \right] = \frac{n^2}{m} \sum_{i,j} \frac{1}{n^2} \|\mathcal{P}_{\mathcal{T}}(e_i e_j^\top)\|_F^2$$
Since $\sum_{i,j} \|\mathcal{P}_{\mathcal{T}}(e_i e_j^\top)\|_F^2 = \text{dim}(\mathcal{T}) = r(n_1 + n_2 - r) \le 2nr$ we get $\sigma^2 \le \frac{2nr}{m}$.
Applying matrix Bernstein gives for $\delta = \sqrt{\frac{4nr \log n}{m}}$, the probability $\mathbb{P}(\mathcal{E}_2^c) \le 2n \exp(\frac{-\delta^2/2}{\sigma^2 + L\delta/3}) \le n^{-3}$.

On the event $\mathcal{E}_1 \cap \mathcal{E}_2$:
$\|(\mathcal{R}_{\mathcal{T}})^{-1}\| \le 2$
$\|\mathcal{R}_{\mathcal{T}, \mathcal{T}^\perp}\|_{\text{op} \to \text{F}} \le \sqrt{\frac{4nr \log n}{m}}$
We substitute these into the inequality~\eqref{eq:Hbound1} and obtain
$$\|H_{\mathcal{T}}\|_F \le 4 \sqrt{\frac{nr \log n}{m}}  \|H_{\mathcal{T}^\perp}\|.$$
By the union bound, $\mathbb{P}(\mathcal{E}_1 \cap \mathcal{E}_2) \ge 1 - 2n^{-3}$. 
   
\end{proof}

With these two lemmata in place, we are now ready to prove Theorem~\ref{th:dualnuclear}.

\begin{proof}[Proof of Theorem~\ref{th:matrixcompletion}]
In the proof $C$ denotes a constant that may change throughout  the derivations.
By Theorem~\ref{th:dualnuclear}, $X_0$ is the unique minimizer if $\alpha < \frac{1-\beta}{\Cleak}$, and 
by Lemma~\ref{le:leakagecompletion} we have
$\Cleak \le  4\sqrt{\frac{ n r \log n}{m}}$.

We will now construct an $(\alpha,\beta)$-approximate dual certificate iteratively via the golfing scheme. To that end we divide the sampling set $\Omega$ into $k$ independent batches $\Omega_1, \Omega_2, \dots, \Omega_k$ of about equal size $m/k$. 
On the one hand, the size of the batches will have
to be chosen large enough to allow for the application of
the matrix-valued concentration of measure bounds tailored for independent
random variables. On the other hand, the batch size $m/k$ cannot be too large, since the speed of convergence is exponential in $k$.
We will determine the proper value of $k$ later.

Let $\mathcal{R}_j$ be the normalized sampling operator for batch $j$.
While taking $Y=UV^T$ as the dual certificate does not work since the dual certificate must live in the range of the sampling operator, it still serves as good first initial guess. The idea of the golfing scheme is to iteratively refine this initial guess until condition~\eqref{eq:dualalpha} is fulfilled.
We proceed as follows:
\begin{enumerate}
\item Initialization: $Q_0 := UV^\top$
\item Iteration: $Y_j := \mathcal{R}_j Q_{j-1}$ and $Q_j := Q_{j-1} - \mathcal{P}_{\mathcal{T}} Y_j$, \quad for $j=1,\dots, k$
\item Final approximate certificate: $Y = \sum_{j=1}^k Y_j$
\end{enumerate}
By expanding the recursion, it is easy to see that $Y$ satisfies
$$\mathcal{P}_{\mathcal{T}} Y = UV^\top - Q_k.$$
We want to show that $P_{\mathcal{T}}Y$ is sufficiently close to $UV^\top$, which is equivalent to showing that $Q_k$ is sufficiently small.
From Lemma~\ref{le:matrixinjectivity}, we know that for each batch
$$\|\mathcal{P}_{\mathcal{T}} - \mathcal{R}_j \mathcal{P}_{\mathcal{T}}\| \le \delta$$
with high probability. Therefore,
$$\|Q_j\|_F = \|(\mathcal{P}_{\mathcal{T}} - \mathcal{R}_j \mathcal{P}_{\mathcal{T}}) Q_{j-1}\|_F \le \delta \|Q_{j-1}\|_F.$$
Iterating this $k$ times gives
$$\|Q_k\|_F \le \delta^k \|Q_0\|_F = \delta^k \sqrt{r}.$$
To make $\alpha = \|Q_k\|_F \le n^{-2}$, we set $\delta = 1/2$ and $k \approx \log(n^2 \sqrt{r}) \approx \log n$.

We must also ensure that the ``leakage'' of the approximate certificate into the orthogonal space $\mathcal{T}^\perp$ remains small, i.e., we want $\beta = \|P_{\mathcal{T}^\perp} Y\| < 1/2$.
Expanding $Y$ gives
$$P_{\mathcal{T}^\perp} Y = \sum_{j=1}^k \mathcal{P}_{\mathcal{T}^\perp} \mathcal{R}_j Q_{j-1}.$$
Each term $W_j = \mathcal{P}_{\mathcal{T}^\perp} \mathcal{R}_j Q_{j-1}$ is a random matrix. 
 By  induced norms,
$$\|W_j\| \le \|\mathcal{P}_{\mathcal{T}^\perp} \mathcal{R}_j \mathcal{P}_{\mathcal{T}}\|_{F \to \text{op}} \|Q_{j-1}\|_F.$$

We can expand $\|\mathcal{P}_{\mathcal{T}^\perp} \mathcal{R}_j \mathcal{P}_{\mathcal{T}}$ into a sum of independent random rank-1 operators and apply the (rectangular) matrix Bernstein inequality. A  standard calculation, that is left to the reader, yields
$$\mathbb{P} \left( \|\mathcal{P}_{\mathcal{T}^\perp} \mathcal{R}_j \mathcal{P}_{\mathcal{T}}\|_{F \to \text{op}} \ge \gamma \right) \le 2n \exp \left( \frac{-\gamma^2 / 2}{\sigma^2 + R\gamma / 3} \right),$$
where $R  \le \frac{n^2}{m_j}$ and $\sigma^2 \le \frac{C \mu n r}{m_j}$.
We solve for $\gamma$ with $m_j \ge C \mu n r \log n$ and get
$$\gamma \le C \sqrt{\frac{\mu n r \log n}{m_j}}.$$
Summing the geometric series gives
$$\|{\mathcal P}_{{\mathcal T}^\perp} Y\| \le \sum_{j=1}^k \|W_j\| \le  \gamma \sum_{j=1}^k \|Q_{j-1}\|_F \le \gamma  \sqrt{r} \frac{1}{1-\delta},$$
with high probability.
Since $\delta = 1/2$ and using our $\gamma$ bound we get
$$\beta \le C\sqrt{\frac{\mu n r^2 \log n}{m_j}} < 1.$$
Choosing  $m \ge C \mu n r \log^2 n$ and using the incoherence assumption implies that $\beta \le 1/2$, and also that
$1-\beta > \alpha \Cleak$. This means all conditions for the approximate dual certificate in Theorem~\ref{th:dualnuclear} are met with high probability, and the nuclear norm minimizer is unique and equal to $X_0$. 

\end{proof}

The extra $\log n$ factor in the number of measurements $m$ (compared to the $\log n$ in standard concentration) arises from the number of iterations $k$ in the golfing scheme needed to make the error $\alpha$ sufficiently small.

One can find many variations and refinements of Theorem~\ref{th:dualnuclear} in the literature, see e.g.~\cite{candes2010matrix,candes2010power,keshavan2010matrix,candes2011robust,fuchs2022proof,WrightMa,gandy2011tensor,liu2012tensor} for a small sample. For instance, recent work~\cite{chen2015incoherence} has shown that the joint coherence condition 
$\|UV^\top\|_\infty \le \sqrt{\frac{\mu r}{n_1 n_2}}$
is not necessary, just the standard incoherence conditions (small row and column coherence) suffice. Moreover,  is not difficult to extend Theorem~\ref{th:matrixcompletion} to the case of noisy measurements. The reader will find various such extensions in the aforementioned literature on matrix completion.

\section{Quantum state tomography}\label{s:quantum}

In quantum mechanics, the state of a system is described by a density matrix $\rho \in \mathbb{C}^{d \times d}$, cf.~\cite{nielsen2010quantum}. Standard Quantum State Tomography (QST) aims to reconstruct this matrix by performing a series of measurements. However, as the number of qubits $q$ increases, the dimension of the Hilbert space $d = 2^q$ grows exponentially, making traditional reconstruction methods computationally and physically prohibitive~\cite{paris2004quantum}.

Low-rank matrix recovery offers a powerful solution to this ``curse of dimensionality'' when the underlying quantum state is pure or nearly pure~\cite{gross2010quantum}.

A pure quantum state is represented by a unit vector $|\psi\rangle$. Its corresponding density matrix is\footnote{In Bra-ket notation we can write the density matrix  as $\rho = |\psi\rangle \langle \psi|$, which is standard in physics. Here, we stick with the linear algebra notation.}
$$\rho = \psi \psi^{\ast}.$$
A state density matrix is Hermitian, positive-semidefinite, and has unit-trace. Pure states correspond to rank-one density matrices. In many practical settings---quantum optics, trapped ions, superconducting qubits---the prepared quantum state is either pure or nearly pure. Even in the presence of environmental noise or decoherence, many physical systems exist in ``mixed states'' that are dominated by a few principal components, making $\rho$ effectively low-rank. This observation places quantum state tomography squarely within the scope of low-rank matrix recovery.

This structural property allows us to apply the principles of compressive sensing: we can reconstruct $\rho$ using far fewer measurements than the $d^2$ required by classical linear inversion~\cite{gross2010quantum}.

In QST, measurements are typically performed using an observable $A$. The expectation value of the measurement is given by the Born rule
$$y = \text{Tr}(A\rho).$$
To recover the state, we collect a set of measurement results $y_1, y_2, \dots, y_m$ from different observables $\{A_i\}_{i=1}^m$. This can be modeled as a linear operator $\mathcal{A}(\rho) = y$.
We thus formulate the QST reconstruction problem as a convex optimization problem
\begin{align}\label{eq:quantumtomo}
\begin{split}
\text{minimize} & \quad  \|\rho\|_*  \\
\text{subject to} & \quad \mathcal{A}(\rho) = y \\
& \quad \rho \succeq 0.
\end{split}
\end{align}
Since $\rho \succeq 0$, the nuclear norm $\|\rho\|_*$ is simply the trace $\text{Tr}(\rho)$. Given that the trace is fixed to 1 for any valid density matrix, the nuclear norm minimization problem~\eqref{eq:quantumtomo} effectively collapses into a feasibility problem, namely the search for a PSD matrix that satisfies the measurement constraints. Surprisingly however, from a theoretical point of view, the feasibility problem is not easier to analyze than the nuclear norm minimization problem.

 A standard choice for QST  are {\em Pauli measurements}, which  are not only theoretically appealing, but also provide a very practical way to extract information from a quantum processor because they align with how quantum hardware is physically built and controlled~\cite{altepeter2005photonic}.

\medskip
\noindent
\textbf{Pauli measurements:} For a single qubit, define the Pauli matrices~\cite{nielsen2010quantum,gross2011recovering}
$$
\sigma_0 = I =
\begin{bmatrix}
1 & 0\\
0 & 1
\end{bmatrix},\quad
\sigma_x =
\begin{bmatrix}
0 & 1\\
1 & 0
\end{bmatrix},\quad
\sigma_y =
\begin{bmatrix}
0 & -i\\
i & 0
\end{bmatrix},\quad
\sigma_z =
\begin{bmatrix}
1 & 0\\
0 & -1
\end{bmatrix}.
$$
For a $q$-qubit system $(d = 2^q)$, the Pauli group
$$
\mathcal{P}_n = \big\{ \sigma_{a_1} \otimes \cdots \otimes \sigma_{a_q} : a_k \in \{0,x,y,z\} \big\}
$$
forms a basis of $\mathbb{C}^{d\times d}$ with cardinality $d^2$ (see Exercise~\ref{ex:pauli}).
We define the normalized Pauli operators by
$$
P_i := \frac{1}{\sqrt{d}} \sigma_{i_1} \otimes \cdots \otimes \sigma_{i_q}.
$$
These operators satisfy
$$
\operatorname{tr}(P_i^\ast P_j) = \delta_{ij},
$$
thus $\{P_i\}_{i=1}^{d^2}$ is an orthonormal basis for the Hilbert–Schmidt inner product (see Exercise~\ref{ex:pauli}).
Therefore, any density matrix admits the expansion
$$
\rho = \sum_{i=1}^{d^2} \langle P_i, \rho \rangle P_i,
\quad
\langle P_i, \rho \rangle = \operatorname{tr}(P_i \rho).
$$

In the noisefree case one measures expectation values of Pauli observables
$$
y_i = \operatorname{tr}(P_{i} \rho),
\quad i=1,\dots,m,
$$
where the $P_i$ are drawn uniformly at random from the Pauli group (and then normalized). Define the sampling operator
$$
\mathcal{A}(\rho)_i := \operatorname{tr}(P_{i} \rho).
$$
This defines a random linear measurement ensemble.

\begin{theorem}\label{th:QST}
Let $\rho \in \mathbb{C}^{d \times d}$ be a density operator of rank $r$, where $d=2^q$. Let $\mathcal{A}: \mathbb{C}^{d \times d} \to \mathbb{R}^m$ be a measurement operator that samples $m$ Pauli observables $\{P_i\}_{i=1}^m$ uniformly at random from the complete Pauli basis (excluding the identity). If
$$m \ge C rd \log^2 d,$$
then with high probability, $\rho$ is the unique minimizer of the nuclear norm minimization problem
$$\min \|X\|_* \quad \text{subject to} \quad \mathcal{A}(X) = \mathcal{A}(\rho).$$    
\end{theorem}

\medskip
\noindent
\textbf{\em Proof sketch.}
The proof follows overall the same architecture as the proof of Theorem~\ref{th:matrixcompletion}. Therefore, instead of giving a detailed proof that would duplicate many derivations of the proof of Theorem~\ref{th:matrixcompletion}, it is more instructive to give a proof sketch which highlights the key differences.

We start with a trivial, but still useful observation.
For a rank-$r$ state $\rho$ with singular value decomposition $U \Sigma V^*$, we have $V=U$, since $\rho$ is Hermitian. Hence, the tangent space $\mathcal{T}$ is defined by the span of matrices $\{Uz^*, wU^*\}$.

Because Pauli matrices are unitary and perfectly ``spread out'' in the Hilbert-Schmidt sense, the incoherence $\mu$ is effectively constant ($\mu = 1$). Specifically, for any Pauli matrix $P$
$$\|\mathcal{P}_{\mathcal{T}}(P)\|_F^2 \le \frac{2rd}{d^2}  \|P\|_F^2 = 2r.$$
This replaces the $\frac{\mu nr}{m}$ scaling in matrix completion with a much more favorable $r/d$ relative scaling.

We define the sampling operator $\mathcal{R}: \mathbb{C}^{d \times d} \to \mathbb{C}^{d \times d}$ as
$$\mathcal{R}(X) = \frac{d}{m} \sum_{k=1}^m \langle P_k, X \rangle P_k.$$
Note the normalization factor $d/m$ arises because 
$$\mathbb{E}[\langle P, X \rangle P] = \frac{1}{d^2-1} \sum \langle P, X \rangle P \approx \frac{1}{d} X.$$

Using the matrix Bernstein inequality as before, the variance $\sigma^2$ is bounded by the incoherence. Since $\|P\|=1$, we find that for $m \ge C r d \log^2 d$
$$\|\mathcal{P}_{\mathcal{T}} \mathcal{R} \mathcal{P}_{\mathcal{T}} - \mathcal{P}_{\mathcal{T}}\| \le \delta$$
with high probability.

The most critical technical change appears in the derivation of the leakage constant $\Cleak$. In matrix completion, $\Cleak$ grew as $\sqrt{n}$.
In QST, the Pauli basis elements $P_i$ are unitaries.
When we compute $\|\mathcal{P}_{\mathcal{T}} \mathcal{R} \mathcal{P}_{\mathcal{T}^\perp}\|_{\text{op} \to F}$, we look at the variance
$$\sigma^2 = \left\| \sum \mathbb{E} [ \mathcal{L}_k^* \mathcal{L}_k ] \right\| \approx \frac{d^2}{m^2} \sum_{P \in \Omega} |\langle P, H \rangle|^2  \|\mathcal{P}_{\mathcal{T}}(P)\|_F^2.$$
Since $\|\mathcal{P}_{\mathcal{T}}(P)\|_F^2 \le 2r$, and $\sum |\langle P, H \rangle|^2 \le d \|H\|_F^2 \le d^2 \|H\|^2$ we have
$$\sigma^2 \le \frac{dr}{m}  \|H\|^2$$
Thus, $\Cleak \approx \sqrt{\frac{d r \log d}{m}}$. Plugging in $m = C d r \log^2 d$, we find
$$\Cleak \approx \sqrt{\frac{dr \log d}{C dr \log^2 d}} \approx C.$$

The key difference is that unlike the entry-wise case in matrix completion, for Pauli measurements, the leakage constant is independent of the dimension $d$ (up to log-factors, which we did hide in the constant).

To construct $Y$ via the golfing scheme, we initialize $Q_0 = U U^{\ast}$ and iterate $Q_j = Q_{j-1} - \mathcal{P}_{\mathcal{T}} \mathcal{R}_j Q_{j-1}$. The key properties of the approximate dual certificate we need to fulfill are:
\begin{itemize}
\item Alignment on $\Tn$  (bound for $\alpha$): After $k = \log d$ steps, $\|Q_k\|_F \le d^{-2}$. 
\item Constraint on $\Tn^\perp$  (bound for $\beta$): We need $\|\mathcal{P}_{\mathcal{T}^\perp} \sum \mathcal{R}_j Q_{j-1}\| < 1/2$. The term $\|\mathcal{P}_{\mathcal{T}^\perp} \mathcal{R}_j \mathcal{P}_{\mathcal{T}}\|_{F \to \text{op}}$ scales as $\sqrt{\frac{d r \log d}{m/k}}$.
With $m \sim d r \log^2 d$, each batch gives $\gamma \approx \frac{1}{\sqrt{r}}$, so $\beta \le \sum \gamma \delta^j \sqrt{r} \approx 1$.
\end{itemize}

Applying the same line of argument as in the matrix completion proof we conclude 
$$\|\rho+H\|_* - \|\rho\|_* \ge (1 - \beta) \|\mathcal{P}_{\mathcal{T}^\perp}H\|_* - \alpha \Cleak \|\mathcal{P}_{\mathcal{T}^\perp}H\|.$$
Because $\Cleak$ is constant and $\alpha \le d^{-2}$, the positivity of $(1-\beta)$ dominates for any $H \in \ker(\mathcal{R})$, which completes our proof sketch. \qed

\medskip

\begin{remark}
The semidefinite program in Theorem~\ref{th:QST} differs from the semidefinite program in~\eqref{eq:quantumtomo} insofar as it does not include the PSD and trace constraints. So, why did we not make use of them in Theorem~\ref{th:QST}?
While including those constraints would make the optimization problem perhaps easier to solve numerically, it does not change the sample complexity $m$ required for unique recovery in the noiseless case. In a nutshell, the bottleneck for recovery is not the ``shape'' of the set (PSD vs.\ general), but the nullspace of the sampling operator ${\mathcal R}$.
\end{remark}

\section{Phase retrieval}

Phase retrieval---the problem of recovering a function from the squared
magnitude of its Fourier transform---arises in numerous applications such as X-ray crystallography~\cite{Mil90,Har93}, diffraction imaging~\cite{GS72}, optics~\cite{Walther_phaseretrievaloptics}, fiber optic communications~\cite{chou2022phase}, quantum mechanics~\cite{Cor06}, astronomy~\cite{DF87},
and biomedical imaging~\cite{dierolf2010ptychographic}.
This problem has confounded engineers, physicists, and mathematicians for over a century. 
Recently, phase retrieval has seen a resurgence in research activity, ignited by new imaging modalities~\cite{neutze2000potential,chapman2007femtosecond} and novel mathematical concepts~\cite{CESV2013,FS2020}. 

There are many ways in which one can pose the phase-retrieval problem,
for instance depending upon whether one assumes a continuous or
discrete-space model for the signal. Here, we consider finite
length signals (one-dimensional or multi-dimensional) for simplicity,
and because numerical algorithms ultimately operate with digital data.
To fix ideas, suppose we have a 1D signal
$x = (x_0, x_1,$ $\ldots, x_{n-1}) \in \C^n$ and write its Fourier transform as
\begin{equation}
  \label{eq:fourier}
  {\hat x}(\omega) =
  \frac{1}{\sqrt{n}} \sum_{t=0}^{n-1} x_t e^{-2\pi i\omega t/n},
\quad \omega \in \Omega.
\end{equation}
Here, $\Omega$ is a grid of sampled frequencies and an important
special case is $\Omega = \{0,1, \ldots, n-1\}$ so that the mapping is
the classical unitary discrete Fourier transform (DFT).   
 The phase retrieval
problem consists in finding $x$ from the intensity measurements
$|{\hat x}(\omega)|^2$, $\omega \in \Omega$.  

Without further information about the
unknown signal $x$, this problem is ill-posed since there are many
different signals whose Fourier transforms have the same
magnitude. Clearly, if $x$ is a solution to the phase retrieval
problem, then (1) $c x$ for any scalar $c \in \C$ obeying $|c| = 1$ is
also solution, (2) the ``mirror function'' or time-reversed signal
$\bar{x}(-t \text{ mod } n)$ where $t = 0,1, \ldots, n-1$ is also
solution, and (3) the shifted signal $x(t - a \text{ mod } n)$ is also
a solution. From a physical viewpoint these ``trivial associates'' of
$x$ are acceptable ambiguities. But in general infinitely many
solutions can be obtained from $\{|\hat{x}[\omega]|^2 : \omega \in
\Omega\}$ beyond these trivial associates, making it a rather challenging ill-posed problem.

For mathematical purposes, it is convenient to formulate the phase retrieval problem in a more general way as follows:
Given information about the squared modulus of the inner product
between the signal $x \in \C^n$ and some vectors $a_k \in \C^n$, namely,
\begin{equation}
\label{eq:data}
y_k = |\langle x,a_k\rangle |^2, \quad k = 1, \ldots, m,
\end{equation}
we want to recover $x$ from $\{y_k\}_{k=1}^m$. Here, successful recovery means that we find a solution $x^*$ which differs from $x$ only by a multiplicative constant of modulus 1, i.e., $x^* = cx$, where $c\in\C$ with $|c|=1$.

The key observation is that quadratic measurements can be lifted up and
interpreted as linear measurements about the rank-one matrix $X = x
x^*$, see~\cite{balan2009painless,CESV2013}. Indeed,
\begin{equation}\label{tracemeasurements}
|\langle a_k, x\rangle |^2 = \trace(x^* a_k a_k^* x) = \trace(a_k a_k^* x x^*) = \trace(A_k X),
\end{equation}
where $A_k := a_k a_k^*$.
Adopting the notational conventions from~\eqref{tracemeasures} and~\eqref{tracemeasureop}, one can
express the data collection $y_k = |\langle x, a_k\rangle |^2 =\trace(A_k X)$, $k=1,\dots,m$ as
\begin{equation}
y = \mathcal{A}(x x^*).
\end{equation}

Hence, the phase retrieval problem can be cast as low-rank matrix recovery problem~\cite{CSV2013,CESV2013}
\begin{equation}\label{eq:phaserank}
  \begin{array}{ll}
    \text{minimize}   & \quad \rank(X)\\
    \text{subject to} & \quad  \cA(X) = y \\
& \quad X \succeq 0.
\end{array}
\end{equation}
Indeed, we know that a rank-one solution exists so the optimal $X$ has
rank one. We then factorize the solution as $x x^*$ in order
to obtain solutions to the phase-retrieval problem. This gives $x$ up
to multiplication by a unit-normed scalar.

As the reader will expect at this point, the idea is now  to solve~\eqref{eq:phaserank} via a convex relaxation. 
Thus,  we solve
\begin{equation}
\label{eq:tracemin}
 \begin{array}{ll}
    \text{minimize}   & \quad \trace(X)\\
    \text{subject to} & \quad  \cA(X) = y\\
& \quad X \succeq 0.
\end{array}
\end{equation}
The program \eqref{eq:tracemin} is a semidefinite program  in
standard form.
If the solution has rank one, we factorize it as above to recover our
signal $x$ (up to a global constant of magnitude 1).
This method which lifts the problem of vector recovery from
quadratic constraints up into that of recovering a rank-one matrix from
affine constraints via semidefinite programming is known under the
name {\em PhaseLift} \cite{CSV2013,CESV2013}.

The phase retrieval problem, when rephrased as rank-one matrix recovery problem, bears many similarities with QST for pure states\footnote{There are of course important and fundamental limitations to this similarity. Phase retrieval is a signal reconstruction problem with algebraic ambiguity, whereas quantum state tomography is a physical inference problem constrained by quantum mechanics. Quantum tomography imposes physical admissibility constraints, involves structured and hardware-limited measurements, admits higher-rank states, and is governed by fundamentally different noise models. These differences necessitate distinct algorithmic and theoretical treatments despite the shared convex-analytic framework.}. But there is a crucial difference, which prevents us to simply incorporate the theoretical treatment of phase retrieval in  Theorem~\ref{th:QST}.
Unlike Pauli measurements which are unitary, the measurement matrices $a_j a_j^*$ are ``spiky'' in the spectral domain (their only eigenvalue is $\|a_j\| \approx n)$. This complicates matters in the theoretical analysis, as we will see below.

In it not difficult to see that we can recover $x$ from $m=n^2$ many measurements (for properly chosen $a_k$). Yet, this is neither interesting nor practical, since $x$ has only $n$ unknowns. It has been shown that for $x \in \C^n$ about $4n$ measurements are necessary in general to recover $x$ \cite{balan2006signal,bandeira2014saving}. But this does not mean that this can be done in a numerically feasible way. In the next theorem will show that under mild conditions  PhaseLift can recover $x$ exactly (up to a global phase factor) with high probability, provided that the number of measurements is on the order of $n \log n$.

\begin{theorem}\cite{CSV2013}
\label{th:phaselift}
Consider an arbitrary signal $x$ in $\RR^d$ or $\C^d$. Let the measurement vectors $a_k$  be sampled independently and uniformly at
random on the unit sphere, and suppose that the number of measurements obeys $m \ge C \, d \log^2 d$, where
$C$ is a sufficiently large constant. Then with high probability the solution to the trace-minimization program is exact
with high probability in the sense that \eqref{eq:tracemin} has a
unique solution obeying
\begin{equation}
  \hat X = x x^\ast.
\end{equation}
\end{theorem}

Before we proceed to the proof of this theorem, a few remarks are in order.
At first glance, the phase retrieval problem with Gaussian measurement vectors $a_k$ appears strikingly similar to the generic low rank matrix recovery problem we studied in Chapter~\ref{ss:matrixrecoverygaussian} successfully via descent cone analysis. However, there is a crucial difference. Here, each measurement matrix $A_k = a_k a_k^*$ is a rank-one matrix, which is far from a generic random matrix (for example a matrix with
standard normal entries is almost surely never rank-deficient).
While a descent-cone based proof is still possible under appropriate assumptions~\cite{fuchs2022proof}, it is considerably more involved than the dual certificate approach.

The original proof of Theorem~\ref{th:phaselift} in~\cite{CSV2013} constructs an approximate dual certificate in a clever way without resorting to the golfing scheme.  However, this construction is rather technical and beyond the scope of our discussion here. We will instead pursue a proof that involves the by now familiar golfing scheme.
In~\cite{candes2014solving} the sampling complexity $m \ge C \, d \log^2 n$ has been improved to the information-theoretic optimum (modulo the constant) $m \ge C \, m$.

Before proceeding to the proof of Theorem~\ref{th:phaselift}, we need some preparation.
We will make use of the following approximate $\ell_1$-isometry bound on the measurement operator $\cA$.

\begin{lemma}[Lemma 3.1 in~\cite{CSV2013}]\label{le:A1}
Fix any $\delta$ in $(0,1/2)$ and assume $m \ge 20 \delta^{-2} \,
  d$. Then for all unit vectors $u$,
\begin{equation}
\label{eq:A1}
(1-\delta) \le \frac{1}{m} \|\cA(uu^*)\|_* \le (1+\delta)
\end{equation}
on an event $E_\delta$ of probability at least $1-2e^{-m
  \epsilon^2/2}$, where $\delta/4 = \epsilon^2 + \epsilon$. On the
same event,
$$
(1-\delta) \|X\|_* \le \frac{1}{m} \|\cA(X)\|_* \le (1+\delta) \|X\|_*
$$
for all positive semidefinite matrices. The right inequality holds for
all Hermitian matrices.
\end{lemma}
\begin{proof}  Let $Z^*$ be the $n \times
  m$ matrix with the measurement vectors $a_i$'s as columns. Then
$$
\|\cA(uu^*)\|_* = \sum_i |\langle a_i, u\rangle|^2 = \|Z u\|^2
$$
so that 
$$
\sigma^2_{\text{min}}(Z) \le \|\cA(uu^*)\|_* \le
\sigma^2_{\text{max}}(Z). 
$$
The claim is a consequence of deviations bounds concerning
the singular values of Gaussian random matrices which we explored in Chapter~\ref{c:probability-matrixconcentration},
namely,
\begin{align*}
\P\left(\sigma_{\text{max}} (Z) > \sqrt{m} + \sqrt{d} + t\right)  & \le e^{- t^2/2}\\
\P\left(\sigma_{\text{min}}(Z) < \sqrt{m} - \sqrt{d} - t\right)  & \le e^{- t^2/2}.
\end{align*}
The conclusion follows from taking $m \ge \epsilon^{-2} \, n$ and $t =
\sqrt{m} \epsilon$ (and from $\epsilon^2 \geq \delta^2/20$ for $0 <
\delta \le 1/2$). For the second part of the lemma, observe that $X
= \sum_j \lambda_j u_j u_j^*$ with nonnegative eigenvalues
$\lambda_j$ so that
\[
\|\cA(X)\|_* = \sum_j \sum_i \lambda_j |\langle u_j,a_i\rangle|^2 = \sum_j
\lambda_j \|\cA(u_j u_j^*)\|_*.
\]
The claim follows from \eqref{eq:A1}. The last claim is a consequence
of $\|\cA(X)\|_* \le \sum_j \sum_i |\lambda_j| |\langle u_j,a_i\rangle|^2$
together with $\sum_j |\lambda_j| = \|X\|_*$. 
\end{proof}

We will also need a bound on the leakage constant $\Cleak$, which is provided by the following lemma.
\begin{lemma}\label{le:phaseleakage}
Let $\mathcal{A}$ be as defined in Theorem~\ref{th:phaselift} and let $\mathcal{T}$ be the tangent space at a rank-1 matrix $X=xx^*$. For $m \ge C d \log^2 d$, the leakage constant $\Cleak$ satisfies
$$\Cleak := \frac{\|\mathcal{P}_{\mathcal{T}} \mathcal{A}^* \mathcal{A} \mathcal{P}_{\mathcal{T}^\perp} \|_{op \to F}}{\lambda_{\min}(\mathcal{P}_{\mathcal{T}} \mathcal{A}^* \mathcal{A} \mathcal{P}_{\mathcal{T}})} \lesssim \sqrt{d}$$
with high probability.
\end{lemma}

\begin{proof}
To bound the minimum eigenvalue, we use Jensen's inequality~\eqref{Jensen},
$$\langle M, \mathcal{A}^* \mathcal{A} M \rangle = \frac{1}{m} \sum_{k=1}^m (a_k^* M a_k)^2 \ge \Big( \frac{1}{m} \sum_{k=1}^m |a_k^* M a_k| \Big)^2.$$
Applying Lemma~\ref{le:A1} gives
$$\langle M, \mathcal{A}^* \mathcal{A} M \rangle \ge (c_0 \|M\|_*)^2.$$
Since $\|M\|_* \ge \|M\|_F$ for any matrix, for a unit-Frobenius matrix $M \in \mathcal{T}$ ($\|M\|_F=1$), we have
$$\lambda_{\min}(\mathcal{P}_{\mathcal{T}} \mathcal{A}^* \mathcal{A} \mathcal{P}_{\mathcal{T}}) \ge c_0^2 > 0.$$
This establishes that the denominator is bounded below by a constant independent of $d$.

The numerator $\|\mathcal{P}_{\mathcal{T}} \mathcal{A}^* \mathcal{A} \mathcal{P}_{\mathcal{T}^\perp} \|_{op \to F}$ requires bounding the Frobenius norm of $Z = \mathcal{P}_{\mathcal{T}} \mathcal{A}^* \mathcal{A} W$ for $W \in \mathcal{T}^\perp$ with $\|W\|=1$.
Let $\{B_i\}_{i=1}^{2d-1}$ be an orthonormal basis for $\mathcal{T}$. We have:
$$\|Z\|_F^2 = \sum_{i=1}^{2d} |\langle B_i, \mathcal{A}^* \mathcal{A} W \rangle|^2.$$
Each term in this sum is $Y_i := \frac{1}{m} \sum_k (a_k^* B_i a_k)(a_k^* W a_k)$. Since $B_i \perp W$, the expectation is $\mathbb{E}[Y_i] = \langle B_i, W \rangle = 0$.
The variance of $Y_i$ is determined by the 4th moments of the Gaussian vectors. For rank-one measurements we have
\begin{equation}\label{eq:Yvar}
\Var(Y_i) \approx \frac{1}{m} \|B_i\|_F^2 \|W\|_F^2.
\end{equation}
For $W \in \mathcal{T}^\perp$ with $\|W\|=1$, the Frobenius norm $\|W\|_F$ can be as large as $\sqrt{d-1} \approx \sqrt{d}$.

Substituting $\|B_i\|_F=1$ and $\|W\|_F \approx \sqrt{d}$ into~\eqref{eq:Yvar} gives
$$\|Z\|_F^2 \approx \sum_{i=1}^{2d} \frac{d}{m} = \frac{2d^2}{m}.$$
Taking the square root and applying an $\epsilon$-net argument, we get the numerator scaling
$$\|\mathcal{P}_{\mathcal{T}} \mathcal{A}^* \mathcal{A} W\|_{\text{op}\to F} \approx \frac{d}{\sqrt{m}}.$$

We combine the constant denominator  and the numerator scaling and obtain
$\Cleak \lesssim \frac{d / \sqrt{m}}{c_0^2} \approx \frac{d}{\sqrt{m}}$. For the regime $m = C d \log^2 d$ this yields 
$$\Cleak \lesssim \frac{d}{\sqrt{d \log^2 d}} = \frac{\sqrt{d}}{\log d} \le C' \sqrt{d}.$$
    
\end{proof}

\begin{proof}[Proof of Theorem~\ref{th:phaselift}]
Recall that to apply Theorem~\ref{th:dualnuclear} we need to verify that $(1-\beta) > \alpha \Cleak$. We already have seen that $\Cleak$ scales like $\sqrt{d}$. This means we need to make $\alpha$ sufficiently small and carefully keep track of this goal when applying the golfing scheme.

We first partition $m$ measurements into $k$ independent batches of size $n = m/k$. For each batch $j$, we define the renormalized sampling operator $\mathcal{R}_j$ given by
$$\mathcal{R}_j(X) = \frac{1}{n} \sum_{i \in \text{Batch } j} \langle a_i a_i^*, X \rangle a_i a_i^*.$$

To recover $X = xx^*$ from $y_j = \text{tr}(a_j a_j^* X)$ using the golfing scheme, we must handle the non-zero mean of the Gaussian measurements.
To that end, we define the centered operator $\tilde{\mathcal{R}}_j$ as the difference between the random sampling and its expectation
$$\tilde{\mathcal{R}}_j(X) = \mathcal{R}_j(X) - \mathbb{E}[\mathcal{R}_j(X)].$$
From the Gaussian moment property $\mathbb{E}[|a^* z|^2 aa^*] = 2zz^* + \|z\|^2 I$, we know
$$\mathbb{E}[\mathcal{R}_j(X)] = 2X + \text{tr}(X)I.$$
The centered operator for a single measurement $a$ is\footnote{Note that in~\eqref{eq:centeredoperator1} $\mathcal{I}$ is the ``operator-valued identity $\mathcal{I}(X)=X$.}
\begin{equation}\label{eq:centeredoperator1}
X_a(\cdot) = \langle aa^*, \cdot \rangle aa^* - (2{\mathcal I}  + \text{tr}(\cdot)I).
\end{equation}
To use the matrix Bernstein inequality, we need the variance of the sum of $n$ independent centered operators: $\sigma^2 = \| \sum_{i=1}^n \mathbb{E}[X_{a,i}^2] \|$.
Since the measurements are i.i.d., $\sigma^2 = n  \| \mathbb{E}[X_a^2] \|$.

To find the variance,  we must compute $\mathbb{E}[X_a^* X_a]$,
which involves the eighth moments of $a$. To see this, note that the term $\langle a a^*, X \rangle$ is a quadratic form in $a$  and the summand $X_a$ involves $\langle a a^*, X \rangle a a^*$, which is a matrix where each entry involves $a$ raised to the fourth power.
Multiplying two of these fourth-power terms together results in an expectation involving the eighth power of the Gaussian entries\footnote{If the eighth moment were infinite (as in some heavy-tailed distributions), the operator would not concentrate, and the golfing scheme would fail or one would have to choose the number of measurements much larger.}.

Since $(\mathbb{E}[X_a])^2$ is a positive semi-definite matrix, subtracting it from $\mathbb{E}[X_a]$ only reduces the eigenvalues. Therefore
$$\|\mathbb{E}[X_a^2] - (\mathbb{E}[X_a])^2\| \le \|\mathbb{E}[X_a^2]\|.$$
Therefore it suffices to bound the second moment of the uncentered operator
$\mathbb{E} \left[ |a^* X a|^2 \|a\|^2 a a^* \right]$.

But we can simplify things further.
Due to the rotation invariance of the normal distribution, we can assume that $X$ is a diagonal rank-1 matrix with $X = e_1 e_1^*$ (so $\|X\|_F = 1$) and let $v = e_1$. In the next derivations, we denote the individual entries of $a$ by $a(1),\dots, a(d).$
Hence, the expression under consideration simplifies to
$$\mathbb{E} \left[ |a(1)|^4 \left( \sum_{k=1}^d |a(k)|^2 \right) |a(1)|^2 \right] = \mathbb{E} \left[ |a(1)|^6 \sum_{k=1}^d |a(k)|^2 \right].$$
We split the sum into two parts: the term where the index $k$ matches the $X$-index ($k=1$) and the terms where it does not ($k > 1$) and get
$$\mathbb{E} \left[ |a(1)|^8 \right] + \sum_{k=2}^d \mathbb{E} \left[ |a(1)|^6 |a(k)|^2 \right].$$

Using the fact that for complex Gaussians $\mathbb{E}|a|^2 = 1$ and the univariate moment formula  $(2p-1)!!$ from Lemma~\ref{lemma:gaussianmoments}, we apply Wick's formula (Lemma~\ref{lemma:Wicksformula}) and get: 
\begin{enumerate}[label=(\roman*)]
\item Term 1 (The $k=1$ case):
This is the 8th moment of a single Gaussian.
$$\mathbb{E}[|a(1)|^8] = 7!! = 105.$$
There is one such term (and its value is independent of $d$). \\
\item Term 2 (The $k \neq 1$ cases):
Because $a(1)$ and $a(k)$ are independent for $k \neq 1$, the expectation factors as
$$\mathbb{E}[|a(1)|^6 |a(k)|^2] = \mathbb{E}[|a(1)|^6] \cdot \mathbb{E}[|a(k)|^2] = 3!! 1!! = 3,$$
where we have again applied the Gaussian moment formula. 
There are $(d-1)$ such terms in the sum.
\end{enumerate}

Adding all terms together 
and normalizing by $1/n^2$ (from the definition of $\mathcal{R}_j$) gives
$$\sigma^2 = n  \frac{1}{n^2} (3(d-1) + 105) \approx \frac{3d}{n},$$
hence $\sigma^2 = \OOO(d/n)$.

\medskip
We now apply the golfing scheme to construct $Y$ iteratively. We initialize $Q_0 = xx^*/\|x\|^2$ and then iterate $Y_j = \mathcal{R}_j(Q_{j-1})$, 
$Q_j = Q_{j-1} - \mathcal{P}_{\mathcal{T}} Y_j$ for $j=1$ to $k$. The final approximate dual certificate is $Y = \sum_{j=1}^k Y_j$.
By construction, the projection onto $\mathcal{T}$ is a telescoping sum
$\mathcal{P}_{\mathcal{T}} Y = \sum (Q_{j-1} - Q_j) = Q_0 - Q_k$. Hence, if $Q_k \to 0$, then $\mathcal{P}_{\mathcal{T}} Y \to Q_0$.

The golfing scheme relies on three critical parameters that govern convergence and certifiability, namely $\alpha, \beta$ and $\Cleak$. 

Recall that the parameter $\alpha \ge \|\mathcal{P}_{\mathcal{T}}(Y) - UV^*\|_F$ tracks how far the certificate is from the ideal subgradient on the tangent space.
Note that 
$$Q_j = Q_{j-1} - \mathcal{P}_{\mathcal{T}} \mathcal{R}_j (Q_{j-1})  = (\mathcal{P}_{\mathcal{T}} - \mathcal{P}_{\mathcal{T}} \mathcal{R}_j \mathcal{P}_{\mathcal{T}}) Q_{j-1}.$$
We will now show  that the operator $\mathcal{R}_j$ satisfies
$$\|\mathcal{P}_{\mathcal{T}} - \mathcal{P}_{\mathcal{T}} \mathcal{R}_j \mathcal{P}_{\mathcal{T}}\| \le \epsilon.$$
By Lemma~\ref{le:A1} for any $M \in \mathcal{T}$ with $\|M\|_F = 1$ we have
$$\lambda_{\min}(\mathcal{P}_{\mathcal{T}} \mathcal{R} \mathcal{P}_{\mathcal{T}}) = \inf_{\|M\|_F=1} \langle M, \mathcal{R} M \rangle = \frac{1}{m} \sum_{k=1}^m (a_k^* M a_k)^2$$
By Jensen's Inequality, we have
$$\frac{1}{m} \sum_{k=1}^m (a_k^* M a_k)^2 \ge \left( \frac{1}{m} \sum_{k=1}^m |a_k^* M a_k| \right)^2 \ge c_0^2$$
This proves that the eigenvalues of $\mathcal{P}_{\mathcal{T}} \mathcal{R} \mathcal{P}_{\mathcal{T}}$ are bounded away from zero by a constant $1 - \delta_{\text{lower}}$.

The upper bound follows from standard sub-exponential concentration. Since $a_k^* M a_k$ are sub-exponential, the sum of their squares concentrates around the mean $\mathbb{E}[\langle M, \mathcal{R} M \rangle] = \langle M, (2 \mathcal{I} + \text{Tr}(\cdot)I) M \rangle$.
For $m \ge C d \log d$, we obtain
$$\lambda_{\max}(\mathcal{P}_{\mathcal{T}} \mathcal{R} \mathcal{P}_{\mathcal{T}}) \le 1 + \delta_{\text{upper}}.$$
We thus have established that the eigenvalues of
 $\mathcal{P}_{\mathcal{T}} \mathcal{R} \mathcal{P}_{\mathcal{T}}$ acting on  $\mathcal{T}$  satisfy
$$(1-\delta) \le \lambda_i \le (1+\delta)$$
where $\delta = \max(\delta_{\text{lower}}, \delta_{\text{upper}})$.
By taking $m_j$ (the size of a single batch) to be a sufficiently large multiple of $d$, we can make $\delta \le \epsilon$, where $\epsilon$ is a desired constant. We set
$\epsilon = 1/2$. 
 
This gives us the recursive contraction
\begin{equation}\label{eq:Qbound1}
\|Q_j\|_F \le \frac{1}{2} \|Q_{j-1}\|_F \implies \|Q_k\|_F \le \left(\frac{1}{2}\right)^k \|Q_0\|_F.
\end{equation}
We know that $\|Q_0\|_F = \|UV^*\|_F = \sqrt{r} = 1$ (for rank-1). We need $\alpha \le \frac{1}{4\sqrt{d}}$ and thus $\left(\frac{1}{2}\right)^k \le \frac{1}{4\sqrt{d}}$. We can achieve this by setting
$$k \gtrsim \log d.$$

The parameter $\beta \ge \|\mathcal{P}_{\mathcal{T}^\perp}(Y)\|$ measures the contribution of the dual certificate in the orthogonal space.  In the orthogonal complement $\mathcal{T}^\perp$, we have
$$\mathcal{P}_{\mathcal{T}^\perp}(Y) = \sum_{j=1}^k \mathcal{P}_{\mathcal{T}^\perp} \mathcal{R}_j(Q_{j-1}).$$
For each batch $j$, $\mathcal{R}_j$ is a random operator. Since $Q_{j-1} \in \mathcal{T}$ and we are projecting onto $\mathcal{T}^\perp$, the expectation $\mathbb{E}[\mathcal{R}_j(Q_{j-1})]$ is zero, i.e.,
$$\mathbb{E}[\mathcal{P}_{\mathcal{T}^\perp} \mathcal{R}_j(Q_{j-1})] = \mathcal{P}_{\mathcal{T}^\perp} \mathbb{E}[\mathcal{R}_j](Q_{j-1}) = \mathcal{P}_{\mathcal{T}^\perp} (Q_{j-1}) = 0.$$
To bound the deviation of the expectation from zero, we again use the matrix Bernstein inequality. For rank-one measurements, each batch must satisfy $m_j \ge C d \log d$ to ensure
$$\|\mathcal{P}_{\mathcal{T}^\perp} \mathcal{R}_j(Q_{j-1})\| \le \delta_j \|Q_{j-1}\|_F.$$
Summing these deviations over $k$ batches gives
$$\beta \le \sum_{j=1}^k \delta_j \|Q_{j-1}\|_F \approx \delta \sum_{j=1}^k  \|Q_{j-1}\|_F .$$
Because $\|Q_j\|_F$ shrinks geometrically (see~\eqref{eq:Qbound1}), this sum is dominated by the first term. By choosing $m_j$ sufficiently large, we ensure $\delta$ is small enough (e.g., $1/4$) so that the total sum $\beta$ never exceeds $1/2$.

The last step of the proof proceeds along a well-traveled path. We need to show that $1-\beta > \alpha \Cleak$ in order to apply Theorem~\ref{th:dualnuclear}. Plugging in our bounds for $\Cleak, \alpha$, and $\beta$ we see that this condition is indeed satisfied and the proof is complete.
    
\end{proof}

\begin{remark}
Because in matrix completion we sample discrete entries, it suffers from the {\em Coupon Collector} problem\footnote{The coupon collector problem asks:
If there are $n$ distinct coupon types and each draw independently yields a uniformly random coupon, how many draws are needed to collect all $n$ types?
The answer is $\OOO(n \log n)$. Collecting the last few missing coupons dominates the total time; the process slows dramatically as the collection nears completion.}.
To ensure every row and column is sampled at least once, we need $\OOO(n \log n)$ measurements. This is why the $\log n$ factors in matrix completion are structurally different from the ``union bound'' $\log n$ factors in phase retrieval. In phase retrieval, the Gaussians are delocalized by default. In matrix completion, we need to use the incoherence property to force the matrix to be delocalized so that random sampling actually works.   
\end{remark}

It is clear that the Gaussian sampling vectors in Theorem~\ref{th:phaselift} are not useful in real-world
applications. But this result can be extended to more practical setting. For example,~\cite{candes2015phase}
studies a physically realistic setup where one can modulate the signal of interest
and then collect the intensity of its diffraction pattern, each modulation thereby
producing a {\em coded diffraction} pattern or {\em random mask}, as suggested in~\cite{CESV2013}. In this setup, a measurement vector $a_k$ is the result of a pointwise multiplication followed by a Fourier transform. 

To be more precise,  let $D_{\ell}$ be a diagonal matrix representing the $\ell$-th random mask (or code) . The entries of $D_{\ell}$ are usually random phases (e.g., $e^{2\pi i \phi}$ where $\phi$ is chosen uniformly from $[0, 1]$).
A single measurement in the $\ell$-th mask at frequency index $\omega$ is given by
$$y_{\ell, \omega} = | \langle a_{\ell, \omega}, x \rangle |^2 = \Big| \sum_{j=0}^{d-1} x_j \cdot [D_\ell]_{j,j}  e^{-2\pi i \omega  j / d} \Big|^2.$$
When moving from Gaussian $a_k$ to coded diffraction Lemmata~\ref{le:A1} and~\ref{le:phaseleakage} still hold, but the constants change. The masks introduce a form of incoherence by ensuring that the signal is spread out in the Fourier domain. One typically needs $L \approx \log^2 d$ or $\log^3 d$ masks, leading to a total $m = L d$ that still scales as $\OOO(d \text{polylog } d)$, see e.g.~\cite{candes2015phase,gross2015partial,huang2025recovery}.

\begin{figure}
\begin{center} 
\subcaptionbox{Original image}{
\includegraphics[width=50mm,height=50mm]{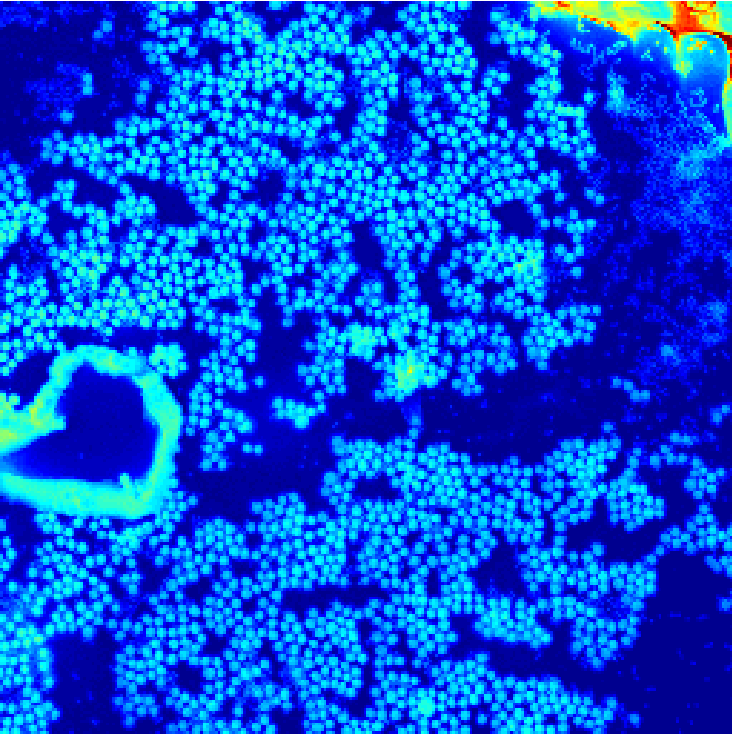}}
\qquad
\subcaptionbox{Reconstruction via PhaseLift}{
\includegraphics[width=50mm,height=50mm]{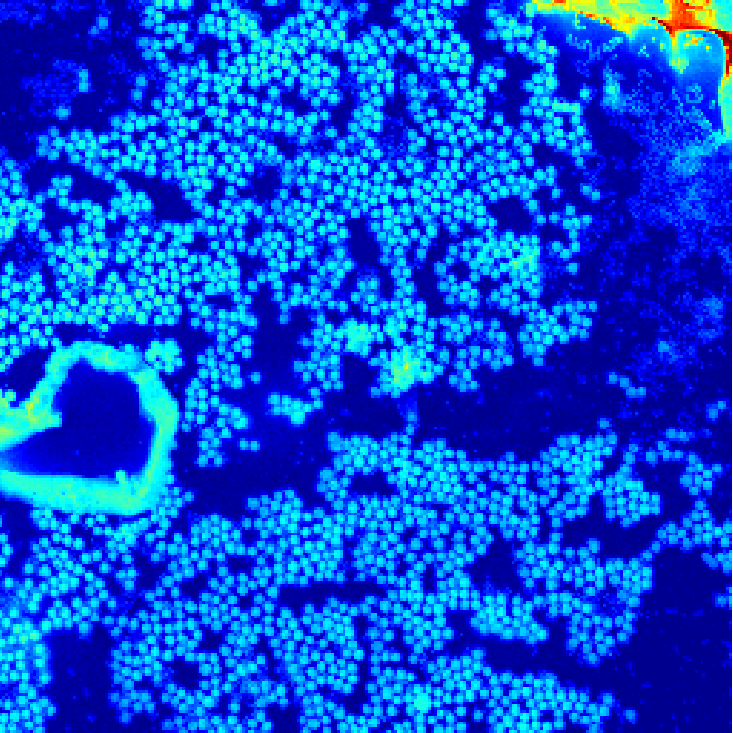}} 
\caption{Original goldballs image and reconstructions via PhaseLift, using three coded diffraction illuminations.}
\label{fig:goldballs}
\end{center}
\end{figure}

Let us explore this phase retrieval framework in a numerical example. We apply  PhaseLift to a stylized version of a setup one encounters in X-ray
crystallography or diffraction imaging.  The test image, shown in
Figure~\ref{fig:goldballs}(a) (magnitude), is a complex-valued image\footnote{Since the original image and the reconstruction are complex-valued, we only display the absolute value of each image.} of size
$256 \times 256$, whose pixel values correspond to the complex
transmission coefficients of a collection of gold balls at nanoscale embedded in a
medium (data courtesy of Stefano Marchesini from Lawrence Berkeley National Laboratory).
We use four coded diffraction illuminations, where the entries of the diffraction matrices  are either
$+1$ or $-1$ with equal probability. We use the TFOCS based implementation of PhaseLift from~\cite{CESV2013} with reweighting. 

The numerical algorithm to solve~\eqref{eq:tracemin} in the example illustrated in Figure~\ref{fig:goldballs} was implemented in Matlab using TFOCS~\cite{becker2011templates}. That implementation avoids setting up the matrix $X$ explicitly and only keeps an $n\times r$ matrix with $r \ll n$ in memory.

\section{First-order and non-convex methods for low-rank models}
\label{s:nonconvexmodels}

The impressive achievements of convex relaxation based methods has also inspired a significant body of research on
non-convex algorithms for low-rank matrix recovery with rigorous performance guarantees.
While SDP based methods provide strong theoretical guarantees, first-order and non-convex algorithms are often preferred in practice because they are more memory-efficient and faster for large-scale data.

The core idea behind most non-convex methods for low-rank recovery is to move from the space of full matrices $X \in \mathbb{R}^{n_1 \times n_2}$ to the space of low-rank factors.
The {\em Burer-Monteiro Factorization}~\cite{burer2003nonlinear} is the most common foundation for non-convex methods in this case. Instead of optimizing over a large matrix $X$, we represent it as the product of two much smaller, thin matrices $X = UV^T$
where $U \in \mathbb{R}^{n_1 \times r}$ and $V \in \mathbb{R}^{n_2 \times r}$. If the rank $r$ of $X$ is small, one only needs to store $(n_1 + n_2)r$ parameters instead of $n_1 \times n_2$. However, this factorization comes with a significant trade-off: While the original problem in $X$ was convex, the product $UV^T$ introduces non-convexity.

{\em Alternating minimization} is one of the standard techniques used in such a setting~\cite{jain2013low,keshavan2010matrix}.  Since the objective is non-convex in $(U, V)$ together but convex in each factor individually, the algorithm alternates between the two steps:
\begin{enumerate}
\setlength{\itemsep}{0.8pt}
\item  Fix $V$: Solve for $U$ (usually a standard least-squares problem).
\item Fix $U$: Solve for $V$.
\end{enumerate}
This method is appealingly simple (although developing rigorous recovery guarantees is far from simple) and was used in early implementations of the Netflix Prize recommendation engine~\cite{zhou2008large}.

{\em Riemannian optimization on fixed-rank manifolds}~\cite{vandereycken2013low,absil2008optimization}
is another common technique. Here, instead of treating $U$ and $V$ as independent variables, we treat the set of all rank-$r$ matrices as a Riemannian manifold.
We perform gradient descent, but after each step, ``project'' the result back onto the manifold of rank-$r$ matrices (often using a truncated SVD).
The updates are restricted to the tangent space $\Tn_{X_0}$ to ensure we stay close to the low-rank structure.

A related (but distinct) idea is {\em Iterative Hard Thresholding} (IHT)~\cite{recht2010guaranteed},
which keeps the problem in the $X$-space but enforces the rank constraint at every step. For example, the core steps of Iterative Hard Thresholding are
\begin{enumerate}
\setlength{\itemsep}{0.8pt}
\item  Gradient step: $\tilde{X}^{(k)} = X^{(k-1)} - \eta \nabla f(X^{(k-1)})$
\item Hard Thresholding: $X^{(k)} = P_r(\tilde{X}^{(k)})$, 
\end{enumerate}
where $P_r$ is the projection operator that keeps only the top $r$ singular values and sets the remaining singular values to zero. 

While Iterative Hard Thresholding and Riemannian manifold-based algorithms  share the same goal of maintaining a low-rank structure, they are fundamentally different in how they treat the geometry of the problem.
Iterative Hard Thresholding takes a step in the ``ambient'' space (the space of all matrices, regardless of rank) and then pulls the result back to the low-rank set. It treats the rank-$r$ constraint as a set to project onto. In contrast, Riemannian algorithms never leave the manifold. The gradient itself is projected onto the tangent space $\Tn_{X_k}$first. The update moves along a ``geodesic'' that follows the curvature of the rank-$r$ surface.

{\em Proximal methods} like {\em Iterative Soft-Thresholding Algorithm} (ISTA,~\cite{cai2010singular}) are also very effective for nuclear minimization problems, or more generally for non-smooth optimization where the objective can be split into a smooth part  $g(X)$ and a non-smooth part $h(X)$. For nuclear norm minimization the proximal method is applied to the objective $f(X) = \frac{1}{2}\|\cA(X)-y\|^2 + \lambda \|X\|_*$). Instead of taking a subgradient step on the whole function, we take a gradient step on the smooth part (i.e.,  $g(X)=\frac{1}{2}\|\cA(X)-y\|^2 $) and then apply a {\em proximal operator} to handle the non-smooth part (i.e., $h(X)=\lambda \|X\|_*$).  How does the proximal operator look like in this case? For a matrix $Y$ and a parameter $\tau > 0$, the proximal operator of the nuclear norm $h(X) = \tau \|X\|_*$ is defined as the solution to
$$\text{prox}_{\tau \|\cdot\|_*}(Y) = \arg\min_X \left( \tau \|X\|_* + \frac{1}{2}\|X - Y\|_F^2 \right).$$
The closed-form solution is obtained by performing an SVD of $Y$ and applying soft-thresholding to its singular values. If the SVD of $Y$ is $U \Sigma V^T$, then
$$\text{prox}_{\tau \|\cdot\|_*}(Y) = U \Sigma_\tau V^T$$
where $\Sigma_\tau$ is a diagonal matrix whose entries are $(\sigma_i - \tau)_+ = \max(\sigma_i - \tau, 0)$.

\begin{figure}[h]
\begin{center}
    \includegraphics[width=.7\textwidth]{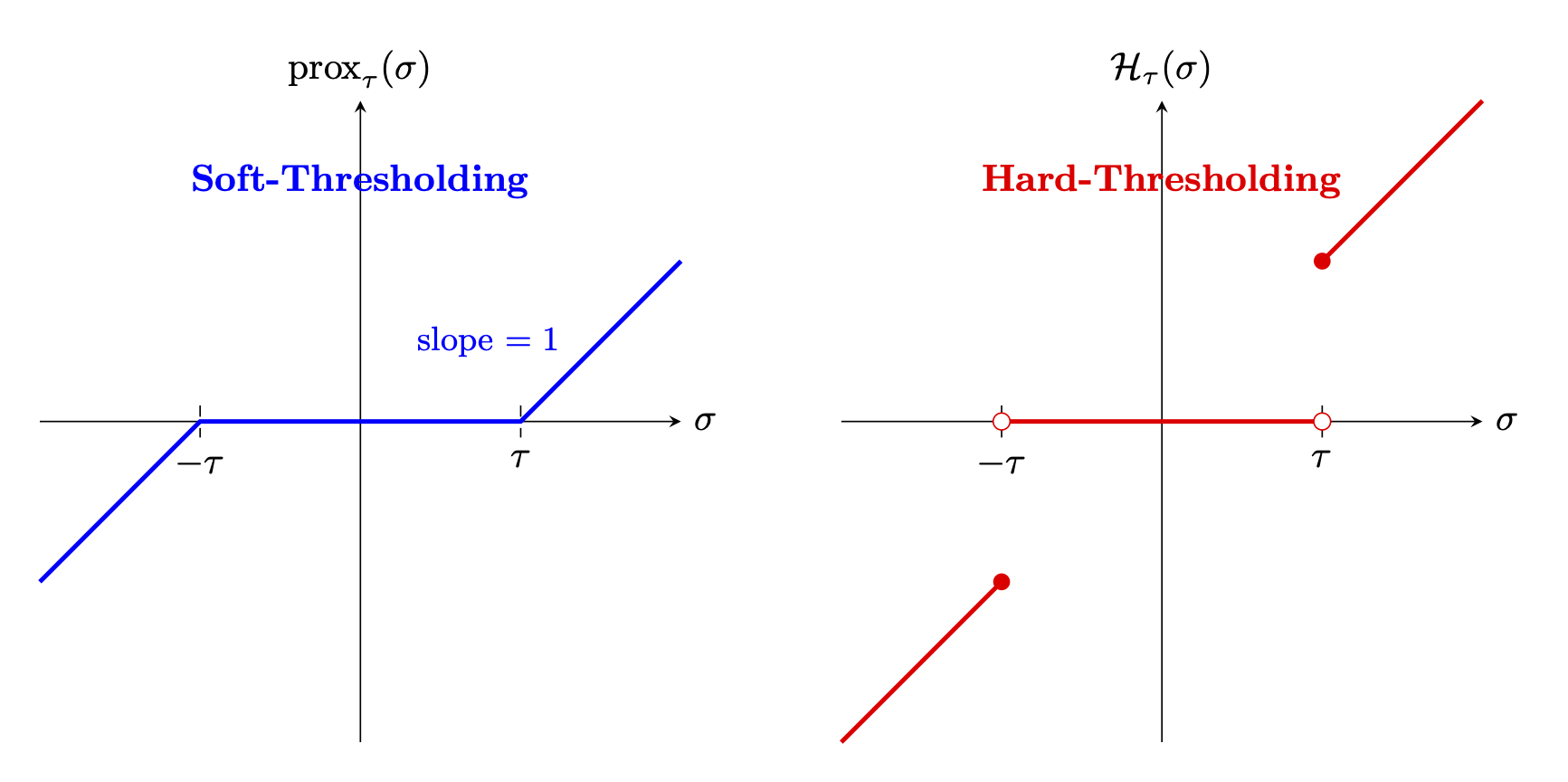}
    \caption{Soft thresholding vs.\ hard thresholding.}
    \label{fig:thresholding}
\end{center}
\end{figure}

While a hard thresholding operator cuts singular values abruptly, a proximal operator shrinks singular values continuously (soft thresholding), see also Figure~\ref{fig:thresholding}.

In both ISTA and IHT, the most computationally expensive step is the full SVD required to find the singular values and vectors. As the matrix dimensions grow, a full SVD becomes a massive bottleneck. For very large-scale matrices, the randomized SVD algorithm of Chapter~\ref{s:RSVD} or Krylov subspace based methods~\cite{Golub_MatrixComputations} provide an effective way to overcome this bottleneck,
especially when we know the target rank is much smaller than the matrix dimensions.

{\em Spectral initialization}~\cite{candes2015wirtinger,Jain:matrixcompletion_altmin_13} has turned out to be an important ingredient in many non-convex iterative methods. To construct a good initializer, one sets
$$X^{(0)}= P_{r}(\cA^{*}y).$$
Under appropriate conditions (such as having a large enough number of random measurements) one can show that this initialization ensures
$$\|X^{(0} - X_0\| \le c \|X_0\|$$
(for some appropriate constant $c>0$) with high probability.
Once an approximation is within this $c$-ball, the objective function behaves locally like a strongly convex function, allowing methods like gradient descent to converge at a linear rate to the global optimum.

A notable development in the last decade~\cite{bhojanapalli2016global,ge2017no,WrightMa} was that for many low-rank problems (like matrix completion or phase retrieval), the non-convex landscape is ``benign'', which means that under certain conditions (like RIP) there are no spurious local minima (every local minimum is also a global minimum) and all other critical points are ``strict saddles,'' meaning there is always a direction of descent to escape them.

We refer the reader to~\cite{WrightMa} for an in-depth analysis of convex and nonconvex methods and a detailed discussion of further interesting applications of low-rank matrix recovery.

\section*{Exercises}
\addcontentsline{toc}{section}{Exercises}

\begin{myexercise}
Let $A$ be a matrix with singular values $\sigma_1, \sigma_2, \dots, \sigma_r > 0$.
Prove that the nuclear norm $\|A\|_*$ is the convex envelope of $\text{rank}(A)$ on the set $\{A : \|A\| \leq 1\}$.
\end{myexercise}

\begin{myexercise}
A linear map $\mathcal{A}: \mathbb{R}^{m \times n} \to \mathbb{R}^p$ satisfies the matrix-RIP with constant $\delta_r$ if for all matrices $X$ of rank at most $r$
$$(1-\delta_r)\|X\|_F^2 \leq \|\mathcal{A}(X)\|^2 \leq (1+\delta_r)\|X\|_F^2.$$
If $\delta_{2r} < 1$, prove that there is at most one matrix of rank $r$ that satisfies $\mathcal{A}(X) = b$.
\end{myexercise}

\begin{myexercise}
Let $C \subset \mathbb{R}^n$ be a {\em closed convex cone}. The {\em statistical dimension} of $C$, denoted $\delta(C)$, is defined as:
$$
\delta(C) := \mathbb{E}\big[ |\Pi_C(g)|_2^2 \big],
$$
where $g \sim \mathcal{N}(0, I_n)$ is a standard Gaussian vector in $\mathbb{R}^n$ and $\Pi_C(g)$ denotes the Euclidean projection of $g$ onto $C$. Intuitively,
$\delta(C)$ measures the ``effective dimension'' of $C$ in $\mathbb{R}^n$ in a probabilistic sense, analogous to the dimension of a subspace. \\
(a) Prove that for a subspace $L \subset \mathbb{R}^n$ of dimension $d$, $\delta(L) = d$.
\\
(b) Let $w$ denote the Gaussian width introduced in Definition~\ref{def:gaussianwidth}. Prove that for a convex cone $C$ we have
$$
w^2(C \cap S^{n-1}) \le \delta(C) \le w^2(C \cap S^{n-1}) + 1.
$$
So the statistical dimension is essentially the squared Gaussian width of the intersection of the cone with the unit sphere.
\end{myexercise}

\begin{myexercise}
Let $M_1, M_2 \in \R^{n_1\times n_2}$ be two matrices, and
let $M=[M_1, M_2]$ be their concatenation. Show that \\
(a) $\|M\|_* \le \|M_1\|_* + \|M_2\|_*$. \\
(b) $\|M\|_* = \|M_1\|_* + \|M_2\|_*$ if $M_1^{T} M_2 = 0$.
\end{myexercise}

\begin{myexercise}
Let $A$ and $B$ be matrices of the same dimensions. Prove the following:
If $A B^T = 0$ and $A^T B = 0$ then 
$\|A+B\|_* = \|A\|_* +\|B\|_*$.
\end{myexercise}

\begin{myexercise}\label{ex:nuclear_SDP}
Show that the nuclear norm minimization problem
$$\min_X \|X\|_* \quad \text{subject to} \quad \mathcal{A}(X) = b$$
can be rewritten  as
$$\min_{X, W_1, W_2} \frac{1}{2}\left(\operatorname{tr}(W_1) + \operatorname{tr}(W_2)\right)$$
$$\text{subject to} \quad \begin{pmatrix} W_1 & X \\ X^\top & W_2 \end{pmatrix} \succeq 0, \quad \mathcal{A}(X) = b,$$
which is a standard SDP in the variables $(X, W_1, W_2)$. \\ (Note: At first glance, the objective $\frac{1}{2}\left(\operatorname{tr}(W_1) + \operatorname{tr}(W_2)\right)$ does not seem to involve $X$, but  the Schur complement of $\begin{pmatrix} W_1 & X \\ X^\top & W_2 \end{pmatrix}$
reveals that $X$ appears implicitly in the objective.)
\end{myexercise}

\begin{myexercise} (Schoenberg's Theorem)
Given $n$ points $x_1,\dots,x_n \in \R^d$,
the {\em Euclidean distance matrix} 
$D = [d_{ij}] \in \mathbb{R}^{n \times n}$ is defined by
$$
d_{ij} := \|x_i - x_j\|^2
\qquad \text{for $i,j = 1,\dots,n$}.
$$
(Note that the distances are squared in the definition of $D$.)\\
(a) Show that $D \in \mathbb{R}^{n \times n}$  is a Euclidean distance matrix if and only if $D=D^T$ and
$H D H^{\ast} \preceq 0$, where $H = I_n - \frac{1}{n}\mathbf{1}\mathbf{1}^T$ is the {\em centering matrix} and
$\mathbf{1}$ is the vector whose entries are all ones.
\\
(b) Show that $\rank(H D H^{\ast}) \le d$.
(This condition implies that $D$ is a low-rank matrix if $d \ll n$).
\end{myexercise}

\begin{myexercise}
Let $X = [x_1, x_2, \dots, x_n]^T \in \mathbb{R}^{n \times d}$ be a matrix where each row $x_i \sim \mathcal{N}(0, I_d)$ is an independent Gaussian random vector in $d=3$ dimensions. We consider $n=200$ points. The full squared Euclidean distance matrix $D$ is defined as
$$D_{ij} = \|x_i - x_j\|^2.$$
The squared distance matrix $D$ can be expressed as
$$D = \mathbf{1}z^T + z\mathbf{1}^T - 2XX^T,$$
where $z$ is a vector containing the squared norms of the rows of $X$. 
\begin{enumerate}[label=(\alph*)]
\item
Based on this decomposition, what is the exact rank of $D$ when $d=3$ and $n=200$?
\item We randomly delete a total of 3000 entries of $D$ (note that entries should be deleted in a symmetric manner since if we do not know the distance from $x_i$ to $x_j$, we also do not know the distance from $x_j$ to $x_i$). Use the convex solver of your choice to reconstruct the missing entries of $D$. 
\item Once you completed the missing entries of the Euclidean distance matrix $D$ you recover the points via a method called {\em multidimensional scaling} which we sketch here: Form the Gram matrix $G=-\frac{1}{2}J D J$,  where $J$ is the centering matrix defined as  $J = I-\frac{1}{n} {\textbf 1} {\textbf 1}^T$, then do an eigenvalue decomposition on $G$. Work out and implement the details of this procedure.
\item This time we delete the 3000 largest entries of $D$ (again, in a symmetric manner). This models the case where a sensor can only measure local distances.
Use again the solver you used in (b) to reconstruct the missing entries of $D$. 
\item How many measurements can you delete on average in each case until the algorithm can no longer accurately (say, up to a relative error of $10^{-2}$ reconstruct the missing entries)?
Compare the performance of the solver in these two settings described in (b) and (d).
\end{enumerate}

\end{myexercise}

\begin{myexercise}\label{ex:pauli}
(a) Show that the Pauli group
$$
\mathcal{P}_{n} = \big\{ 
\sigma_{a_1} \otimes \cdots \otimes \sigma_{a_n} : a_k \in \{0,x,y,z\} \big\}
$$
forms a basis of $\mathbb{C}^{d\times d}$ with cardinality $d^2$. \\
(b) Prove that the normalized Pauli operators satisfy
$$
\operatorname{tr}(P_\alpha^\ast P_\beta) = \delta_{\alpha\beta},
$$
hence $\{P_\alpha\}_{\alpha=1}^{d^2}$ is an orthonormal basis for the Hilbert–Schmidt inner product.
\end{myexercise}

\begin{myexercise}
Low-Rank Quantum State Tomography:
Consider a 5-qubit system ($n=5$).  Assume the quantum system is prepared in a state $\rho$ that is pure. The density matrix $\rho$ is an $32 \times 32$ complex matrix.
You are given the expectation values of $m$ randomly sampled Pauli observables $\{A_k\}_{k=1}^m$. In the noise-free case each measurement $y_k$ is given by
$$y_k = \text{Tr}(A_k \rho).$$
The observables $A_k$ are chosen from the set of 5-qubit Pauli strings $\{I, X, Y, Z\}^{\otimes 5}$.
Since $\rho$ is known to be PSD and has a trace of 1, we use the following convex optimization problem to recover $\rho$ from the measurements $y$:
\begin{align*}
\min_{\hat{\rho}} & \quad \sum_{k=1}^m |\text{Tr}(A_k \hat{\rho}) - y_k|^2 + \lambda \|\hat{\rho}\|_* \\ \text{s.t.} & \quad \hat{\rho} \succeq 0 \quad \text{and} \quad \text{Tr}(\hat{\rho}) = 1.
\end{align*}
(Note that because $\rho$ is PSD, the nuclear norm is simply the trace: $\|\rho\|_* = \text{Tr}(\rho)$. Since we already constrain $\text{Tr}(\rho) = 1$, the nuclear norm is constant. The ``low-rank'' promotion actually comes from the PSD constraint combined with the objective function.)

\begin{enumerate}[label=(\alph*)]
\item Sample only $m = 250$ random Pauli measurements (note that a full basis requires $4^5 = 1024$ measurements) and use a Semidefinite Programming  solver (like CVX or SCS) to recover $\hat{\rho}$.
\item Calculate the {\em quantum fidelity} $F(\rho, \hat{\rho}) = \left( \text{Tr} \sqrt{\sqrt{\rho} \hat{\rho} \sqrt{\rho}} \right)^2$. How close is your recovery to the true state?
\item Repeat the experiment with a reduced the number of measurements $m$. How small can you make $m$ so that recovery of $\rho$ is numerically still possible in your simulations?
\end{enumerate}
\end{myexercise}

\begin{myexercise}
Consider the phase retrieval problem where we want to reconstruct a signal $x\in \C^n$ from intensity measurements of the form $\{|\langle x,a_k \rangle|^2\}_{k=1}^m$.
\begin{enumerate}[label=(\alph*)]
\item Assume the measurement vectors $\{a_k\}_{k=1}^m$ form an equiangular tight frame for $\C^n$ with $m=n^2$. 
Let $X=x x^\top$. Show that in this case the solution takes on the very simple form
\begin{equation}\label{eq:phaseequiframe}
X = \frac{n+1}{n} \sum_{k=1}^{n^2} |\langle x,a_k \rangle|^2 a_k a_k^\top - \|x\|^2 I_n.
\end{equation}
\item In this reconstruction formula, how can we address the issue that we do not know $\|x\|$?
\item How does the reconstruction formula~\eqref{eq:phaseequiframe}
change if instead of an equiangular tight frame $\{a_k\}_{k=1}^{n^2}$ we use a mutually unbiased basis with a total of $n^2+n$ elements?
\end{enumerate}

\end{myexercise}



\backmatter
\printindex

\def\refname{}
\def\bibtitle#1{\def\refname{#1}}
\bibtitle{{\bf Bibliography}}
\bibliography{mathbook}
\bibliographystyle{plain}


\end{document}